\pdfoutput=1
\documentclass[11pt]{article}

\usepackage[preprint]{acl}
\usepackage{times}
\usepackage{latexsym}

\usepackage[T1]{fontenc}
\usepackage[utf8]{inputenc}

\usepackage{microtype}

\usepackage{inconsolata}

\usepackage{graphicx}

\usepackage{subcaption}

\usepackage{bm}
\usepackage{amsfonts}
\usepackage{amsmath}

\usepackage{heatmap}
\usepackage{booktabs}

\usepackage{tabularx}
\usepackage{tcolorbox}

\usepackage{CJKutf8}

\title{The Transfer Neurons Hypothesis: An Underlying Mechanism for Language Latent Space Transitions in Multilingual LLMs}

\author{
 \textbf{Hinata Tezuka\textsuperscript{1}} \quad
 \textbf{Naoya Inoue\textsuperscript{1,2}} \quad
\\
 \textsuperscript{1}Japan Advanced Institute of Science and Technology \quad
 \textsuperscript{2}RIKEN
\\
\texttt{\{hinata-t, naoya-i\}@jaist.ac.jp}
}

\begin{document}
\maketitle
\begin{abstract}
% This document is a supplement to the general instructions for *ACL authors. It contains instructions for using the \LaTeX{} style files for ACL conferences.
% The document itself conforms to its own specifications, and is therefore an example of what your manuscript should look like.
% These instructions should be used both for papers submitted for review and for final versions of accepted papers.
Recent studies have suggested a processing framework for multilingual inputs in decoder-based LLMs: early layers convert inputs into English-centric and language-agnostic representations; middle layers perform reasoning within an English-centric latent space; and final layers generate outputs by transforming these representations back into language-specific latent spaces.
However, the internal dynamics of such transformation and the underlying mechanism remain underexplored.
Towards a deeper understanding of this framework, we propose and empirically validate \textbf{The Transfer Neurons Hypothesis}: certain neurons in the MLP module are responsible for transferring representations between language-specific latent spaces and a shared semantic latent space.
Furthermore, we show that one function of language-specific neurons, as identified in recent studies, is to facilitate movement between latent spaces.
Finally, we show that transfer neurons are critical for reasoning in multilingual LLMs\footnote{Our codes are available at \url{https://github.com/HinaTezuka/emnlp2025-transfer-neurons}.}.
\end{abstract}

\section{Introduction}
\label{sec:introduction}
Multilingual \textbf{L}arge \textbf{L}anguage \textbf{M}odels (LLMs), pretrained on multilingual corpora, can process multiple languages within a single model. Recent studies suggest that decoder-based multilingual LLMs operate on a semantic latent space that is independent of the input language, e.g., language-agnostic latent space~\cite{zhang-etal-2024-unveiling-linguistic, coling2025-converging, bandarkar2025layer-swapping} and English latent space~\cite{neurips2024-multi,acl2024-llamas,schut2025multilingualllmsthinkenglish}.

\begin{figure}[t]
  \includegraphics[width=\columnwidth]{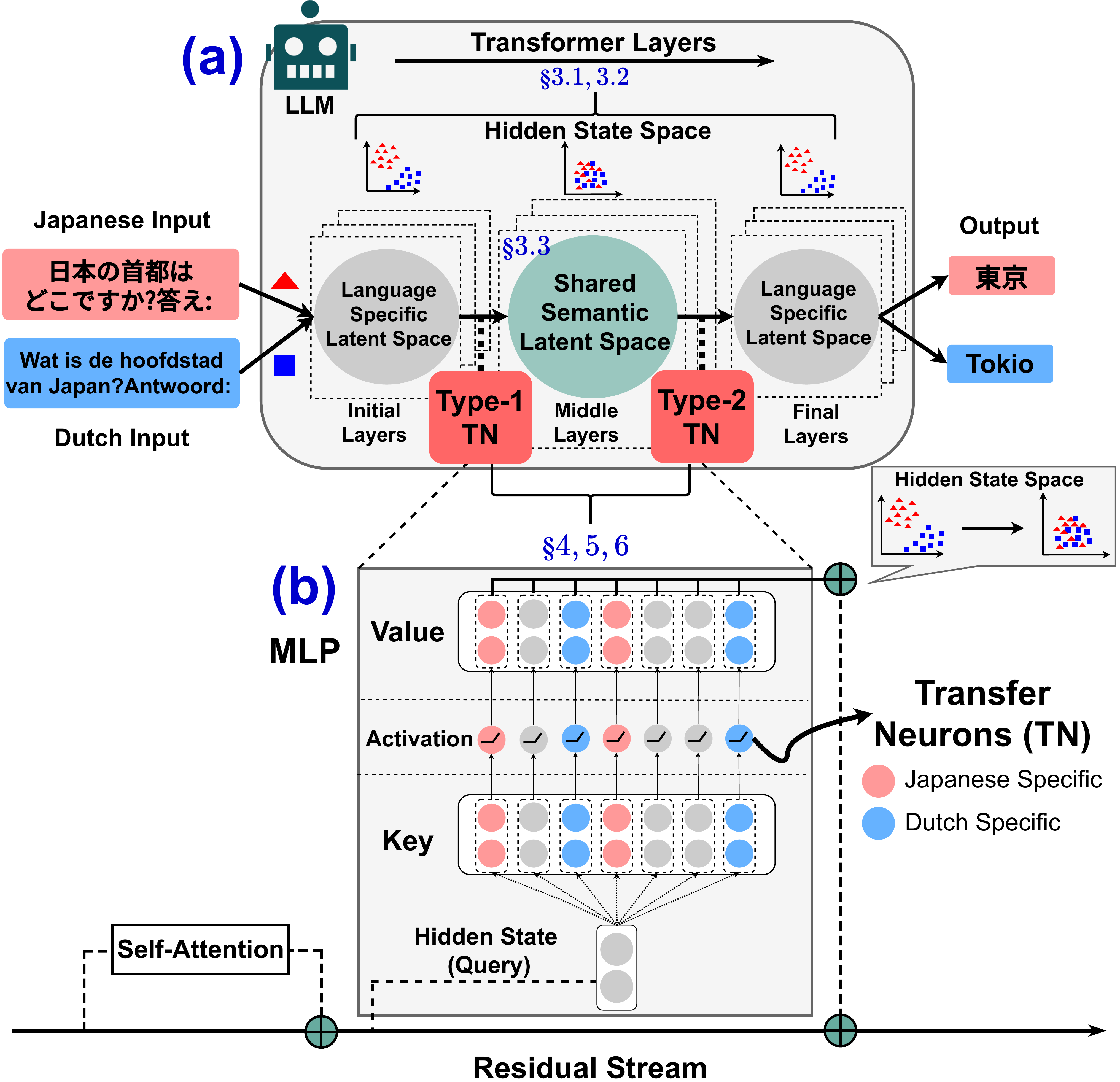}
  \caption{\textbf{An Overview of the Transfer Neurons Hypothesis}. We hypothesize that specific neurons in MLP module move representations from language-specific latent spaces to a shared semantic latent space, and vice versa.}
  \label{fig:figure1}
\end{figure}

The previous findings imply that multilingual LLMs potentially process inputs by (i) transferring input into the shared latent space, (ii) performing semantic processing, and (iii) mapping results back to the input language, as shown in Fig.~\ref{fig:figure1}~(a). However, the dynamics of internal representation transformation and the underlying mechanism of the representation transfer remain underexplored, preventing a comprehensive understanding of semantic processing in multilingual LLMs.

% figure1

To address these issues, first, we study how internal representations evolve across layers and examine the existence of a shared semantic latent space in the middle layers.
Second, as a plausible explanation for the representation transfer mechanism, we propose the \textit{Transfer Neurons Hypothesis}: specific neurons in the \textbf{M}ulti \textbf{L}ayer \textbf{P}erceptron (MLP) module are responsible for this transfer. As shown in Fig.~\ref{fig:figure1}~(b), we hypothesize that certain neurons, named \textit{transfer neurons}, are sensitive to the input language, and that their firing results in adding corresponding value vectors to the residual stream, thereby facilitating the representation transfer.

\paragraph{Our Contributions are summarized as follows: }
\begin{itemize}
%     \item We reconstruct the inference framework of LLMs for handling multilingualism by unifying findings from previous studies and contributing additional evidence.
    \item We uncover the dynamics of internal representations in multilingual LLMs: initial representations are language-specific, subsequently converge into a shared latent space across languages, and finally diverge again (\S\ref{sec:The Dynamics of Spatial Transition}).
    \item We empirically demonstrate the existence of transfer neurons that mediate transitions between language-specific latent spaces and the shared semantic latent space. We further show their necessity for the representation transfer through intervention experiments (\S\ref{sec:methodology for identifying transfer neurons}, \S\ref{sec:detecting and controlling transfer neurons}).
    \item We suggest that one of the key role of language-specific neurons identified in recent studies is to facilitate movement between latent spaces, and show that transfer neurons are essential for downstream tasks (\S\ref{sec:nature of transfer neurons}).
\end{itemize}

\section{Background}
\label{sec:background}

\subsection{Related Works}
\label{sec:Relatedworks}
\paragraph{The Framework for Multilingual Processing.} By detecting and controlling language-specific neurons, \citet{neurips2024-multi} introduced the \texttt{MWork} framework, arguing that LLMs process multilingual queries in the first few layers, reason in English in the middle layers, and subsequently generate responses in input language. Although they do not explicitly define it as English-centric, \citet{coling2025-converging} proposed a similar framework by introducing the concept of a \texttt{Lingua Franca} — a common semantic latent space that facilitates language-agnostic processing. Assuming language-agnostic reasoning occurs in the middle layers, \citet{bandarkar2025layer-swapping} improved multilingual reasoning performance using a layer swapping technique. They trained language-expert and reasoning-expert models separately, then merged them by assigning the language-expert to the first and last few layers, and the reasoning-expert to the middle layers. \citet{acl2024-llamas, schut2025multilingualllmsthinkenglish} observed that when non-English queries are presented, the models often identify an answer in English internally, and then translate it into the target output language.
\paragraph{Language-Agnostic Representations.}
It has been shown that knowledge neurons cross-lingually encoding specific concepts exist \citep{journey_knowledge_neuron, tracing_facts_mllms, multilingual_knowledge_neurons_colling2025}. Building on findings that encoder-based LLMs exhibit a degree of language-agnosticism \citep{2019-multilingual-bert, libovicky-etal-2020-language}, \citet{yoon-etal-2024-langbridge} developed the LangBridge model, which connects a multilingual encoder to LLM, and improved performance on reasoning tasks even in low-resource languages.
\paragraph{Language-Specific Regions.}
Recent studies \citep{kojima2024, tang_lang_specific_neurons, neurips2024-multi, coling2025-converging, duan-etal-2025-unveiling} have revealed the existence of language-specific regions or neurons in decoder-based LLMs, which respond strongly to inputs in a particular language. These regions are primarily distributed in the initial and final few layers of the models, a finding that is also supported by our experiment in Appendix~\ref{sec:appendix:language specific neurons}. \citet{mondal-etal-2025-language-specific} demonstrated that fine-tuning only these neurons is not sufficient for cross-lingual improvements on downstream tasks. The role of these regions or neurons, however, has not yet been sufficiently revealed.

Although these studies offer valuable yet fragmentary evidence, the overall process of the latent space transitions still remains unclear and lack quantitative demonstration.

% figure: pca llama3.
\begin{figure*}[t]
    \centering
    
    \includegraphics[width=0.32\linewidth]{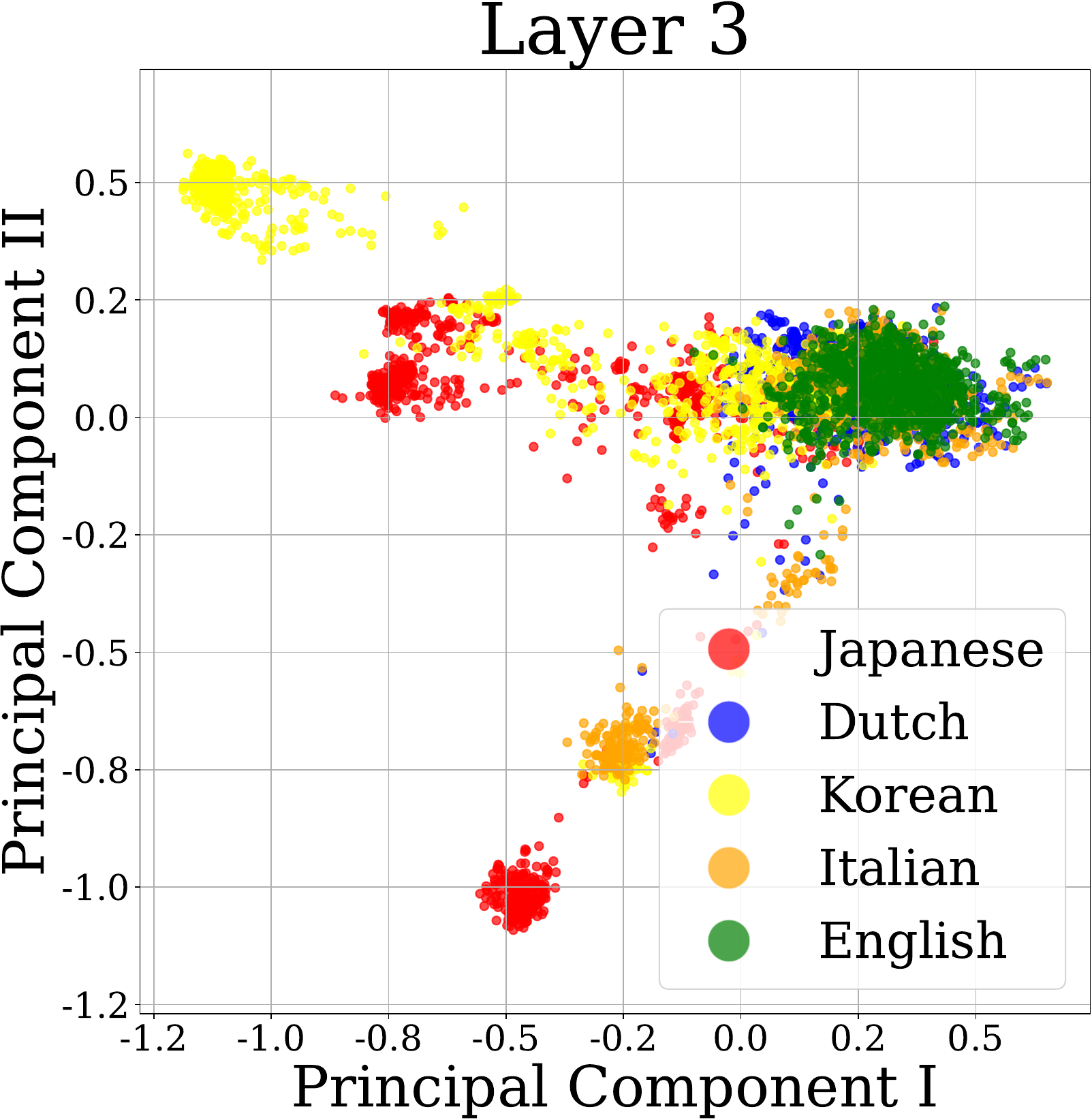}
    \includegraphics[width=0.32\linewidth]{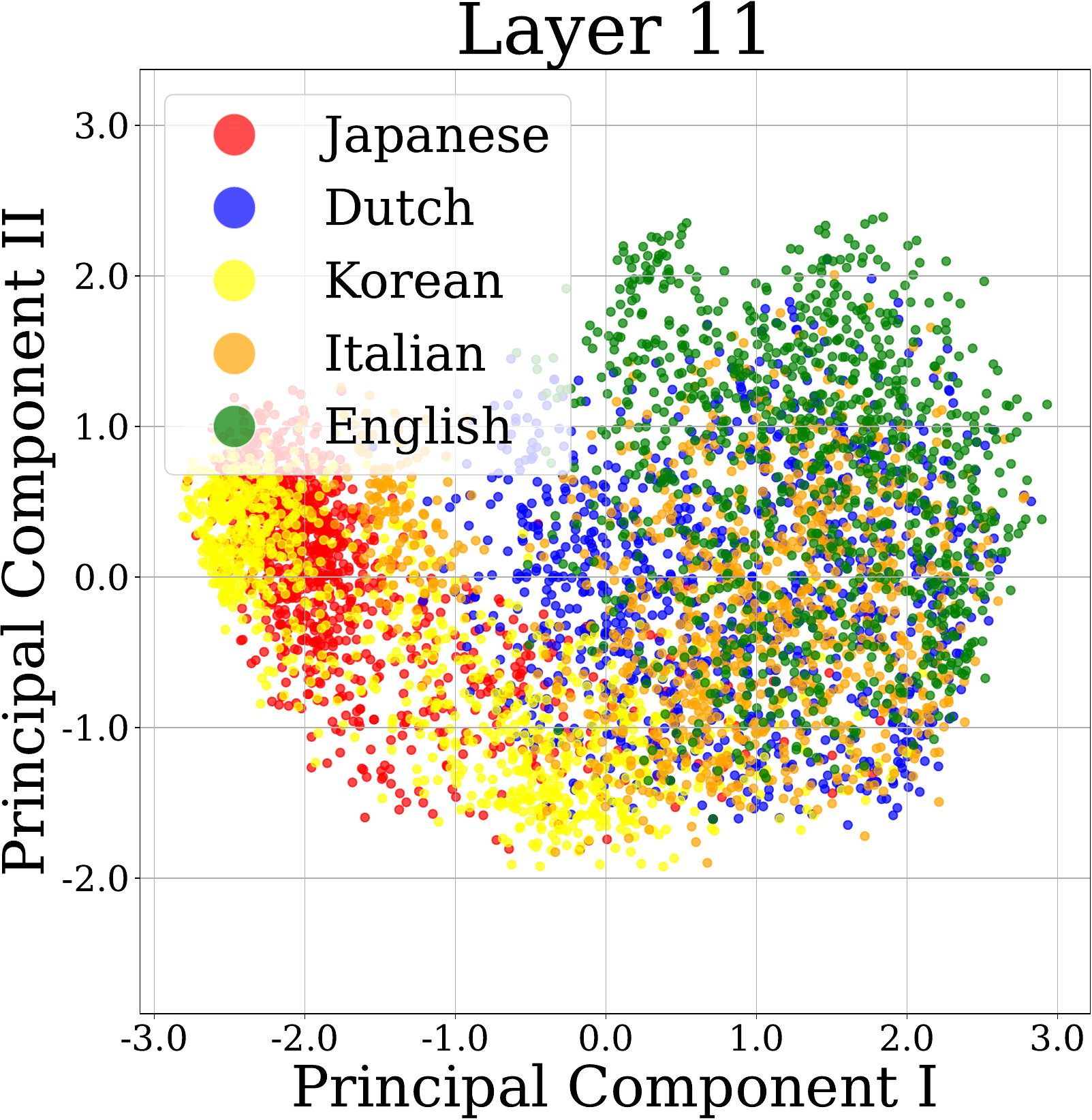}
    \includegraphics[width=0.32\linewidth]{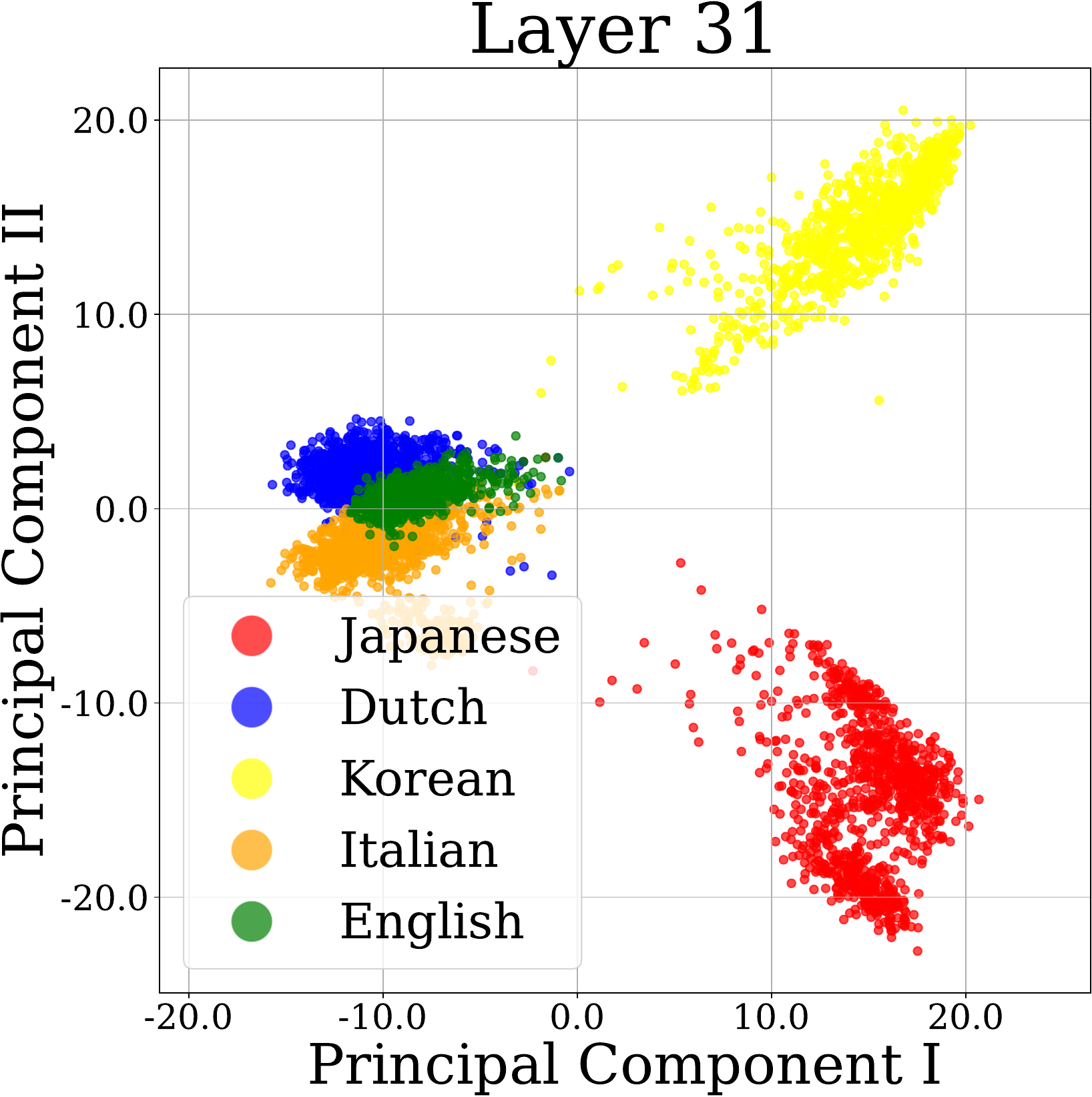}
    
    \caption{\textbf{The results of PCA applied to hidden language representations for capturing spatial transitions (LLaMA3-8B).}
    \textcolor{red}{Japanese}, \textcolor{orange}{Italian}, \textcolor{blue}{Duch}, \textcolor[HTML]{228B22}{English} and \textcolor[HTML]{CCCC00}{Korean} hidden representations, respectively. Each input sentence from each language has the same meaning. The results for all layers and models can be found in Appendix~\ref{sec:appendix:visualization language subspaces}.
    }
    \label{fig:pca llama first-middle-last layers}
\end{figure*}

\subsection{Neuronic View of MLPs in Transformers}
\label{computation of mlp}
Given an input vector $\bm{x} \in \mathbb{R}^d$, the MLP module in Transformer at layer $l$ operates as follows:
\begin{equation}
\mathrm{MLP}^{l}(\bm{x}) = a(\bm{x}M_{\mathrm{up}}^{l}) M_{\mathrm{down}}^{l}
\label{eq:mlp}
\end{equation}
where $a$ represents a non-linear activation function, and $M_{\mathrm{up}}^{l} \in \mathbb{R}^{d\times d_m}, M_{\mathrm{down}}^{l} \in \mathbb{R}^{d_m\times d}$ refer to the up and down projection matrices, respectively. \newcite{geva2021} proposed a vector-based key-value store view of the MLP module. Here, the keys are $d_m$ column vectors in $M_{\mathrm{up}}^{l}$, the values are $d_m$ row vectors in $M_{\mathrm{down}}$, and the query is $\bm{x}$.
Viewing $\bm{\alpha}^{l} = a(\bm{x}M_{\mathrm{up}}^{l})$ as the relevance scores between the keys and query, the MLP module can be rewritten as follows:
\begin{equation}
\mathrm{MLP}^{l}(\bm{x}) = \bm{\alpha}^{l}M_{\mathrm{down}}^{l} = \sum_{i=1}^{d_m} \alpha_i^{l}\bm{v}_i^{l}
\label{eq:weighted_sum}
\end{equation}
where $\bm{v}_i^{l}$ represents $i$-th row vector in $M_{\mathrm{down}}^l$.
In our experiments, we adopt the MLP module with a gating mechanism~\citep{gluvariantsimprovetransformer, gated_mlp_paper}, namely $\bm{\alpha}^l = a(\bm{x}M_{\mathrm{gate}}^{l}) \odot \bm{x}M_{\mathrm{up}}^{l}$, where $\odot, M_{\mathrm{gate}}^{l} \in \mathbb{R}^{d\times d_m}$ represents element-wise multiplication and gated projection, respectively.
Following \citet{knowledge_neurons}, we call each $\alpha_i^l$ \emph{neuron}.

\section{The Dynamics of Hidden States}
Based on the findings of prior work~\cite{neurips2024-multi, acl2024-llamas, schut2025multilingualllmsthinkenglish}, we assume that a shared semantic latent space is predominantly English-centric. In other words, \textcolor[HTML]{228B22}{the English latent space} effectively serves as the shared semantic space across languages. Accordingly, we adopt English as the core language within this shared latent space throughout the experiments in this study.

In this section, we provide several evidence supporting the latent space transitions across layers and the existence of the shared semantic latent space in middle layers, as illustrated in Fig.~\ref{fig:figure1}(a).
\label{sec:The Dynamics of Spatial Transition}
\subsection{Spatial Transition Phenomenon}
\paragraph{Representations Form Language-Specific Latent Spaces and a Shared Latent Space in Hidden State Space.}
\label{Visualizing the Hidden language Subspace with PCA}
We encode 5k sentences from MKQA dataset~\cite{longpre-etal-2021-mkqa}, a multilingual dataset of parallel QA sentences in five languages, using LLaMA3-8B (1k sentences per language). We extract the hidden states corresponding to the final token of each sentence from each layer, apply PCA, and plot the results along the top two principal components in Fig.~\ref{fig:pca llama first-middle-last layers}. It shows that in the initial layers, each language forms its own latent space, which gradually converges into a shared latent space towards the middle layers. In the final layers, each language re-establishes its distinct latent space, which confirms our hypothesis. An unexpected finding, however, is that two latent spaces specific to \textcolor{blue}{Dutch} and \textcolor{orange}{Italian}, which have linguistic proximity to \textcolor[HTML]{228B22}{English}, remain close to the English latent space even in the initial and final layers, in contrast to \textcolor{red}{Japanese} and \textcolor[HTML]{CCCC00}{Korean}. Additionally, Japanese and Korean, which are linguistically more distant from English, do not fully converge into the shared latent space even in the middle layers. These trends are consistent across models (see Appendix~\ref{sec:appendix:visualization language subspaces}). These results suggests that while representations tend to converge into the English-centric shared semantic latent space, the degree of convergence varies across languages.

%--- NEW (Subspace Property Section) ---%
\subsection{Geometry of Language-Specific Latent Spaces}
\label{sec:Subspace Property of Language-Specific Hidden States}
\paragraph{Hidden States Reside in Language Latent Spaces.}To investigate whether hidden states for each language can be captured by low dimensional latent space, we conduct a \textbf{S}ingular \textbf{V}alue \textbf{D}ecomposition (SVD) to a matrix formed by hidden states of each language and compute the cumulative explained variance ratio, to estimate the number of basis vectors necessary to form each language latent space. Let $M^l_{L_i} \in \mathbb{R}^{n \times d}$ be a matrix containing hidden state vectors at layer $l$ (each of dimension $d$) for $n$ sentences in language $L_i$. Then, $M^l_{L_i}$ is decomposed as follows:
\begin{equation}
    M^l_{L_i} = U \Sigma V^T
\end{equation}
where $U$ and $V$ are orthogonal matrices that can be interpreted as a set of orthonormal basis for row space and column space, respectively. $\Sigma$ is a diagonal matrix containing the singular values of $M^l_{L_i}$, which represent the magnitude of each corresponding component in the lower dimensional latent space in hidden state space. These singular values are ordered in descending order. Finally, we compute the cumulative explained variance ratio as follows:

\begin{equation}
    \text{CEVR}^{l}_{L_i} = \sum^k_{i=1}\frac{\sigma^2_i}{\Sigma^r_{j=1}\sigma^2_j}
\end{equation}
where $\sigma^2_i$ denotes the $i$-th singular value, $k$ is the number of components retained, and $r$ is the total number of non-zero singular values. The $\text{CEVR}^l_{L_i}$ thus represents the proportion of the total variance explained by the top-$k$ singular values.

\begin{figure}[t]
  \centering

  \includegraphics[width=\linewidth]{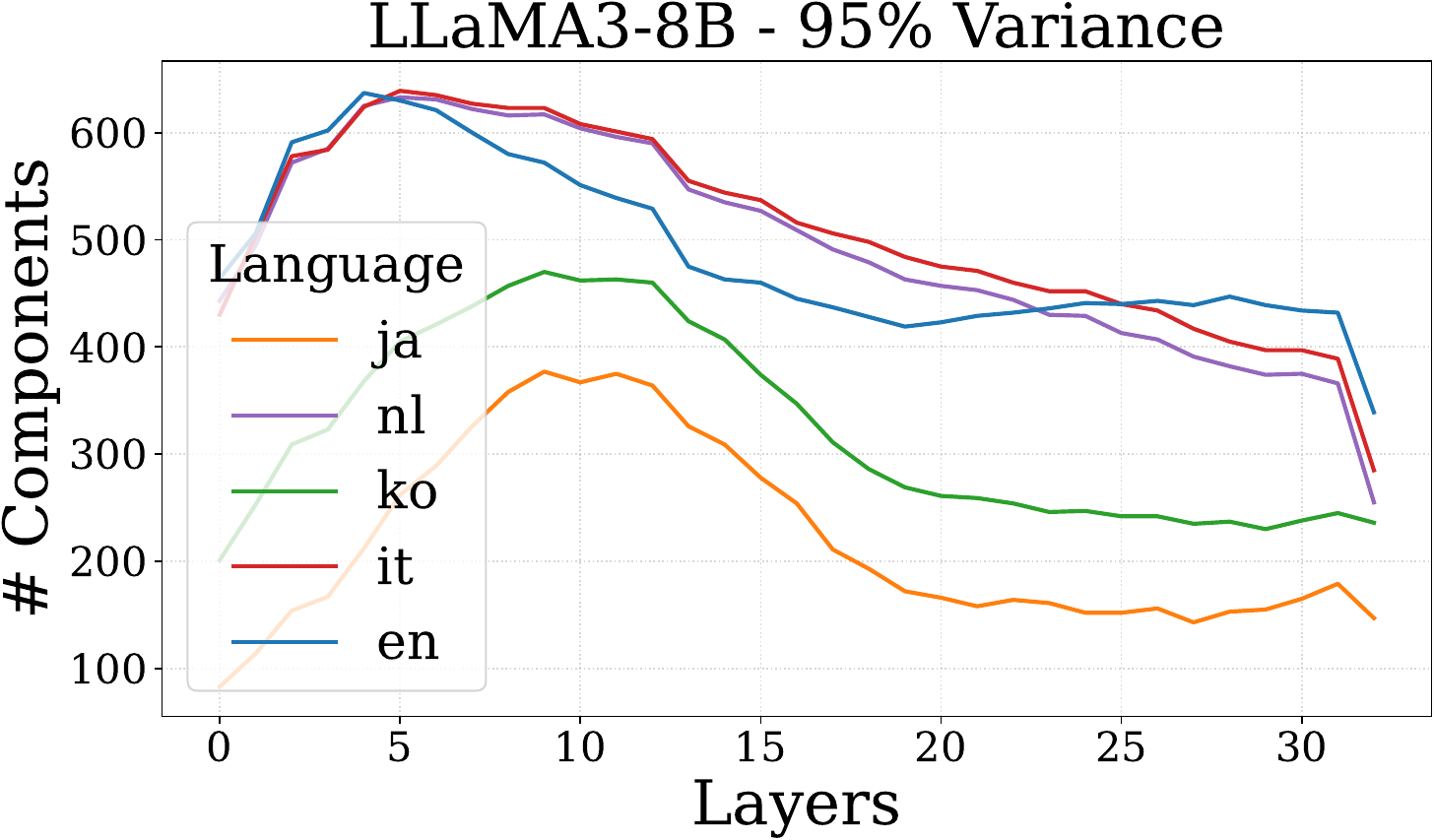}

  \caption{\textbf{Dimensionality of language latent spaces across layers estimated via SVD and CEVR (LLaMA3-8B).}
  }
  \label{fig:cumulative_explained_ratio_llama_mistral_95}
\end{figure}

The results are shown in Fig.~\ref{fig:cumulative_explained_ratio_llama_mistral_95}. As indicated, in initial and final few layers where hidden states specific to each language form their own latent space, most (95\%) of representations lie in a relatively low-dimensional latent space, a tendency that is expecially pronounced for languages that are more distant from English (i.e., Japanese and Korean). On the other hand, representations in the middle layers, where semantic processing and inference mainly occur, form a relatively high-dimensional latent space.
Additionally, as quantitatively demonstrated in Appendix~\ref{sec:appendix:distance among each language subspaces}, the transitions of the distance among the centroids of language-specific latent spaces across layers are well aligned with the observation of spatial transition in \S\ref{Visualizing the Hidden language Subspace with PCA} and Fig.~\ref{fig:pca llama first-middle-last layers}. The results for other variance thresholds and models can be found in Appendix~\ref{sec:appendix:dimensionality of subspaces across layers}.

These results indicate that the hidden states of each language shown in Fig.~\ref{fig:pca llama first-middle-last layers} lie in a relatively low dimensional latent space of the hidden state space.

% figure: SVD cumulative explained variance raio llama3/mistral.

% % figure2
% \begin{figure}[t]
%   \centering
%   % 1行目
%   \begin{minipage}[b]{0.49\linewidth}
%     \centering
%     \includegraphics[width=\linewidth]{figures/llama3/hs_sim/ja.pdf}
%     % \subcaption{Japanese}
%   \end{minipage}
%   \begin{minipage}[b]{0.49\linewidth}
%     \centering
%     \includegraphics[width=\linewidth]{figures/llama3/hs_sim/nl.pdf}
%     % \subcaption{Dutch}
%   \end{minipage}

%   % 2行目
%   \begin{minipage}[b]{0.49\linewidth}
%     \centering
%     \includegraphics[width=\linewidth]{figures/llama3/act_sim/ja.pdf}
%     \subcaption{en-ja}
%   \end{minipage}
%   \begin{minipage}[b]{0.49\linewidth}
%     \centering
%     \includegraphics[width=\linewidth]{figures/llama3/act_sim/nl.pdf}
%     \subcaption{en-nl}
%   \end{minipage}

%   \caption{\textbf{Similarity of hidden states and activations (LLaMA3-8B).}
%   The upper figures present the similarity of hidden states, and the lower ones present the similarity of activation patterns. \textcolor{blue}{Blue} indicates parallel sentence pairs, whereas \textcolor{red}{red} and \textcolor{orange}{orange} indicate non-parallel ones. en: English, ja: Japanese, nl: Dutch.}
%   \label{fig:llama_sim_hs_act}
% \end{figure}

\subsection{The Existence of a Shared Semantic Latent Space in Middle Layers}
\label{sec:mutual knn}
\paragraph{Sentence Representations with Similar Meanings across Languages Converge to Similar Locations in Middle Layers.}
To numerically demonstrate whether language-agnostic semantic processing indeed occurs in the shared semantic latent space shown in Fig.~\ref{fig:figure1}(a), we measure \textit{Mutual k-Nearest Neighbor Alignment Metric} proposed in \citep{icml2024platonic}, between a set of parallel sentence representations across language pairs. If similarities peak in middle layers, it suggests that semantically similar or equivalent sentence representations converge to similar positions across languages in hidden state space. This increases shared nearest neighbors, indicating aligned latent spaces and language-agnostic semantic processing. The computation is formalized in Appendix~\ref{sec:appendix:formalizing mutual_knn}.
% figure: mutual knn llama3/aya.
\begin{figure}[t]
  \centering

  \includegraphics[width=0.48\linewidth]{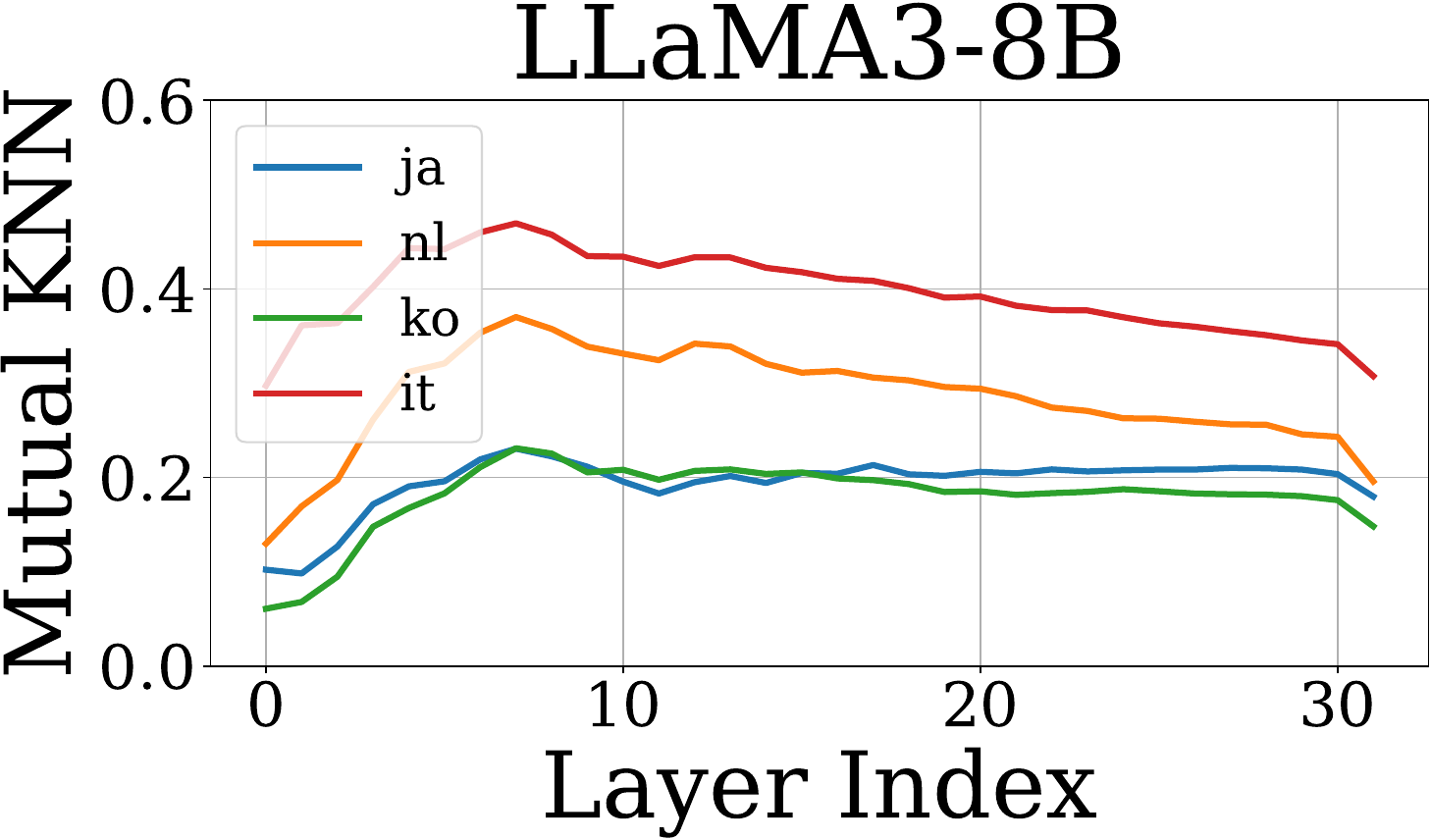}
  \includegraphics[width=0.48\linewidth]{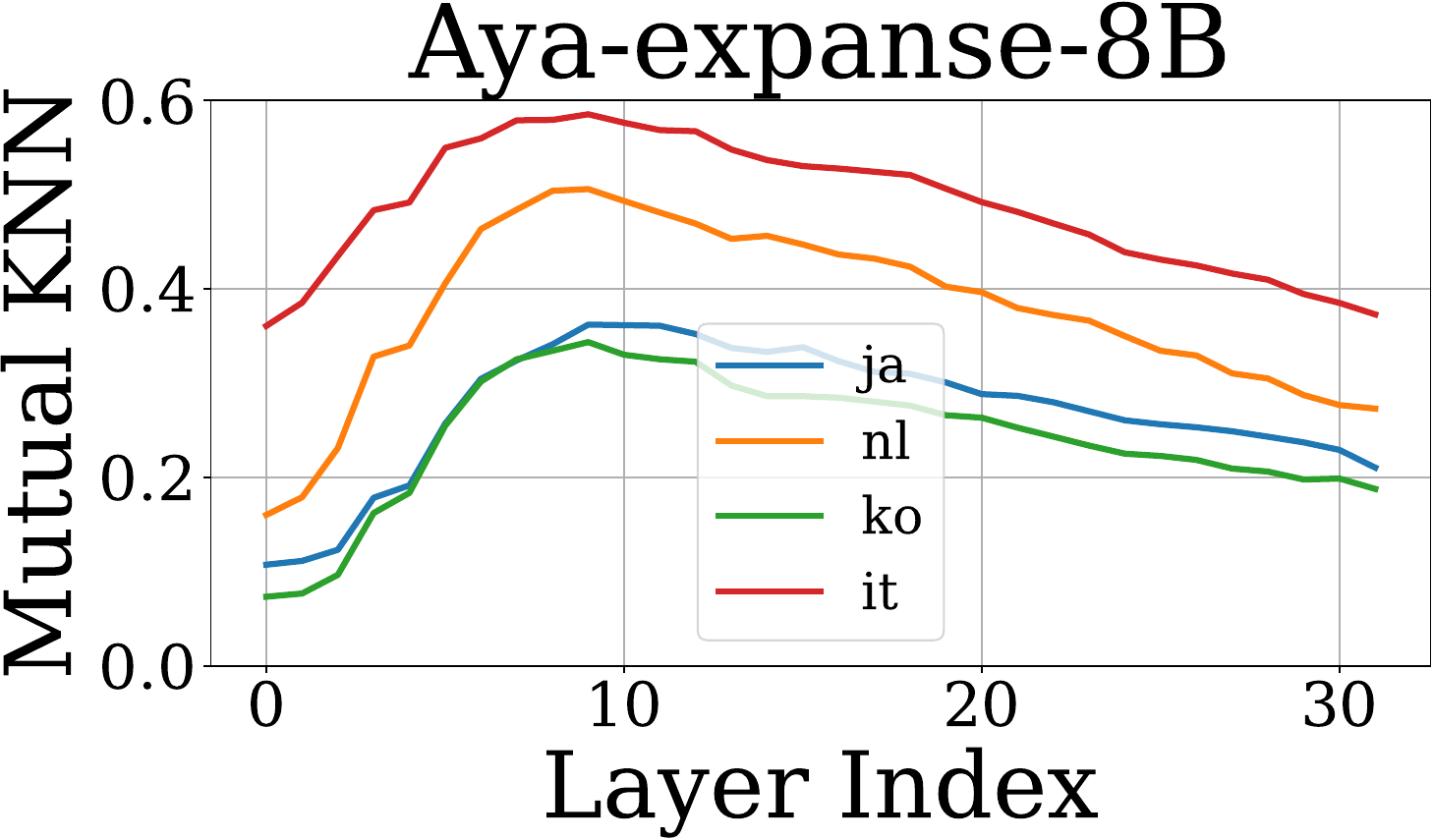}

  \caption{\textbf{Kernel-based similarity between language latent spaces across layers (LLaMA3-8B and Aya expanse-8B, k=5).}
  English-\textcolor{red}{Italian}, English-\textcolor{orange}{Dutch}, English-\textcolor{blue}{Japanese}, English-\textcolor[HTML]{228B22}{Korean} pairs, respectively.
  }
  \label{fig:mutual_knn}
\end{figure}

Fig.~\ref{fig:mutual_knn} shows that latent space similarity between English and other languages peaks in the middle layers, suggesting shared, language-agnostic semantic processing. As noted in Appendix~\ref{sec:appendixB.1}, this pattern holds across models.

Additionally, in Appendix~\ref{sec:appendix: Similarity of Internal Representations}, we measured the similarity of hidden states and activation patterns for parallel sentences across languages, as well as the linear separability of representations for parallel and non-parallel sentences, further supporting the existence of this processing.

\section{Identifying Transfer Neurons}
\label{sec:methodology for identifying transfer neurons}

\subsection{Hypothesis: Specific Activations and Value Vectors Facilitate a Parallel Shift of Representations to the Target Latent Space}
Based on the Eqs.~\ref{eq:mlp} and \ref{eq:weighted_sum}, we hypothesize that certain activations and their corresponding value vectors $\alpha^{l}_i \bm{v}_i^{l}$ are responsible for shifting internal representations between language-specific latent spaces and the shared semantic latent space, as described in Fig.~\ref{fig:figure1}(b). These activation units are what we define as the \textit{Transfer Neurons}.

\subsection{Preparation: Two Types of Transfer Neurons}
\label{sec:two types of tn}
As discussed in \S\ref{sec:introduction}, we consider two types of neurons in the context of transfer neurons:
\begin{enumerate}
    \item Neurons that transfer input representations from the language-specific latent space to the shared semantic latent space, located in the initial layers (\textbf{Type-1 Transfer Neurons}).
    \item Neurons that transfer reasoned representations from the shared semantic latent space to the language-specific latent space for output generation, located in the final layers (\textbf{Type-2 Transfer Neurons}).
\end{enumerate}
Based on the observation that the difference in representation similarity between parallel and non-parallel sentence pairs widens in specific layers (Fig.~\ref{fig:appendix:hs_sim_all_models} in Appendix~\ref{sec:appendix: similarity of hs and act patterns}), we target layers 1–20 for detecting Type-1 neurons and layers 21–32 (up to the final layer) for detecting Type-2 neurons\footnote{All LLMs adopted in this study have 32 decoder layers.}.

\subsection{Scoring the Candidate Neurons}
\label{sec:scoring the candidate neurons}
Our goal is to assign higher scores to candidate neurons whose activations and value vectors move the representation closer to the desired latent space, when the model is given an input sentence in the target language.
\paragraph{Centroids Estimation for Latent Spaces.}
\label{centroids estimation}
We begin by estimating the centroids of each representational latent space, which allows us to compute the distances to these latent spaces.
Let $\bm{h}_{\mathrm{L2}, k}^l$ denote the hidden state from the $l$-th layer corresponding to the $k$-th sample sentence in the particular non-English language “L2”. The centroids of the respective latent spaces are estimated as follows:
\begin{equation}
\bm{C}^l_{\mathrm{L2}} = \frac{1}{n}\sum_{k=1}^n \bm{h}_{\mathrm{L2}, k}^l
\label{eq:centroids_lang_specific_subspace}
\end{equation}
\begin{equation}
\bm{C}^l_{\mathrm{shared}} = \frac{1}{n}\sum_{k=1}^n \mathrm{mean}(\bm{h}_{\mathrm{en}, k}^l, \bm{h}_{\mathrm{L2}, k}^l)
\label{eq:centroids_shared_subspace}
\end{equation}
where $\bm{C}^l_{\mathrm{L2}}$ denotes the centroid of the L2-specific latent space in the $l$-th layer, whereas $\bm{C}^l_{\mathrm{shared}}$ is the centroid of the shared semantic latent space in the $l$-th layer, and $n$ is the total number of sample sentences. To compute Eq.~\ref{eq:centroids_shared_subspace}, we use parallel sentence pairs from an English–L2 pair. This is because, as mentioned in \S\ref{sec:The Dynamics of Spatial Transition}, we assume that the English latent space serves as the shared semantic latent space\footnote{The rationale for using the centroid for every candidate layer is explained in Appendix~\ref{sec:appendix:moving trajectory for english subspace}.}.

% figure: deactivating Type-1 TN.
\begin{figure*}[t]
  \centering
  \includegraphics[width=0.24\linewidth]{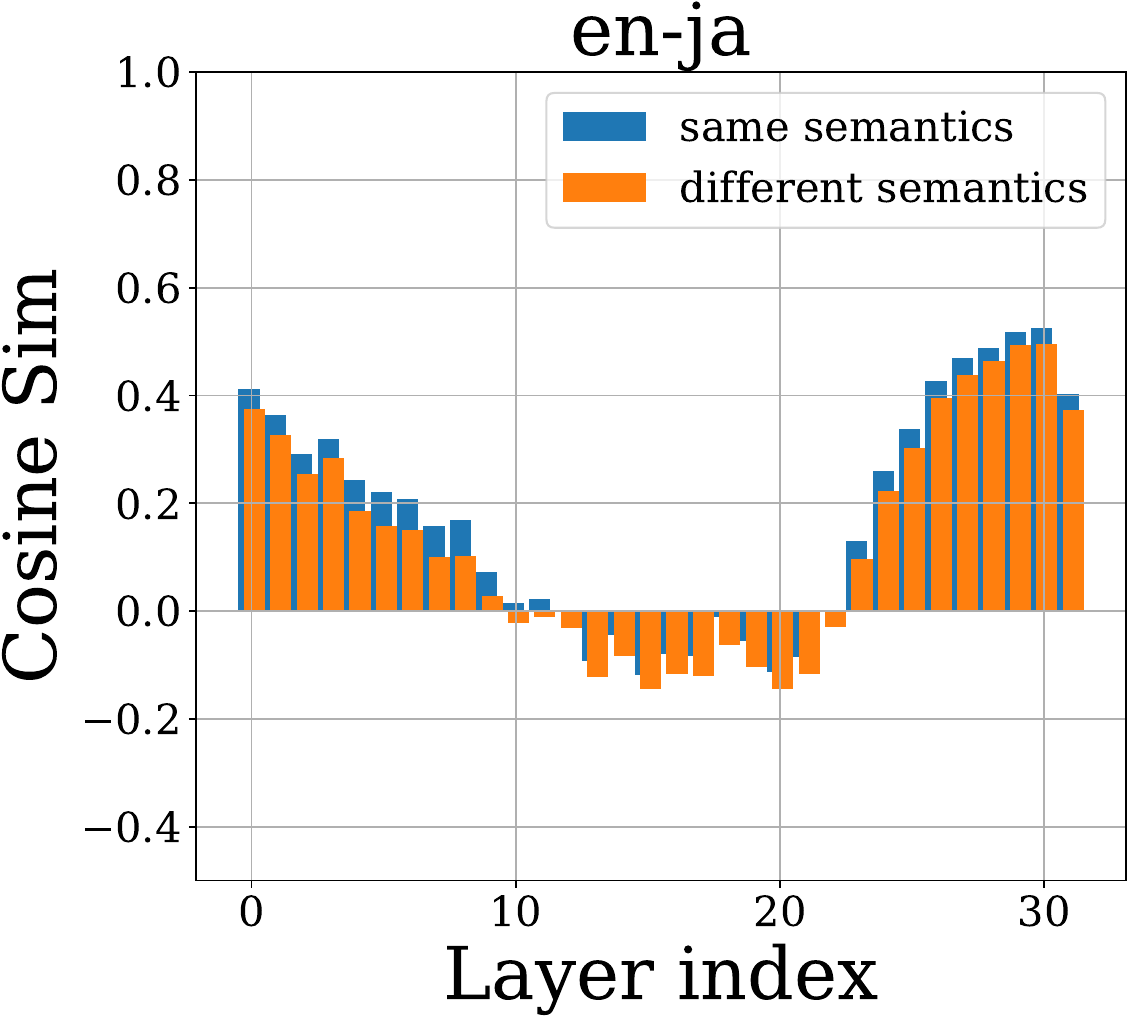}
  \includegraphics[width=0.24\linewidth]{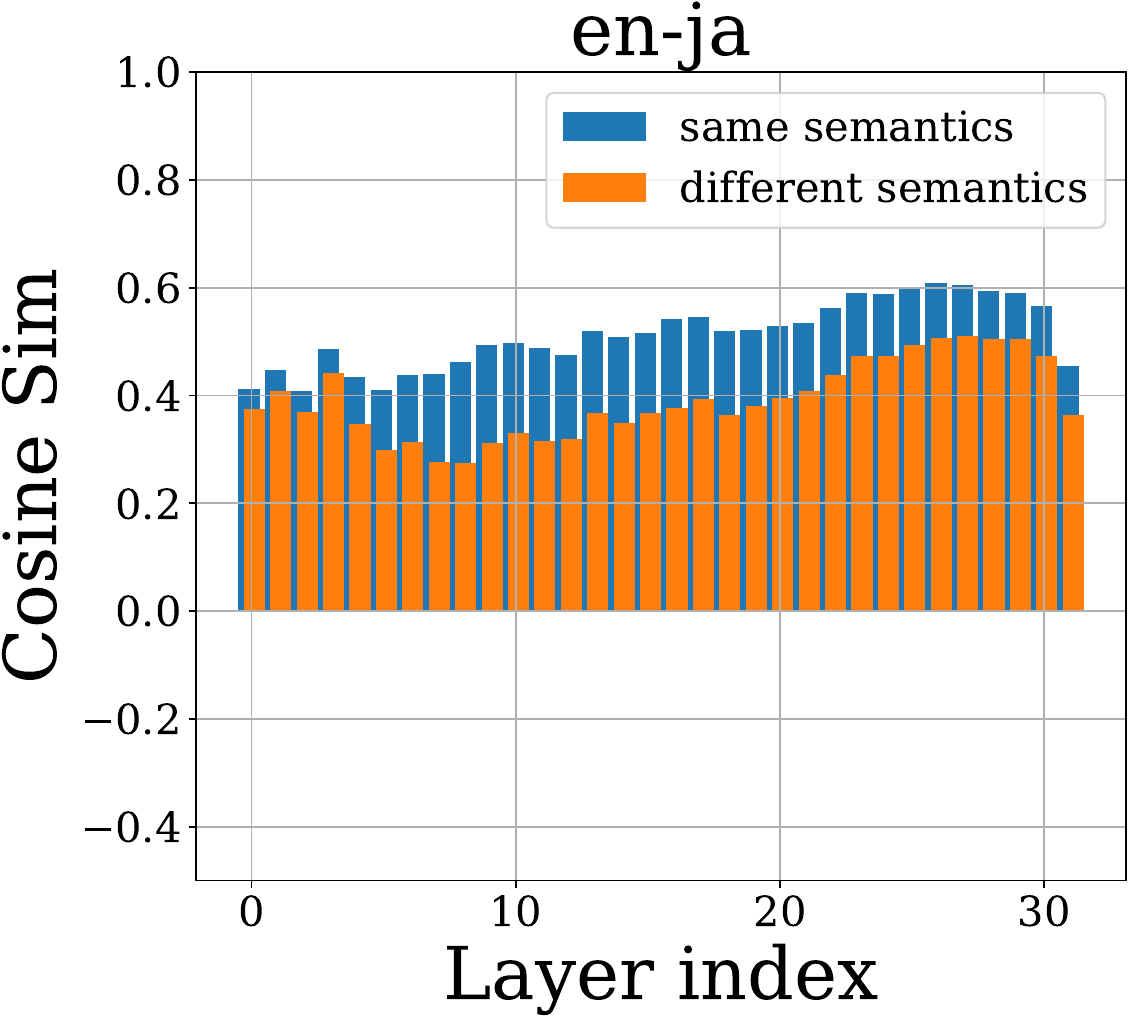}
  \includegraphics[width=0.24\linewidth]{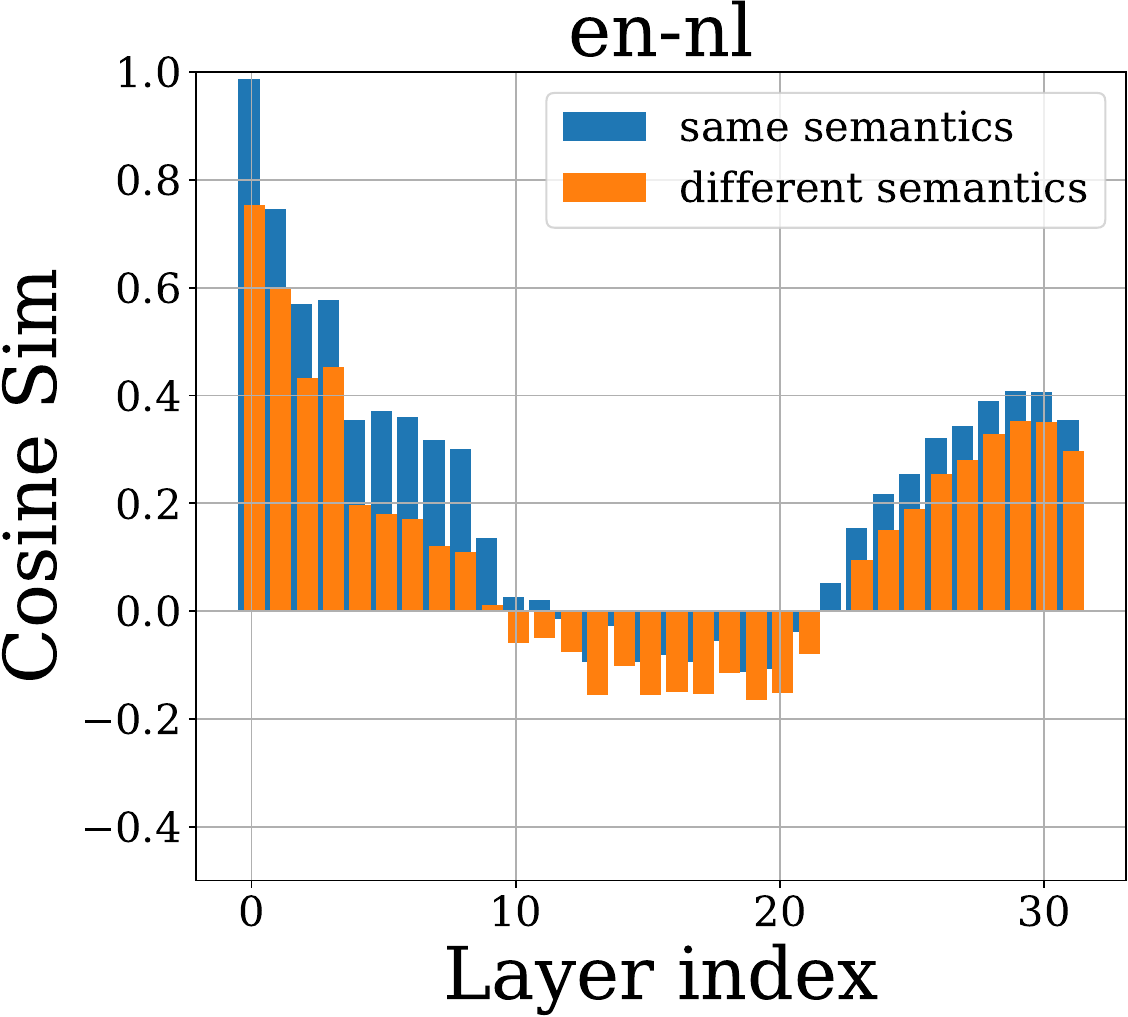}
  \includegraphics[width=0.24\linewidth]{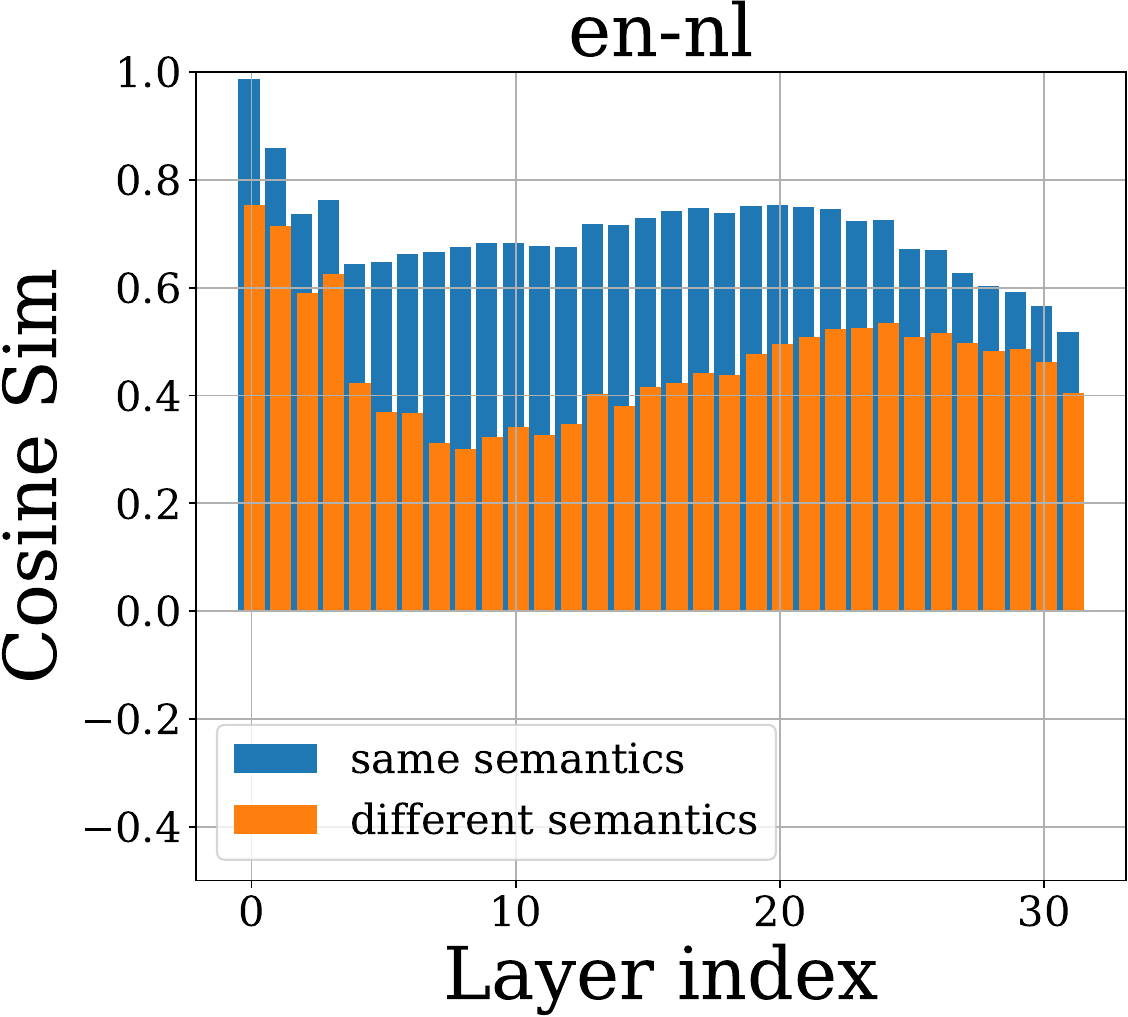}

  \includegraphics[width=0.24\linewidth]{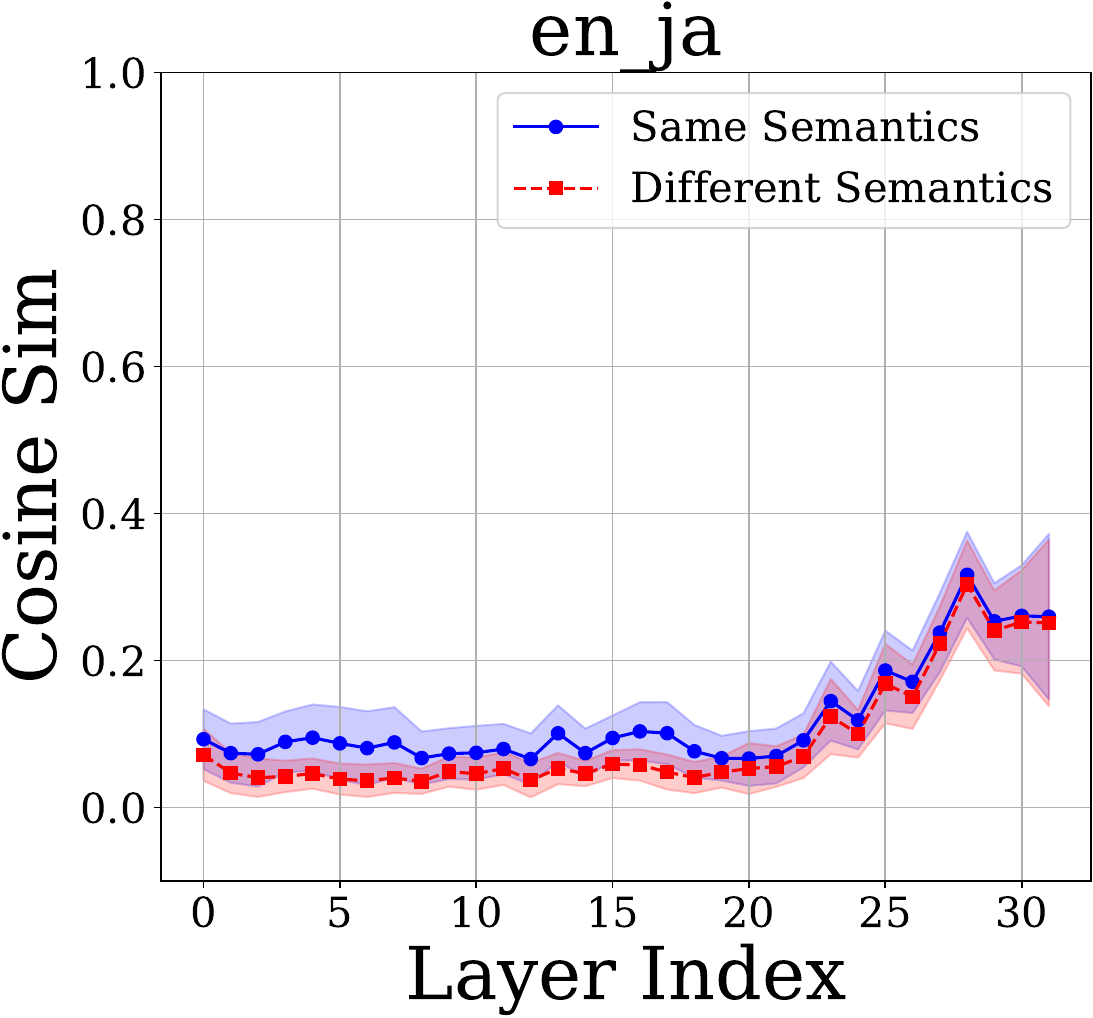}
  \includegraphics[width=0.24\linewidth]{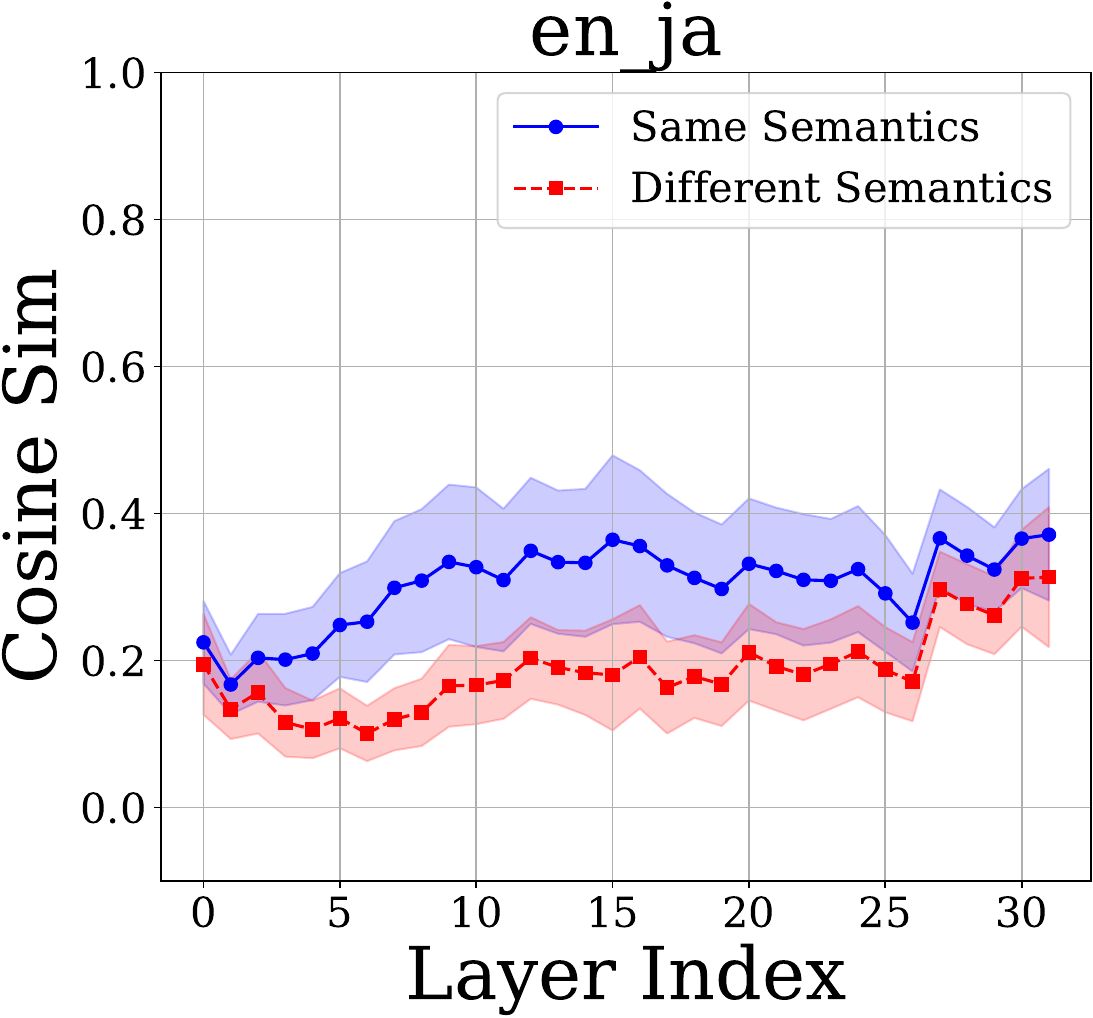}
  \includegraphics[width=0.24\linewidth]{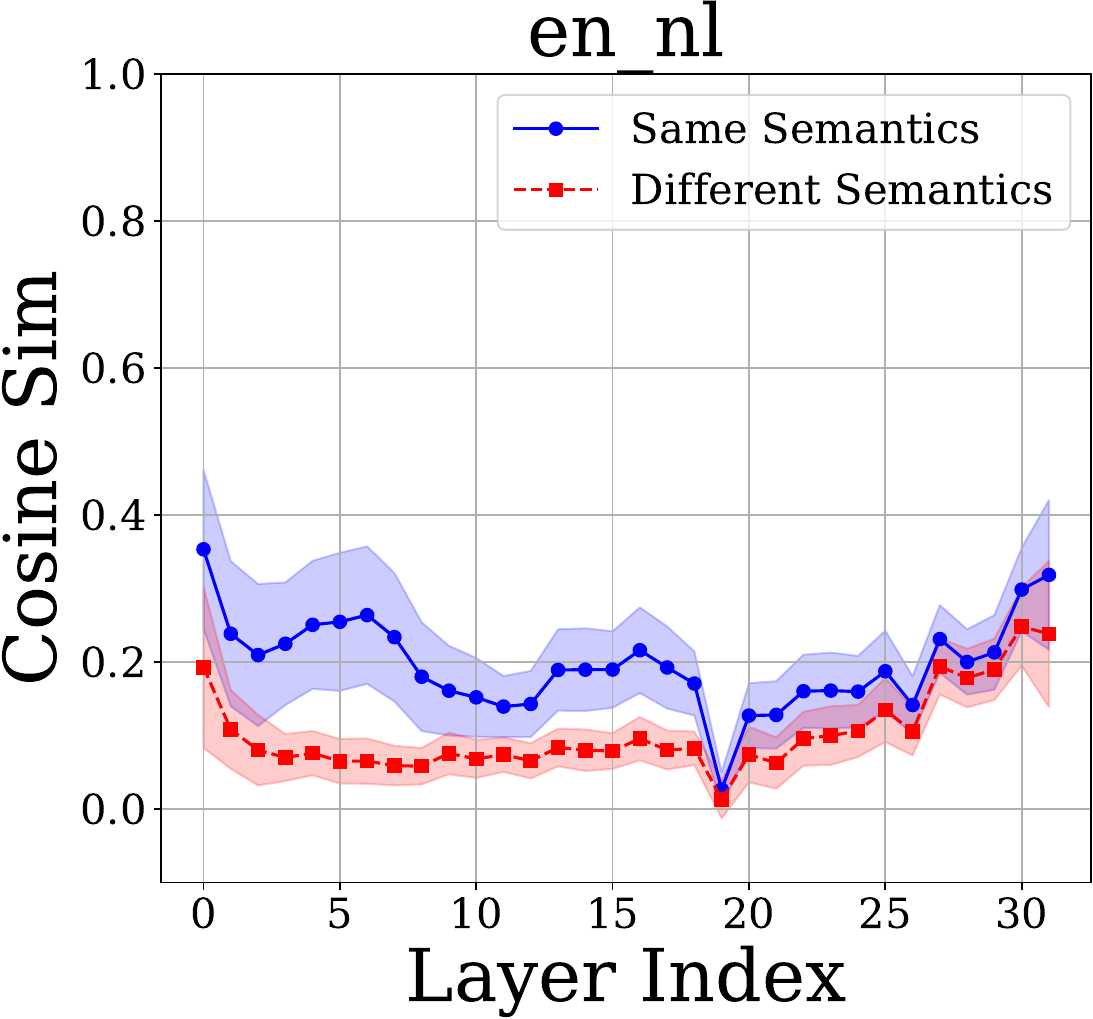}
  \includegraphics[width=0.24\linewidth]{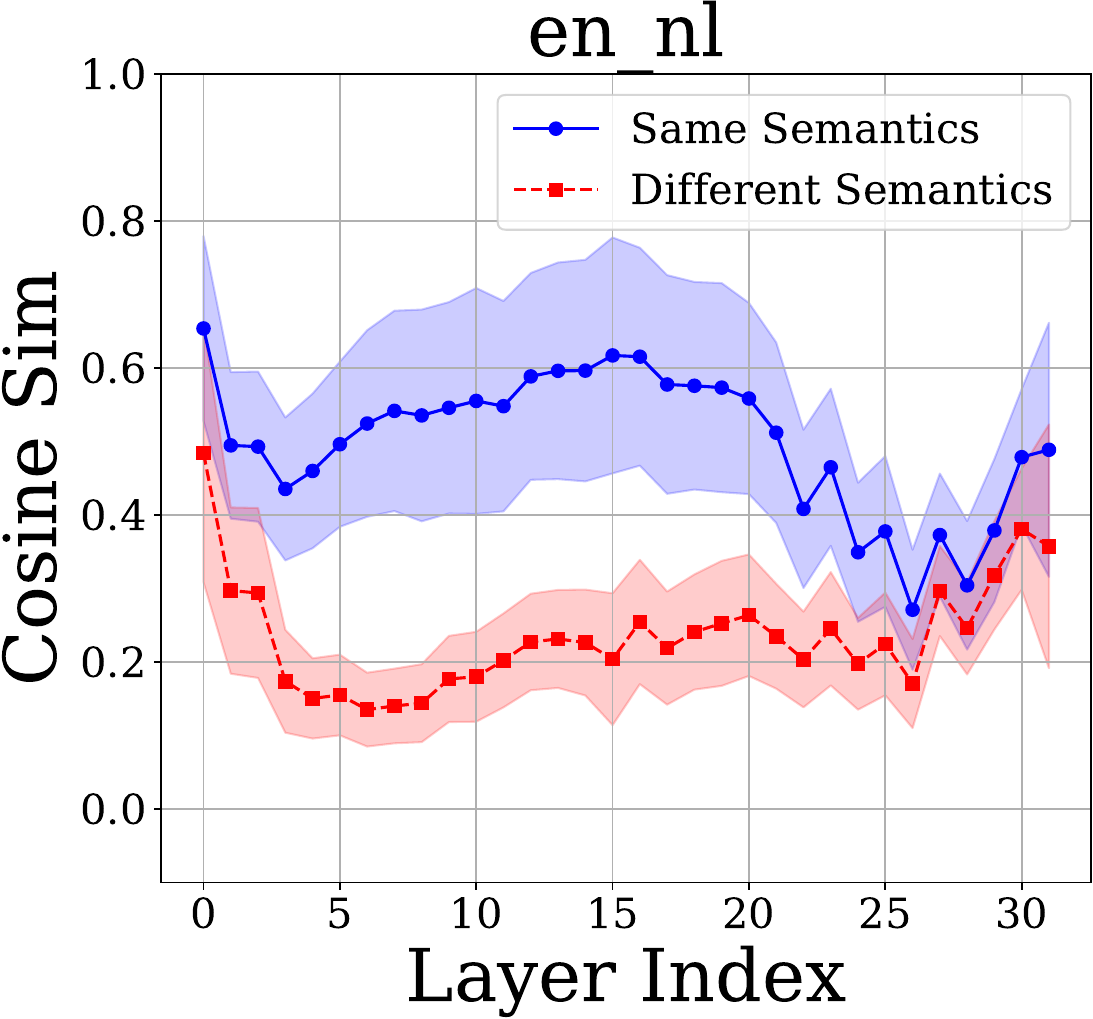}

  \begin{minipage}{0.24\linewidth}\centering en-ja\end{minipage}
  \begin{minipage}{0.24\linewidth}\centering en-ja (baseline)\end{minipage}
  \begin{minipage}{0.24\linewidth}\centering en-nl\end{minipage}
  \begin{minipage}{0.24\linewidth}\centering en-nl (baseline)\end{minipage}

  \caption{\textbf{Similarity of hidden states and activation patterns across layers while deactivating top-1k Type-1 Transfer Neurons (LLaMA3-8B).}
  The upper figures present the similarity of hidden states, and the lower ones present the similarity of activation values.}
  \label{fig:hs_and_act_sim_while_deactivating_llama_ja_nl}
\end{figure*}

\paragraph{Scoring Methodology.}
We score each candidate neuron in the target layers with the following computations:
\begin{equation}
L^l_k = \mathrm{dist}\left ( (\bm{h}^{l-1}_k + \bm{A}^l_k), \bm{C}^l \right )
\label{eq:layer score shared}
\end{equation}
\begin{equation}
N^l_{i,k} = \mathrm{dist}\left( (\bm{h}^{l-1}_k + \bm{A}^l_k + \alpha^l_{i,k} \bm{v}^l_{i,k}), \bm{C}^l \right)
\label{eq:neuron score}
\end{equation}
\begin{equation}
\mathrm{Score}^l_i = \frac{1}{n}\sum^n_{k=1} (N^l_{i,k} - L^l_k)
\label{eq:transfer score}
\end{equation}
where $\bm{h}^{l-1}_k$ is the hidden state from the previous layer, and $\bm{A}^l_k$ is the output of the self-attention in the $l$-th (current) layer. Here, $k$ refers to the index of the sample sentence. Therefore, $(\bm{h}^{l-1}_k + \bm{A}^l_k)$ represents the hidden state at the $l$-th layer immediately before the MLP module for the $k$-th input sample. The \textit{dist} function measures how close two vectors are\footnote{We adopt Cosine similarity as a \textit{dist} function in this study; however, we also experimented with Euclidean distance, which yielded similar results.}. $L^l_k$ denotes the layer score, which indicates how close the hidden state before the MLP is to the $l$-th layer centroid of the target latent space. $N^l_{i,k}$ denotes the neuron score, which expresses how effectively the neuron and its corresponding value vector, on their own, bring the representation closer to the centroid of the target latent space, in comparison to $L^l_k$.

If the score of the $i$-th neuron in the $l$-th layer, $\mathrm{Score}^l_i$ > 0, it indicates that the neuron and value vector brings representations closer to the centroid of the target latent space for most of the samples. Conversely, if $\mathrm{Score}^l_i$ < 0, it suggests that the neuron pushes the hidden states further apart from the centroid across various samples. We set $\bm{C}^l$ to $\bm{C}^l_{\mathrm{shared}}$ for detecting Type-1 neurons, and to $\bm{C}^l_{\mathrm{L2}}$ for detecting Type-2 neurons. The larger the $\mathrm{Score}^l_i$ is, the more effectively the neuron and its corresponding value vector bring the representations closer to the target latent space. Finally, we sort all candidate neurons in descending order based on $\mathrm{Score}^l_i$, and extract the top-$n$ neurons.

\section{Detecting and Controlling Transfer Neurons}
\label{sec:detecting and controlling transfer neurons}

\subsection{Distribution}
\label{sec:distribution of transfer neurons}
\begin{figure}[t]
  \centering

  \includegraphics[width=0.49\linewidth]{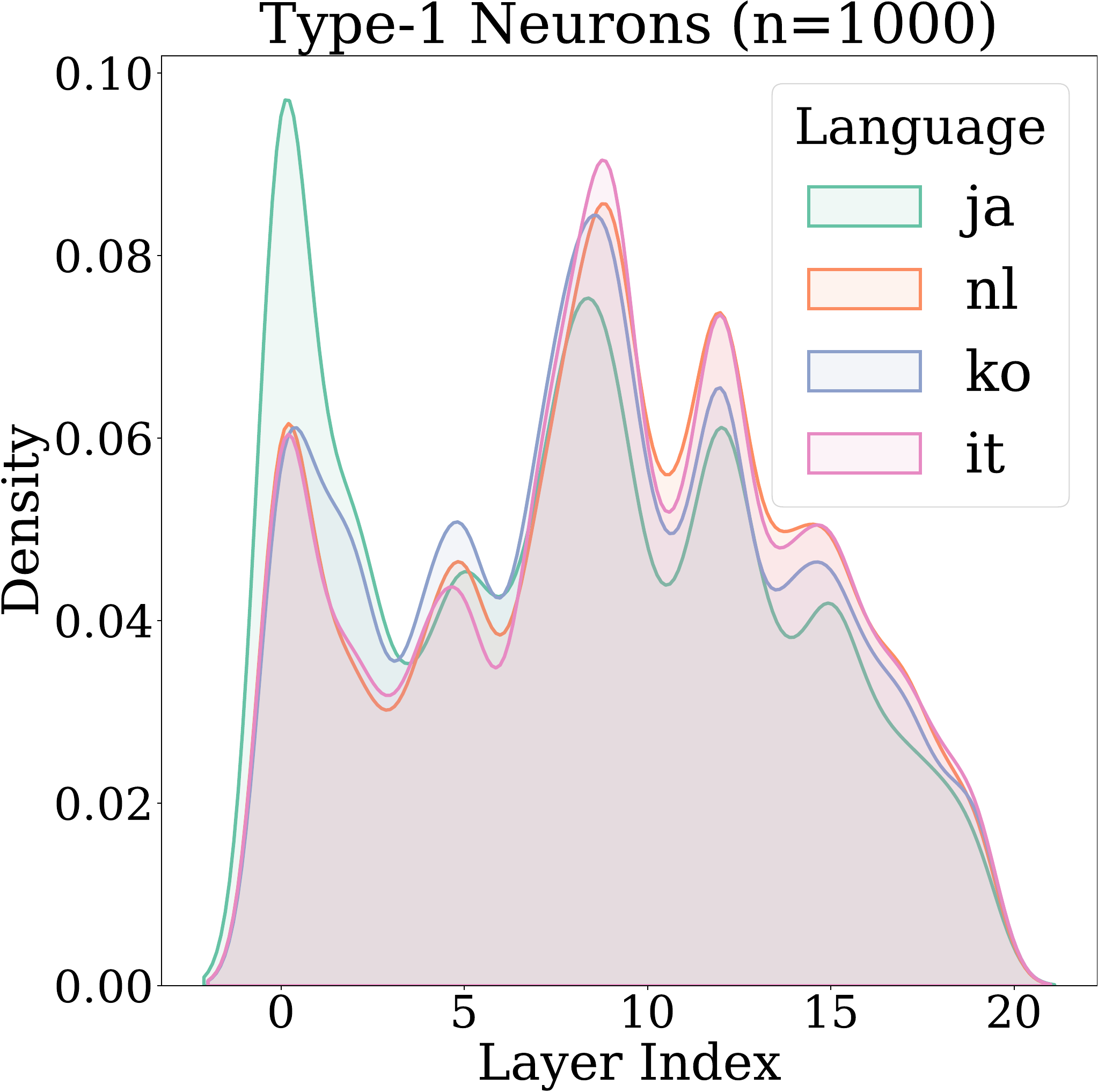}
  \includegraphics[width=0.49\linewidth]{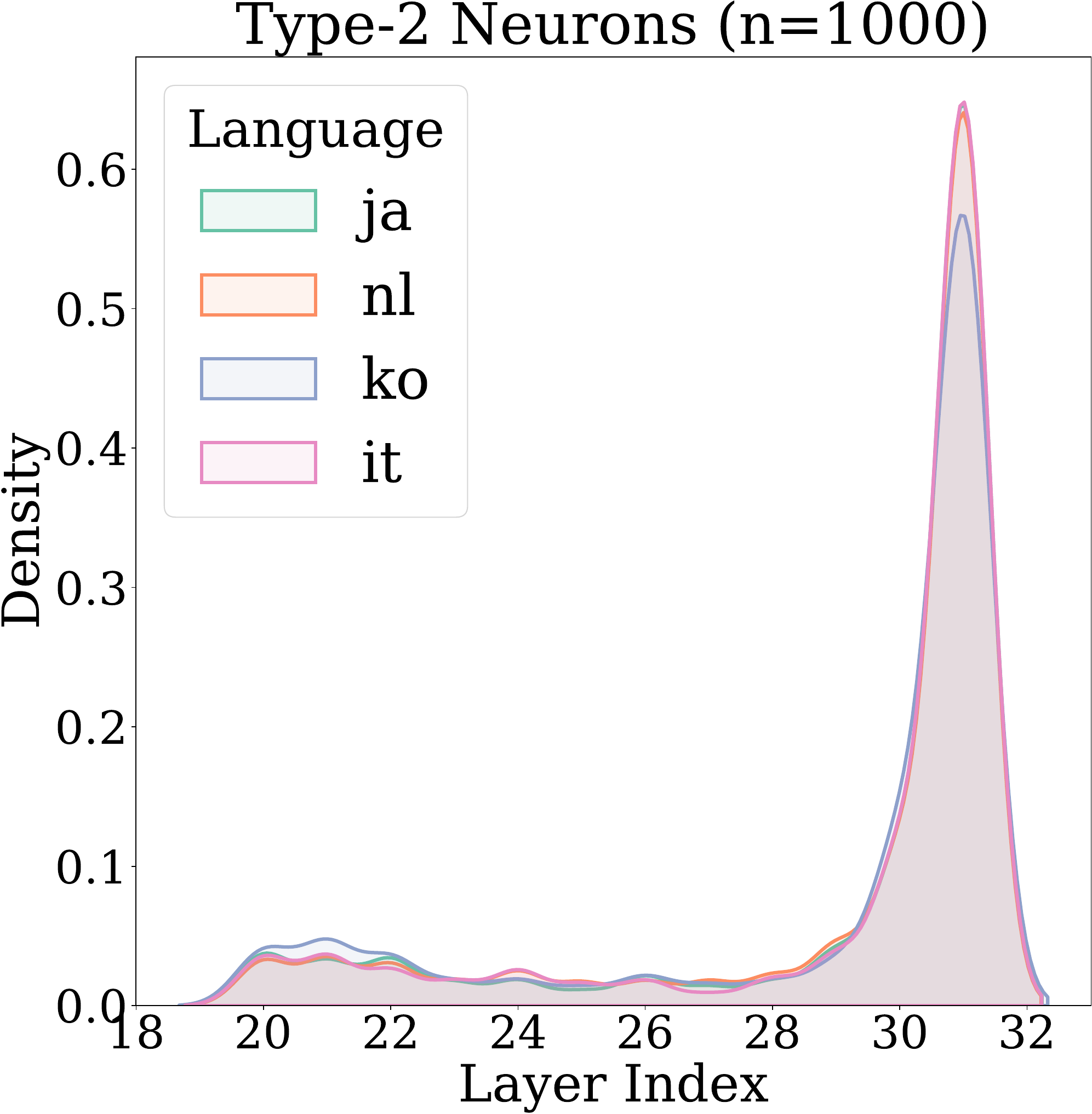}

  \caption{\textbf{Distribution of top-1k Transfer Neurons (LLaMA3-8B).} Left shows Type-1 neurons, whereas right shows Type-2 neurons.
  }
  \label{fig:distribution transfer neurons llama ja}
\end{figure}

Fig.~\ref{fig:distribution transfer neurons llama ja} and Appendix~\ref{sec:appendix:distribution} shows the distributions of two types of transfer neurons across layers. For Type-1 neurons, the first layers and the middle layers contain a noticeable amount. In contrast, Type-2 neurons are predominantly found in the final layer, which aligns with the observation that representations exhibit the largest shift in the final layers, as shown in Appendix~\ref{sec:appendix:visualization language subspaces}. Additionally, these distributions closely resemble those of language-specific neurons, as shown in Appendix~\ref{sec:appendix:language specific neurons}. These tendencies are consistent across languages and models.

\subsection{Representation Similarity Measurement while Deactivating Transfer Neurons}
\label{sec:Similarity Measurement While Deactivating Transfer Neurons}

\paragraph{Type-1 Neurons Facilitate the Mapping of Representations to the Shared Semantic Latent Space.}
Fig.~\ref{fig:hs_and_act_sim_while_deactivating_llama_ja_nl} presents the similarity of hidden states and MLP activation patterns for parallel and non-parallel English–L2 sentence pairs. The measurements are taken while deactivating the top-1k Type-1 neurons, which account for only \textbf{0.2\%} of the total neuron population\footnote{Here, "deactivating" refers to setting the activation values of the target neurons to zero, thereby nullifying their effect.}. Although only a very small proportion of neurons were deactivated, we observe a sharp reduction in the difference in similarity between parallel and non-parallel sentence pairs for both activation patterns and hidden states. By contrast, the baseline — defined as deactivating 1k randomly sampled neurons from the same layers as the Type-1 neurons — has almost no effect on the similarity compared to the original state (Fig.~\ref{fig:appendix:hs_sim_all_models} and~\ref{fig:appendix:act_sim_all_models} in Appendix~\ref{sec:appendix: similarity of hs and act patterns}). This tendency is consistent across other language pairs and models.

% correlation ratio for language-specificity.
\begin{table*}[t]
    \centering
    \small
    \setlength{\tabcolsep}{4pt}
    \renewcommand{\arraystretch}{1.2}
    \setContinuousHeatMapMax{0.3}
    \setContinuousHeatMapMin{0.0}
    
    \begin{tabularx}{0.9\linewidth}{l c c c c l c c c c}
        \hline
                {\textbf{Top-1000}} & \textbf{ja} & \textbf{nl} & \textbf{ko} & \textbf{it} & {\textbf{Top-100}} & \textbf{ja} & \textbf{nl} & \textbf{ko} & \textbf{it} \\
        \hline
        \textbf{Type-1 TN} & 
        \begin{tabular}{V} 0.16 \end{tabular} & 
        \begin{tabular}{V} 0.03 \end{tabular} & 
        \begin{tabular}{V} 0.05 \end{tabular} & 
        \begin{tabular}{V} 0.03 \end{tabular} & 
        \textbf{Type-1 TN} & 
        \begin{tabular}{V} 0.16 \end{tabular} & 
        \begin{tabular}{V} 0.02 \end{tabular} & 
        \begin{tabular}{V} 0.05 \end{tabular} & 
        \begin{tabular}{V} 0.03 \end{tabular} \\
        \textbf{Type-2 TN} & 
        \begin{tabular}{V} 0.25 \end{tabular} & 
        \begin{tabular}{V} 0.22 \end{tabular} & 
        \begin{tabular}{V} 0.17 \end{tabular} & 
        \begin{tabular}{V} 0.16 \end{tabular} & 
        \textbf{Type-2 TN} & 
        \begin{tabular}{V} 0.40 \end{tabular} & 
        \begin{tabular}{V} 0.27 \end{tabular} & 
        \begin{tabular}{V} 0.33 \end{tabular} & 
        \begin{tabular}{V} 0.19 \end{tabular} \\
        \hline
    \end{tabularx}

    \caption{\textbf{Correlation ratio of top-100 and top-1k Transfer Neurons for language specificity (LLaMA3-8B).}
    Typically, a correlation ratio above \textbf{0.1} suggests a correlation, above \textbf{0.25} suggests a moderately strong correlation, and above \textbf{0.5} indicates a strong correlation.
    }
    \label{table:corr ratio}
\end{table*}

These results indicate that the models deactivated Type-1 neurons are unable to map input representations to the correct positions in the shared semantic latent space in a way that reflects sentence meaning, since the similarity remains nearly unchanged regardless of whether the pairs are parallel or non-parallel. This indicate that Type-1 neurons we detected play a critical role in this mapping process.

% figure: distribution of TN (both Type-1 and 2, llama3).

Also, we surmise that the relatively remaining difference of similarity between parallel and non-parallel sentence pairs in the en-nl (Dutch) and en-it (Italian) settings across models (Fig.~\ref{fig:hs_and_act_sim_while_deactivating_llama_ja_nl} and Appendix~\ref{sec:sim_deactivating_Type-1_neurons_appendix}) — even after the deactivation of the top-1k Type-1 neurons — is due to the linguistic and spatial proximity of these languages to English. That is, even if we disturb representations from shifting into the shared semantic latent space by deactivating Type-1 neurons, they are already very close to the english-centric shared semantic latent space (see \S\ref{Visualizing the Hidden language Subspace with PCA} and Appendix~\ref{sec:appendix:visualization language subspaces}).

Appendix~\ref{sec:appendix:kernel-based sim between spaces while deactivating Type-1} shows that deactivating Type-1 neurons significantly reduces kernel-based similarity, underscoring their role in aligning representations to a shared semantic latent space.

Ablation studies, including experiments with different numbers of deactivated neurons, are provided in Appendix~\ref{sec:sim_deactivating_Type-1_neurons_appendix}.

The results of deactivating Type-2 neurons are presented in Appendix~\ref{sec:sim_deactivating_Type-2_neurons_appendix}, including evidence of a significant inhibition in latent space transitions when the neurons are deactivated.

\section{The Nature of Transfer Neurons}
\label{sec:nature of transfer neurons}

\subsection{Language and Language-Family Specificity}
\label{sec:language-specific nature of transfer neurons}

% tab: language-family specificity
\begin{table}[t]
    \small
    \centering
    \renewcommand{\arraystretch}{1.2}
    
    \begin{tabular}{l c c c c}
        \hline
        \textbf{Top-1000} & \textbf{ja-nl} & \textbf{ja-ko} & \textbf{nl-it} & \textbf{ja-it} \\
        \hline
        \textbf{Type-1 TN} & 0.39 & 0.51 & 0.75 & 0.41 \\
        \textbf{Type-2 TN} & 0.23 & 0.37 & 0.51 & 0.26 \\
        \hline
    \end{tabular}

    \caption{\textbf{Jaccard index for top-1k Transfer Neurons across language pairs. (LLaMA3-8B).} A score closer to 1 indicates a greater overlap between the neurons of each language pair.}
    \label{table:jaccard_index_llama3}
\end{table}

\paragraph{Language Specificity.}
To investigate the language specificity of transfer neurons, we measure the correlation ratio between neuron activations and sentence labels\footnote{Detailed explanation about the correlation ratio is given in Appendix~\ref{sec:appendix:corr_ratio}}, assigning \texttt{label1} to sentences in the target language and \texttt{label0} to all others. This allows us to measure the strength of the correlation between the activations of the transfer neurons and inputs in a specific language.

As shown in Tab.~\ref{table:corr ratio}, Type-1 neurons generally do not exhibit correlation, except for those involved in shifting representations from the Japanese latent space to the shared semantic latent space. We surmise that this is because the Japanese-specific latent space in first few layers is highly distant from the English-centric shared latent space (see PCA results in Appendix~\ref{sec:appendix:visualization language subspaces}), which is why certain Type-1 neurons must undergo a considerable shift to align with the shared latent space.

On the other hand, it turns out that \textbf{for Type-2 neurons, the more their activations and corresponding value vectors move representations towards the language-specific latent spaces, the stronger the correlation with the input sentences of the target language}. In other words, the neurons that strongly shift representations from the shared semantic latent space to each language-specific latent space for output generation can be considered language-specific.
This is further supported by the high similarity in the distributions between language-specific neurons and Type-2 neurons, as demonstrated in Appendix~\ref{sec:appendix:distribution} and~\ref{sec:appendix:language specific neurons}.

For both Type-1 and Type-2 neurons, the strength of the correlation ratio score correlates with the linguistic distance from English, suggesting that the greater the distance neurons must shift to the target latent space, the more language-specific those neurons are likely to be (As shown in Tab.~\ref{table:corr ratio}, \S\ref{Visualizing the Hidden language Subspace with PCA}, Appendices~\ref{sec:appendix:visualization language subspaces}, and~\ref{sec:appendix:language specificity}). 

Additionally, when comparing the distributions of the language-specific neurons (detected in Appendix~\ref{sec:appendix:language specific neurons}) with PCA results of language-specific latent spaces (Appendix~\ref{sec:appendix:visualization language subspaces}) and the correlation ratio scores, we observe that \textbf{languages whose original latent spaces are closer to the English-specific latent space in the initial layers (such as Dutch and Italian) tend to exhibit significantly sparser distributions of language-specific neurons in those layers, along with relatively low correlation ratio scores}. We surmise that this is because these languages require fewer language-specific neurons, as they are already closely aligned with the English-centric shared semantic latent space.

These results suggest one of the key roles of language-specific neurons, as identified in recent studies: \textbf{facilitating the movement of internal representations between language-specific latent spaces and a shared semantic latent space}.

The results for other values of $n$ (i.e., top-$n$ neurons) and models can be found in Appendix~\ref{sec:appendix:language specificity}.

\paragraph{Language-Family Specificity.}
Tab.~\ref{table:jaccard_index_llama3} presents the Jaccard Index measurements for transfer neurons specific to each language, providing insight into the extent of overlap among transfer neurons. As shown, language pairs that are linguistically similar tend to exhibit relatively greater overlap, with this effect being particularly pronounced in Type-2 neurons. The results suggest that, from the perspective of the shared semantic latent space, the locations of each language-specific latent space within the hidden state space of the model are likely to be similar, which explains why there are many overlapping neurons used to shift to the target latent space. This is further supported by the visualization of each language latent space (Fig.~\ref{fig:pca llama first-middle-last layers}, Appendix~\ref{sec:appendix:visualization language subspaces}). In Appendix~\ref{sec:appendix:language-family specificity}, we show the results of other models and correlation ratio measurement to further verify language-family specificity of these neurons.

The results of the hypothesis testing conducted to verify the reliability of the correlation scores reported above are provided in Appendix~\ref{sec:appendix:hypothesis testing for correlation metric}.

\subsection{Assessing the Importance of Transfer Neurons in Reasoning}
\label{sec:reasoning}
In this section, we examine the role of Type-1 neurons in enabling reasoning. Specifically, we test whether disrupting the shift of input representations to the shared semantic latent space during inference degrades performance. A notable drop when deactivating these neurons — thereby disrupting the inference framework of the hypothesis in Fig.~\ref{fig:figure1} — would suggest that the model struggles to generate appropriate responses when inputs are misaligned with the shared semantic latent space.

\paragraph{Task and Evaluation Metric.}
We use a simple multilingual knowledge QA task called MKQA \citep{longpre-etal-2021-mkqa}, which consists of 10k knowledge question-answer pairs aligned across 26 languages. Following the original work, we adopt the token-based F1 score as the evaluation metric. 

We conduct the following three experimental setups for the same questions to investigate the influence of Type-1 neurons: (a) Performance without any intervention. (b) Performance with deactivation in the top-1k Type-1 neurons. (c) Performance with deactivation in 1k randomly sampled neurons from the same layers as Type-1 neurons (baseline neurons).
Answer generation was performed under a zero-shot setting.

\begin{figure}[t]
  \centering

  \includegraphics[width=0.49\linewidth]{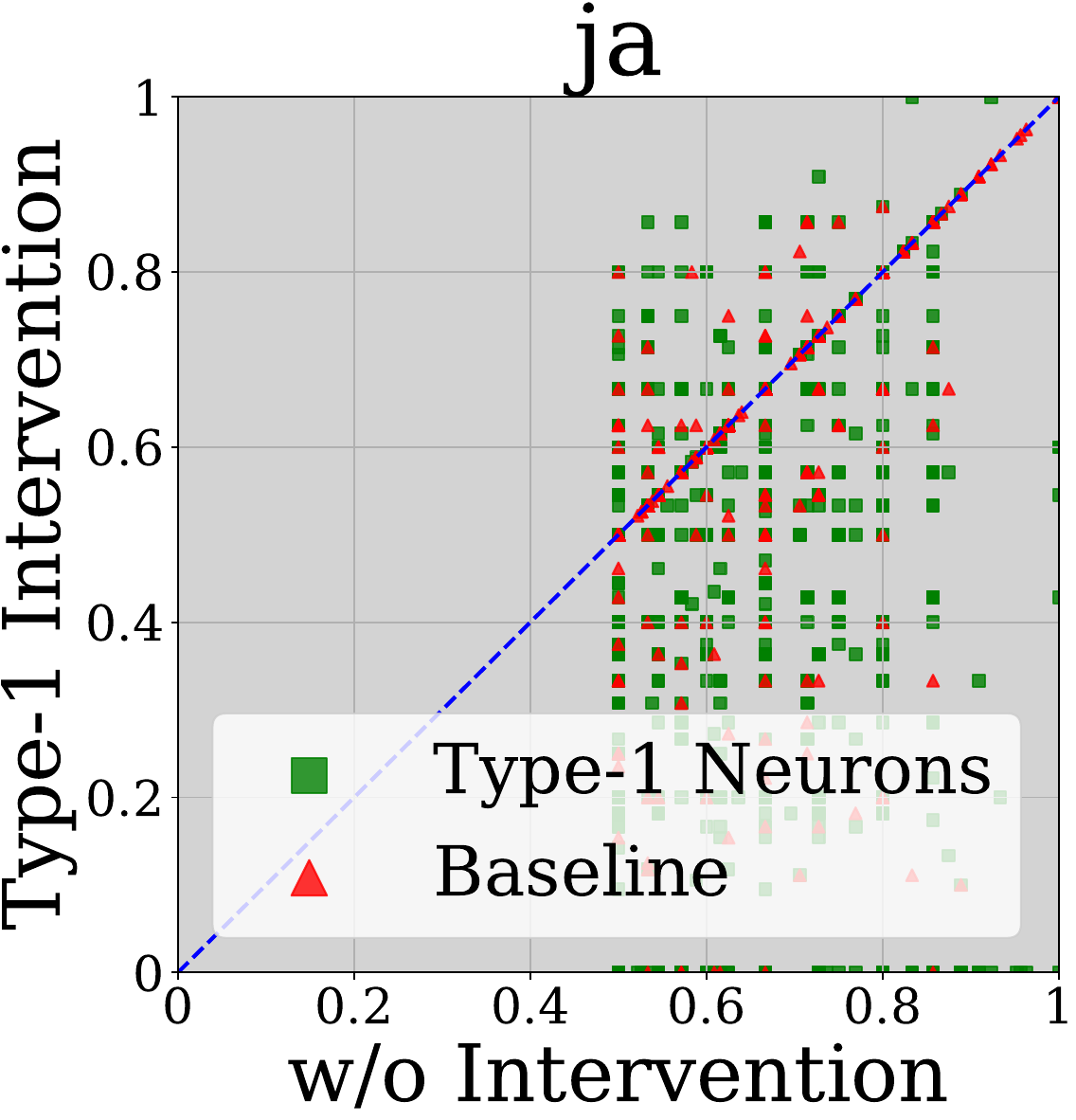}
  \includegraphics[width=0.49\linewidth]{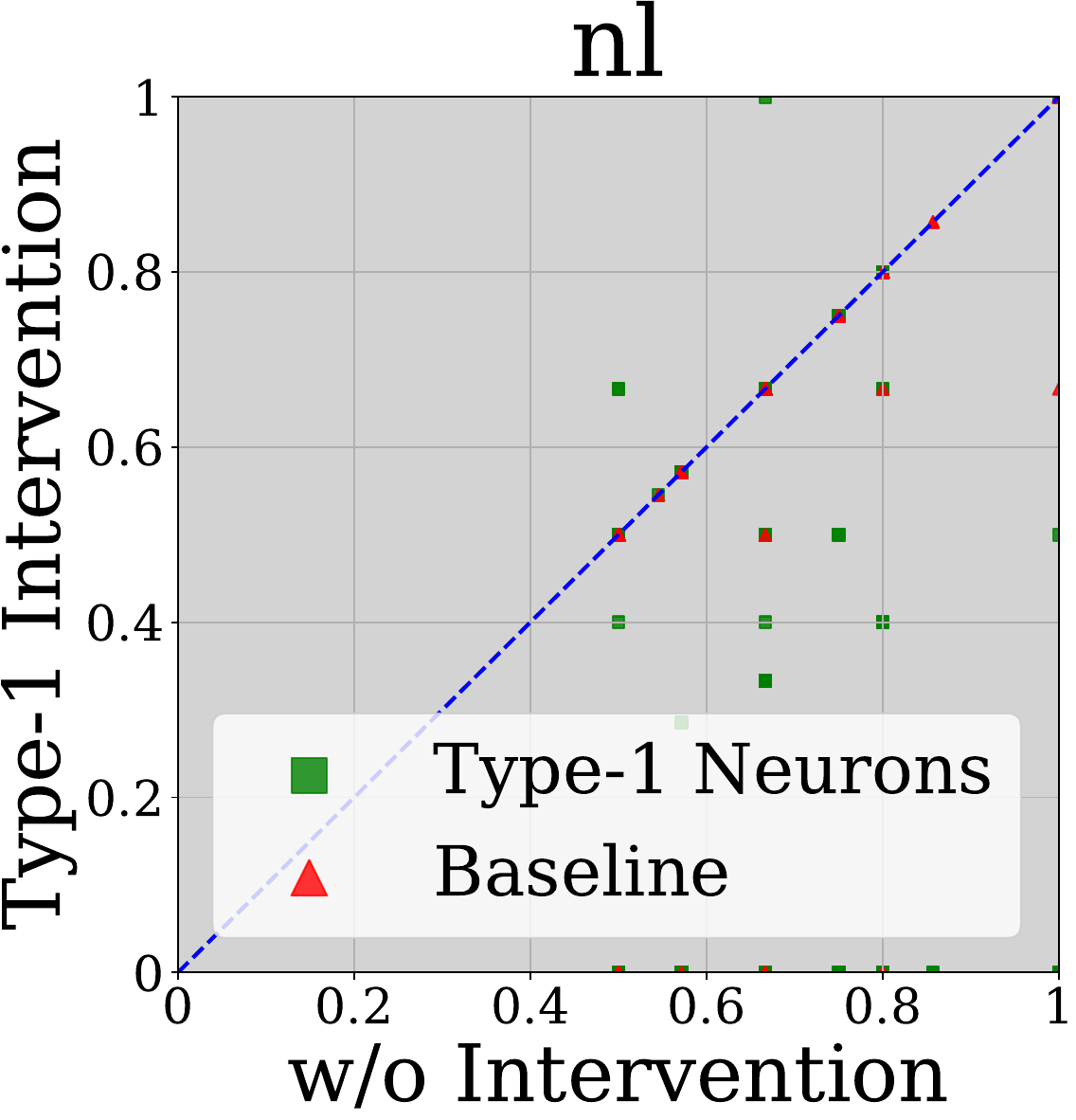}
  \includegraphics[width=0.49\linewidth]{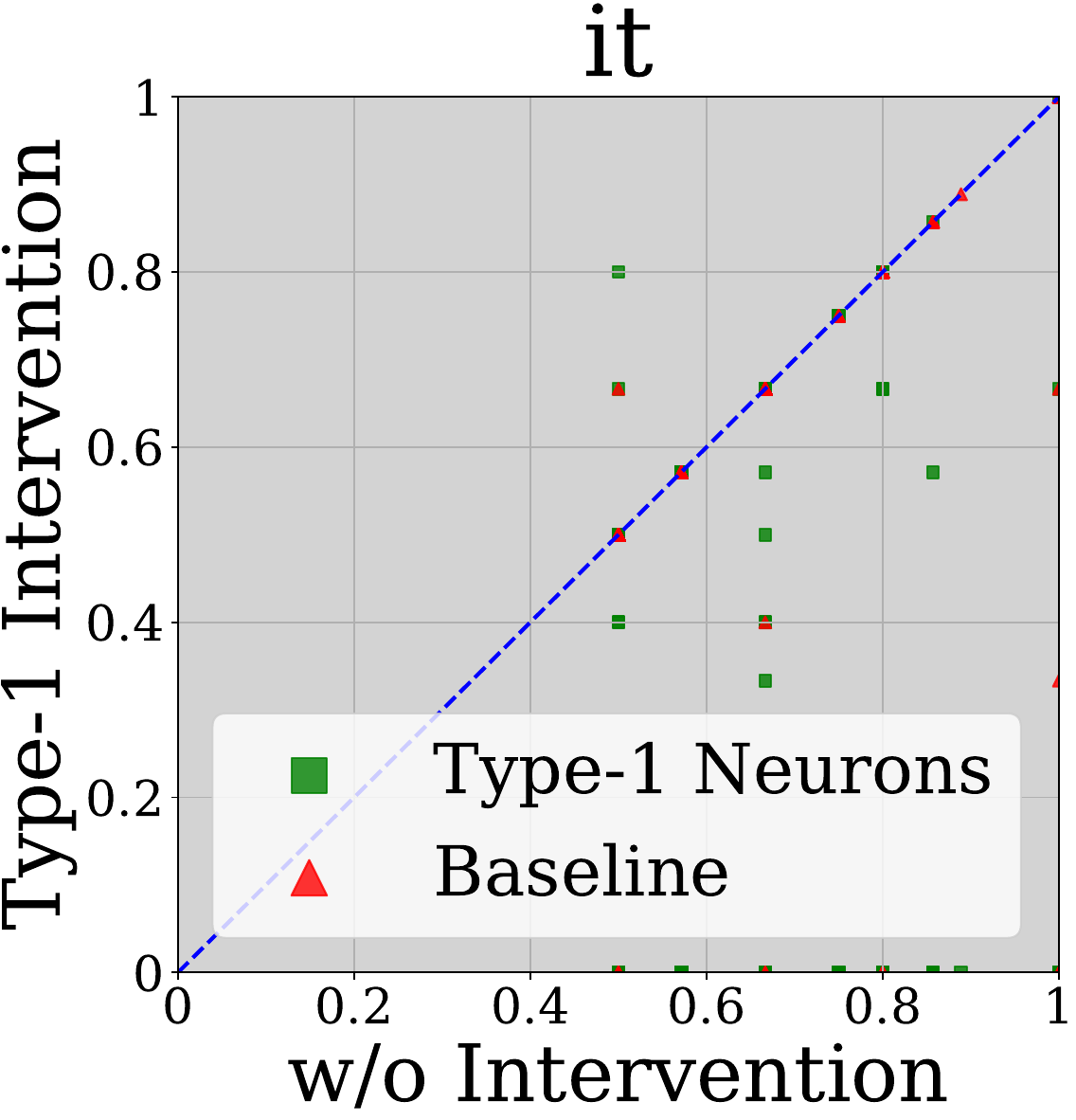}
  \includegraphics[width=0.49\linewidth]{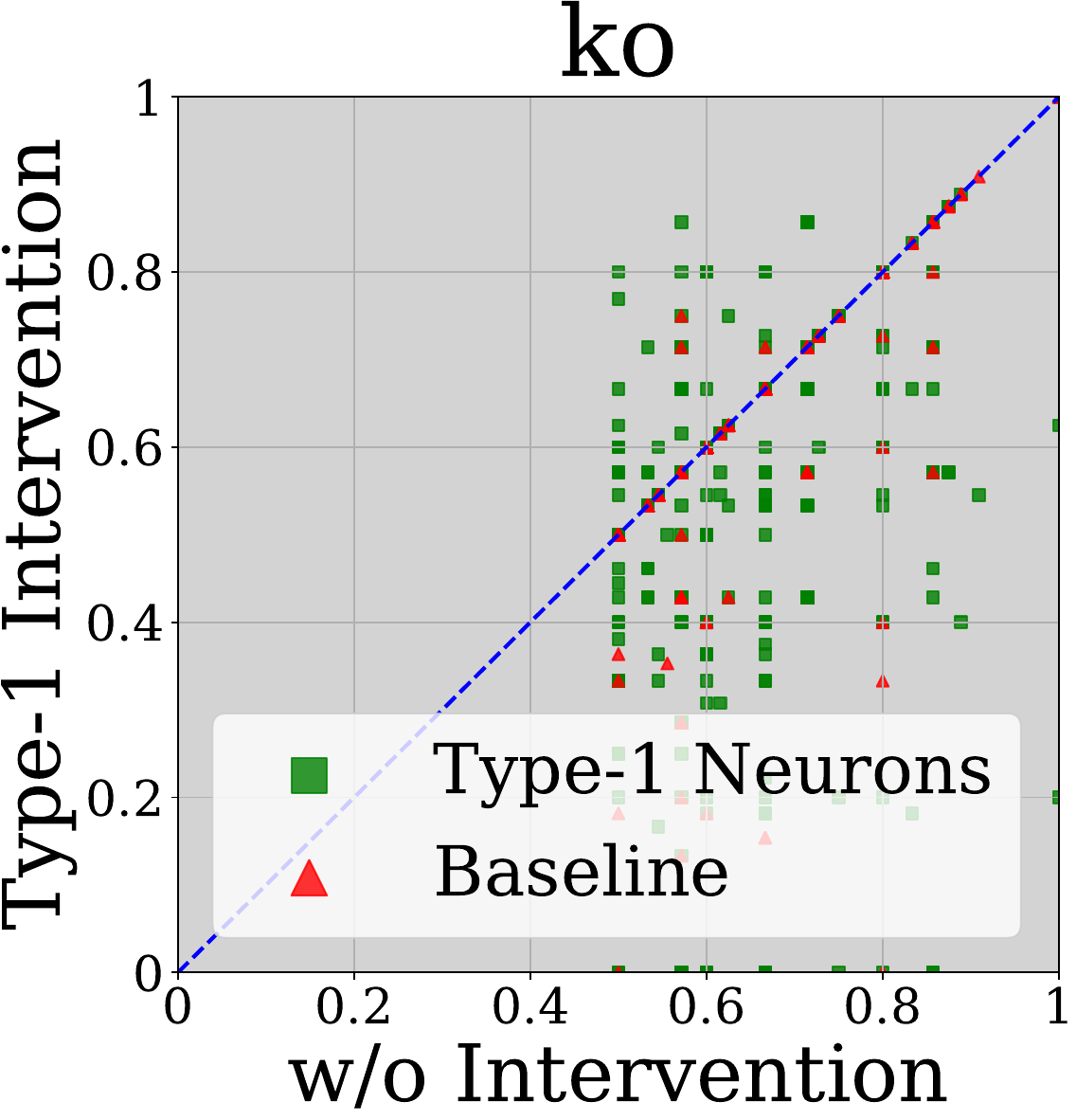}

  \caption{\textbf{Token-based F1 score when deactivating top-1k Type-1 transfer neurons (LLaMA3-8B).}
  \textcolor[HTML]{228B22}{Green} denotes the score with intervention in \textcolor[HTML]{228B22}{the Type-1 neurons}, while \textcolor{red}{red} represents the score with intervention in \textcolor{red}{the randomly sampled neurons} (i.e., baseline). Points below the \textcolor{blue}{$y=x$ line} indicate a decrease in the score due to the intervention, whereas points above the $y=x$ line indicate an increase in the score. Points on the $y=x$ line denote no change in the score before and after the intervention.
  }
  \label{fig:mkqa llama3 above 0.5}
\end{figure}
% tab: mkqa f1 socre changes above 0.5
\begin{table}[t]
    \small
    \centering
    \renewcommand{\arraystretch}{1.2}
    \begin{tabularx}{\linewidth}{l *{4}{>{\centering\arraybackslash}X}}
        \hline
        \textbf{MKQA (F1)} & \textbf{ja} & \textbf{nl} & \textbf{ko} & \textbf{it} \\
        \hline
        \textbf{(a) w/o Intervention} & 0.66 & 0.64 & 0.64 & 0.64 \\
        \textbf{(b) Type-1 $\Delta$} & \textbf{-0.15} & \textbf{-0.41} & \textbf{-0.06} & \textbf{-0.49} \\
        \textbf{(c) Baseline $\Delta$} & -0.01 & -0.03 & -0.01 & -0.04 \\
        \hline
    \end{tabularx}
    \caption{\textbf{Changes in token-based F1 scores for questions with original scores above 0.5 (LLaMA3-8B).}}
    \label{tab:mkqa_f1}
\end{table}

\paragraph{Type-1 Transfer Neurons are Critical for Reasoning.}
Fig.~\ref{fig:mkqa llama3 above 0.5} and Tab.~\ref{tab:mkqa_f1} shows the intervention results for questions of each language where the model’s generated answer, under setup (a), exceeds an F1 score of 0.5. This allows us to examine how interventions in neurons affect questions that the model is originally able to answer to some extent. As indicated, although the number of deactivated Type-1 neurons is very small (only 0.2\% of all neurons), deactivating them causes a significant degradation, while deactivating baseline neurons has almost no effect. These results suggest that Type-1 neurons play an essential role in eliciting an answer. This tendency was consistent across models. The results for other models and F1 threshold can be found in Appendix~\ref{sec:appendix: mkqa}. 

To further validate the influence of deactivating Type-1 neurons, we conducted the same experiments on MMLU-ProX~\citep{mmluprox}, a multilingual benchmark designed to evaluate comprehensive reasoning ability. The results are highly consistent with the MKQA results reported above and are provided in Appendix~\ref{mmlu-prox}.

% mkqa f1 above 0.5
% \begin{figure}[t]
%   \includegraphics[width=\columnwidth]{figures/llama3/qa/llama3_above0.5.pdf}
%   \caption{\textbf{Token-based F1 score when deactivating top-1k Type-1 transfer neurons (LLaMA3-8B).}
%   \textcolor[HTML]{228B22}{Green} denotes the score with intervention in \textcolor[HTML]{228B22}{the Type-1 neurons}, while \textcolor{red}{red} represents the score with intervention in \textcolor{red}{the randomly sampled neurons}. Points below the \textcolor{blue}{$y=x$ line} indicate a decrease in the score due to the intervention, whereas points above the $y=x$ line indicate an increase in the score. Points on the $y=x$ line denote no change in the score before and after the intervention.
%   }
%   \label{fig:mkqa llama3 above 0.5}
% \end{figure}

\section{Discussion, Future Studies and Conclusion}
\label{conclusion}

\begin{figure}[t]
  \includegraphics[width=\columnwidth, scale=0.7]{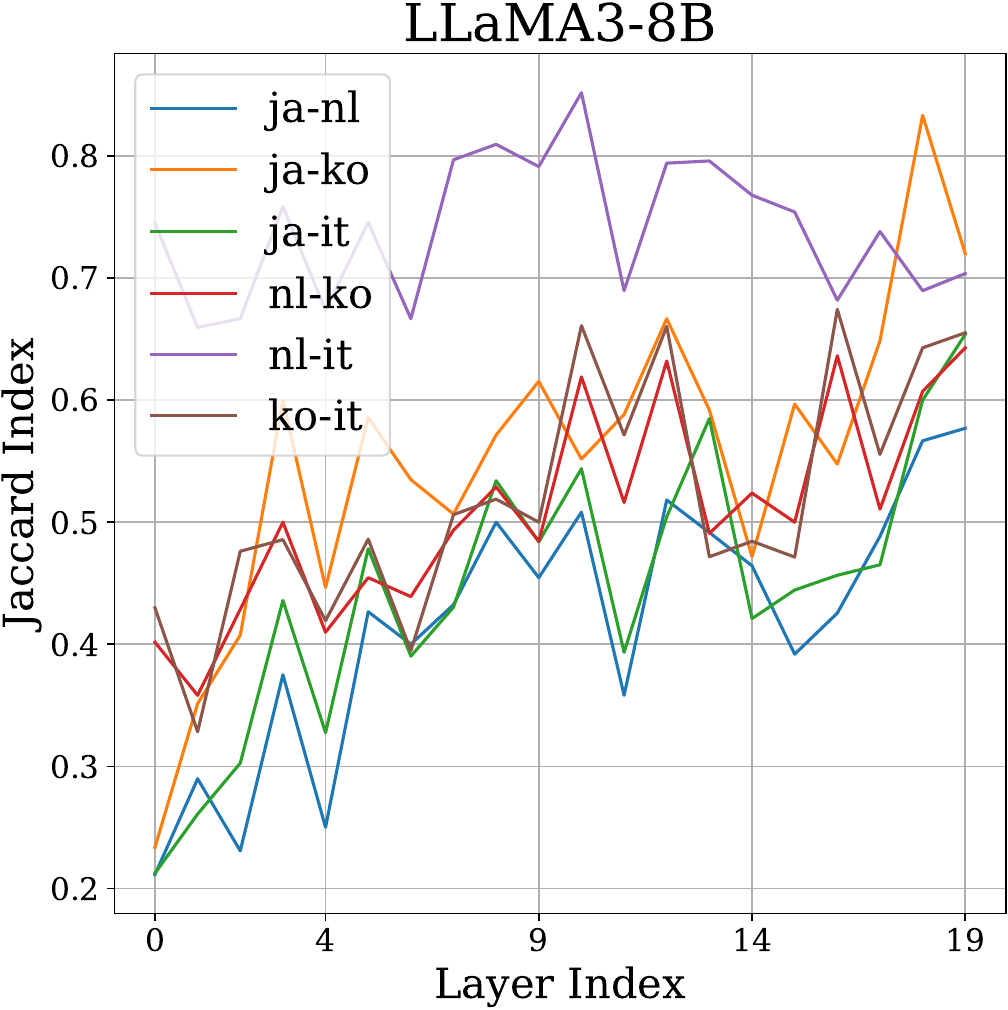}
  \caption{\textbf{Overlap ratio of Type-1 transfer neurons across language pairs and decoder layers (LLaMA3-8B)}. The higher the value on the vertical axis, the greater the overlap of neurons across languages.}
  \label{fig:type1_overlap_llama}
\end{figure}

\paragraph{Discussion 1: The Direction of Representational Transfer Suggests Language-Specificity of Transfer Neurons.}We surmise that the unique direction of the representational shift is highly correlate with the language-specificity of the transfer neurons. That is, if the direction of the transfer (i.e., the direction of the target latent space) is language-specific, the neurons which strongly facilitate the representational shift towards the direction tend to be language-specific. This explains why most Type-1 neurons are less language-specific, while Japanese Type-1 neurons and Type-2 neurons are more likely to be language-specific, as discussed in \S\ref{sec:language-specific nature of transfer neurons} and Appendix~\ref{sec:appendix:language specificity}. In our hypothesis illustrated in Fig.~\ref{fig:figure1}, Type-1 neurons share the same target latent space (i.e., the English latent space). Consequently, as representations move closer to the English latent space towards the middle layers, languages exhibit greater overlap in Type-1 neurons as indicated in Figs.~\ref{fig:type1_overlap_llama} and~\ref{fig:appendix:type1_overlap_mistral_aya}, revealing their language-agnostic nature.

\paragraph{Discussion 2: The Difference between Transfer Neurons and Language-Specific Neurons.}As mentioned in \S\ref{sec:Relatedworks}, although language-specific neurons and regions were discovered in previous studies, their functional role was not clearly identified. In contrast, our work empirically proves (i) certain neurons facilitate the movement of hidden states between language latent spaces, and (ii) some of the transfer neurons we identify tend to be language-specific (e.g., most Type-2 neurons), while others tend not to be (e.g., most Type-1 neurons).
Therefore, it is highly likely that some of the language-specific neurons discovered in previous studies correspond to the Type-2 neurons identified in our work. However, our analysis goes further by also identifying Type-1 neurons and non-language-specific Type-2 neurons, which were not captured in the previous studies. 
The reason why not all transfer neurons are fully language-specific, is that, as discussed in Discussion 1 above, the trajectories for representational transfer can be partly shared by languages especially those with linguistic and spacial proximity. This might be the rationale for language-family specificity of transfer neurons.

\paragraph{Future Studies.}
% We believe that transfer neurons are likely to play a key role in reasoning and cross-lingual transfer between high resource languages and middle- and low-resource languages in multilingual LLMs. Therefore, one potential use case of our findings is to leverage transfer neurons as a mechanism to enhance multilingual ability of LLMs—particularly for languages with limited resources. Specifically, after multilingual pretraining, further fine-tuning of transfer neurons could help align the hidden states of the target language with a English-centric shared latent space where reasoning is performed. This better alignment may enable lower-resource languages to use better-organized conceptual and structural representations learned from high-resource languages such as English, potentially improving the multilingual ability of LLMs.
We argue that transfer neurons play a central role in reasoning and cross-lingual transfer between high- and low-resource languages in multilingual LLMs. A practical implication is to exploit these neurons to enhance multilingual ability: after multilingual pretraining, selectively fine-tuning transfer neurons could better align hidden states of low-resource languages with the English-centric latent space where reasoning occurs. Such alignment may allow low-resource languages to leverage conceptual and structural representations learned from high-resource languages, thereby improving multilingual performance.

\paragraph{Conclusion.}
% In this study, we proposed and empirically validated the Transfer Neurons Hypothesis. First, we showed evidence of latent space transitions and the existence of not only language-specific latent spaces, but also a shared semantic latent space in LLMs. We then identified transfer neurons, highlighting their language and language family specificity and their role in representational transfer to the target latent spaces. Finally, we demonstrated their impact on reasoning performance. 

In this study, we proposed and empirically validated the Transfer Neurons Hypothesis. We provided evidence for language-specific latent spaces and a shared semantic latent space, identified transfer neurons highlighting their language and language family specificity, and showed that transfer neurons are essential for reasoning.

% We believe this study sheds light on a deeper understanding of multilingual processing in LLMs.

\section*{Limitations}
\paragraph{Linear Transformation as Representational Transfer.}Throughout the series of experiments, we hypothesize that neurons in the MLP module are responsible for transferring internal representations to and from a shared semantic latent space. As described in Eqs.~\ref{eq:mlp} and~\ref{eq:weighted_sum}, the MLP computation ultimately reduces to a weighted sum of activations and their corresponding value vectors in down projection matrix. Therefore, when considering only the MLP module, the transformation between latent spaces can be interpreted as a parallel shift (i.e., vector summation). This implies that our analysis is limited to linear transformations in the representational space. We did not consider other modules that might induce alternative types of vector movements,  which may represent a limitation of this study.
\paragraph{Other Languages and Datasets.} In this study, we used three LLMs and validated four languages for each model, including both those with linguistic proximity to English (i.e., Dutch and Italian) and those without such proximity (i.e., Japanese and Korean). As sentence datasets, we primarily employed tatoeba~\cite{tatoeba} and MKQA~\cite{longpre-etal-2021-mkqa} for neuron detection and validation. Although we believe the settings demonstrate the generalizability of our findings well, other types of languages and datasets (e.g., those involving longer input sentences) warrant further investigation.

% Since December 2023, a "Limitations" section has been required for all papers submitted to ACL Rolling Review (ARR). This section should be placed at the end of the paper, before the references. The "Limitations" section (along with, optionally, a section for ethical considerations) may be up to one page and will not count toward the final page limit. Note that these files may be used by venues that do not rely on ARR so it is recommended to verify the requirement of a "Limitations" section and other criteria with the venue in question.

\section*{Acknowledgments}
This work was supported by The Nakajima Foundation.
The authors would like to express their gratitude to Mr. Hakaze Cho, Mr. Kenshiro Tanaka, and Mr. Koga Kobayashi of the Japan Advanced Institute of Science and Technology, for providing valuable comments for this work.

% \begin{figure}[t]
%   \includegraphics[width=\columnwidth, scale=0.7]{figures/llama3/type1_jaccard_revised.pdf}
%   \caption{\textbf{Overlap ratio of Type-1 transfer neurons across language pairs and decoder layers (LLaMA3-8B)}. The higher the value on the vertical axis, the greater the overlap of neurons across languages.}
%   \label{fig:type1_overlap_llama}
% \end{figure}

\bibliography{custom}

\begin{thebibliography}{32}
\providecommand{\natexlab}[1]{#1}

\bibitem[{Bandarkar et~al.(2025)Bandarkar, Muller, Yuvraj, Hou, Singhal, Lv, and Liu}]{bandarkar2025layer-swapping}
Lucas Bandarkar, Benjamin Muller, Pritish Yuvraj, Rui Hou, Nayan Singhal, Hongjiang Lv, and Bing Liu. 2025.
\newblock \href {https://openreview.net/forum?id=vQhn4wrQ6j} {Layer swapping for zero-shot cross-lingual transfer in large language models}.
\newblock In \emph{The Thirteenth International Conference on Learning Representations}.

\bibitem[{Chen et~al.(2024)Chen, Cao, Chen, Liu, and Zhao}]{journey_knowledge_neuron}
Yuheng Chen, Pengfei Cao, Yubo Chen, Kang Liu, and Jun Zhao. 2024.
\newblock \href {https://arxiv.org/abs/2308.13198} {Journey to the center of the knowledge neurons: Discoveries of language-independent knowledge neurons and degenerate knowledge neurons}.
\newblock In \emph{AAAI Conference on Artificial Intelligence}.

\bibitem[{Dai et~al.(2022)Dai, Dong, Hao, Sui, Chang, and Wei}]{knowledge_neurons}
Damai Dai, Li~Dong, Yaru Hao, Zhifang Sui, Baobao Chang, and Furu Wei. 2022.
\newblock \href {https://aclanthology.org/2022.acl-long.581/} {Knowledge neurons in pretrained transformers}.
\newblock In \emph{Proceedings of the 60th Annual Meeting of the Association for Computational Linguistics (Volume 1: Long Papers)}, pages 8493--8502, Dublin, Ireland. Association for Computational Linguistics.

\bibitem[{Dang et~al.(2024)Dang, Singh, D'souza, Ahmadian, Salamanca, Smith, Peppin, Hong, Govindassamy, Zhao, Kublik, Amer, Aryabumi, Campos, Tan, Kocmi, Strub, Grinsztajn, Flet-Berliac, Locatelli, Lin, Talupuru, Venkitesh, Cairuz, Yang, Chung, Ko, Shi, Shukayev, Bae, Piktus, Castagné, Cruz-Salinas, Kim, Crawhall-Stein, Morisot, Roy, Blunsom, Zhang, Gomez, Frosst, Fadaee, Ermis, Üstün, and Hooker}]{aya}
John Dang, Shivalika Singh, Daniel D'souza, Arash Ahmadian, Alejandro Salamanca, Madeline Smith, Aidan Peppin, Sungjin Hong, Manoj Govindassamy, Terrence Zhao, Sandra Kublik, Meor Amer, Viraat Aryabumi, Jon~Ander Campos, Yi-Chern Tan, Tom Kocmi, Florian Strub, Nathan Grinsztajn, Yannis Flet-Berliac, and 26 others. 2024.
\newblock \href {https://arxiv.org/abs/2412.04261} {Aya expanse: Combining research breakthroughs for a new multilingual frontier}.
\newblock \emph{arXiv Preprint}, arXiv:2412.04261.

\bibitem[{Duan et~al.(2025)Duan, Zhou, Xiao, and Cai}]{duan-etal-2025-unveiling}
Xufeng Duan, Xinyu Zhou, Bei Xiao, and Zhenguang Cai. 2025.
\newblock \href {https://aclanthology.org/2025.coling-main.677/} {Unveiling language competence neurons: A psycholinguistic approach to model interpretability}.
\newblock In \emph{Proceedings of the 31st International Conference on Computational Linguistics}, pages 10148--10157, Abu Dhabi, UAE. Association for Computational Linguistics.

\bibitem[{Geva et~al.(2022)Geva, Caciularu, Wang, and Goldberg}]{geva2022}
Mor Geva, Avi Caciularu, Kevin Wang, and Yoav Goldberg. 2022.
\newblock \href {https://aclanthology.org/2022.emnlp-main.3/} {Transformer feed-forward layers build predictions by promoting concepts in the vocabulary space}.
\newblock In \emph{Proceedings of the 2022 Conference on Empirical Methods in Natural Language Processing}, pages 30--45, Abu Dhabi, United Arab Emirates. Association for Computational Linguistics.

\bibitem[{Geva et~al.(2021)Geva, Schuster, Berant, and Levy}]{geva2021}
Mor Geva, Roei Schuster, Jonathan Berant, and Omer Levy. 2021.
\newblock \href {https://aclanthology.org/2021.emnlp-main.446/} {Transformer feed-forward layers are key-value memories}.
\newblock In \emph{Proceedings of the 2021 Conference on Empirical Methods in Natural Language Processing}, pages 5484--5495, Online and Punta Cana, Dominican Republic. Association for Computational Linguistics.

\bibitem[{Grattafiori et~al.(2024)Grattafiori, Dubey, Jauhri, Pandey, Kadian, Al-Dahle, Letman, Mathur, Schelten, Vaughan, Yang, Fan, Goyal, Hartshorn, Yang, Mitra, Sravankumar, Korenev, Hinsvark, Rao, Zhang, Rodriguez, Gregerson, Spataru, Roziere, Biron, Tang, Chern, Caucheteux, Nayak, Bi, Marra, McConnell, Keller, Touret, Wu, Wong, Ferrer, Nikolaidis, Allonsius, Song, Pintz, Livshits, Wyatt, Esiobu, Choudhary, Mahajan, Garcia-Olano, Perino, Hupkes, Lakomkin, AlBadawy, Lobanova, Dinan, Smith, Radenovic, Guzmán, Zhang, Synnaeve, Lee, Anderson, Thattai, Nail, Mialon, Pang, Cucurell, Nguyen, Korevaar, Xu, Touvron, Zarov, Ibarra, Kloumann, Misra, Evtimov, Zhang, Copet, Lee, Geffert, Vranes, Park, Mahadeokar, Shah, van~der Linde, Billock, Hong, Lee, Fu, Chi, Huang, Liu, Wang, Yu, Bitton, Spisak, Park, Rocca, Johnstun, Saxe, Jia, Alwala, Prasad, Upasani, Plawiak, Li, Heafield, Stone, El-Arini, Iyer, Malik, Chiu, Bhalla, Lakhotia, Rantala-Yeary, van~der Maaten, Chen, Tan, Jenkins, Martin, Madaan, Malo, Blecher,
  Landzaat, de~Oliveira, Muzzi, Pasupuleti, Singh, Paluri, Kardas, Tsimpoukelli, Oldham, Rita, Pavlova, Kambadur, Lewis, Si, Singh, Hassan, Goyal, Torabi, Bashlykov, Bogoychev, Chatterji, Zhang, Duchenne, Çelebi, Alrassy, Zhang, Li, Vasic, Weng, Bhargava, Dubal, Krishnan, Koura, Xu, He, Dong, Srinivasan, Ganapathy, Calderer, Cabral, Stojnic, Raileanu, Maheswari, Girdhar, Patel, Sauvestre, Polidoro, Sumbaly, Taylor, Silva, Hou, Wang, Hosseini, Chennabasappa, Singh, Bell, Kim, Edunov, Nie, Narang, Raparthy, Shen, Wan, Bhosale, Zhang, Vandenhende, Batra, Whitman, Sootla, Collot, Gururangan, Borodinsky, Herman, Fowler, Sheasha, Georgiou, Scialom, Speckbacher, Mihaylov, Xiao, Karn, Goswami, Gupta, Ramanathan, Kerkez, Gonguet, Do, Vogeti, Albiero, Petrovic, Chu, Xiong, Fu, Meers, Martinet, Wang, Wang, Tan, Xia, Xie, Jia, Wang, Goldschlag, Gaur, Babaei, Wen, Song, Zhang, Li, Mao, Coudert, Yan, Chen, Papakipos, Singh, Srivastava, Jain, Kelsey, Shajnfeld, Gangidi, Victoria, Goldstand, Menon, Sharma, Boesenberg,
  Baevski, Feinstein, Kallet, Sangani, Teo, Yunus, Lupu, Alvarado, Caples, Gu, Ho, Poulton, Ryan, Ramchandani, Dong, Franco, Goyal, Saraf, Chowdhury, Gabriel, Bharambe, Eisenman, Yazdan, James, Maurer, Leonhardi, Huang, Loyd, Paola, Paranjape, Liu, Wu, Ni, Hancock, Wasti, Spence, Stojkovic, Gamido, Montalvo, Parker, Burton, Mejia, Liu, Wang, Kim, Zhou, Hu, Chu, Cai, Tindal, Feichtenhofer, Gao, Civin, Beaty, Kreymer, Li, Adkins, Xu, Testuggine, David, Parikh, Liskovich, Foss, Wang, Le, Holland, Dowling, Jamil, Montgomery, Presani, Hahn, Wood, Le, Brinkman, Arcaute, Dunbar, Smothers, Sun, Kreuk, Tian, Kokkinos, Ozgenel, Caggioni, Kanayet, Seide, Florez, Schwarz, Badeer, Swee, Halpern, Herman, Sizov, Guangyi, Zhang, Lakshminarayanan, Inan, Shojanazeri, Zou, Wang, Zha, Habeeb, Rudolph, Suk, Aspegren, Goldman, Zhan, Damlaj, Molybog, Tufanov, Leontiadis, Veliche, Gat, Weissman, Geboski, Kohli, Lam, Asher, Gaya, Marcus, Tang, Chan, Zhen, Reizenstein, Teboul, Zhong, Jin, Yang, Cummings, Carvill, Shepard, McPhie,
  Torres, Ginsburg, Wang, Wu, U, Saxena, Khandelwal, Zand, Matosich, Veeraraghavan, Michelena, Li, Jagadeesh, Huang, Chawla, Huang, Chen, Garg, A, Silva, Bell, Zhang, Guo, Yu, Moshkovich, Wehrstedt, Khabsa, Avalani, Bhatt, Mankus, Hasson, Lennie, Reso, Groshev, Naumov, Lathi, Keneally, Liu, Seltzer, Valko, Restrepo, Patel, Vyatskov, Samvelyan, Clark, Macey, Wang, Hermoso, Metanat, Rastegari, Bansal, Santhanam, Parks, White, Bawa, Singhal, Egebo, Usunier, Mehta, Laptev, Dong, Cheng, Chernoguz, Hart, Salpekar, Kalinli, Kent, Parekh, Saab, Balaji, Rittner, Bontrager, Roux, Dollar, Zvyagina, Ratanchandani, Yuvraj, Liang, Alao, Rodriguez, Ayub, Murthy, Nayani, Mitra, Parthasarathy, Li, Hogan, Battey, Wang, Howes, Rinott, Mehta, Siby, Bondu, Datta, Chugh, Hunt, Dhillon, Sidorov, Pan, Mahajan, Verma, Yamamoto, Ramaswamy, Lindsay, Lindsay, Feng, Lin, Zha, Patil, Shankar, Zhang, Zhang, Wang, Agarwal, Sajuyigbe, Chintala, Max, Chen, Kehoe, Satterfield, Govindaprasad, Gupta, Deng, Cho, Virk, Subramanian, Choudhury,
  Goldman, Remez, Glaser, Best, Koehler, Robinson, Li, Zhang, Matthews, Chou, Shaked, Vontimitta, Ajayi, Montanez, Mohan, Kumar, Mangla, Ionescu, Poenaru, Mihailescu, Ivanov, Li, Wang, Jiang, Bouaziz, Constable, Tang, Wu, Wang, Wu, Gao, Kleinman, Chen, Hu, Jia, Qi, Li, Zhang, Zhang, Adi, Nam, Yu, Wang, Zhao, Hao, Qian, Li, He, Rait, DeVito, Rosnbrick, Wen, Yang, Zhao, and Ma}]{llama3}
Aaron Grattafiori, Abhimanyu Dubey, Abhinav Jauhri, Abhinav Pandey, Abhishek Kadian, Ahmad Al-Dahle, Aiesha Letman, Akhil Mathur, Alan Schelten, Alex Vaughan, Amy Yang, Angela Fan, Anirudh Goyal, Anthony Hartshorn, Aobo Yang, Archi Mitra, Archie Sravankumar, Artem Korenev, Arthur Hinsvark, and 542 others. 2024.
\newblock \href {https://arxiv.org/abs/2407.21783} {The llama 3 herd of models}.
\newblock \emph{arXiv Preprint}, arXiv:2407.21783.

\bibitem[{Hiraoka and Inui(2025)}]{hiraoka-inui-2025-repetition}
Tatsuya Hiraoka and Kentaro Inui. 2025.
\newblock \href {https://aclanthology.org/2025.naacl-short.41/} {Repetition neurons: How do language models produce repetitions?}
\newblock In \emph{Proceedings of the 2025 Conference of the Nations of the Americas Chapter of the Association for Computational Linguistics: Human Language Technologies (Volume 2: Short Papers)}, pages 483--495, Albuquerque, New Mexico. Association for Computational Linguistics.

\bibitem[{Huh et~al.(2024)Huh, Cheung, Wang, and Isola}]{icml2024platonic}
Minyoung Huh, Brian Cheung, Tongzhou Wang, and Phillip Isola. 2024.
\newblock \href {https://arxiv.org/abs/2405.07987} {The platonic representation hypothesis}.
\newblock In \emph{International Conference on Machine Learning}.

\bibitem[{Ji et~al.(2025)Ji, Li, Paul, Paavola, Lin, Chen, O'Brien, Luo, Schütze, Tiedemann, and Haddow}]{polywrite}
Shaoxiong Ji, Zihao Li, Indraneil Paul, Jaakko Paavola, Peiqin Lin, Pinzhen Chen, Dayyán O'Brien, Hengyu Luo, Hinrich Schütze, Jörg Tiedemann, and Barry Haddow. 2025.
\newblock \href {https://arxiv.org/abs/2409.17892v2} {Emma-500: Enhancing massively multilingual adaptation of large language models}.
\newblock \emph{arXiv Preprint}, arXiv:2409.17892.

\bibitem[{Jiang et~al.(2023)Jiang, Sablayrolles, Mensch, Bamford, Chaplot, de~las Casas, Bressand, Lengyel, Lample, Saulnier, Lavaud, Lachaux, Stock, Scao, Lavril, Wang, Lacroix, and Sayed}]{mistral}
Albert~Q. Jiang, Alexandre Sablayrolles, Arthur Mensch, Chris Bamford, Devendra~Singh Chaplot, Diego de~las Casas, Florian Bressand, Gianna Lengyel, Guillaume Lample, Lucile Saulnier, Lélio~Renard Lavaud, Marie-Anne Lachaux, Pierre Stock, Teven~Le Scao, Thibaut Lavril, Thomas Wang, Timothée Lacroix, and William~El Sayed. 2023.
\newblock \href {https://arxiv.org/abs/2310.06825} {Mistral 7b}.
\newblock \emph{arXiv Preprint}, arXiv:2310.06825.

\bibitem[{Kojima et~al.(2024)Kojima, Okimura, Iwasawa, Yanaka, and Matsuo}]{kojima2024}
Takeshi Kojima, Itsuki Okimura, Yusuke Iwasawa, Hitomi Yanaka, and Yutaka Matsuo. 2024.
\newblock \href {https://aclanthology.org/2024.naacl-long.384/} {On the multilingual ability of decoder-based pre-trained language models: Finding and controlling language-specific neurons}.
\newblock In \emph{Proceedings of the 2024 Conference of the North American Chapter of the Association for Computational Linguistics: Human Language Technologies (Volume 1: Long Papers)}, pages 6919--6971, Mexico City, Mexico. Association for Computational Linguistics.

\bibitem[{Libovick{\'y} et~al.(2020)Libovick{\'y}, Rosa, and Fraser}]{libovicky-etal-2020-language}
Jind{\v{r}}ich Libovick{\'y}, Rudolf Rosa, and Alexander Fraser. 2020.
\newblock \href {https://aclanthology.org/2020.findings-emnlp.150/} {On the language neutrality of pre-trained multilingual representations}.
\newblock In \emph{Findings of the Association for Computational Linguistics: EMNLP 2020}, pages 1663--1674, Online. Association for Computational Linguistics.

\bibitem[{Liu et~al.(2021)Liu, Dai, So, and Le}]{gated_mlp_paper}
Hanxiao Liu, Zihang Dai, David So, and Quoc~V Le. 2021.
\newblock \href {https://proceedings.neurips.cc/paper_files/paper/2021/file/4cc05b35c2f937c5bd9e7d41d3686fff-Paper.pdf} {Pay attention to mlps}.
\newblock In \emph{Advances in Neural Information Processing Systems}, volume~34, pages 9204--9215. Curran Associates, Inc.

\bibitem[{Longpre et~al.(2021)Longpre, Lu, and Daiber}]{longpre-etal-2021-mkqa}
Shayne Longpre, Yi~Lu, and Joachim Daiber. 2021.
\newblock \href {https://aclanthology.org/2021.tacl-1.82/} {{MKQA}: A linguistically diverse benchmark for multilingual open domain question answering}.
\newblock \emph{Transactions of the Association for Computational Linguistics}, 9:1389--1406.

\bibitem[{Mondal et~al.(2025)Mondal, Sen, Singhania, and Jyothi}]{mondal-etal-2025-language-specific}
Soumen~Kumar Mondal, Sayambhu Sen, Abhishek Singhania, and Preethi Jyothi. 2025.
\newblock \href {https://aclanthology.org/2025.insights-1.6/} {Language-specific neurons do not facilitate cross-lingual transfer}.
\newblock In \emph{The Sixth Workshop on Insights from Negative Results in NLP}, pages 46--62, Albuquerque, New Mexico. Association for Computational Linguistics.

\bibitem[{Pires et~al.(2019)Pires, Schlinger, and Garrette}]{2019-multilingual-bert}
Telmo Pires, Eva Schlinger, and Dan Garrette. 2019.
\newblock \href {https://aclanthology.org/P19-1493/} {How multilingual is multilingual {BERT}?}
\newblock In \emph{Proceedings of the 57th Annual Meeting of the Association for Computational Linguistics}, pages 4996--5001, Florence, Italy. Association for Computational Linguistics.

\bibitem[{Schut et~al.(2025)Schut, Gal, and Farquhar}]{schut2025multilingualllmsthinkenglish}
Lisa Schut, Yarin Gal, and Sebastian Farquhar. 2025.
\newblock \href {https://arxiv.org/abs/2502.15603} {Do multilingual llms think in english?}
\newblock \emph{arXiv Preprint}, arXiv:2502.15603.

\bibitem[{Shazeer(2020)}]{gluvariantsimprovetransformer}
Noam Shazeer. 2020.
\newblock \href {https://arxiv.org/abs/2002.05202} {Glu variants improve transformer}.
\newblock \emph{arXiv Preprint}, arXiv:2002.05202.

\bibitem[{Suau et~al.(2022)Suau, Zappella, and Apostoloff}]{suau2022selfcond}
Xavier Suau, Luca Zappella, and Nicholas Apostoloff. 2022.
\newblock \href {https://arxiv.org/abs/2110.02802} {Self-conditioning pre-trained language models}.
\newblock \emph{International Conference on Machine Learning}.

\bibitem[{Tang et~al.(2024)Tang, Luo, Huang, Zhang, Wang, Zhao, Wei, and Wen}]{tang_lang_specific_neurons}
Tianyi Tang, Wenyang Luo, Haoyang Huang, Dongdong Zhang, Xiaolei Wang, Xin Zhao, Furu Wei, and Ji-Rong Wen. 2024.
\newblock \href {https://aclanthology.org/2024.acl-long.309/} {Language-specific neurons: The key to multilingual capabilities in large language models}.
\newblock In \emph{Proceedings of the 62nd Annual Meeting of the Association for Computational Linguistics (Volume 1: Long Papers)}, pages 5701--5715, Bangkok, Thailand. Association for Computational Linguistics.

\bibitem[{Tiedemann(2020)}]{tatoeba}
J{\"o}rg Tiedemann. 2020.
\newblock \href {https://aclanthology.org/2020.wmt-1.139/} {The tatoeba translation challenge {--} realistic data sets for low resource and multilingual {MT}}.
\newblock In \emph{Proceedings of the Fifth Conference on Machine Translation}, pages 1174--1182. Association for Computational Linguistics.

\bibitem[{Wang et~al.(2024)Wang, Haddow, Peng, and Birch}]{wang2024sharing}
Weixuan Wang, Barry Haddow, Wei Peng, and Alexandra Birch. 2024.
\newblock \href {https://arxiv.org/abs/2406.09265} {Sharing matters: Analysing neurons across languages and tasks in llms}.
\newblock \emph{arXiv Preprint}, arXiv:2406.09265.

\bibitem[{Wendler et~al.(2024)Wendler, Veselovsky, Monea, and West}]{acl2024-llamas}
Chris Wendler, Veniamin Veselovsky, Giovanni Monea, and Robert West. 2024.
\newblock \href {https://aclanthology.org/2024.acl-long.820/} {Do llamas work in {E}nglish? on the latent language of multilingual transformers}.
\newblock In \emph{Proceedings of the 62nd Annual Meeting of the Association for Computational Linguistics (Volume 1: Long Papers)}, pages 15366--15394, Bangkok, Thailand. Association for Computational Linguistics.

\bibitem[{Xuan et~al.(2025)Xuan, Yang, Qi, Zeng, Xiao, Feng, Liu, Xing, Wang, Gao, Lu, Jiang, Li, Li, Yu, Dong, Gu, Li, Xie, Juefei-Xu, Khomh, Yoshie, Chen, Teodoro, Liu, Goebel, Ma, Marrese-Taylor, Lu, Iwasawa, Matsuo, and Li}]{mmluprox}
Weihao Xuan, Rui Yang, Heli Qi, Qingcheng Zeng, Yunze Xiao, Aosong Feng, Dairui Liu, Yun Xing, Junjue Wang, Fan Gao, Jinghui Lu, Yuang Jiang, Huitao Li, Xin Li, Kunyu Yu, Ruihai Dong, Shangding Gu, Yuekang Li, Xiaofei Xie, and 13 others. 2025.
\newblock \href {https://arxiv.org/abs/2503.10497} {Mmlu-prox: A multilingual benchmark for advanced large language model evaluation}.
\newblock \emph{arXiv Preprint}, arXiv:2503.10497.

\bibitem[{Yoon et~al.(2024)Yoon, Jang, Kim, Kim, Shafayat, and Seo}]{yoon-etal-2024-langbridge}
Dongkeun Yoon, Joel Jang, Sungdong Kim, Seungone Kim, Sheikh Shafayat, and Minjoon Seo. 2024.
\newblock \href {https://aclanthology.org/2024.acl-long.405/} {{L}ang{B}ridge: Multilingual reasoning without multilingual supervision}.
\newblock In \emph{Proceedings of the 62nd Annual Meeting of the Association for Computational Linguistics (Volume 1: Long Papers)}, pages 7502--7522, Bangkok, Thailand. Association for Computational Linguistics.

\bibitem[{Zeng et~al.(2025)Zeng, Han, Chen, and Yu}]{coling2025-converging}
Hongchuan Zeng, Senyu Han, Lu~Chen, and Kai Yu. 2025.
\newblock \href {https://aclanthology.org/2025.coling-main.707/} {Converging to a lingua franca: Evolution of linguistic regions and semantics alignment in multilingual large language models}.
\newblock In \emph{Proceedings of the 31st International Conference on Computational Linguistics}, pages 10602--10617, Abu Dhabi, UAE. Association for Computational Linguistics.

\bibitem[{Zhang et~al.(2025)Zhang, Liang, Meng, Zhang, Chen, Xu, and Zhou}]{multilingual_knowledge_neurons_colling2025}
Xue Zhang, Yunlong Liang, Fandong Meng, Songming Zhang, Yufeng Chen, Jinan Xu, and Jie Zhou. 2025.
\newblock \href {https://aclanthology.org/2025.coling-main.385/} {Multilingual knowledge editing with language-agnostic factual neurons}.
\newblock In \emph{Proceedings of the 31st International Conference on Computational Linguistics}, pages 5775--5788, Abu Dhabi, UAE. Association for Computational Linguistics.

\bibitem[{Zhang et~al.(2024)Zhang, Zhao, Zhang, Gui, and Huang}]{zhang-etal-2024-unveiling-linguistic}
Zhihao Zhang, Jun Zhao, Qi~Zhang, Tao Gui, and Xuanjing Huang. 2024.
\newblock \href {https://doi.org/10.18653/v1/2024.acl-long.338} {Unveiling linguistic regions in large language models}.
\newblock In \emph{Proceedings of the 62nd Annual Meeting of the Association for Computational Linguistics (Volume 1: Long Papers)}, pages 6228--6247, Bangkok, Thailand. Association for Computational Linguistics.

\bibitem[{Zhao et~al.(2024{\natexlab{a}})Zhao, Yoshinaga, and Oba}]{tracing_facts_mllms}
Xin Zhao, Naoki Yoshinaga, and Daisuke Oba. 2024{\natexlab{a}}.
\newblock \href {https://doi.org/10.18653/v1/2024.eacl-long.127} {Tracing the roots of facts in multilingual language models: Independent, shared, and transferred knowledge}.
\newblock In \emph{Proceedings of the 18th Conference of the European Chapter of the Association for Computational Linguistics (Volume 1: Long Papers)}, pages 2088--2102, St. Julian{'}s, Malta. Association for Computational Linguistics.

\bibitem[{Zhao et~al.(2024{\natexlab{b}})Zhao, Zhang, Chen, Kawaguchi, and Bing}]{neurips2024-multi}
Yiran Zhao, Wenxuan Zhang, Guizhen Chen, Kenji Kawaguchi, and Lidong Bing. 2024{\natexlab{b}}.
\newblock \href {https://neurips.cc/virtual/2024/poster/94375} {How do large language models handle multilingualism?}
\newblock In \emph{Advances in Neural Information Processing Systems (NeurIPS)}.

\end{thebibliography}

% ---------------- Appendix ---------------- %
\appendix

\section{Experimental Settings}
\label{sec:appendix: experimental settings}
\paragraph{Models.} Throughout the experiments in this study, we used LLaMA3-8B\citep{llama3}, Mistral-7B\citep{mistral}, and Aya expanse-8B\citep{aya}.
\paragraph{Datasets.}
For the similarity measurements in \S\ref{sec:detecting and controlling transfer neurons}, we used the Tatoeba parallel corpus \citep{tatoeba}. This dataset was also used to investigate the language and language-family specificity of transfer neurons in \S\ref{sec:nature of transfer neurons}, by randomly sampling sentences for each language.

For the QA task and the dimensionality reduction presented in \S\ref{Visualizing the Hidden language Subspace with PCA} and Appendix~\ref{sec:appendix:visualization language subspaces}, we used the MKQA dataset \citep{longpre-etal-2021-mkqa}.

For similarity measurements of hidden states and activation patterns, computation of centroids of each latent spaces, identifying transfer neurons, applying PCA to hidden representations, and investigating language specificity of transfer neurons, we sampled 1k sentences per language. Finally, for SVD experiments described in \S\ref{sec:Subspace Property of Language-Specific Hidden States}, we sampled 1k hidden states per layer and per language using MKQA sentences.

\paragraph{Train-Test Split for Identifying and Controlling Transfer Neurons.}
We split the sentence data into a 50:50 ratio (train:test), using 1k samples for each split (2k samples in total) to identify and deactivate transfer neurons. In other words, the similarity re-measurements described in \S\ref{sec:detecting and controlling transfer neurons} were conducted using sentences different from those used to identify the transfer neurons, in order to prove their generalization performance. We believe that this setting further strengthens the reliability of our findings.

\paragraph{Representations.}
Throughout the experiments, we extracted representations corresponding to the final token of each input, reflecting our focus on sentence-level rather than word-level semantic processing.

\paragraph{Computational Resources.}
We conducted the experiments using NVIDIA A40 (48 GB), A100 (40 GB), and H200 (141 GB) GPUs. However, all experiments are reproducible on a single A100, and the majority can be executed with a single A40.

% % figure: subspace trajectory
% \begin{figure}[t]
%   \includegraphics[width=\columnwidth]{figures/subspace_trajectory.pdf}
%   \caption{\textbf{Possible Trajectories of the English Subspace Movements towards Middle Layers}. When we consider the movement trajectories of language subspaces across layers, at least two patterns emerge. One is a linear movement towards a target convergence point (\textcolor{red}{\textbf{a}}), whereas the other follows a non-linear trajectory (\textcolor{blue}{\textbf{b}}). The investigation about English subspace trajectories towards middle layers can be found in Appendix~\ref{sec:appendix:moving trajectory for english subspace}.}
%   \label{fig:subspace_trajectory}
% \end{figure}
% figure: subspace trajectory
\begin{figure}[t]
  \includegraphics[width=\columnwidth]{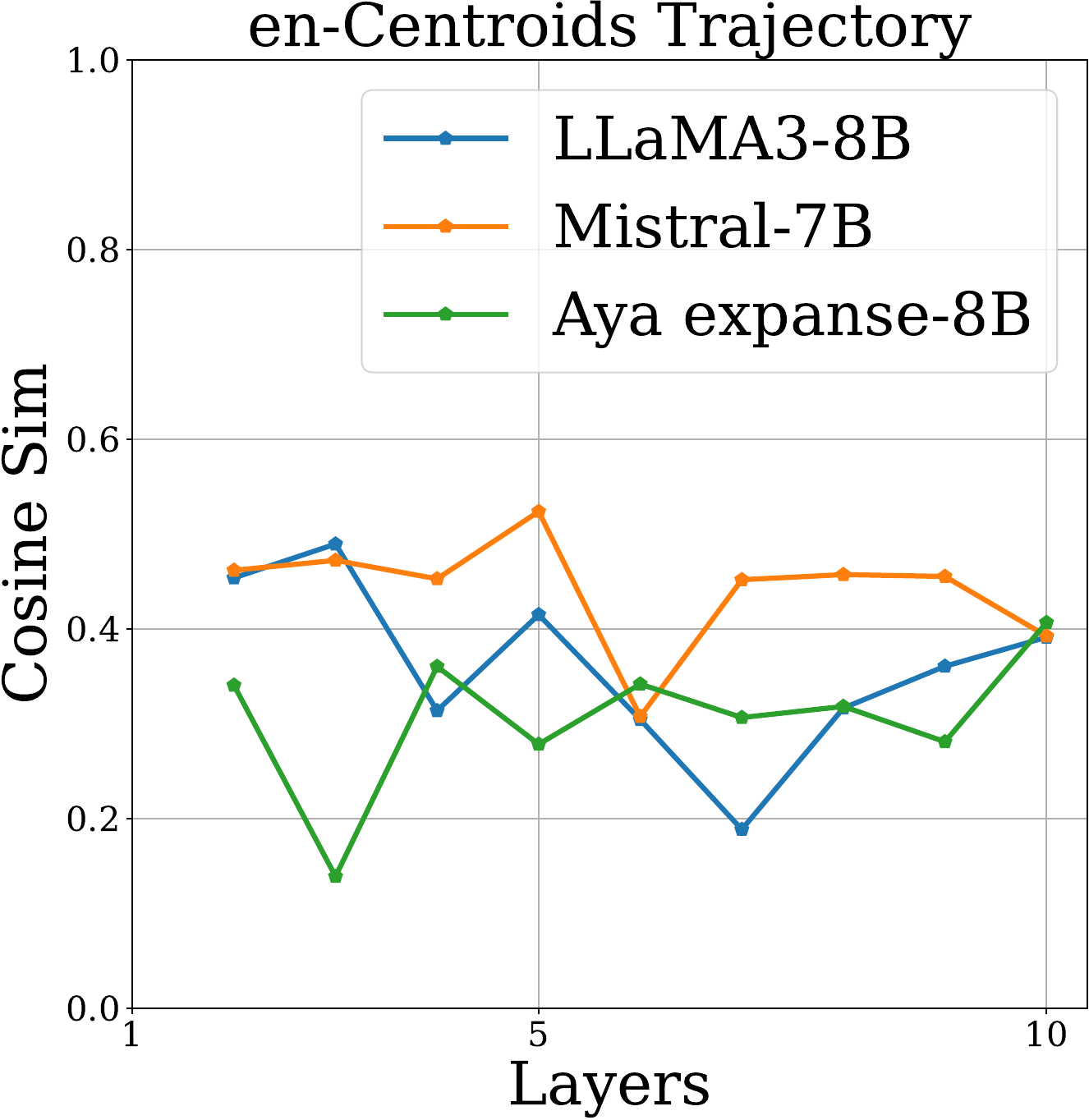}
  \caption{\textbf{Similarity between layer-wise trajectories and linear path from decoder layer 1 to the middle layers for the centroid of the English latent space}.}
  \label{fig:appendix:subspace_trajectory_mid_10}
\end{figure}

\section{Trajectory of the English Latent Space Moving towards Middle Layers}
\label{sec:appendix:moving trajectory for english subspace}
\paragraph{Computational Settings.}
Does English latent space move towards the middle layers in a linear or non-linear manner? To answer this question, we computed the similarity  between a linear trajectory and actual layer-wise trajectory of centroid of the English latent space across layers.

If the similarity is high, it indicates that the layer-wise trajectory closely follows a linear path, suggesting that the representations move almost linearly towards a target convergence point in the middle layers, where semantic processing and reasoning primarily occur. On the other hand, if the similarity is relatively low, it indicates that the representations move non-linearly.

Computations are as follows:
\begin{equation}
    \bm{P}_\mathrm{en} = \bm{C}^{m}_\mathrm{en} - \bm{C}^1_\mathrm{en}
\end{equation}
\begin{equation}
    \bm{T}^l_\mathrm{en} = \bm{C}^{l}_\mathrm{en} - \bm{C}^{l-1}_\mathrm{en}, \quad l \in \{2, \dotsc, m\}
\end{equation}
\begin{equation}
    \mathrm{Sim\_score}^l = \mathrm{cos}(\bm{T}^l_\mathrm{en}, \bm{P}_\mathrm{en})
\end{equation}
where $\bm{P}_\mathrm{en}$ denotes a vector expressing the linear path of the centroid of the English latent space from decoder layer 1 to layer $m$ (the middle layer). It is defined as the difference vector between the centroid at layer $m$, $\bm{C}^m_\mathrm{en}$, and the centroid at layer 1, $\bm{C}^1_\mathrm{en}$. A vector $\bm{T}^l_\mathrm{en}$ represents the actual trajectory step of the centroid of the English latent space between layer $l-1$ and layer $l$. As we mentioned, the higher the similarity score ($\mathrm{Sim\_score}^l$), the more \textbf{linearly} the centroid moves in hidden state space from layer $l-1$ to layer $l$, indicating that the internal representations move linearly towards the target convergence point in the middle layers. To the contrary, lower scores across layers indicates that the representations move \textbf{non-linearly}.

Based on the observations from hidden states and activation patterns similarity (Figs.~\ref{fig:appendix:hs_sim_all_models} and~\ref{fig:appendix:act_sim_all_models}), kernel-based similarity (Fig.~\ref{fig:mutual_knn} in \S\ref{sec:mutual knn} and Fig.~\ref{fig:appendix:kernel-based sim while deactivating Type-1 k=5} in Appendix~\ref{sec:appendix:mutual knn alignment metric}), and PCA visualization of language latent spaces (Figs.~\ref{fig:appendix:pca_all_layers_llama3},~\ref{fig:appendix:pca_all_layers_mistral} and~\ref{fig:appendix:pca_all_layers_aya}), we set $m$ to $10$ representing middle layer where each language latent spaces aligns well.

\paragraph{English Latent Space Transitions Non-Linearly towards Middle Layers.}
Fig.~\ref{fig:appendix:subspace_trajectory_mid_10} present the results. As indicated, English latent space move non-linearly, and this tendency is highly consistent with all the models adopted in this paper.

Based on this observation, we adopt the layer-wise centroid, $\bm{C}^l$, as the target centroid to detect transfer neurons with Eqs.~\ref{eq:layer score shared} and~\ref{eq:neuron score} in \S\ref{sec:scoring the candidate neurons}. This is because using the centroid of a specific layer instead of layer-wise centroids could significantly hinder the effective identification of neurons that contribute to the shift of representations towards the moving English latent space, given that English latent space moves non-linearly.

\section{Similarity and Linear Separability of Internal Representations}
\label{sec:appendix: Similarity of Internal Representations}

\subsection{Similarity of Hidden States and Activation Patterns for Parallel Sentences across Languages}
\label{sec:appendix: similarity of hs and act patterns}
% figure: hs_sim all langs all models.
\begin{figure*}[t]
  \centering
  \includegraphics[width=0.23\linewidth]{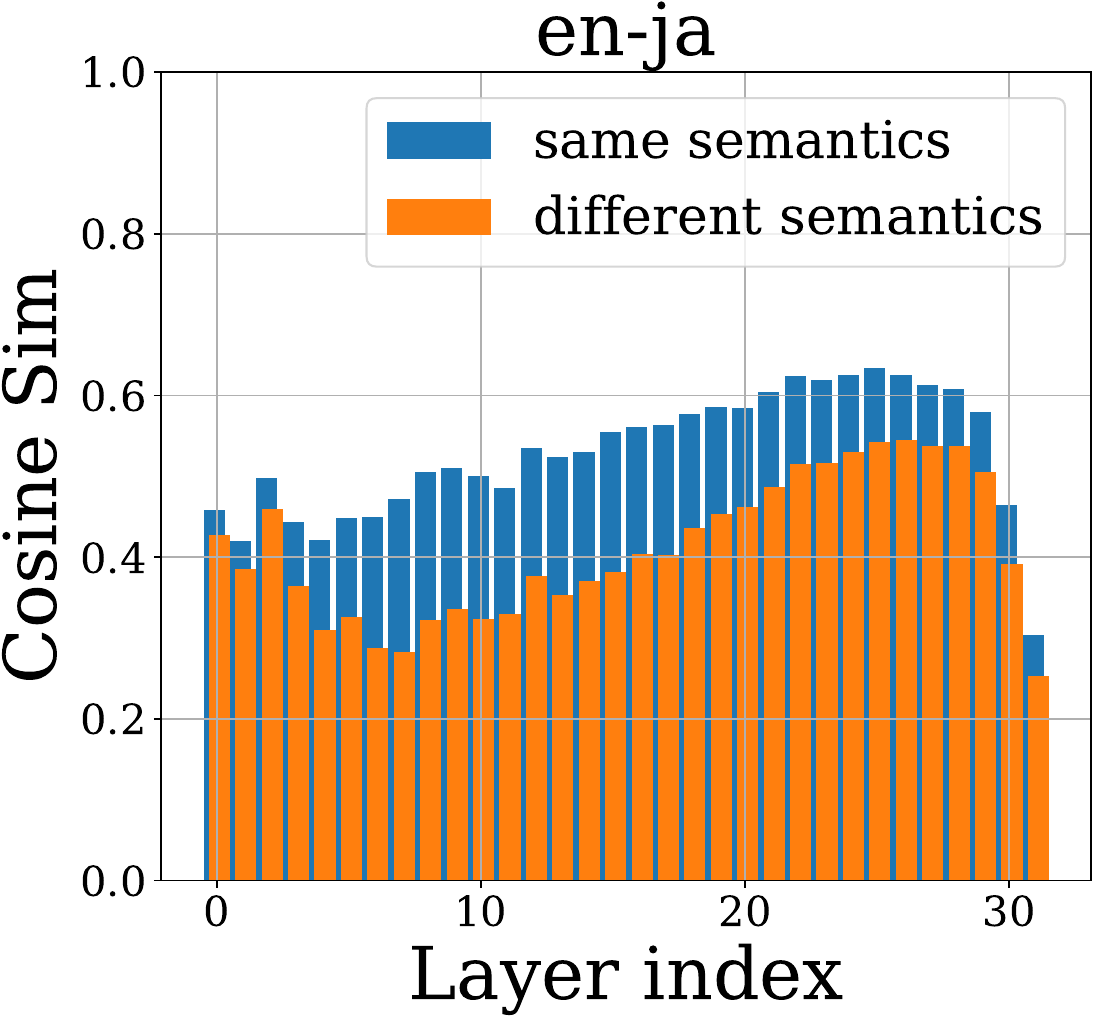}
  \includegraphics[width=0.23\linewidth]{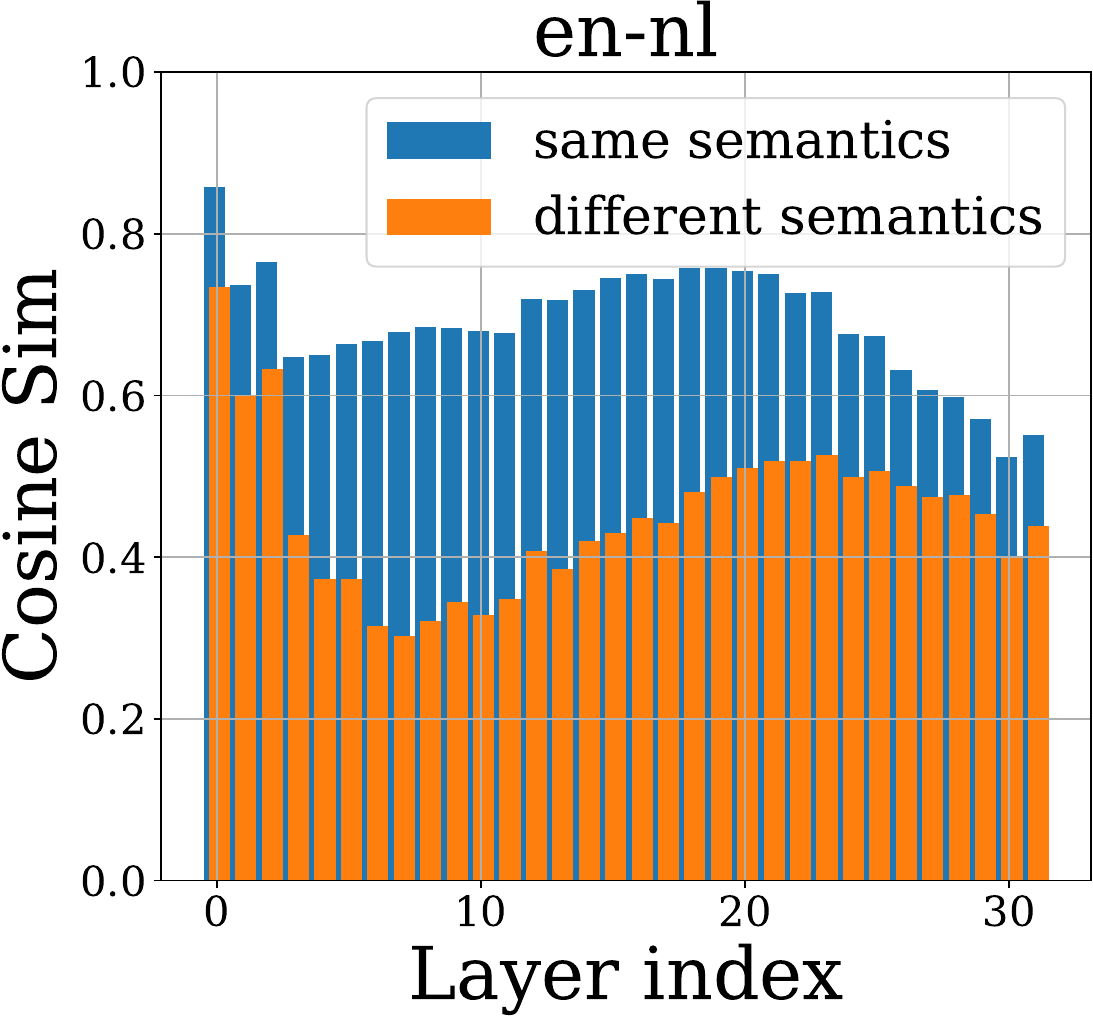}
  \includegraphics[width=0.23\linewidth]{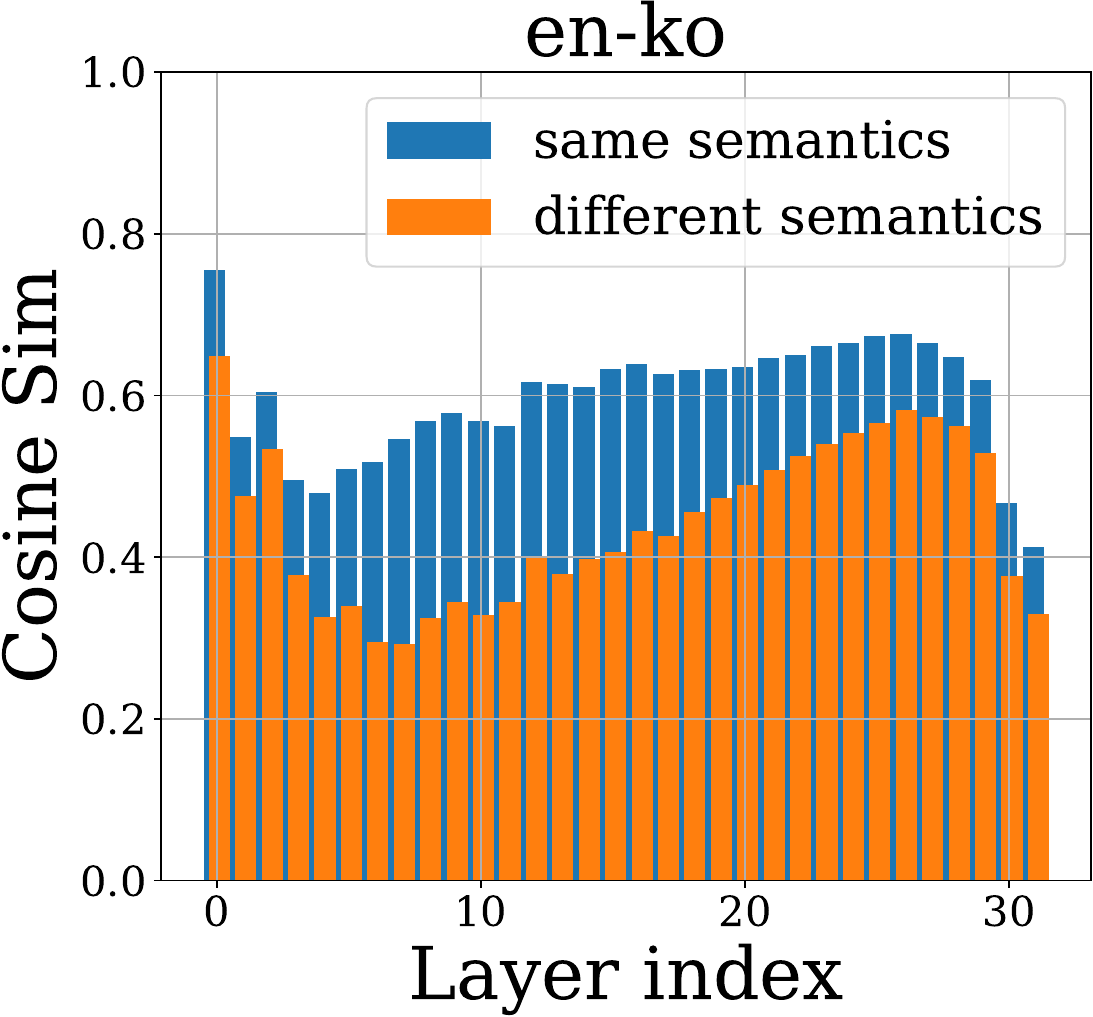}
  \includegraphics[width=0.23\linewidth]{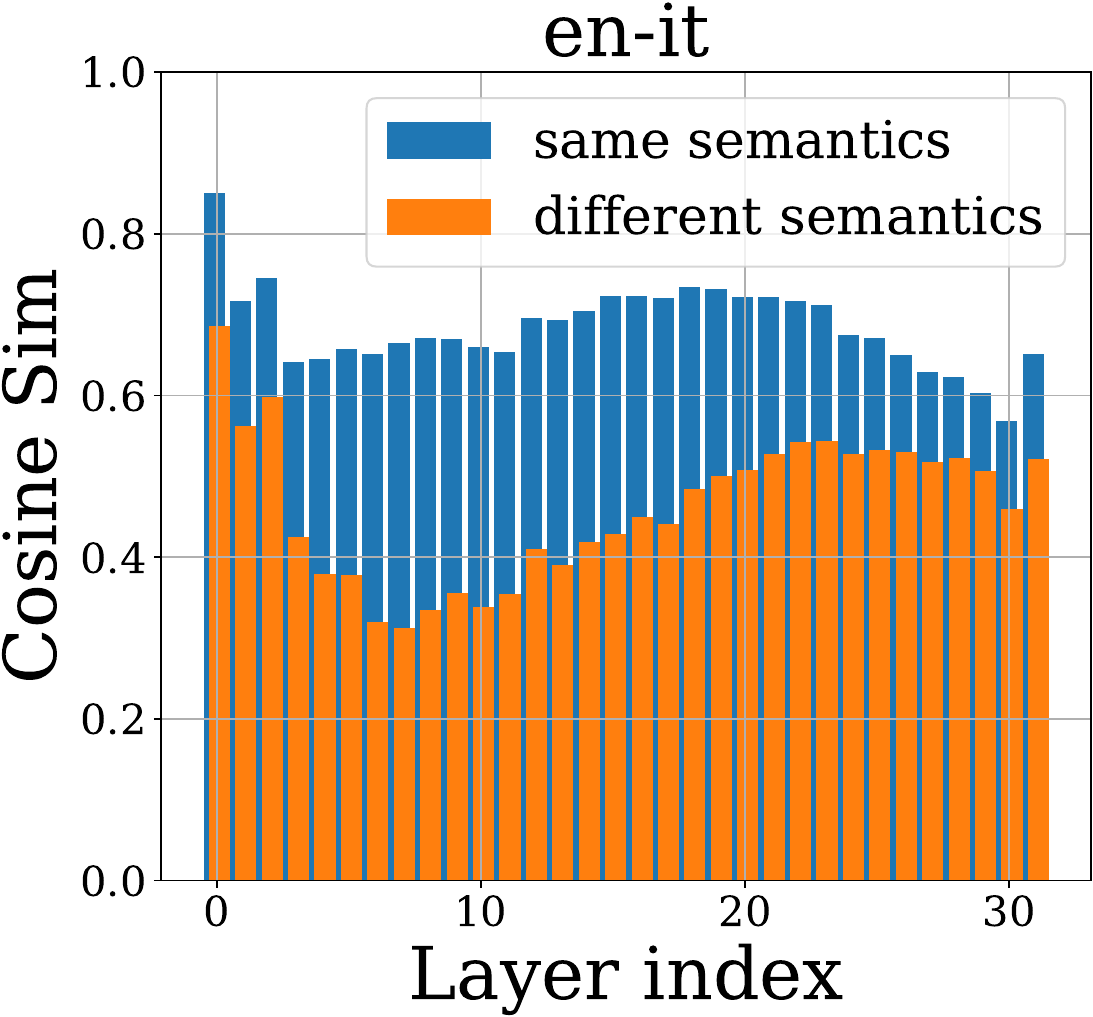}

  \begin{minipage}{0.23\linewidth}\centering en-ja (llama3)\end{minipage}
  \begin{minipage}{0.23\linewidth}\centering en-nl (llama3)\end{minipage}
  \begin{minipage}{0.23\linewidth}\centering en-ko (llama3)\end{minipage}
  \begin{minipage}{0.23\linewidth}\centering en-it (llama3)\end{minipage}

  \includegraphics[width=0.23\linewidth]{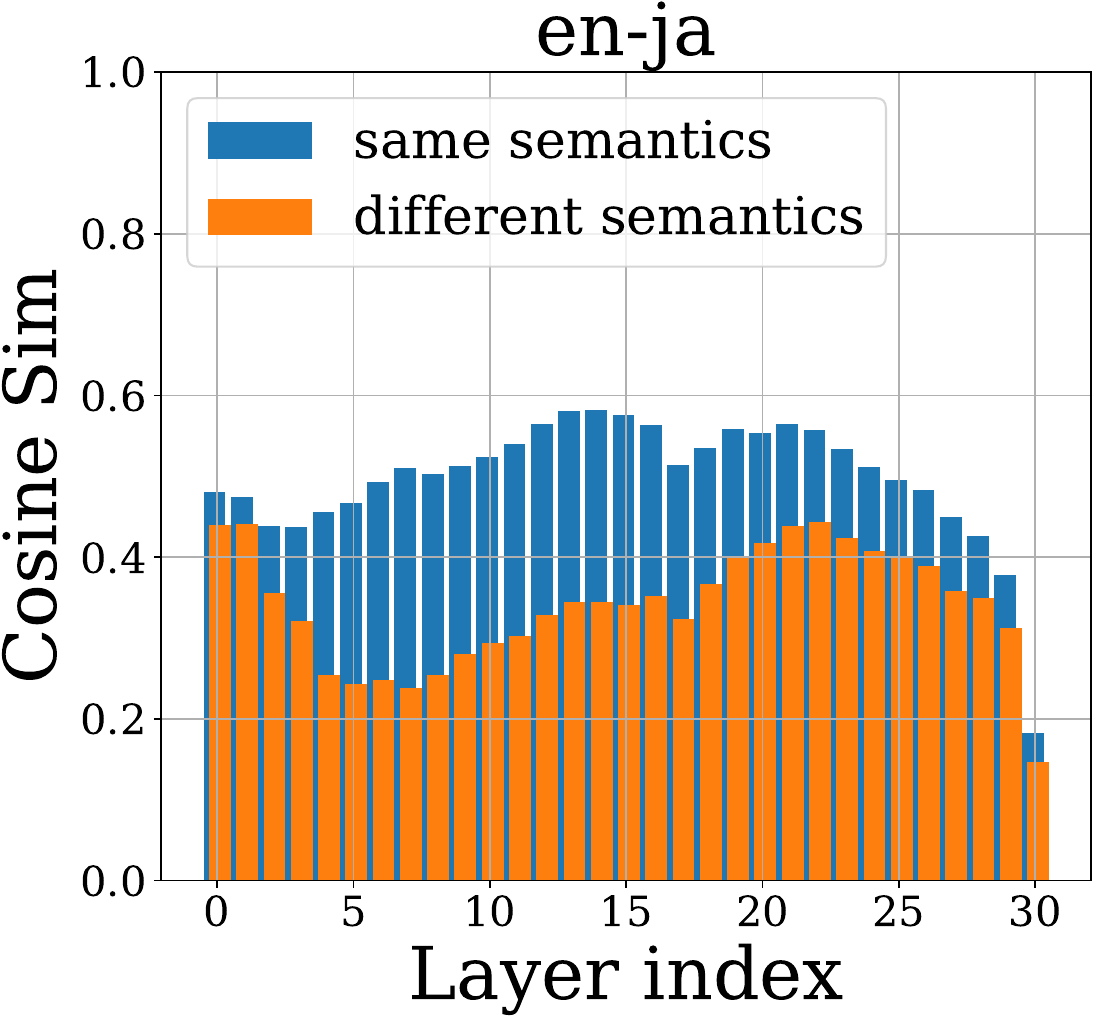}
  \includegraphics[width=0.23\linewidth]{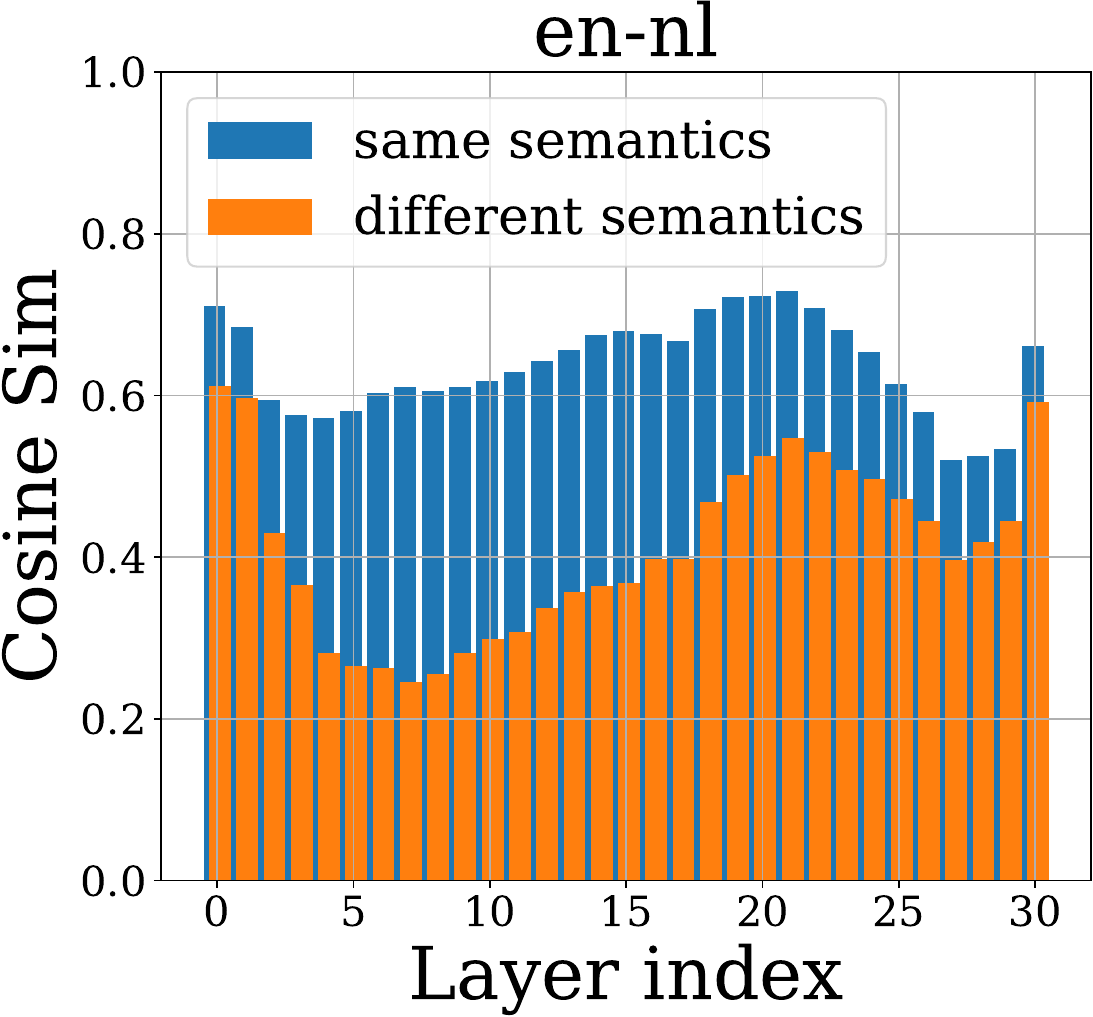}
  \includegraphics[width=0.23\linewidth]{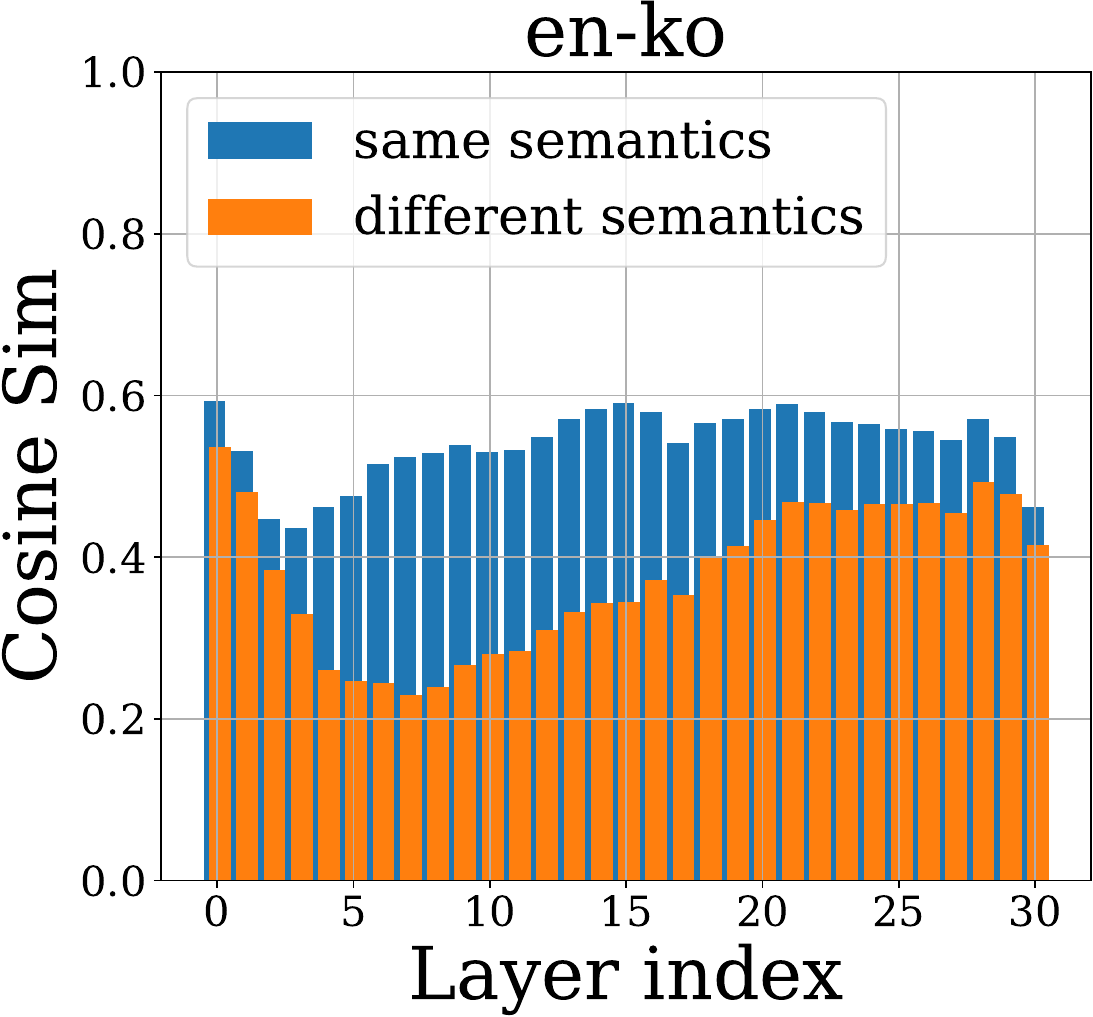}
  \includegraphics[width=0.23\linewidth]{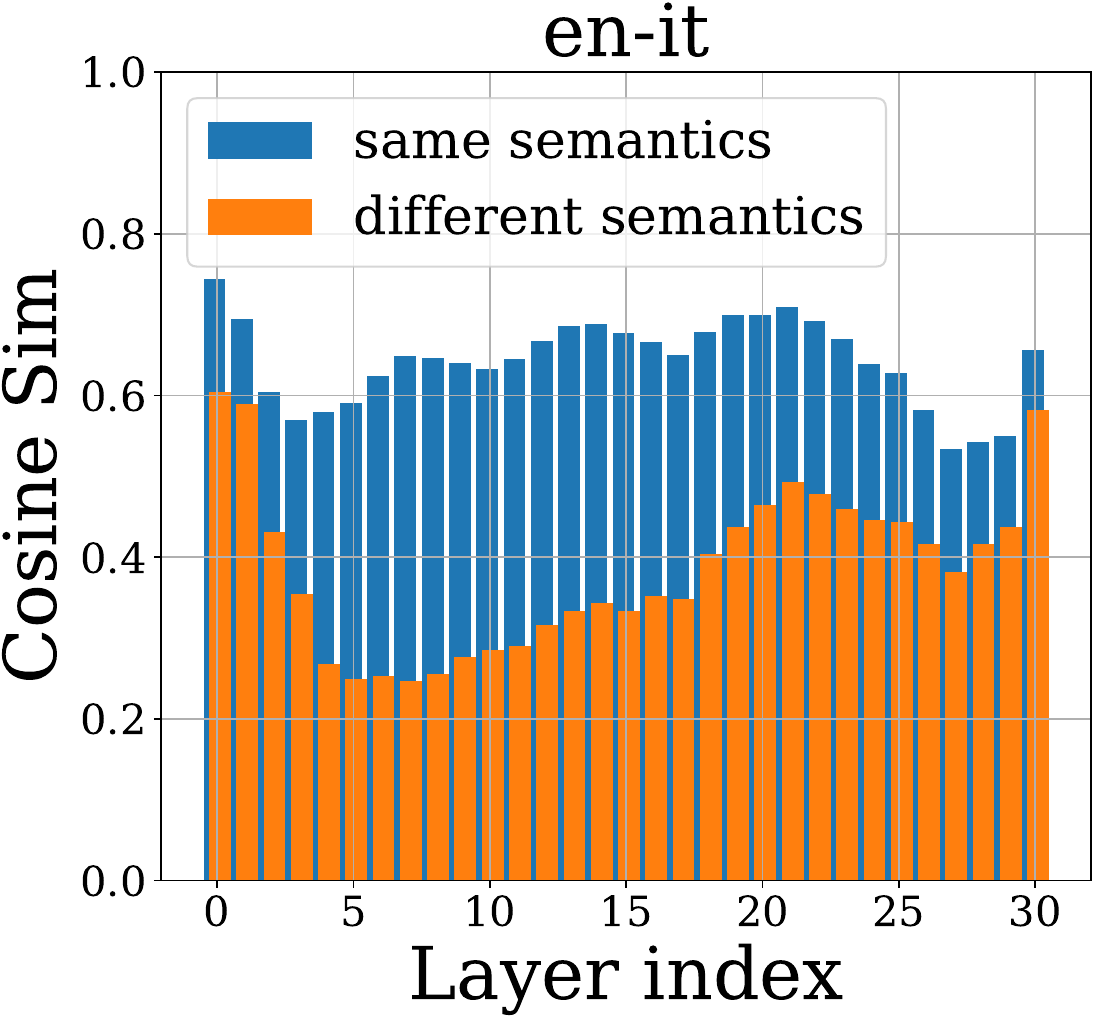}

  \begin{minipage}{0.23\linewidth}\centering en-ja (mistral)\end{minipage}
  \begin{minipage}{0.23\linewidth}\centering en-nl (mistral)\end{minipage}
  \begin{minipage}{0.23\linewidth}\centering en-ko (mistral)\end{minipage}
  \begin{minipage}{0.23\linewidth}\centering en-it (mistral)\end{minipage}

  \includegraphics[width=0.23\linewidth]{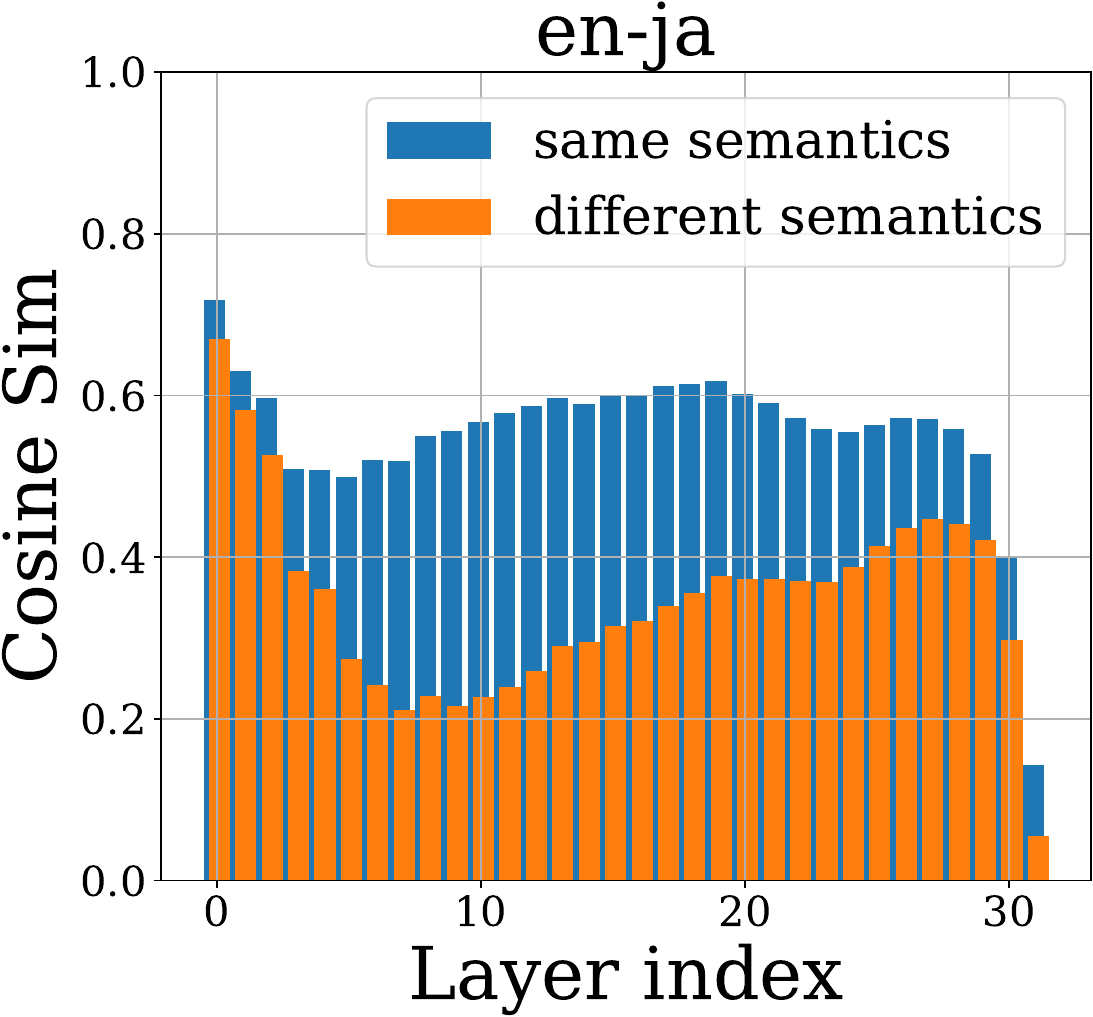}
  \includegraphics[width=0.23\linewidth]{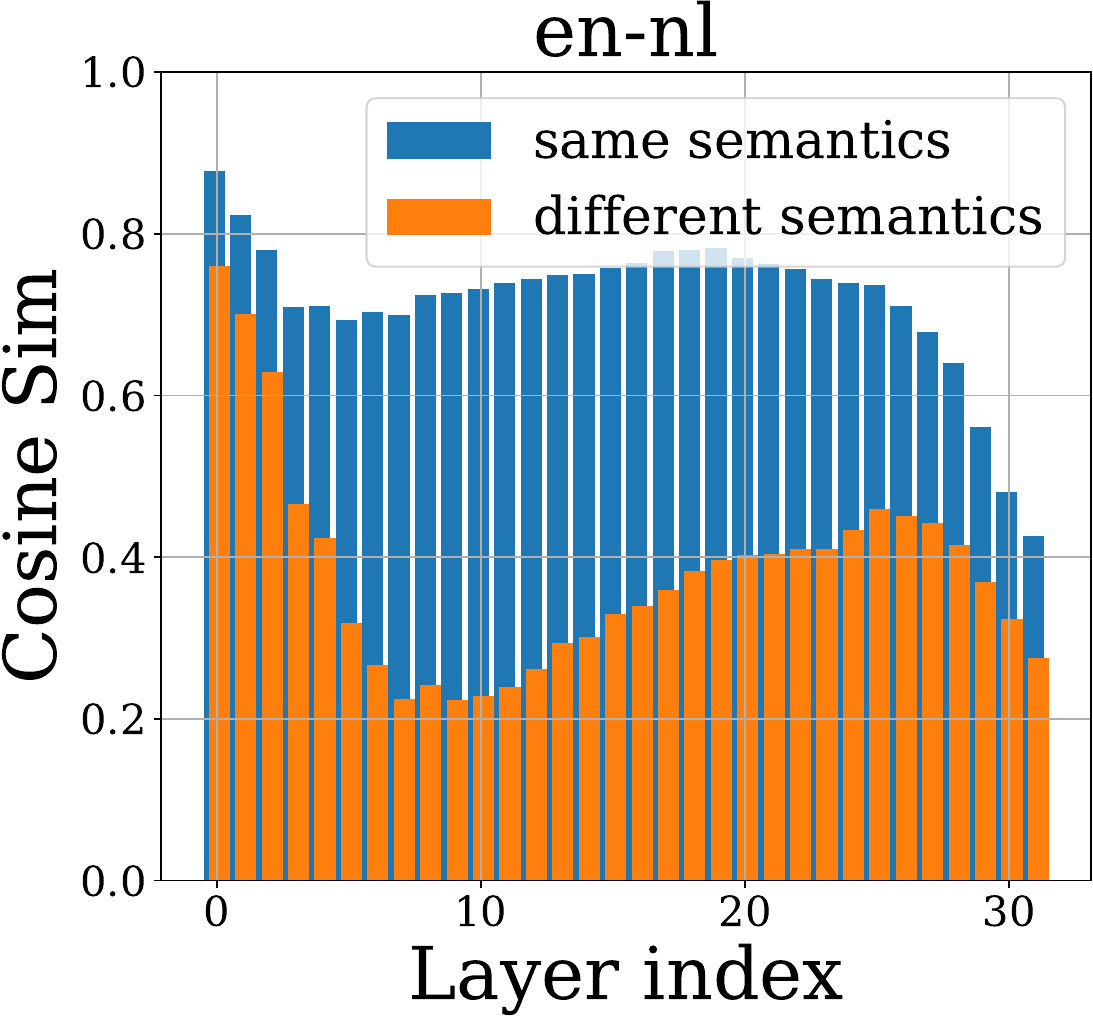}
  \includegraphics[width=0.23\linewidth]{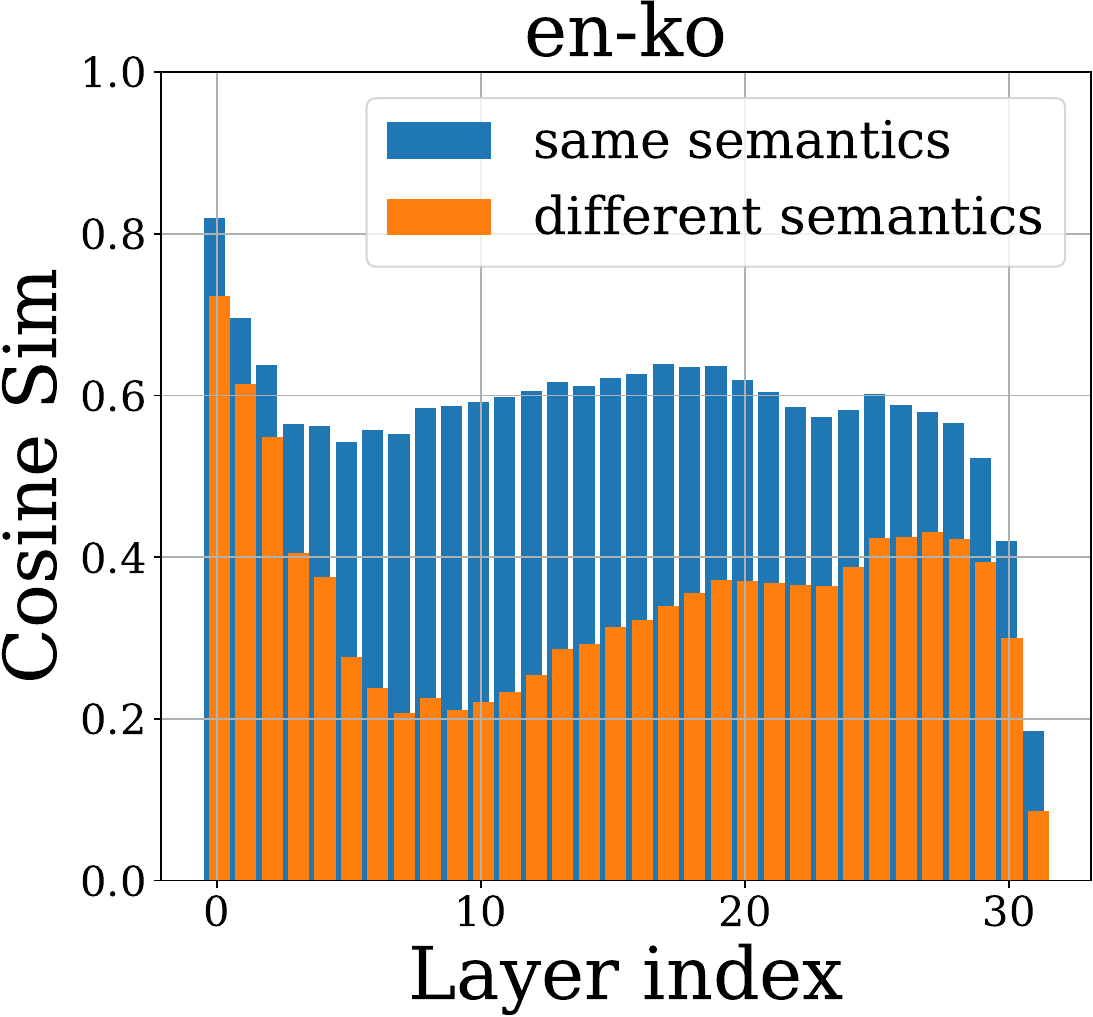}
  \includegraphics[width=0.23\linewidth]{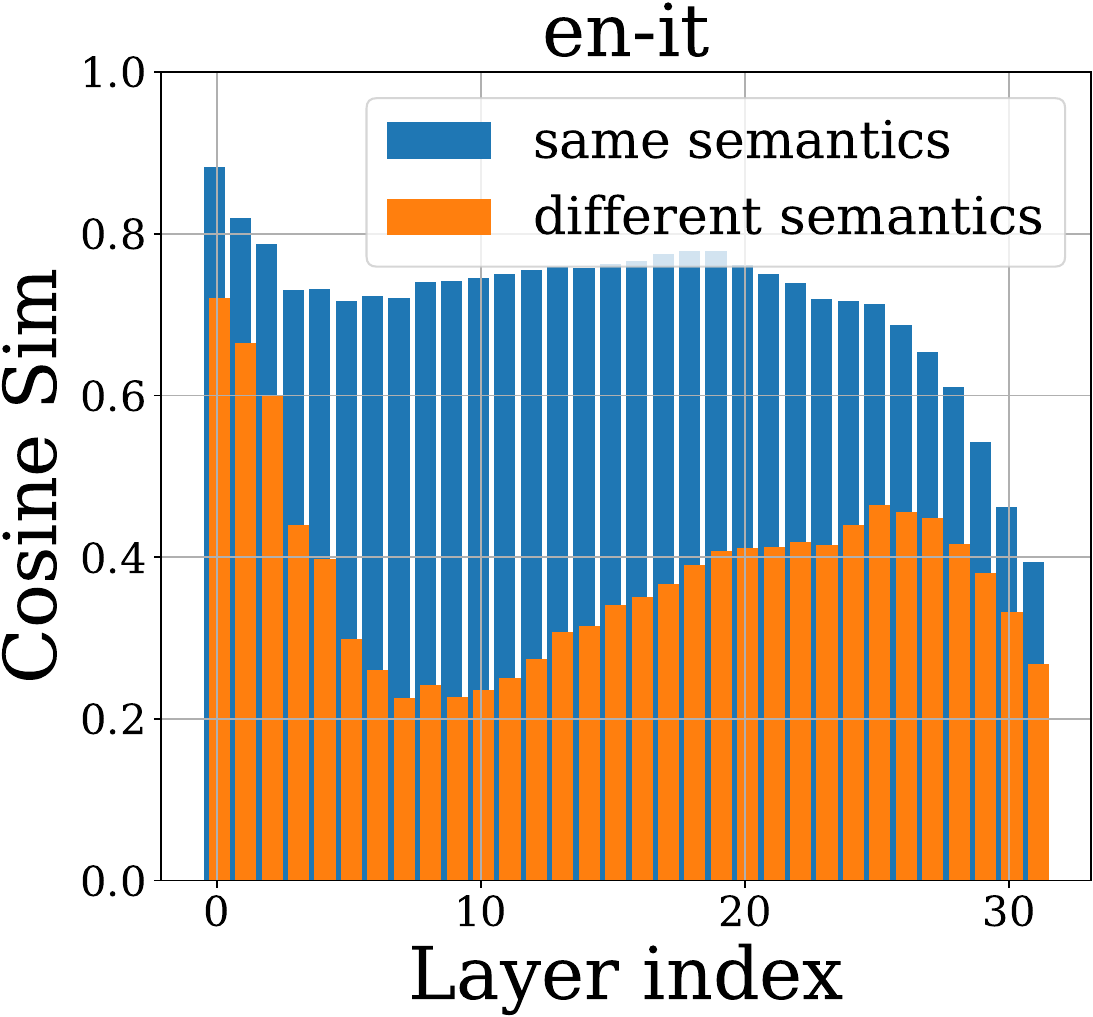}

  \begin{minipage}{0.23\linewidth}\centering en-ja (aya)\end{minipage}
  \begin{minipage}{0.23\linewidth}\centering en-nl (aya)\end{minipage}
  \begin{minipage}{0.23\linewidth}\centering en-ko (aya)\end{minipage}
  \begin{minipage}{0.23\linewidth}\centering en-it (aya)\end{minipage}

  \caption{\textbf{Similarity of hidden states across layers.}}
  \label{fig:appendix:hs_sim_all_models}
\end{figure*}
% figure: act_sim all langs all models.
\begin{figure*}[t]
  \centering

  \includegraphics[width=0.23\linewidth]{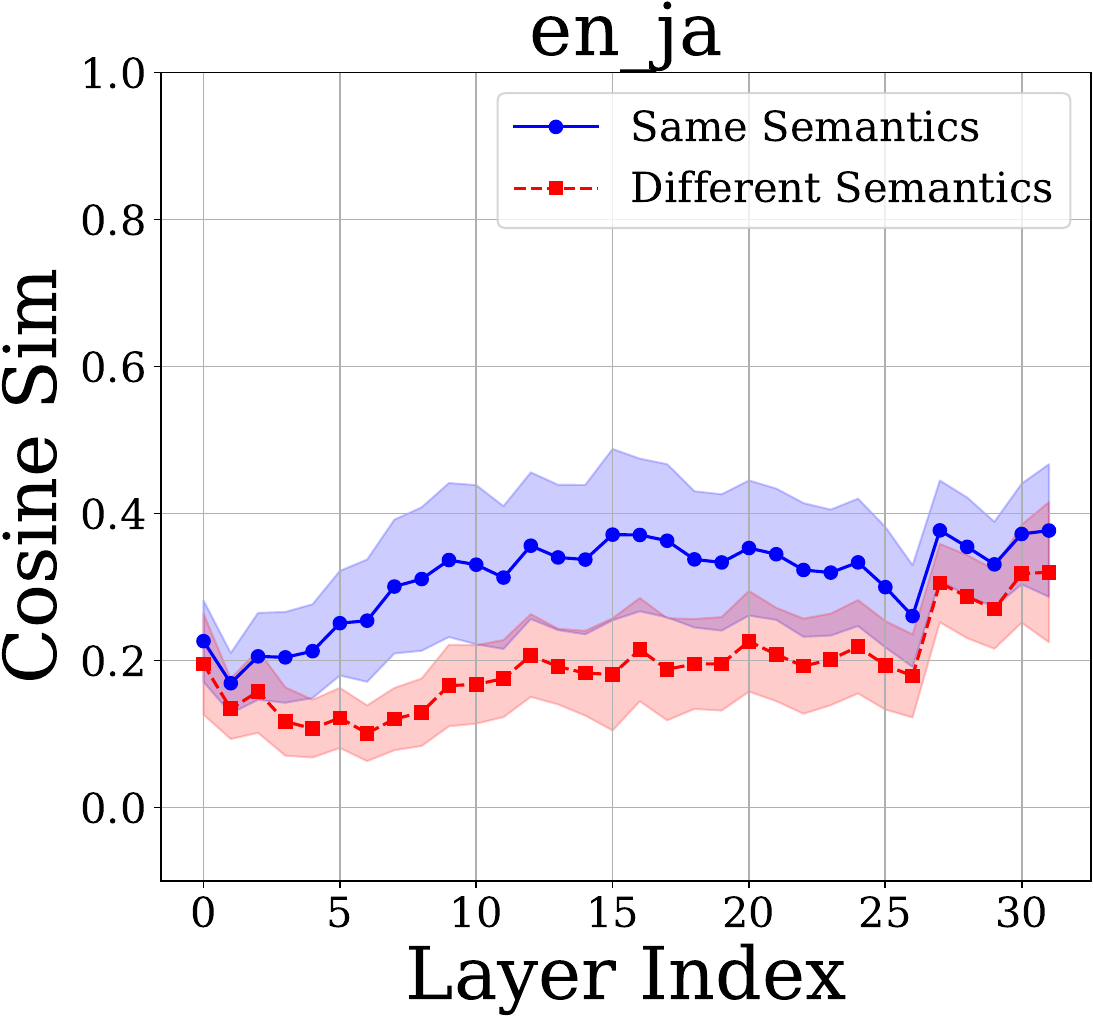}
  \includegraphics[width=0.23\linewidth]{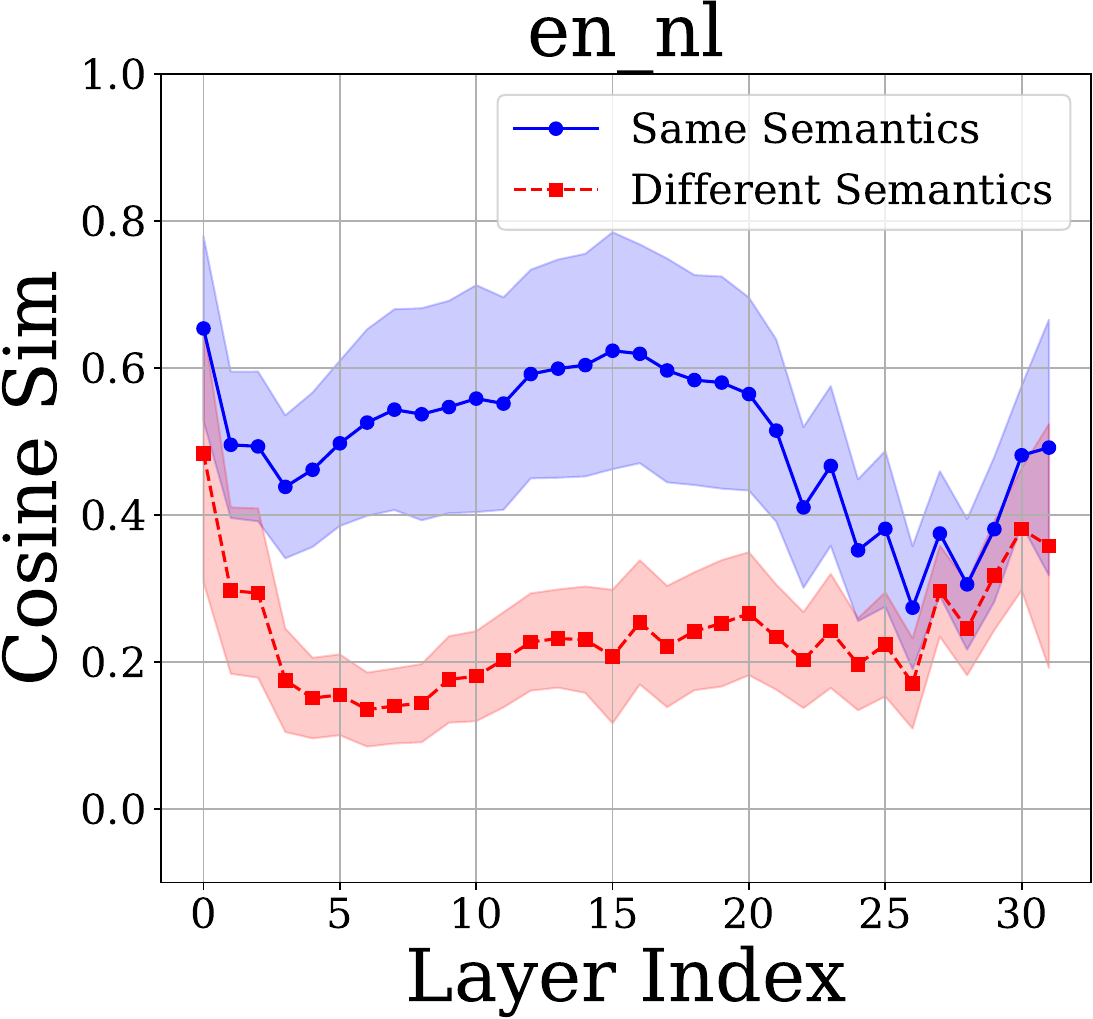}
  \includegraphics[width=0.23\linewidth]{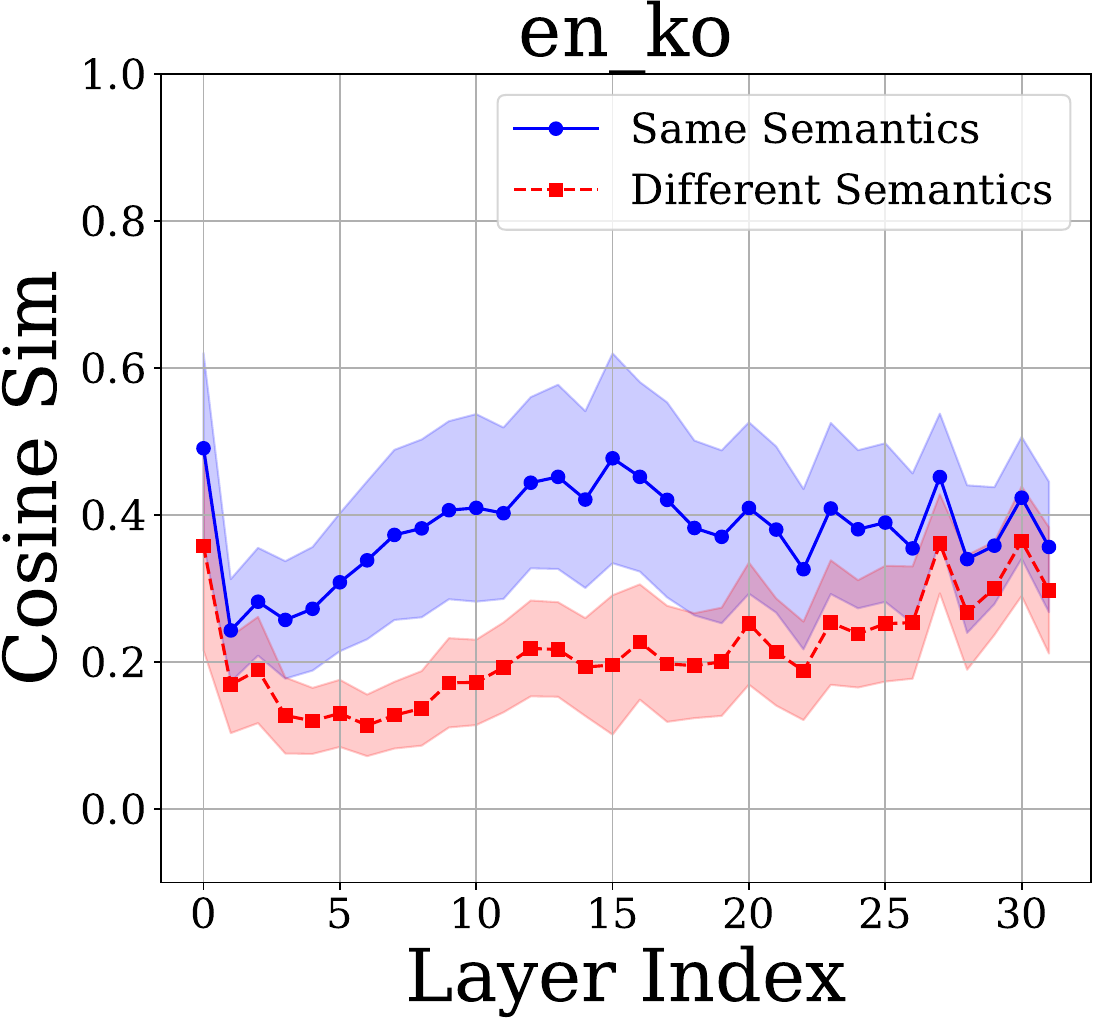}
  \includegraphics[width=0.23\linewidth]{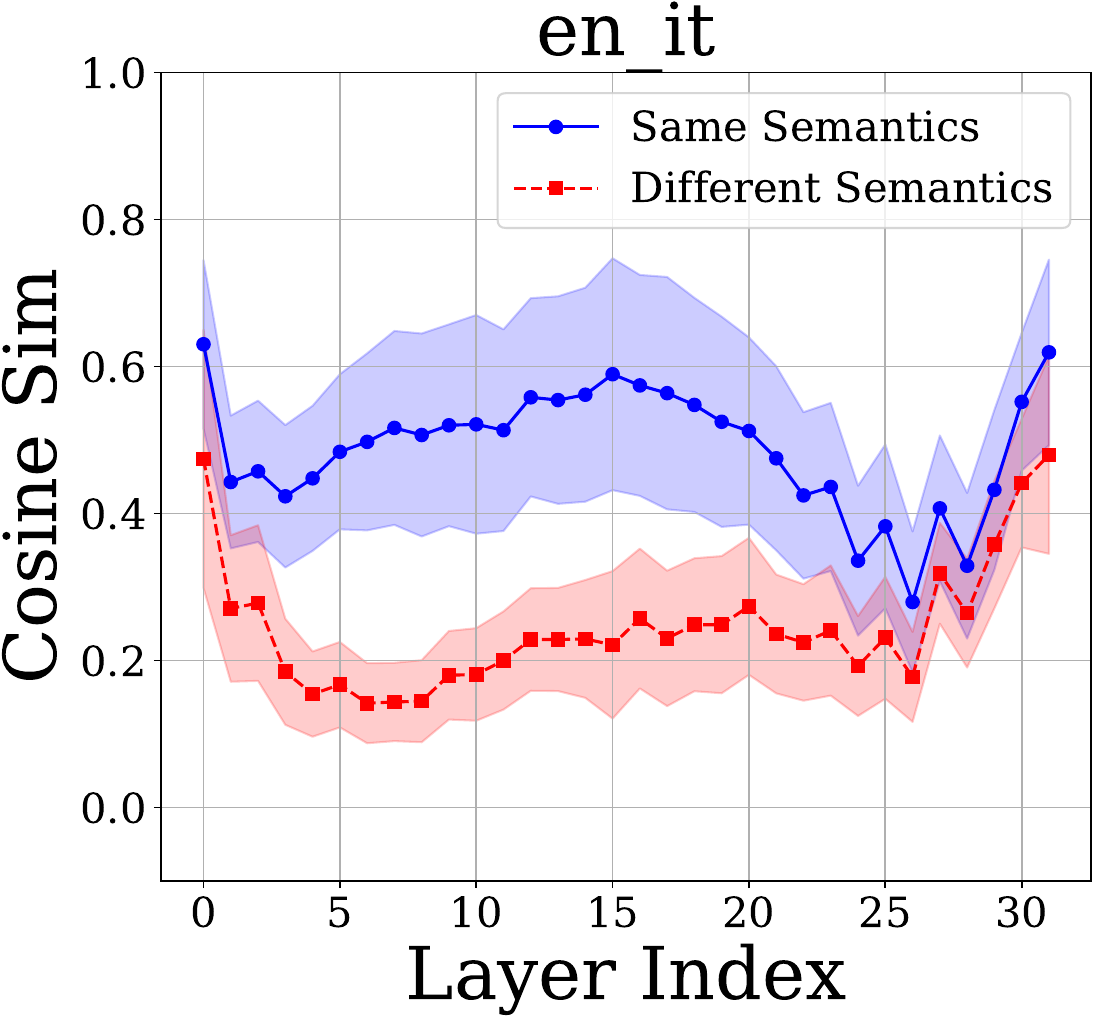}

  \begin{minipage}{0.23\linewidth}\centering en-ja (llama3)\end{minipage}
  \begin{minipage}{0.23\linewidth}\centering en-nl (llama3)\end{minipage}
  \begin{minipage}{0.23\linewidth}\centering en-ko (llama3)\end{minipage}
  \begin{minipage}{0.23\linewidth}\centering en-it (llama3)\end{minipage}

  \includegraphics[width=0.23\linewidth]{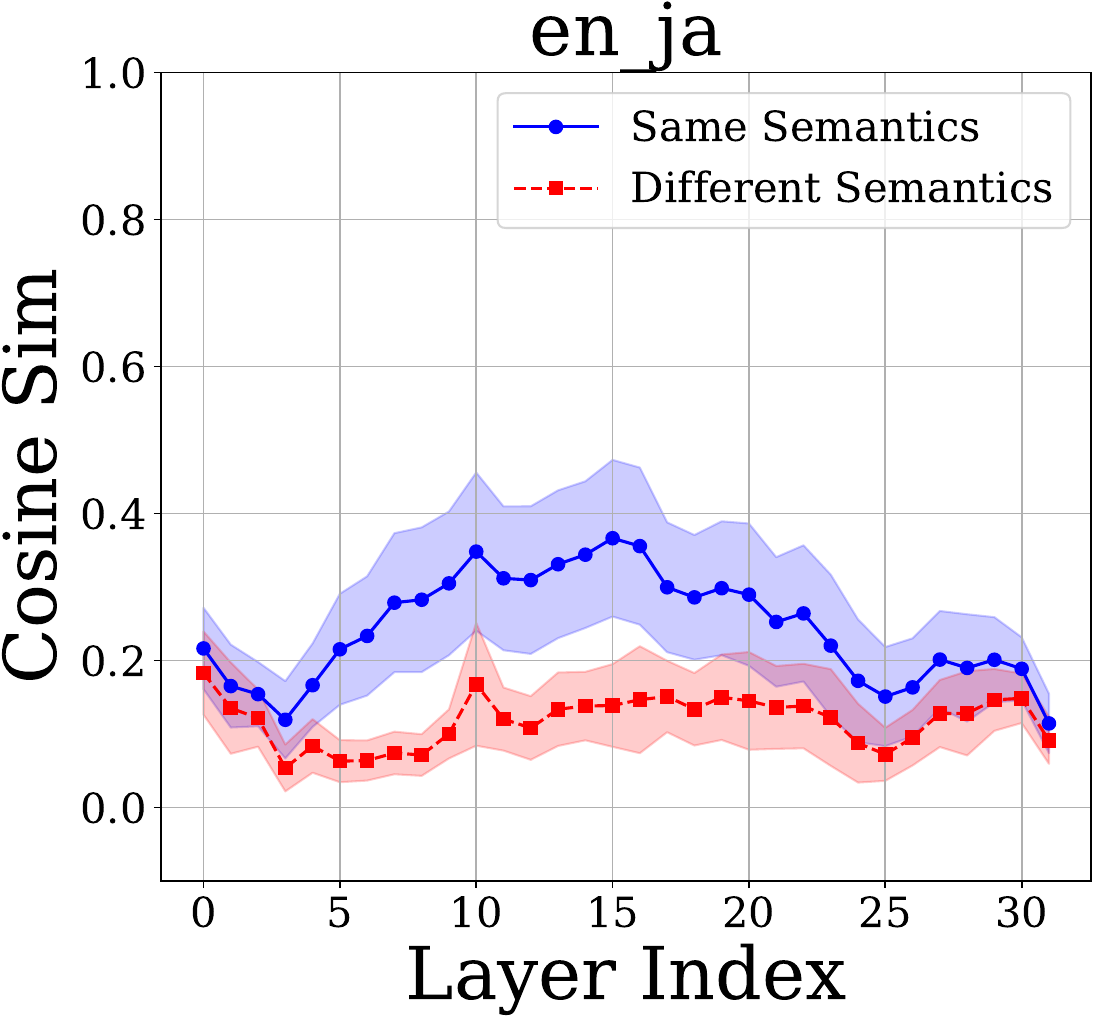}
  \includegraphics[width=0.23\linewidth]{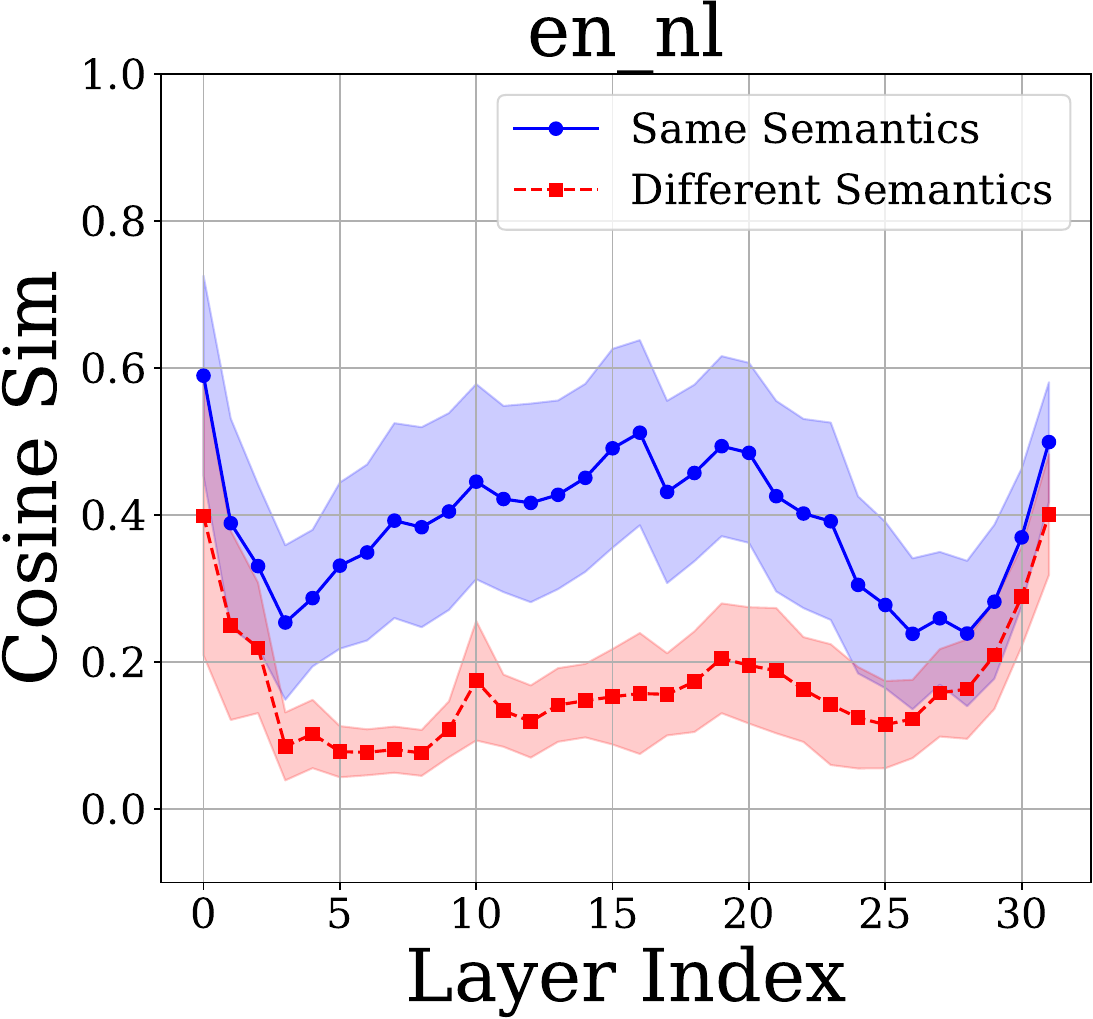}
  \includegraphics[width=0.23\linewidth]{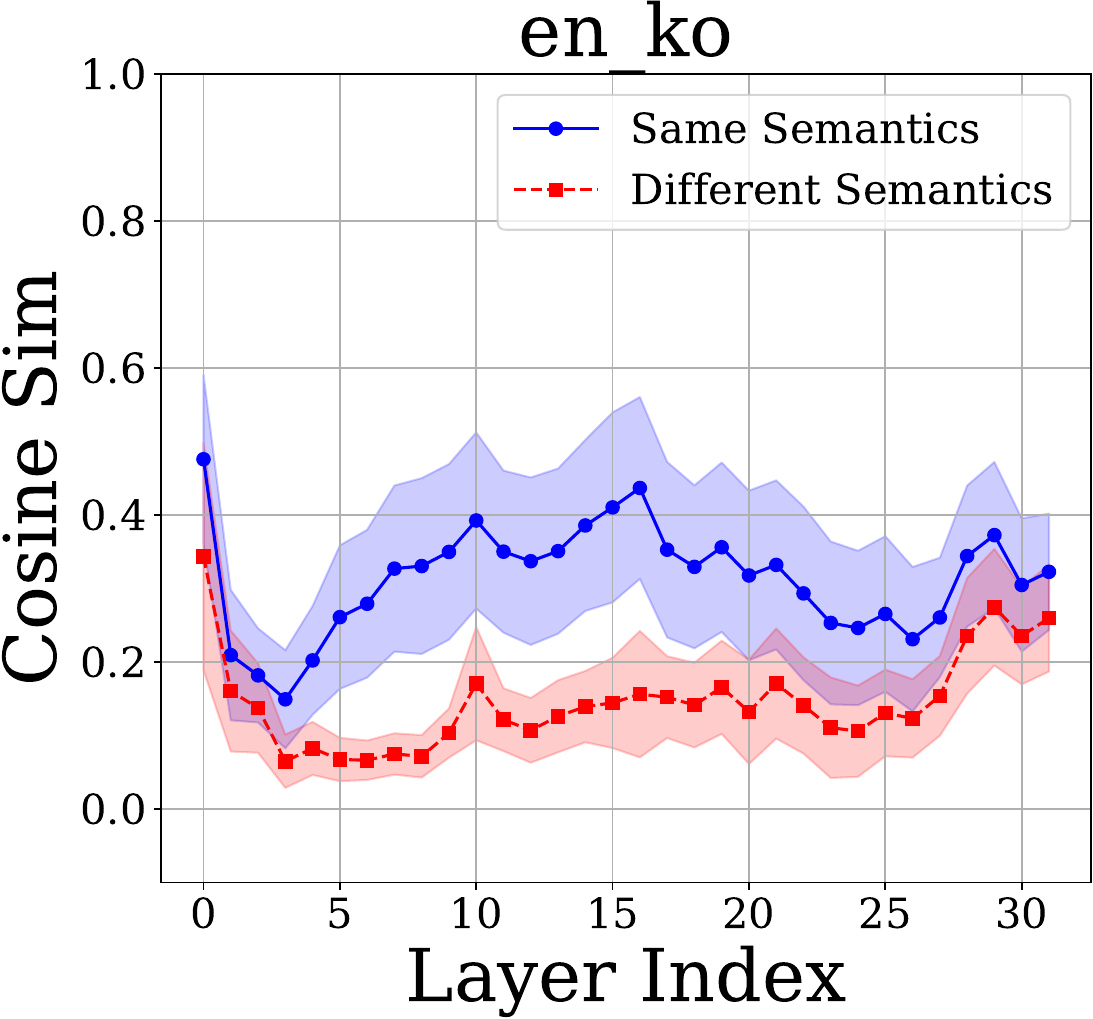}
  \includegraphics[width=0.23\linewidth]{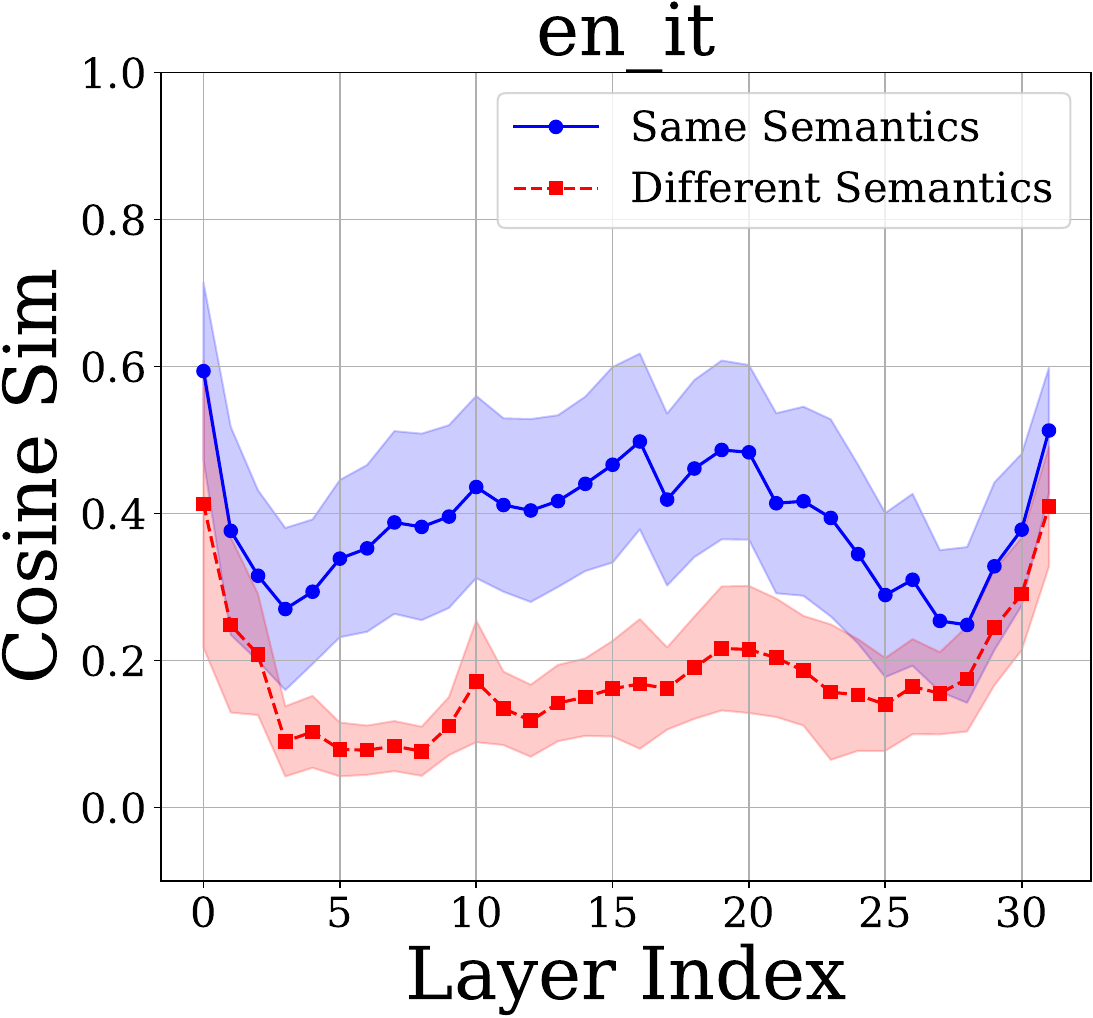}

  \begin{minipage}{0.23\linewidth}\centering en-ja (mistral)\end{minipage}
  \begin{minipage}{0.23\linewidth}\centering en-nl (mistral)\end{minipage}
  \begin{minipage}{0.23\linewidth}\centering en-ko (mistral)\end{minipage}
  \begin{minipage}{0.23\linewidth}\centering en-it (mistral)\end{minipage}

  \includegraphics[width=0.23\linewidth]{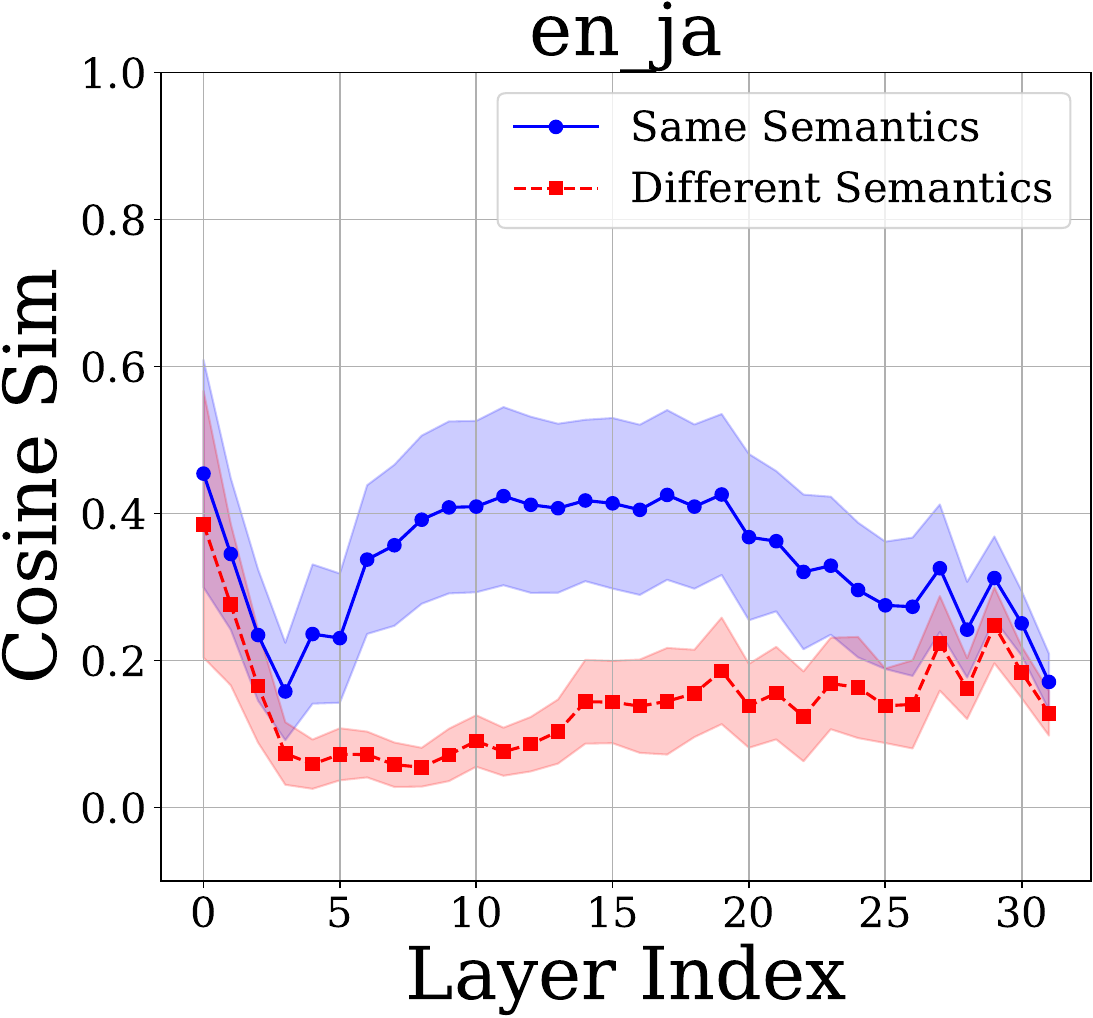}
  \includegraphics[width=0.23\linewidth]{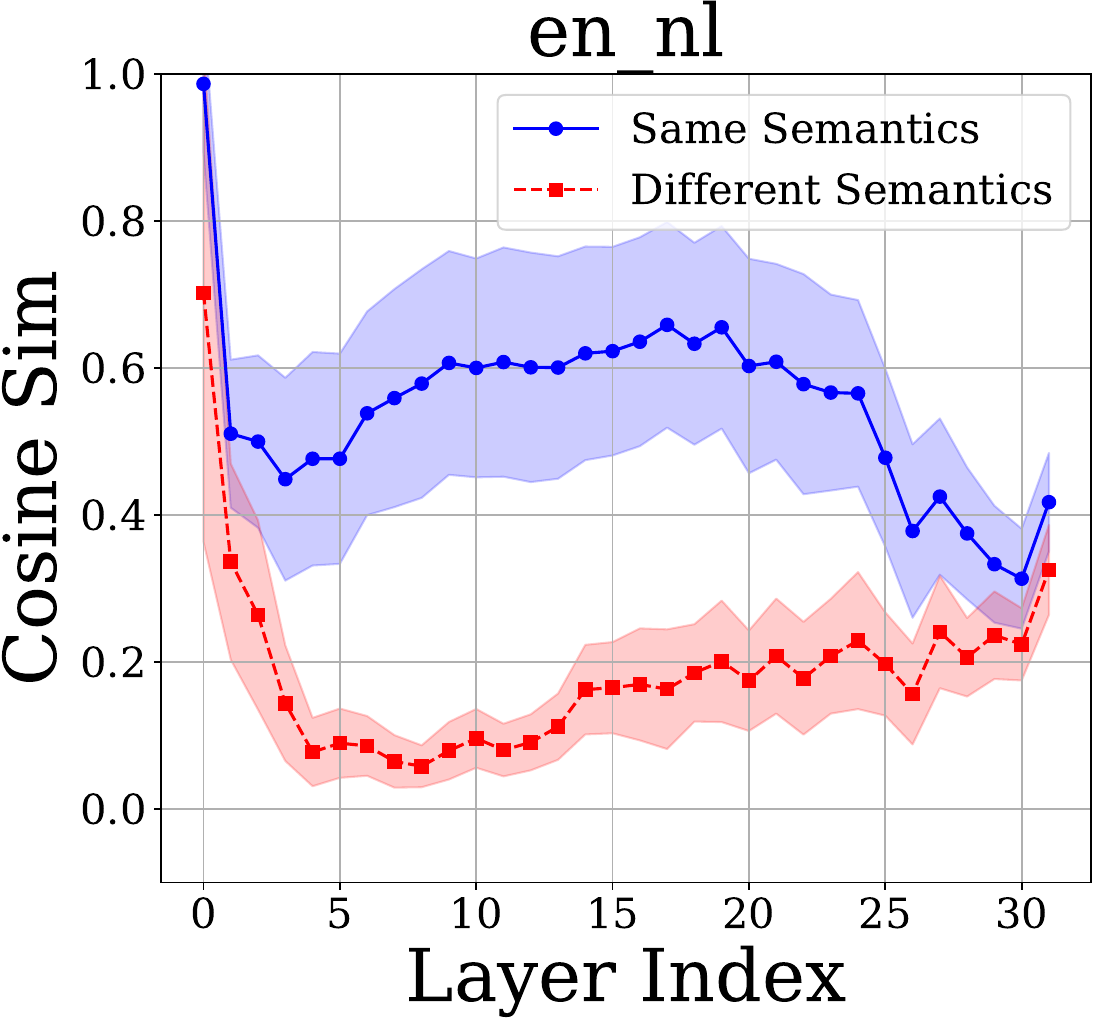}
  \includegraphics[width=0.23\linewidth]{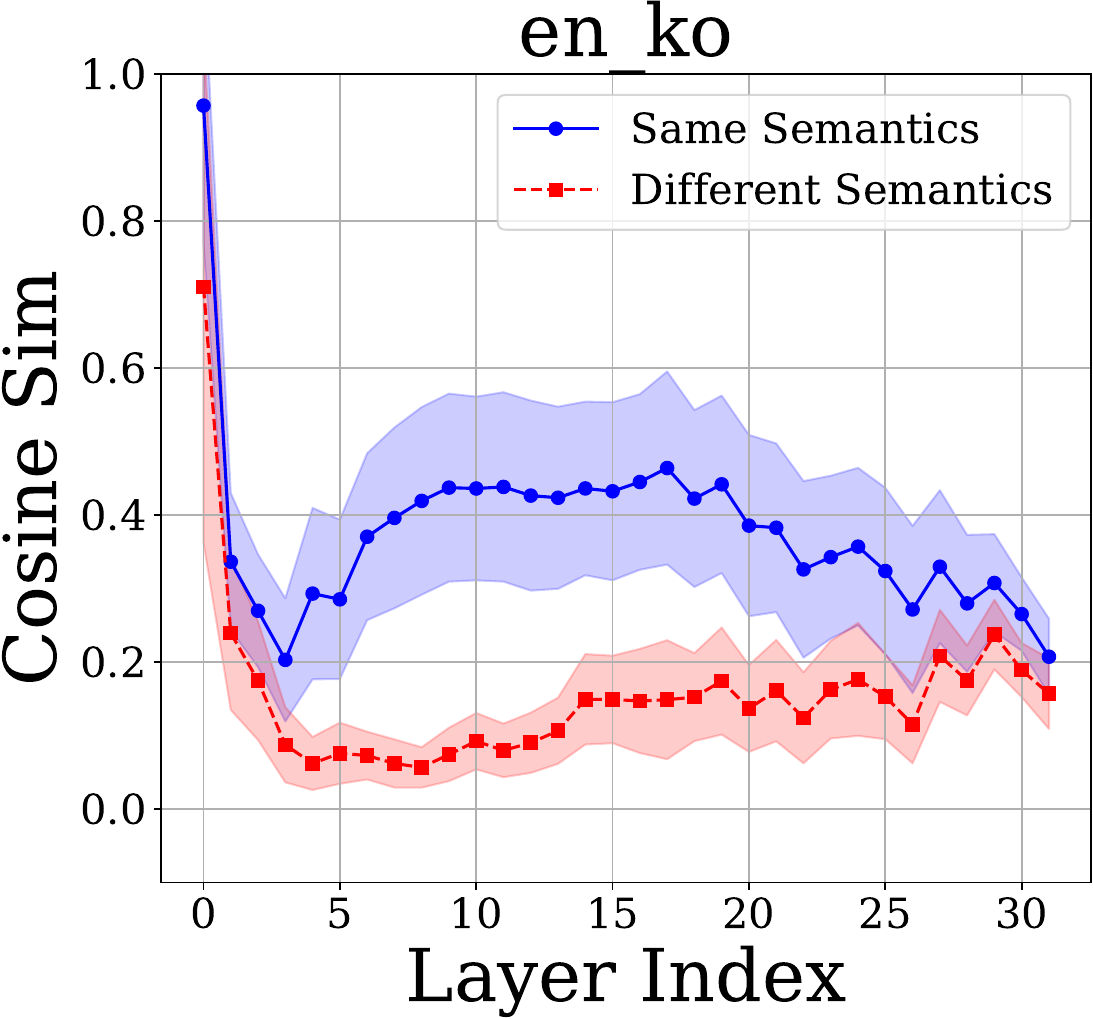}
  \includegraphics[width=0.23\linewidth]{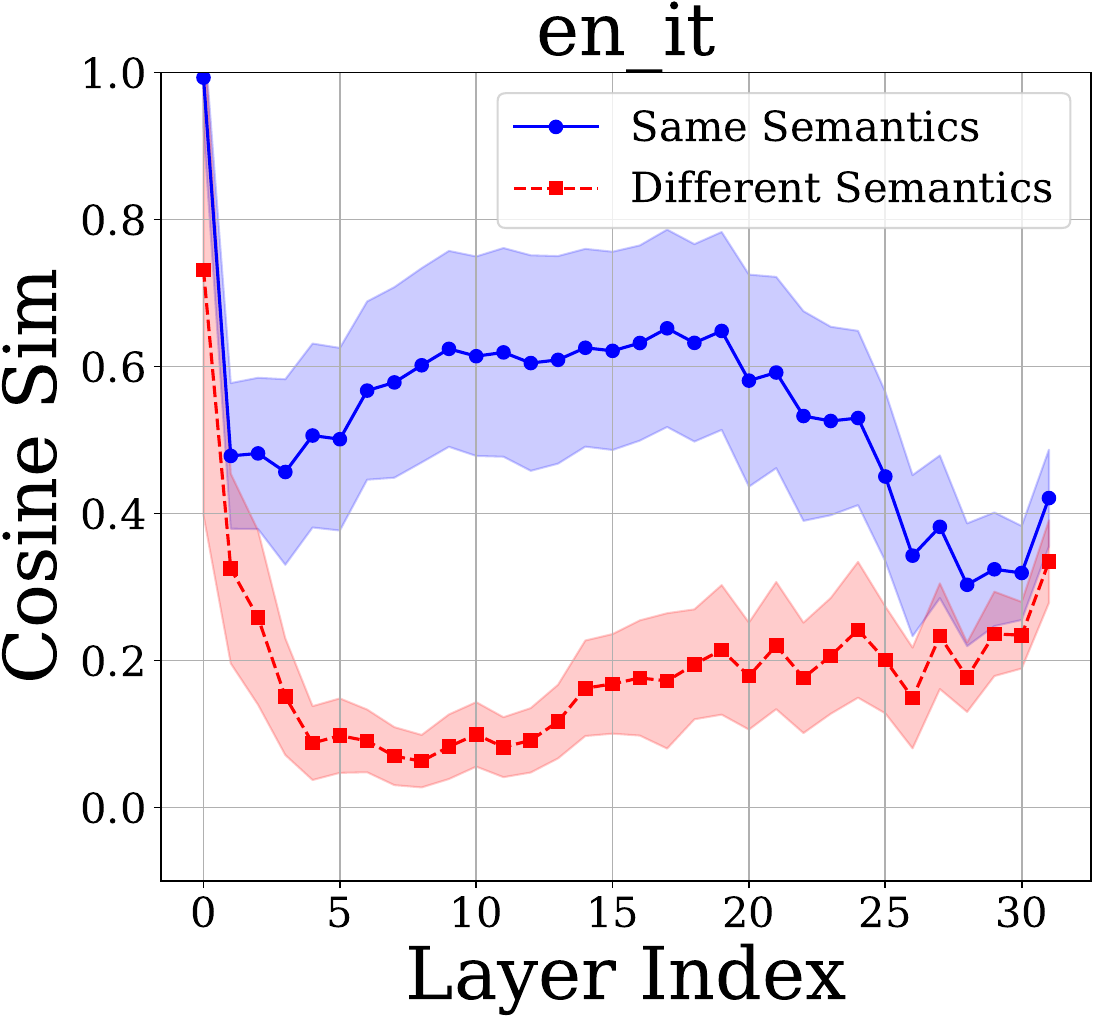}

  \begin{minipage}{0.23\linewidth}\centering en-ja (aya)\end{minipage}
  \begin{minipage}{0.23\linewidth}\centering en-nl (aya)\end{minipage}
  \begin{minipage}{0.23\linewidth}\centering en-ko (aya)\end{minipage}
  \begin{minipage}{0.23\linewidth}\centering en-it (aya)\end{minipage}

  \caption{\textbf{Similarity of activation patterns in MLP module across layers.}}
  \label{fig:appendix:act_sim_all_models}
\end{figure*}
\paragraph{Similar Hidden States for Similar Semantics in Middle Layers.}
\label{sec:appendix: Cosine Similarities of Internal Representations for Other Language Pairs and Models}
To quantitatively investigate the similarity of hidden states across languages, given $\{(\bm{s}_i^{l},\bm{t}_i^{l})\}_{i=1}^N$, a set of $l$-th layer hidden states for parallel sentences, and $\{(\bm{u}_i^{l},\bm{v}_i^{l})\}_{i=1}^N$, the ones for \emph{non}-parallel sentences, we show the difference between $\frac{1}{N}\sum_i \mathrm{cos}(\bm{s}_i^{l}, \bm{t}_i^{l})$ and $\frac{1}{N}\sum_i \mathrm{cos}(\bm{u}_i^{l}, \bm{v}_i^{l})$ at each layer in Fig.~\ref{fig:appendix:hs_sim_all_models}. The $\bm{u}_i$ are fixed to the set of hidden states for English sentences.
Clearly, models process similarly for parallel sentence pairs compared to non-parallel ones, especially in the middle layers. The relatively pronounced divergence in similarity between parallel and non-parallel ones observed in the middle layers suggests that, at these stages, the inputs are processed within a shared semantic latent space, wherein semantically similar sentences are more likely to be mapped to proximate locations. On the other hand, the relatively small divergence observed in the initial and final layers might be attributed to the model operating within language-specific representational latent spaces at these stages, where the primary focus is on processing the input and generating the output in the specific language. We found that this tendencies are consistent across other pair patterns and models.

\paragraph{Similar Activations for Similar Semantics in Middle Layers.}
Interpreting the MLP as key-value memory, activation values reflect how strongly the model accesses value vectors, which encode concepts across languages \citep{geva2021, geva2022, knowledge_neurons, journey_knowledge_neuron}. From this perspective, relatively high similarity in activation patterns for parallel sentences compared to non-parallel ones suggests the model encodes similar concepts across languages, supporting the existence of a shared semantic latent space.

Fig.~\ref{fig:appendix:act_sim_all_models} show the similarity of the activation patterns across layers with the same inputs as similarity measurement of hidden states, which is calculated as $\frac{1}{N}\sum_i \mathrm{cos}(\bm{\alpha}_{\mathrm{en},i}^l, \bm{\alpha}_{\mathrm{L2},i}^l)$ 
where $\bm{\alpha}^{l}$ denotes the activation values vector in $l$-th layer described in Eq.~\ref{eq:weighted_sum}, and $n$ represents the number of sample sentence pairs. L2 denotes the languages other than English.
Similar to the results for hidden states similarity, the difference in similarity between parallel and non-parallel inputs is particularly prominent in the middle layers, supporting the existence of a shared semantic latent space. This tendency is consistent across other language pairs and models, and it also aligns with the findings of~\citet{coling2025-converging}, despite differences in experimental settings.

\subsection{Investigating Linear Separability between Parallel and Non-Parallel Inner Representations with Logistic Regression Model}
\label{sec:appendix: logistic_regression}
\begin{figure*}
  \centering
  % 1行目
  \begin{minipage}[b]{\linewidth}
    \centering
    \includegraphics[width=\linewidth]{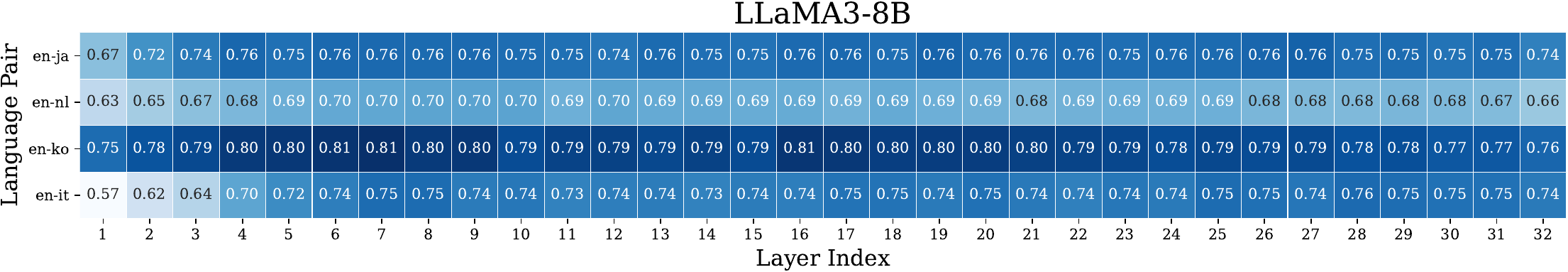}
    % \subcaption{Japanese}
  \end{minipage}
  \begin{minipage}[b]{\linewidth}
    \centering
    \includegraphics[width=\linewidth]{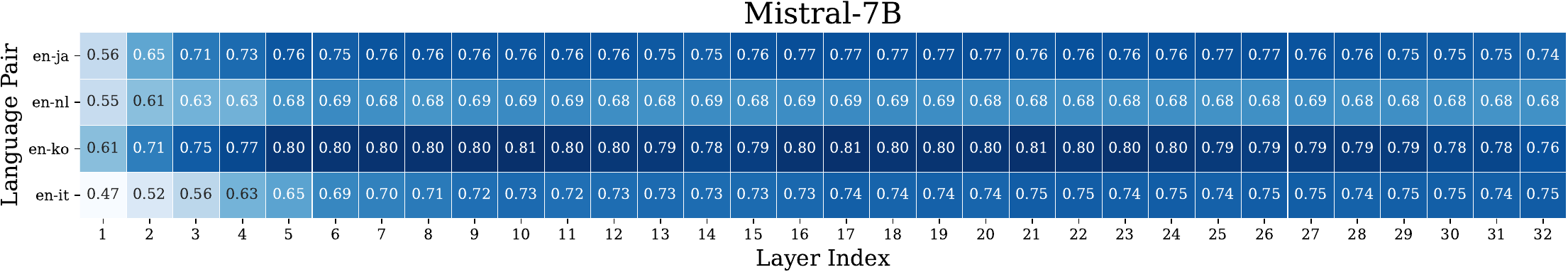}
    % \subcaption{Dutch}
  \end{minipage}
  \begin{minipage}[b]{\linewidth}
    \centering
    \includegraphics[width=\linewidth]{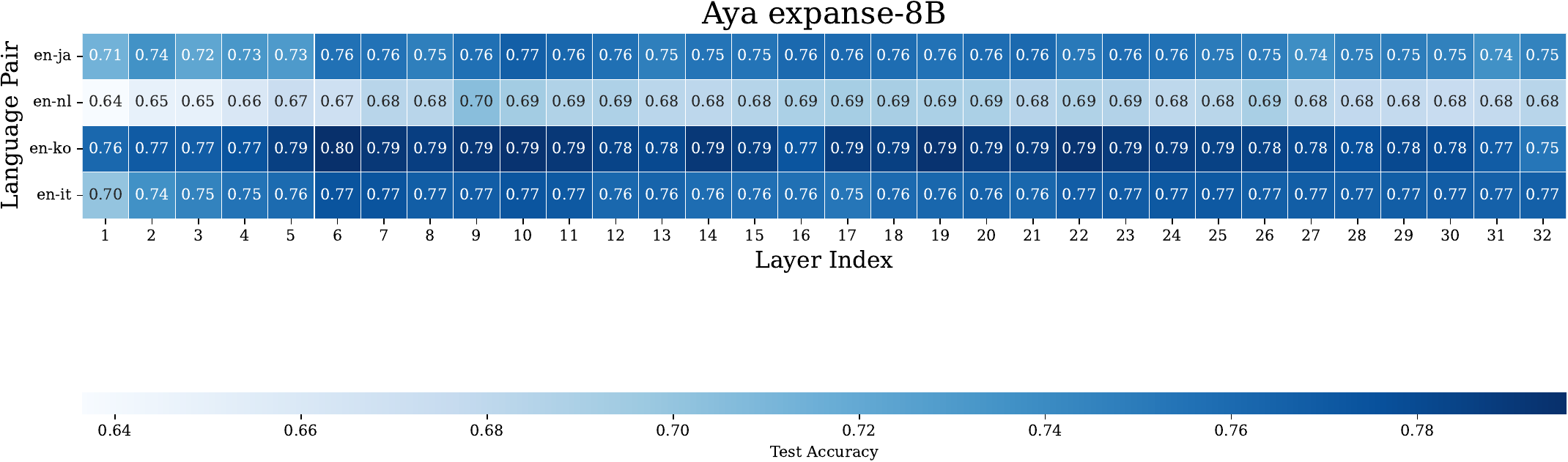}
    % \subcaption{Dutch}
  \end{minipage}

  \caption{\textbf{Test accuracy of logistic regression model trained on parallel semantic features and non-parallel semantic features for each layer.}
  The y-axis indicates the language pairs, and the x-axis denotes the layer indices.
  }
  \label{fig:logistic_regression}
\end{figure*}

To investigate linear separability between parallel and non-parallel pairs of representations, we trained a logistic regression model on the hidden representations from each layer. For each sentence pair (parallel or non-parallel), the input features were constructed by concatenating their hidden states. Parallel pairs were labeled as \texttt{1}, and non-parallel pairs as \texttt{0} (1000 samples for \texttt{label 1} and \texttt{label 0}, respectively). We employed stratified 10-fold cross-validation to ensure balanced evaluation across classes.
As shown in Fig.~\ref{fig:logistic_regression}, except for the initial few layers, most layers achieve an test accuracy of approximately over 70\%. Moreover, we observe that layers with higher similarity between hidden language latent spaces — where the shared semantic latent space appears to exist, as shown in Figs.~\ref{fig:mutual_knn},~\ref{fig:appendix:kernel-based sim while deactivating Type-1 k=10}, and~\ref{fig:appendix:kernel-based sim while deactivating Type-1 k=5} — tend to yield higher classification accuracy. This result suggests that the model can effectively classify whether a pair of input sentences in two languages is parallel or non-parallel in the middle layers. This results also support the existence of a shared semantic latent space in middle layers.

% figure: logistic regression (all models).

% Appendix: Subspace property via SVD.
\section{Latent Space Property of Hidden States}
\label{sec:appendix:subspace property of hidden states}

\subsection{Dimensionality of Latent Spaces Across Layers}
\label{sec:appendix:dimensionality of subspaces across layers}
% figure: Dimensionality of Subspaces across Layers.
\begin{figure*}[t]
    \centering
    % llama3
    \includegraphics[width=0.32\linewidth]{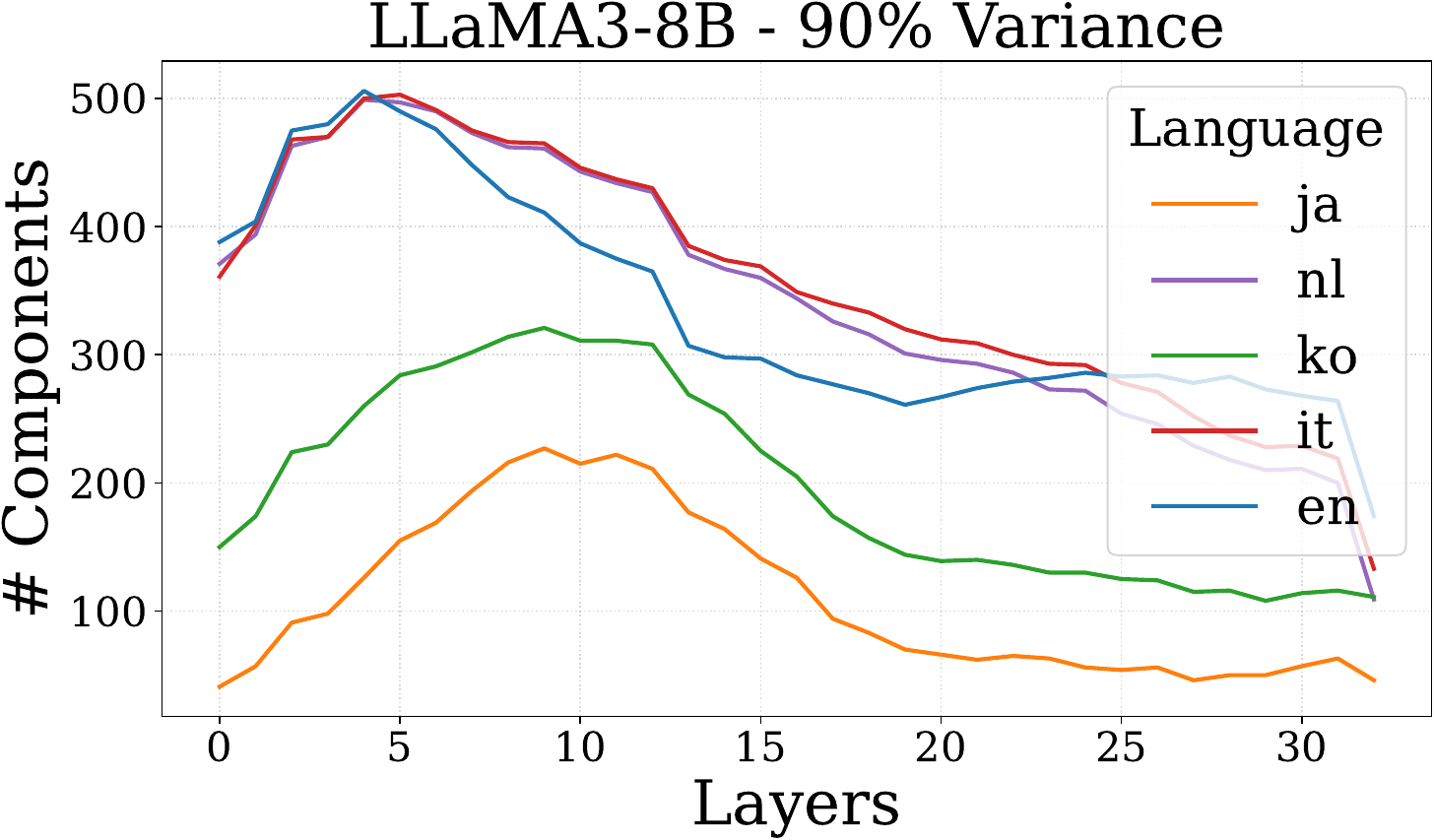}
    \includegraphics[width=0.32\linewidth]{figures/llama3/subspace/llama3_95.pdf}
    \includegraphics[width=0.32\linewidth]{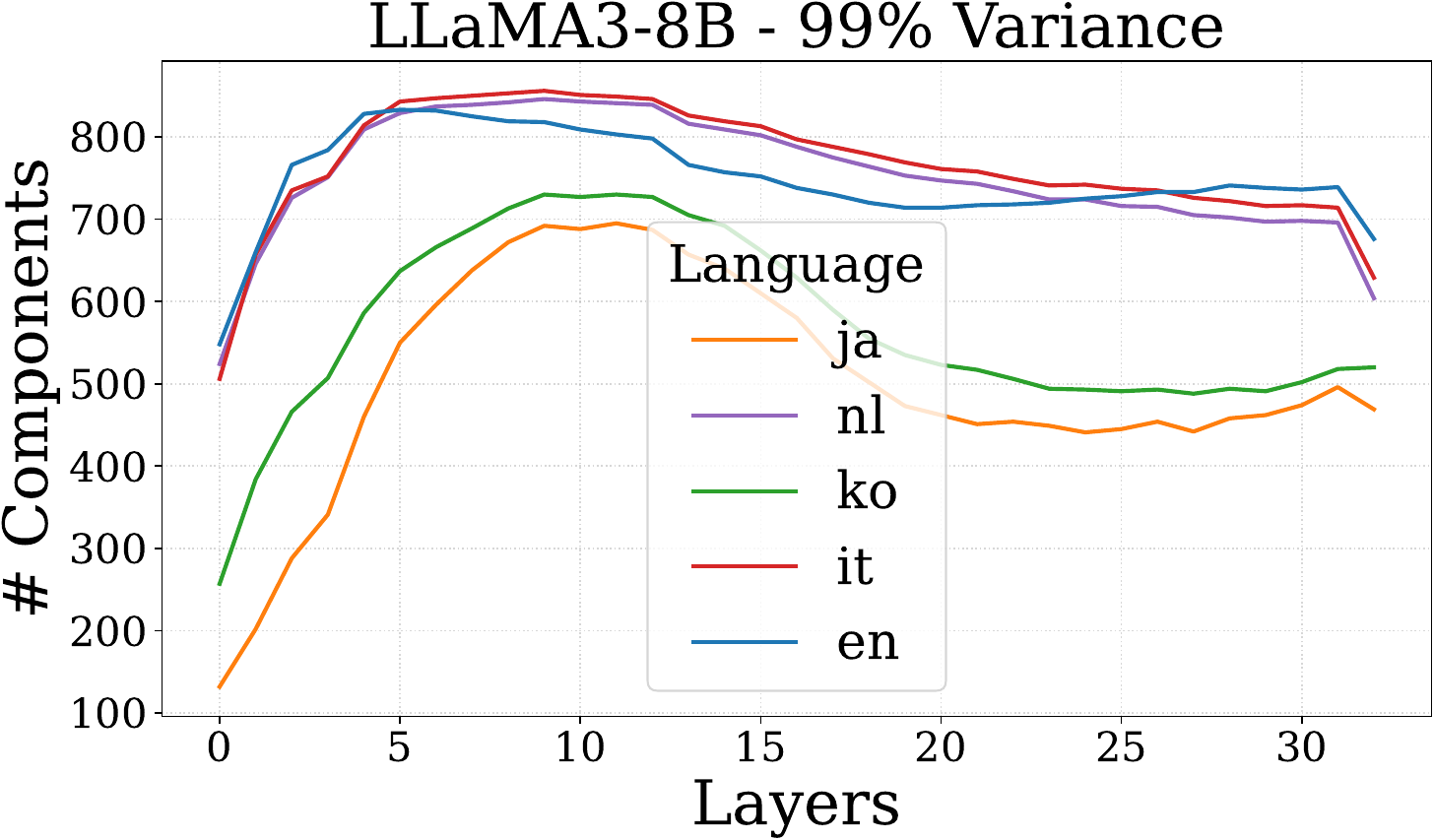}
    
    % mistral
    \includegraphics[width=0.32\linewidth]{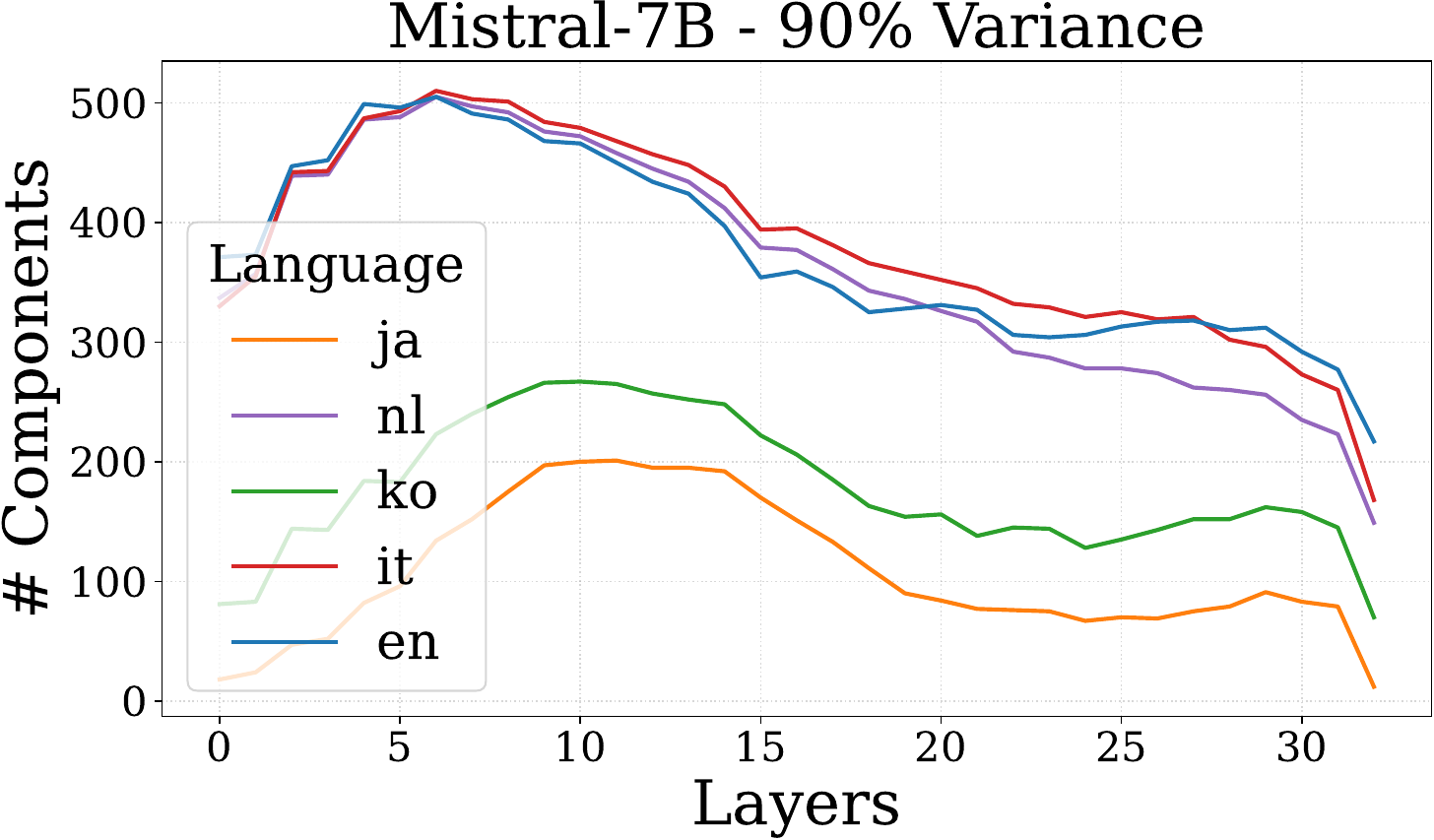}
    \includegraphics[width=0.32\linewidth]{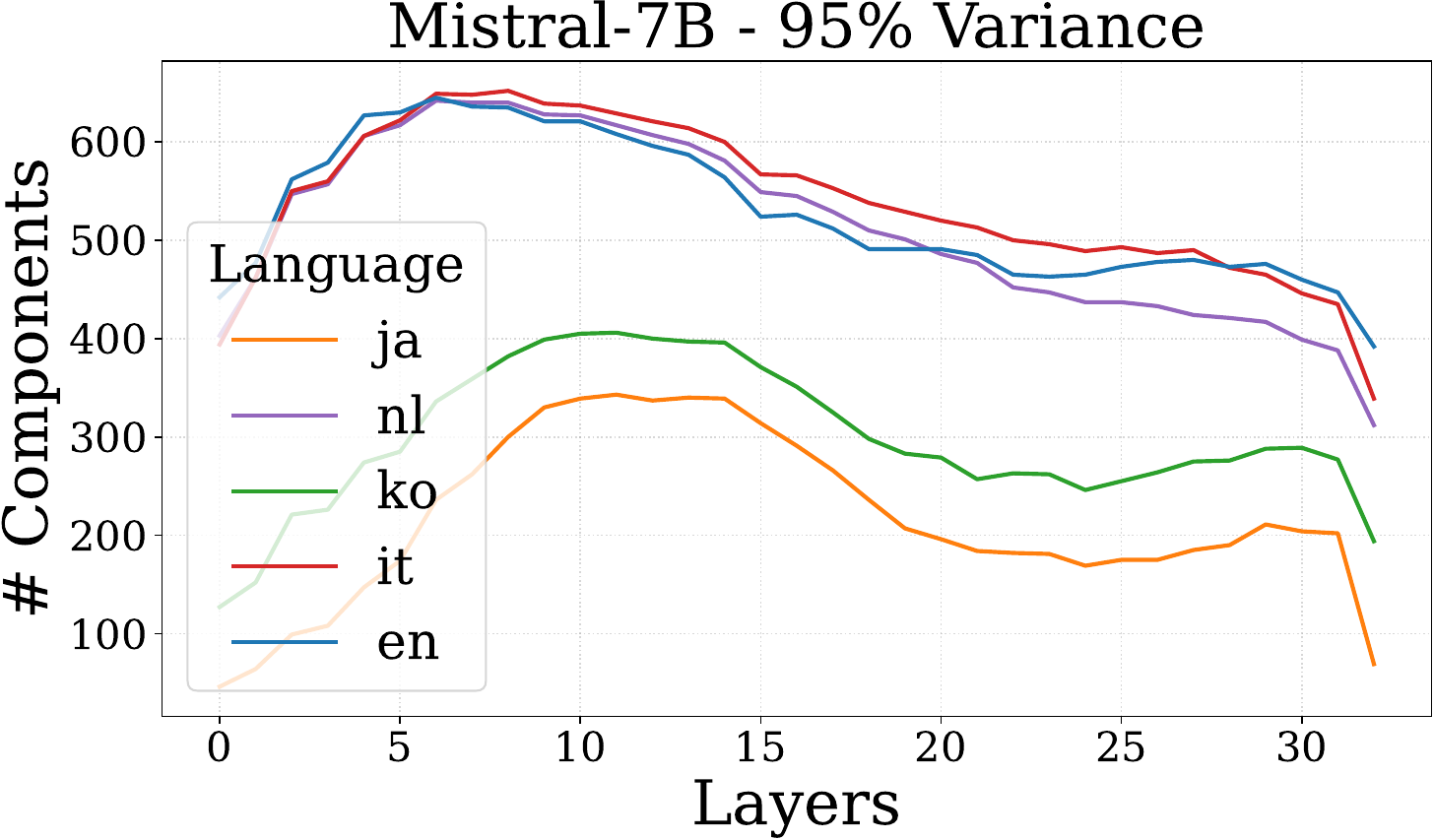}
    \includegraphics[width=0.32\linewidth]{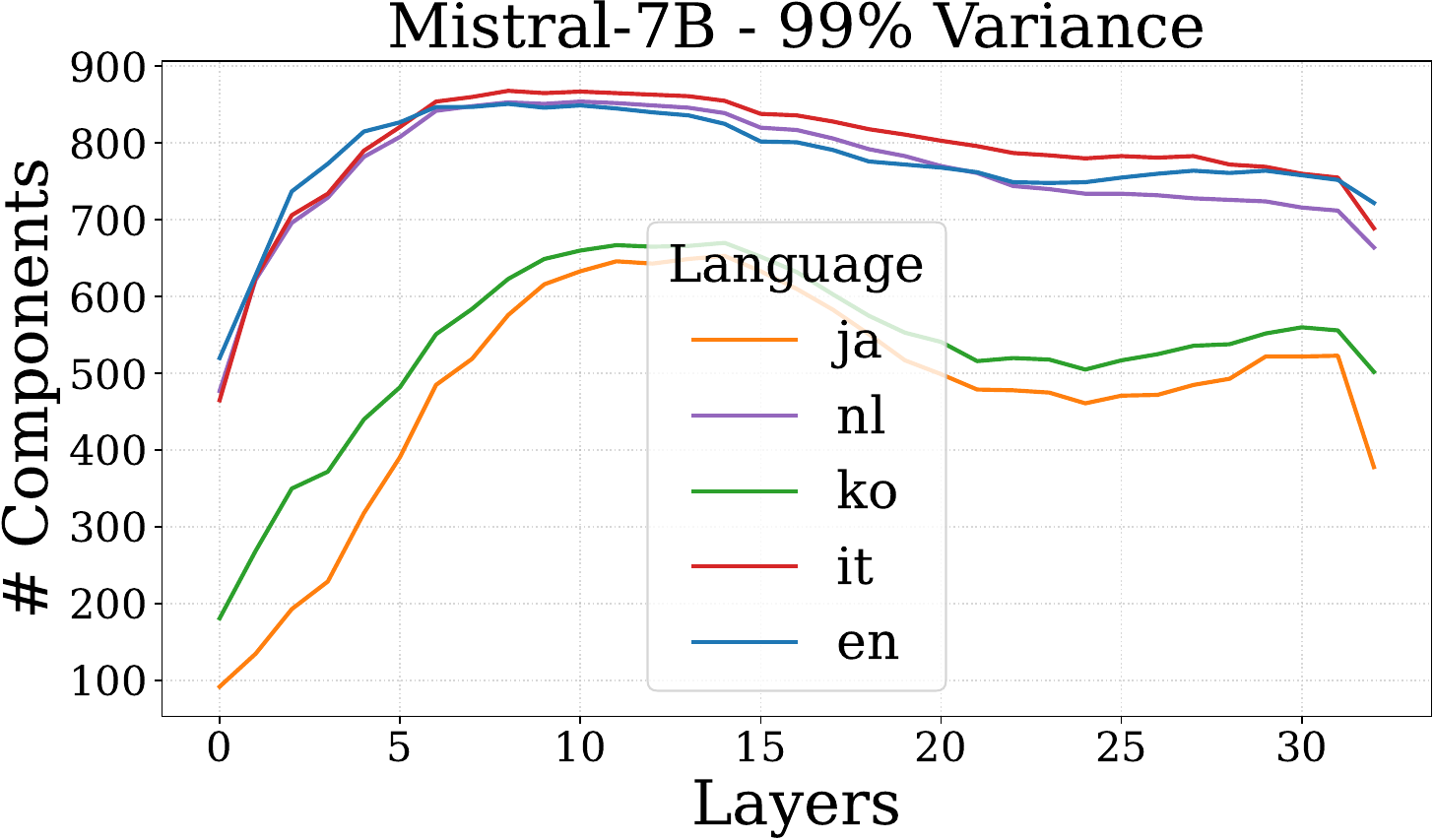}

    % aya
    \includegraphics[width=0.32\linewidth]{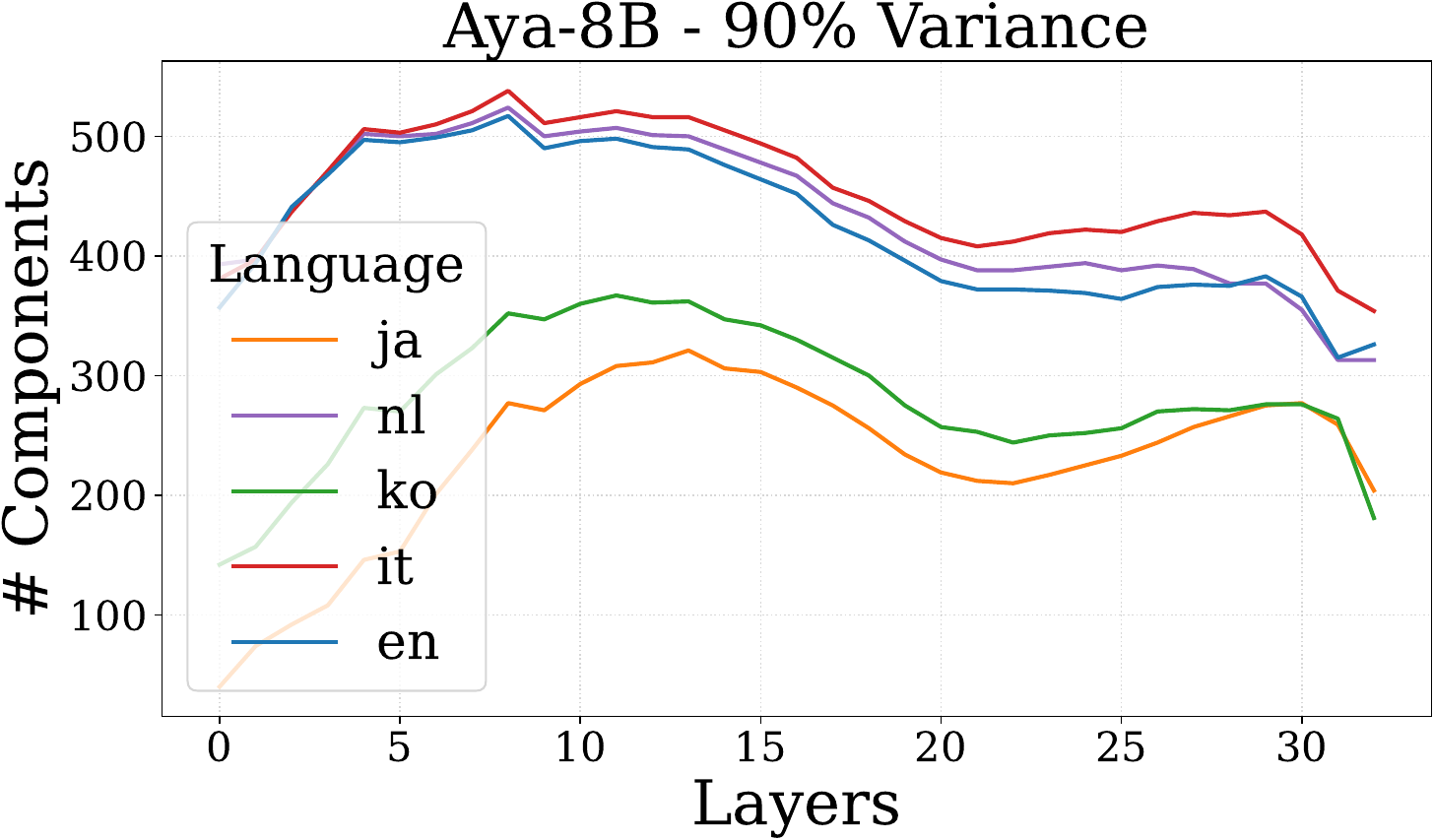}
    \includegraphics[width=0.32\linewidth]{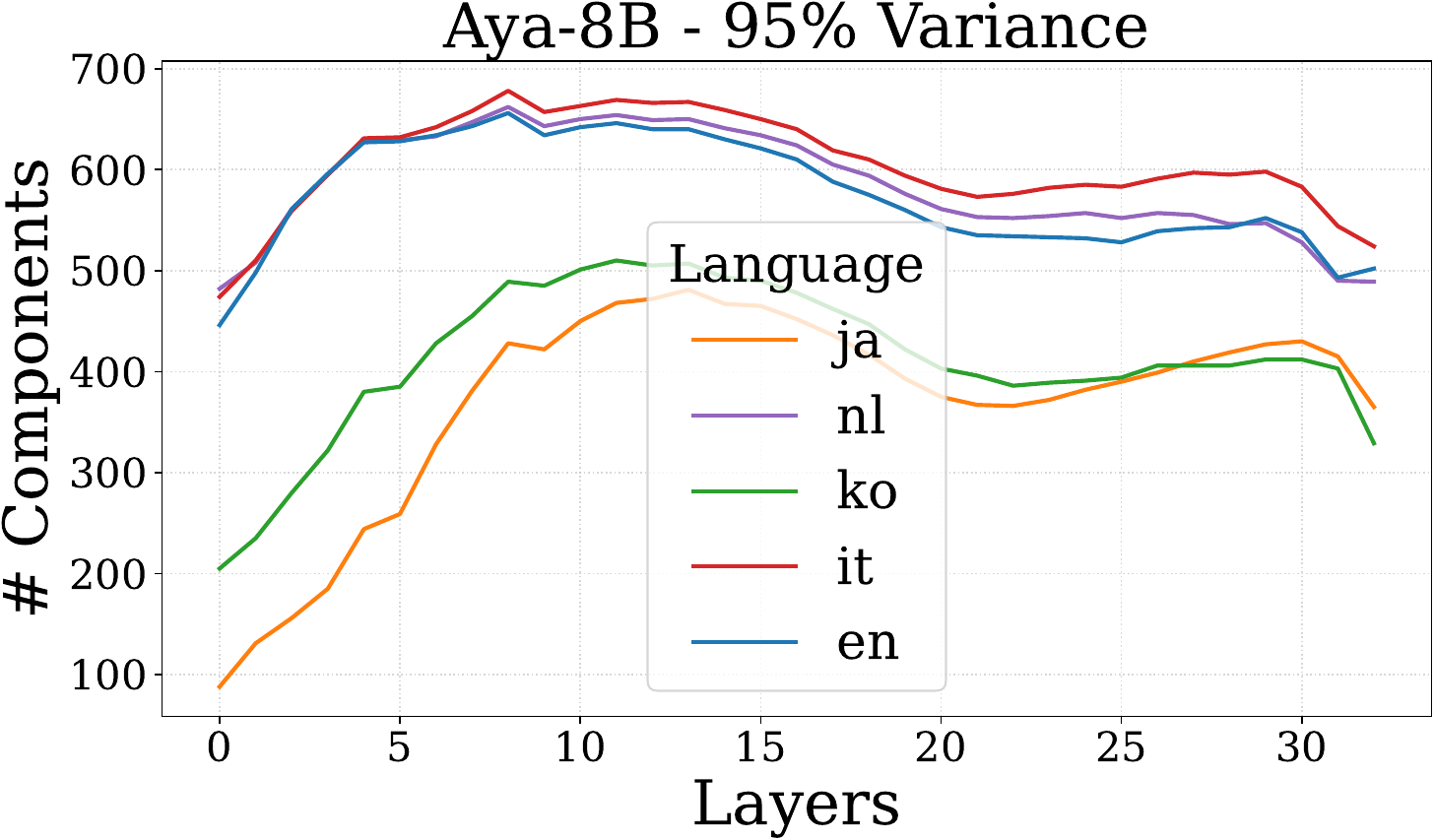}
    \includegraphics[width=0.32\linewidth]{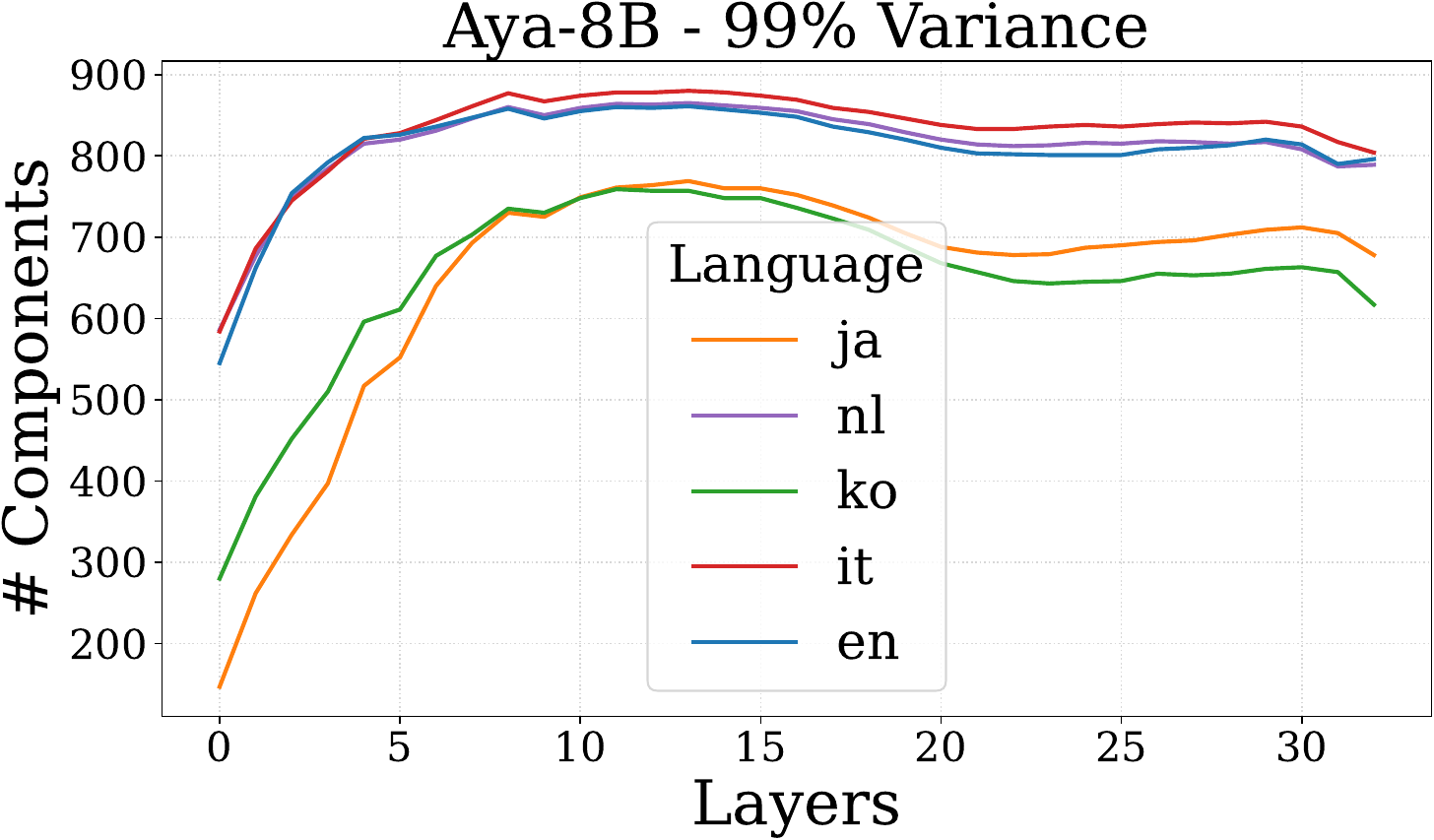}
    
    \caption{\textbf{Estimated dimensionality of latent spaces across layers (LLaMA3-8B, Mistral-7B, and Aya expanse-8B).} The vertical-axis represents the number of singular values (i.e., the number of orthonormal basis) required to explain 90, 95, and 99\% of the variance. It is computed via SVD and cumulative explained variance ratio.}
    \label{fig:appendix:dimensionality of subspaces}
\end{figure*}

Fig.~\ref{fig:appendix:dimensionality of subspaces} shows the estimated dimensionality (i.e, the number of orthonormal basis) of the latent space across languages and layers. As shown, each language latent spaces has lower dimensionality than that of hidden states, indicating they are latent spaces in the hidden state space.
\subsection{Distance among Language Latent Spaces}
\label{sec:appendix:distance among each language subspaces}

% llama3, distance among subspaces, centroids
\begin{figure*}[t]
  \centering

  \includegraphics[width=0.19\linewidth]{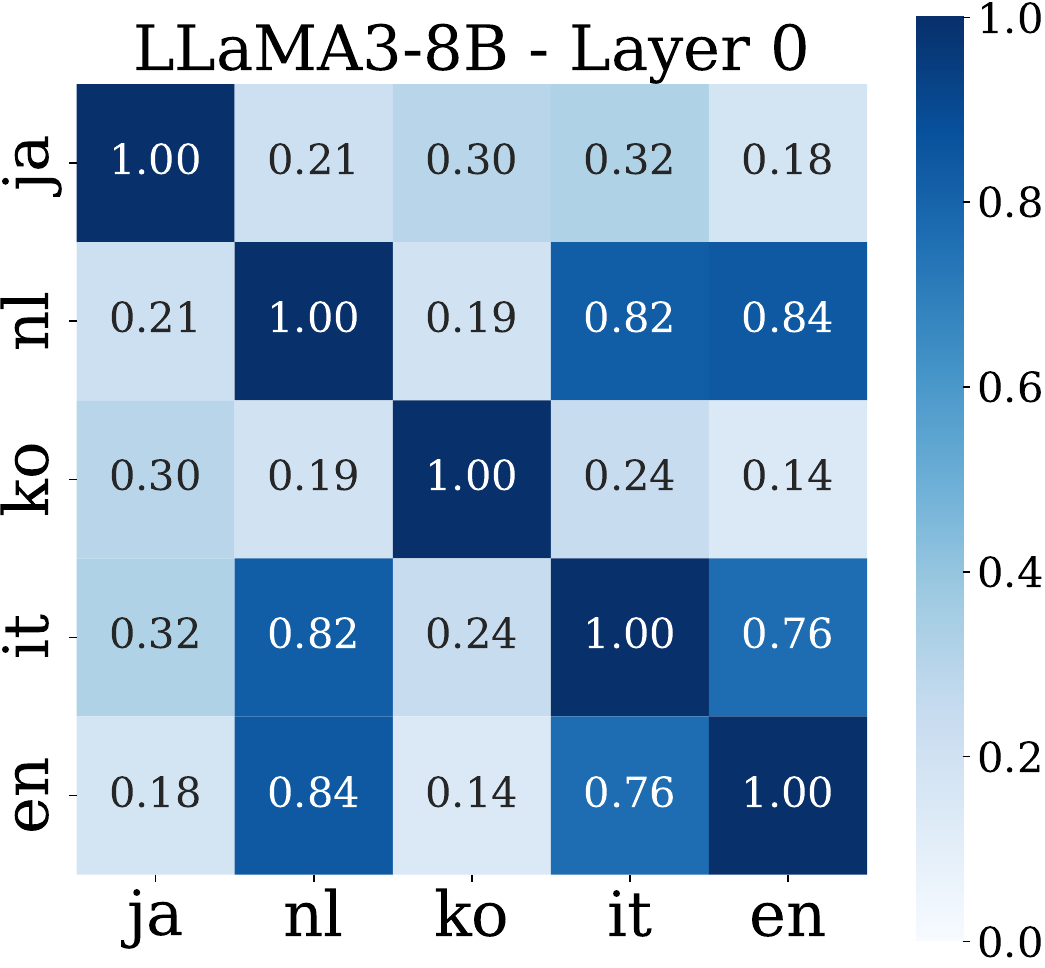}
  \includegraphics[width=0.19\linewidth]{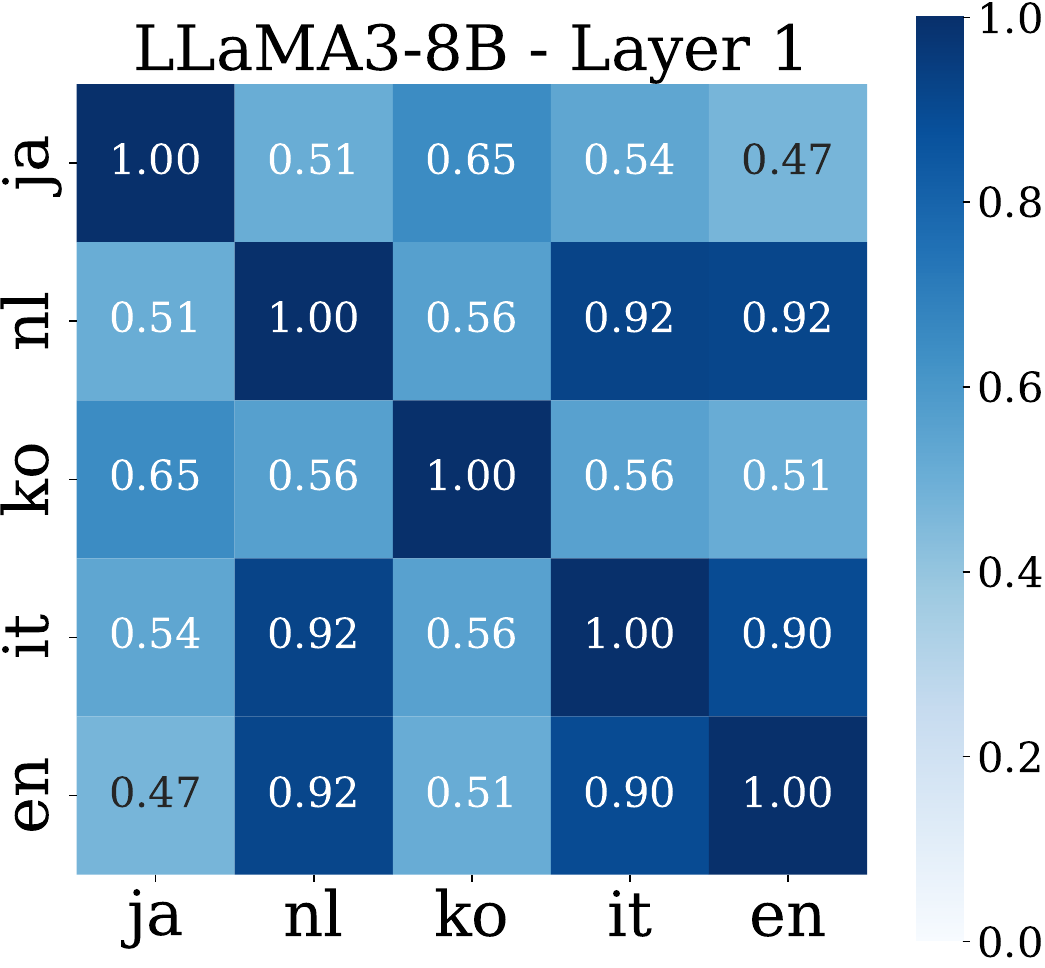}
  \includegraphics[width=0.19\linewidth]{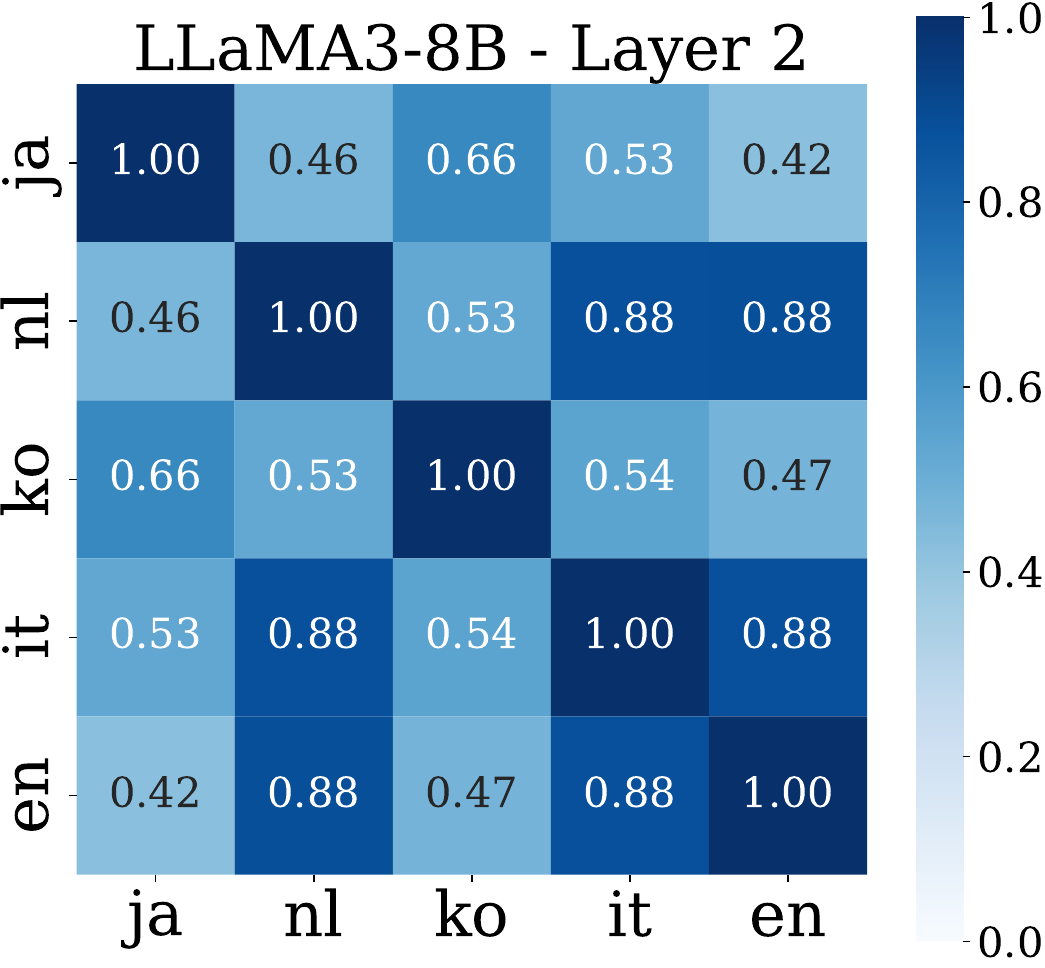}
  \includegraphics[width=0.19\linewidth]{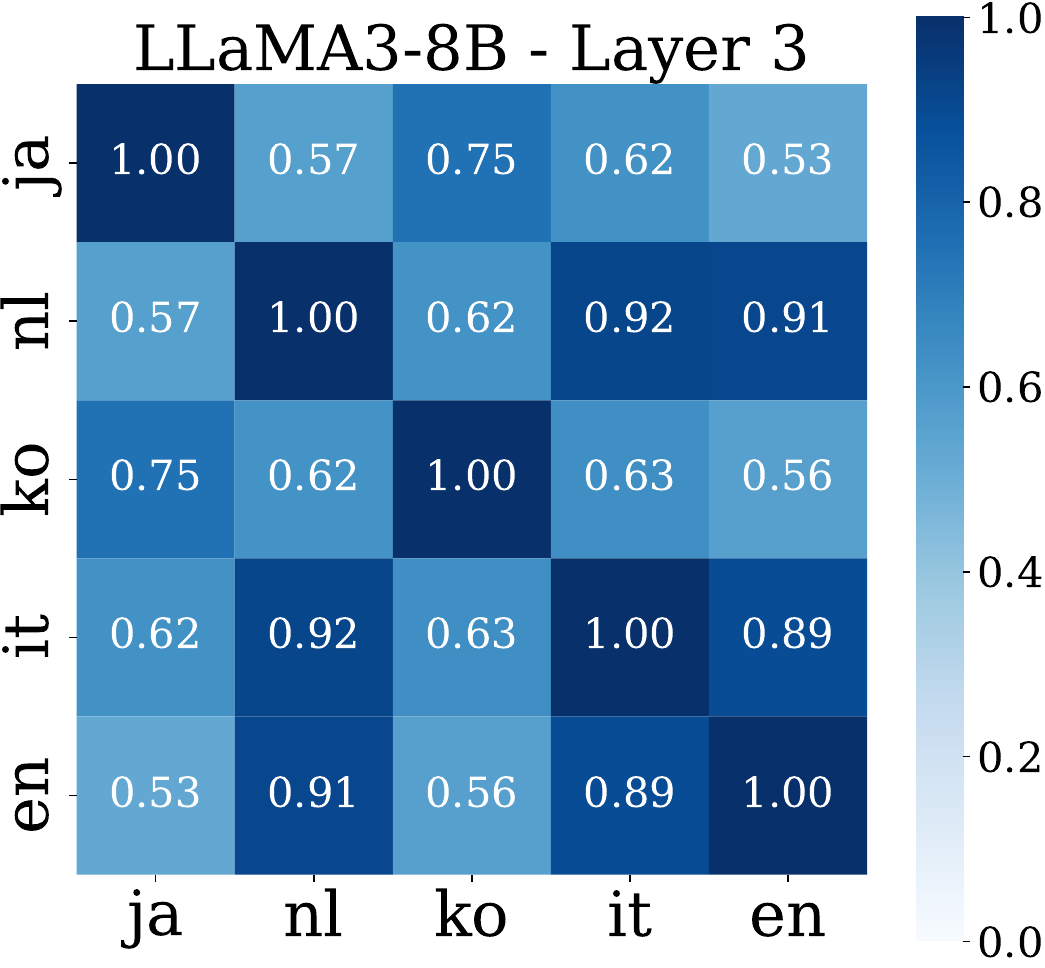}
  \includegraphics[width=0.19\linewidth]{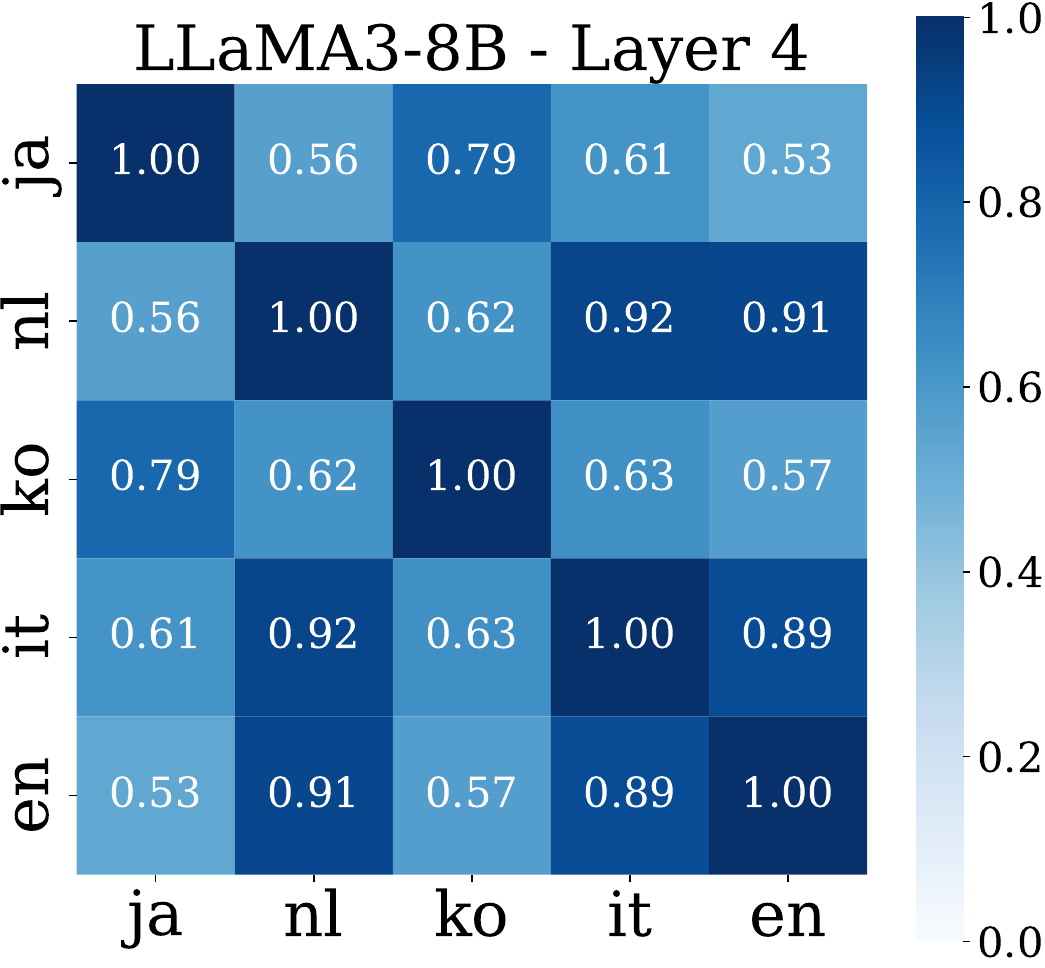}

  \includegraphics[width=0.19\linewidth]{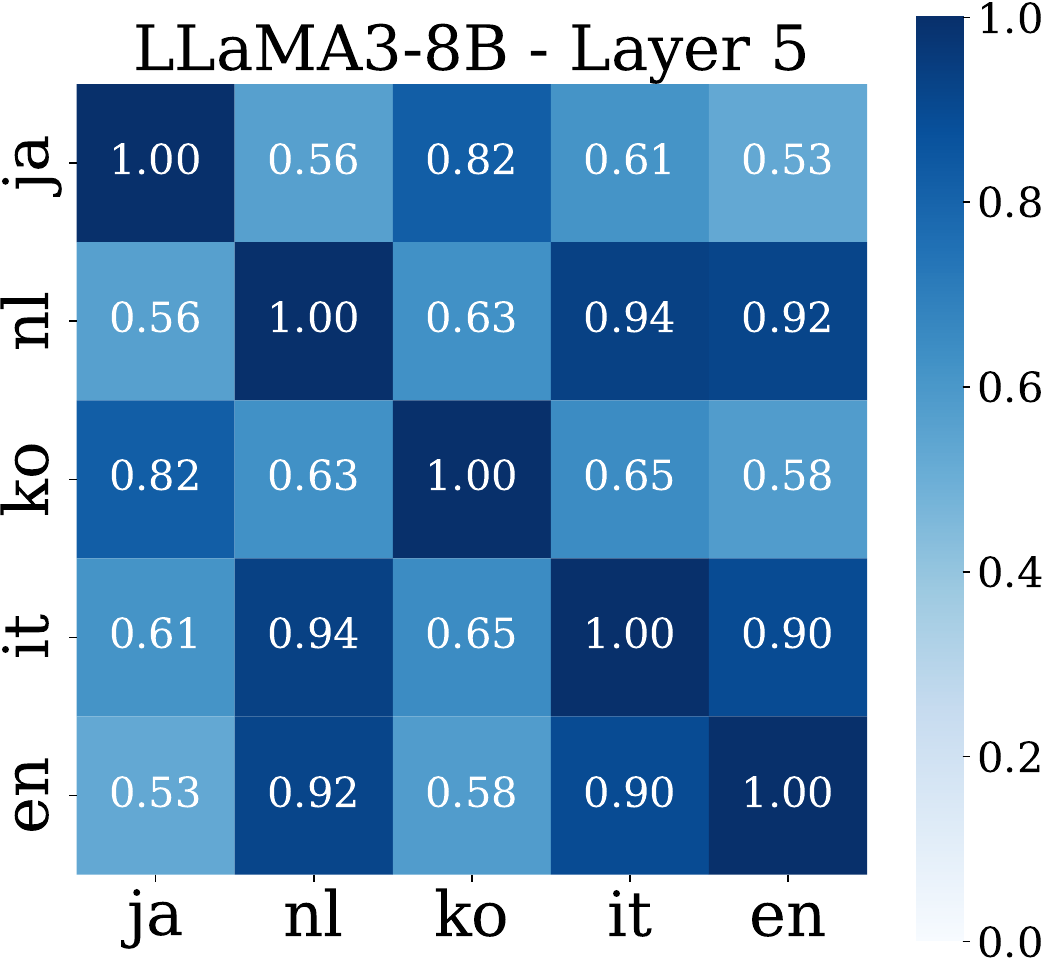}
  \includegraphics[width=0.19\linewidth]{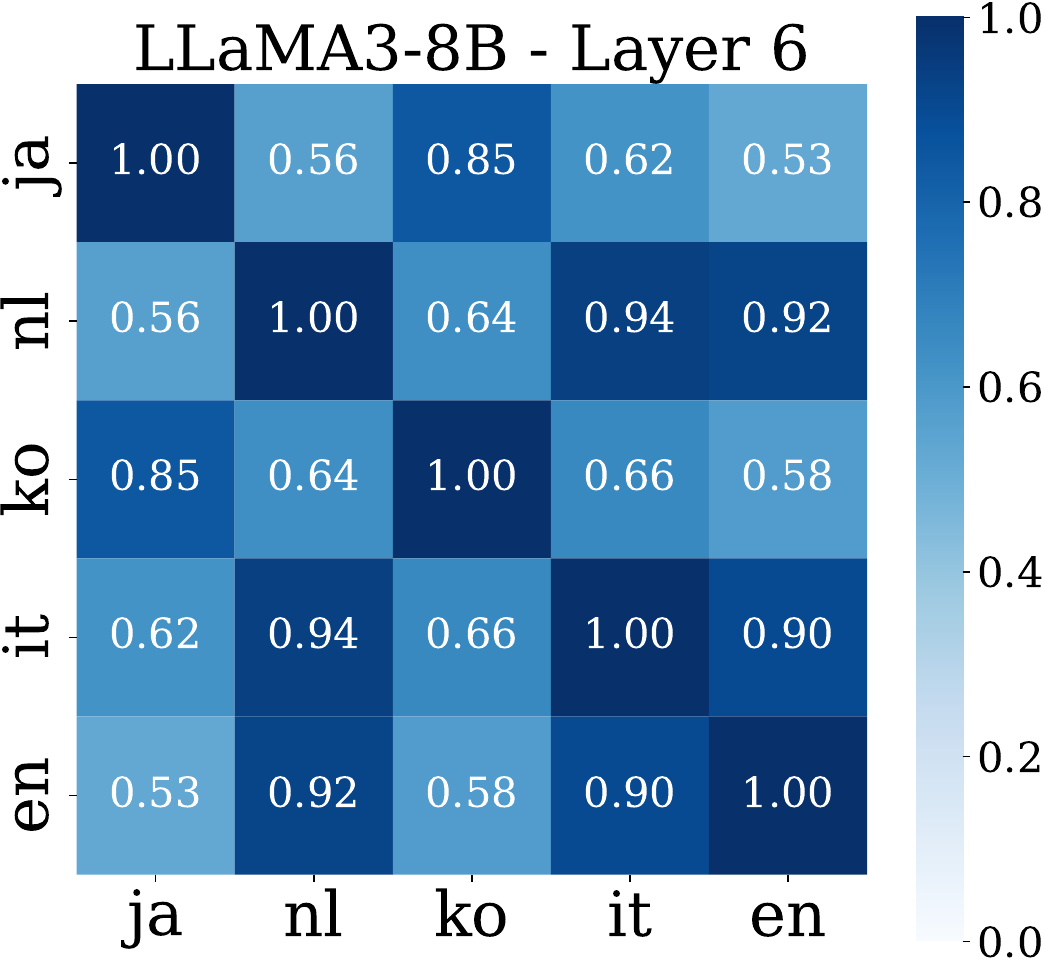}
  \includegraphics[width=0.19\linewidth]{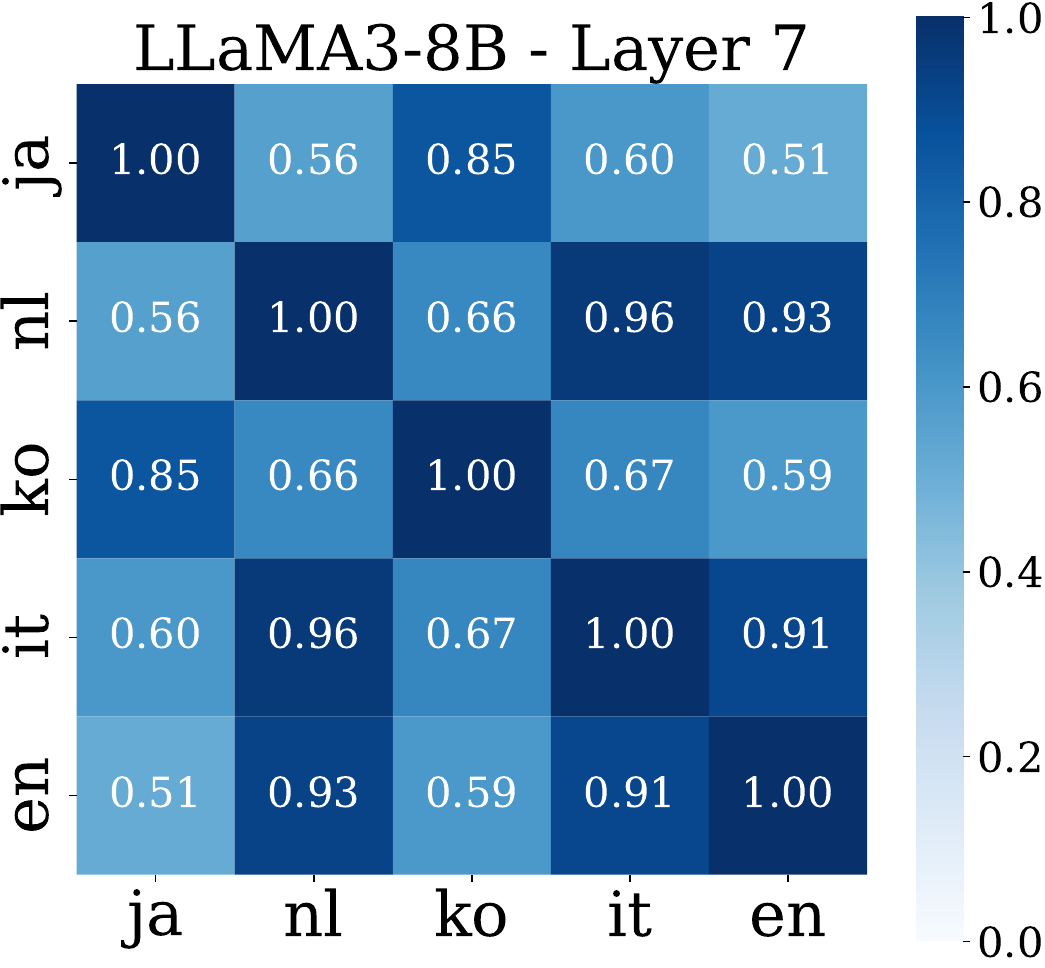}
  \includegraphics[width=0.19\linewidth]{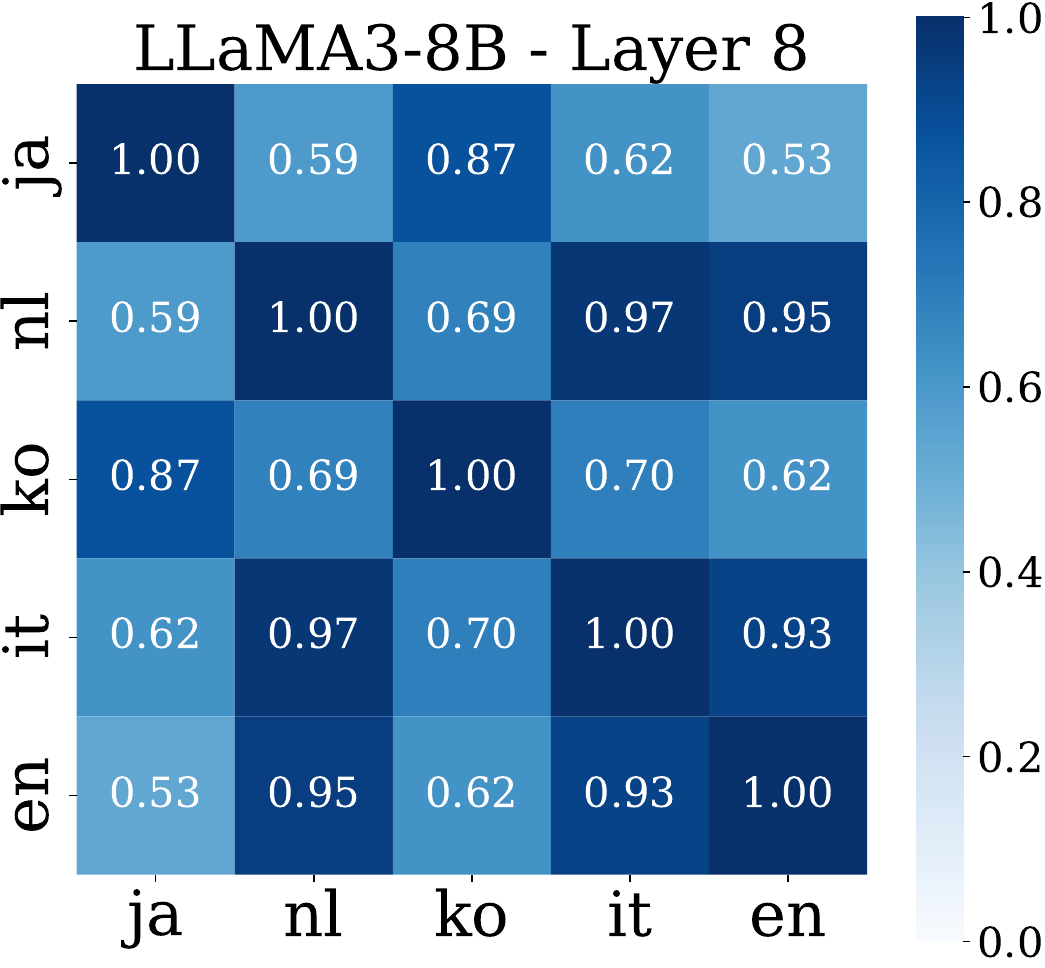}
  \includegraphics[width=0.19\linewidth]{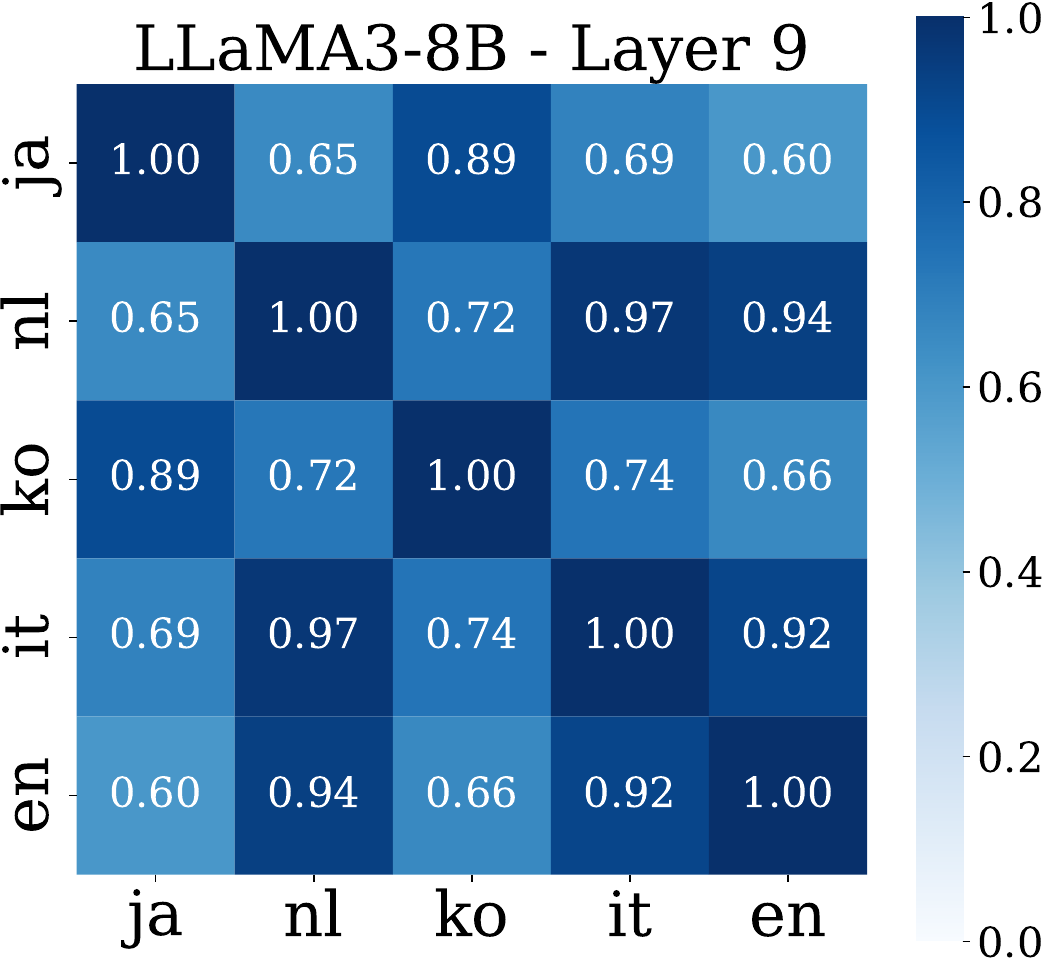}

  \includegraphics[width=0.19\linewidth]{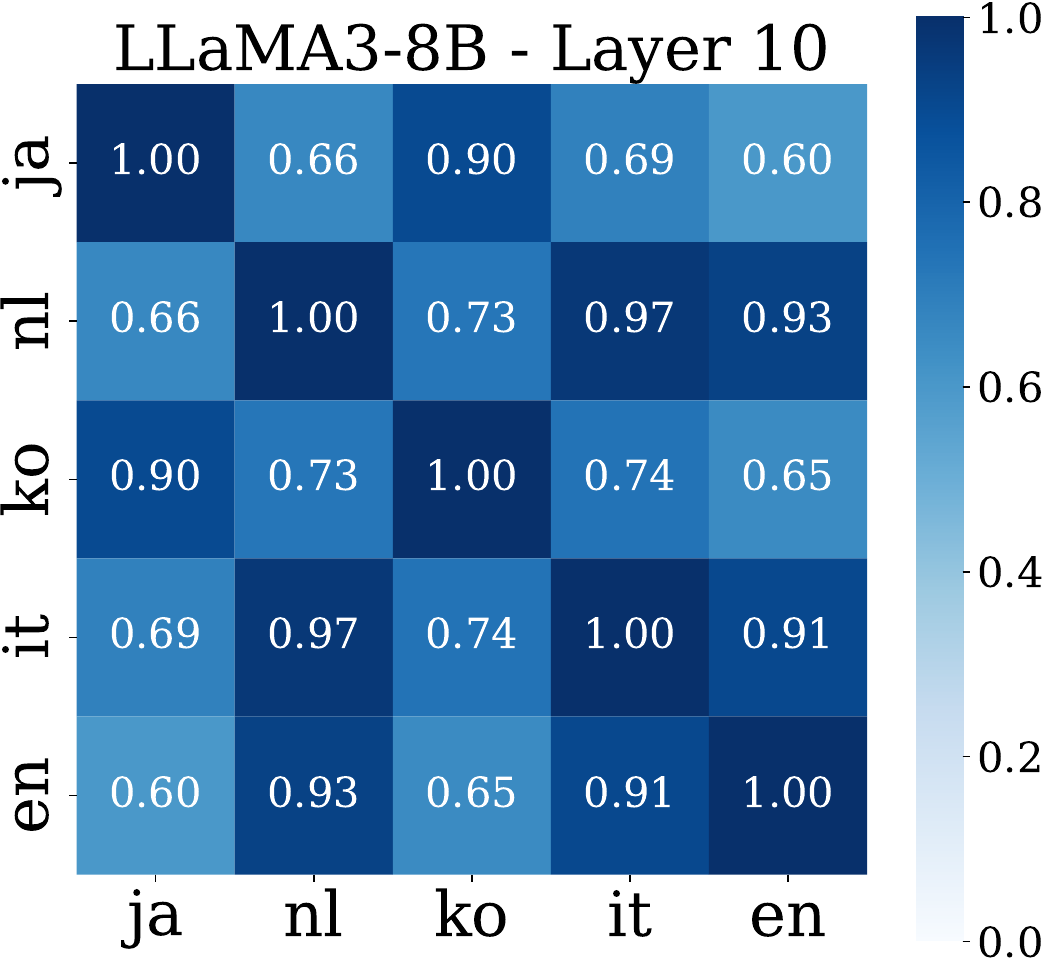}
  \includegraphics[width=0.19\linewidth]{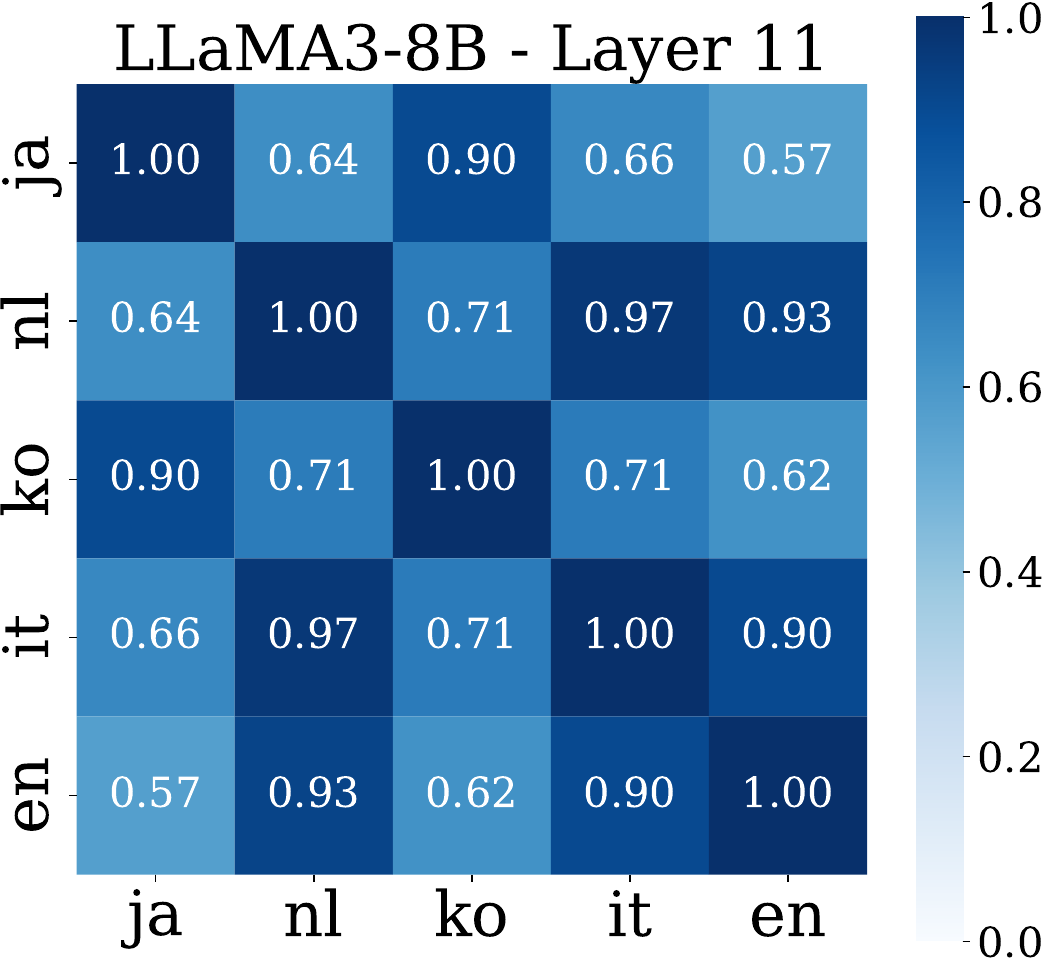}
  \includegraphics[width=0.19\linewidth]{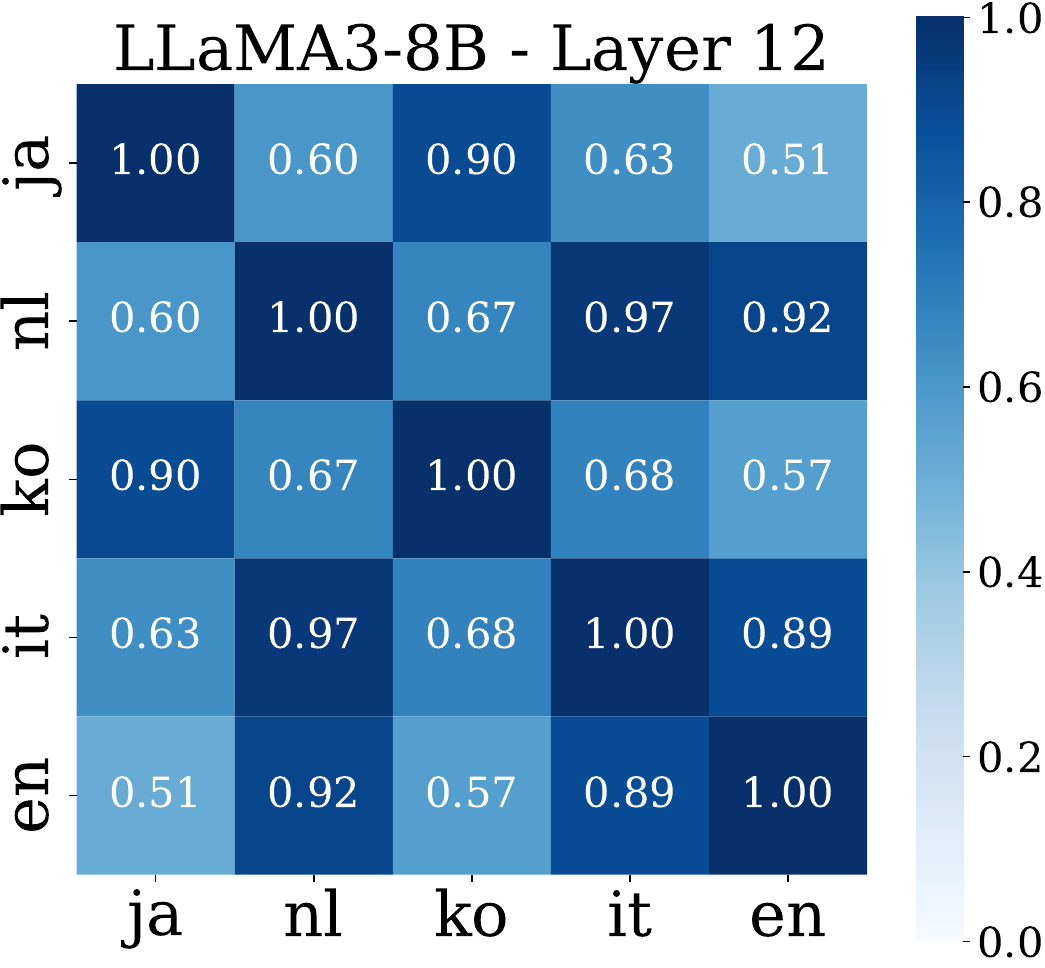}
  \includegraphics[width=0.19\linewidth]{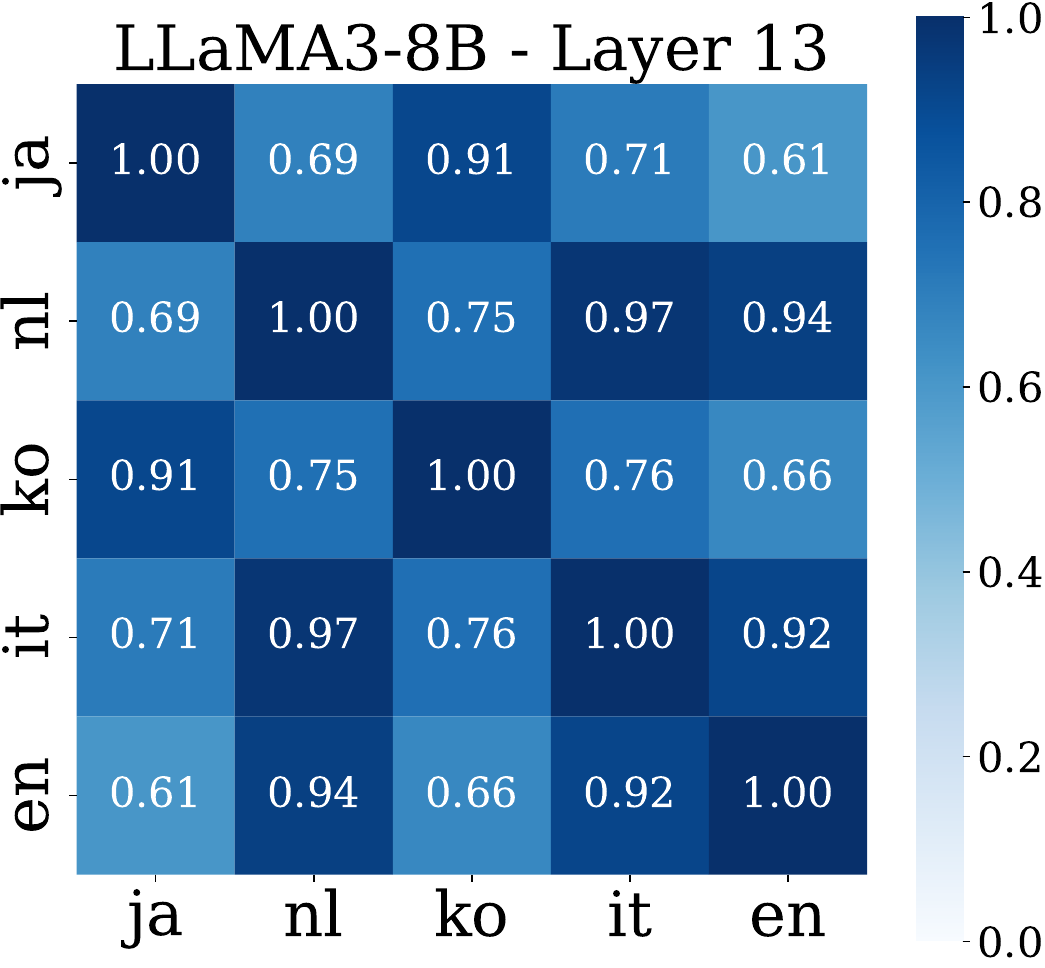}
  \includegraphics[width=0.19\linewidth]{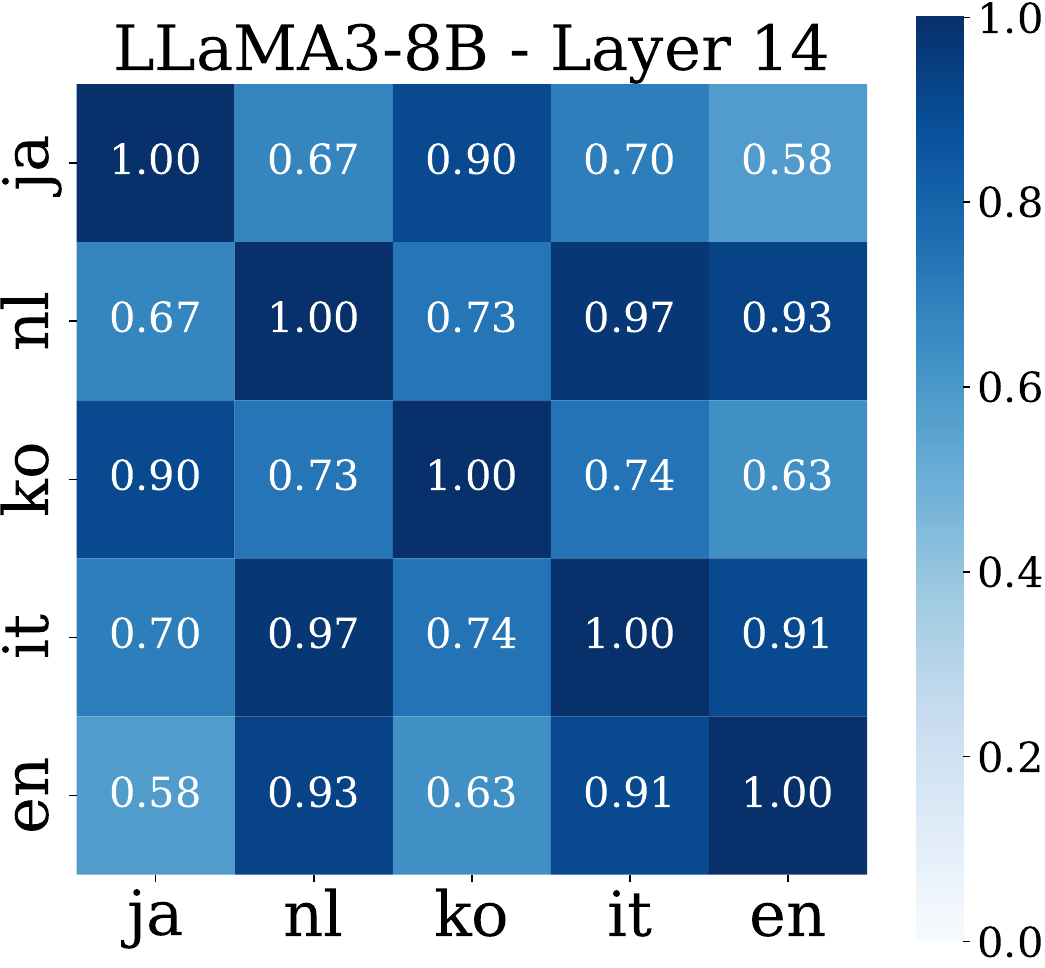}

  \includegraphics[width=0.19\linewidth]{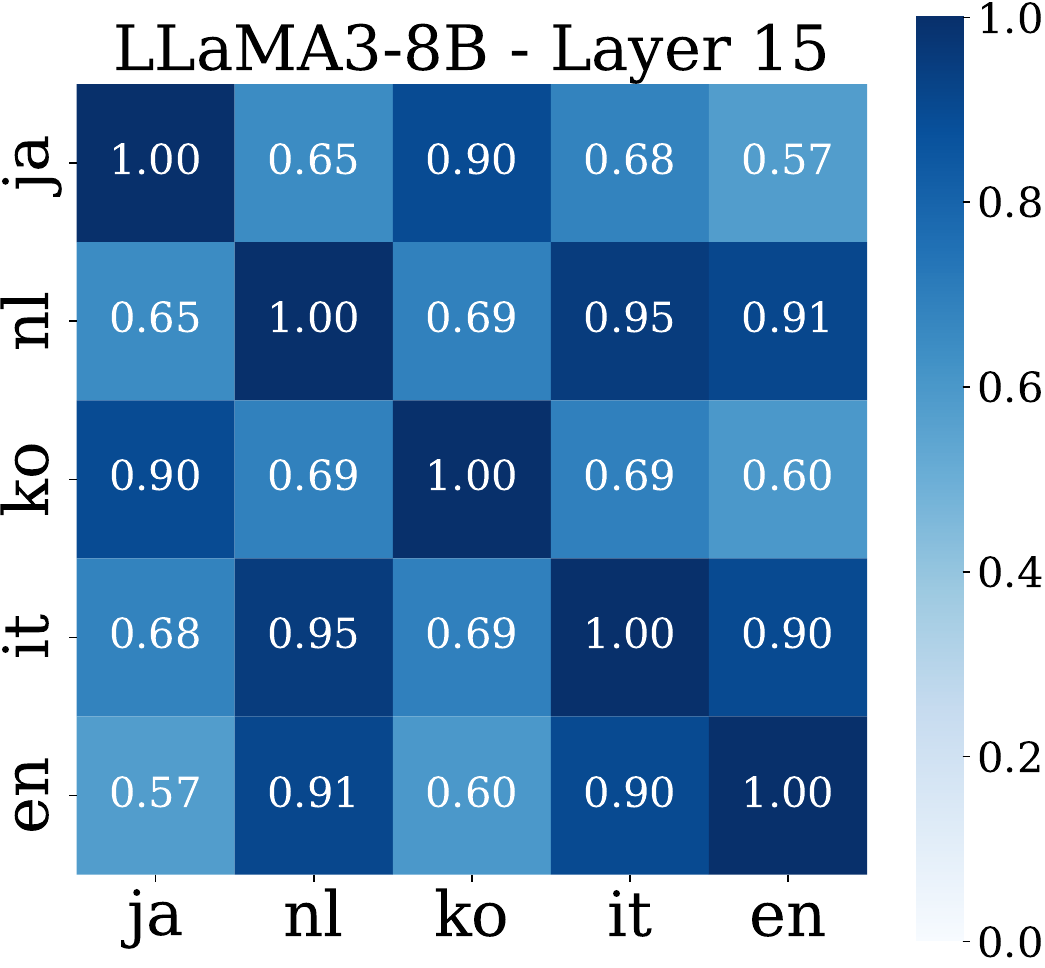}
  \includegraphics[width=0.19\linewidth]{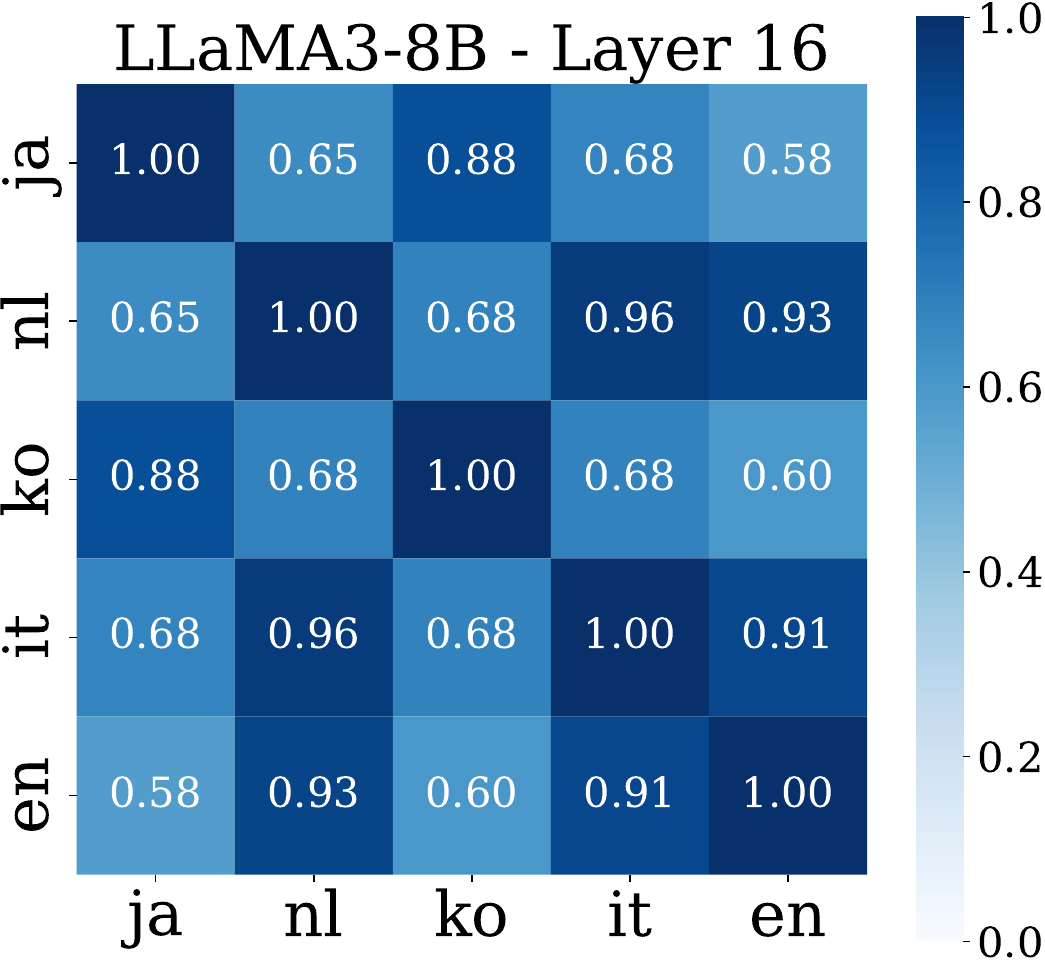}
  \includegraphics[width=0.19\linewidth]{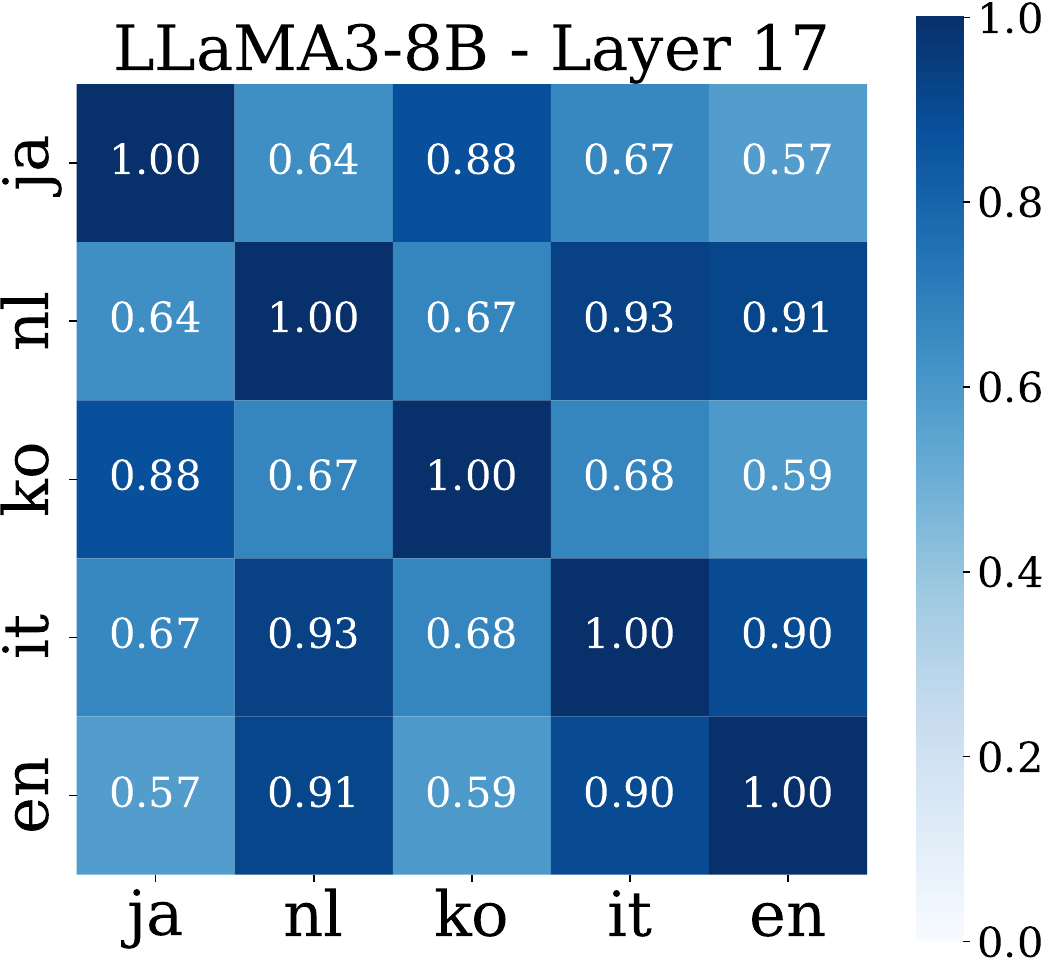}
  \includegraphics[width=0.19\linewidth]{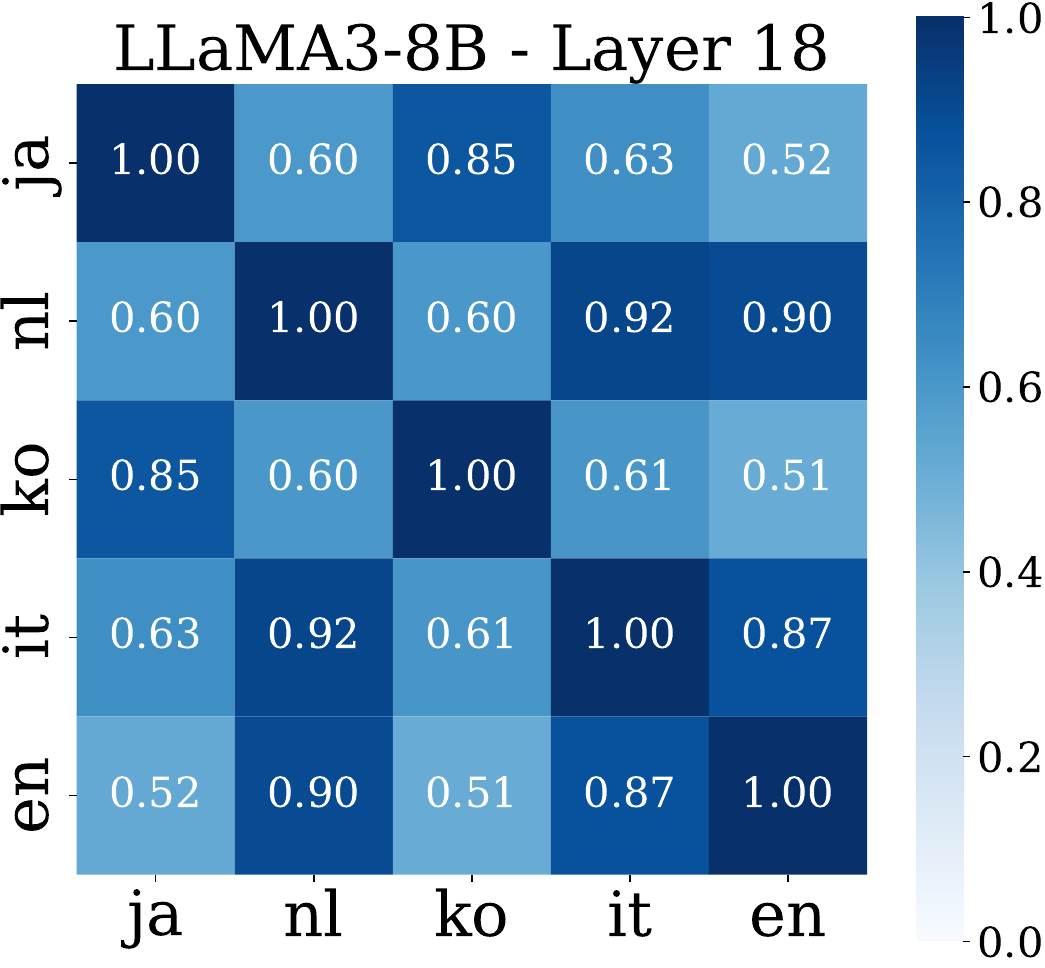}
  \includegraphics[width=0.19\linewidth]{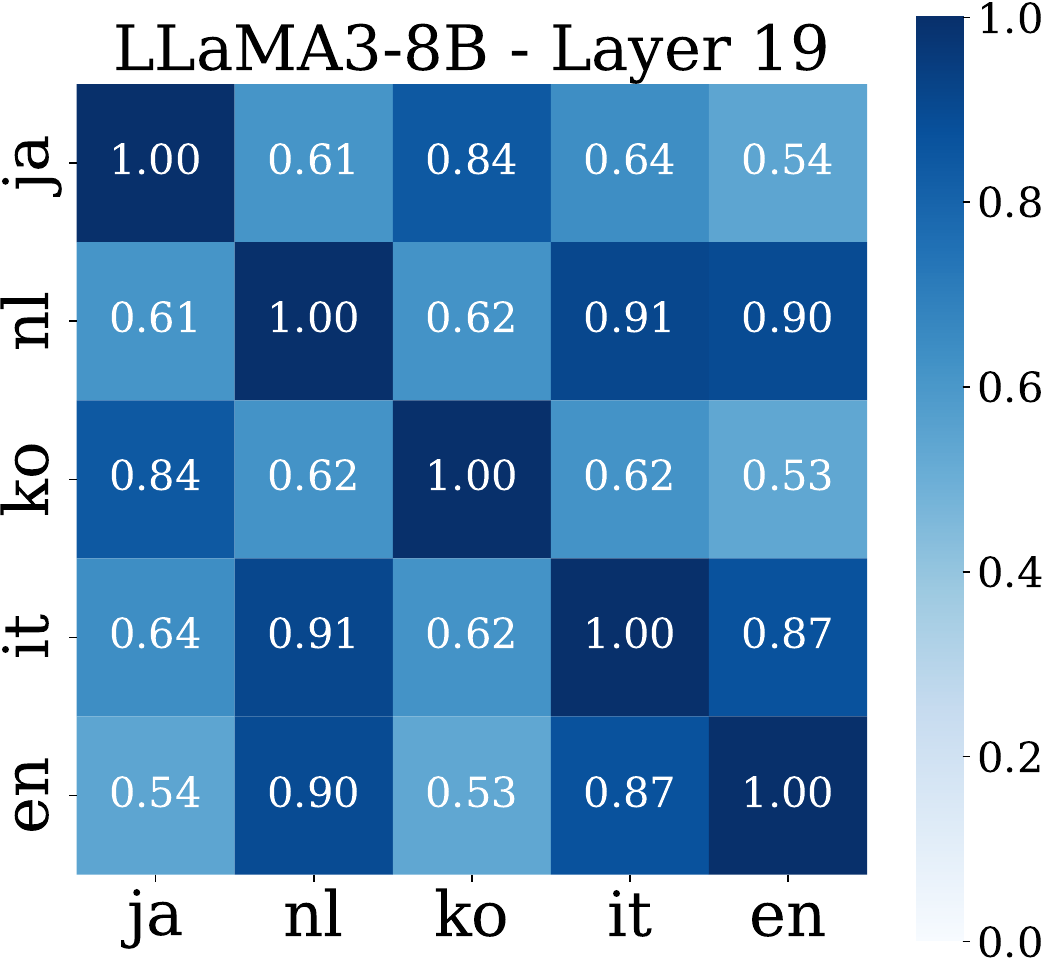}

  \includegraphics[width=0.19\linewidth]{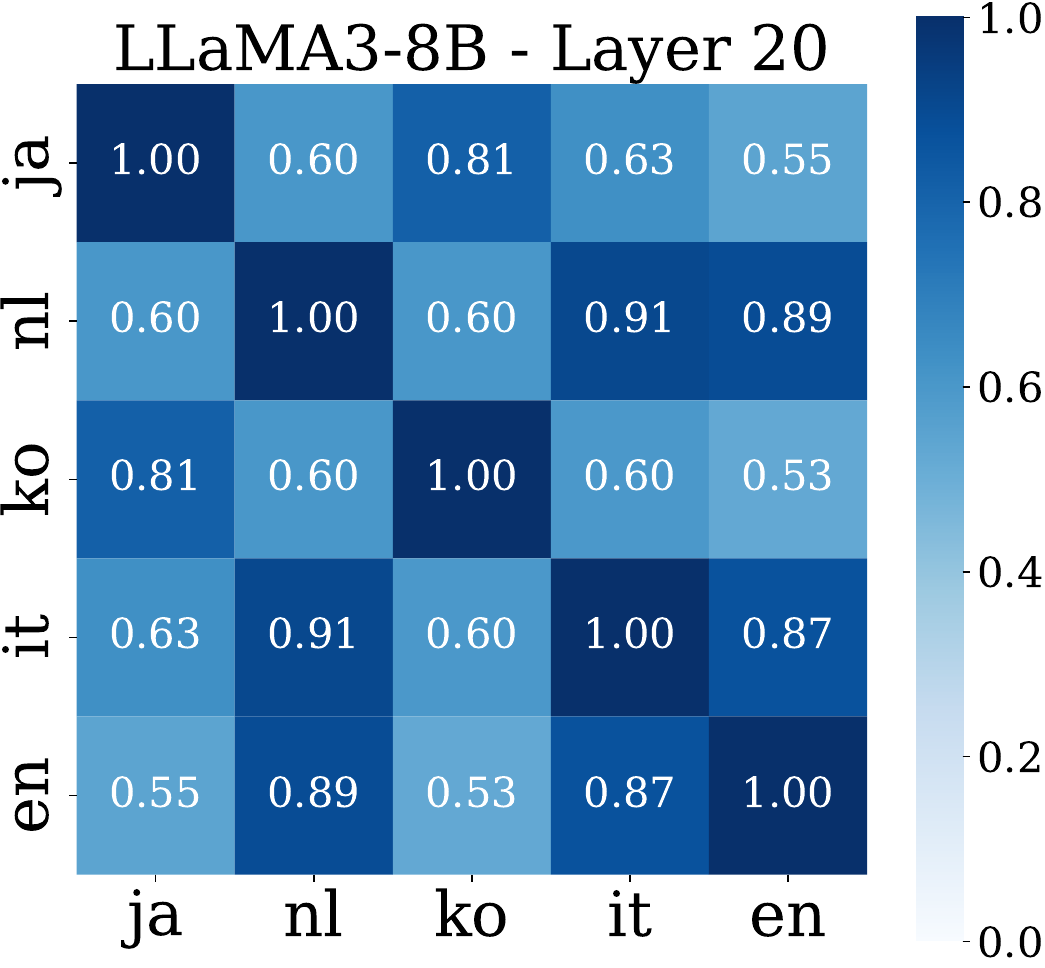}
  \includegraphics[width=0.19\linewidth]{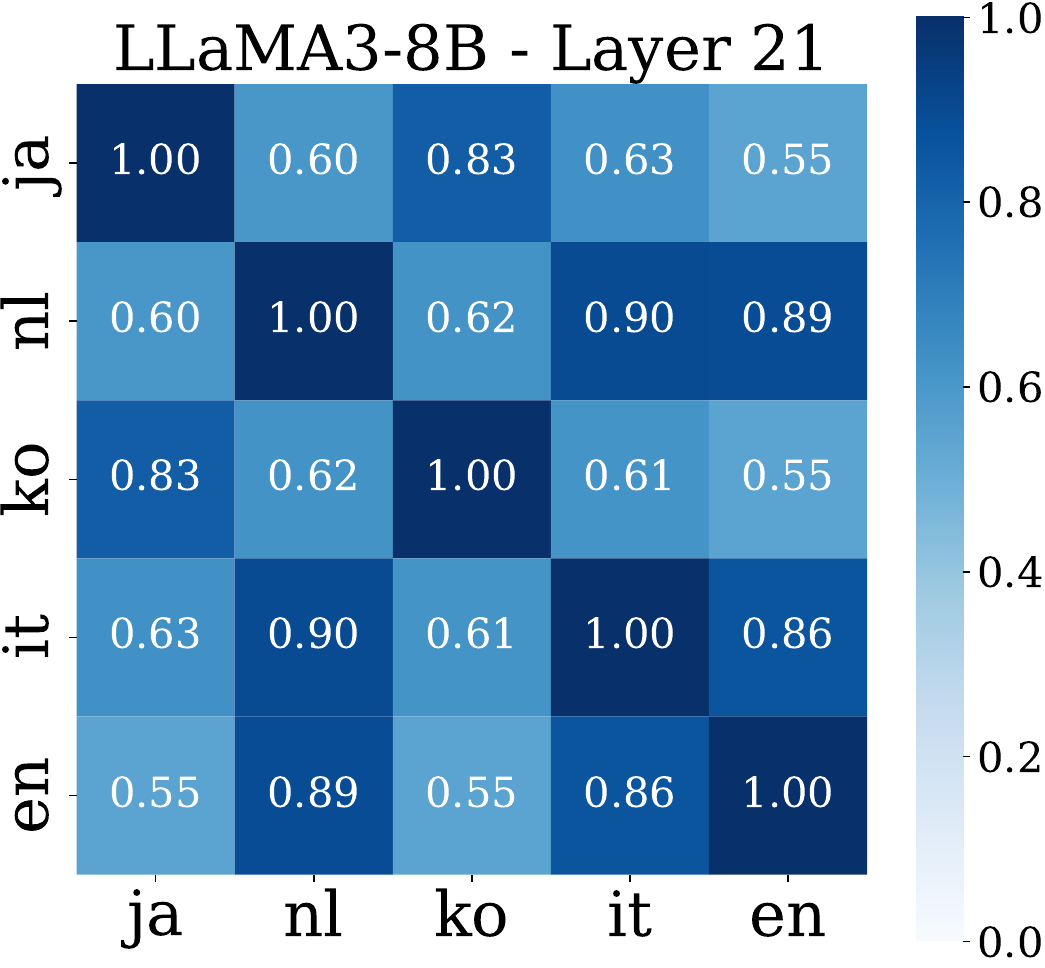}
  \includegraphics[width=0.19\linewidth]{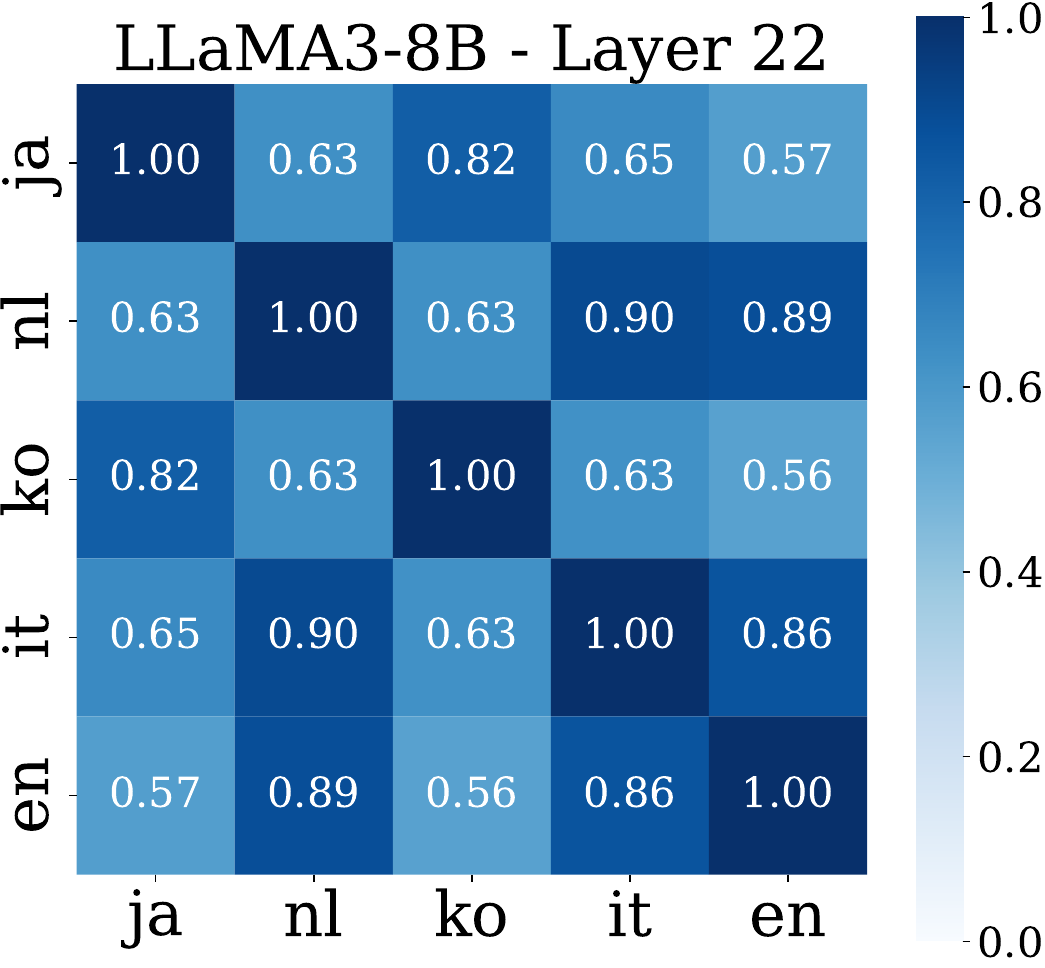}
  \includegraphics[width=0.19\linewidth]{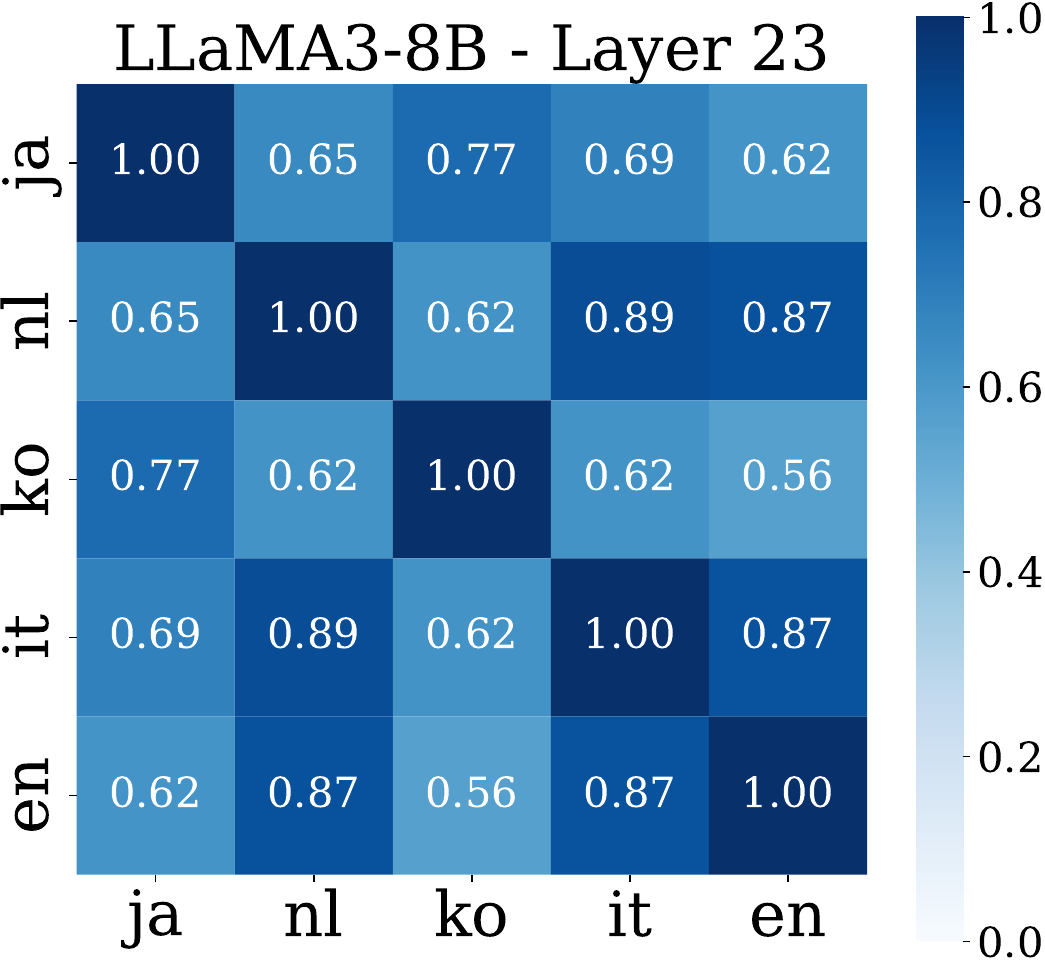}
  \includegraphics[width=0.19\linewidth]{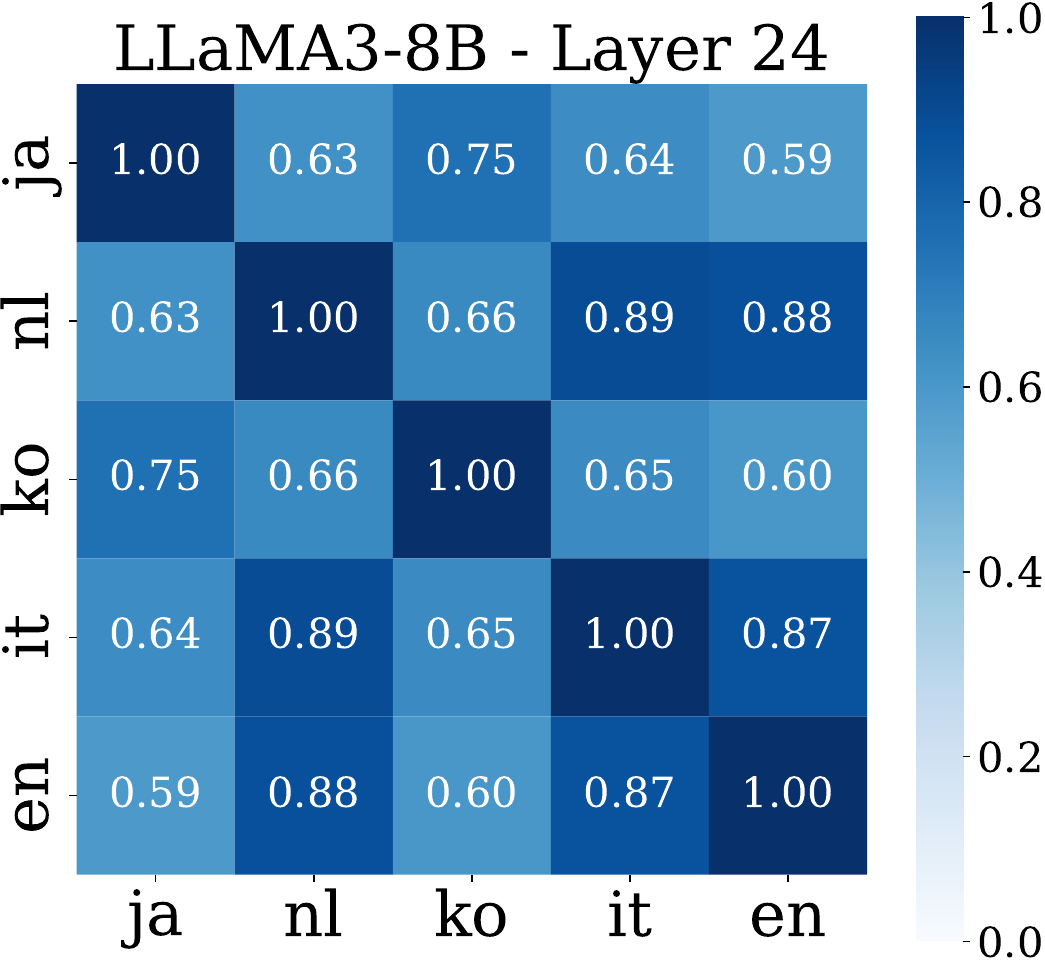}

  \includegraphics[width=0.19\linewidth]{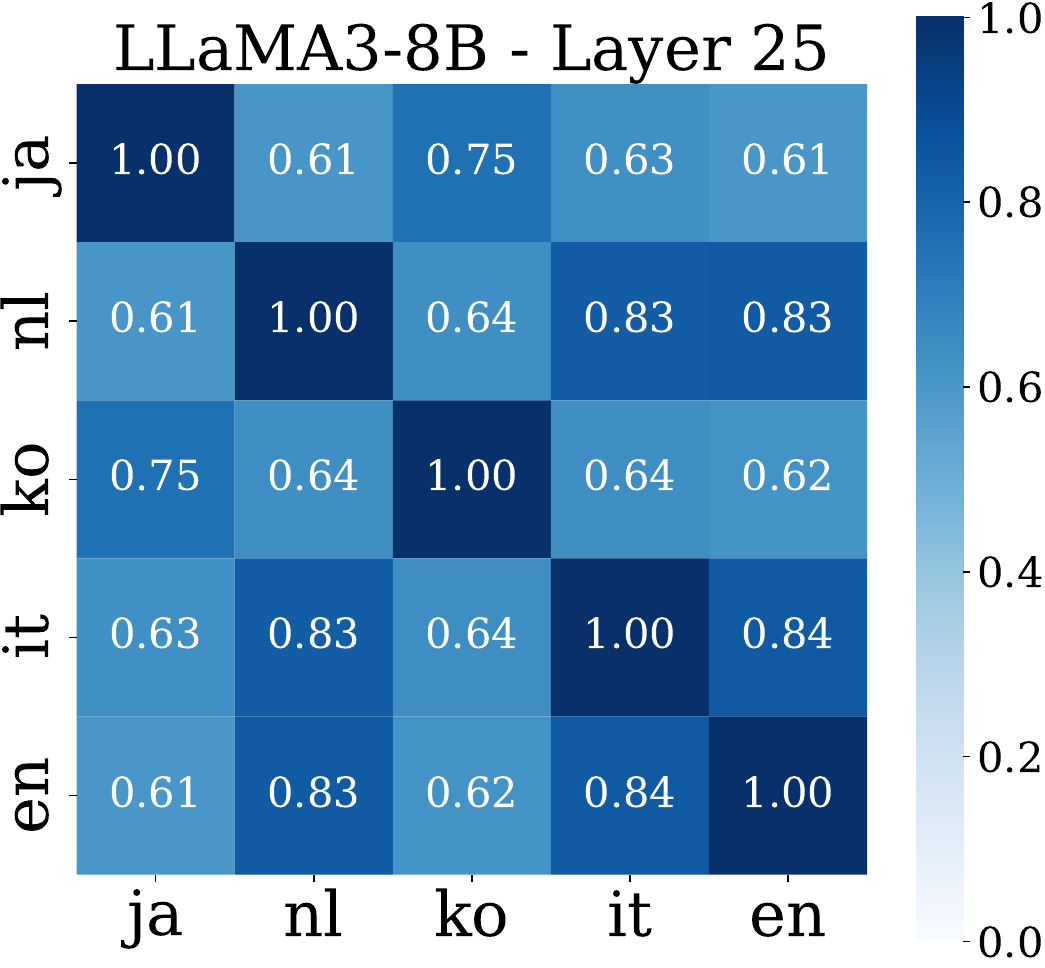}
  \includegraphics[width=0.19\linewidth]{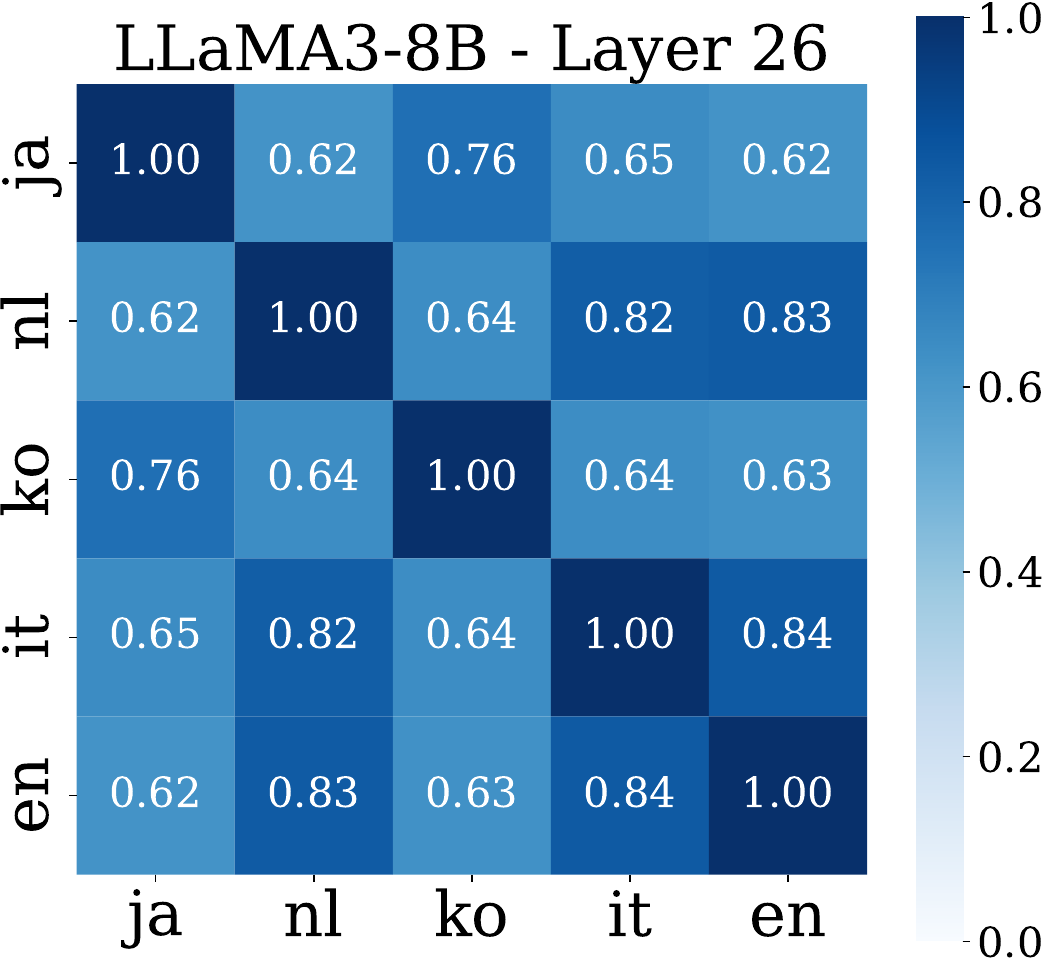}
  \includegraphics[width=0.19\linewidth]{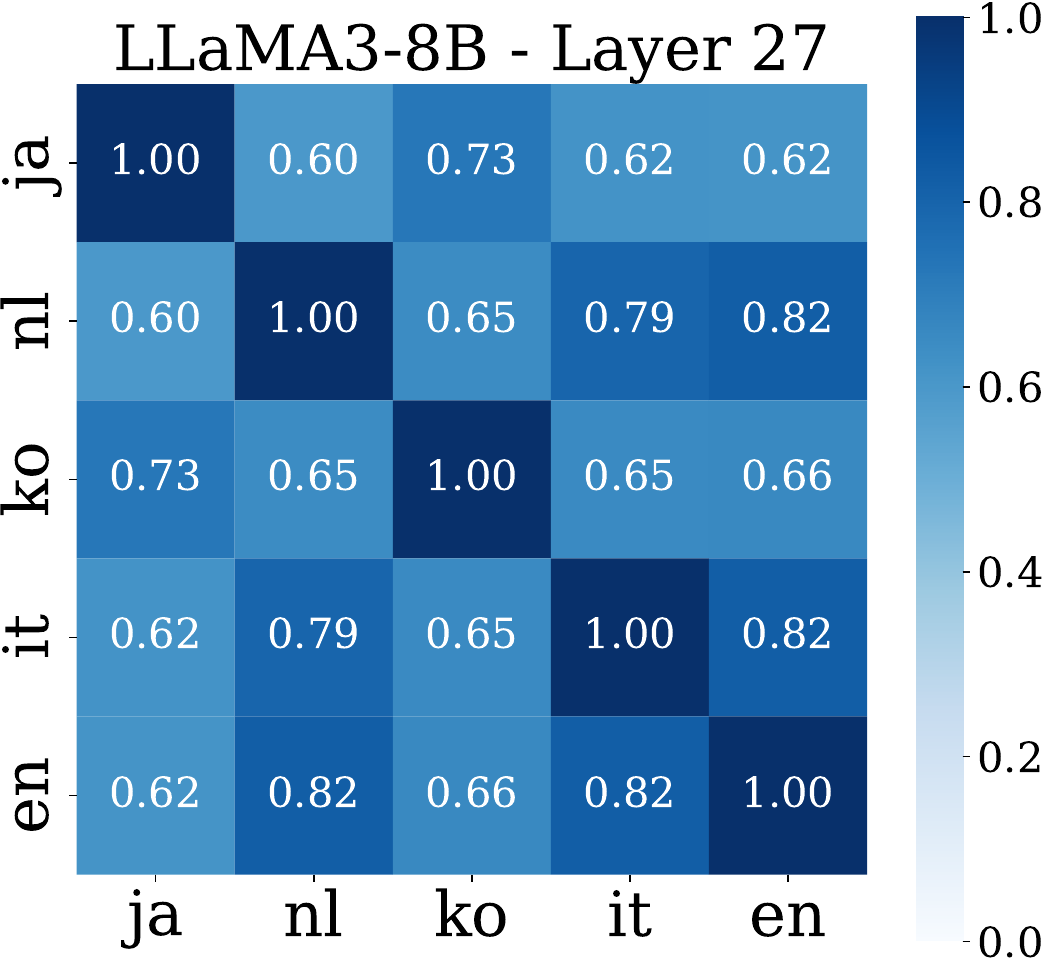}
  \includegraphics[width=0.19\linewidth]{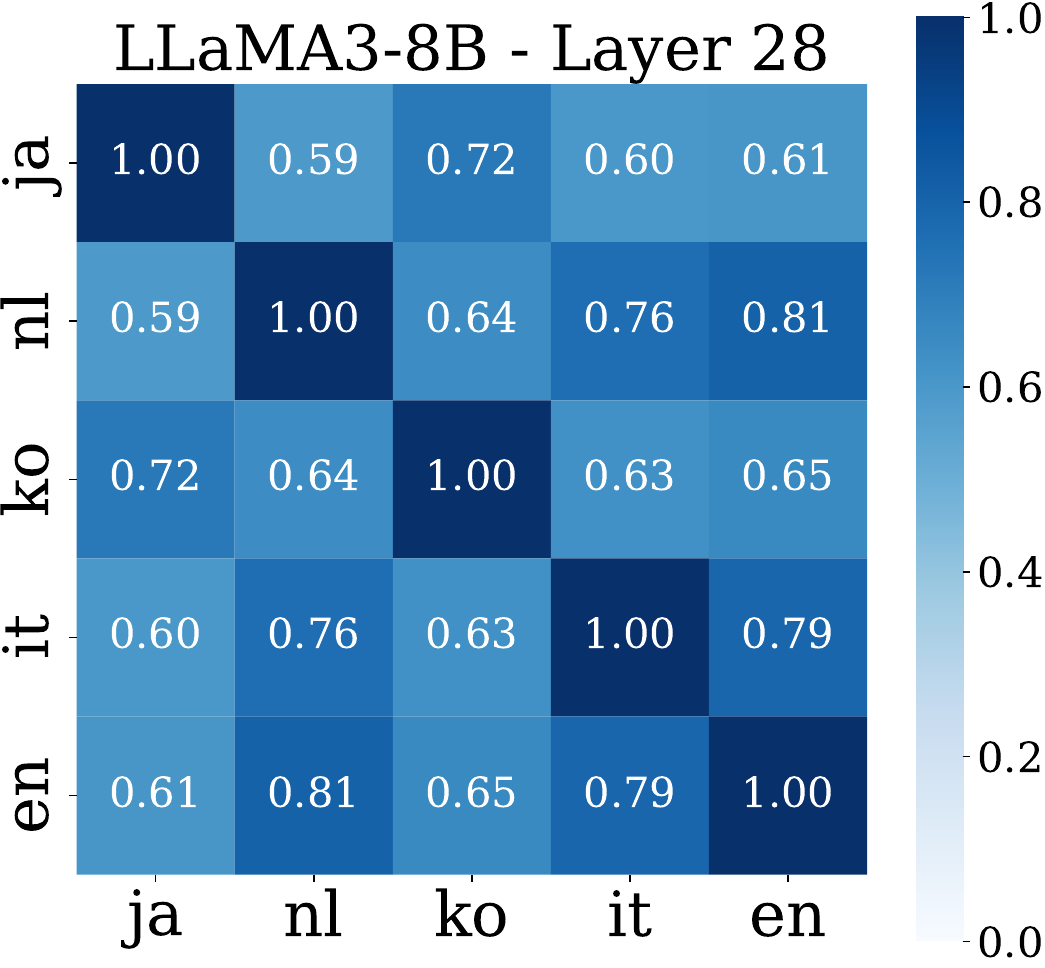}
  \includegraphics[width=0.19\linewidth]{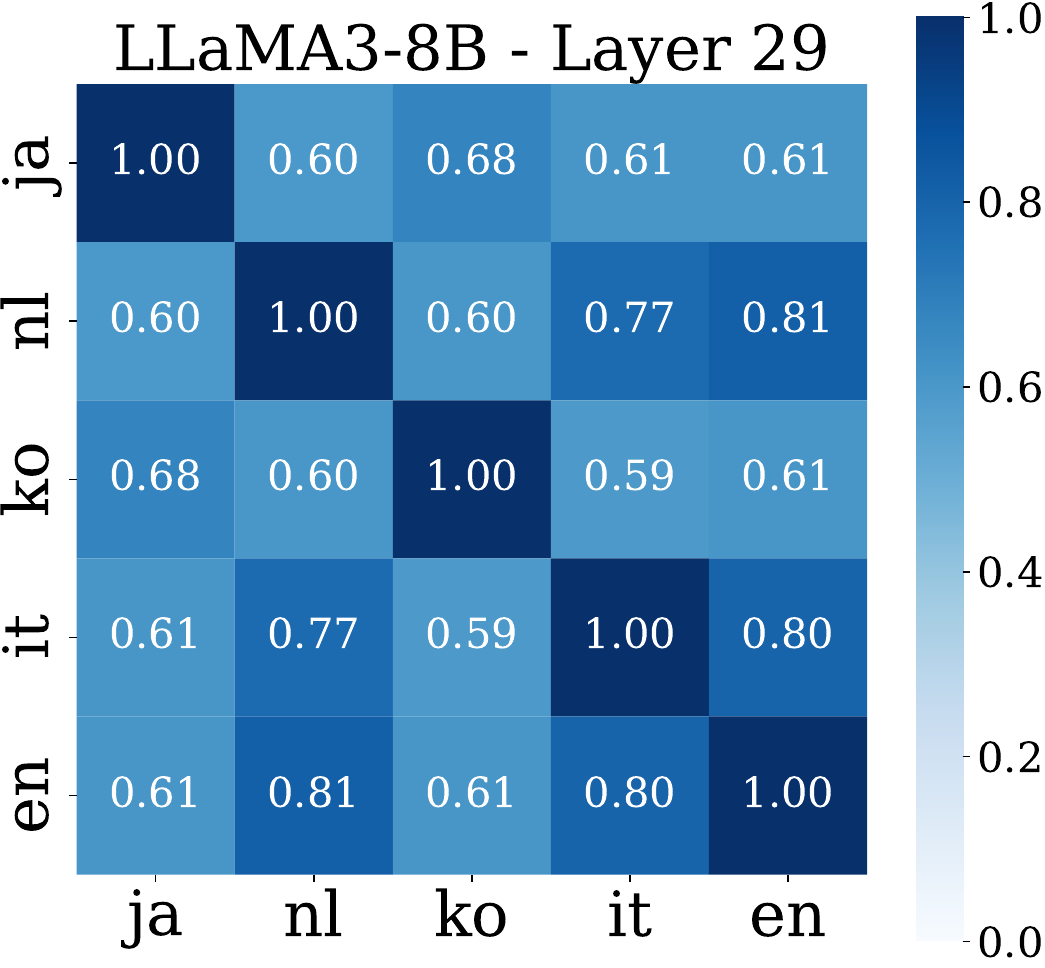}

  \includegraphics[width=0.19\linewidth]{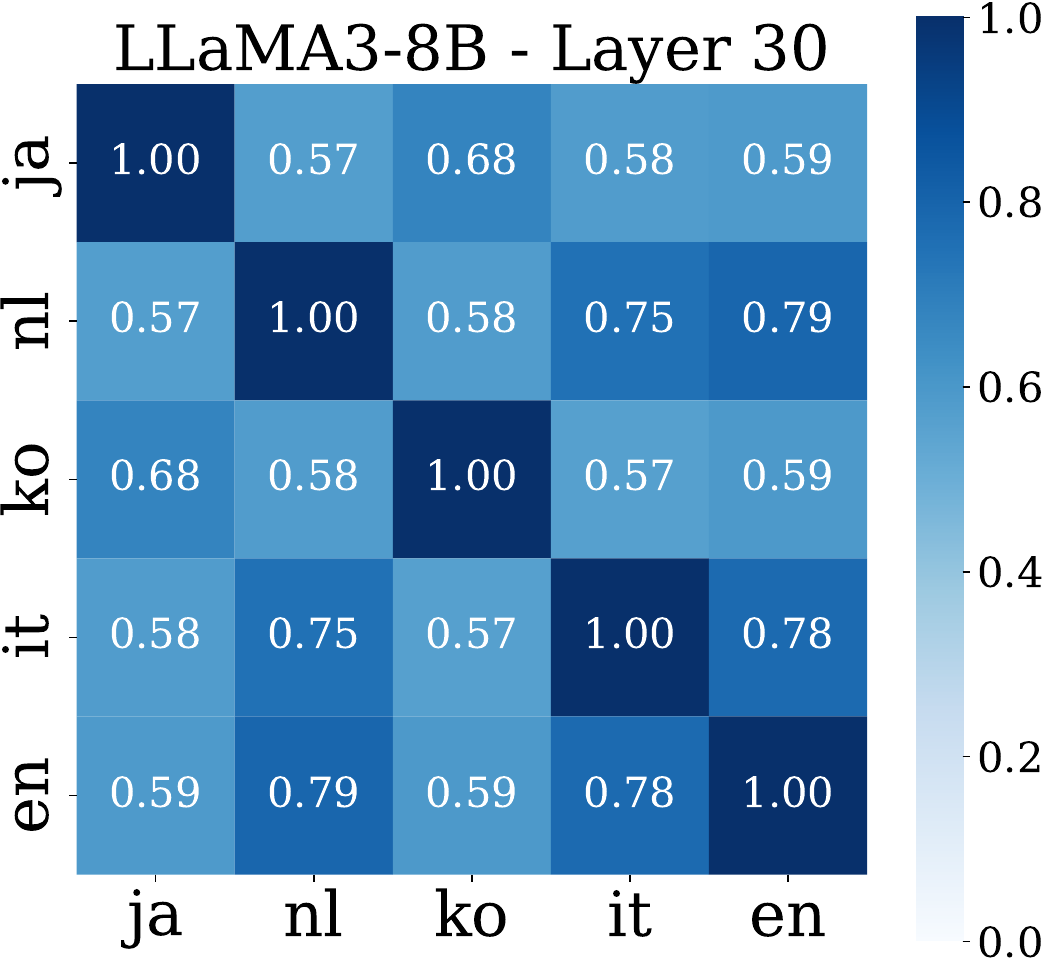}
  \includegraphics[width=0.19\linewidth]{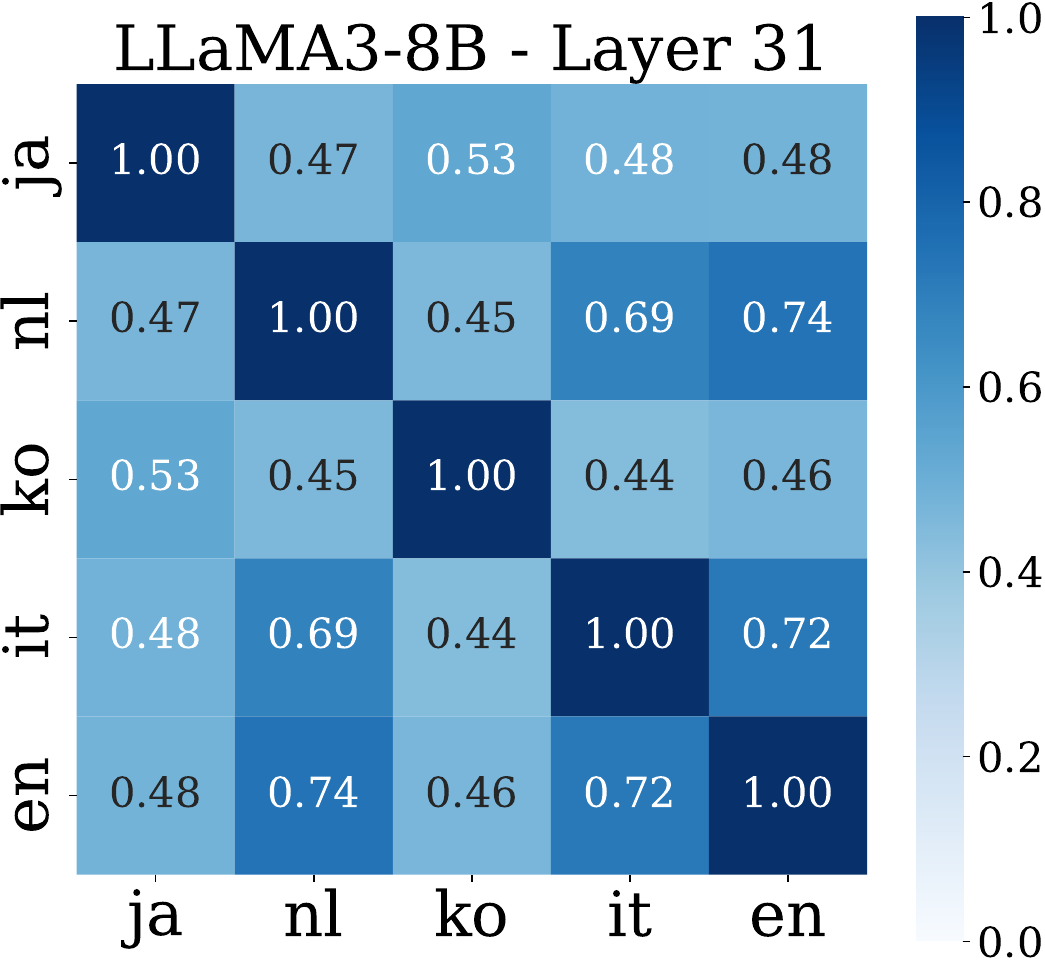}
  \includegraphics[width=0.19\linewidth]{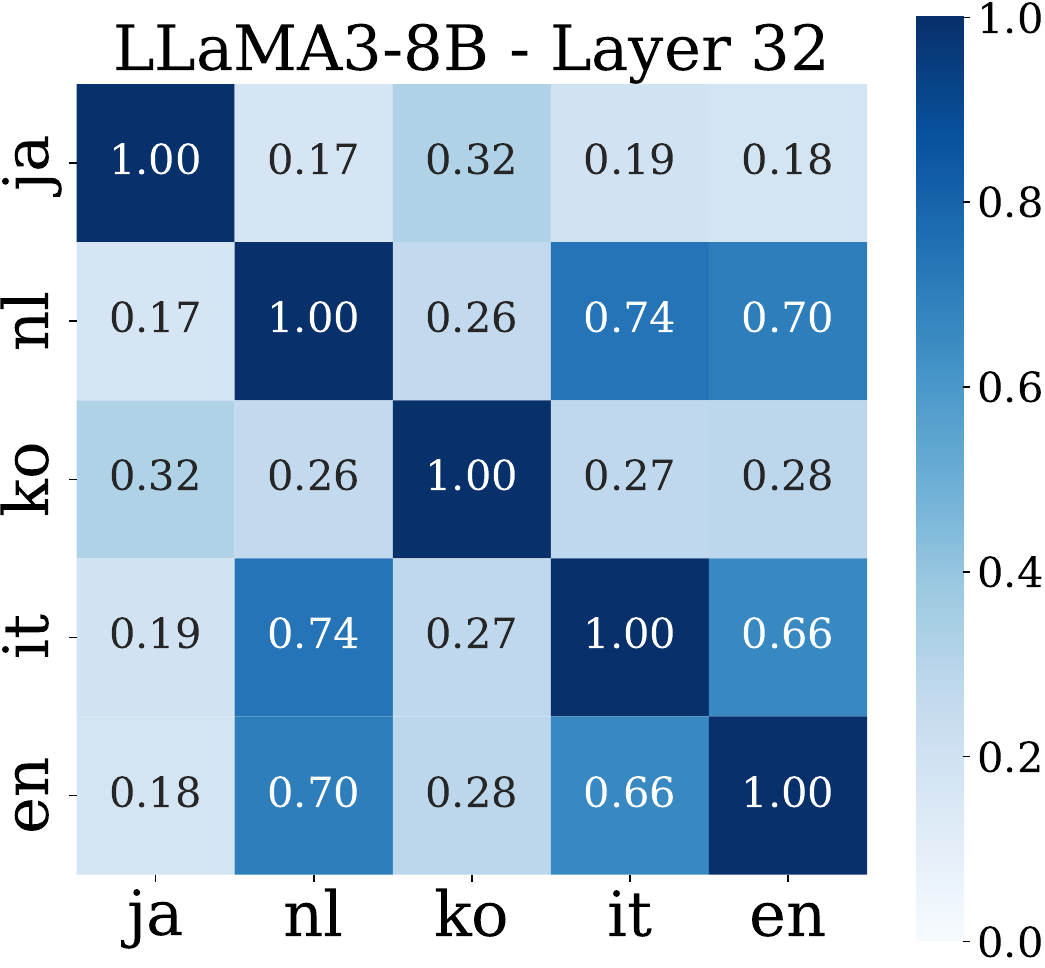}

  \caption{\textbf{The distance among centroids of language latent spaces (LLaMA3-8B).}}
  \label{fig:appendix:distance among subspaces centroids llama3}
\end{figure*}
% mistral, distance among subspaces, centroids
\begin{figure*}[t]
  \centering

  \includegraphics[width=0.19\linewidth]{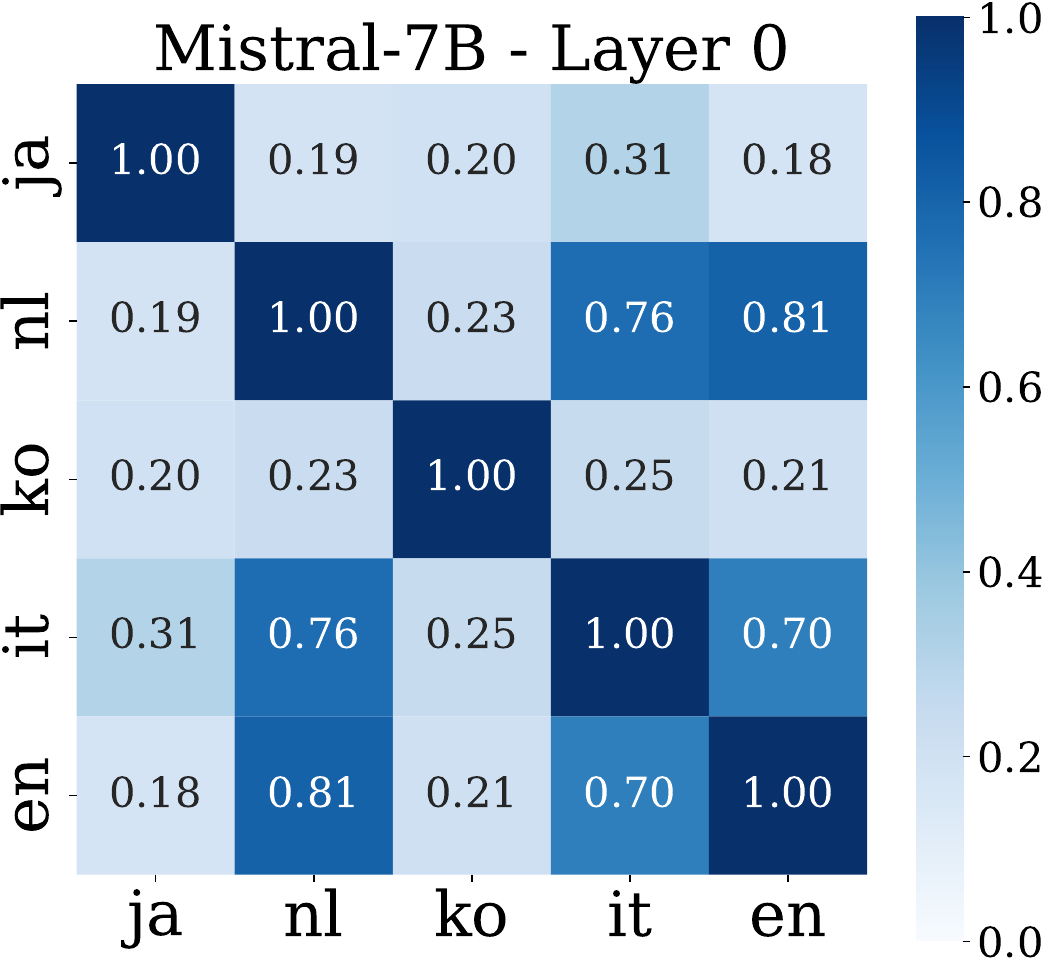}
  \includegraphics[width=0.19\linewidth]{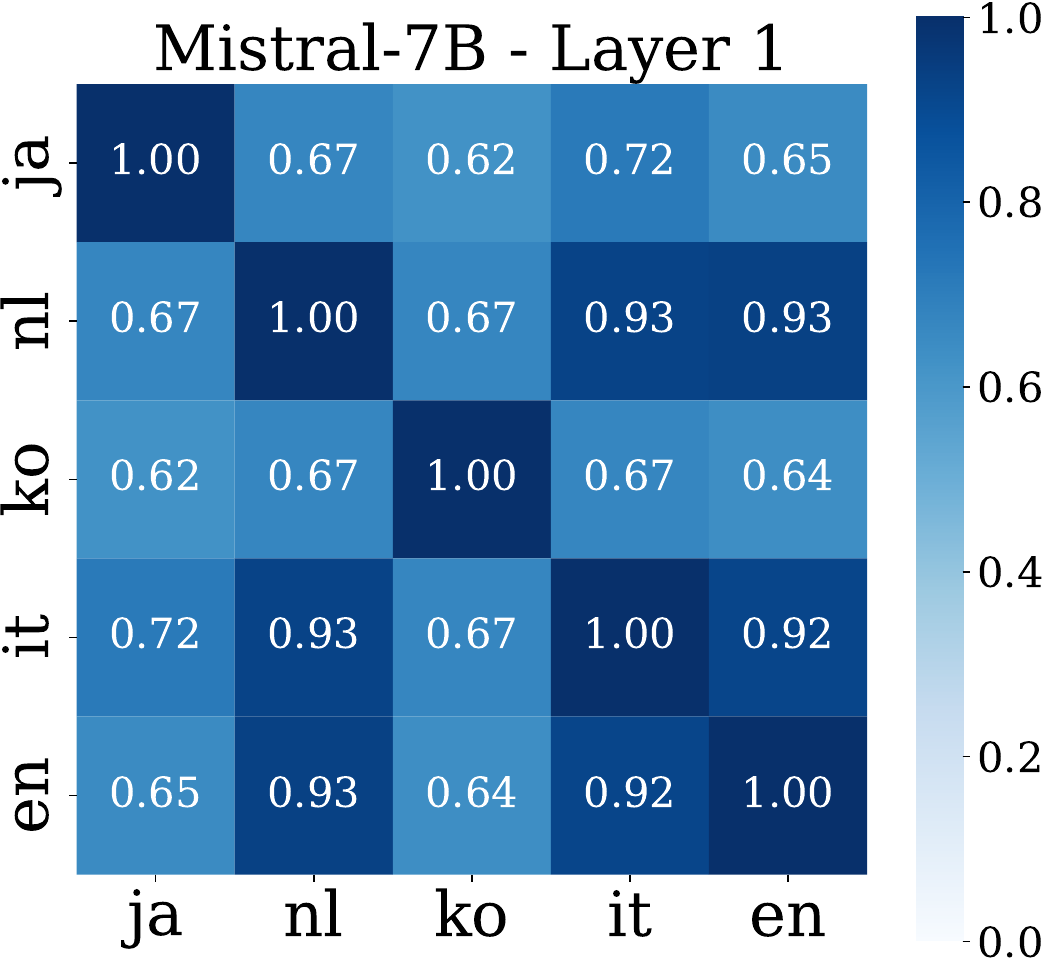}
  \includegraphics[width=0.19\linewidth]{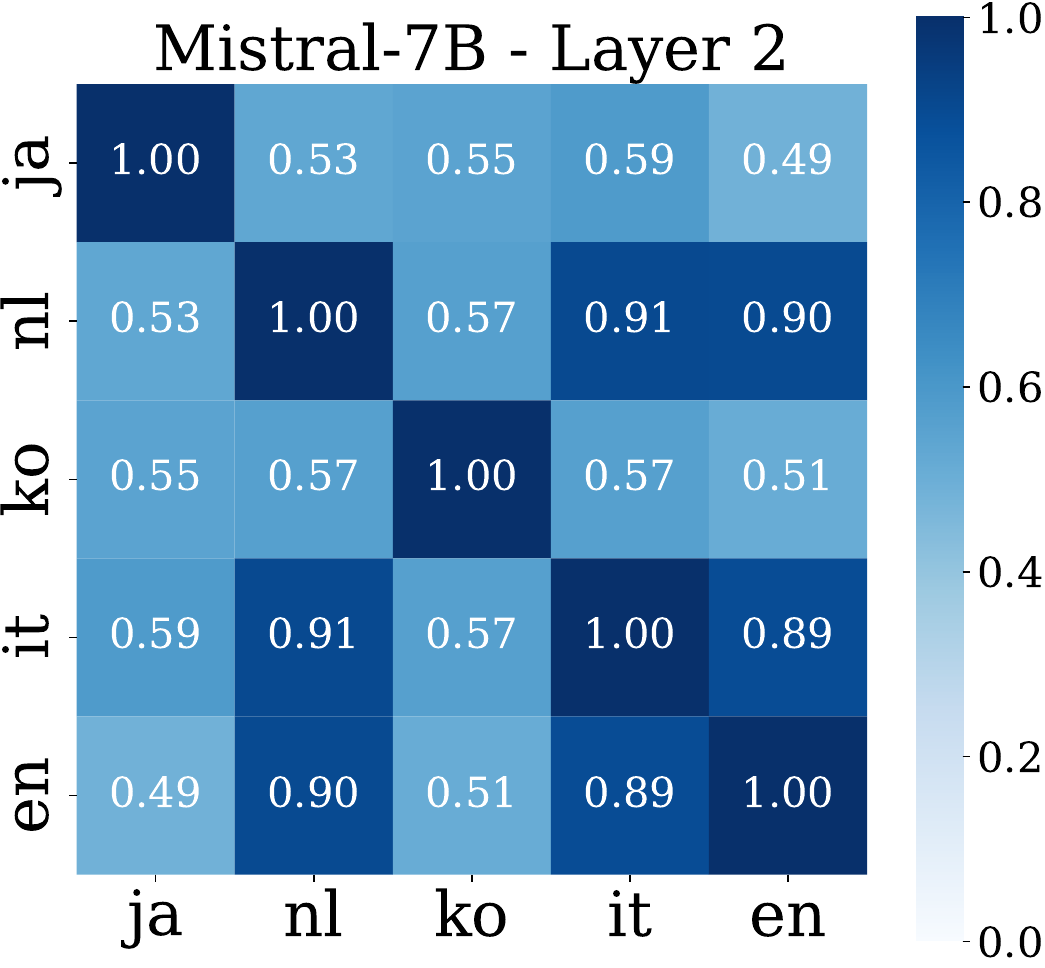}
  \includegraphics[width=0.19\linewidth]{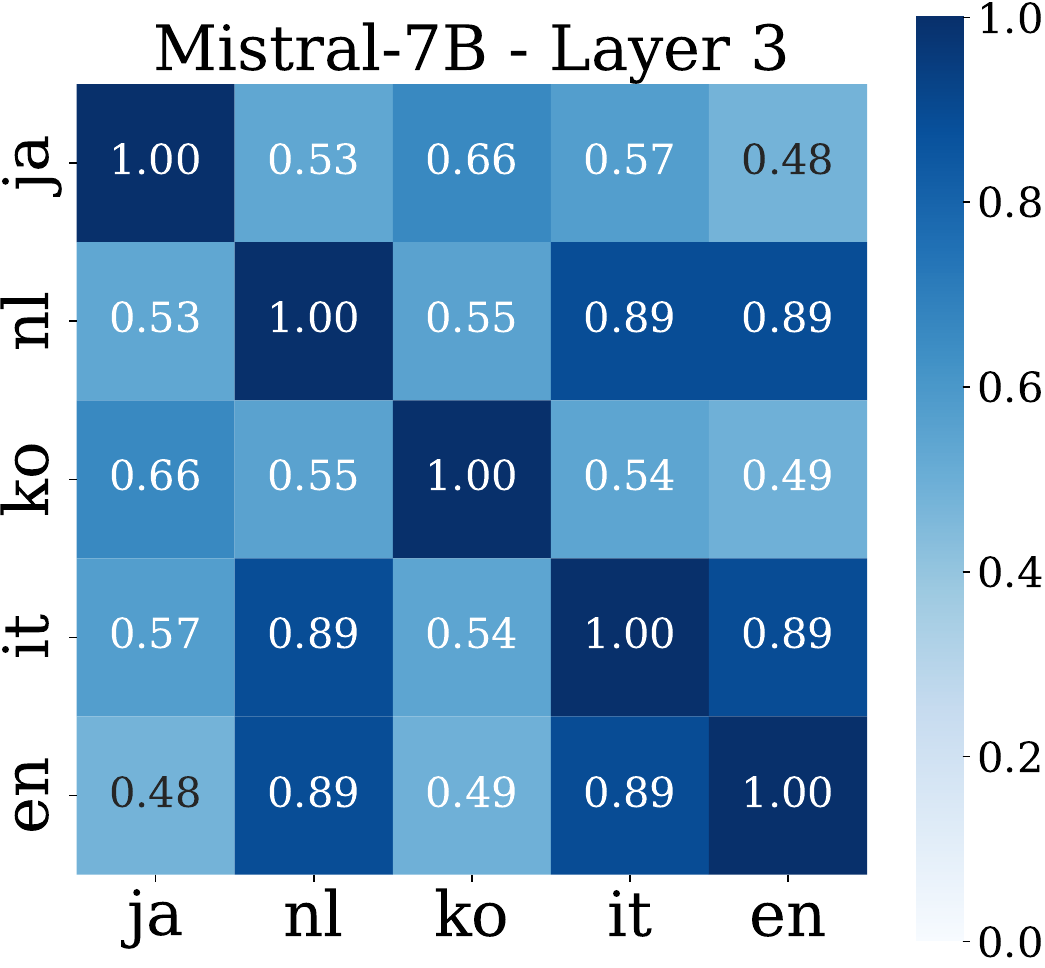}
  \includegraphics[width=0.19\linewidth]{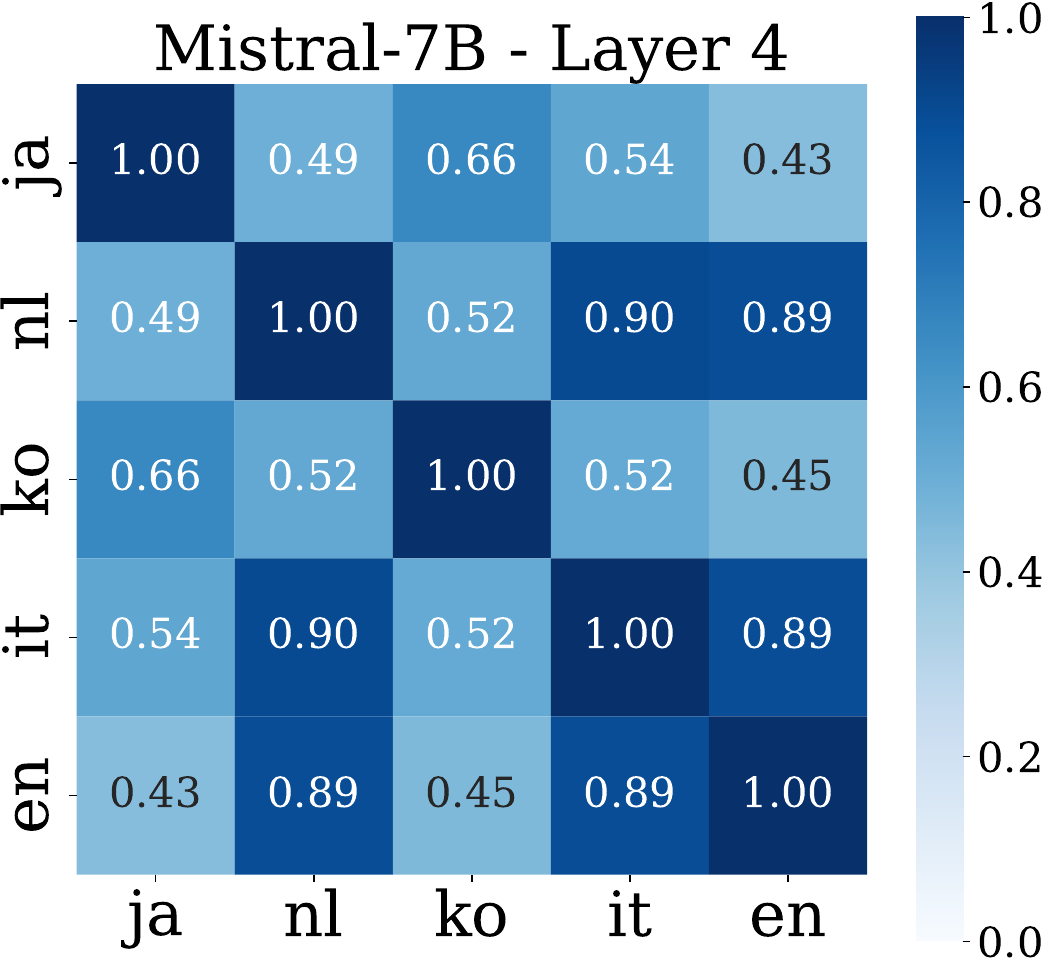}

  \includegraphics[width=0.19\linewidth]{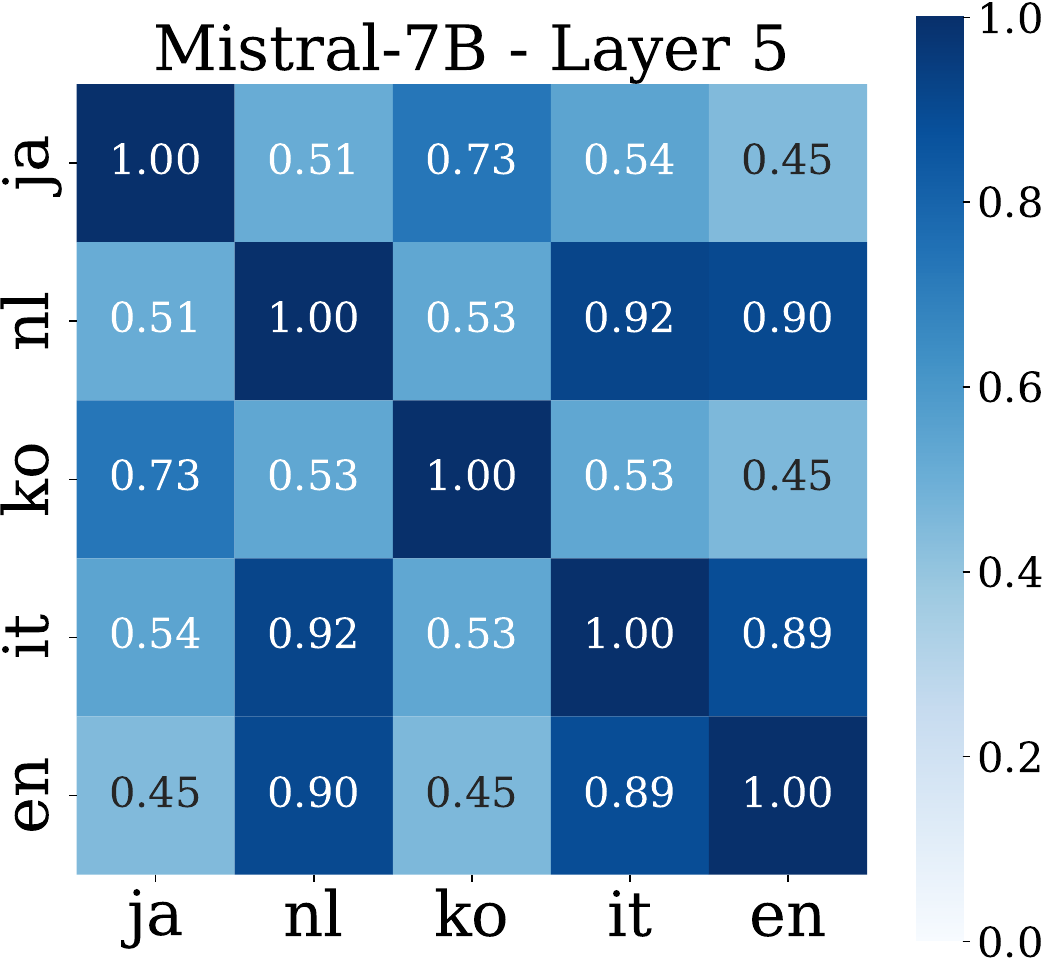}
  \includegraphics[width=0.19\linewidth]{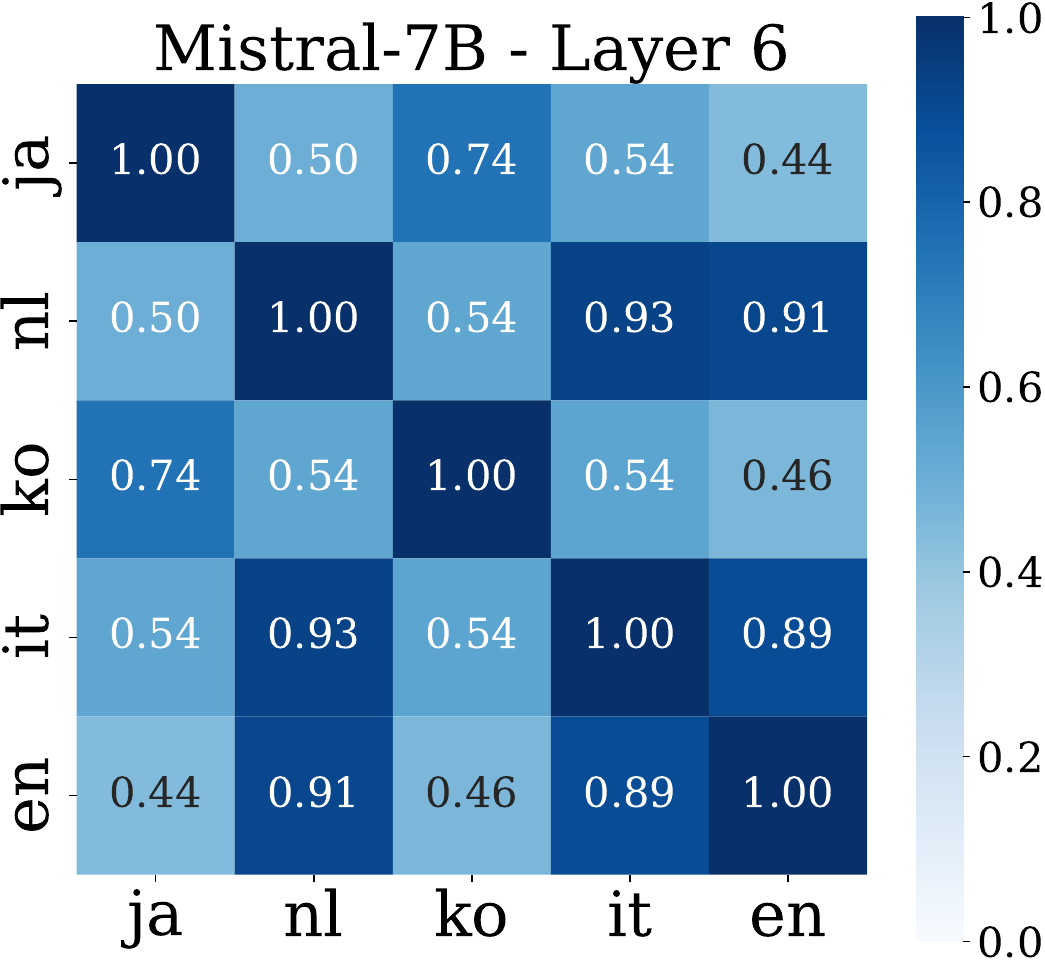}
  \includegraphics[width=0.19\linewidth]{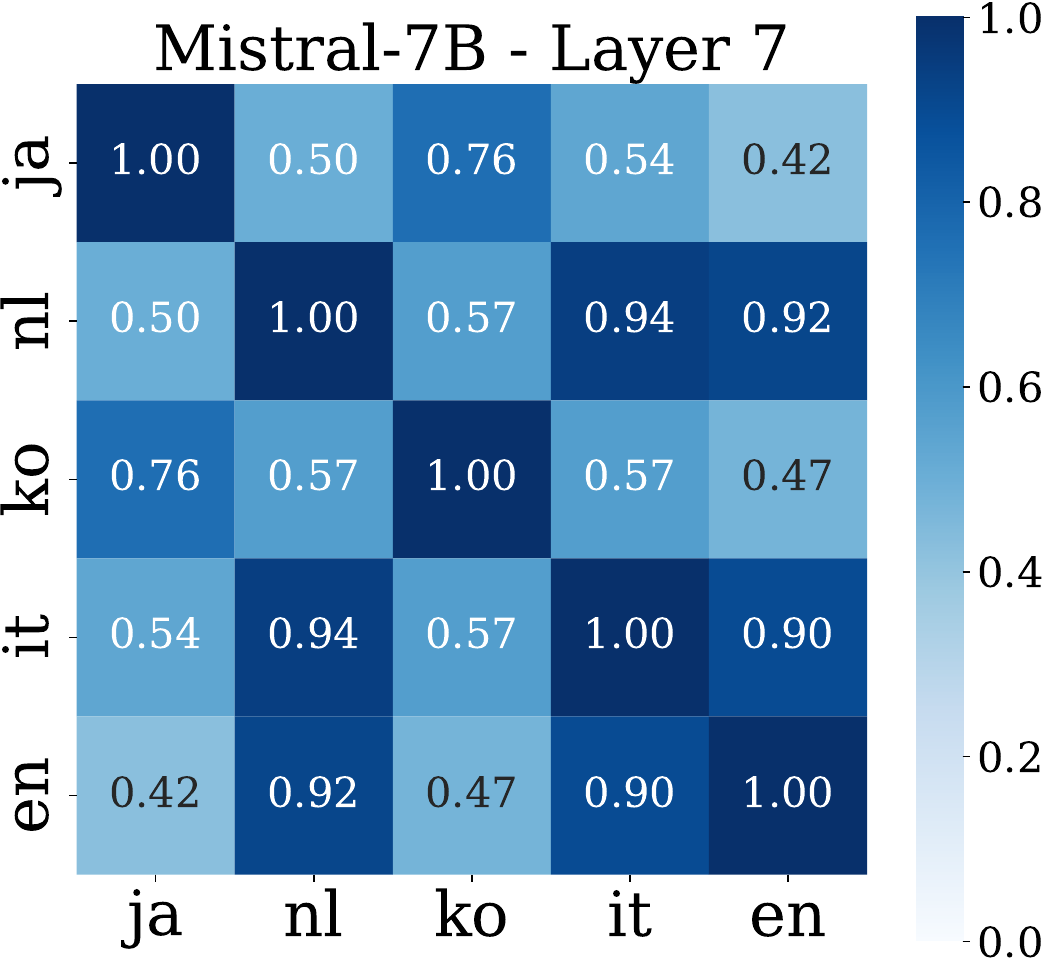}
  \includegraphics[width=0.19\linewidth]{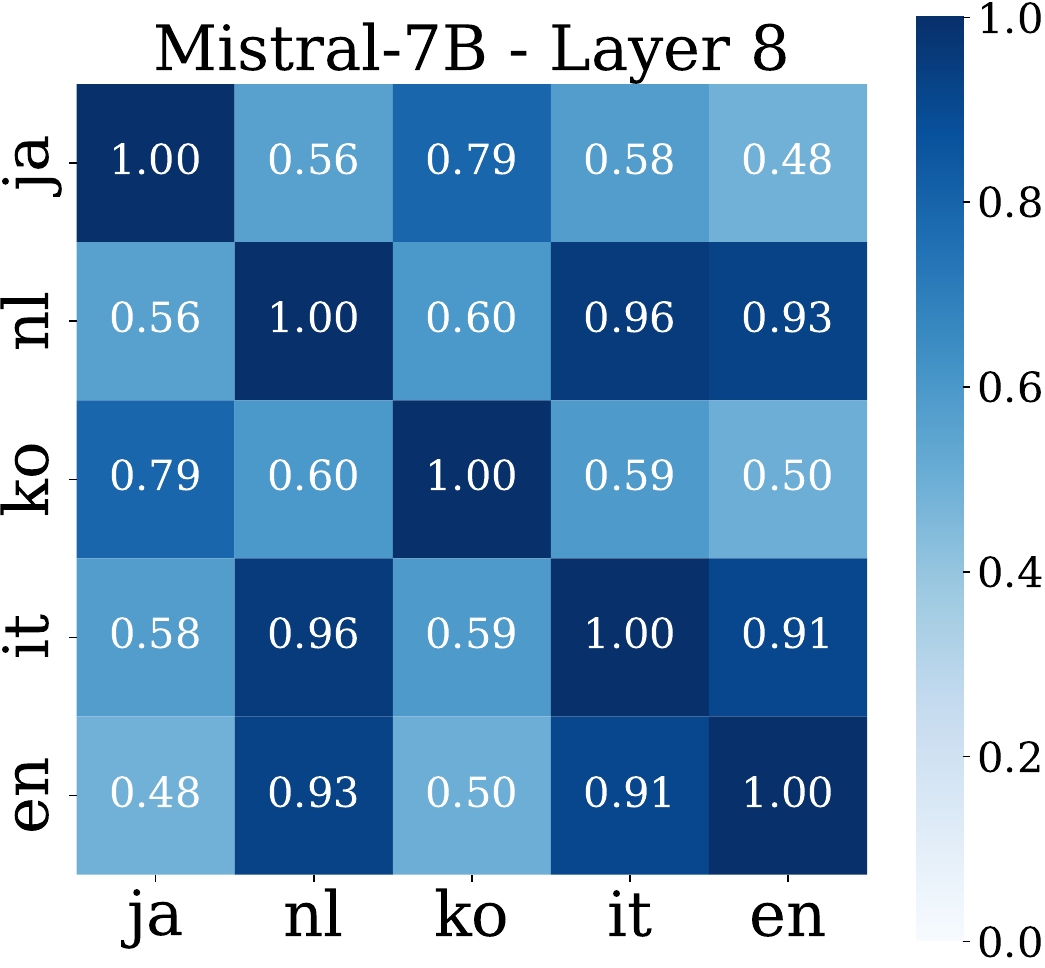}
  \includegraphics[width=0.19\linewidth]{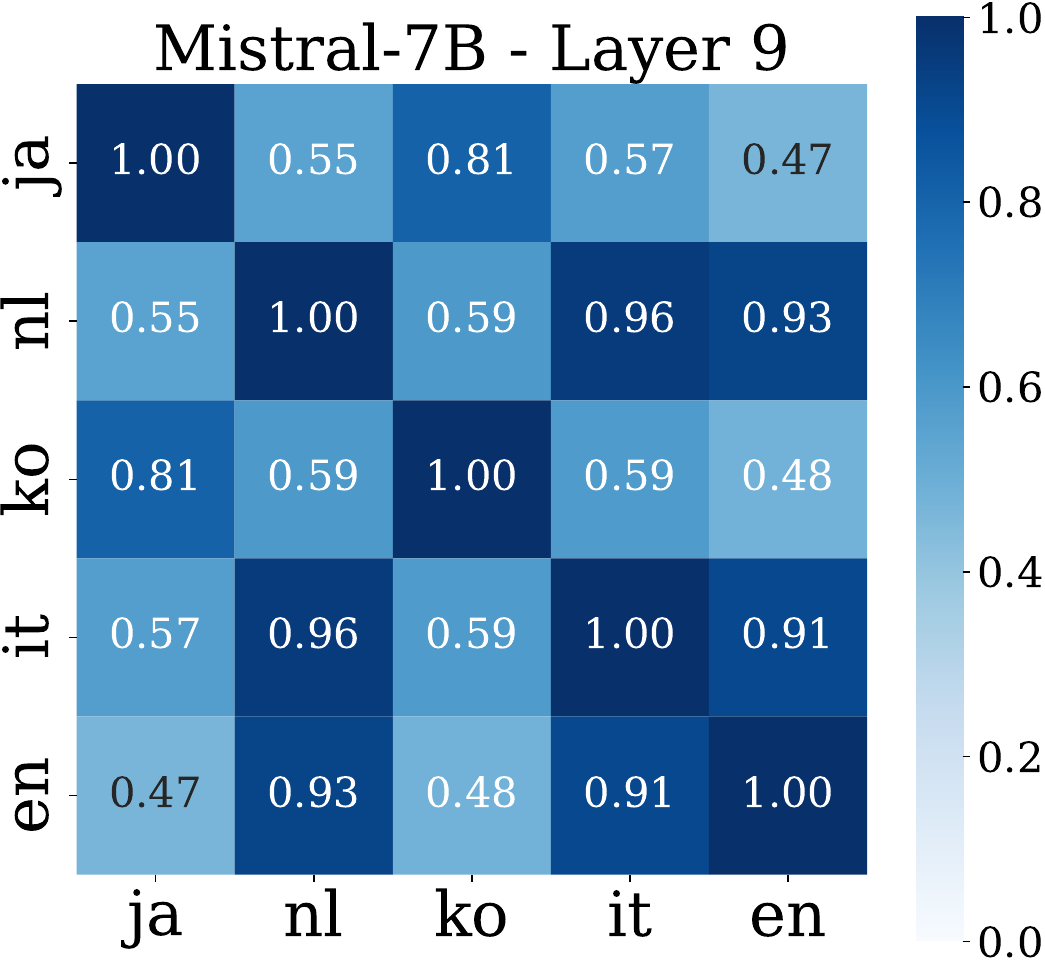}

  \includegraphics[width=0.19\linewidth]{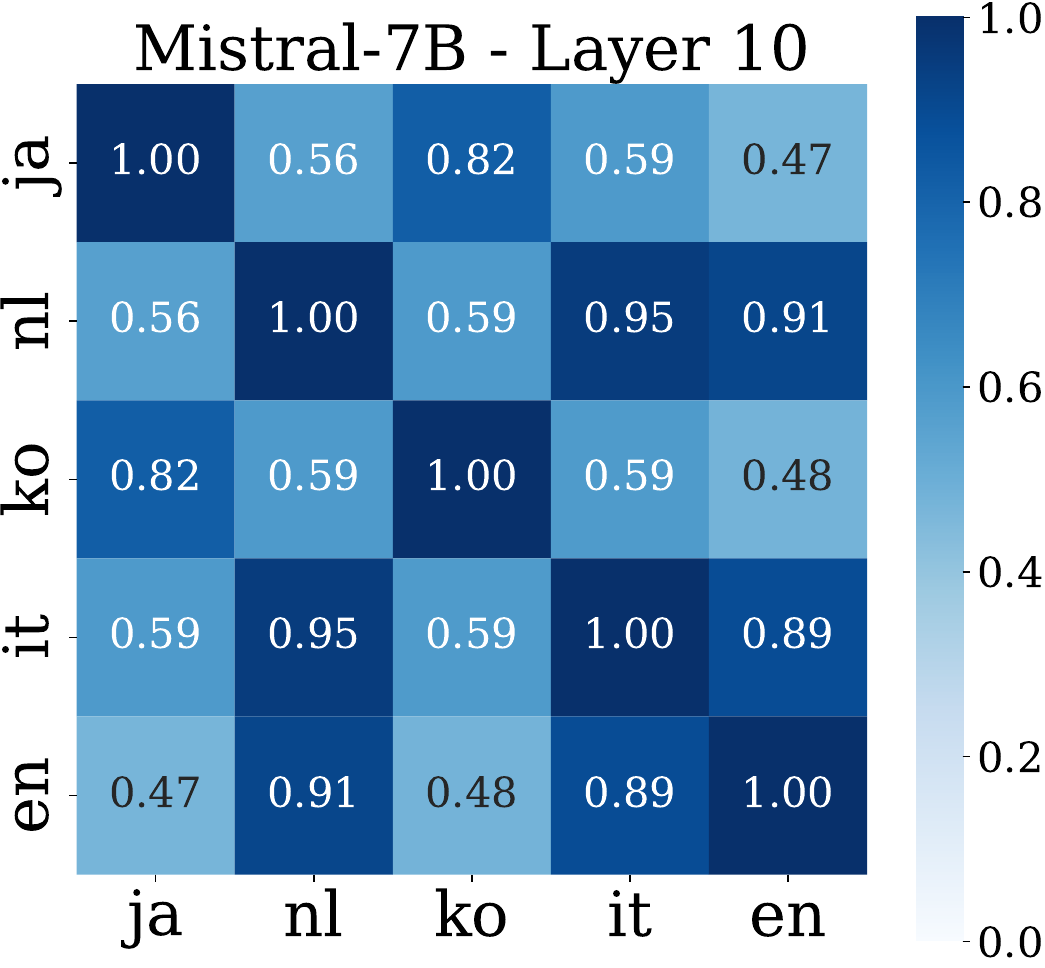}
  \includegraphics[width=0.19\linewidth]{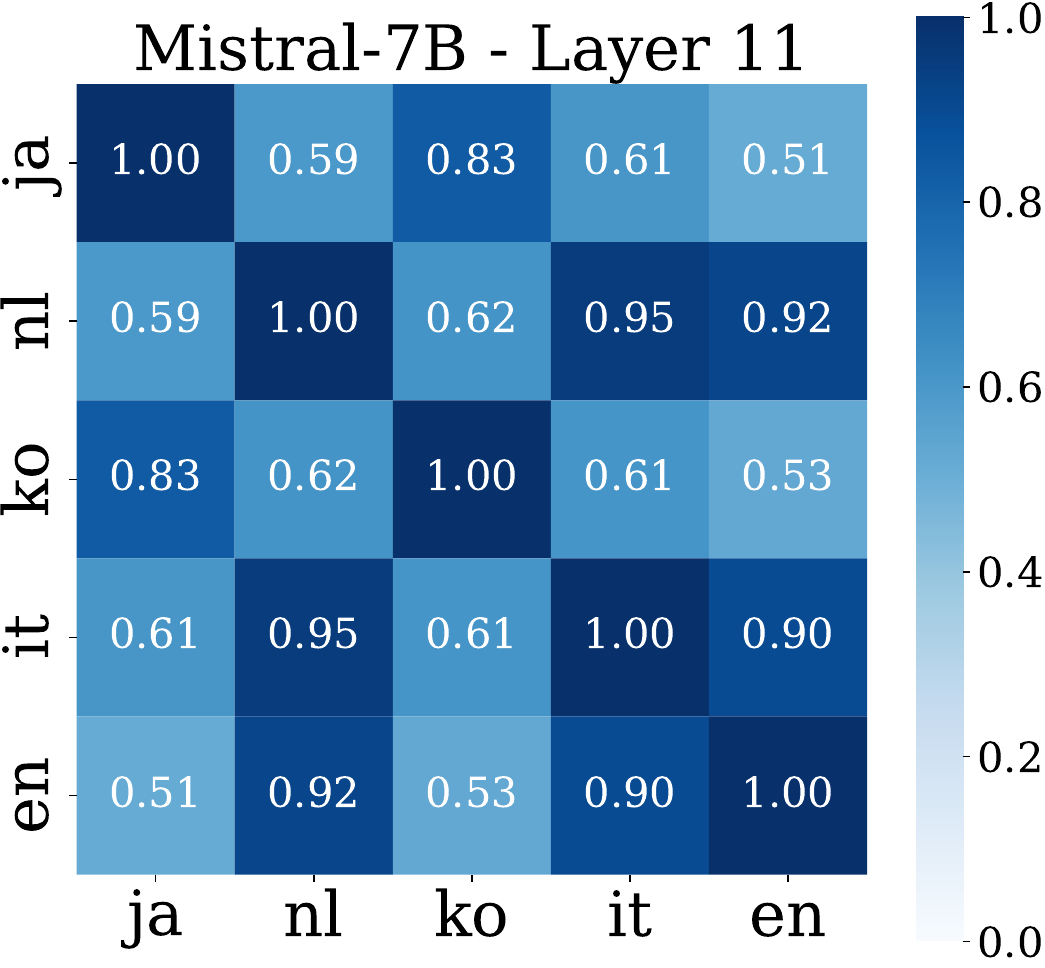}
  \includegraphics[width=0.19\linewidth]{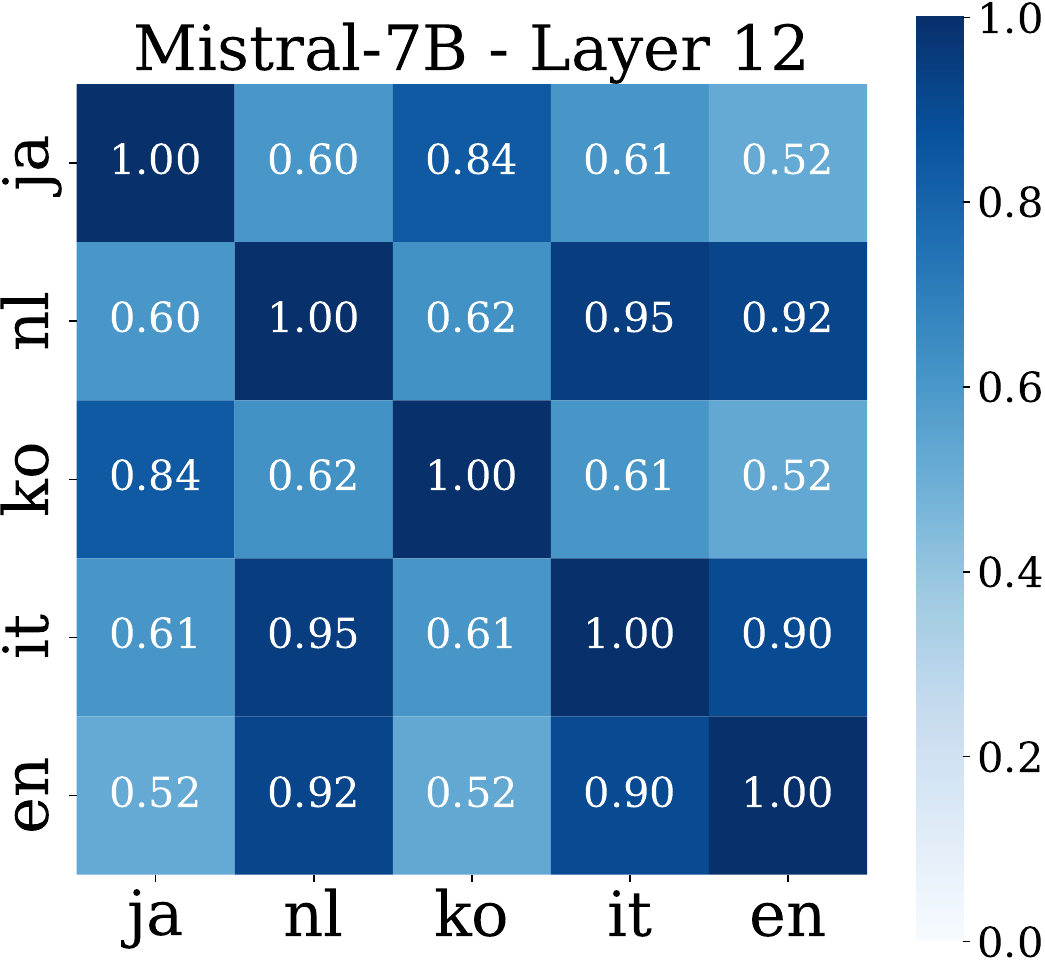}
  \includegraphics[width=0.19\linewidth]{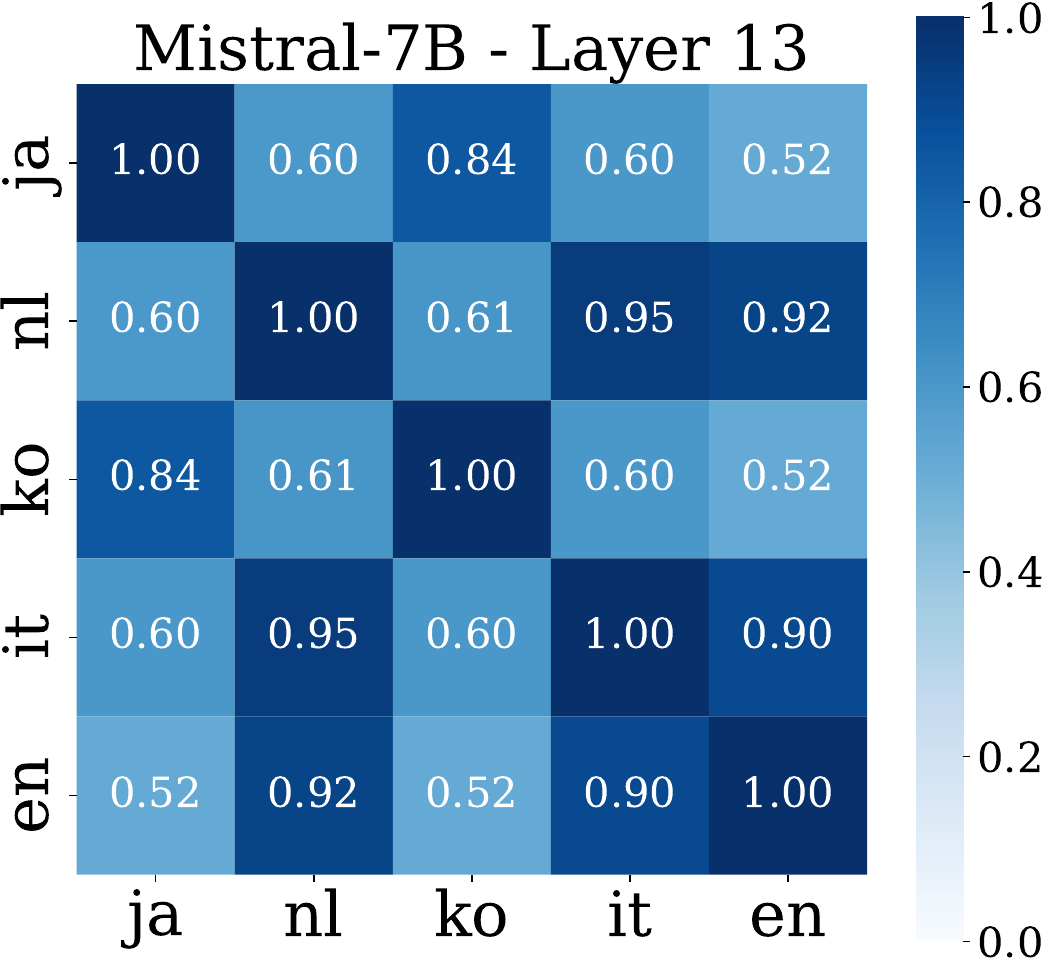}
  \includegraphics[width=0.19\linewidth]{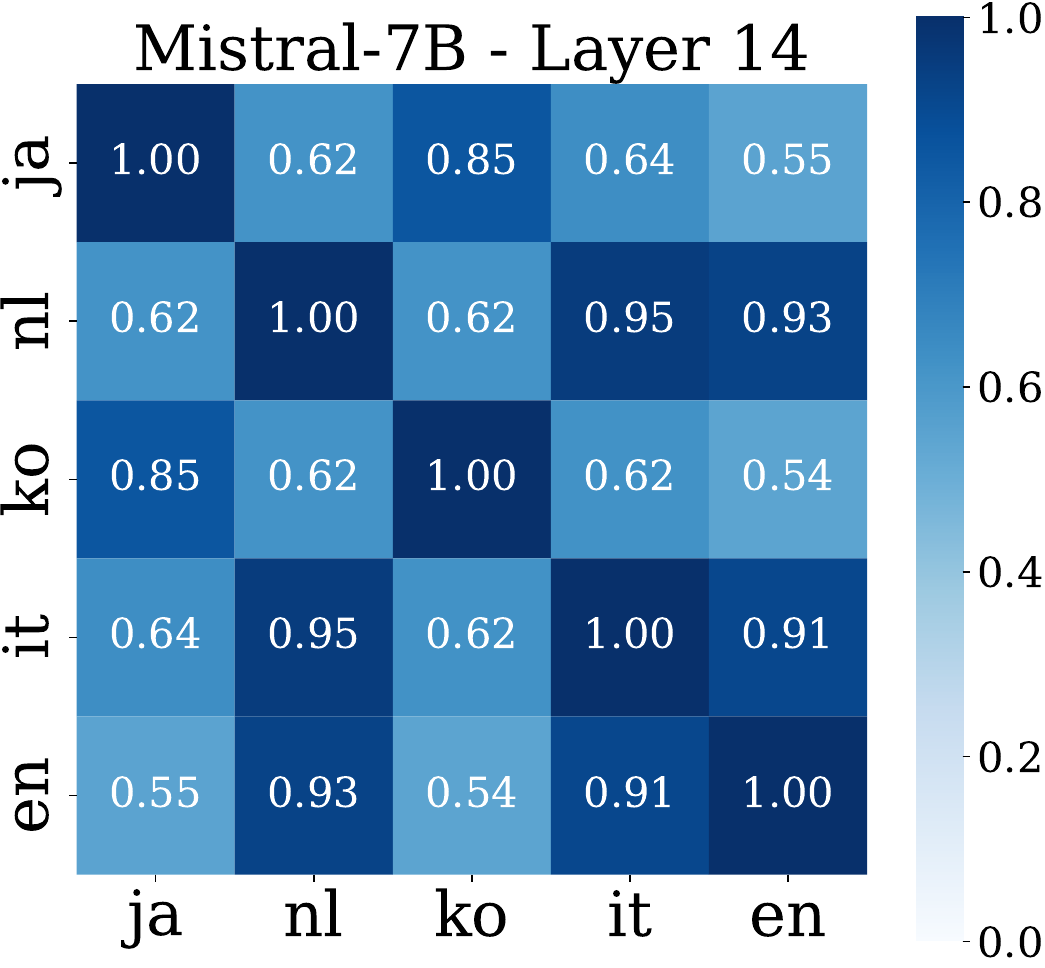}

  \includegraphics[width=0.19\linewidth]{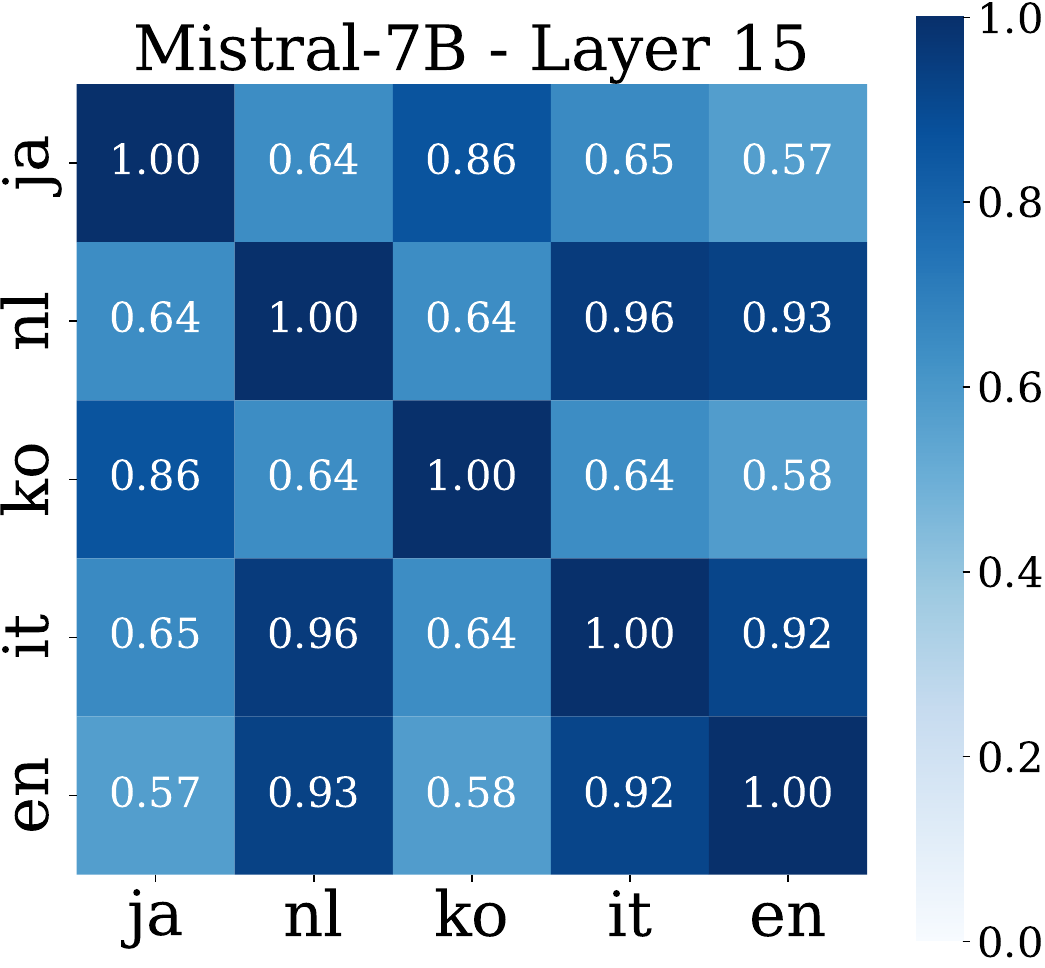}
  \includegraphics[width=0.19\linewidth]{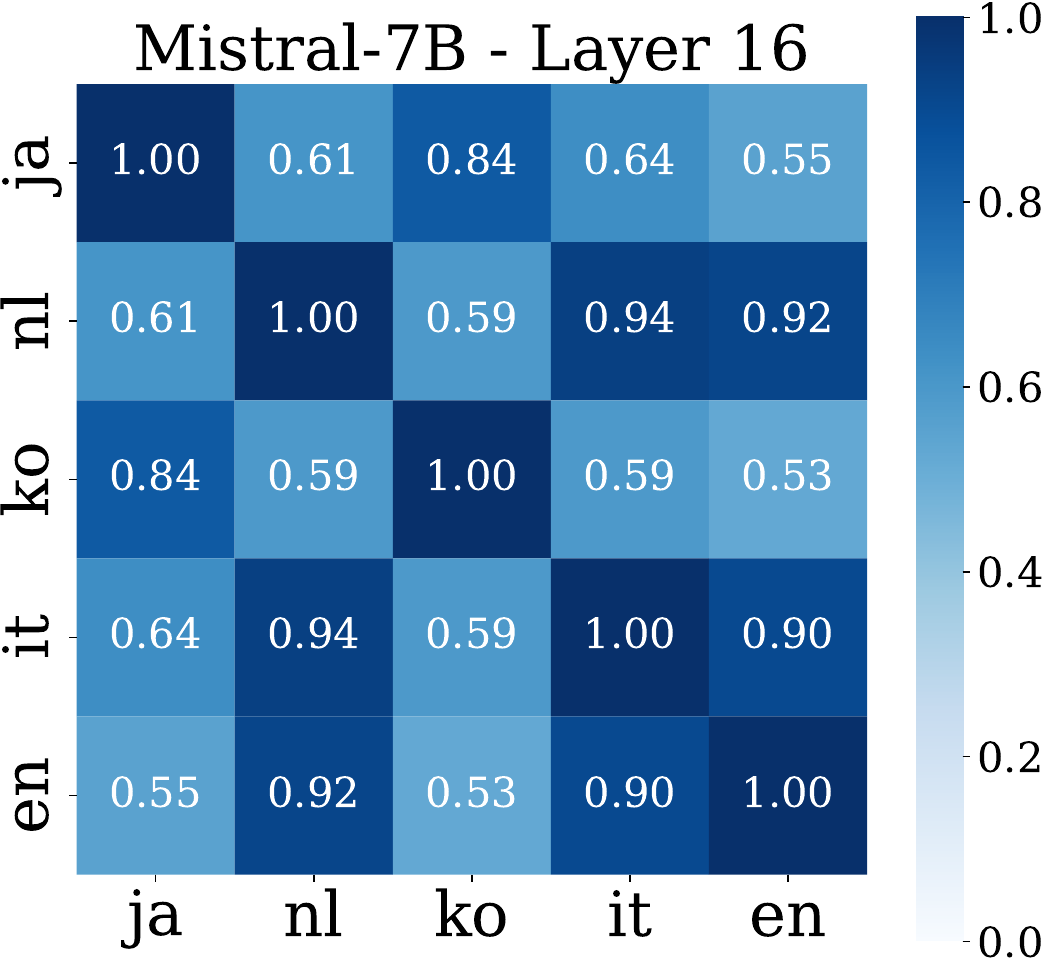}
  \includegraphics[width=0.19\linewidth]{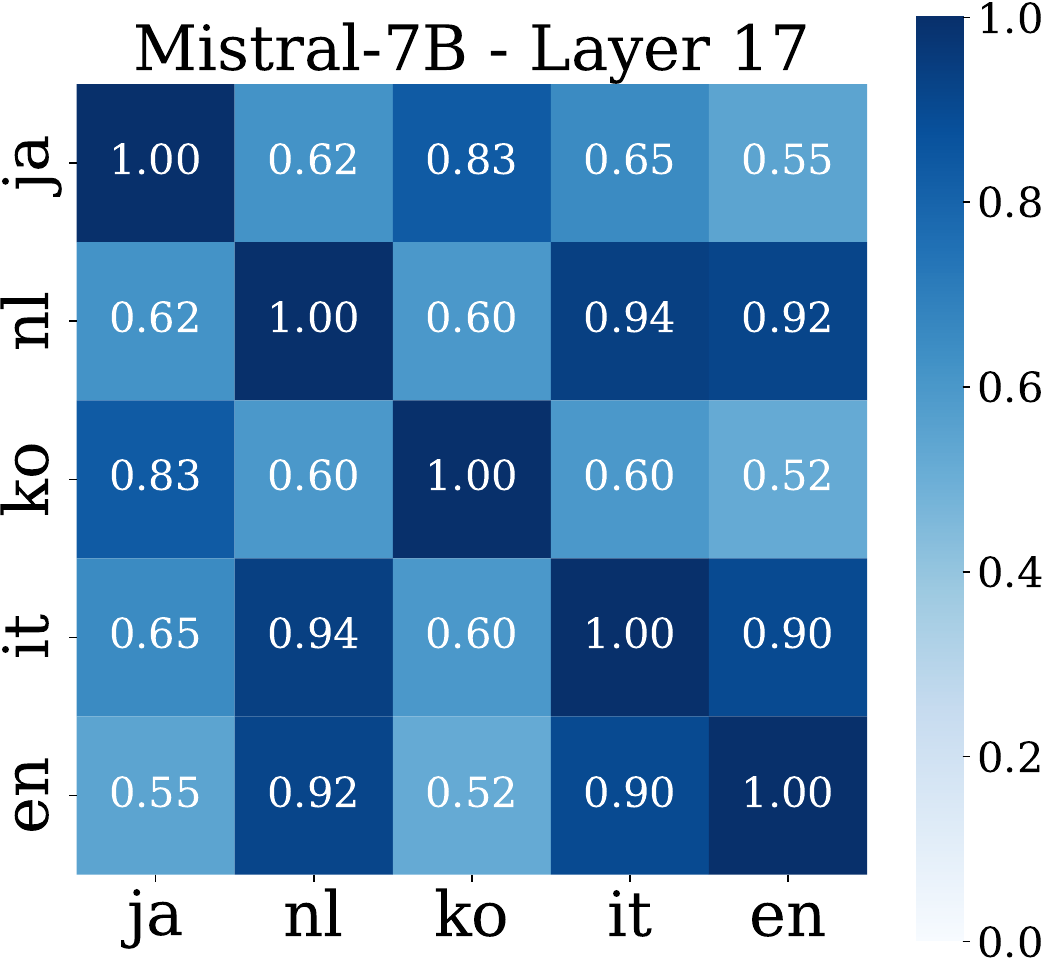}
  \includegraphics[width=0.19\linewidth]{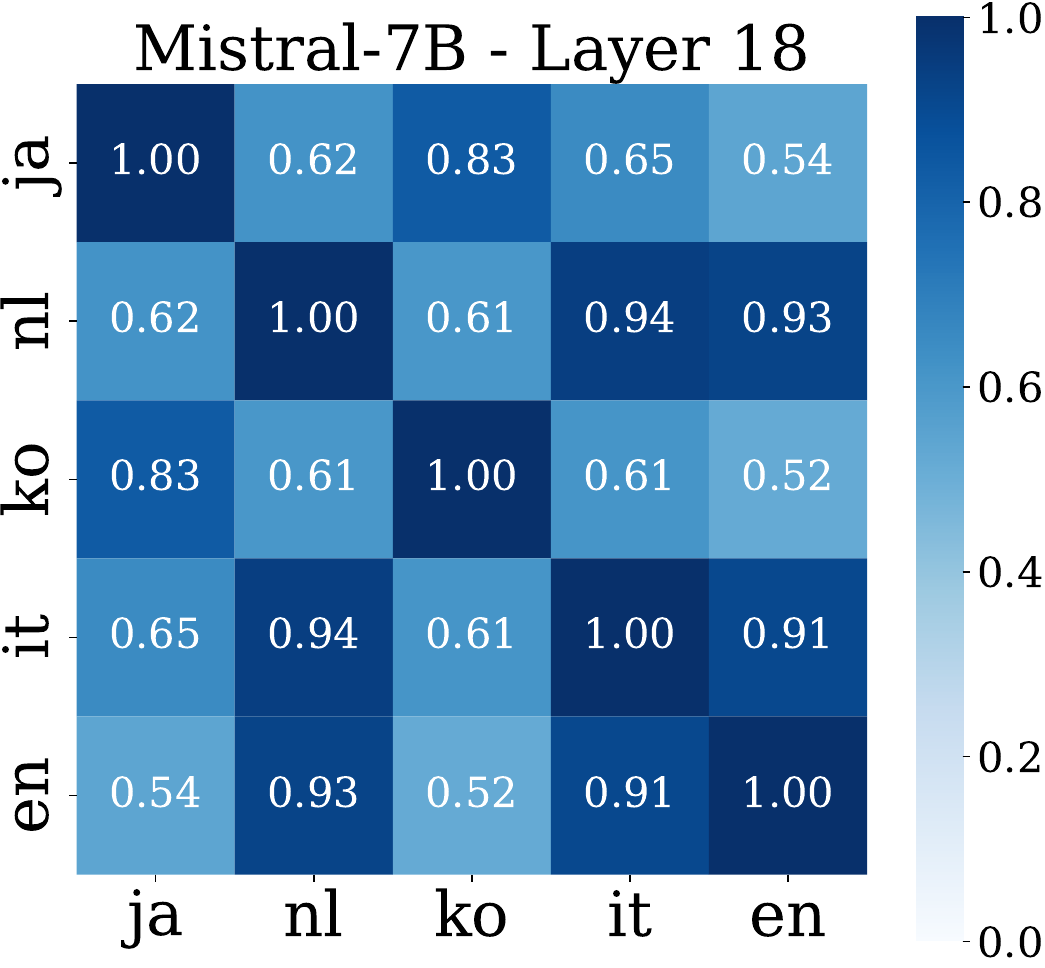}
  \includegraphics[width=0.19\linewidth]{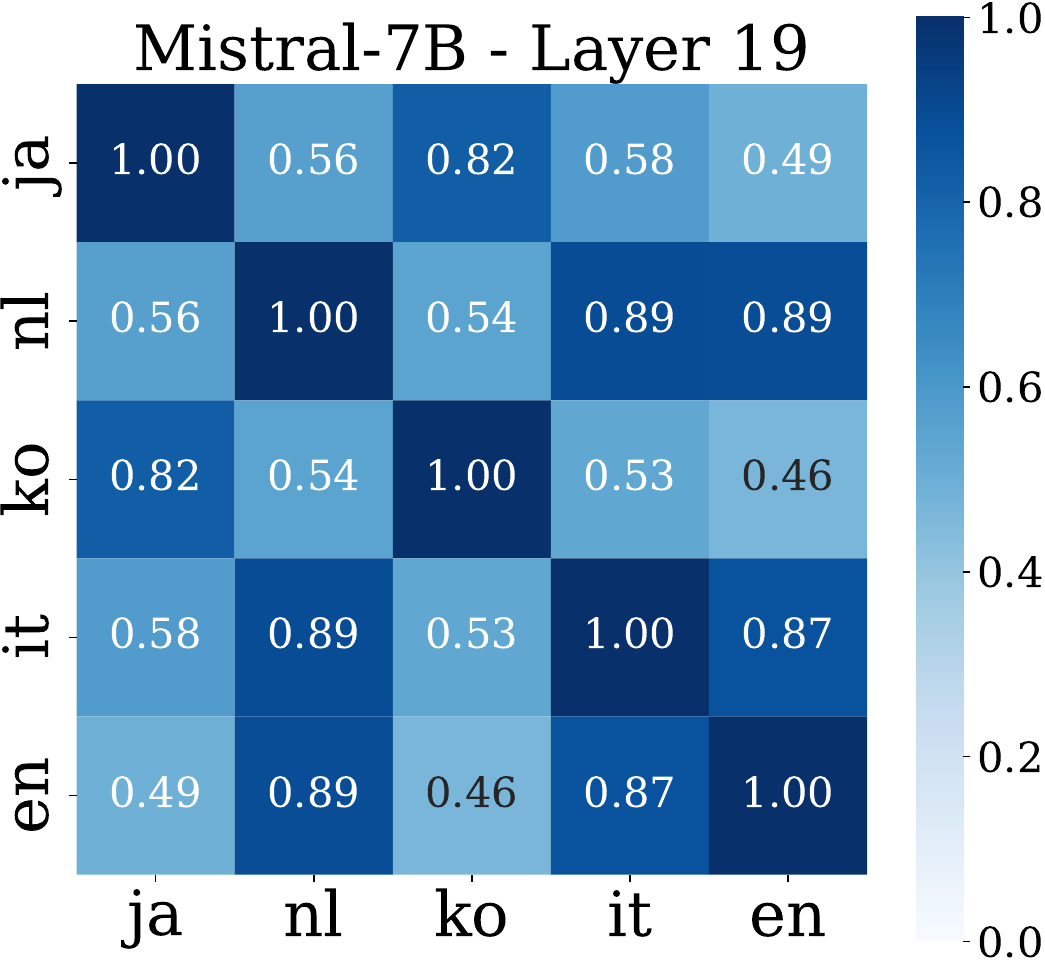}

  \includegraphics[width=0.19\linewidth]{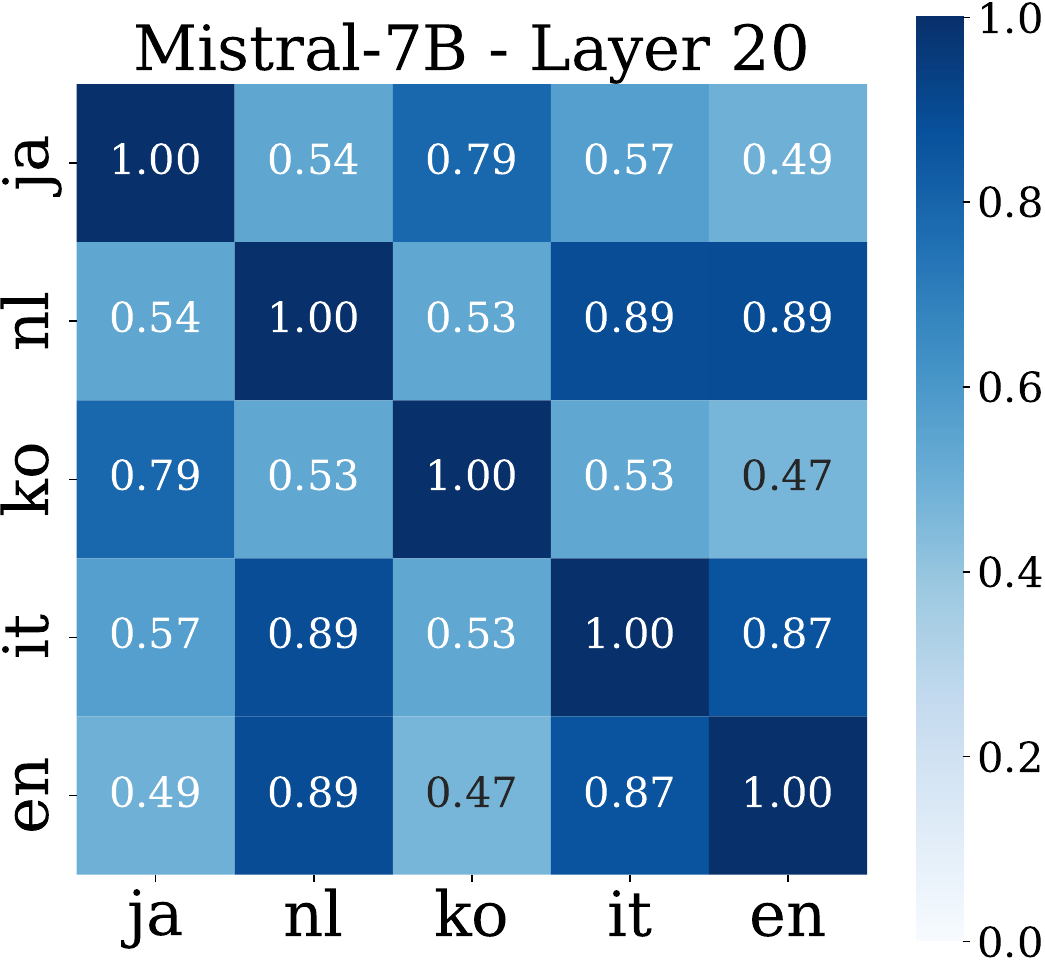}
  \includegraphics[width=0.19\linewidth]{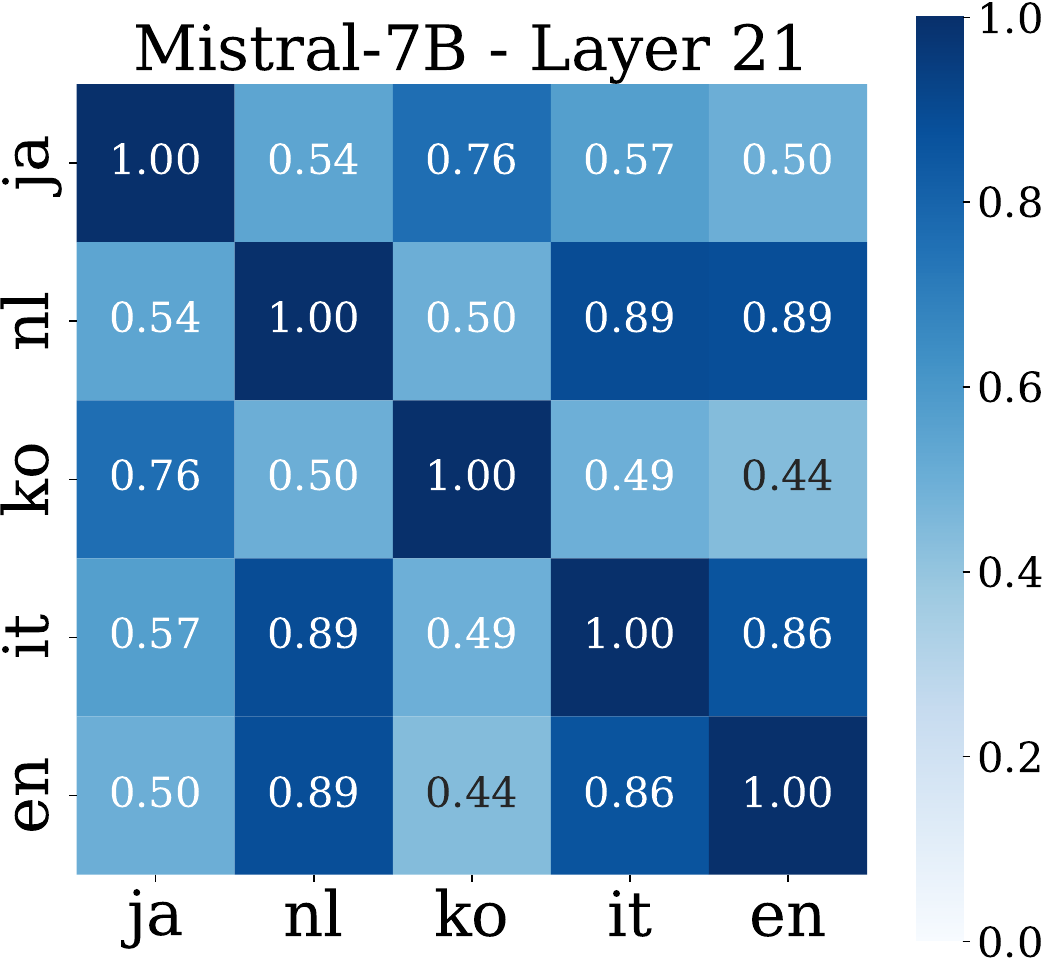}
  \includegraphics[width=0.19\linewidth]{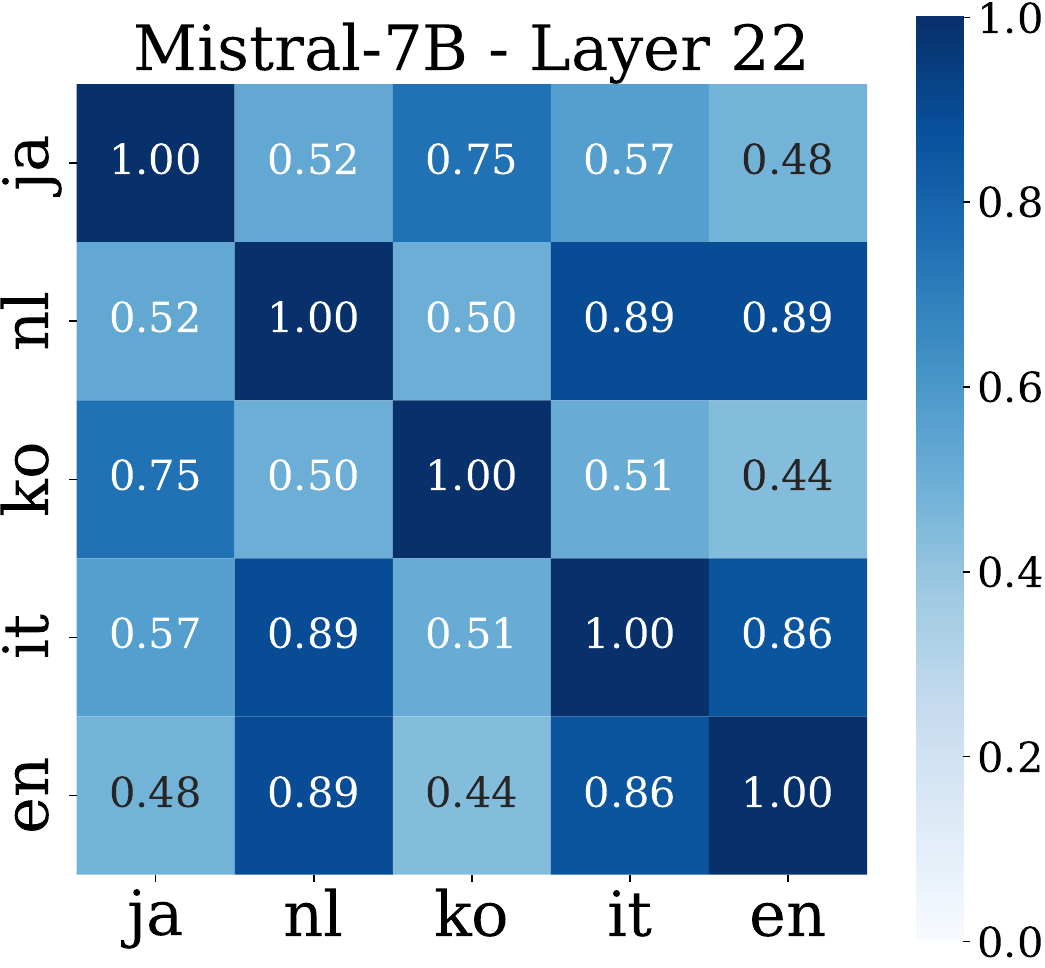}
  \includegraphics[width=0.19\linewidth]{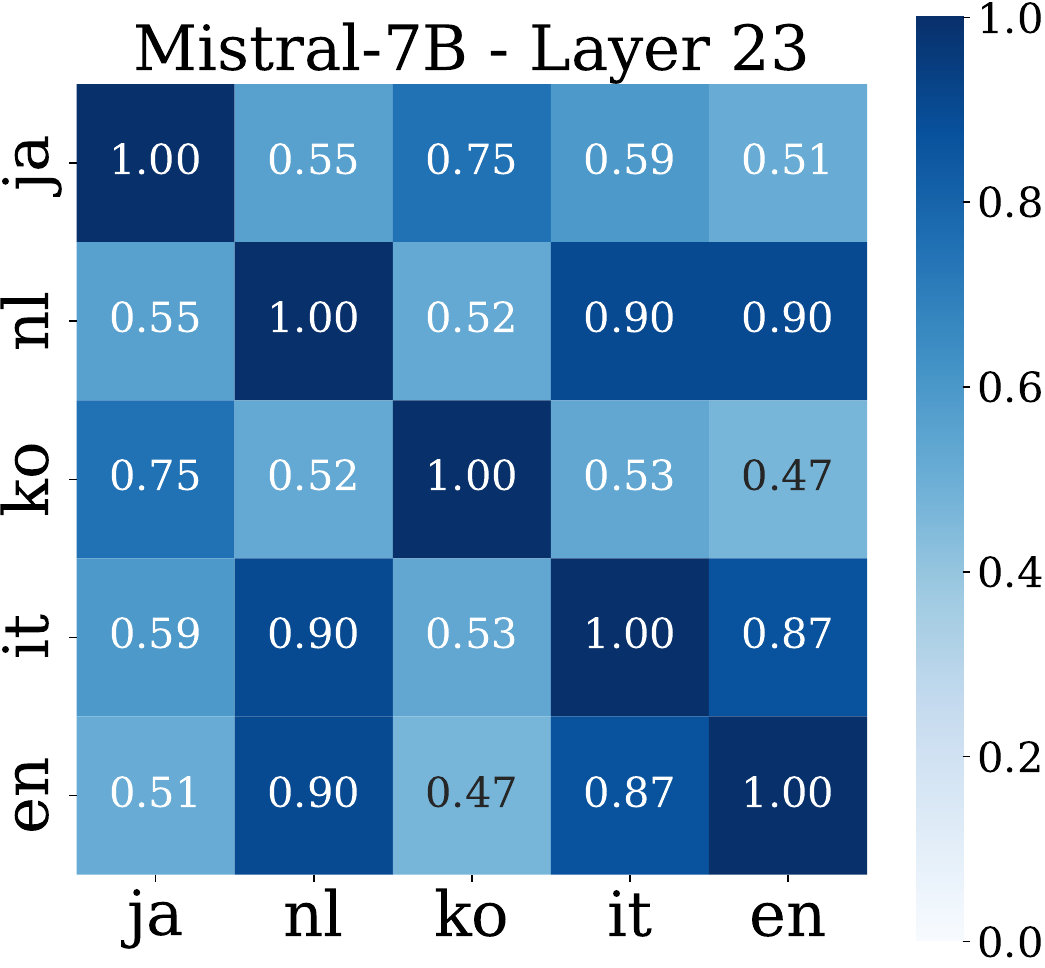}
  \includegraphics[width=0.19\linewidth]{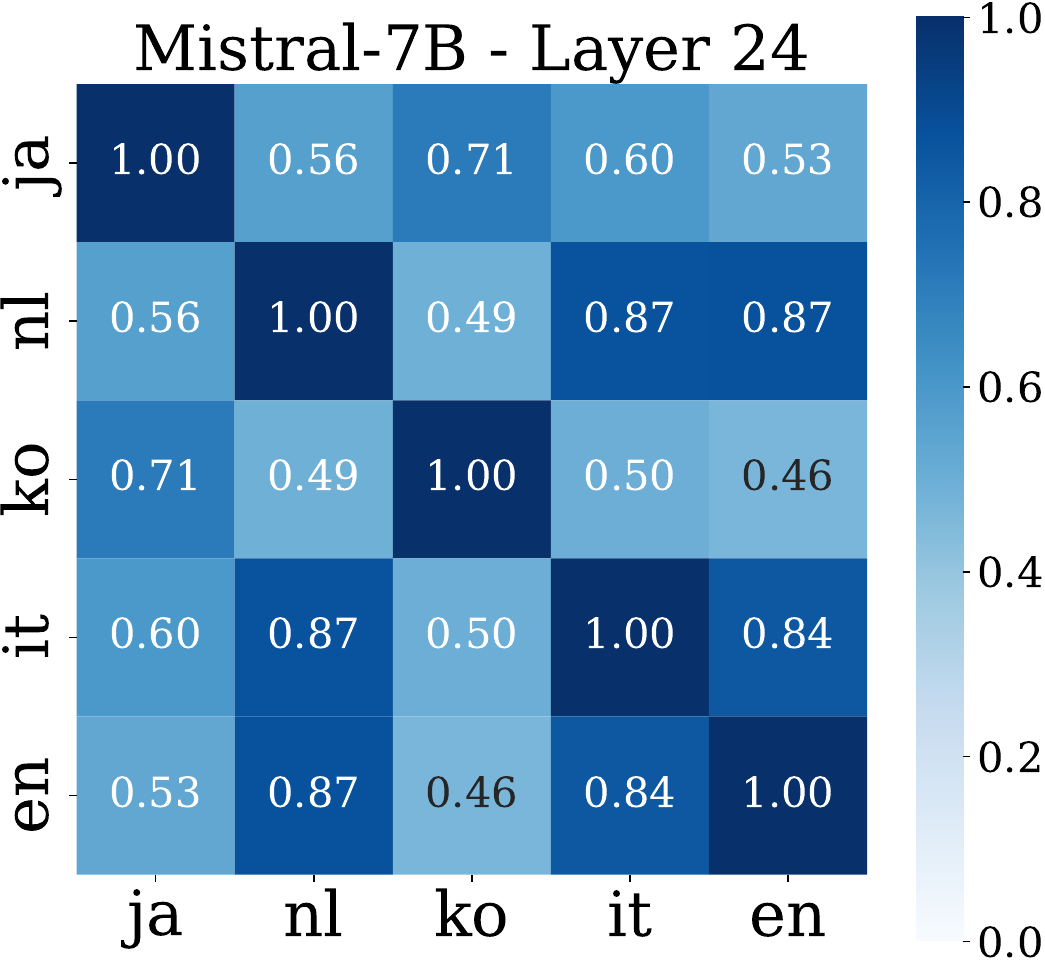}

  \includegraphics[width=0.19\linewidth]{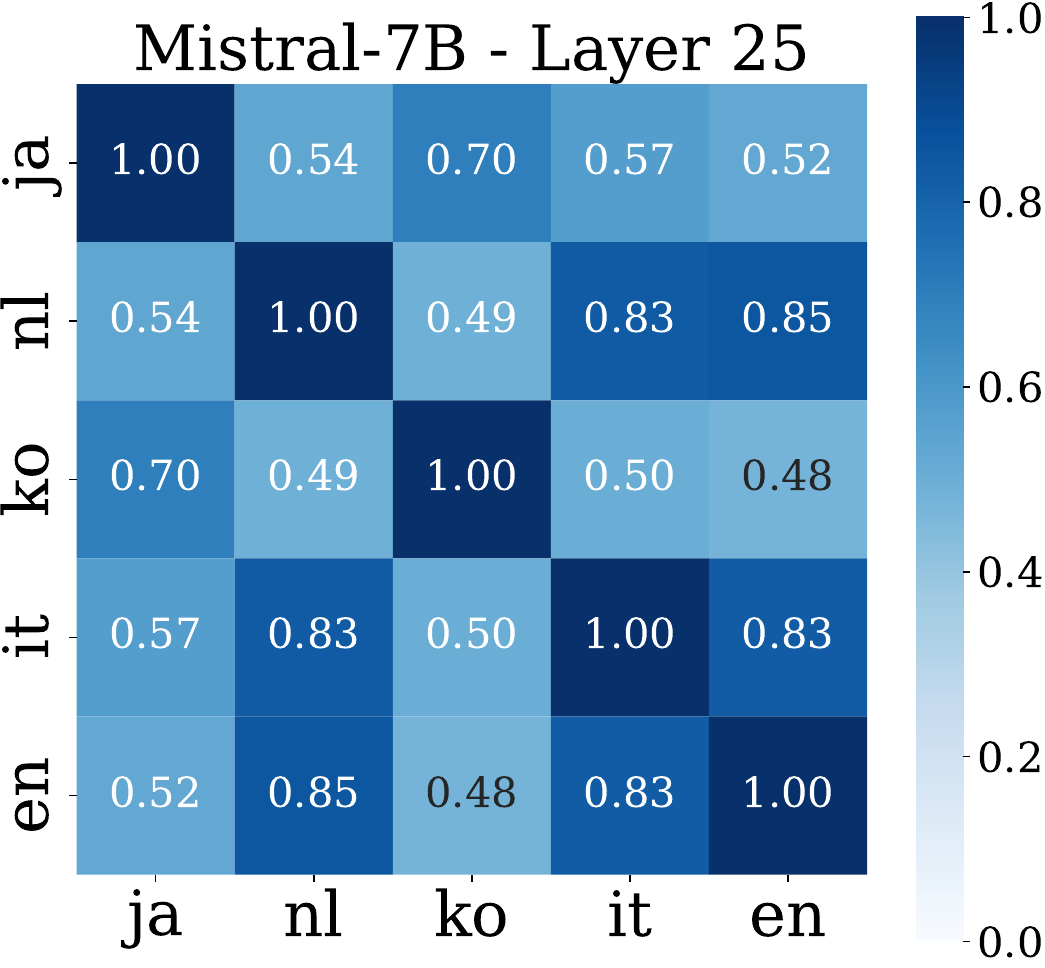}
  \includegraphics[width=0.19\linewidth]{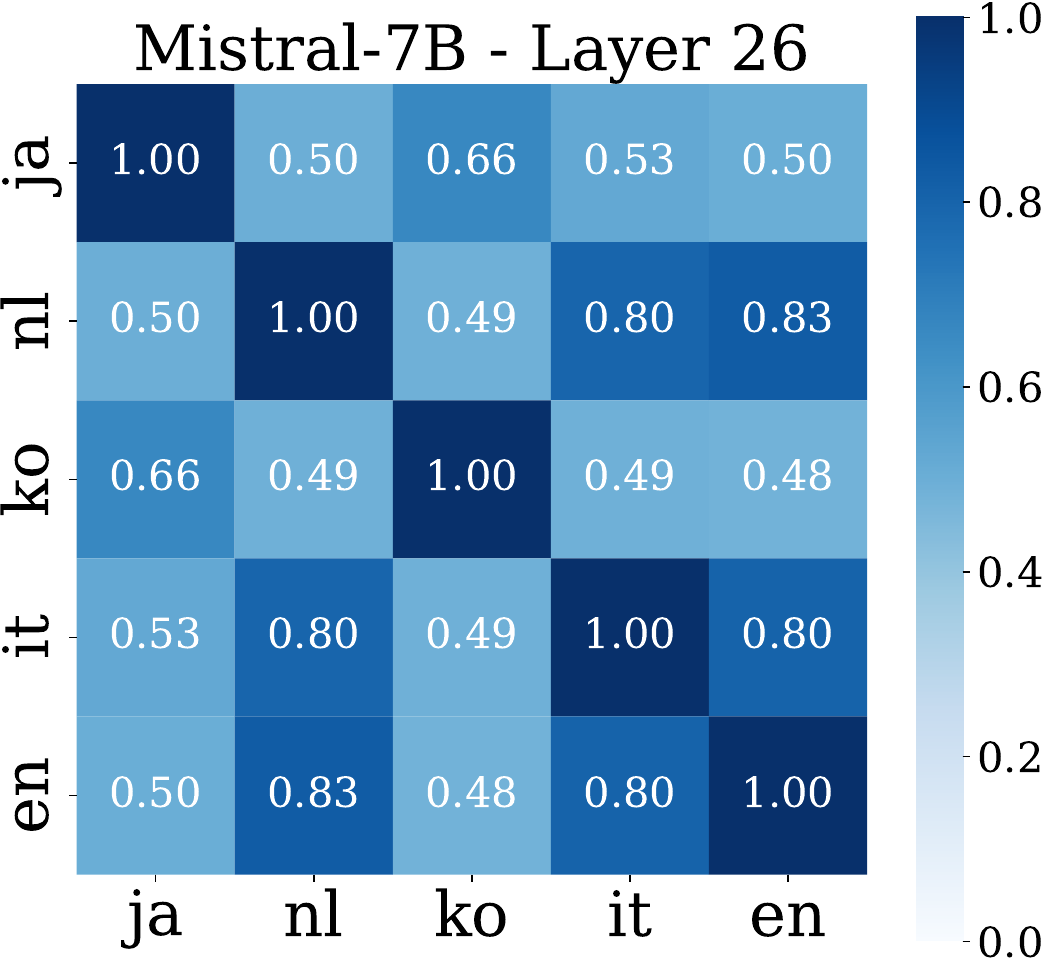}
  \includegraphics[width=0.19\linewidth]{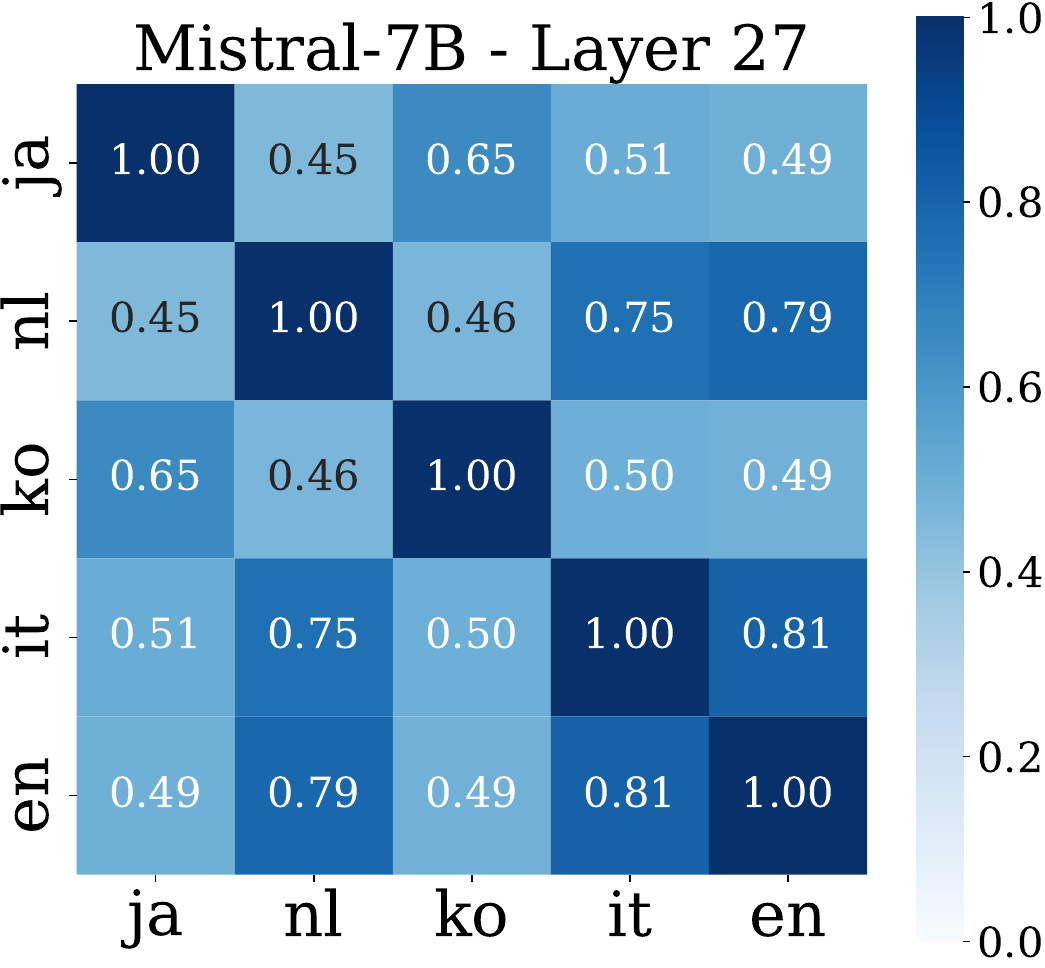}
  \includegraphics[width=0.19\linewidth]{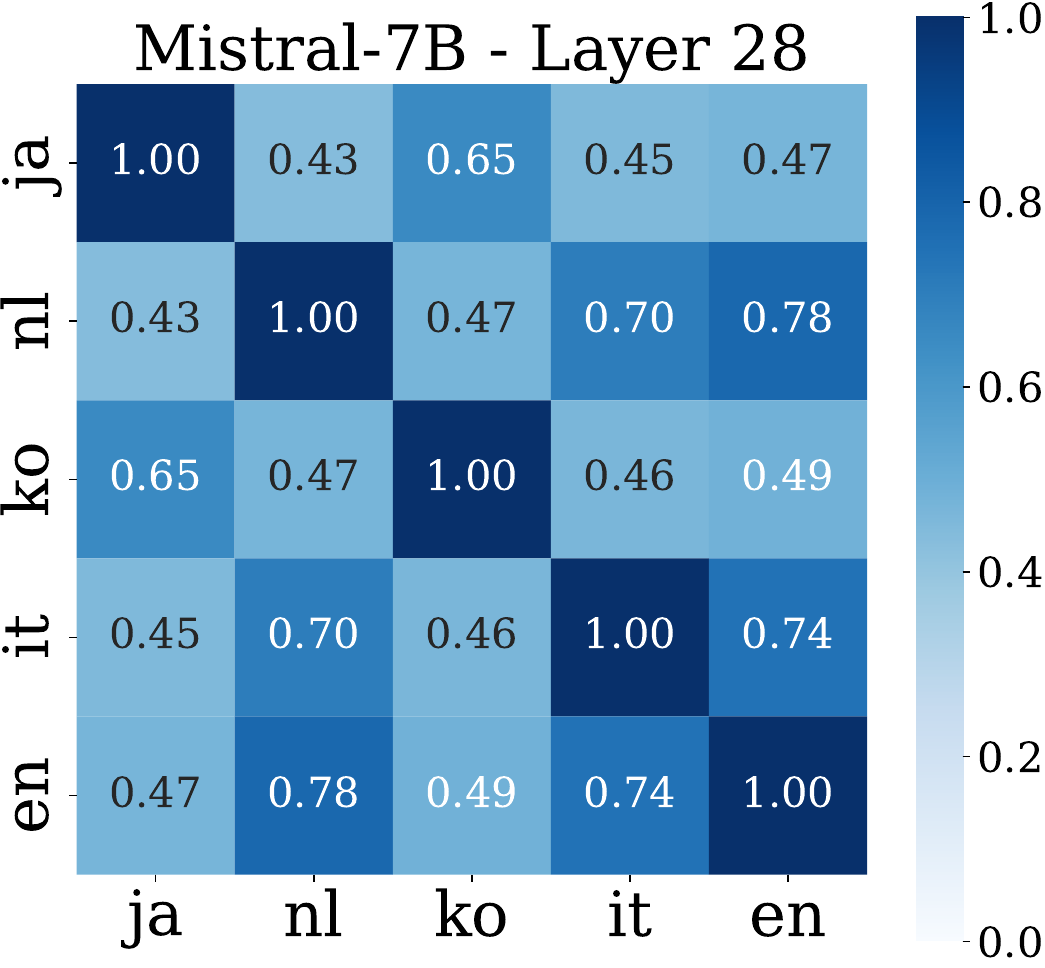}
  \includegraphics[width=0.19\linewidth]{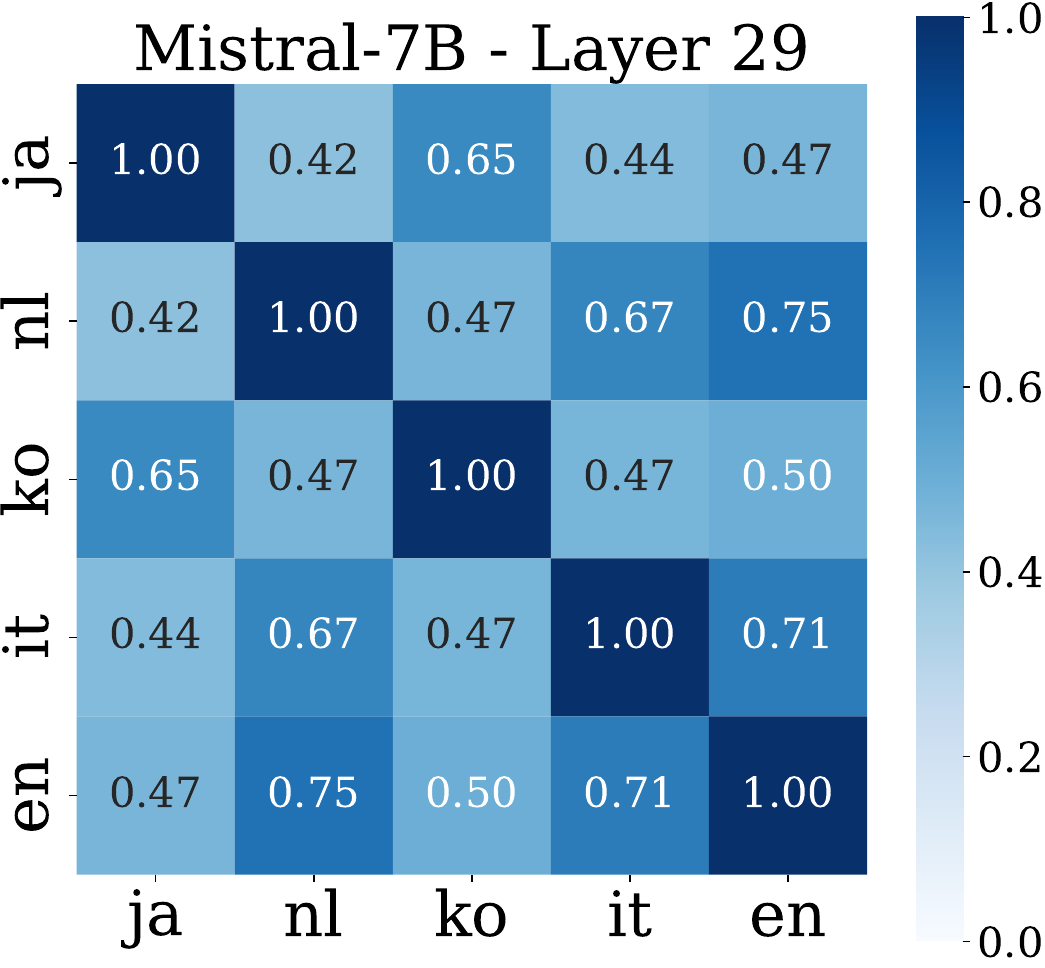}

  \includegraphics[width=0.19\linewidth]{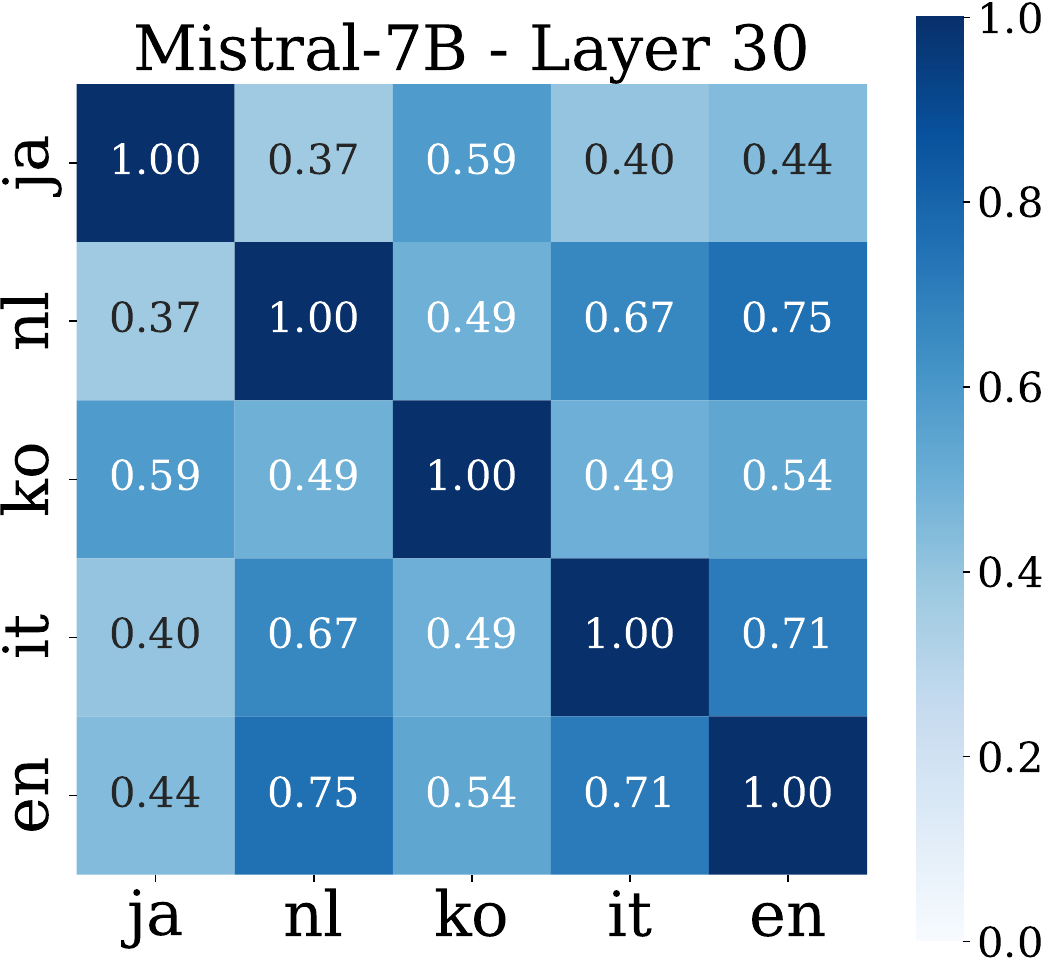}
  \includegraphics[width=0.19\linewidth]{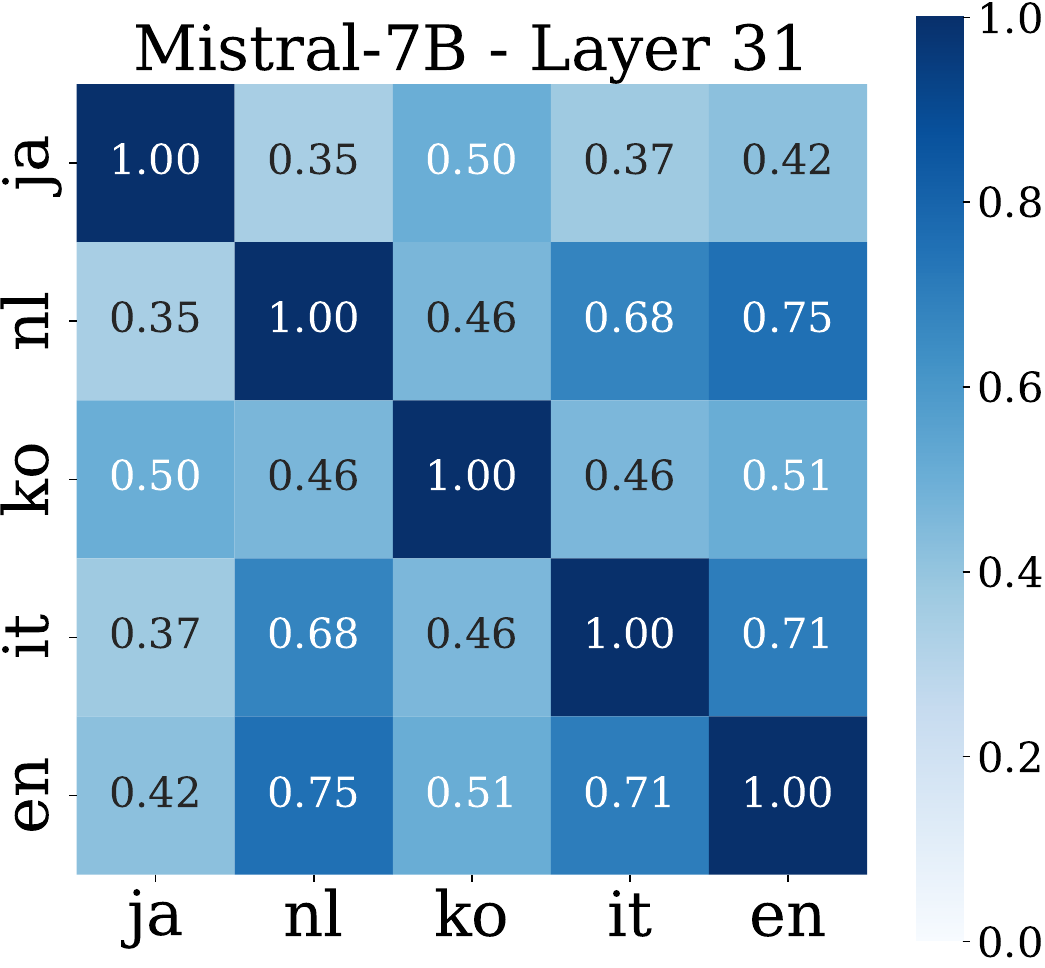}
  \includegraphics[width=0.19\linewidth]{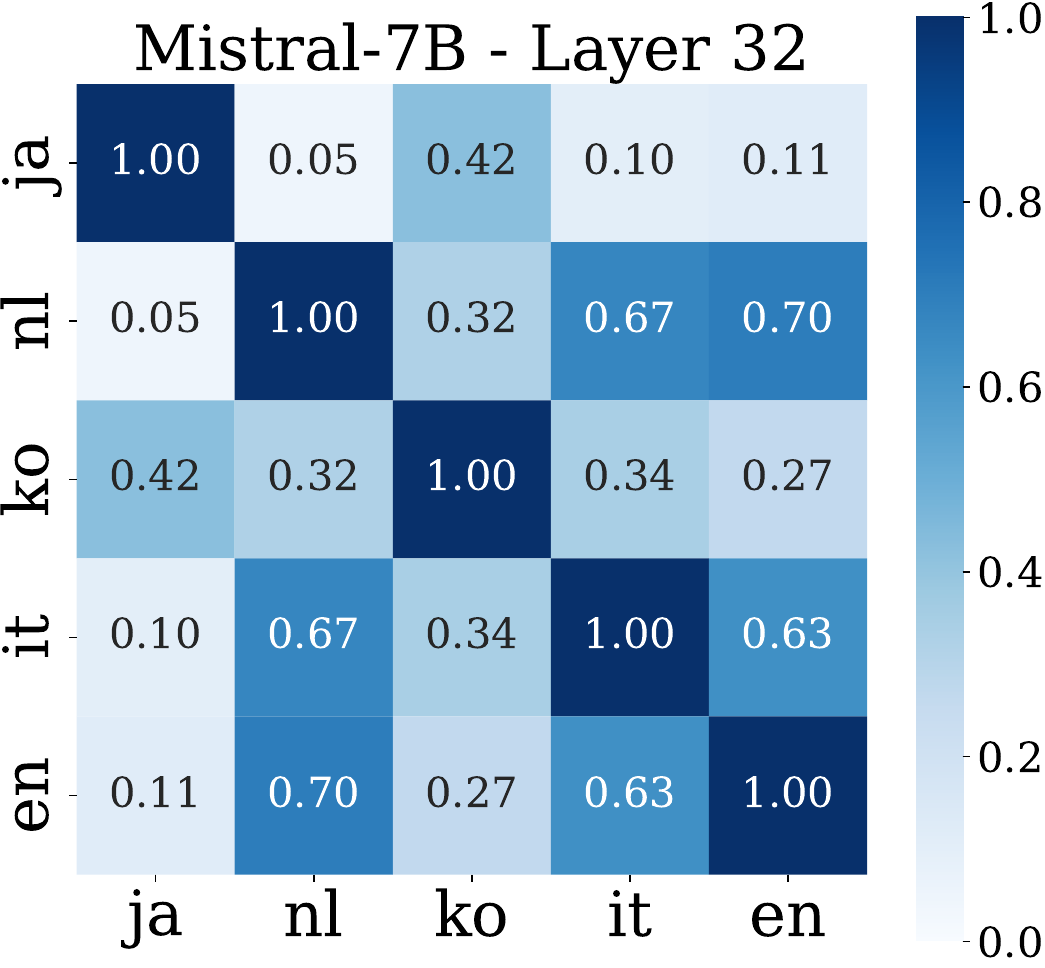}

  \caption{\textbf{The distance among centroids of language latent spaces (Mistral-7B).}}
  \label{fig:appendix:distance among subspaces centoids mistral}
\end{figure*}
% aya, distance among subspaces, centroids
\begin{figure*}[t]
  \centering

  \includegraphics[width=0.19\linewidth]{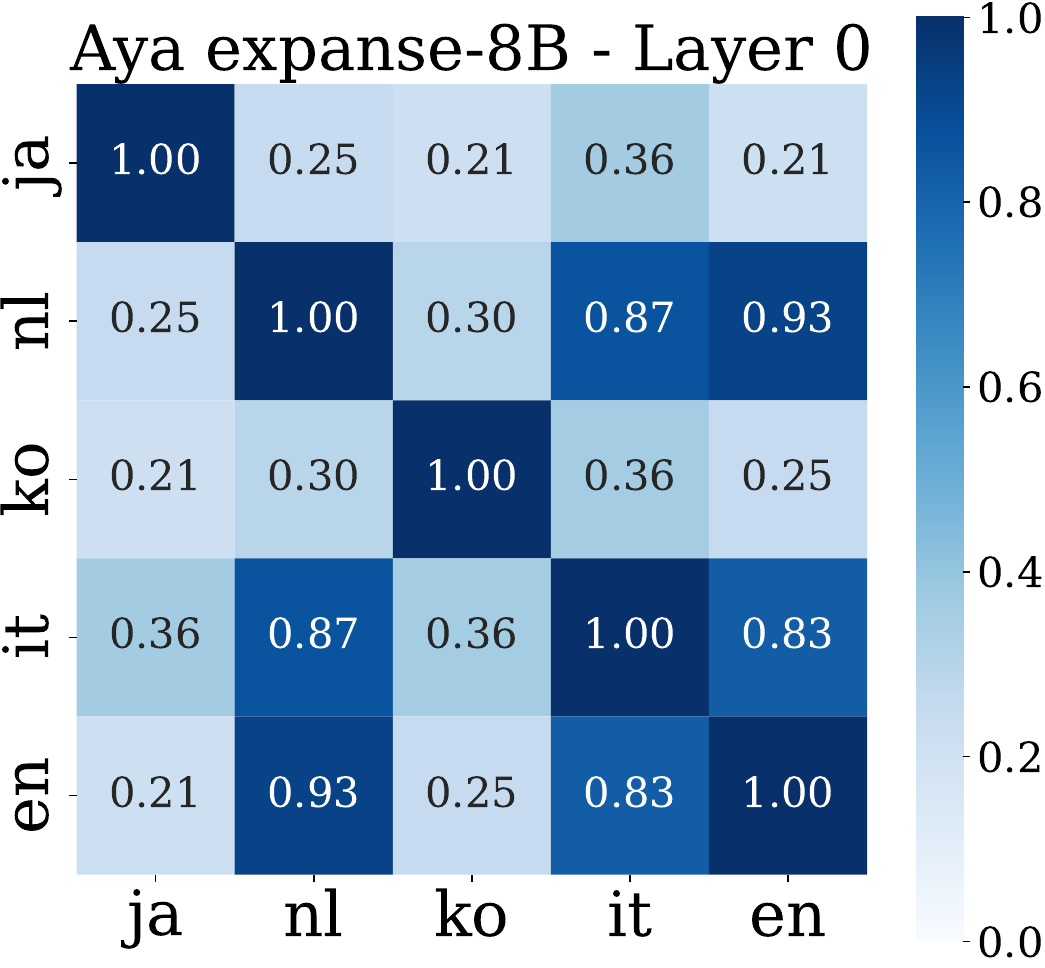}
  \includegraphics[width=0.19\linewidth]{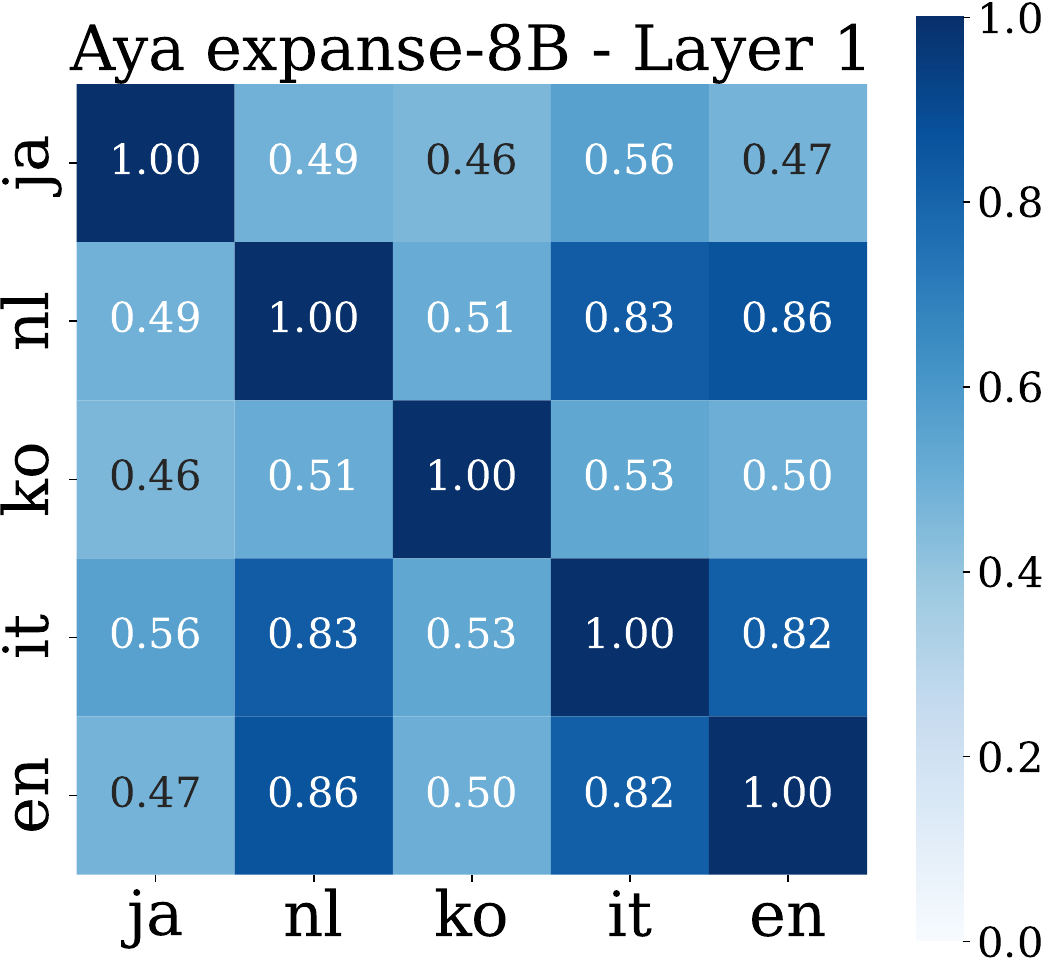}
  \includegraphics[width=0.19\linewidth]{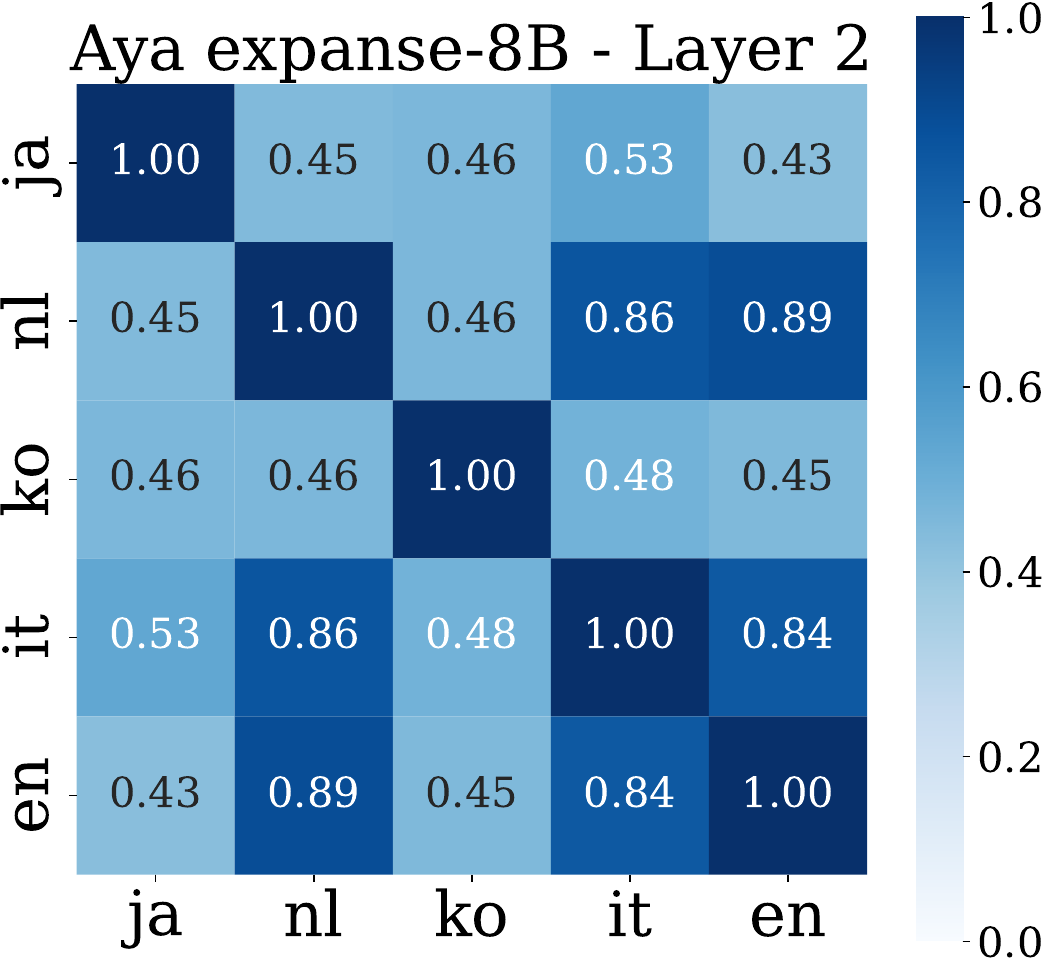}
  \includegraphics[width=0.19\linewidth]{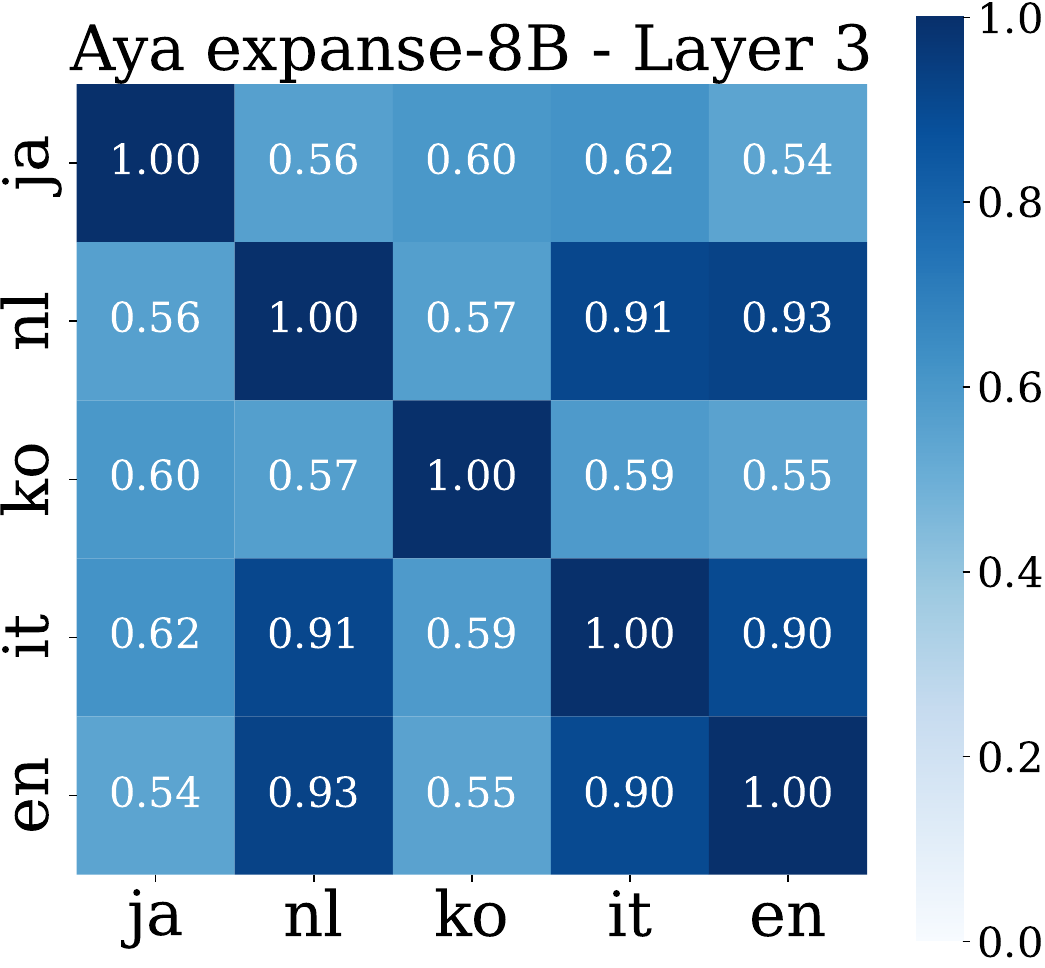}
  \includegraphics[width=0.19\linewidth]{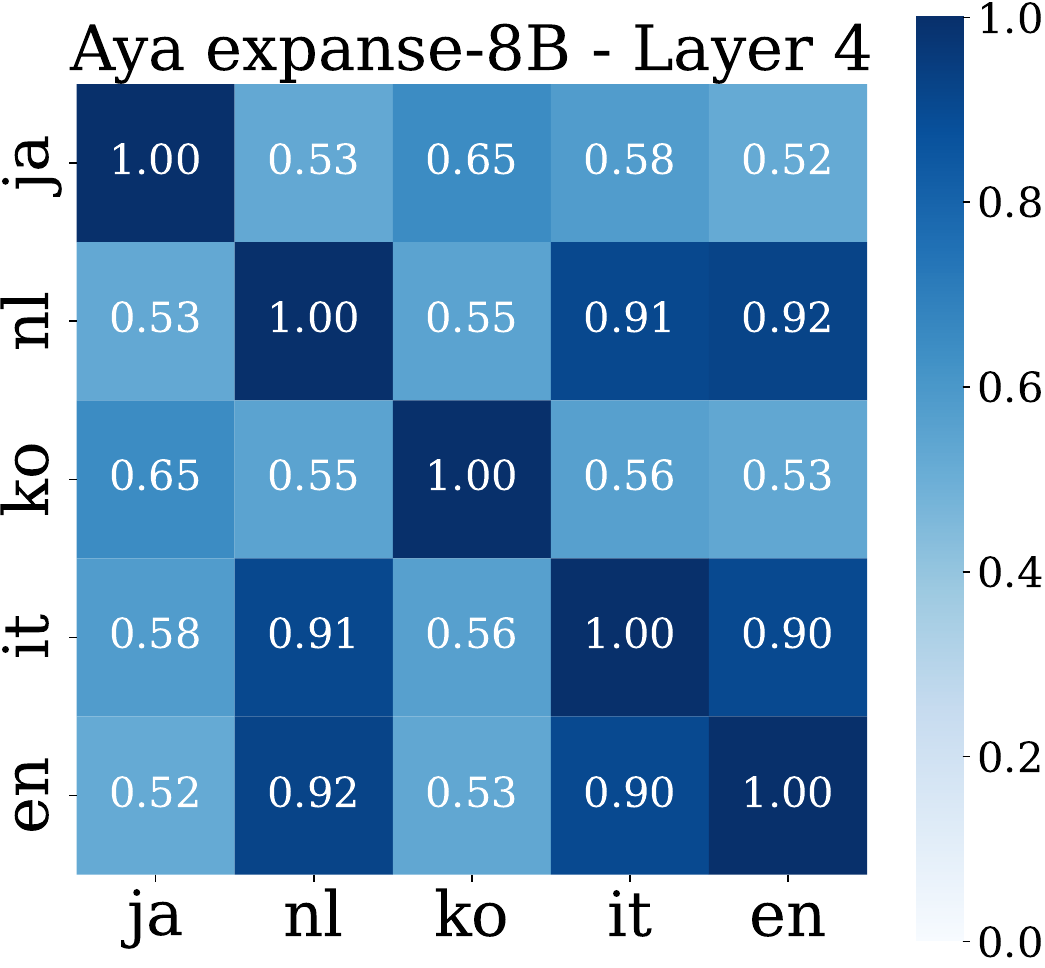}

  \includegraphics[width=0.19\linewidth]{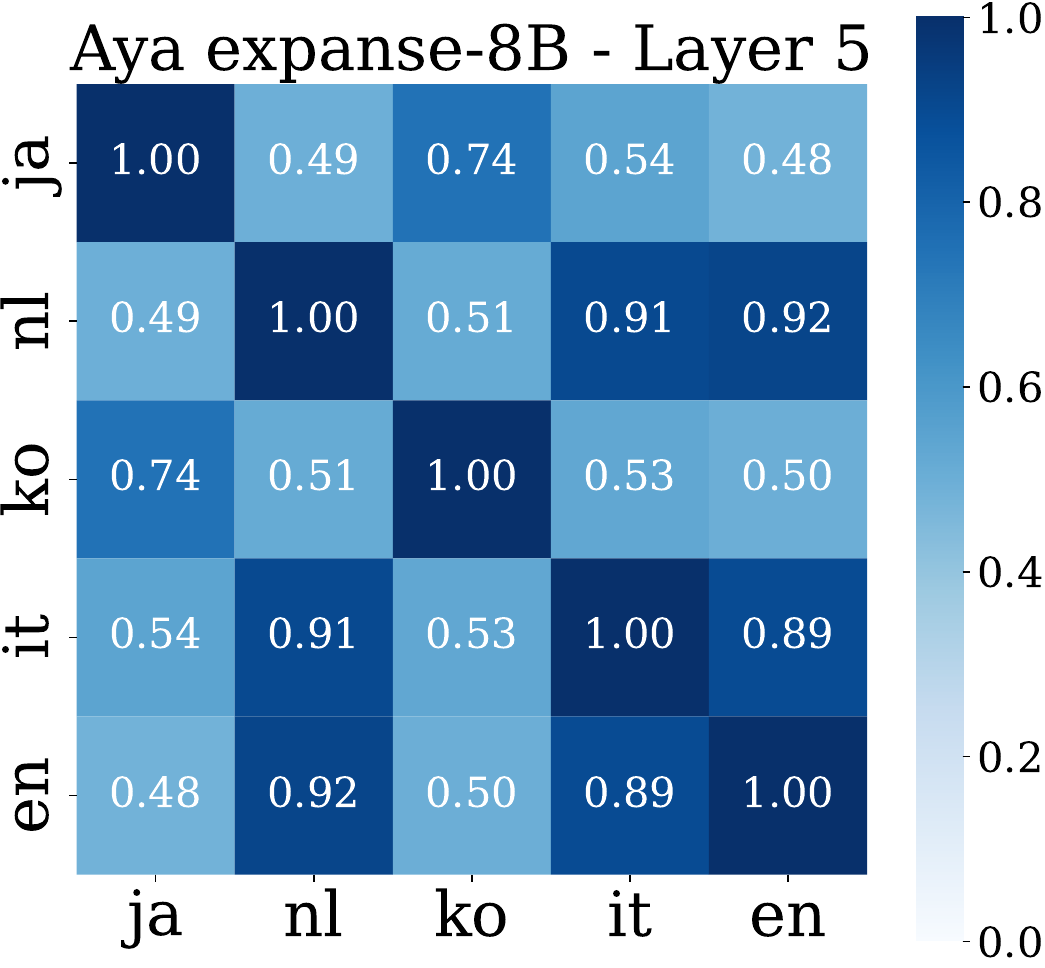}
  \includegraphics[width=0.19\linewidth]{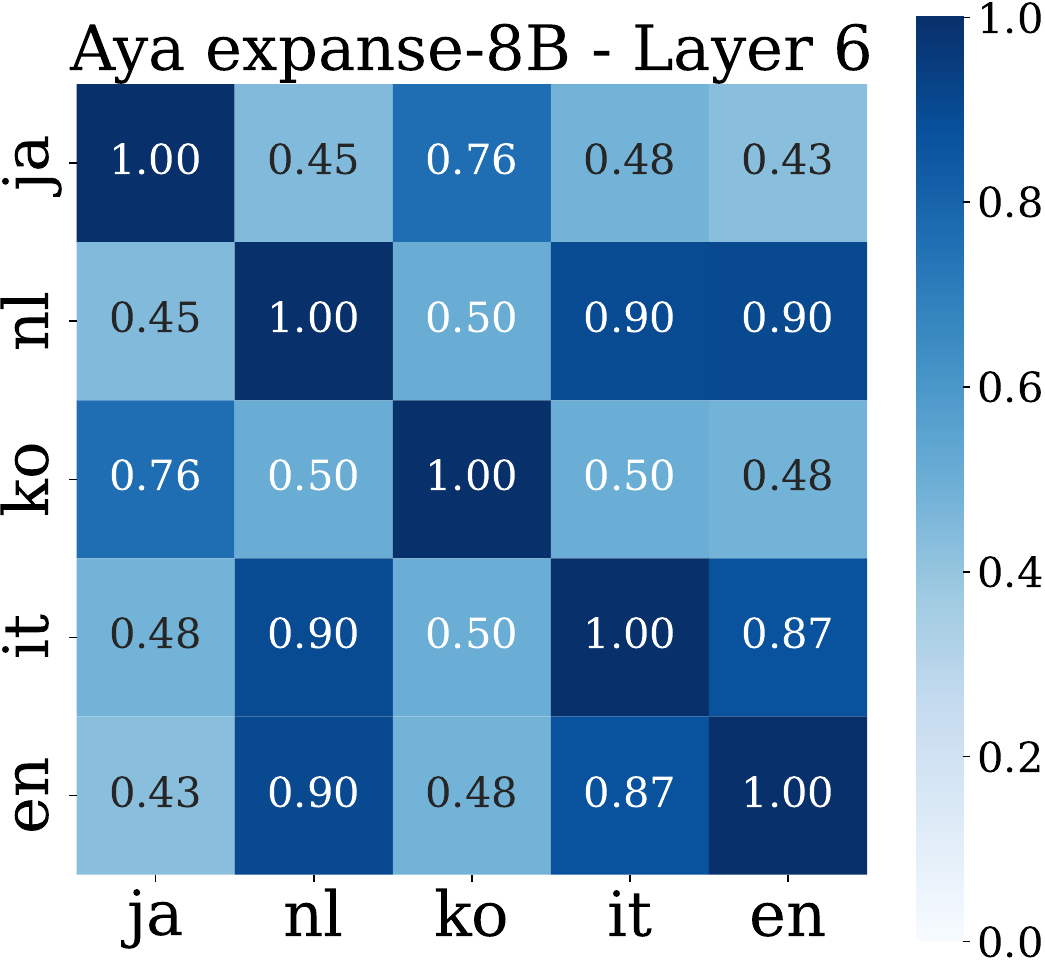}
  \includegraphics[width=0.19\linewidth]{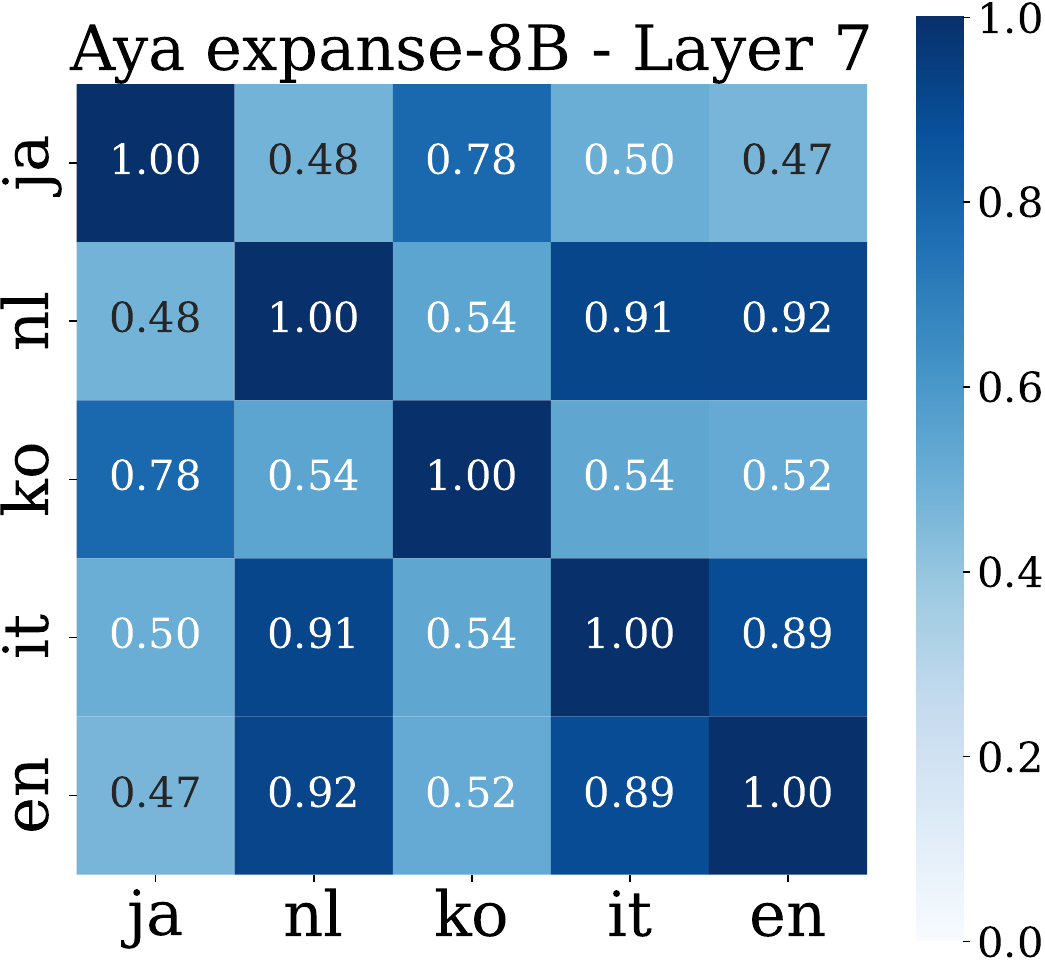}
  \includegraphics[width=0.19\linewidth]{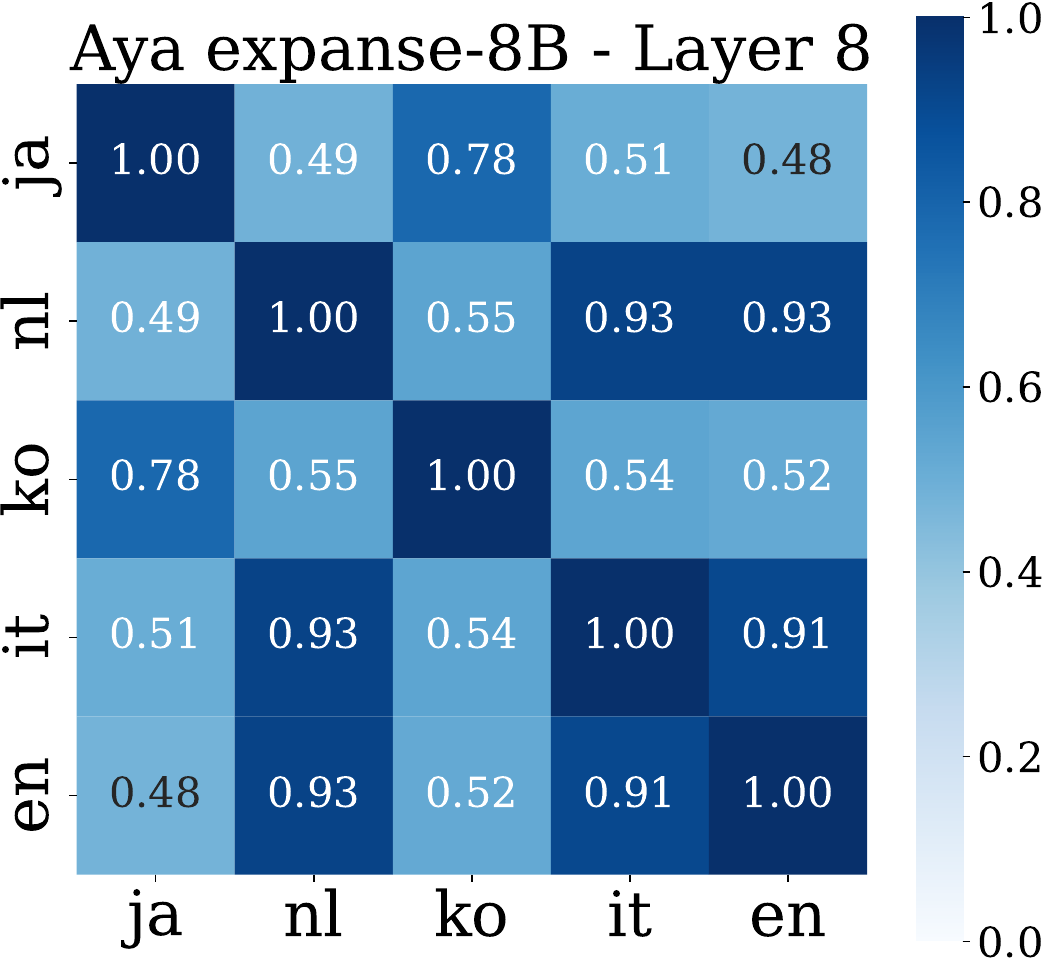}
  \includegraphics[width=0.19\linewidth]{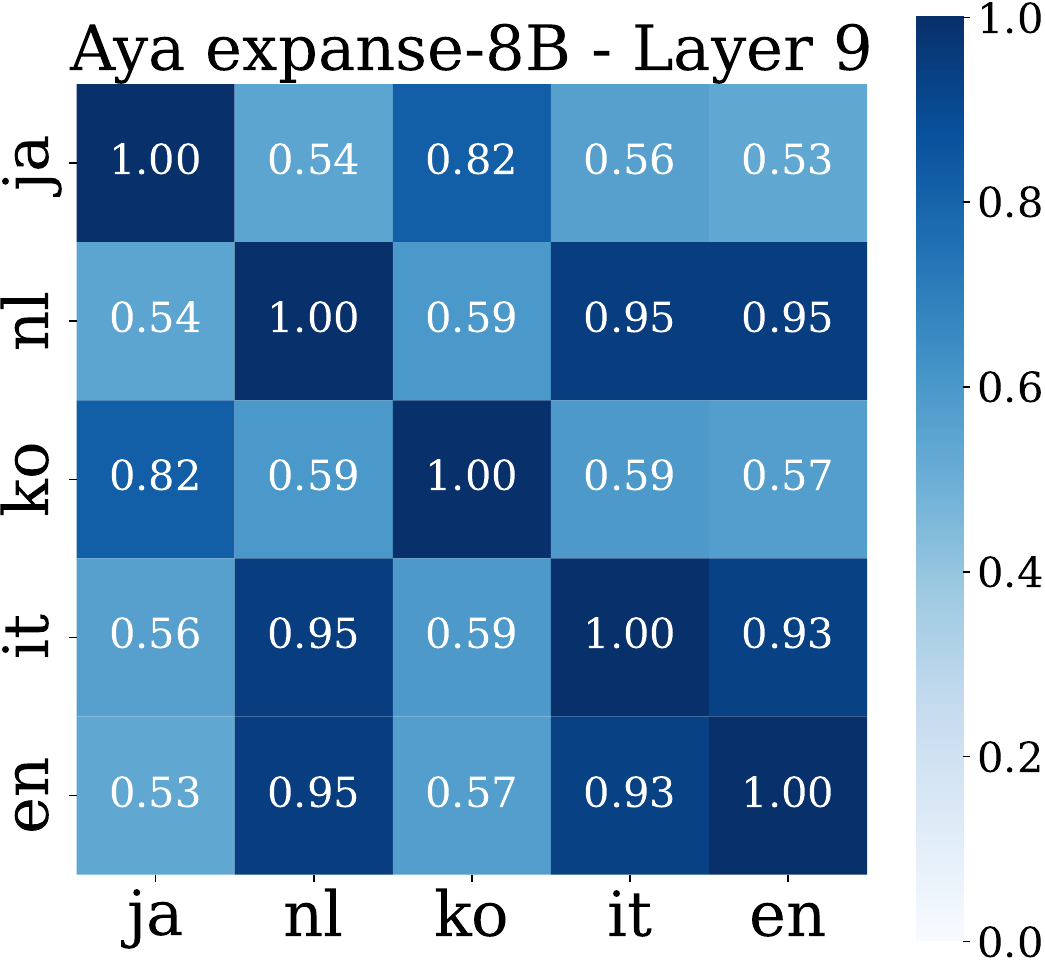}

  \includegraphics[width=0.19\linewidth]{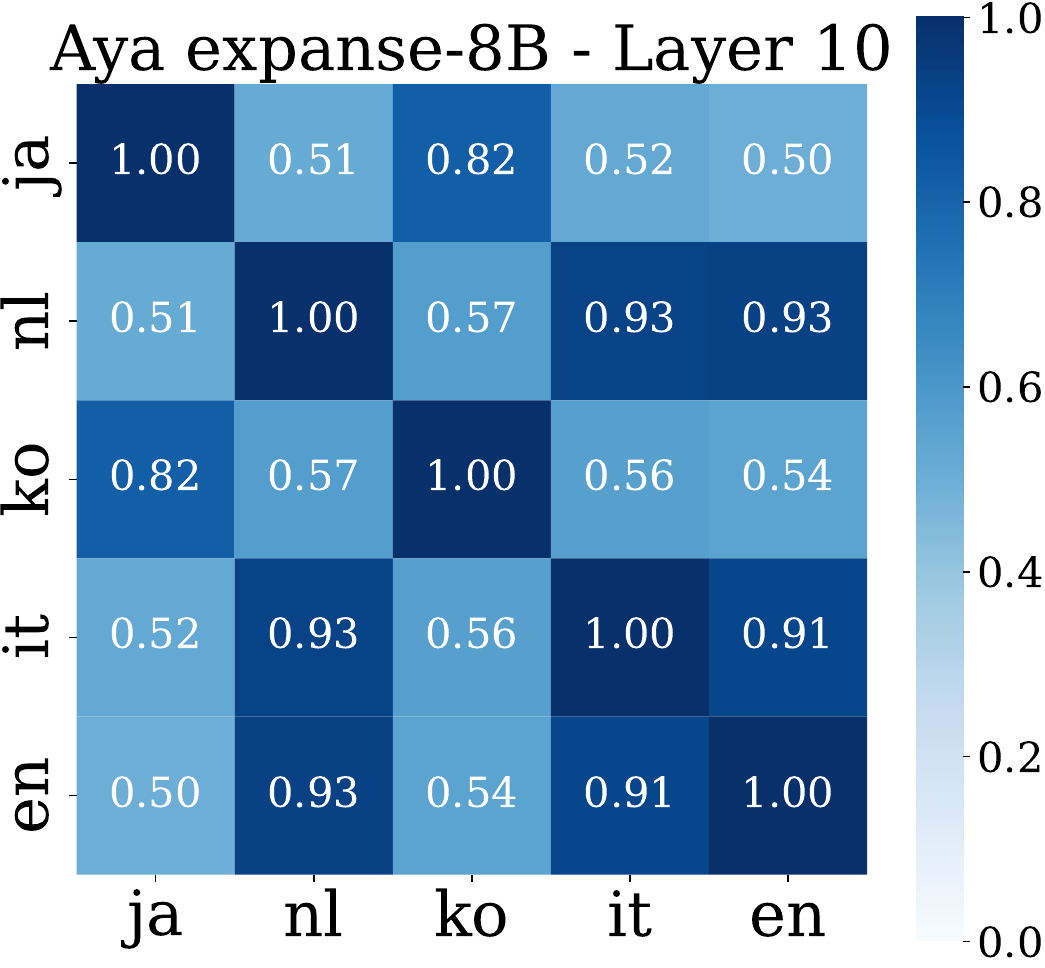}
  \includegraphics[width=0.19\linewidth]{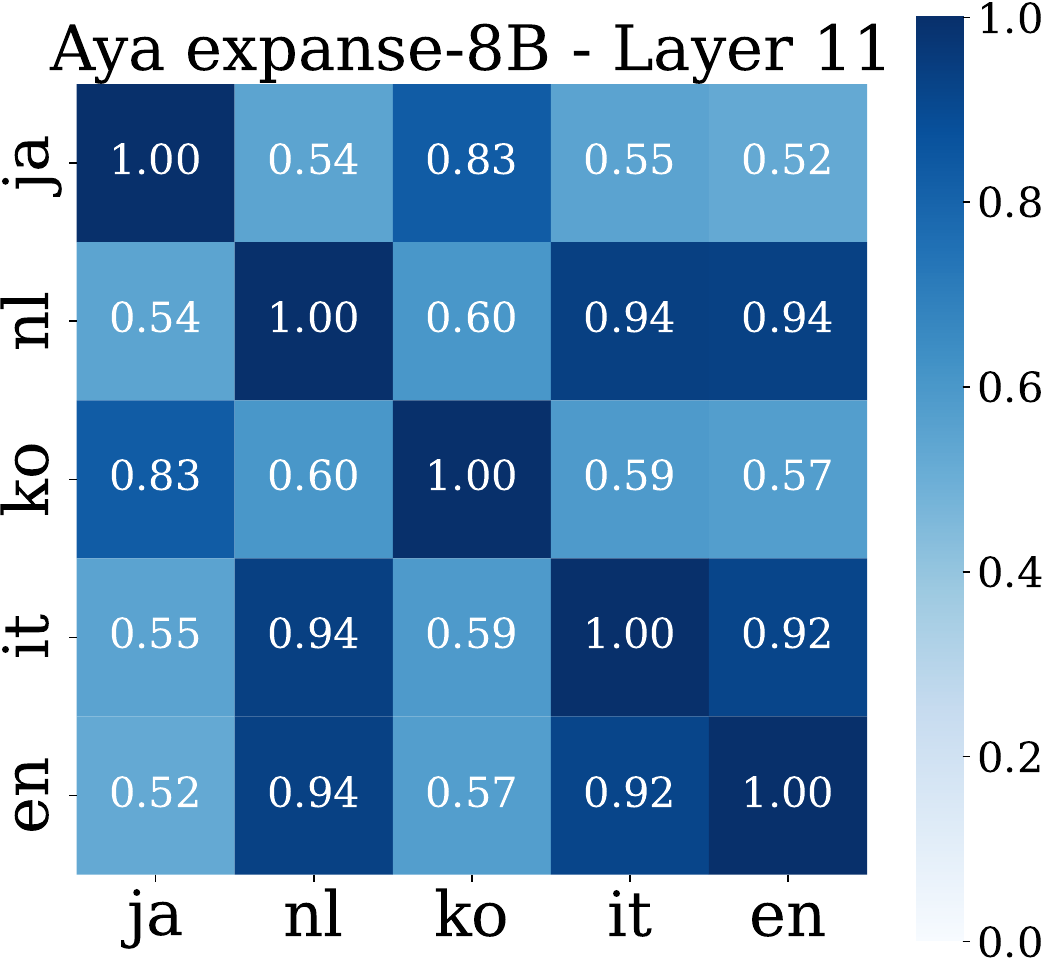}
  \includegraphics[width=0.19\linewidth]{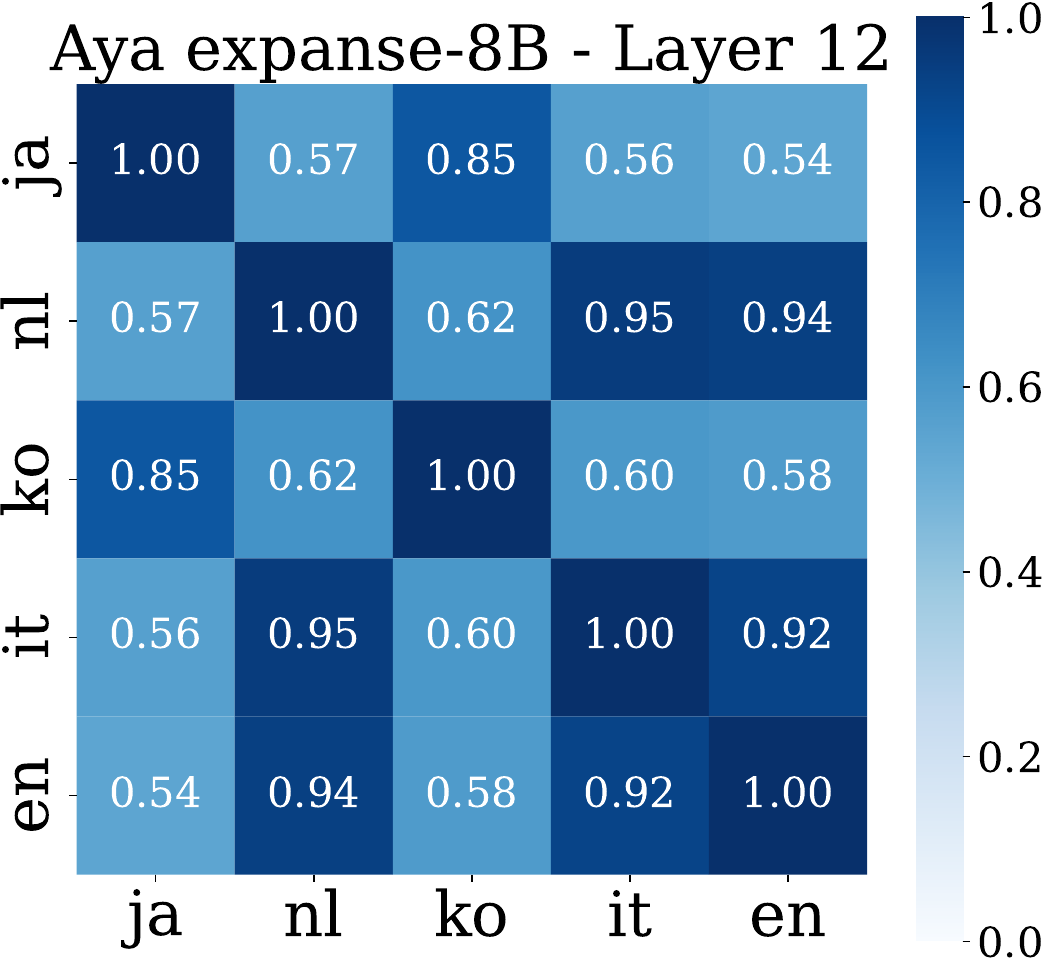}
  \includegraphics[width=0.19\linewidth]{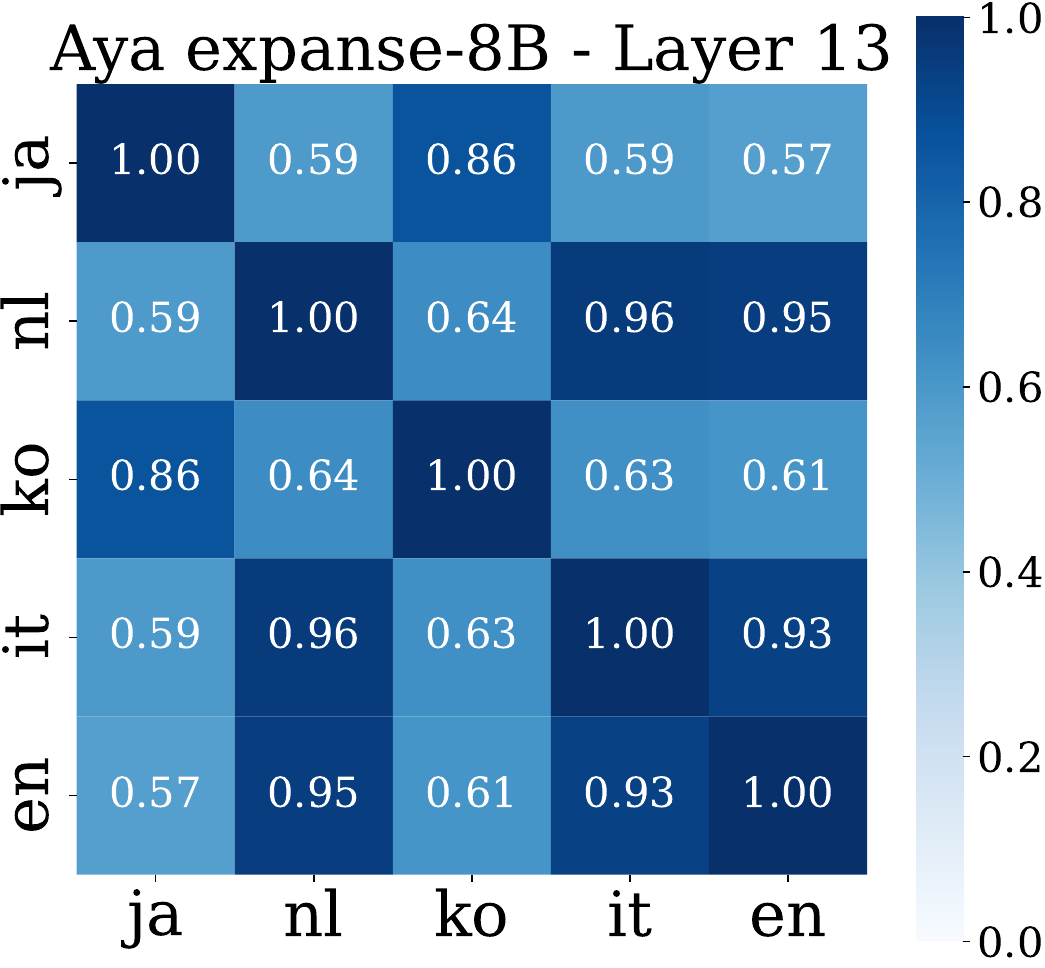}
  \includegraphics[width=0.19\linewidth]{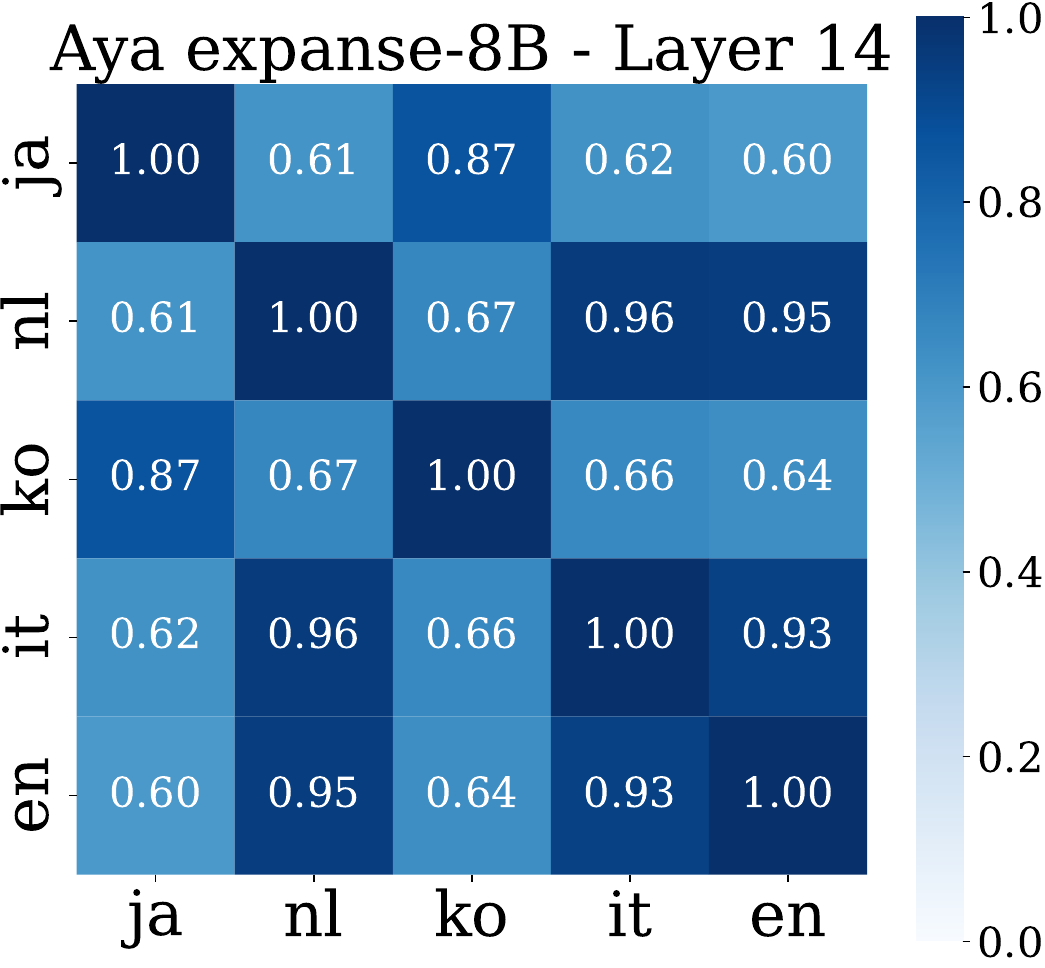}

  \includegraphics[width=0.19\linewidth]{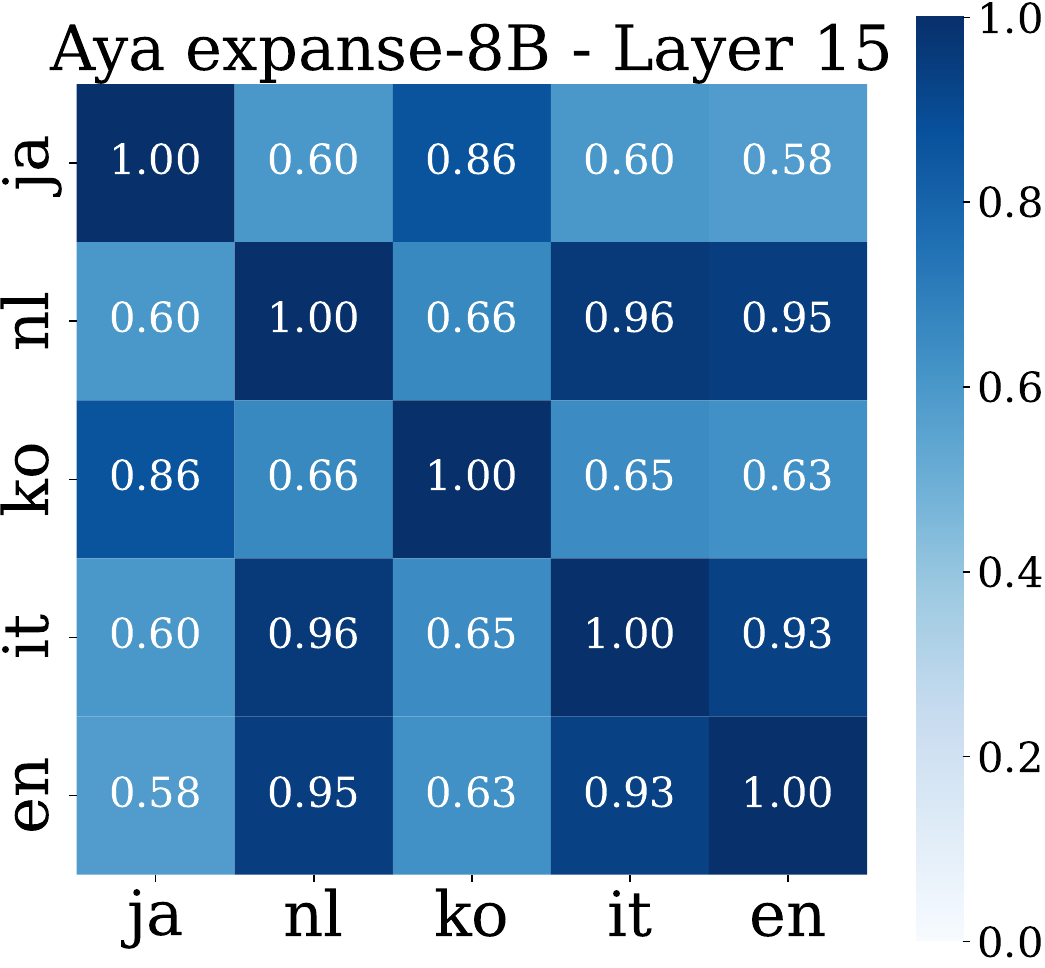}
  \includegraphics[width=0.19\linewidth]{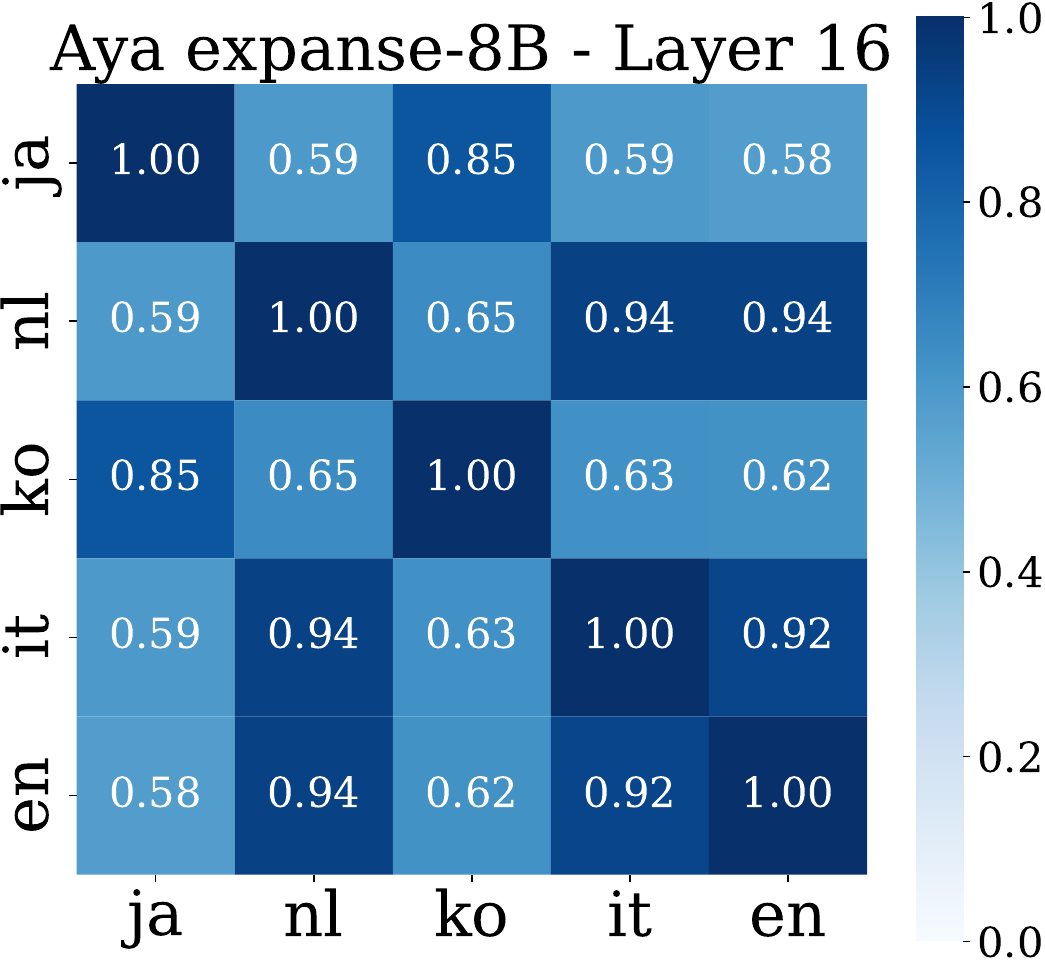}
  \includegraphics[width=0.19\linewidth]{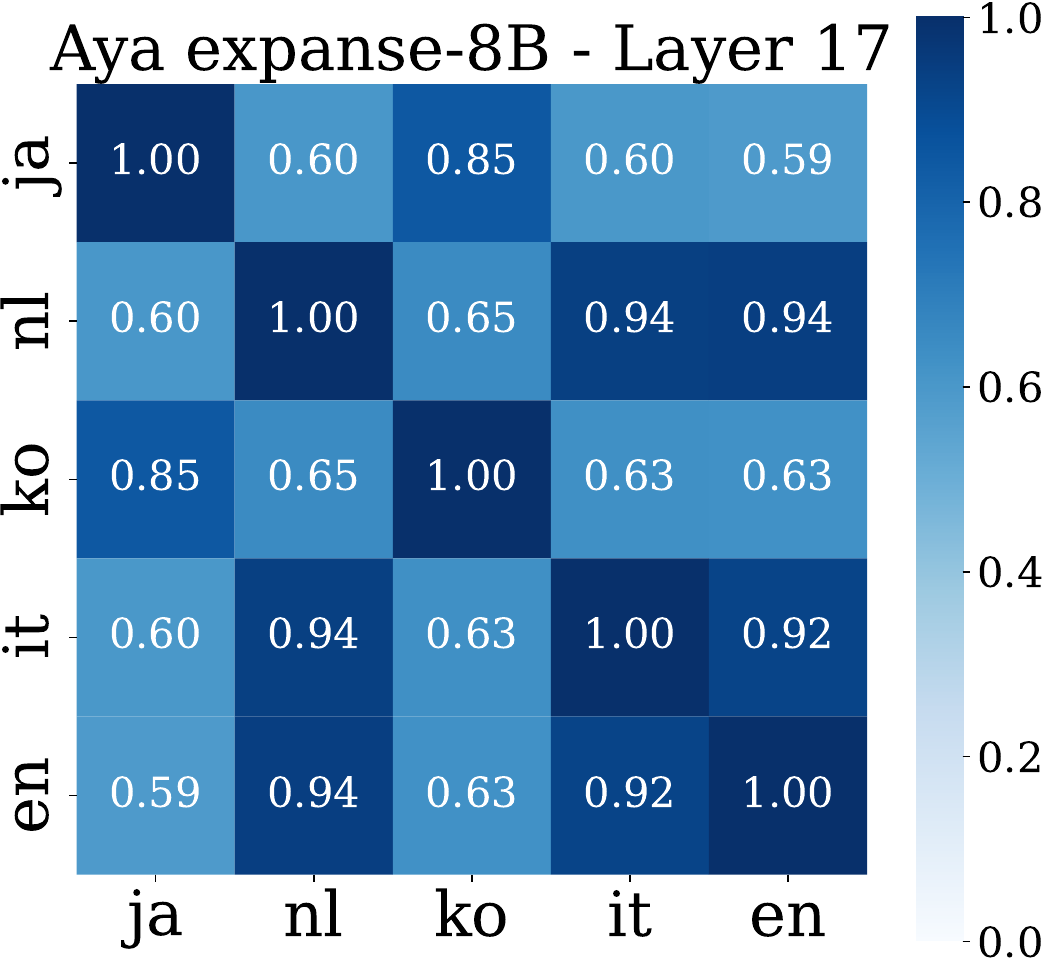}
  \includegraphics[width=0.19\linewidth]{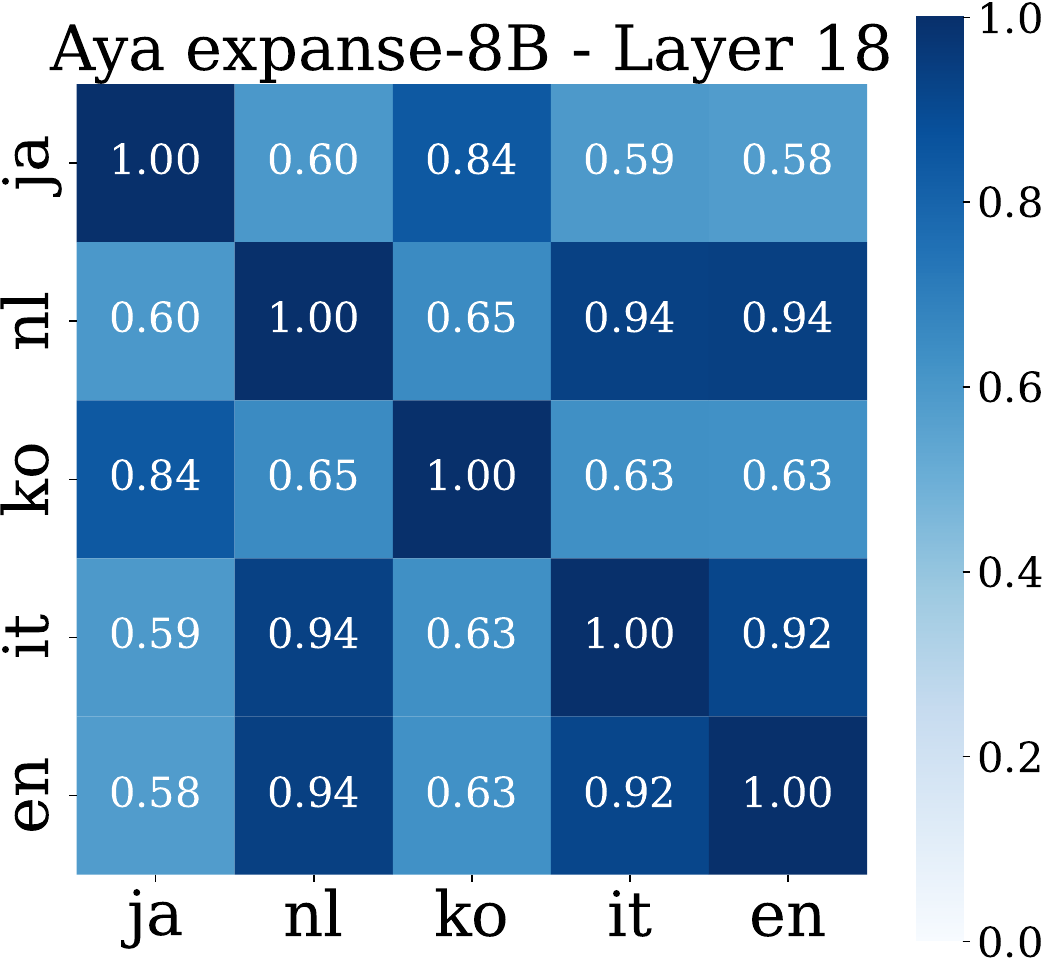}
  \includegraphics[width=0.19\linewidth]{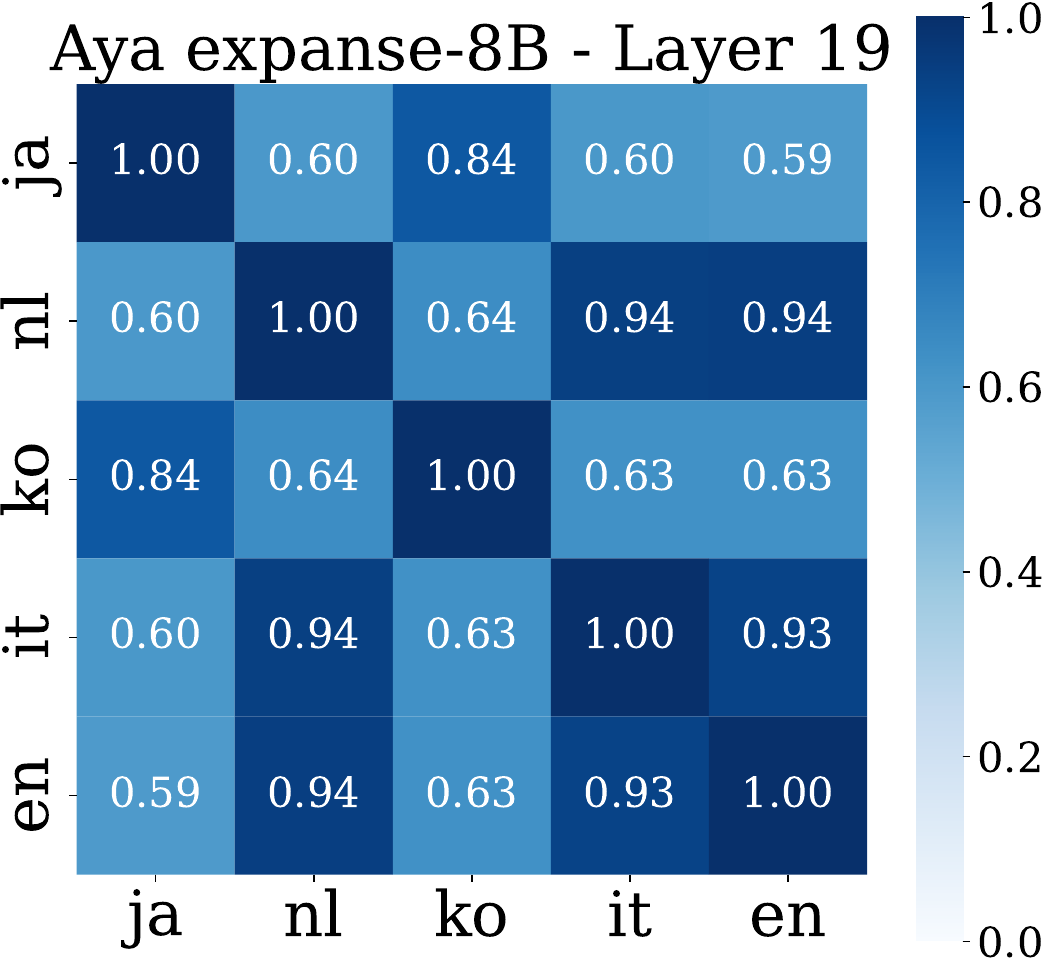}

  \includegraphics[width=0.19\linewidth]{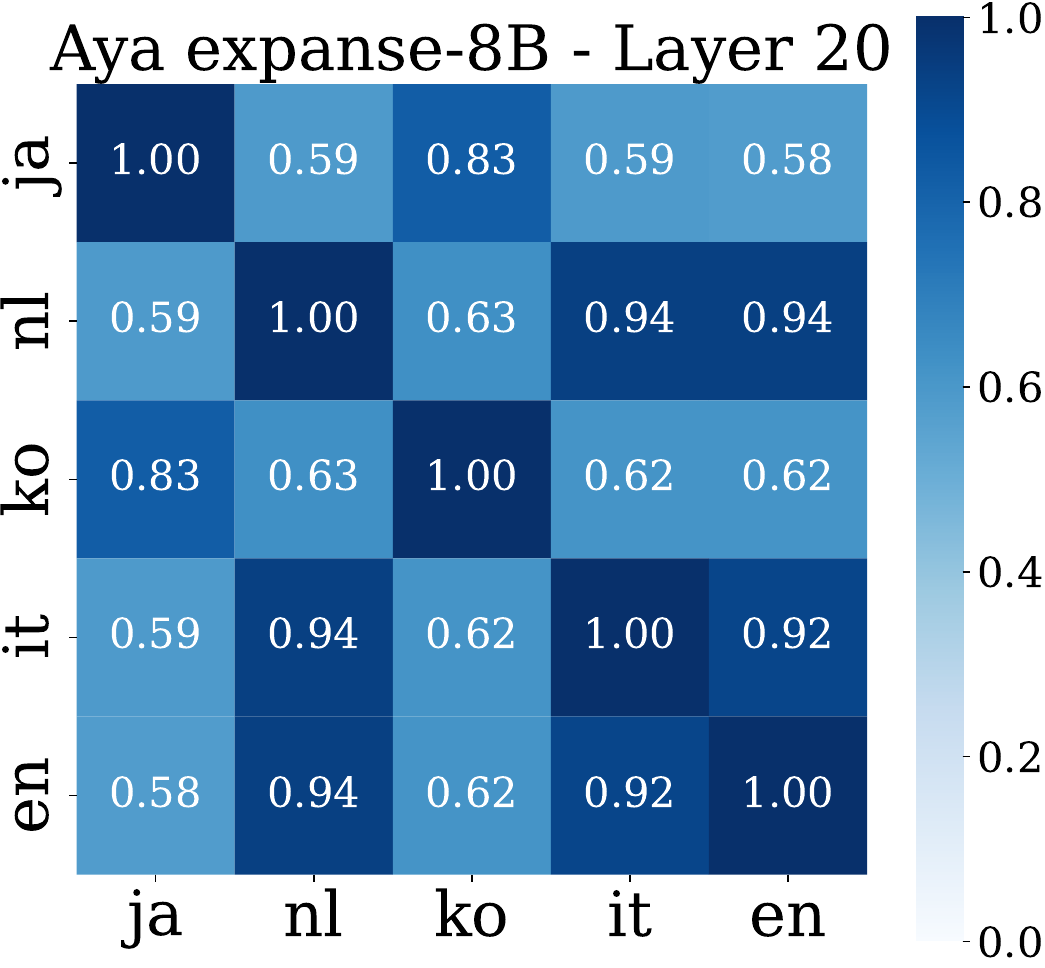}
  \includegraphics[width=0.19\linewidth]{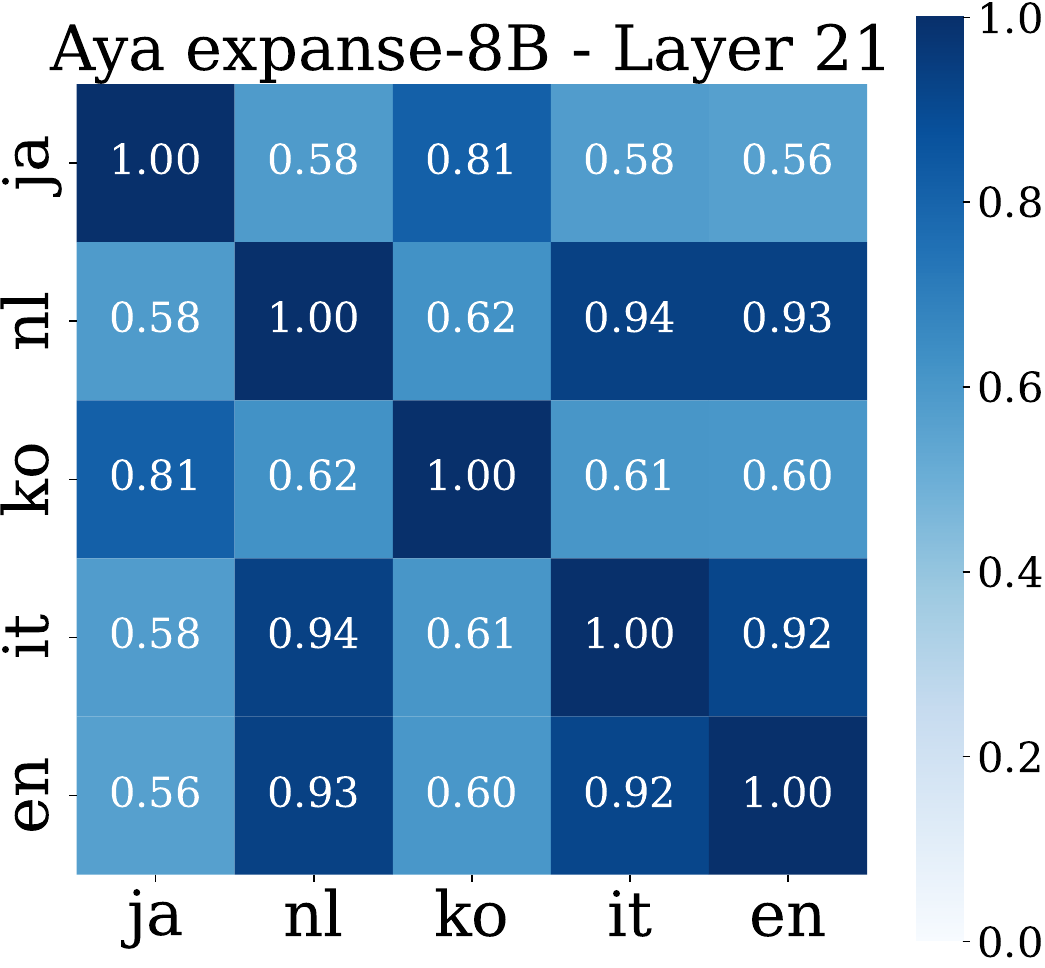}
  \includegraphics[width=0.19\linewidth]{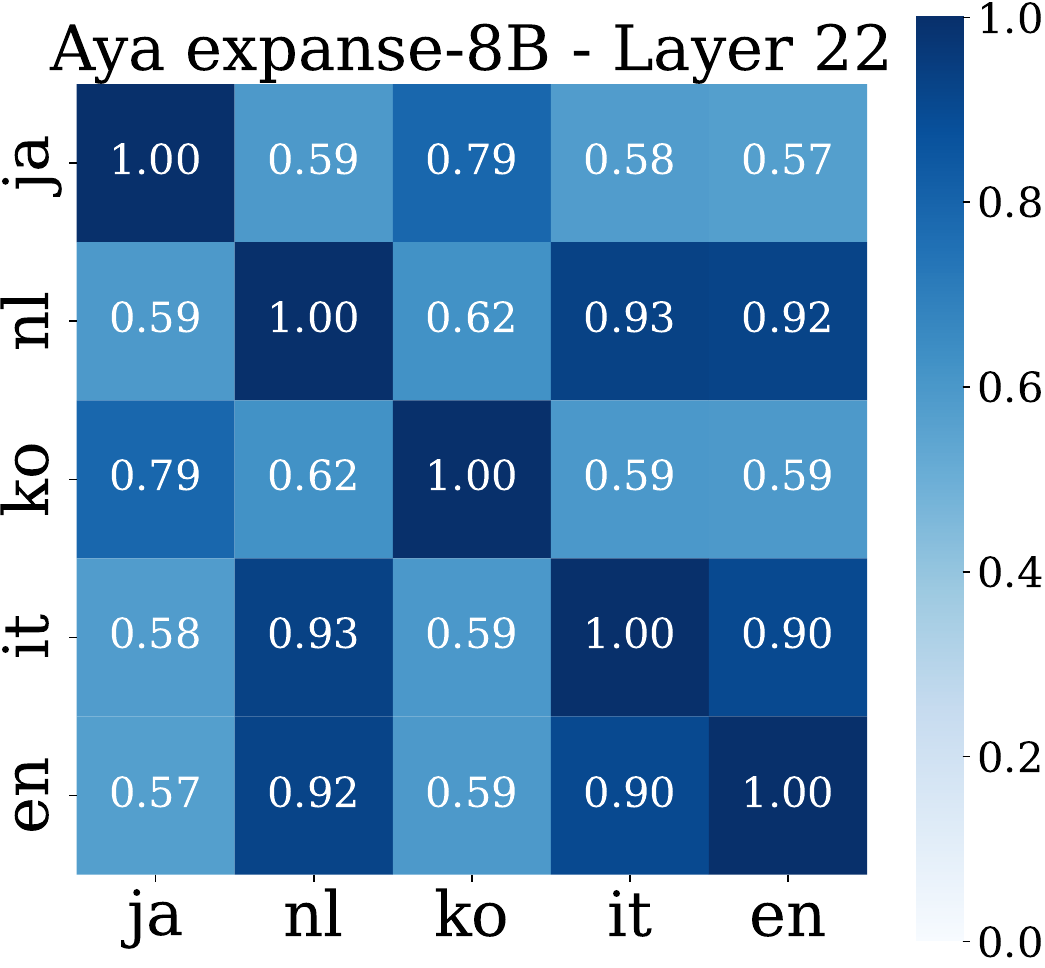}
  \includegraphics[width=0.19\linewidth]{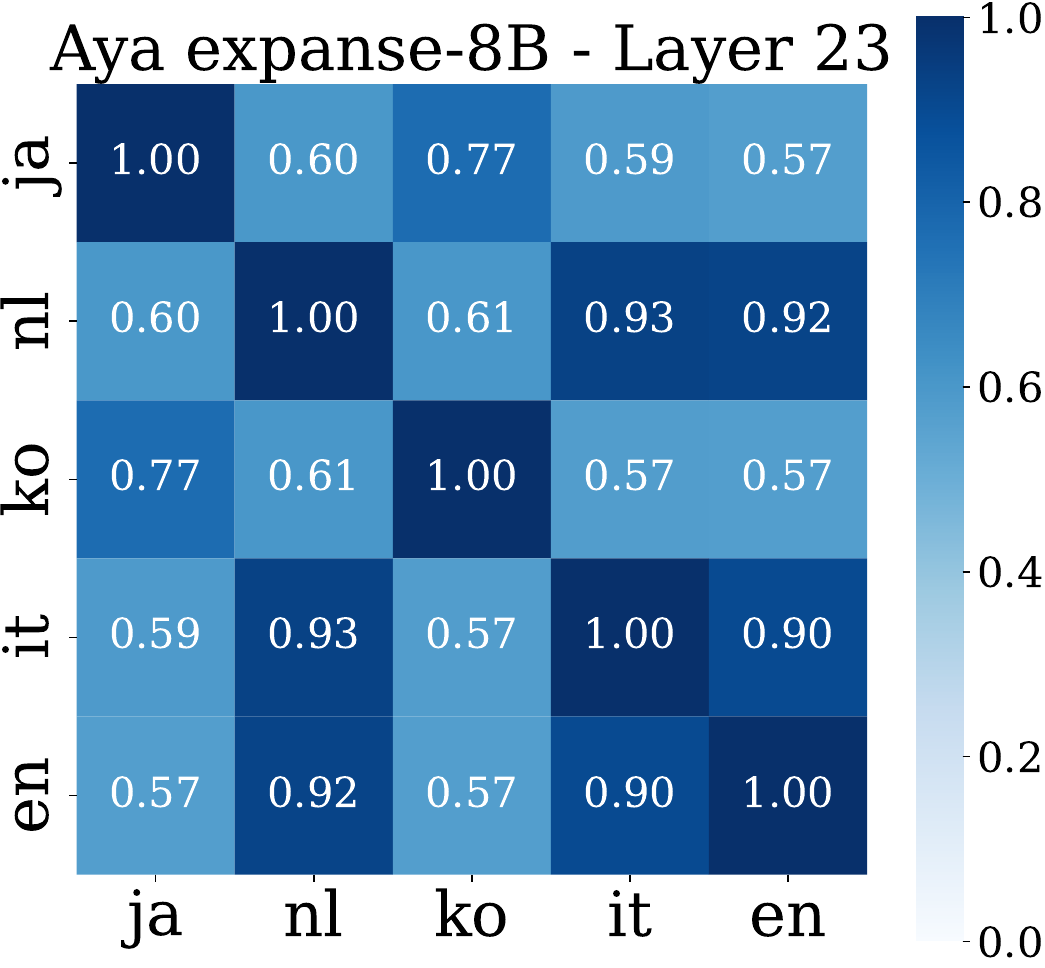}
  \includegraphics[width=0.19\linewidth]{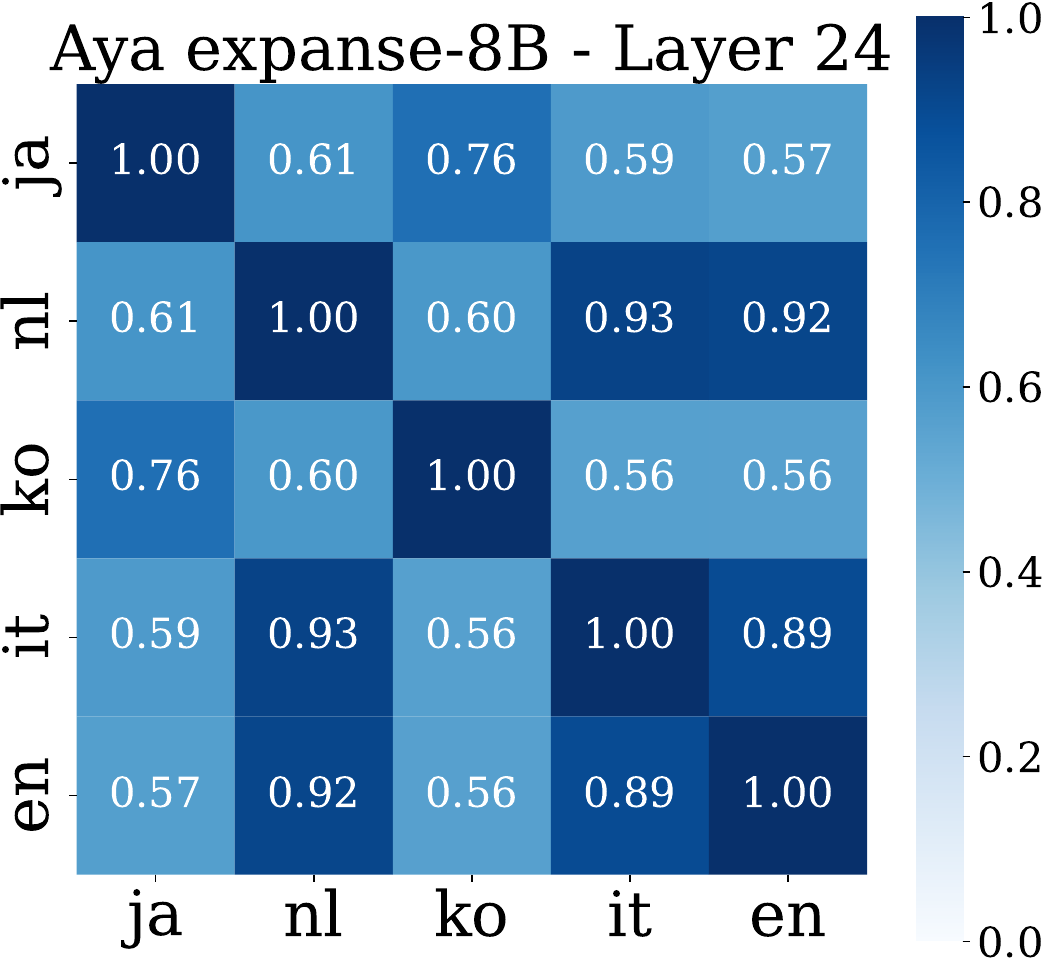}

  \includegraphics[width=0.19\linewidth]{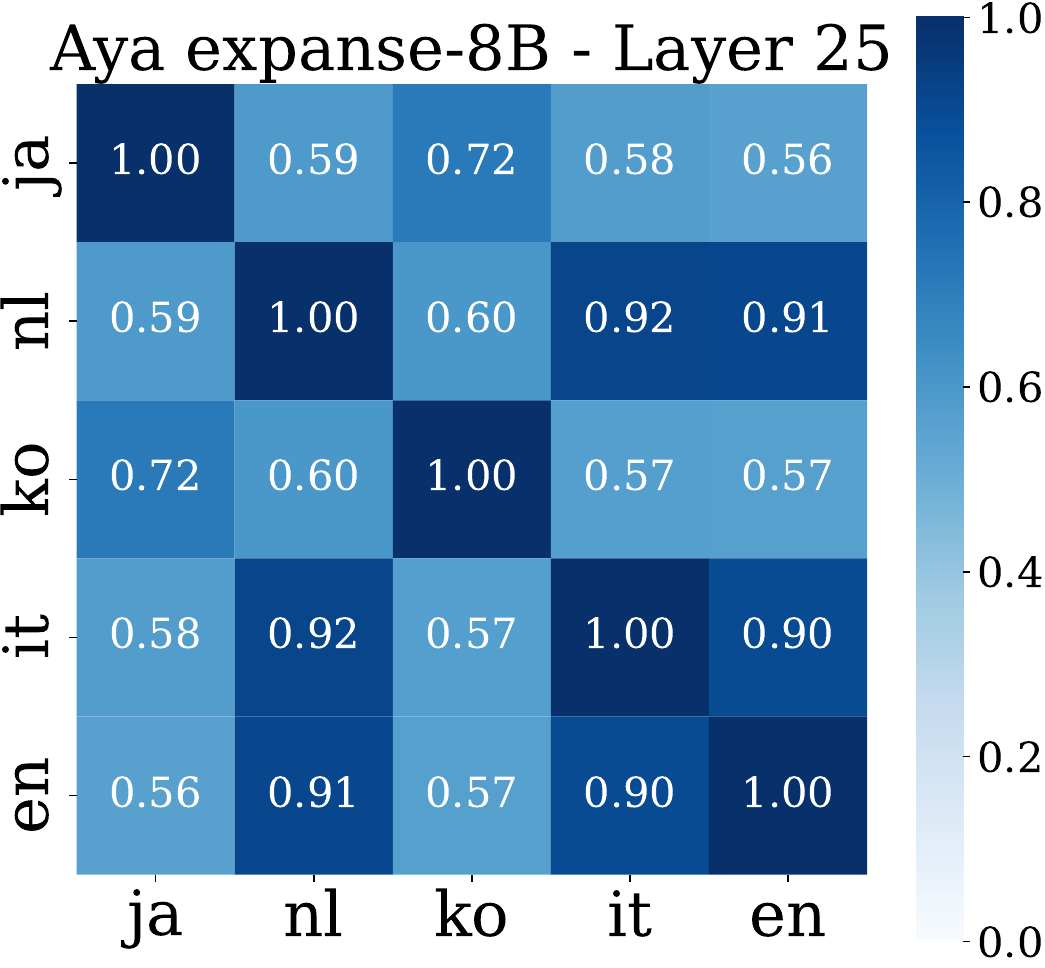}
  \includegraphics[width=0.19\linewidth]{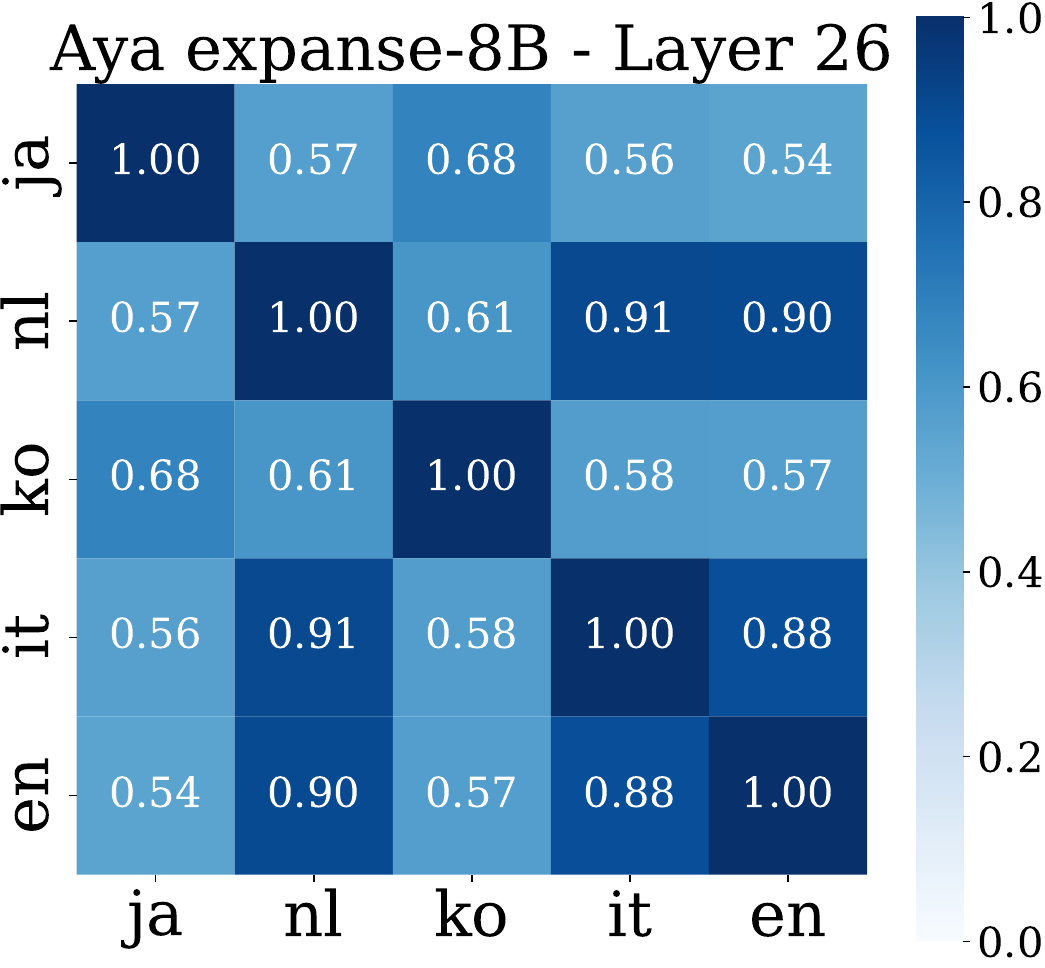}
  \includegraphics[width=0.19\linewidth]{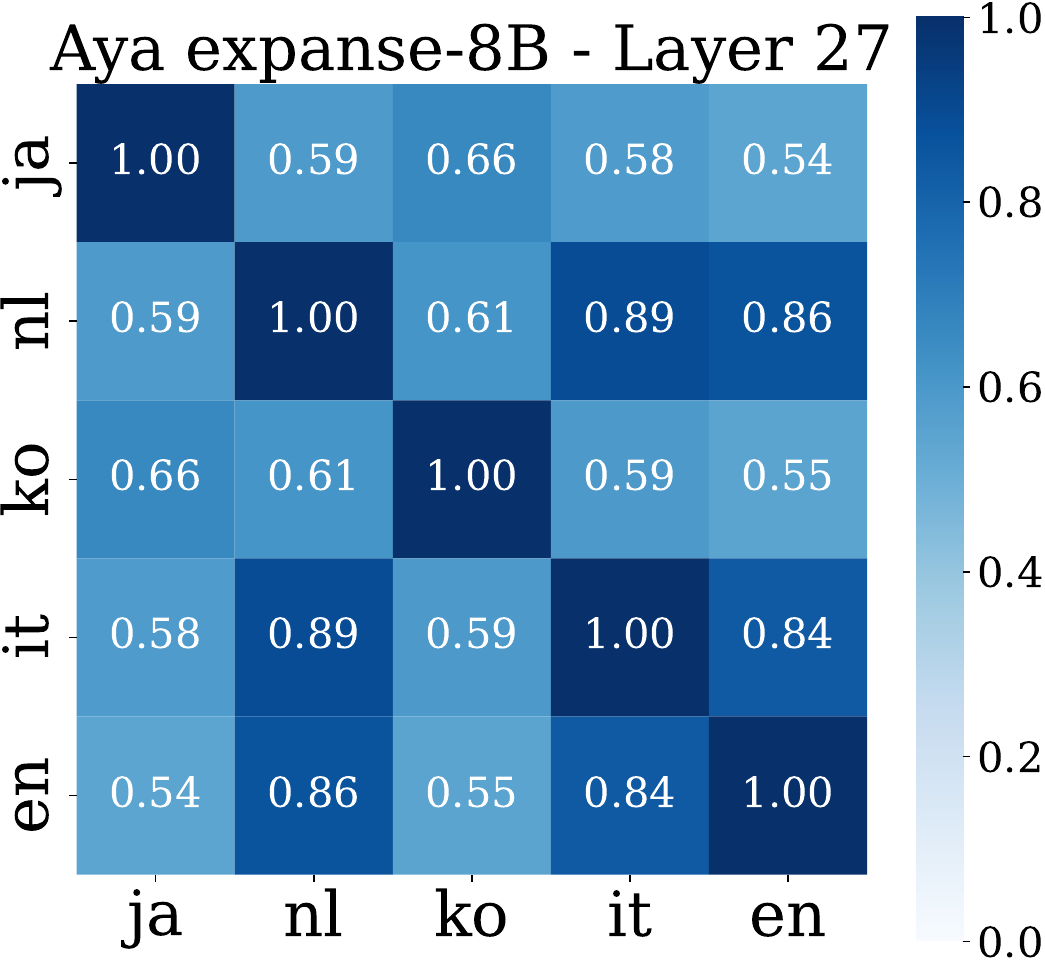}
  \includegraphics[width=0.19\linewidth]{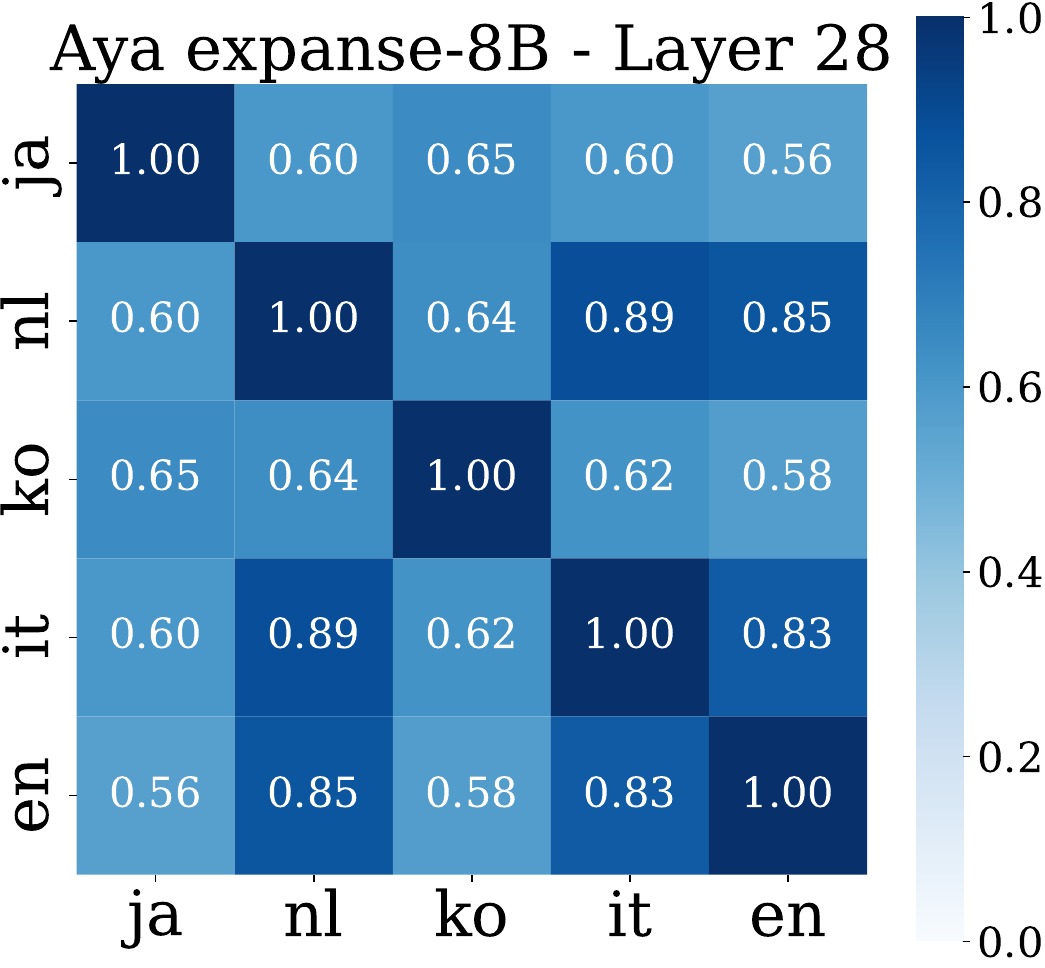}
  \includegraphics[width=0.19\linewidth]{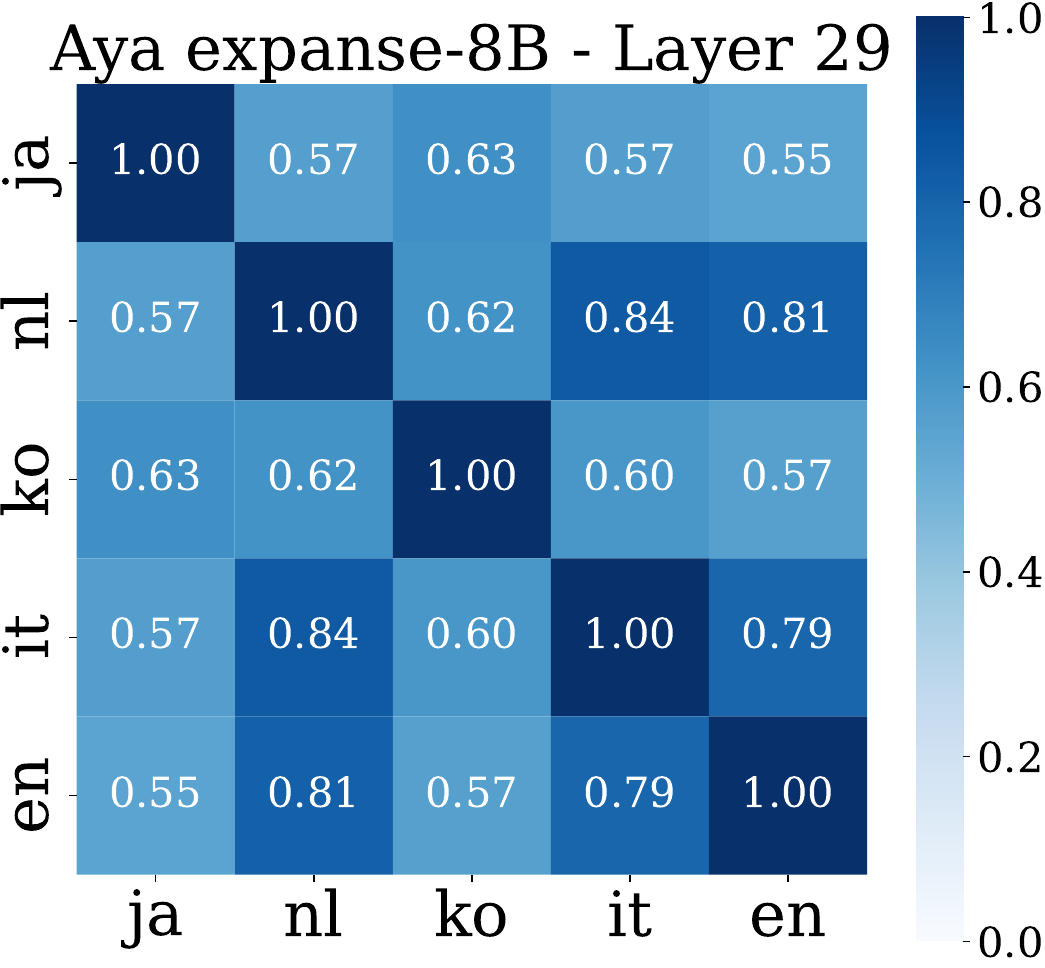}

  \includegraphics[width=0.19\linewidth]{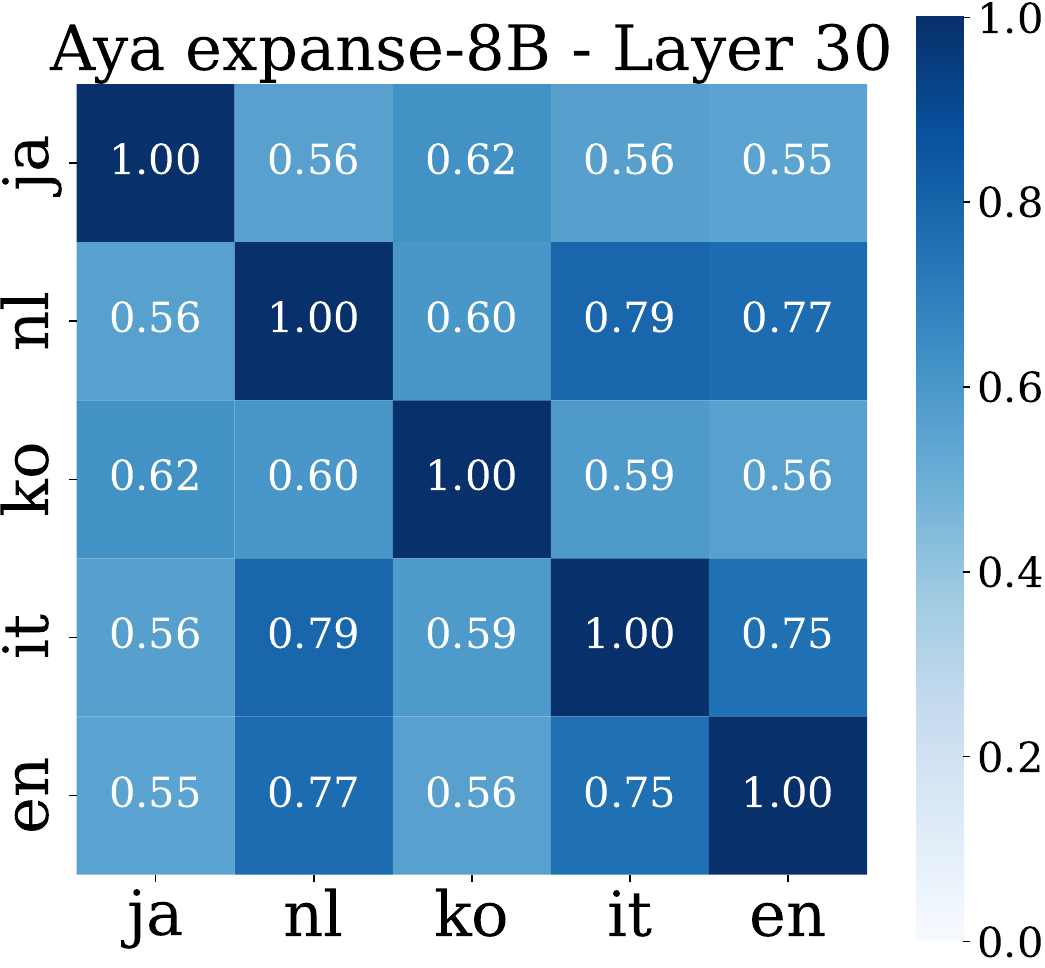}
  \includegraphics[width=0.19\linewidth]{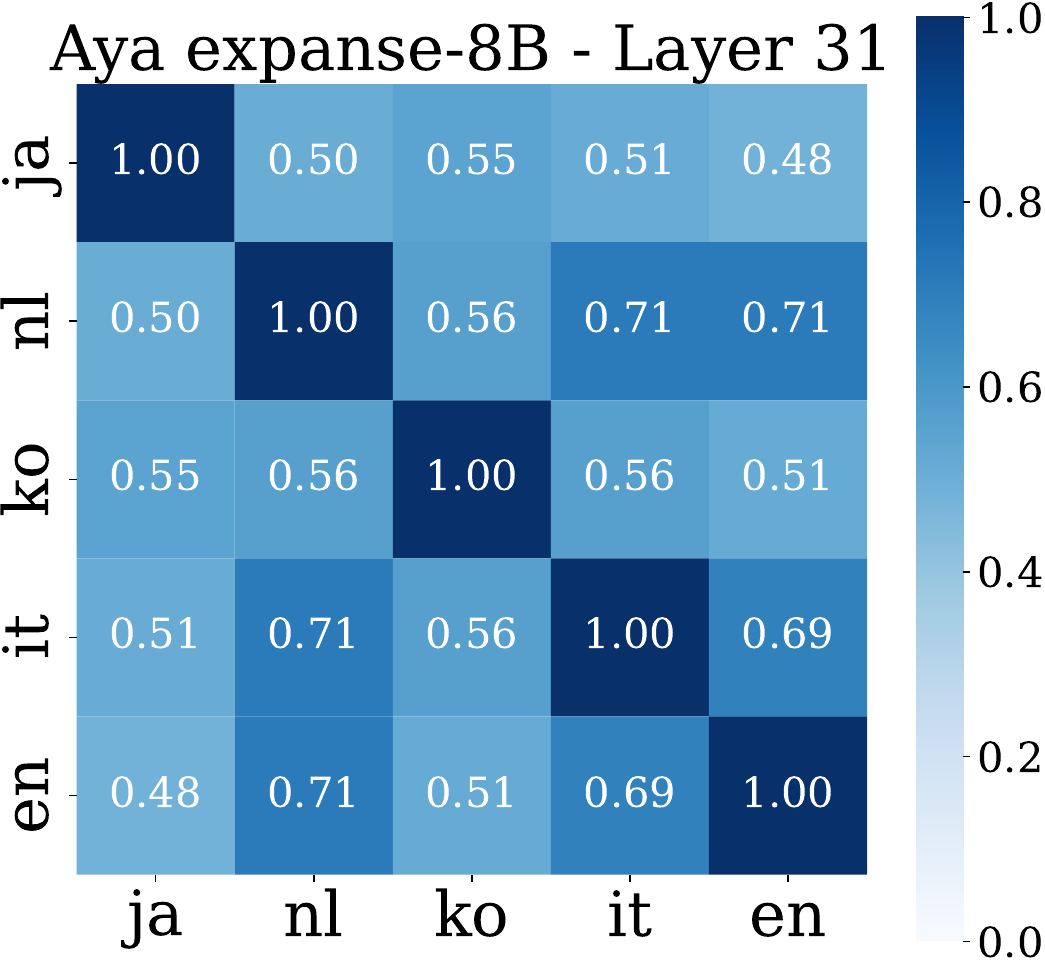}
  \includegraphics[width=0.19\linewidth]{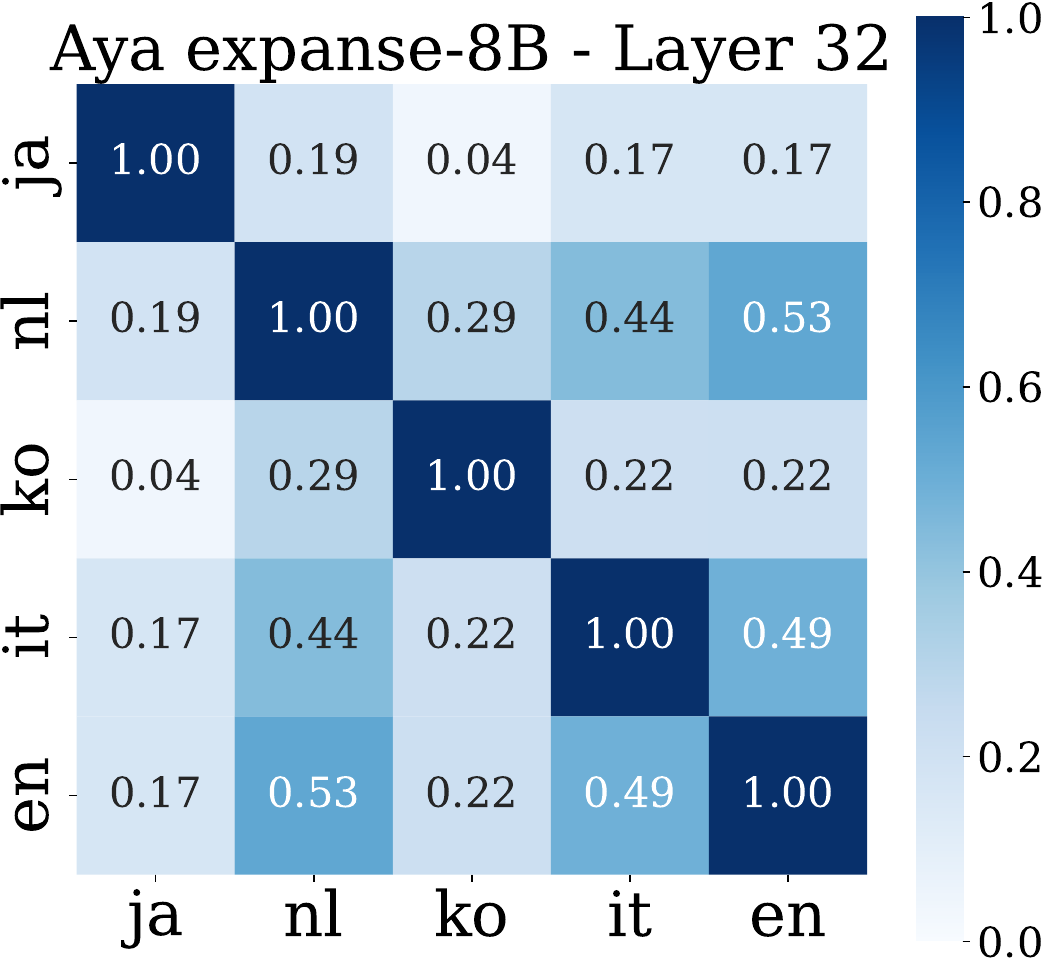}

  \caption{\textbf{The distance among centroids of language latent spaces (Aya expanse-8B).}}
  \label{fig:appendix:distance among subspaces centroids aya}
\end{figure*}

To quantitatively demonstrate the spacial transition phenomena and measure distance among language latent spaces, we compute $\mathrm{cos}(\bm{C}^l_{\mathrm{L_1}}, \bm{C}^l_{\mathrm{L_2}})$, where both $\bm{C}^l_{\mathrm{L_1}}$ and $\bm{C}^l_{\mathrm{L_2}}$ are the centroid of each latent space computed by Eq.~\ref{eq:centroids_lang_specific_subspace}.

Figs.~\ref{fig:appendix:distance among subspaces centroids llama3},~\ref{fig:appendix:distance among subspaces centoids mistral}, and~\ref{fig:appendix:distance among subspaces centroids aya} show the results. As indicated, these results are well align with the hypothesis shown in Fig.~\ref{fig:figure1} (a): in the initial and final few layers, language latent spaces remain relatively distant from each other, while they become closer in the middle layers where language-agnostic semantic processing and reasoning are mainly occurred.

% Appendix: Mutual KNN.
\section{Mutual \textit{k}-Nearest Neighbor Alignment Metric}
\label{sec:appendix:mutual knn alignment metric}

% B.2: formalization of the computation of mutual knn.
\subsection{Formalization of the Computation for \textit{Mutual k-NN Alignment Metric} across Layers and Input Language Pairs}
\label{sec:appendix:formalizing mutual_knn}
This metric was originally for computing the similarity between representations formed by semantically equivalent inputs from different modalities, each residing in separate model spaces. In our setting, however, we adapt it to compare hidden latent spaces formed by different input languages within a single model, allowing us to assess kernel-based layer-wise similarity between language-specific latent spaces (English–L2). Building on \citet{icml2024platonic}, we adjust the computation as follows:

Let \( f \) be a single model (LLM), and let \(x_i\) and \(y_i\) denote an English sentence and its corresponding sentence in another language (L2), respectively, both expressing the same meaning.
% m_knn
\begin{equation}
    \phi_i^l = f(x_i^l)
    \label{eq:phi}
\end{equation}
\begin{equation}
    \psi_i^l = f(y_i^l)
    \label{eq:psi}
\end{equation}
where \(\phi_i^l\), \(\psi_i^l\) are hidden states in \(l\)-th layer. Collection of these representations are denoted as:
\begin{equation}
    \Phi^l = \{ \phi_1^l,...,\phi_b^l \}
    \label{}
\end{equation}
\begin{equation}
    \Psi^l = \{ \psi_1^l, ...., \psi_b^l \}
\end{equation}
Then for each representations pair $\{\phi_i^l, \psi_i^l\}$, we compute the respective nearest neighbor sets $\mathcal{S}(\phi_i^l)$ and $\mathcal{S}(\psi_i^l)$:
\begin{equation}
    d_\mathrm{knn}^l(\phi_i^l, \Phi^l \backslash \phi_i^l) = \mathcal{S}(\phi_i^l)
\end{equation}
\begin{equation}
    d_\mathrm{knn}^l(\psi_i^l, \Psi^l \backslash \psi_i^l ) = \mathcal{S}(\psi_i^l)
\end{equation}
where $d_\mathrm{knn}^l$ returns the set of indices of its \textit{k}-nearest neighbors. Finally, we measure its average intersection via:
\begin{equation}
    m_\mathrm{nn}^l(\phi_i^l, \psi_i^l) = \frac{1}k{| \mathcal{S}(\phi_i^l) \cap \mathcal{S}(\psi_i^l)} |
\end{equation}
where $|\cdot|$ is the size of intersection.

% B.1: results of Mistral-7B
\subsection{Result of Mistral-7B}
\label{sec:appendixB.1}
The left column of Figs.~\ref{fig:appendix:kernel-based sim while deactivating Type-1 k=5} and~\ref{fig:appendix:kernel-based sim while deactivating Type-1 k=10} show the result of computing the mutual \textit{k}-NN alignment metric across layers of the models including Mistra-7B.
As shown, we obtain results of Mistral-7B similar to those of LLaMA3-8B and Aya-expanse-8B (Fig.~\ref{fig:mutual_knn}). Additionally, inputs from languages that belong to language families relatively close to English (\textcolor{orange}{Dutch}, \textcolor{red}{Italian}) tend to form more similar latent spaces compared to those that do not.

% Appendix: Visualization of Language Subspaces.
\section{Visualization of Language Latent Spaces}
\label{sec:appendix:visualization language subspaces}
\begin{figure*}[t]
  \centering

  \includegraphics[width=0.19\linewidth]{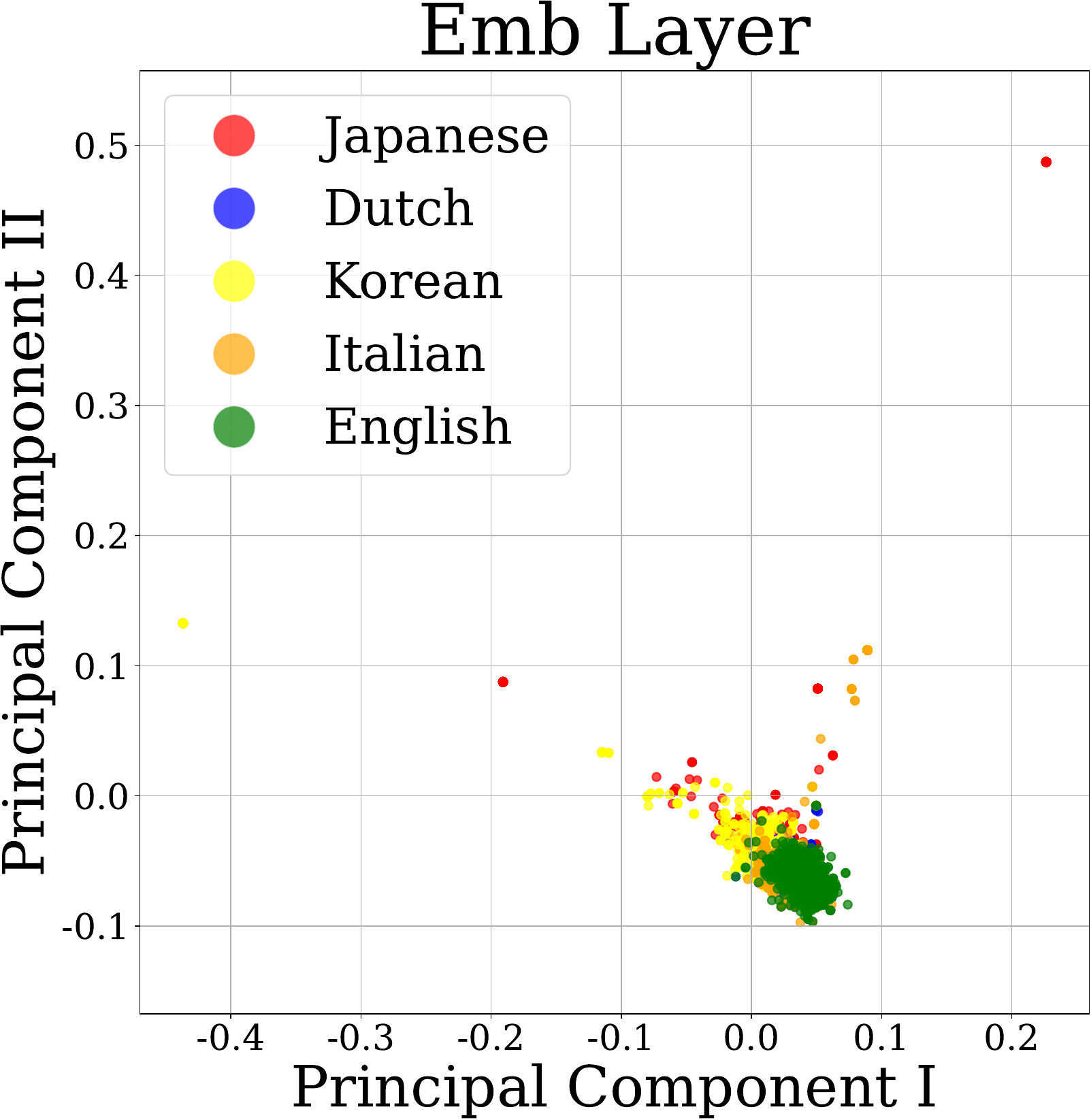}
  \includegraphics[width=0.19\linewidth]{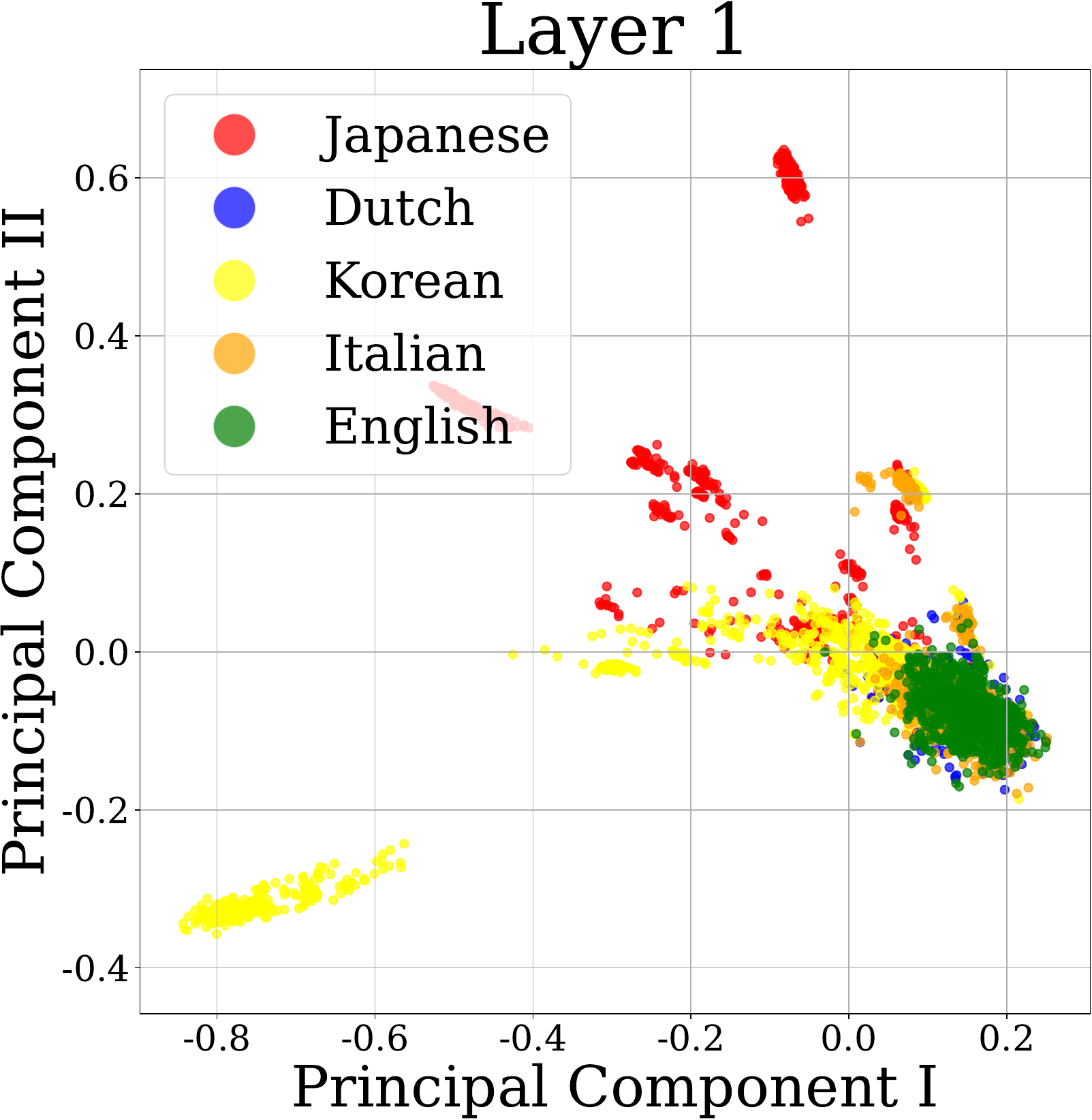}
  \includegraphics[width=0.19\linewidth]{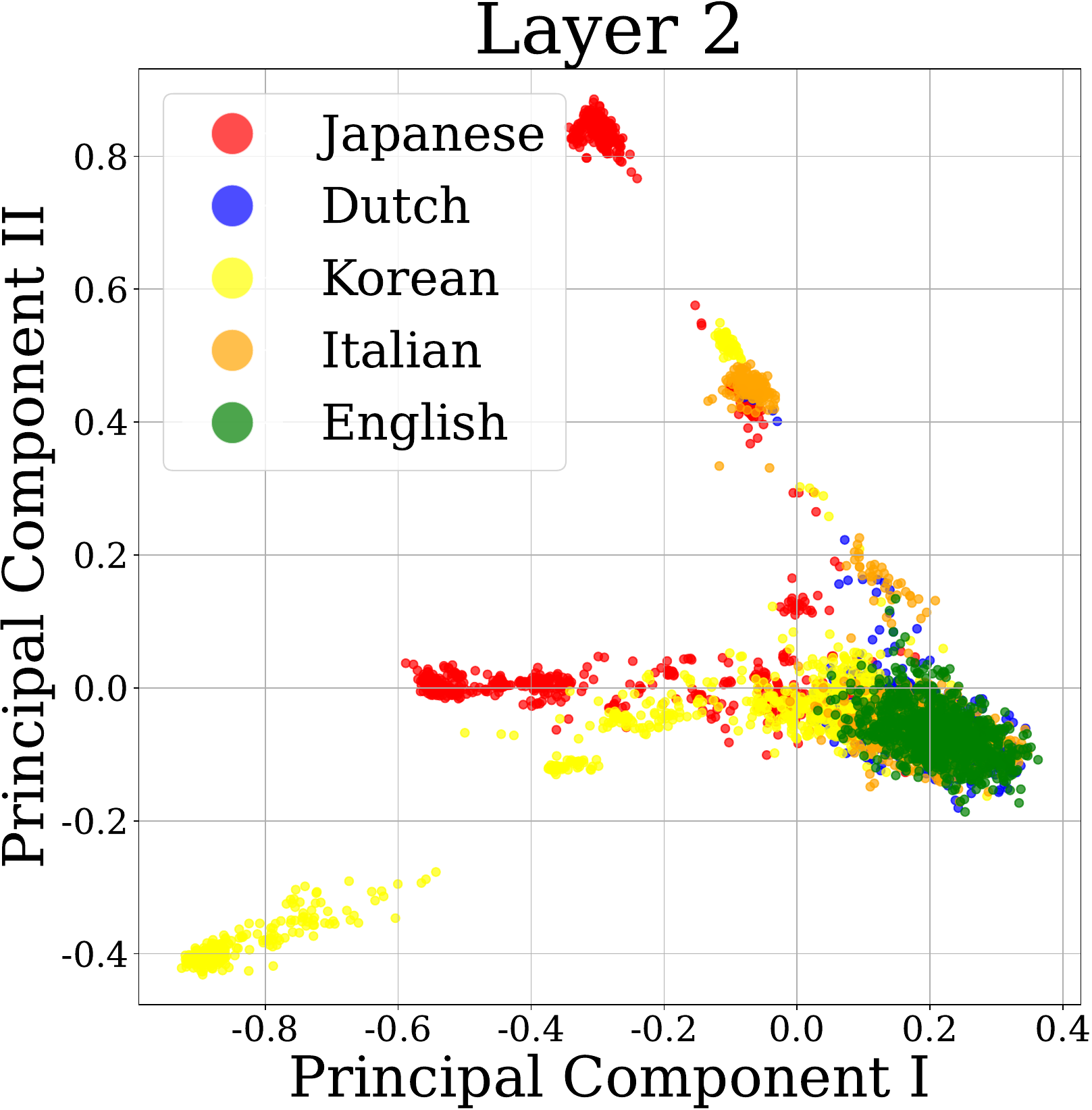}
  \includegraphics[width=0.19\linewidth]{figures/llama3/pca/3.pdf}
  \includegraphics[width=0.19\linewidth]{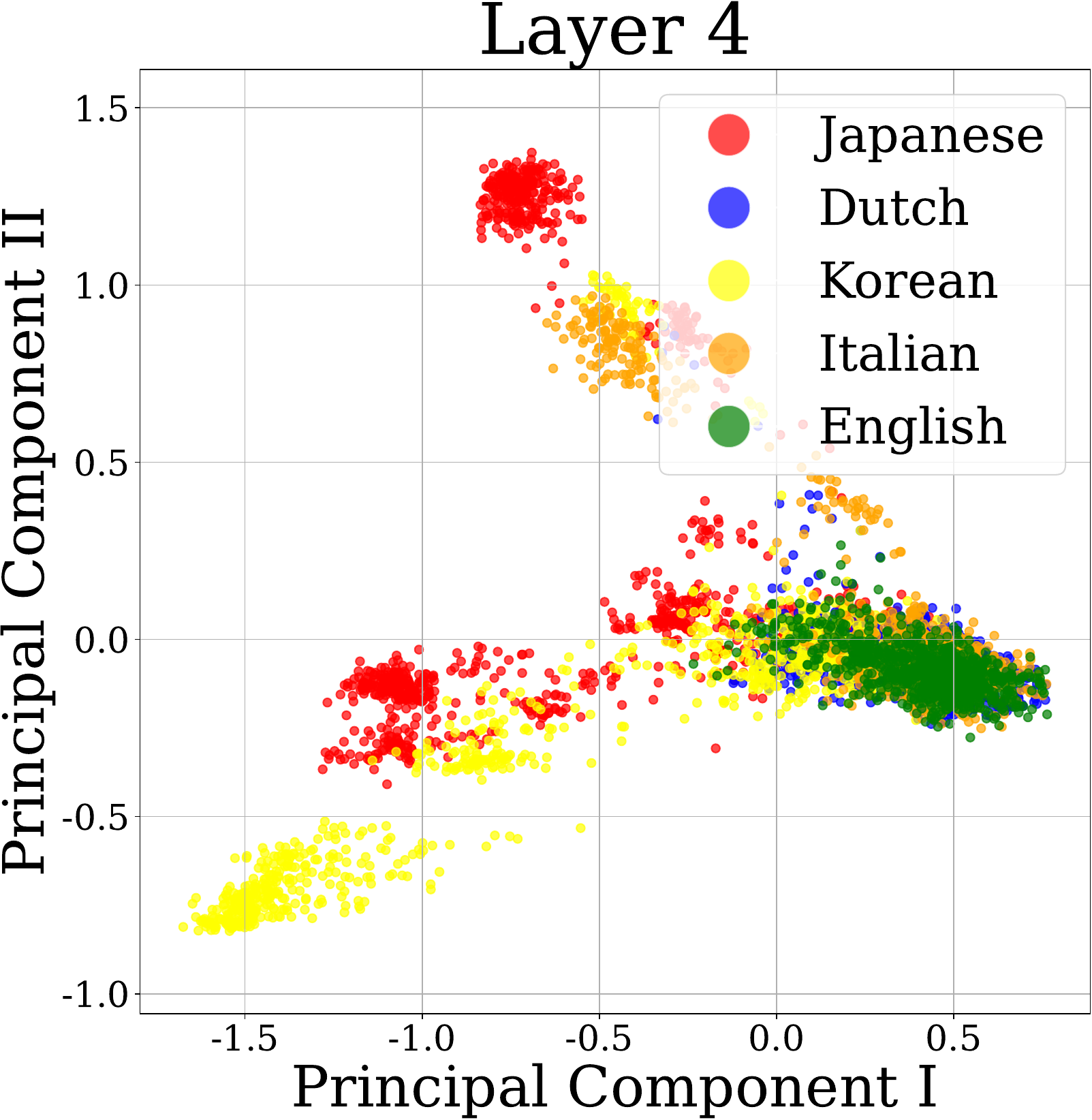}

  \includegraphics[width=0.19\linewidth]{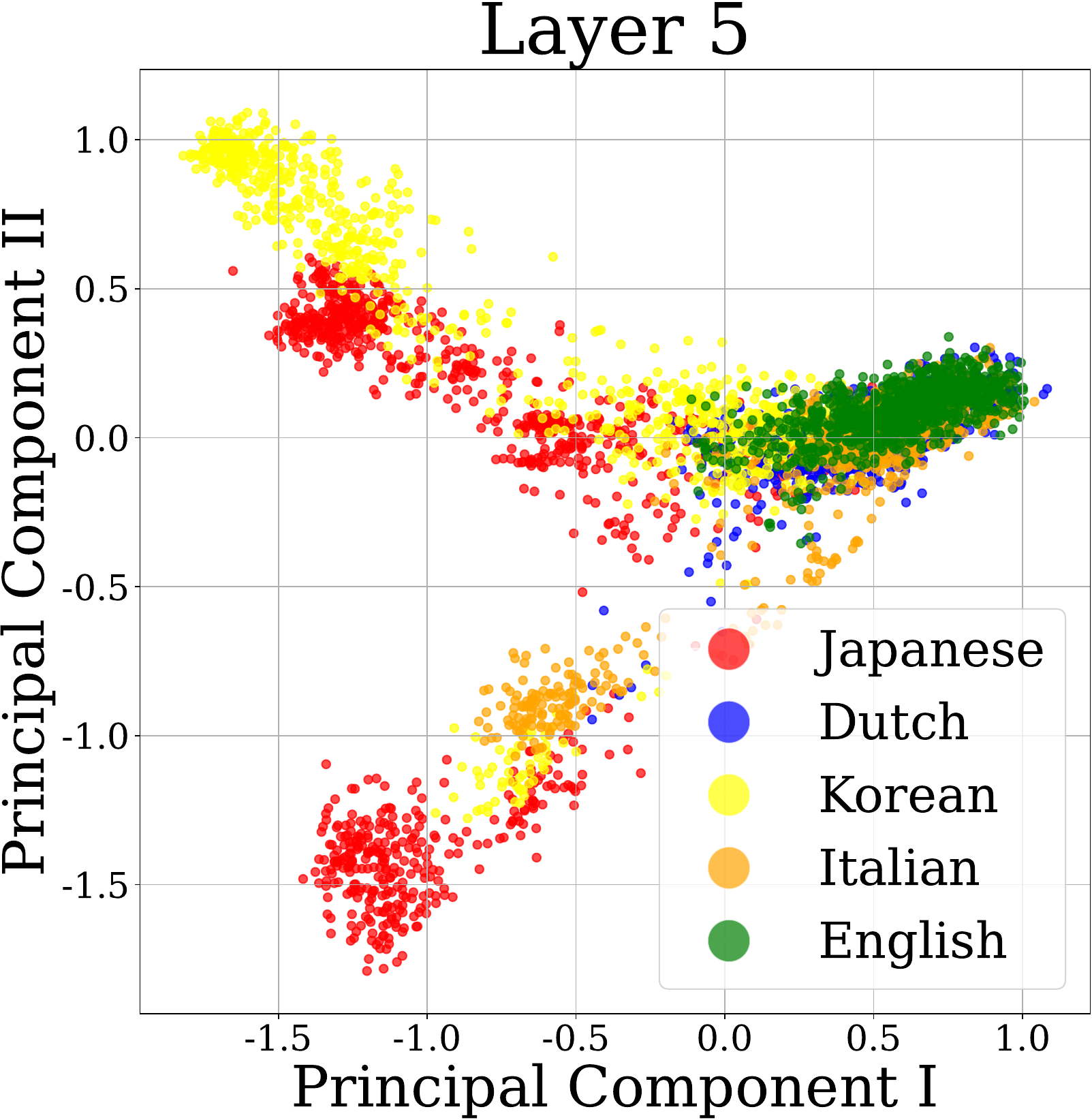}
  \includegraphics[width=0.19\linewidth]{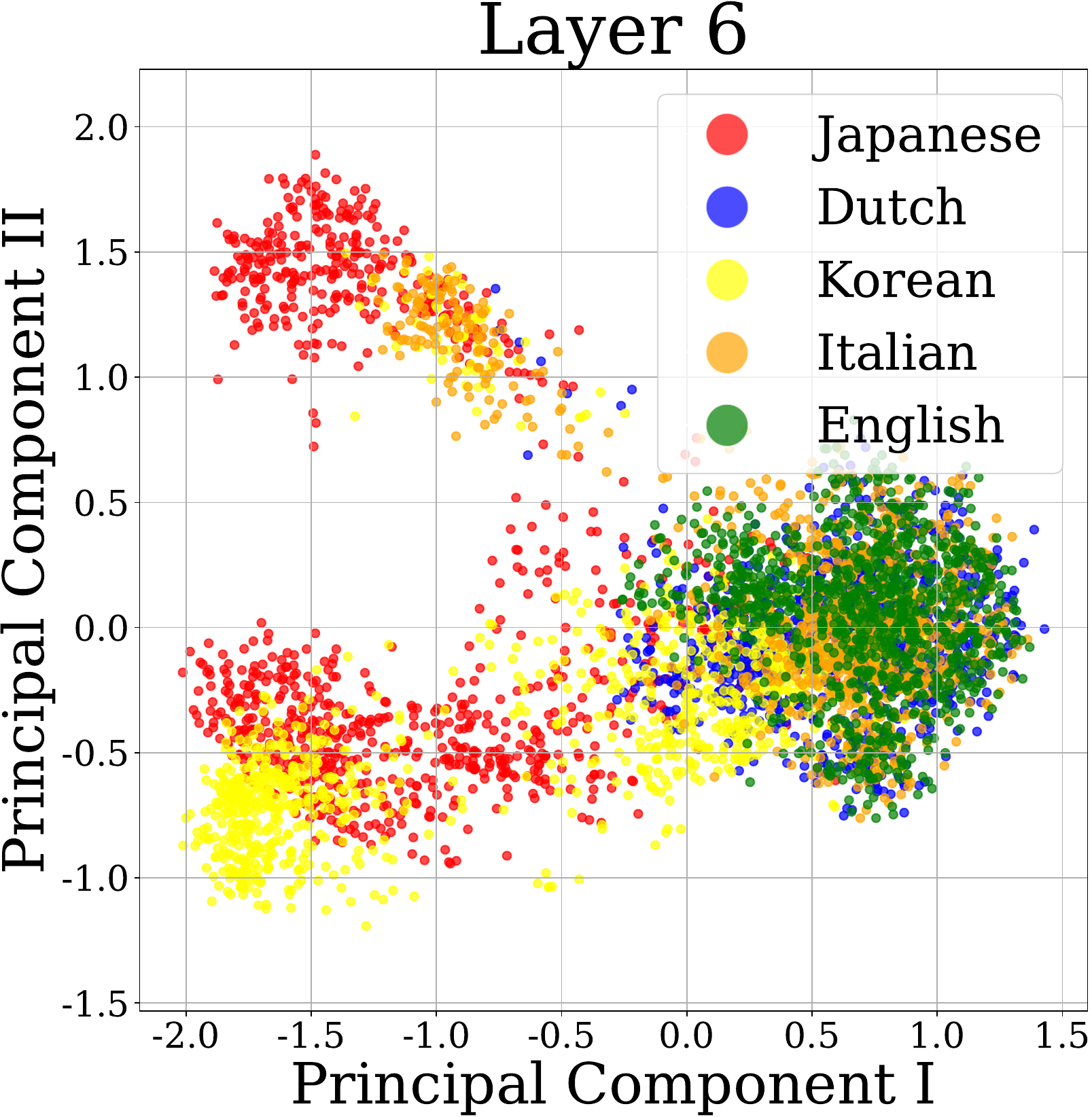}
  \includegraphics[width=0.19\linewidth]{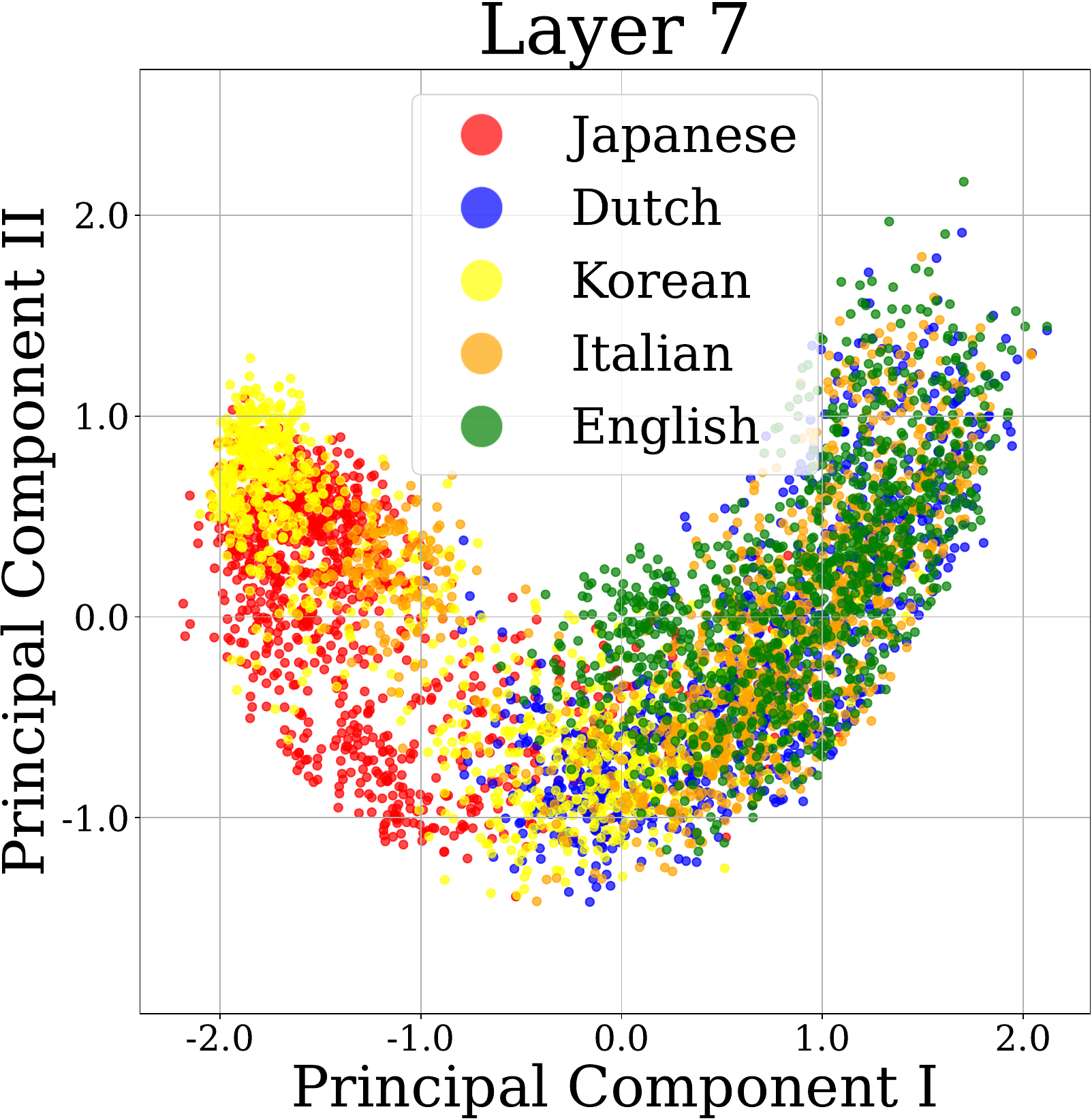}
  \includegraphics[width=0.19\linewidth]{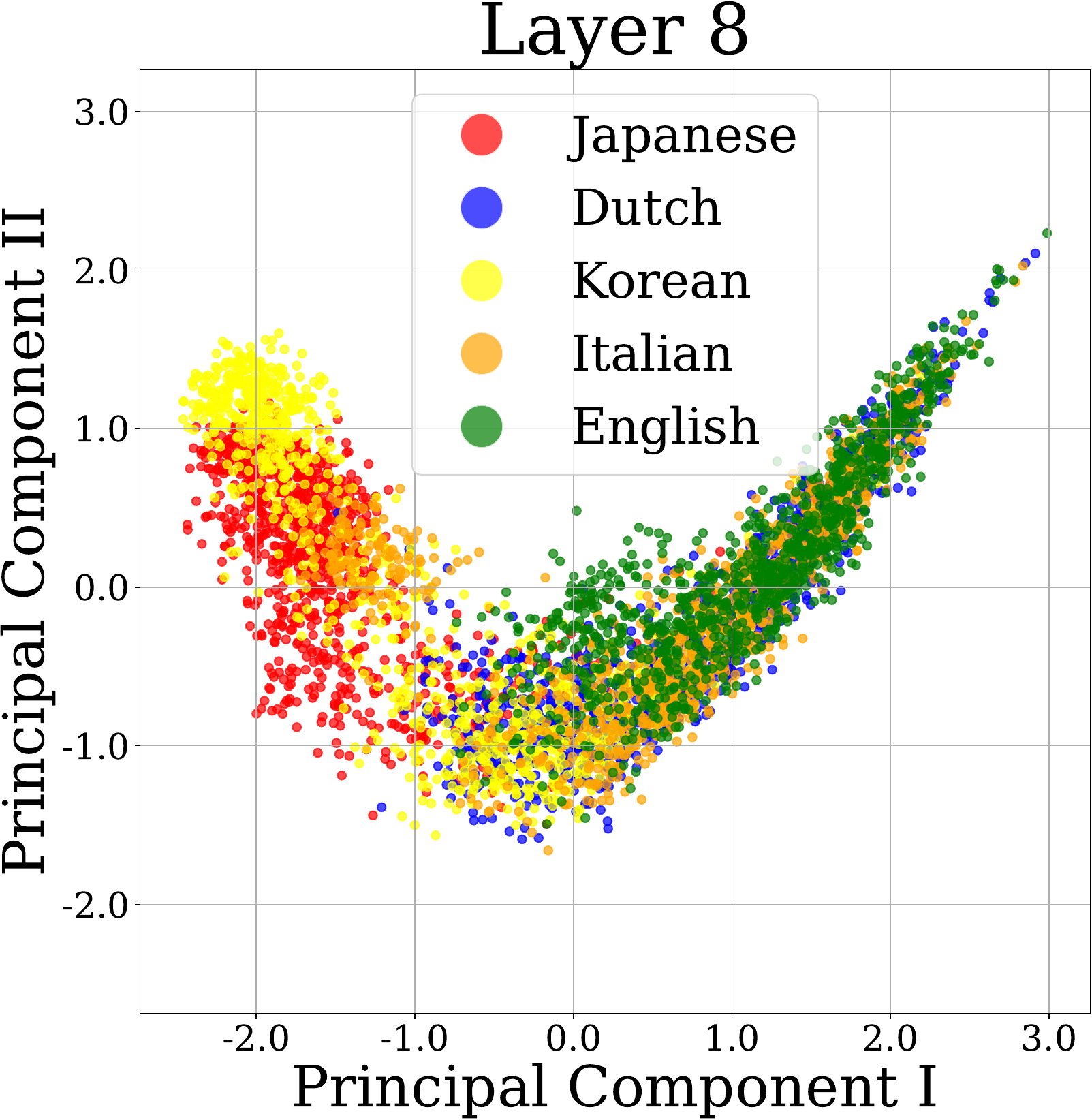}
  \includegraphics[width=0.19\linewidth]{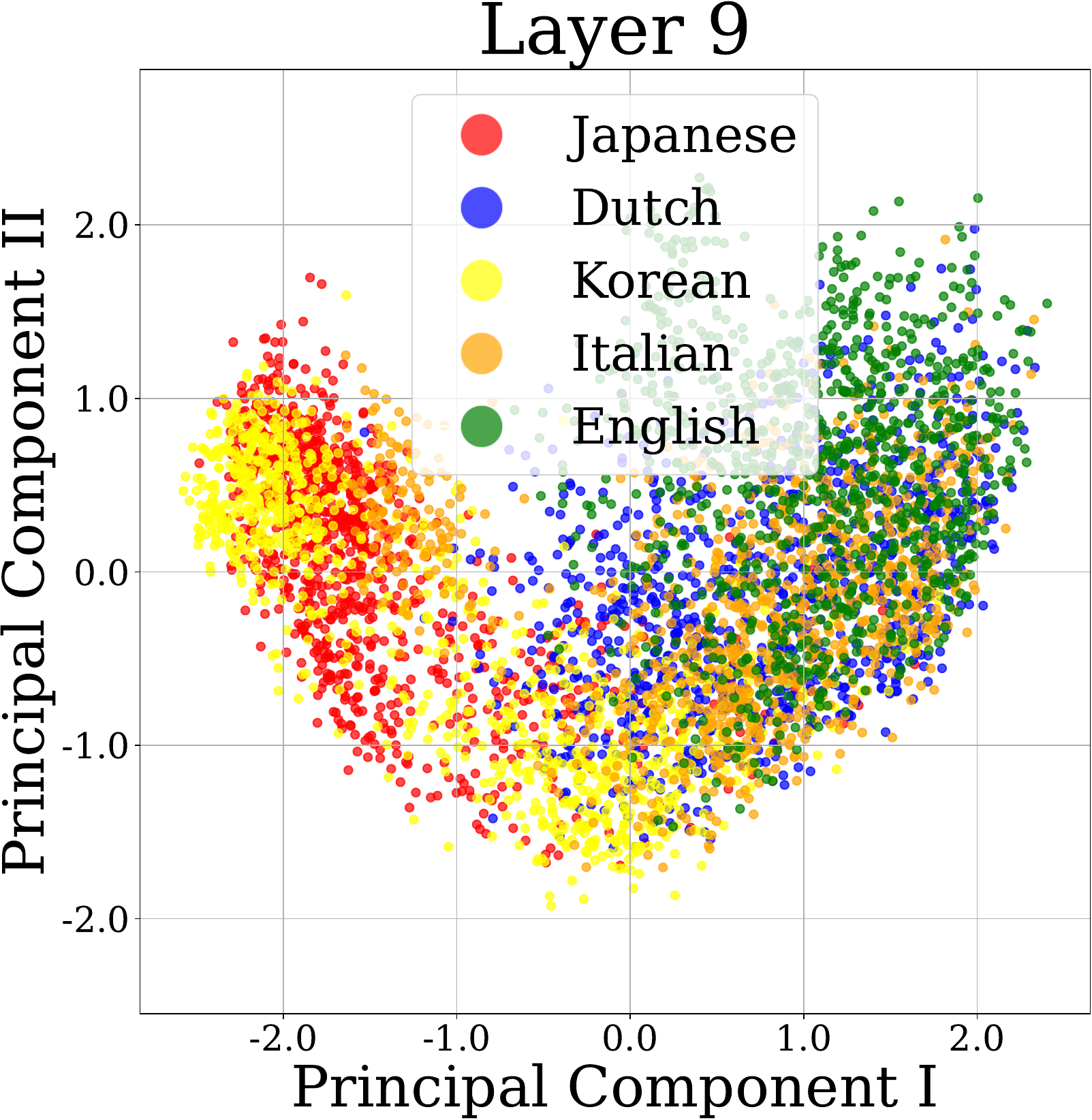}

  \includegraphics[width=0.19\linewidth]{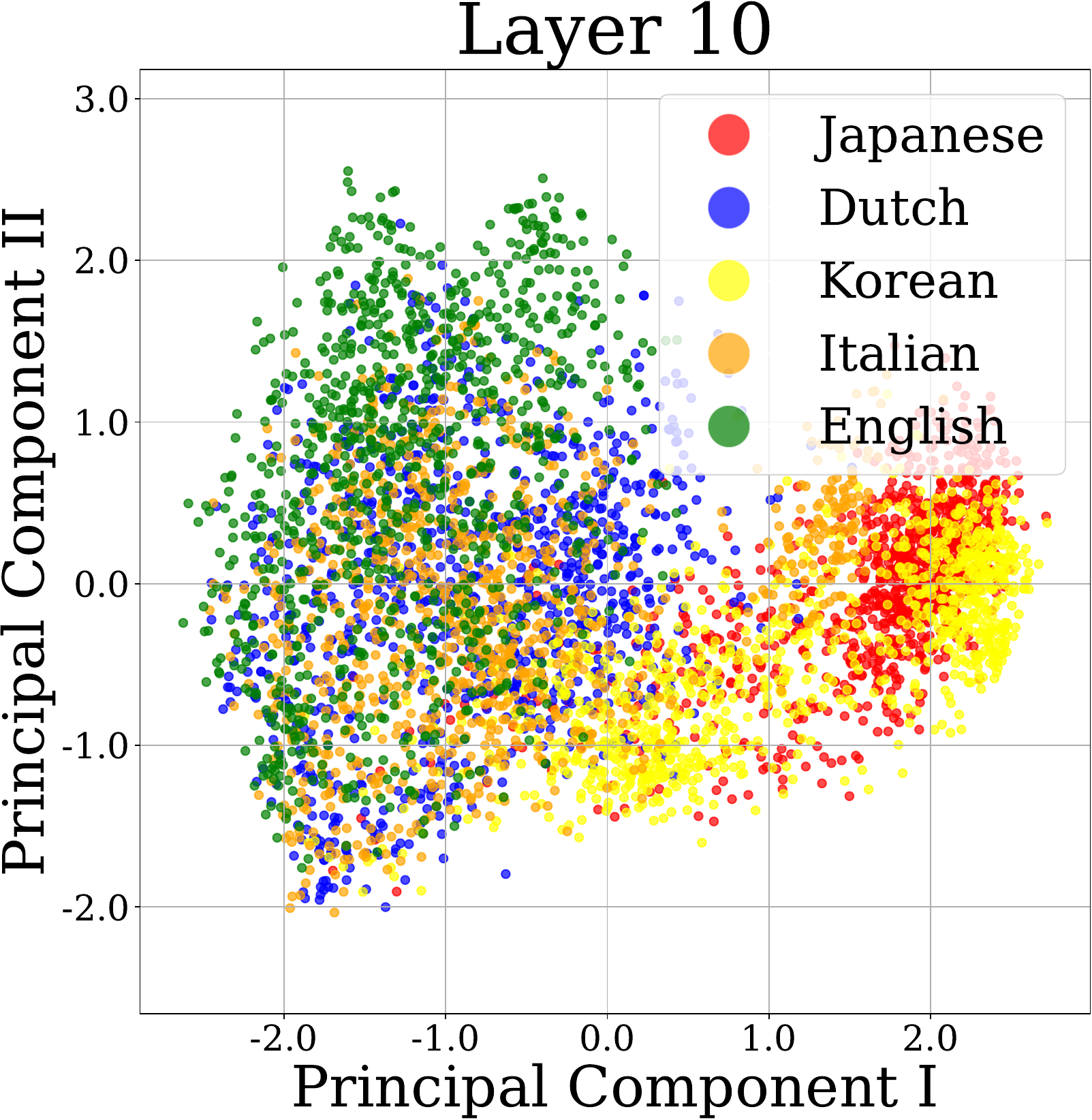}
  \includegraphics[width=0.19\linewidth]{figures/llama3/pca/11.pdf}
  \includegraphics[width=0.19\linewidth]{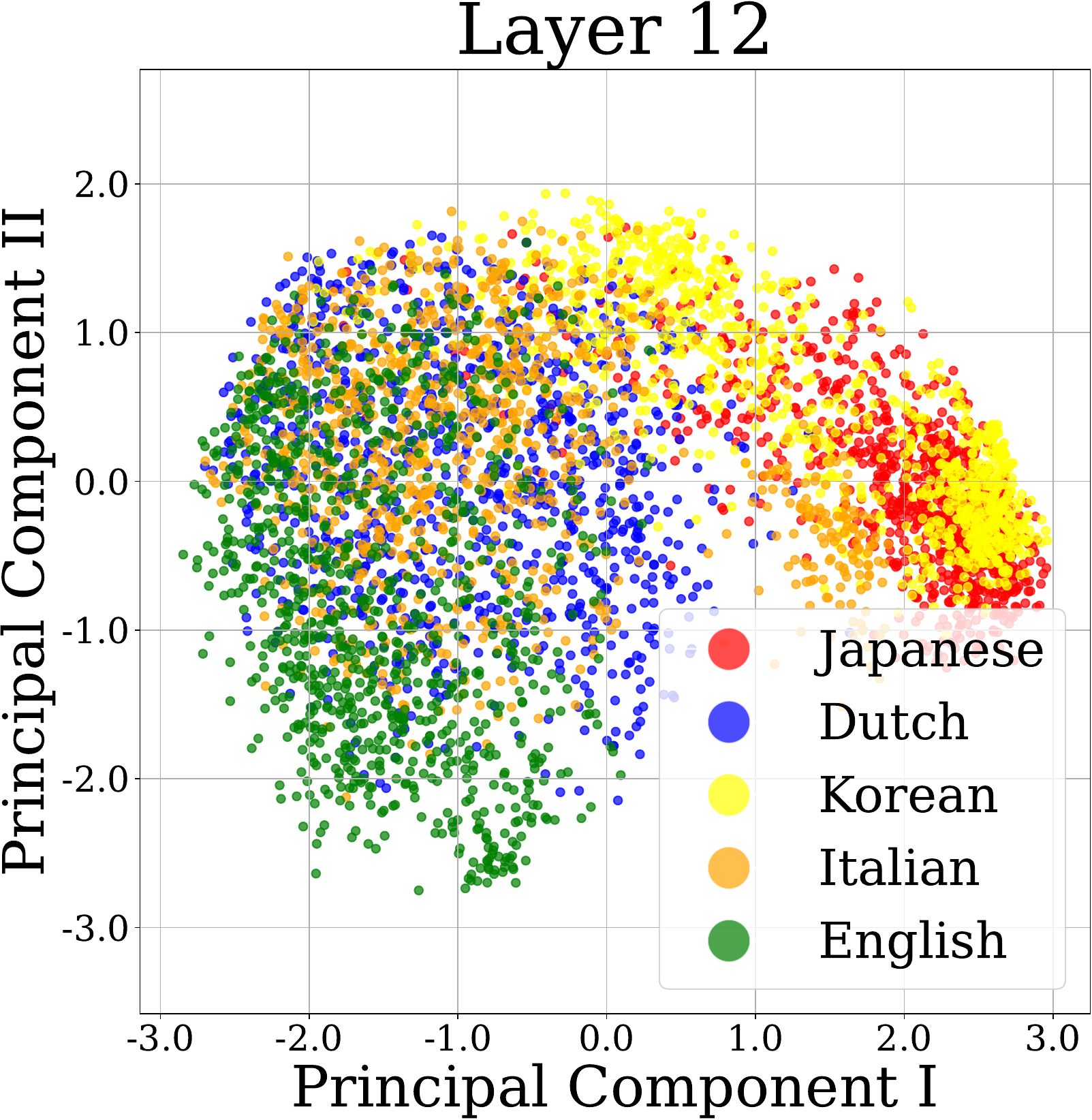}
  \includegraphics[width=0.19\linewidth]{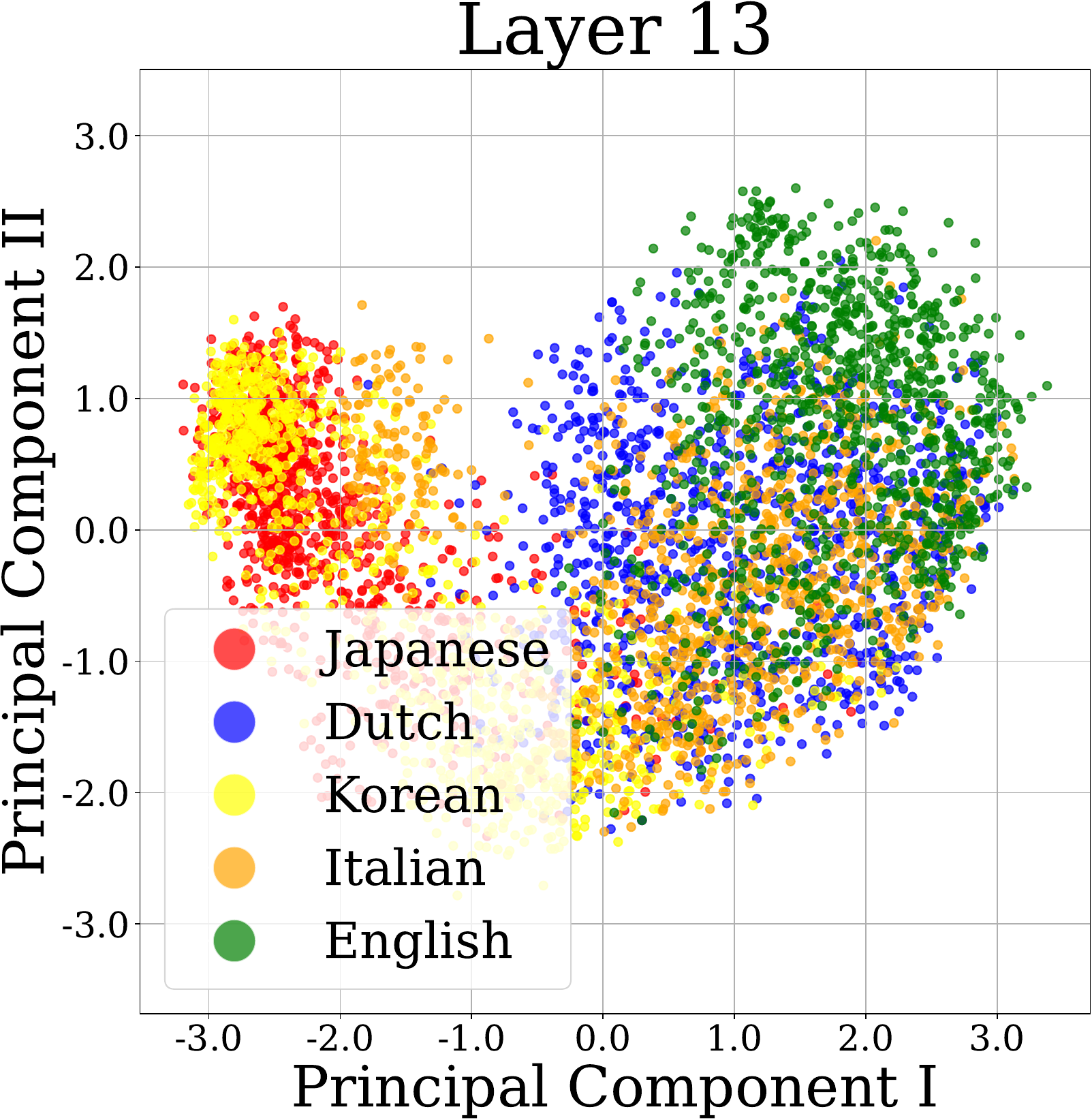}
  \includegraphics[width=0.19\linewidth]{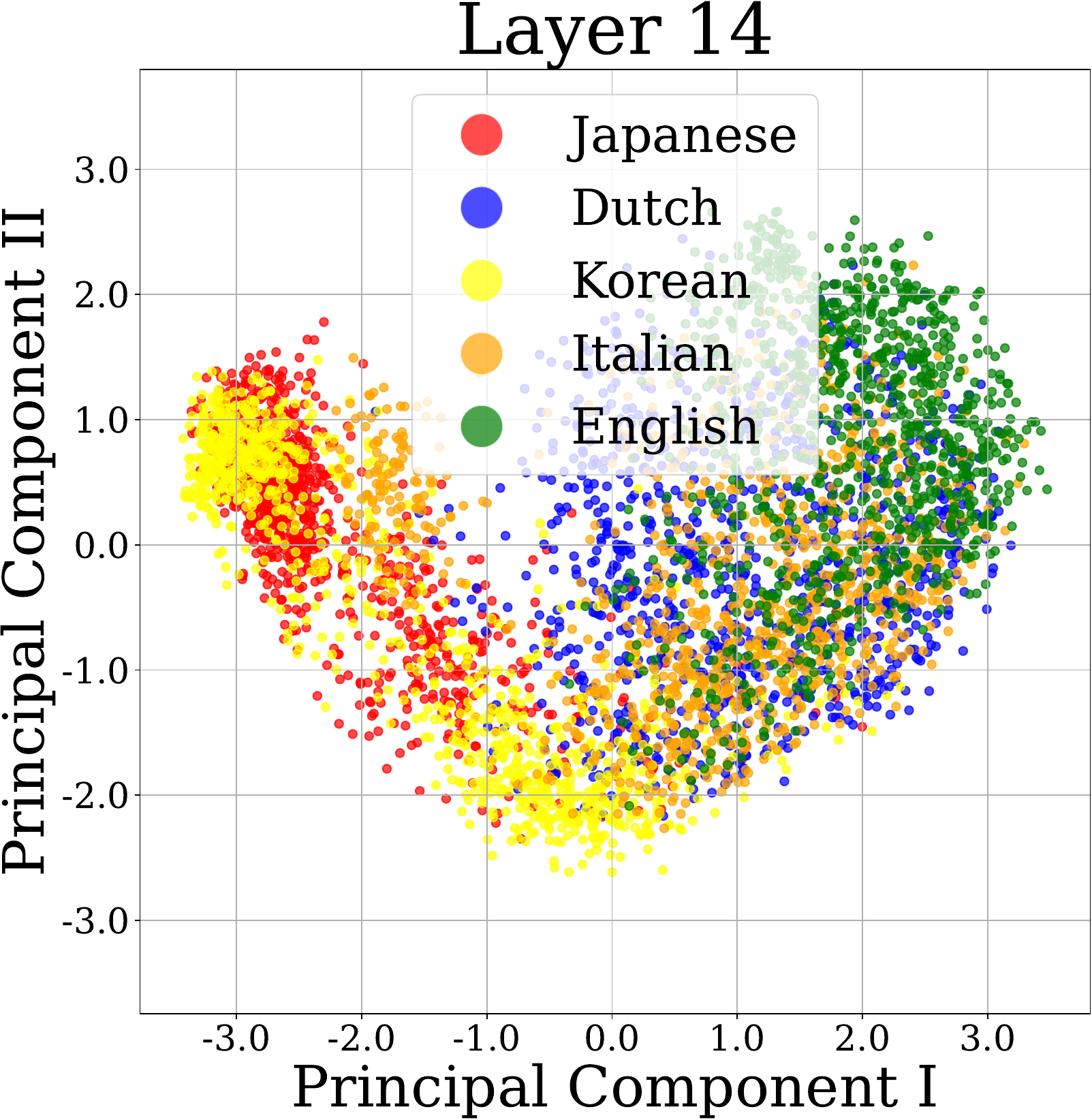}

  \includegraphics[width=0.19\linewidth]{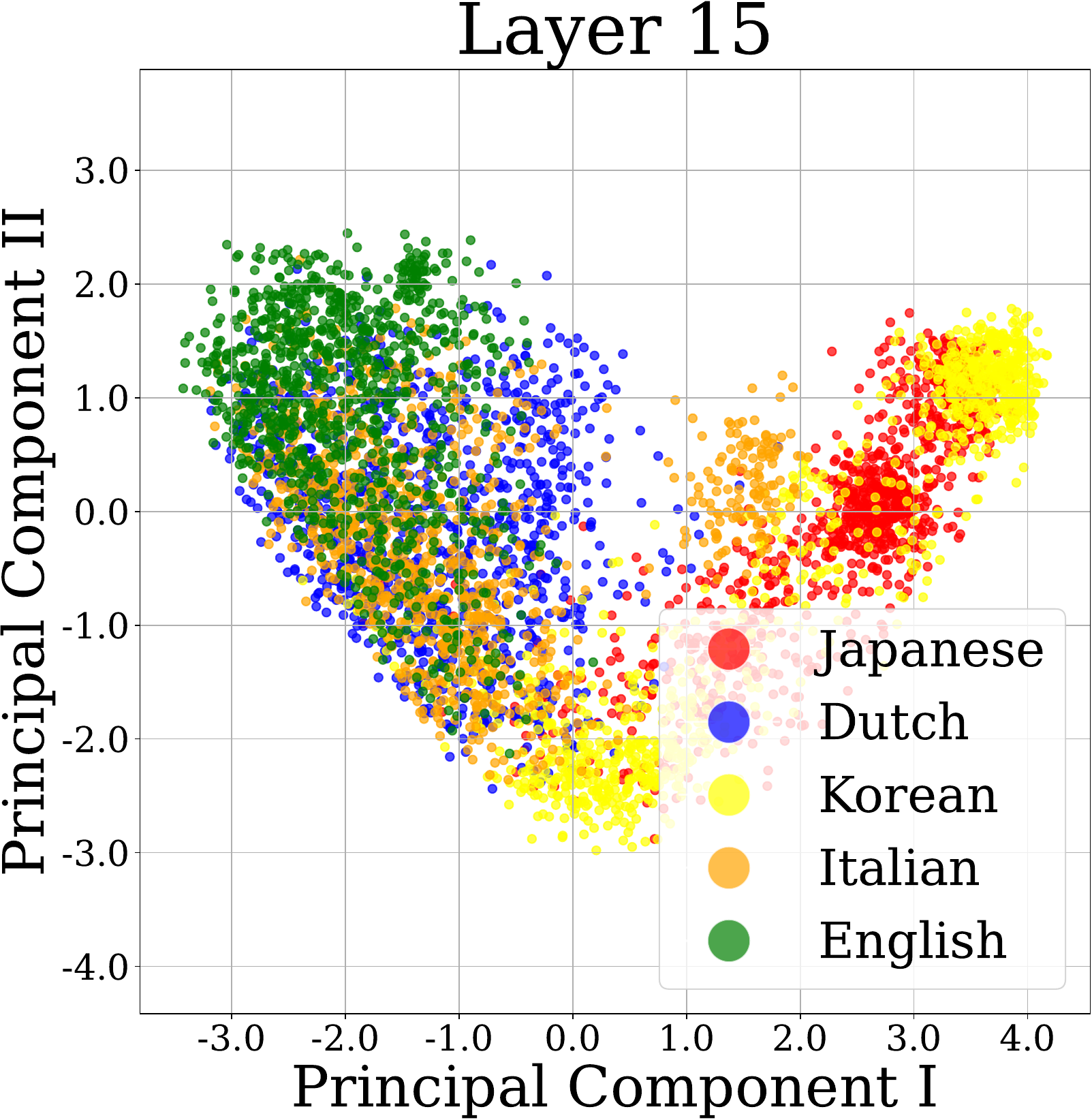}
  \includegraphics[width=0.19\linewidth]{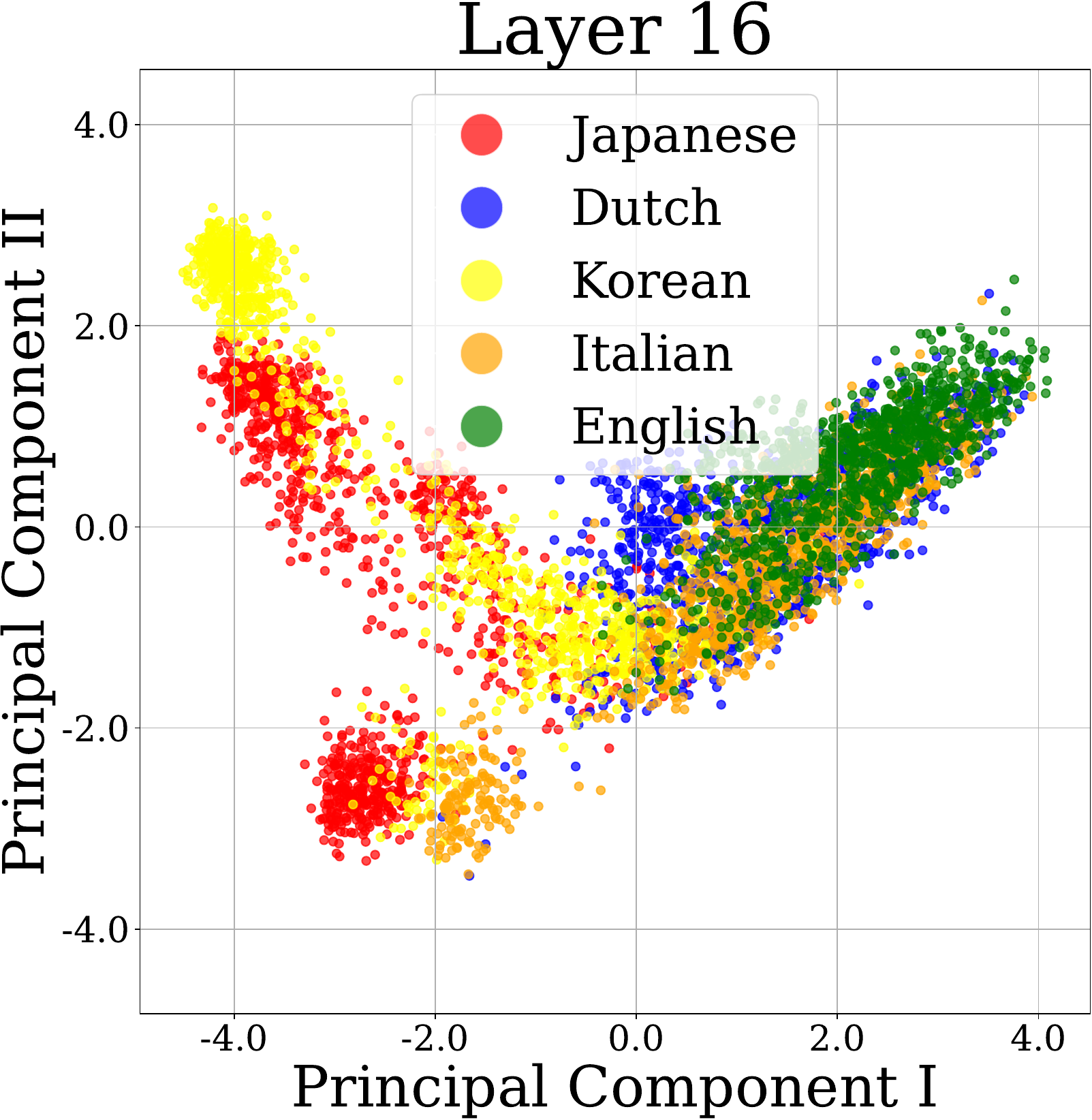}
  \includegraphics[width=0.19\linewidth]{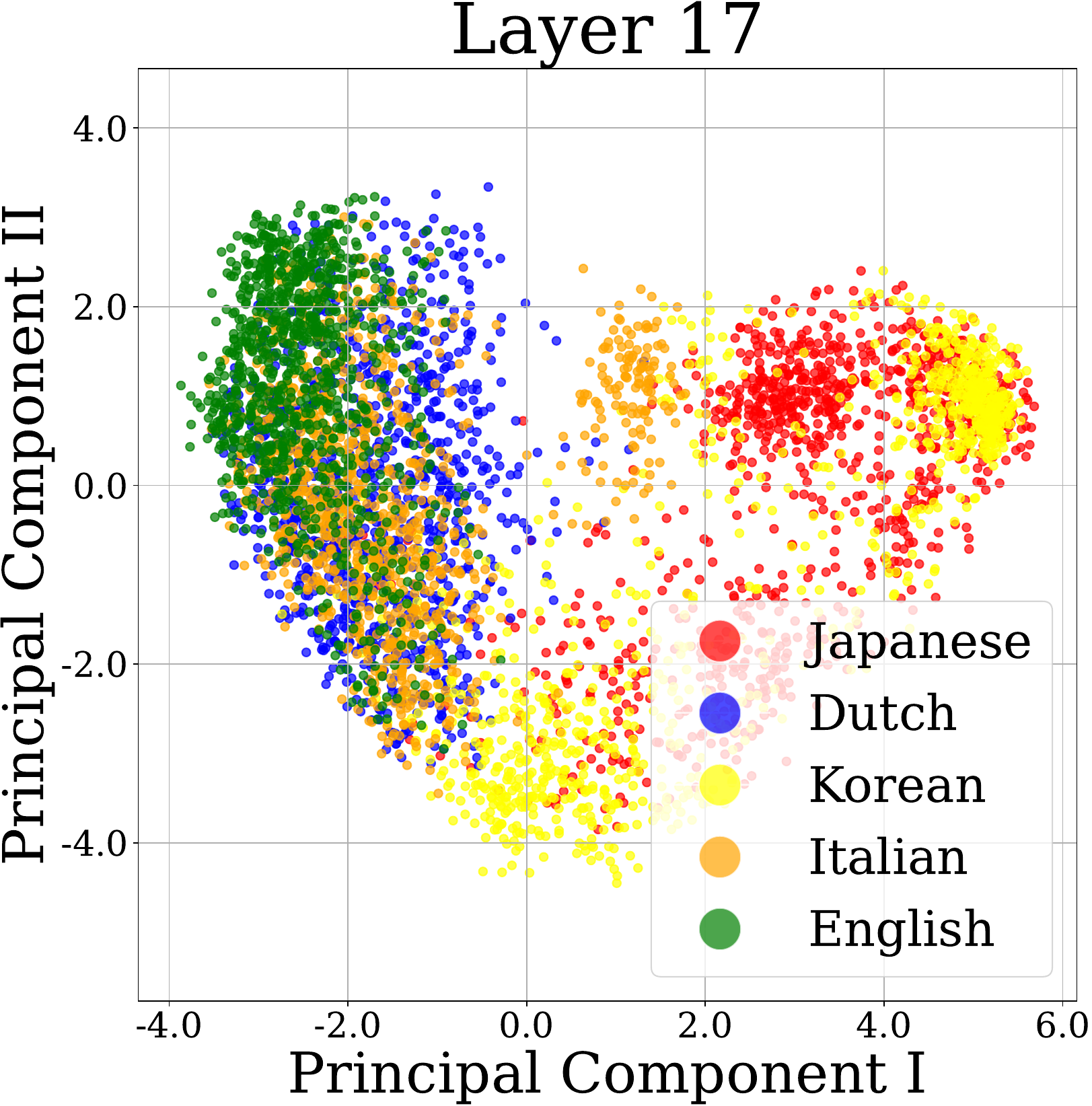}
  \includegraphics[width=0.19\linewidth]{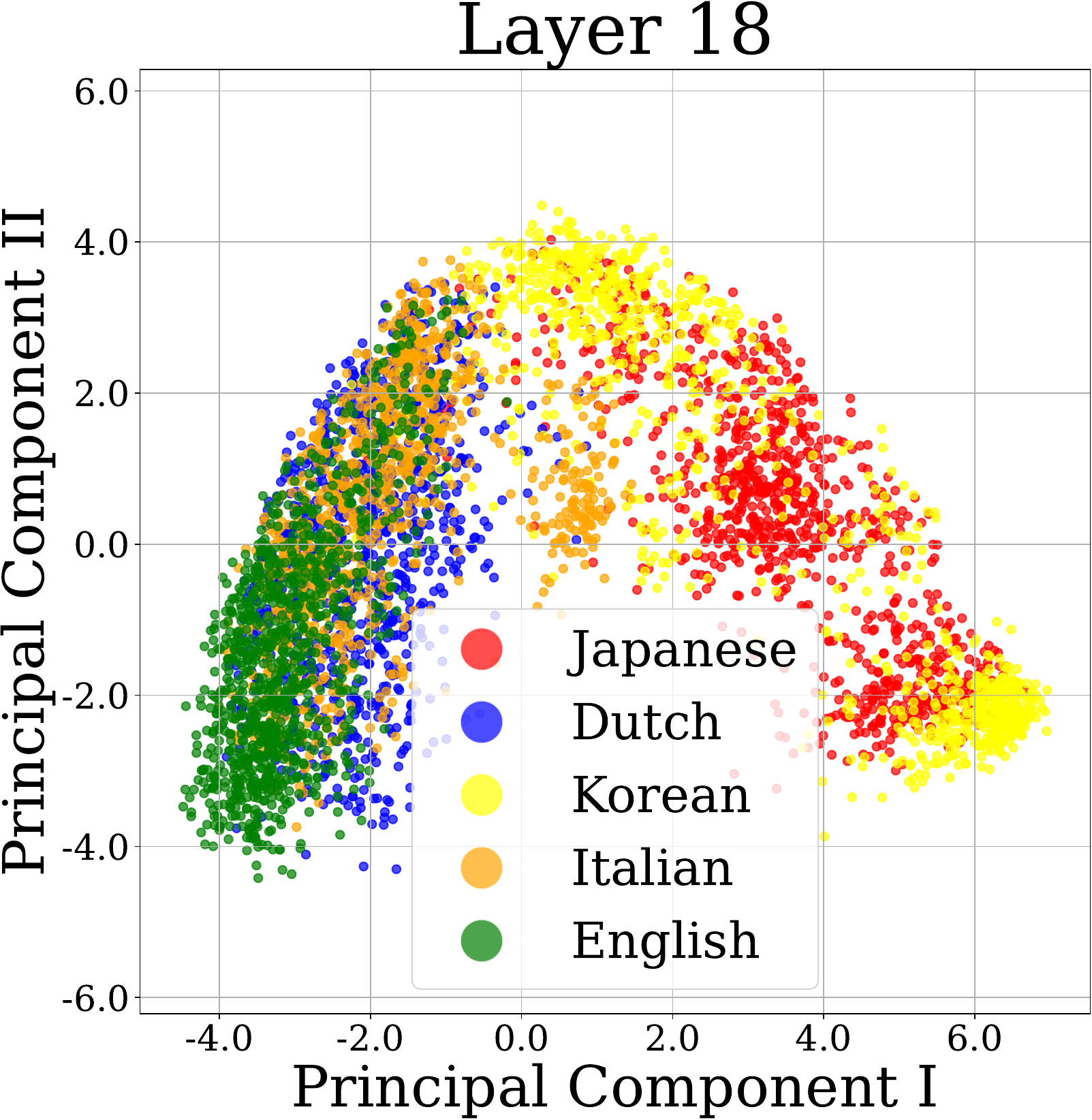}
  \includegraphics[width=0.19\linewidth]{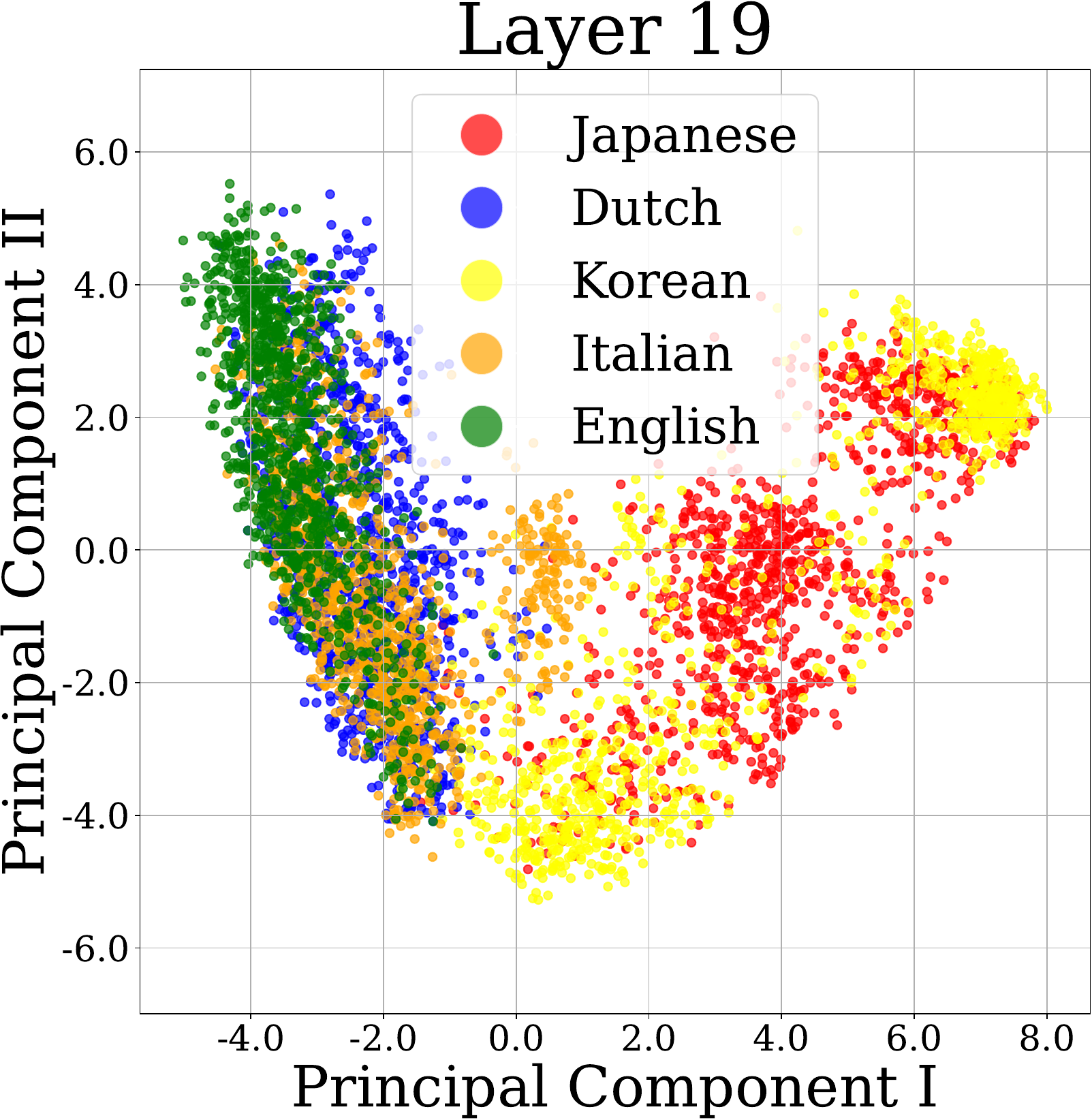}

  \includegraphics[width=0.19\linewidth]{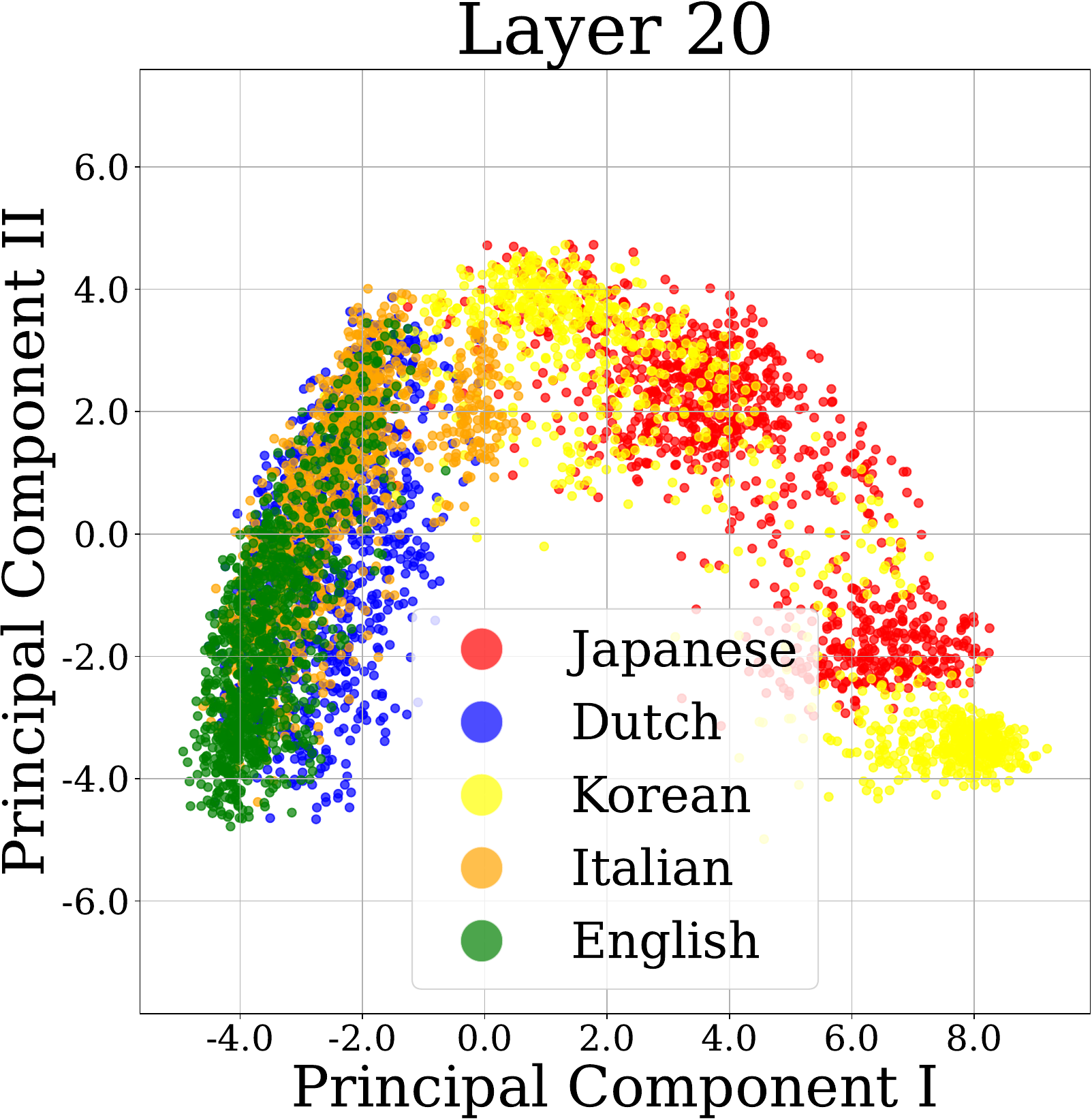}
  \includegraphics[width=0.19\linewidth]{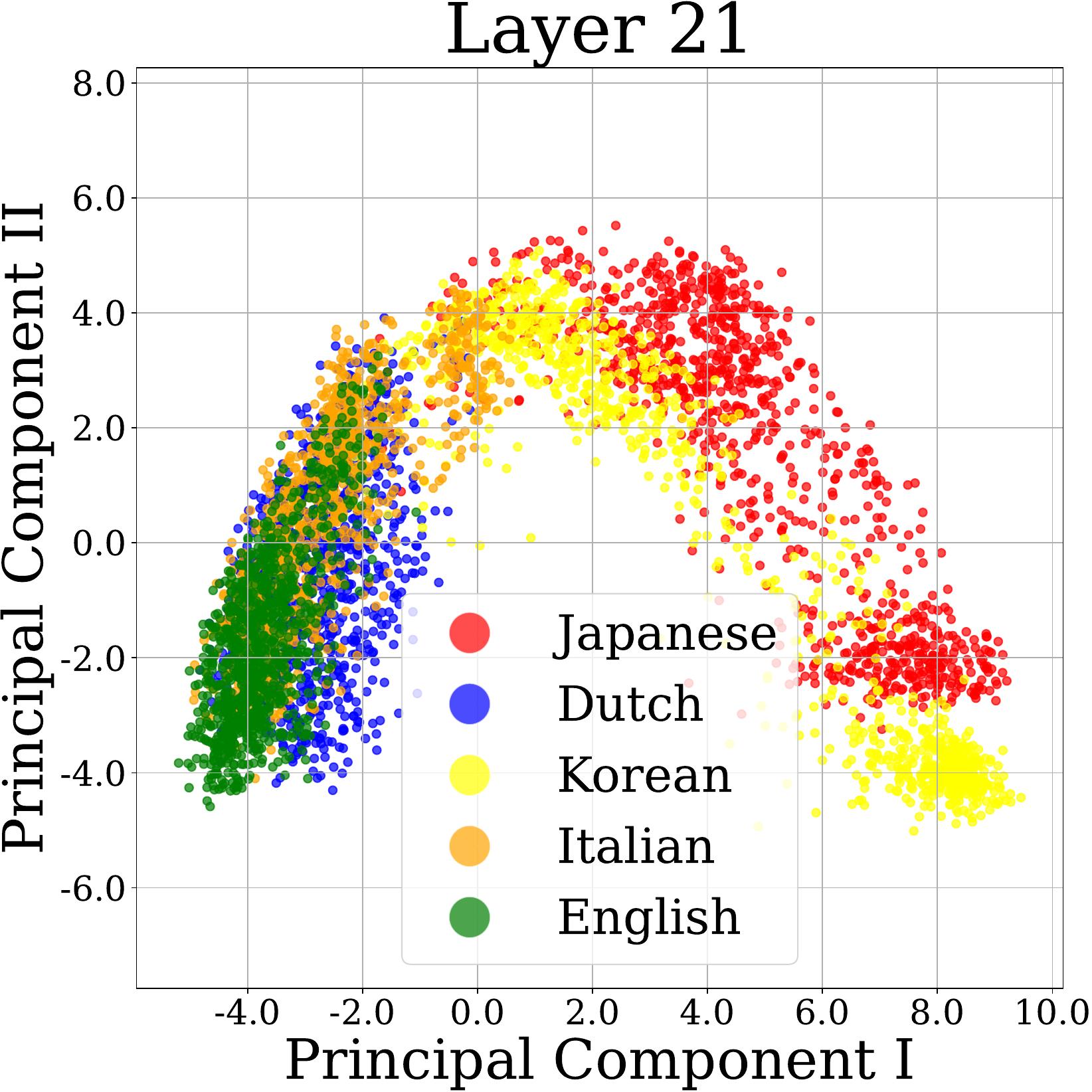}
  \includegraphics[width=0.19\linewidth]{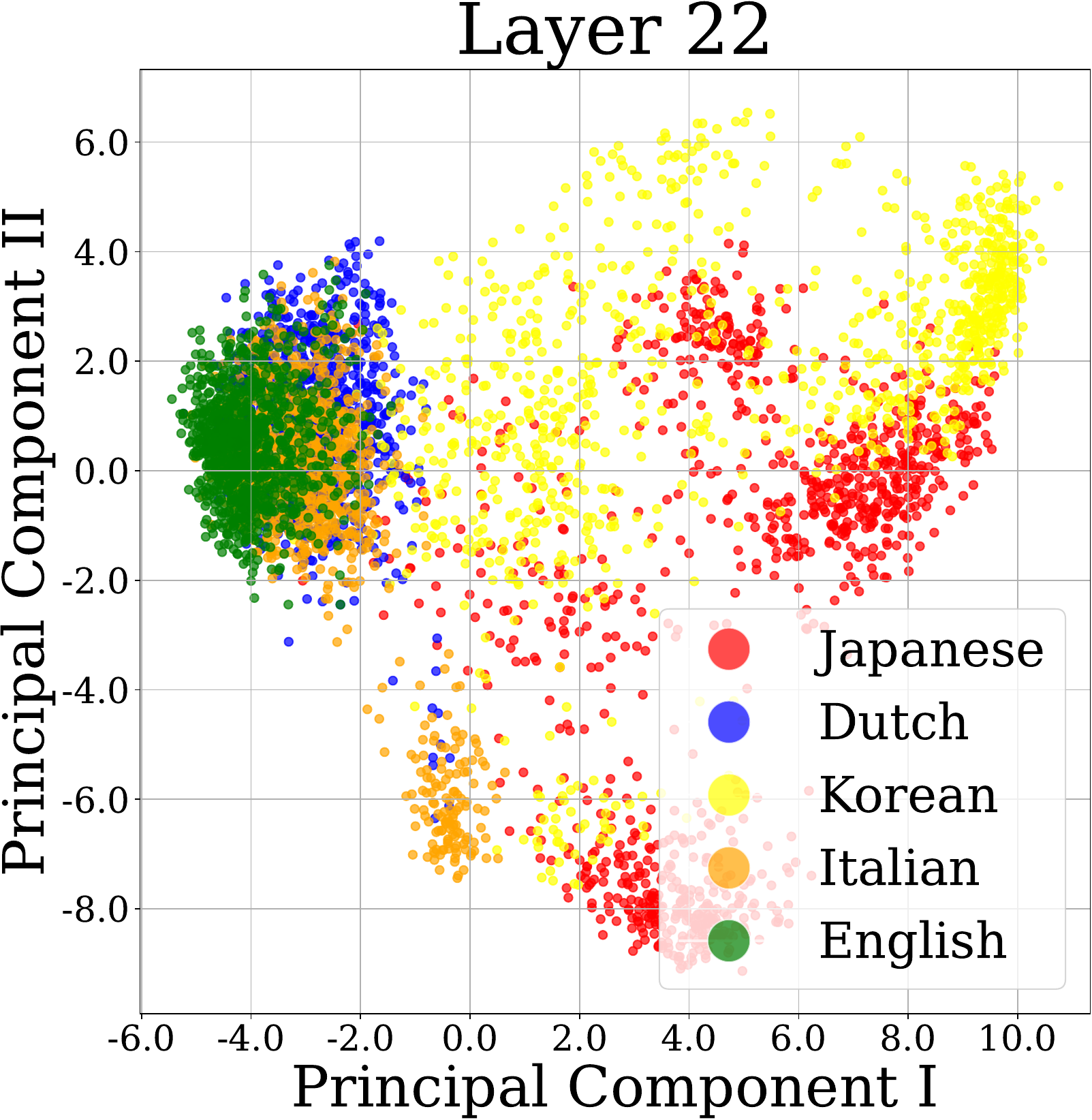}
  \includegraphics[width=0.19\linewidth]{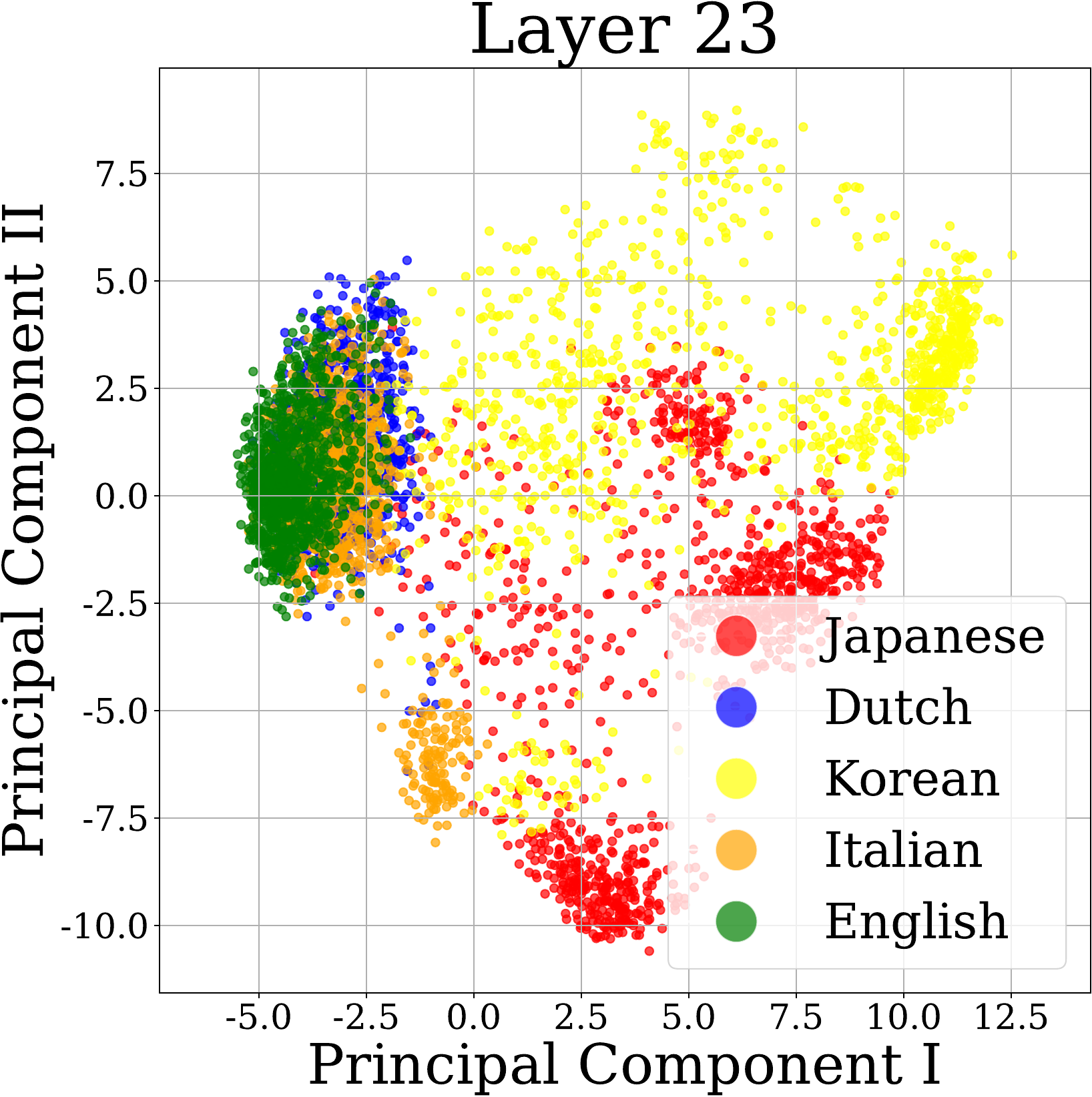}
  \includegraphics[width=0.19\linewidth]{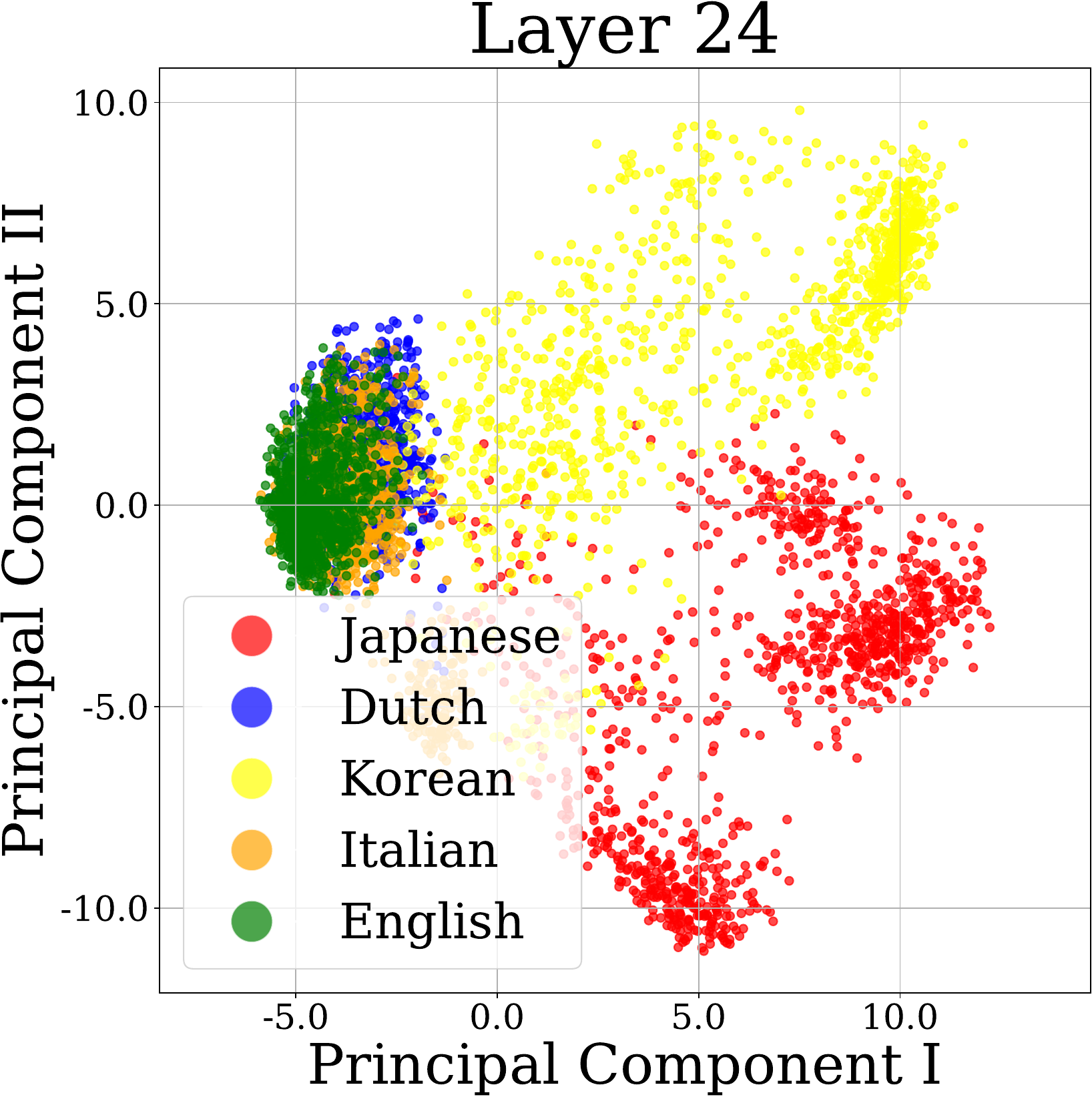}

  \includegraphics[width=0.19\linewidth]{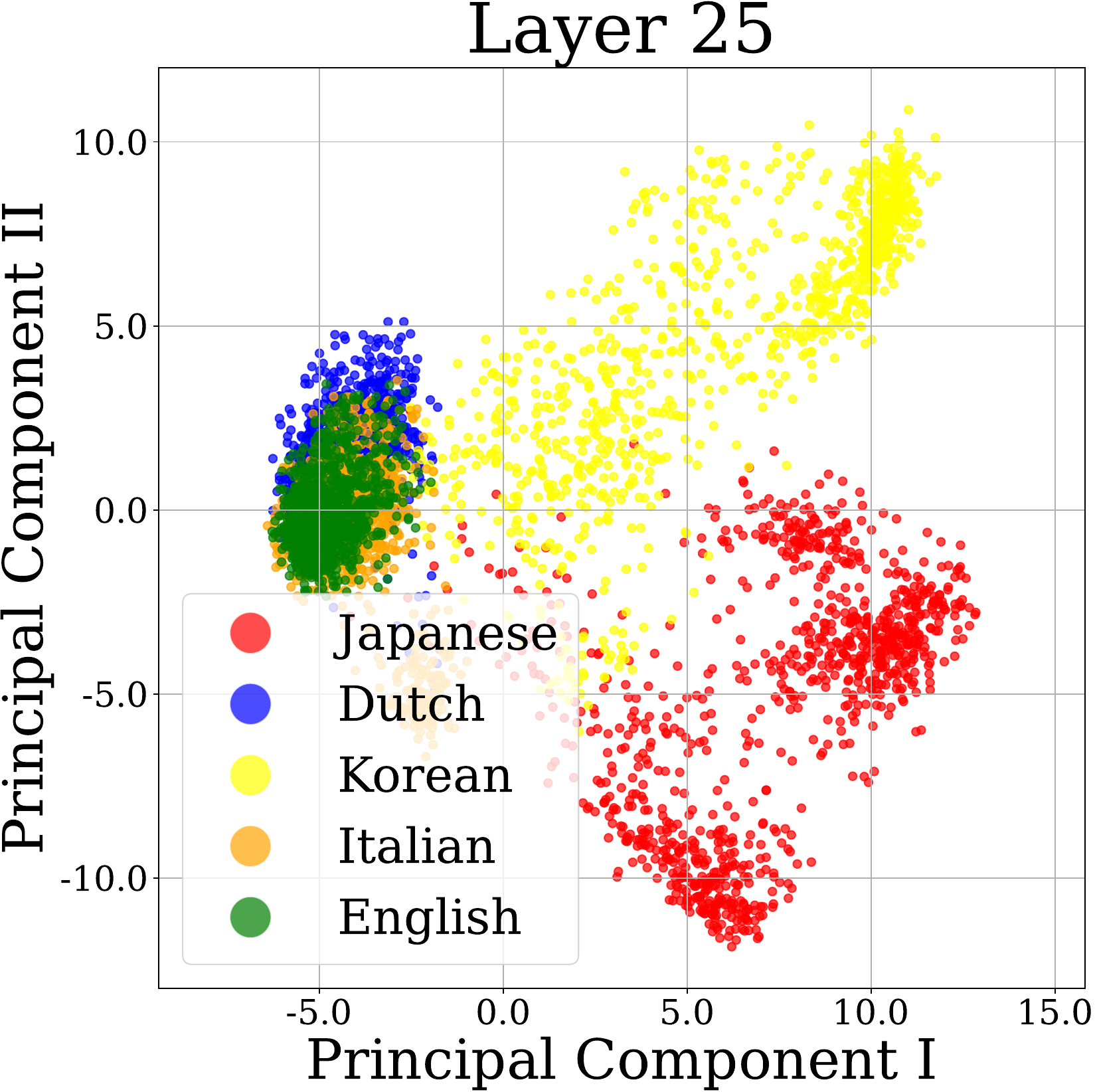}
  \includegraphics[width=0.19\linewidth]{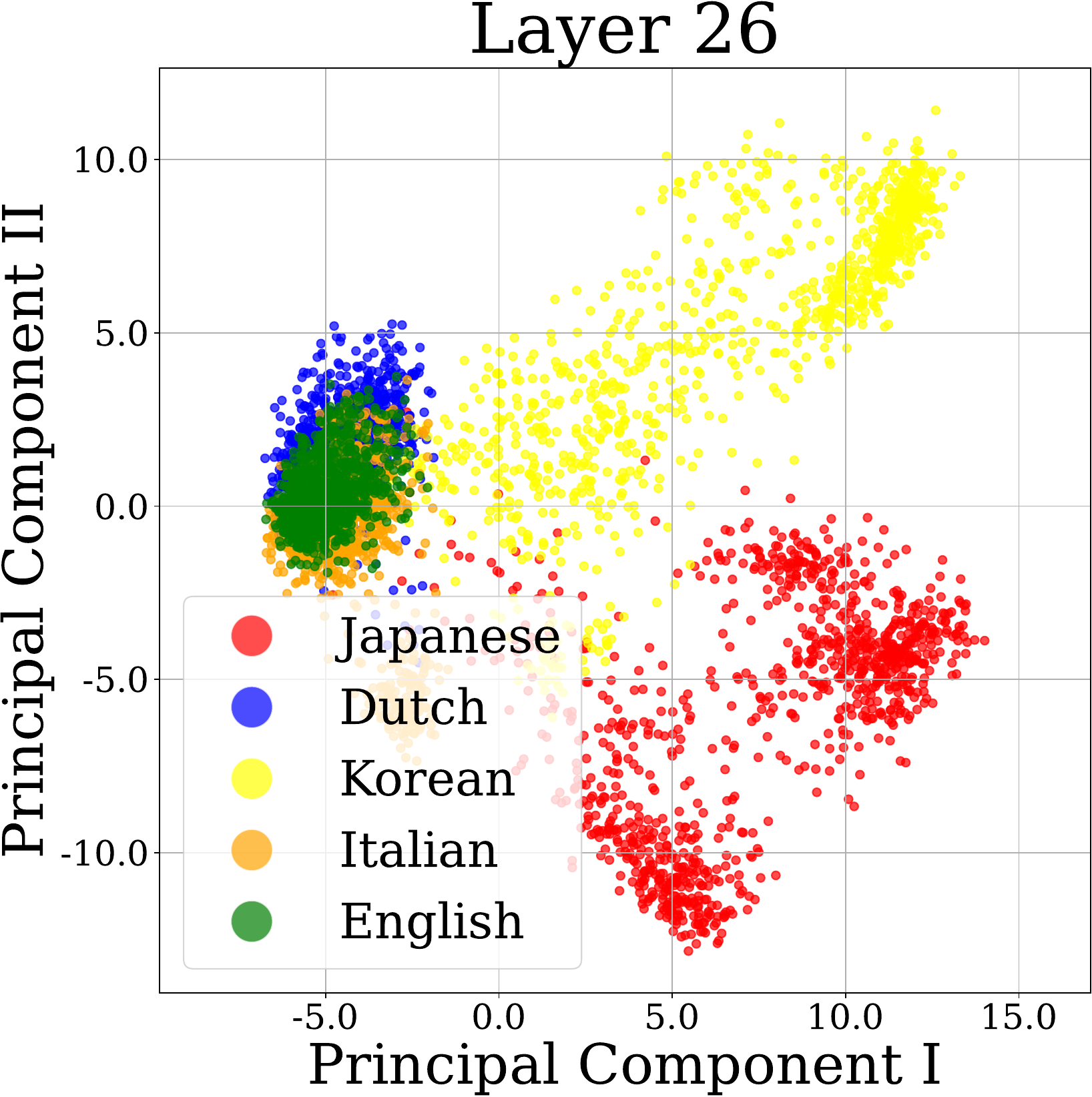}
  \includegraphics[width=0.19\linewidth]{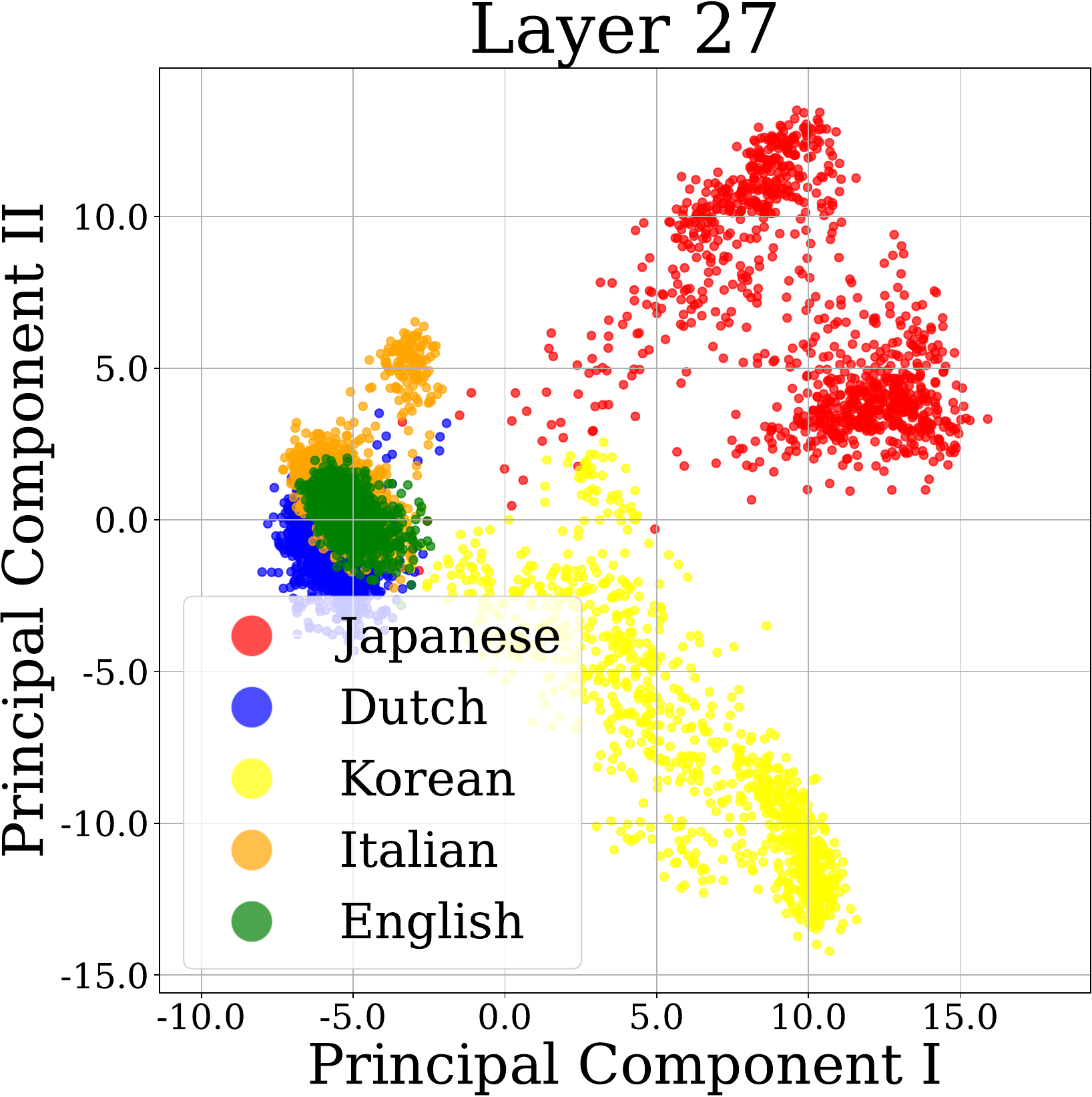}
  \includegraphics[width=0.19\linewidth]{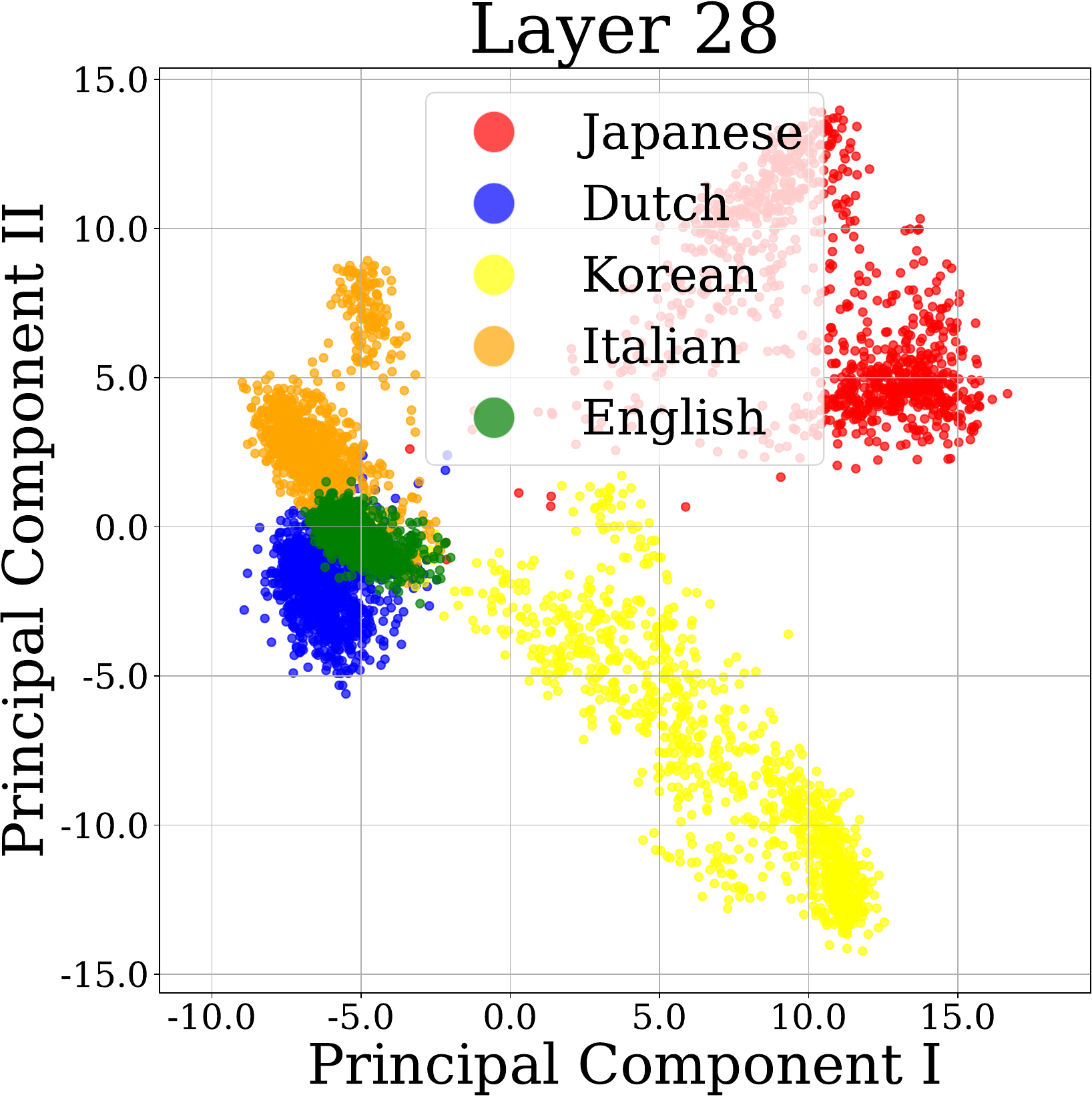}
  \includegraphics[width=0.19\linewidth]{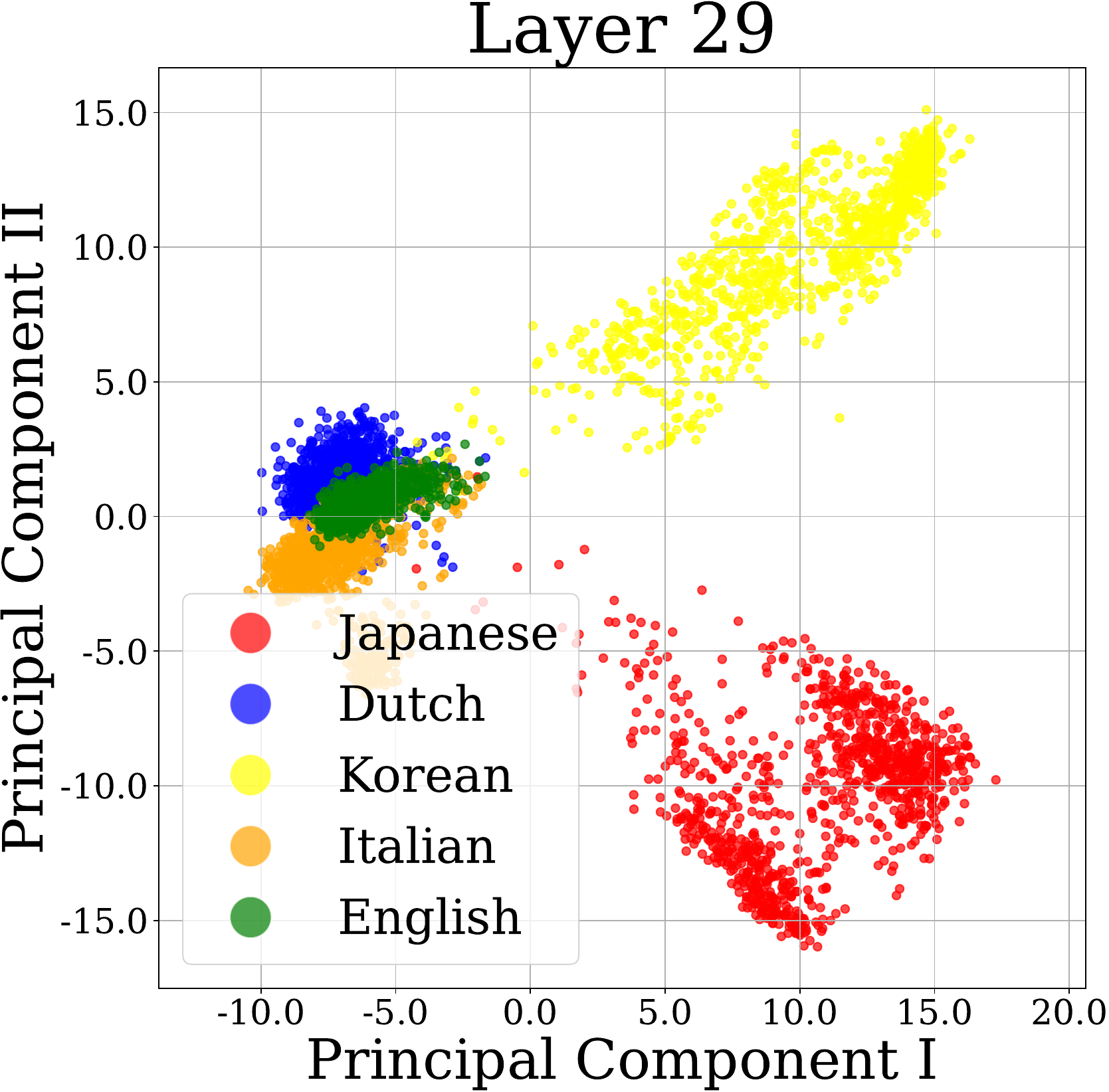}

  \includegraphics[width=0.19\linewidth]{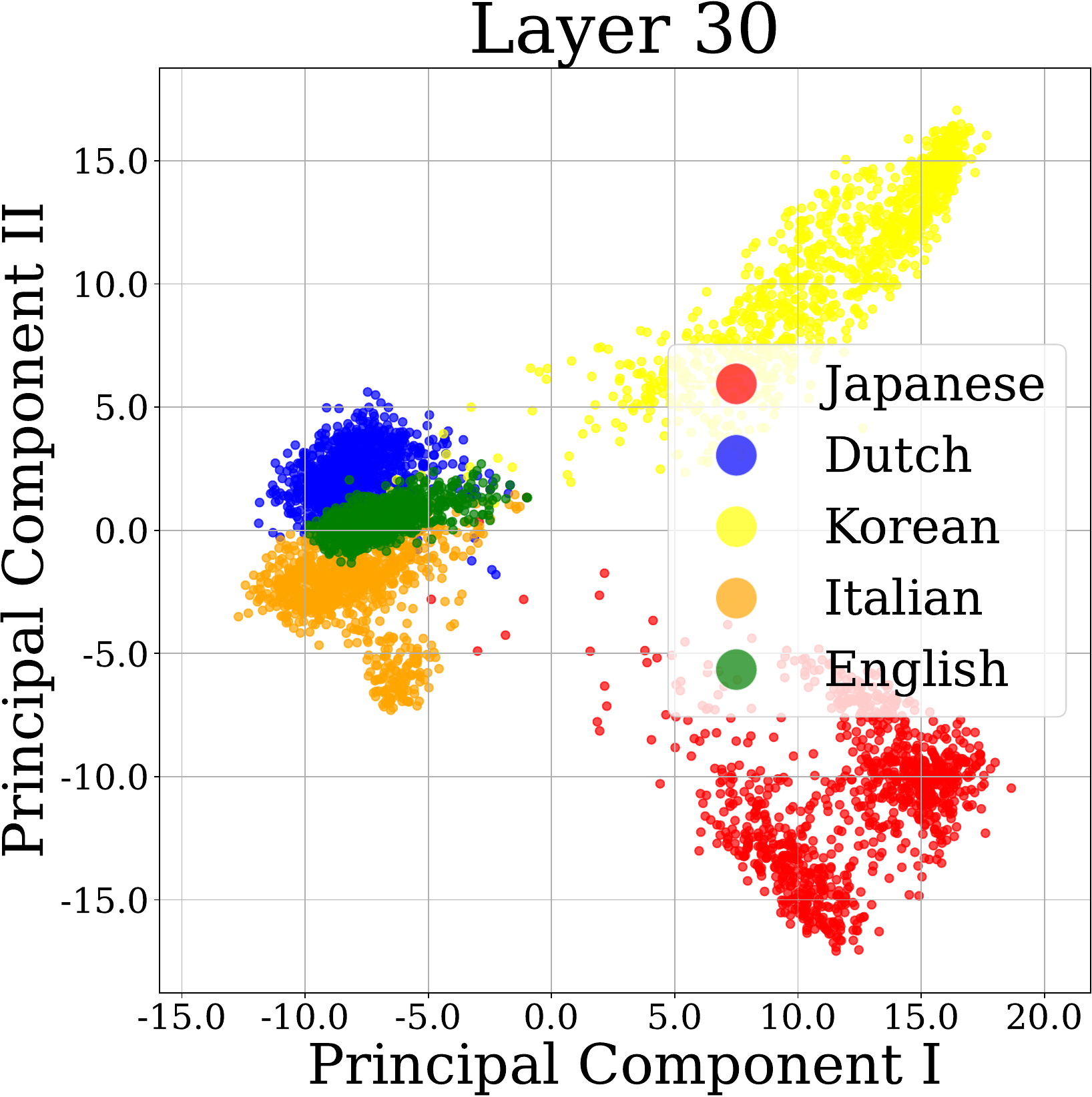}
  \includegraphics[width=0.19\linewidth]{figures/llama3/pca/31.pdf}
  \includegraphics[width=0.19\linewidth]{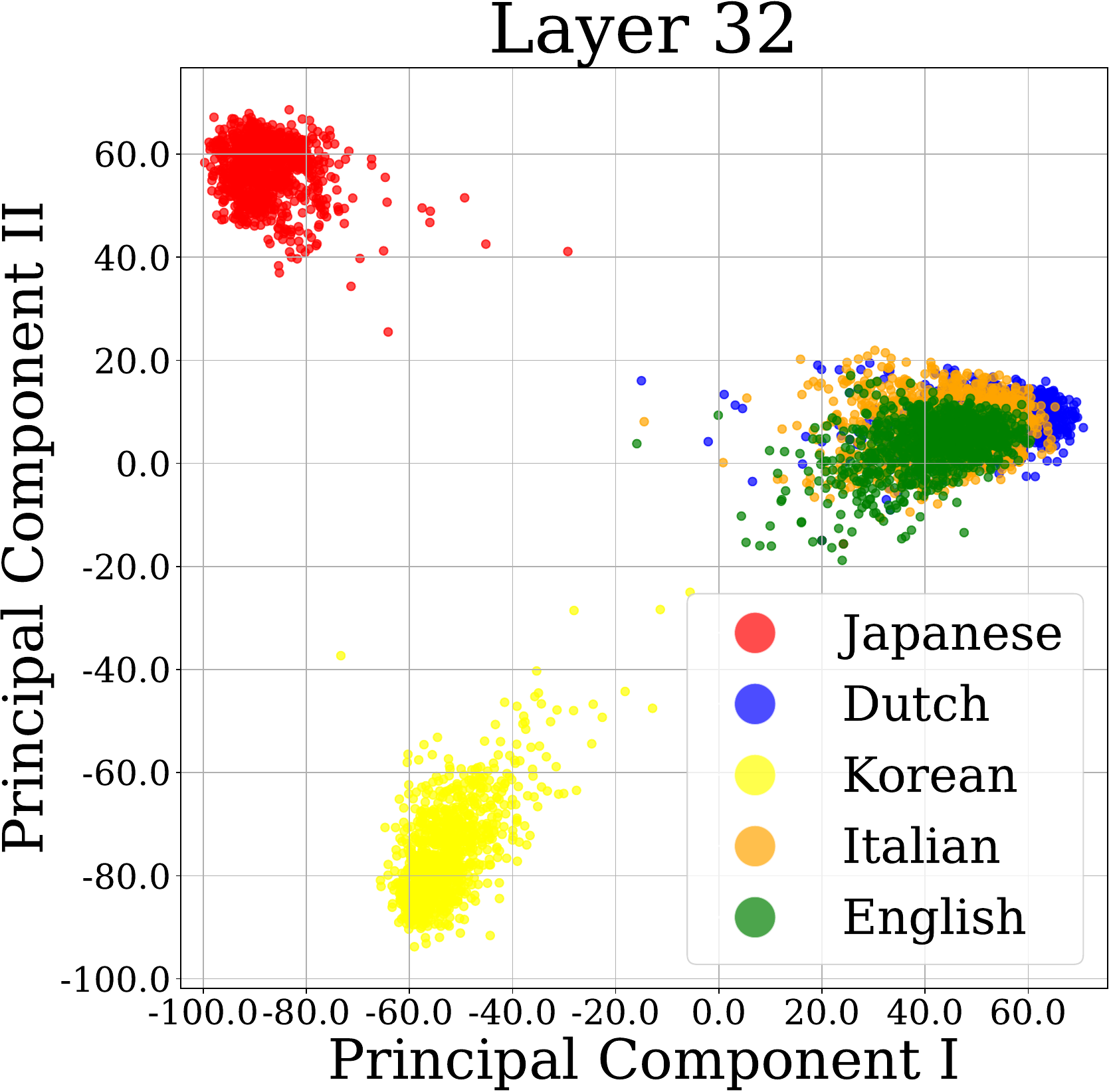}

  \caption{\textbf{The resutls of PCA applied to the hidden representations across layers (LLaMA3-8B).}}
  \label{fig:appendix:pca_all_layers_llama3}
\end{figure*}

\begin{figure*}[t]
  \centering

  \includegraphics[width=0.19\linewidth]{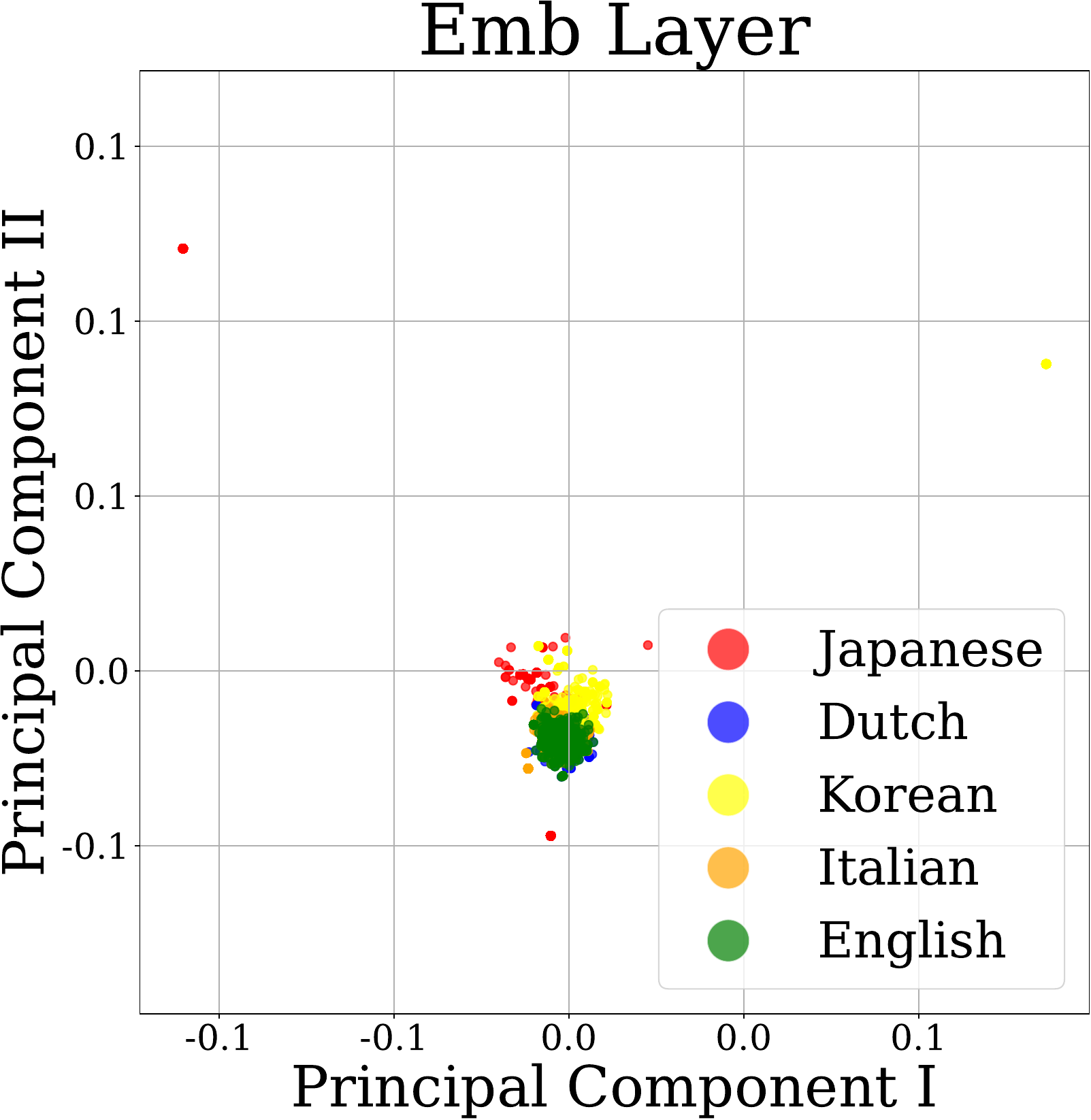}
  \includegraphics[width=0.19\linewidth]{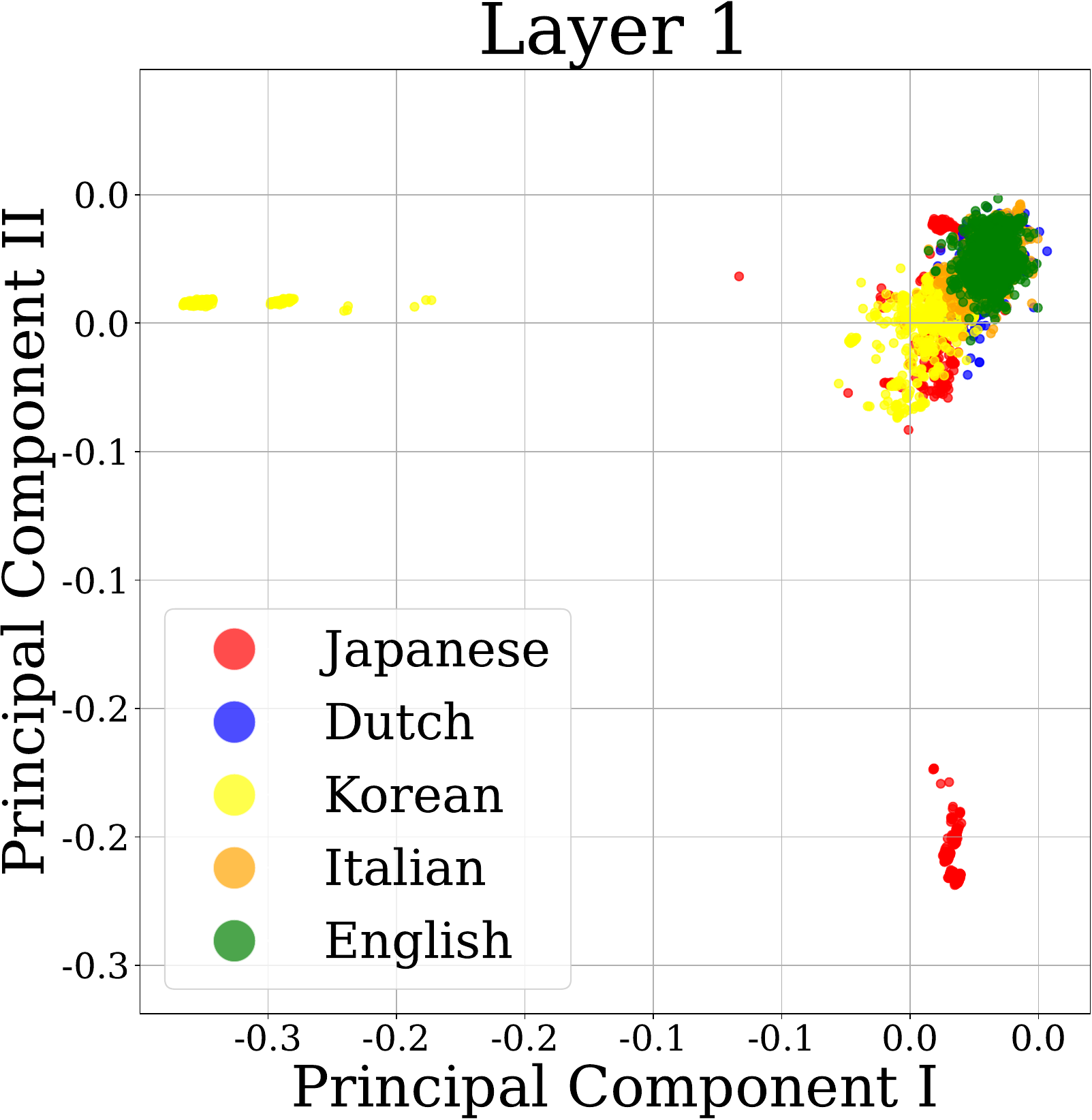}
  \includegraphics[width=0.19\linewidth]{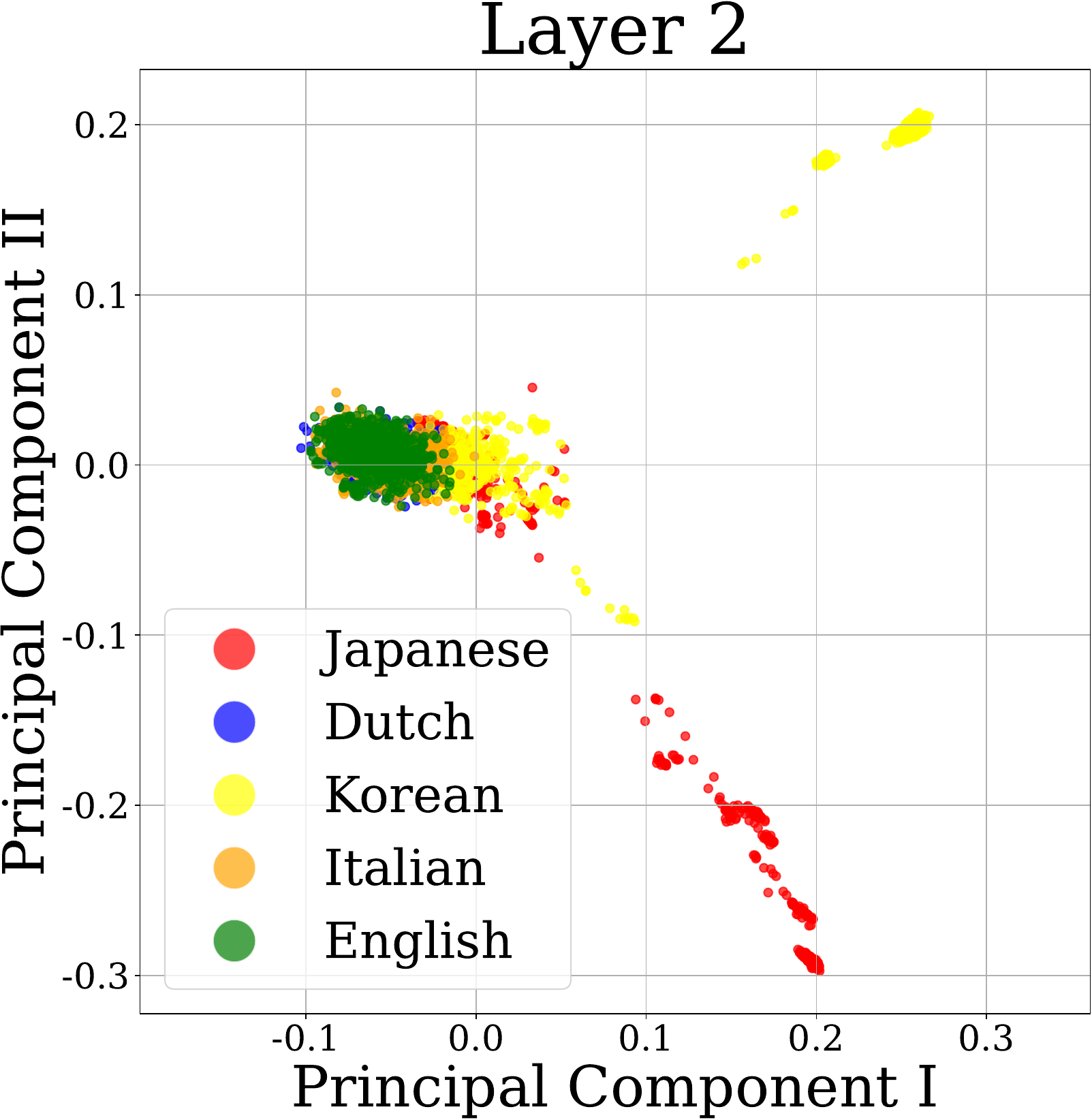}
  \includegraphics[width=0.19\linewidth]{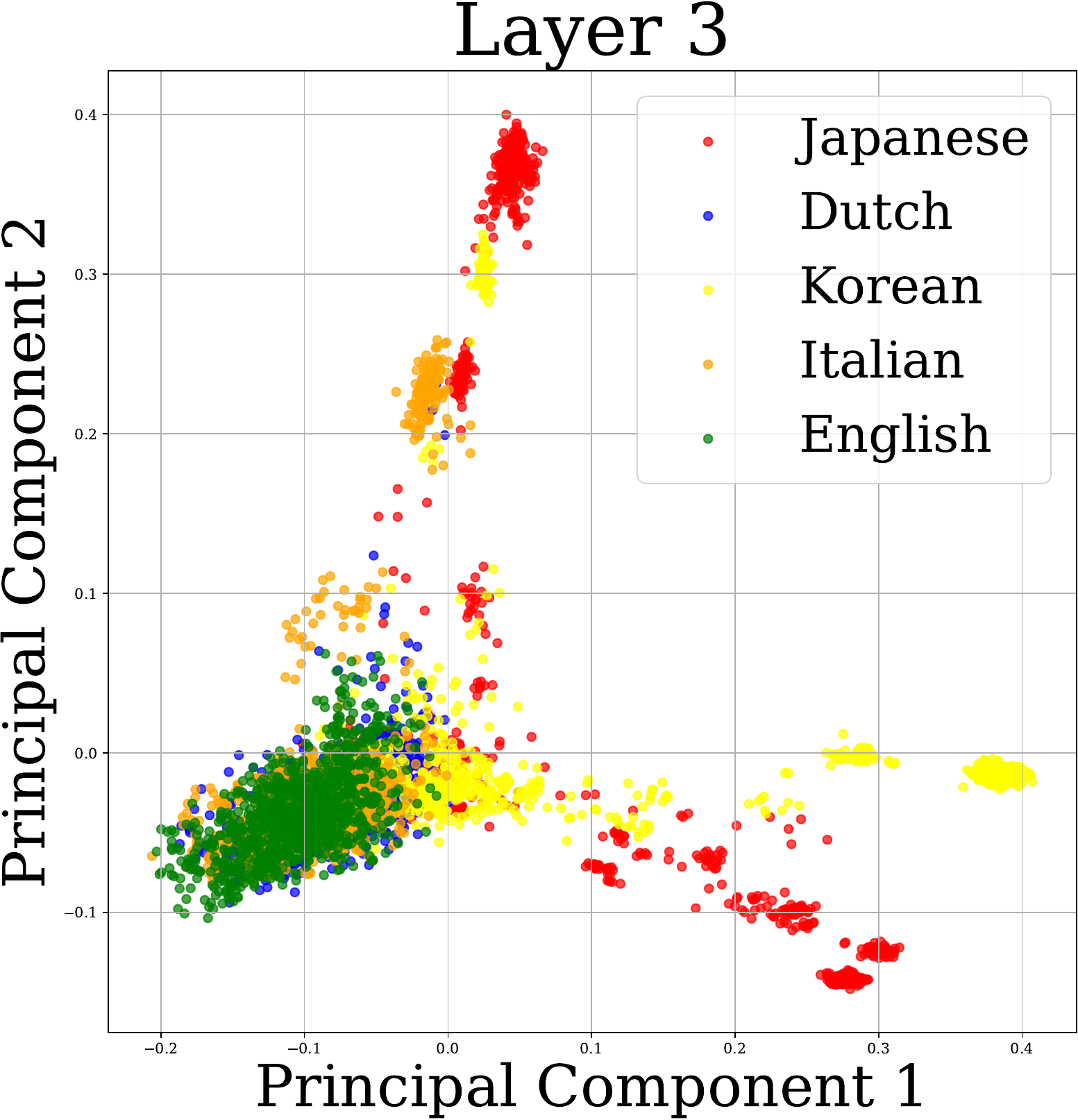}
  \includegraphics[width=0.19\linewidth]{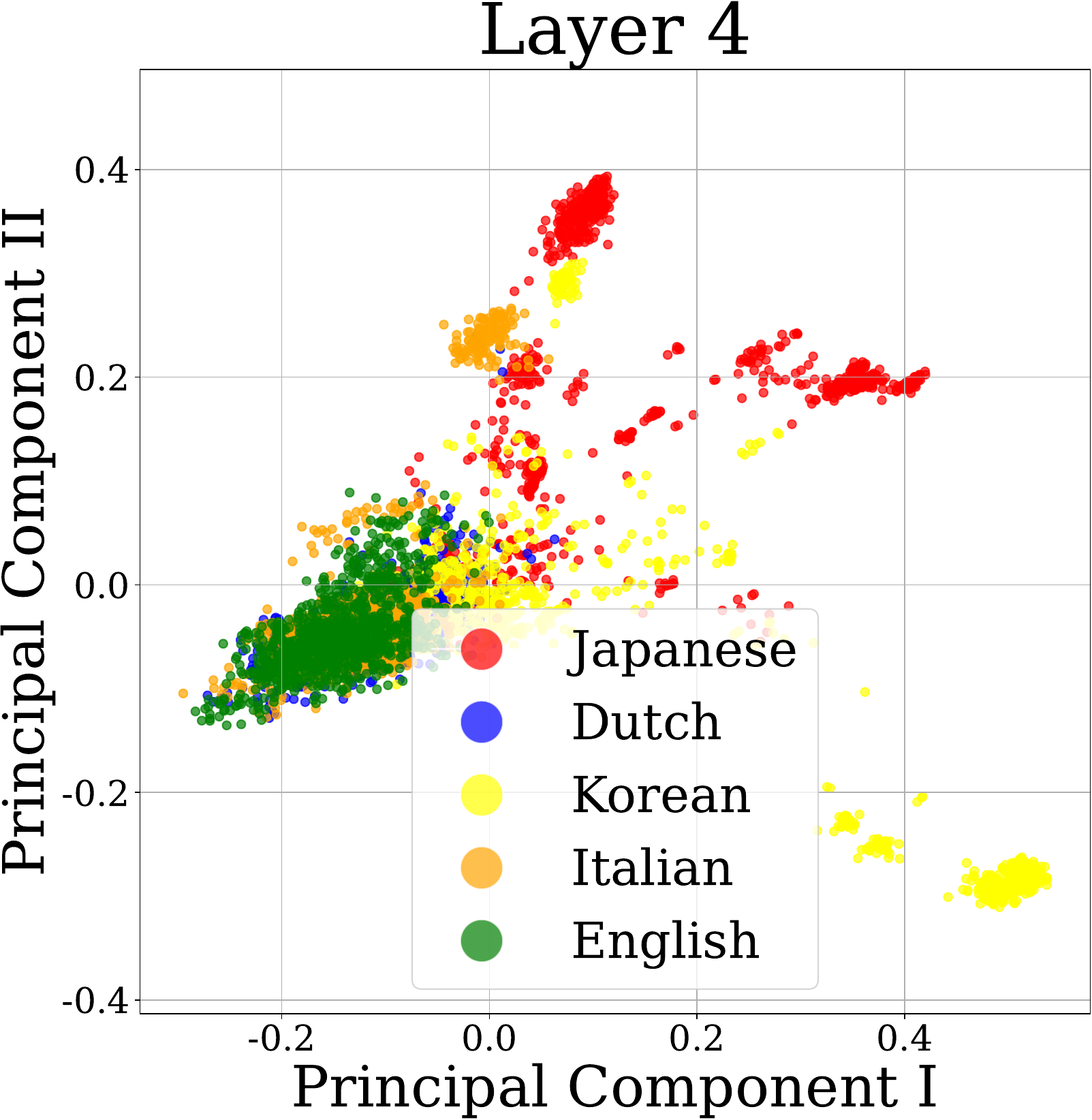}

  \includegraphics[width=0.19\linewidth]{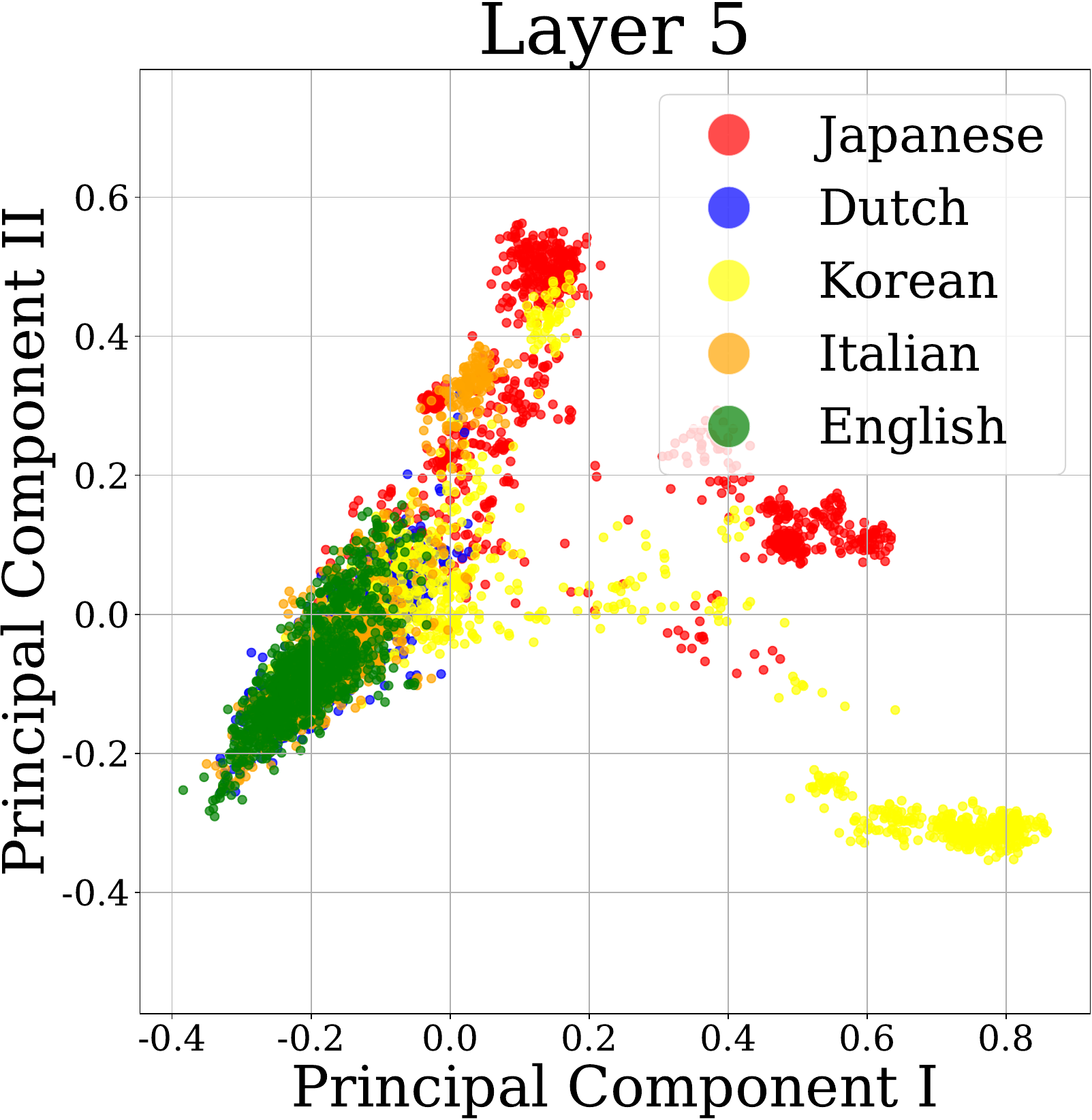}
  \includegraphics[width=0.19\linewidth]{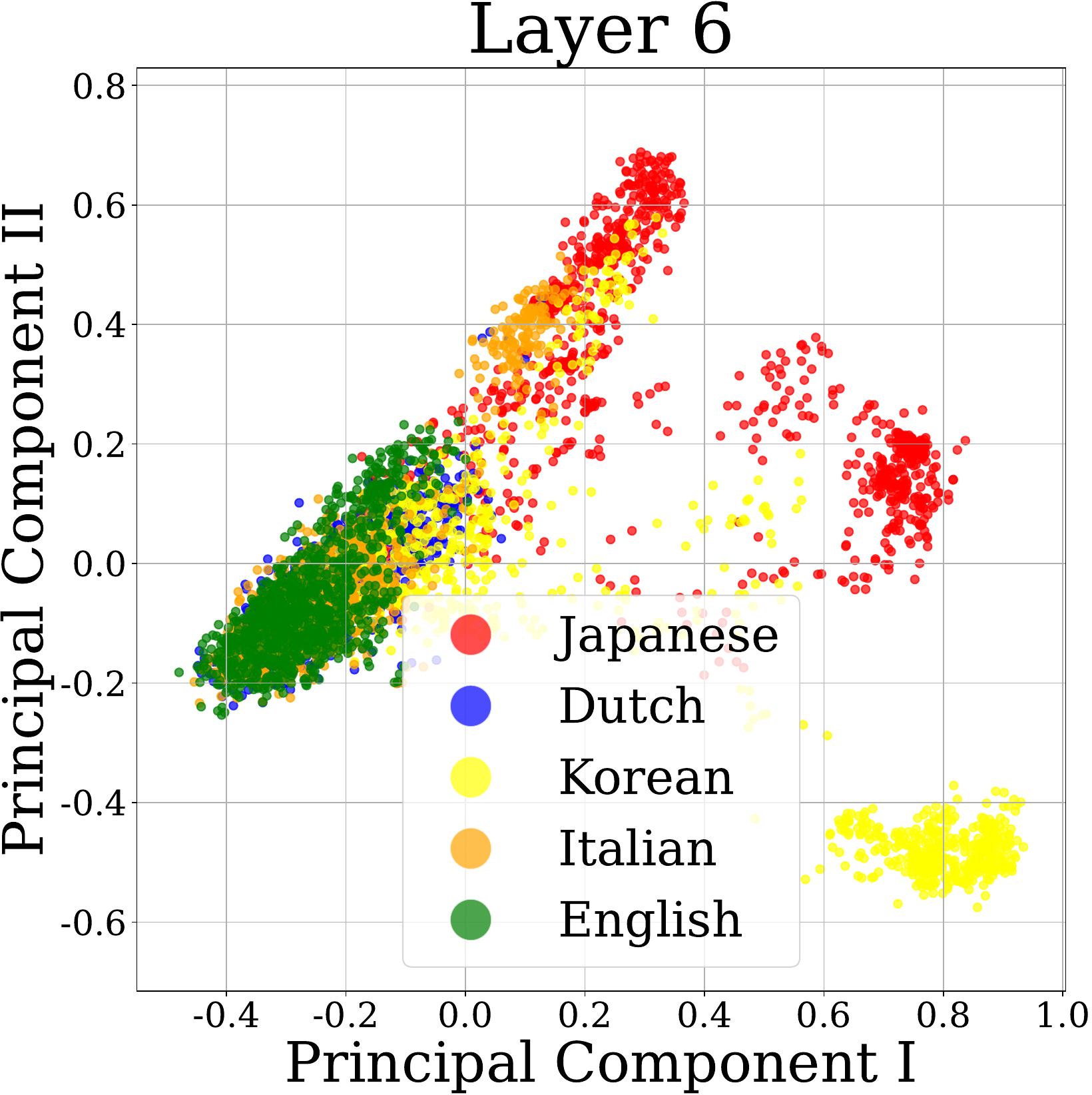}
  \includegraphics[width=0.19\linewidth]{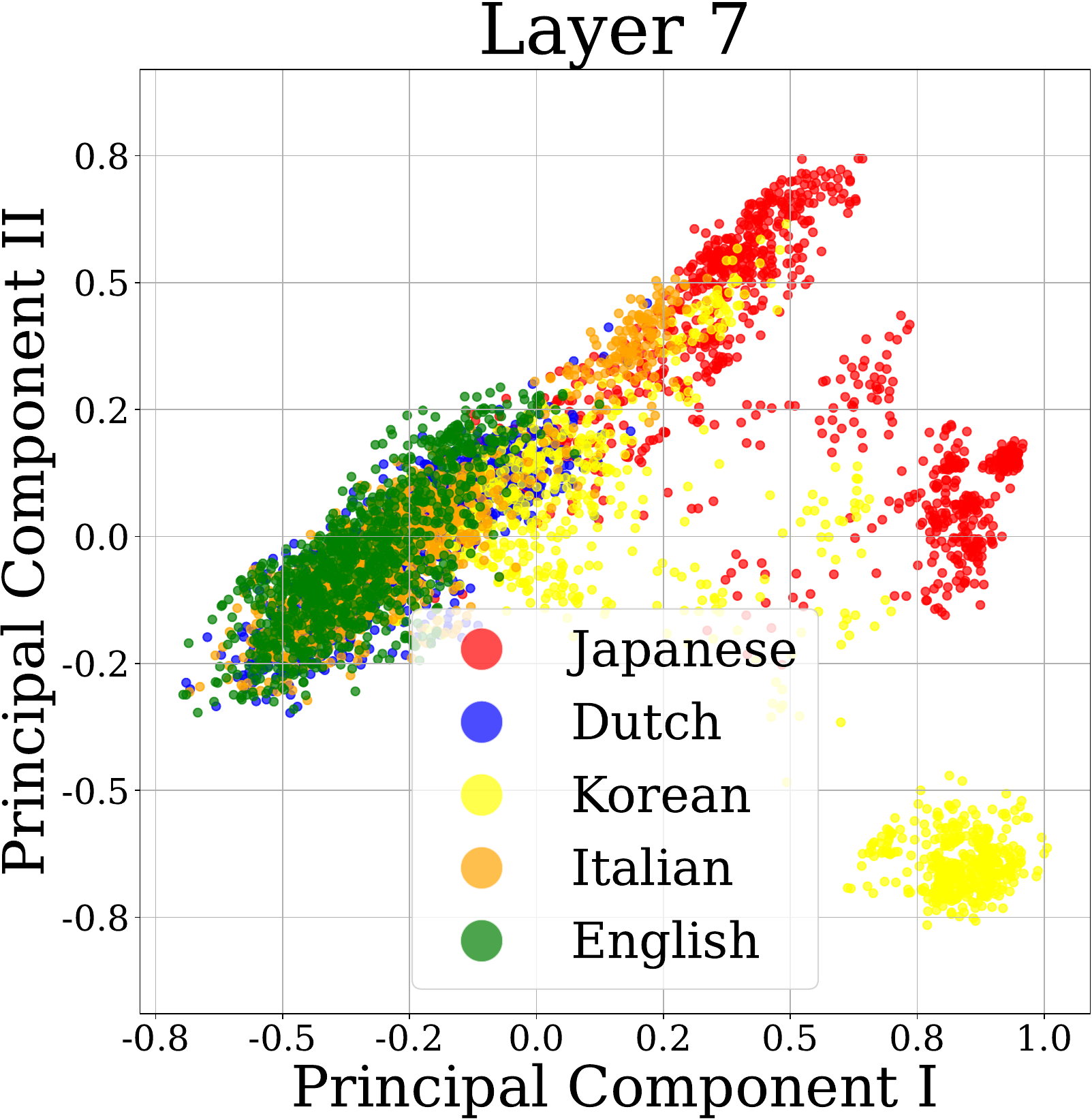}
  \includegraphics[width=0.19\linewidth]{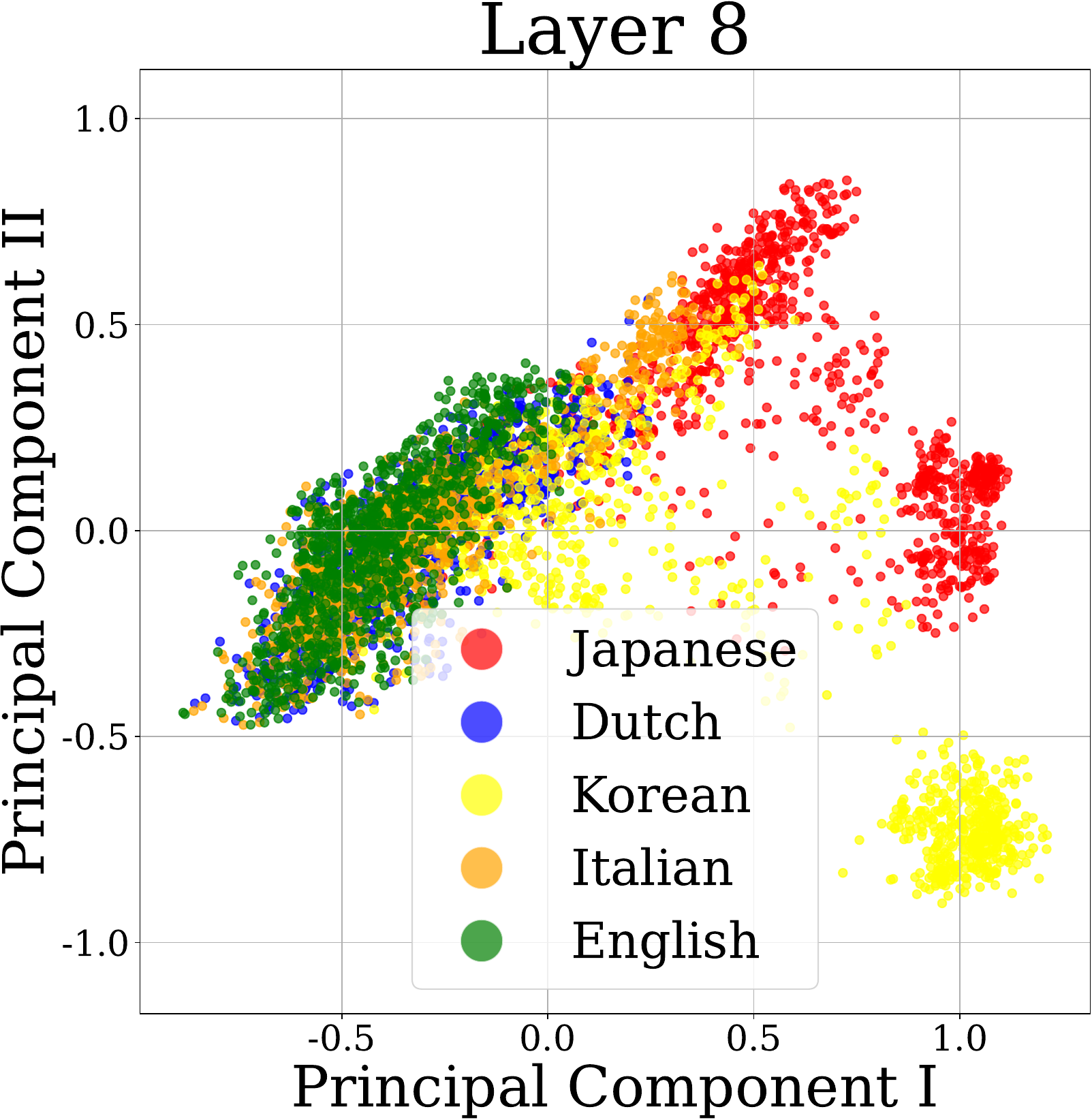}
  \includegraphics[width=0.19\linewidth]{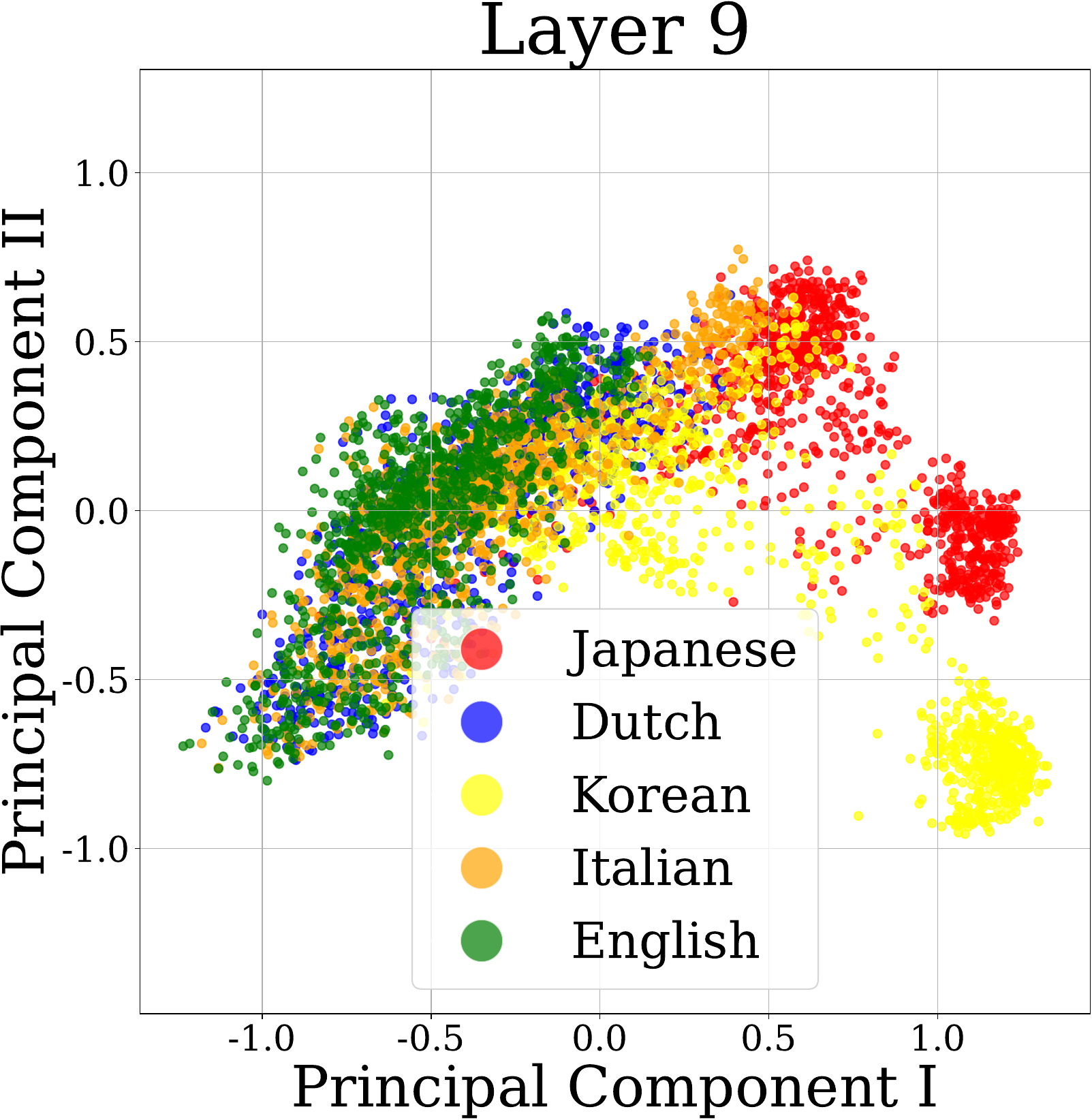}

  \includegraphics[width=0.19\linewidth]{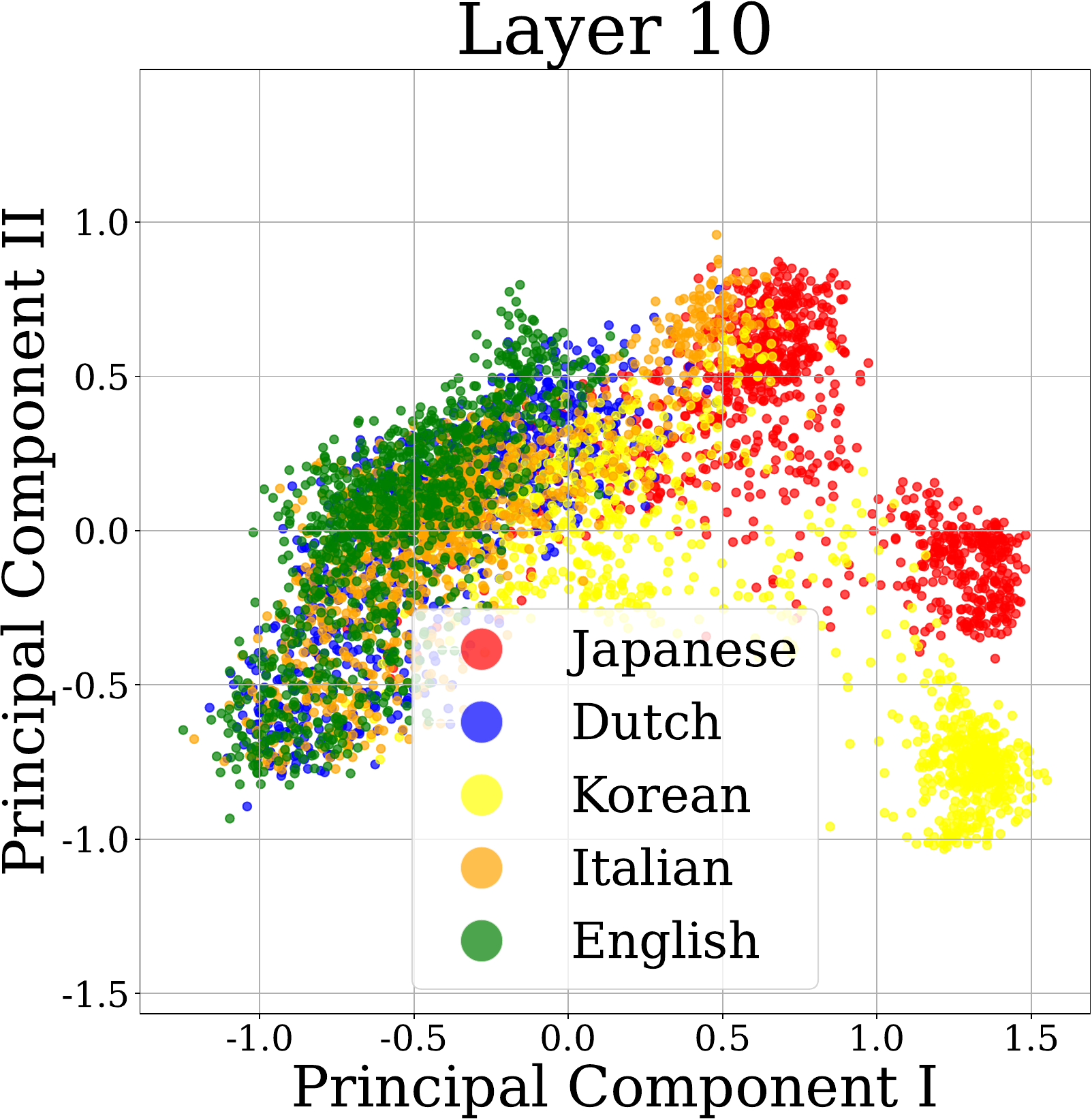}
  \includegraphics[width=0.19\linewidth]{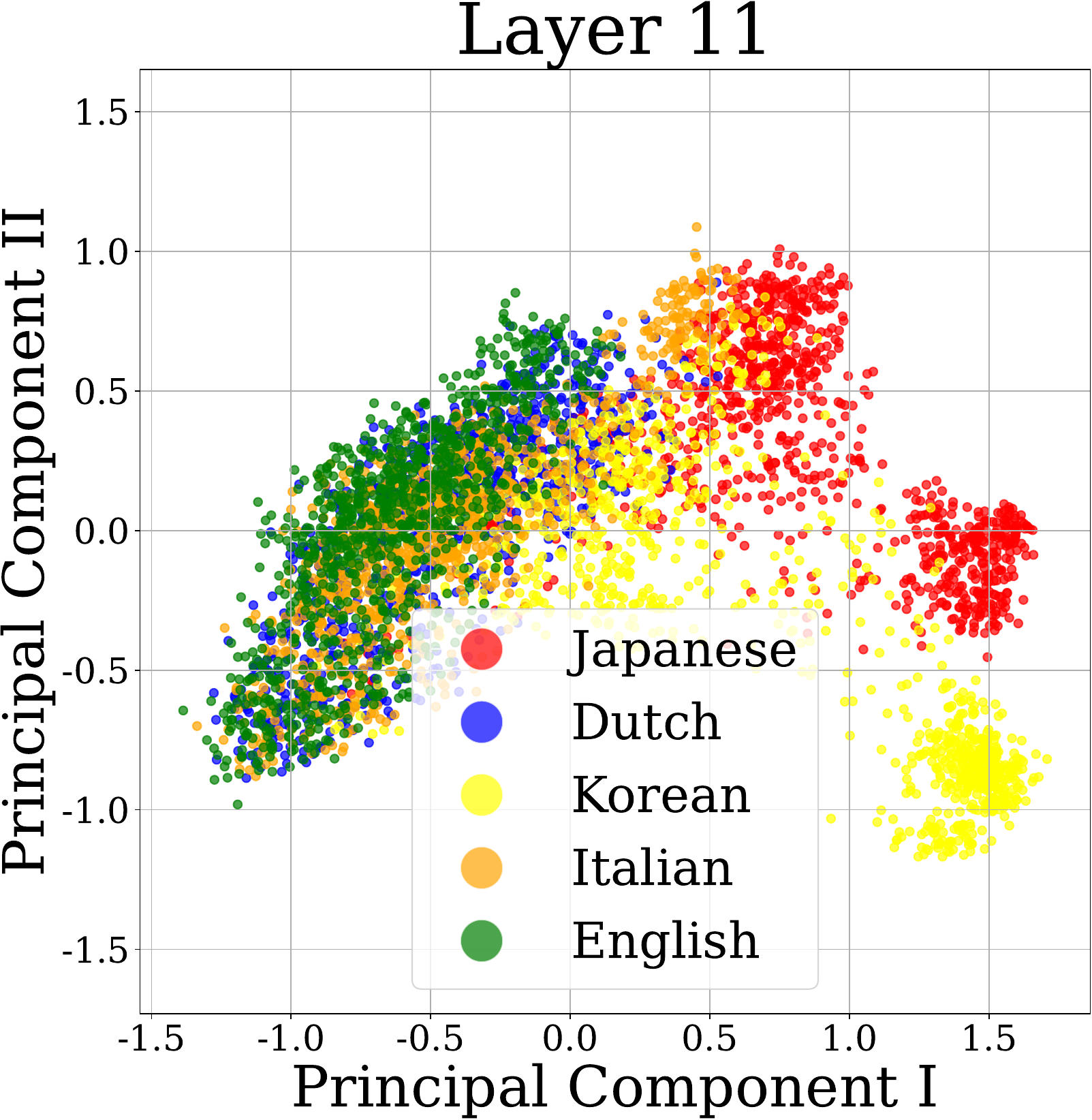}
  \includegraphics[width=0.19\linewidth]{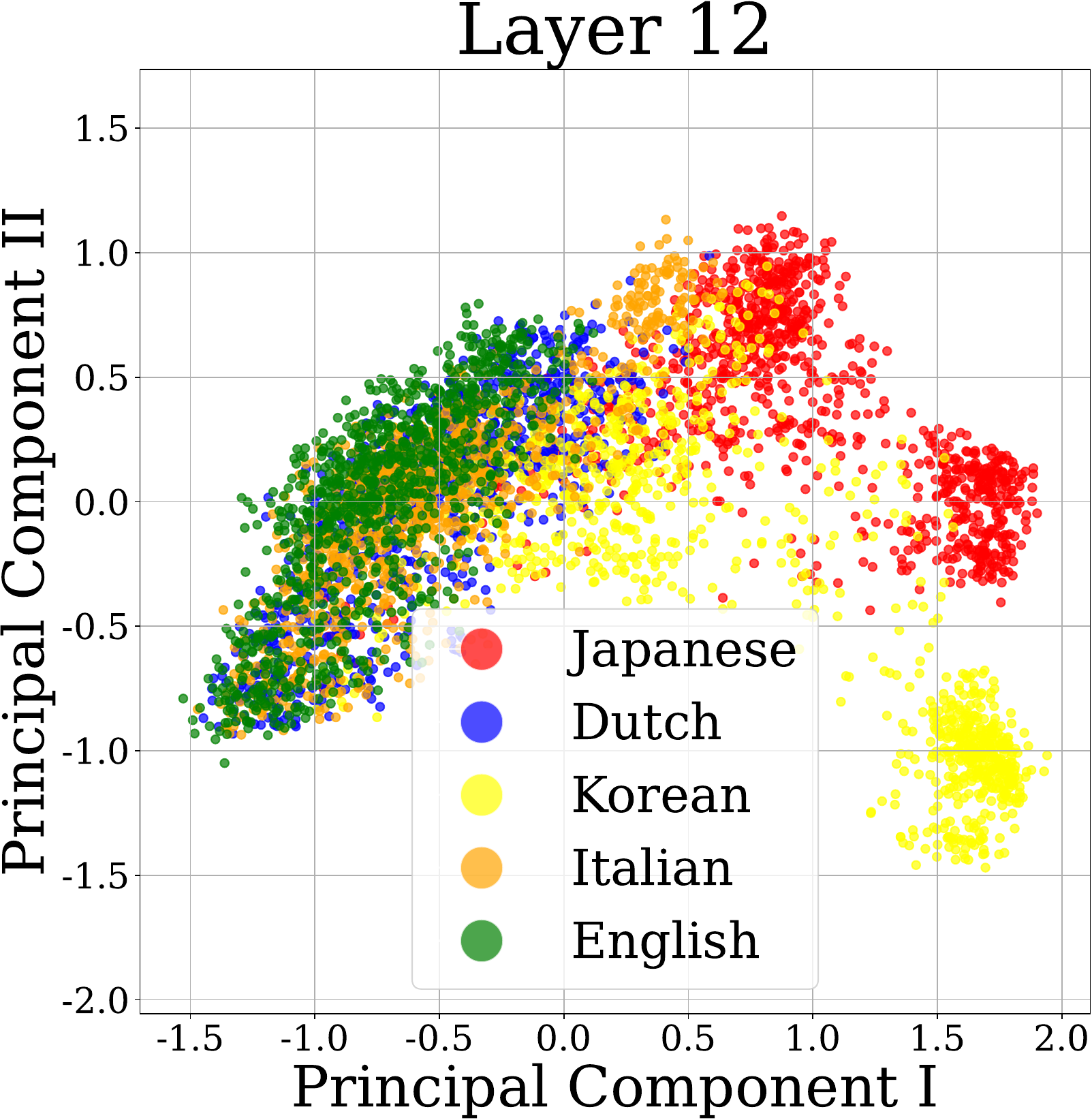}
  \includegraphics[width=0.19\linewidth]{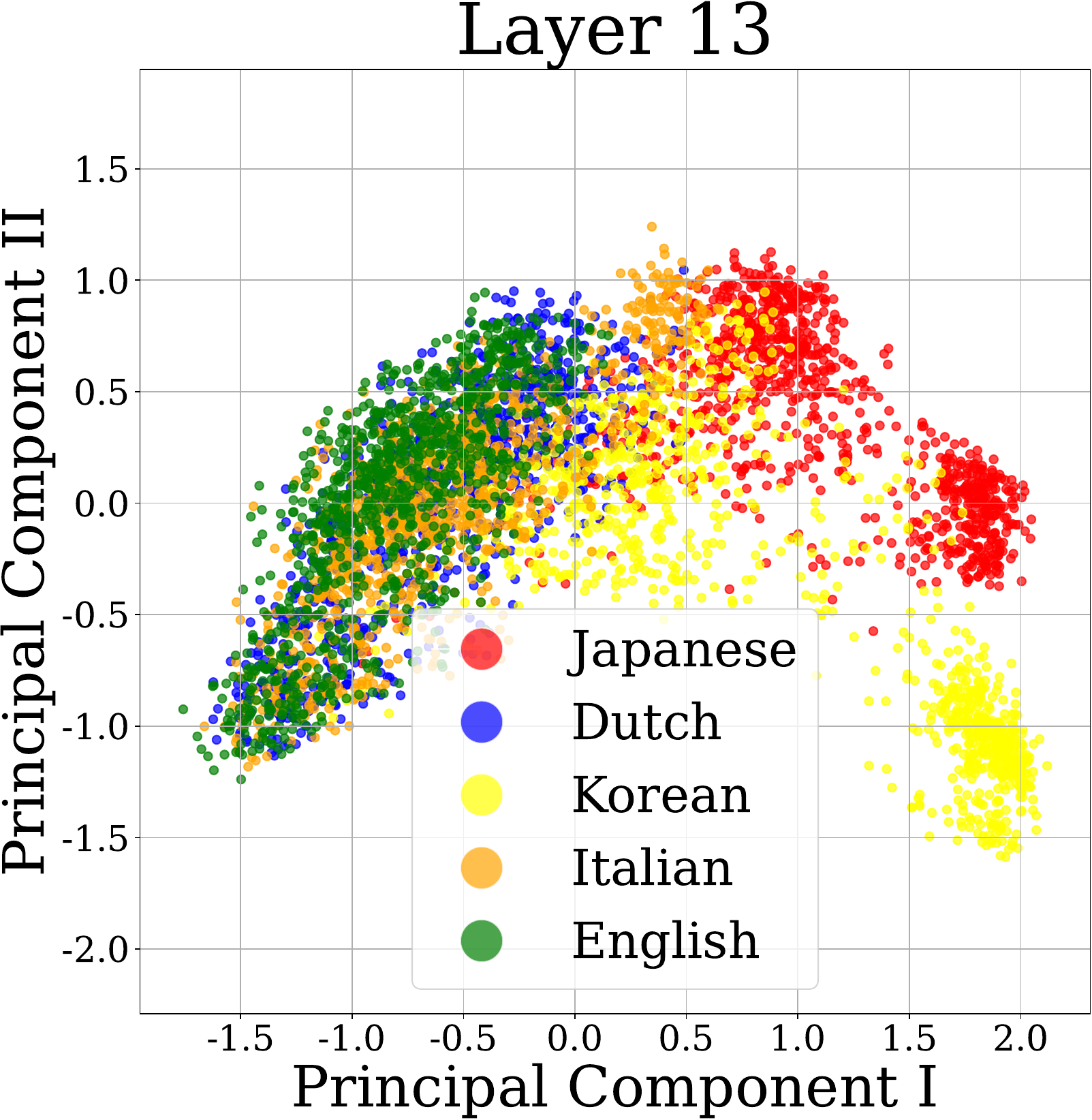}
  \includegraphics[width=0.19\linewidth]{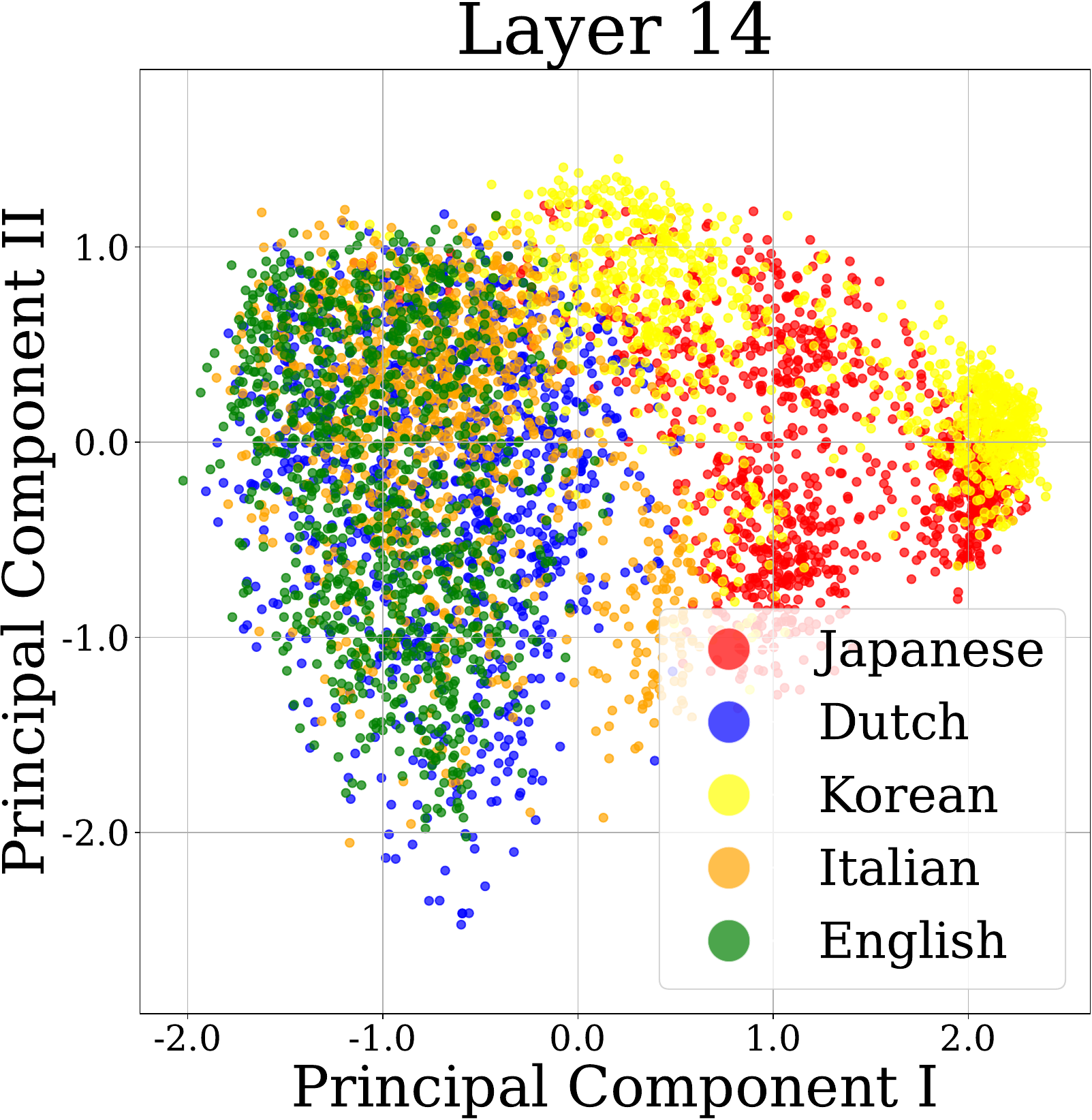}

  \includegraphics[width=0.19\linewidth]{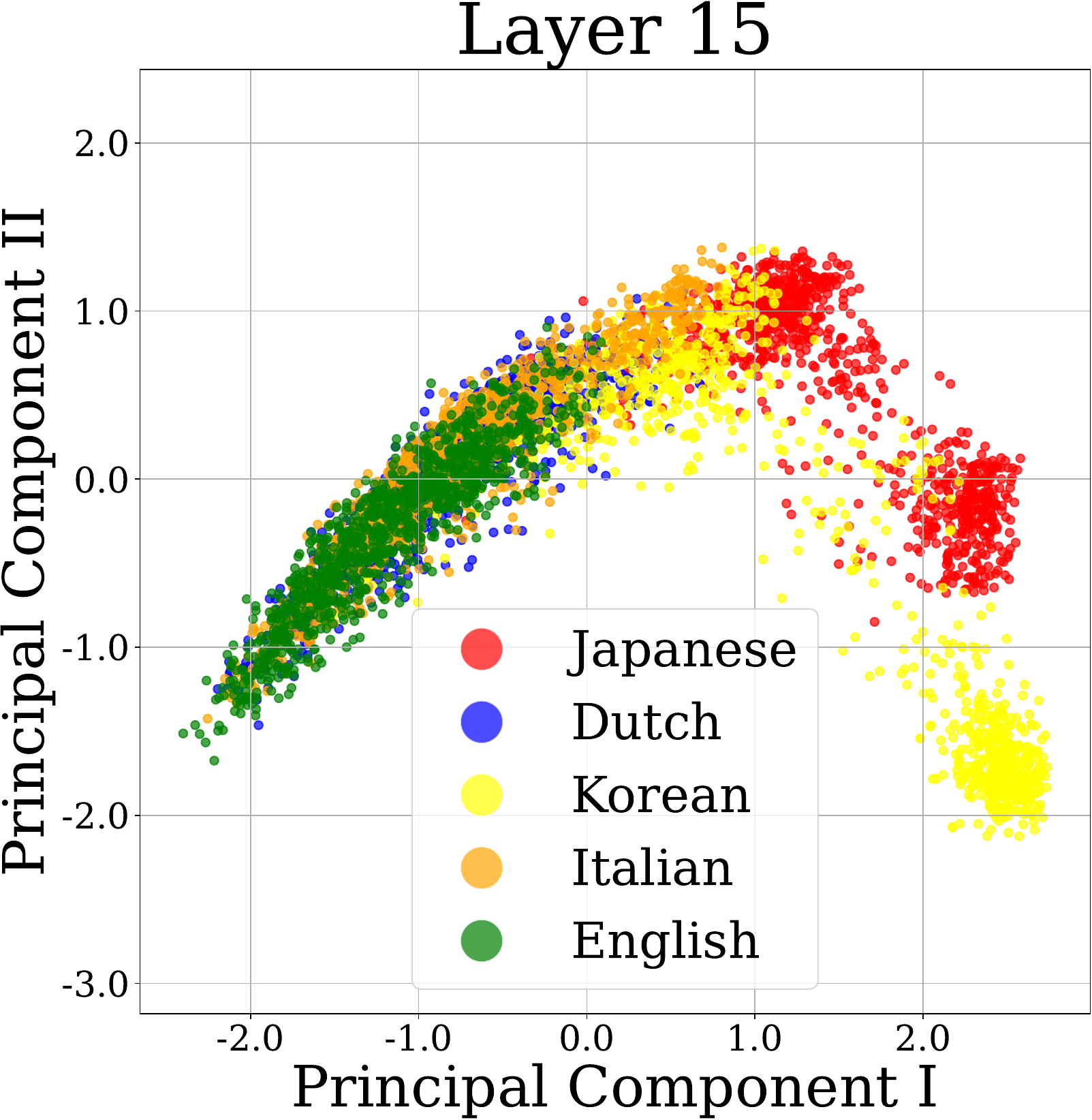}
  \includegraphics[width=0.19\linewidth]{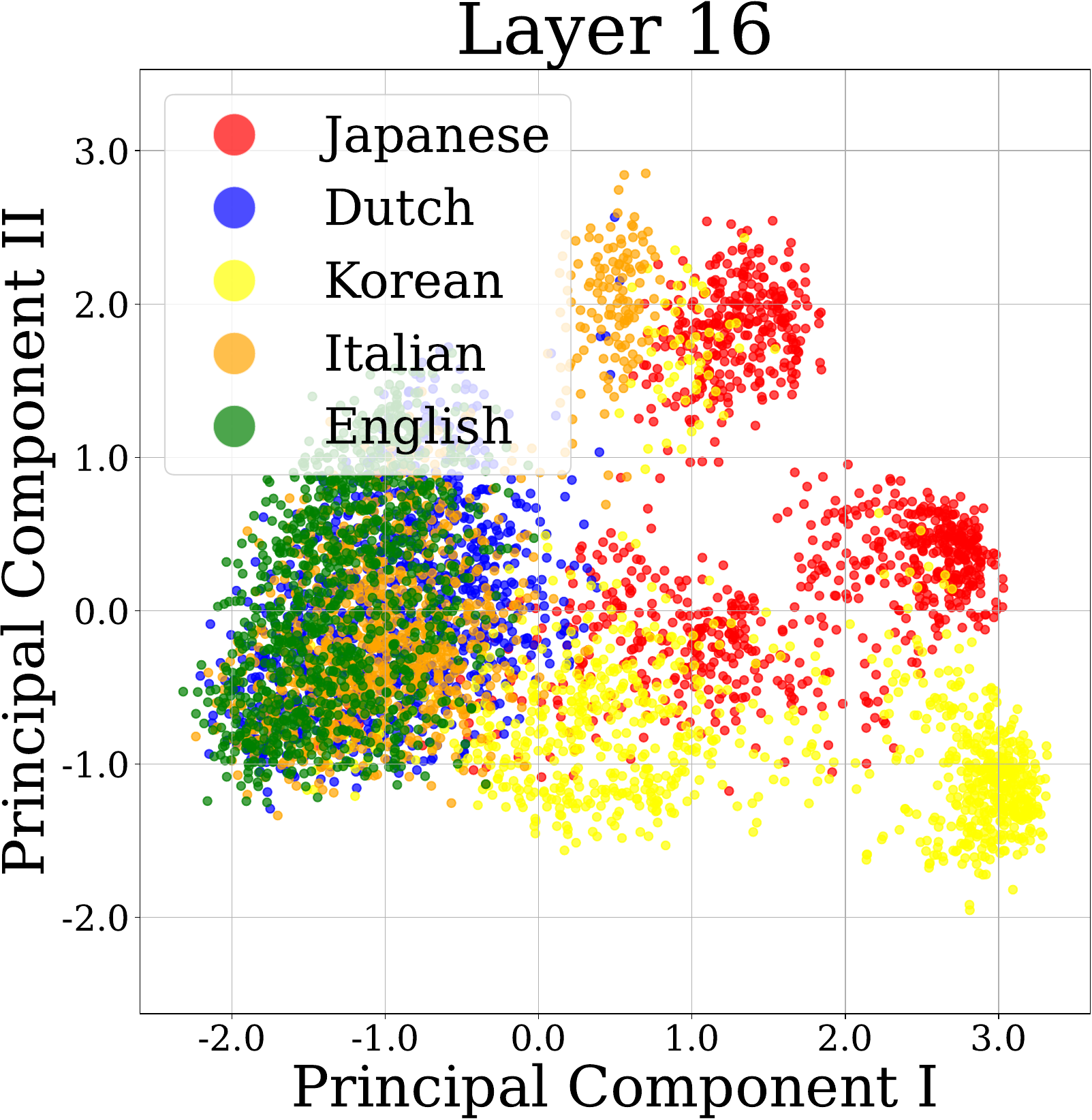}
  \includegraphics[width=0.19\linewidth]{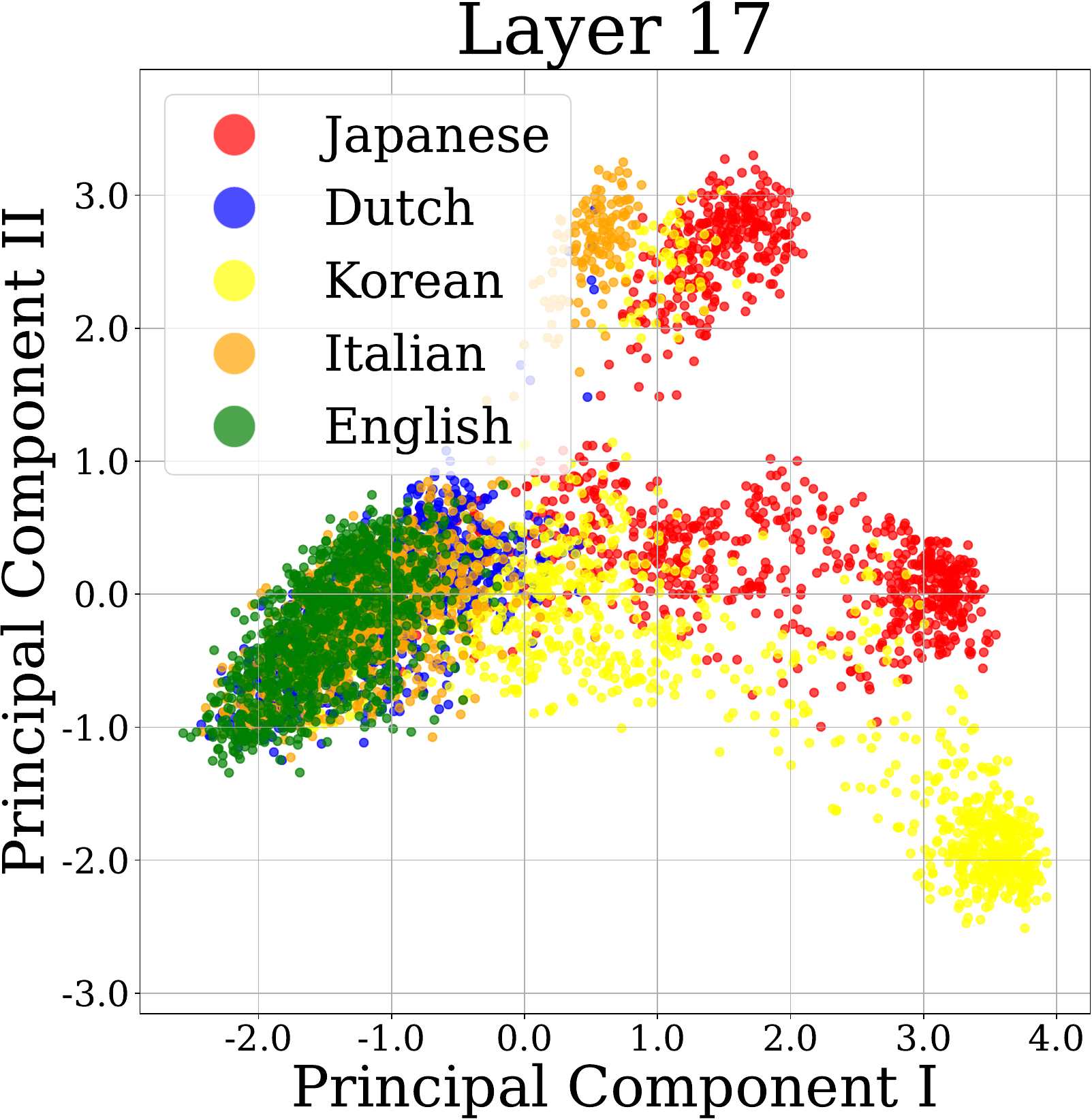}
  \includegraphics[width=0.19\linewidth]{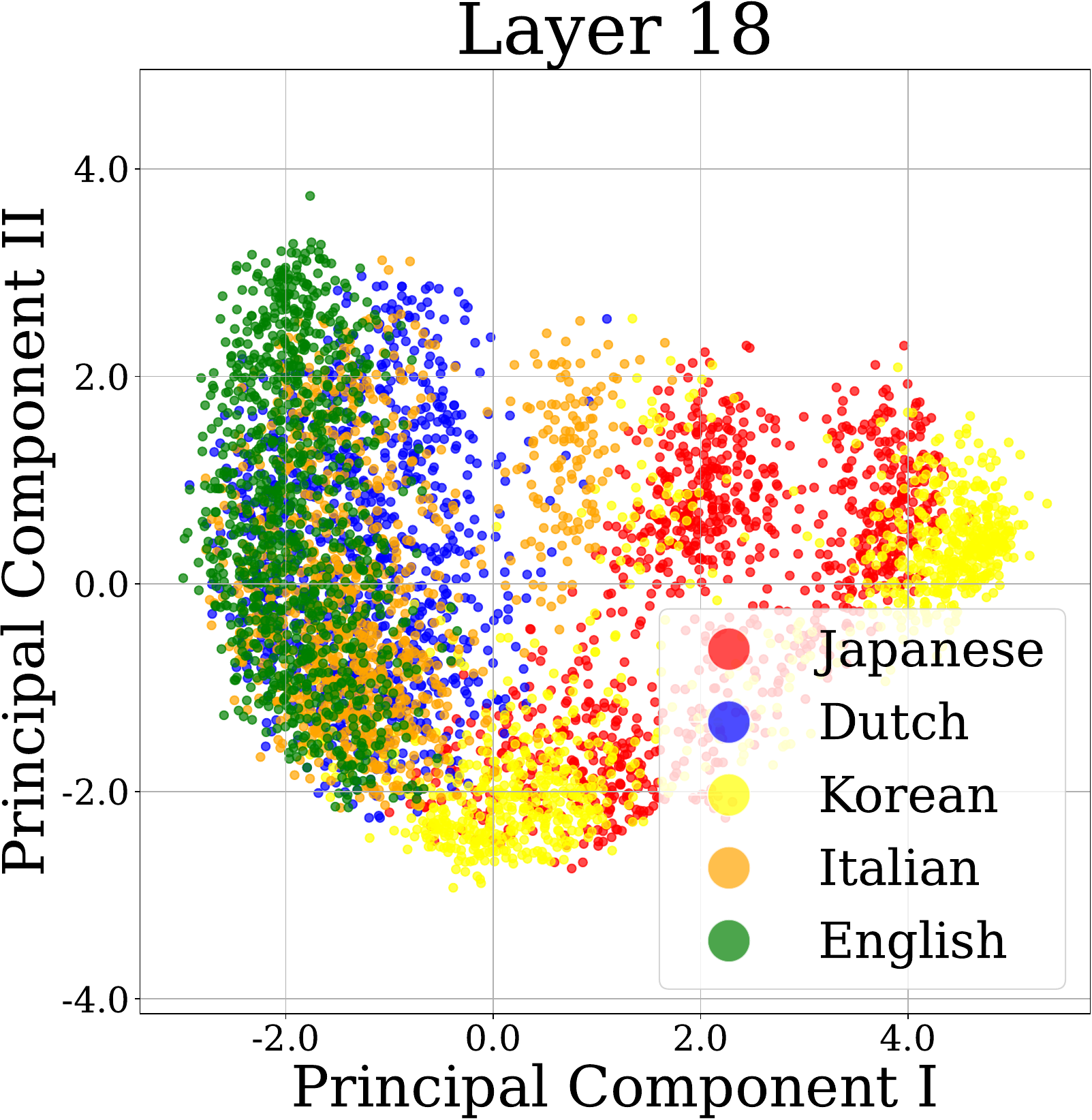}
  \includegraphics[width=0.19\linewidth]{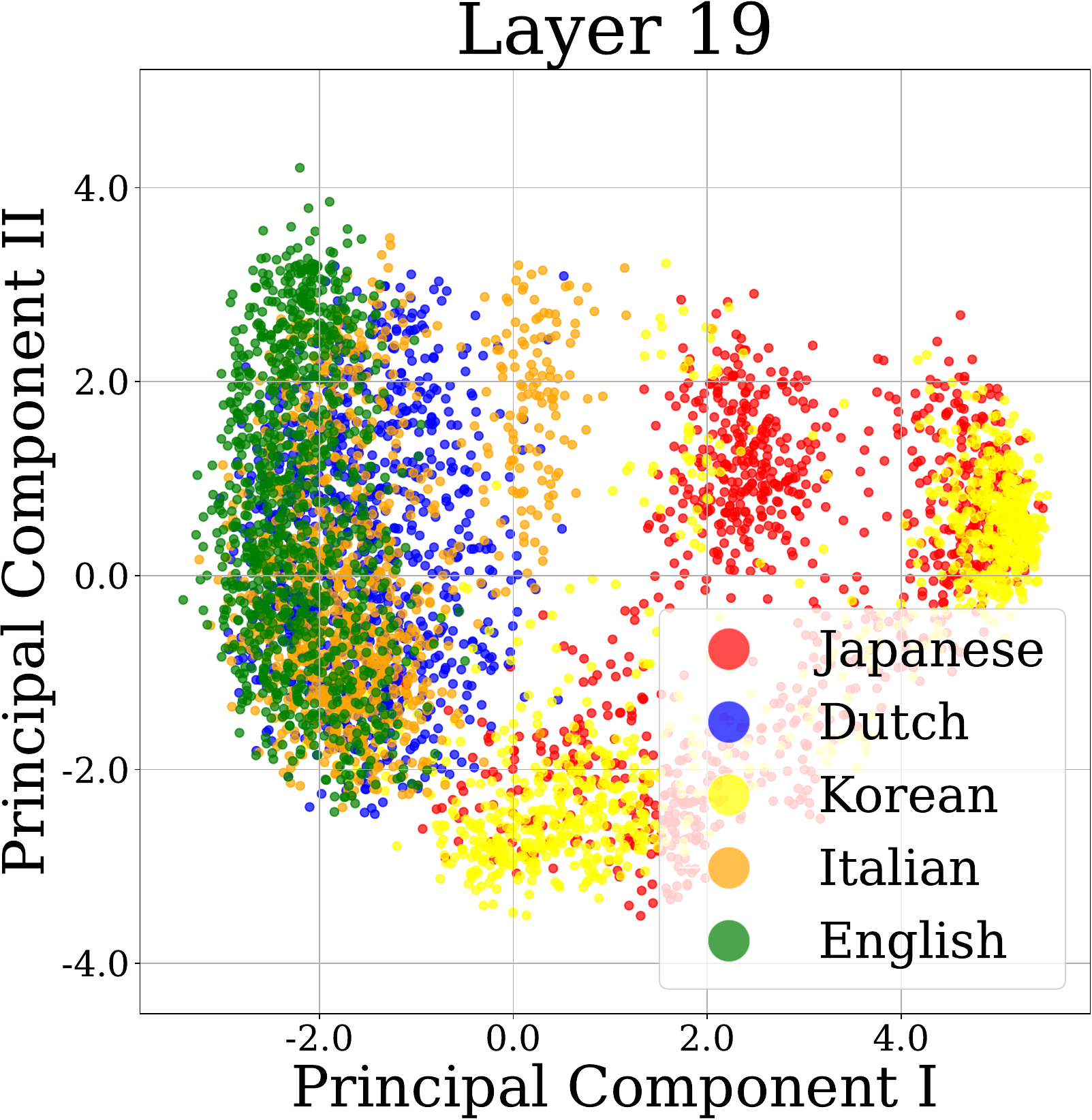}

  \includegraphics[width=0.19\linewidth]{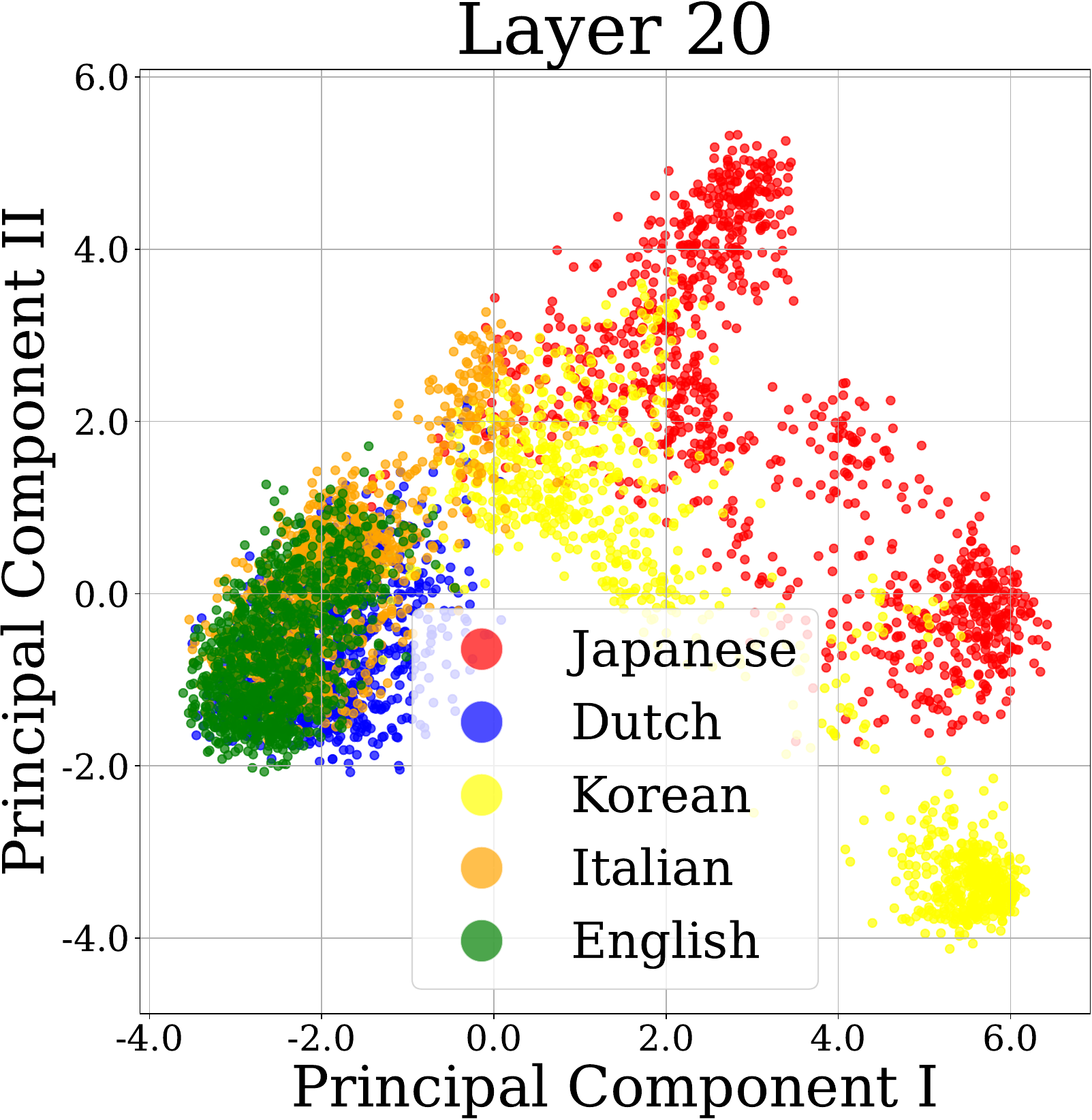}
  \includegraphics[width=0.19\linewidth]{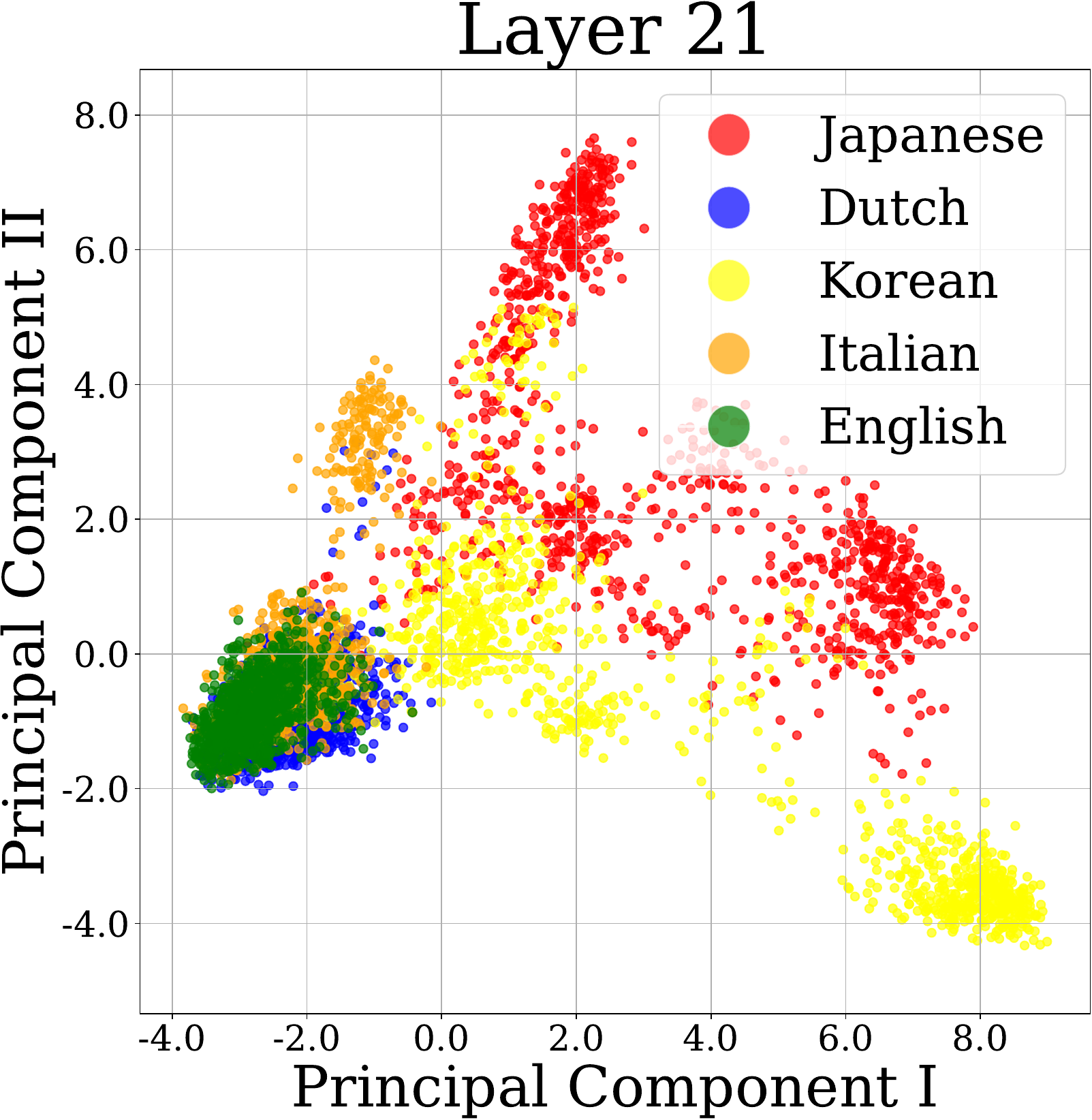}
  \includegraphics[width=0.19\linewidth]{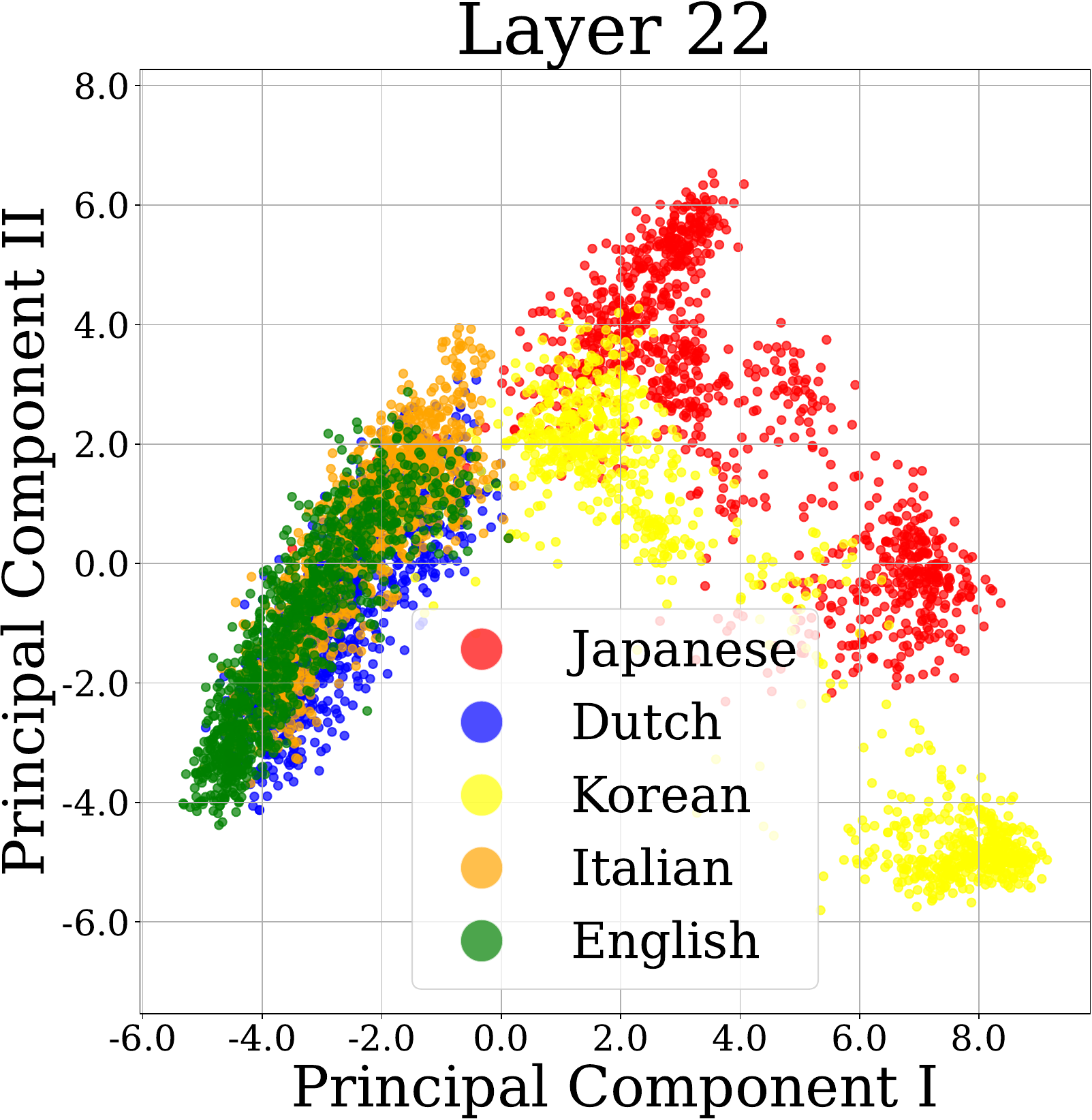}
  \includegraphics[width=0.19\linewidth]{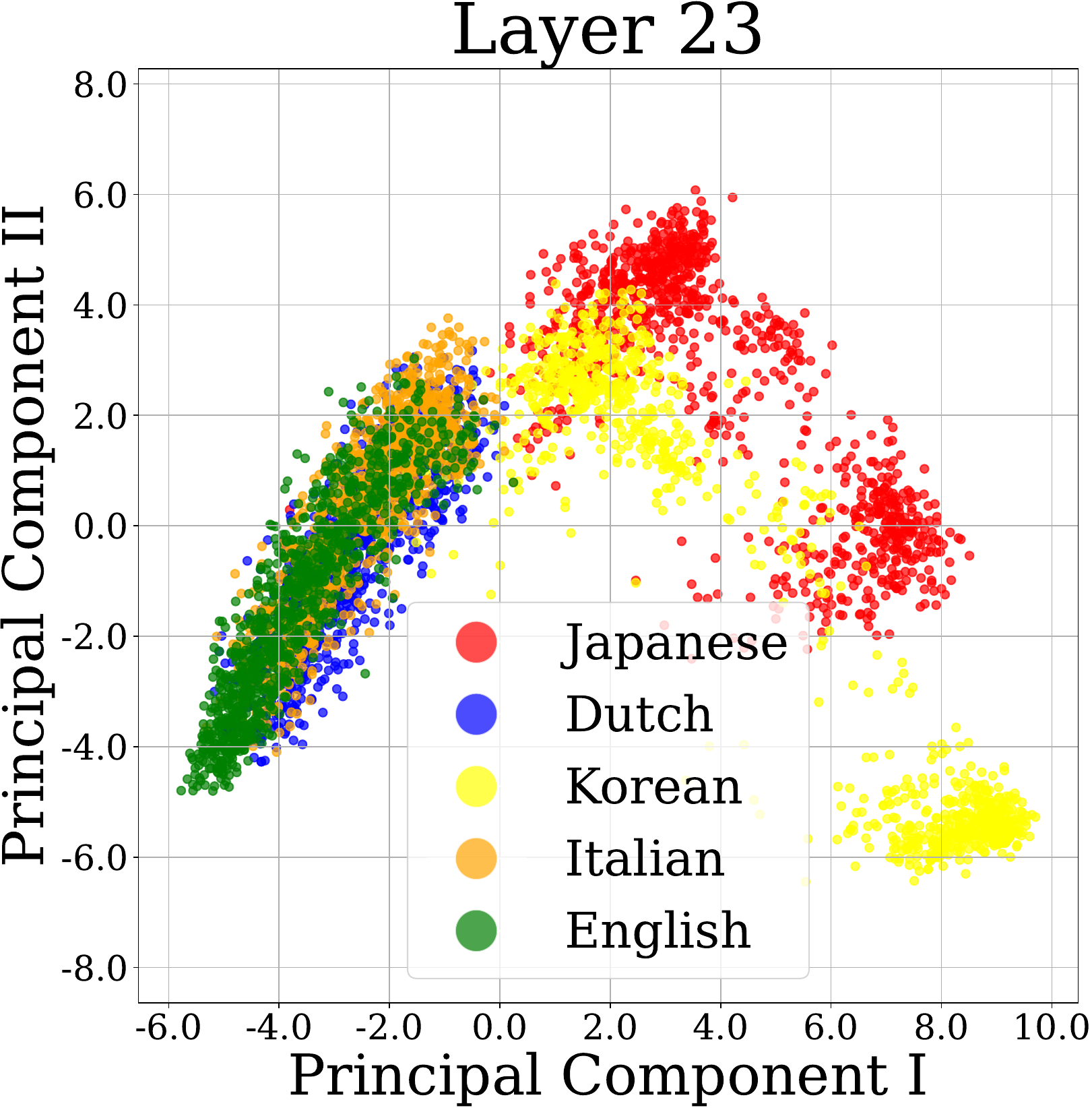}
  \includegraphics[width=0.19\linewidth]{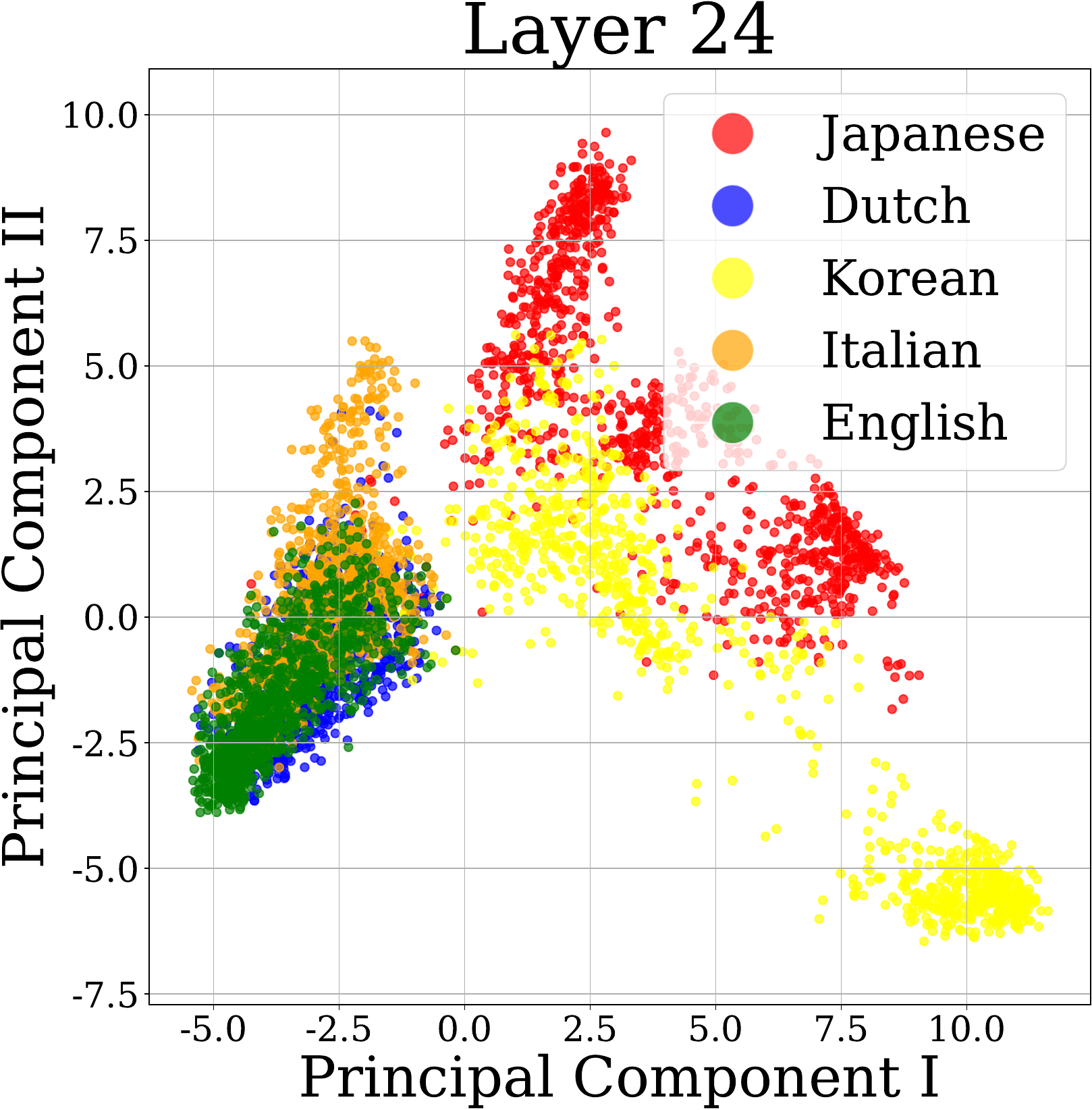}

  \includegraphics[width=0.19\linewidth]{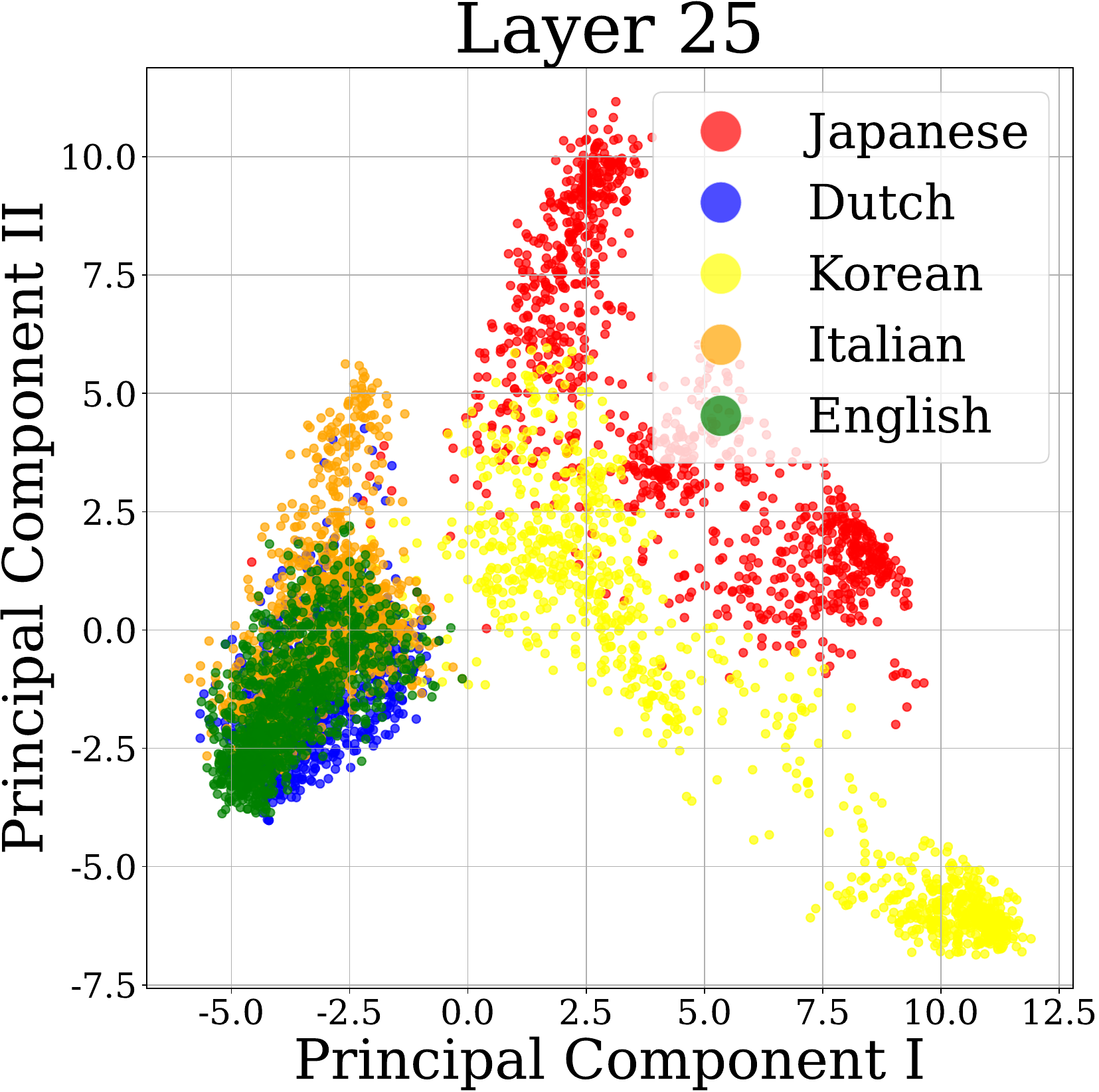}
  \includegraphics[width=0.19\linewidth]{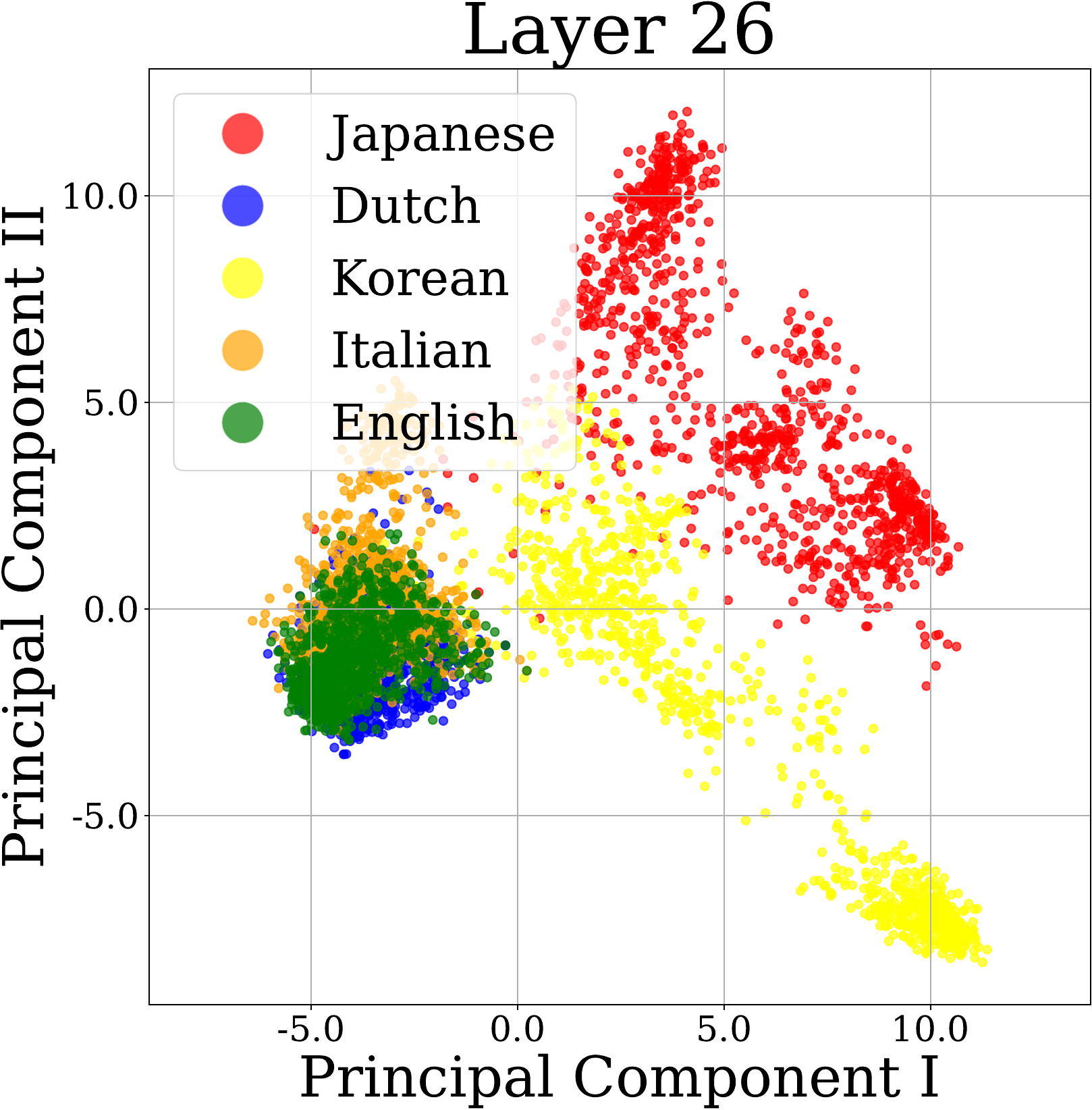}
  \includegraphics[width=0.19\linewidth]{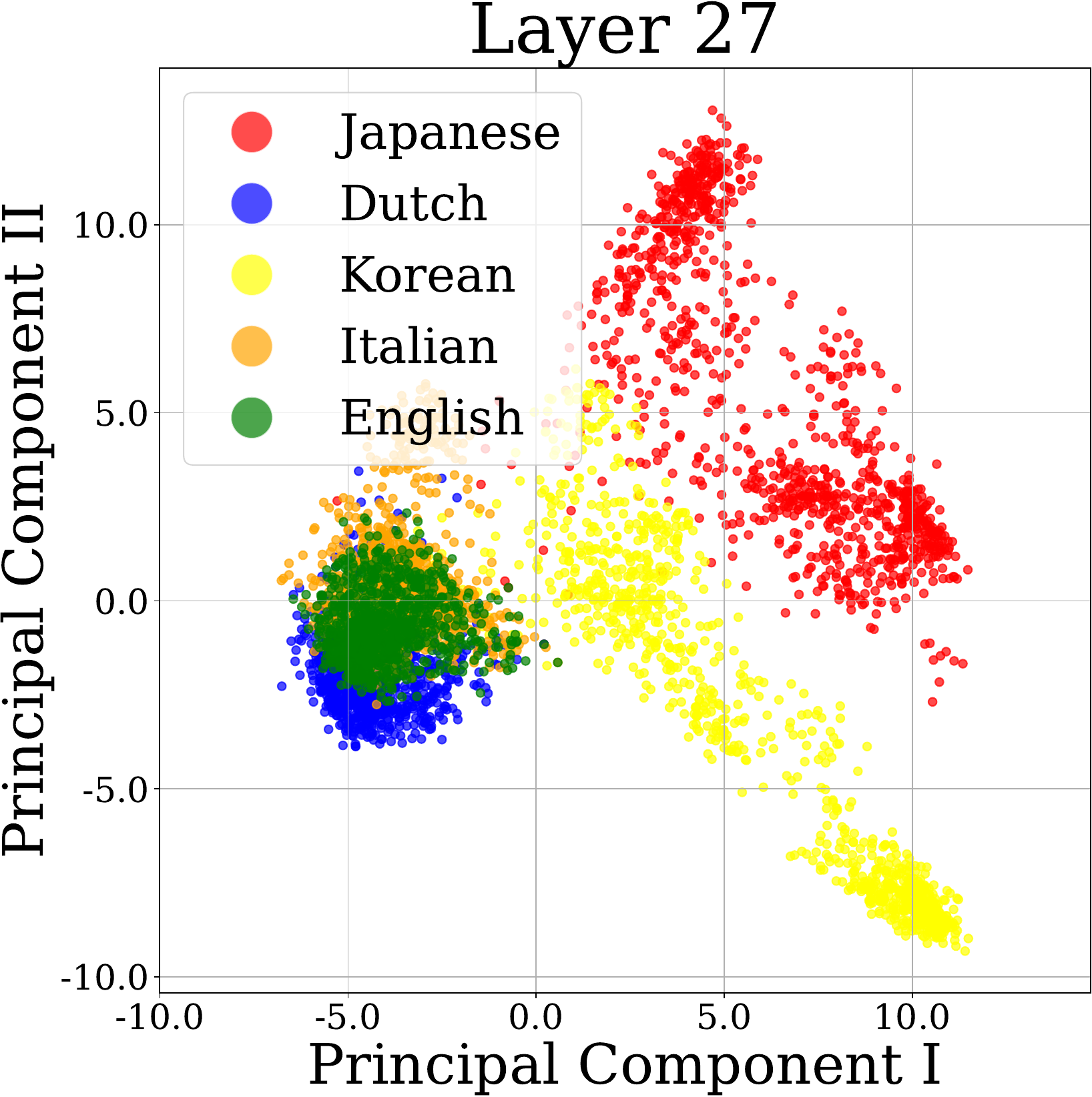}
  \includegraphics[width=0.19\linewidth]{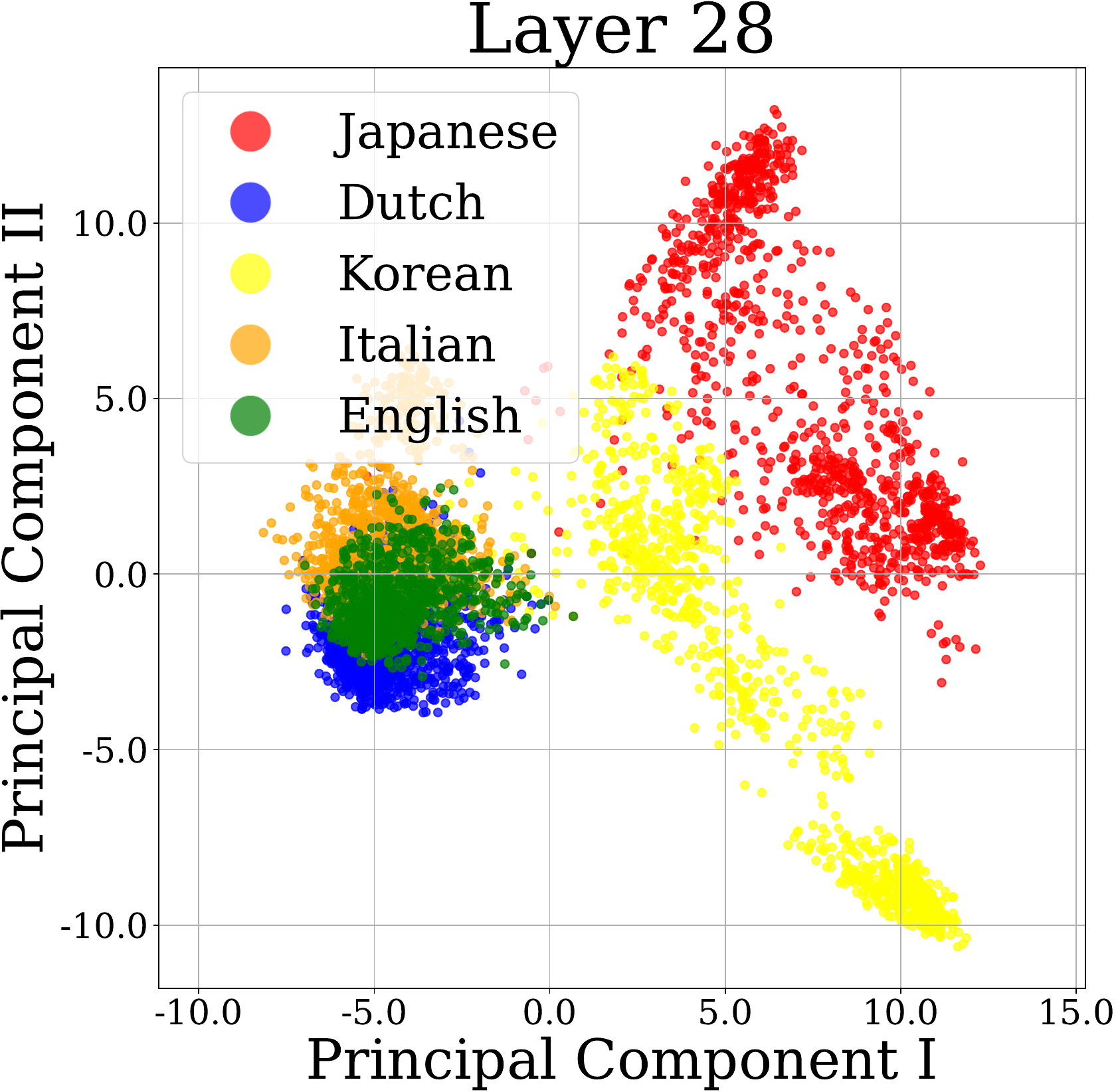}
  \includegraphics[width=0.19\linewidth]{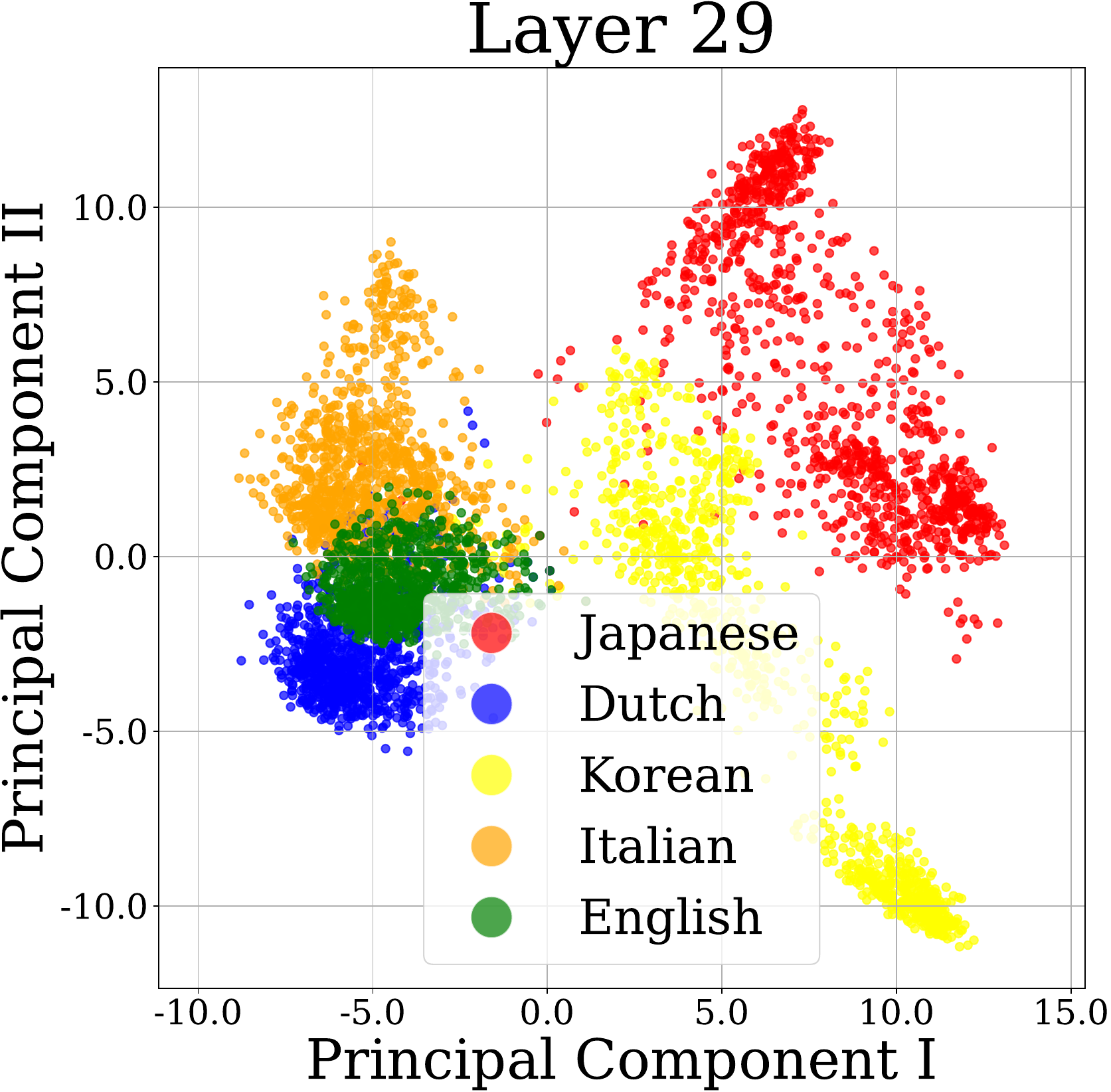}

  \includegraphics[width=0.19\linewidth]{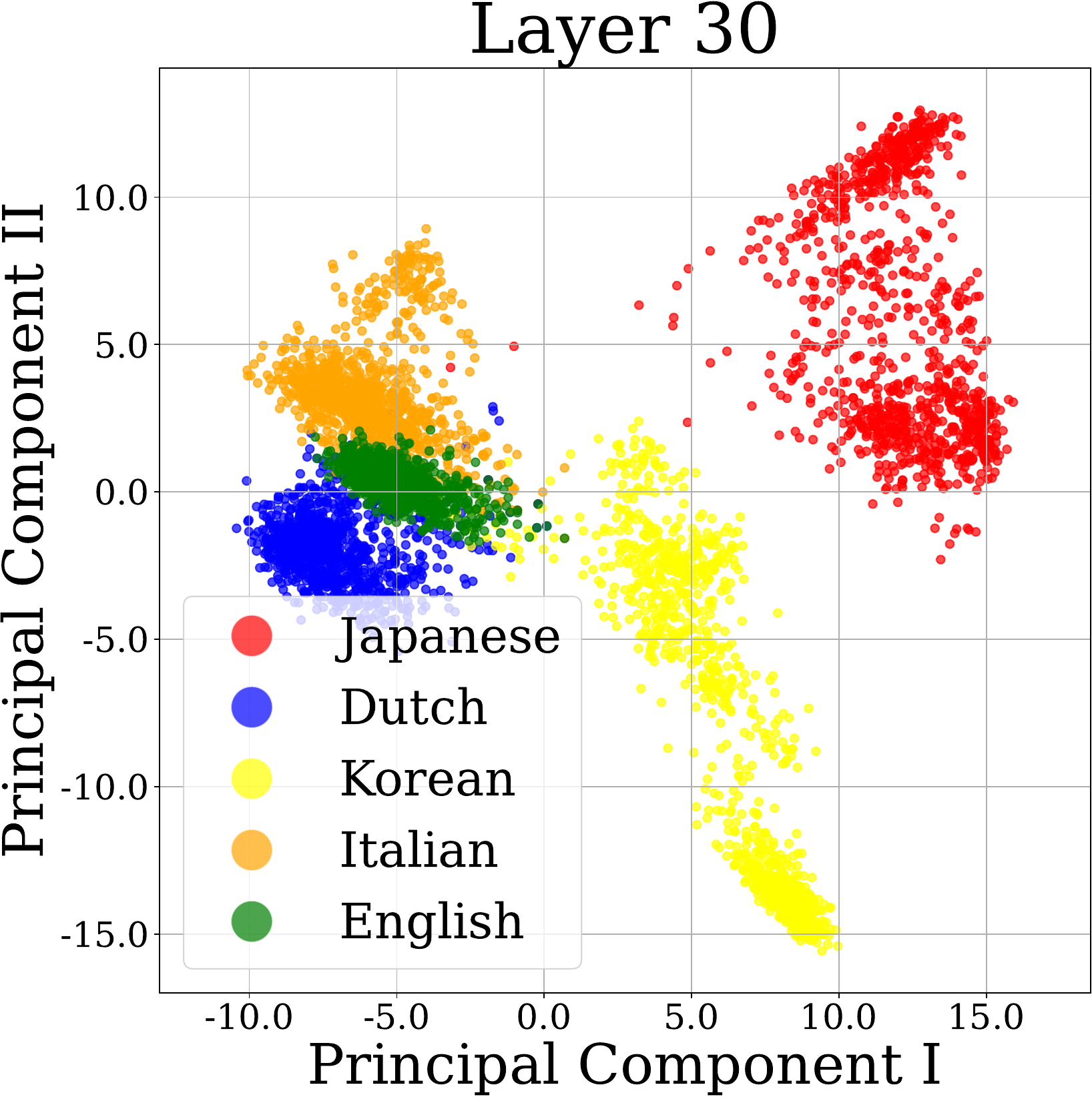}
  \includegraphics[width=0.19\linewidth]{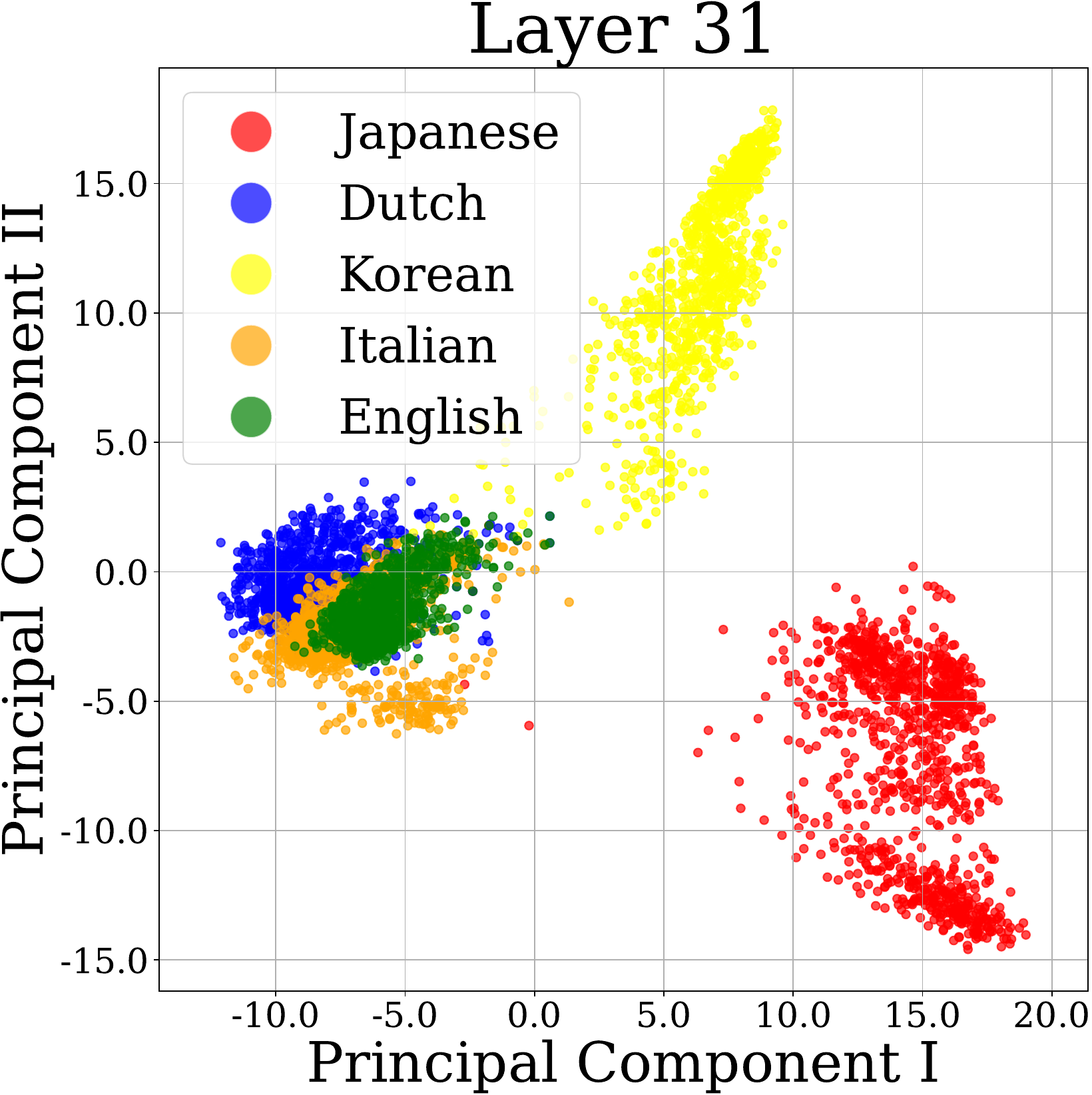}
  \includegraphics[width=0.19\linewidth]{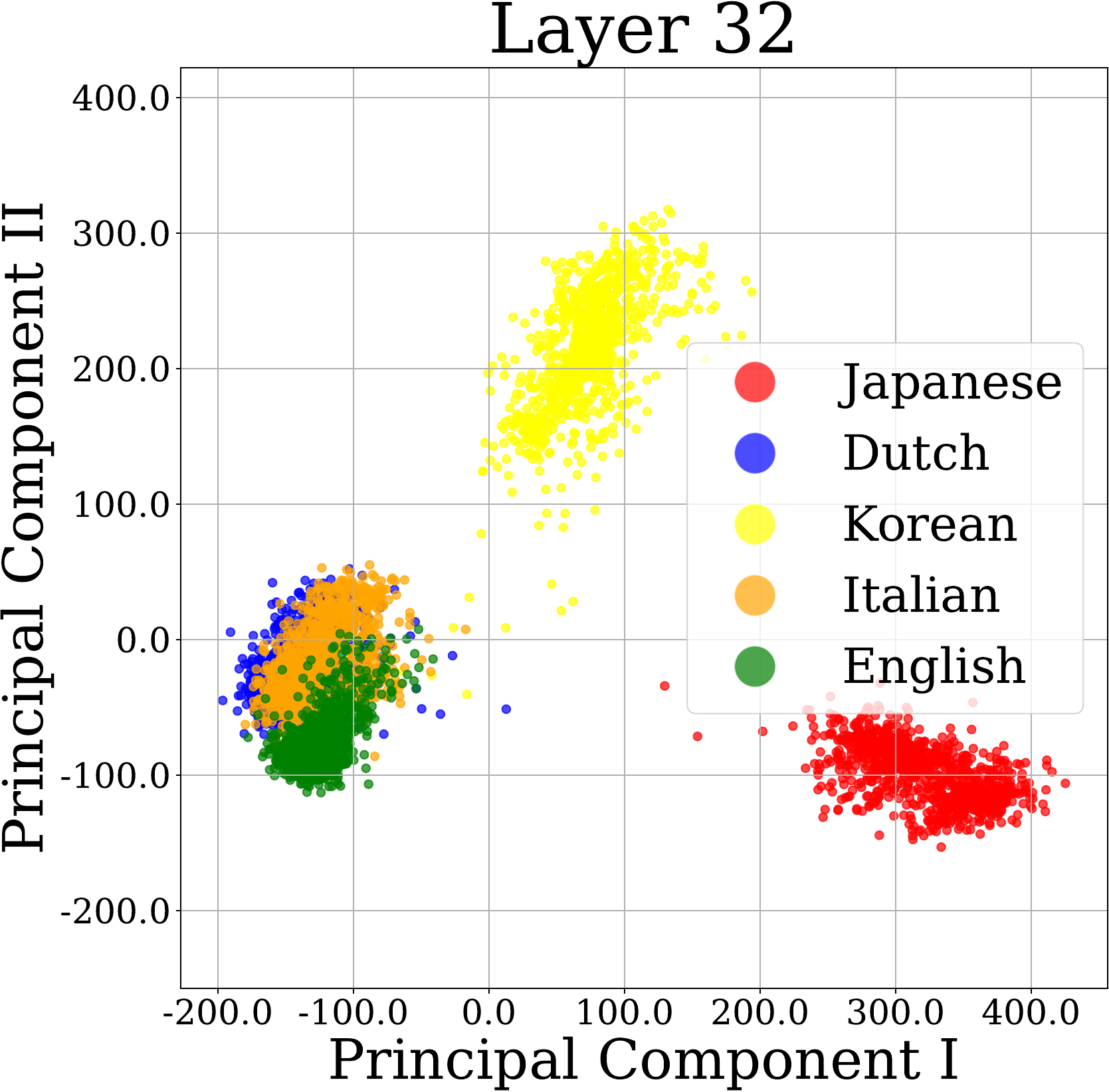}

  \caption{\textbf{The resutls of PCA applied to the hidden representations across layers (Mistral-7B).}}
  \label{fig:appendix:pca_all_layers_mistral}
\end{figure*}

\begin{figure*}[t]
  \centering

  \includegraphics[width=0.19\linewidth]{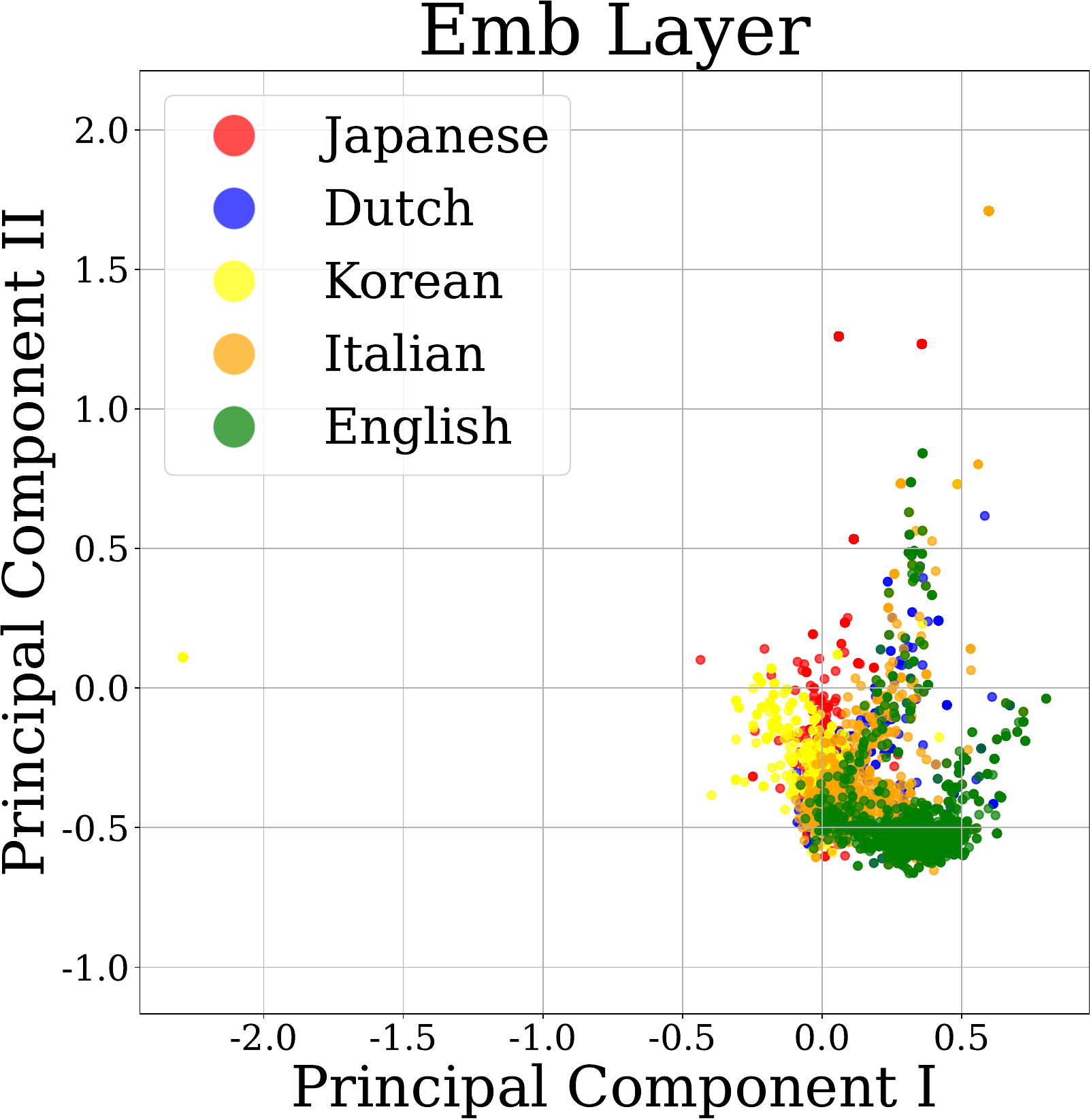}
  \includegraphics[width=0.19\linewidth]{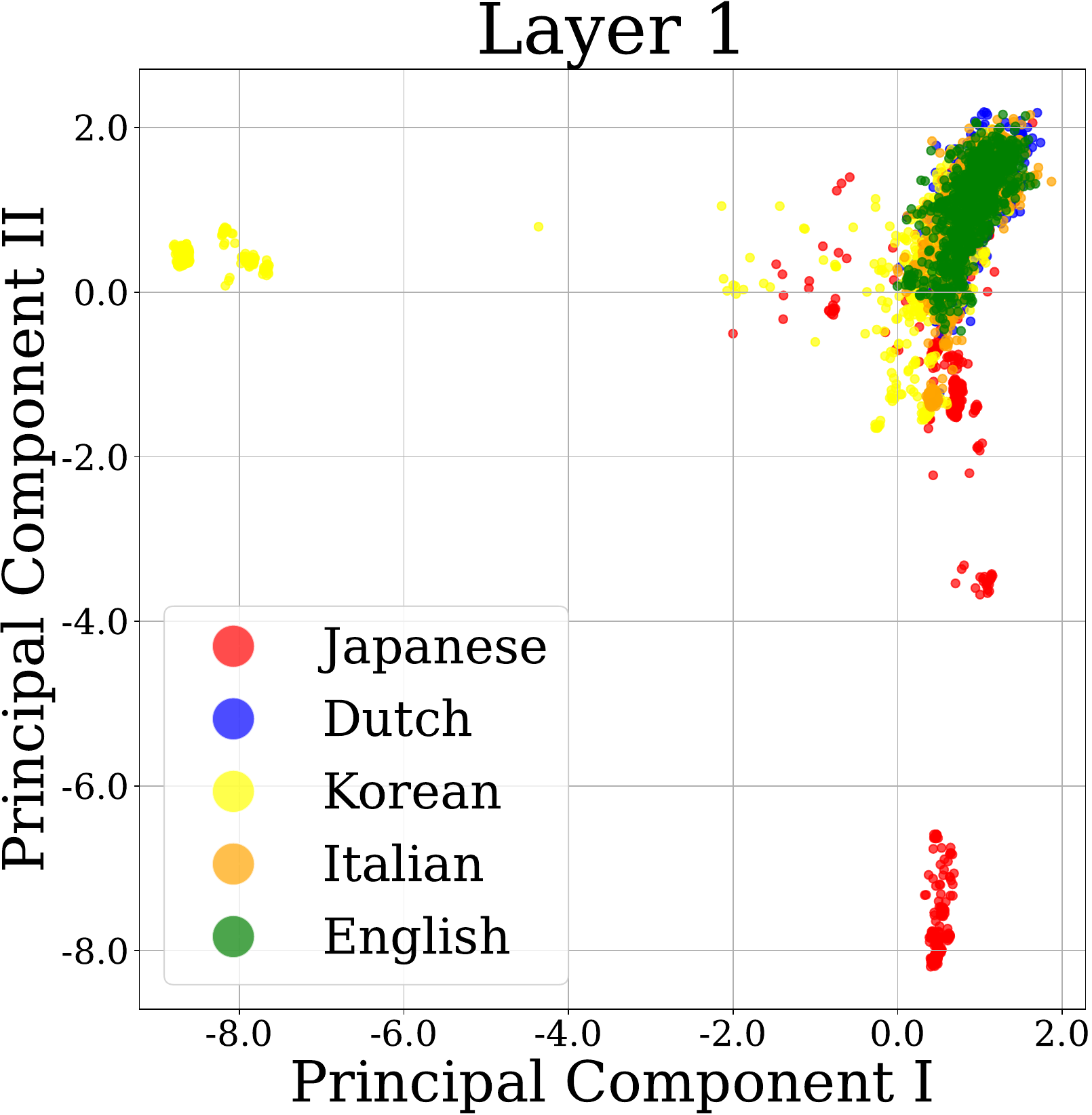}
  \includegraphics[width=0.19\linewidth]{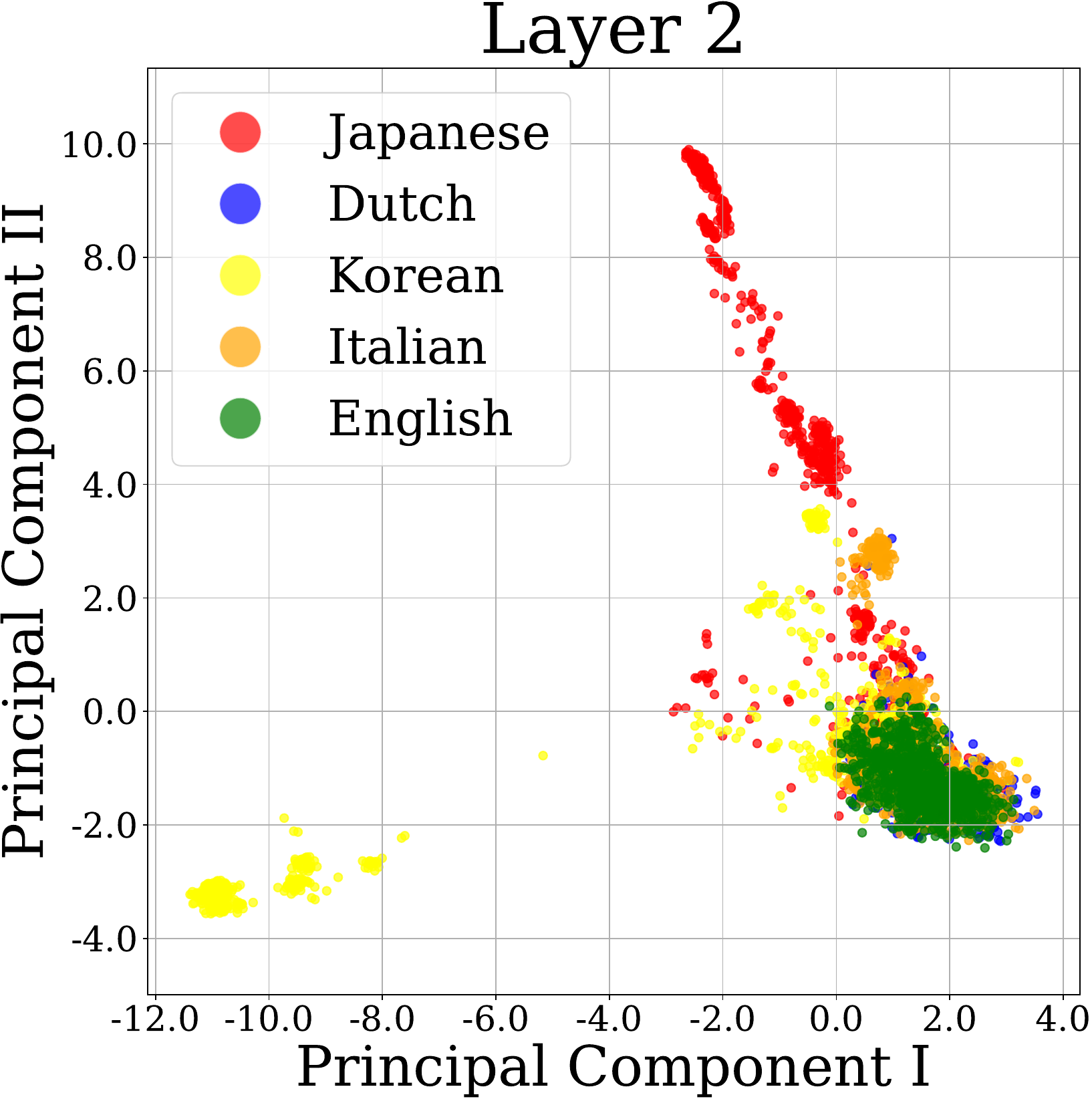}
  \includegraphics[width=0.19\linewidth]{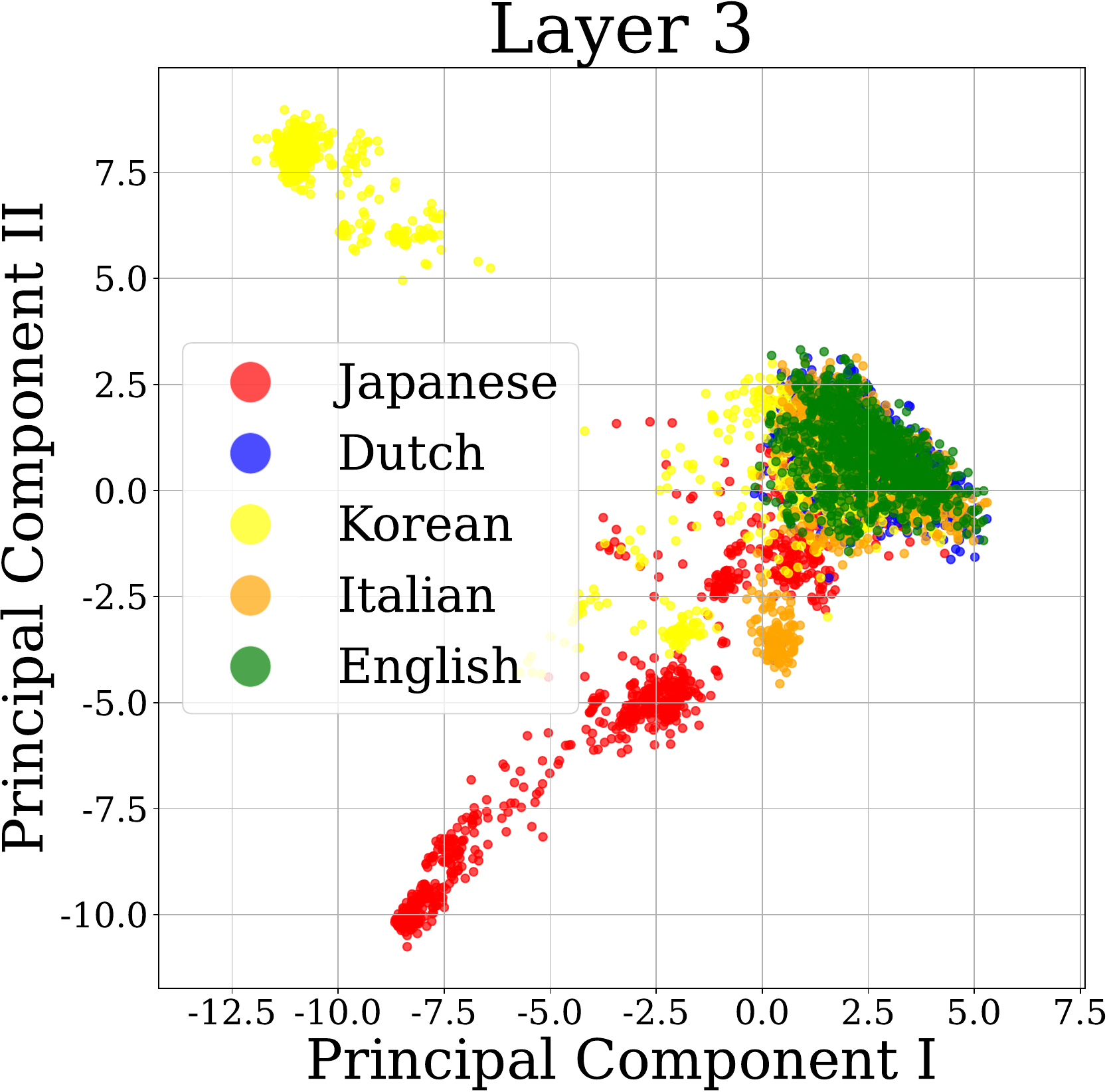}
  \includegraphics[width=0.19\linewidth]{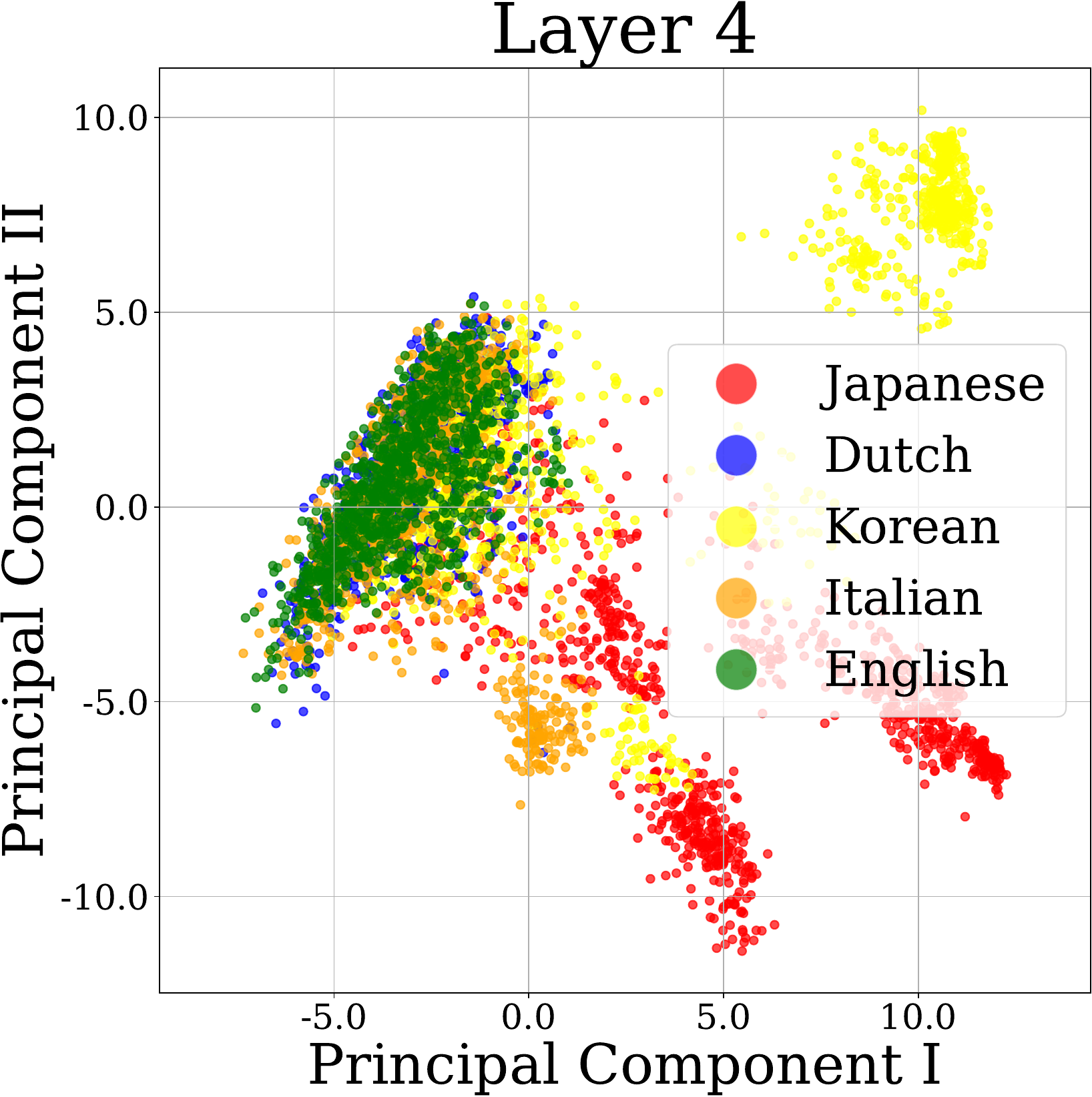}

  \includegraphics[width=0.19\linewidth]{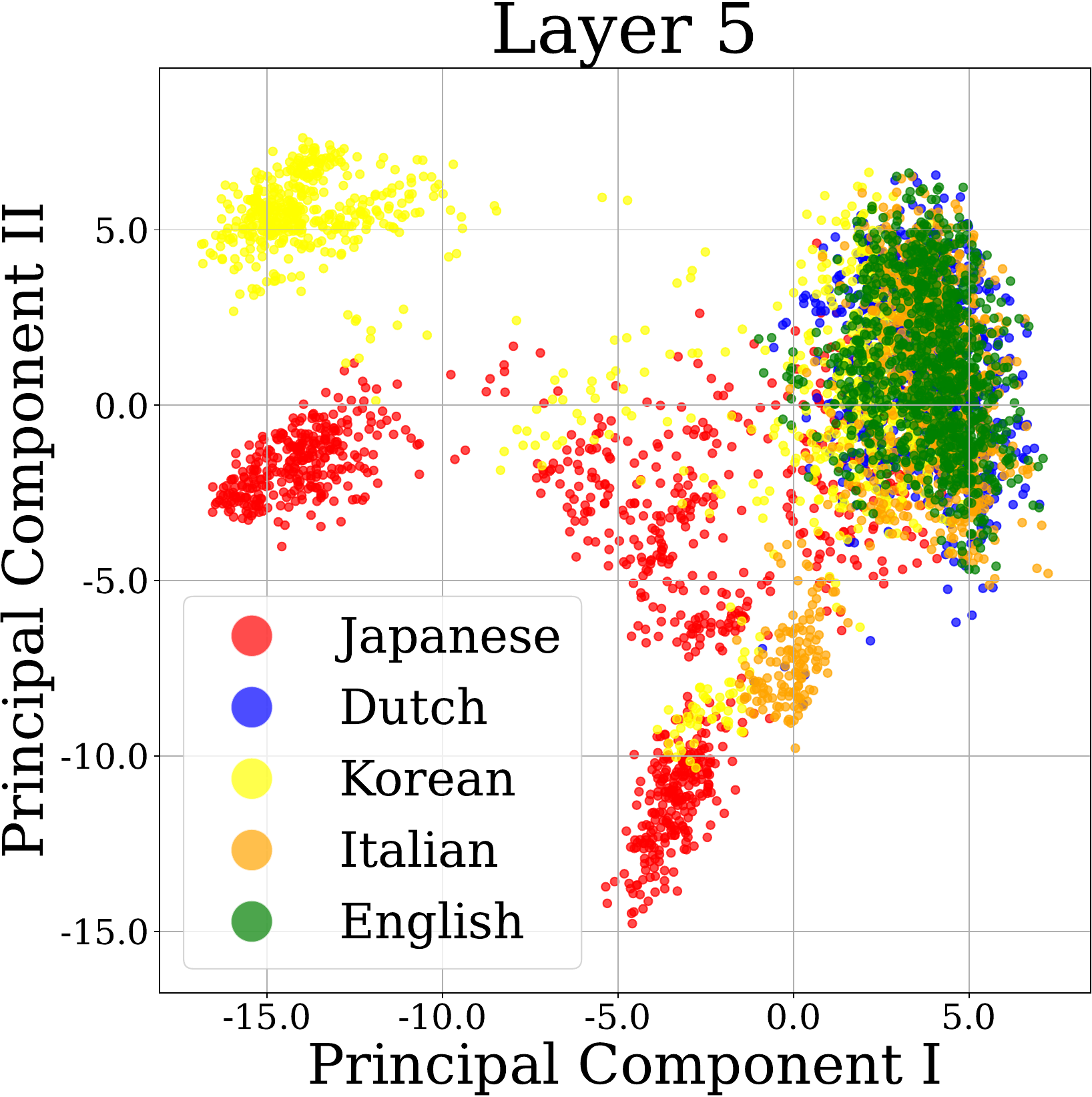}
  \includegraphics[width=0.19\linewidth]{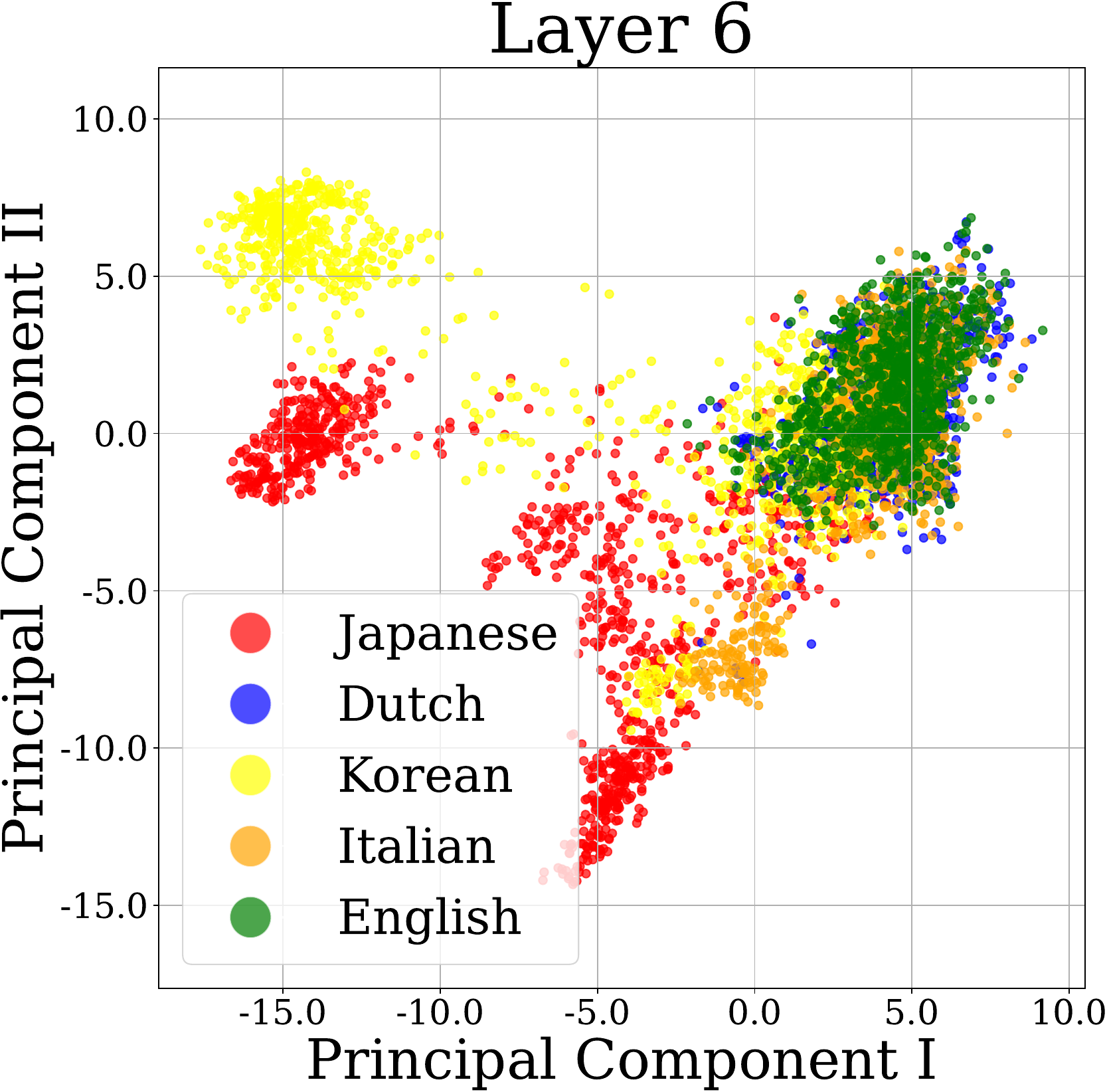}
  \includegraphics[width=0.19\linewidth]{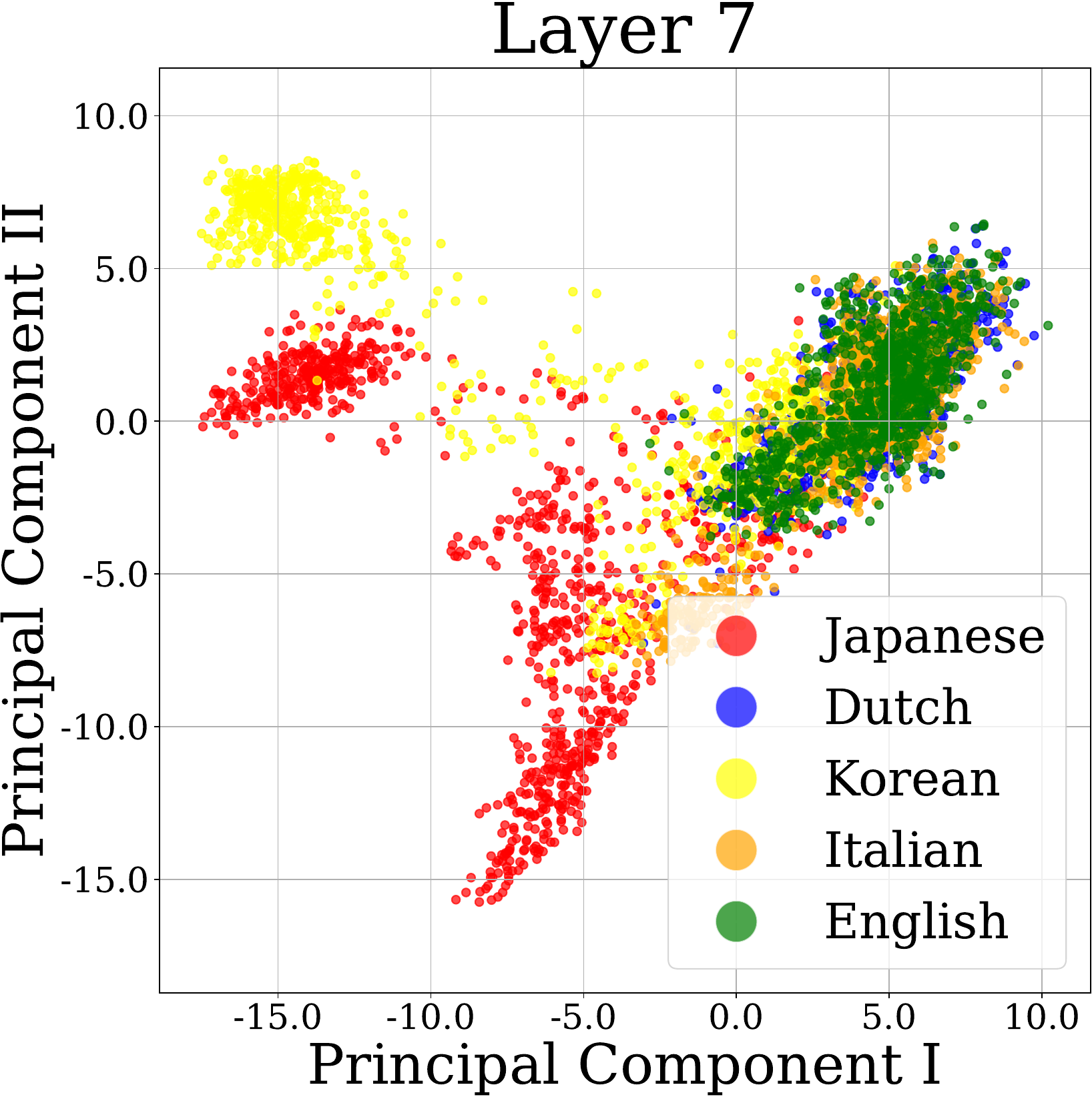}
  \includegraphics[width=0.19\linewidth]{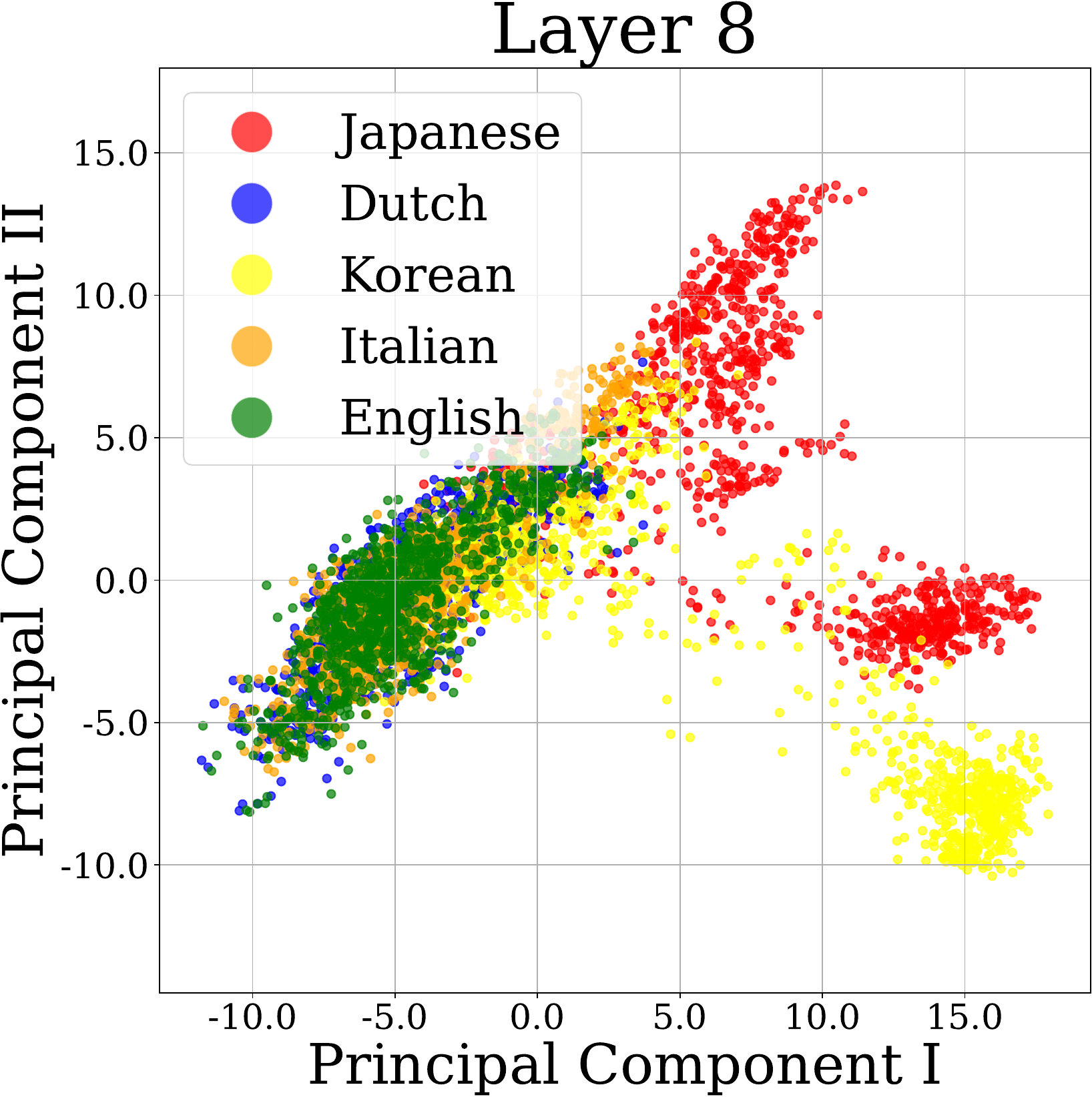}
  \includegraphics[width=0.19\linewidth]{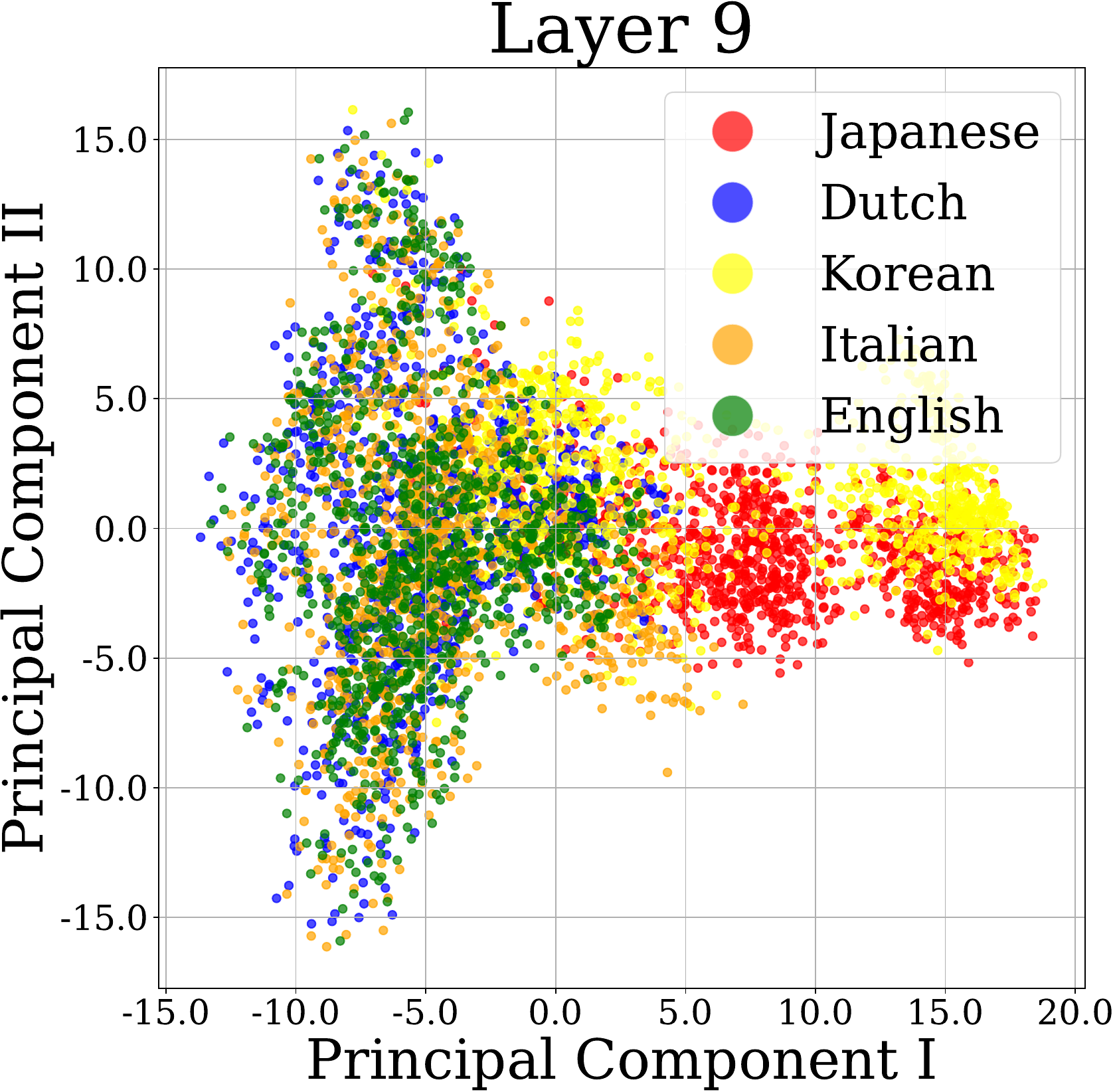}

  \includegraphics[width=0.19\linewidth]{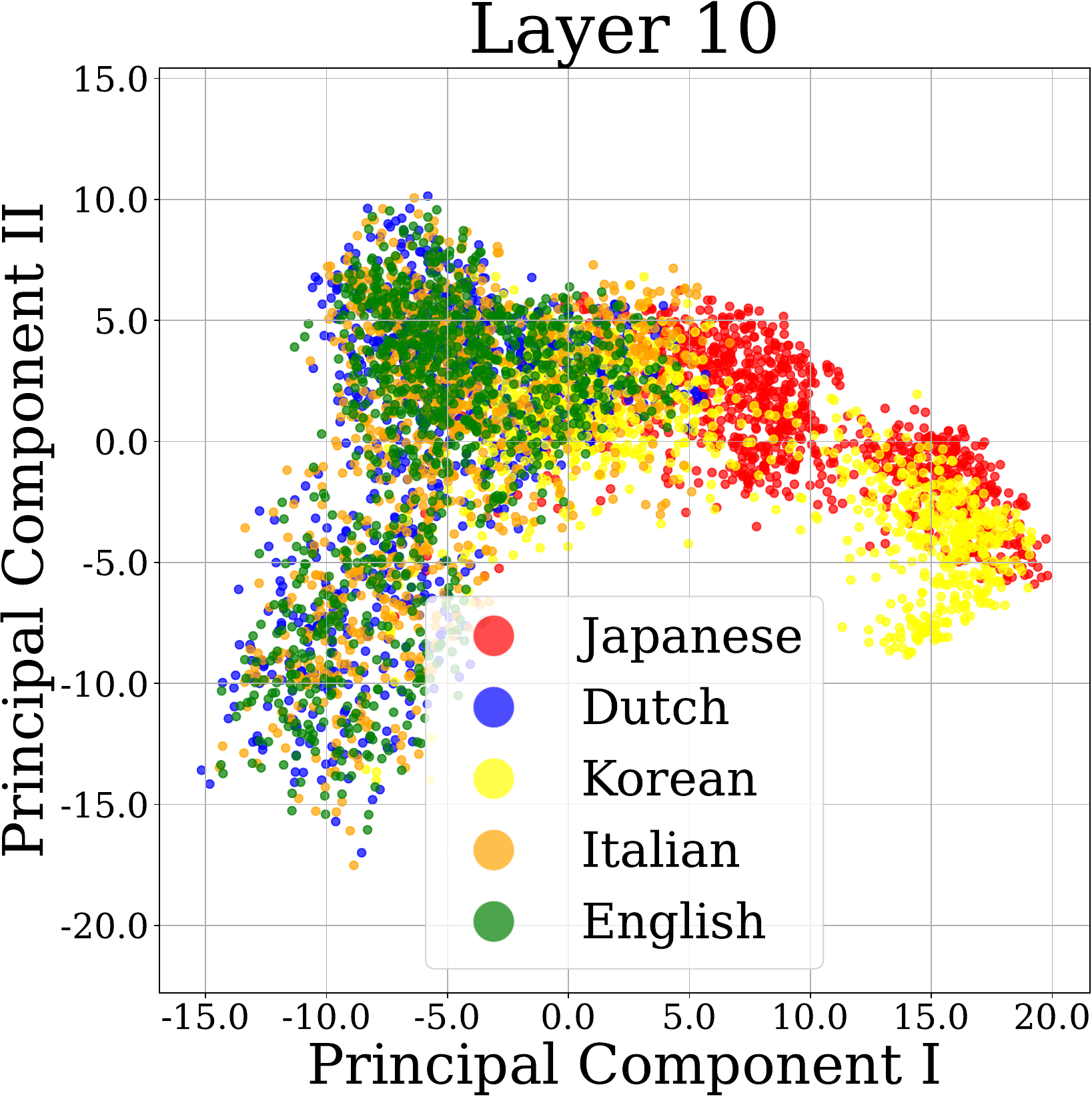}
  \includegraphics[width=0.19\linewidth]{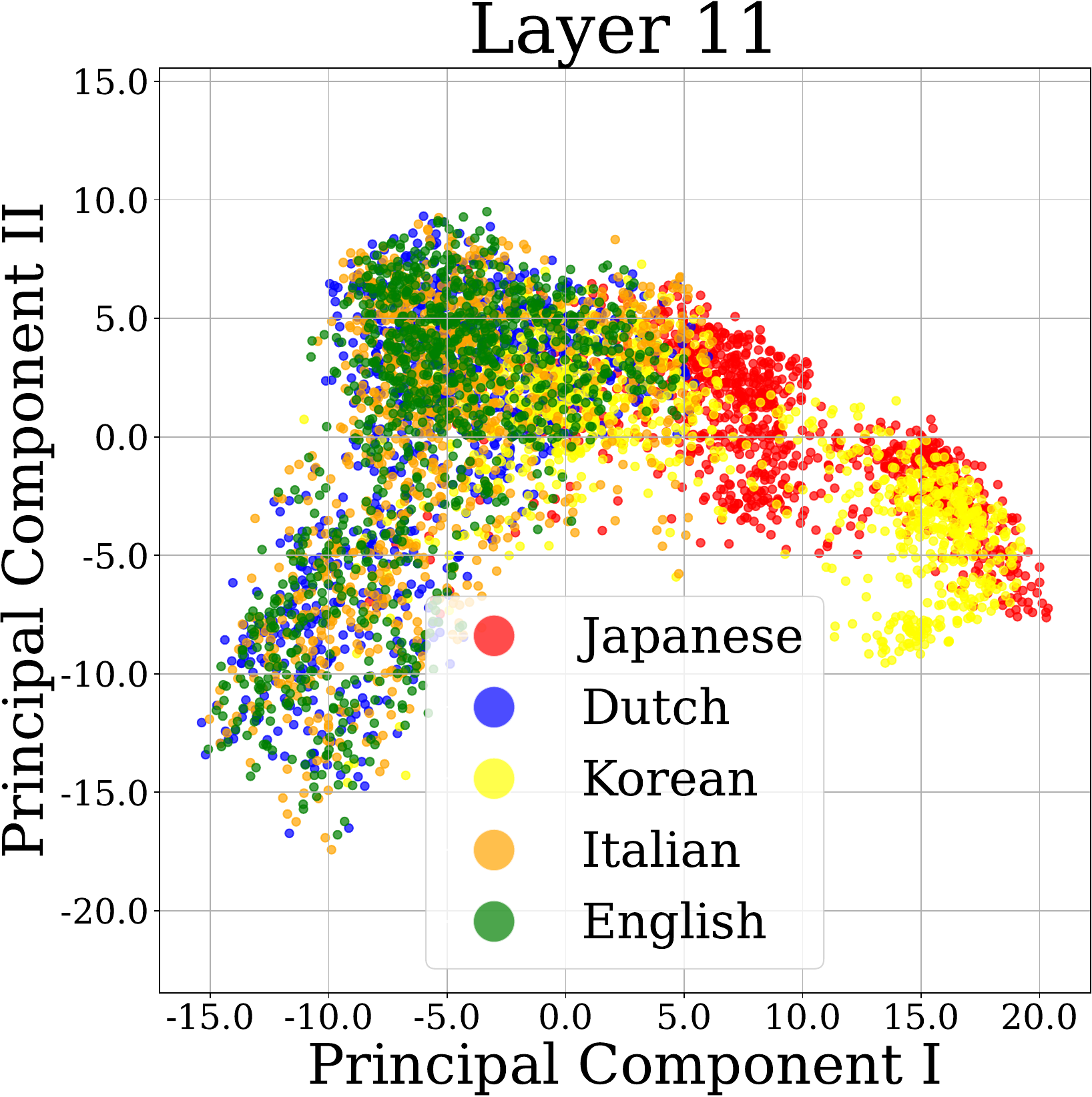}
  \includegraphics[width=0.19\linewidth]{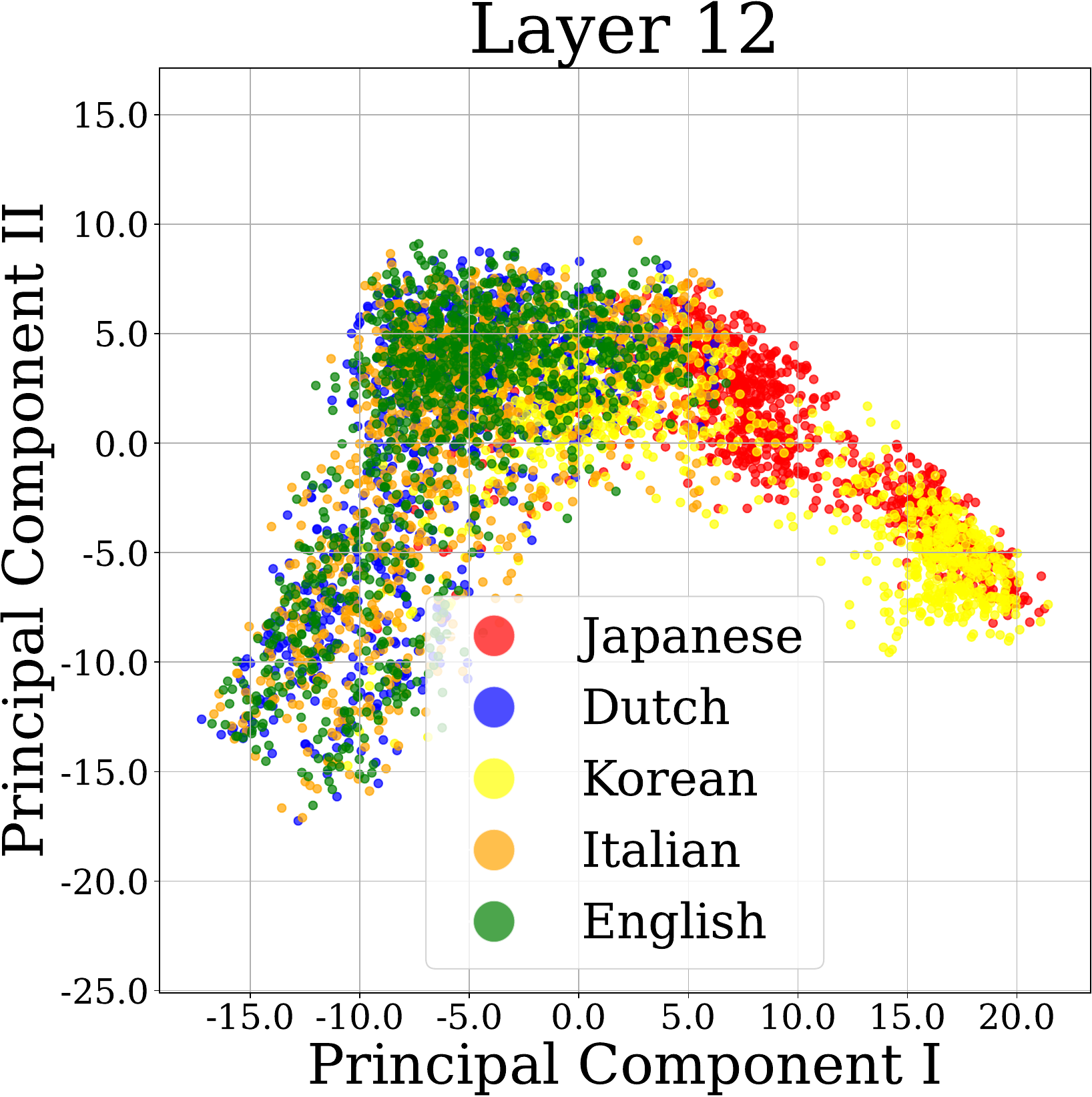}
  \includegraphics[width=0.19\linewidth]{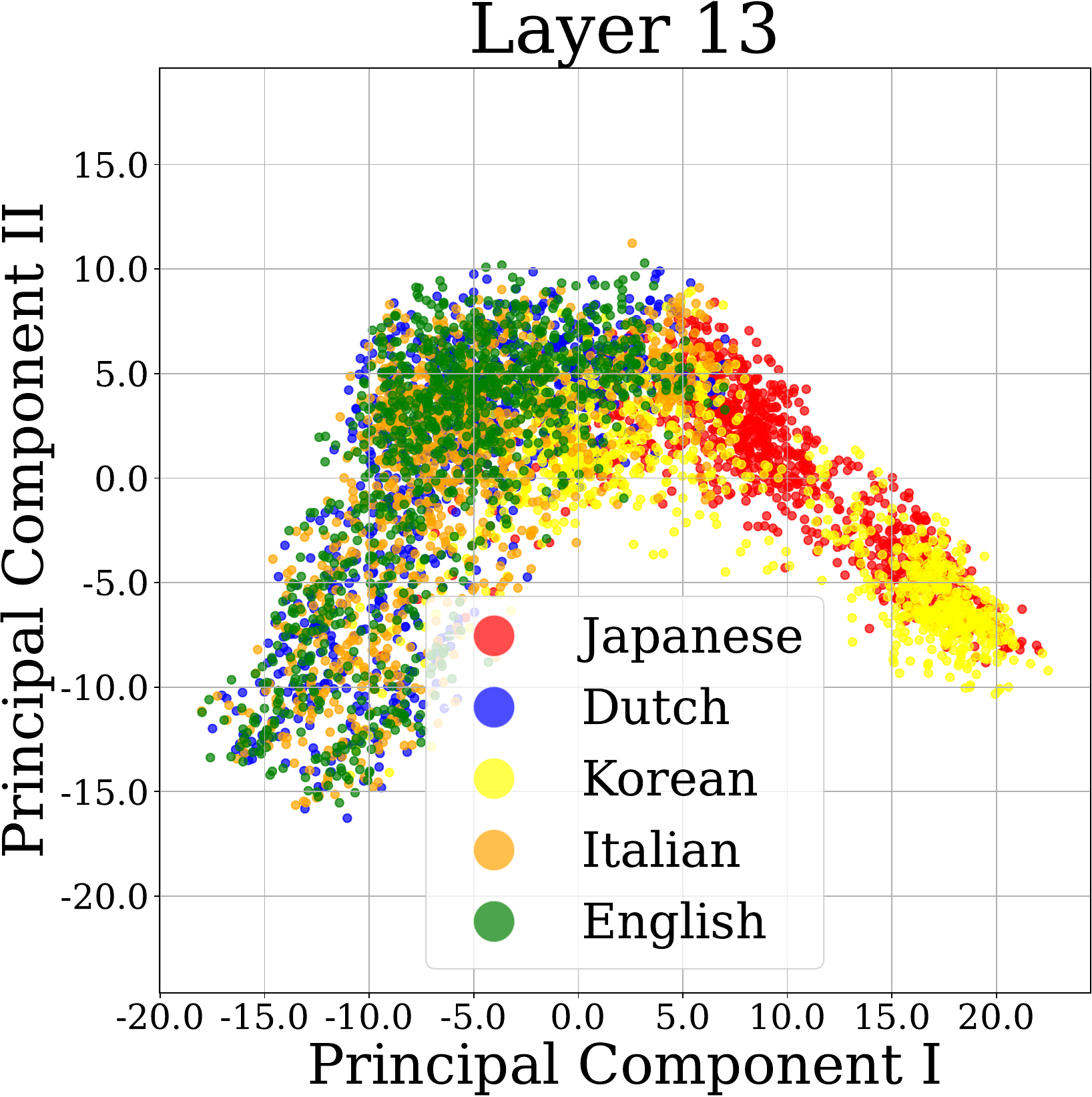}
  \includegraphics[width=0.19\linewidth]{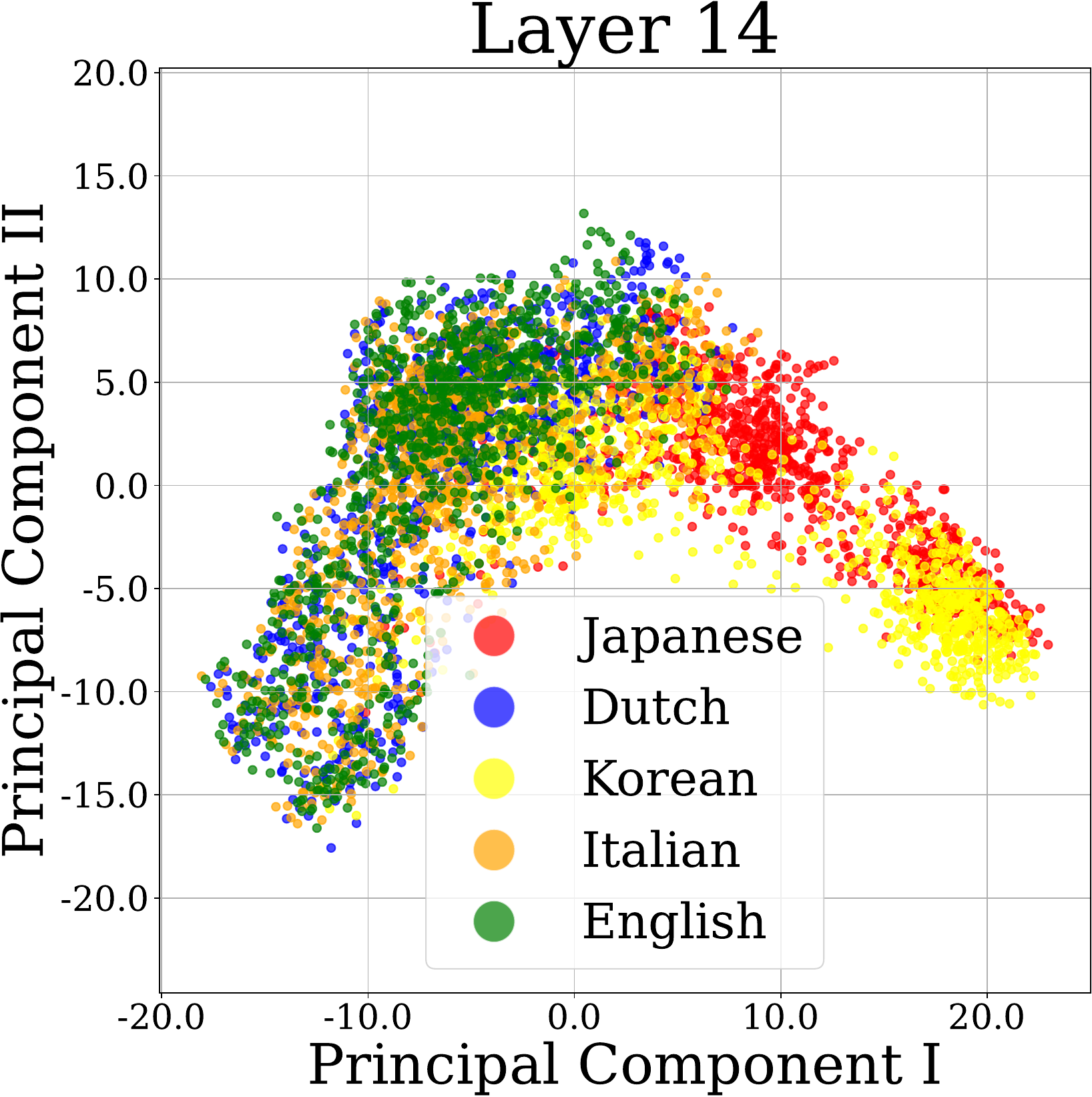}

  \includegraphics[width=0.19\linewidth]{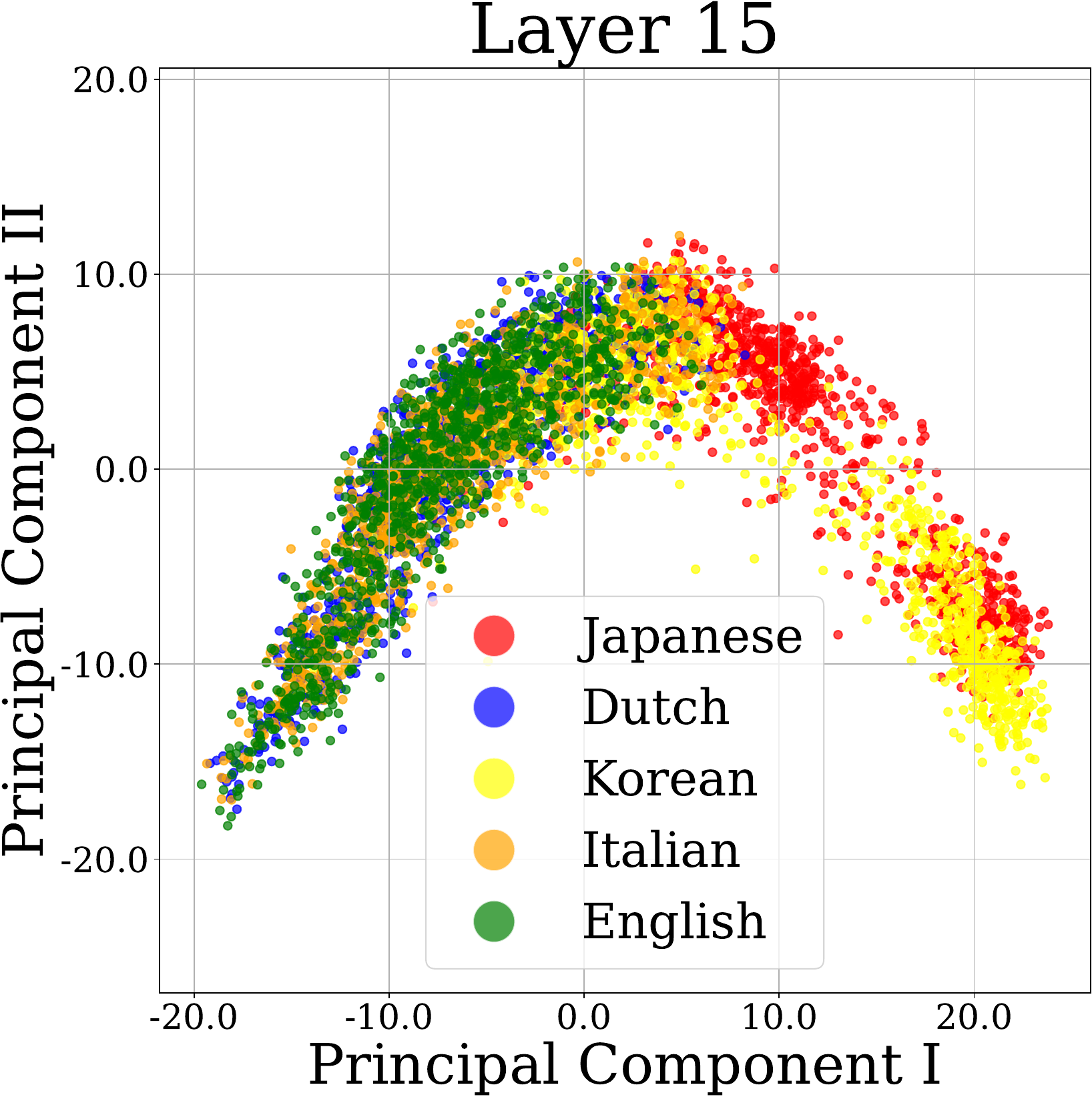}
  \includegraphics[width=0.19\linewidth]{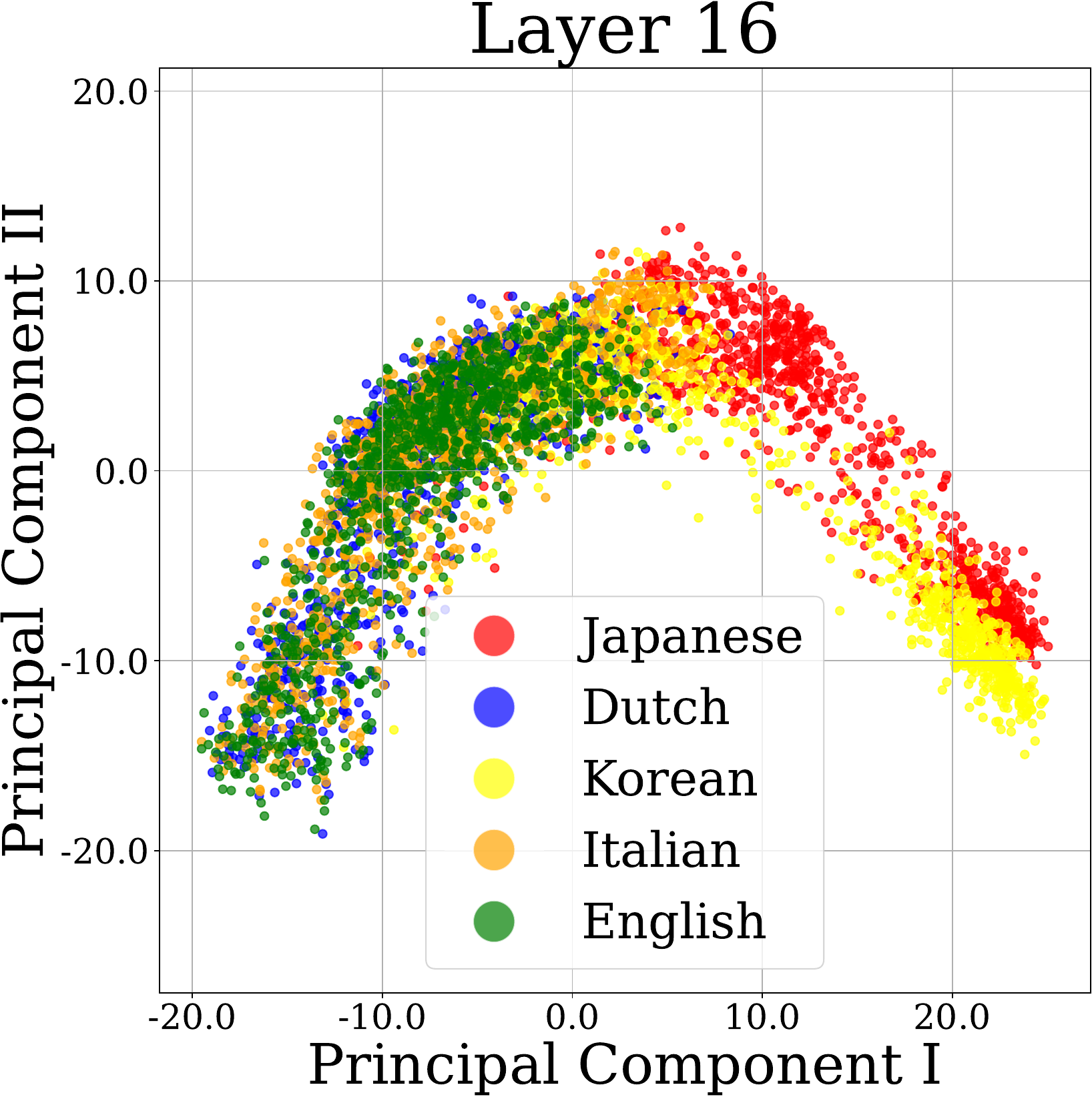}
  \includegraphics[width=0.19\linewidth]{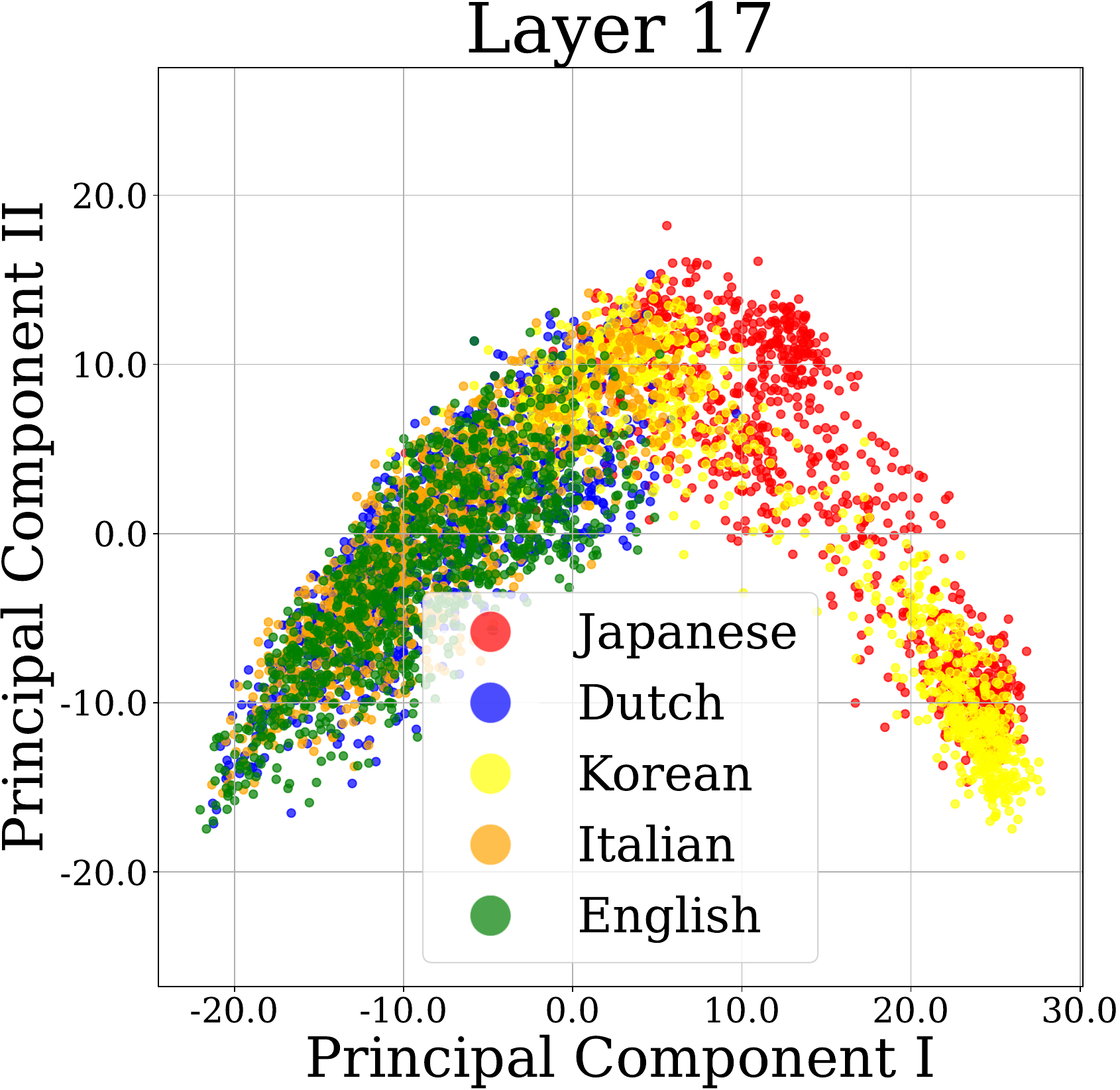}
  \includegraphics[width=0.19\linewidth]{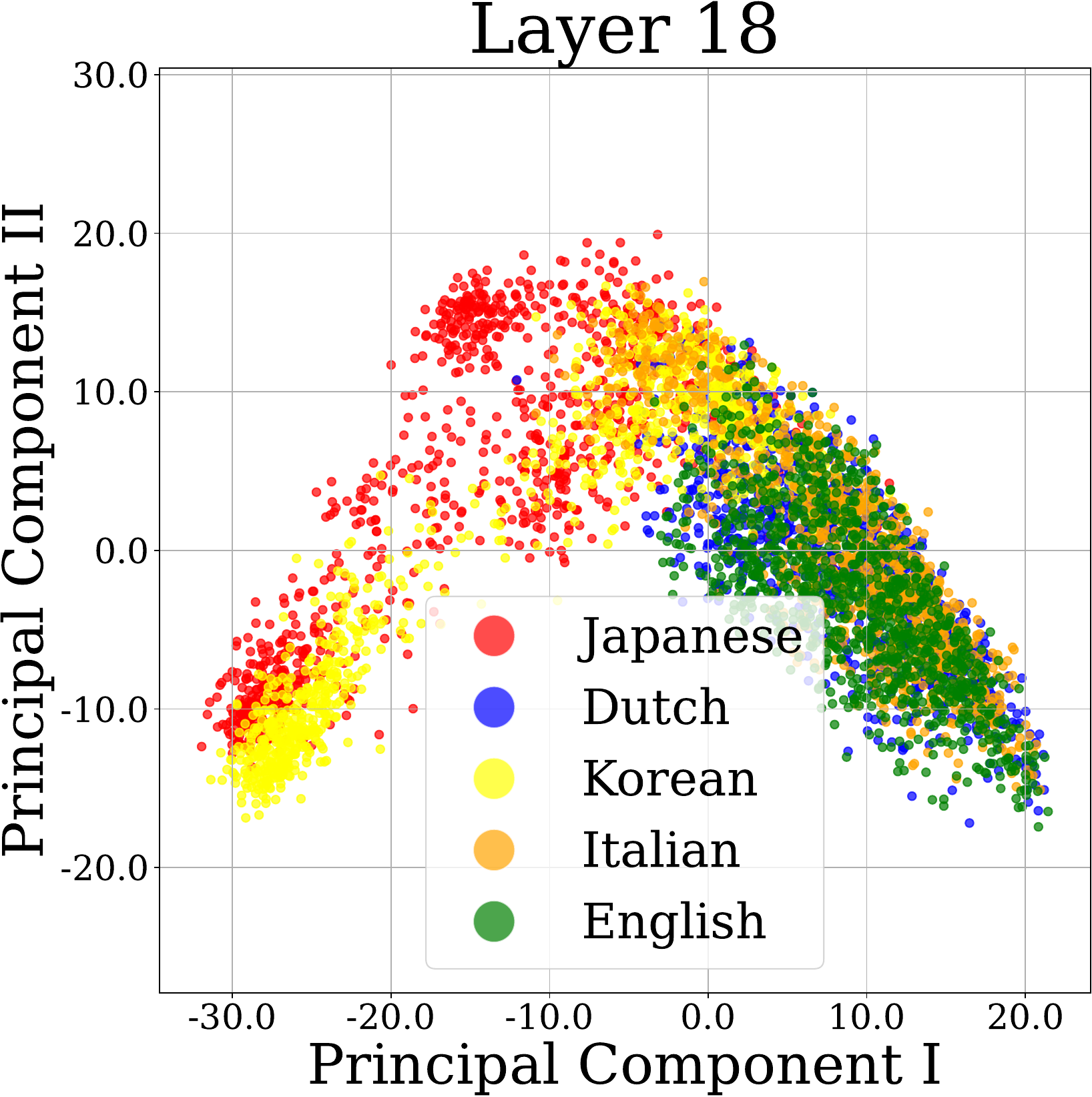}
  \includegraphics[width=0.19\linewidth]{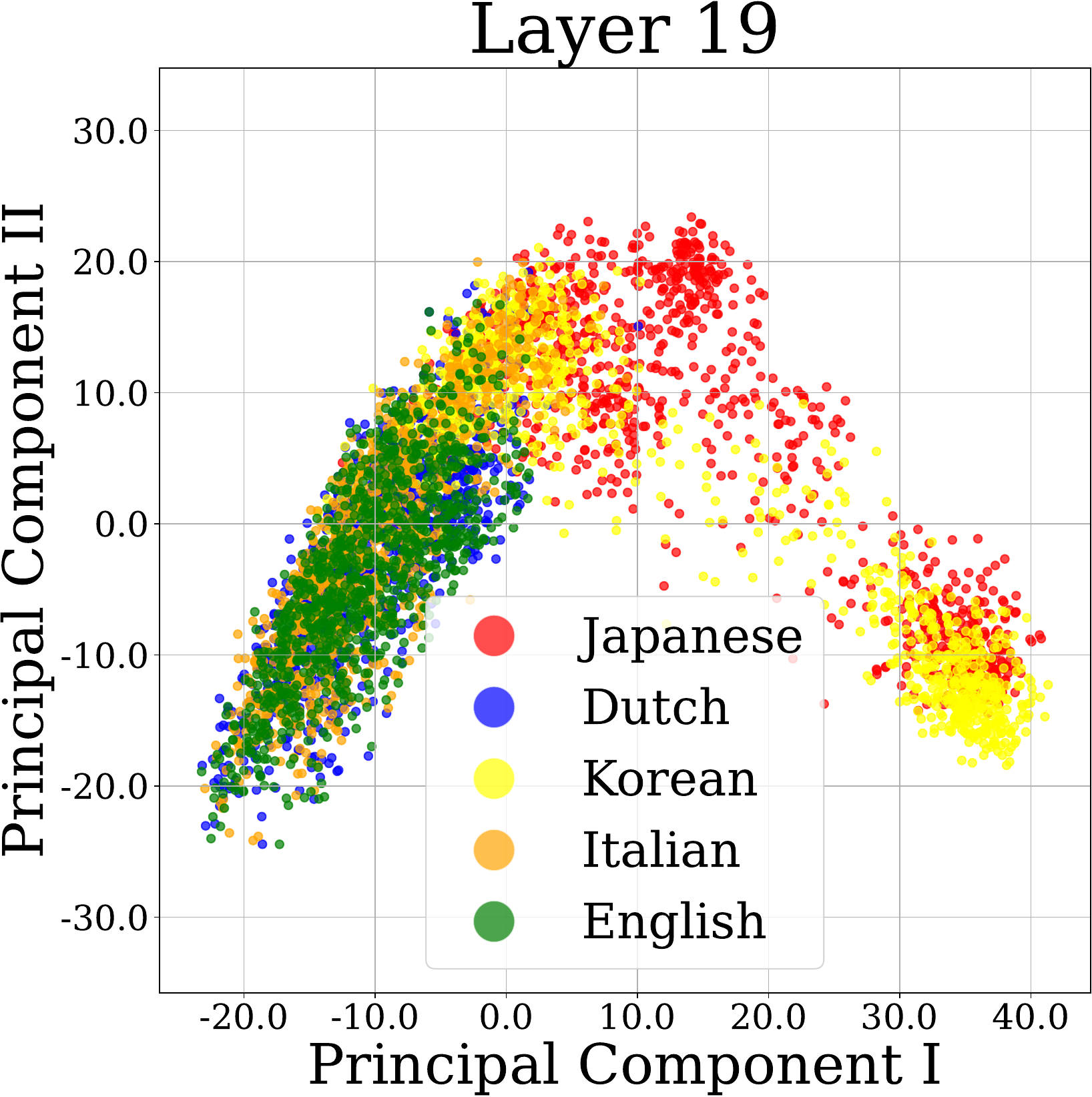}

  \includegraphics[width=0.19\linewidth]{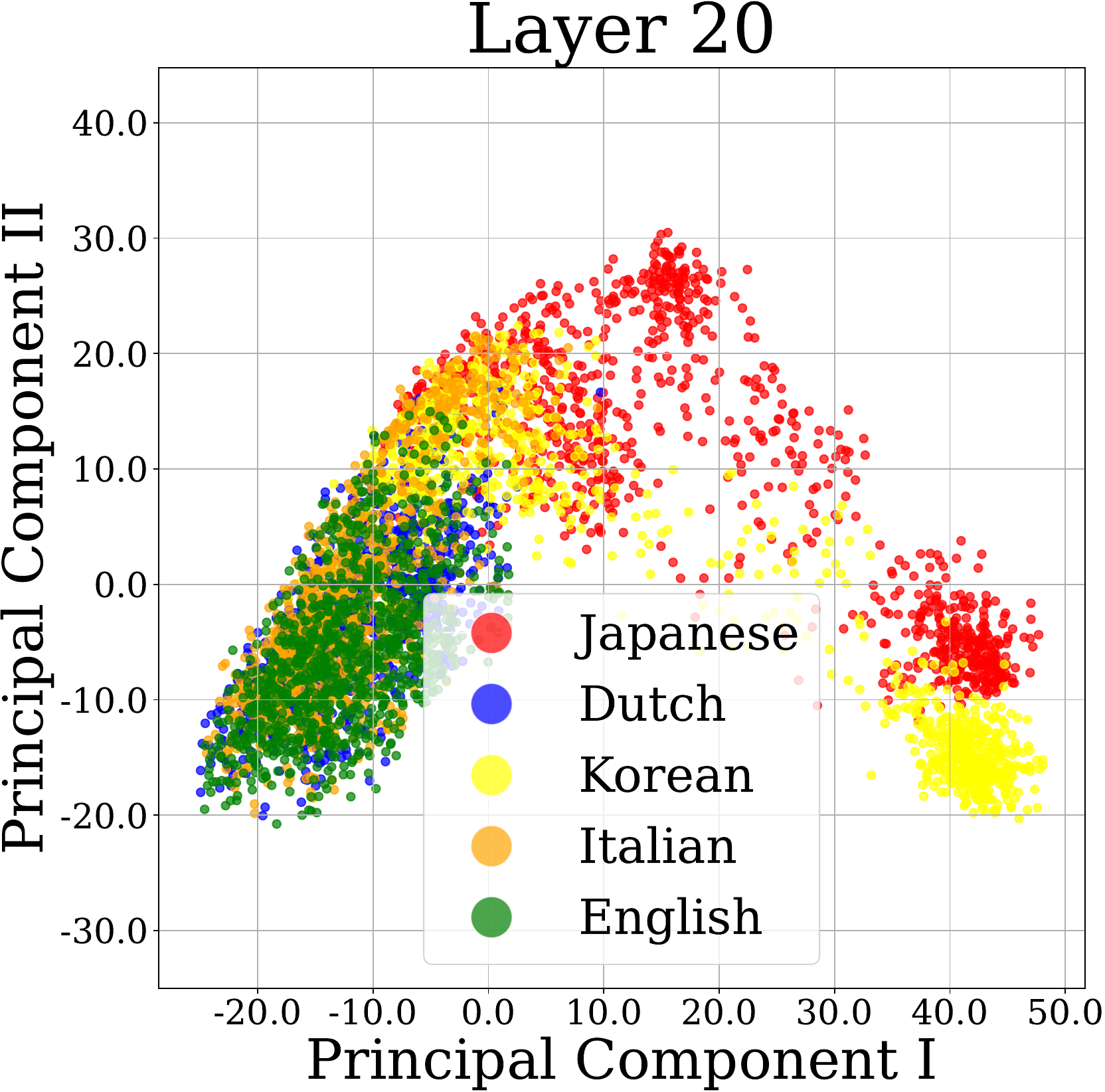}
  \includegraphics[width=0.19\linewidth]{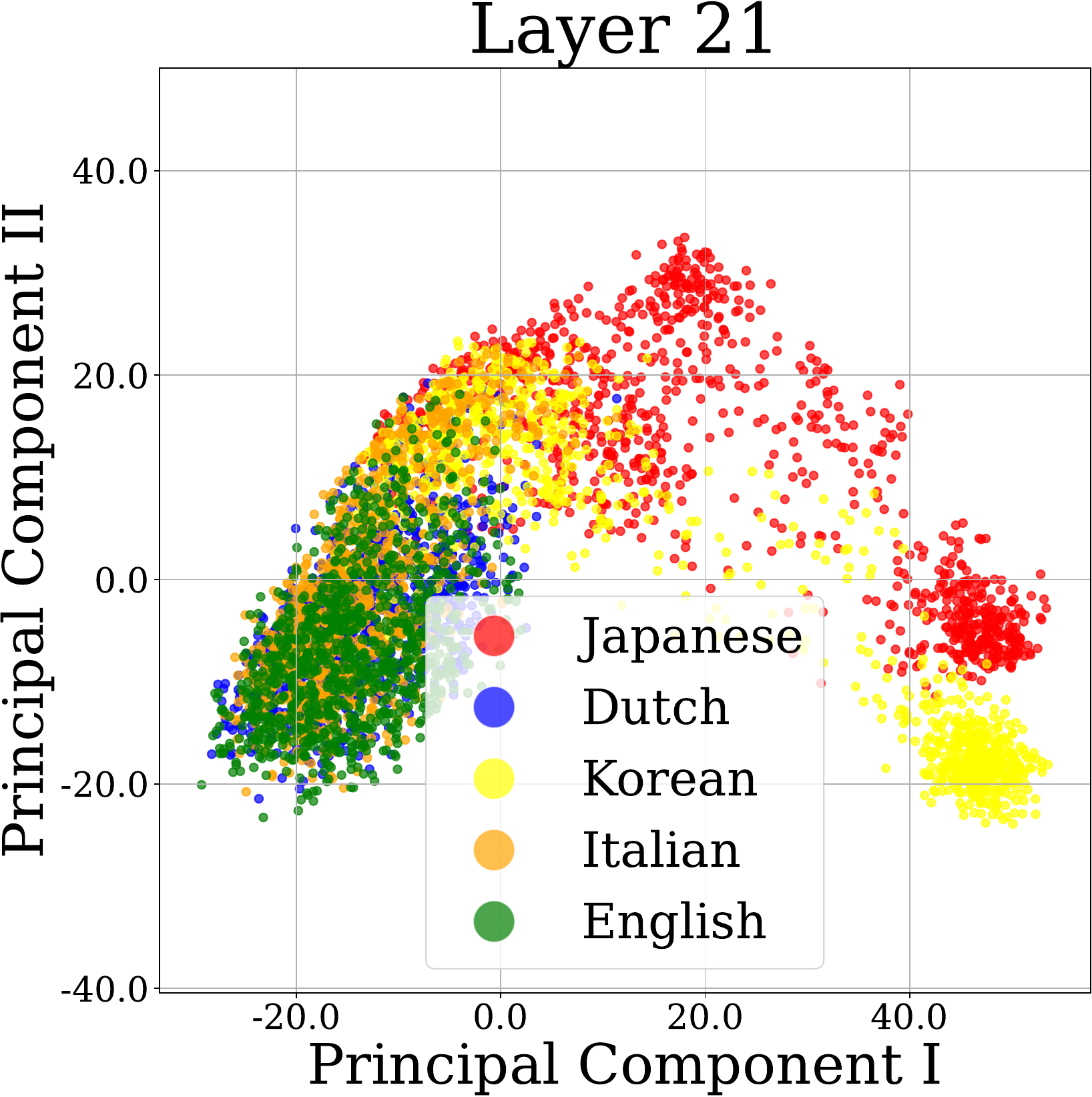}
  \includegraphics[width=0.19\linewidth]{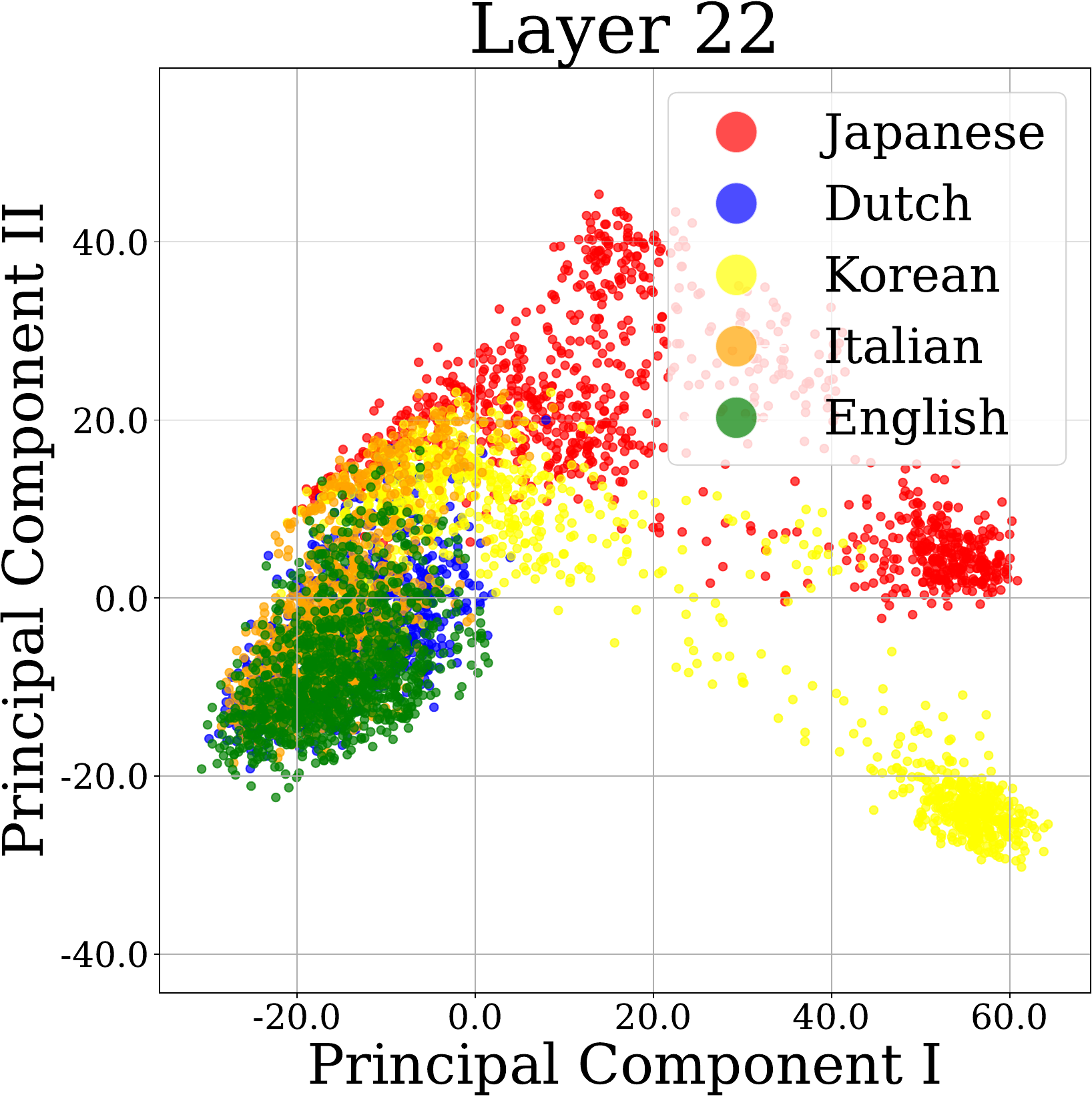}
  \includegraphics[width=0.19\linewidth]{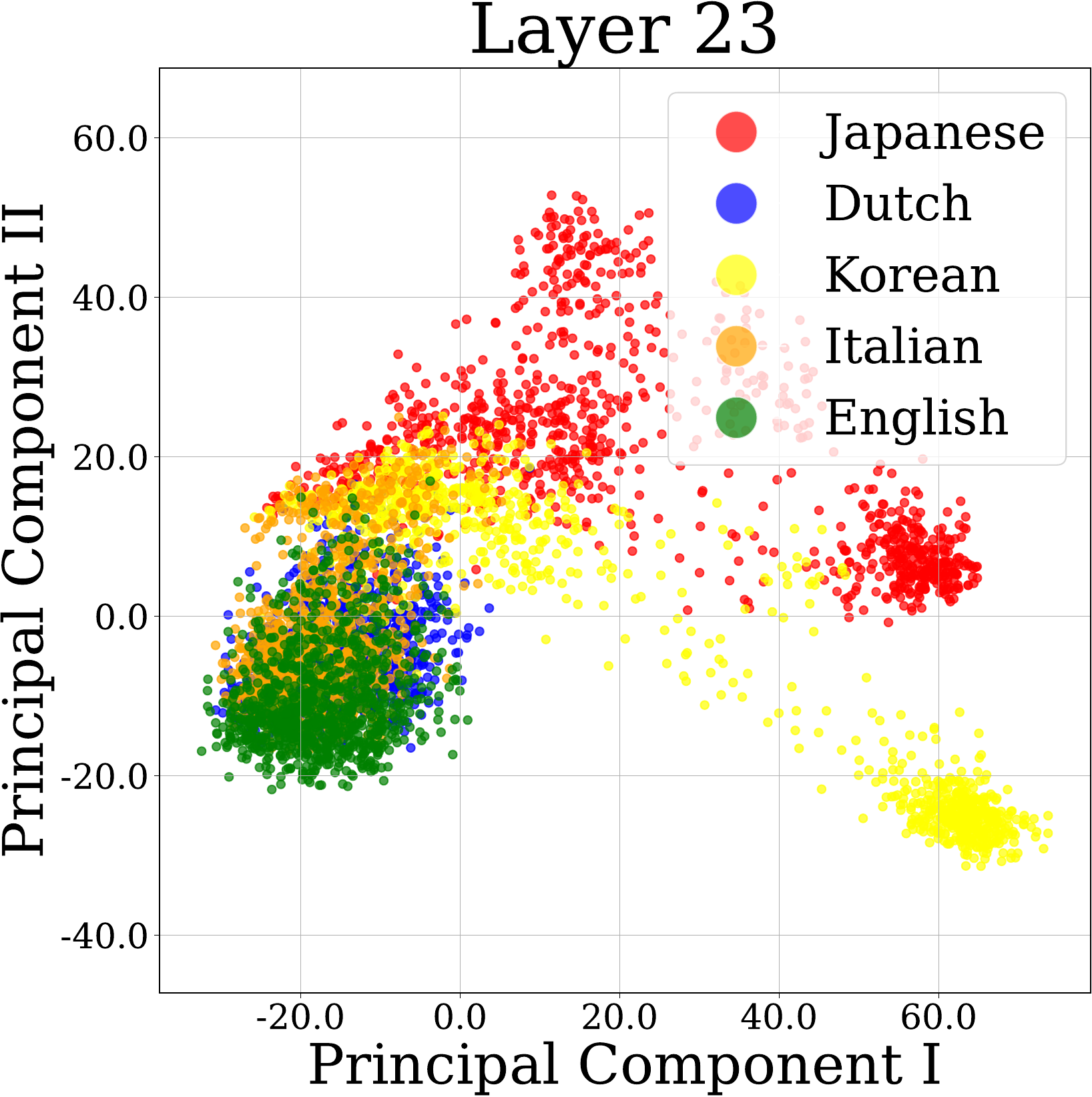}
  \includegraphics[width=0.19\linewidth]{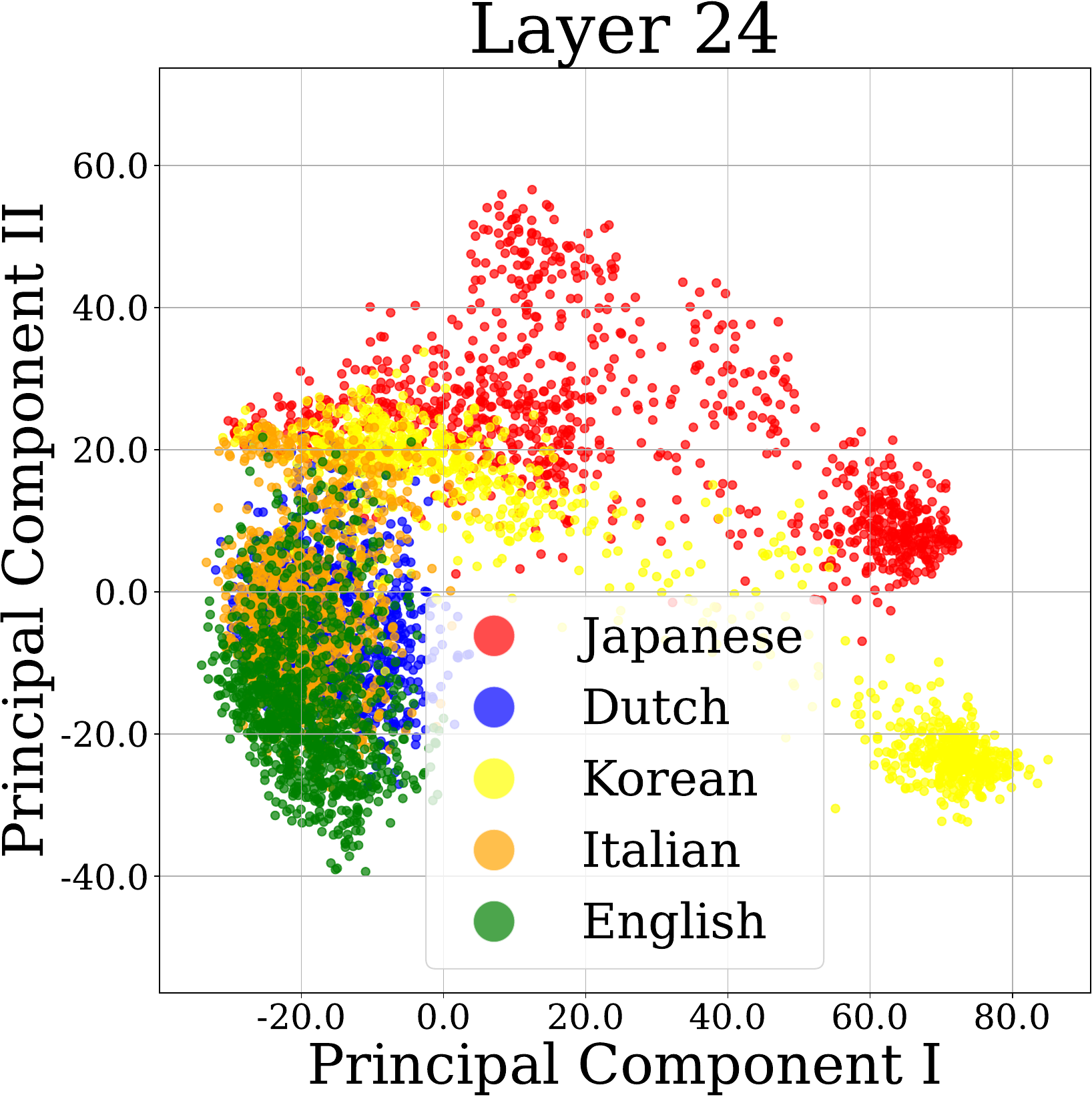}

  \includegraphics[width=0.19\linewidth]{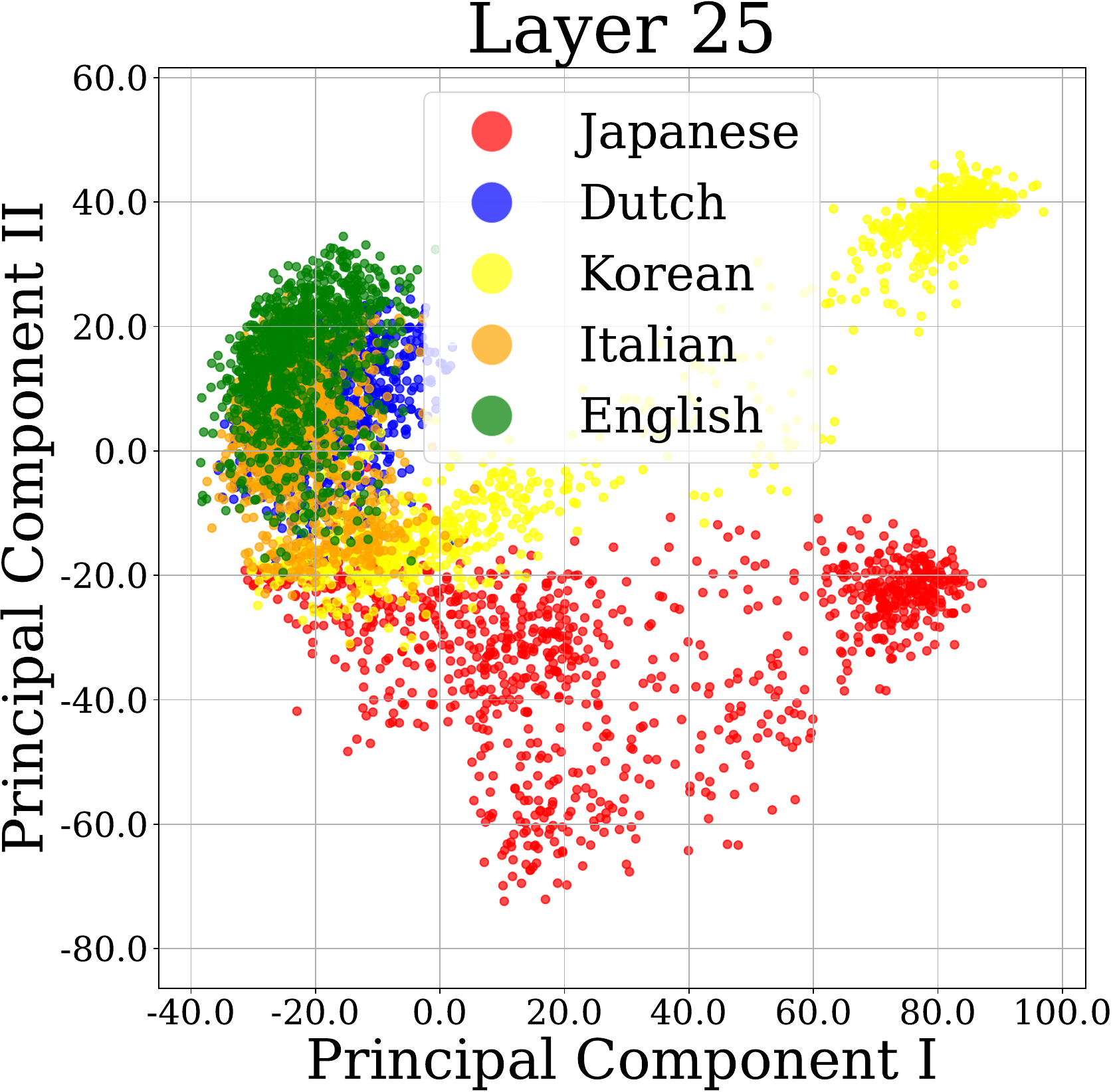}
  \includegraphics[width=0.19\linewidth]{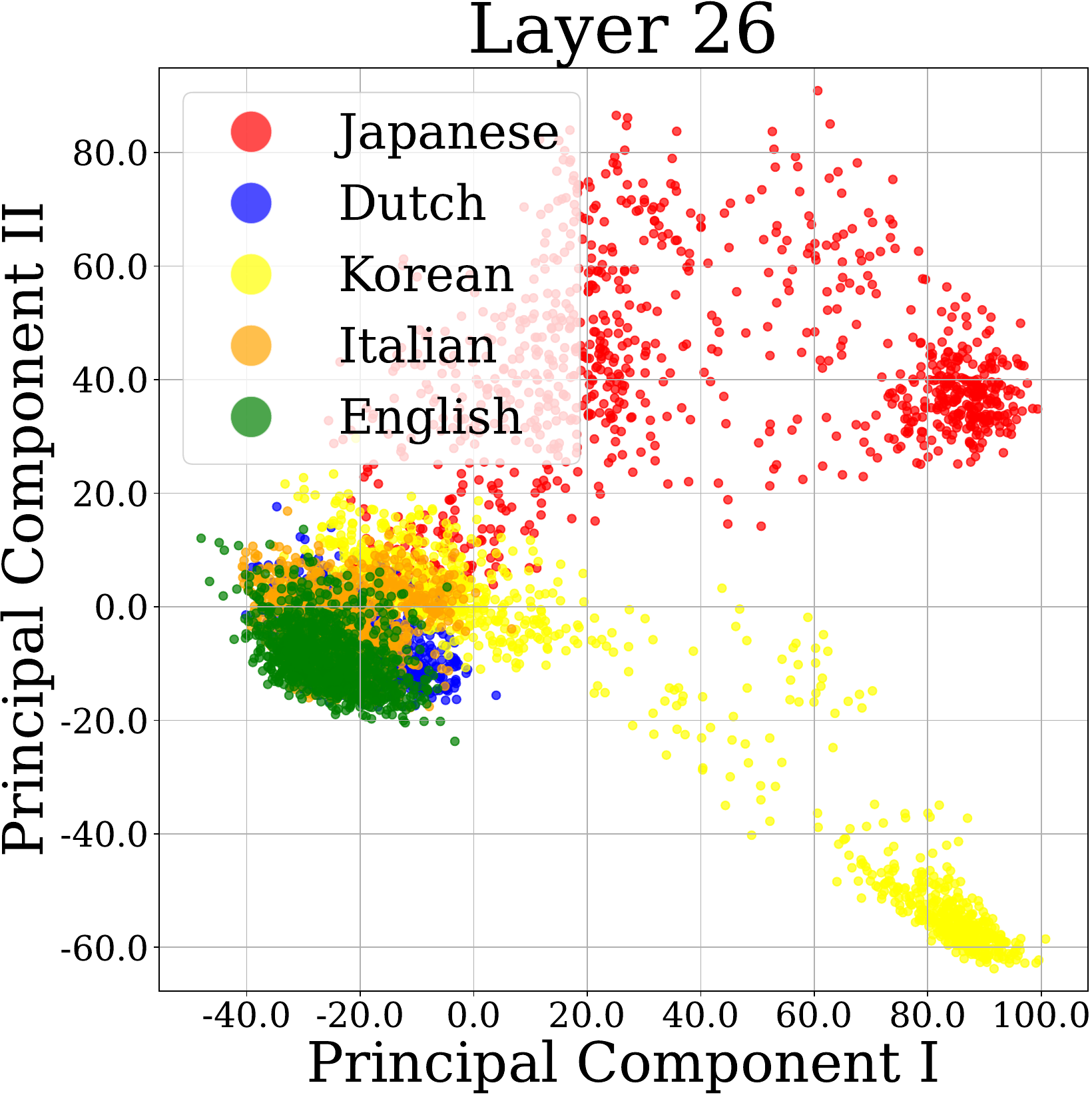}
  \includegraphics[width=0.19\linewidth]{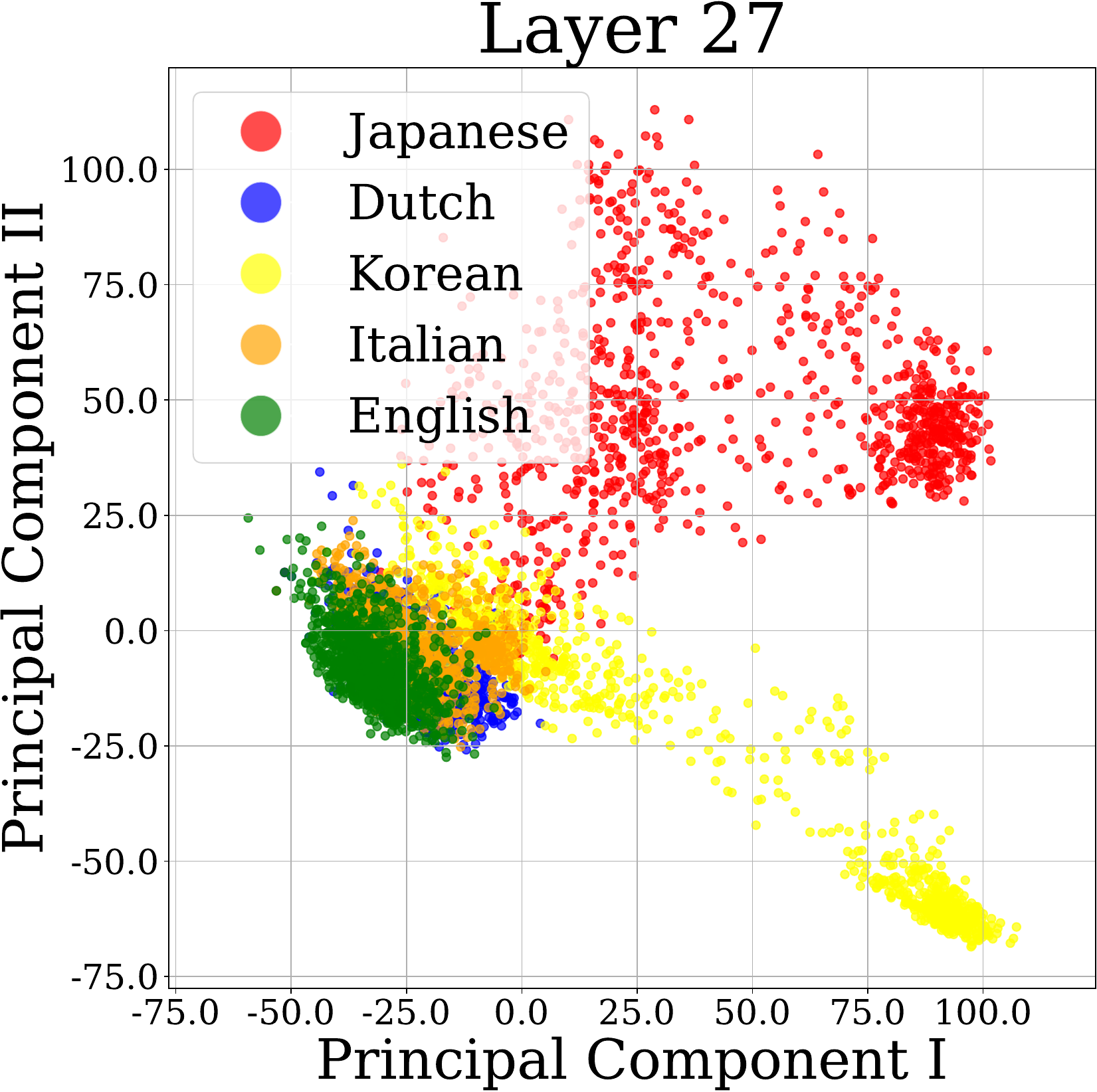}
  \includegraphics[width=0.19\linewidth]{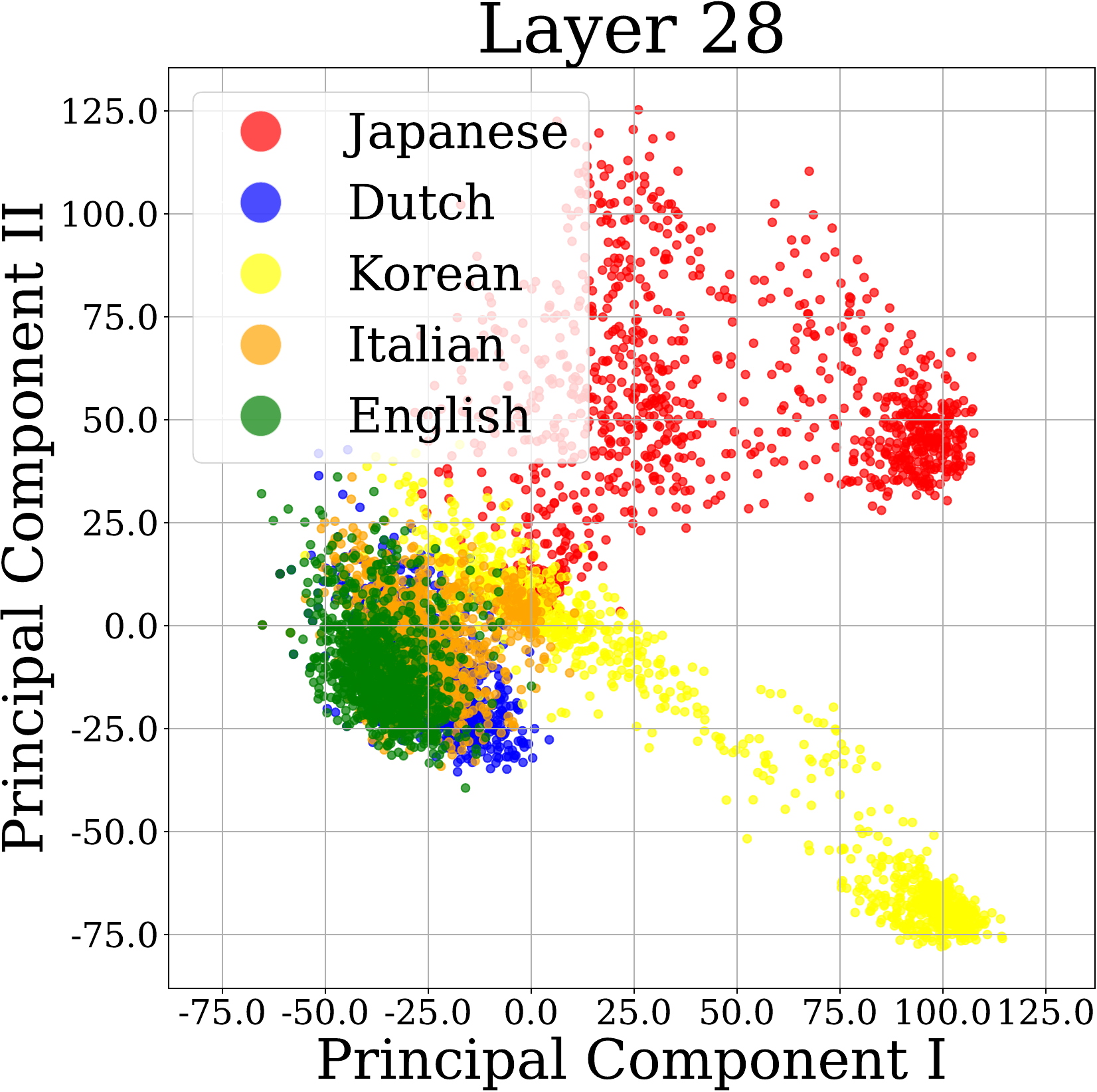}
  \includegraphics[width=0.19\linewidth]{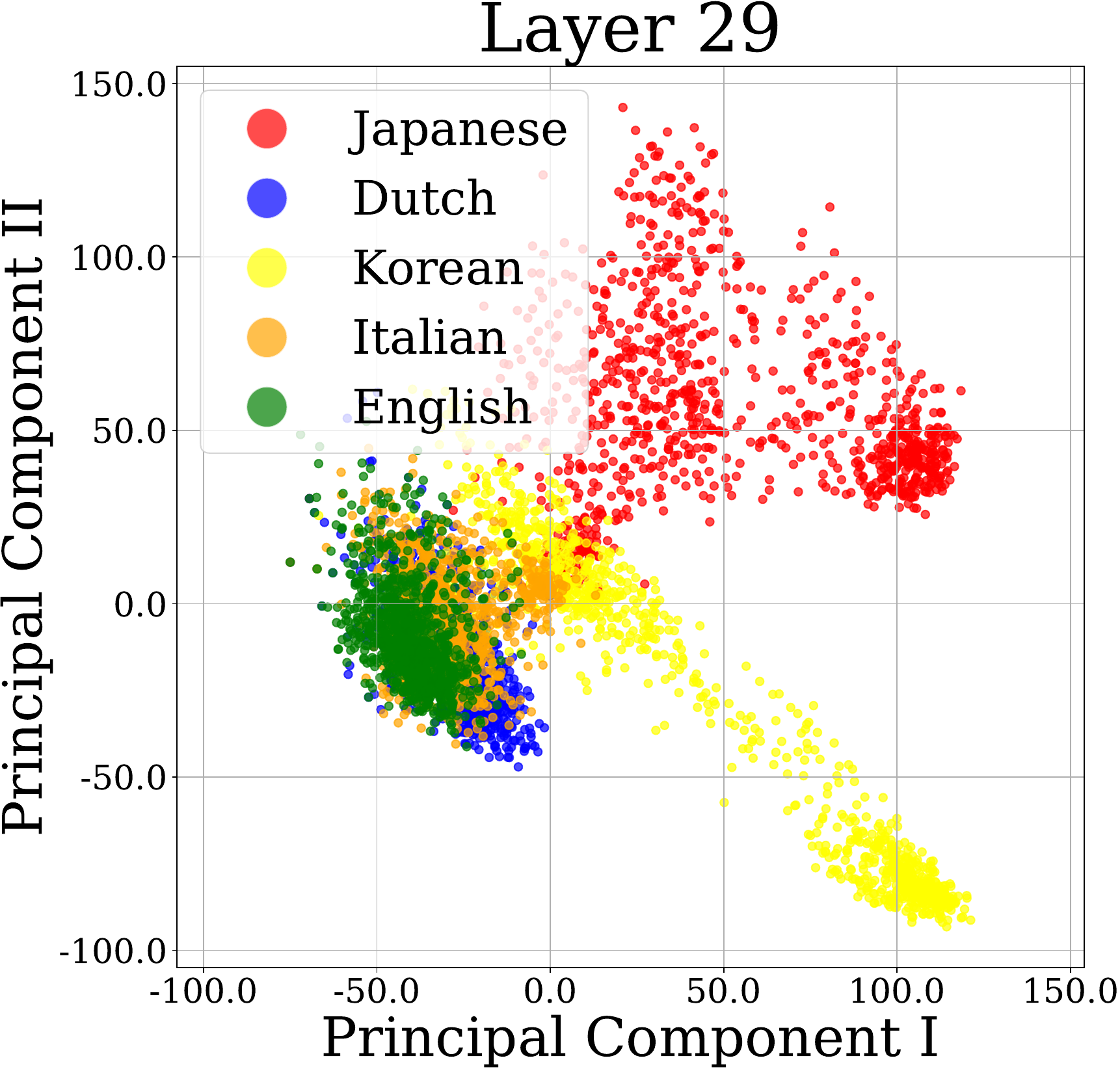}

  \includegraphics[width=0.19\linewidth]{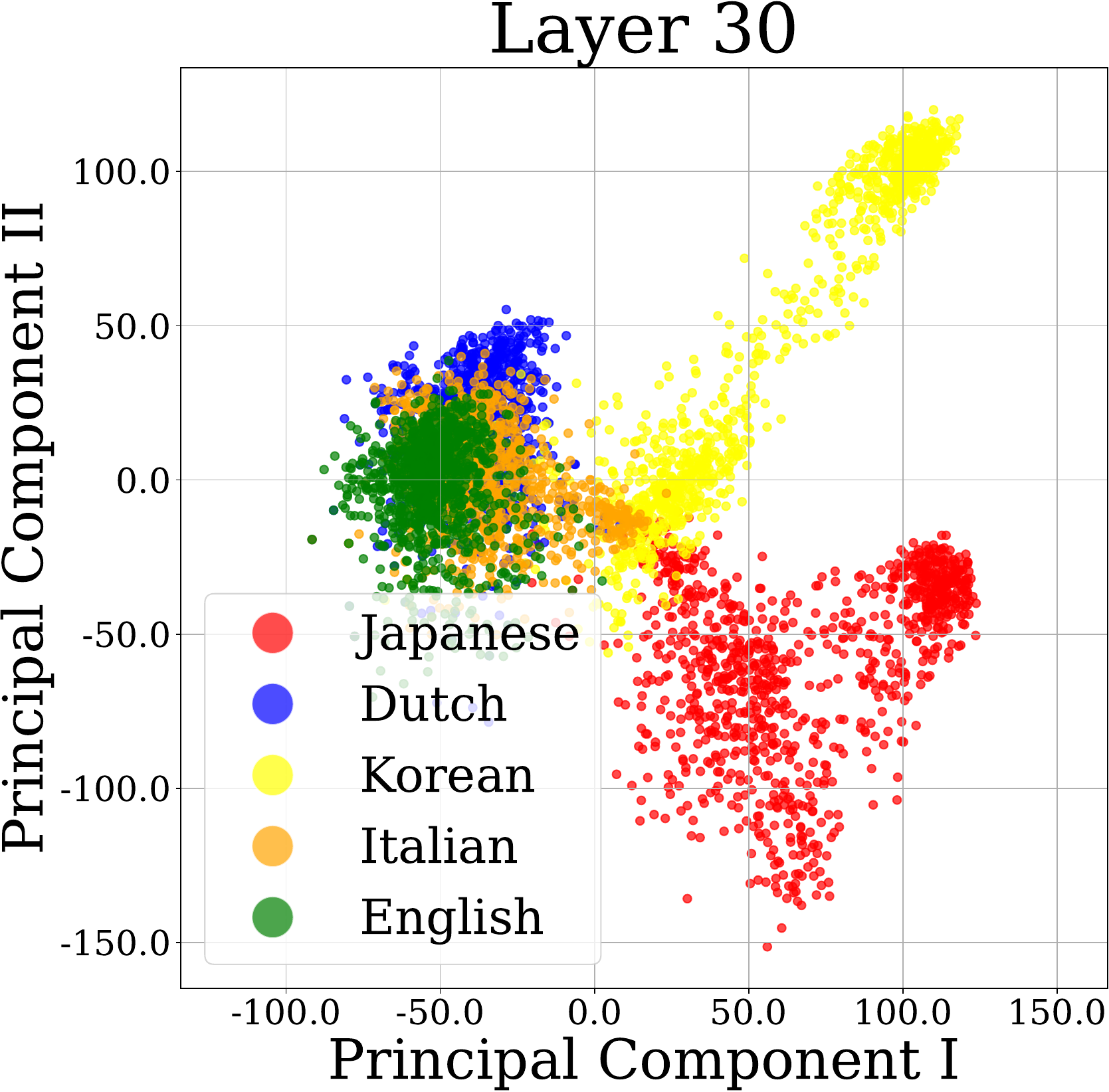}
  \includegraphics[width=0.19\linewidth]{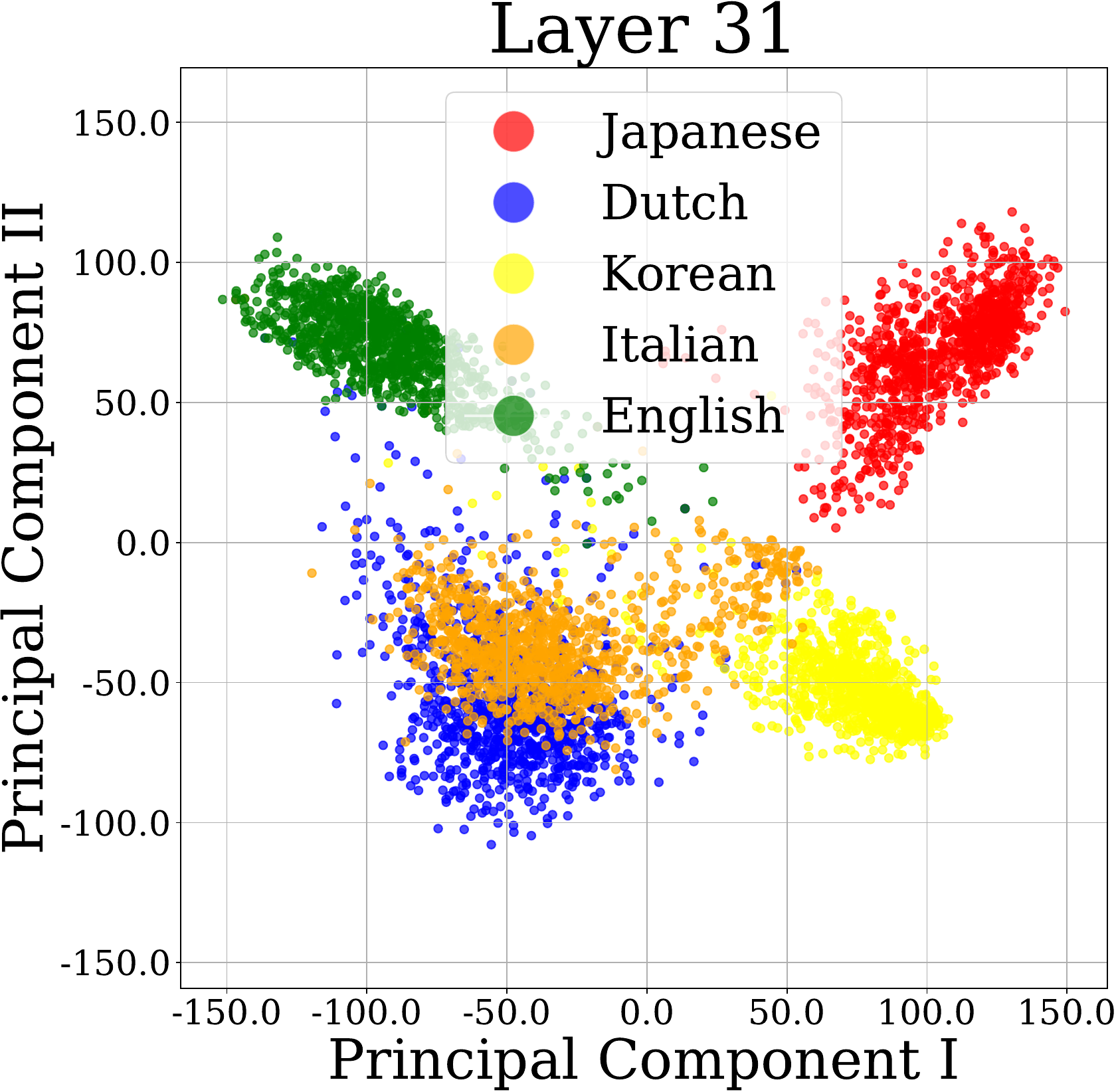}
  \includegraphics[width=0.19\linewidth]{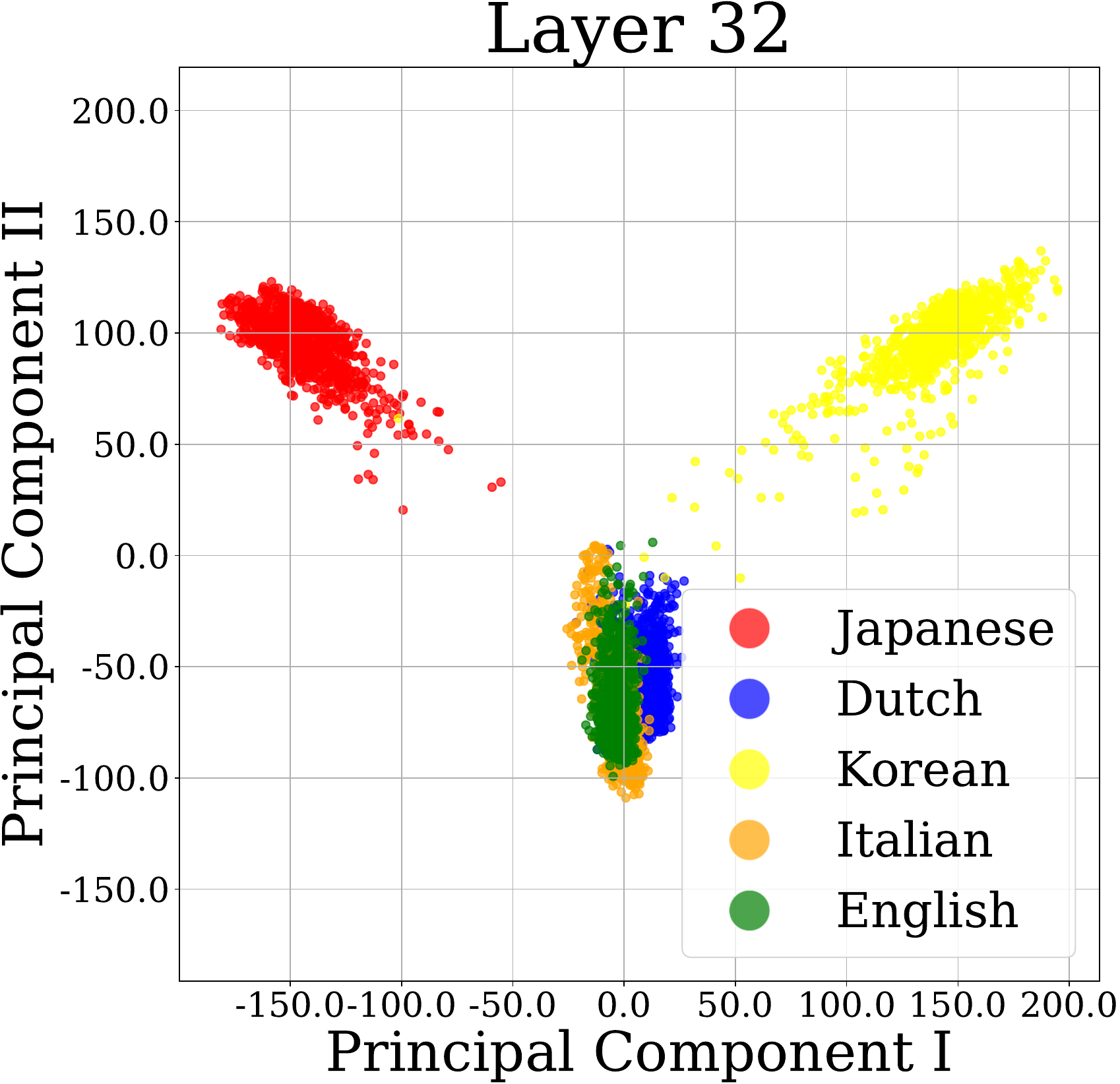}

  \caption{\textbf{The resutls of PCA applied to the hidden representations across layers (Aya expanse-8B).}}
  \label{fig:appendix:pca_all_layers_aya}
\end{figure*}

Figs.~\ref{fig:appendix:pca_all_layers_llama3}, \ref{fig:appendix:pca_all_layers_mistral}, and \ref{fig:appendix:pca_all_layers_aya} present the results of PCA applied to the hidden language representations across all layers of each model.
% PCA results, all layers, llama3
% PCA results, all layers, mistral
% PCA results, all layers, aya

\section{Detecting and Controlling Transfer Neurons}
\subsection{The Definition of Neurons in This Study}
\label{sec:appendix:the definition of neurons}
While several studies \citep{suau2022selfcond, tang_lang_specific_neurons, wang2024sharing, hiraoka-inui-2025-repetition, mondal-etal-2025-language-specific} have investigated neuron-level analysis or identification in LLMs, these studies primarily regard the output of the non-linear activation function as activation values and treat them as the fundamental unit of neurons.
Throughout the experiments in this study, however, we defined neurons as the activation values $\alpha^{l}_i$ denoted in Eqs.~\ref{eq:mlp} and~\ref{eq:weighted_sum} (i.e., the output of element-wise product). We adopted this definition because, even when the output of the activation function is large, the actual activation value $\alpha_i^{l}$ can be small depending on the output of the up projection, given that the models we adopted has a gated-MLP. As a result, the impact on subsequent neurons may be attenuated, which renders $\alpha_i^{l}$ a more appropriate indicator of neuron activity in the context of our analysis.

\subsection{Deactivating Type-1 Transfer Neurons}

\subsubsection{Kernel-Based Similarity between Latent Spaces}
\label{sec:appendix:kernel-based sim between spaces while deactivating Type-1}
\begin{figure*}[t]
    \centering
    % llama3, n1000, Type-1
    \includegraphics[width=0.32\linewidth]{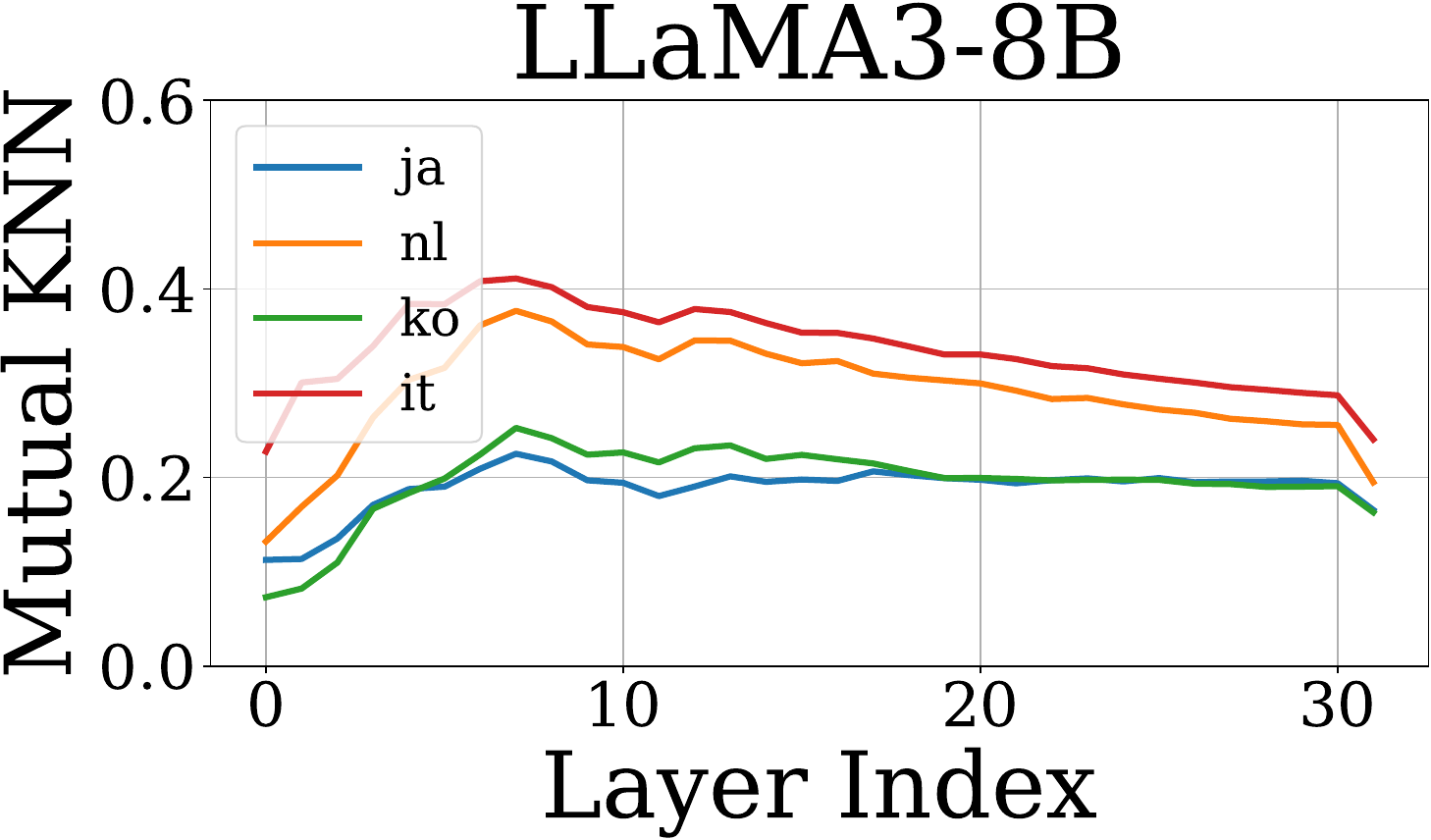}
    \includegraphics[width=0.32\linewidth]{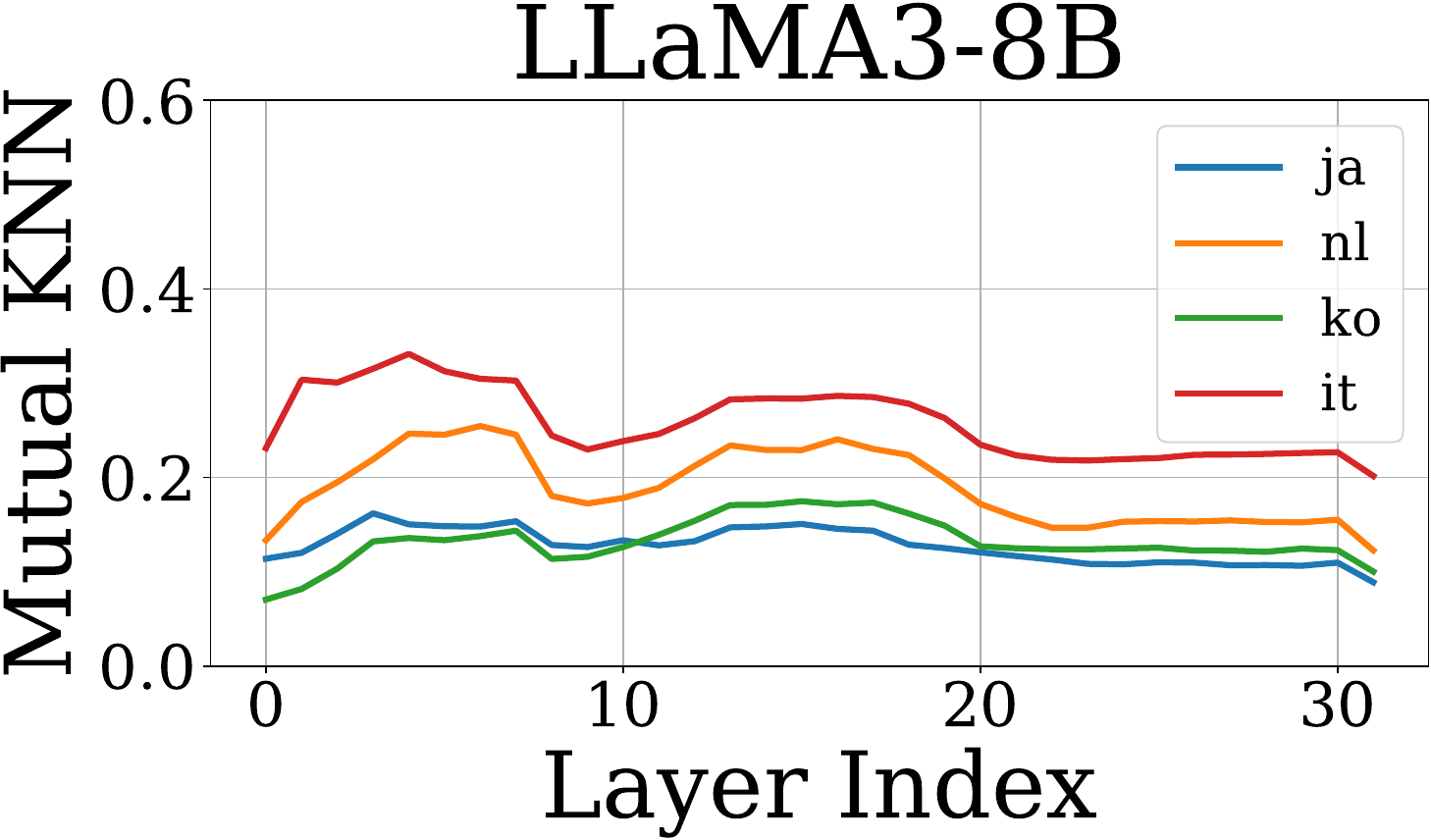}
    \includegraphics[width=0.32\linewidth]{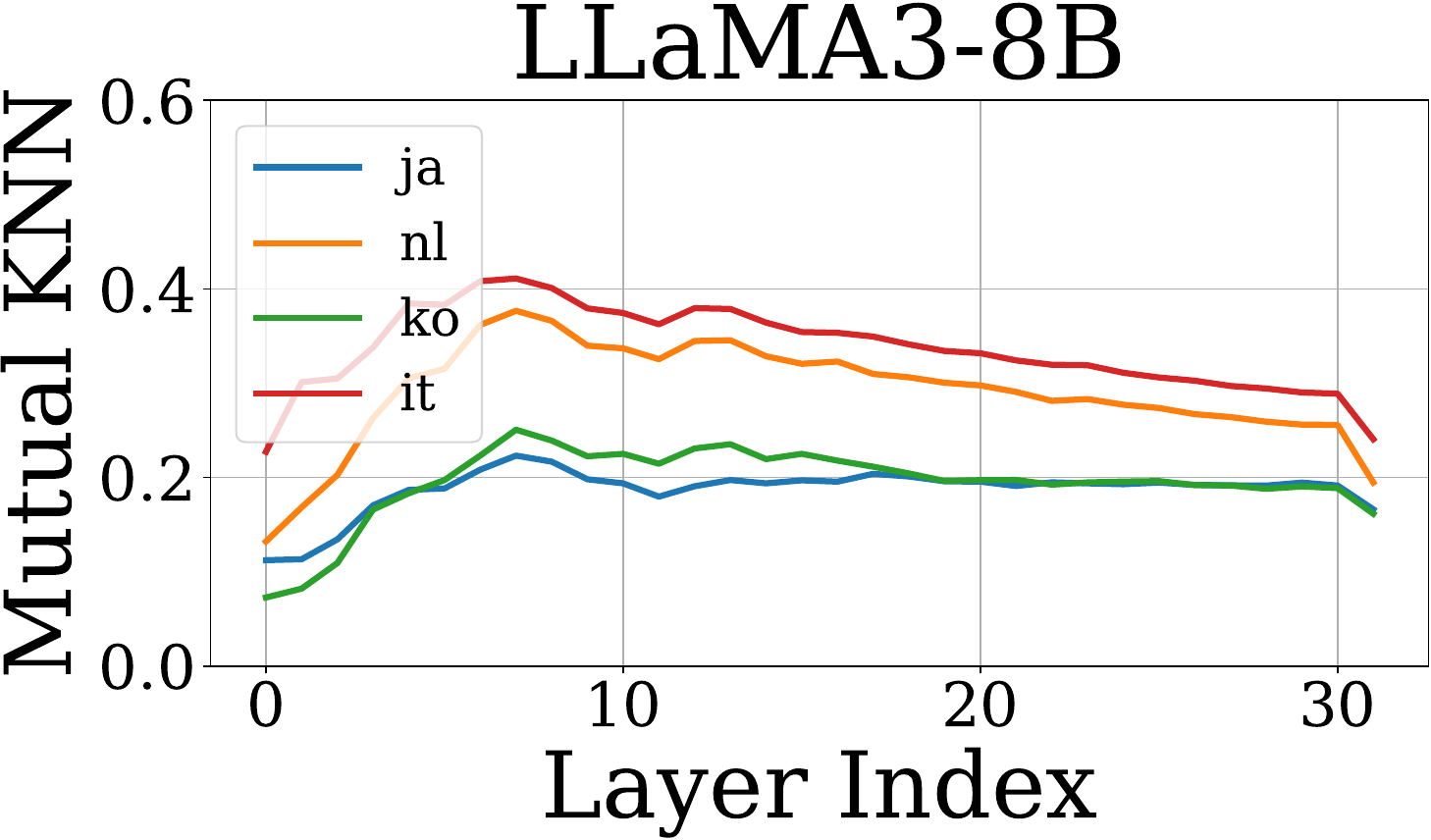}
    \begin{minipage}{0.32\linewidth}\centering w/o intervention\end{minipage}
    \begin{minipage}{0.32\linewidth}\centering \textbf{top-1k deactivated}\end{minipage}
    \begin{minipage}{0.32\linewidth}\centering baseline\end{minipage}
    
    % mistral, n1000, Type-1
    \includegraphics[width=0.32\linewidth]{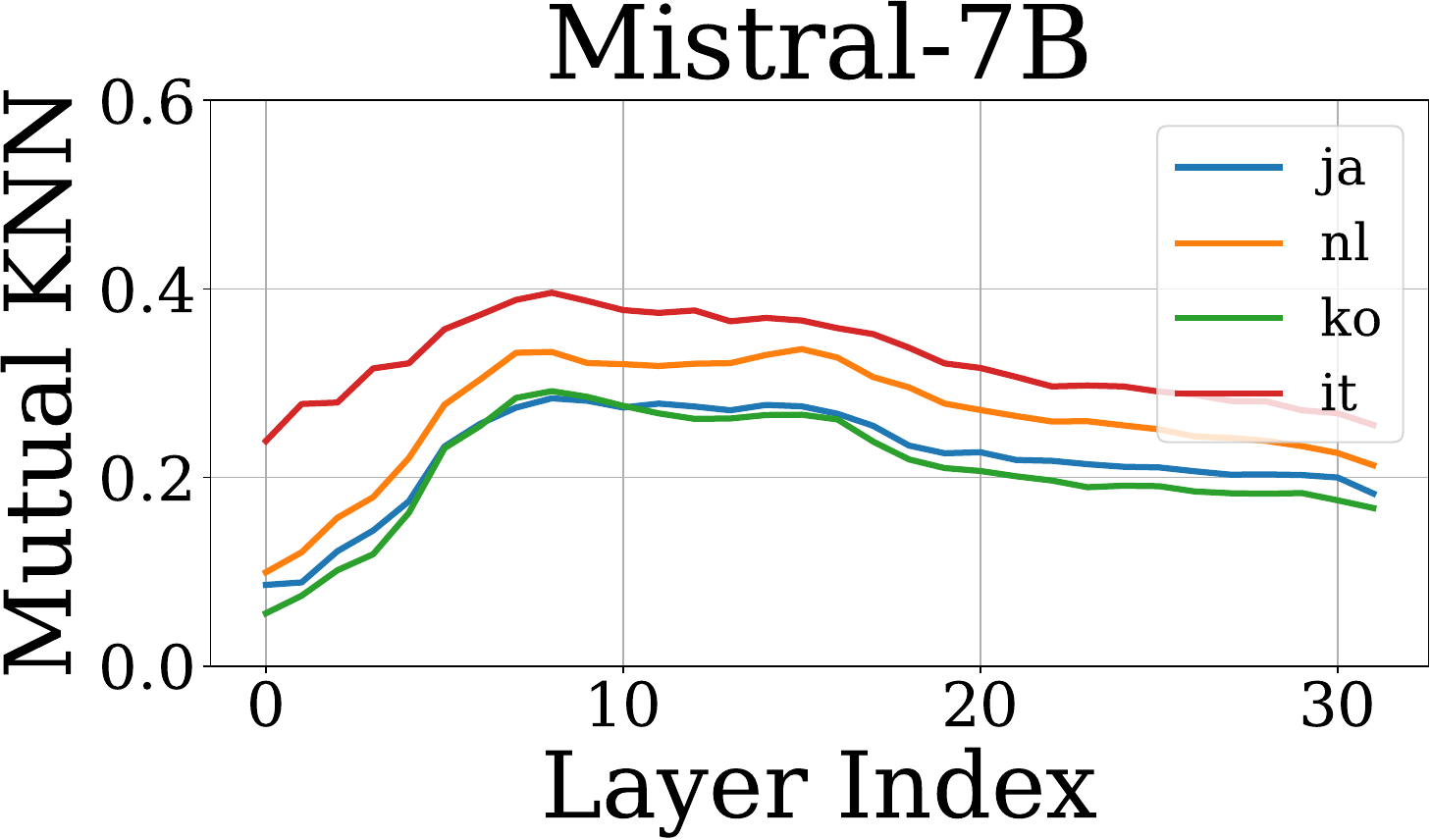}
    \includegraphics[width=0.32\linewidth]{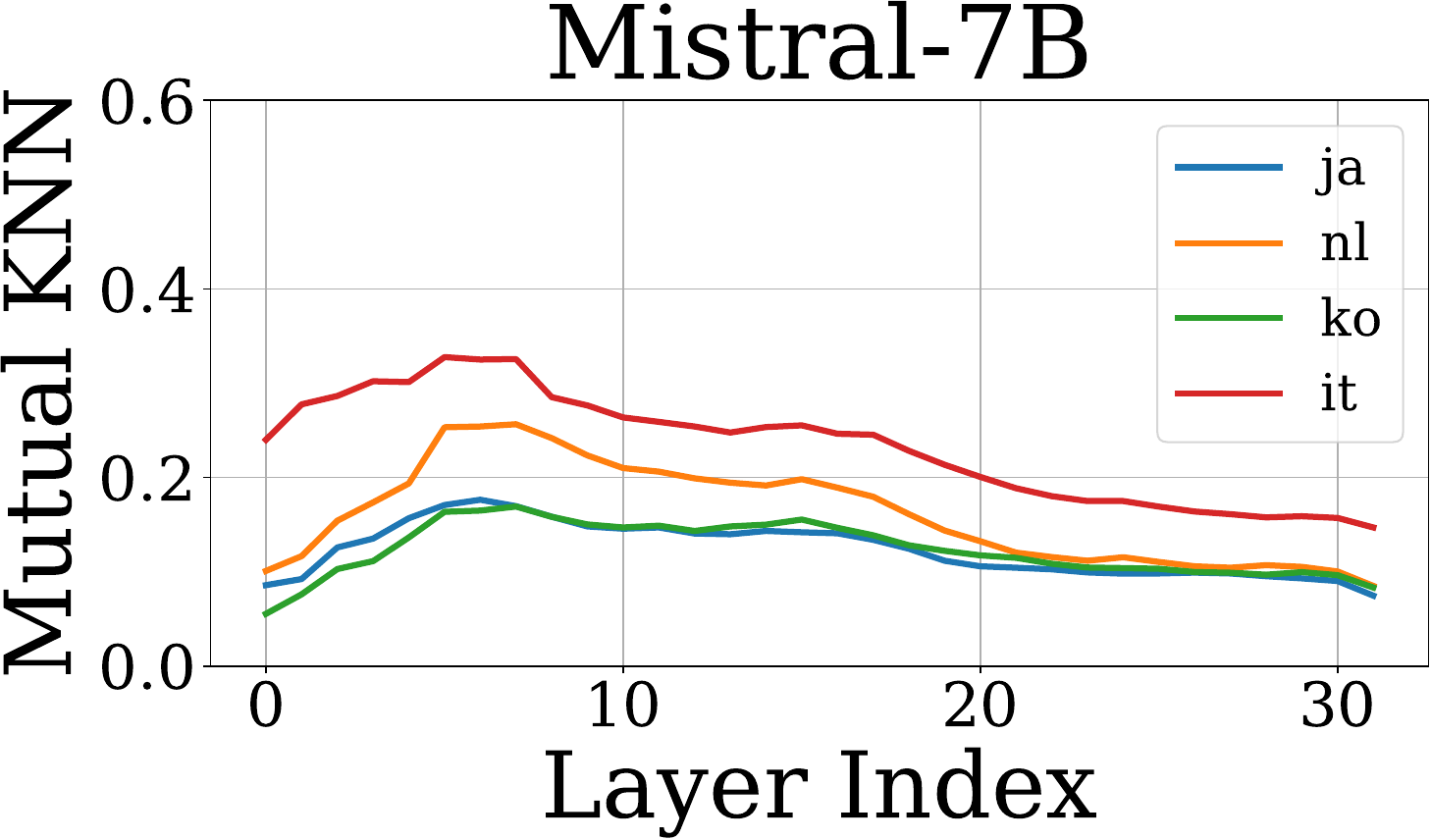}
    \includegraphics[width=0.32\linewidth]{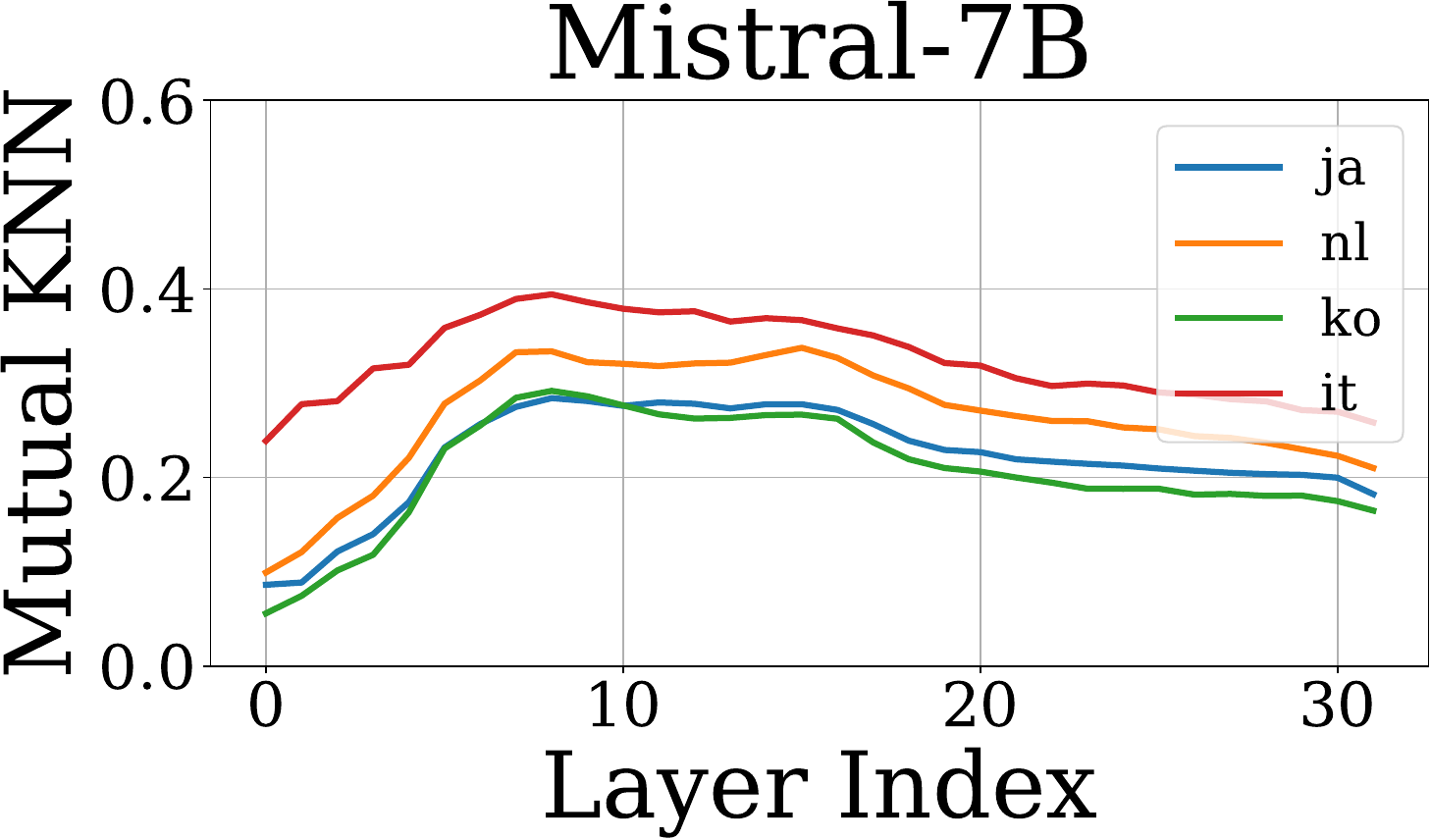}
    \begin{minipage}{0.32\linewidth}\centering w/o intervention\end{minipage}
    \begin{minipage}{0.32\linewidth}\centering \textbf{top-1k deactivated}\end{minipage}
    \begin{minipage}{0.32\linewidth}\centering baseline\end{minipage}

    % aya, n1000, Type-1
    \includegraphics[width=0.32\linewidth]{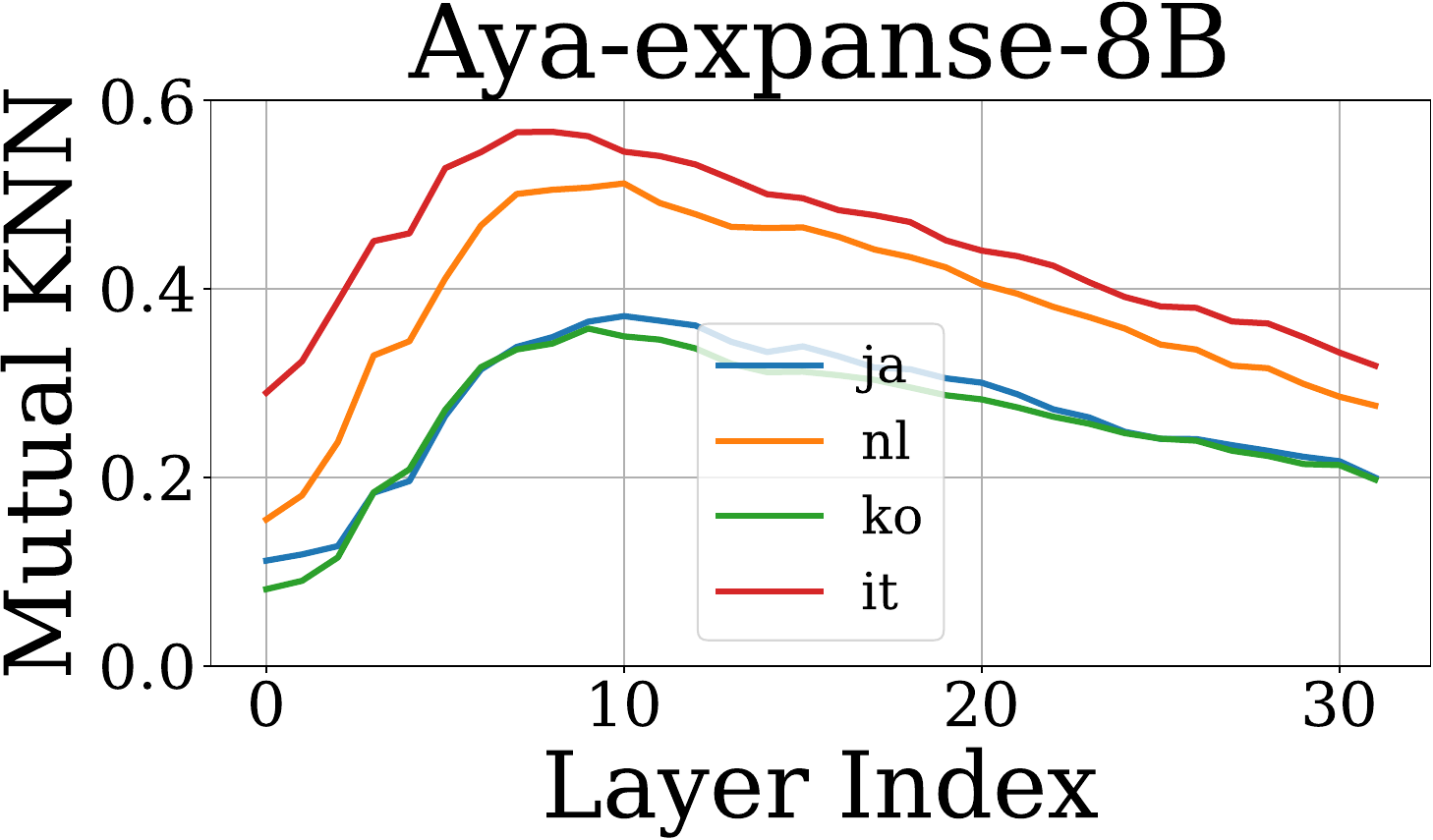}
    \includegraphics[width=0.32\linewidth]{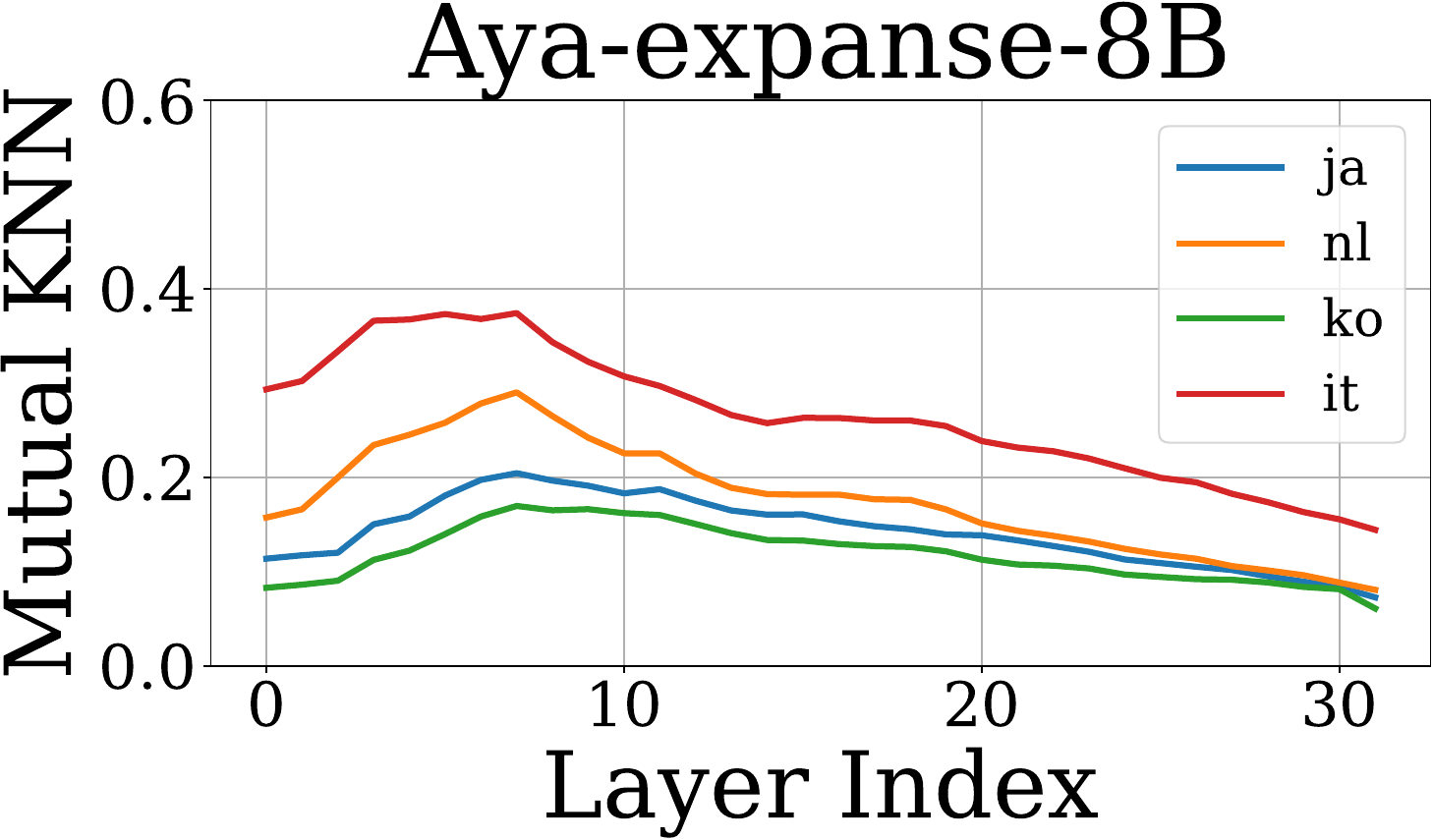}
    \includegraphics[width=0.32\linewidth]{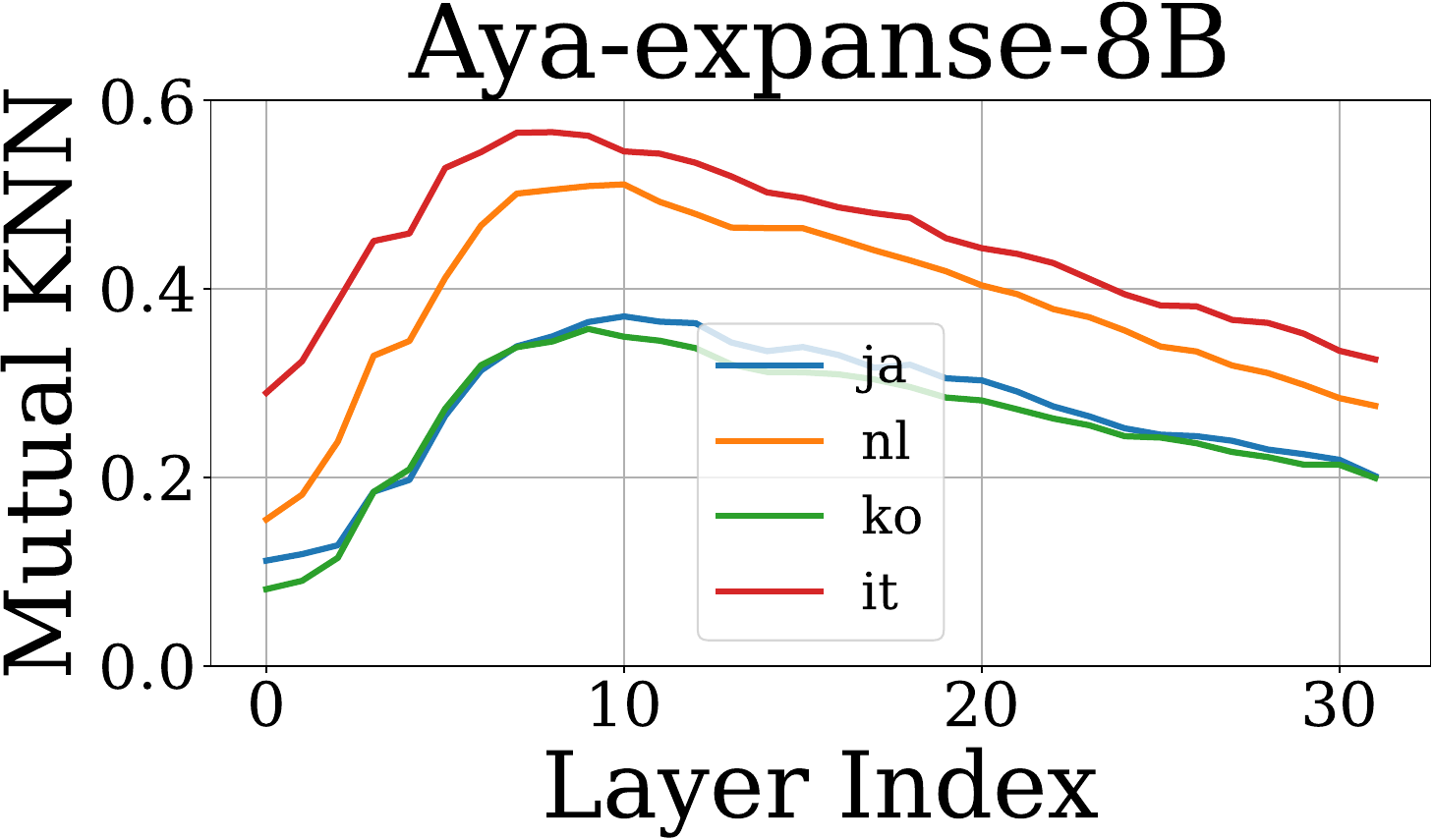}
    \begin{minipage}{0.32\linewidth}\centering w/o intervention\end{minipage}
    \begin{minipage}{0.32\linewidth}\centering \textbf{top-1k deactivated}\end{minipage}
    \begin{minipage}{0.32\linewidth}\centering baseline\end{minipage}
    
    \caption{\textbf{Kernel-Based Similarity with Deactivation of Top-1k Type-1 Transfer Neurons (k=10).} The middle column presents the results after deactivating the top-1k Type-1 Transfer Neurons. The right column shows the results after deactivating 1k randomly sampled neurons from the same layers as Type-1 neurons for a baseline, while the left column shows the original results without any intervention.}
    \label{fig:appendix:kernel-based sim while deactivating Type-1 k=10}
\end{figure*}

\begin{figure*}[t]
    \centering
    % llama3, n1000, Type-1
    \includegraphics[width=0.32\linewidth]{figures/llama3/knn/k=5/normal.pdf}
    \includegraphics[width=0.32\linewidth]{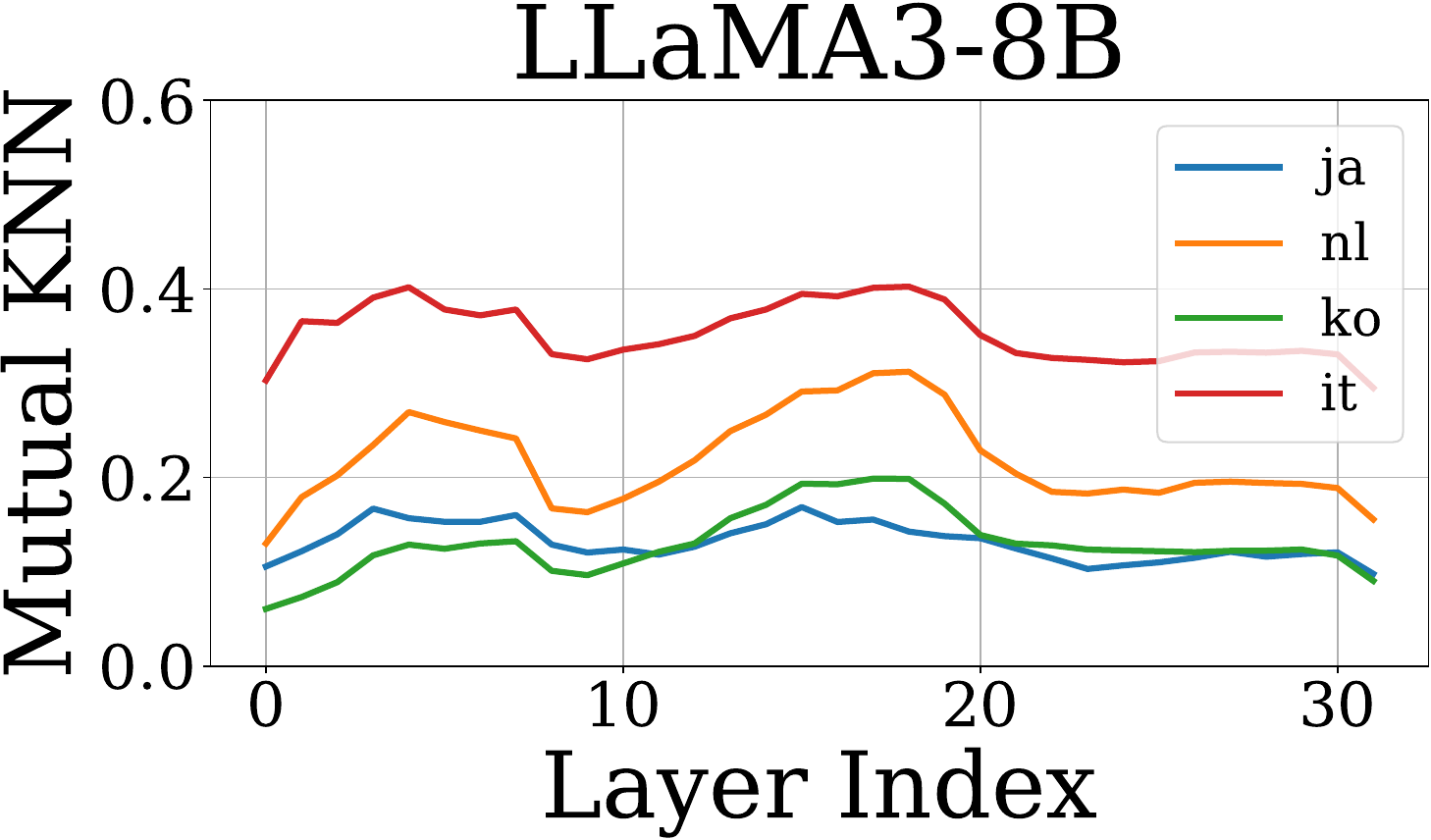}
    \includegraphics[width=0.32\linewidth]{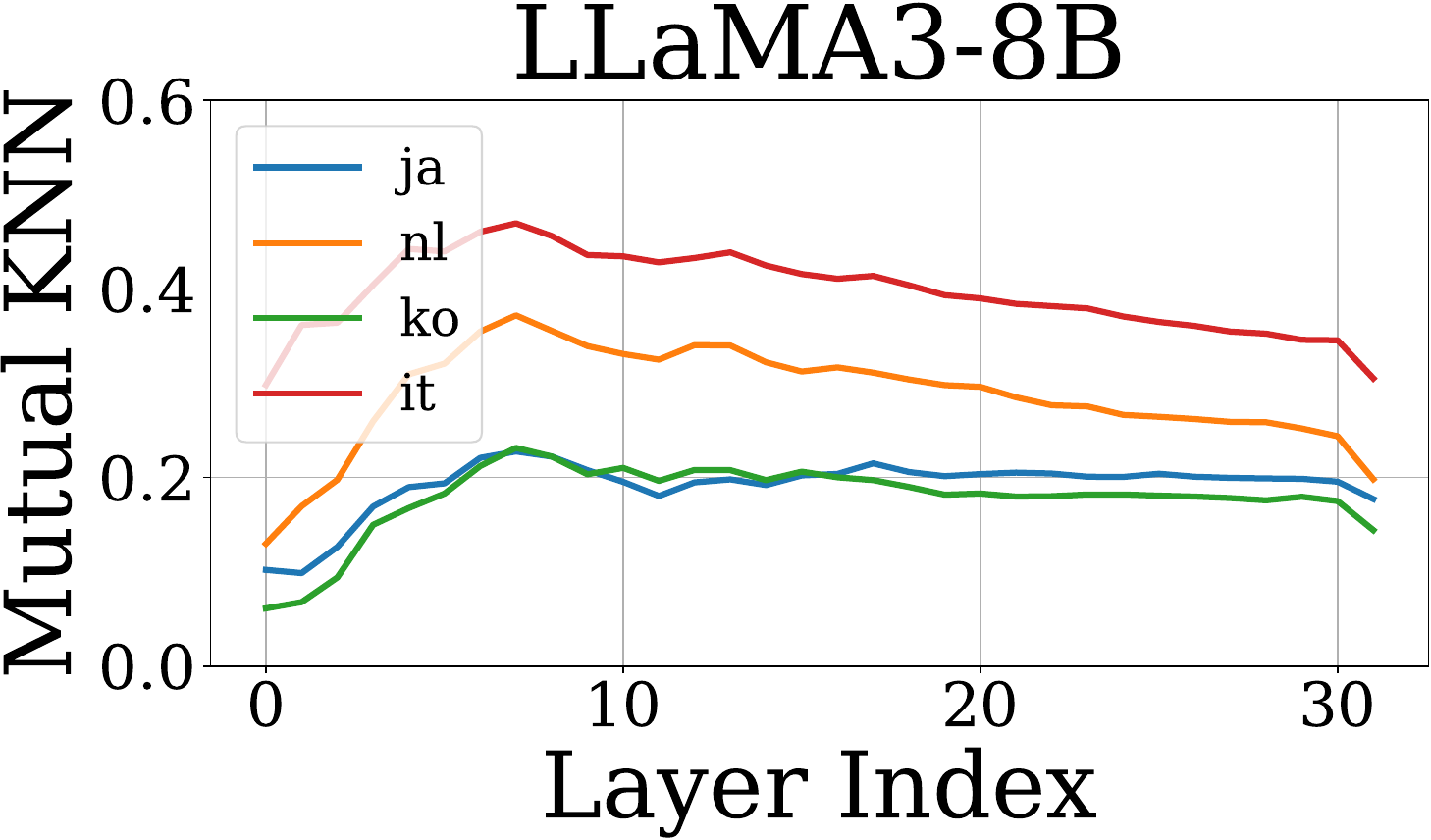}
    \begin{minipage}{0.32\linewidth}\centering w/o intervention\end{minipage}
    \begin{minipage}{0.32\linewidth}\centering \textbf{top-1k deactivated}\end{minipage}
    \begin{minipage}{0.32\linewidth}\centering baseline\end{minipage}
    
    % mistral, n1000, Type-1
    \includegraphics[width=0.32\linewidth]{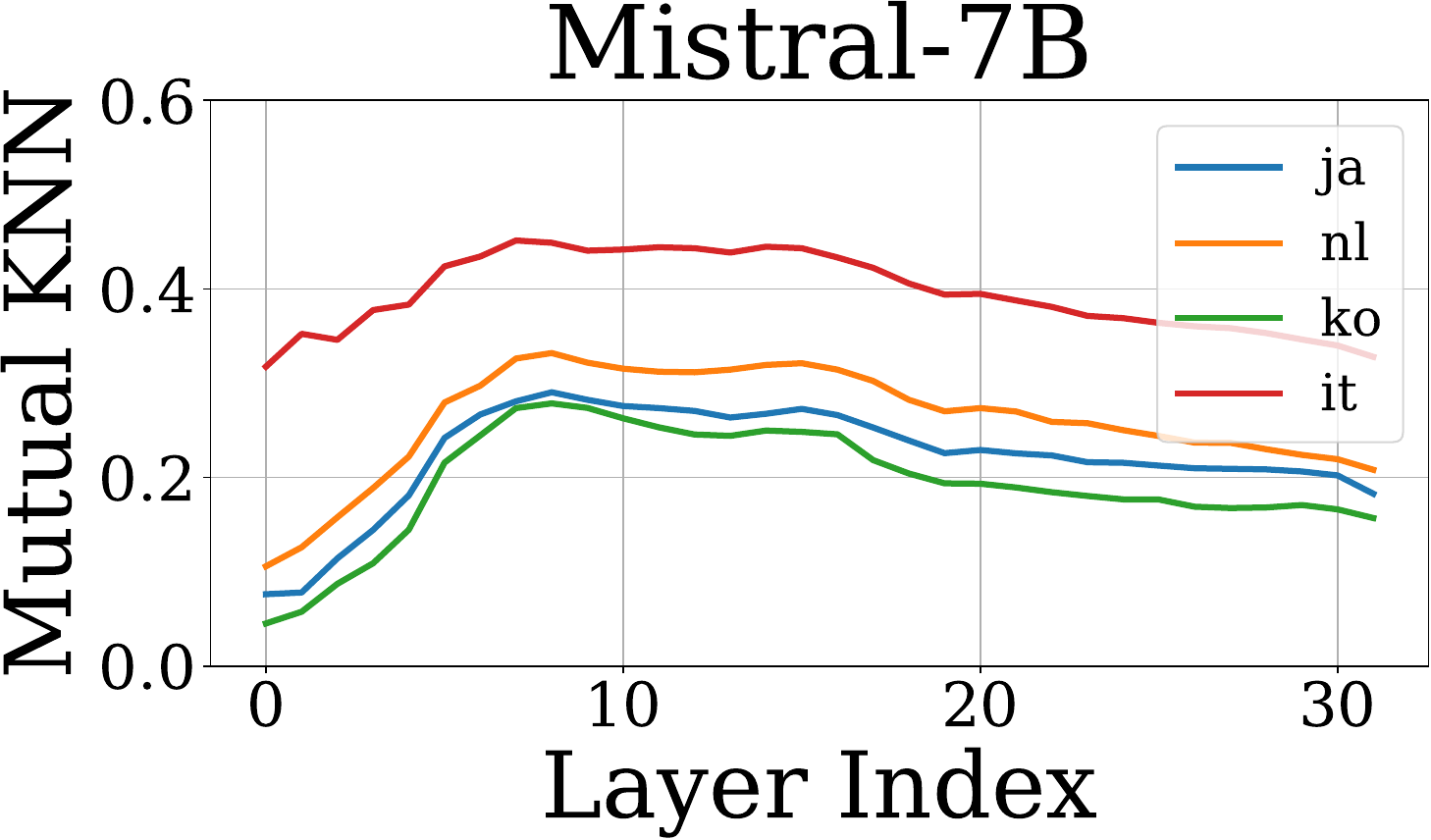}
    \includegraphics[width=0.32\linewidth]{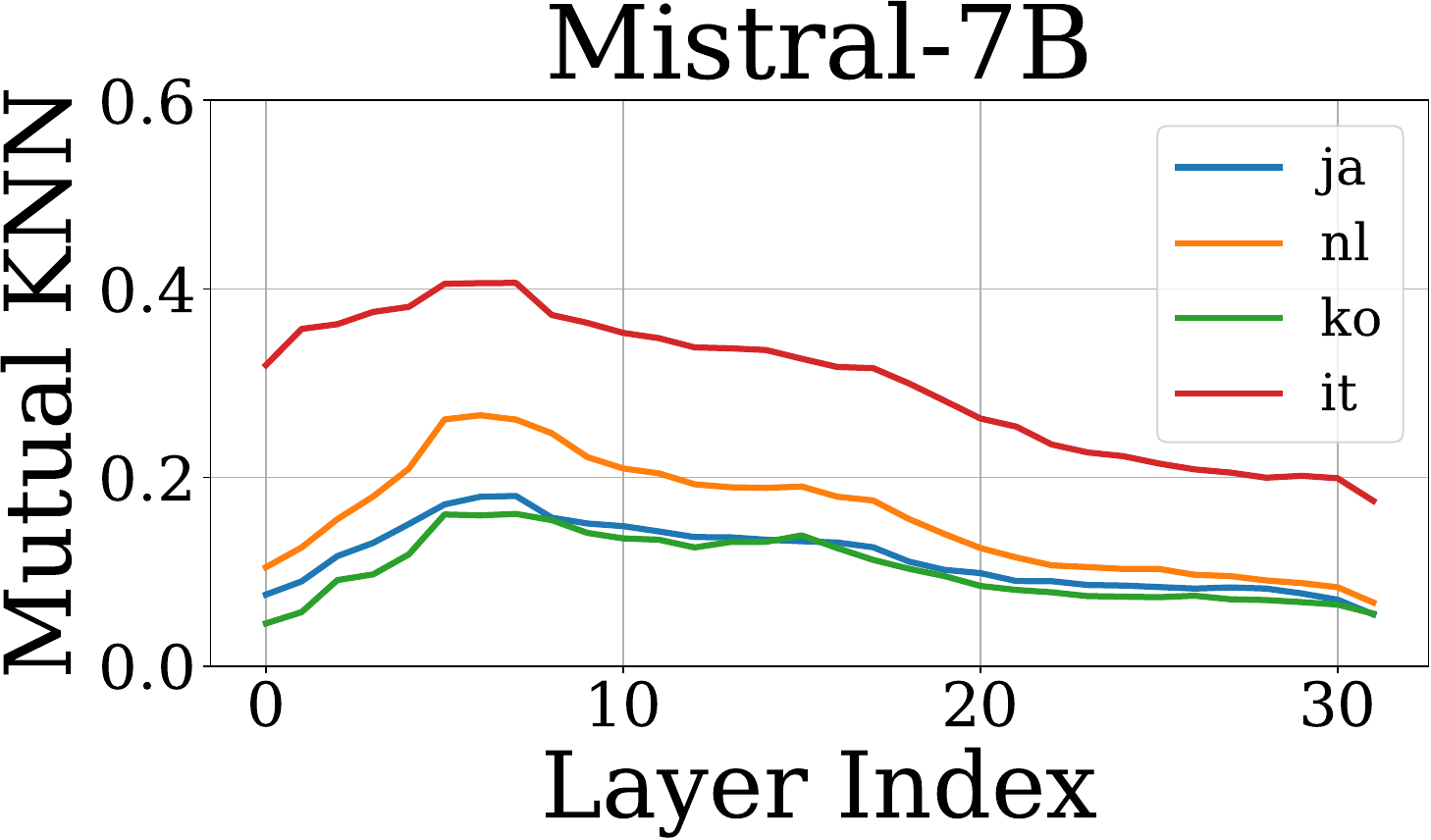}
    \includegraphics[width=0.32\linewidth]{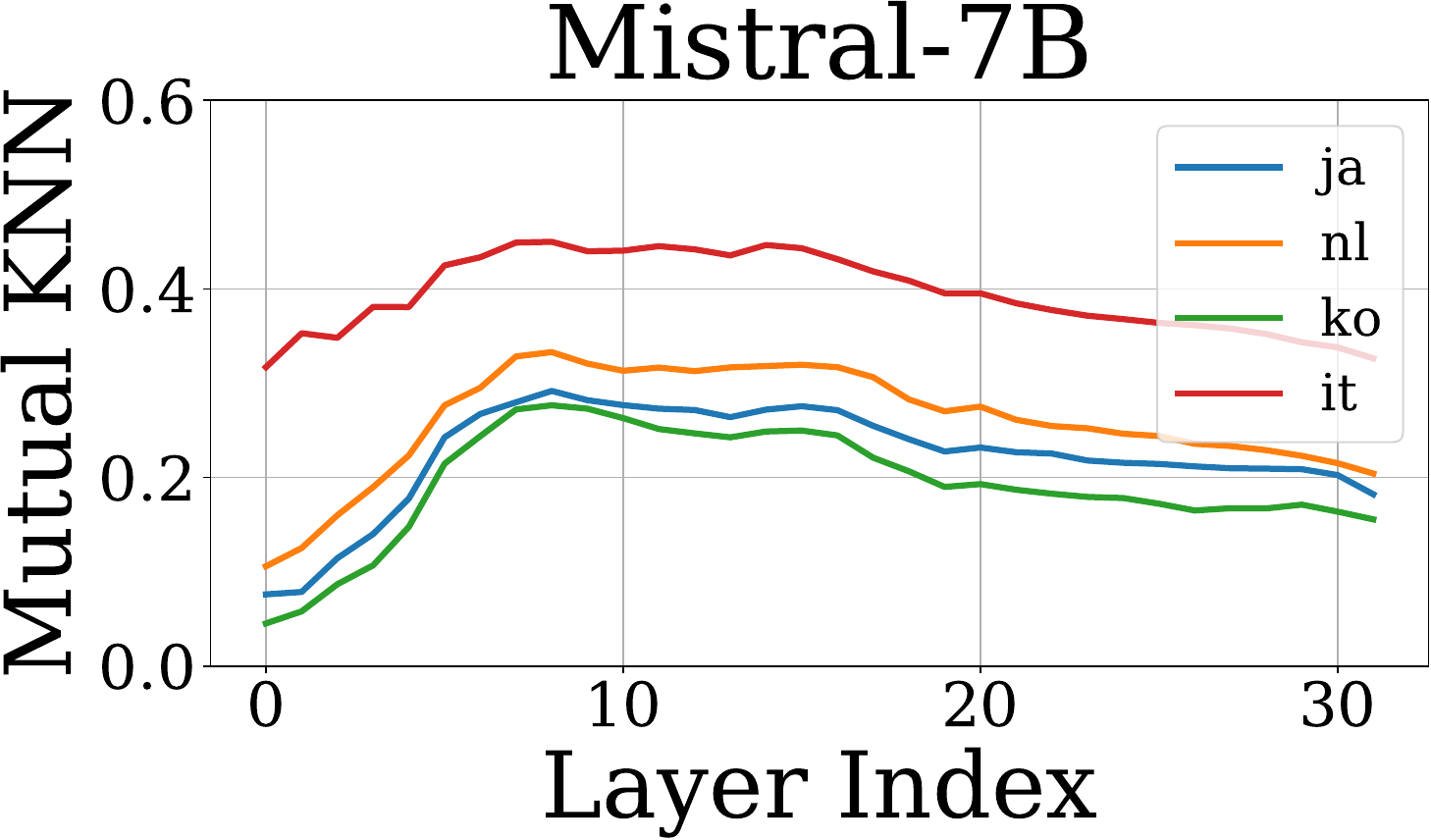}
    \begin{minipage}{0.32\linewidth}\centering w/o intervention\end{minipage}
    \begin{minipage}{0.32\linewidth}\centering \textbf{top-1k deactivated}\end{minipage}
    \begin{minipage}{0.32\linewidth}\centering baseline\end{minipage}

    % aya, n1000, Type-1
    \includegraphics[width=0.32\linewidth]{figures/aya/knn/k=5/normal.pdf}
    \includegraphics[width=0.32\linewidth]{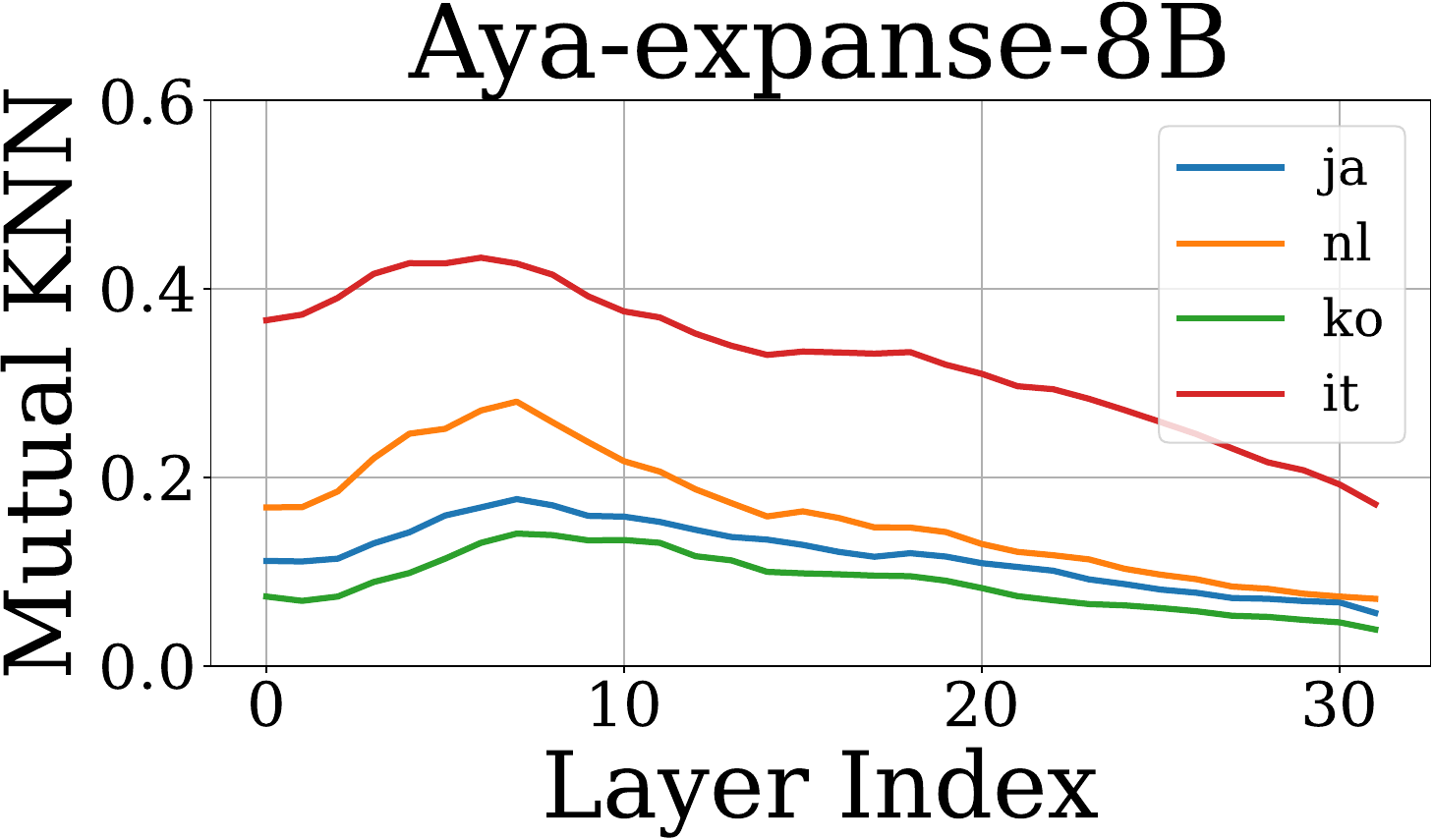}
    \includegraphics[width=0.32\linewidth]{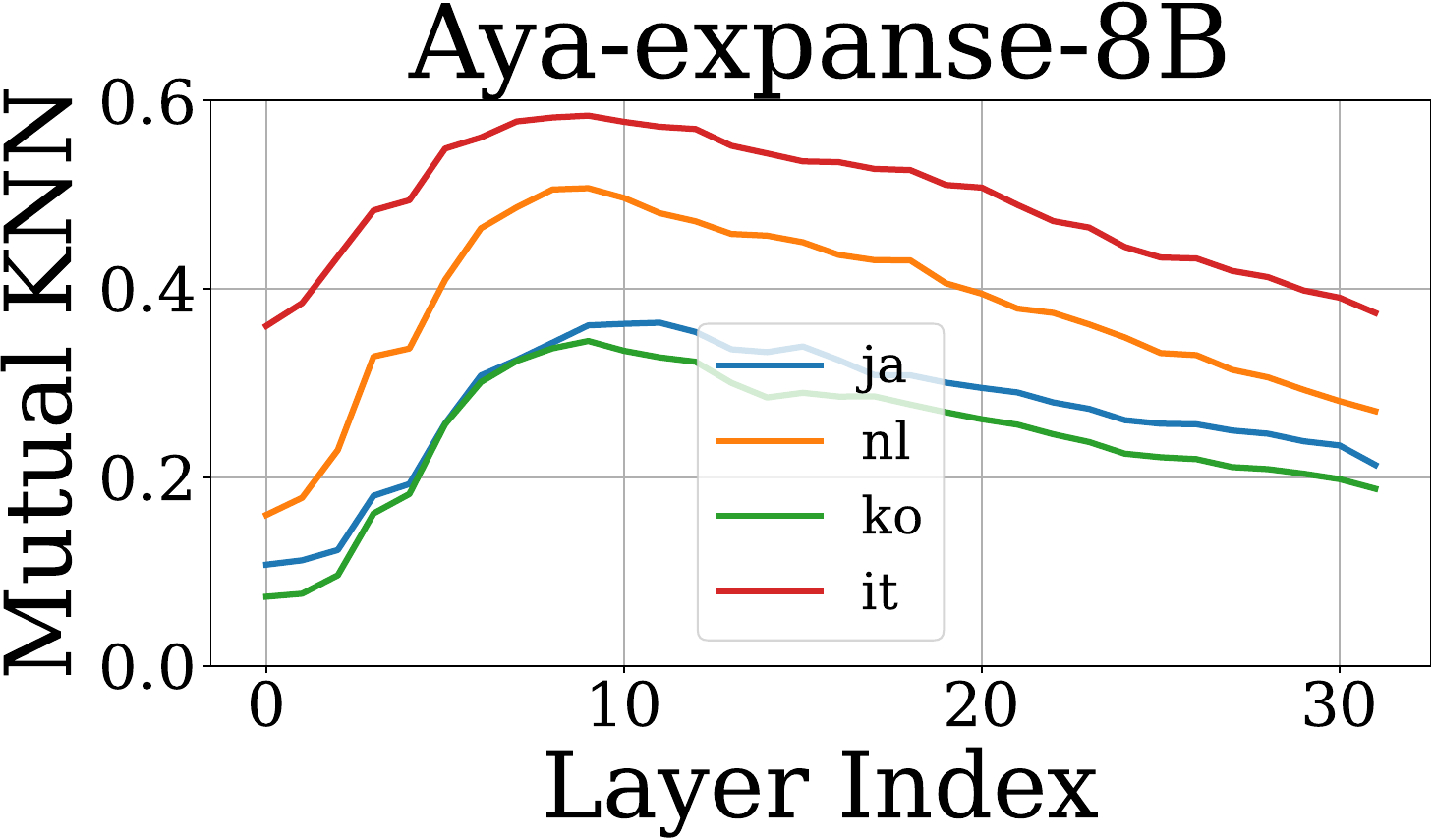}
    \begin{minipage}{0.32\linewidth}\centering w/o intervention\end{minipage}
    \begin{minipage}{0.32\linewidth}\centering \textbf{top-1k deactivated}\end{minipage}
    \begin{minipage}{0.32\linewidth}\centering baseline\end{minipage}
    
    \caption{\textbf{Kernel-Based Similarity with Deactivation of Top-1k Type-1 Transfer Neurons (k=5).}}
    \label{fig:appendix:kernel-based sim while deactivating Type-1 k=5}
\end{figure*}

The middle columns of Figs.~\ref{fig:appendix:kernel-based sim while deactivating Type-1 k=10} and~\ref{fig:appendix:kernel-based sim while deactivating Type-1 k=5} presents the results of kernel-based similarity between English and L2 latent spaces while deactivating top-1k Type-1 neurons, computed using the Mutual \textit{k}-NN Alignment Metric formalized in Appendix~\ref{sec:appendix:formalizing mutual_knn}. As shown, deactivating the top-1k Type-1 neurons leads to a substantial drop in similarity, particularly in the middle layers where similarity originally peaked — suggesting the presence of a shared semantic latent space. This also suggests that the Type-1 neurons identified in \S\ref{sec:detecting and controlling transfer neurons} play a pivotal role in shifting representations towards this shared latent space. As explained in \S\ref{sec:Similarity Measurement While Deactivating Transfer Neurons}, we surmise that the relatively high remaining similarity in the en-nl (Dutch) and en-it (Italian) settings — even after deactivating Type-1 neurons — suggests that their specific latent spaces are already close to each other in even early layers, as shown in the PCA results (\S\ref{Visualizing the Hidden language Subspace with PCA}).

% figure: mutual knn (k=10) while deactivating top-1k Type-1 TN.
% figure: mutual knn (k=5) while deactivating top-1k Type-1 TN.

\subsubsection{Hidden States and Activation Patterns Similarity}
\label{sec:sim_deactivating_Type-1_neurons_appendix}
\begin{figure*}[t]
    \centering

    \begin{minipage}{0.23\linewidth}
      \centering
      \includegraphics[width=\linewidth]{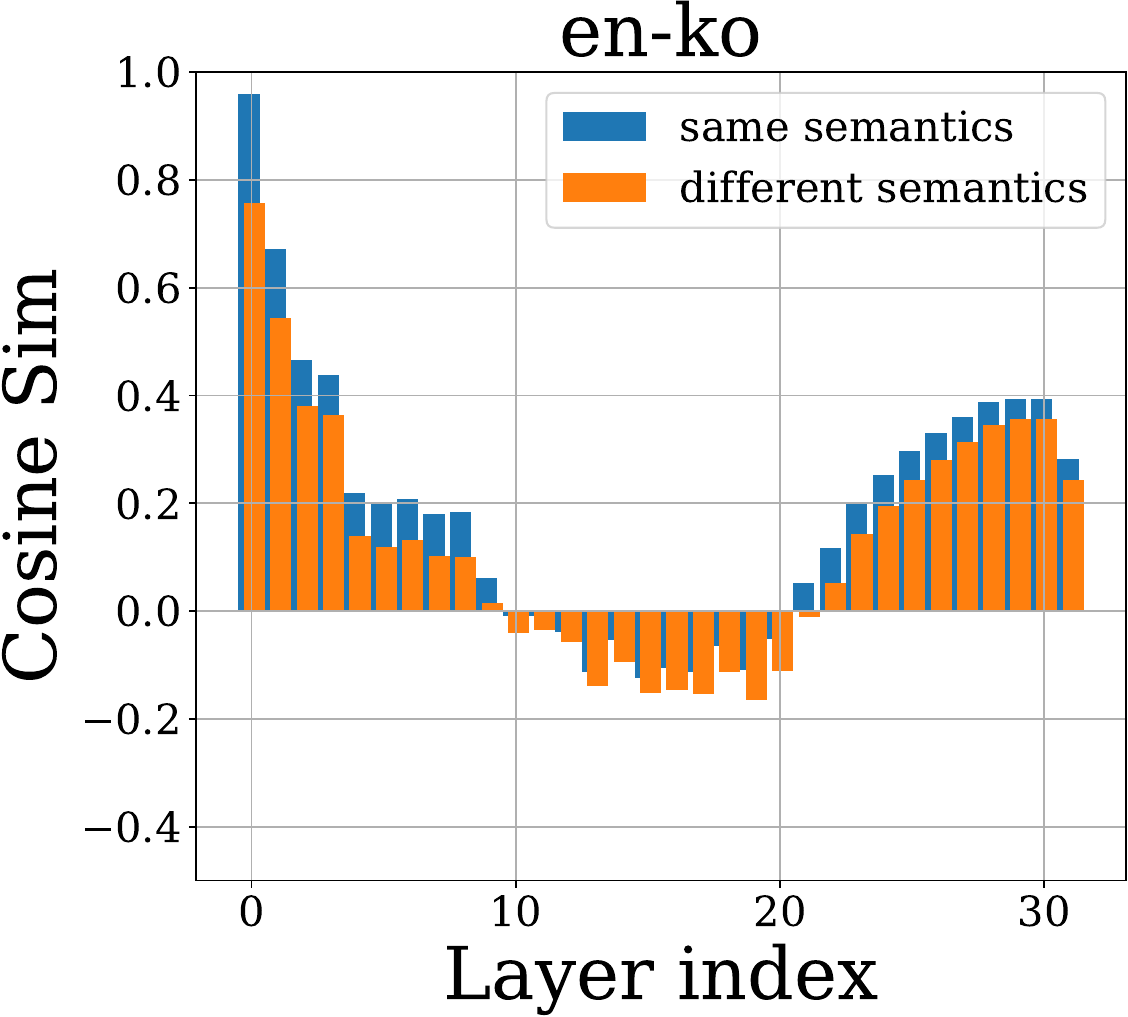}
      \subcaption{en-ko}
    \end{minipage}
    \begin{minipage}{0.23\linewidth}
      \centering
      \includegraphics[width=\linewidth]{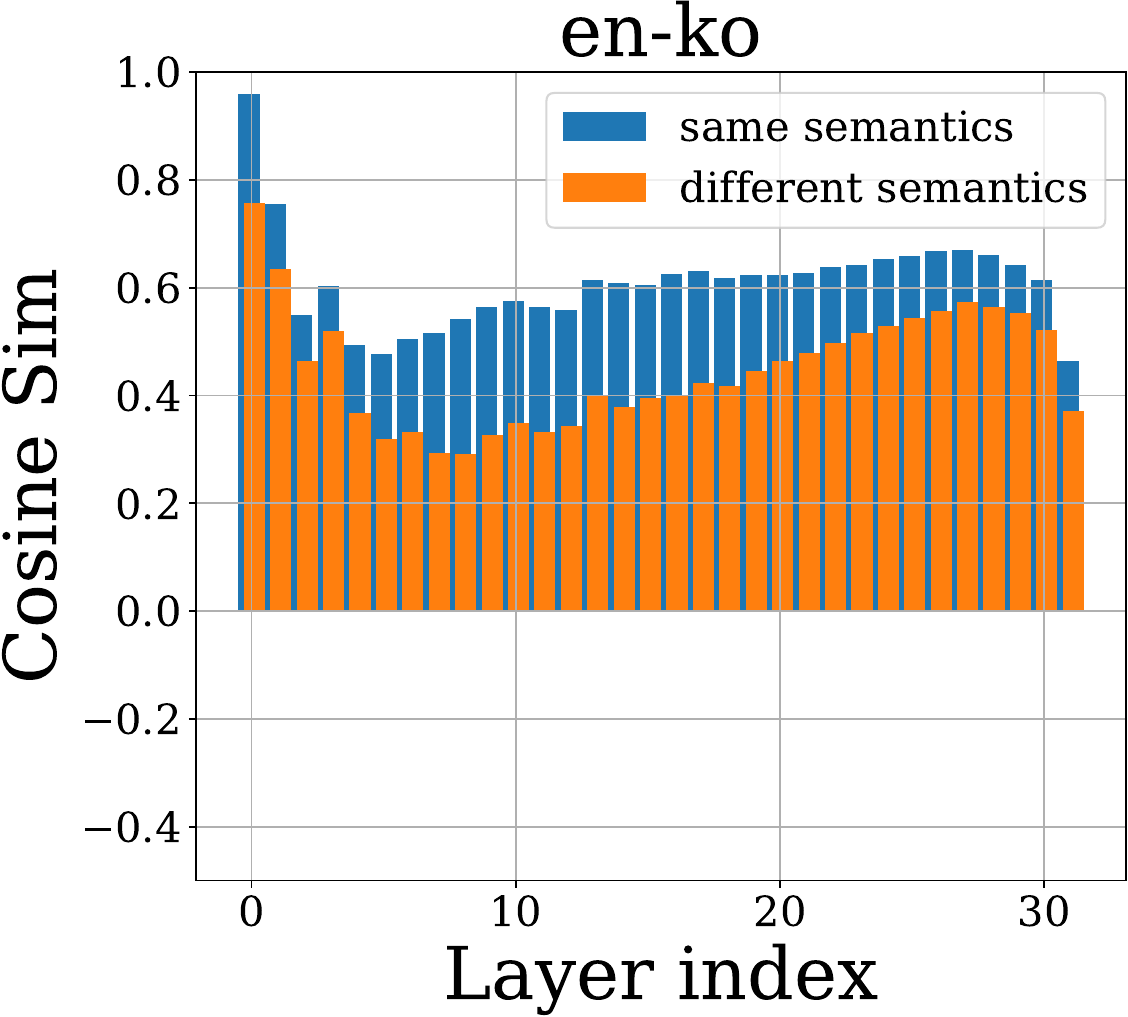}
      \subcaption{en-ko (baseline)}
    \end{minipage}
    \begin{minipage}{0.23\linewidth}
      \centering
      \includegraphics[width=\linewidth]{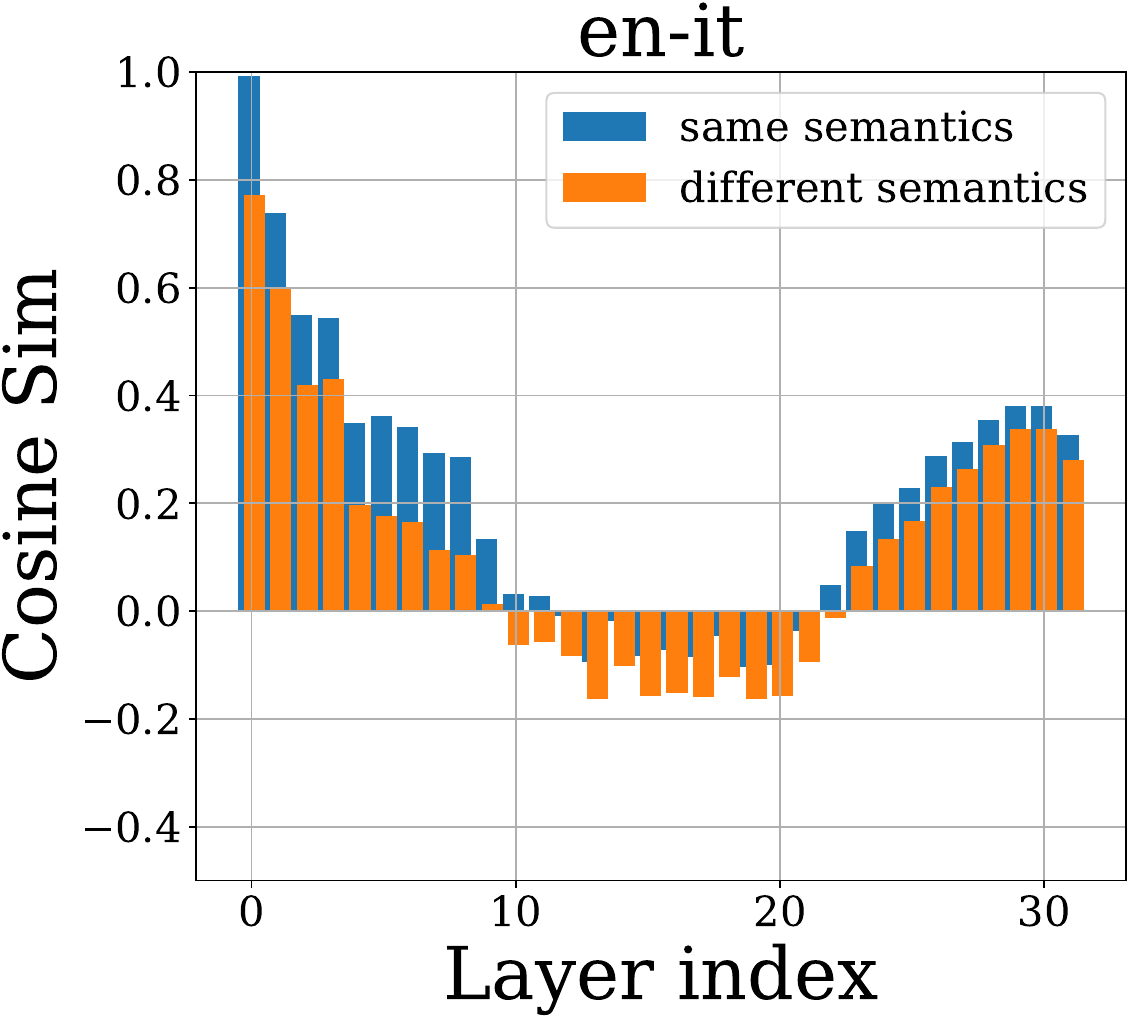}
      \subcaption{en-it}
    \end{minipage}
    \begin{minipage}{0.23\linewidth}
      \centering
      \includegraphics[width=\linewidth]{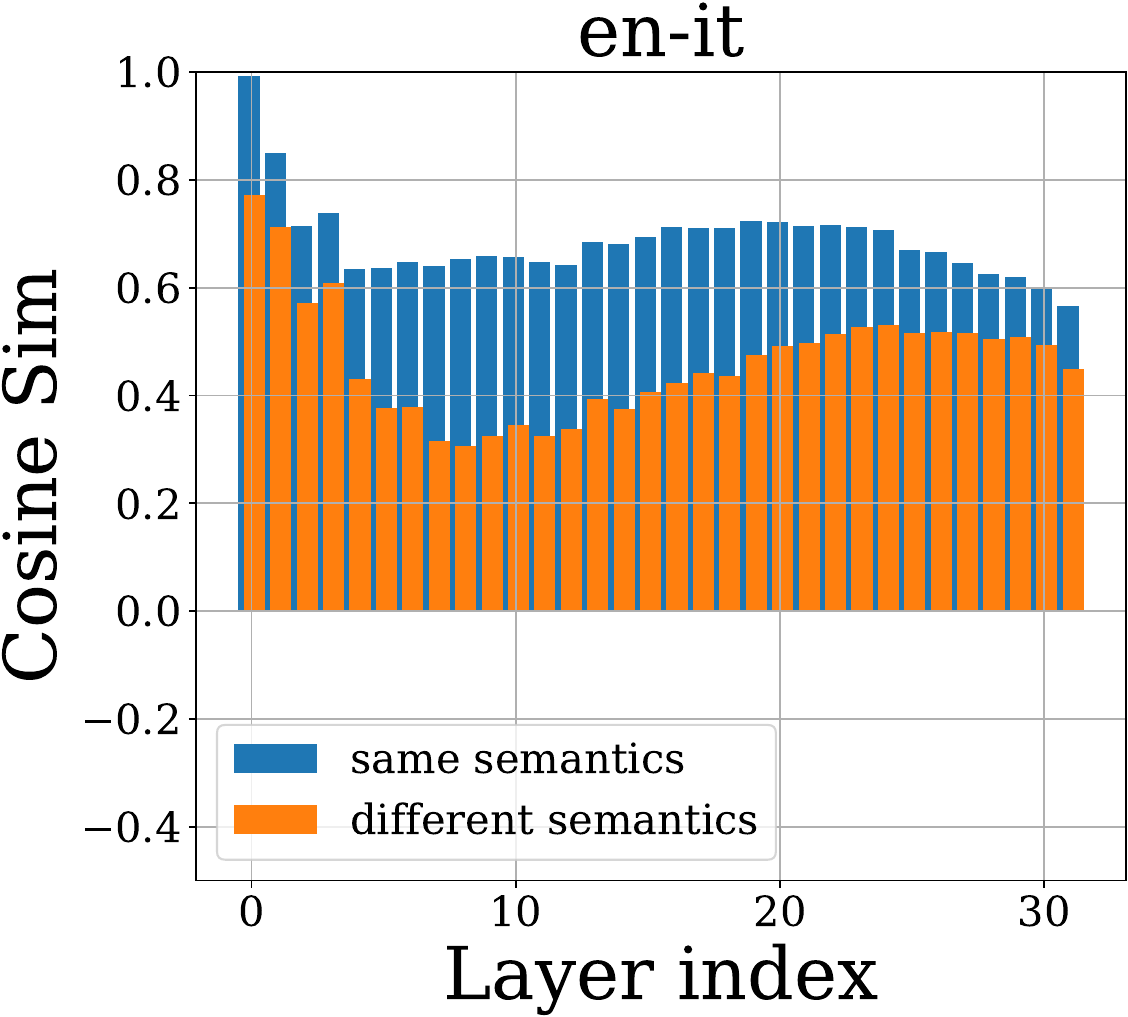}
      \subcaption{en-it (baseline)}
    \end{minipage}
    
      % First row label
      \begin{minipage}{\linewidth}
        \centering
        \small \textbf{(a) top-1000  (representing 0.2\% of all neurons)}
      \end{minipage}

    % second row
    \begin{minipage}{0.23\linewidth}
      \centering
      \includegraphics[width=\linewidth]{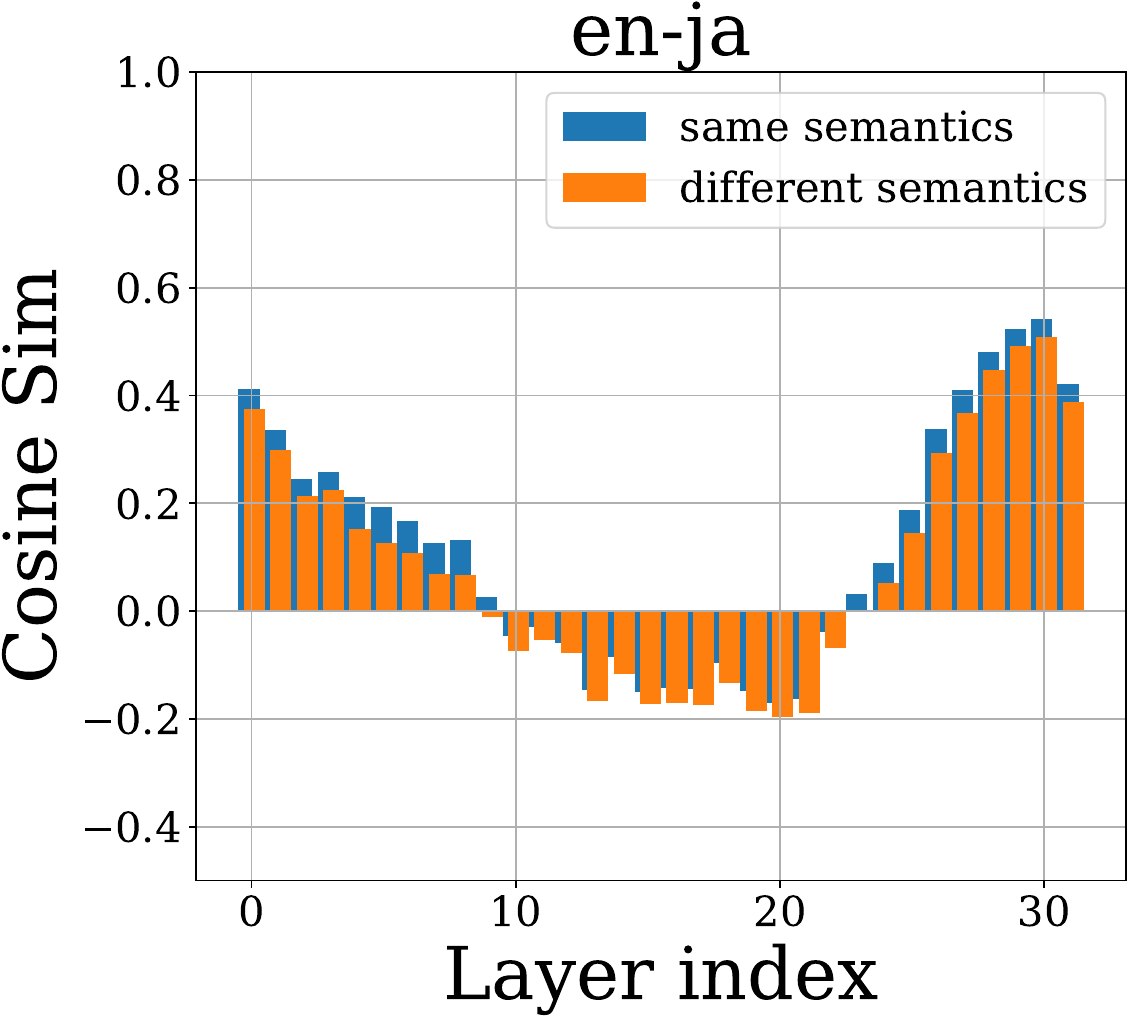}
      \subcaption{en-ja}
    \end{minipage}
    \begin{minipage}{0.23\linewidth}
      \centering
      \includegraphics[width=\linewidth]{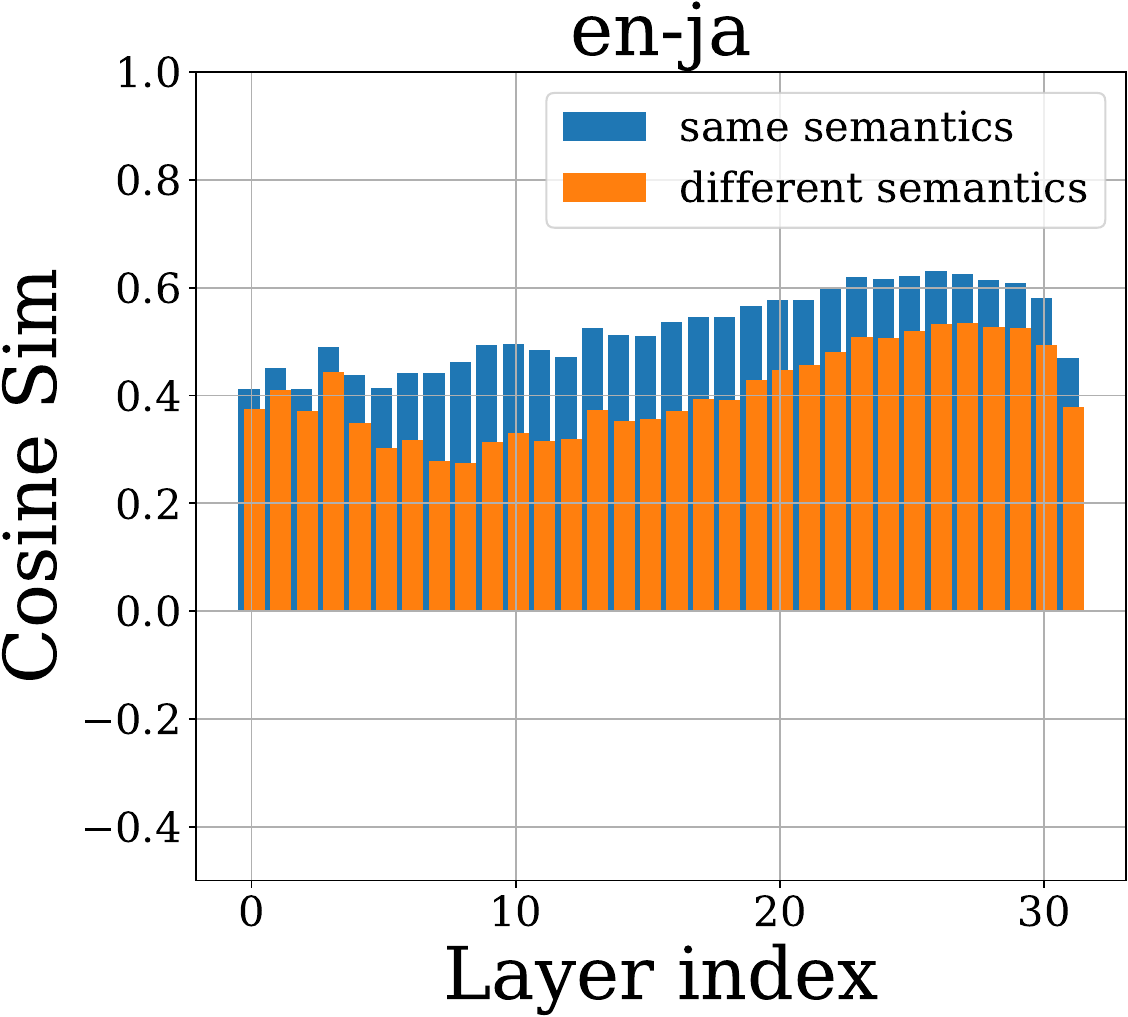}
      \subcaption{en-ja (baseline)}
    \end{minipage}
    \begin{minipage}{0.23\linewidth}
      \centering
      \includegraphics[width=\linewidth]{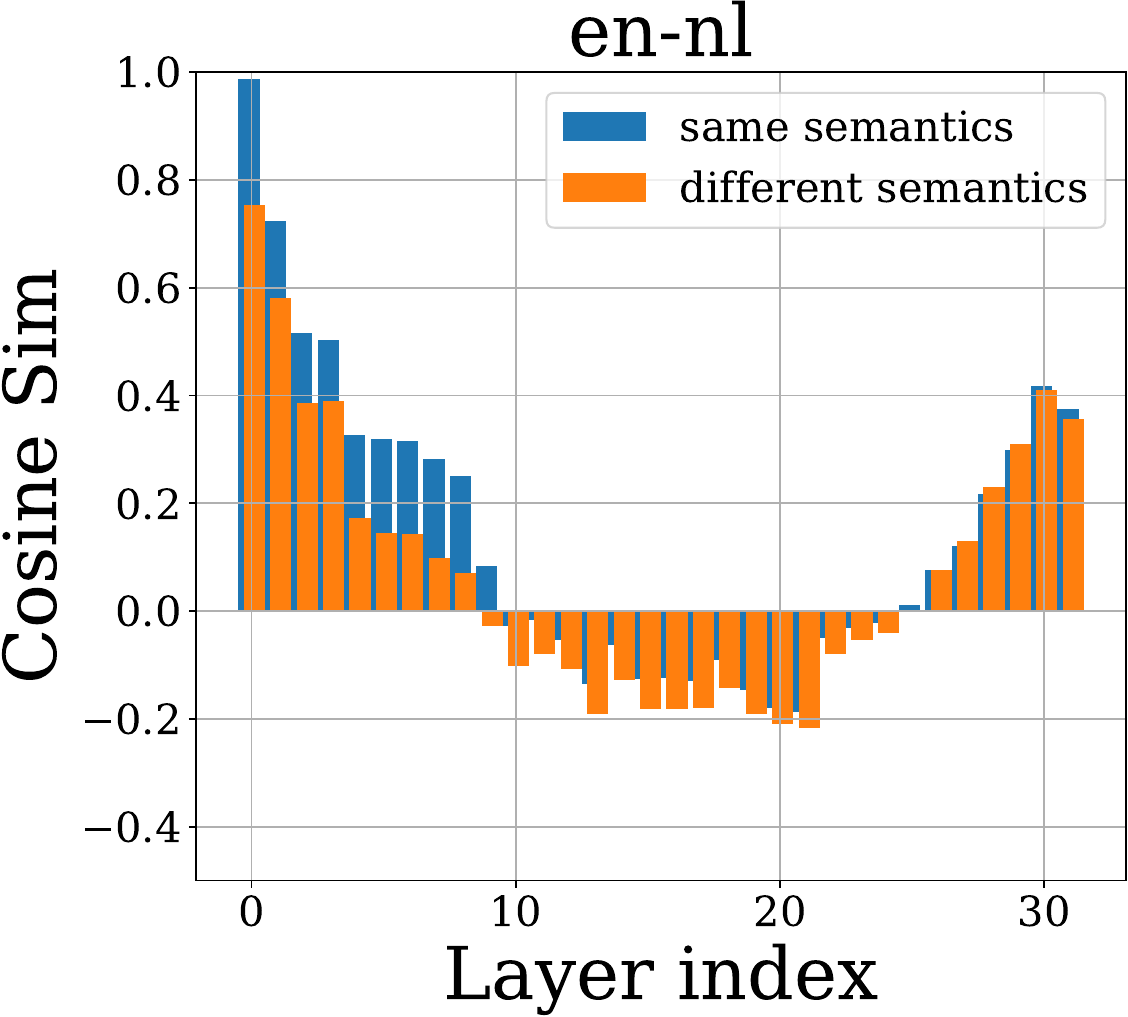}
      \subcaption{en-nl}
    \end{minipage}
    \begin{minipage}{0.23\linewidth}
      \centering
      \includegraphics[width=\linewidth]{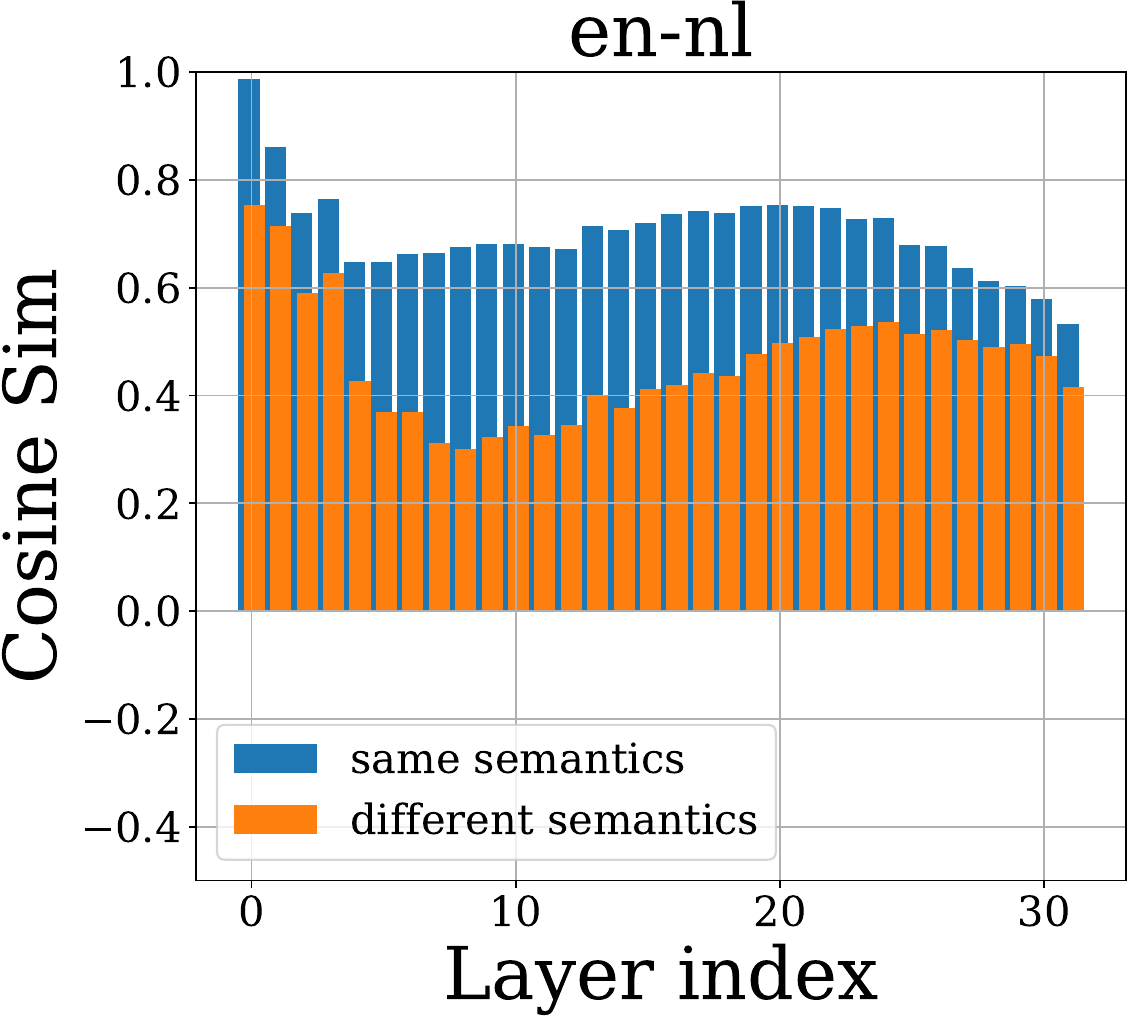}
      \subcaption{en-nl (baseline)}
    \end{minipage}

    \begin{minipage}{0.23\linewidth}
      \centering
      \includegraphics[width=\linewidth]{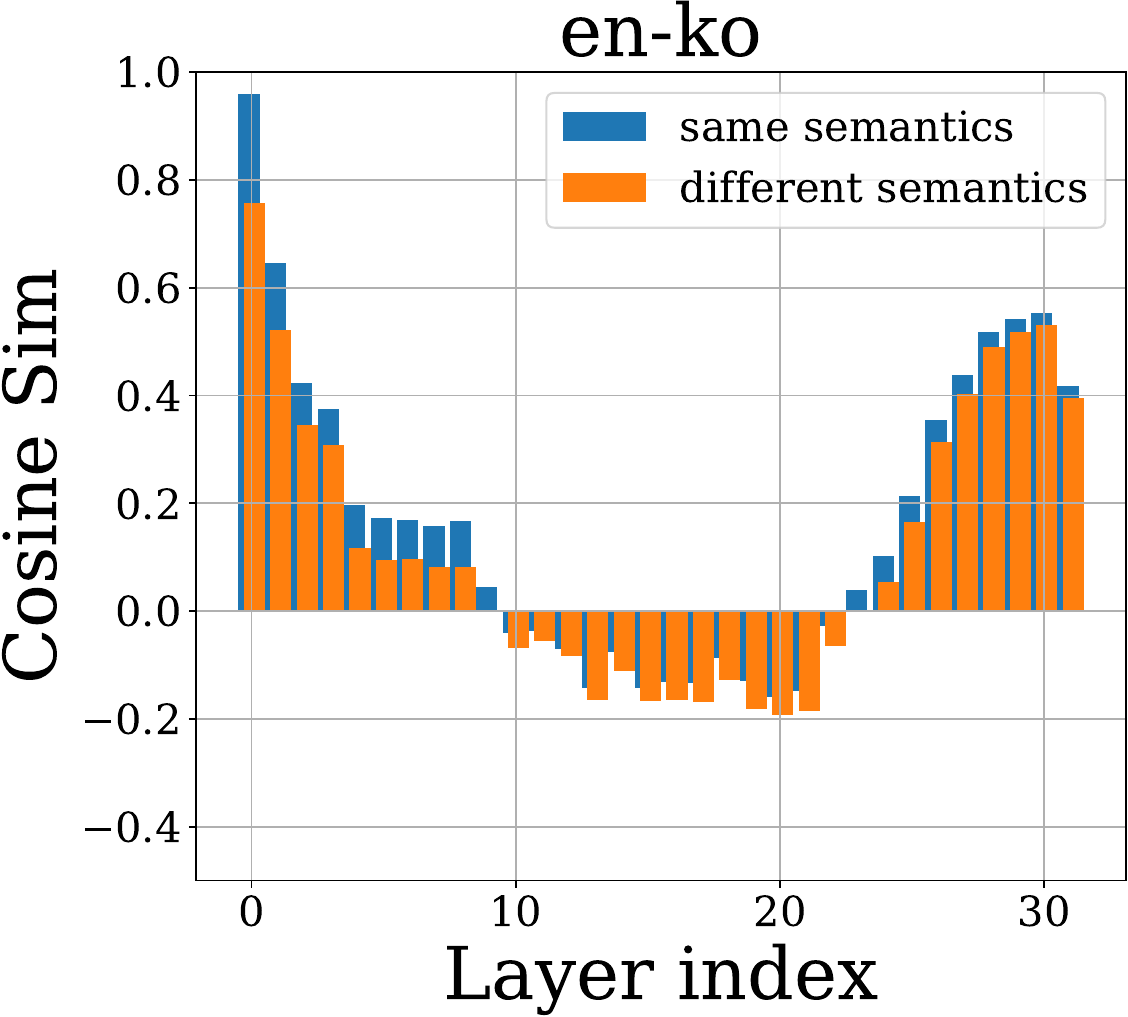}
      \subcaption{en-ko}
    \end{minipage}
    \begin{minipage}{0.23\linewidth}
      \centering
      \includegraphics[width=\linewidth]{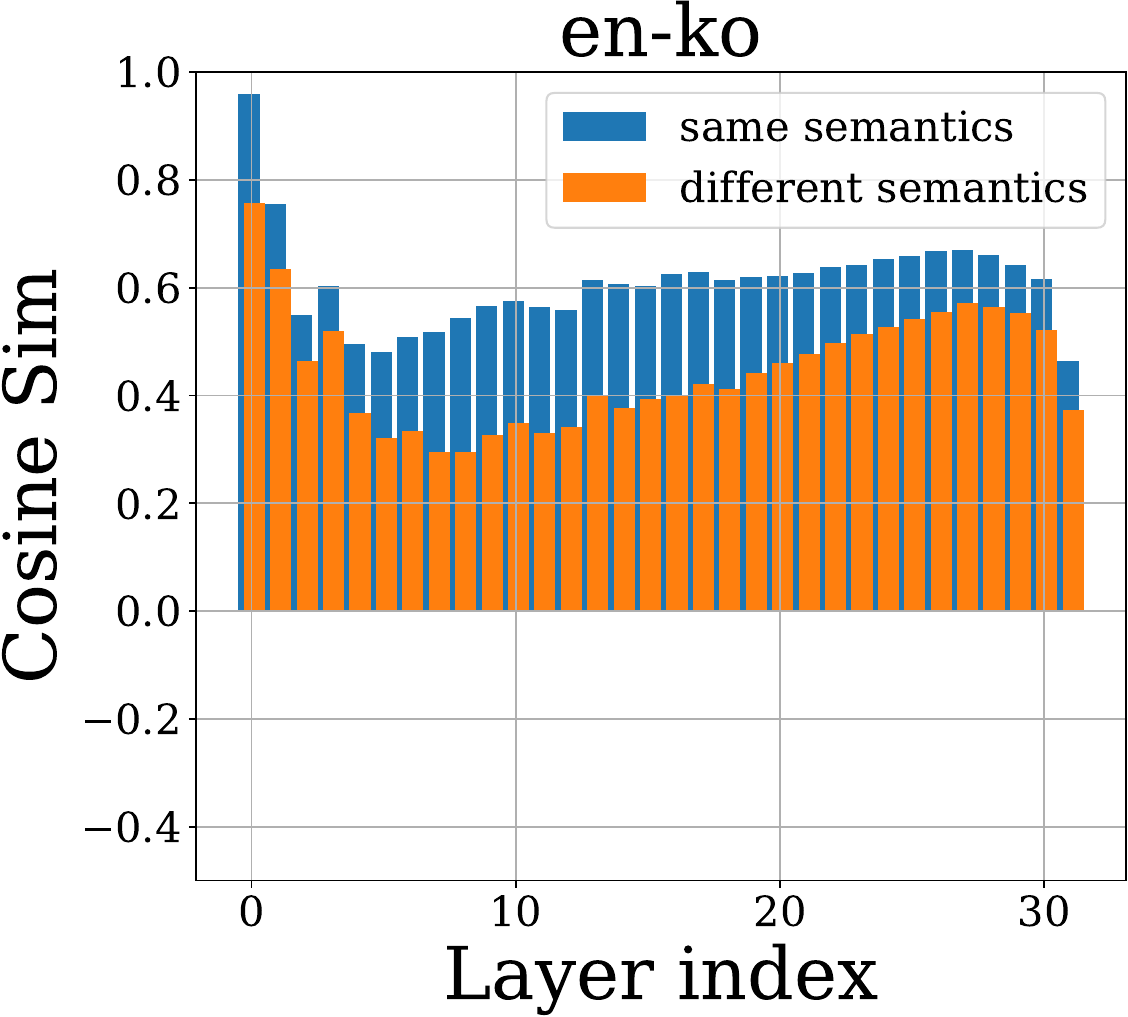}
      \subcaption{en-ko (baseline)}
    \end{minipage}
    \begin{minipage}{0.23\linewidth}
      \centering
      \includegraphics[width=\linewidth]{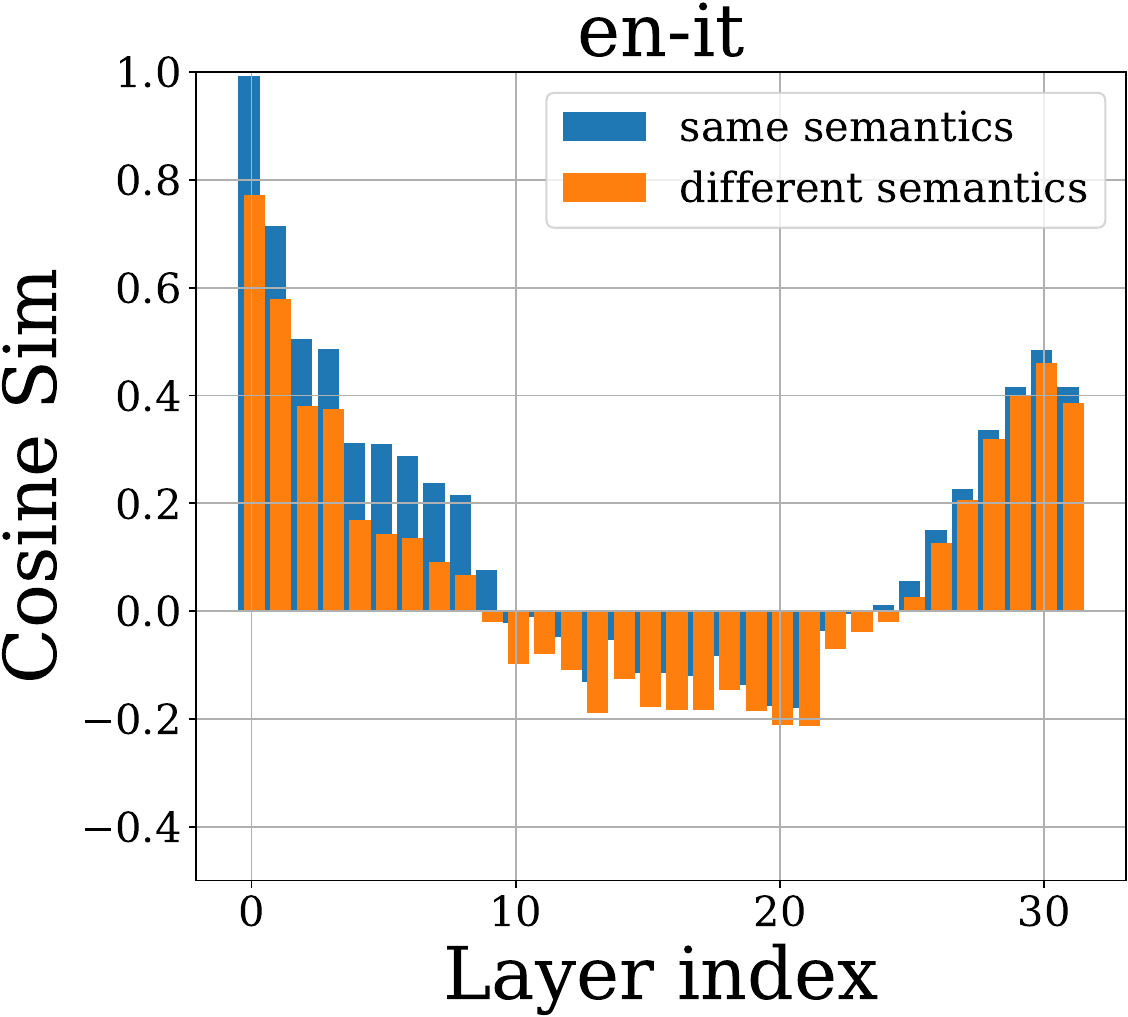}
      \subcaption{en-it}
    \end{minipage}
    \begin{minipage}{0.23\linewidth}
      \centering
      \includegraphics[width=\linewidth]{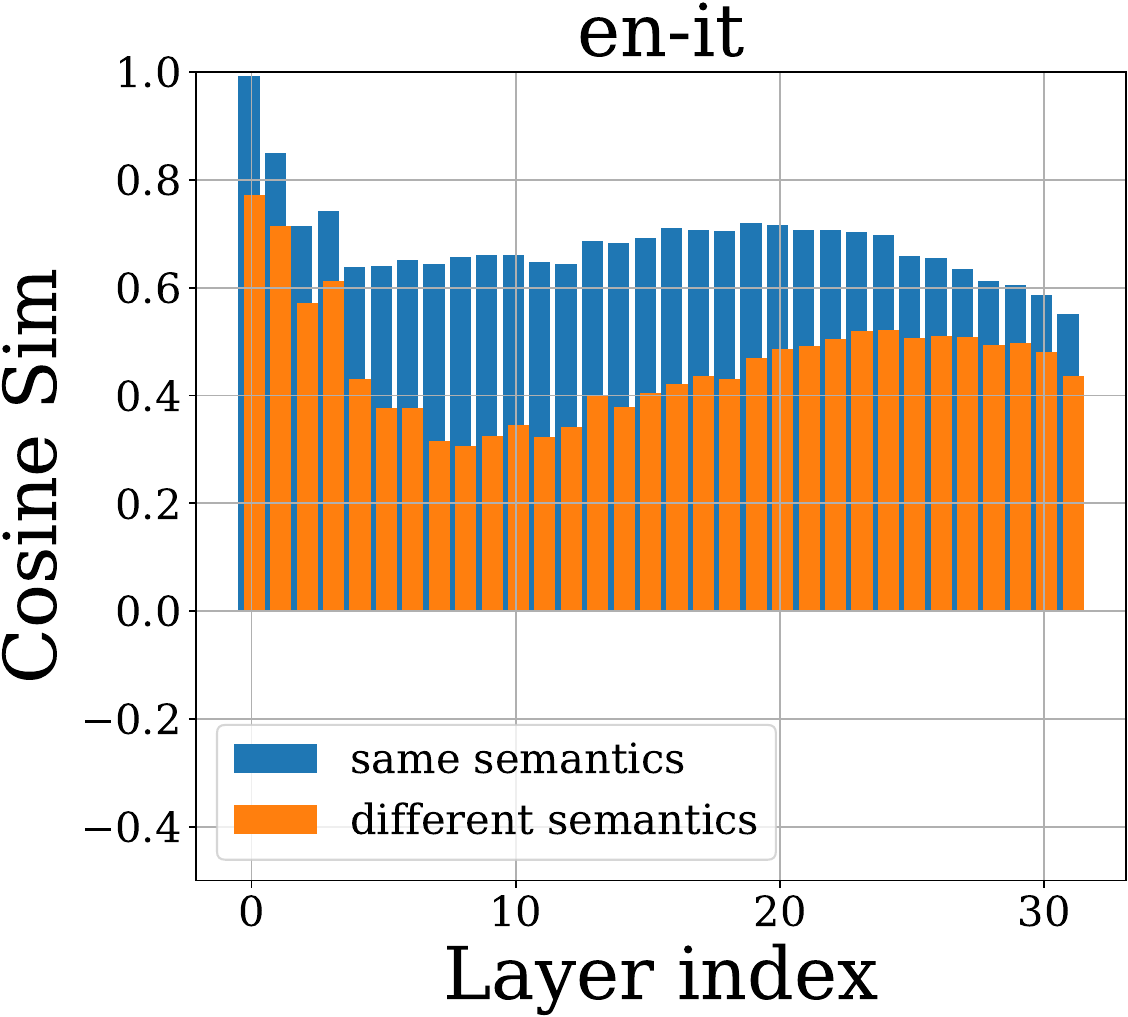}
      \subcaption{en-it (baseline)}
    \end{minipage}
    
      \begin{minipage}{\linewidth}
        \centering
        \small \textbf{(b) top-3000 (representing 0.6\% of all neurons)}
      \end{minipage}

    \begin{minipage}{0.23\linewidth}
      \centering
      \includegraphics[width=\linewidth]{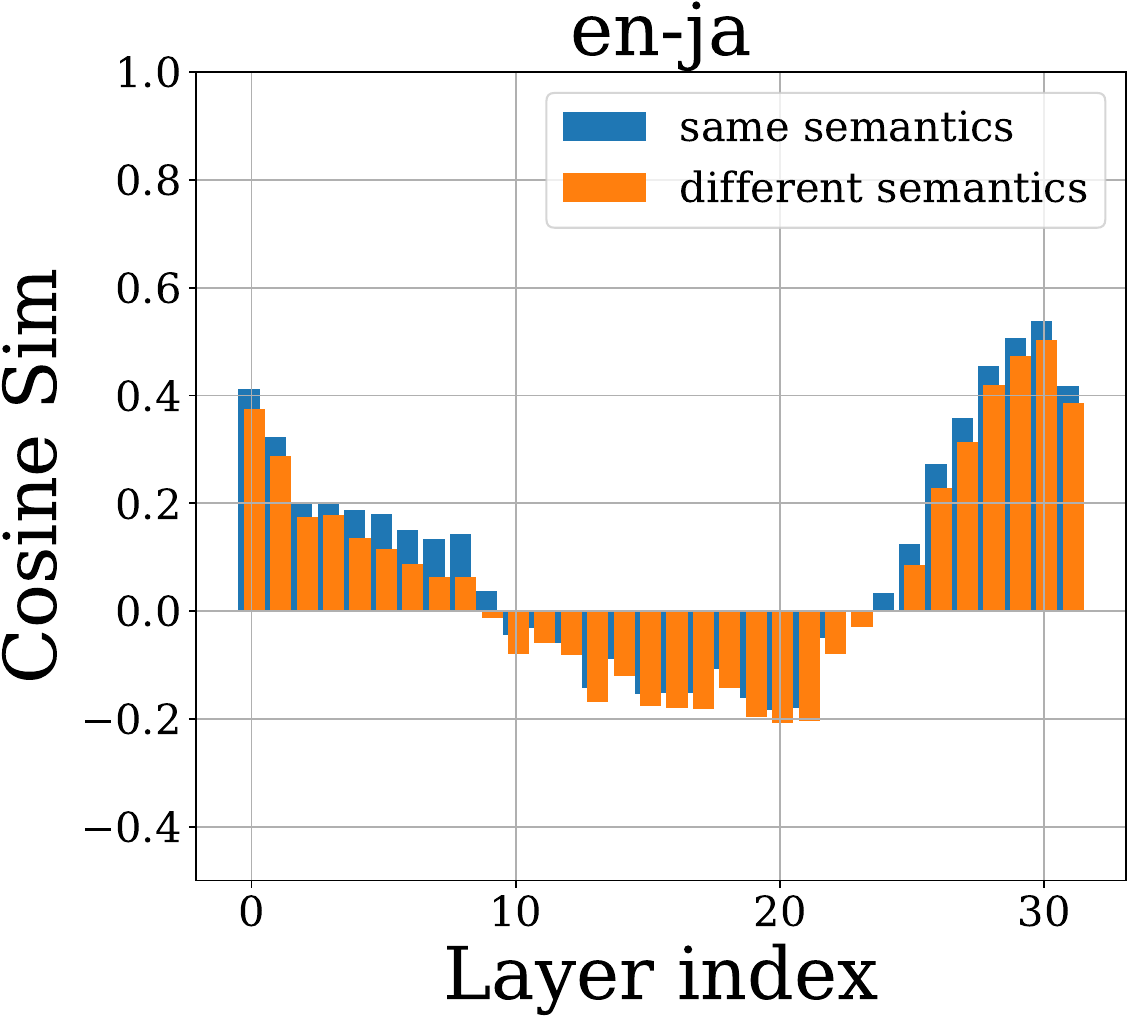}
      \subcaption{en-ja}
    \end{minipage}
    \begin{minipage}{0.23\linewidth}
      \centering
      \includegraphics[width=\linewidth]{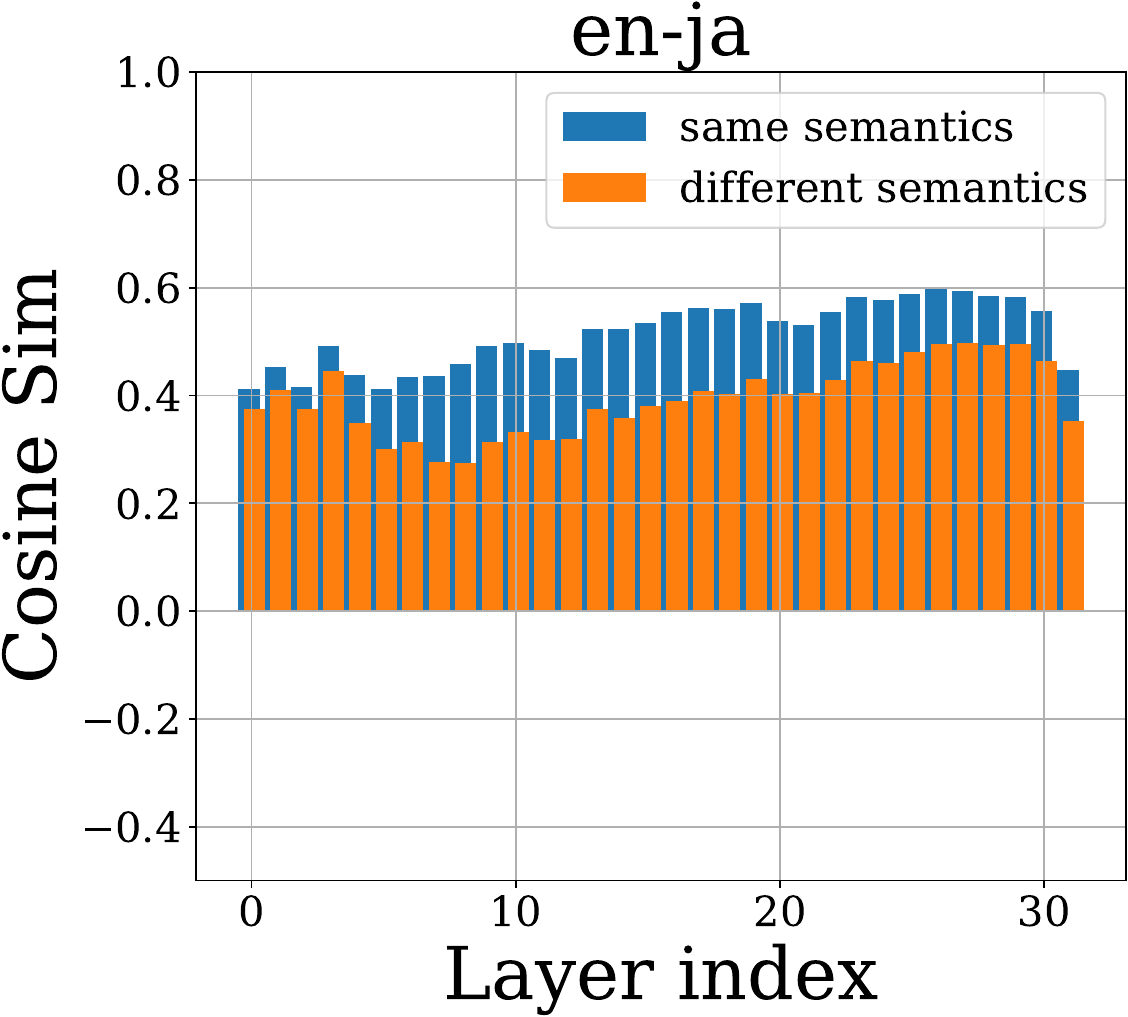}
      \subcaption{en-ja (baseline)}
    \end{minipage}
    \begin{minipage}{0.23\linewidth}
      \centering
      \includegraphics[width=\linewidth]{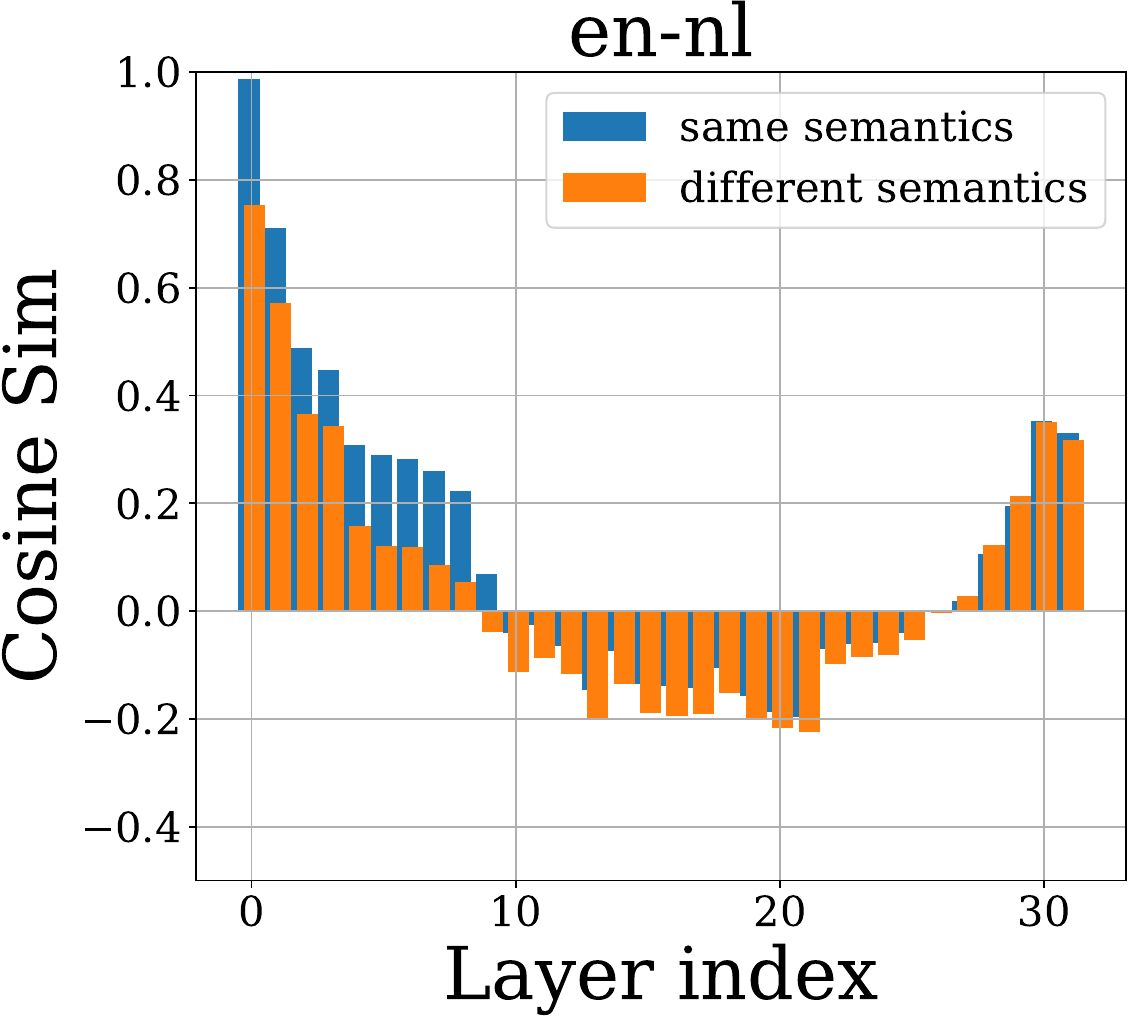}
      \subcaption{en-nl}
    \end{minipage}
    \begin{minipage}{0.23\linewidth}
      \centering
      \includegraphics[width=\linewidth]{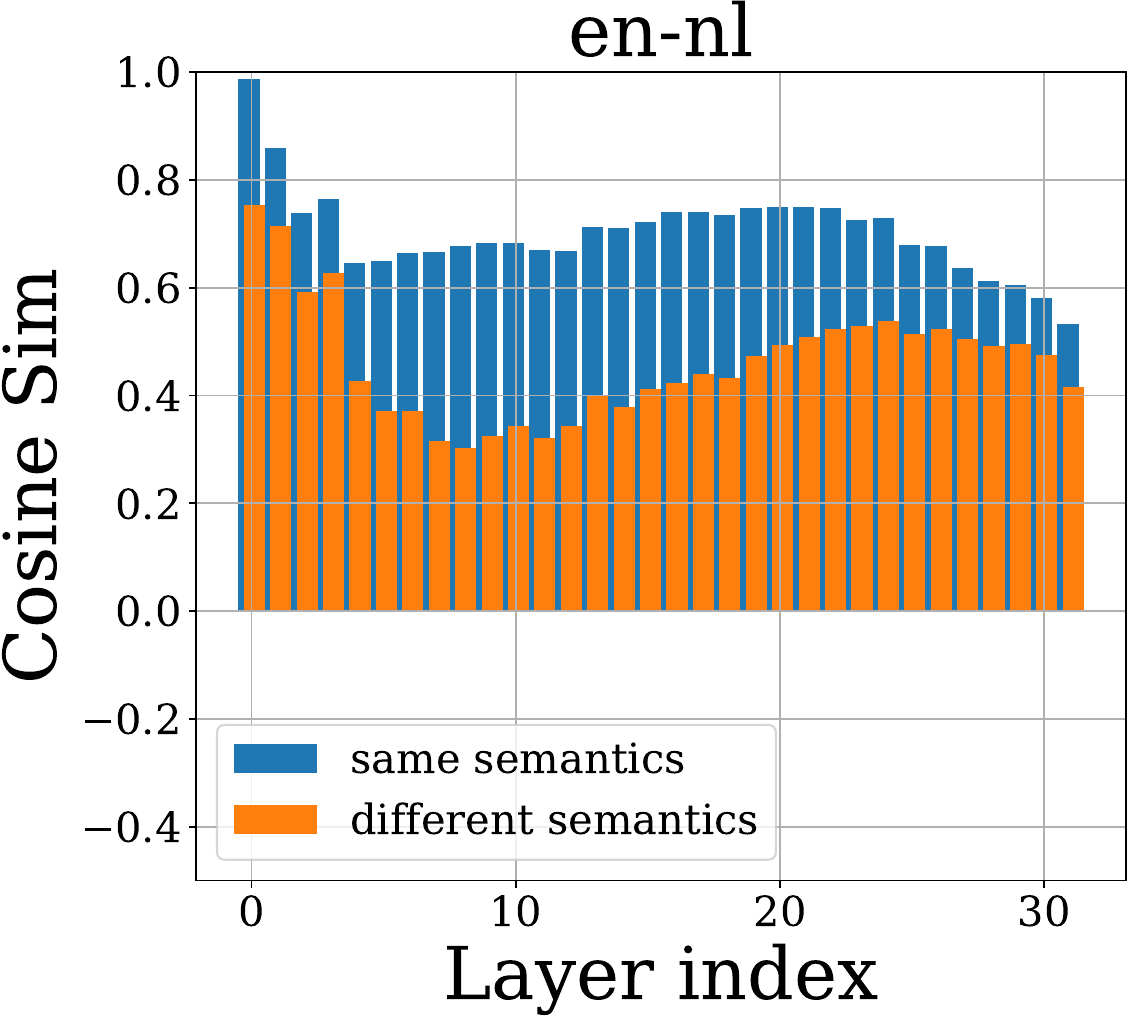}
      \subcaption{en-nl (baseline)}
    \end{minipage}

    \begin{minipage}{0.23\linewidth}
      \centering
      \includegraphics[width=\linewidth]{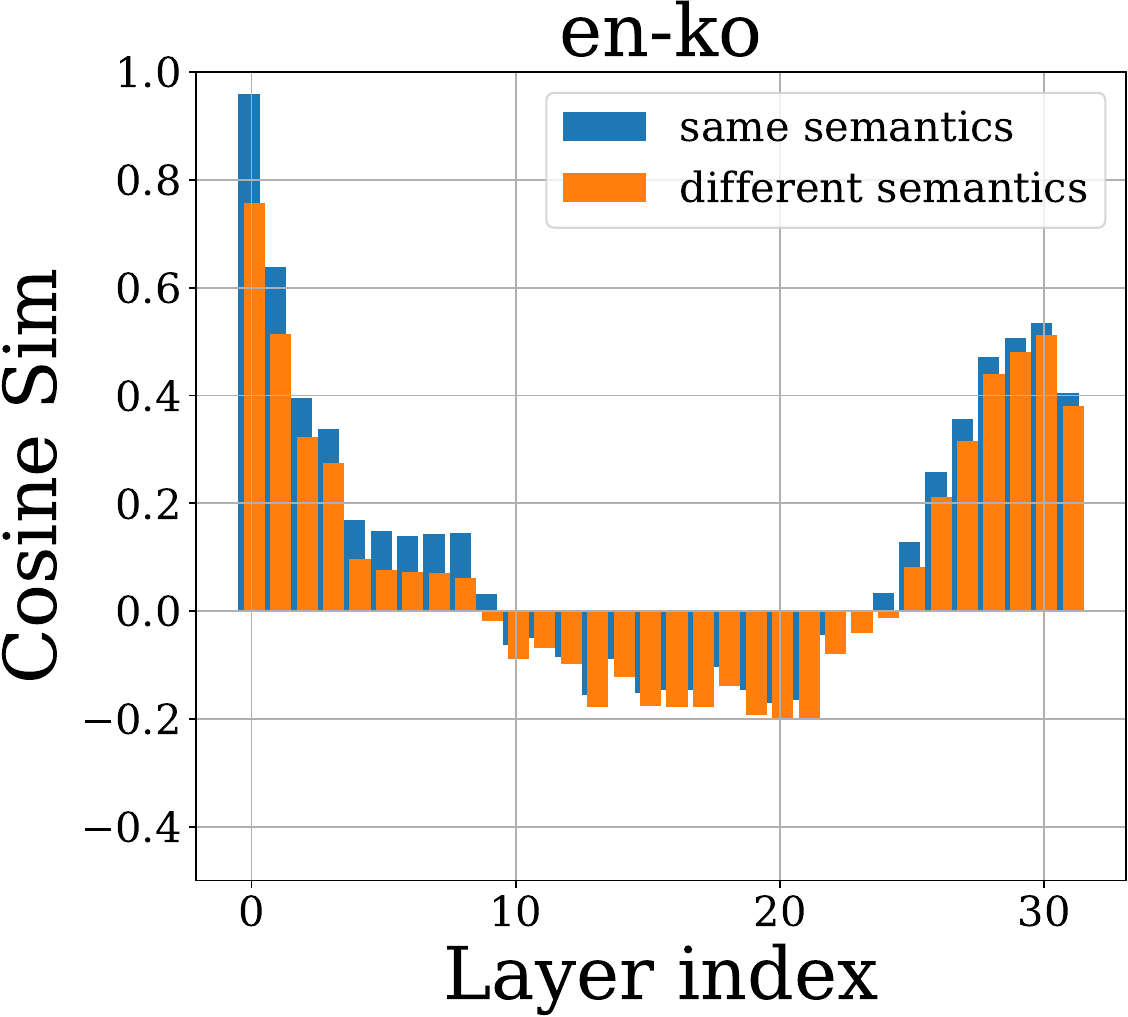}
      \subcaption{en-ko}
    \end{minipage}
    \begin{minipage}{0.23\linewidth}
      \centering
      \includegraphics[width=\linewidth]{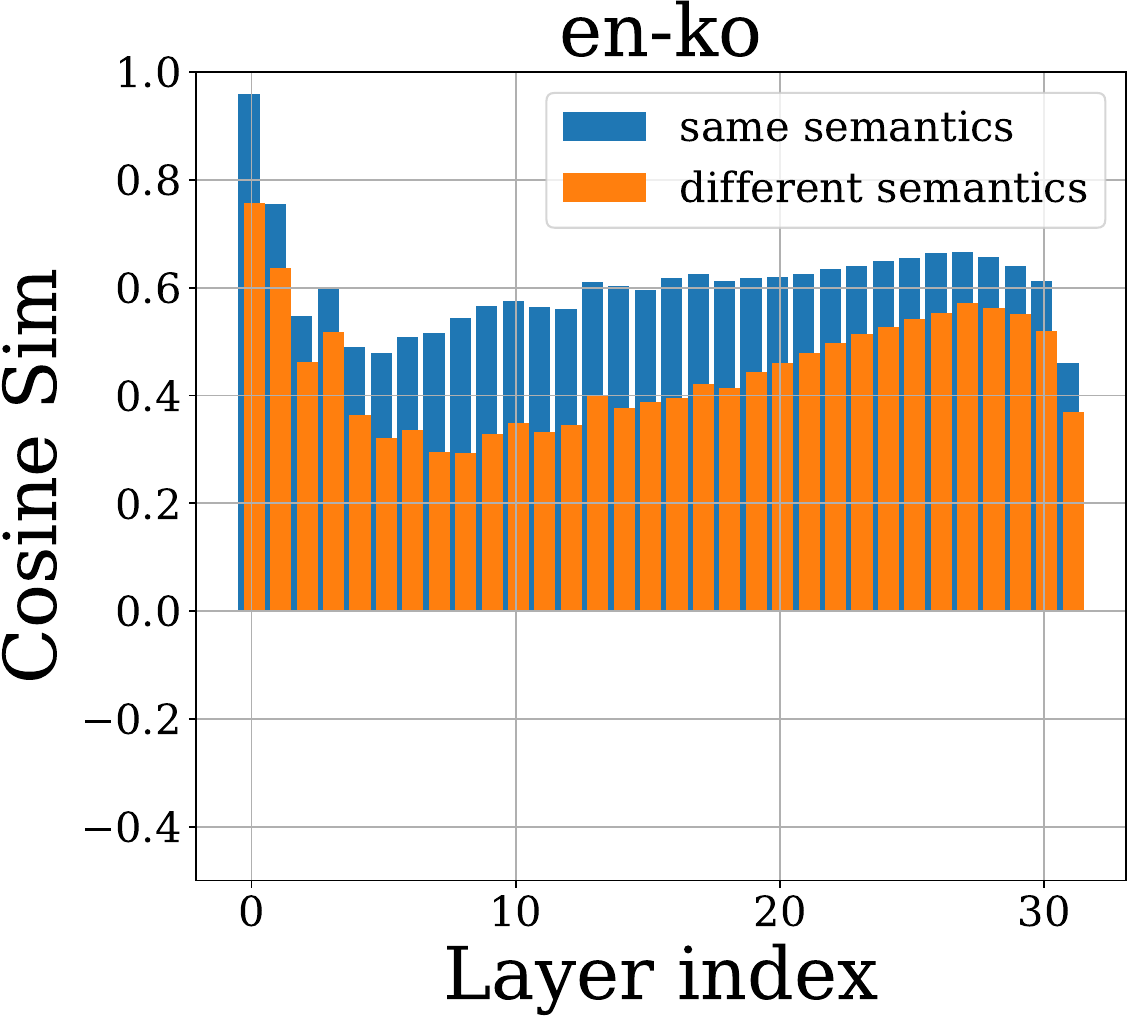}
      \subcaption{en-ko (baseline)}
    \end{minipage}
    \begin{minipage}{0.23\linewidth}
      \centering
      \includegraphics[width=\linewidth]{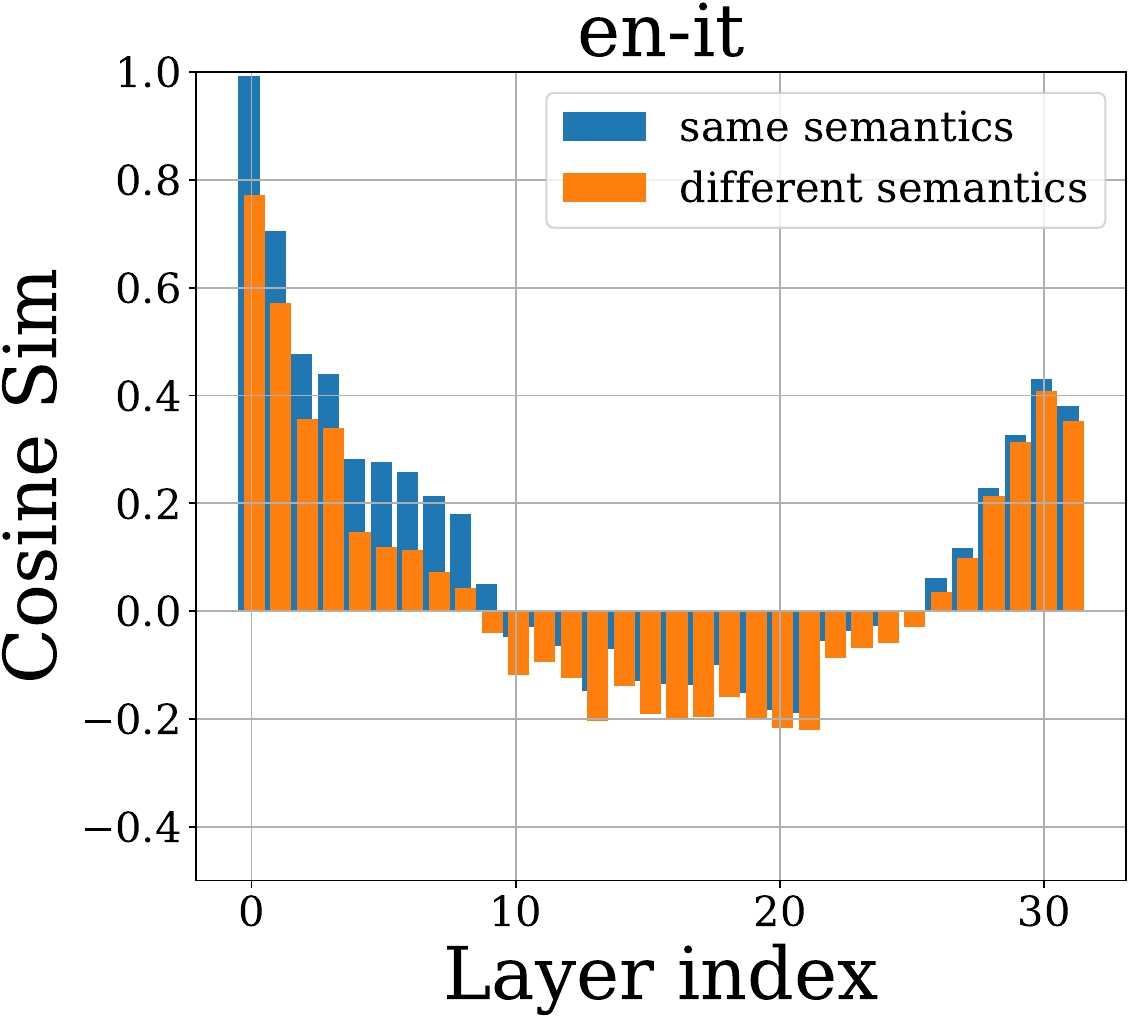}
      \subcaption{en-it}
    \end{minipage}
    \begin{minipage}{0.23\linewidth}
      \centering
      \includegraphics[width=\linewidth]{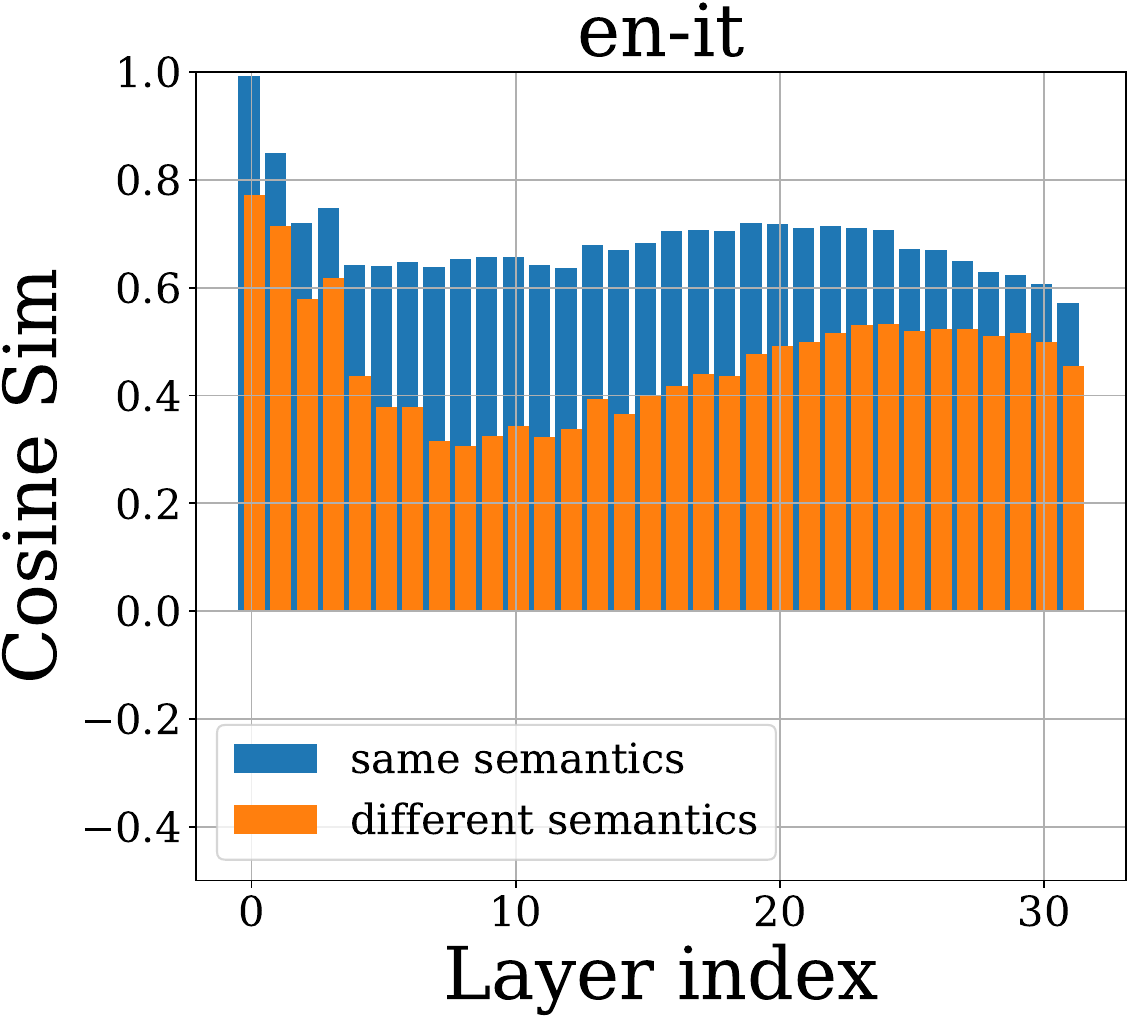}
      
      \subcaption{en-it (baseline)}
    \end{minipage}
    
      \begin{minipage}{\linewidth}
        \centering
        \small \textbf{(b) top-5000 (representing 1\% of all neurons)}
      \end{minipage}

  \caption{\textbf{Similarity of hidden states across layers while deactivating Type-1 Transfer Neurons (LLaMA3-8B).}}
  \label{fig:appendix:hs_sim_llama_deactivating_top-1k_Type-1}
\end{figure*}

\begin{figure*}[t]
    \centering

    \begin{minipage}{0.20\linewidth}
      \centering
      \includegraphics[width=\linewidth]{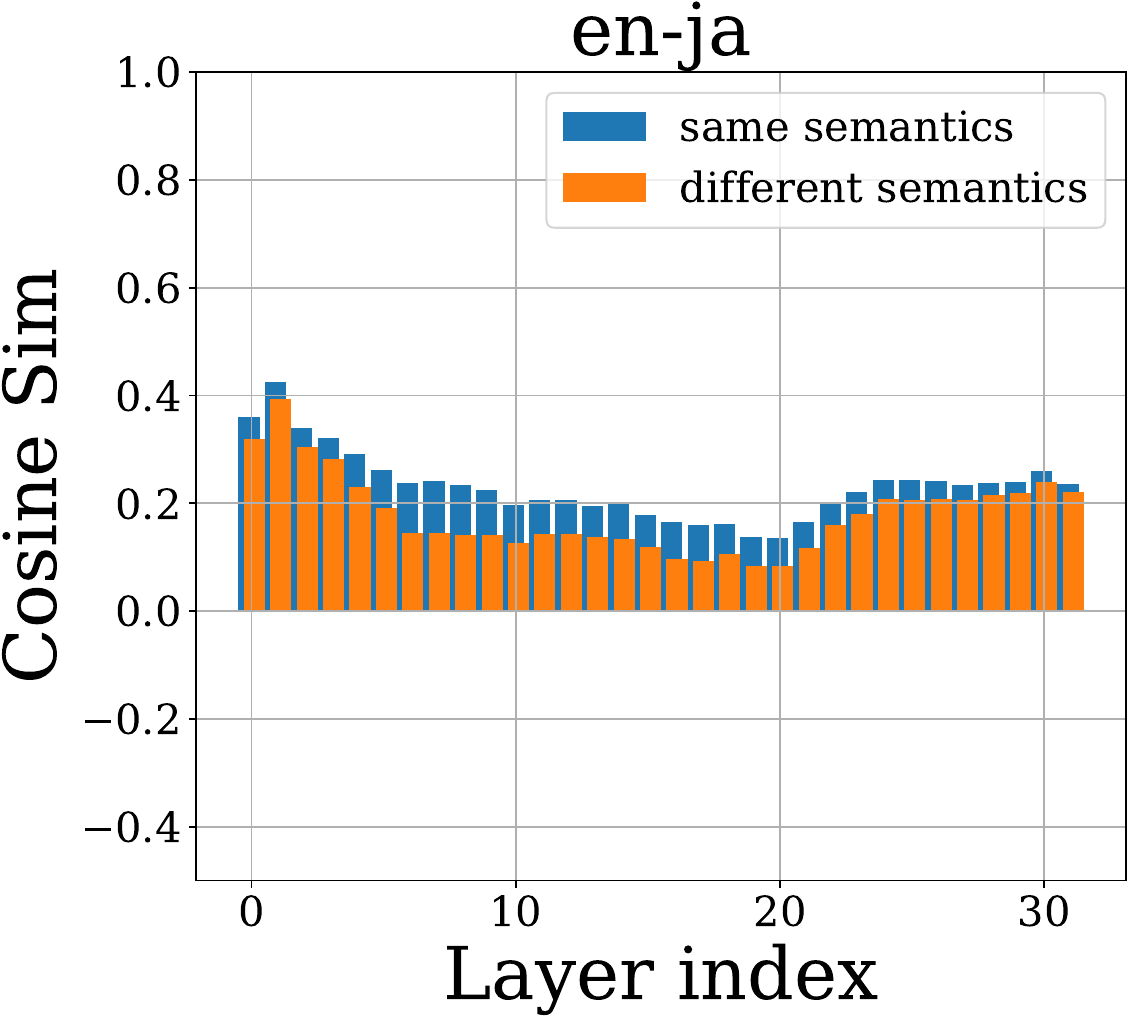}
      \subcaption{en-ja}
    \end{minipage}
    \begin{minipage}{0.20\linewidth}
      \centering
      \includegraphics[width=\linewidth]{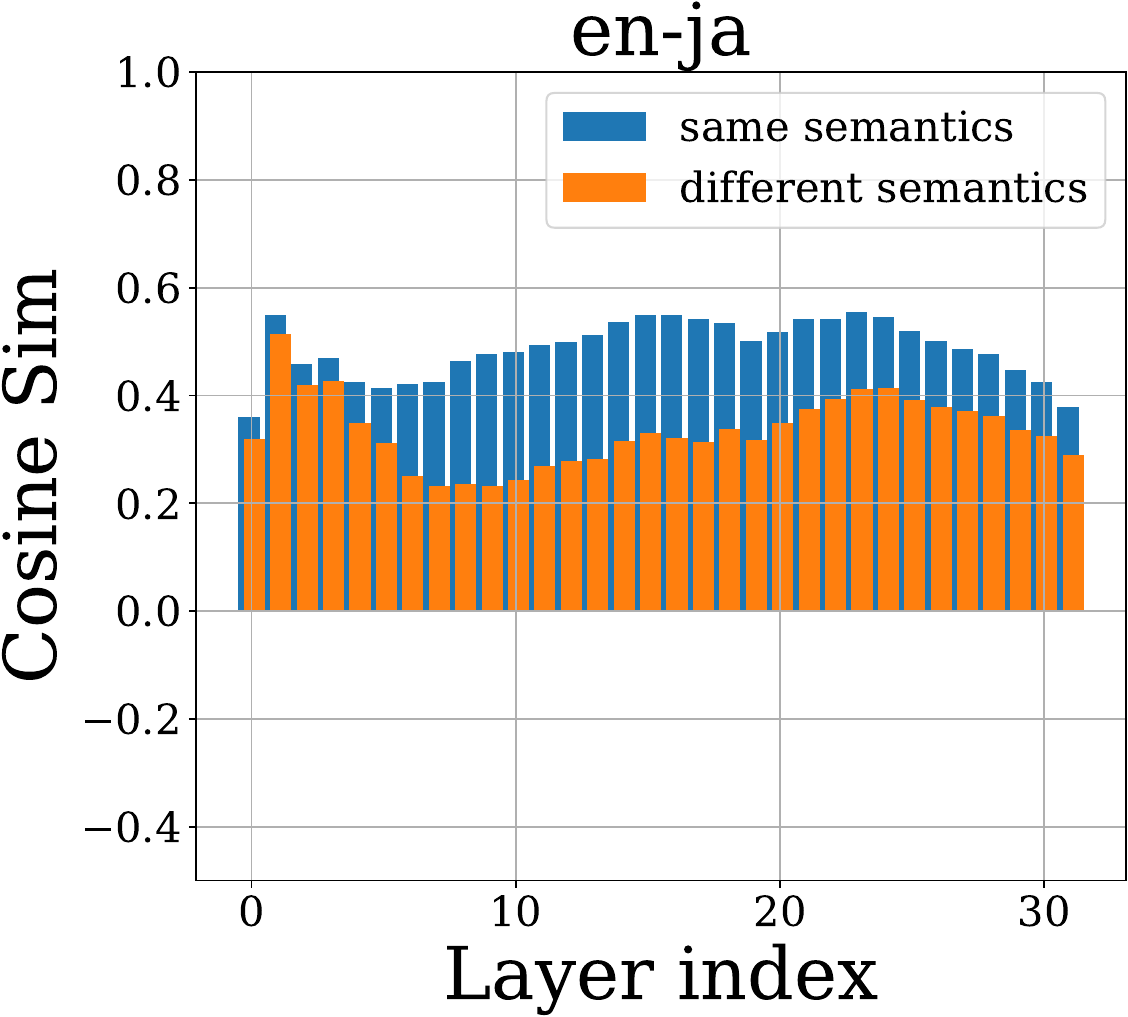}
      \subcaption{en-ja (baseline)}
    \end{minipage}
    \begin{minipage}{0.20\linewidth}
      \centering
      \includegraphics[width=\linewidth]{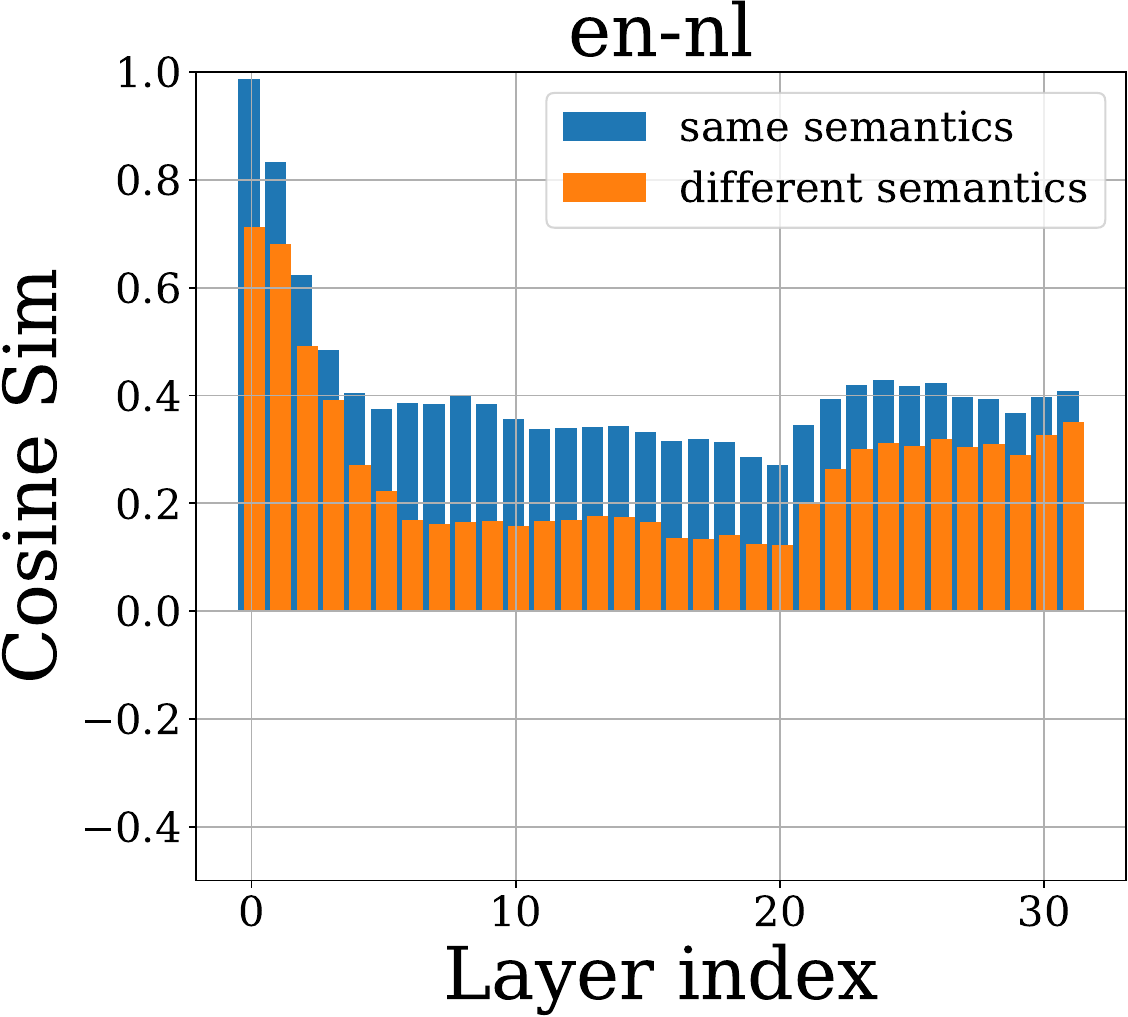}
      \subcaption{en-nl}
    \end{minipage}
    \begin{minipage}{0.20\linewidth}
      \centering
      \includegraphics[width=\linewidth]{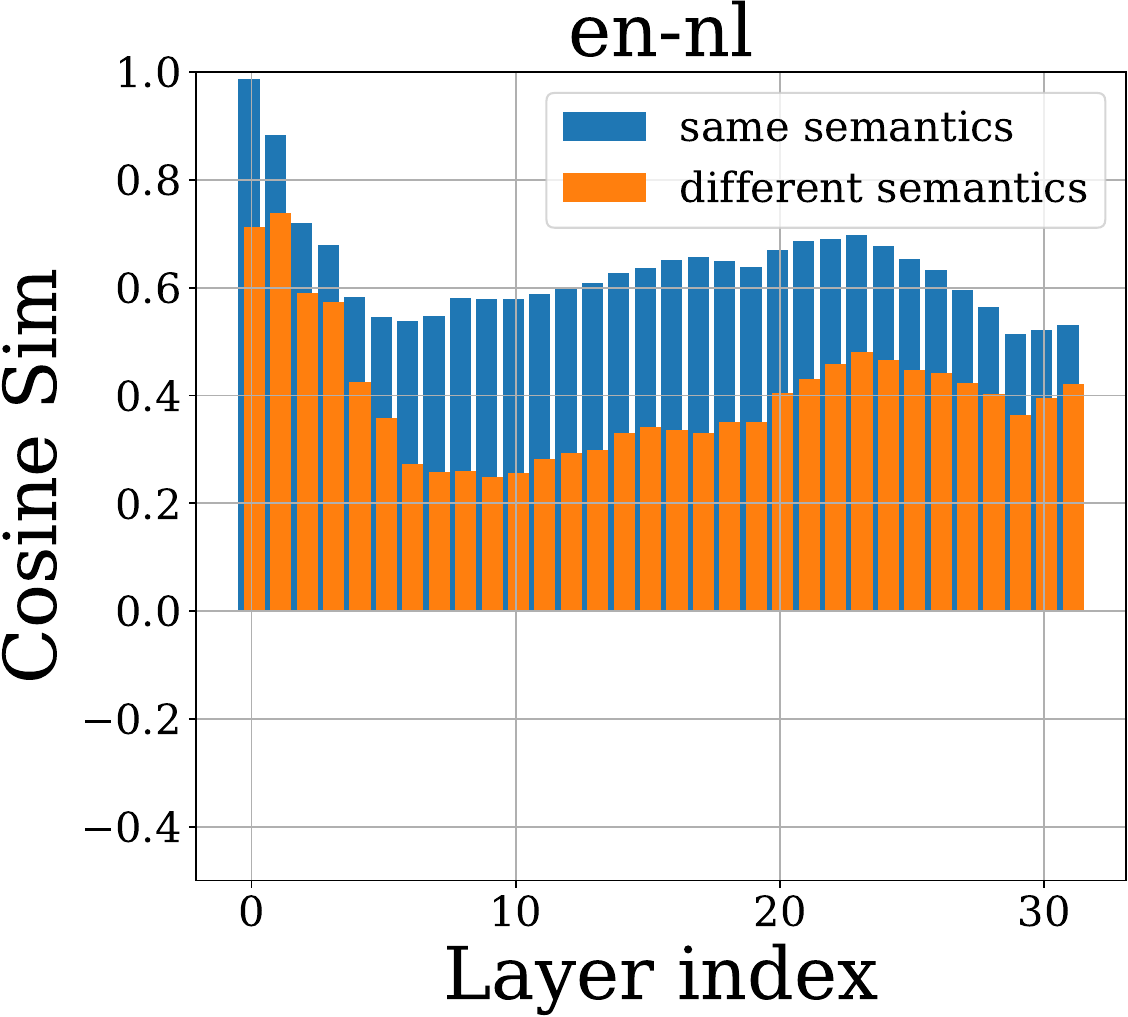}
      \subcaption{en-nl (baseline)}
    \end{minipage}

    \begin{minipage}{0.20\linewidth}
      \centering
      \includegraphics[width=\linewidth]{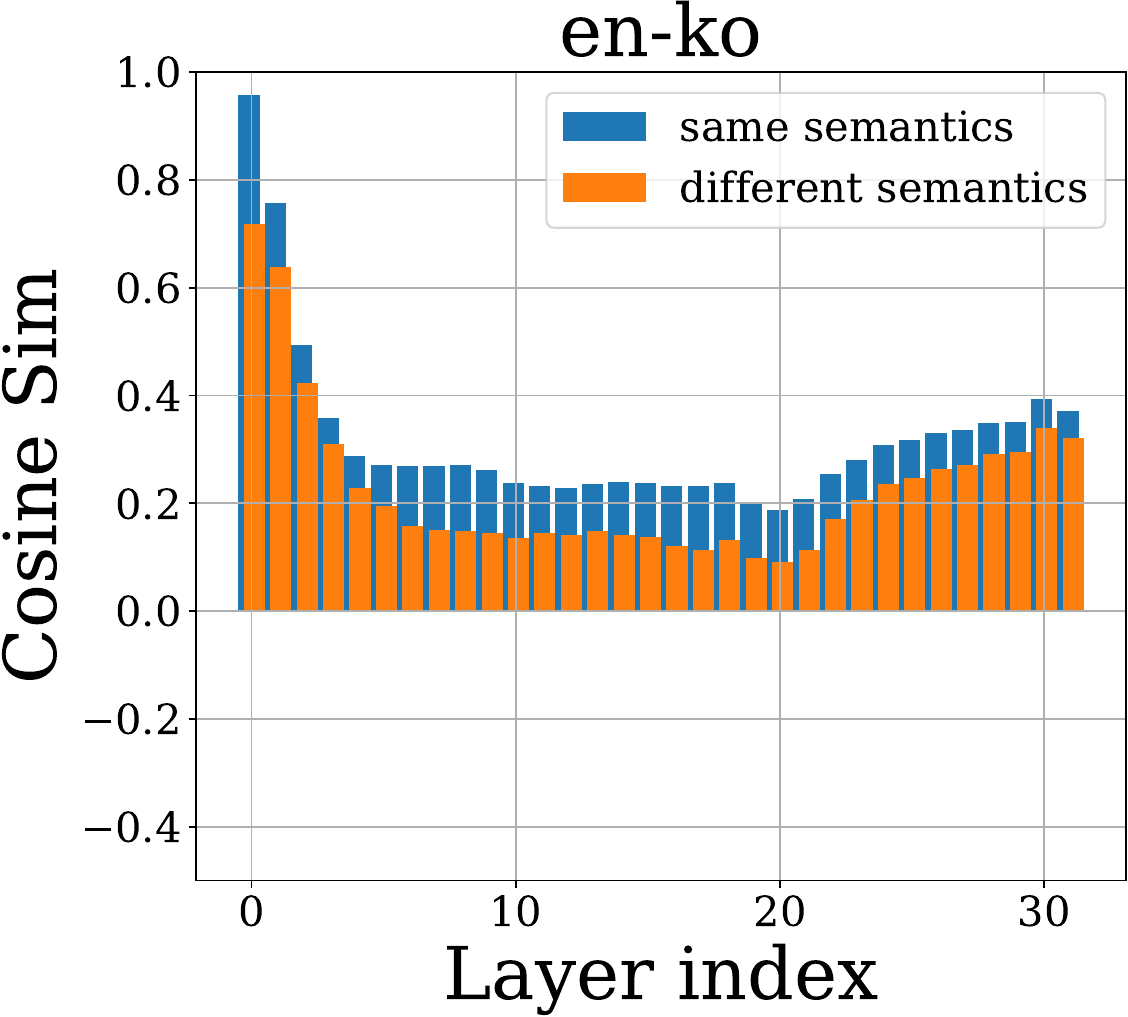}
      \subcaption{en-ko}
    \end{minipage}
    \begin{minipage}{0.20\linewidth}
      \centering
      \includegraphics[width=\linewidth]{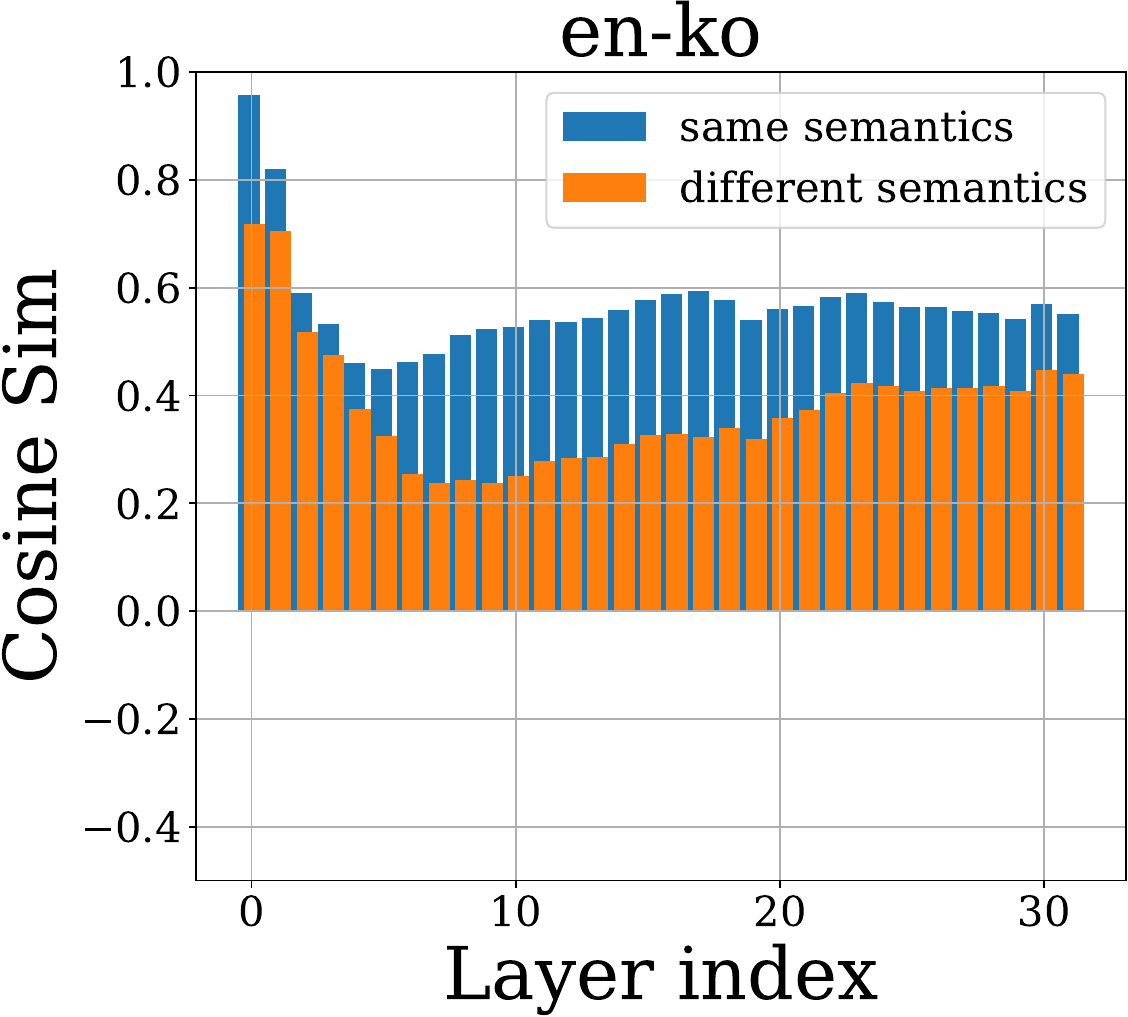}
      \subcaption{en-ko (baseline)}
    \end{minipage}
    \begin{minipage}{0.20\linewidth}
      \centering
      \includegraphics[width=\linewidth]{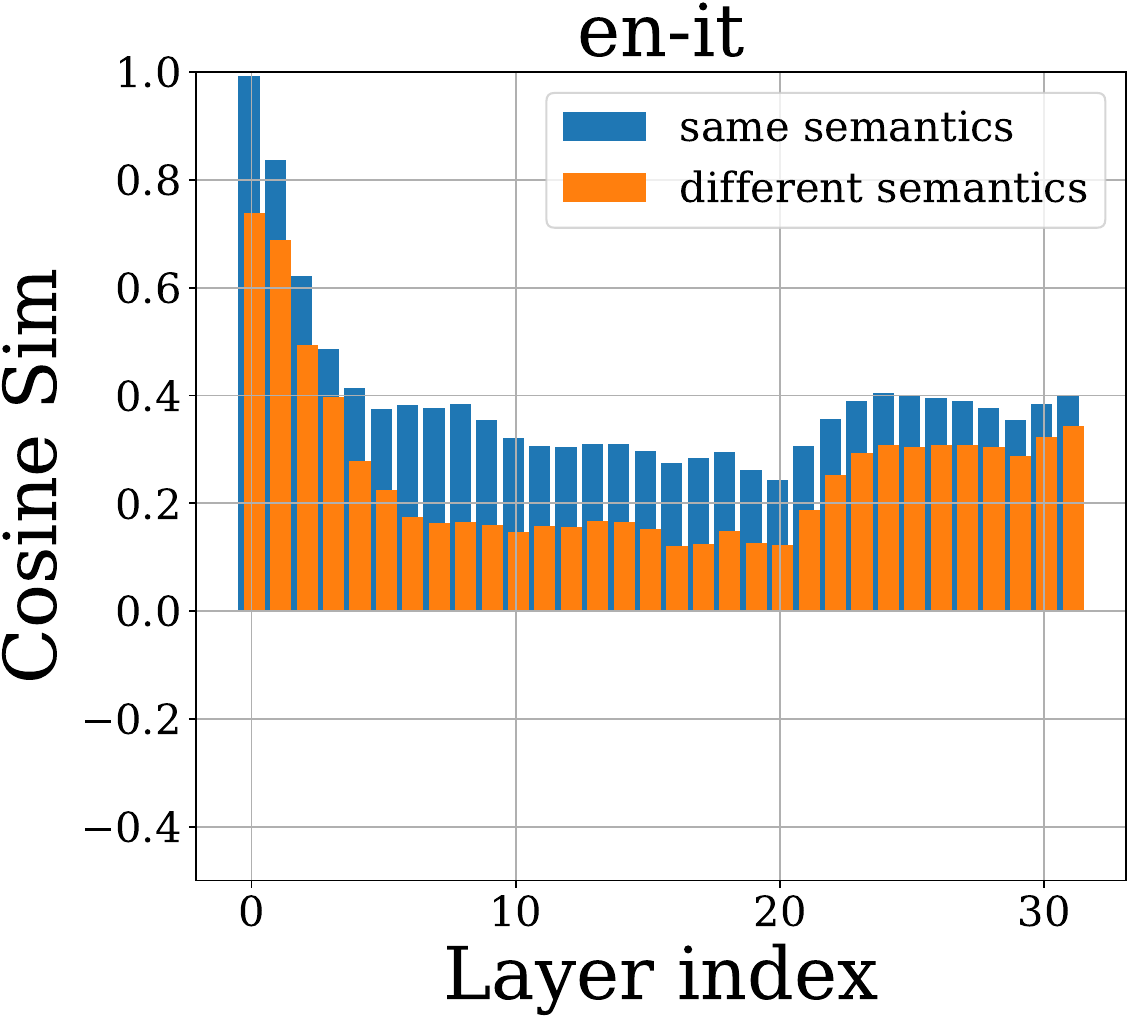}
      \subcaption{en-it}
    \end{minipage}
    \begin{minipage}{0.20\linewidth}
      \centering
      \includegraphics[width=\linewidth]{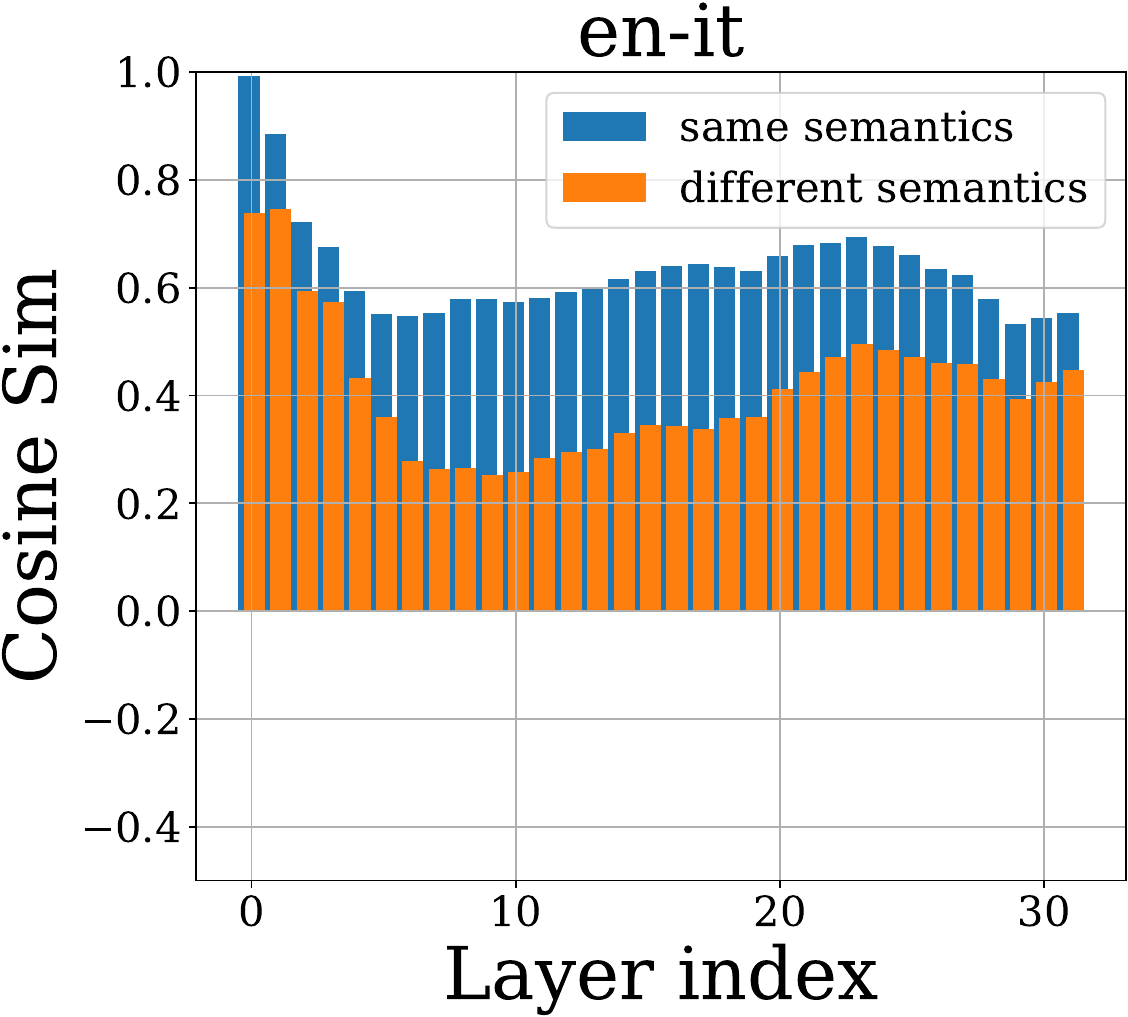}
      \subcaption{en-it (baseline)}
    \end{minipage}
    
      % First row label
      \begin{minipage}{\linewidth}
        \centering
        \small \textbf{(a) top-1000 (representing 0.2\% of all neurons)}
      \end{minipage}

    % second row
    \begin{minipage}{0.20\linewidth}
      \centering
      \includegraphics[width=\linewidth]{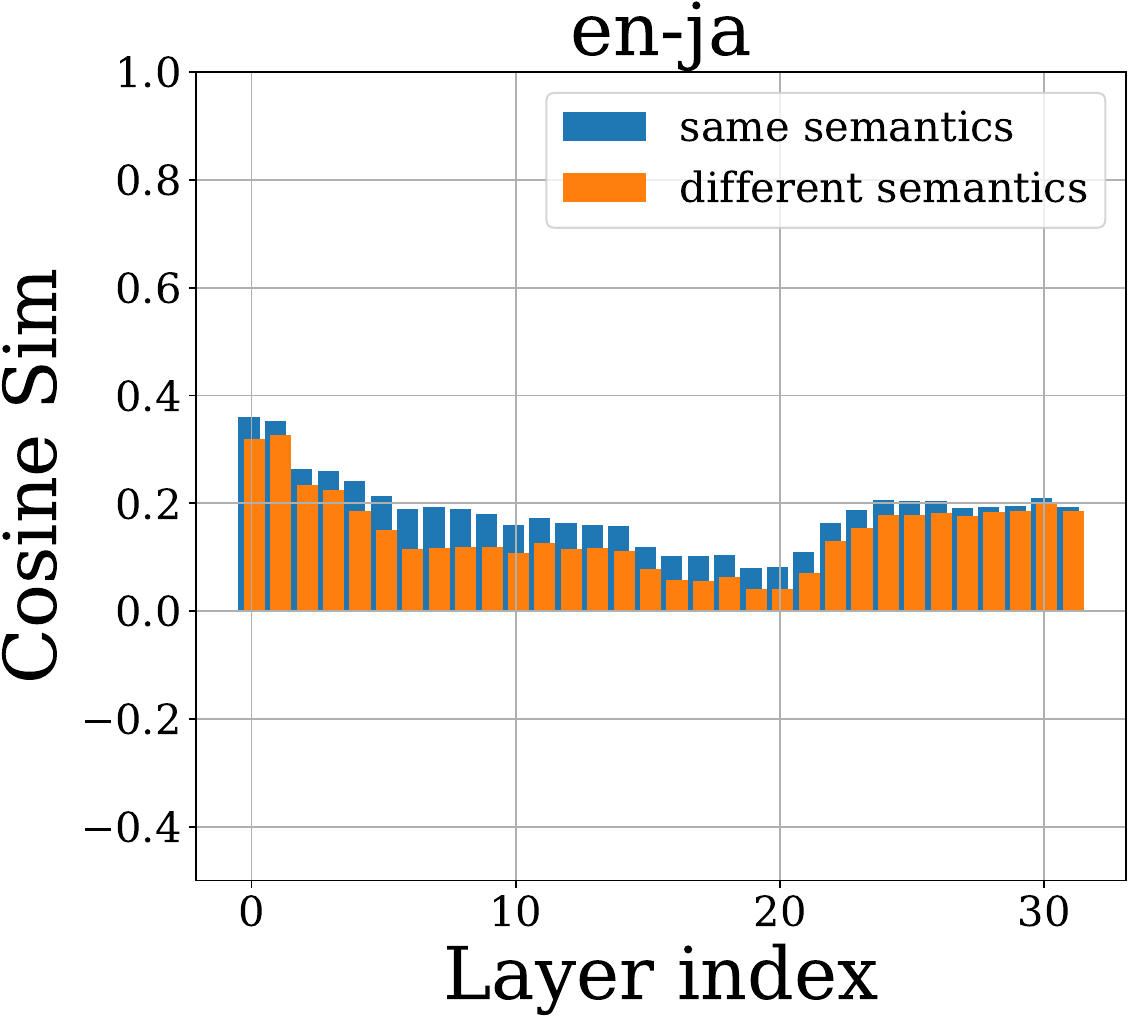}
      \subcaption{en-ja}
    \end{minipage}
    \begin{minipage}{0.20\linewidth}
      \centering
      \includegraphics[width=\linewidth]{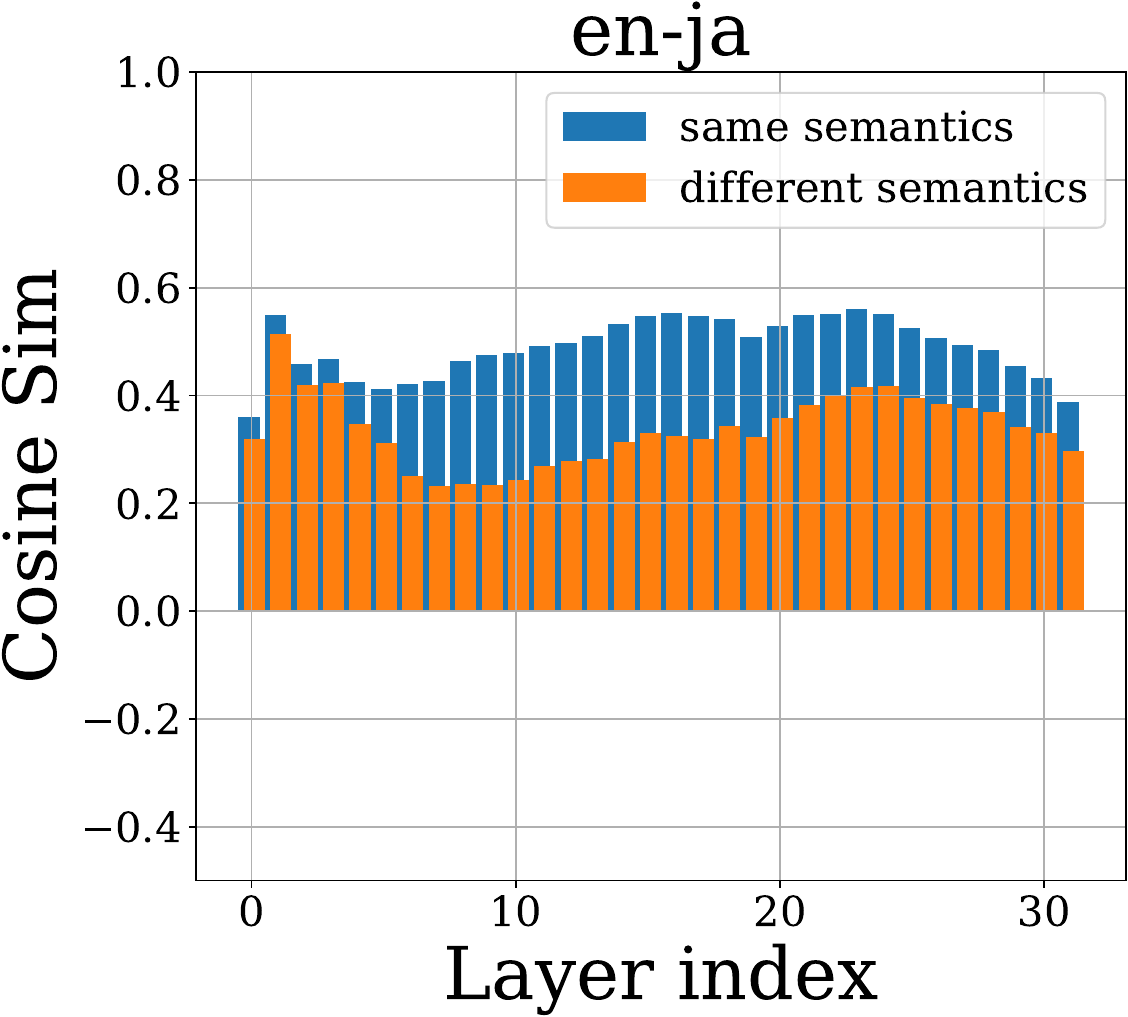}
      \subcaption{en-ja (baseline)}
    \end{minipage}
    \begin{minipage}{0.20\linewidth}
      \centering
      \includegraphics[width=\linewidth]{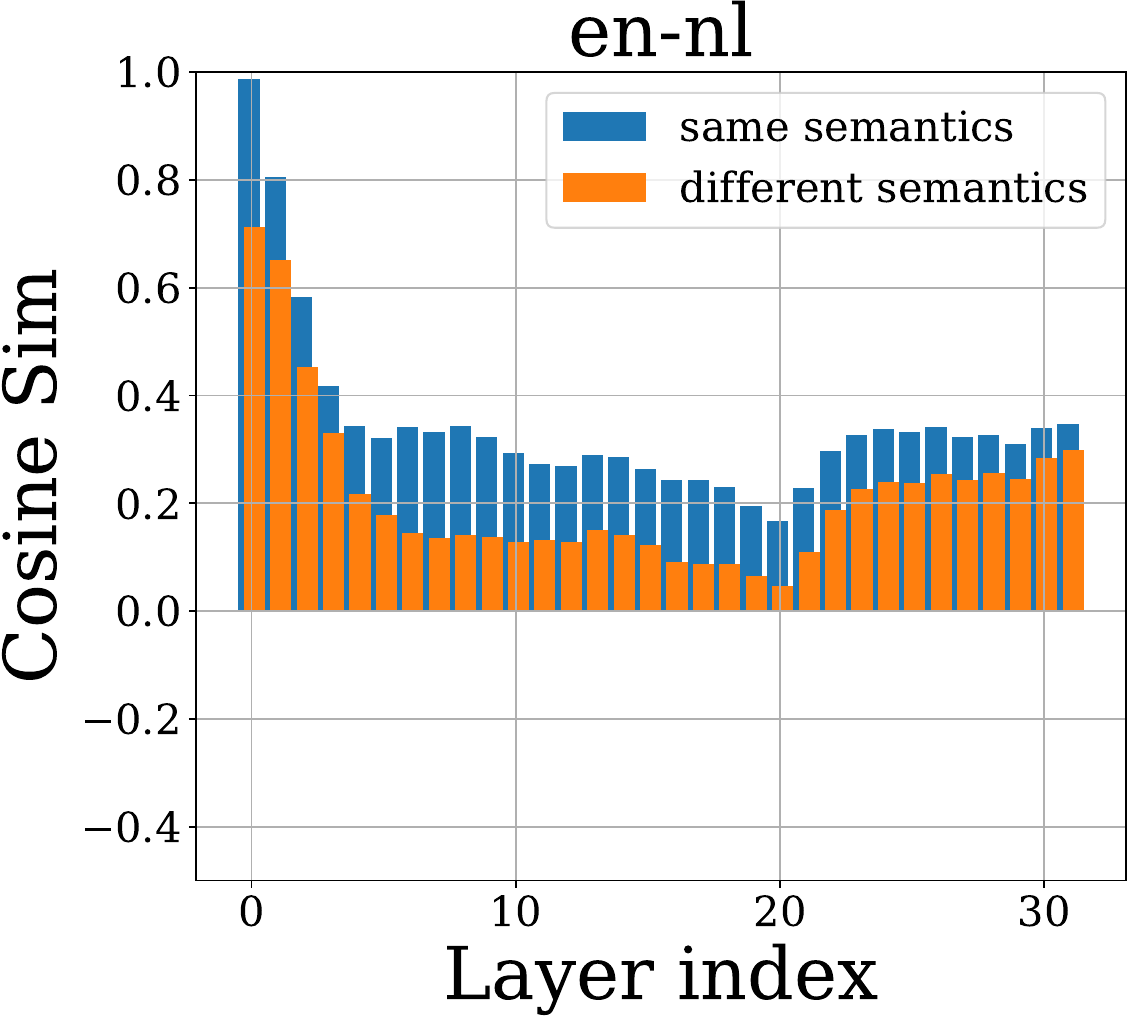}
      \subcaption{en-nl}
    \end{minipage}
    \begin{minipage}{0.20\linewidth}
      \centering
      \includegraphics[width=\linewidth]{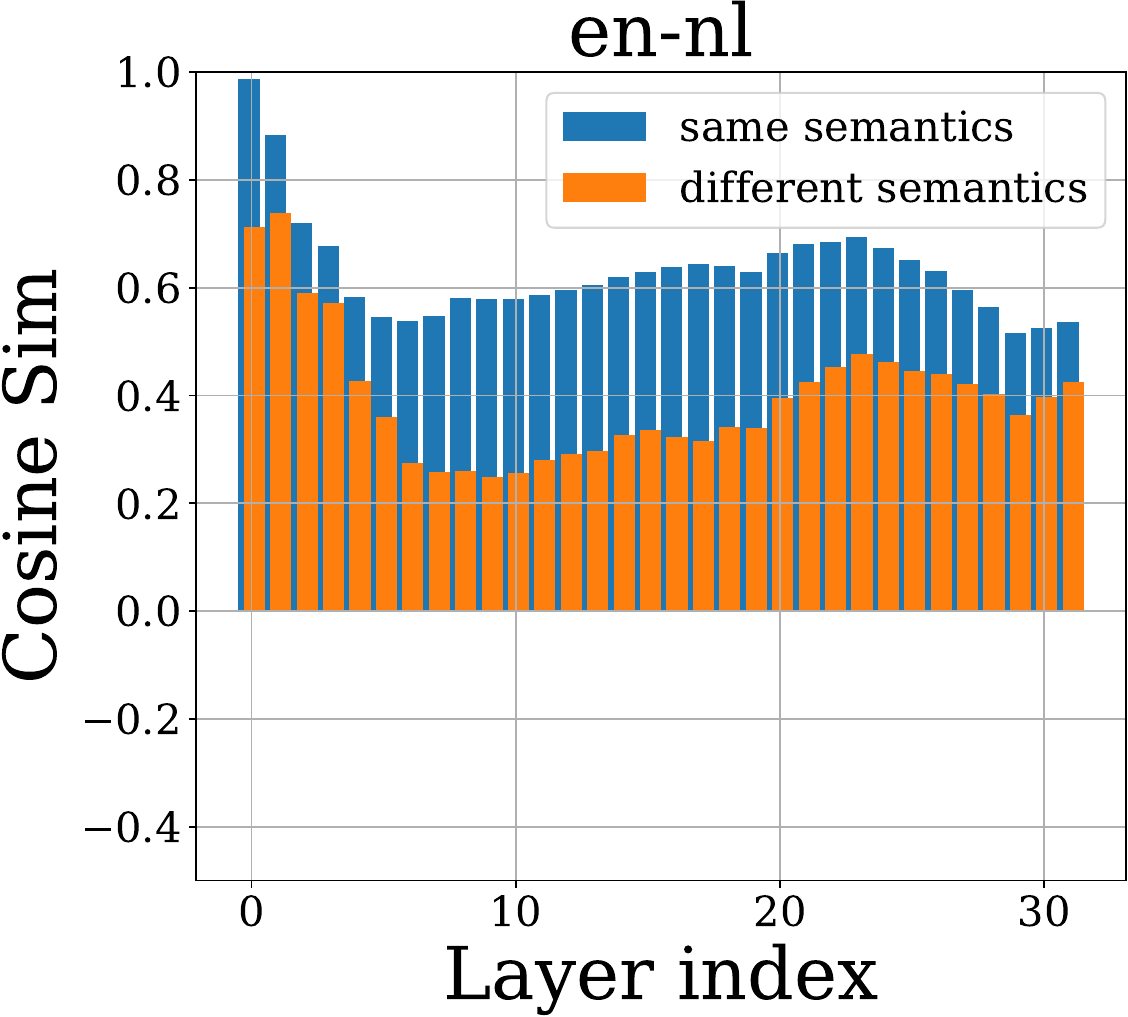}
      \subcaption{en-nl (baseline)}
    \end{minipage}

    \begin{minipage}{0.20\linewidth}
      \centering
      \includegraphics[width=\linewidth]{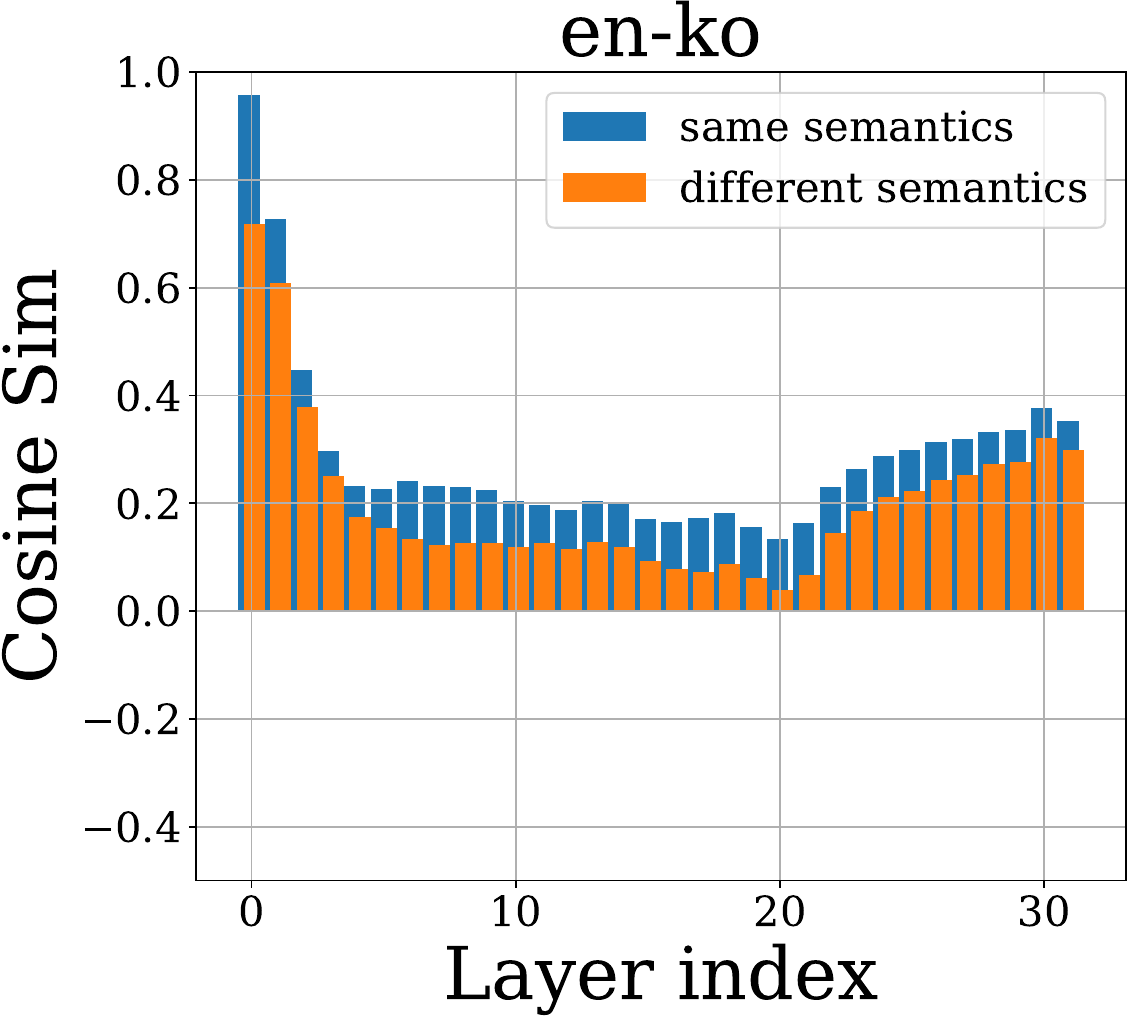}
      \subcaption{en-ko}
    \end{minipage}
    \begin{minipage}{0.20\linewidth}
      \centering
      \includegraphics[width=\linewidth]{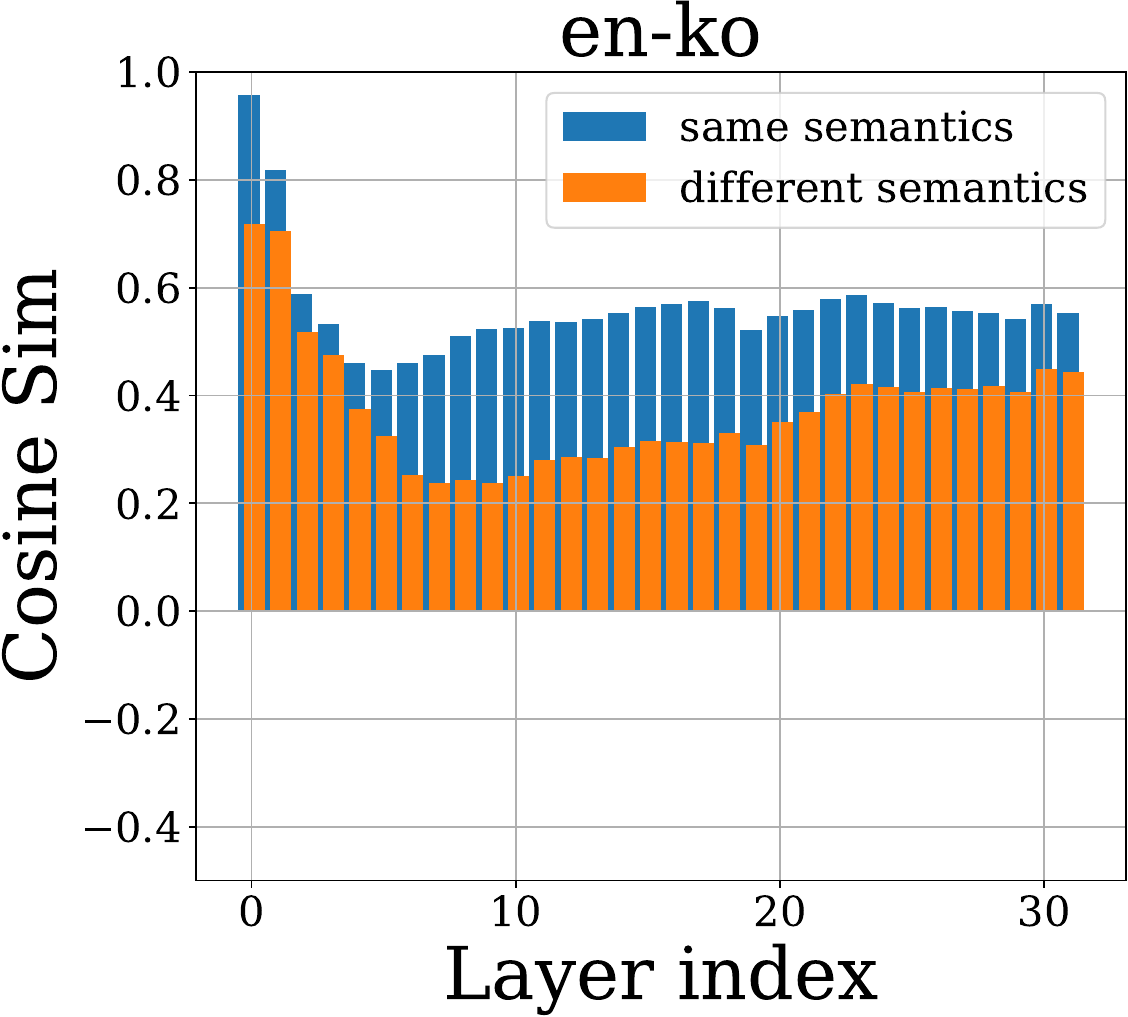}
      \subcaption{en-ko (baseline)}
    \end{minipage}
    \begin{minipage}{0.20\linewidth}
      \centering
      \includegraphics[width=\linewidth]{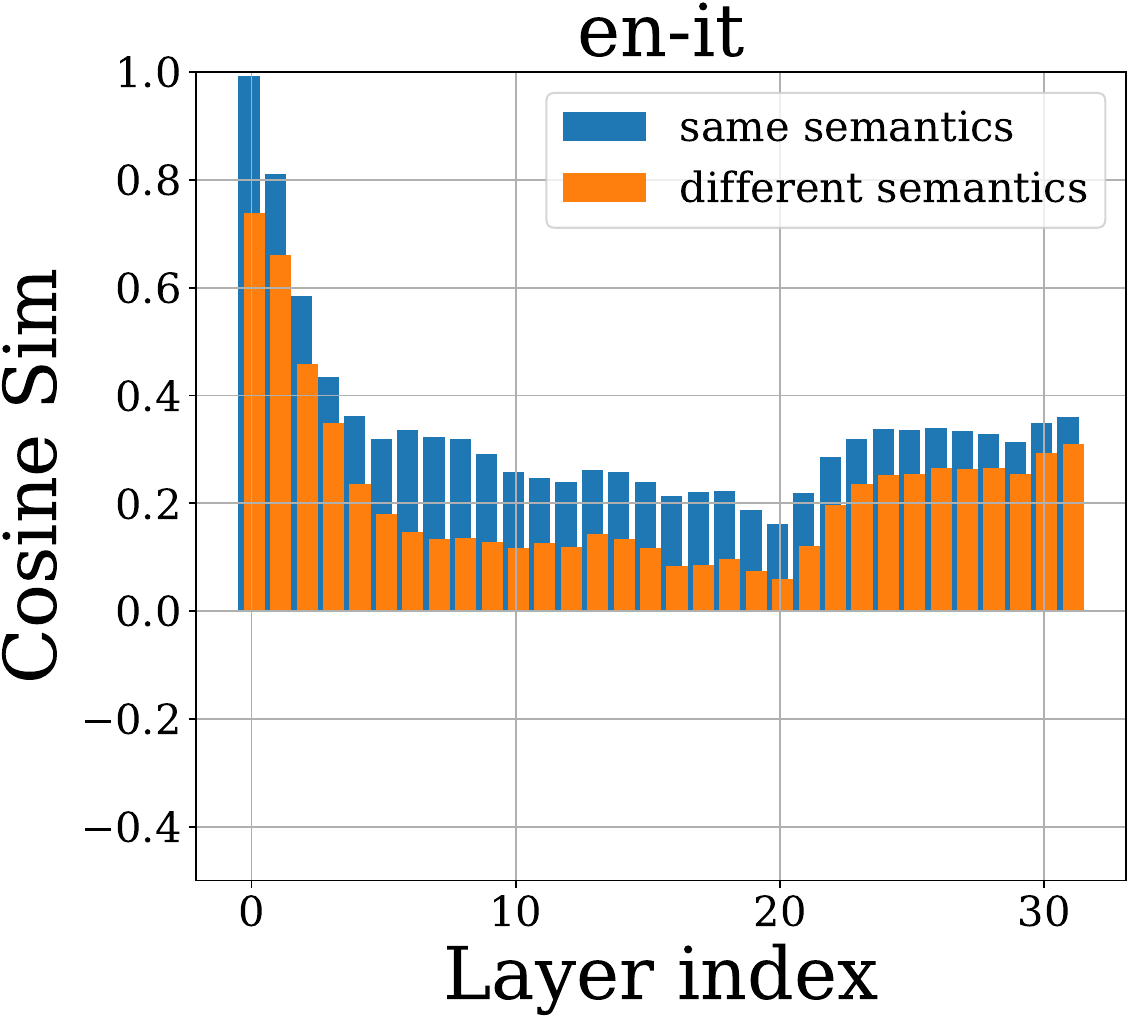}
      \subcaption{en-it}
    \end{minipage}
    \begin{minipage}{0.20\linewidth}
      \centering
      \includegraphics[width=\linewidth]{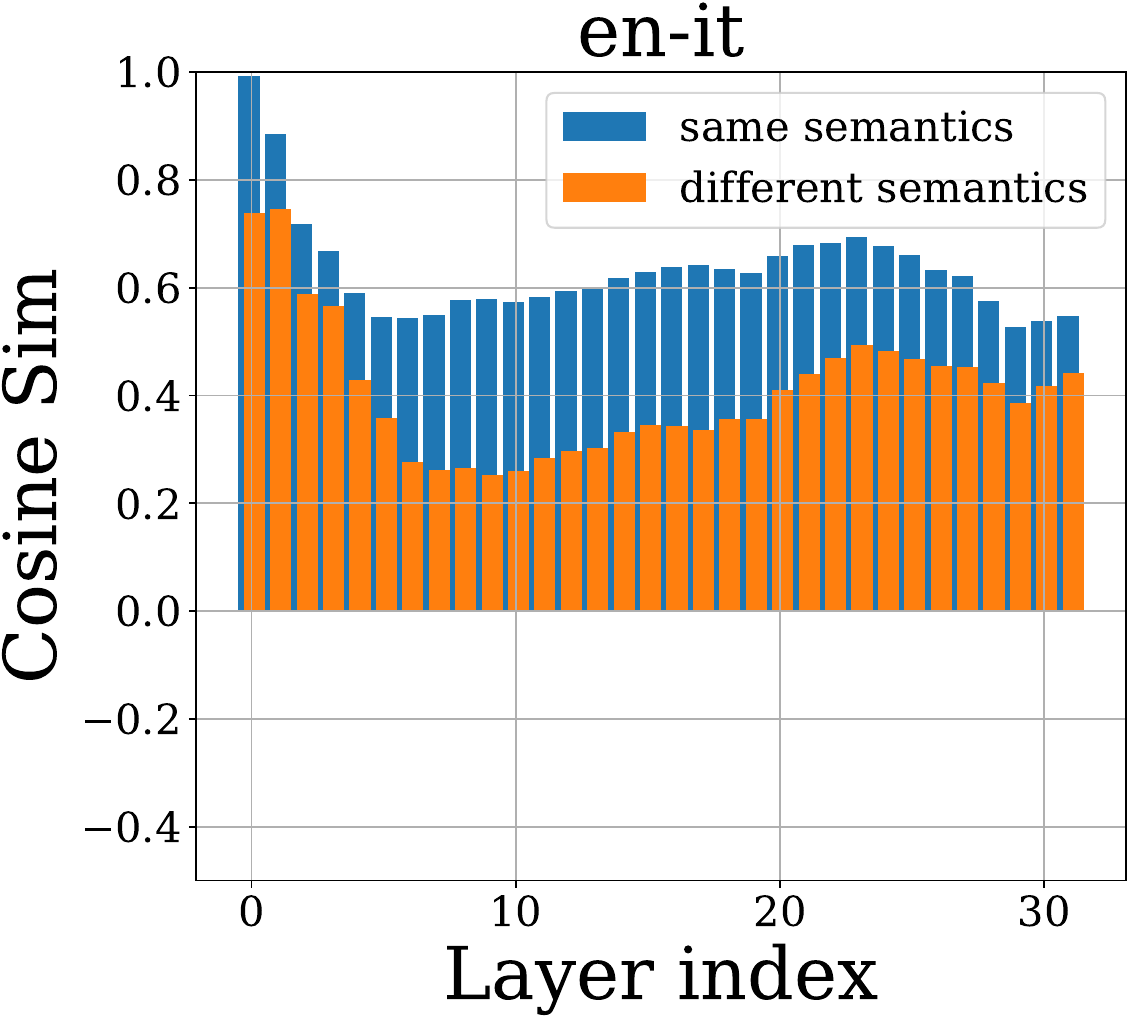}
      \subcaption{en-it (baseline)}
    \end{minipage}
    
      \begin{minipage}{\linewidth}
        \centering
        \small \textbf{(b) top-3000 (representing 0.6\% of all neurons)}
      \end{minipage}

    \begin{minipage}{0.20\linewidth}
      \centering
      \includegraphics[width=\linewidth]{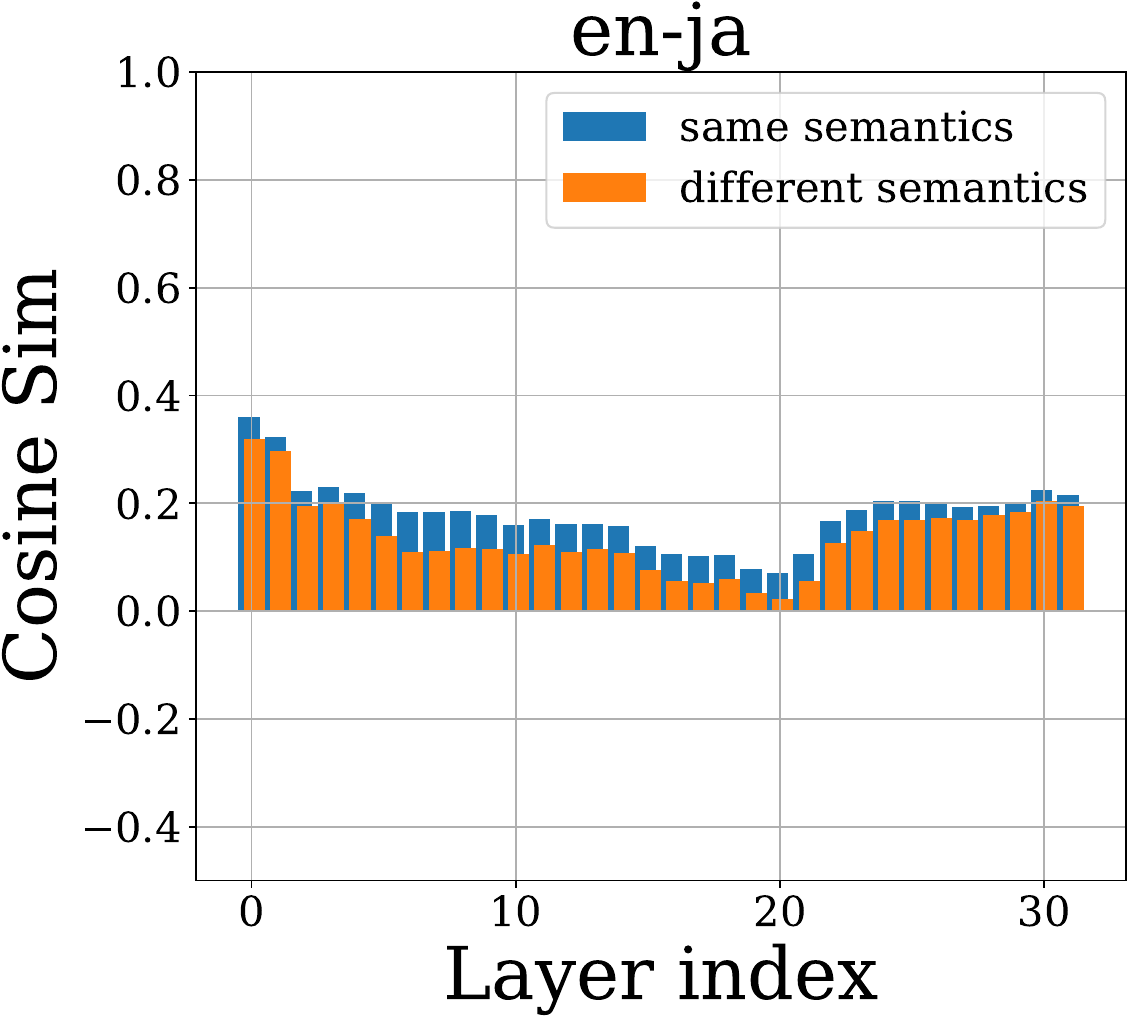}
      \subcaption{en-ja}
    \end{minipage}
    \begin{minipage}{0.20\linewidth}
      \centering
      \includegraphics[width=\linewidth]{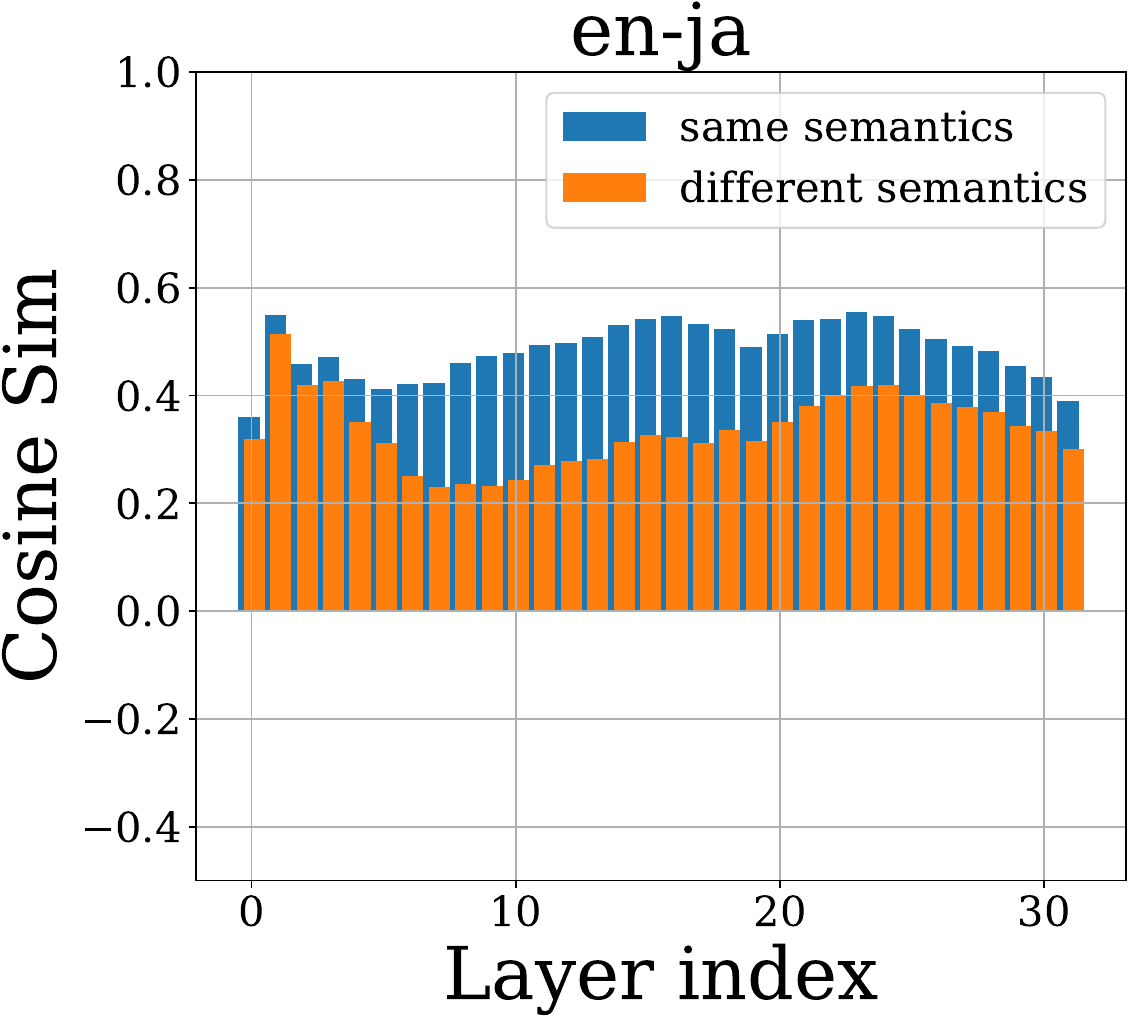}
      \subcaption{en-ja (baseline)}
    \end{minipage}
    \begin{minipage}{0.20\linewidth}
      \centering
      \includegraphics[width=\linewidth]{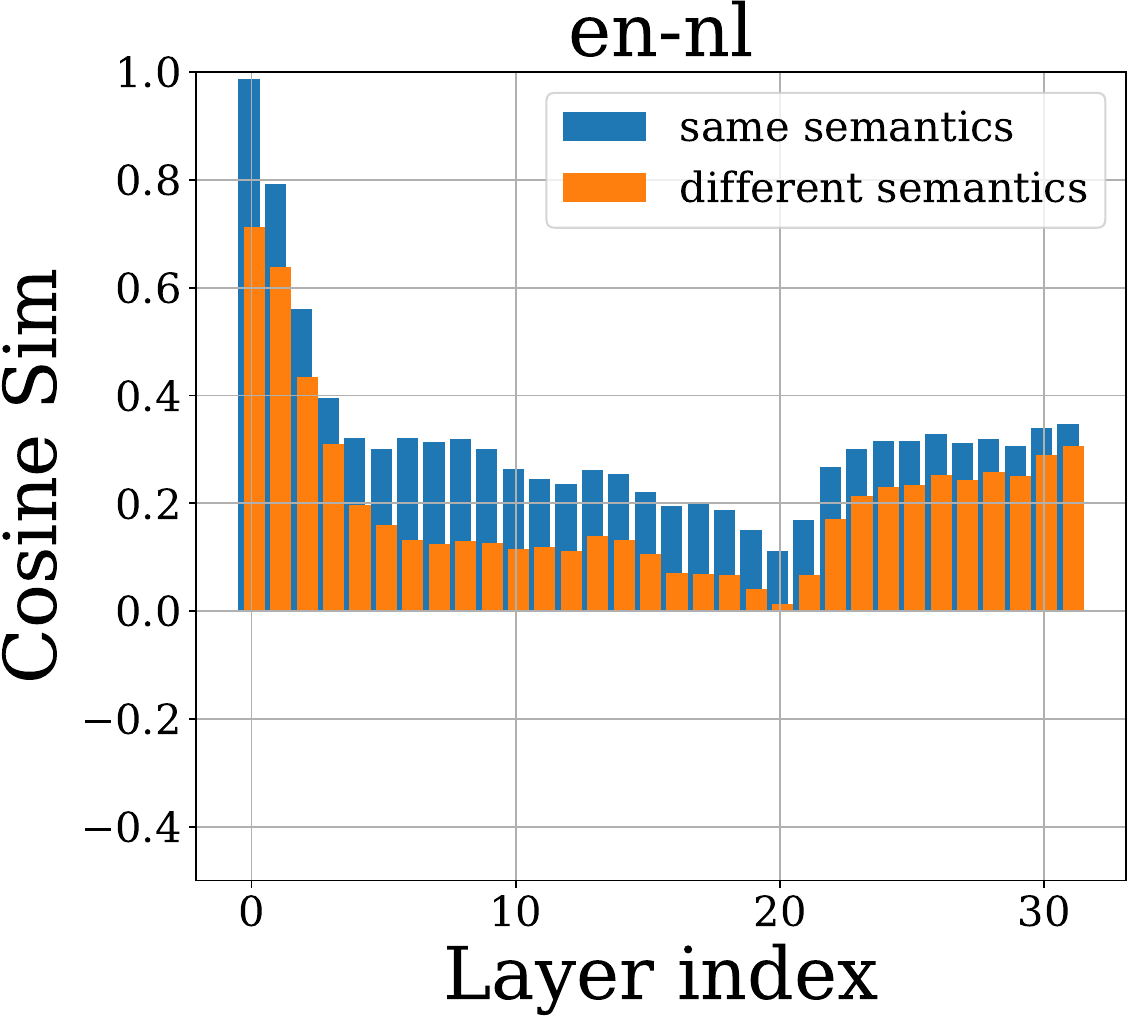}
      \subcaption{en-nl}
    \end{minipage}
    \begin{minipage}{0.20\linewidth}
      \centering
      \includegraphics[width=\linewidth]{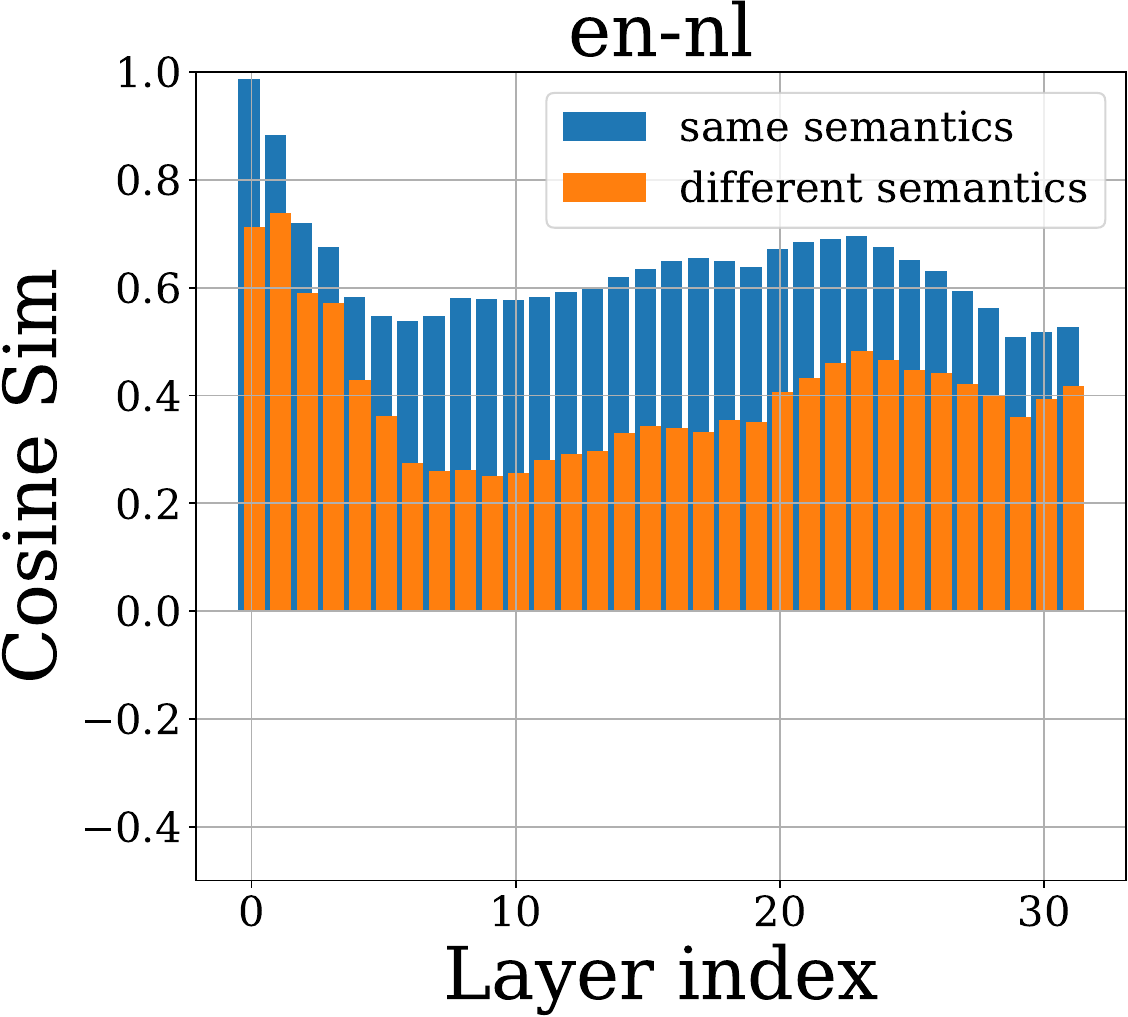}
      \subcaption{en-nl (baseline)}
    \end{minipage}

    \begin{minipage}{0.20\linewidth}
      \centering
      \includegraphics[width=\linewidth]{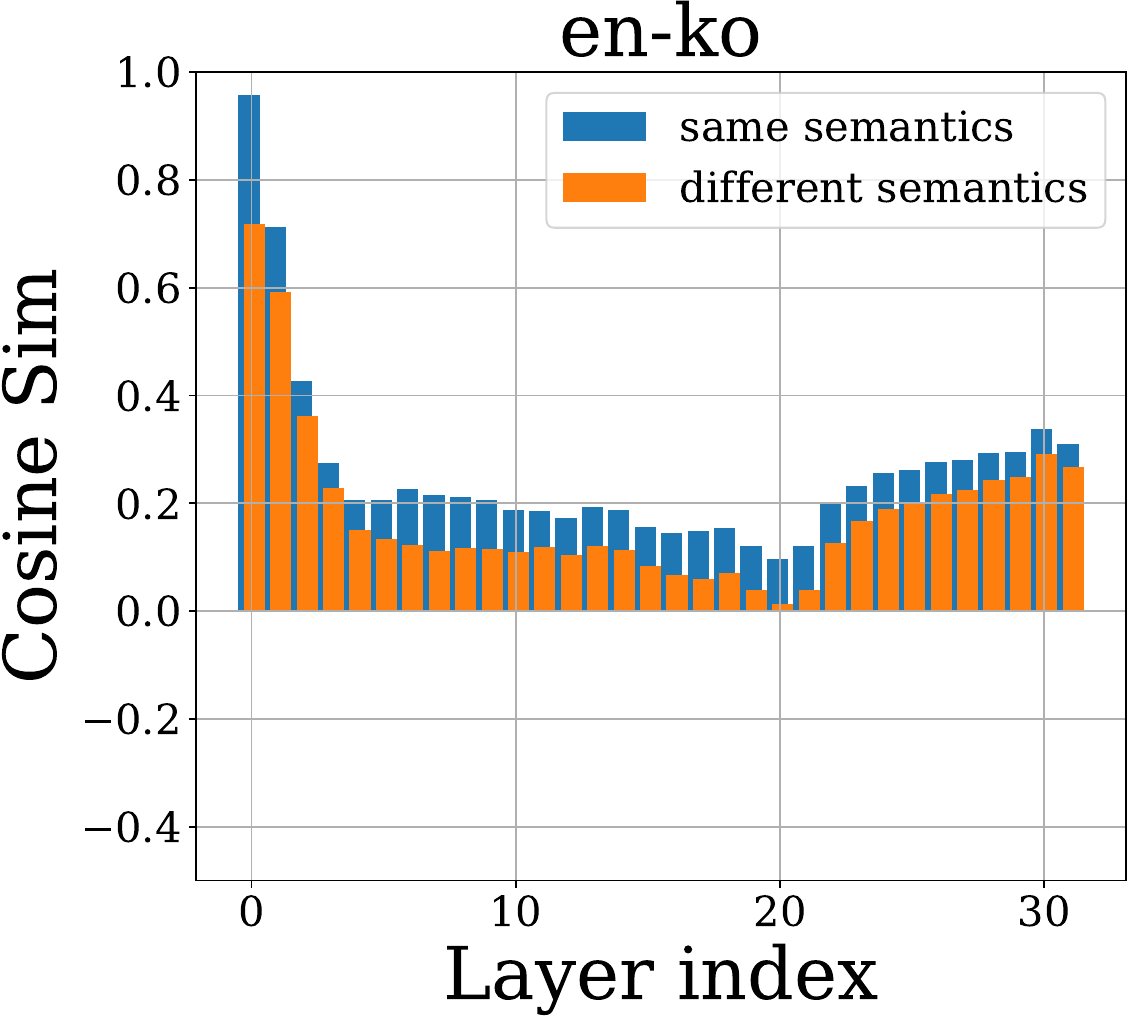}
      \subcaption{en-ko}
    \end{minipage}
    \begin{minipage}{0.20\linewidth}
      \centering
      \includegraphics[width=\linewidth]{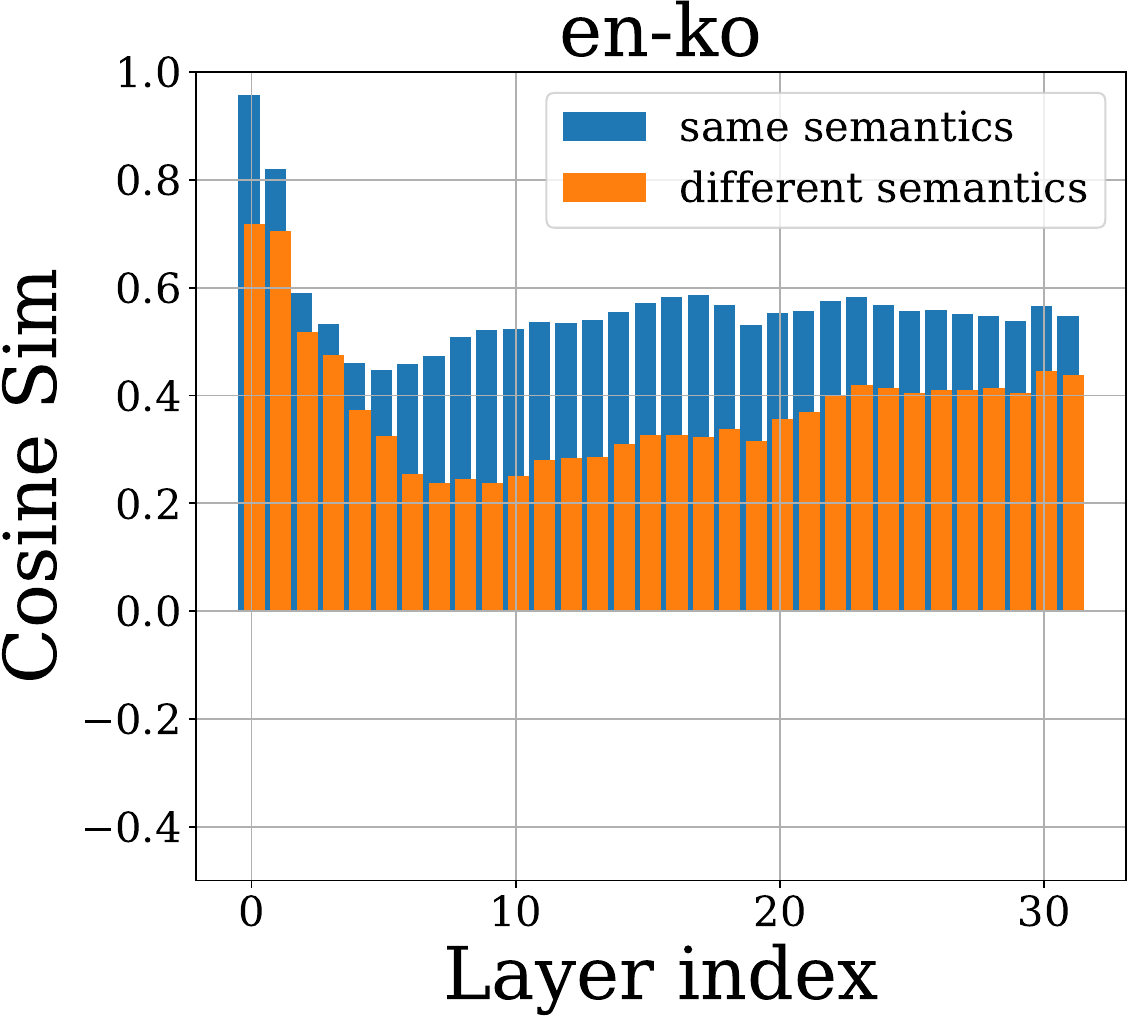}
      \subcaption{en-ko (baseline)}
    \end{minipage}
    \begin{minipage}{0.20\linewidth}
      \centering
      \includegraphics[width=\linewidth]{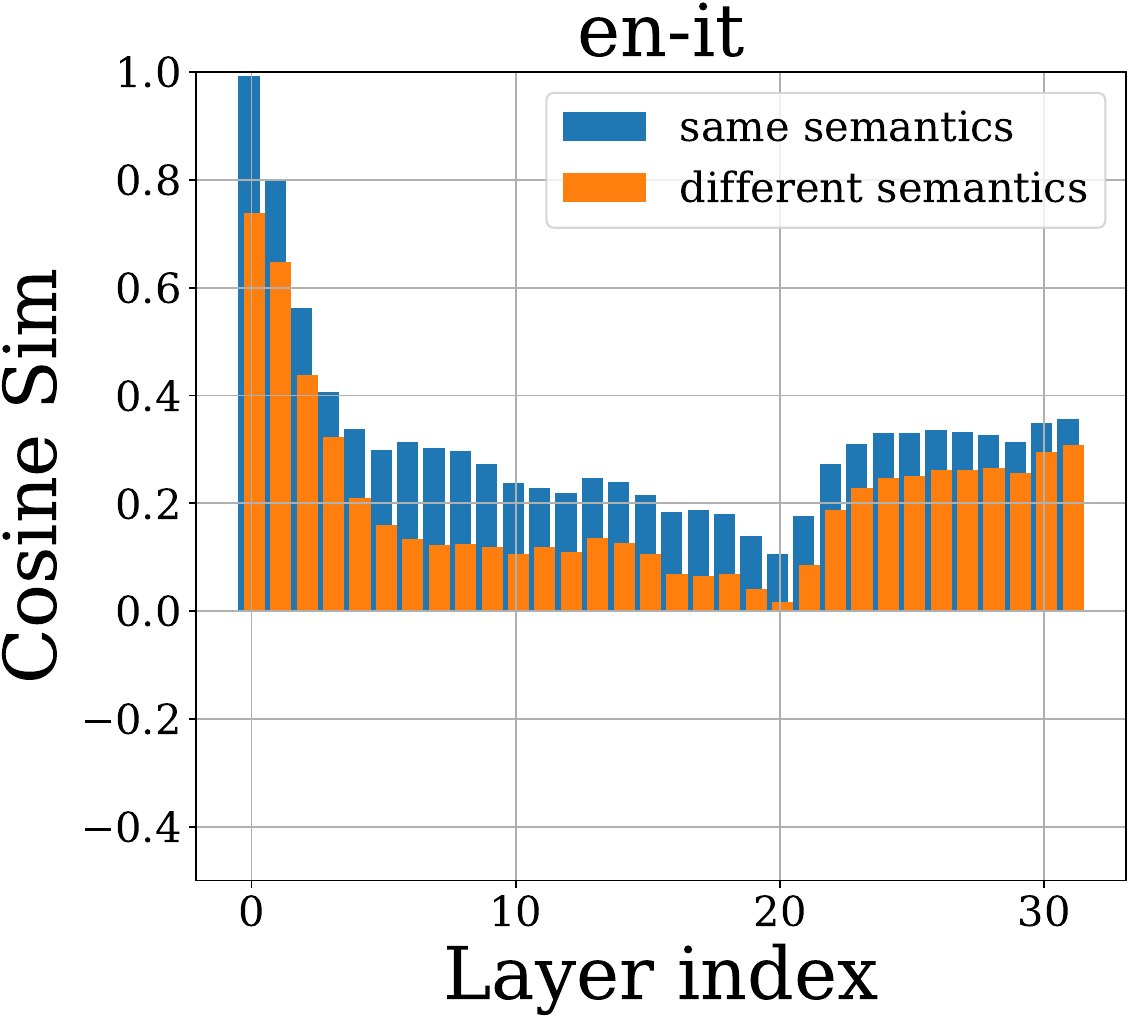}
      \subcaption{en-it}
    \end{minipage}
    \begin{minipage}{0.20\linewidth}
      \centering
      \includegraphics[width=\linewidth]{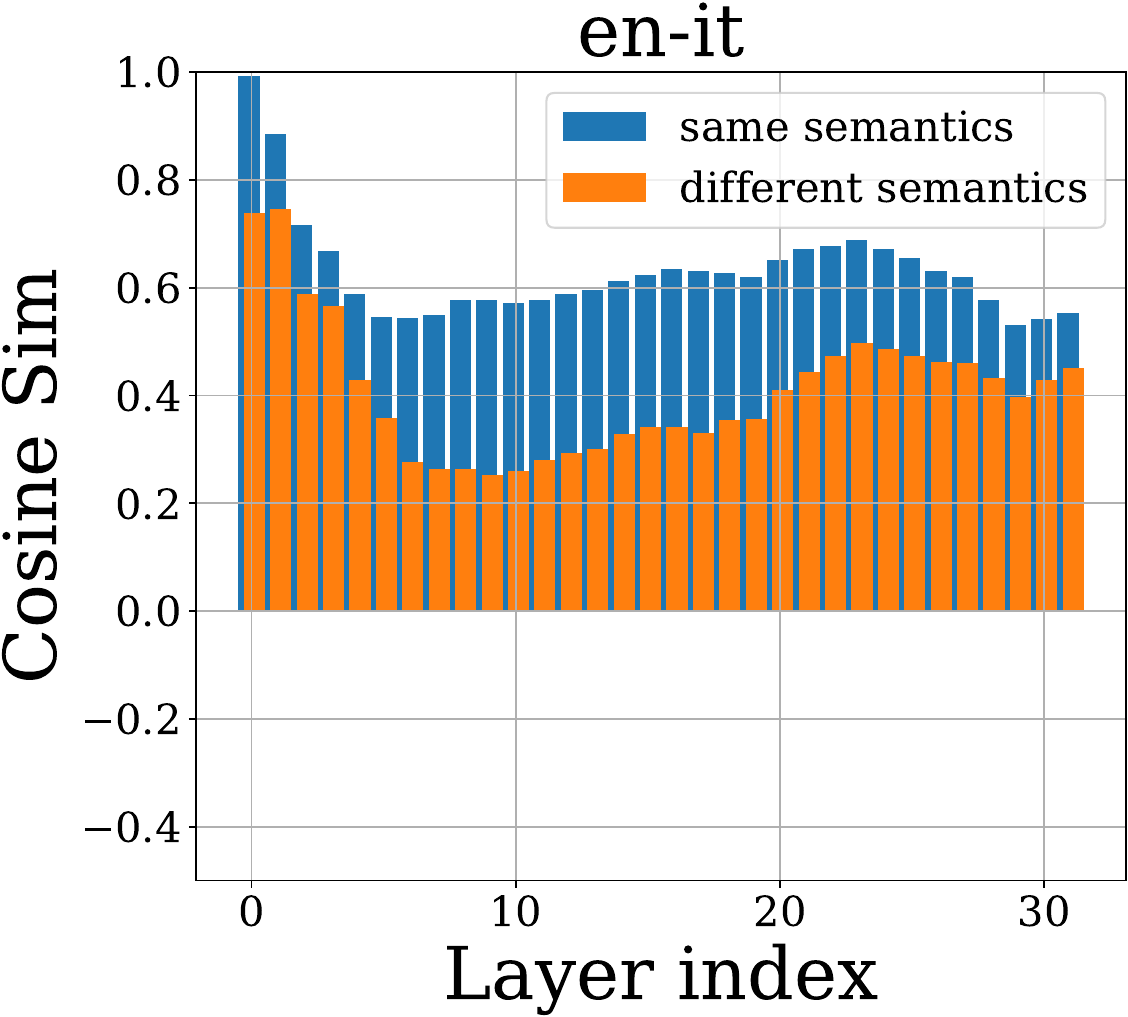}
      
      \subcaption{en-it (baseline)}
    \end{minipage}
    
      \begin{minipage}{\linewidth}
        \centering
        \small \textbf{(b) top-5000 (representing 1\% of all neurons)}
      \end{minipage}

  \caption{\textbf{Similarity of hidden states across layers while deactivating Type-1 Transfer Neurons (Mistral-7B).}}
  \label{fig:appendix:hs_sim_mistral_deactivating_top-1k_Type-1}
\end{figure*}

\begin{figure*}[t]
    \centering

    \begin{minipage}{0.20\linewidth}
      \centering
      \includegraphics[width=\linewidth]{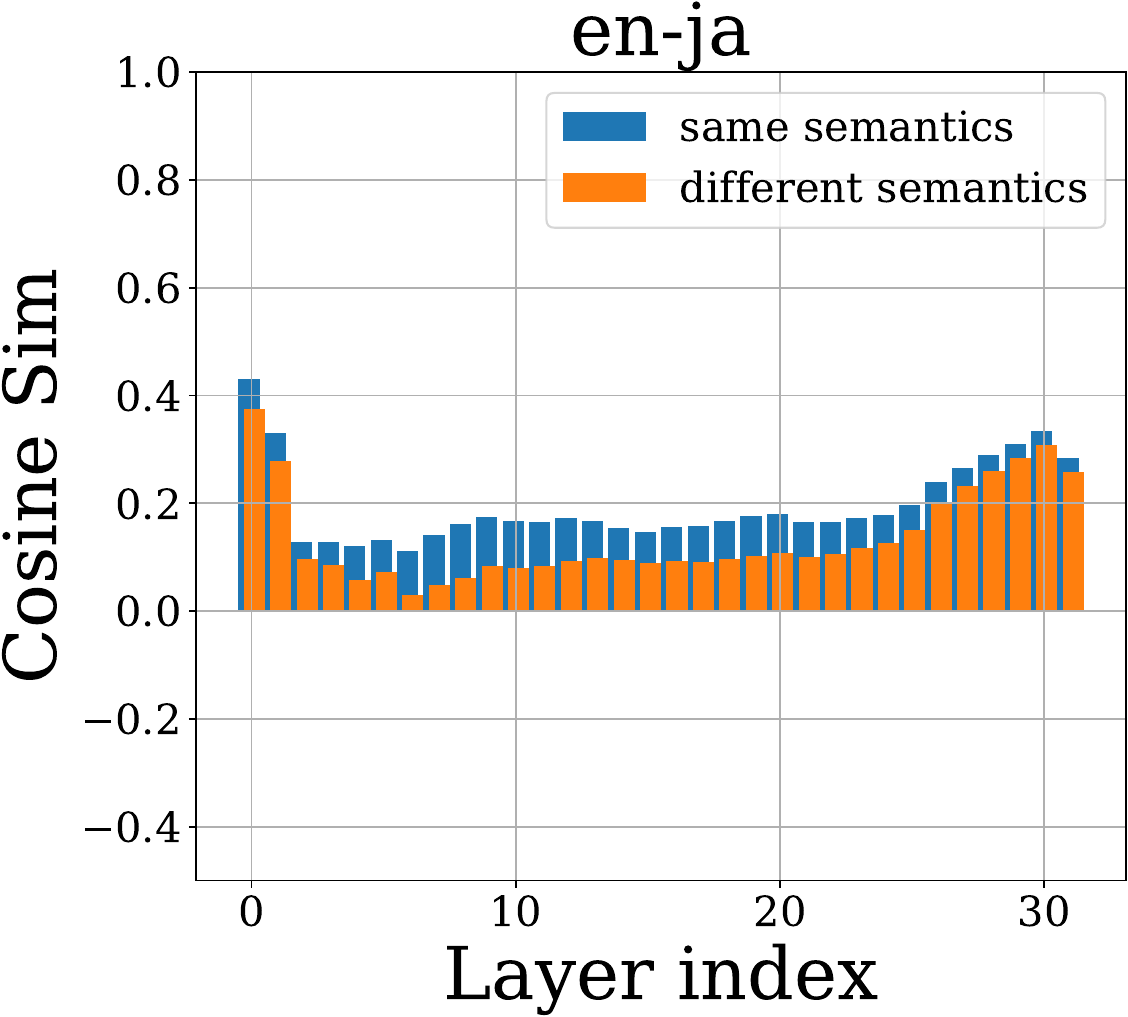}
      \subcaption{en-ja}
    \end{minipage}
    \begin{minipage}{0.20\linewidth}
      \centering
      \includegraphics[width=\linewidth]{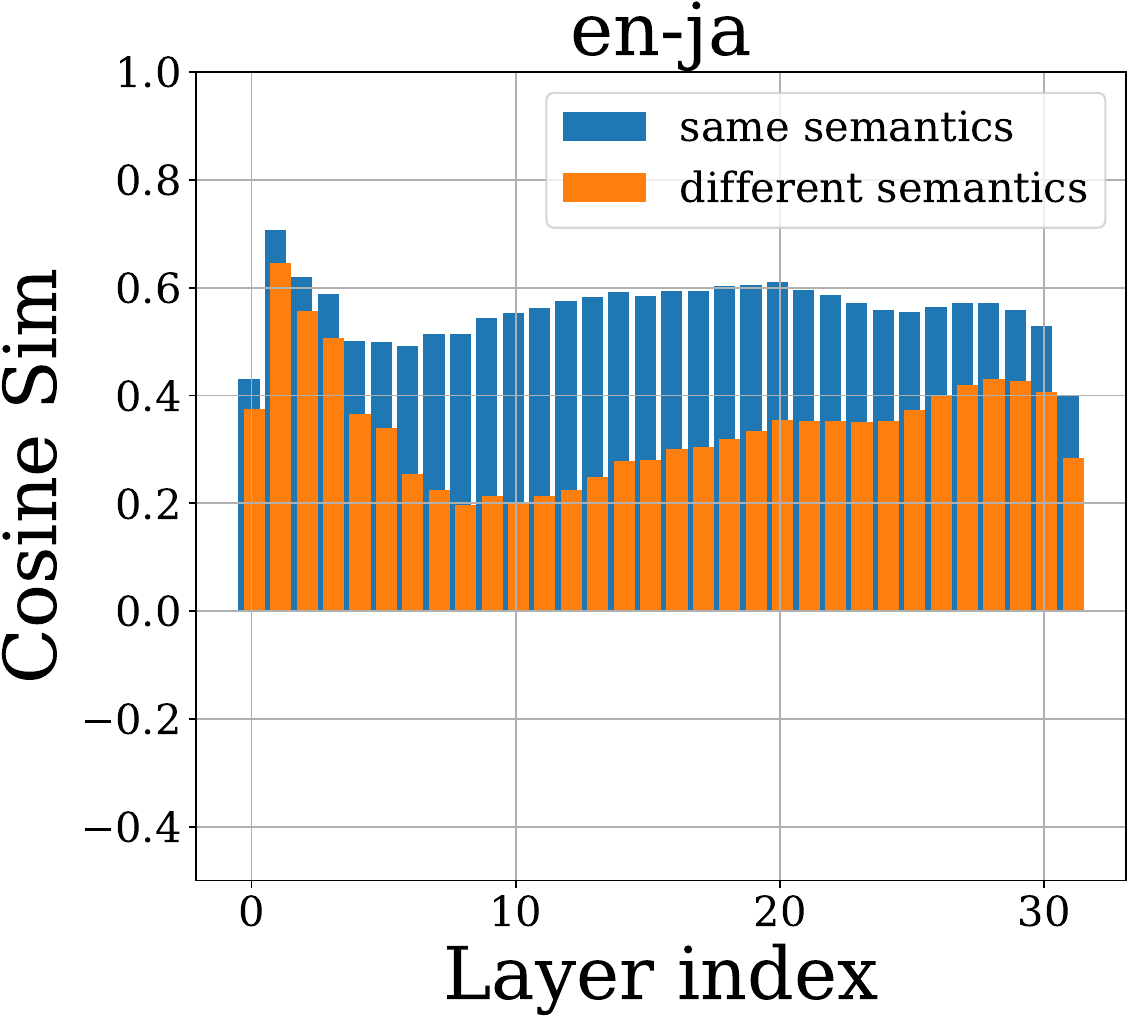}
      \subcaption{en-ja (baseline)}
    \end{minipage}
    \begin{minipage}{0.20\linewidth}
      \centering
      \includegraphics[width=\linewidth]{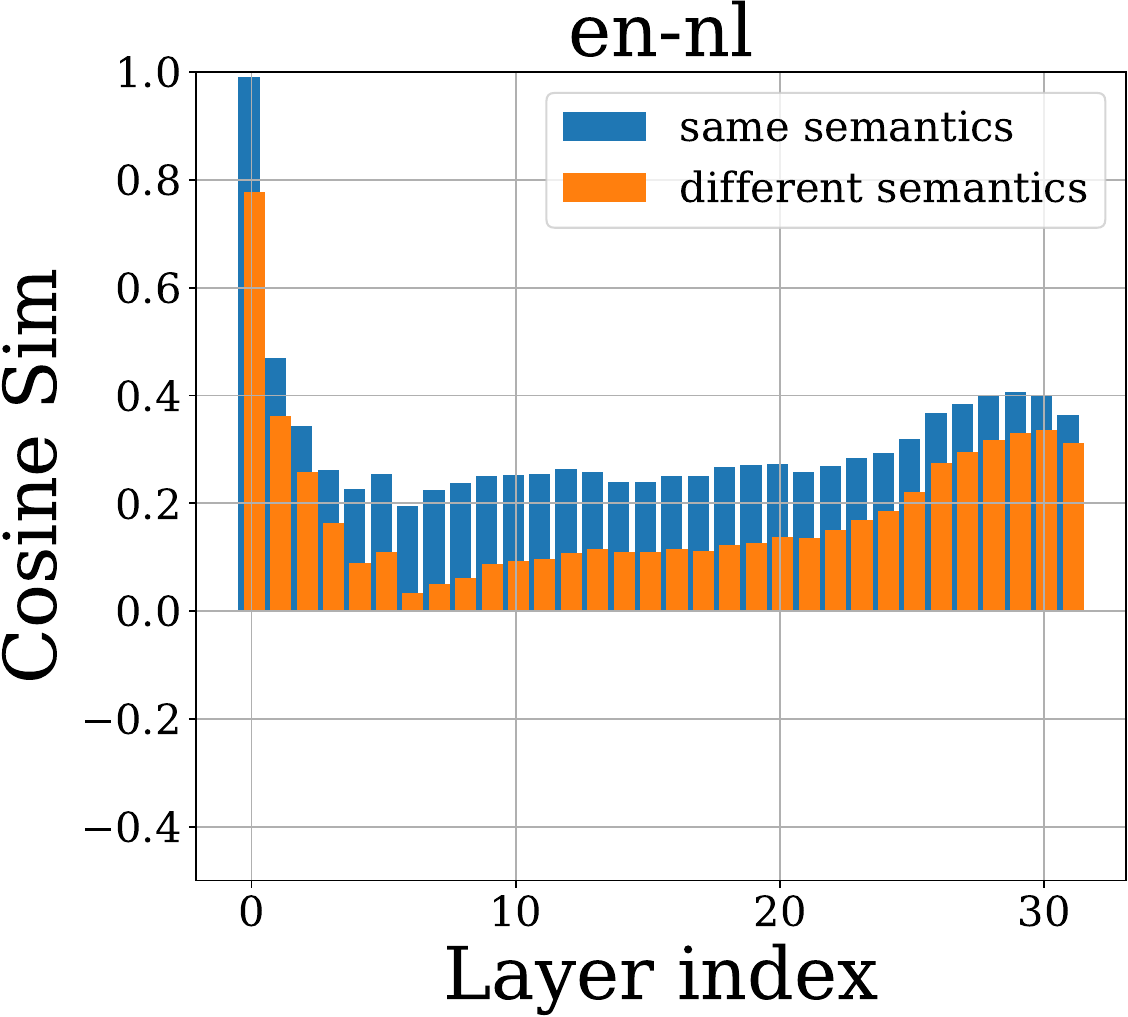}
      \subcaption{en-nl}
    \end{minipage}
    \begin{minipage}{0.20\linewidth}
      \centering
      \includegraphics[width=\linewidth]{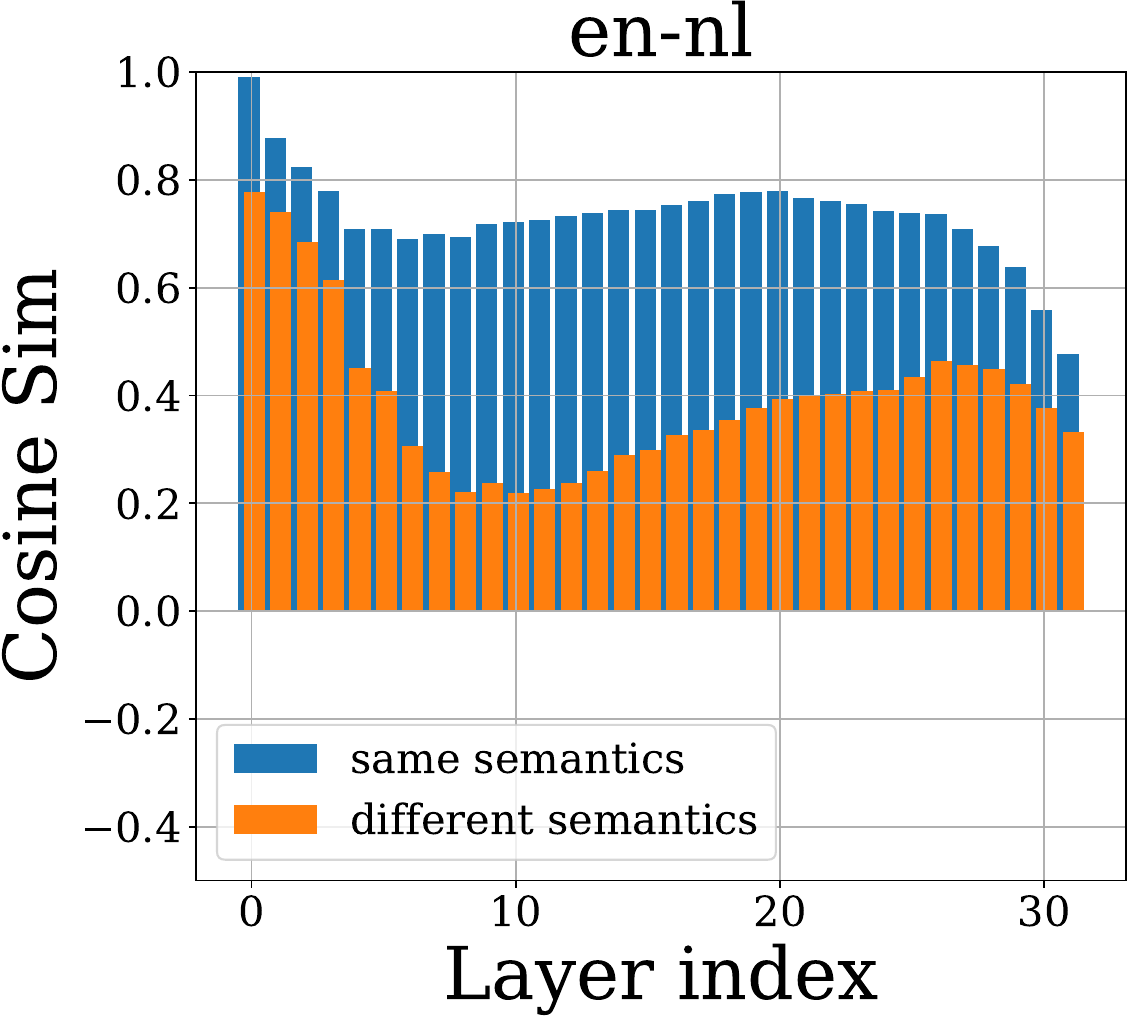}
      \subcaption{en-nl (baseline)}
    \end{minipage}

    \begin{minipage}{0.20\linewidth}
      \centering
      \includegraphics[width=\linewidth]{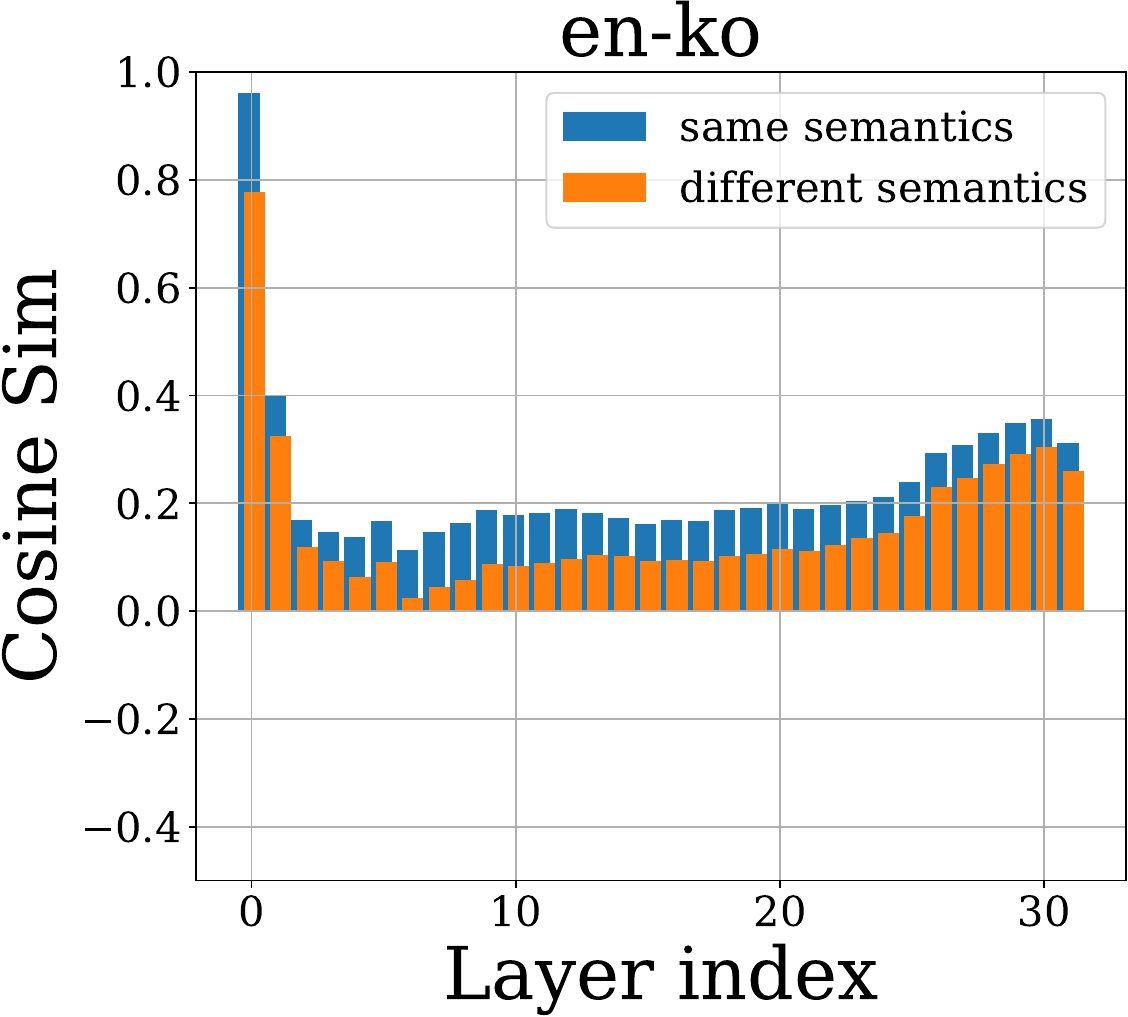}
      \subcaption{en-ko}
    \end{minipage}
    \begin{minipage}{0.20\linewidth}
      \centering
      \includegraphics[width=\linewidth]{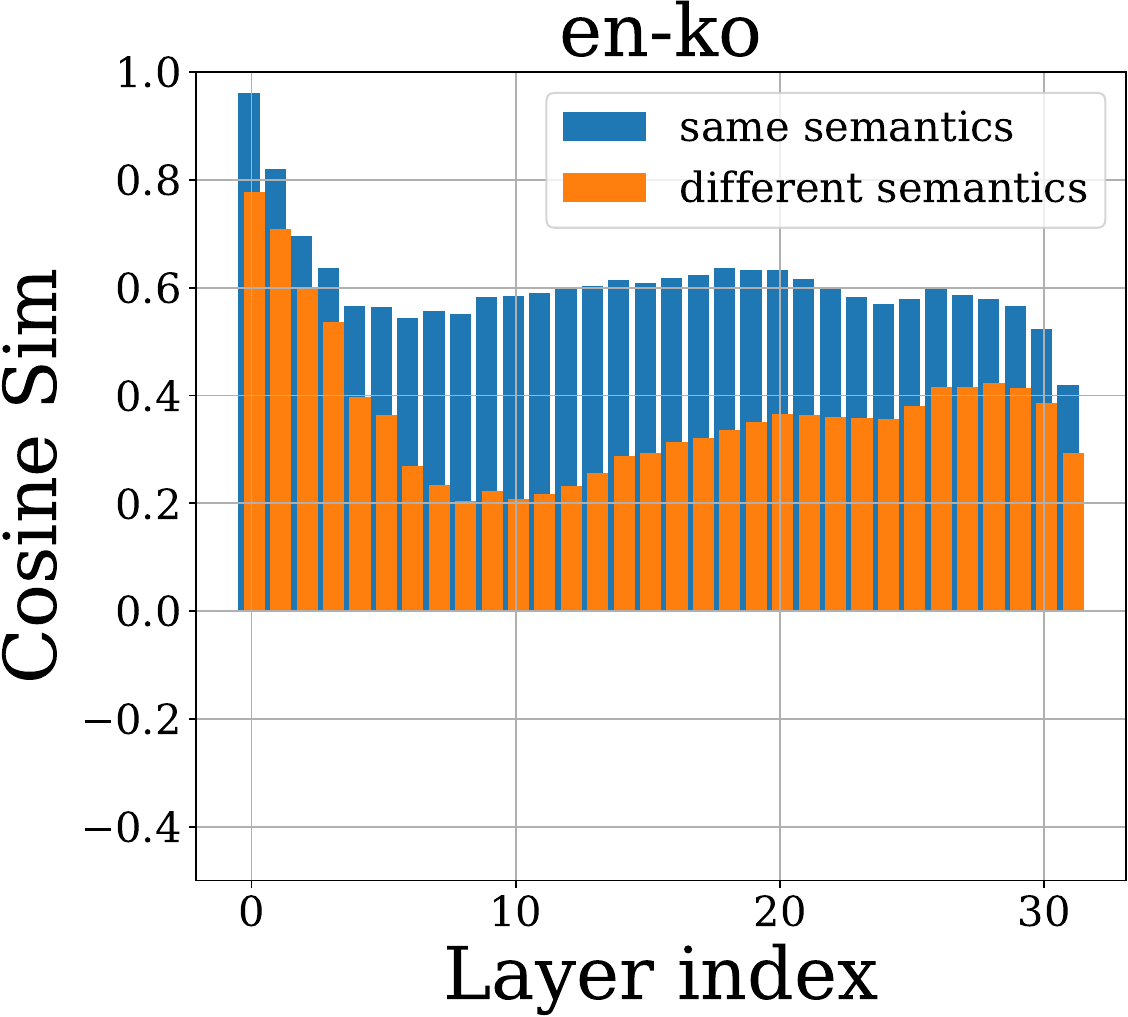}
      \subcaption{en-ko (baseline)}
    \end{minipage}
    \begin{minipage}{0.20\linewidth}
      \centering
      \includegraphics[width=\linewidth]{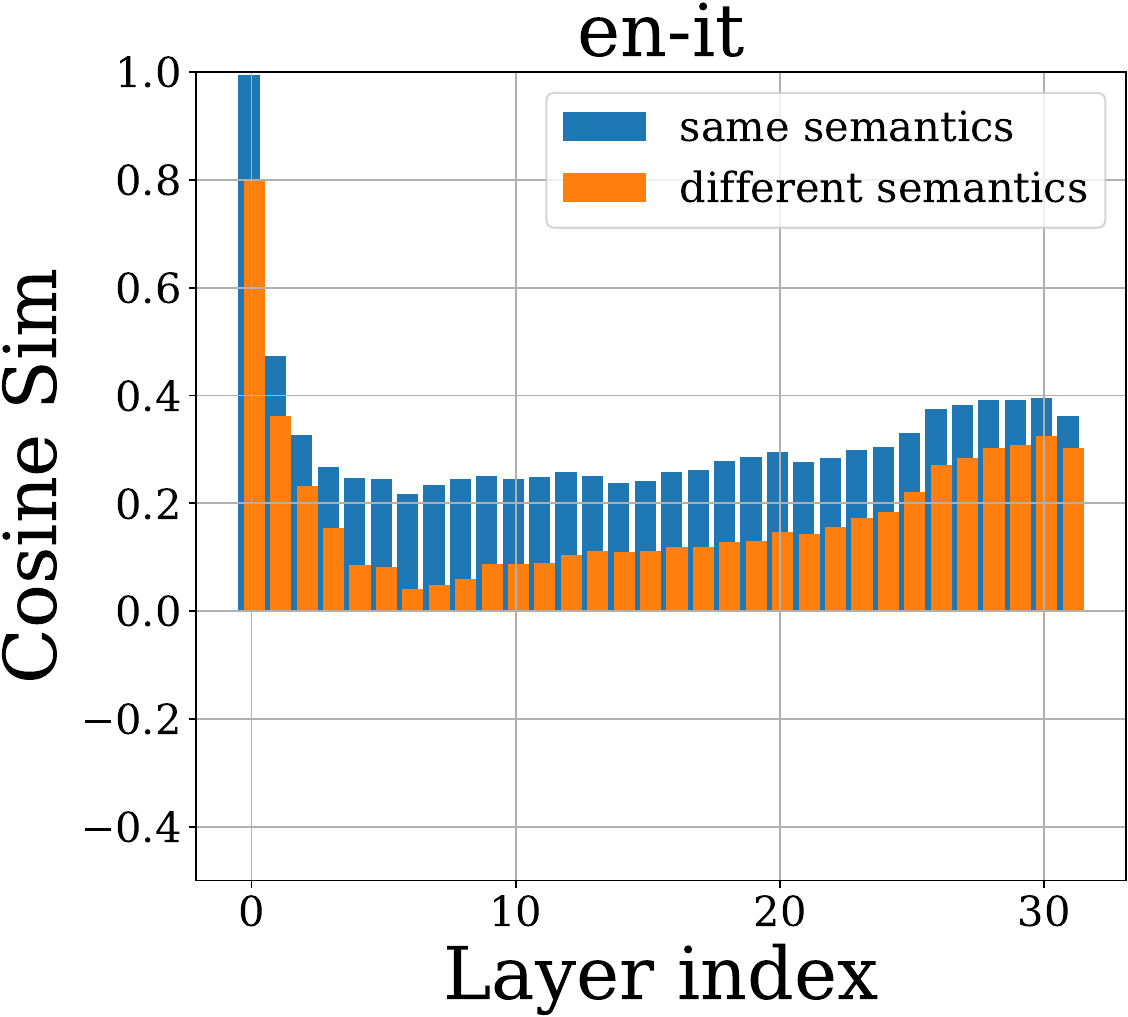}
      \subcaption{en-it}
    \end{minipage}
    \begin{minipage}{0.20\linewidth}
      \centering
      \includegraphics[width=\linewidth]{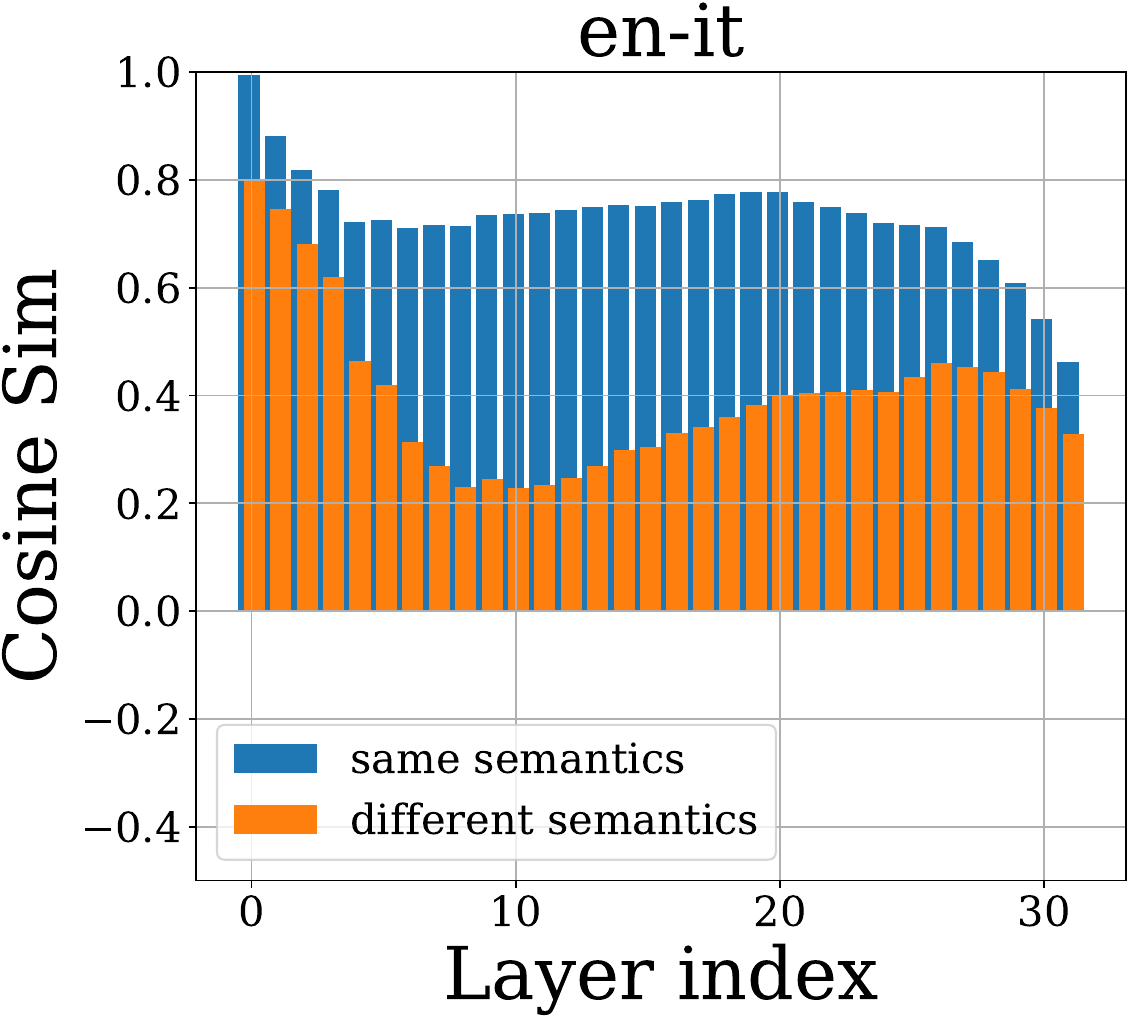}
      \subcaption{en-it (baseline)}
    \end{minipage}
    
      % First row label
      \begin{minipage}{\linewidth}
        \centering
        \small \textbf{(a) top-1000 (representing 0.2\% of all neurons)}
      \end{minipage}

    % second row
    \begin{minipage}{0.20\linewidth}
      \centering
      \includegraphics[width=\linewidth]{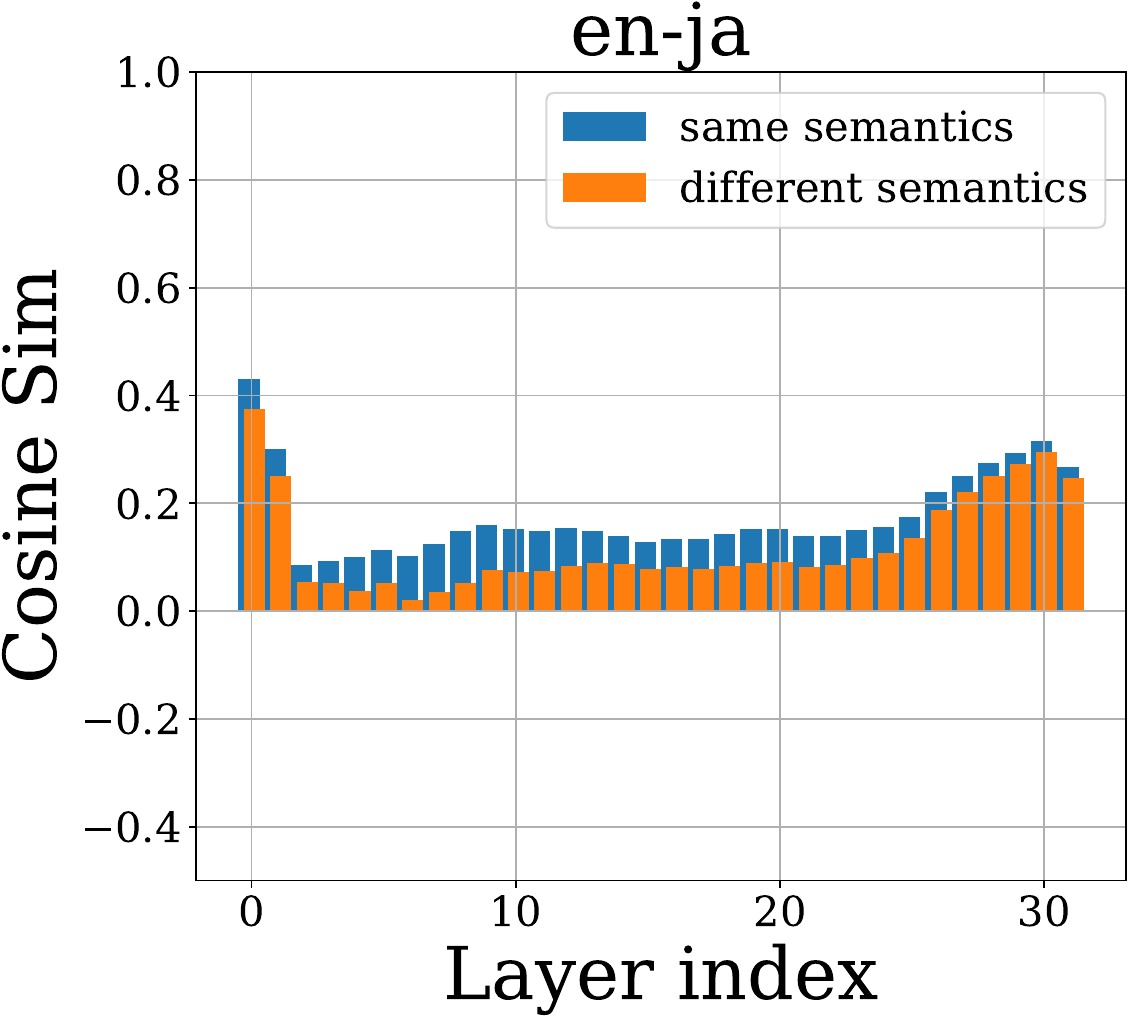}
      \subcaption{en-ja}
    \end{minipage}
    \begin{minipage}{0.20\linewidth}
      \centering
      \includegraphics[width=\linewidth]{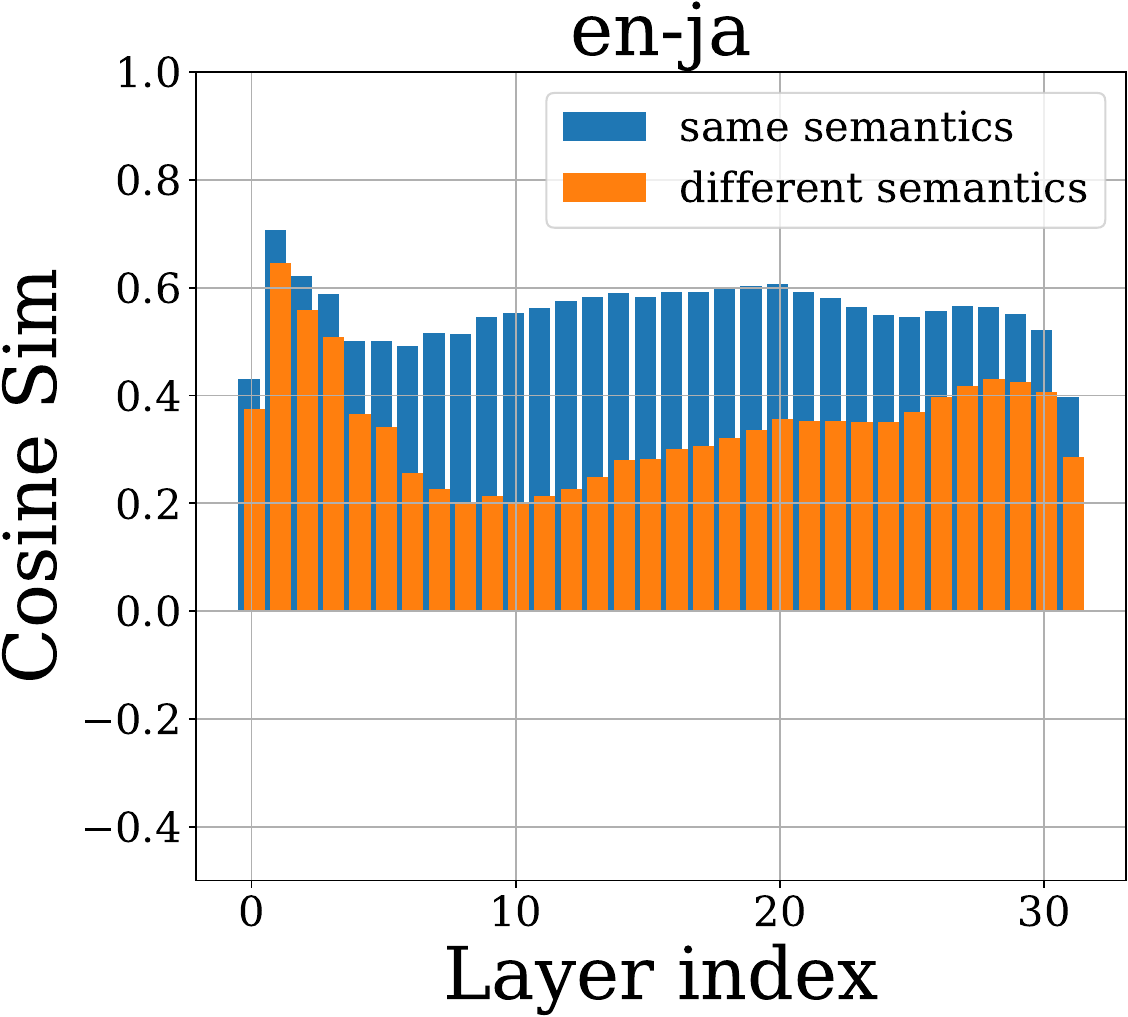}
      \subcaption{en-ja (baseline)}
    \end{minipage}
    \begin{minipage}{0.20\linewidth}
      \centering
      \includegraphics[width=\linewidth]{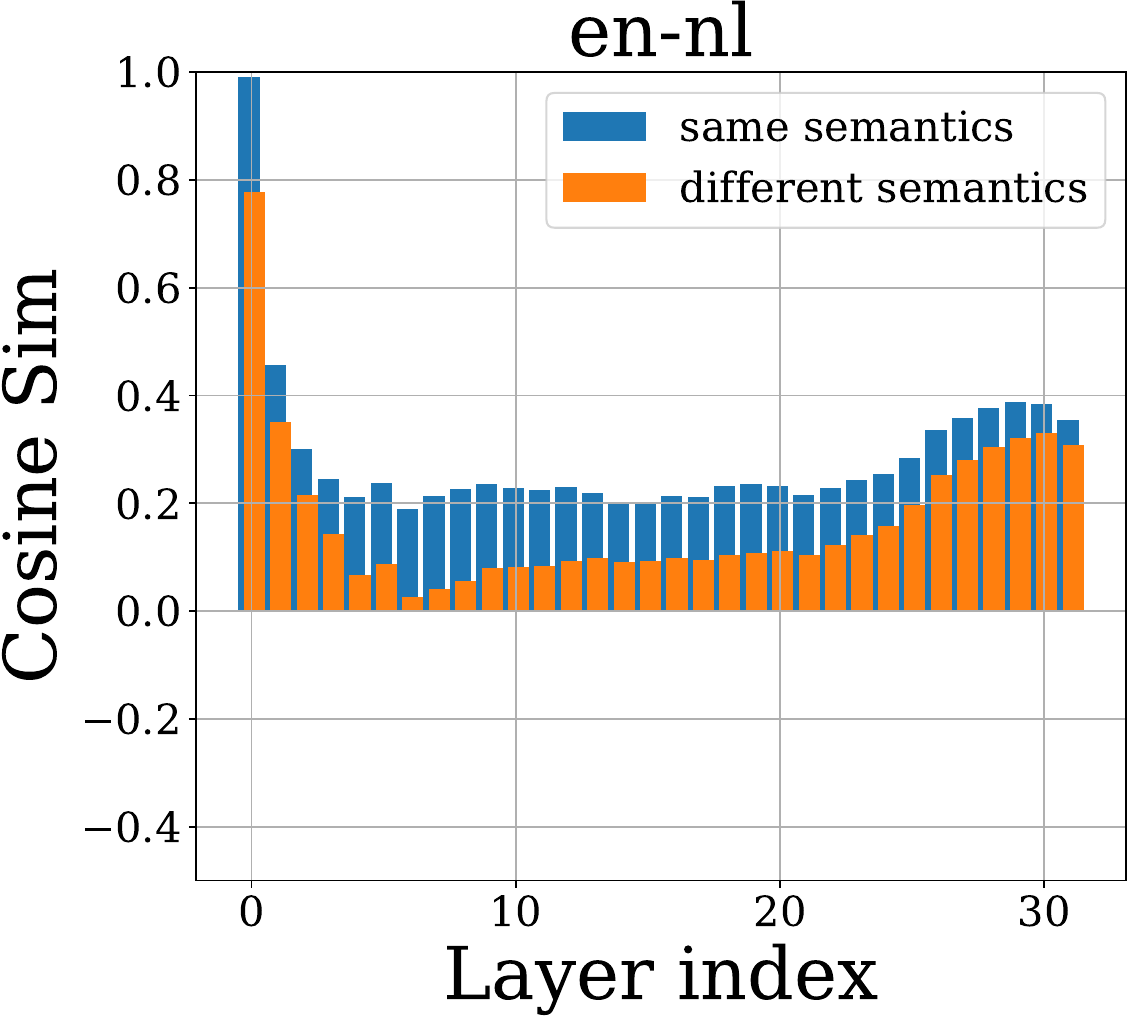}
      \subcaption{en-nl}
    \end{minipage}
    \begin{minipage}{0.20\linewidth}
      \centering
      \includegraphics[width=\linewidth]{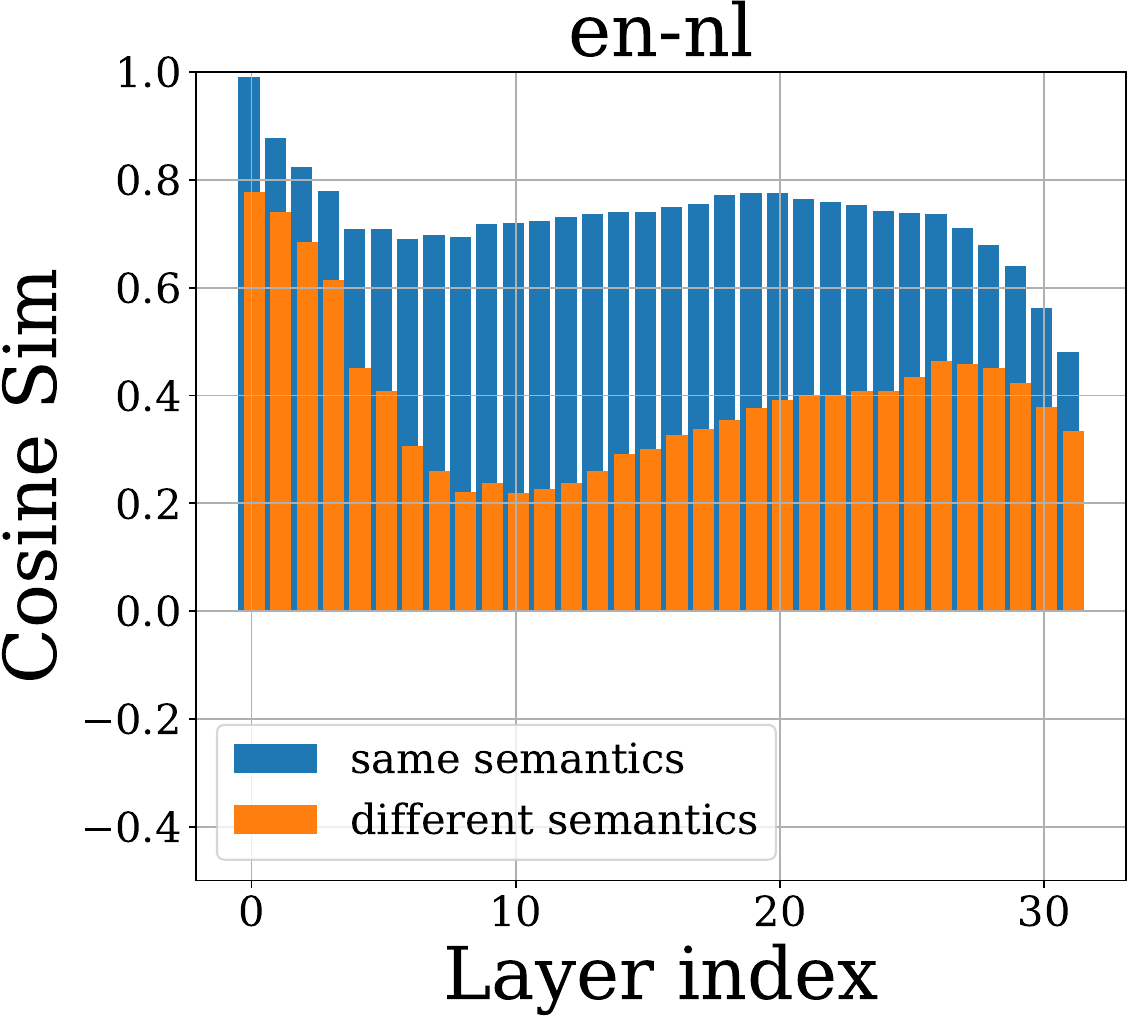}
      \subcaption{en-nl (baseline)}
    \end{minipage}

    \begin{minipage}{0.20\linewidth}
      \centering
      \includegraphics[width=\linewidth]{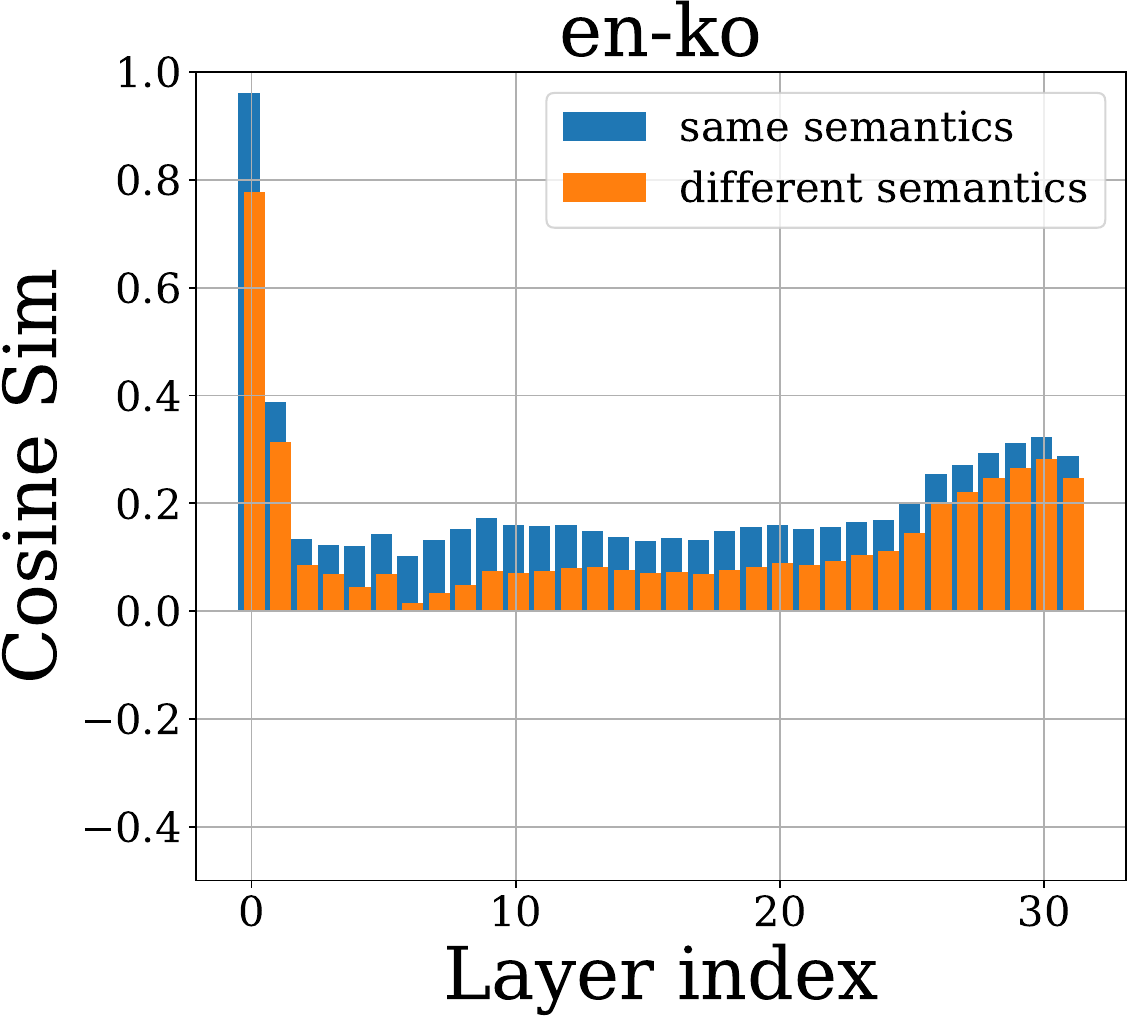}
      \subcaption{en-ko}
    \end{minipage}
    \begin{minipage}{0.20\linewidth}
      \centering
      \includegraphics[width=\linewidth]{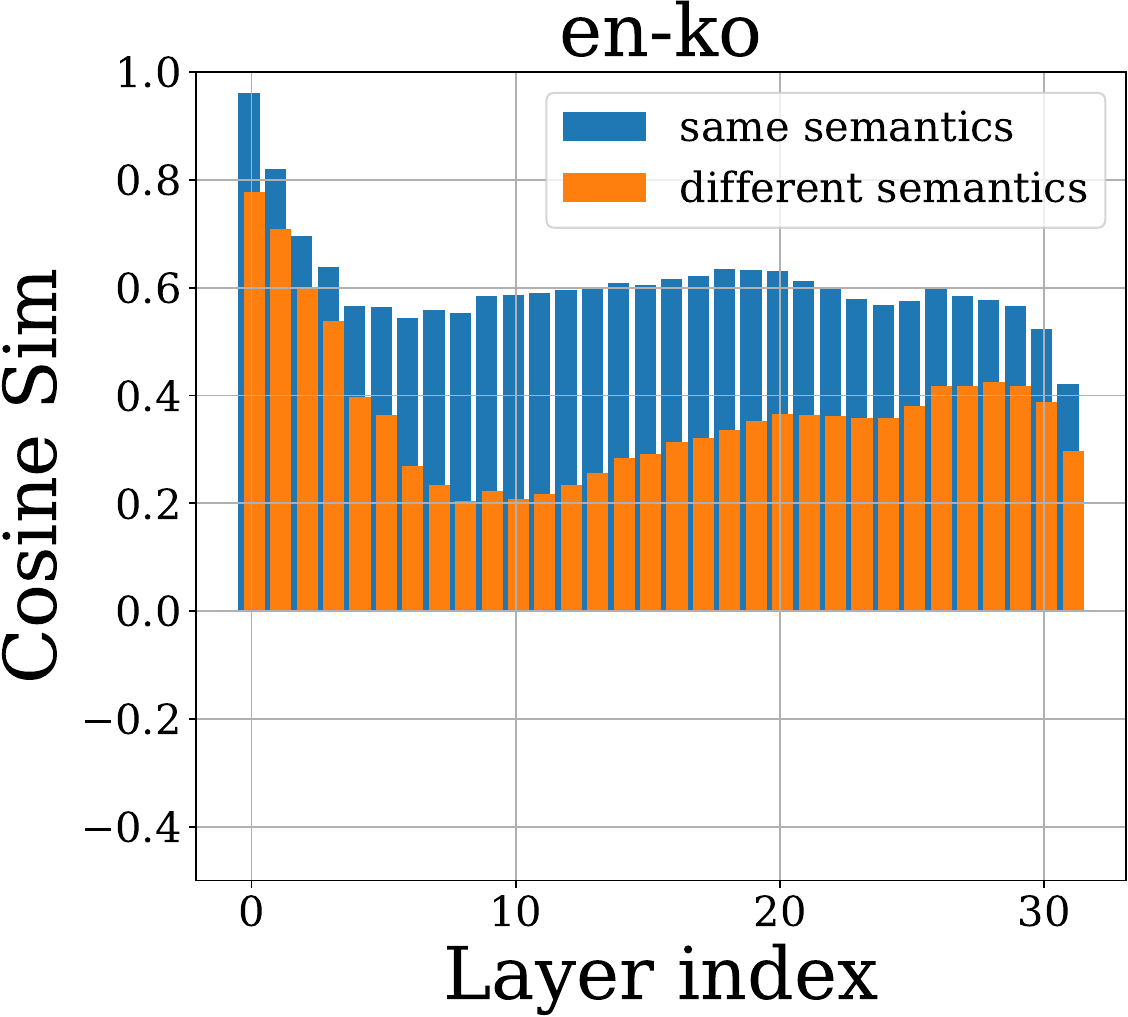}
      \subcaption{en-ko (baseline)}
    \end{minipage}
    \begin{minipage}{0.20\linewidth}
      \centering
      \includegraphics[width=\linewidth]{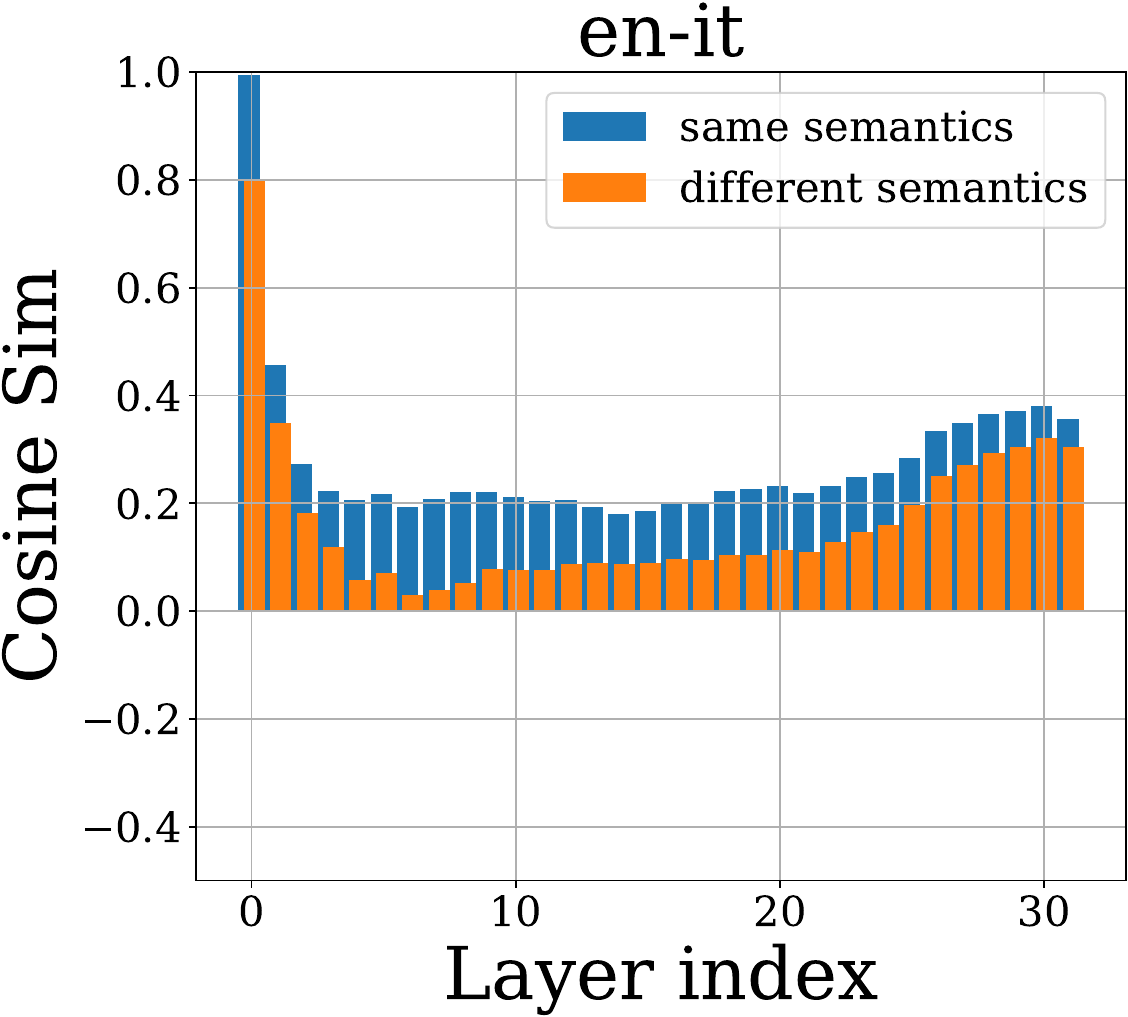}
      \subcaption{en-it}
    \end{minipage}
    \begin{minipage}{0.20\linewidth}
      \centering
      \includegraphics[width=\linewidth]{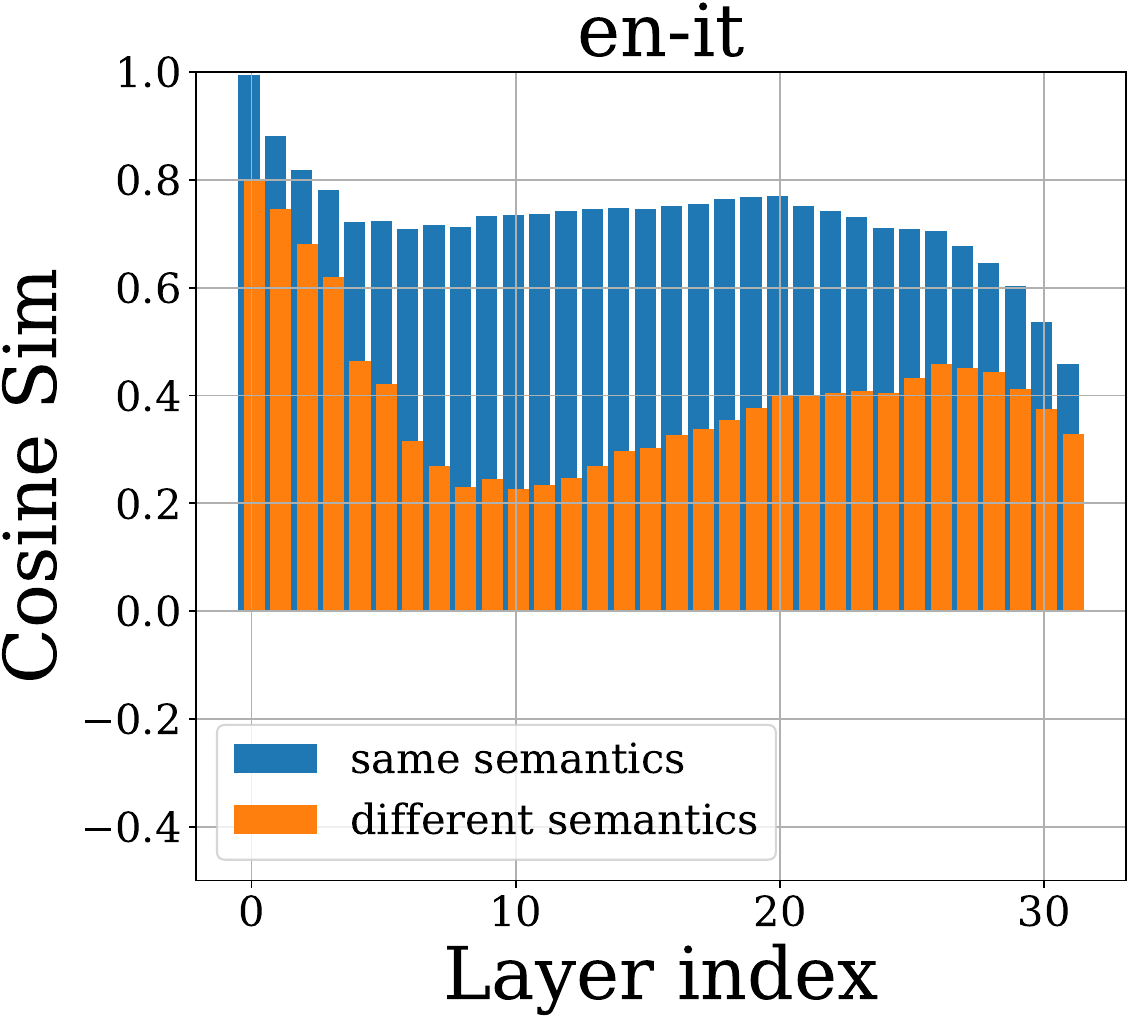}
      \subcaption{en-it (baseline)}
    \end{minipage}
    
      \begin{minipage}{\linewidth}
        \centering
        \small \textbf{(b) top-3000 (representing 0.6\% of all neurons)}
      \end{minipage}

    \begin{minipage}{0.20\linewidth}
      \centering
      \includegraphics[width=\linewidth]{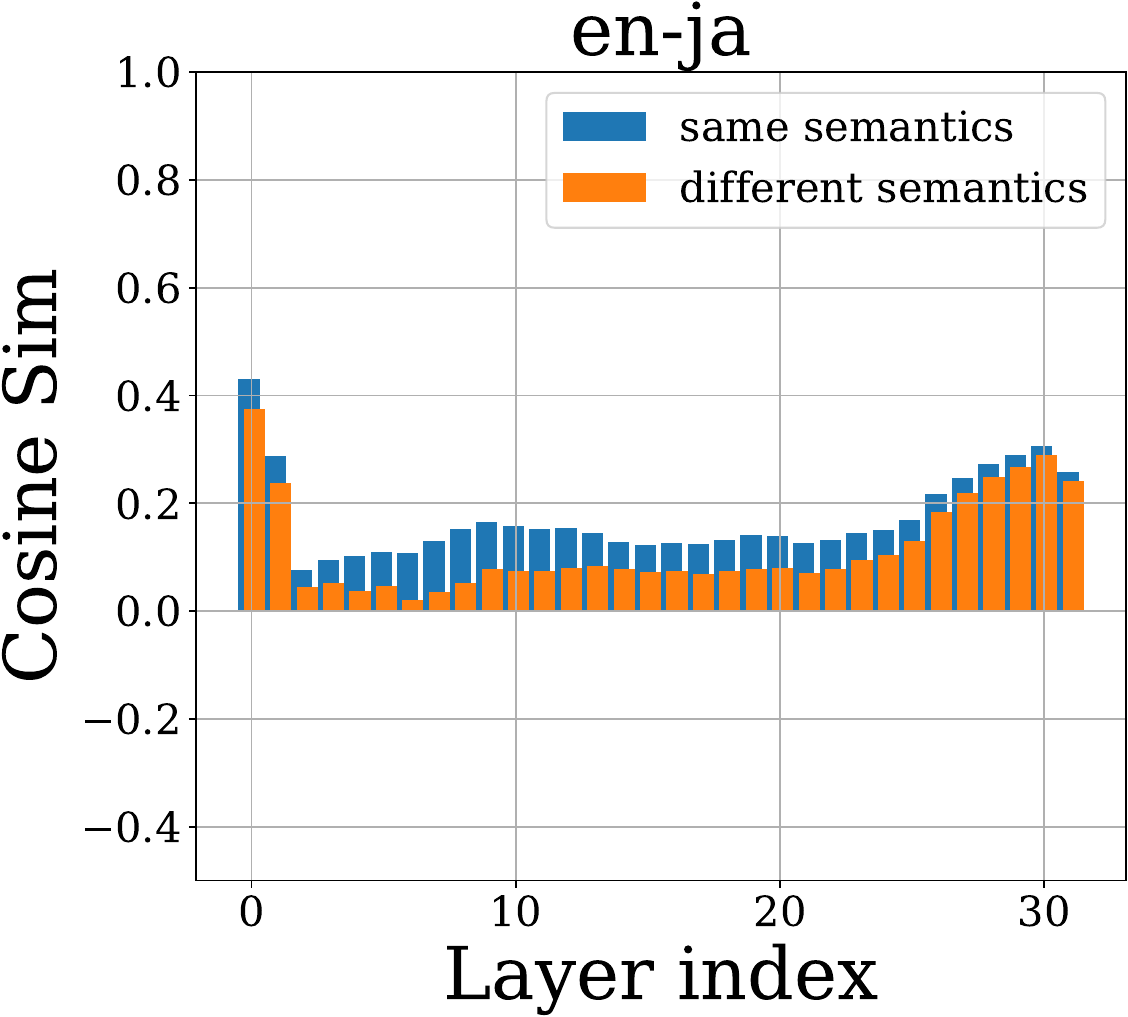}
      \subcaption{en-ja}
    \end{minipage}
    \begin{minipage}{0.20\linewidth}
      \centering
      \includegraphics[width=\linewidth]{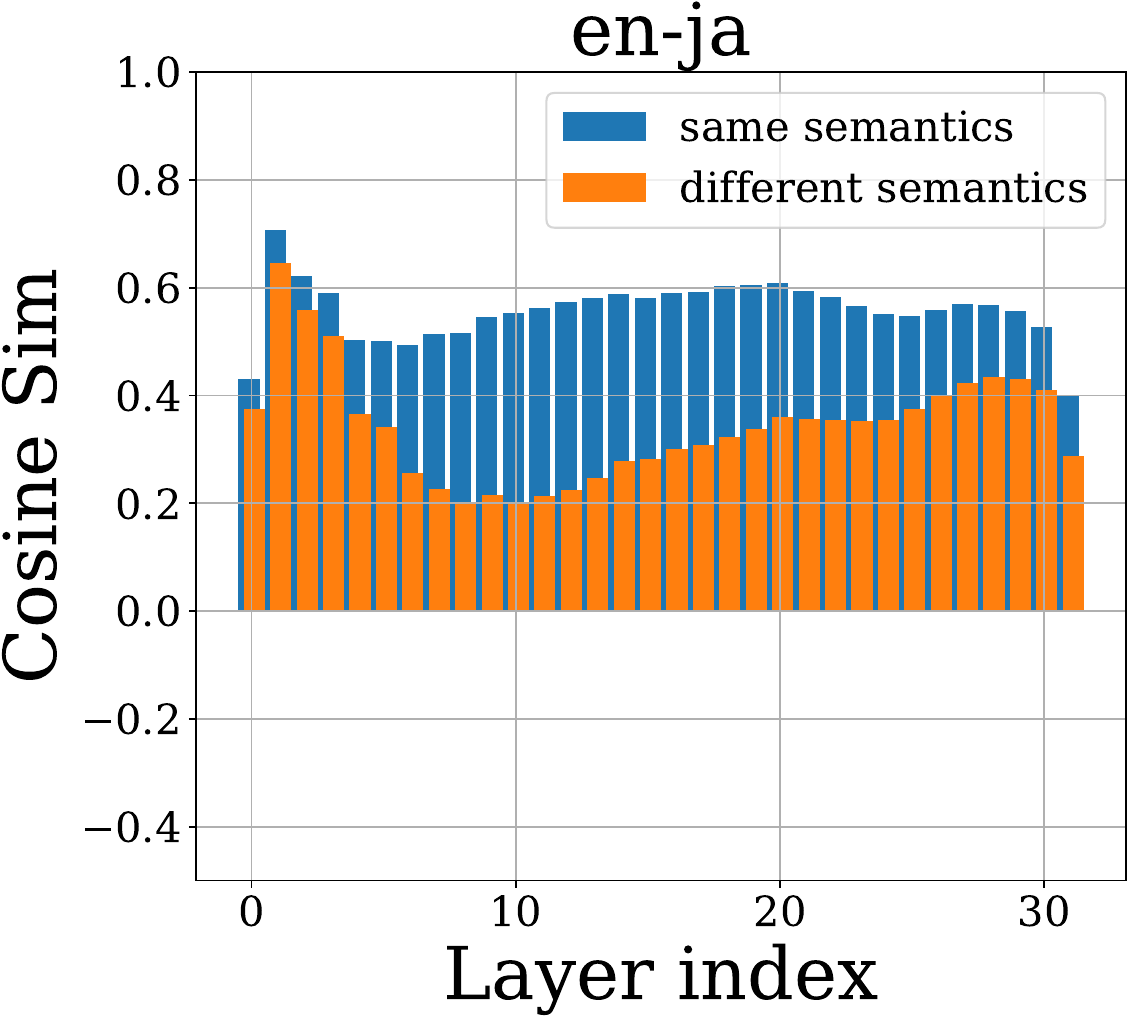}
      \subcaption{en-ja (baseline)}
    \end{minipage}
    \begin{minipage}{0.20\linewidth}
      \centering
      \includegraphics[width=\linewidth]{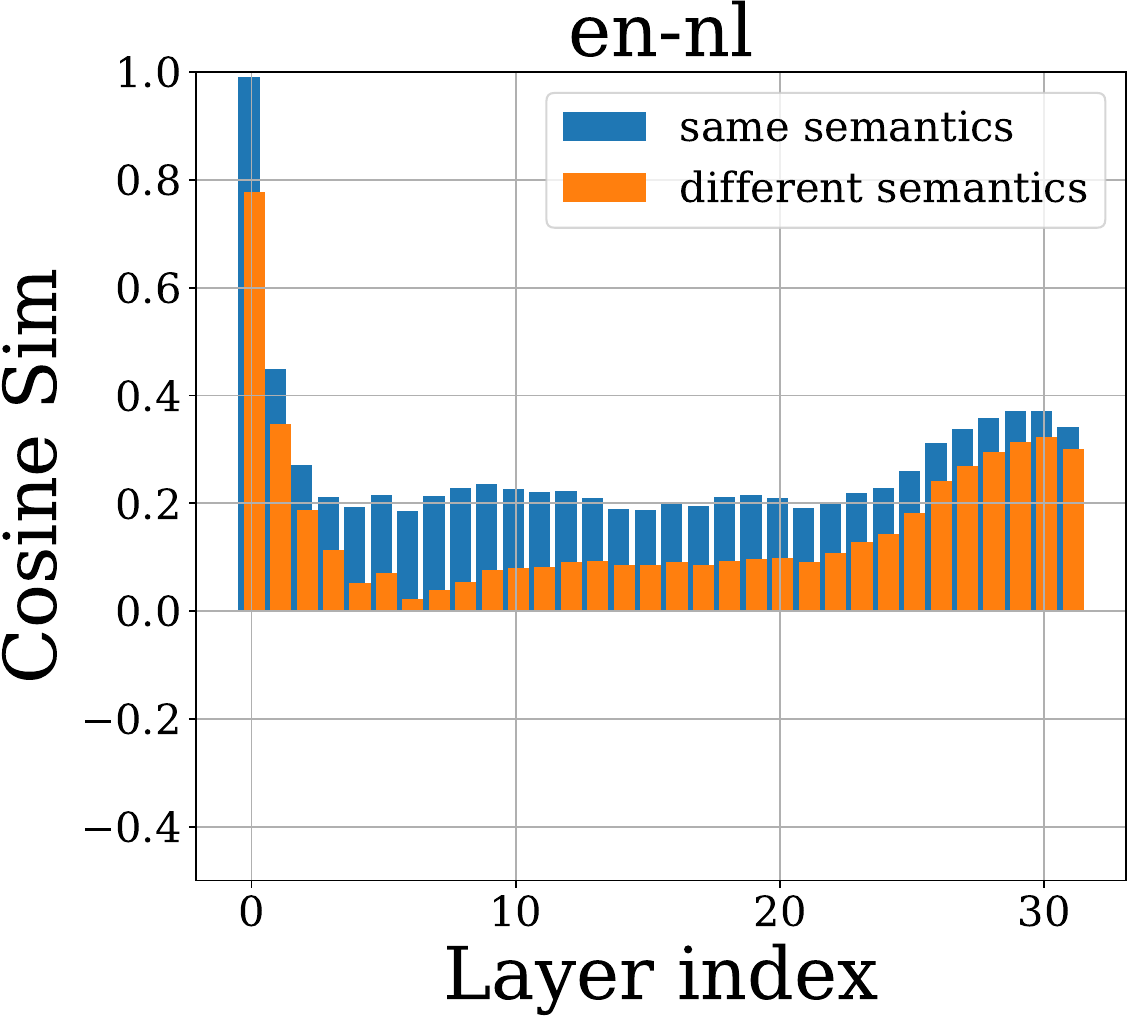}
      \subcaption{en-nl}
    \end{minipage}
    \begin{minipage}{0.20\linewidth}
      \centering
      \includegraphics[width=\linewidth]{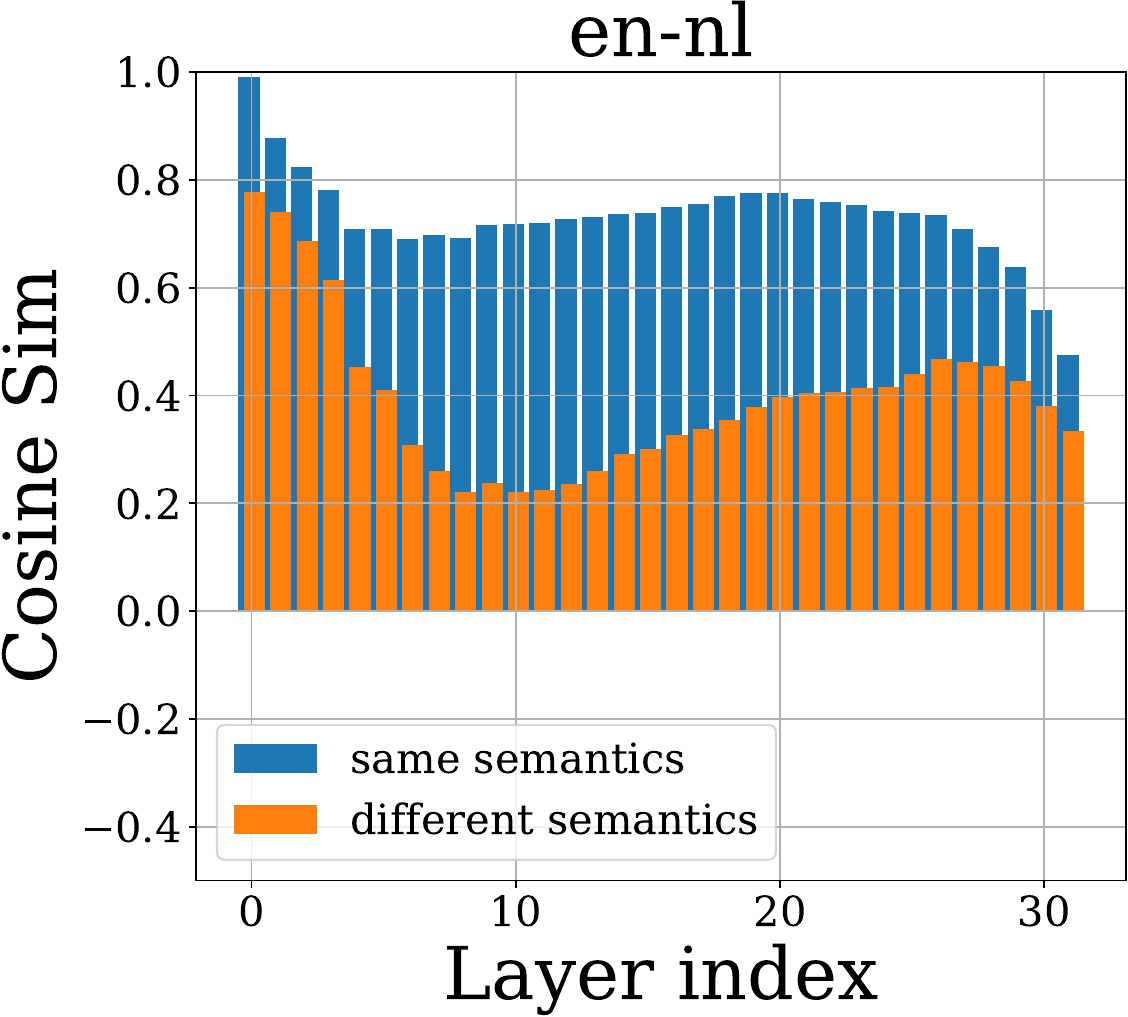}
      \subcaption{en-nl (baseline)}
    \end{minipage}

    \begin{minipage}{0.20\linewidth}
      \centering
      \includegraphics[width=\linewidth]{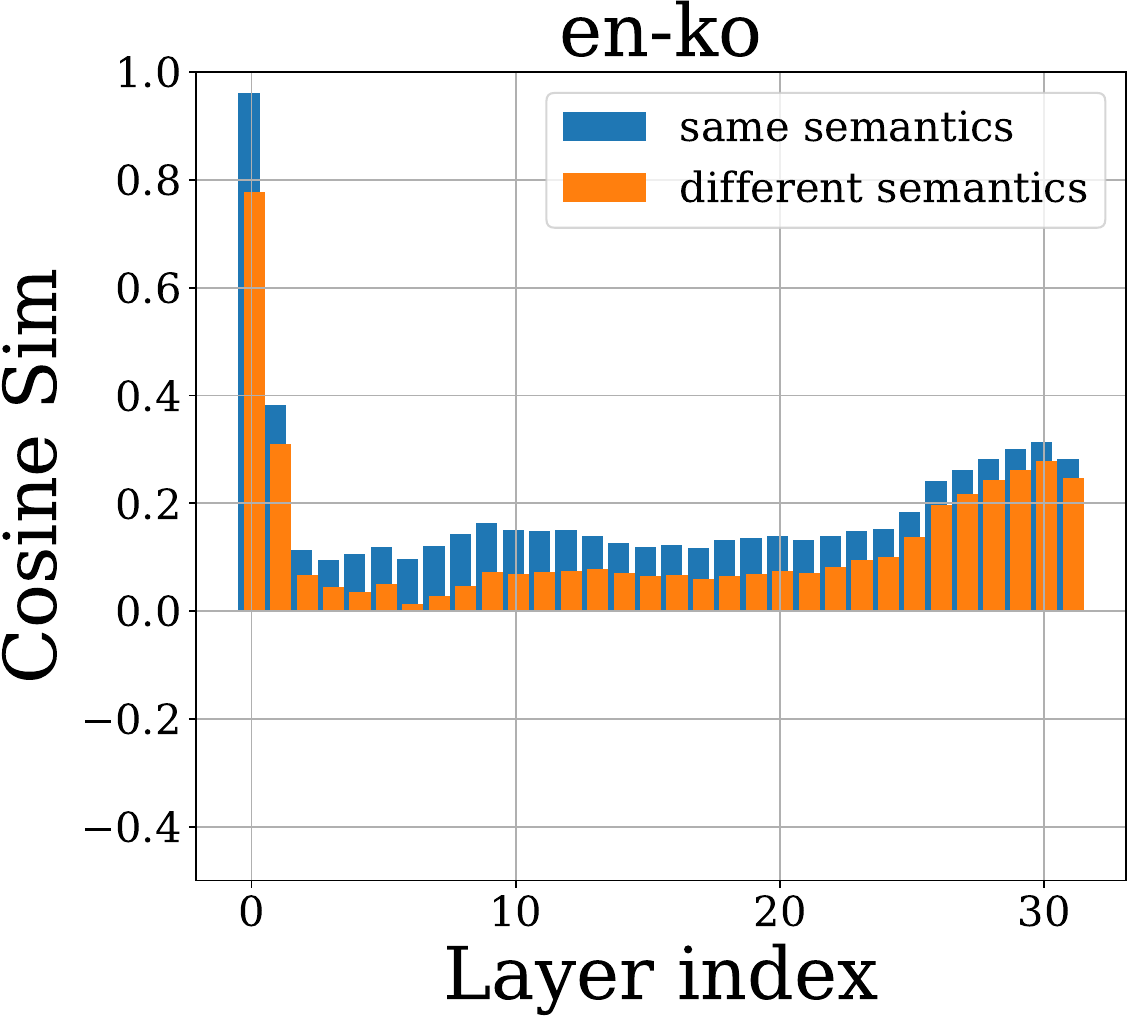}
      \subcaption{en-ko}
    \end{minipage}
    \begin{minipage}{0.20\linewidth}
      \centering
      \includegraphics[width=\linewidth]{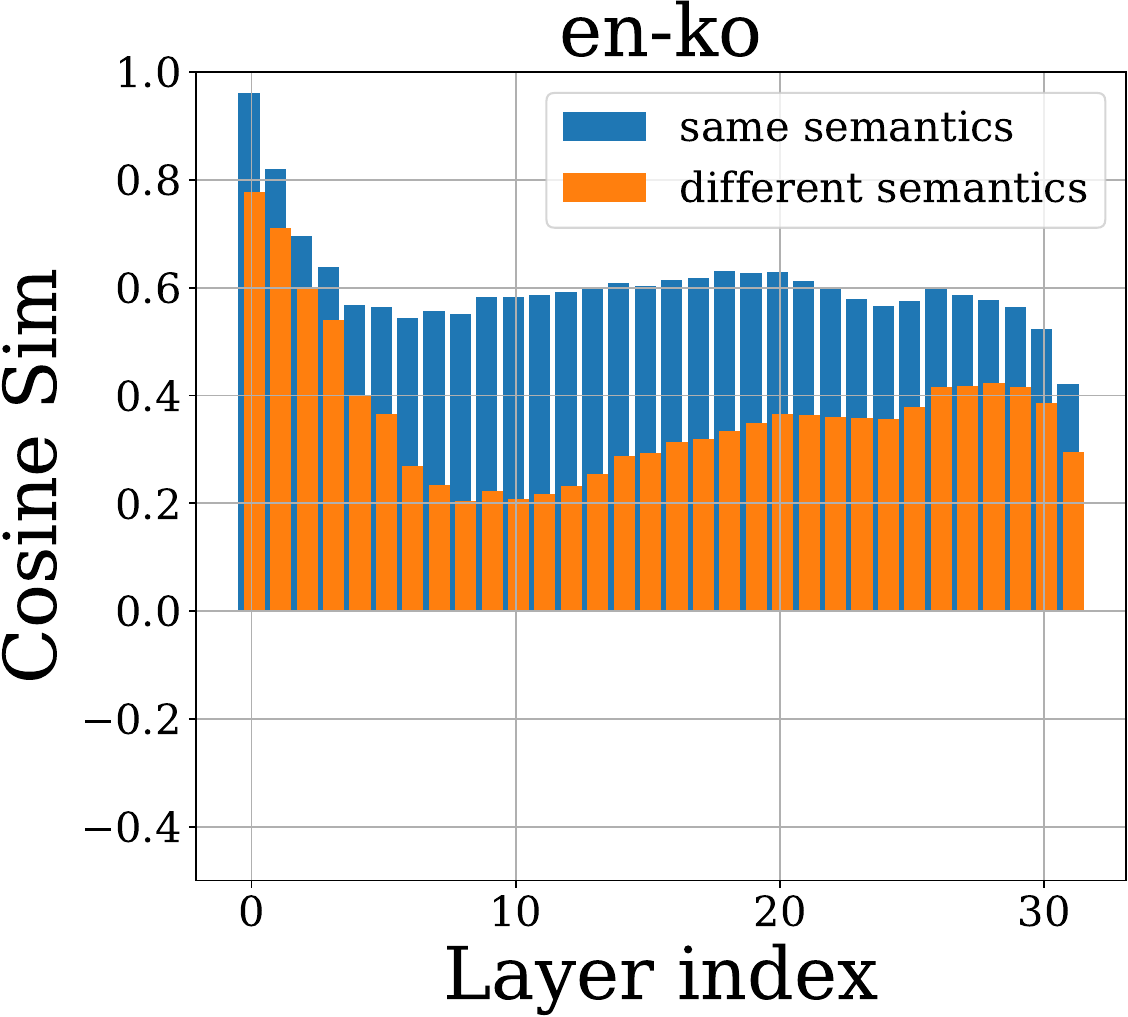}
      \subcaption{en-ko (baseline)}
    \end{minipage}
    \begin{minipage}{0.20\linewidth}
      \centering
      \includegraphics[width=\linewidth]{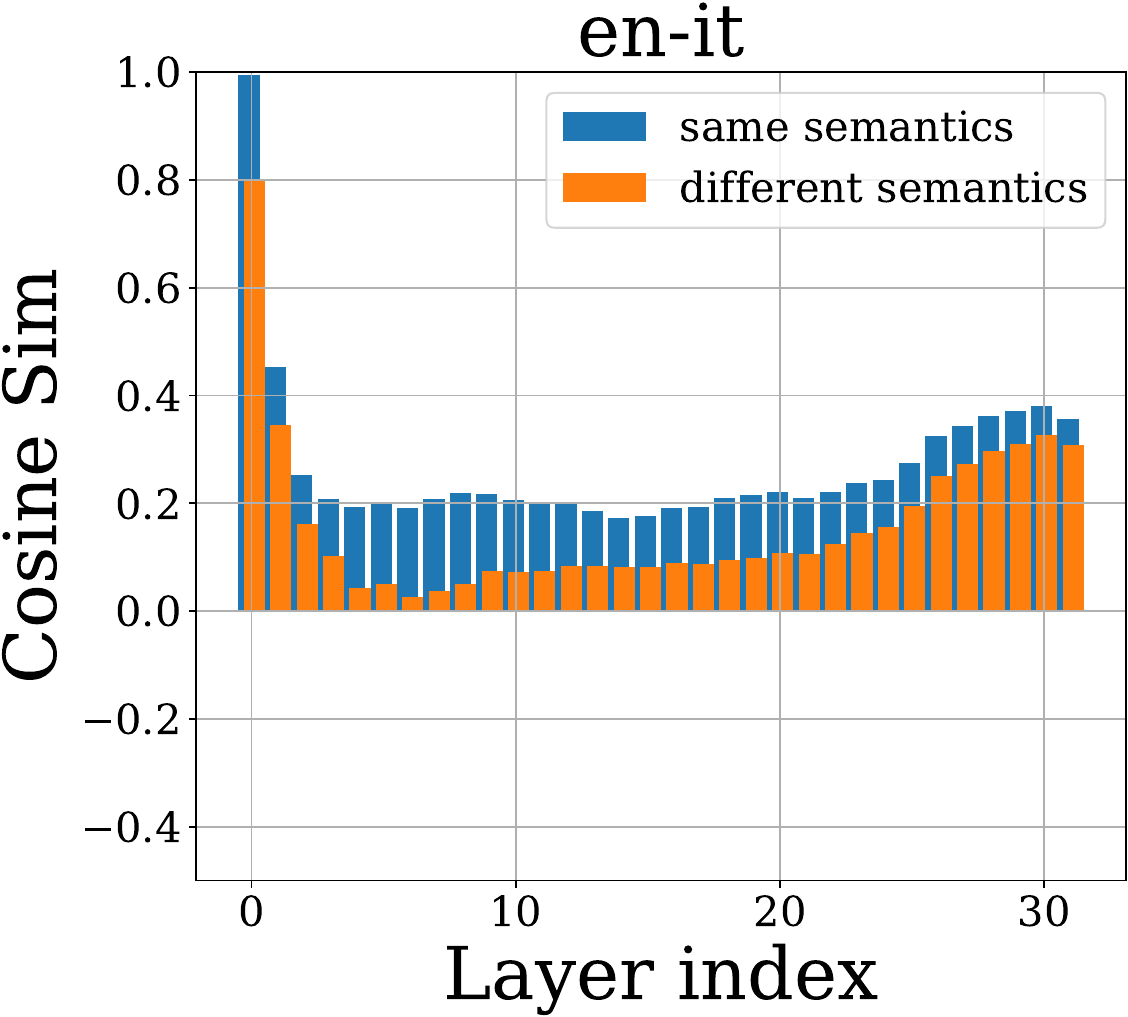}
      \subcaption{en-it}
    \end{minipage}
    \begin{minipage}{0.20\linewidth}
      \centering
      \includegraphics[width=\linewidth]{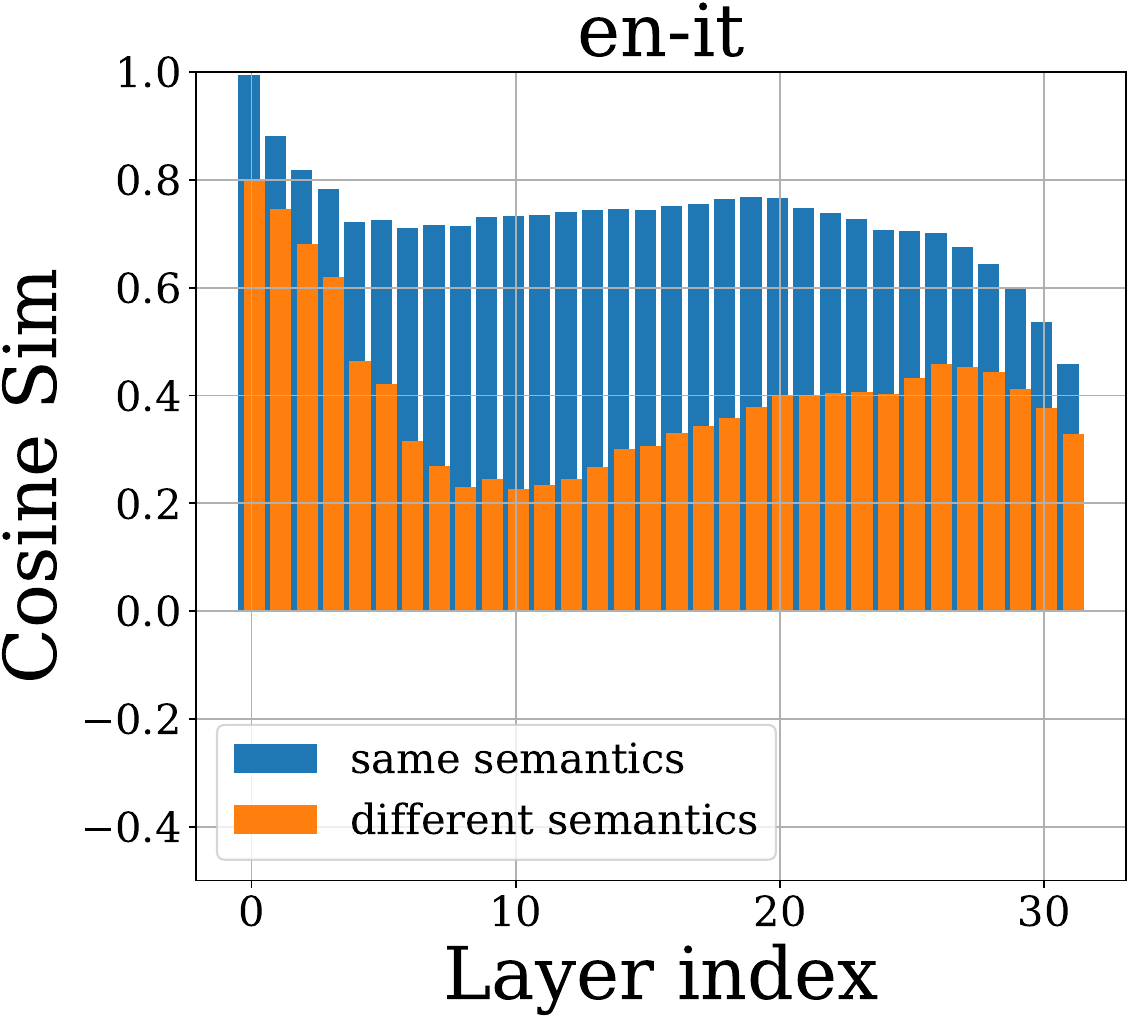}
      
      \subcaption{en-it (baseline)}
    \end{minipage}
    
      \begin{minipage}{\linewidth}
        \centering
        \small \textbf{(b) top-5000 (representing 1\% of all neurons)}
      \end{minipage}

  \caption{\textbf{Similarity of hidden states across layers while deactivating Type-1 Transfer Neurons (Aya expanse-8B).}}
  \label{fig:appendix:hs_sim_aya_deactivating_top-1k_Type-1}
\end{figure*}

\begin{figure*}[t]
    \centering

    \begin{minipage}{0.23\linewidth}
      \centering
      \includegraphics[width=\linewidth]{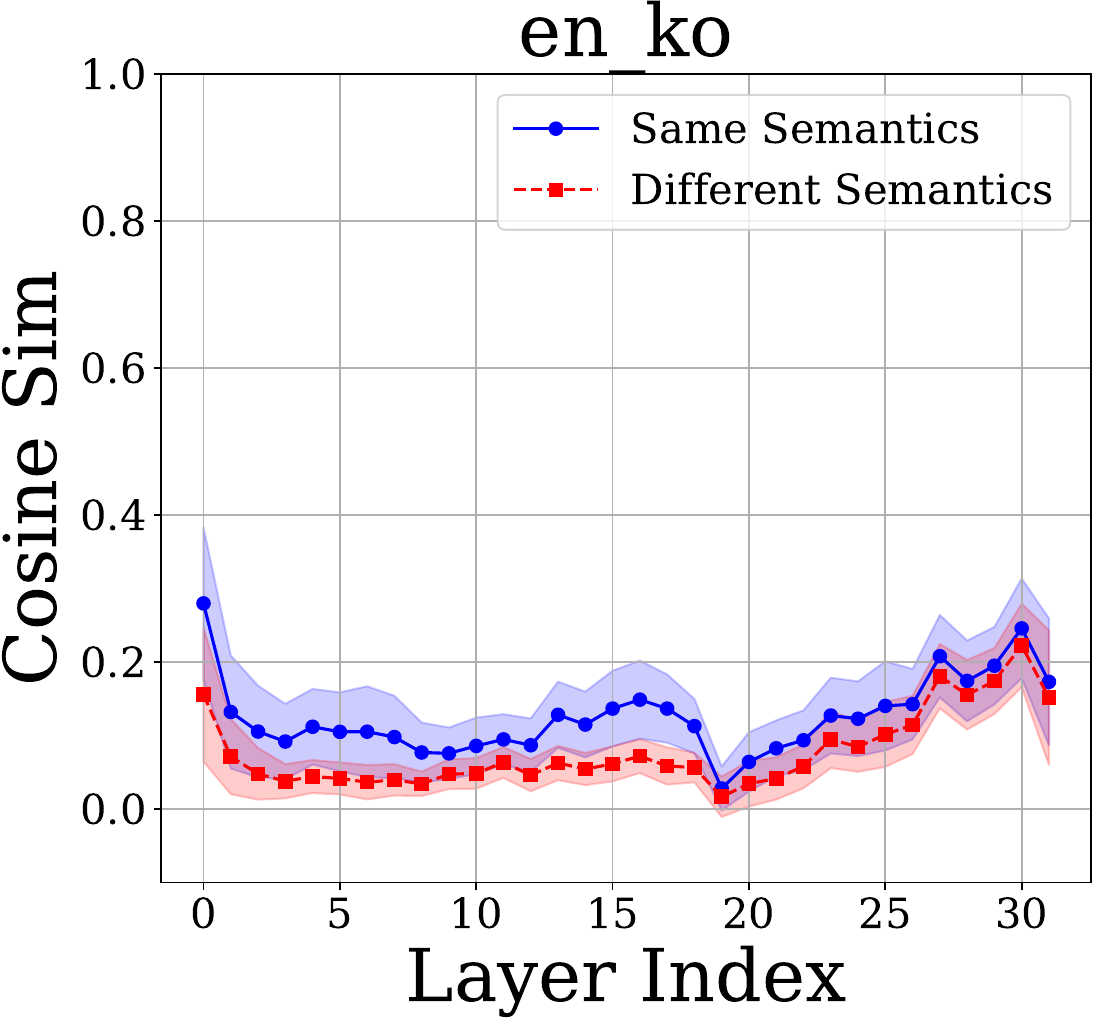}
      \subcaption{en-ko}
    \end{minipage}
    \begin{minipage}{0.23\linewidth}
      \centering
      \includegraphics[width=\linewidth]{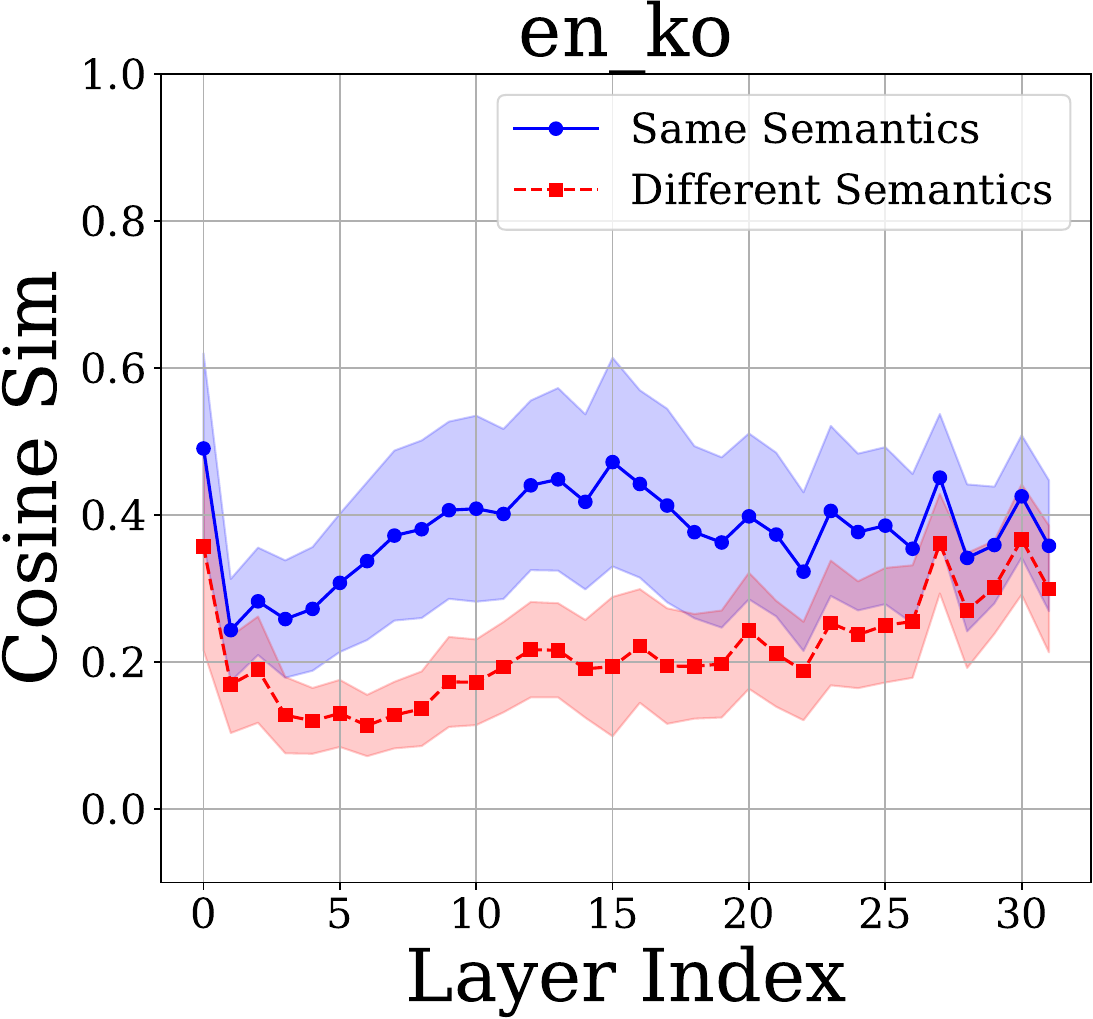}
      \subcaption{en-ko (baseline)}
    \end{minipage}
    \begin{minipage}{0.23\linewidth}
      \centering
      \includegraphics[width=\linewidth]{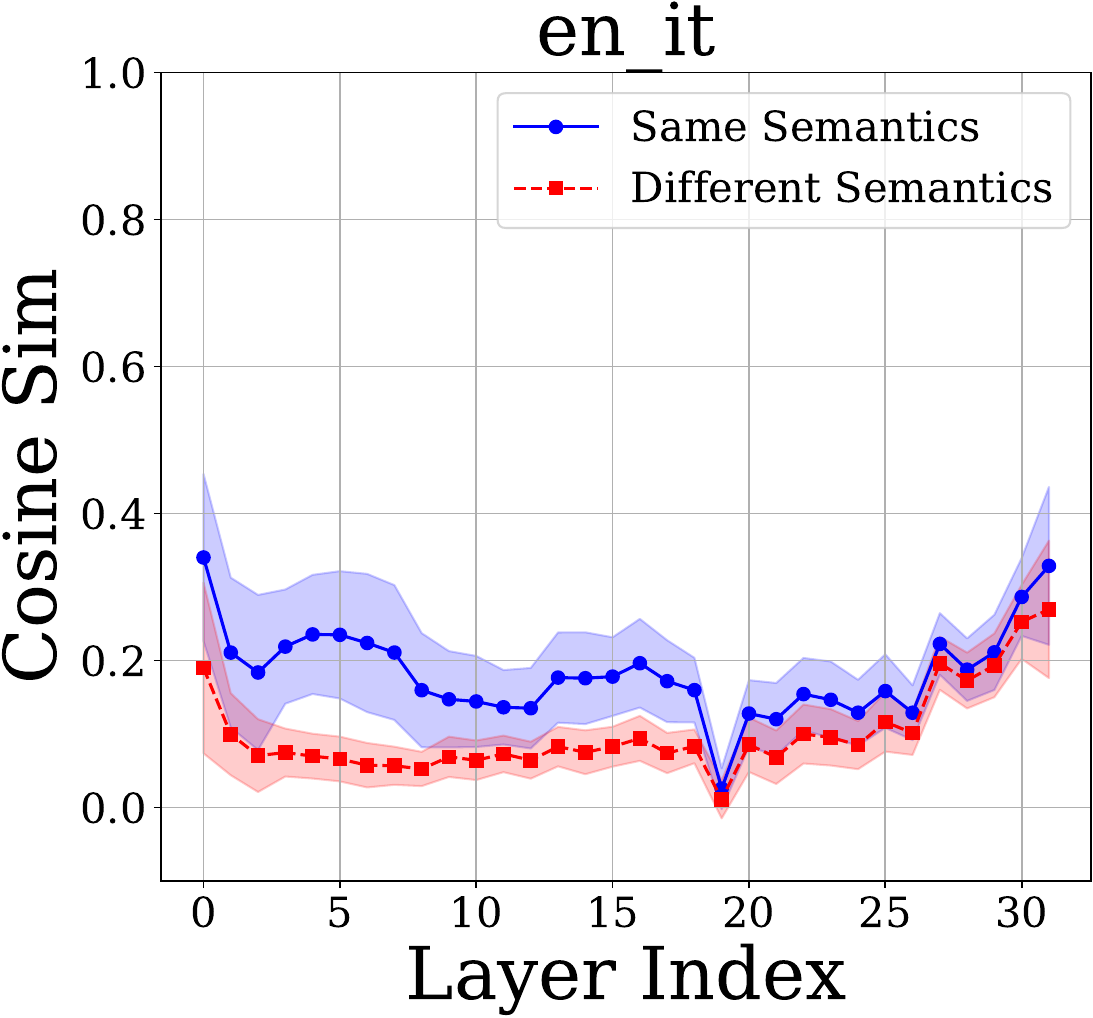}
      \subcaption{en-it}
    \end{minipage}
    \begin{minipage}{0.23\linewidth}
      \centering
      \includegraphics[width=\linewidth]{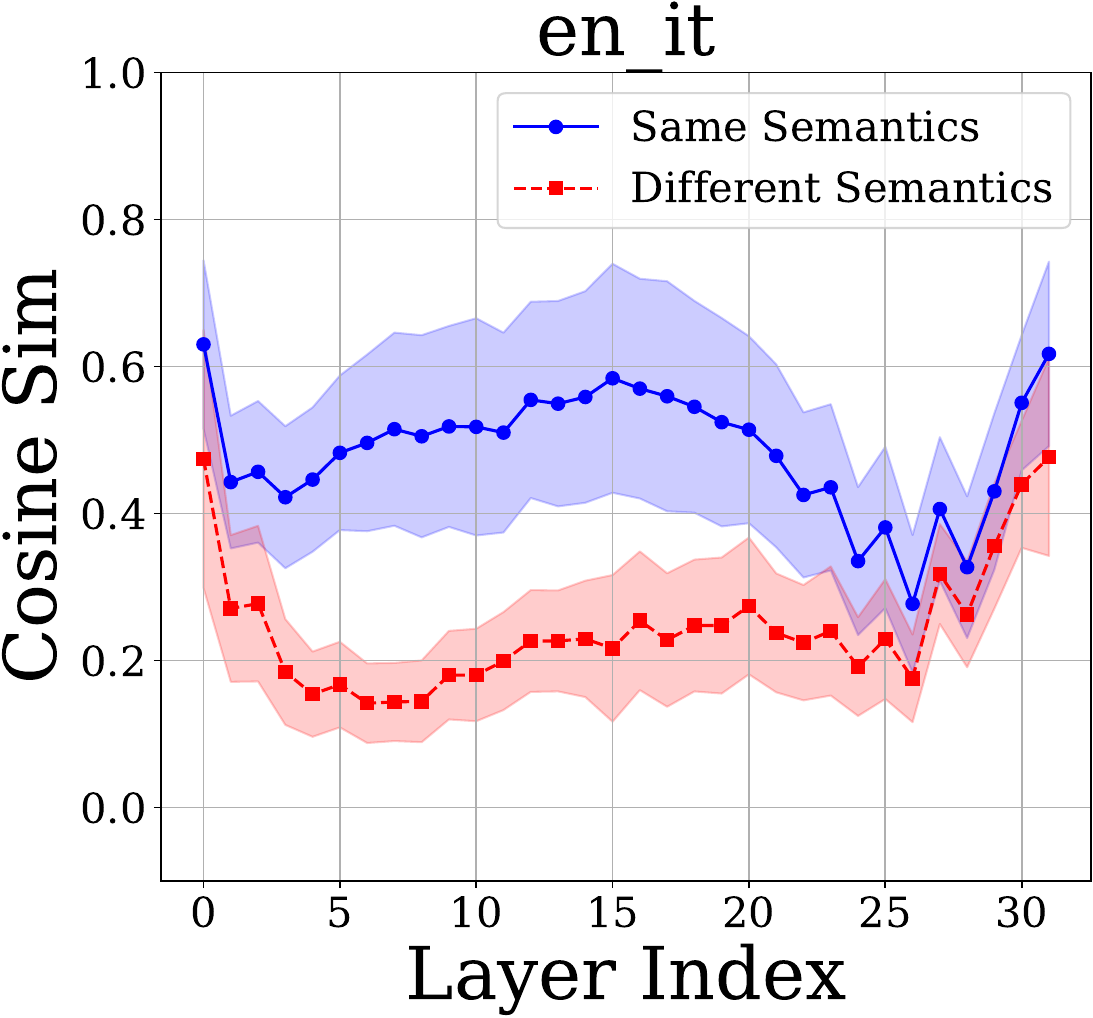}
      \subcaption{en-it (baseline)}
    \end{minipage}
    
      % First row label
      \begin{minipage}{\linewidth}
        \centering
        \small \textbf{(a) top-1000  (representing 0.2\% of all neurons)}
      \end{minipage}

    % second row
    \begin{minipage}{0.23\linewidth}
      \centering
      \includegraphics[width=\linewidth]{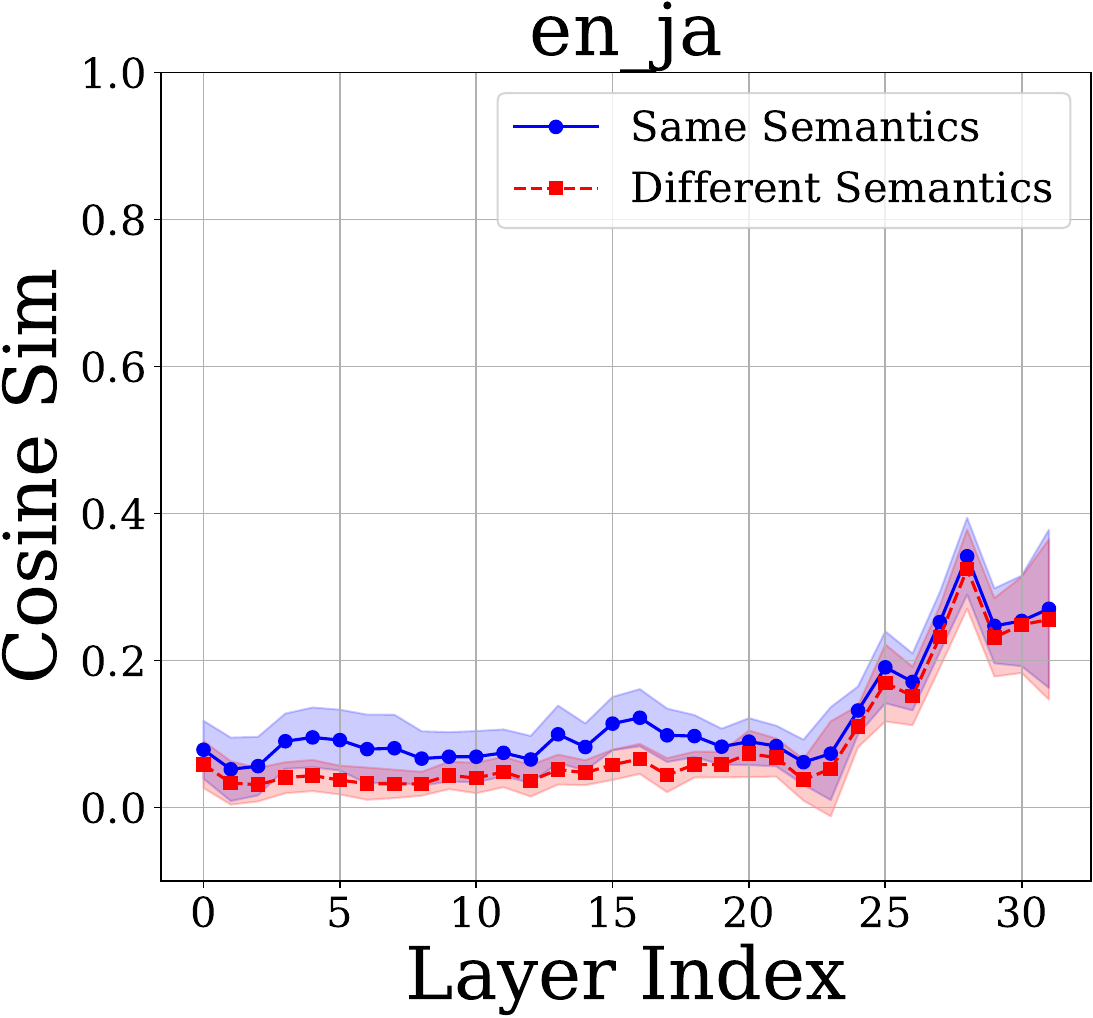}
      \subcaption{en-ja}
    \end{minipage}
    \begin{minipage}{0.23\linewidth}
      \centering
      \includegraphics[width=\linewidth]{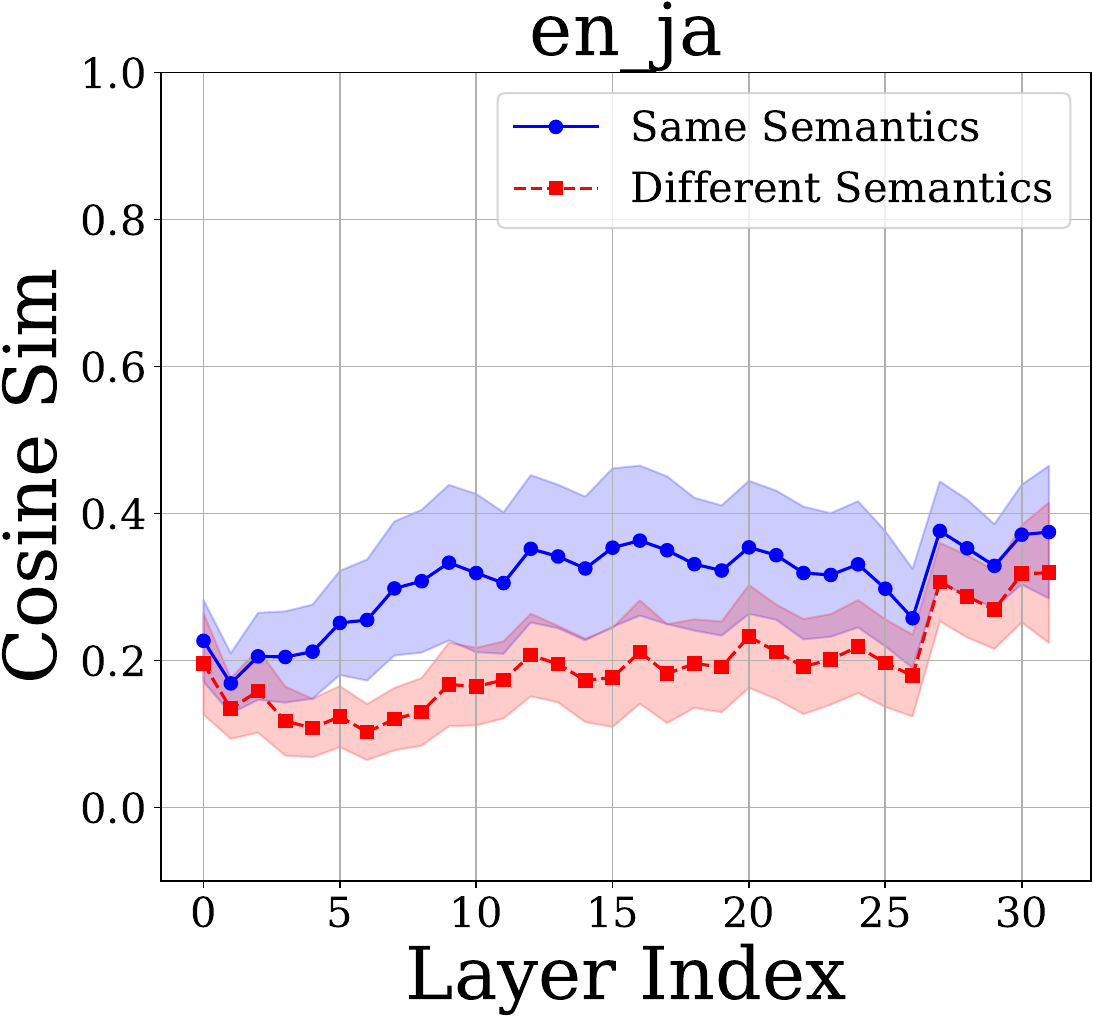}
      \subcaption{en-ja (baseline)}
    \end{minipage}
    \begin{minipage}{0.23\linewidth}
      \centering
      \includegraphics[width=\linewidth]{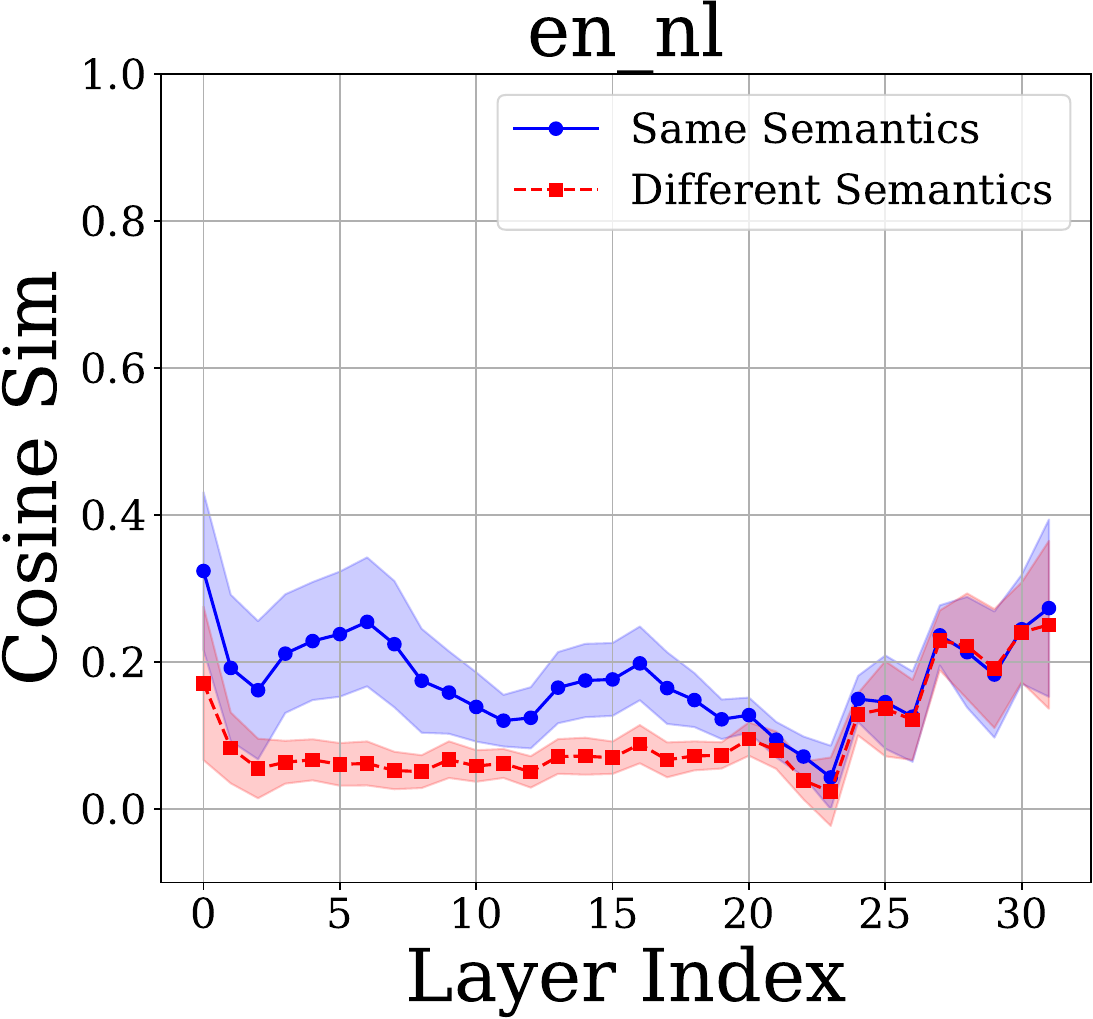}
      \subcaption{en-nl}
    \end{minipage}
    \begin{minipage}{0.23\linewidth}
      \centering
      \includegraphics[width=\linewidth]{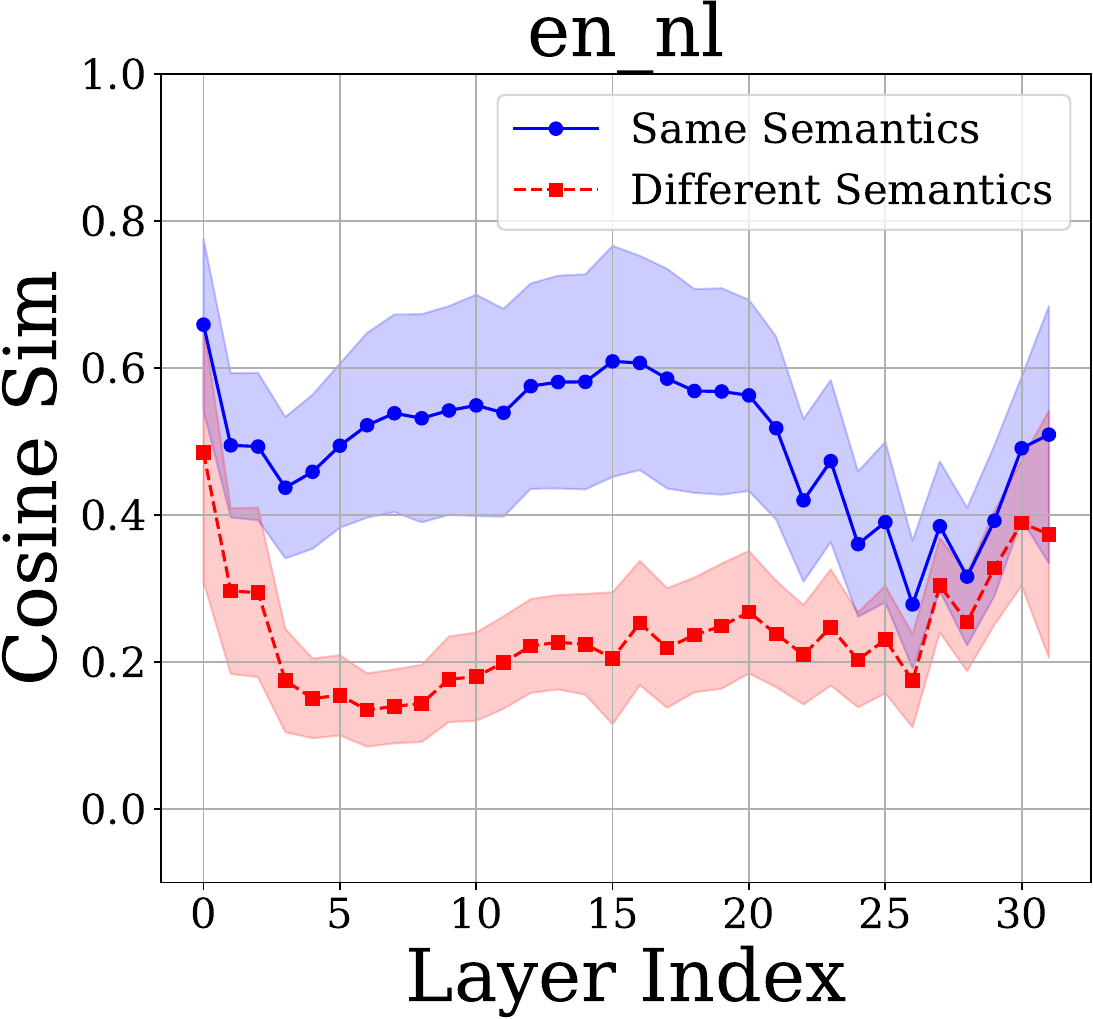}
      \subcaption{en-nl (baseline)}
    \end{minipage}

    \begin{minipage}{0.23\linewidth}
      \centering
      \includegraphics[width=\linewidth]{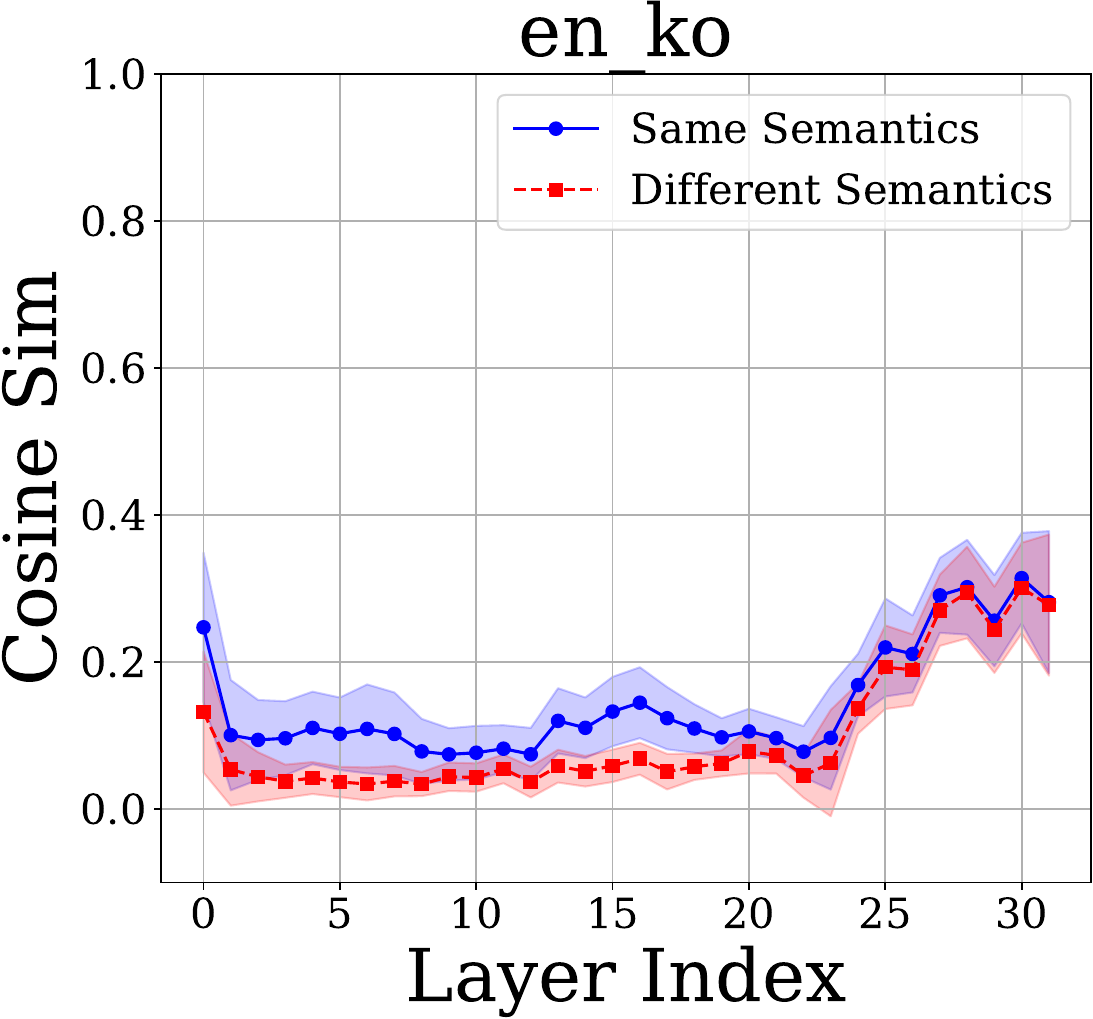}
      \subcaption{en-ko}
    \end{minipage}
    \begin{minipage}{0.23\linewidth}
      \centering
      \includegraphics[width=\linewidth]{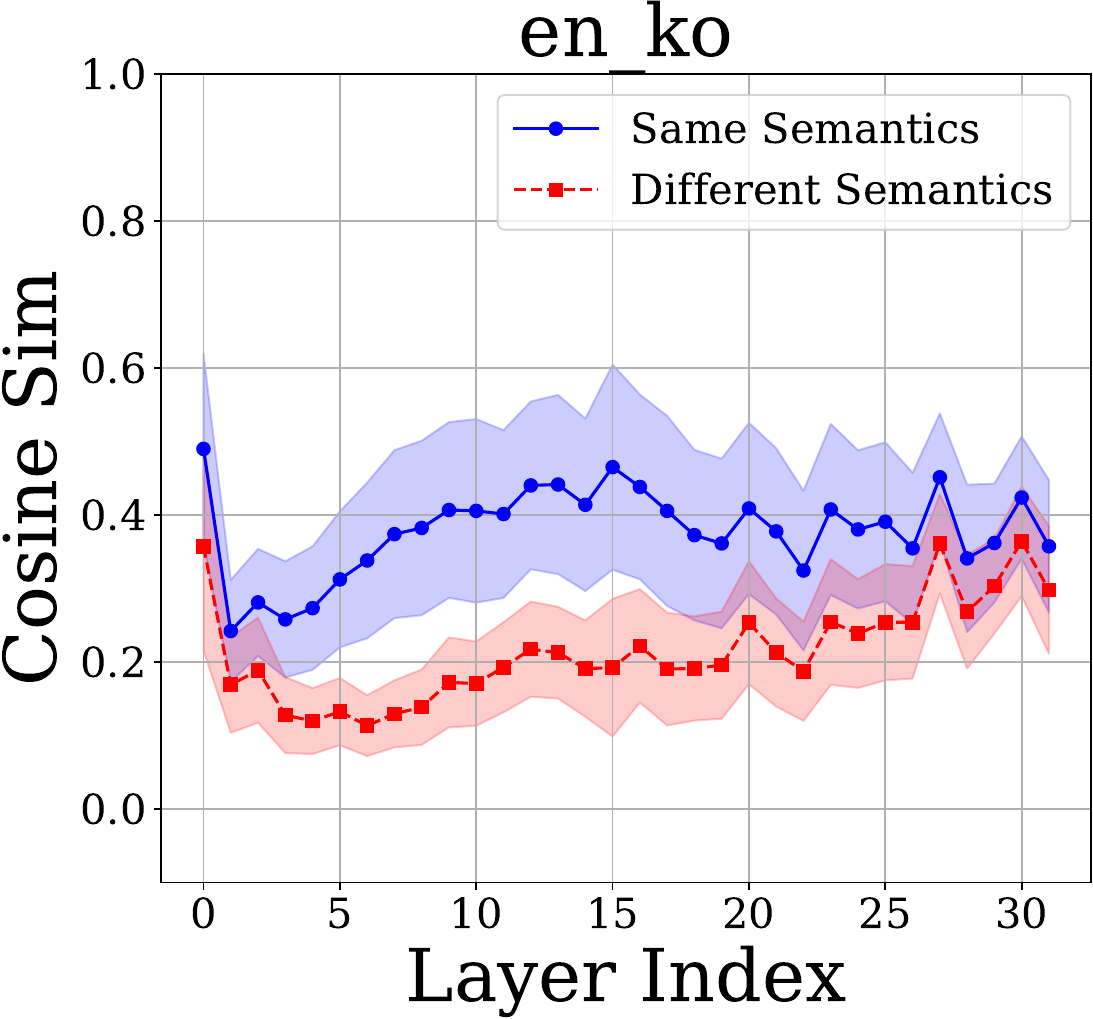}
      \subcaption{en-ko (baseline)}
    \end{minipage}
    \begin{minipage}{0.23\linewidth}
      \centering
      \includegraphics[width=\linewidth]{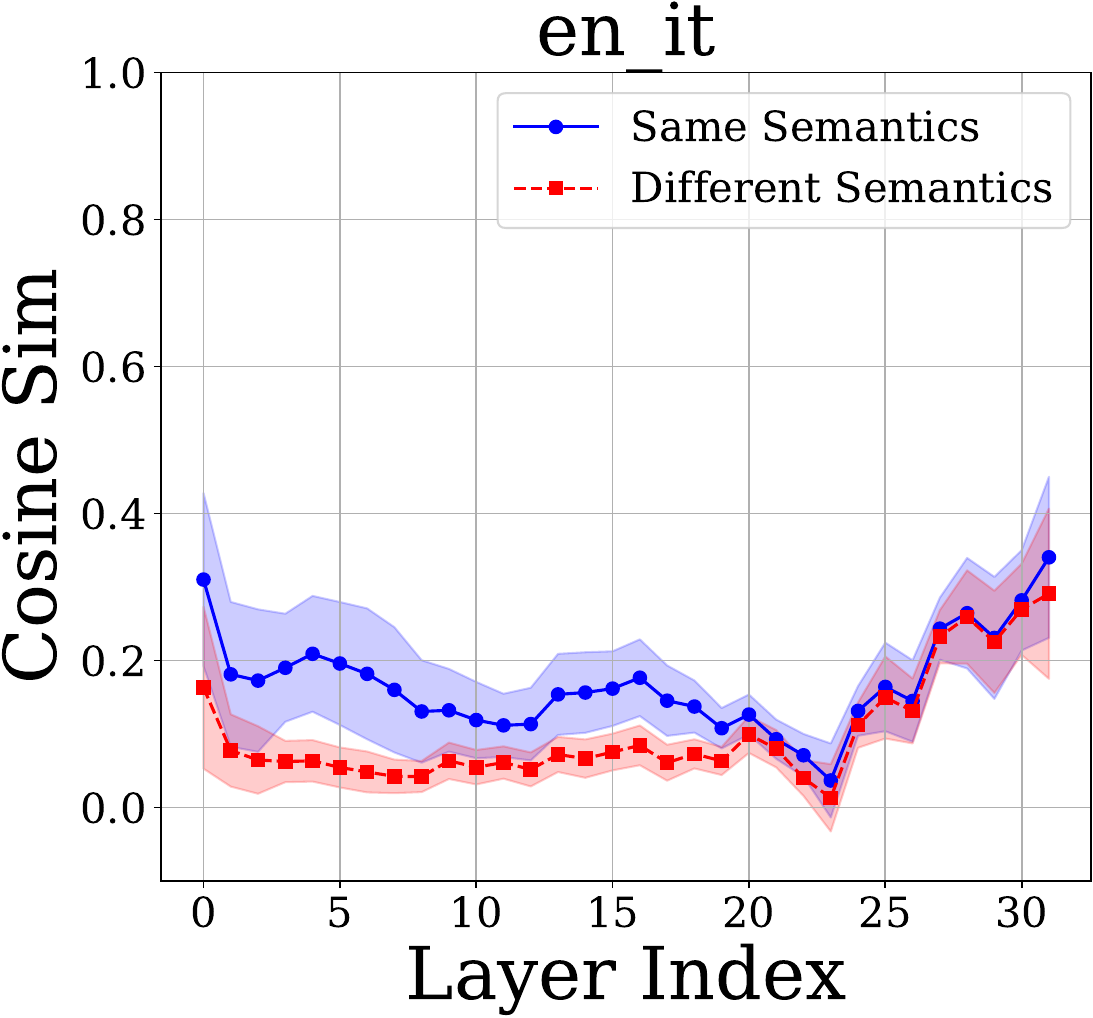}
      \subcaption{en-it}
    \end{minipage}
    \begin{minipage}{0.23\linewidth}
      \centering
      \includegraphics[width=\linewidth]{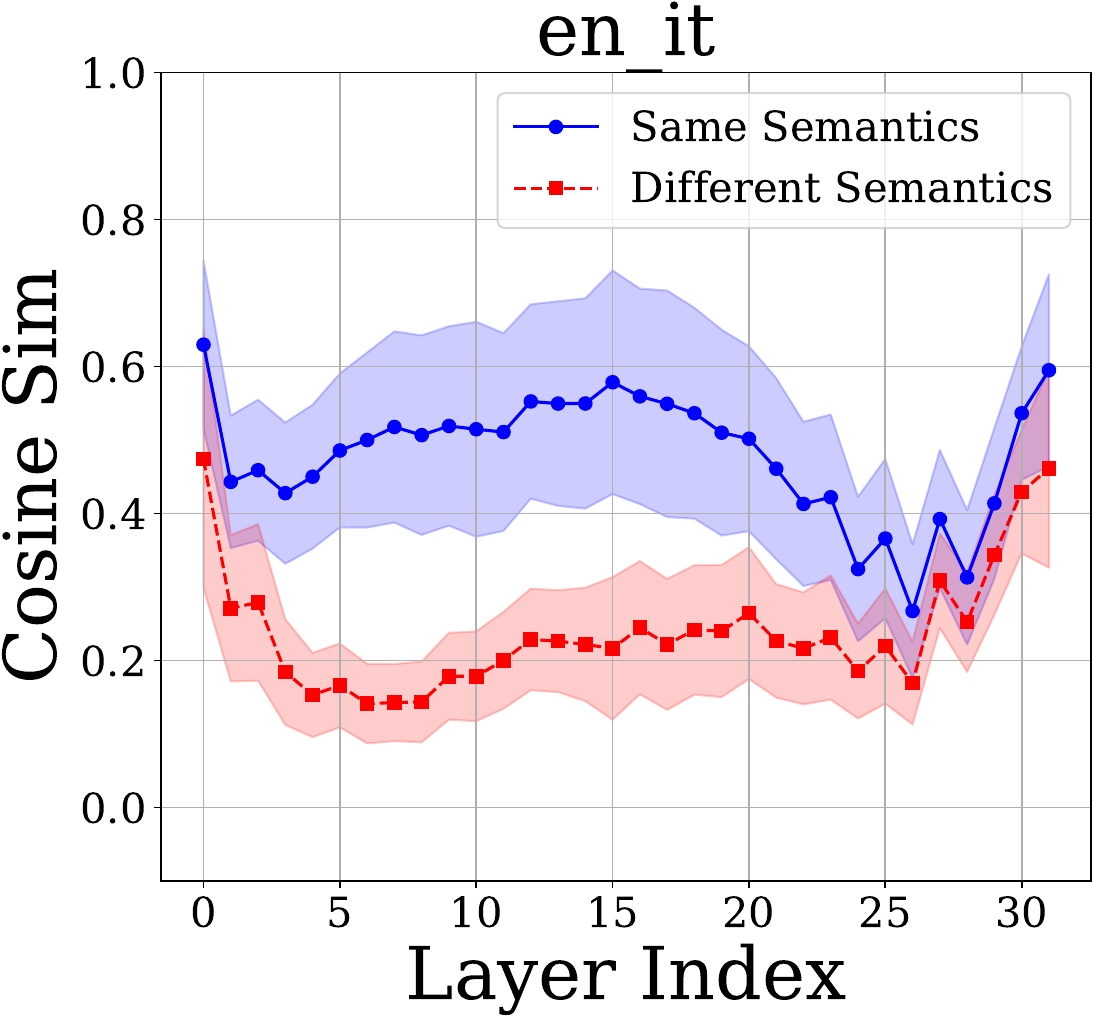}
      \subcaption{en-it (baseline)}
    \end{minipage}
    
      \begin{minipage}{\linewidth}
        \centering
        \small \textbf{(b) top-3000 (representing 0.6\% of all neurons)}
      \end{minipage}

    \begin{minipage}{0.23\linewidth}
      \centering
      \includegraphics[width=\linewidth]{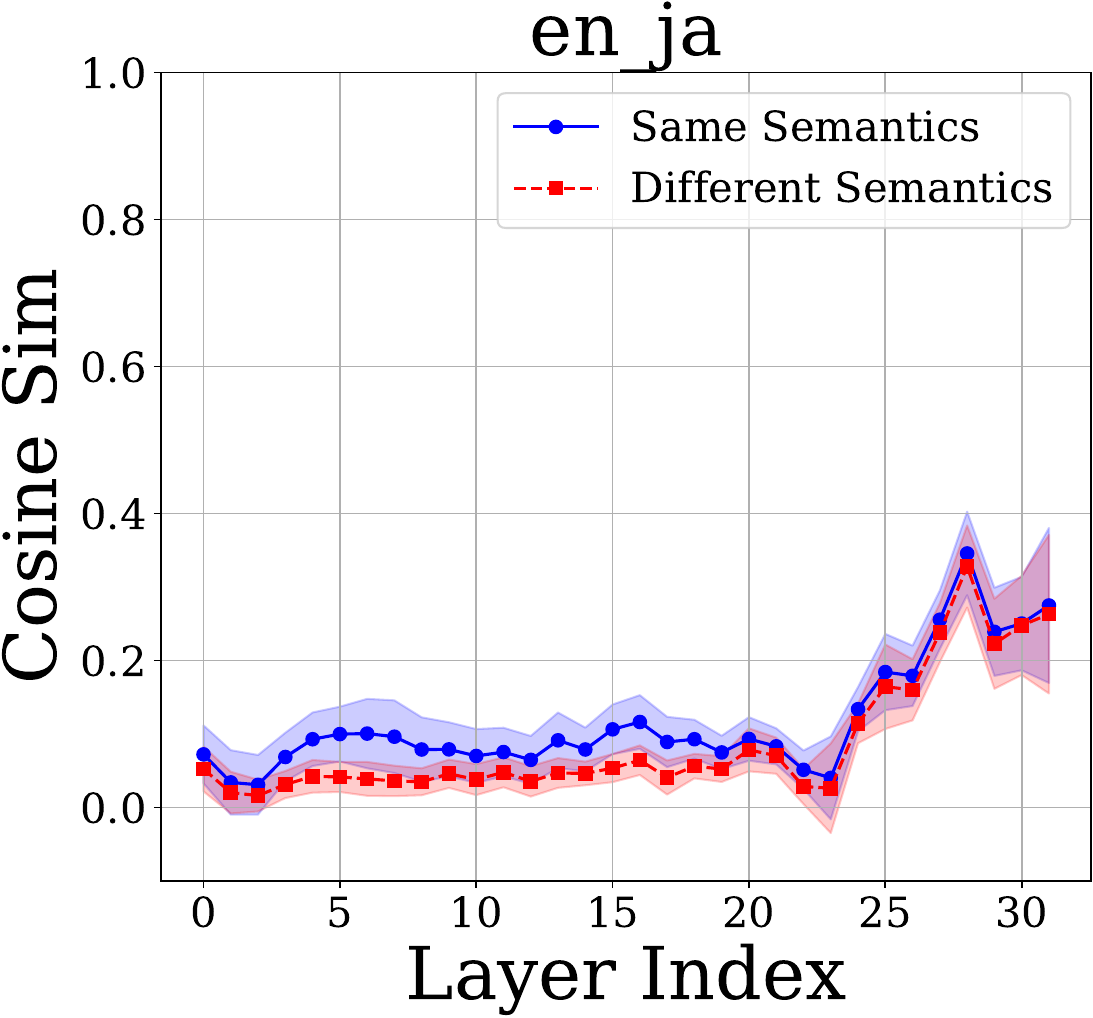}
      \subcaption{en-ja}
    \end{minipage}
    \begin{minipage}{0.23\linewidth}
      \centering
      \includegraphics[width=\linewidth]{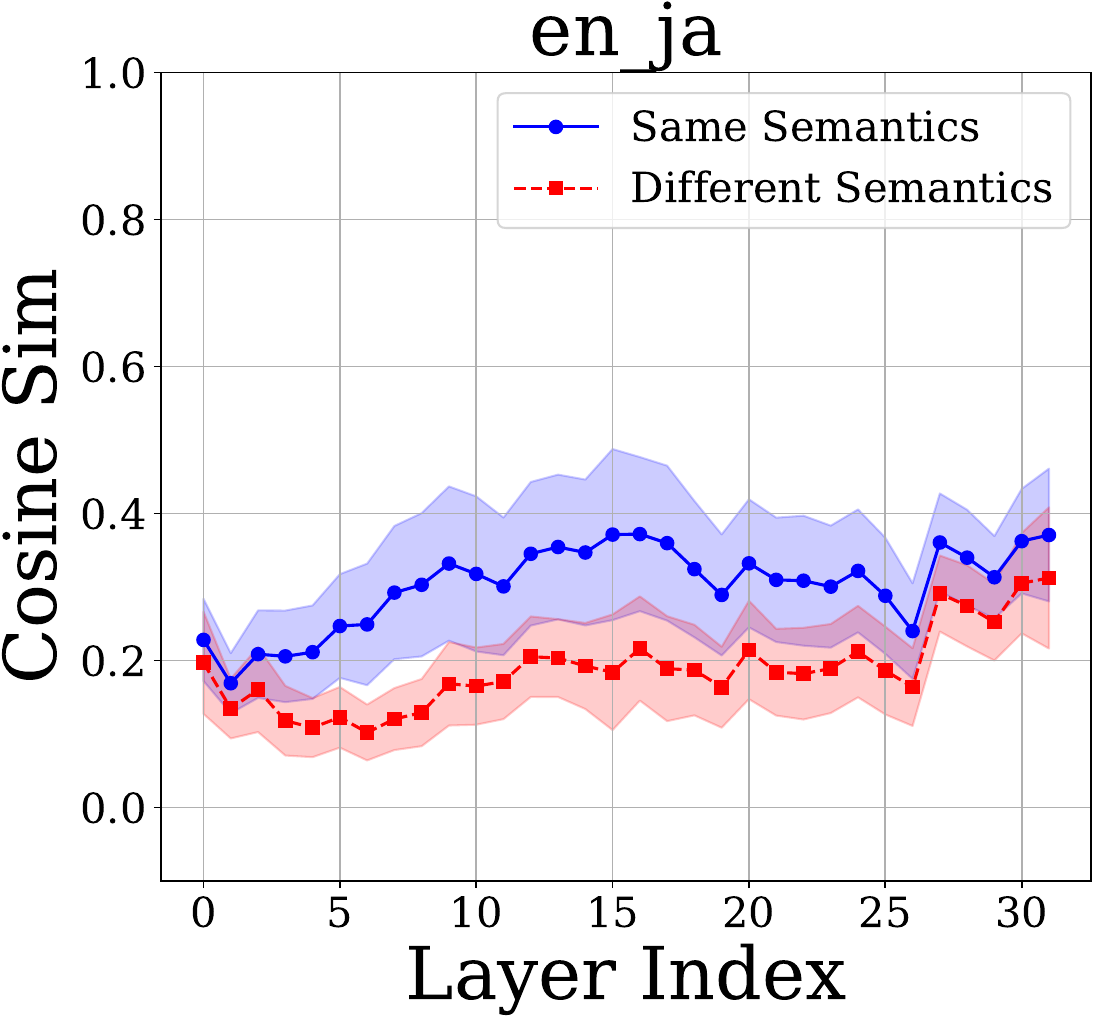}
      \subcaption{en-ja (baseline)}
    \end{minipage}
    \begin{minipage}{0.23\linewidth}
      \centering
      \includegraphics[width=\linewidth]{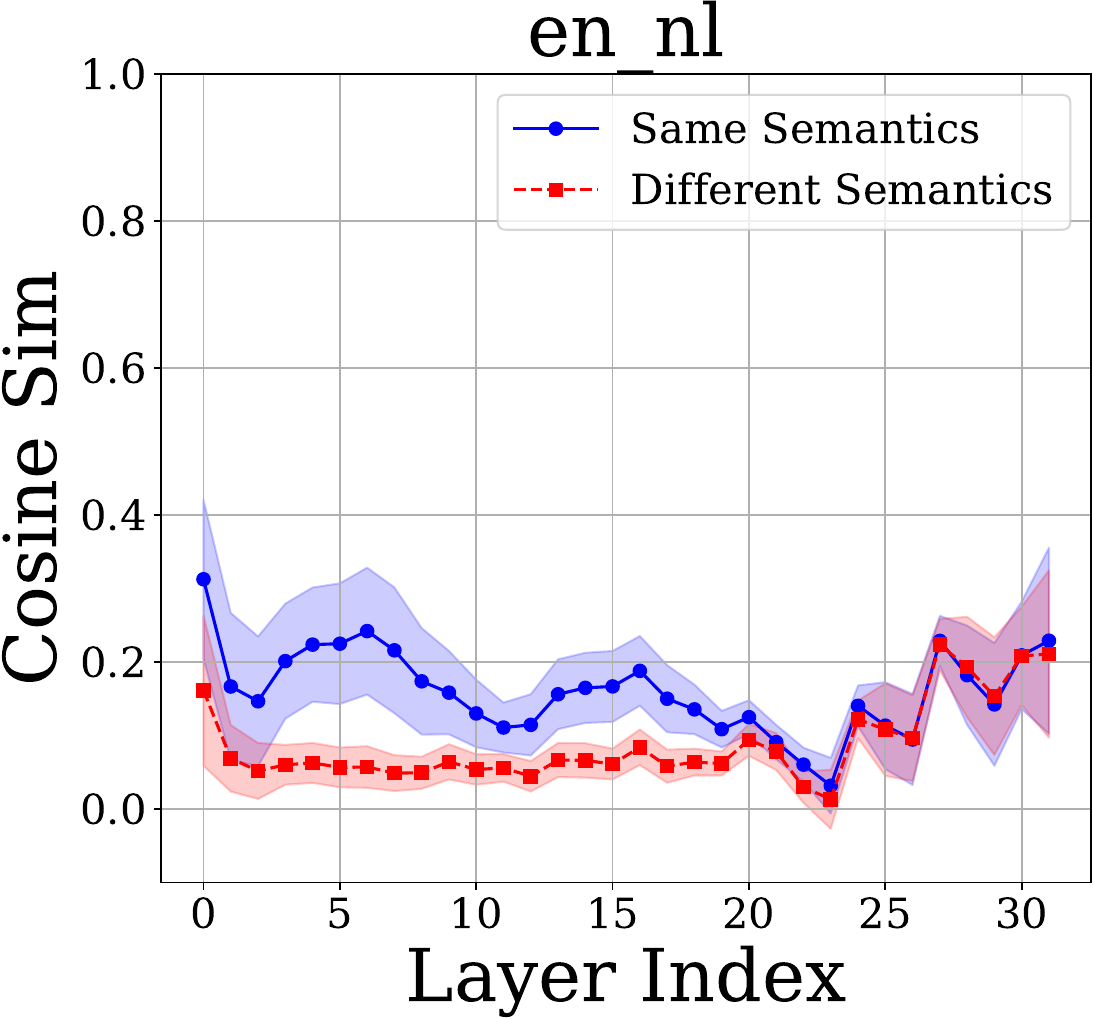}
      \subcaption{en-nl}
    \end{minipage}
    \begin{minipage}{0.23\linewidth}
      \centering
      \includegraphics[width=\linewidth]{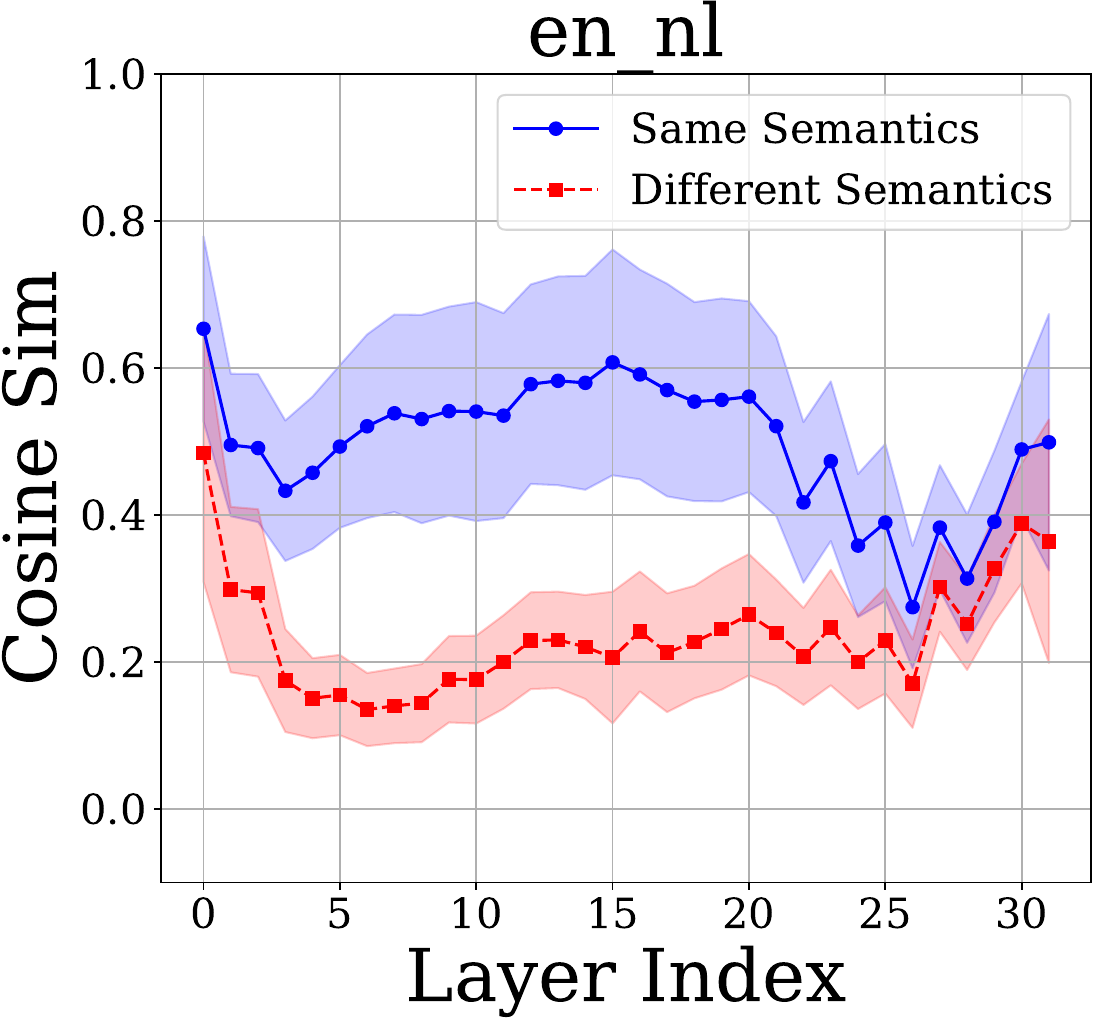}
      \subcaption{en-nl (baseline)}
    \end{minipage}

    \begin{minipage}{0.23\linewidth}
      \centering
      \includegraphics[width=\linewidth]{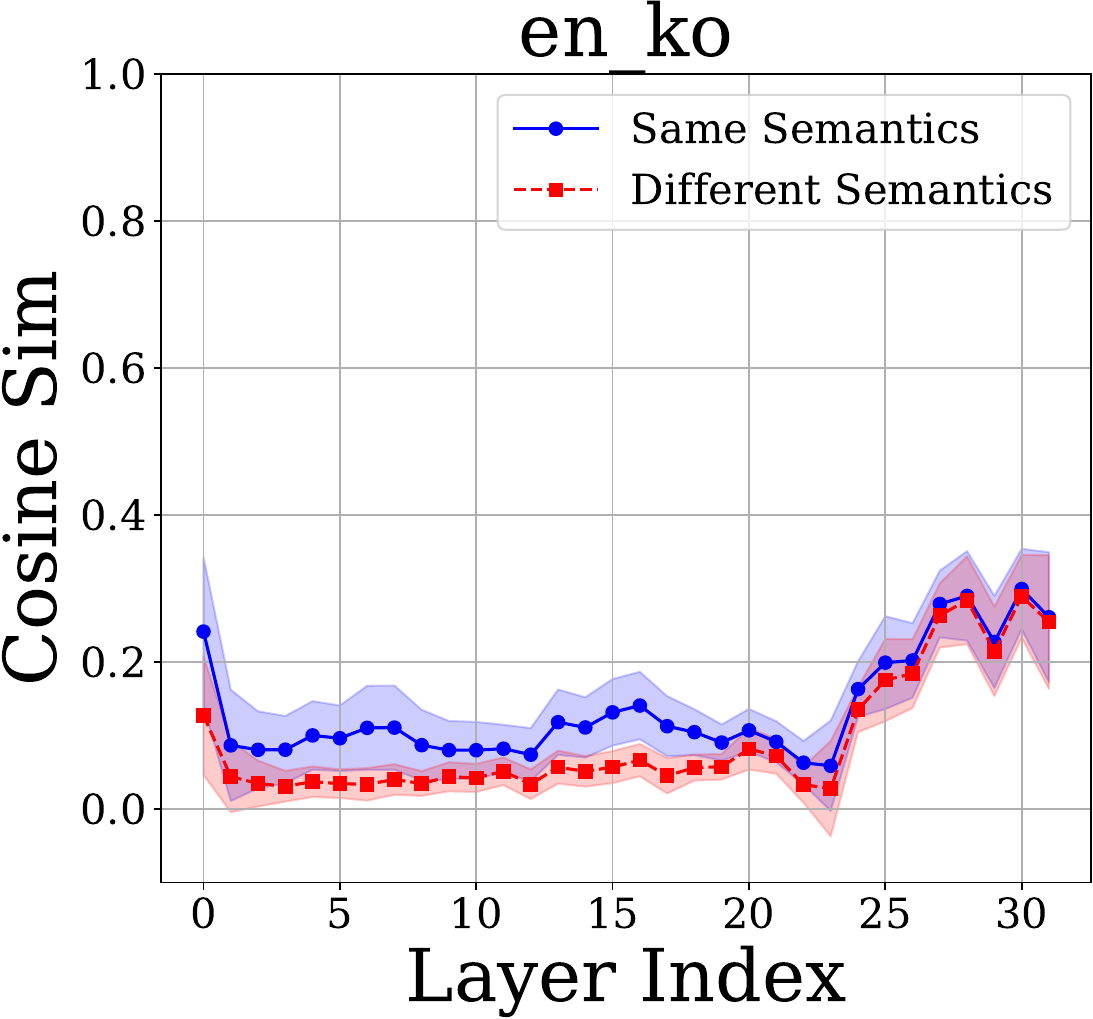}
      \subcaption{en-ko}
    \end{minipage}
    \begin{minipage}{0.23\linewidth}
      \centering
      \includegraphics[width=\linewidth]{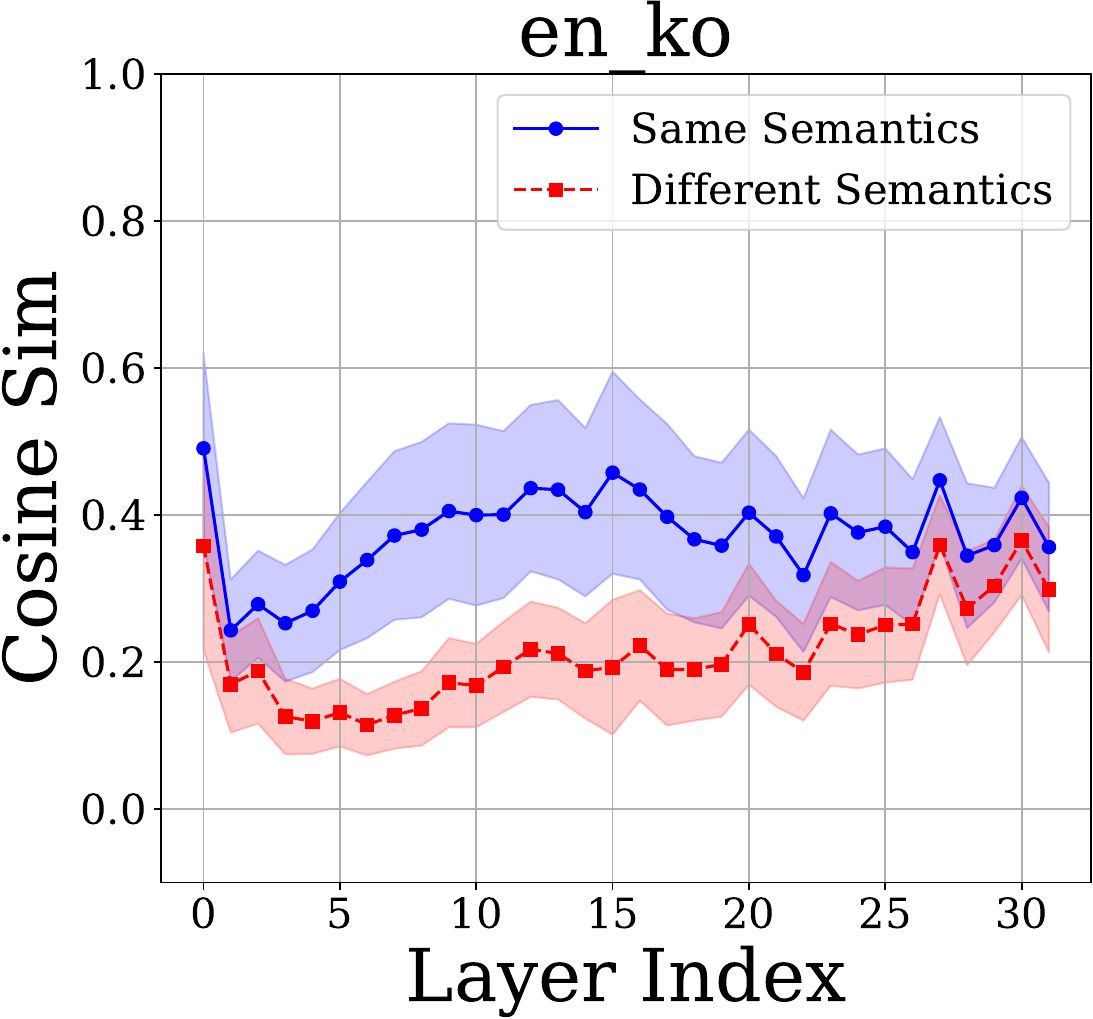}
      \subcaption{en-ko (baseline)}
    \end{minipage}
    \begin{minipage}{0.23\linewidth}
      \centering
      \includegraphics[width=\linewidth]{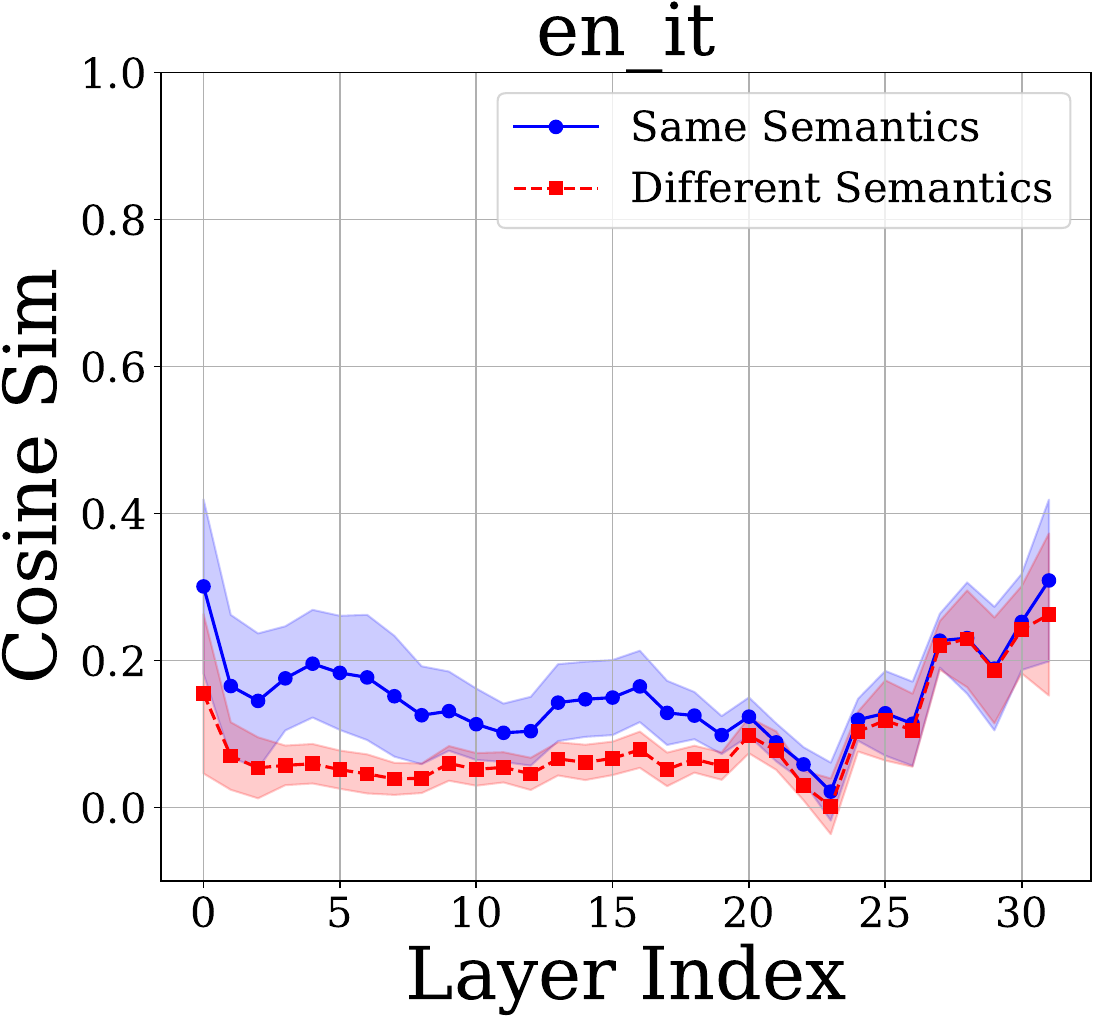}
      \subcaption{en-it}
    \end{minipage}
    \begin{minipage}{0.23\linewidth}
      \centering
      \includegraphics[width=\linewidth]{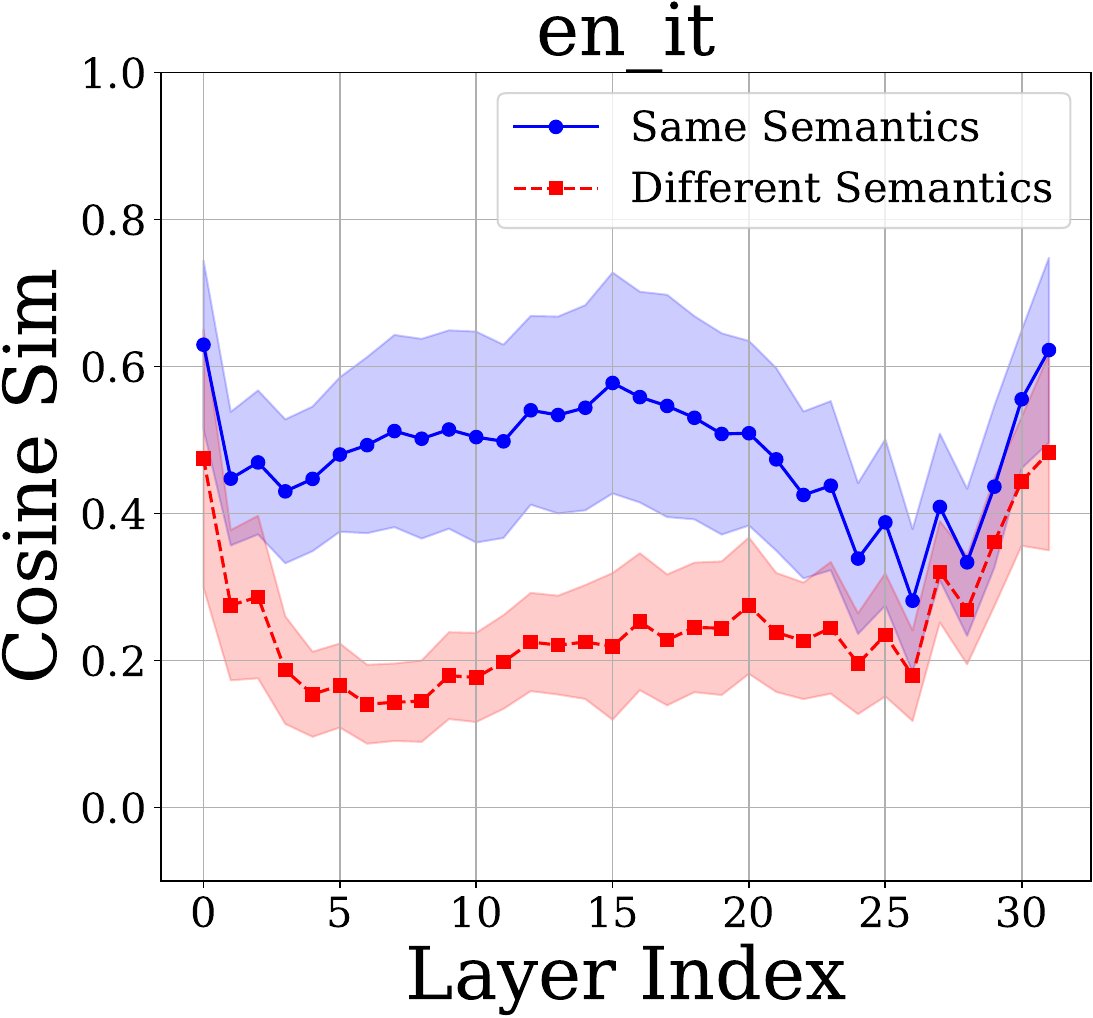}
      
      \subcaption{en-it (baseline)}
    \end{minipage}
    
      \begin{minipage}{\linewidth}
        \centering
        \small \textbf{(b) top-5000 (representing 1\% of all neurons)}
      \end{minipage}

  \caption{\textbf{Similarity of activation patterns across layers while deactivating Type-1 Transfer Neurons (LLaMA3-8B).}}
  \label{fig:appendix:act_sim_llama_deactivating_top-1k_Type-1}
\end{figure*}

\begin{figure*}[t]
    \centering

    \begin{minipage}{0.20\linewidth}
      \centering
      \includegraphics[width=\linewidth]{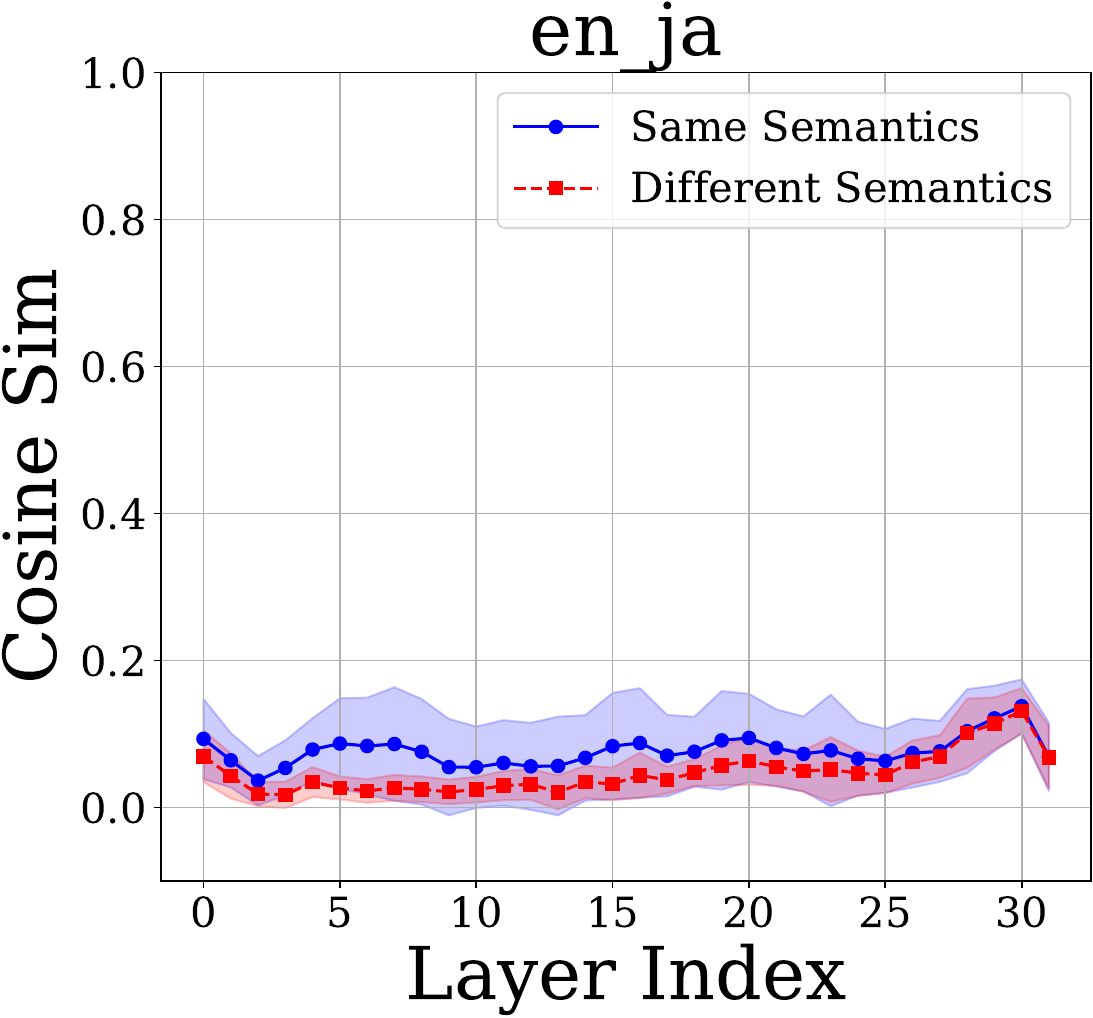}
      \subcaption{en-ja}
    \end{minipage}
    \begin{minipage}{0.20\linewidth}
      \centering
      \includegraphics[width=\linewidth]{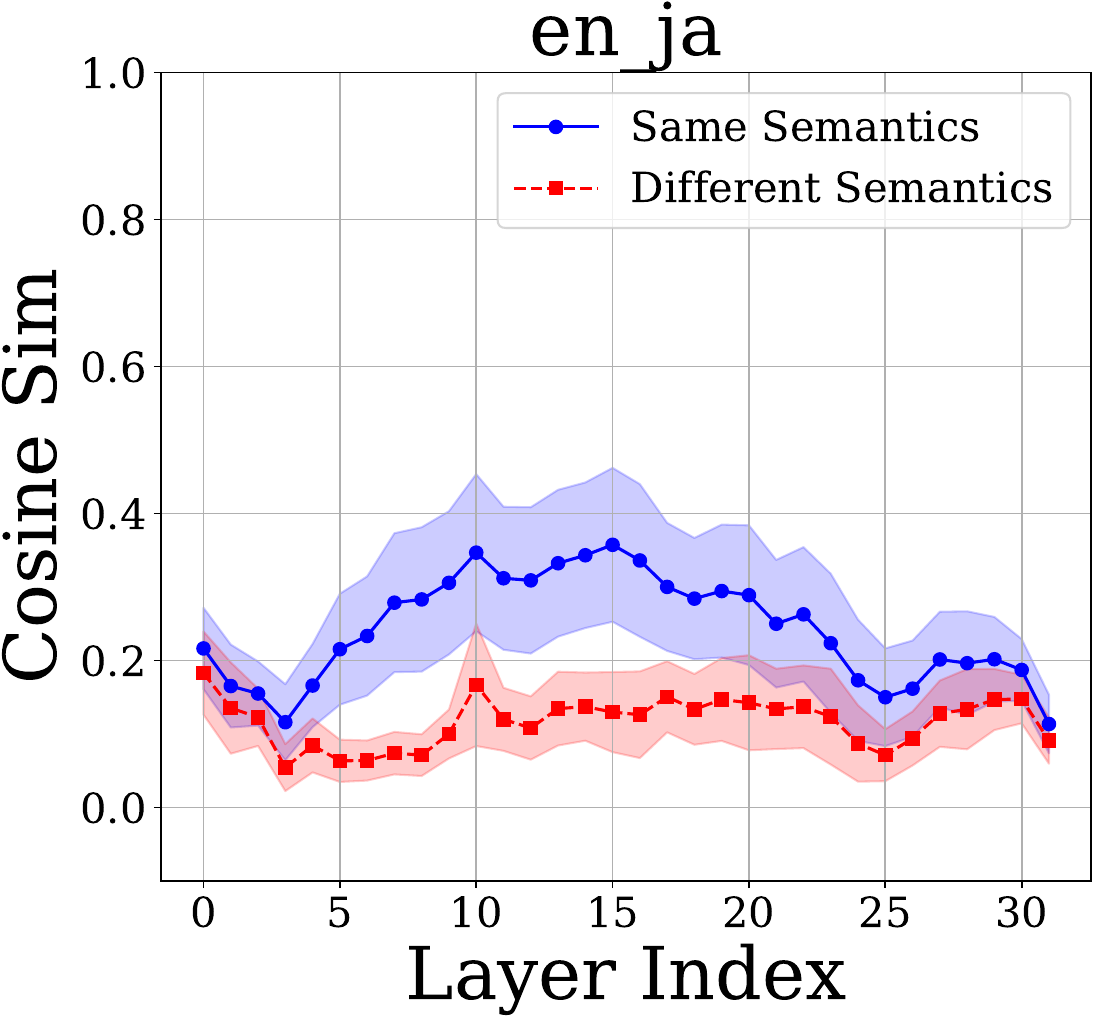}
      \subcaption{en-ja (baseline)}
    \end{minipage}
    \begin{minipage}{0.20\linewidth}
      \centering
      \includegraphics[width=\linewidth]{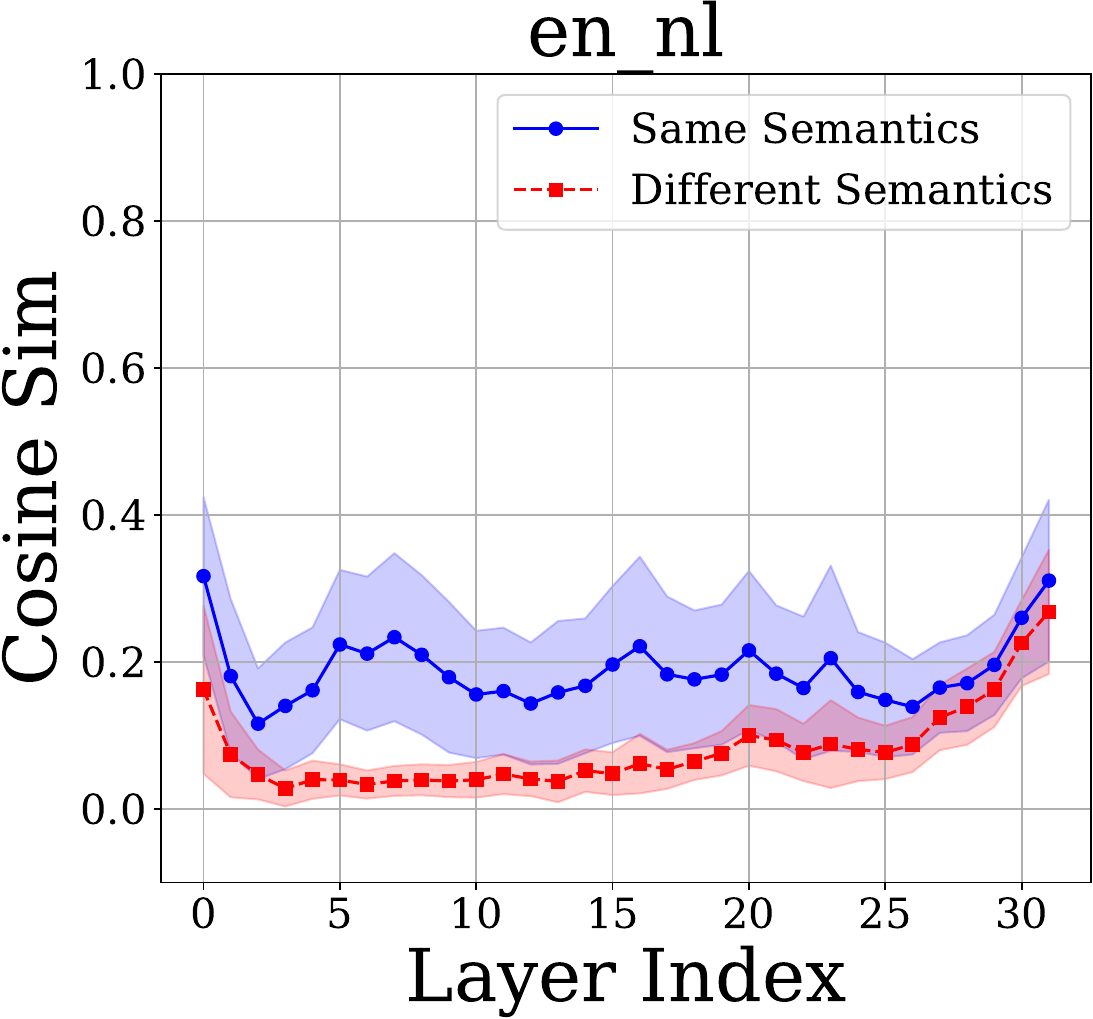}
      \subcaption{en-nl}
    \end{minipage}
    \begin{minipage}{0.20\linewidth}
      \centering
      \includegraphics[width=\linewidth]{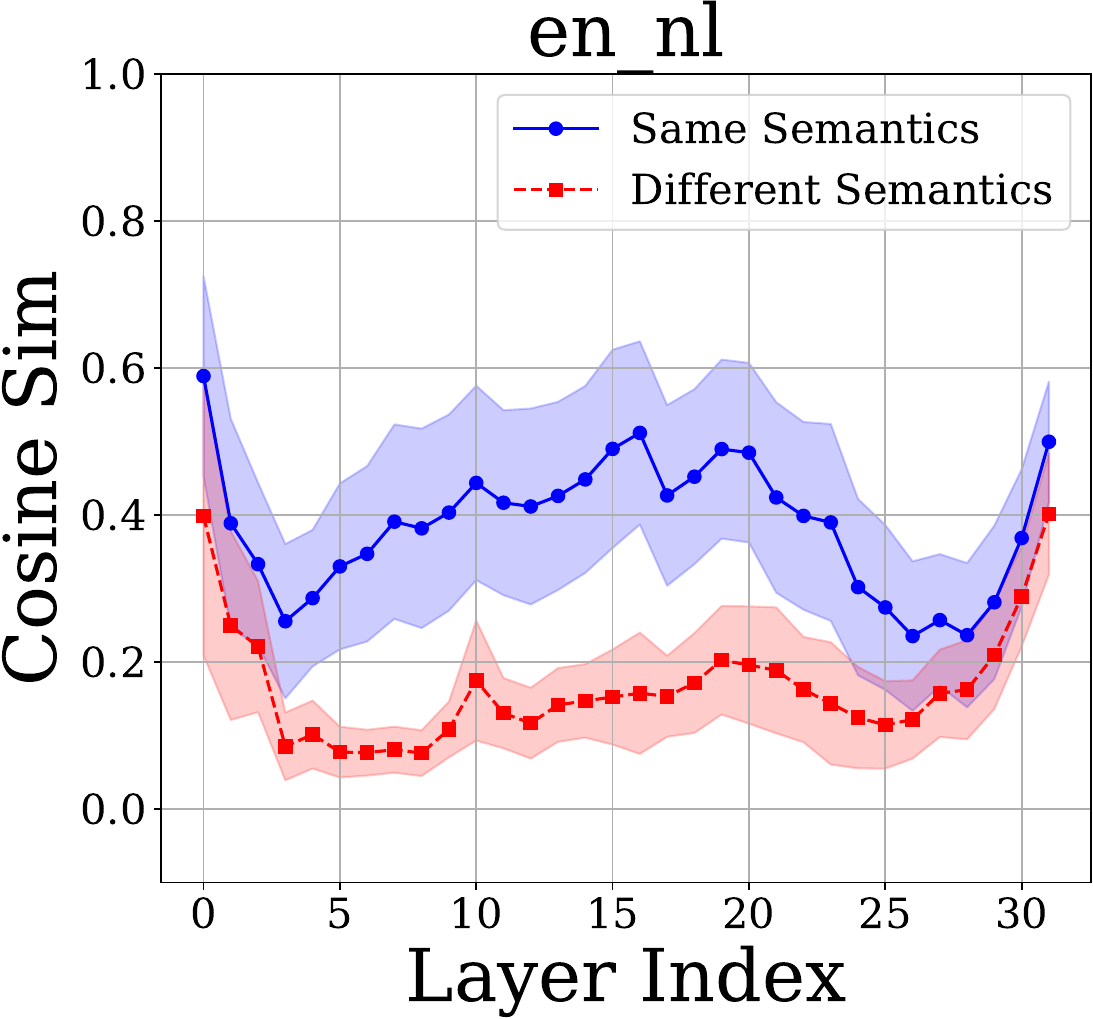}
      \subcaption{en-nl (baseline)}
    \end{minipage}

    \begin{minipage}{0.20\linewidth}
      \centering
      \includegraphics[width=\linewidth]{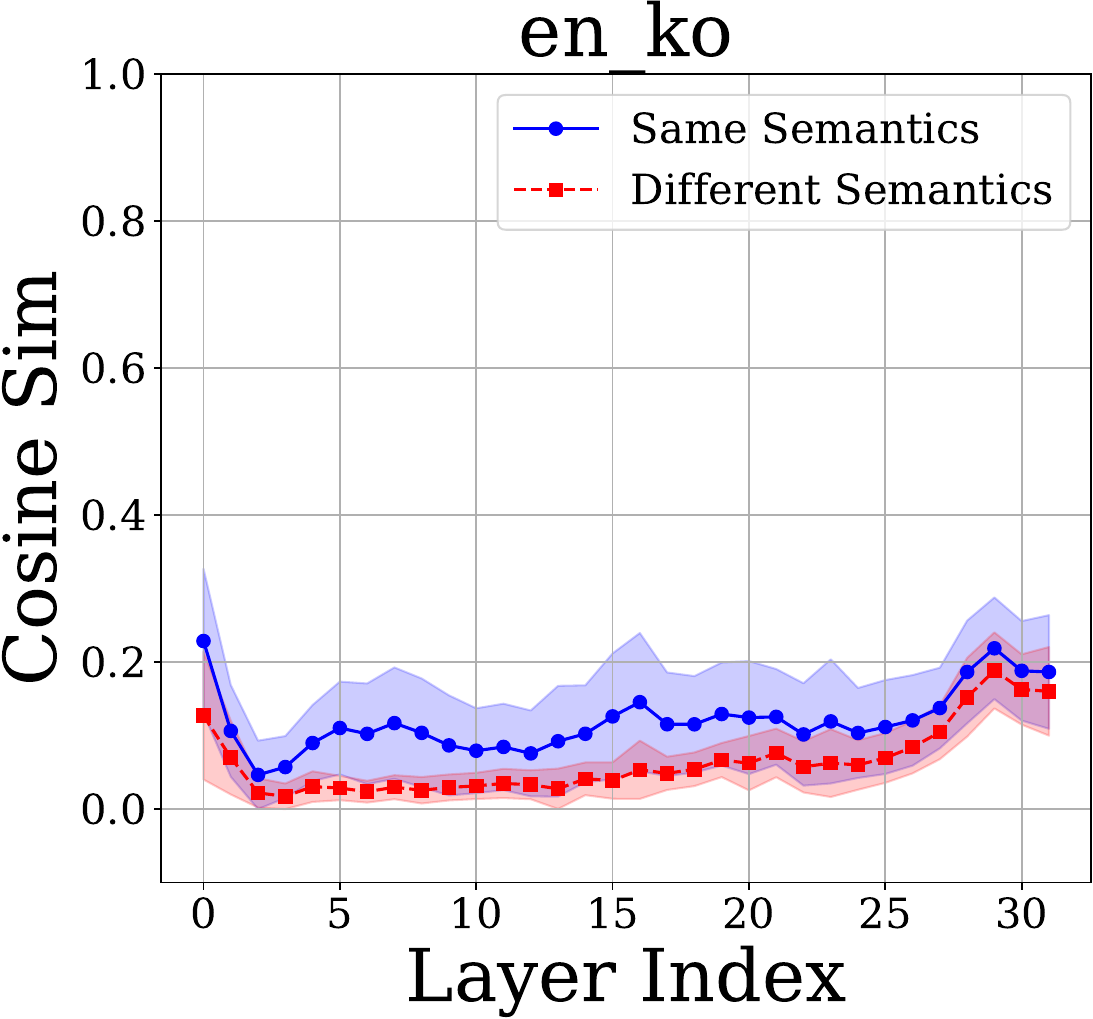}
      \subcaption{en-ko}
    \end{minipage}
    \begin{minipage}{0.20\linewidth}
      \centering
      \includegraphics[width=\linewidth]{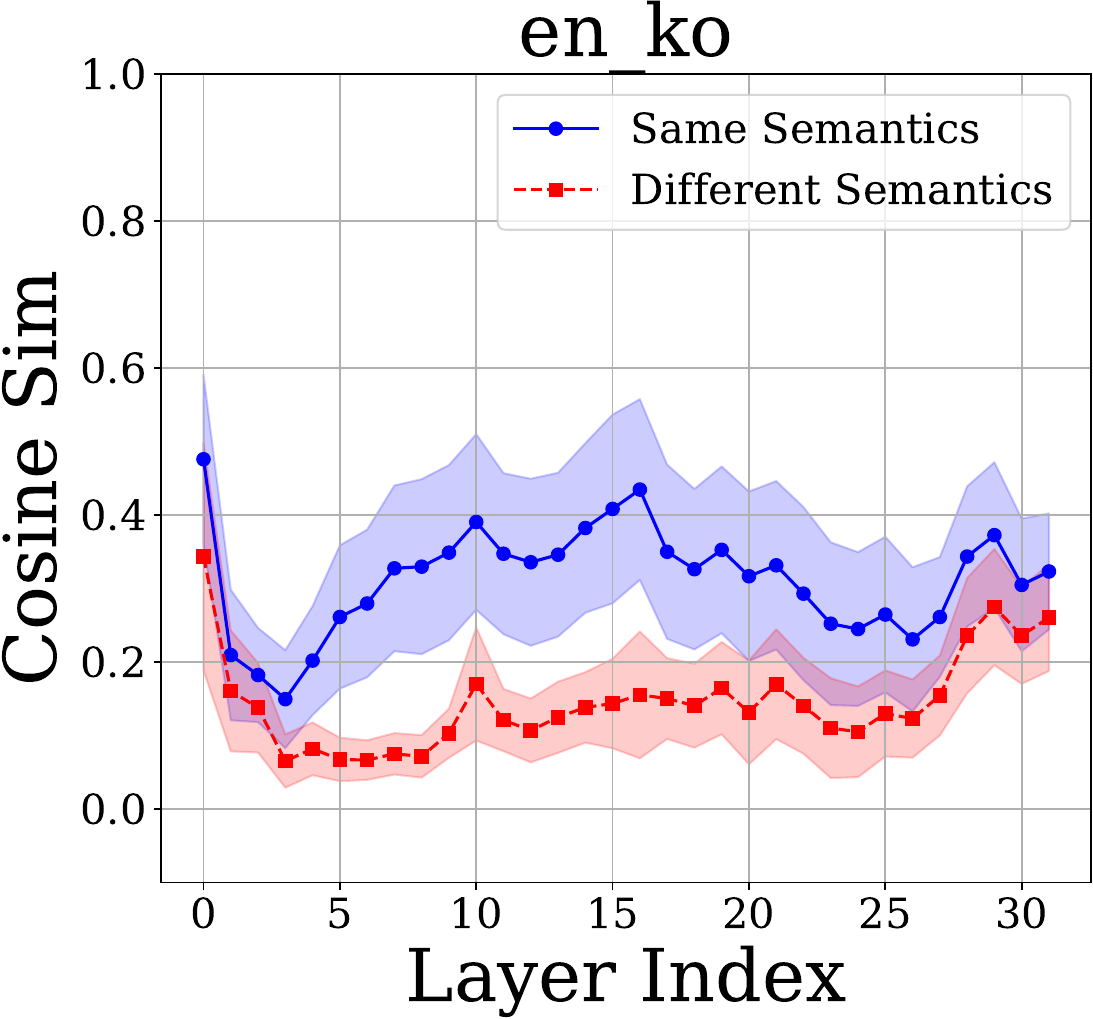}
      \subcaption{en-ko (baseline)}
    \end{minipage}
    \begin{minipage}{0.20\linewidth}
      \centering
      \includegraphics[width=\linewidth]{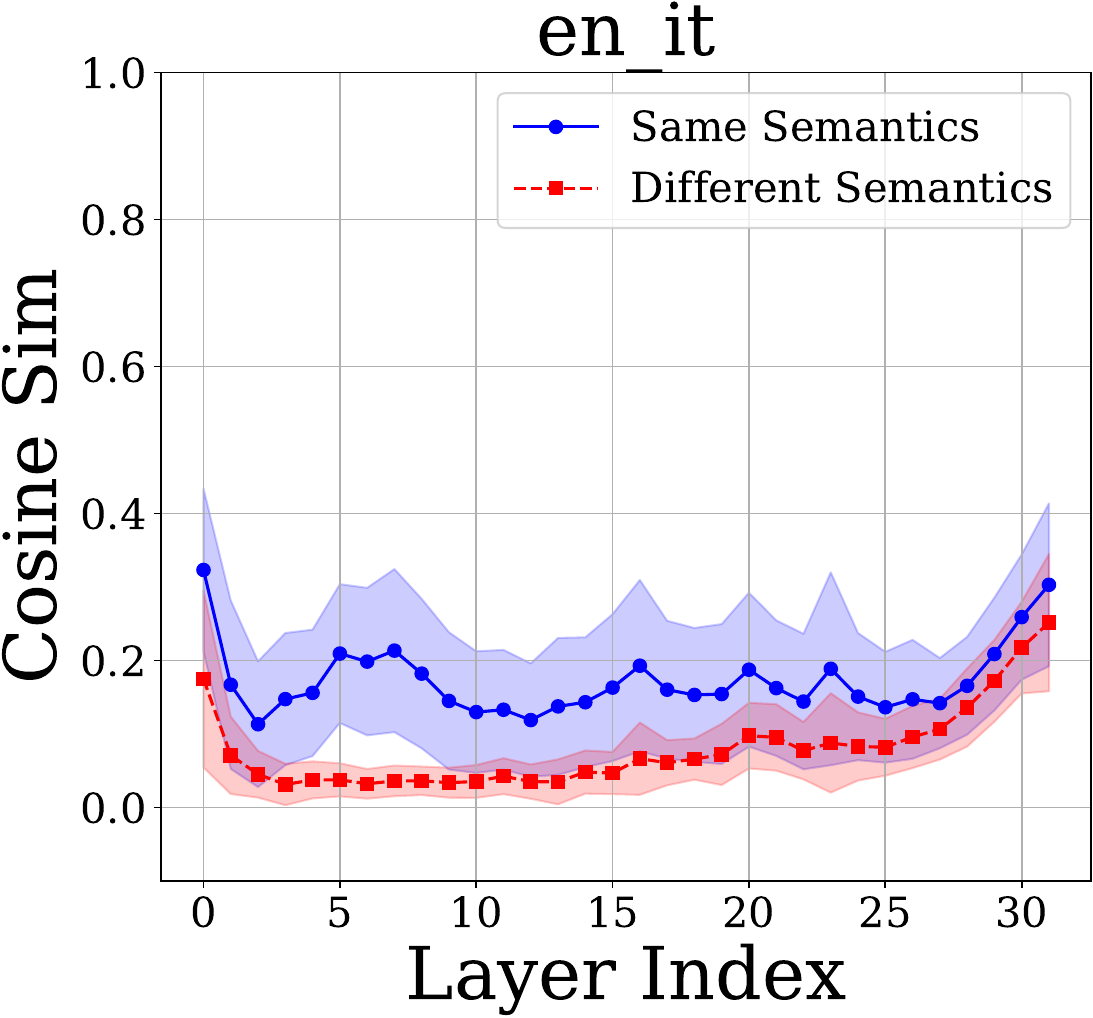}
      \subcaption{en-it}
    \end{minipage}
    \begin{minipage}{0.20\linewidth}
      \centering
      \includegraphics[width=\linewidth]{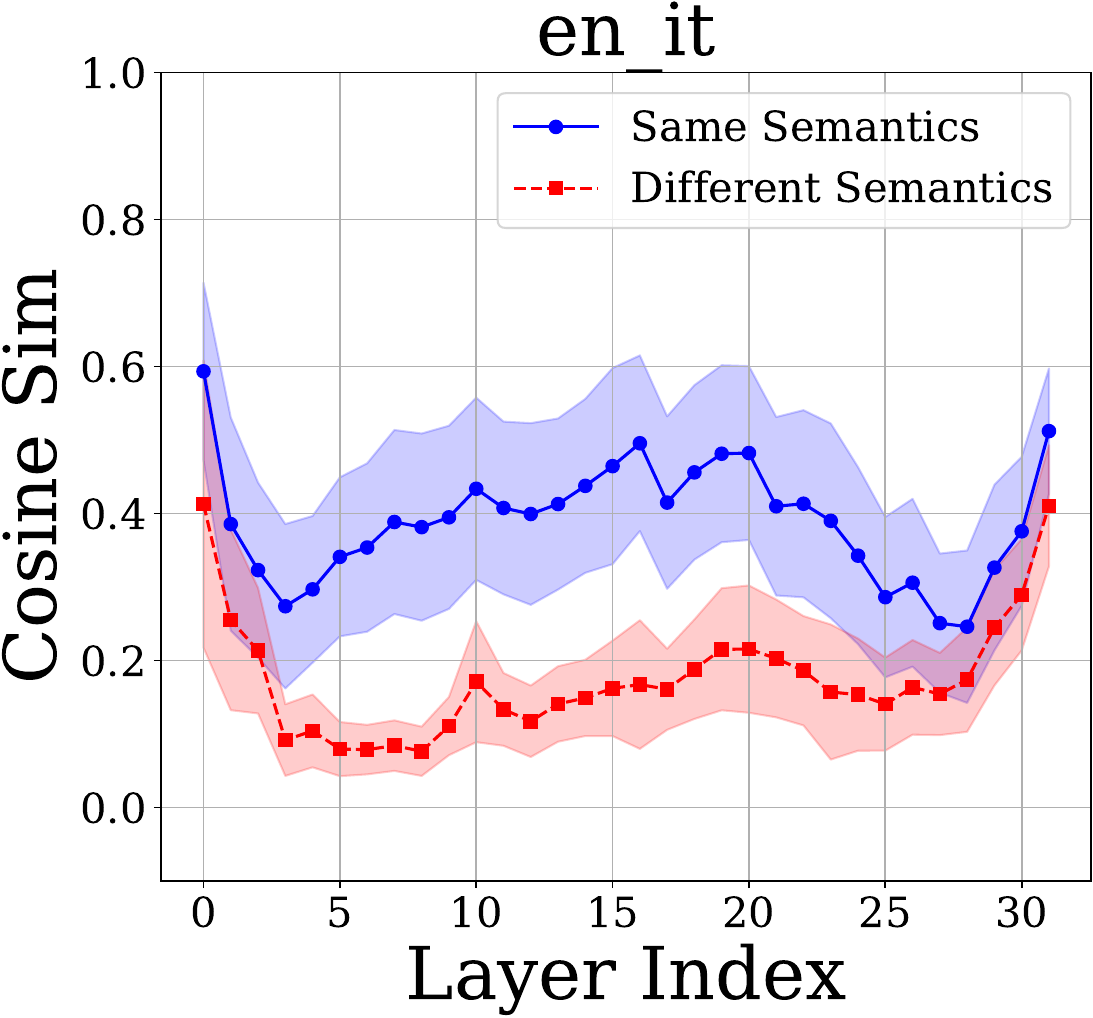}
      \subcaption{en-it (baseline)}
    \end{minipage}
    
      % First row label
      \begin{minipage}{\linewidth}
        \centering
        \small \textbf{(a) top-1000 (representing 0.2\% of all neurons)}
      \end{minipage}

    % second row
    \begin{minipage}{0.20\linewidth}
      \centering
      \includegraphics[width=\linewidth]{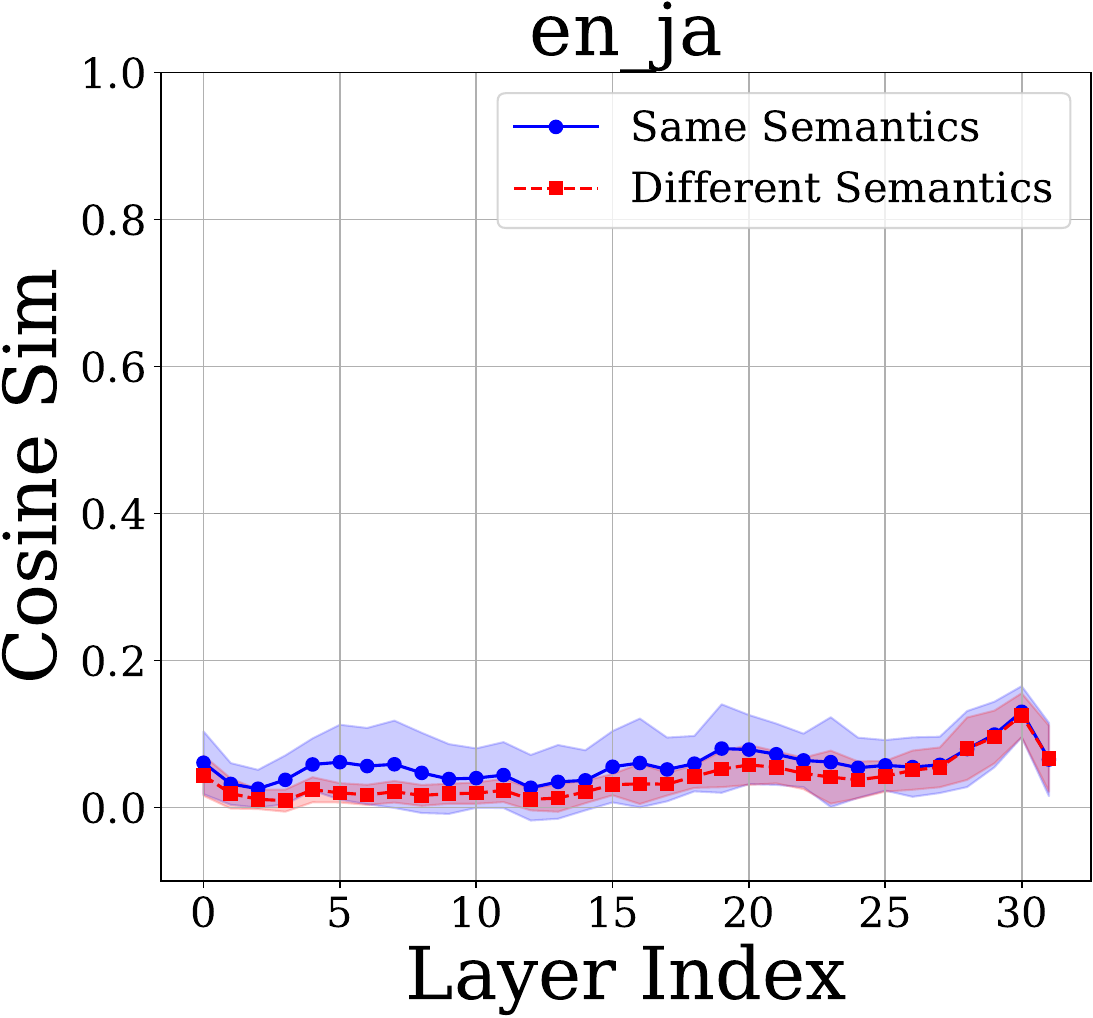}
      \subcaption{en-ja}
    \end{minipage}
    \begin{minipage}{0.20\linewidth}
      \centering
      \includegraphics[width=\linewidth]{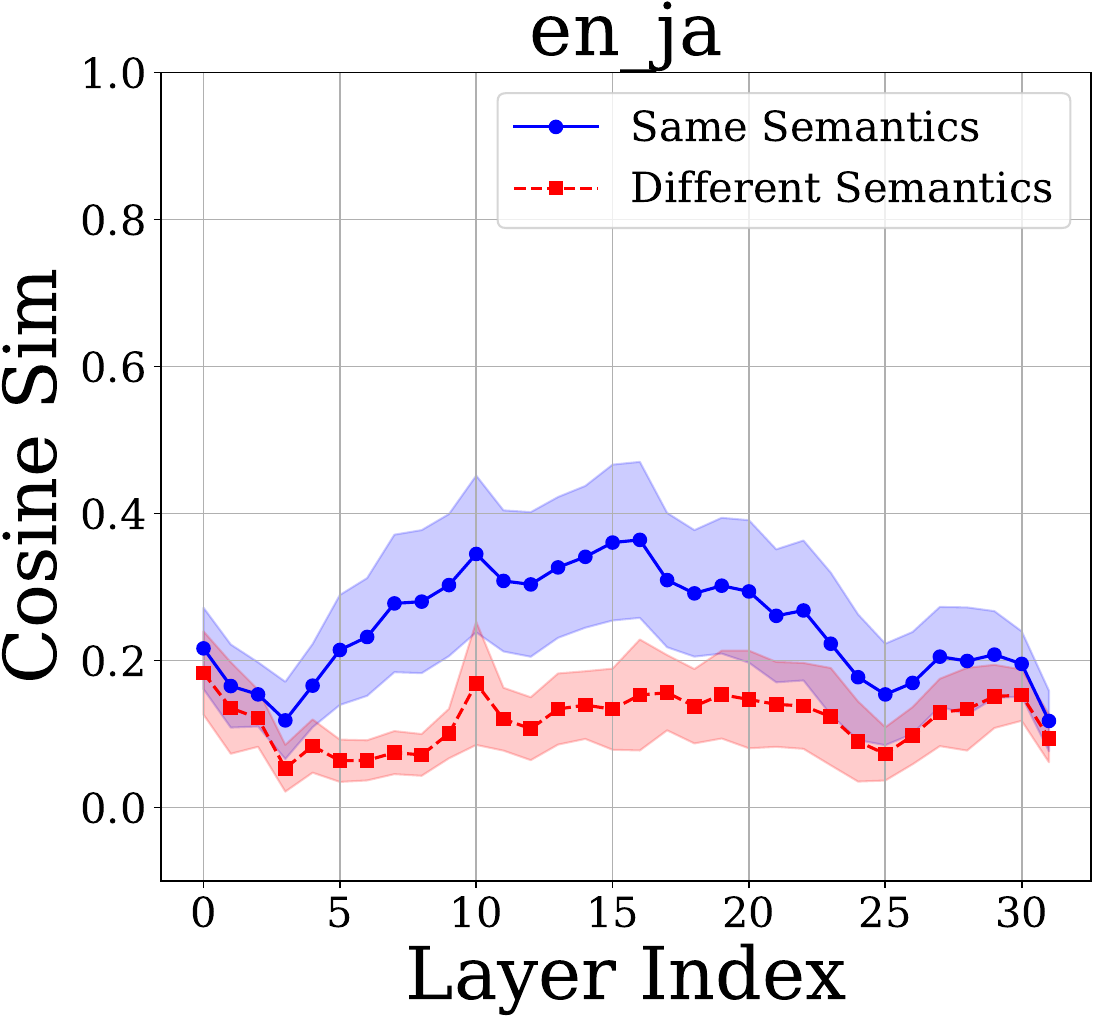}
      \subcaption{en-ja (baseline)}
    \end{minipage}
    \begin{minipage}{0.20\linewidth}
      \centering
      \includegraphics[width=\linewidth]{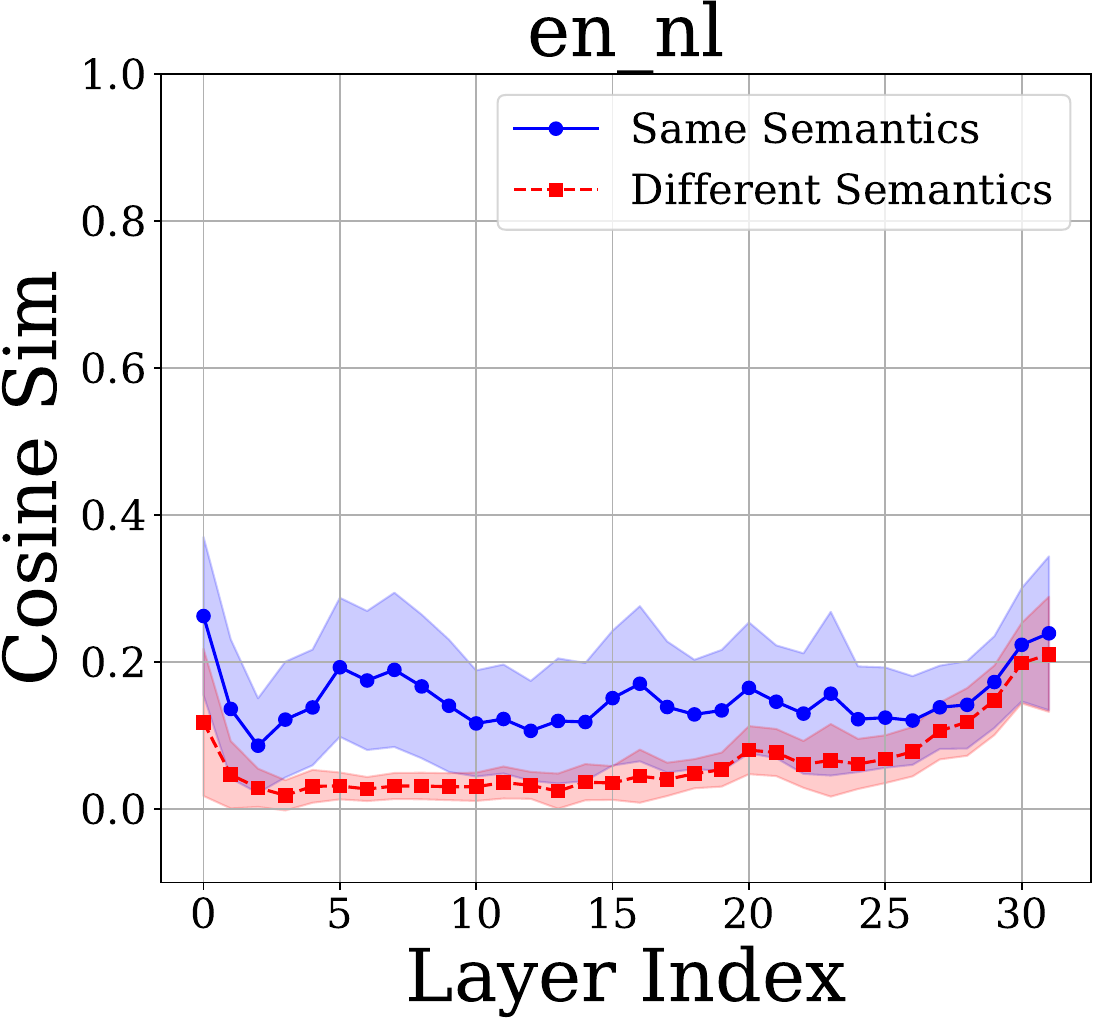}
      \subcaption{en-nl}
    \end{minipage}
    \begin{minipage}{0.20\linewidth}
      \centering
      \includegraphics[width=\linewidth]{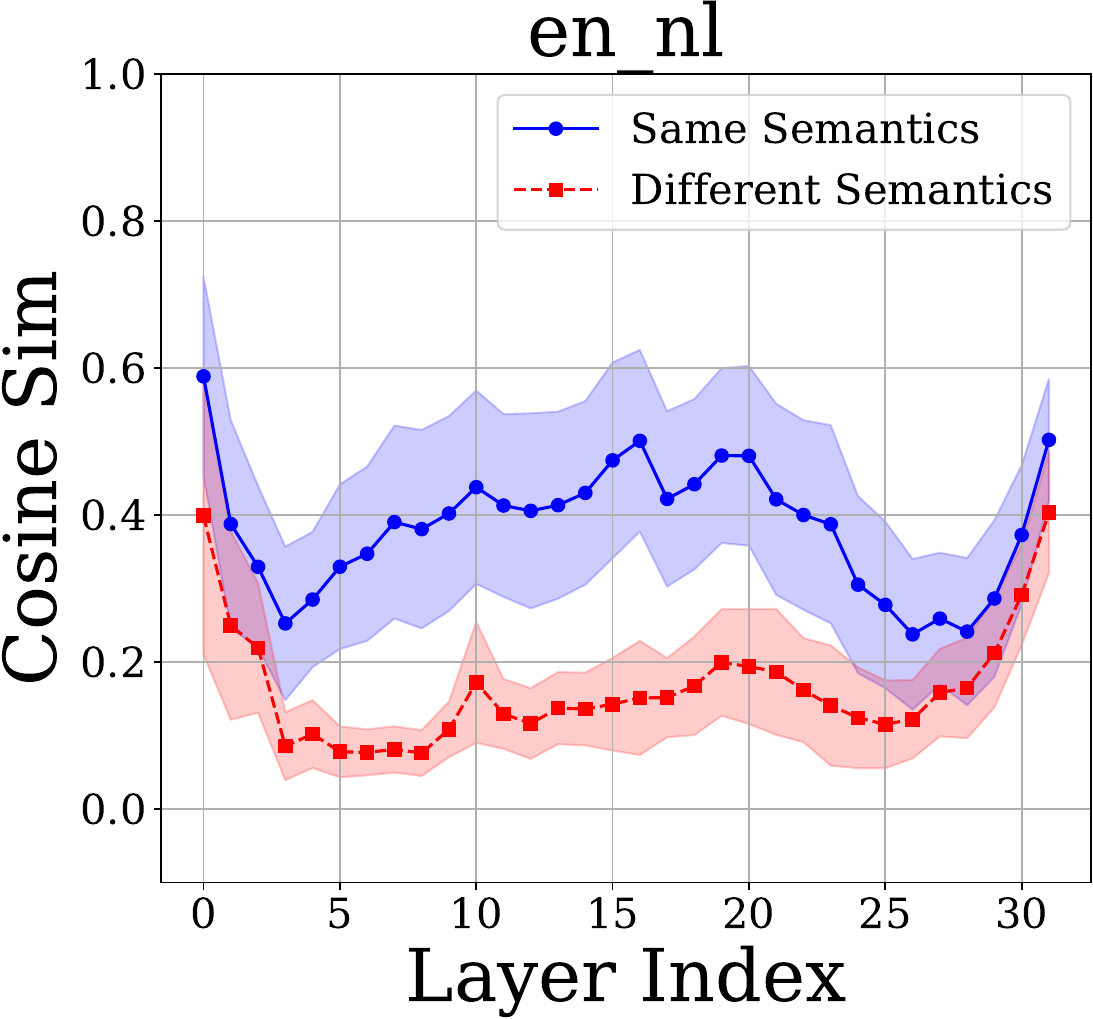}
      \subcaption{en-nl (baseline)}
    \end{minipage}

    \begin{minipage}{0.20\linewidth}
      \centering
      \includegraphics[width=\linewidth]{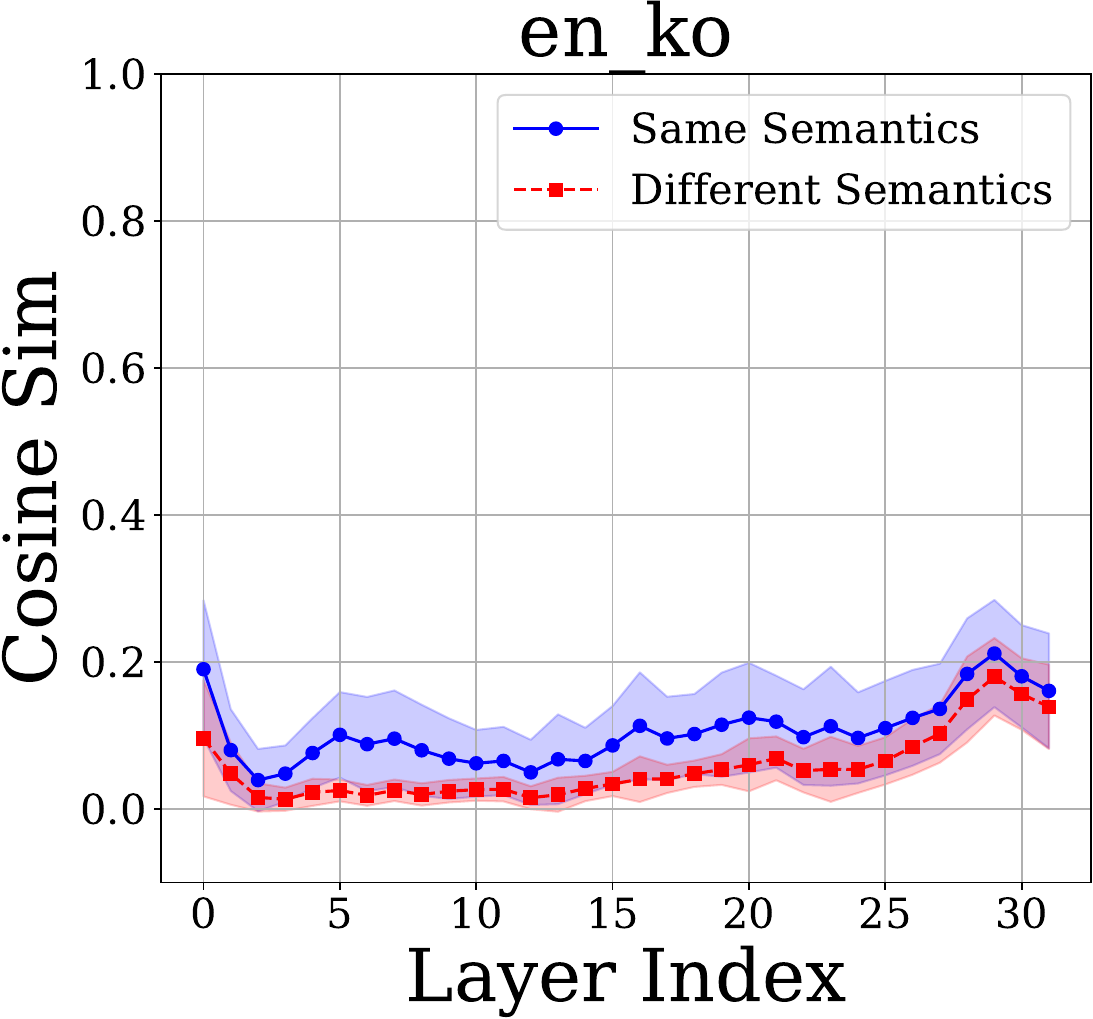}
      \subcaption{en-ko}
    \end{minipage}
    \begin{minipage}{0.20\linewidth}
      \centering
      \includegraphics[width=\linewidth]{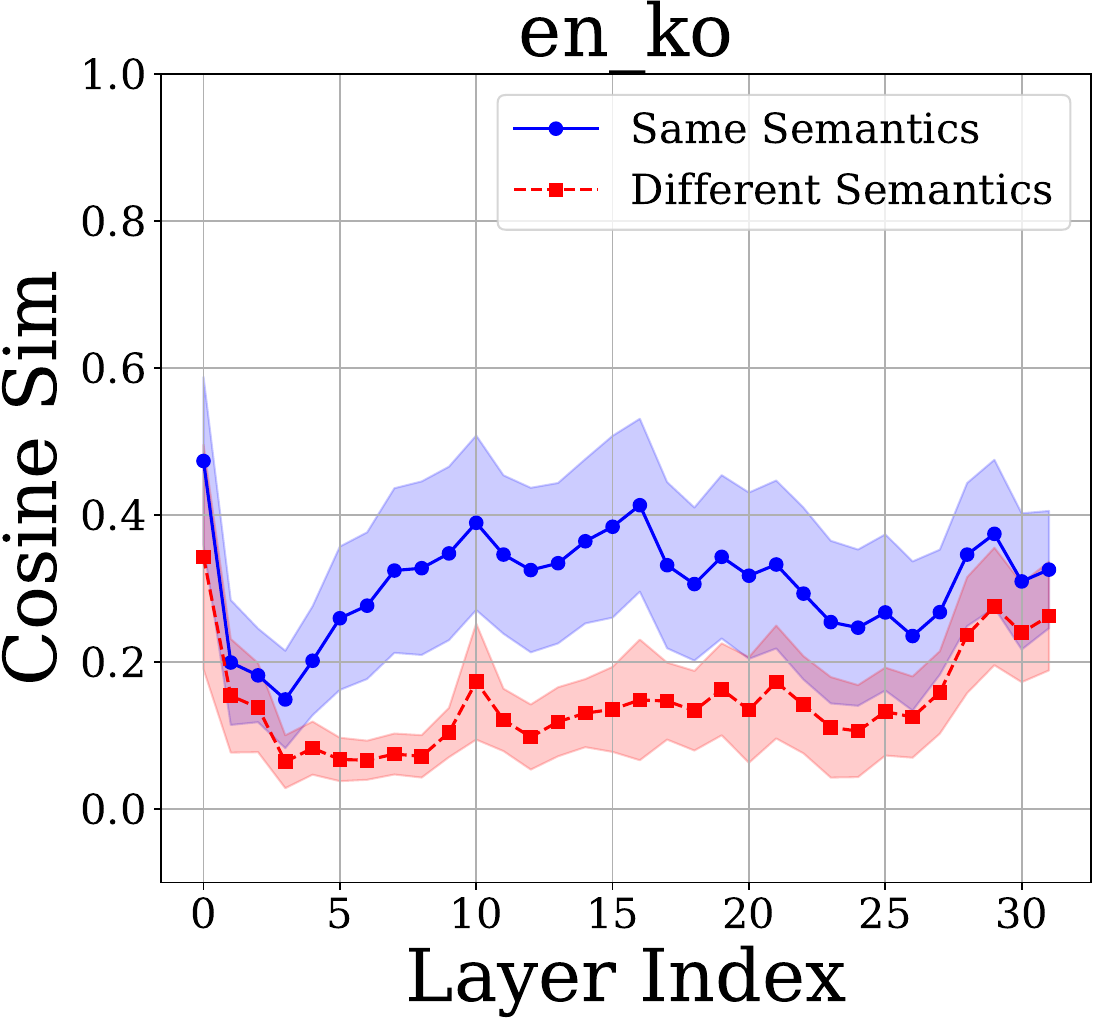}
      \subcaption{en-ko (baseline)}
    \end{minipage}
    \begin{minipage}{0.20\linewidth}
      \centering
      \includegraphics[width=\linewidth]{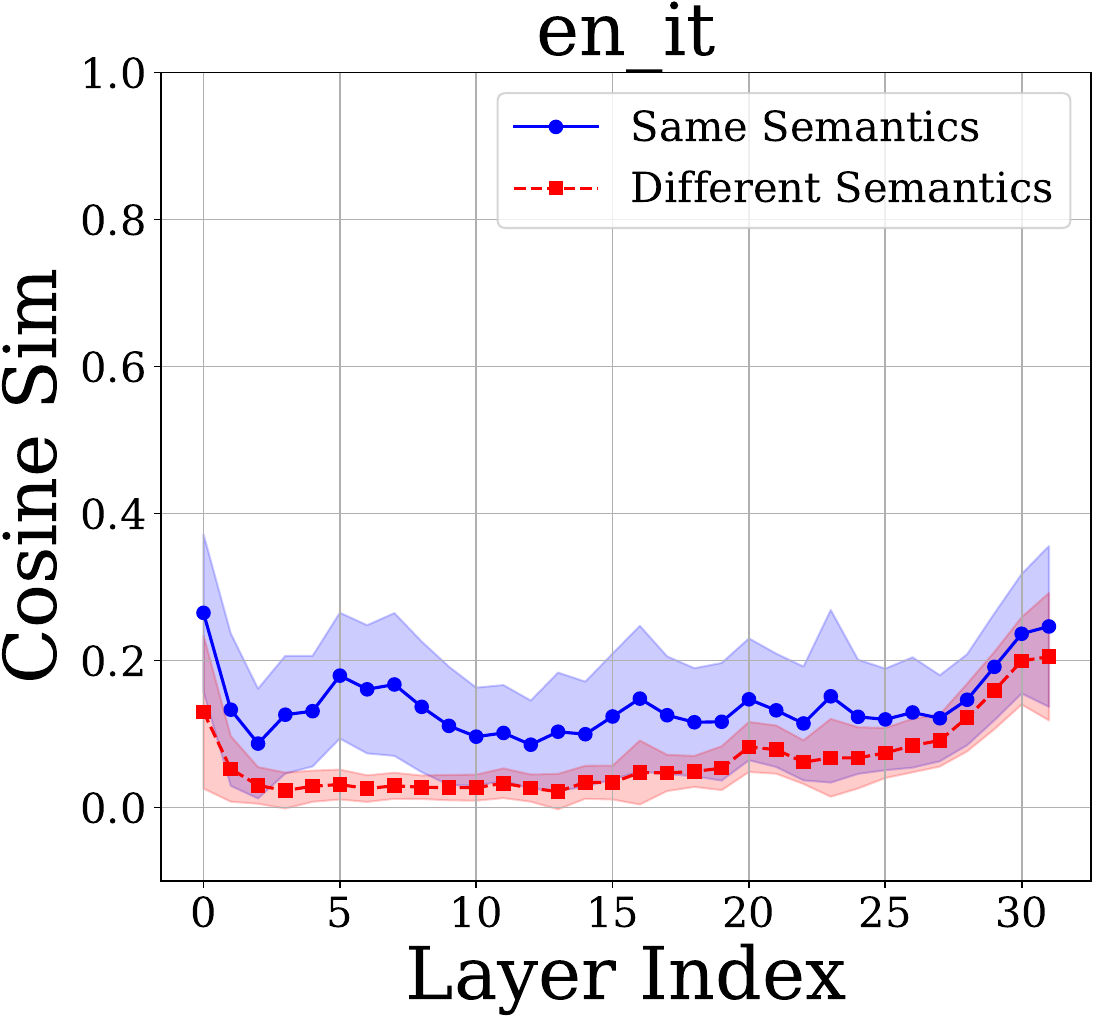}
      \subcaption{en-it}
    \end{minipage}
    \begin{minipage}{0.20\linewidth}
      \centering
      \includegraphics[width=\linewidth]{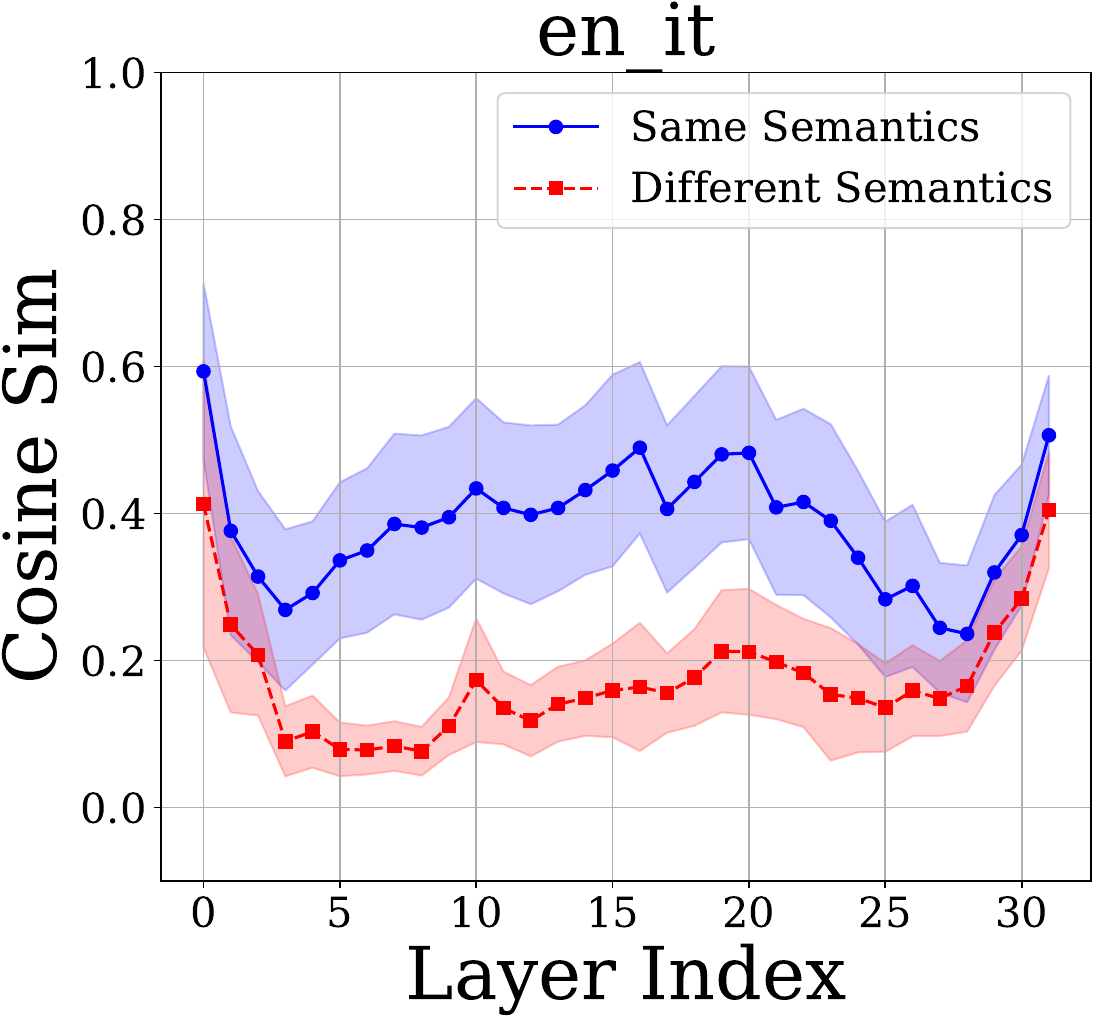}
      \subcaption{en-it (baseline)}
    \end{minipage}
    
      \begin{minipage}{\linewidth}
        \centering
        \small \textbf{(b) top-3000 (representing 0.6\% of all neurons)}
      \end{minipage}

    \begin{minipage}{0.20\linewidth}
      \centering
      \includegraphics[width=\linewidth]{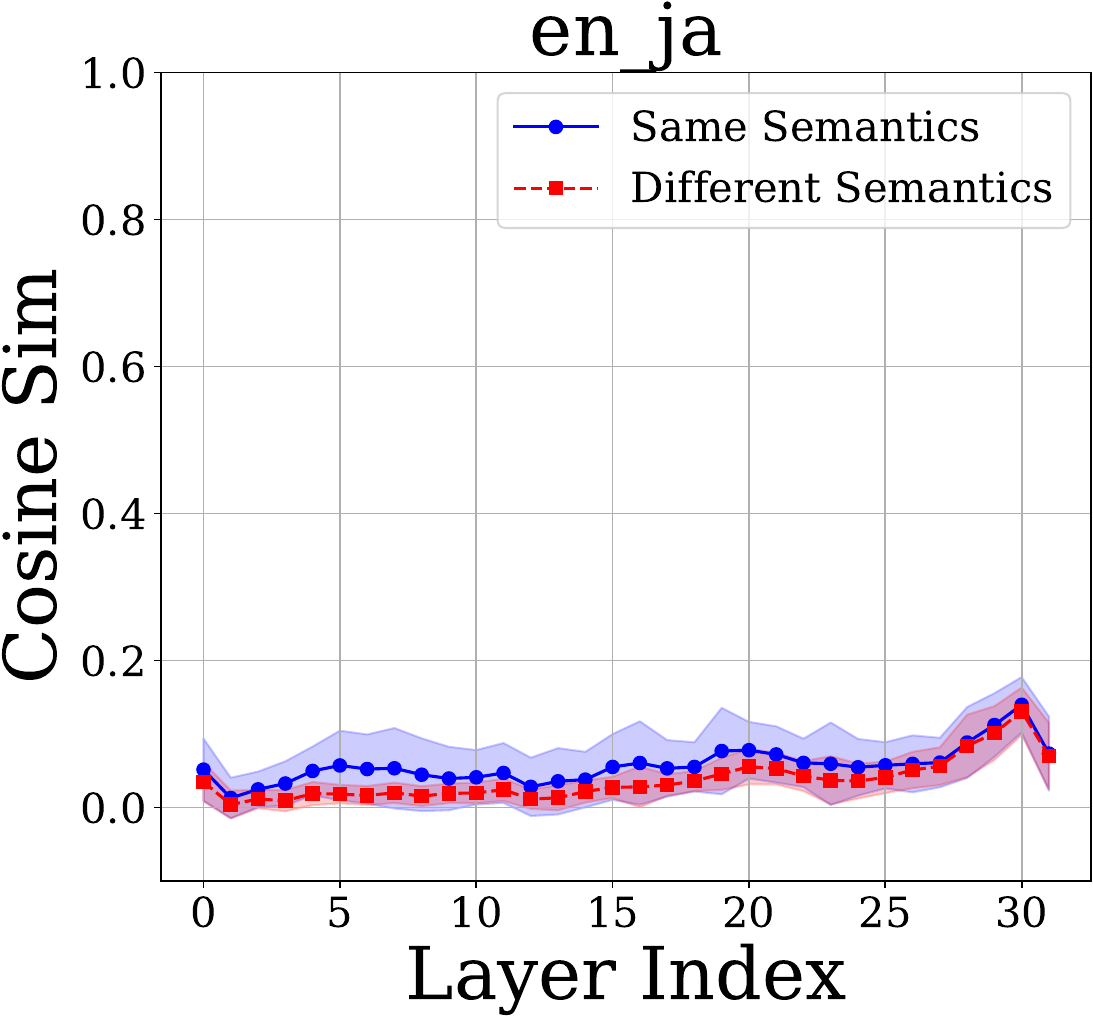}
      \subcaption{en-ja}
    \end{minipage}
    \begin{minipage}{0.20\linewidth}
      \centering
      \includegraphics[width=\linewidth]{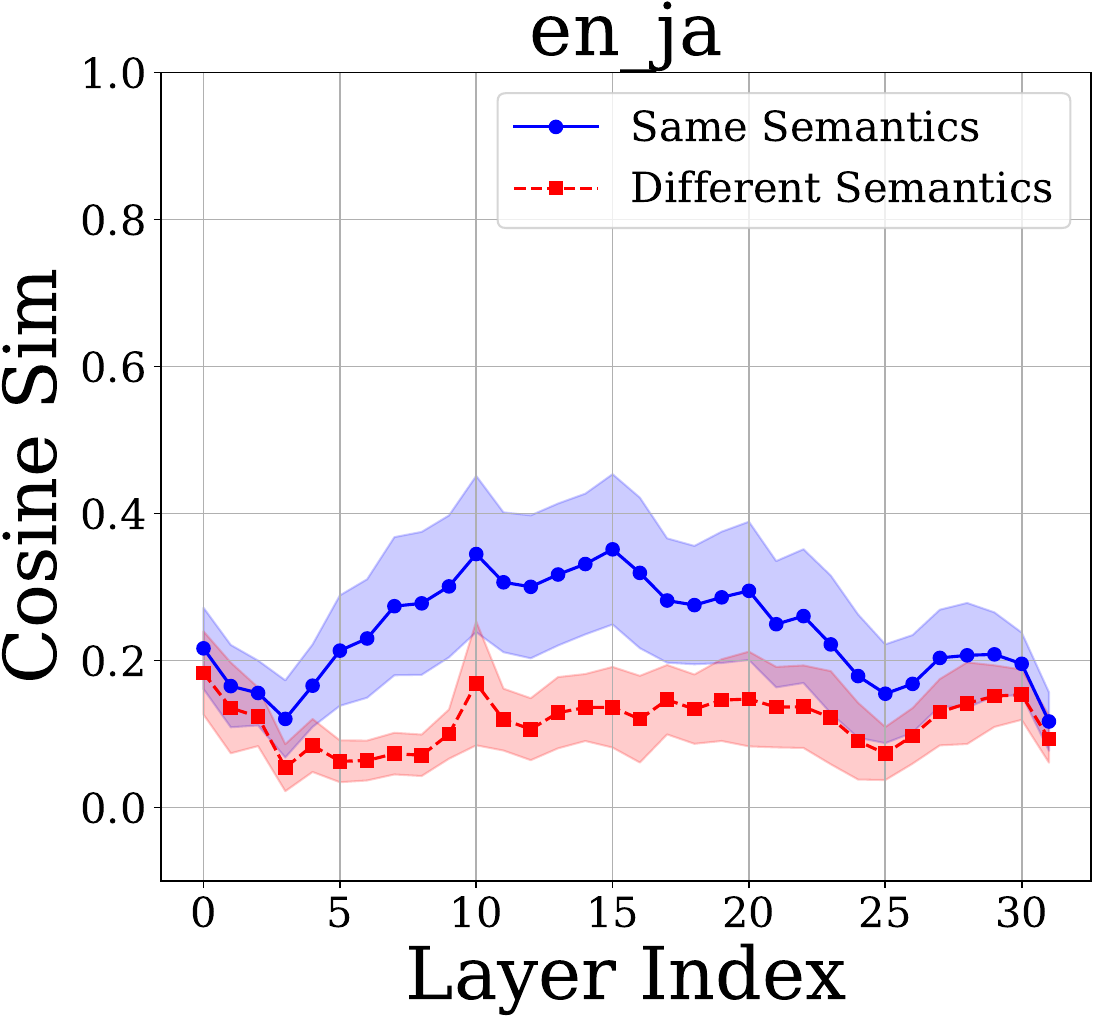}
      \subcaption{en-ja (baseline)}
    \end{minipage}
    \begin{minipage}{0.20\linewidth}
      \centering
      \includegraphics[width=\linewidth]{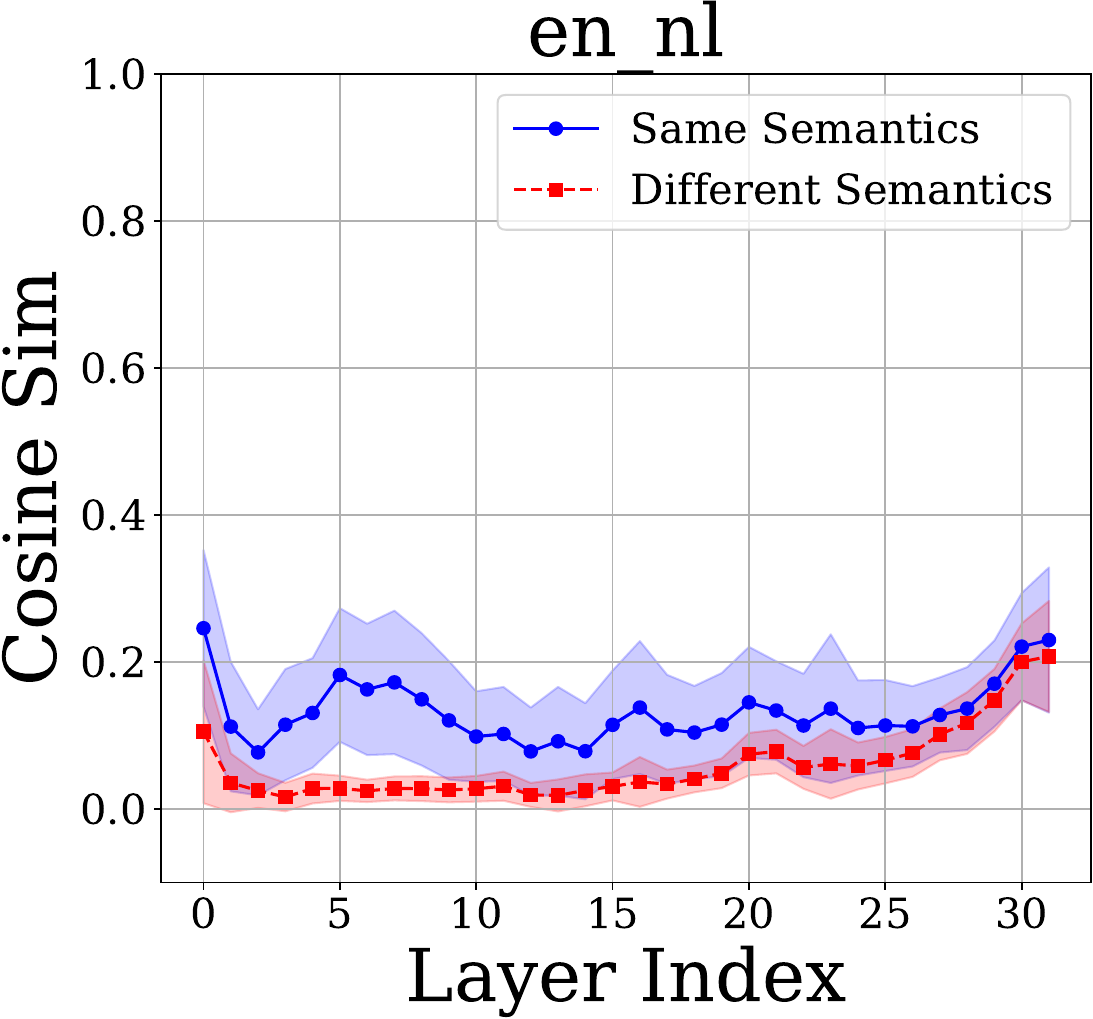}
      \subcaption{en-nl}
    \end{minipage}
    \begin{minipage}{0.20\linewidth}
      \centering
      \includegraphics[width=\linewidth]{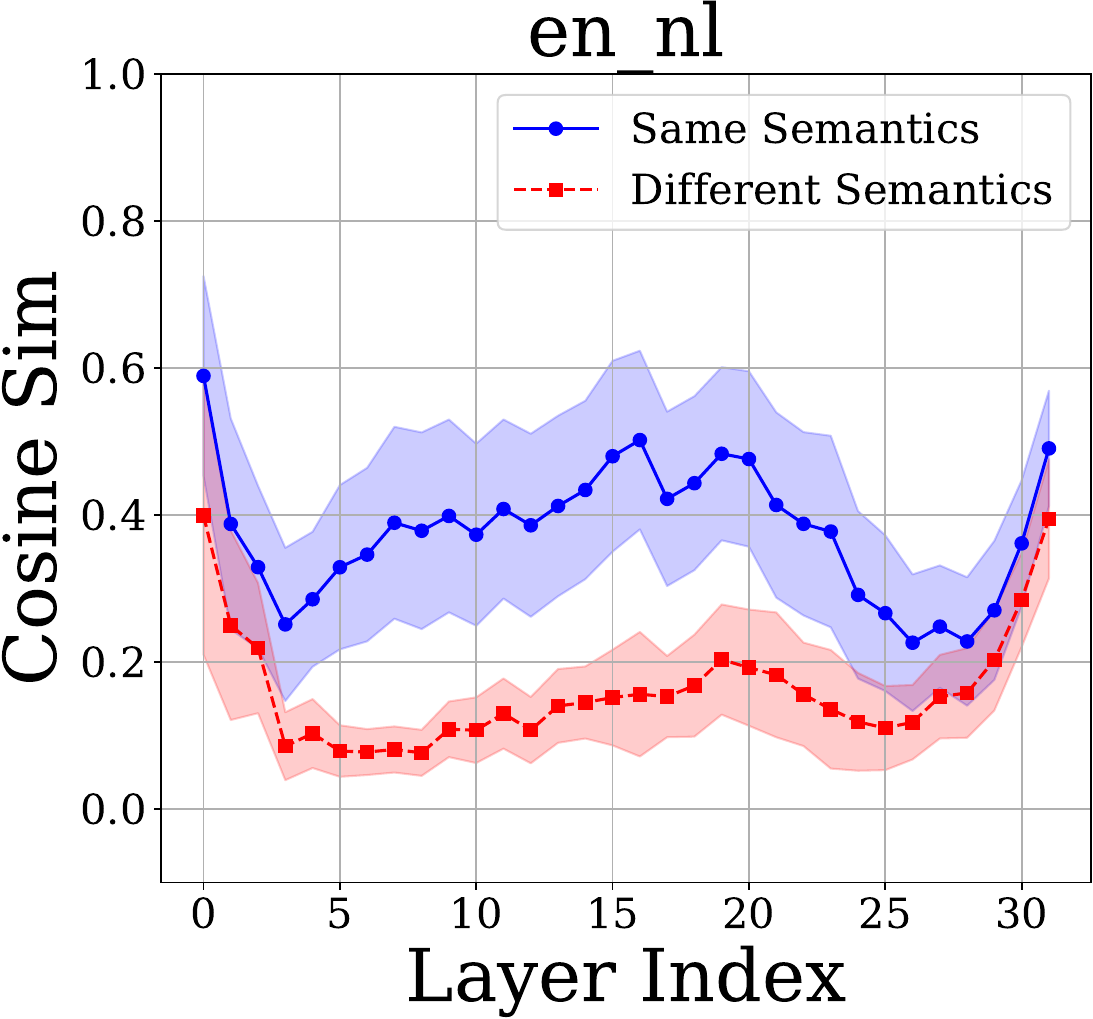}
      \subcaption{en-nl (baseline)}
    \end{minipage}

    \begin{minipage}{0.20\linewidth}
      \centering
      \includegraphics[width=\linewidth]{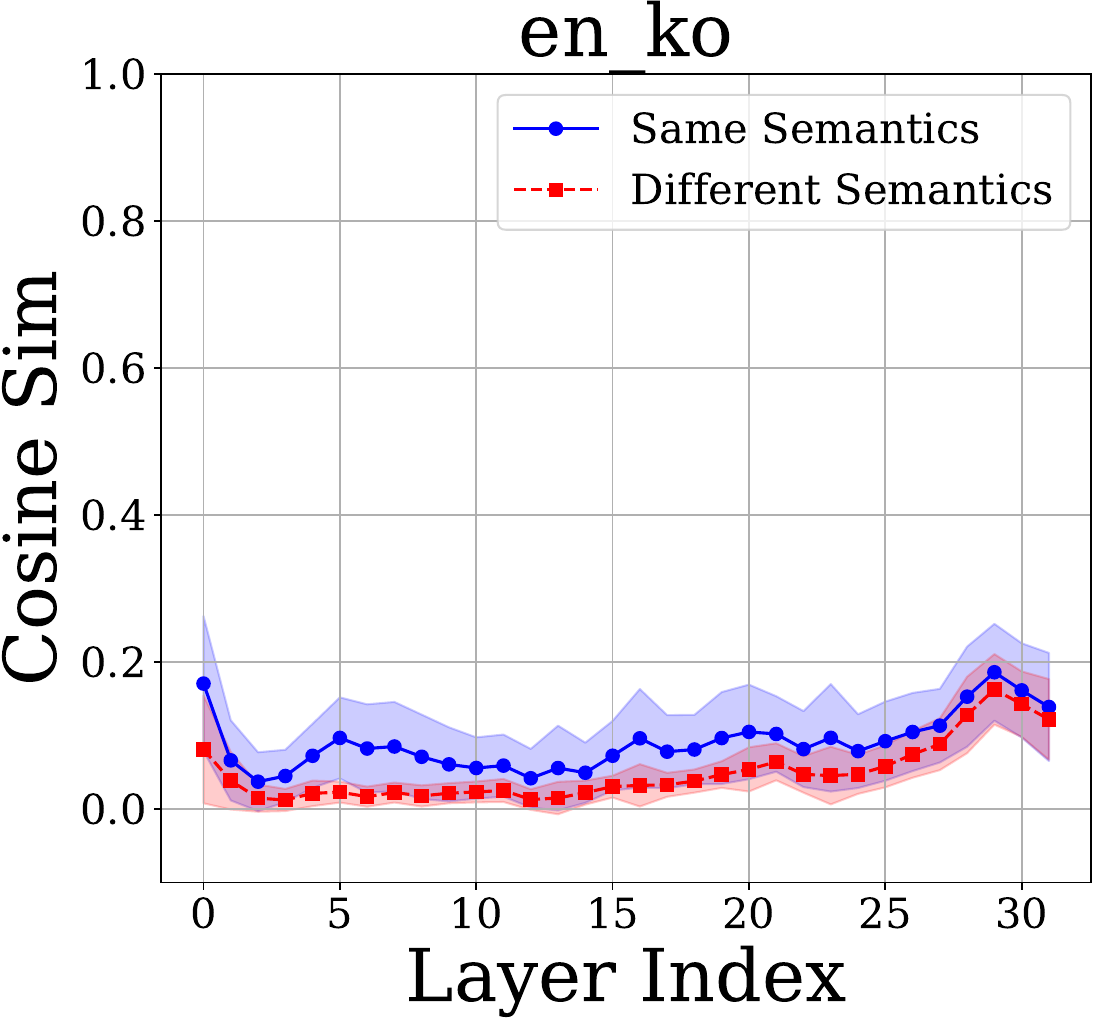}
      \subcaption{en-ko}
    \end{minipage}
    \begin{minipage}{0.20\linewidth}
      \centering
      \includegraphics[width=\linewidth]{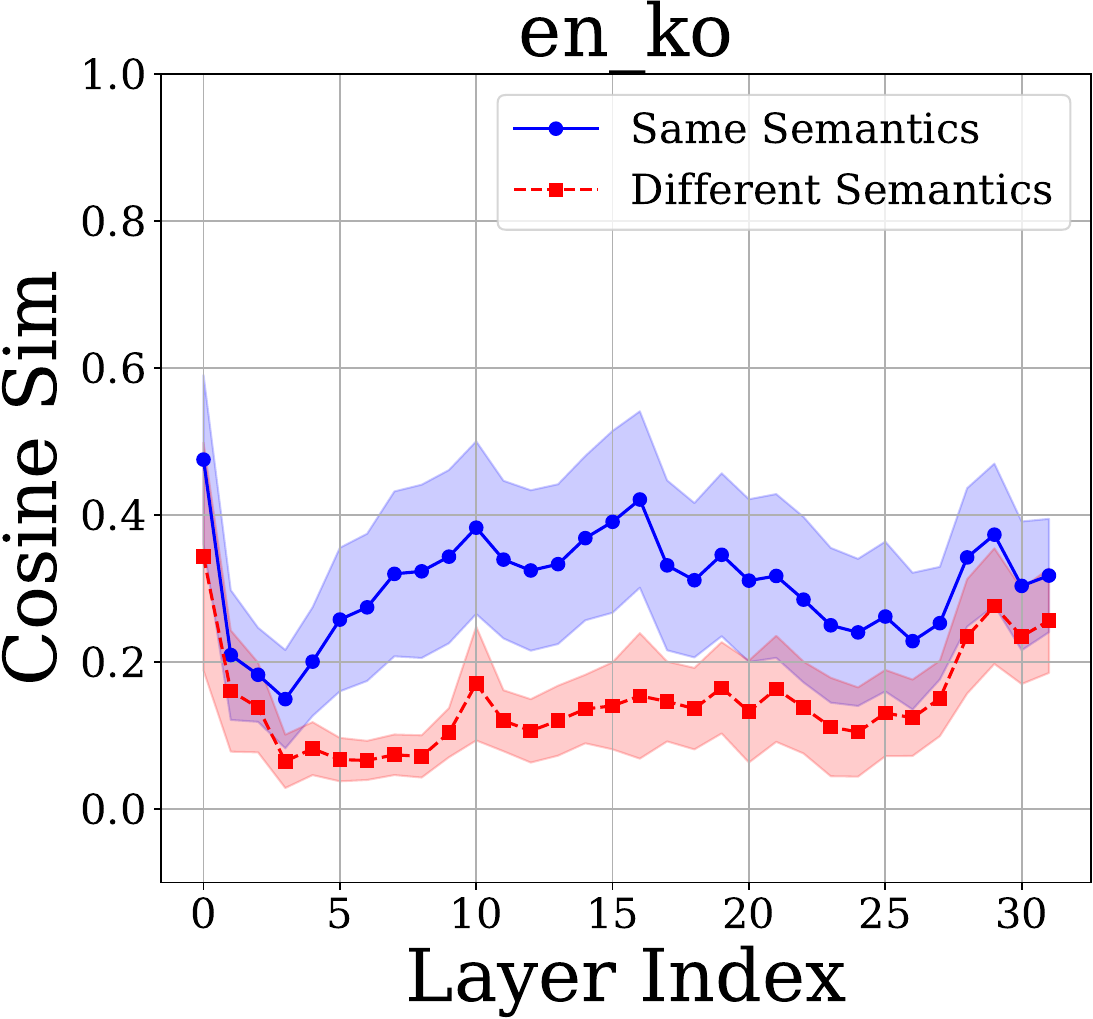}
      \subcaption{en-ko (baseline)}
    \end{minipage}
    \begin{minipage}{0.20\linewidth}
      \centering
      \includegraphics[width=\linewidth]{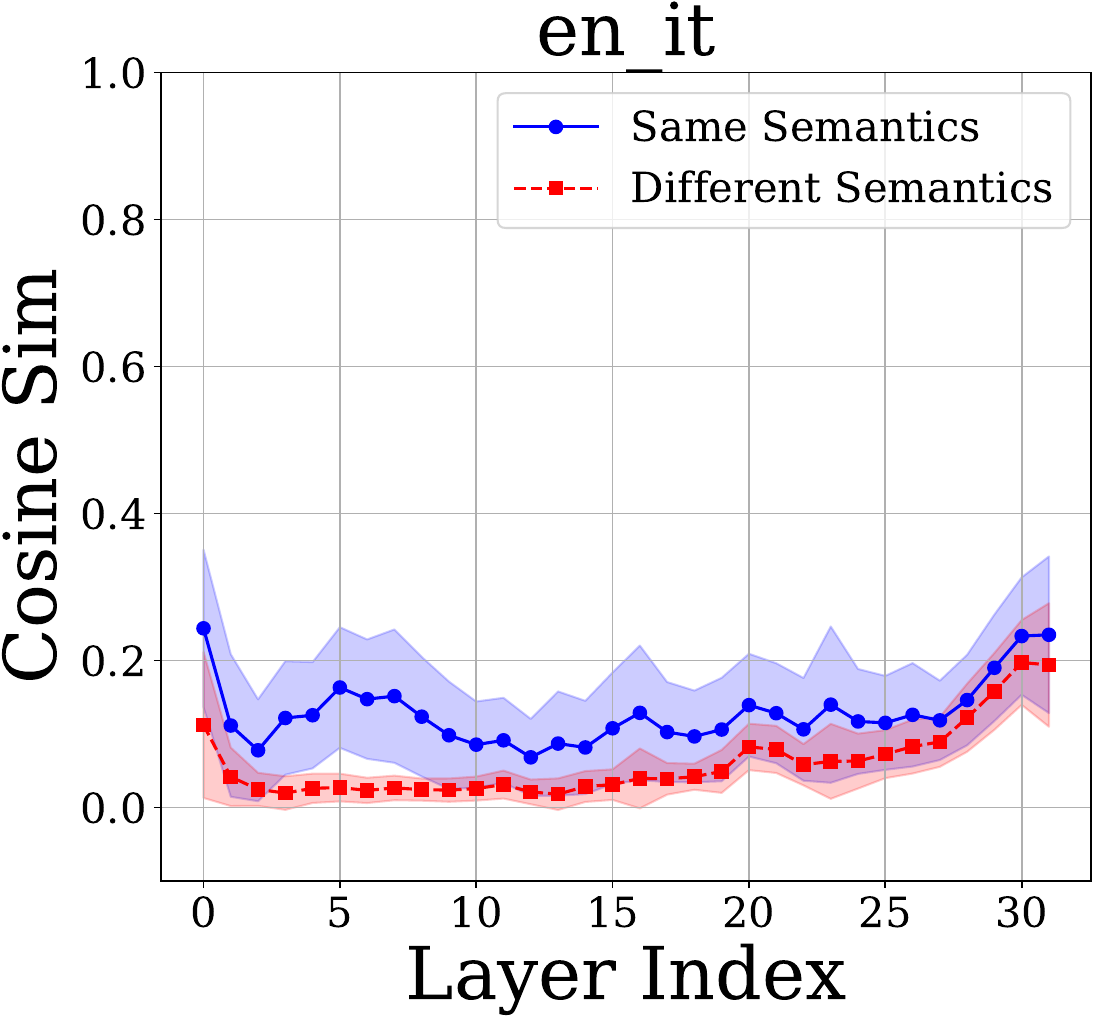}
      \subcaption{en-it}
    \end{minipage}
    \begin{minipage}{0.20\linewidth}
      \centering
      \includegraphics[width=\linewidth]{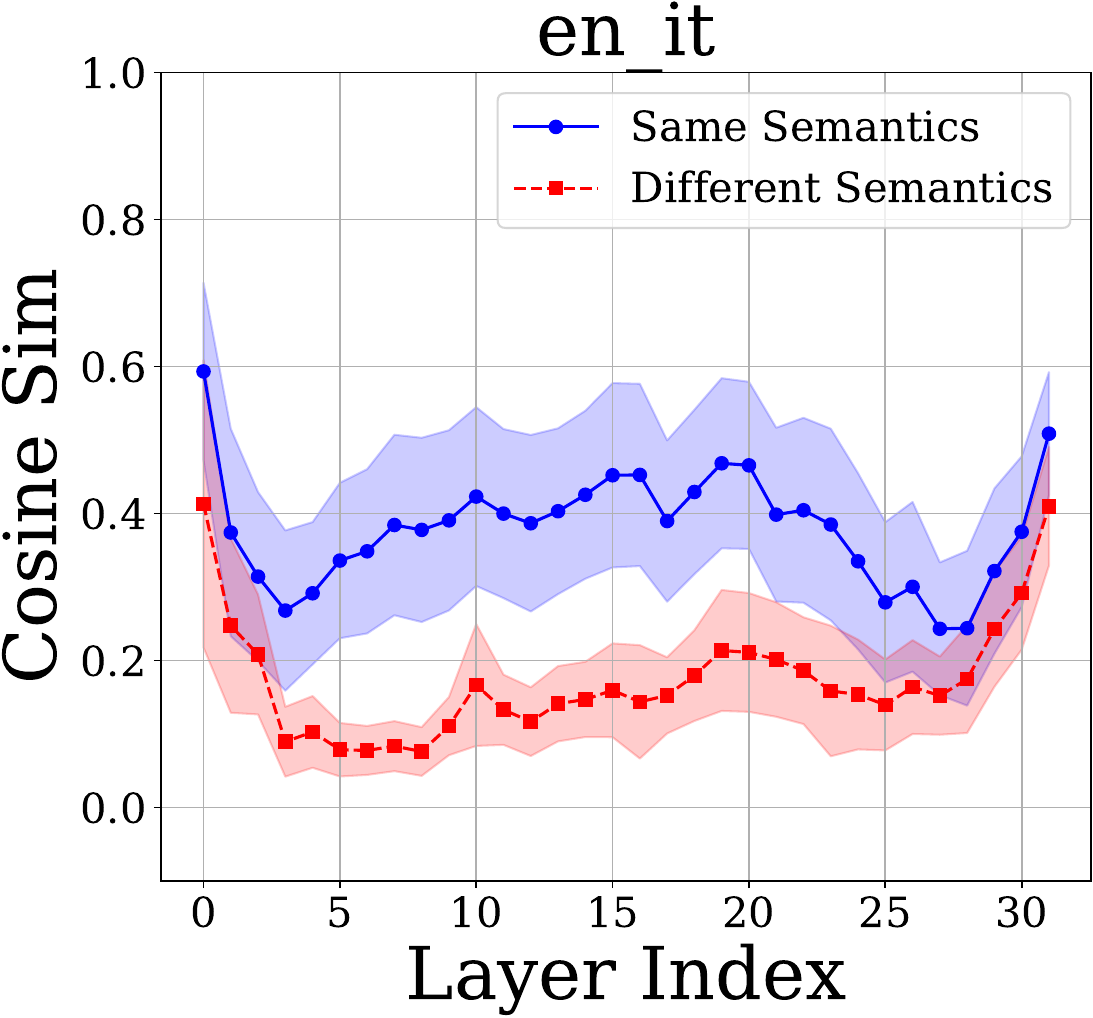}
      
      \subcaption{en-it (baseline)}
    \end{minipage}
    
      \begin{minipage}{\linewidth}
        \centering
        \small \textbf{(b) top-5000 (representing 1\% of all neurons)}
      \end{minipage}

  \caption{\textbf{Similarity of activation patterns across layers while deactivating Type-1 Transfer Neurons (Mistral-7B).}}
  \label{fig:appendix:act_sim_mistral_deactivating_top-1k_Type-1}
\end{figure*}

\begin{figure*}[t]
    \centering

    \begin{minipage}{0.20\linewidth}
      \centering
      \includegraphics[width=\linewidth]{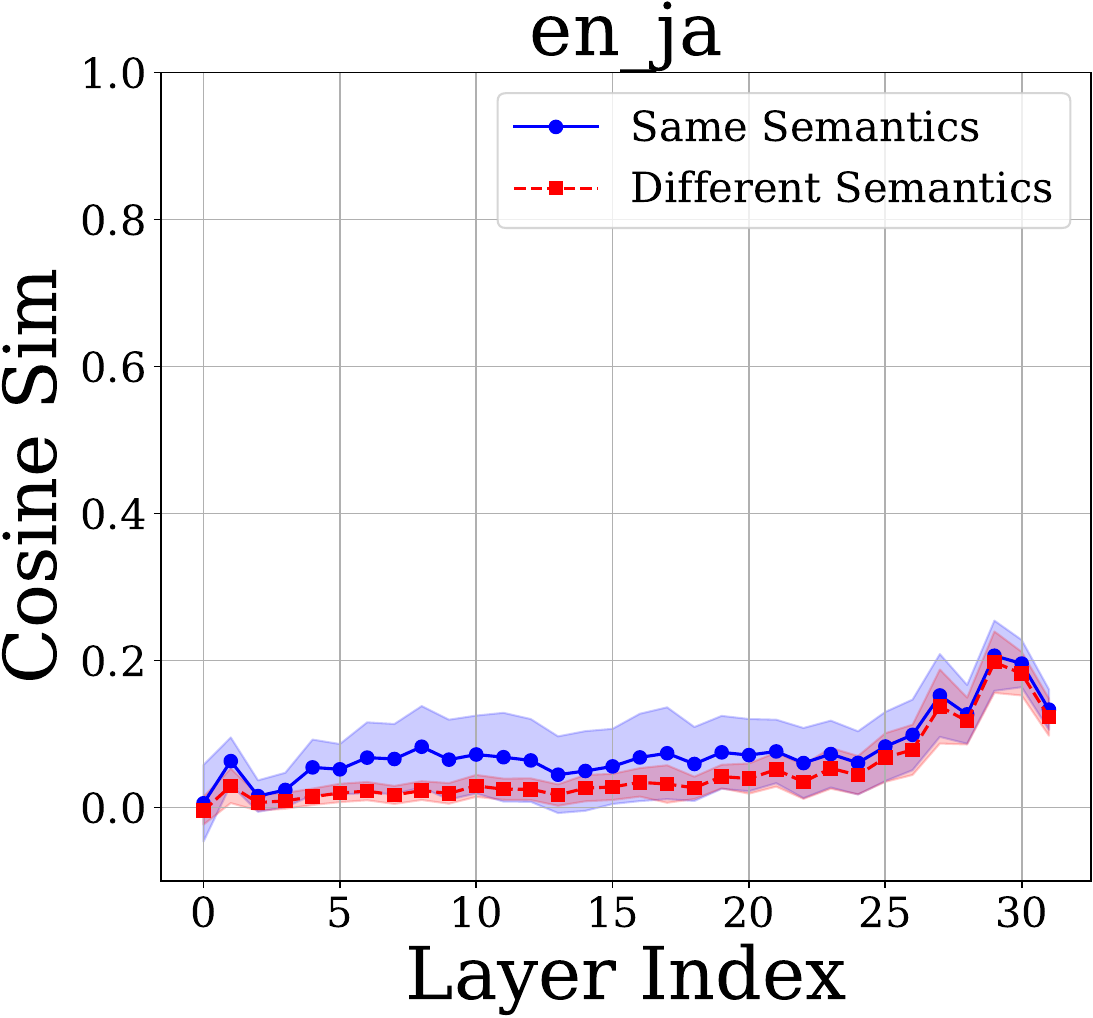}
      \subcaption{en-ja}
    \end{minipage}
    \begin{minipage}{0.20\linewidth}
      \centering
      \includegraphics[width=\linewidth]{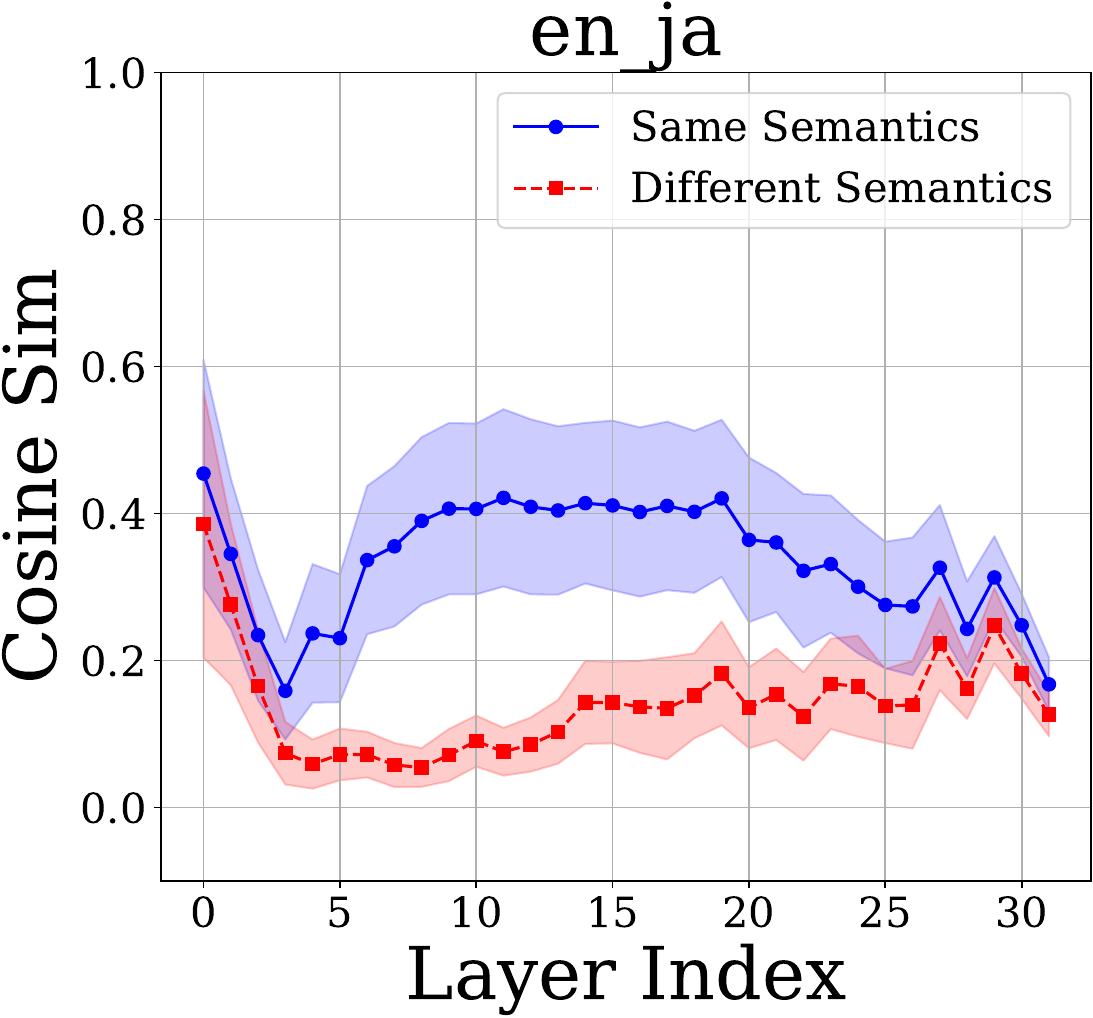}
      \subcaption{en-ja (baseline)}
    \end{minipage}
    \begin{minipage}{0.20\linewidth}
      \centering
      \includegraphics[width=\linewidth]{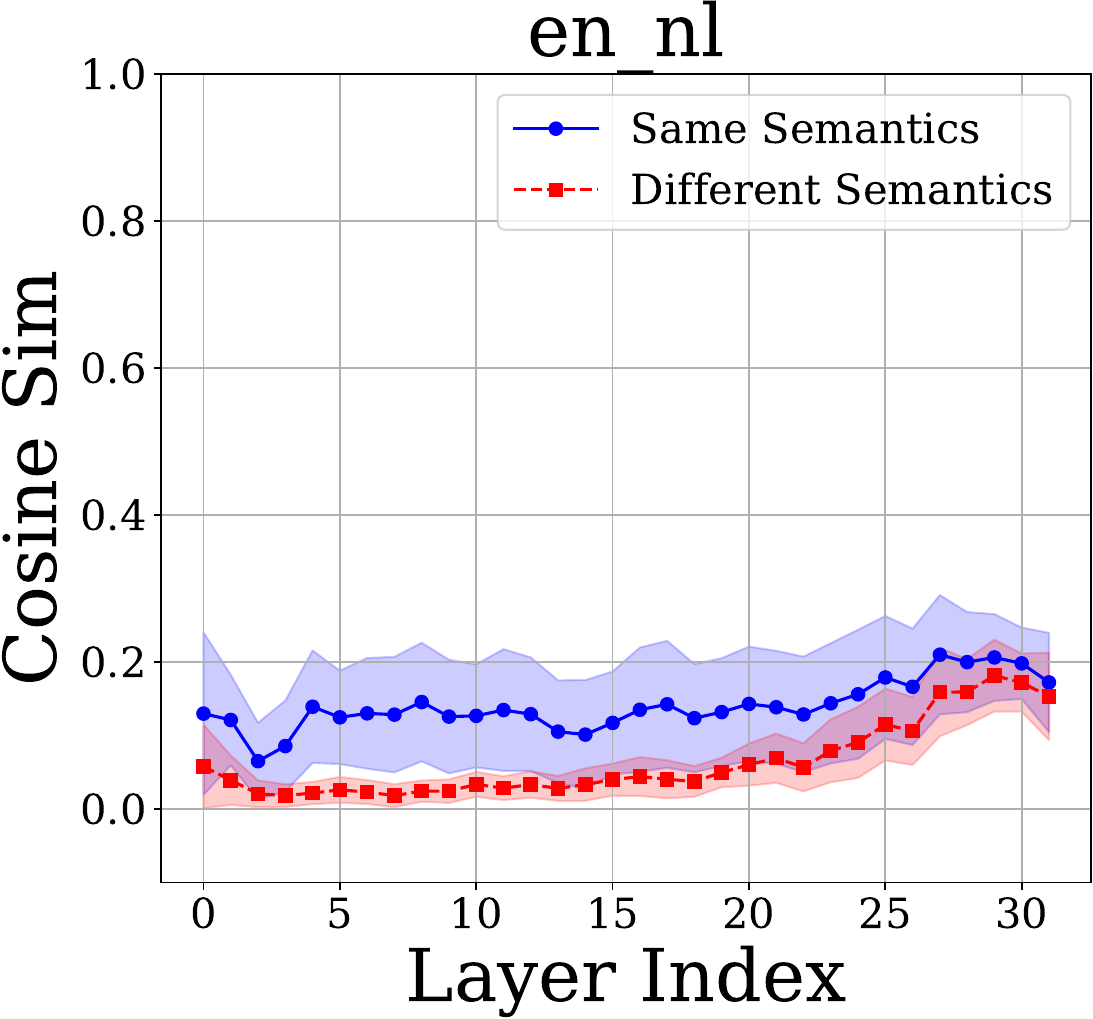}
      \subcaption{en-nl}
    \end{minipage}
    \begin{minipage}{0.20\linewidth}
      \centering
      \includegraphics[width=\linewidth]{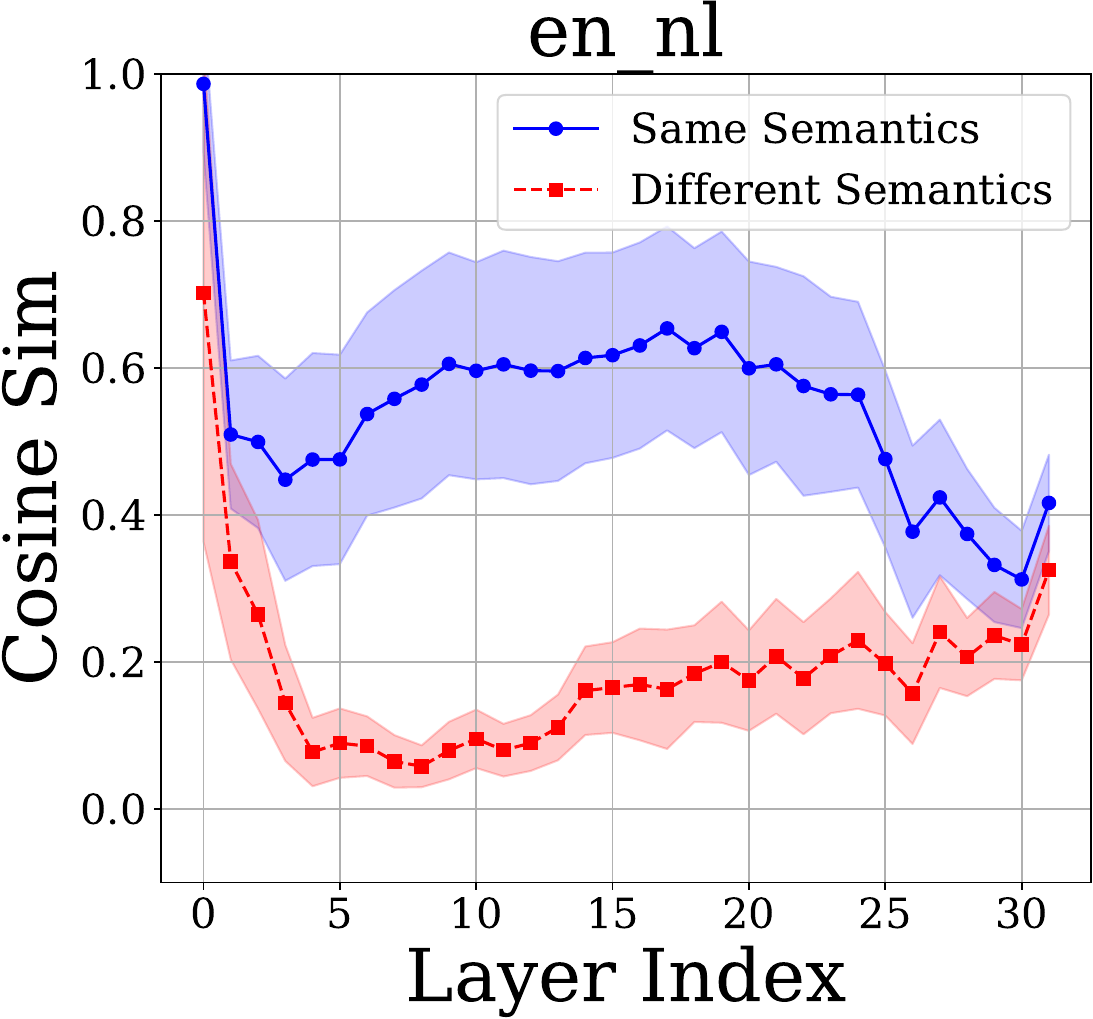}
      \subcaption{en-nl (baseline)}
    \end{minipage}

    \begin{minipage}{0.20\linewidth}
      \centering
      \includegraphics[width=\linewidth]{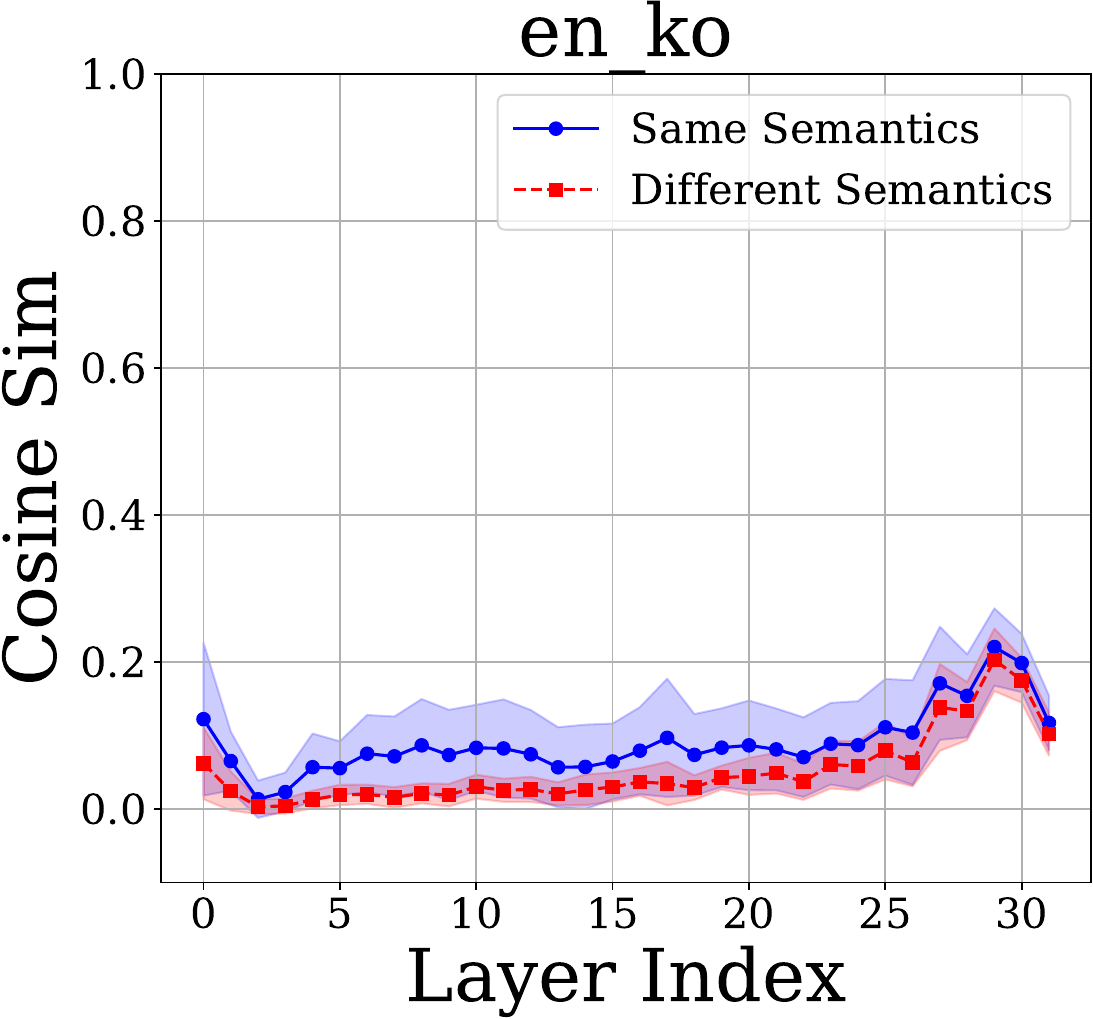}
      \subcaption{en-ko}
    \end{minipage}
    \begin{minipage}{0.20\linewidth}
      \centering
      \includegraphics[width=\linewidth]{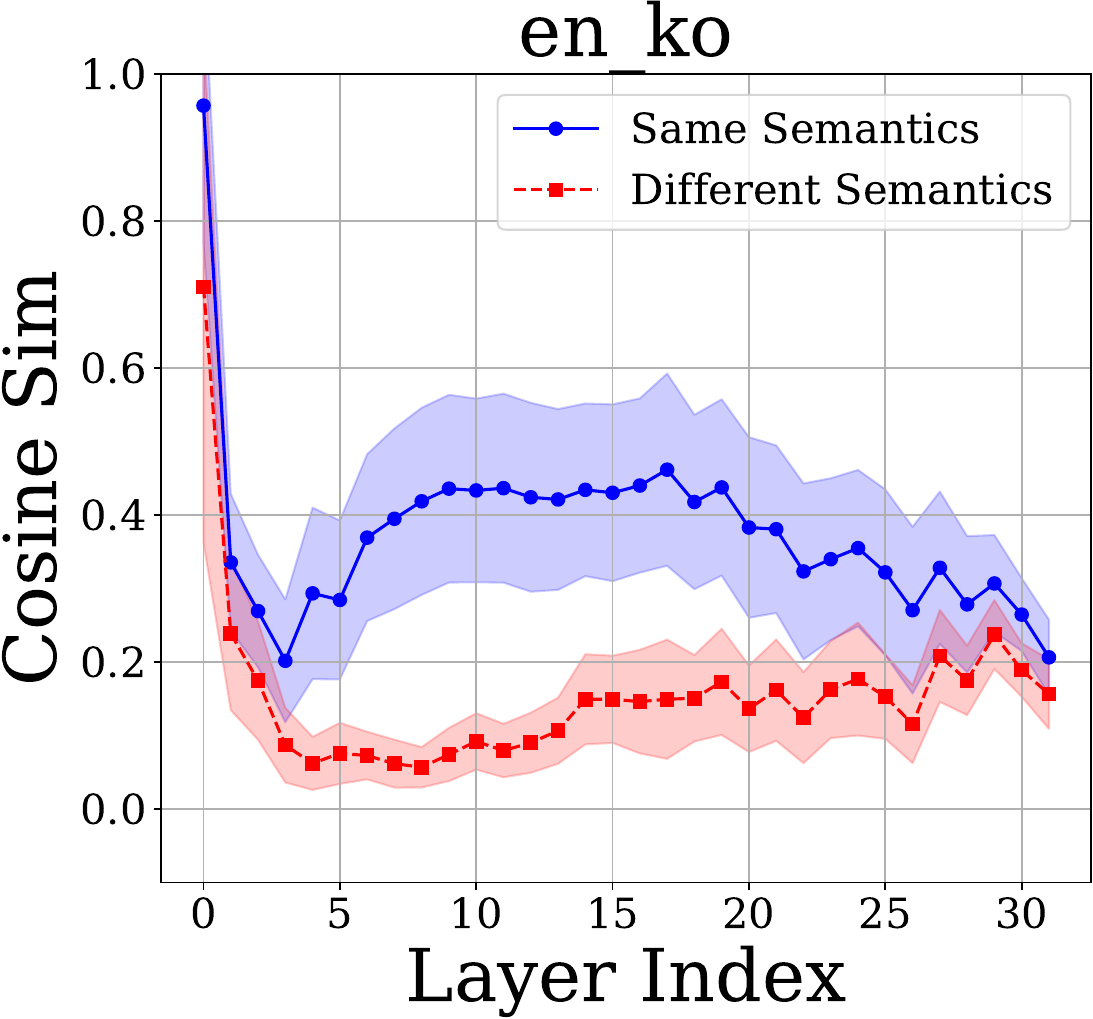}
      \subcaption{en-ko (baseline)}
    \end{minipage}
    \begin{minipage}{0.20\linewidth}
      \centering
      \includegraphics[width=\linewidth]{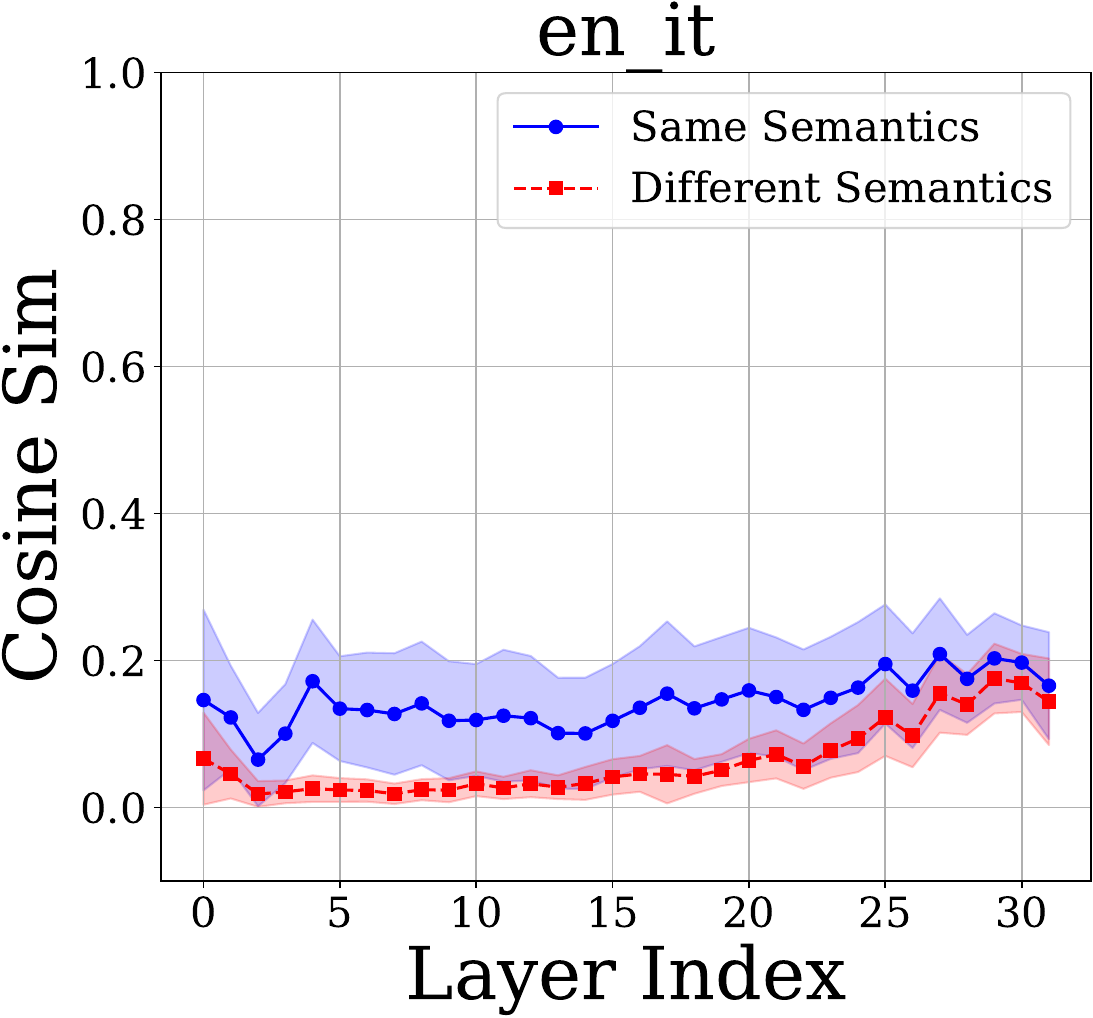}
      \subcaption{en-it}
    \end{minipage}
    \begin{minipage}{0.20\linewidth}
      \centering
      \includegraphics[width=\linewidth]{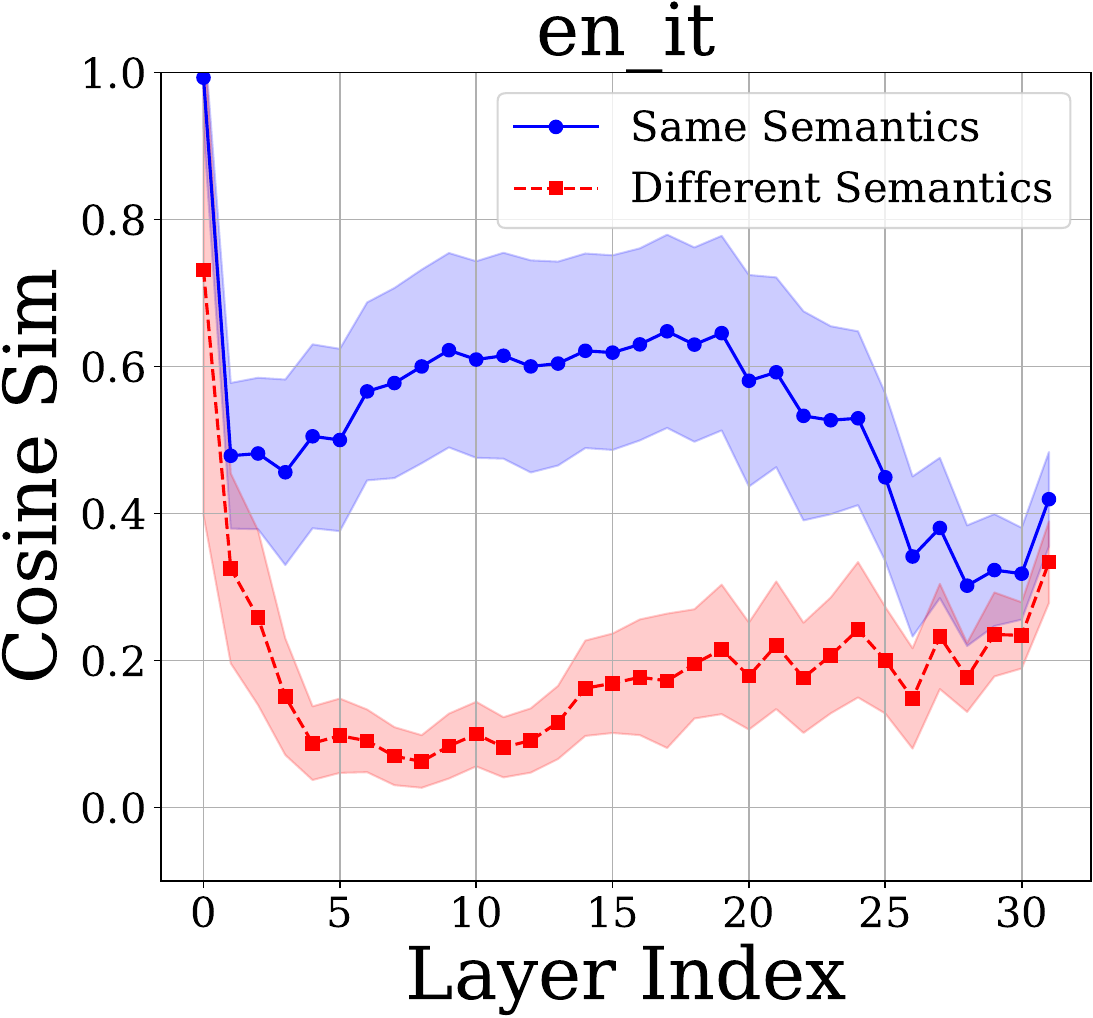}
      \subcaption{en-it (baseline)}
    \end{minipage}
    
      % First row label
      \begin{minipage}{\linewidth}
        \centering
        \small \textbf{(a) top-1000 (representing 0.2\% of all neurons)}
      \end{minipage}

    % second row
    \begin{minipage}{0.20\linewidth}
      \centering
      \includegraphics[width=\linewidth]{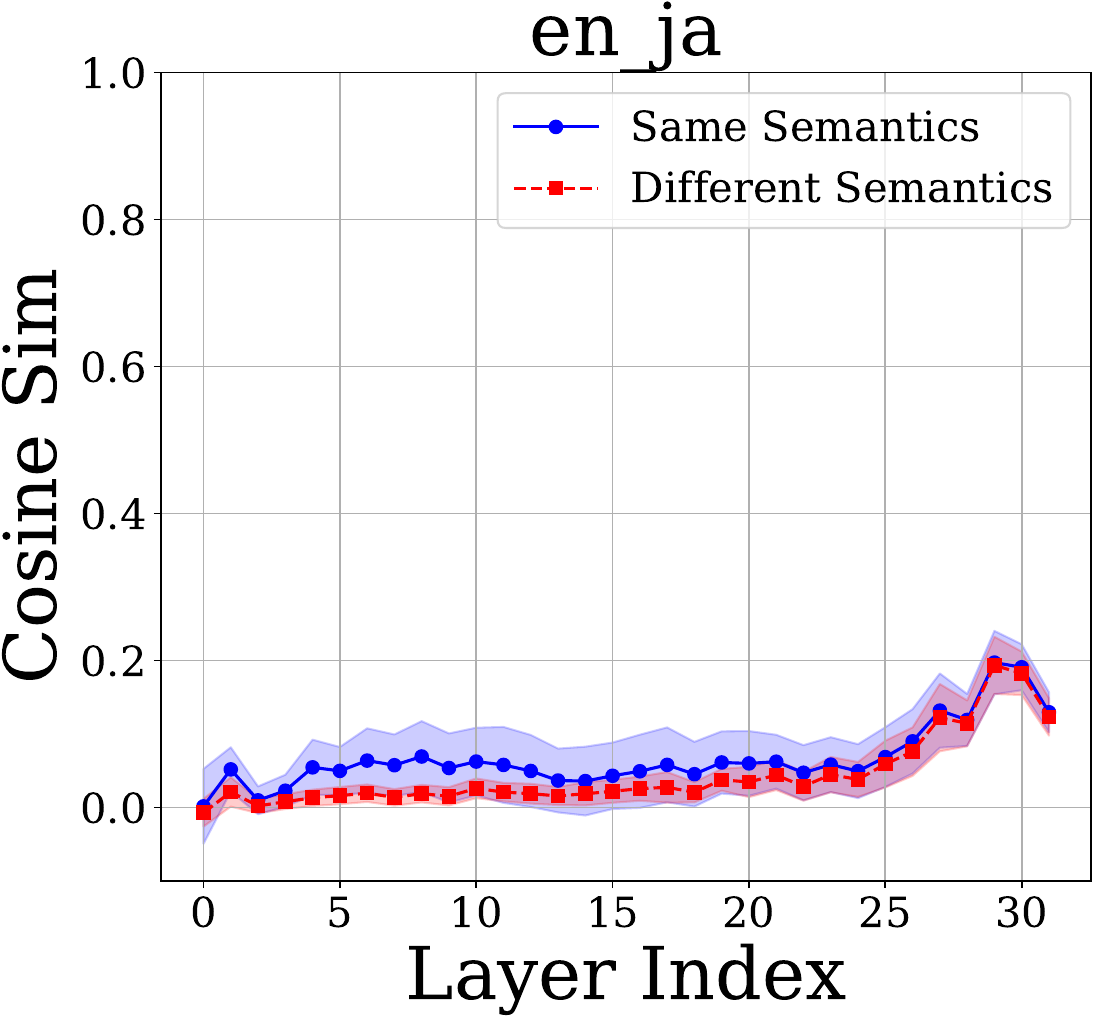}
      \subcaption{en-ja}
    \end{minipage}
    \begin{minipage}{0.20\linewidth}
      \centering
      \includegraphics[width=\linewidth]{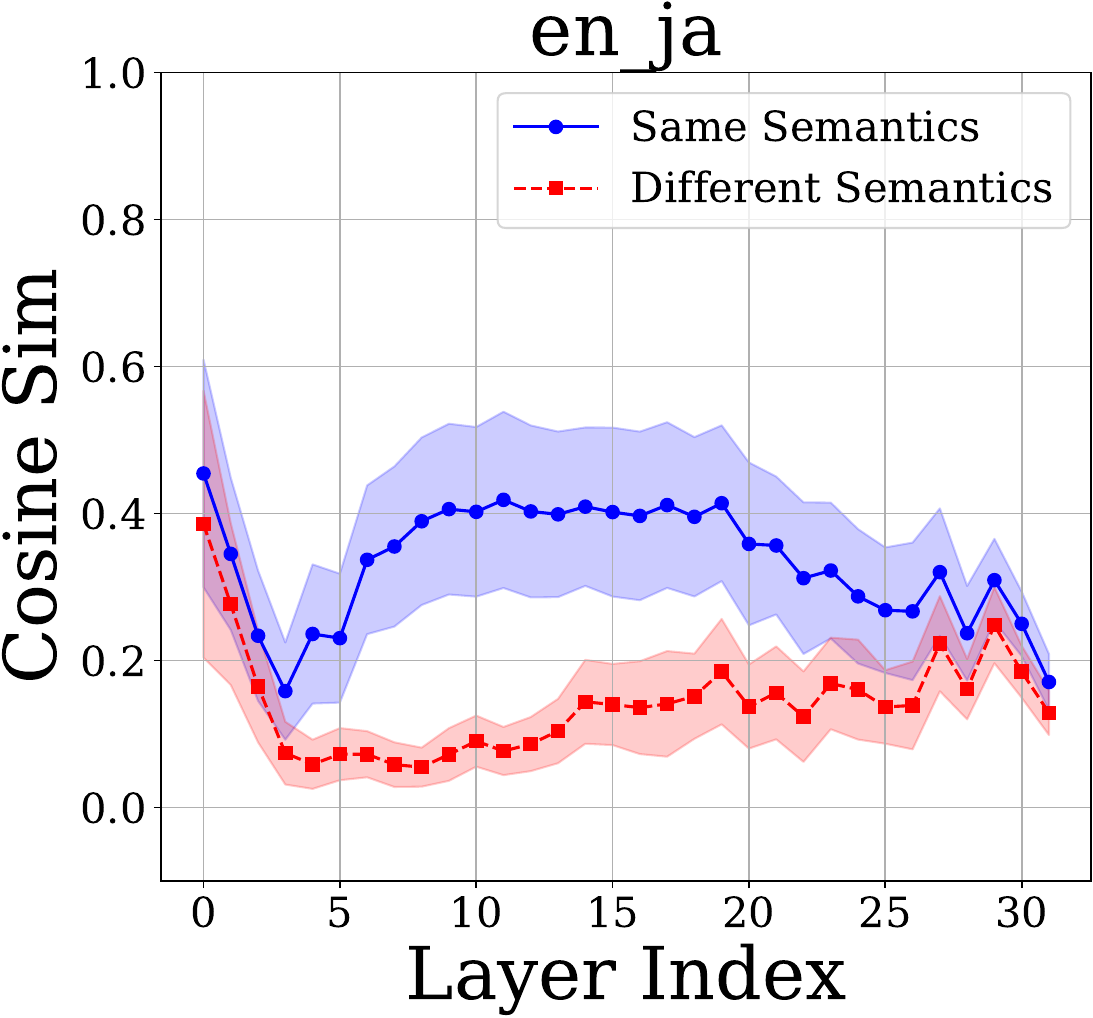}
      \subcaption{en-ja (baseline)}
    \end{minipage}
    \begin{minipage}{0.20\linewidth}
      \centering
      \includegraphics[width=\linewidth]{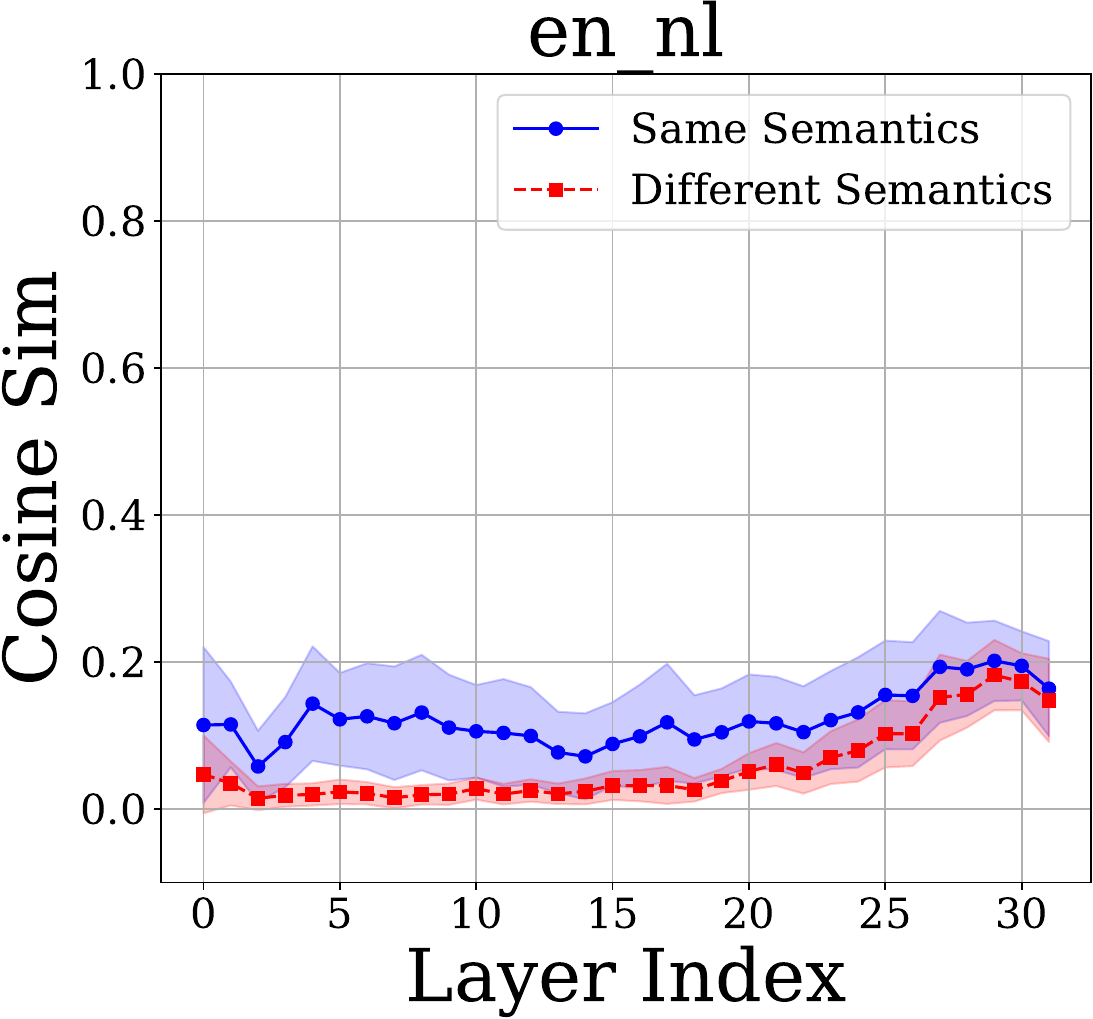}
      \subcaption{en-nl}
    \end{minipage}
    \begin{minipage}{0.20\linewidth}
      \centering
      \includegraphics[width=\linewidth]{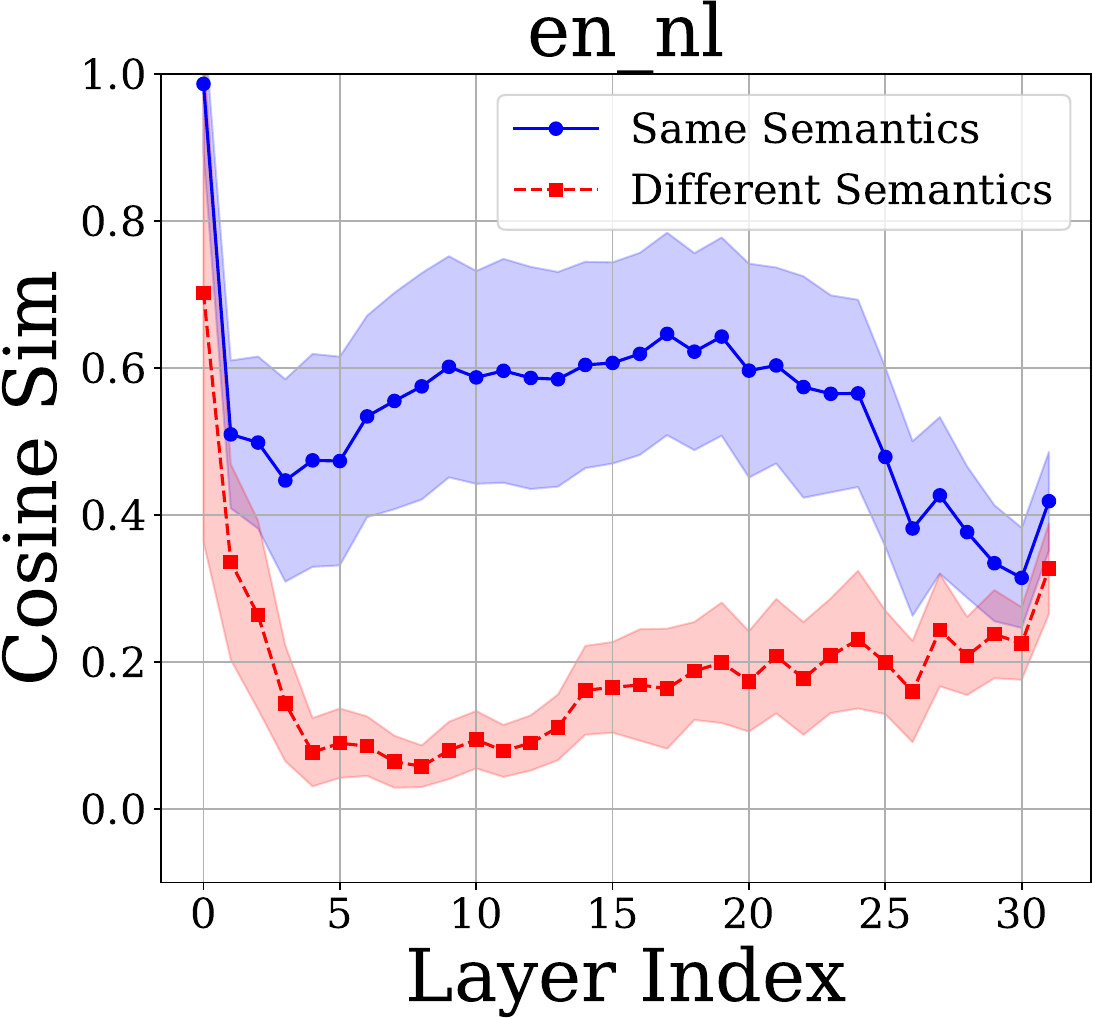}
      \subcaption{en-nl (baseline)}
    \end{minipage}

    \begin{minipage}{0.20\linewidth}
      \centering
      \includegraphics[width=\linewidth]{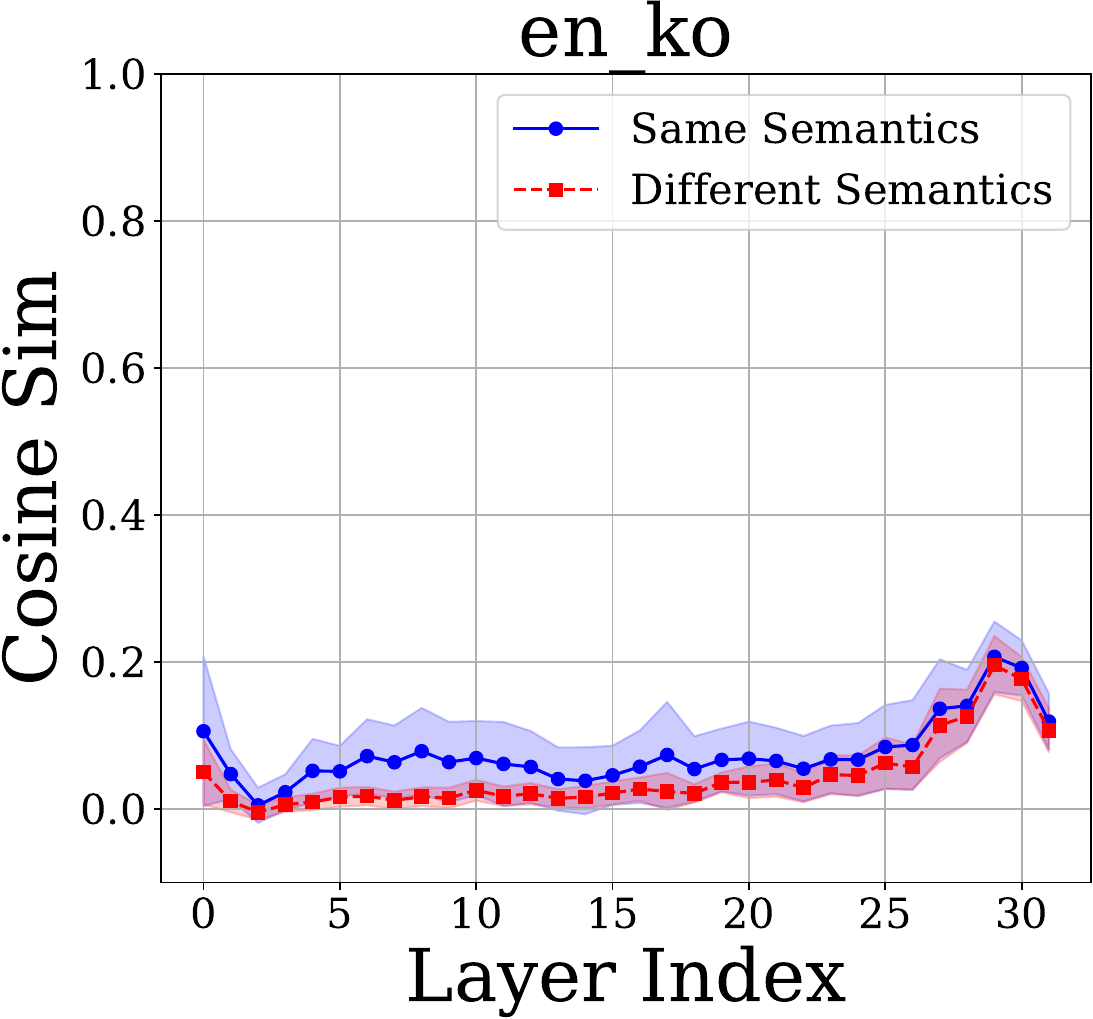}
      \subcaption{en-ko}
    \end{minipage}
    \begin{minipage}{0.20\linewidth}
      \centering
      \includegraphics[width=\linewidth]{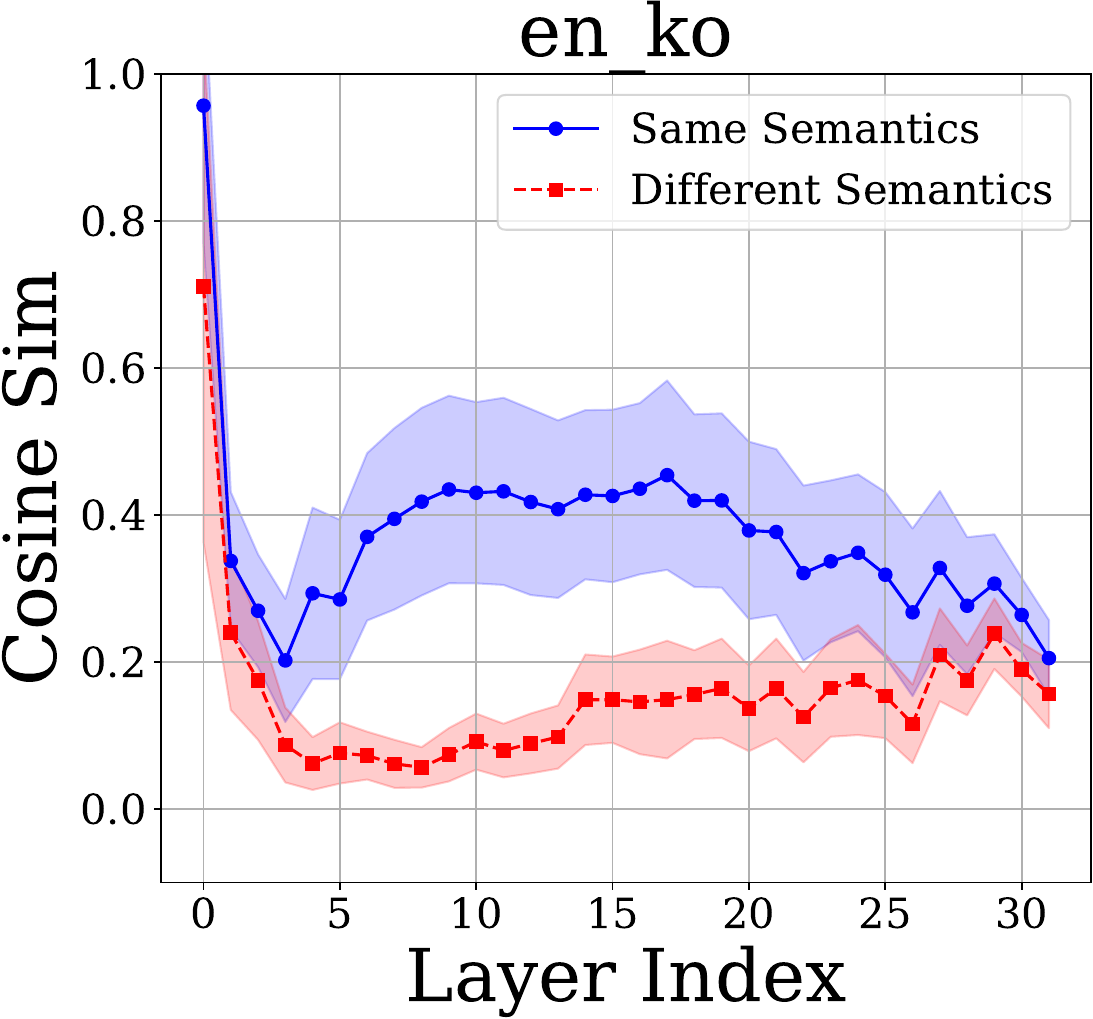}
      \subcaption{en-ko (baseline)}
    \end{minipage}
    \begin{minipage}{0.20\linewidth}
      \centering
      \includegraphics[width=\linewidth]{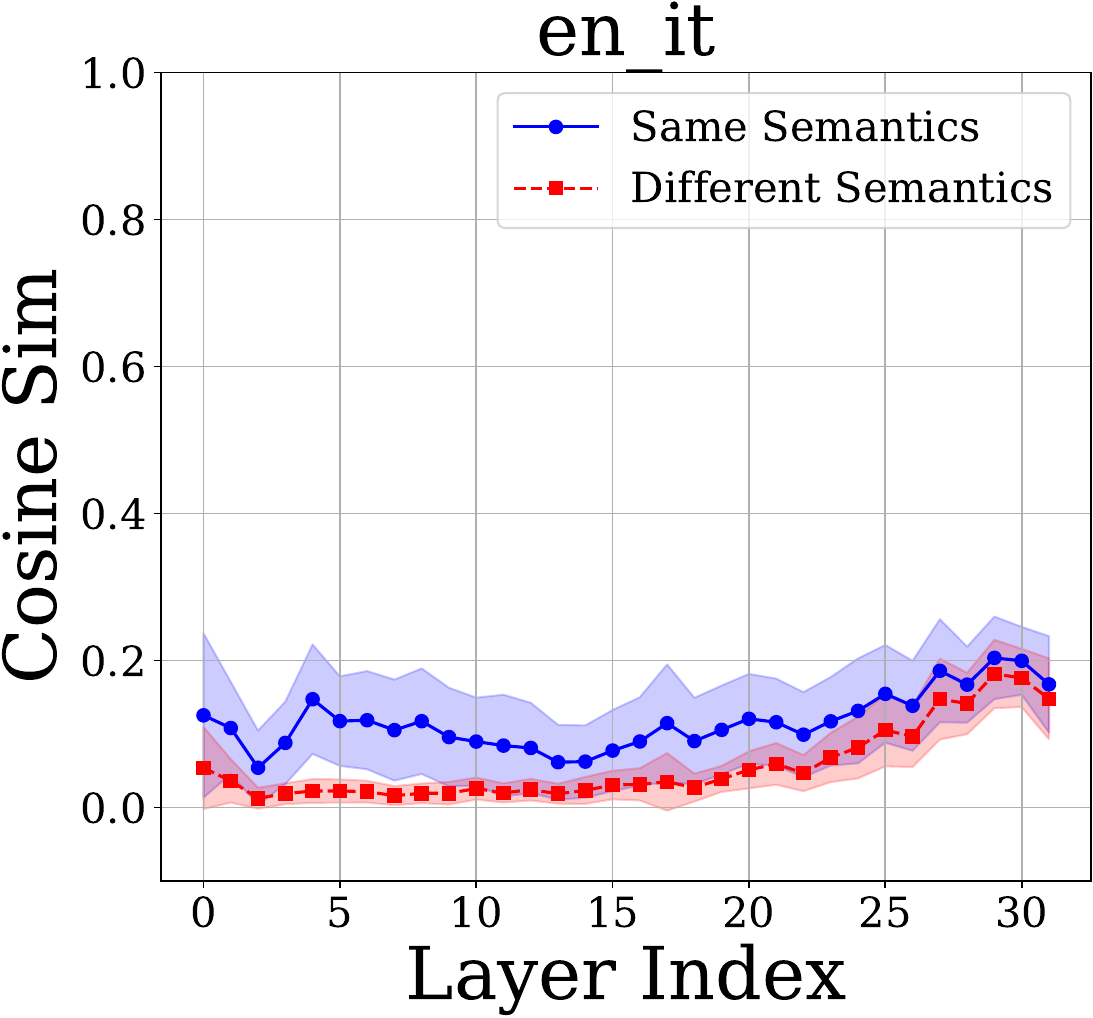}
      \subcaption{en-it}
    \end{minipage}
    \begin{minipage}{0.20\linewidth}
      \centering
      \includegraphics[width=\linewidth]{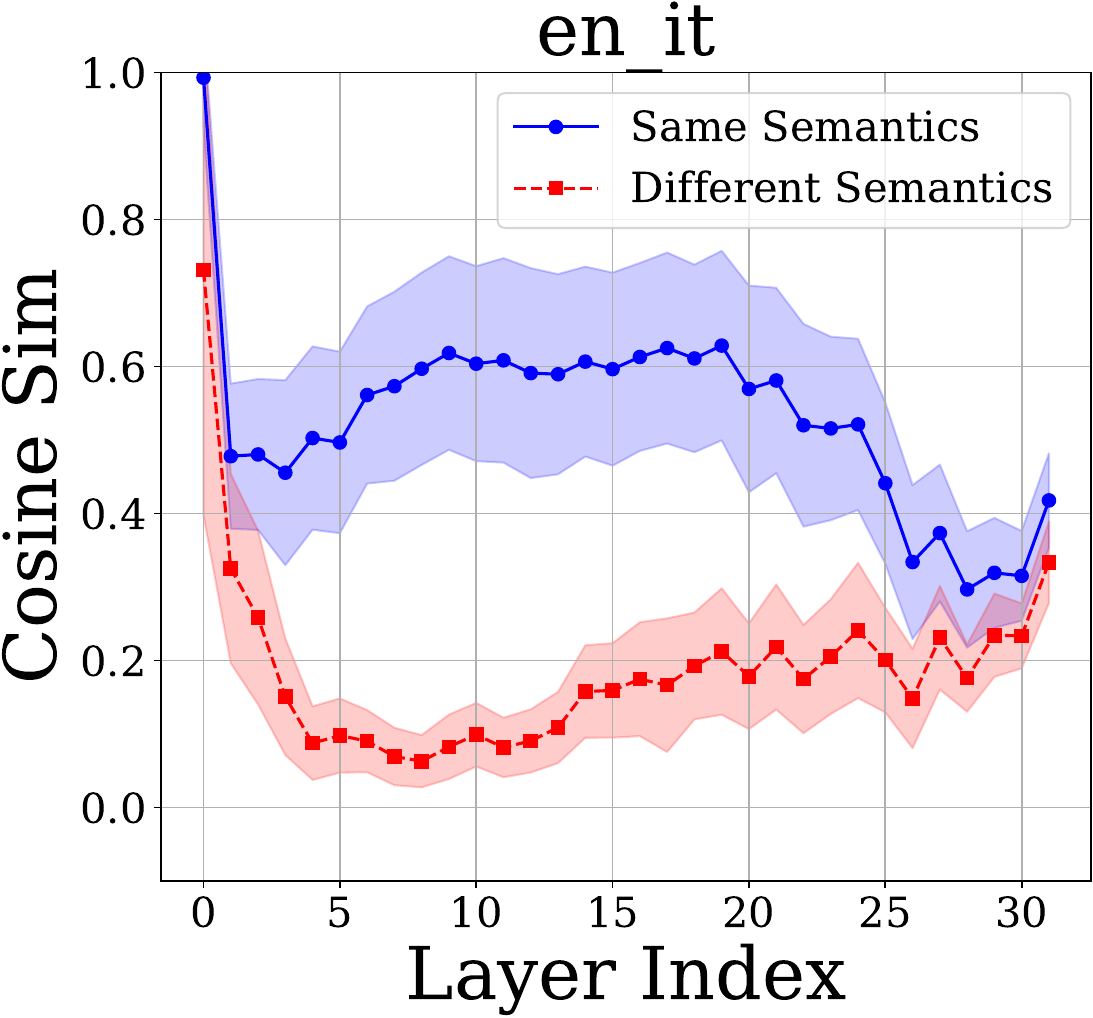}
      \subcaption{en-it (baseline)}
    \end{minipage}
    
      \begin{minipage}{\linewidth}
        \centering
        \small \textbf{(b) top-3000 (representing 0.6\% of all neurons)}
      \end{minipage}

    \begin{minipage}{0.20\linewidth}
      \centering
      \includegraphics[width=\linewidth]{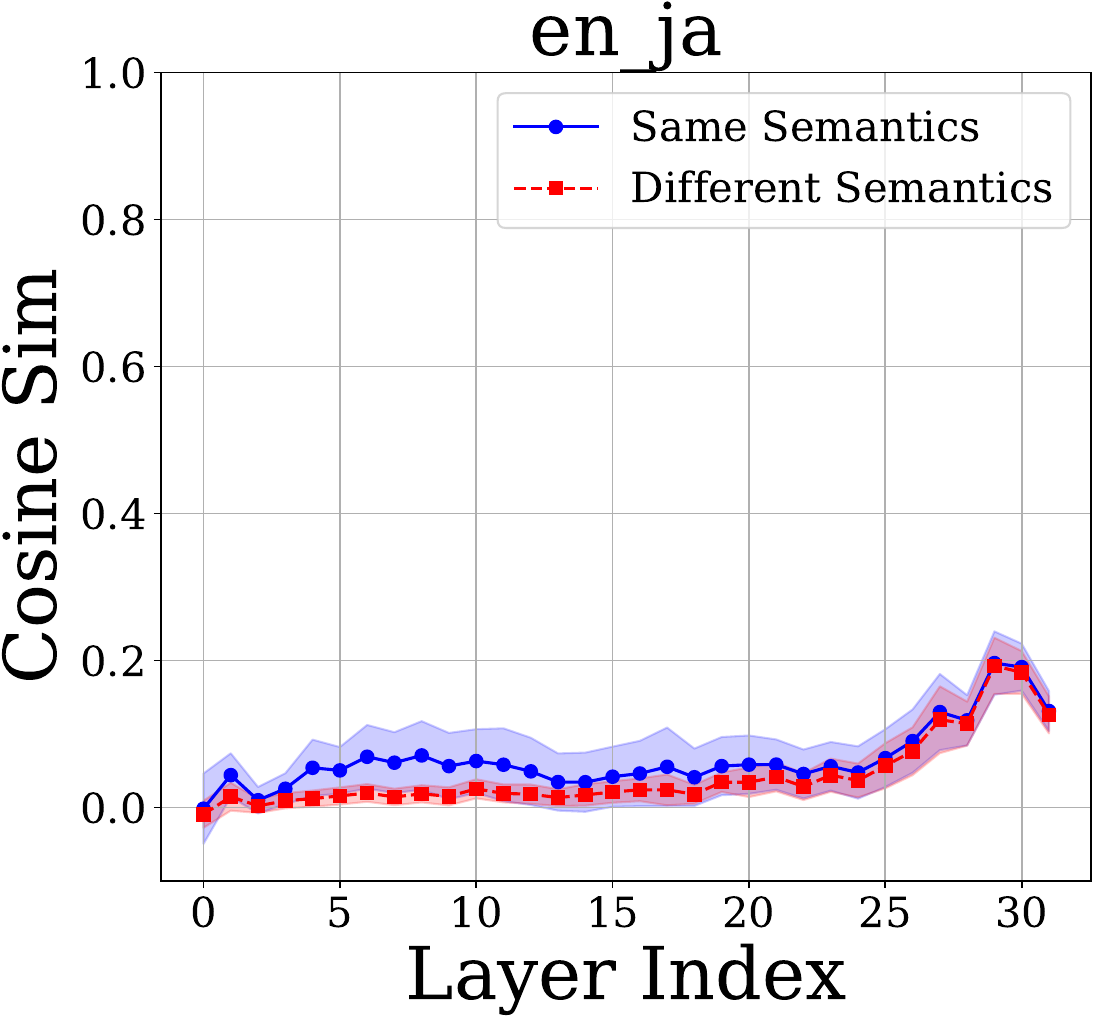}
      \subcaption{en-ja}
    \end{minipage}
    \begin{minipage}{0.20\linewidth}
      \centering
      \includegraphics[width=\linewidth]{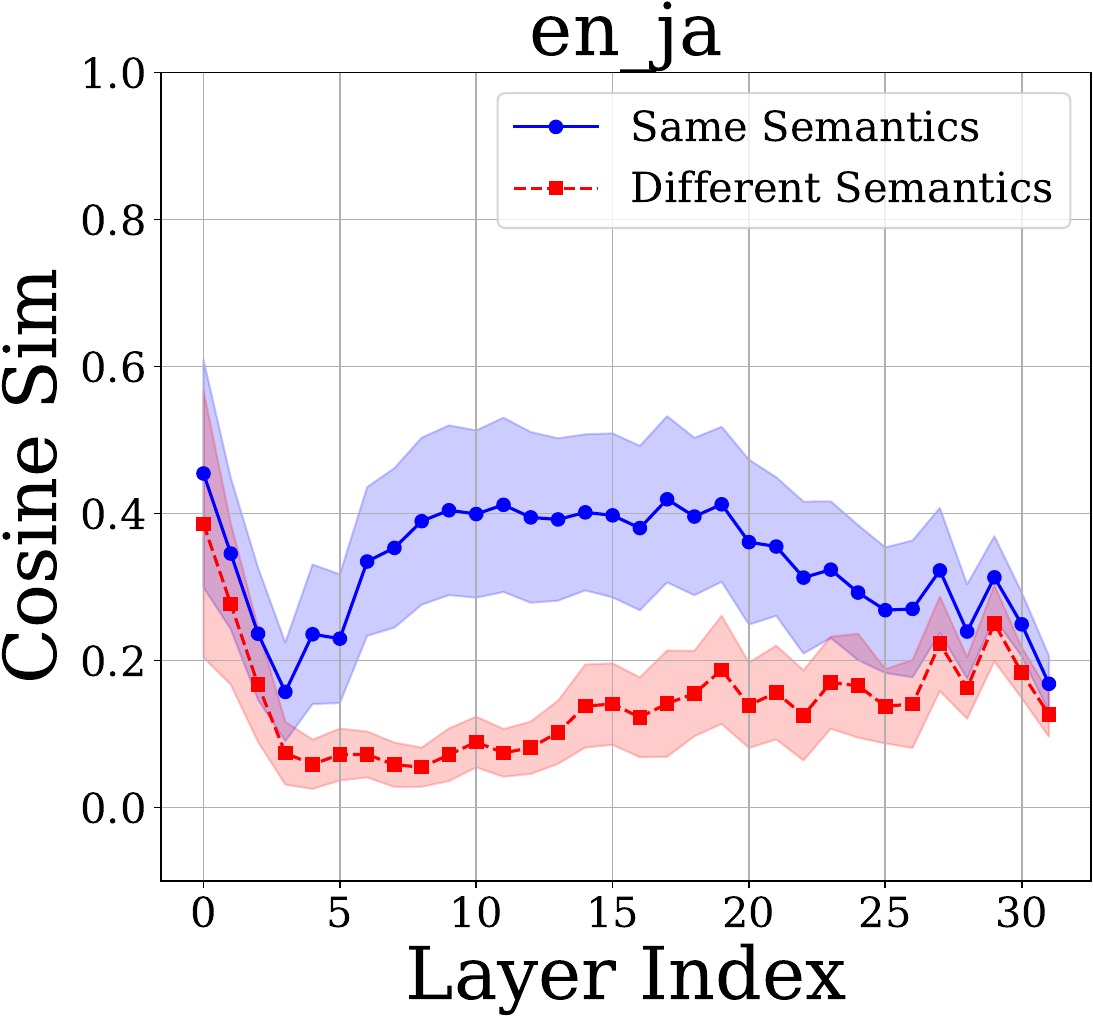}
      \subcaption{en-ja (baseline)}
    \end{minipage}
    \begin{minipage}{0.20\linewidth}
      \centering
      \includegraphics[width=\linewidth]{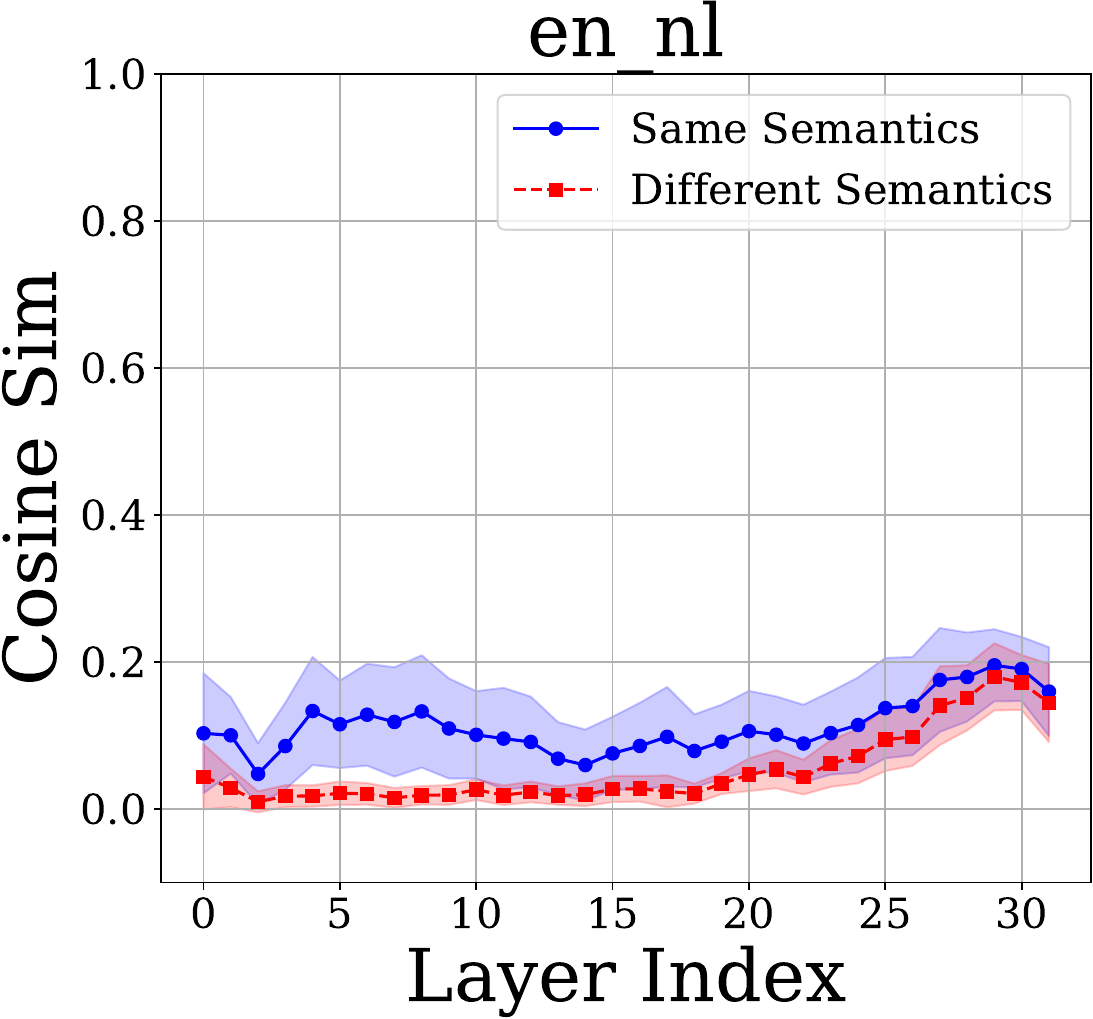}
      \subcaption{en-nl}
    \end{minipage}
    \begin{minipage}{0.20\linewidth}
      \centering
      \includegraphics[width=\linewidth]{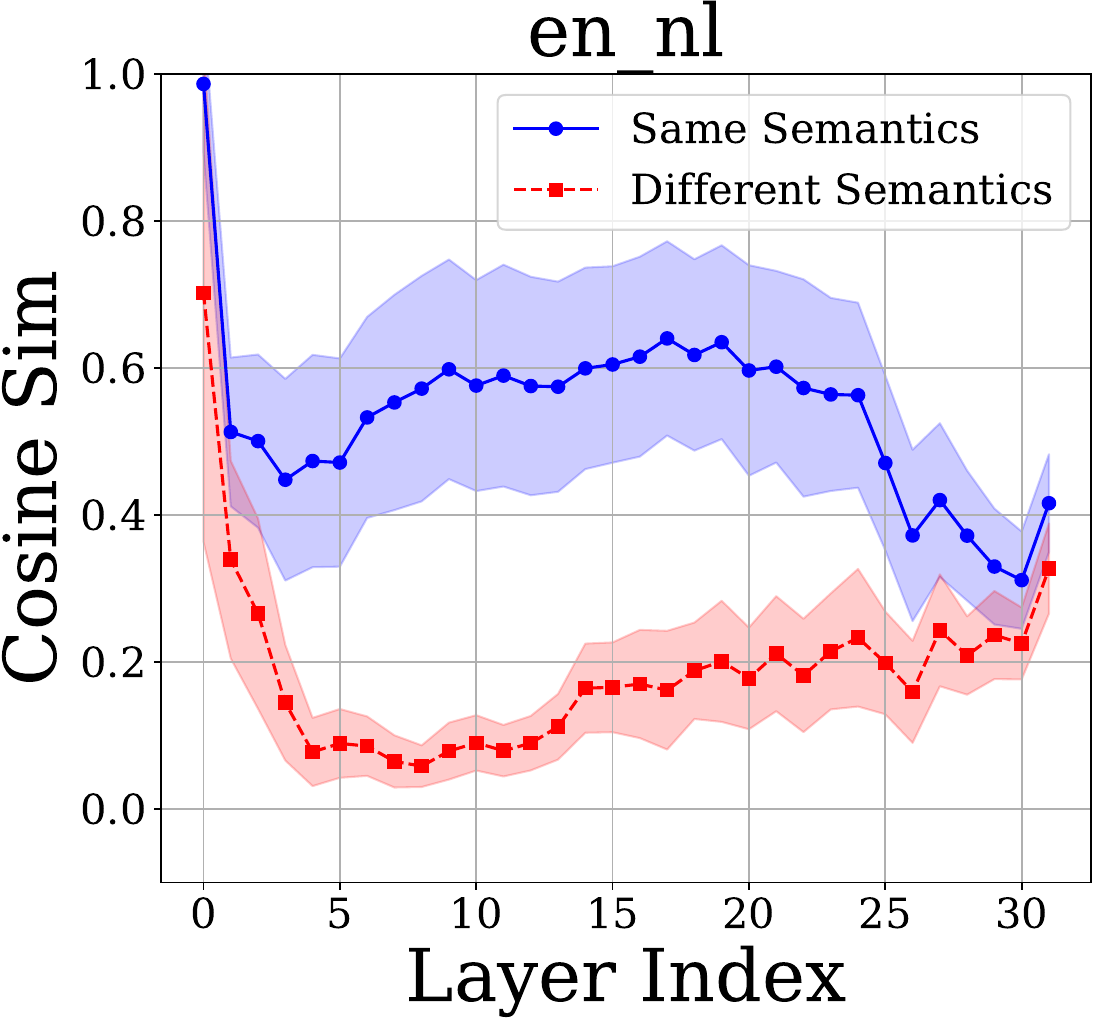}
      \subcaption{en-nl (baseline)}
    \end{minipage}

    \begin{minipage}{0.20\linewidth}
      \centering
      \includegraphics[width=\linewidth]{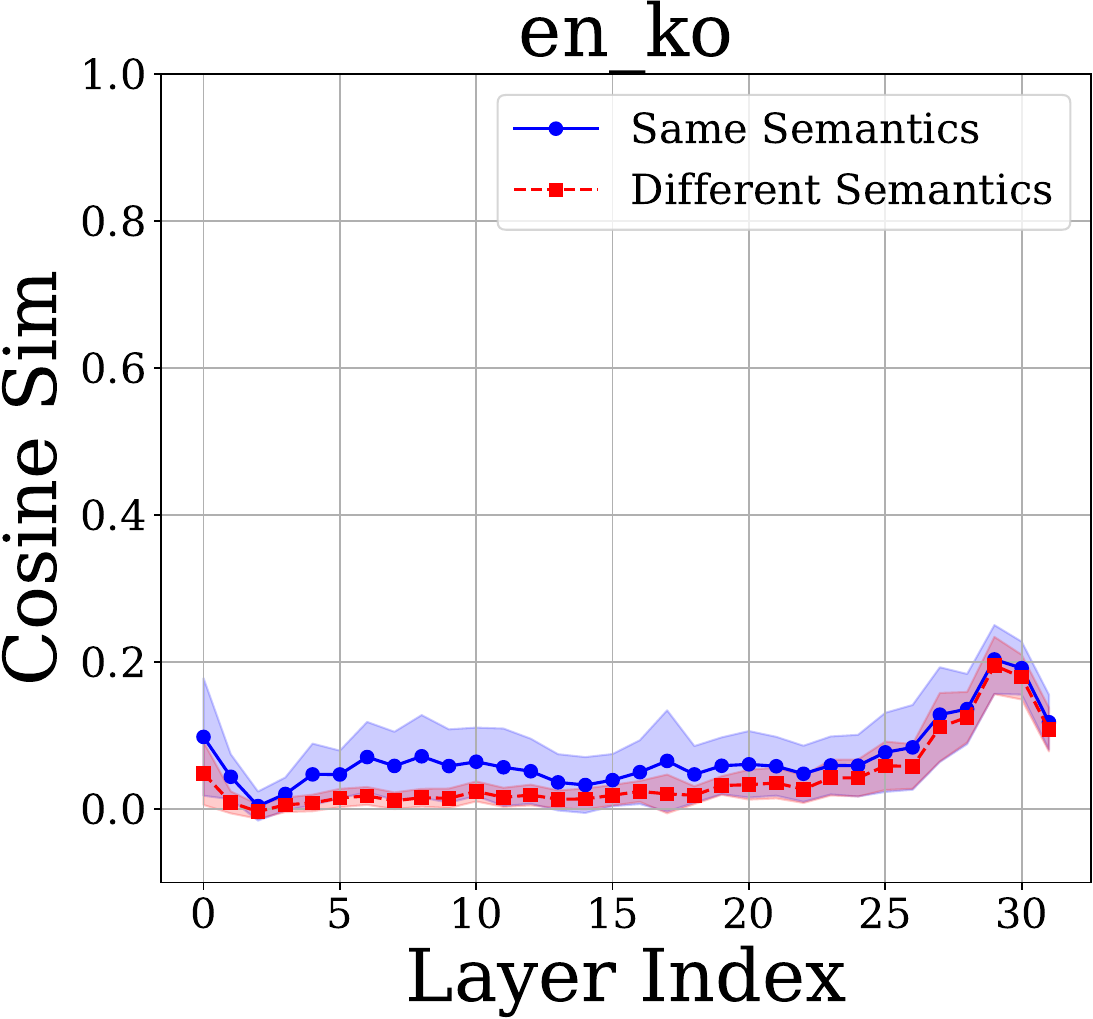}
      \subcaption{en-ko}
    \end{minipage}
    \begin{minipage}{0.20\linewidth}
      \centering
      \includegraphics[width=\linewidth]{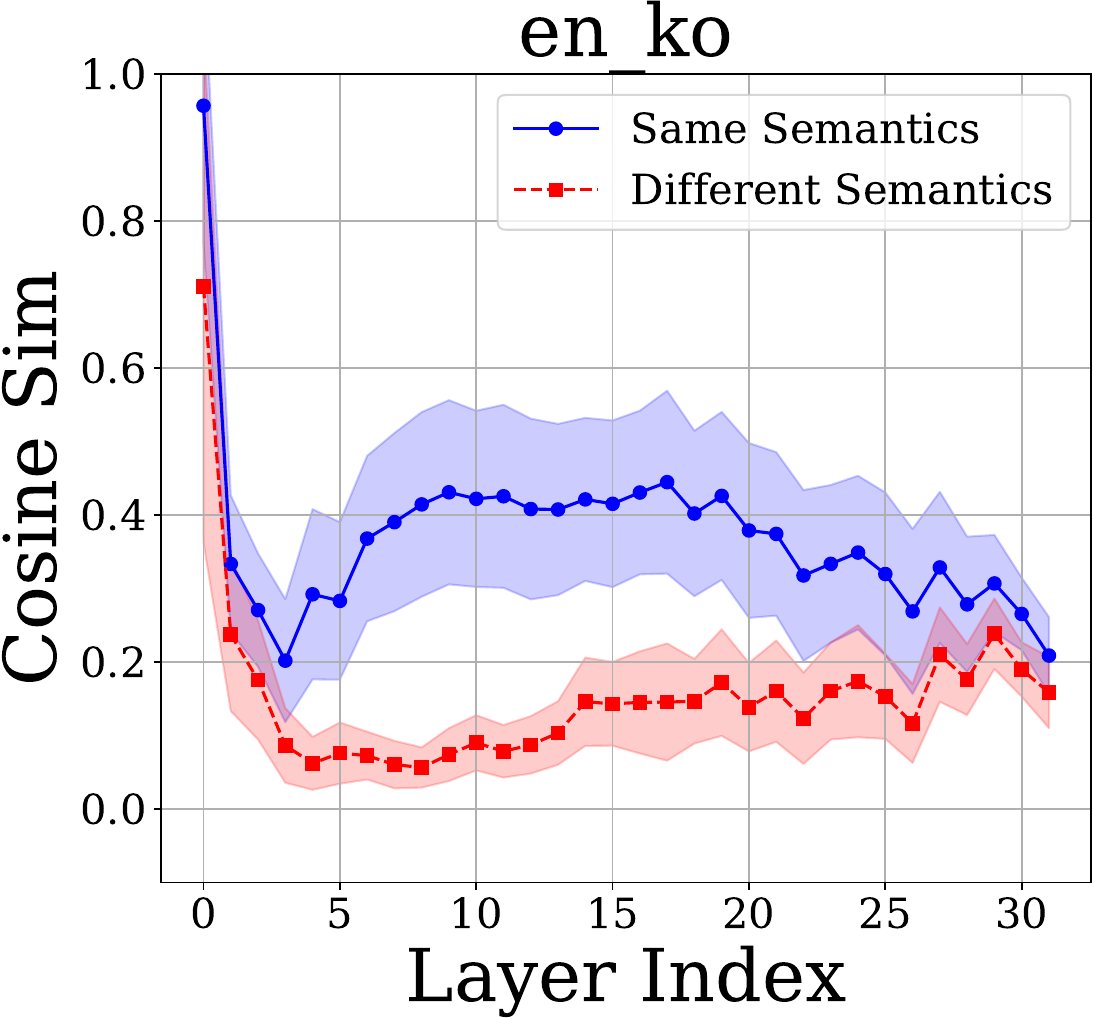}
      \subcaption{en-ko (baseline)}
    \end{minipage}
    \begin{minipage}{0.20\linewidth}
      \centering
      \includegraphics[width=\linewidth]{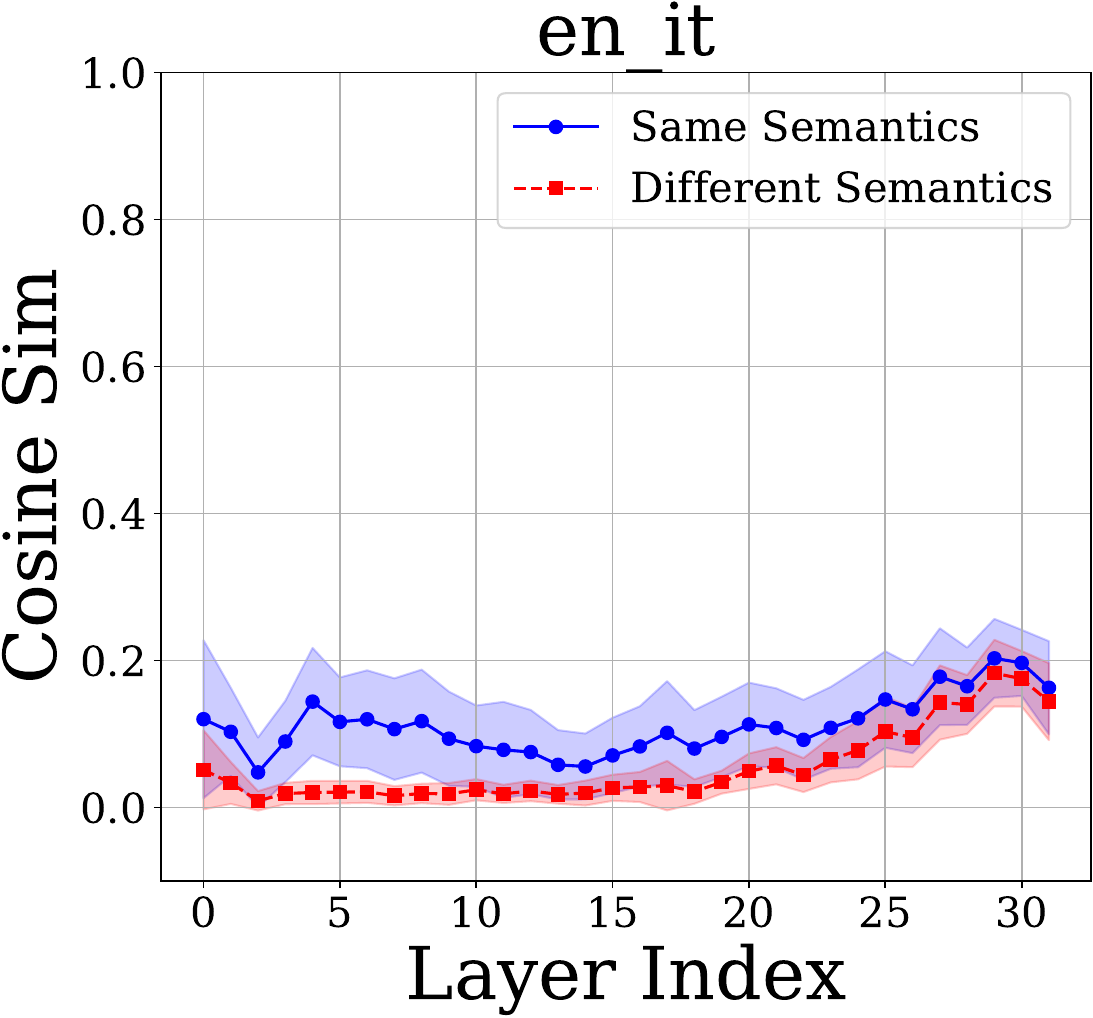}
      \subcaption{en-it}
    \end{minipage}
    \begin{minipage}{0.20\linewidth}
      \centering
      \includegraphics[width=\linewidth]{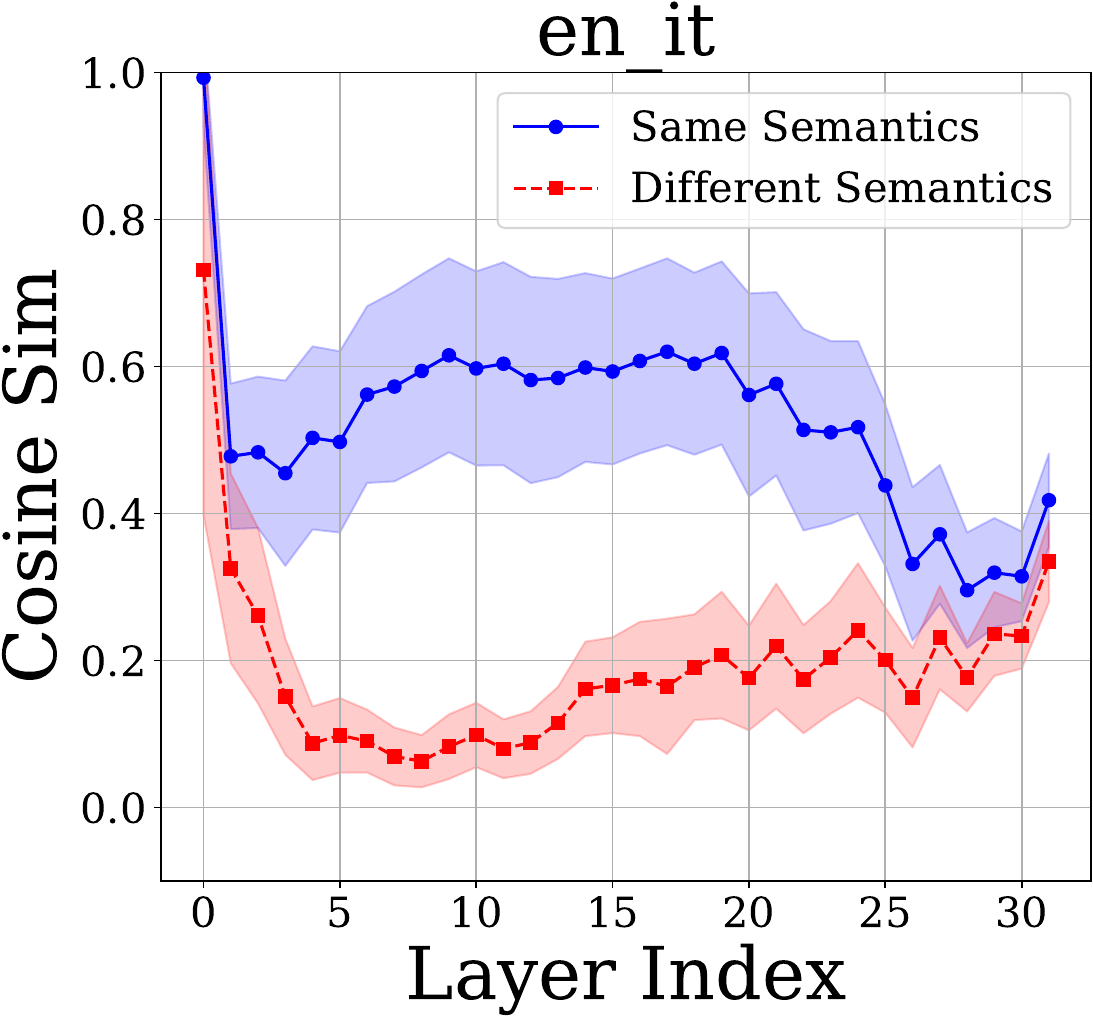}
      
      \subcaption{en-it (baseline)}
    \end{minipage}
    
      \begin{minipage}{\linewidth}
        \centering
        \small \textbf{(b) top-5000 (representing 1\% of all neurons)}
      \end{minipage}

  \caption{\textbf{Similarity of activation patterns across layers while deactivating Type-1 Transfer Neurons (Aya expanse-8B).}}
  \label{fig:appendix:act_sim_aya_deactivating_top-1k_Type-1}
\end{figure*}

Figs.~\ref{fig:appendix:hs_sim_llama_deactivating_top-1k_Type-1}, \ref{fig:appendix:hs_sim_mistral_deactivating_top-1k_Type-1}, and~\ref{fig:appendix:hs_sim_aya_deactivating_top-1k_Type-1} illustrate the similarity of hidden states across all models when the top 1k, 3k, and 5k Type-1 neurons are deactivated. Similarly, Figs.~\ref{fig:appendix:act_sim_llama_deactivating_top-1k_Type-1}, \ref{fig:appendix:act_sim_mistral_deactivating_top-1k_Type-1}, and~\ref{fig:appendix:act_sim_aya_deactivating_top-1k_Type-1} present the corresponding similarities in activation patterns.

%--- hs sim ---%
% llama3, top1k-5k
% mistral, top1k-5k
% aya, top1k-5k

%--- act patterns sim ---%
% llama3, top1k-5k
% mistral, top1k-5k
% aya, top1k-5k

\subsubsection{Quantitative Distance among Language Latent Spaces}
\label{sec:appendix:distance between subspaces while deactivating type-1}
\begin{figure*}[t]
  \centering

  \includegraphics[width=0.15\linewidth]{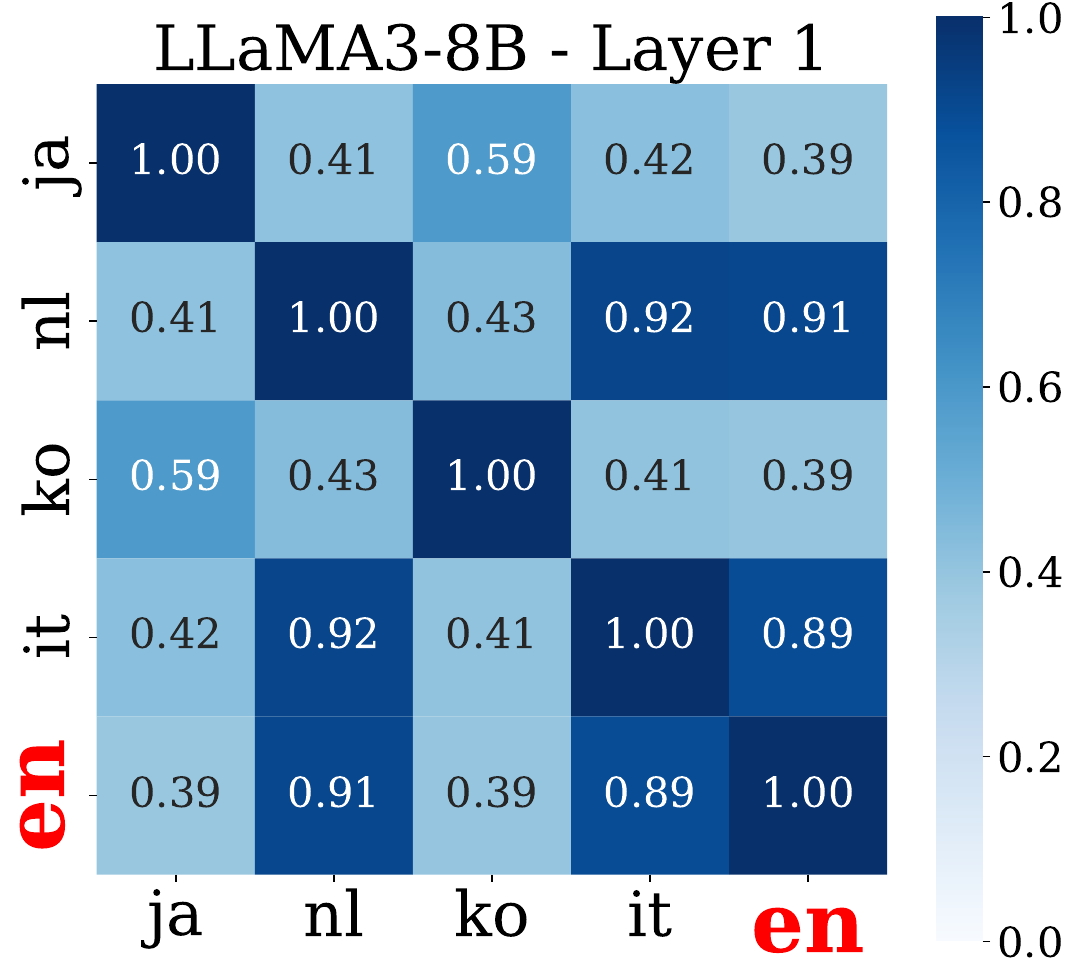}
  \includegraphics[width=0.15\linewidth]{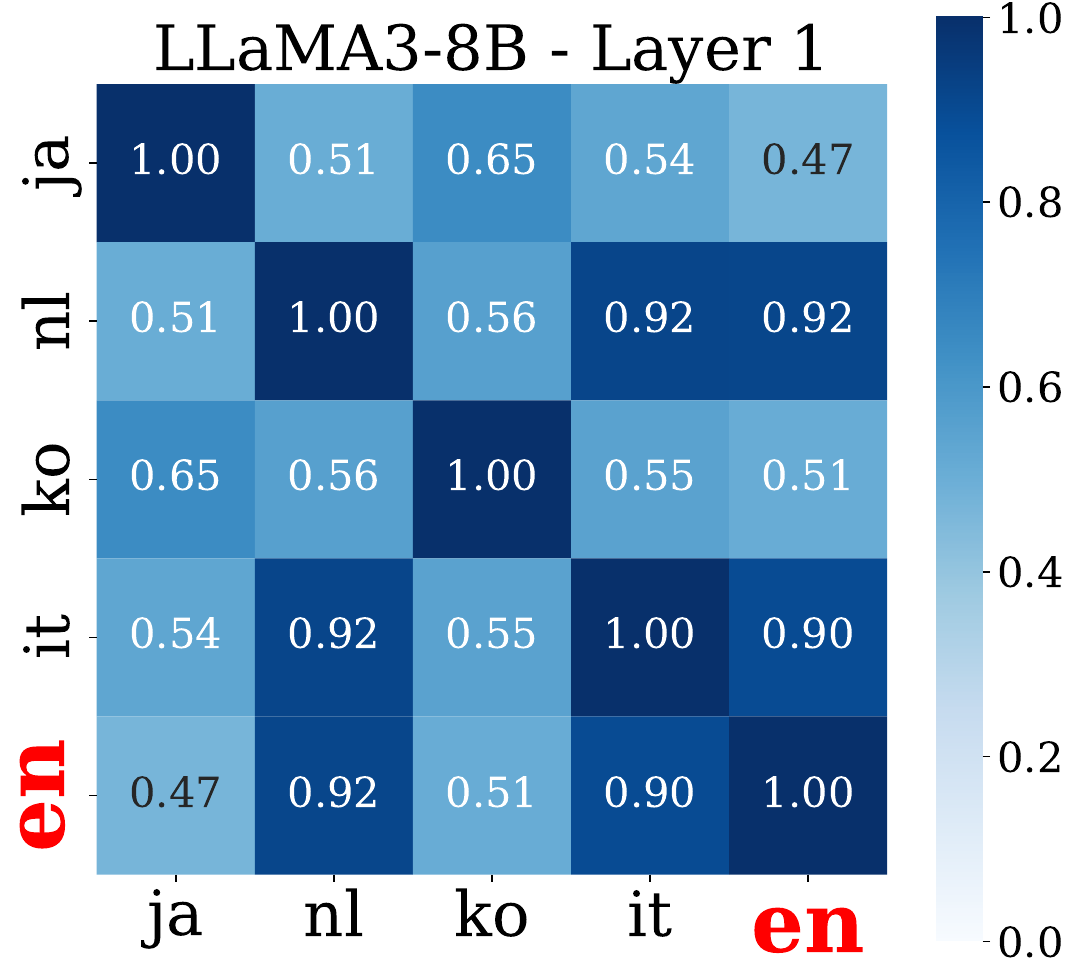}
  \includegraphics[width=0.15\linewidth]{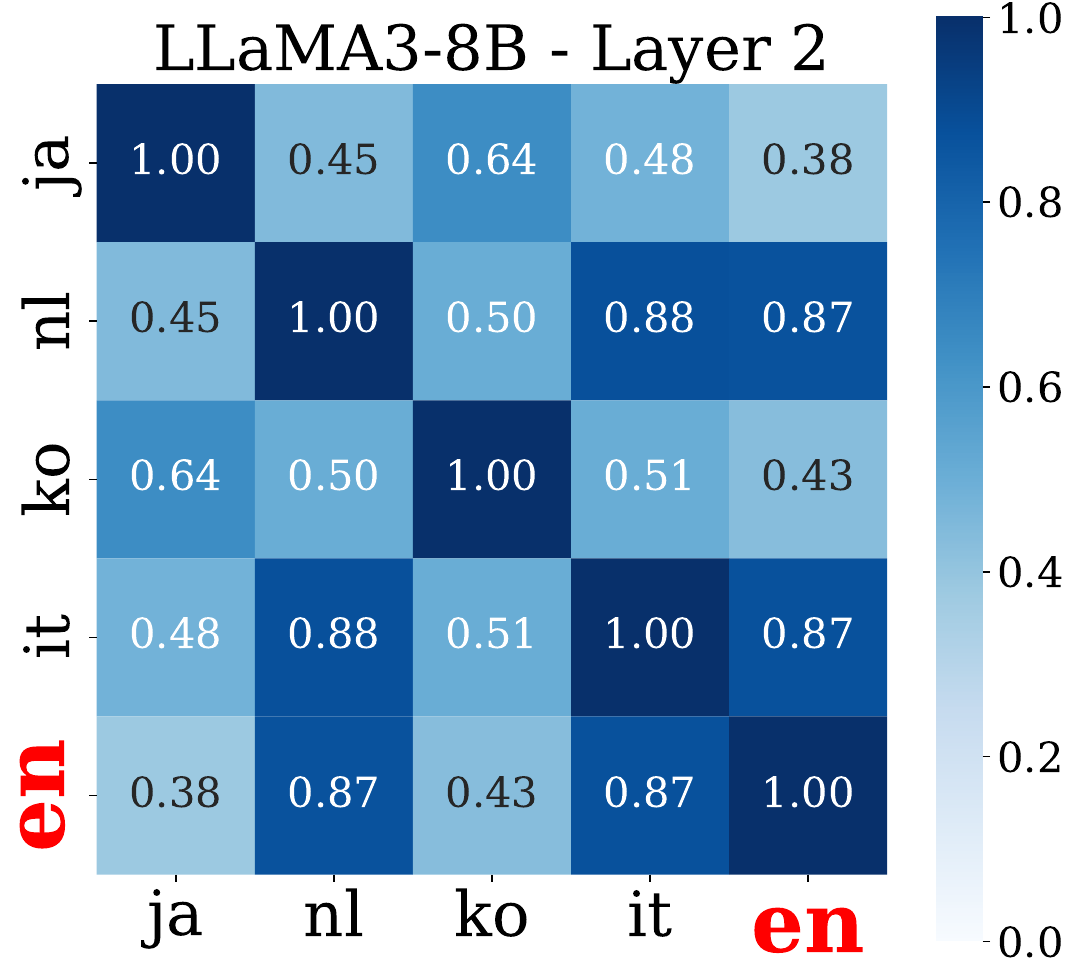}
  \includegraphics[width=0.15\linewidth]{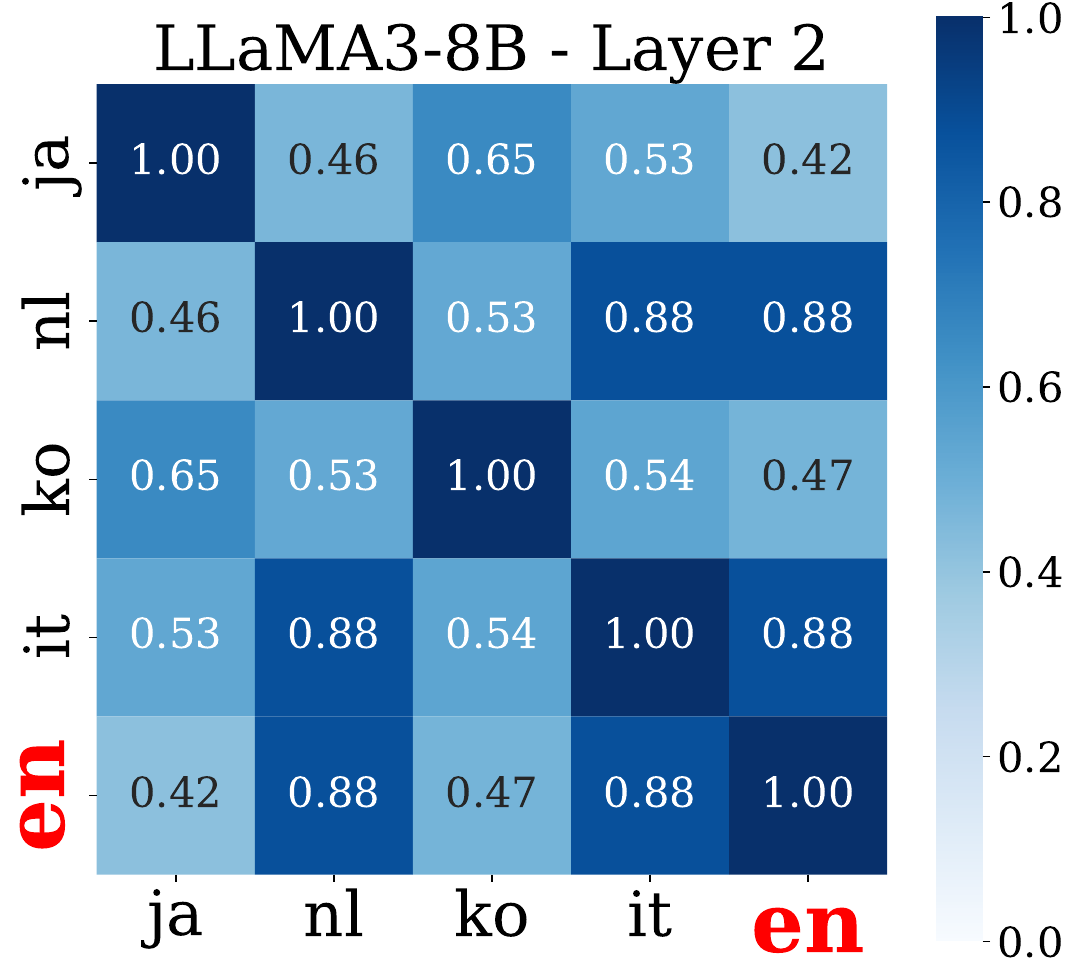}
  \includegraphics[width=0.15\linewidth]{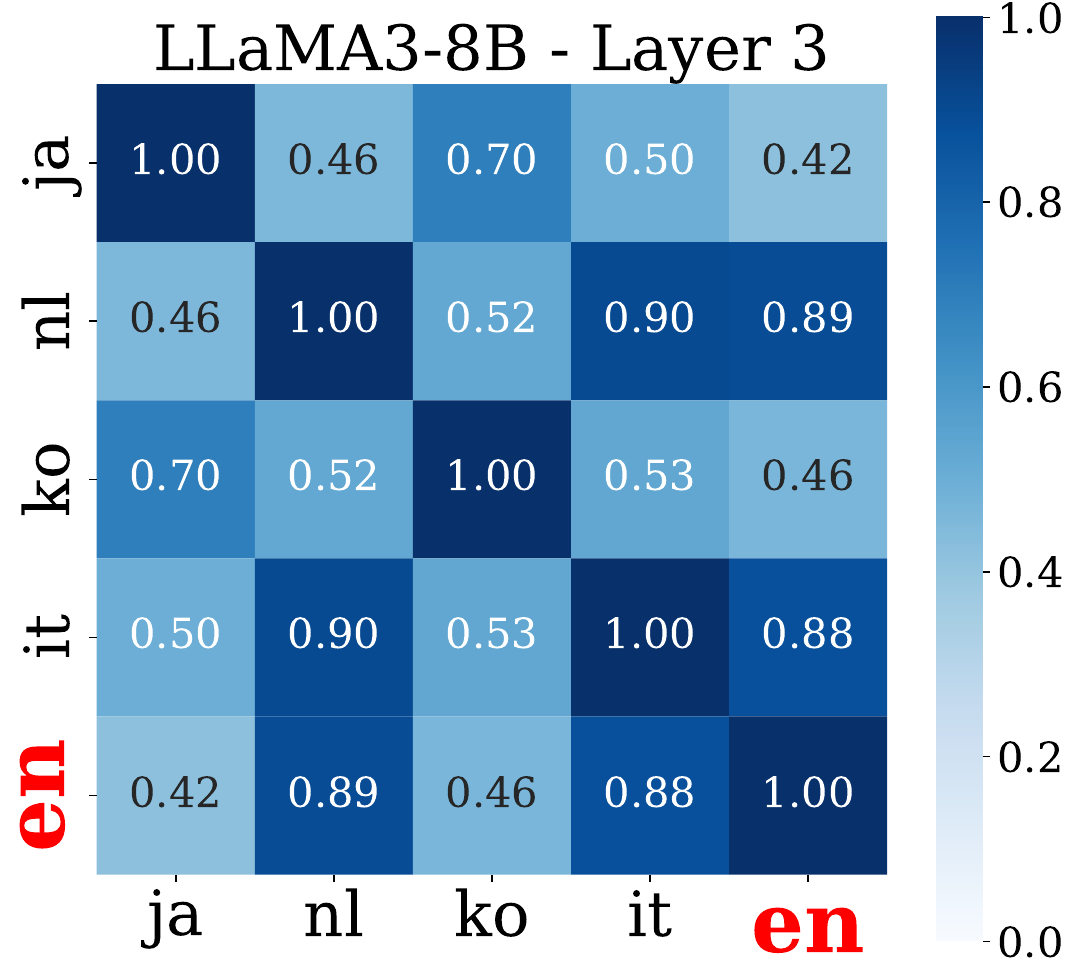}
  \includegraphics[width=0.15\linewidth]{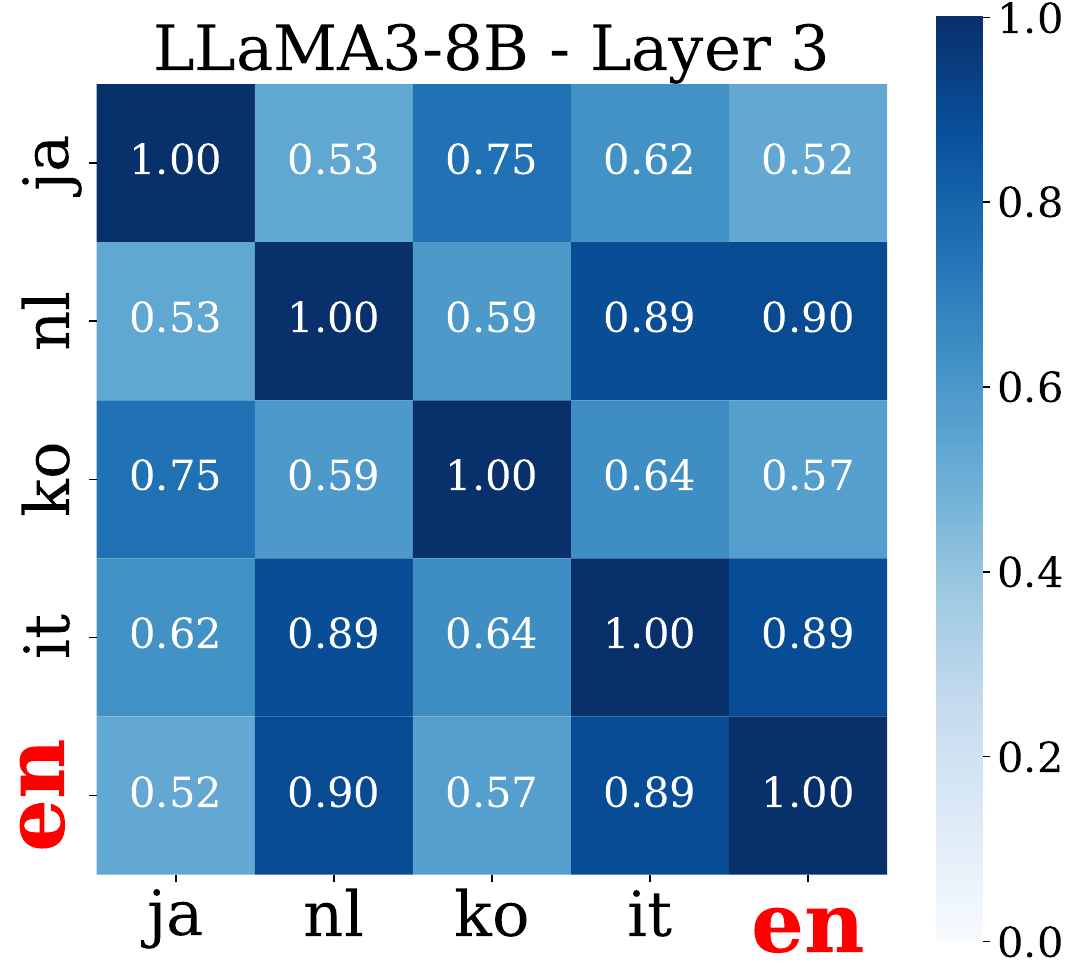}

  \begin{minipage}{0.15\linewidth}\centering \textbf{\textcolor{red}{layer 1 (Type-1)}}\end{minipage}
  \begin{minipage}{0.15\linewidth}\centering layer 1 (baseline)\end{minipage}
  \begin{minipage}{0.15\linewidth}\centering \textbf{\textcolor{red}{layer 2 (Type-1)}}\end{minipage}
  \begin{minipage}{0.15\linewidth}\centering layer 2 (baseline)\end{minipage}
  \begin{minipage}{0.15\linewidth}\centering \textbf{\textcolor{red}{layer 3 (Type-1)}}\end{minipage}
  \begin{minipage}{0.15\linewidth}\centering layer 3 (baseline)\end{minipage}

  \includegraphics[width=0.15\linewidth]{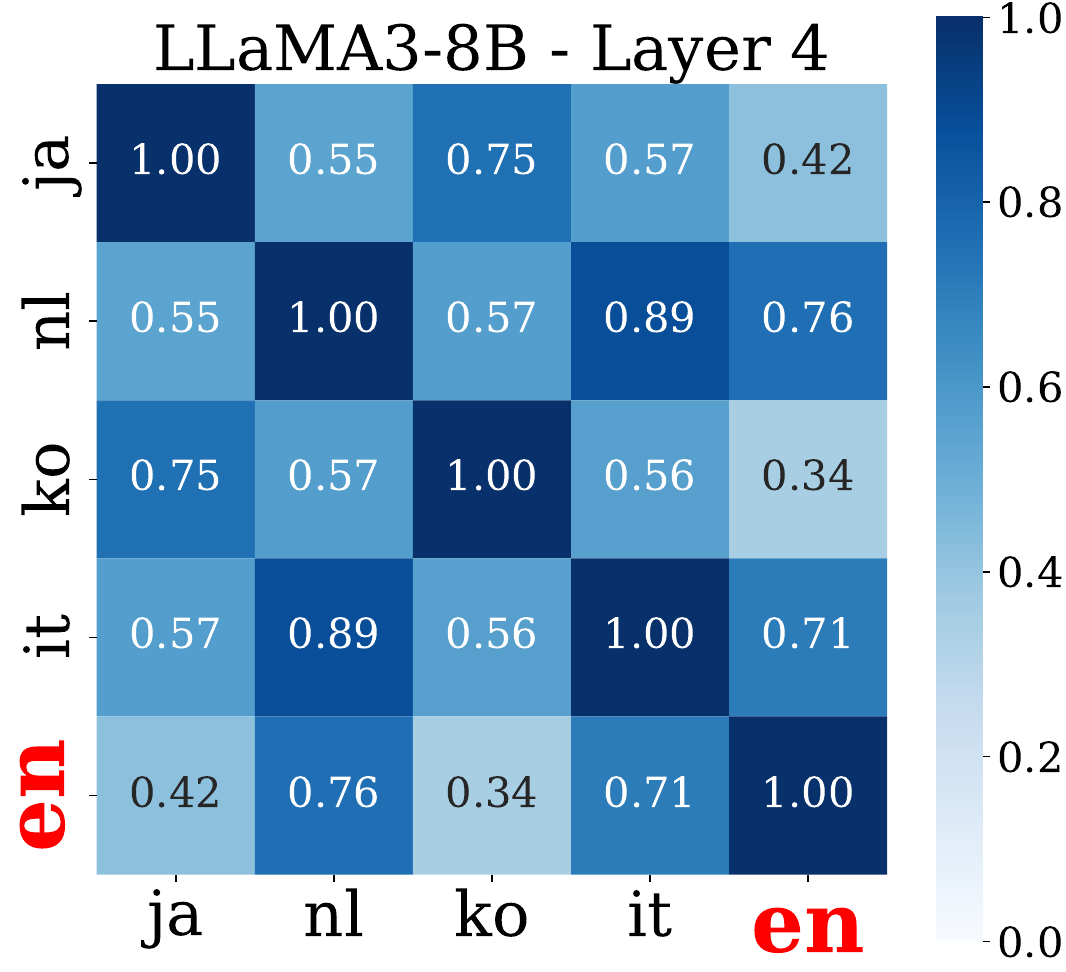}
  \includegraphics[width=0.15\linewidth]{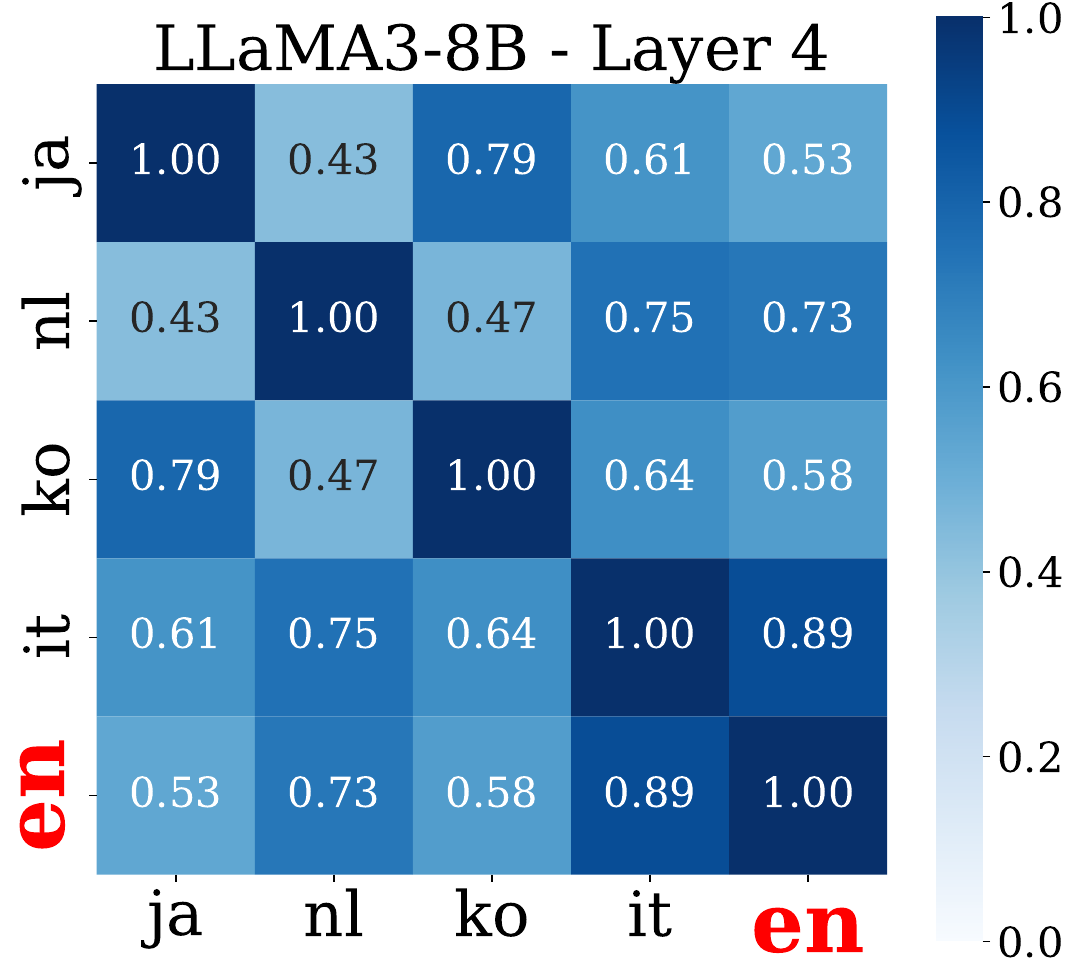}
  \includegraphics[width=0.15\linewidth]{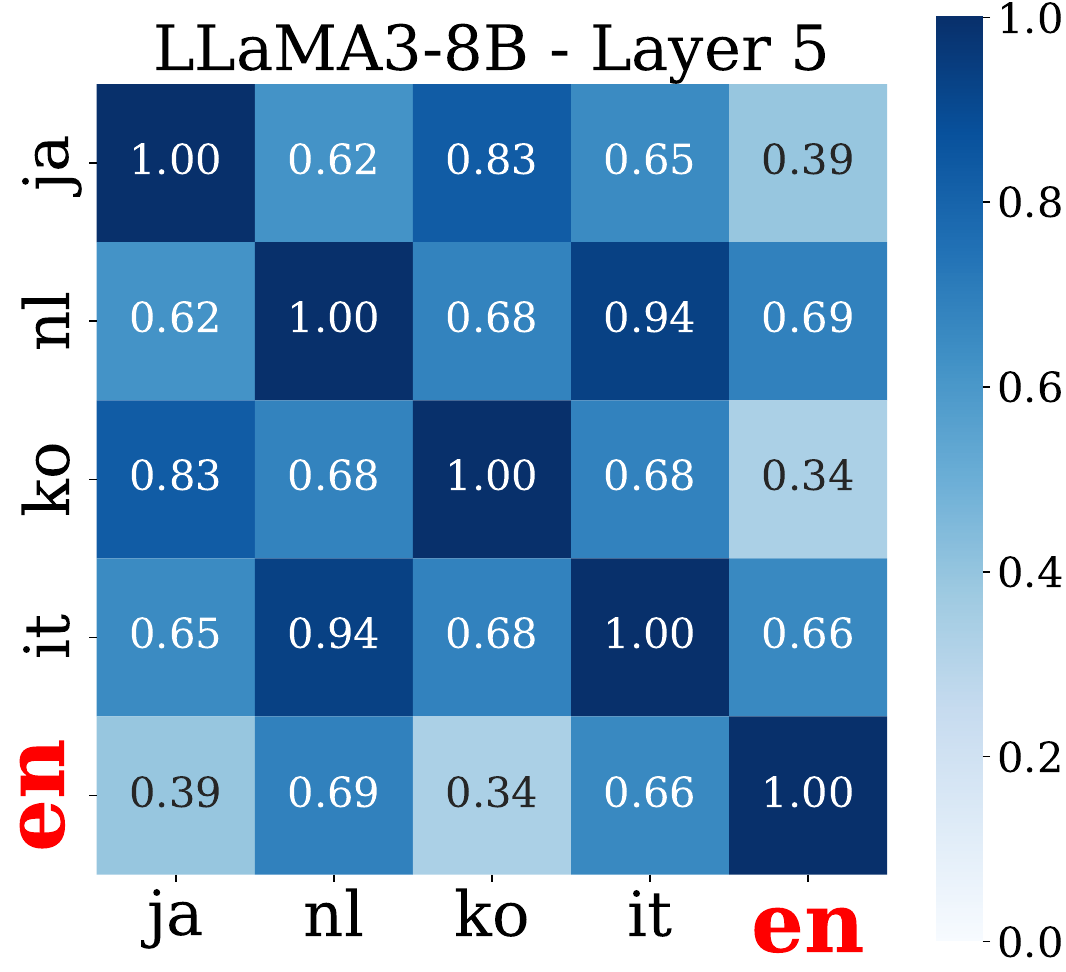}
  \includegraphics[width=0.15\linewidth]{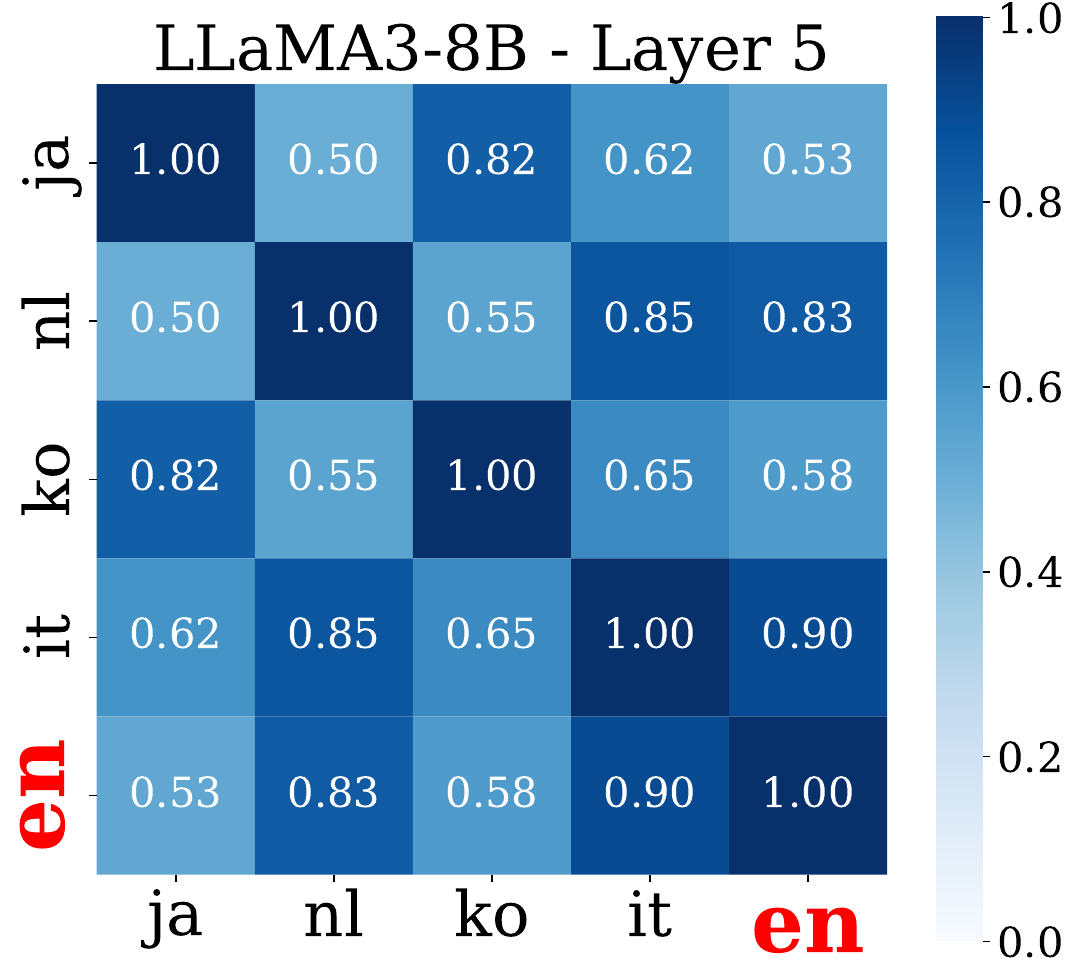}
  \includegraphics[width=0.15\linewidth]{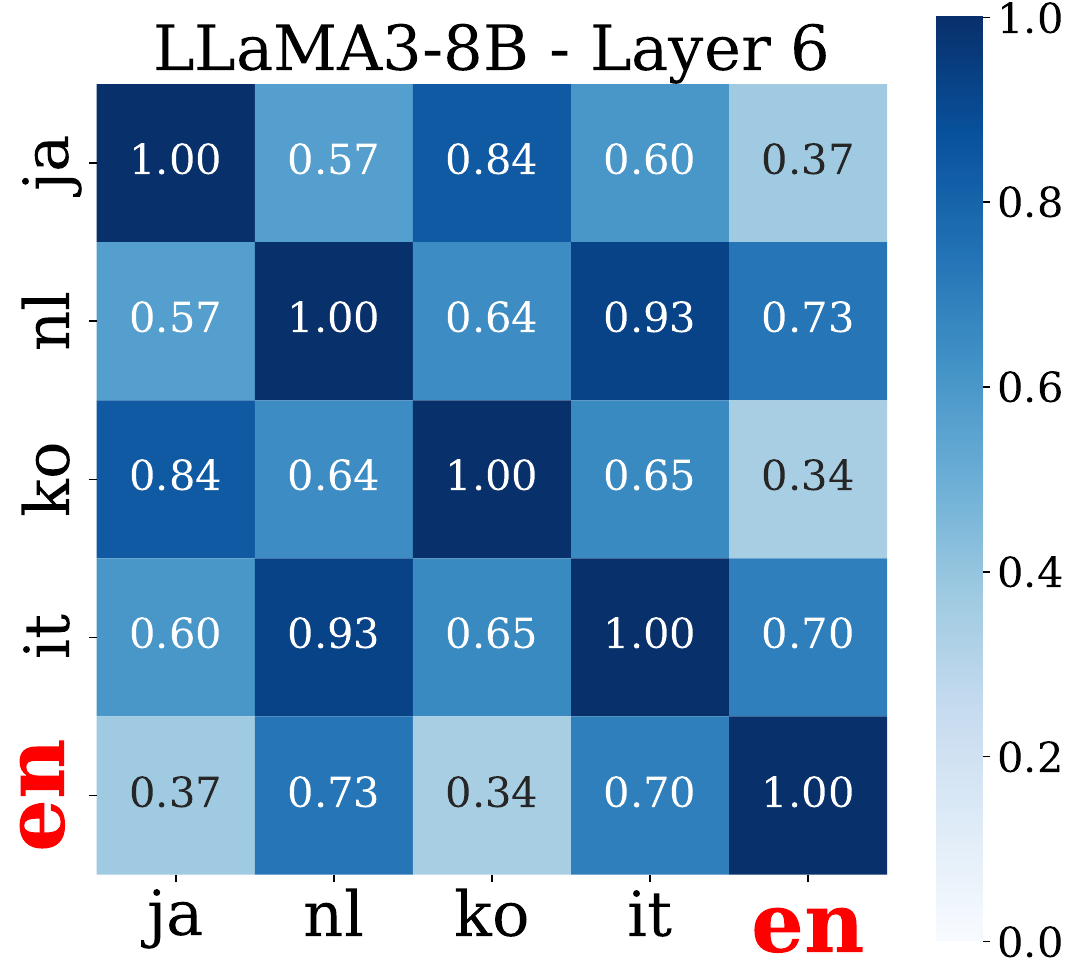}
  \includegraphics[width=0.15\linewidth]{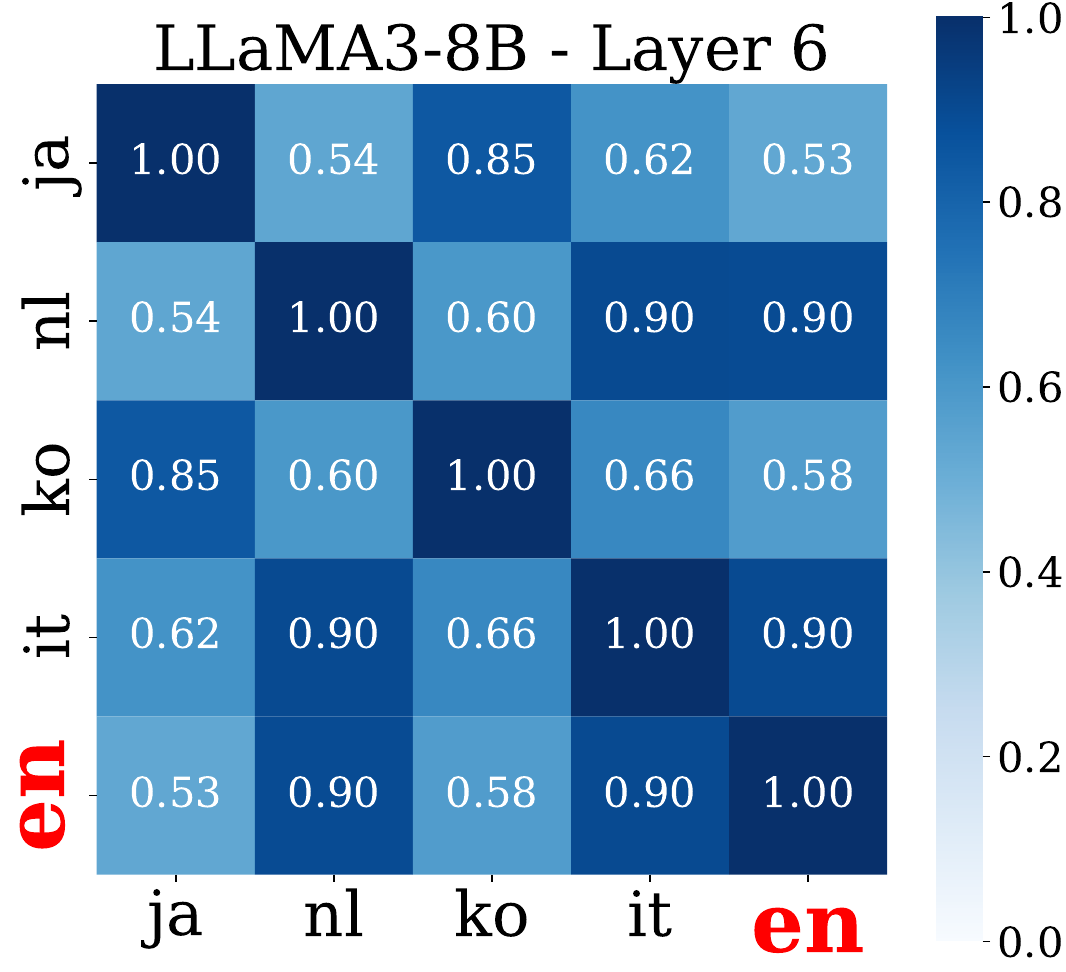}

  \begin{minipage}{0.15\linewidth}\centering \textbf{\textcolor{red}{layer 4 (Type-1)}}\end{minipage}
  \begin{minipage}{0.15\linewidth}\centering layer 4 (baseline)\end{minipage}
  \begin{minipage}{0.15\linewidth}\centering \textbf{\textcolor{red}{layer 5 (Type-1)}}\end{minipage}
  \begin{minipage}{0.15\linewidth}\centering layer 5 (baseline)\end{minipage}
  \begin{minipage}{0.15\linewidth}\centering \textbf{\textcolor{red}{layer 6 (Type-1)}}\end{minipage}
  \begin{minipage}{0.15\linewidth}\centering layer 6 (baseline)\end{minipage}

  \includegraphics[width=0.15\linewidth]{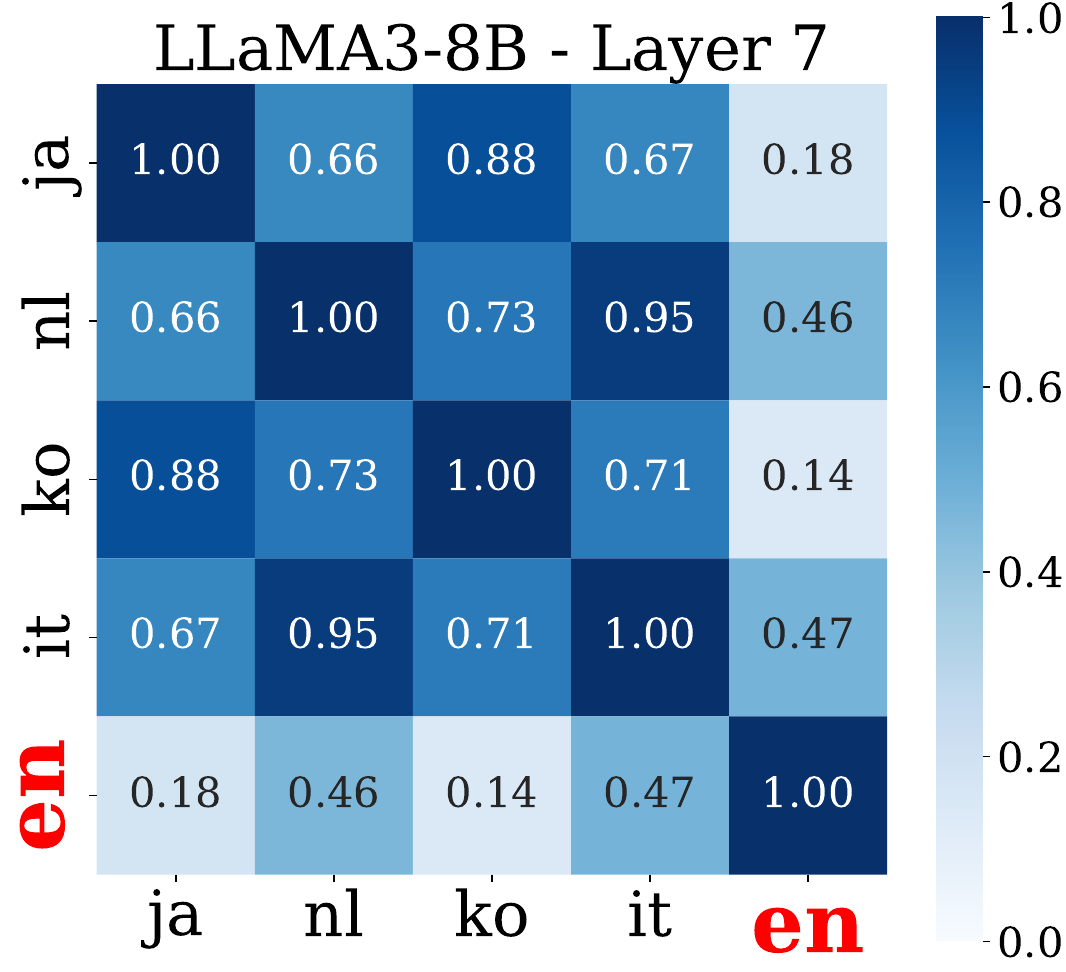}
  \includegraphics[width=0.15\linewidth]{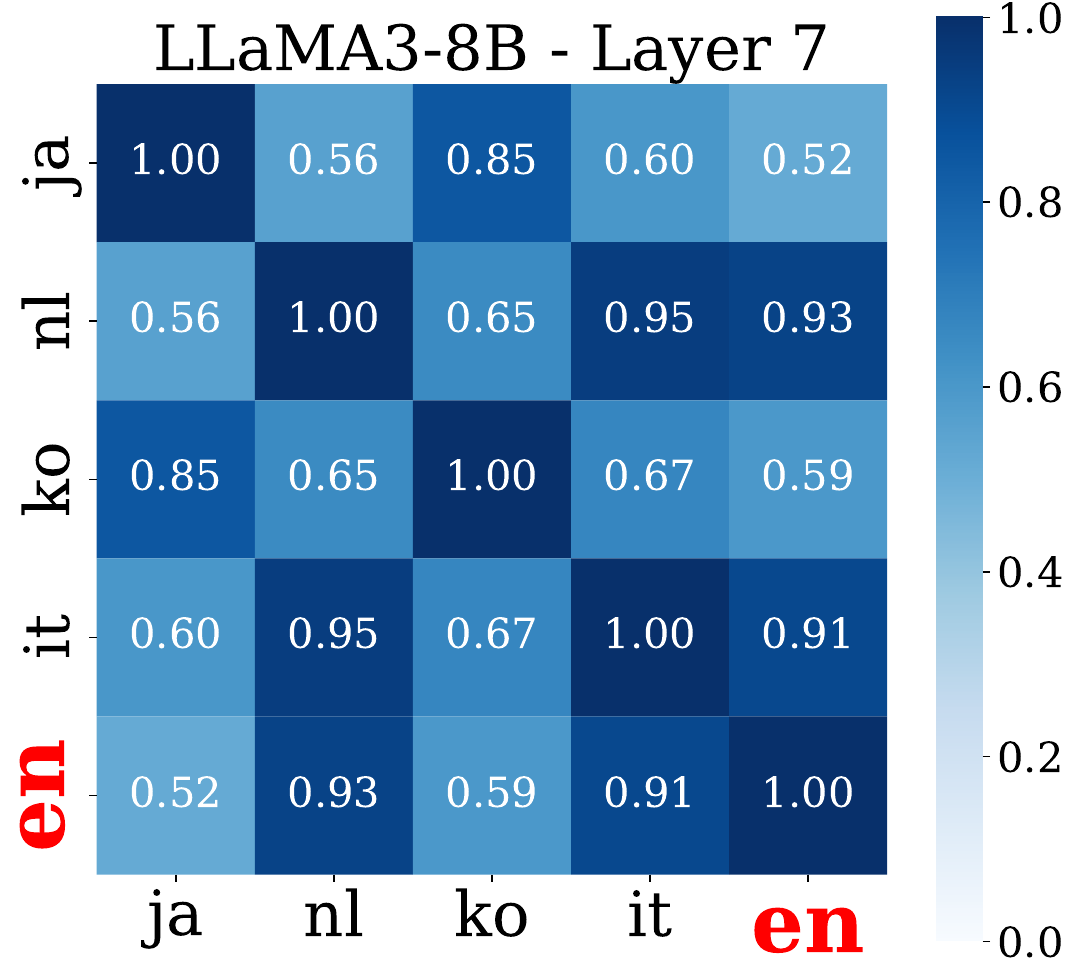}
  \includegraphics[width=0.15\linewidth]{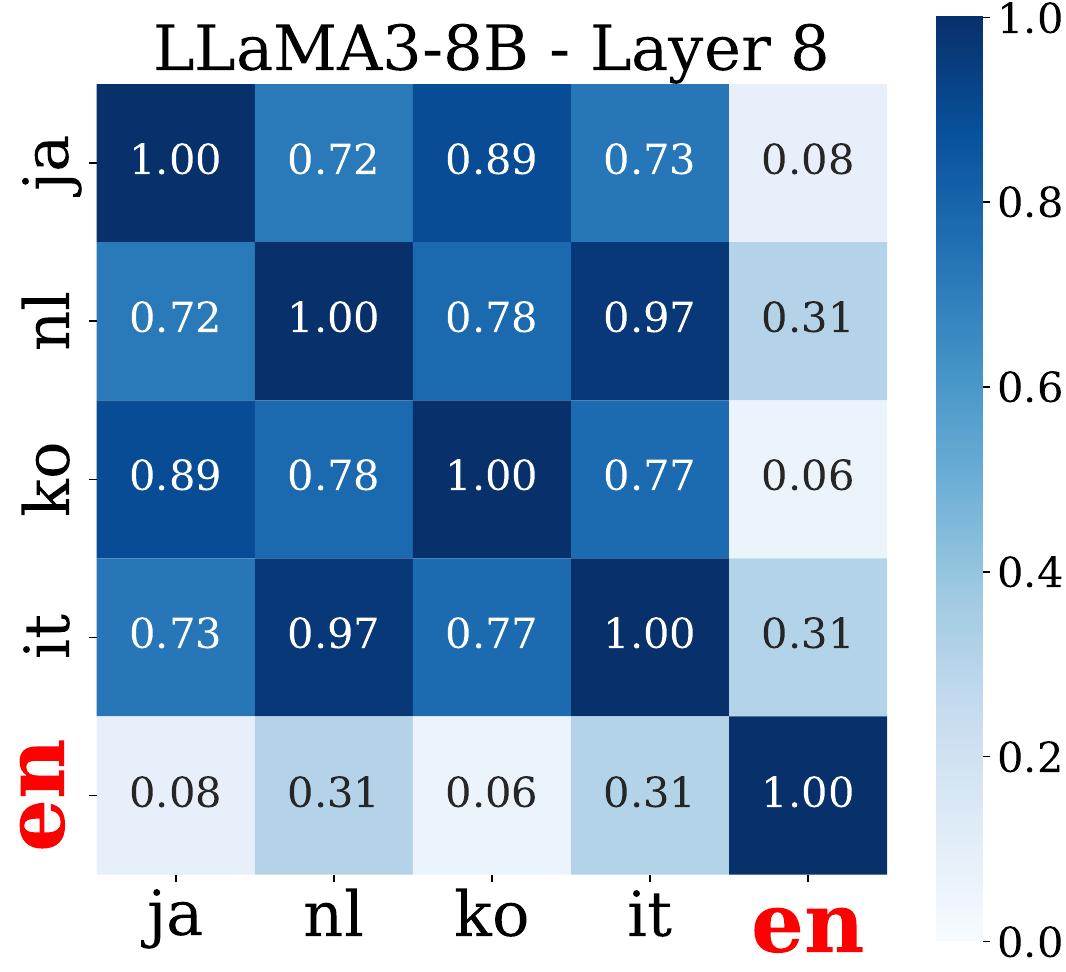}
  \includegraphics[width=0.15\linewidth]{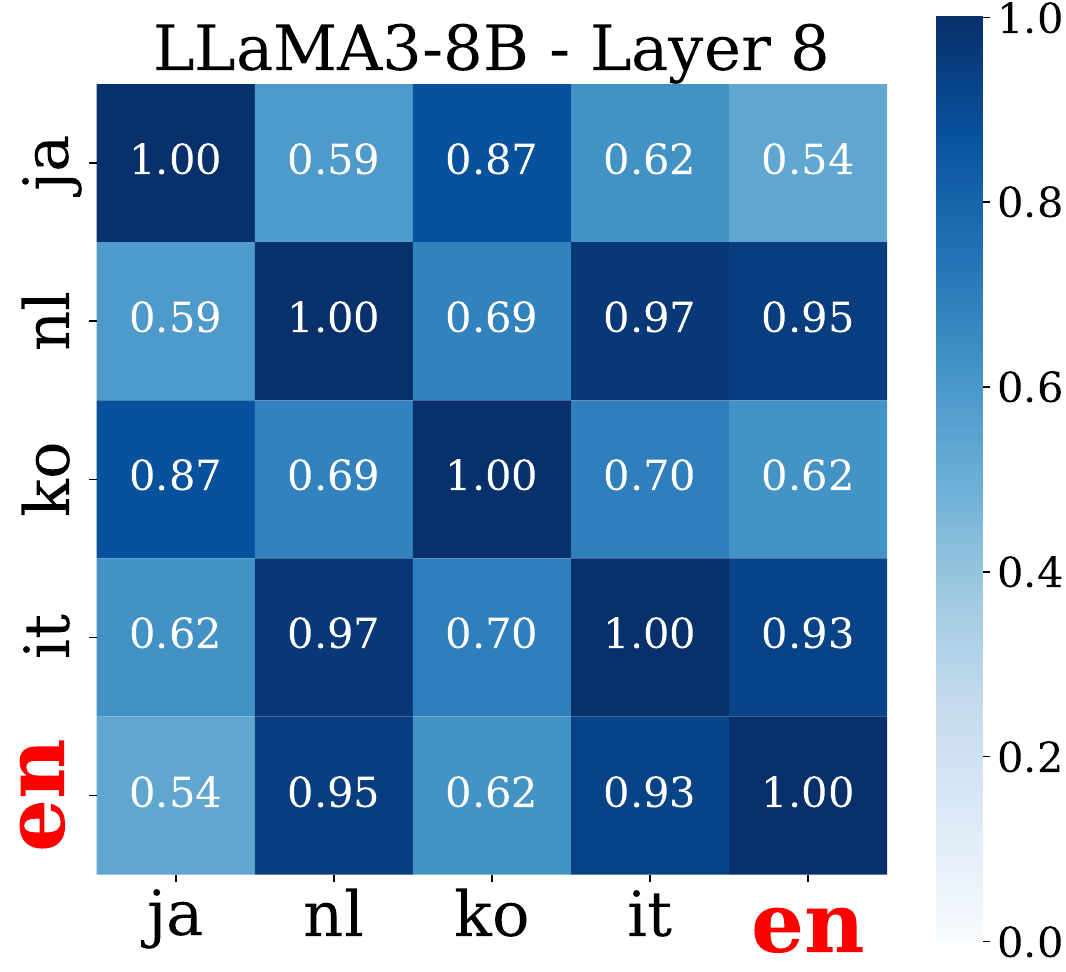}
  \includegraphics[width=0.15\linewidth]{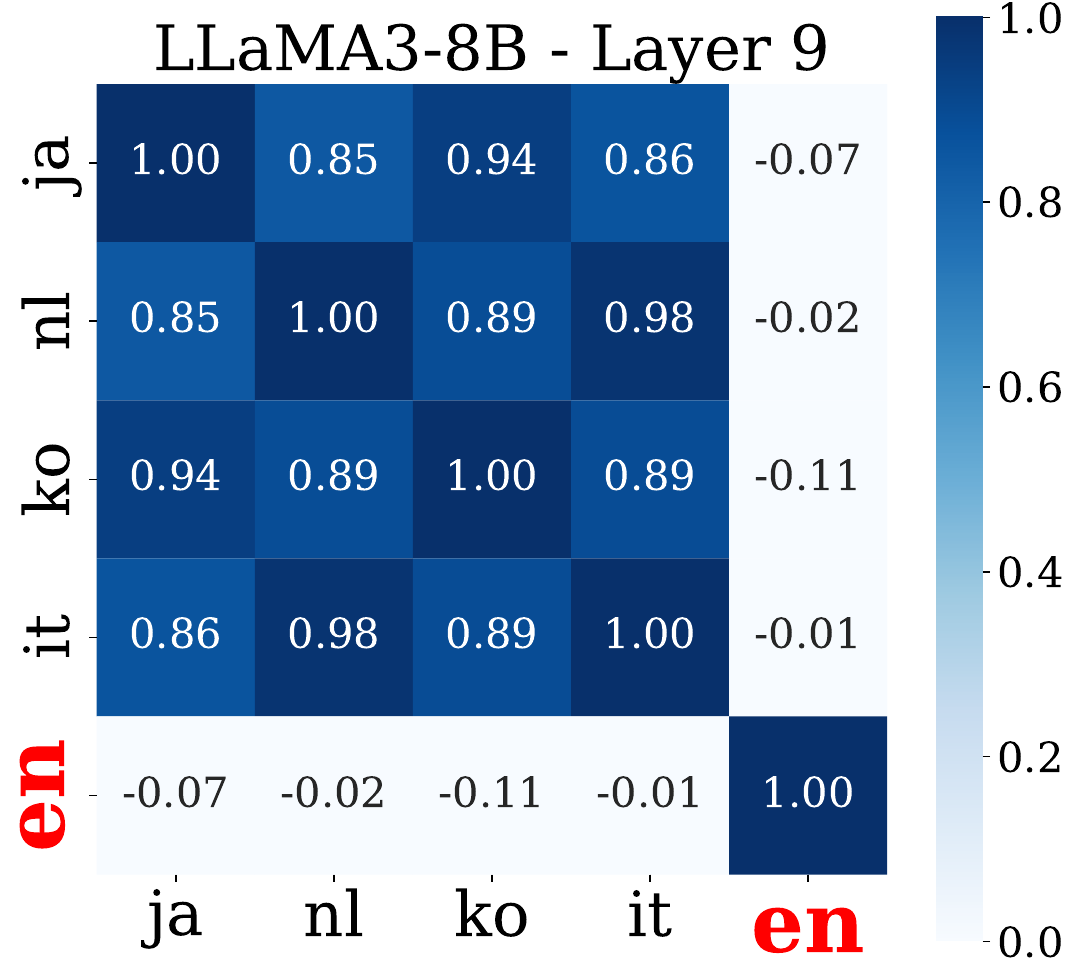}
  \includegraphics[width=0.15\linewidth]{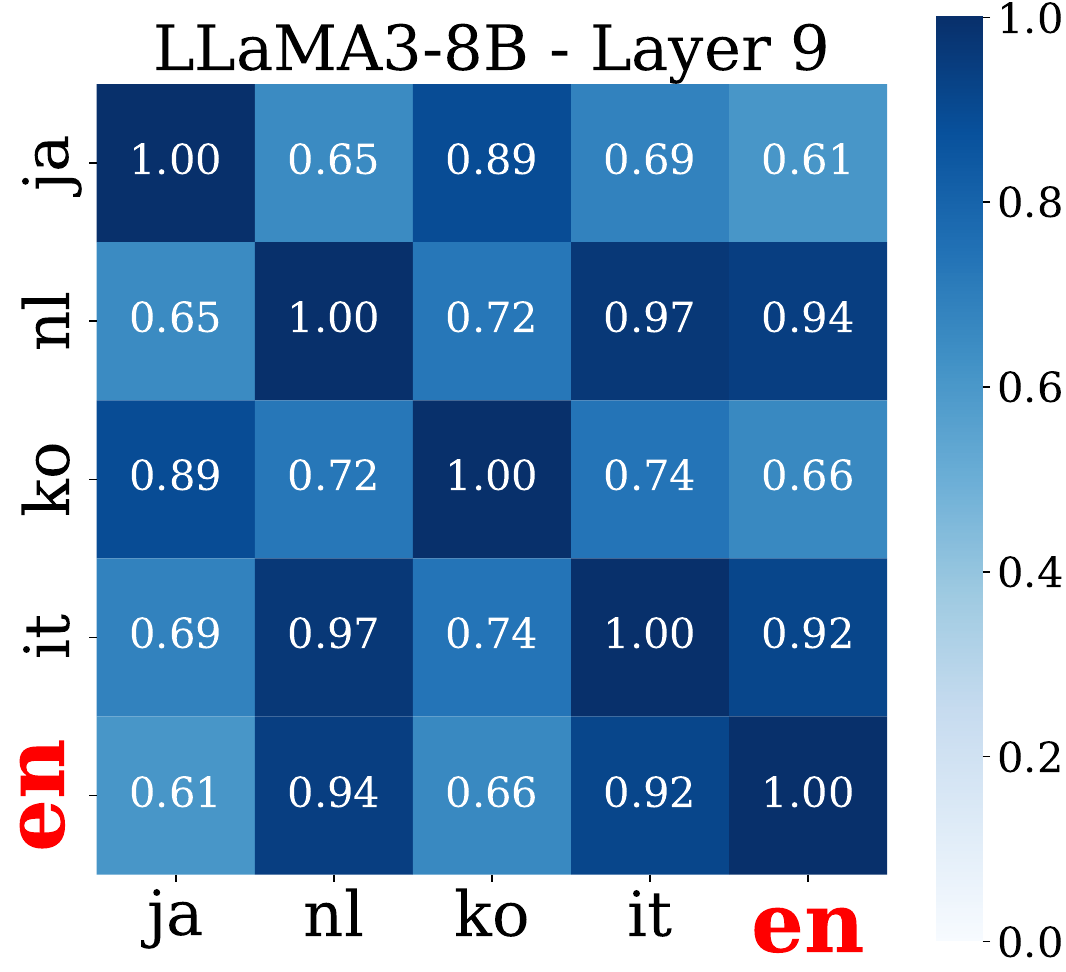}

  \begin{minipage}{0.15\linewidth}\centering \textbf{\textcolor{red}{layer 7 (Type-1)}}\end{minipage}
  \begin{minipage}{0.15\linewidth}\centering layer 7 (baseline)\end{minipage}
  \begin{minipage}{0.15\linewidth}\centering \textbf{\textcolor{red}{layer 8 (Type-1)}}\end{minipage}
  \begin{minipage}{0.15\linewidth}\centering layer 8 (baseline)\end{minipage}
  \begin{minipage}{0.15\linewidth}\centering \textbf{\textcolor{red}{layer 9 (Type-1)}}\end{minipage}
  \begin{minipage}{0.15\linewidth}\centering layer 9 (baseline)\end{minipage}

  \includegraphics[width=0.15\linewidth]{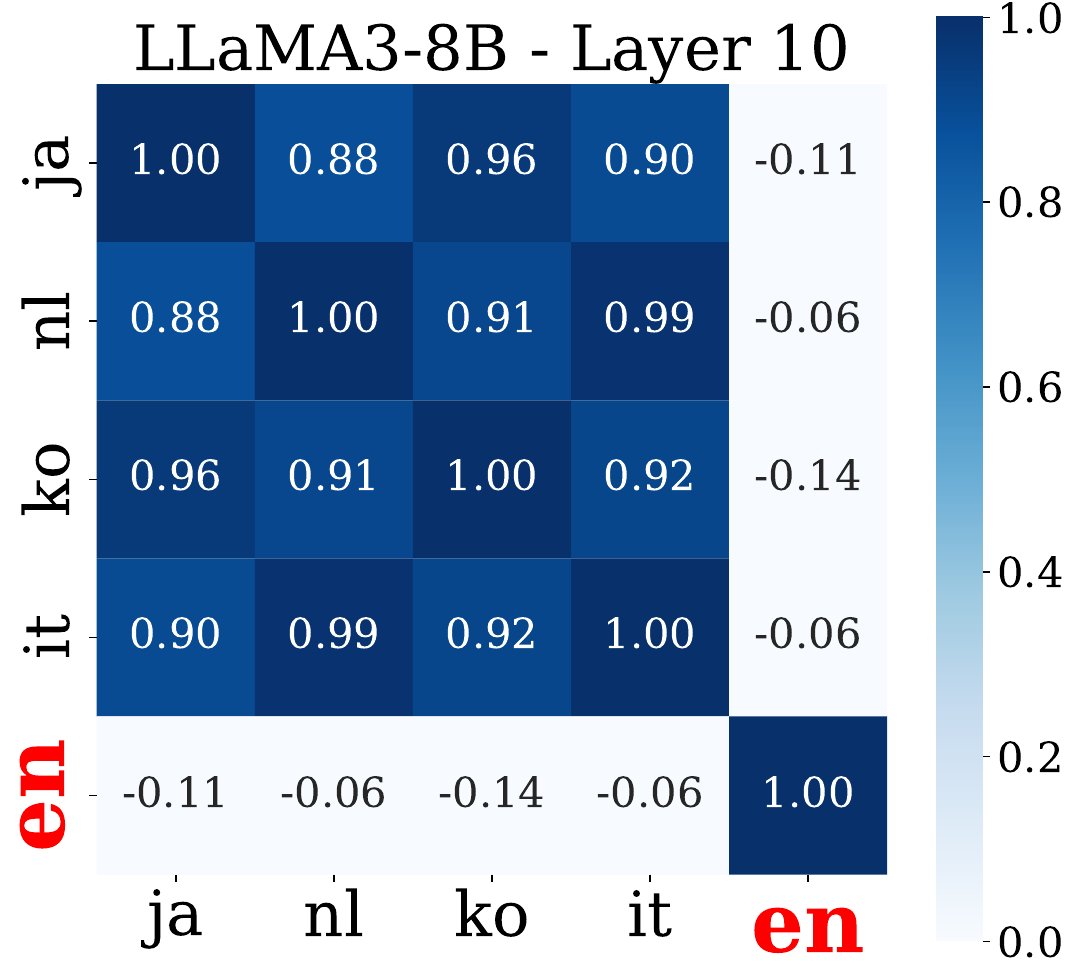}
  \includegraphics[width=0.15\linewidth]{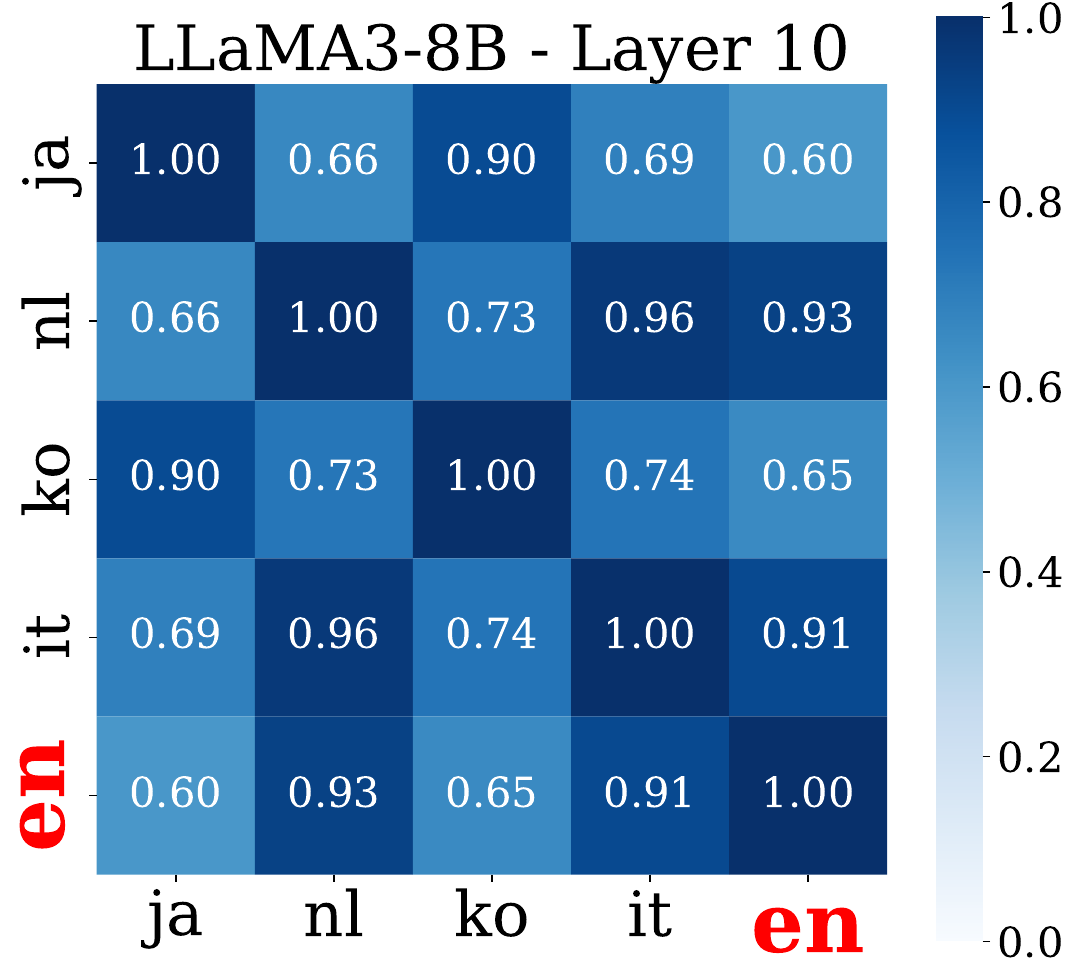}
  \includegraphics[width=0.15\linewidth]{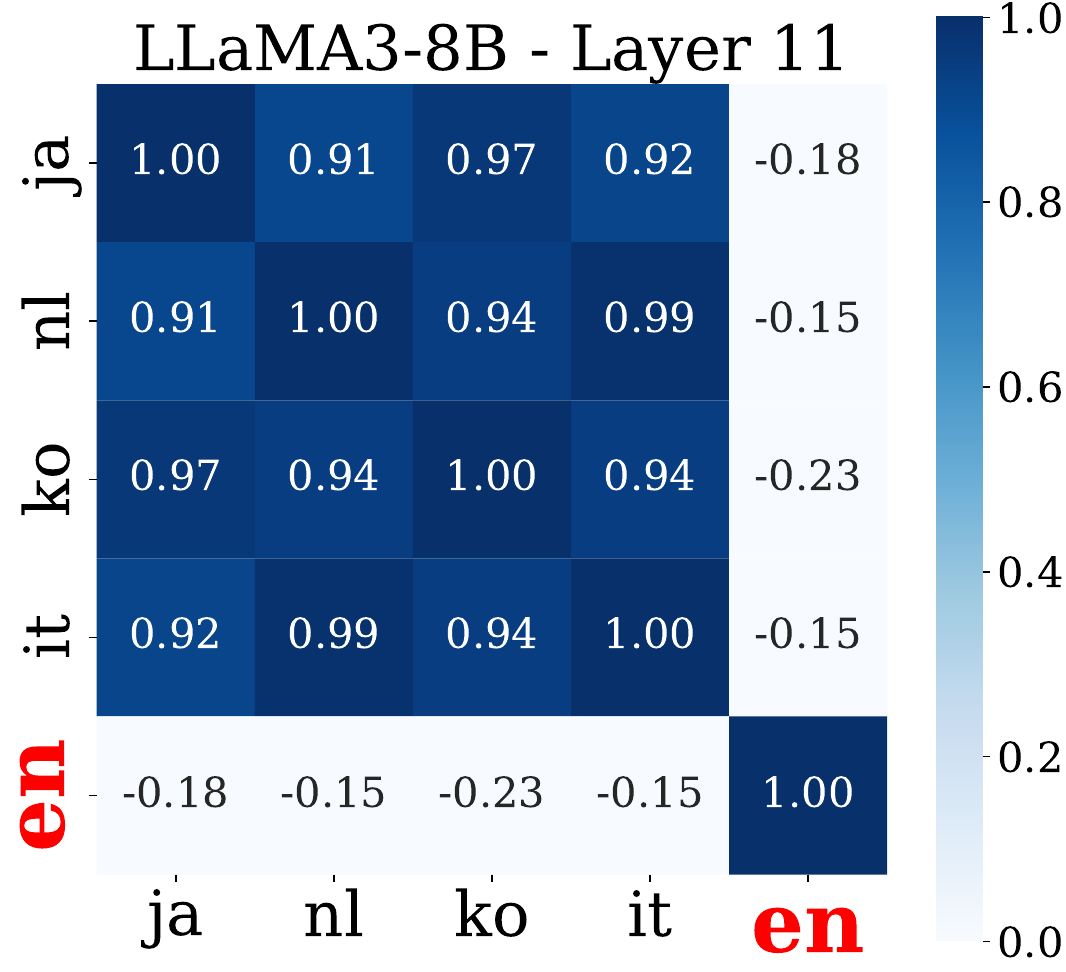}
  \includegraphics[width=0.15\linewidth]{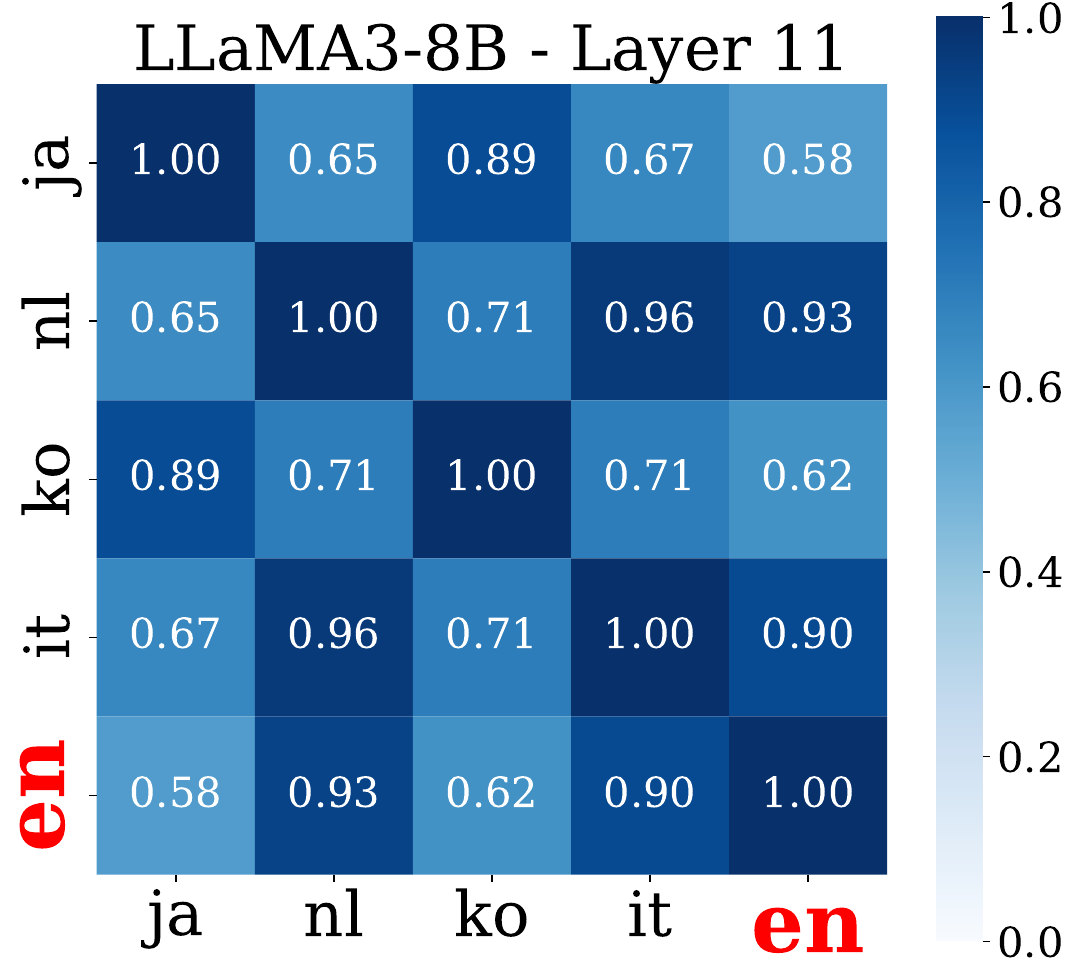}
  \includegraphics[width=0.15\linewidth]{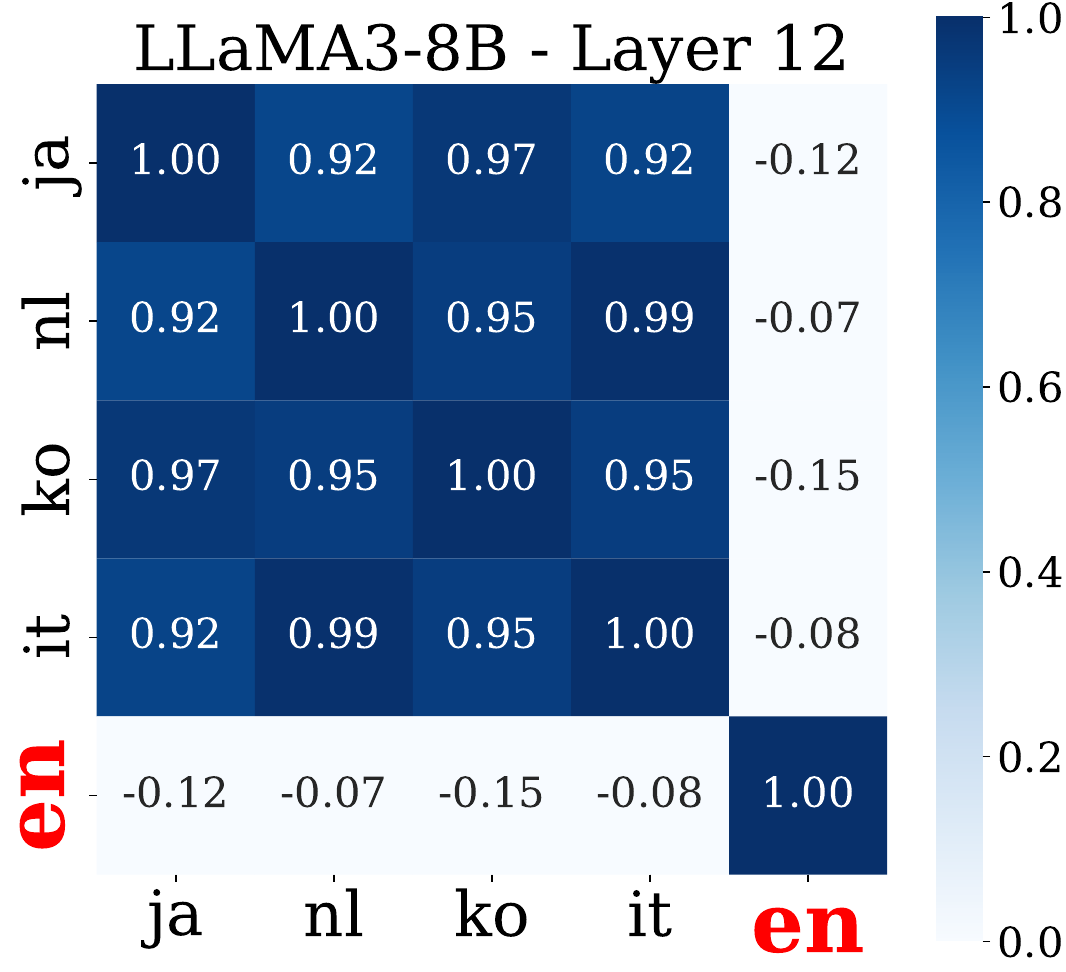}
  \includegraphics[width=0.15\linewidth]{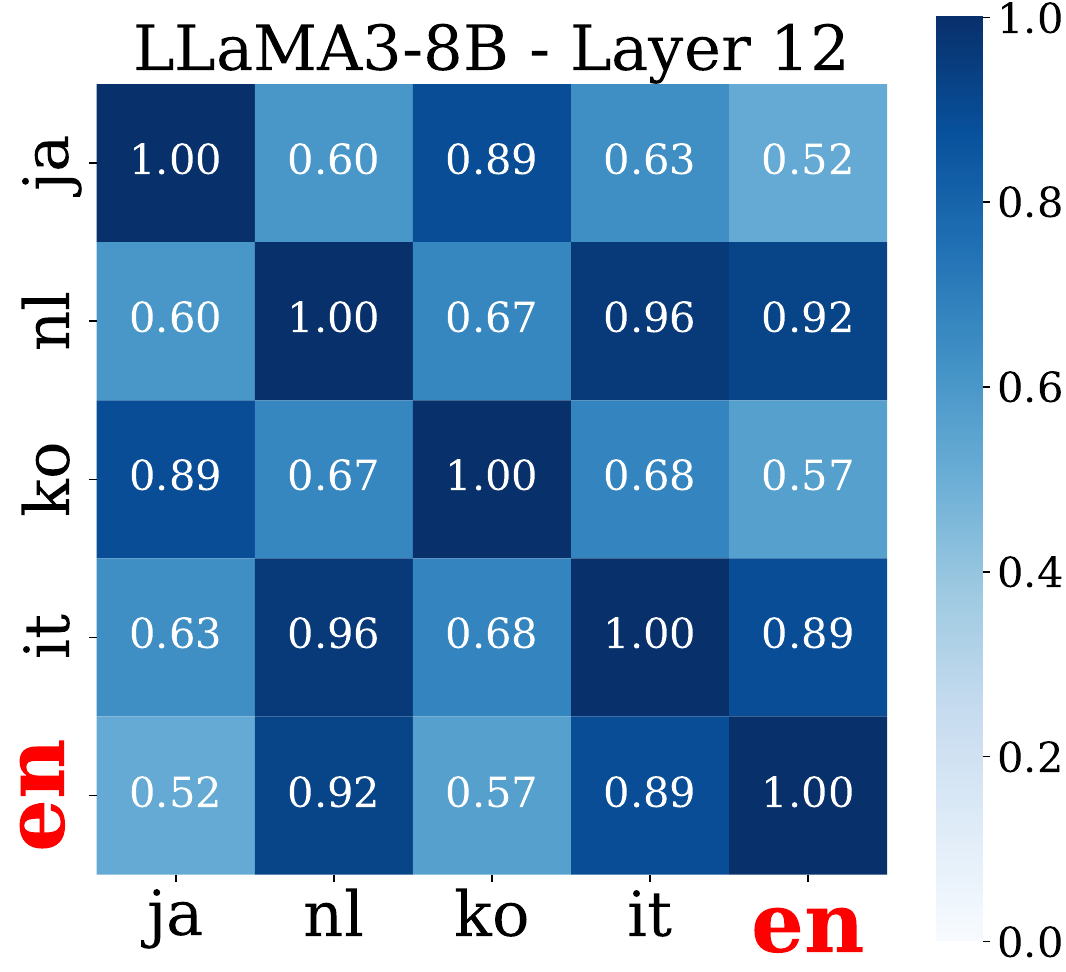}

  \begin{minipage}{0.15\linewidth}\centering \textbf{\textcolor{red}{layer 10 (Type-1)}}\end{minipage}
  \begin{minipage}{0.15\linewidth}\centering layer 10 (baseline)\end{minipage}
  \begin{minipage}{0.15\linewidth}\centering \textbf{\textcolor{red}{layer 11 (Type-1)}}\end{minipage}
  \begin{minipage}{0.15\linewidth}\centering layer 11 (baseline)\end{minipage}
  \begin{minipage}{0.15\linewidth}\centering \textbf{\textcolor{red}{layer 12 (Type-1)}}\end{minipage}
  \begin{minipage}{0.15\linewidth}\centering layer 12 (baseline)\end{minipage}

  \includegraphics[width=0.15\linewidth]{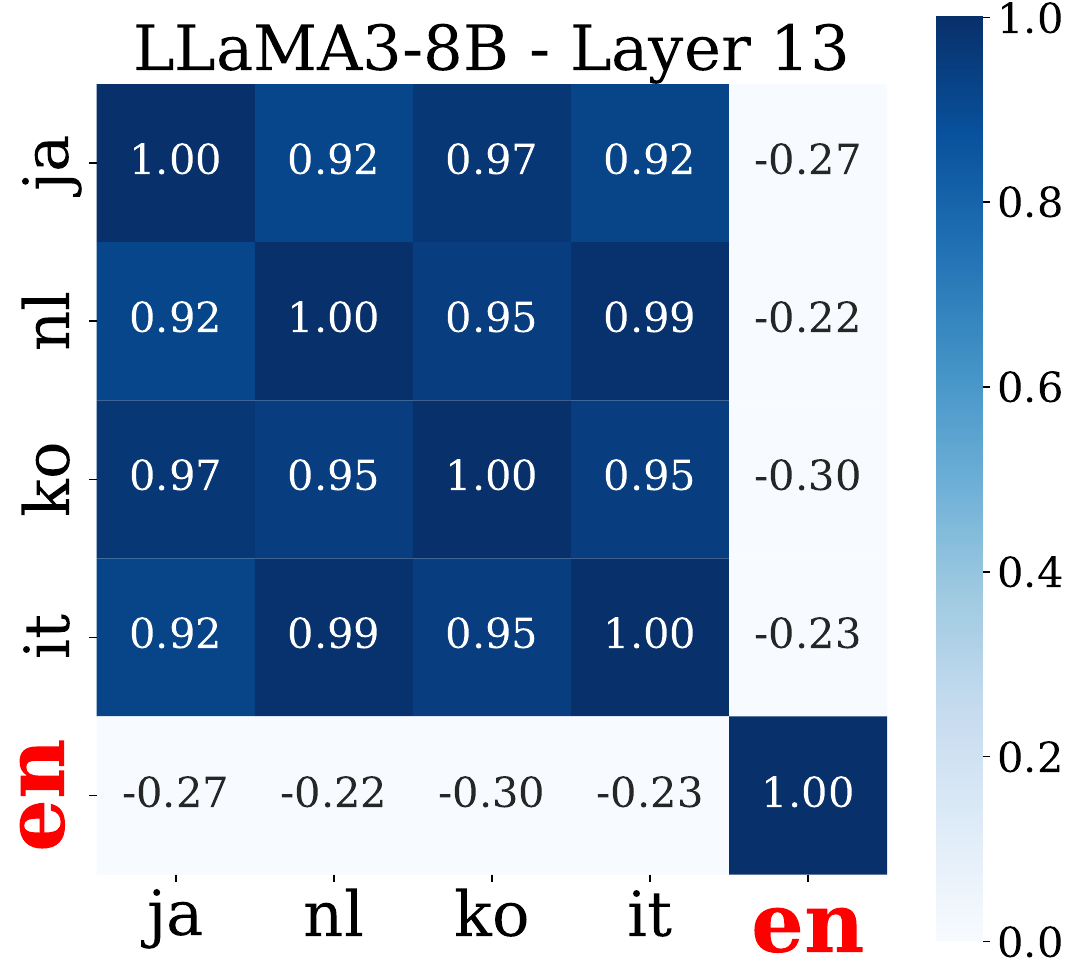}
  \includegraphics[width=0.15\linewidth]{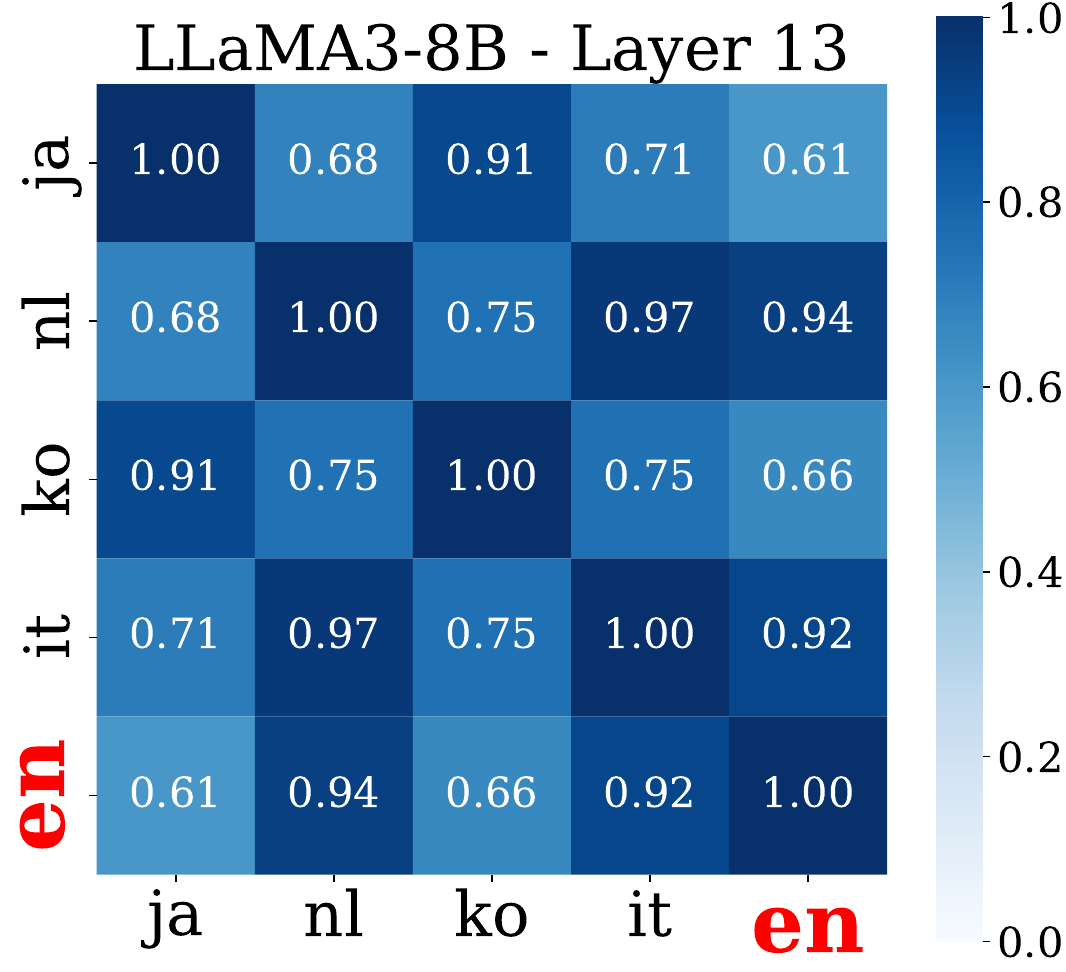}
  \includegraphics[width=0.15\linewidth]{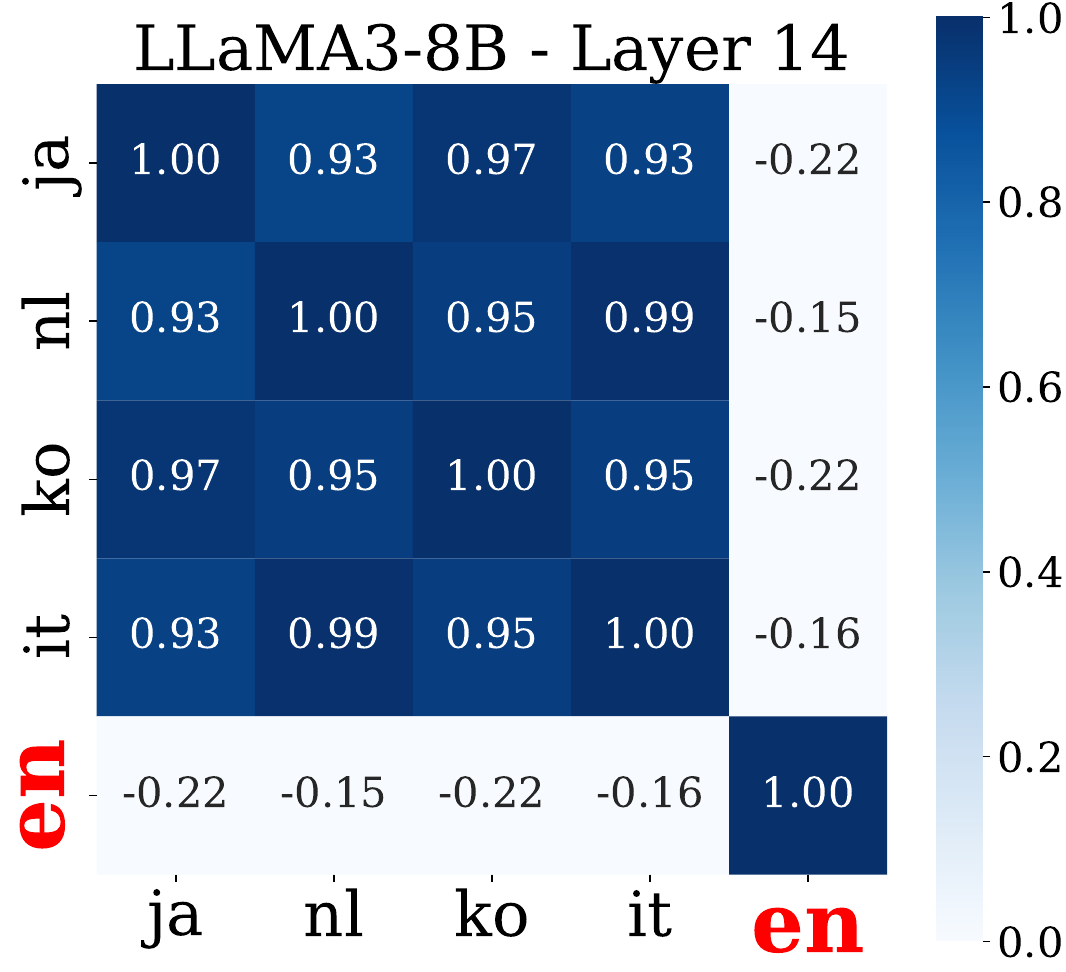}
  \includegraphics[width=0.15\linewidth]{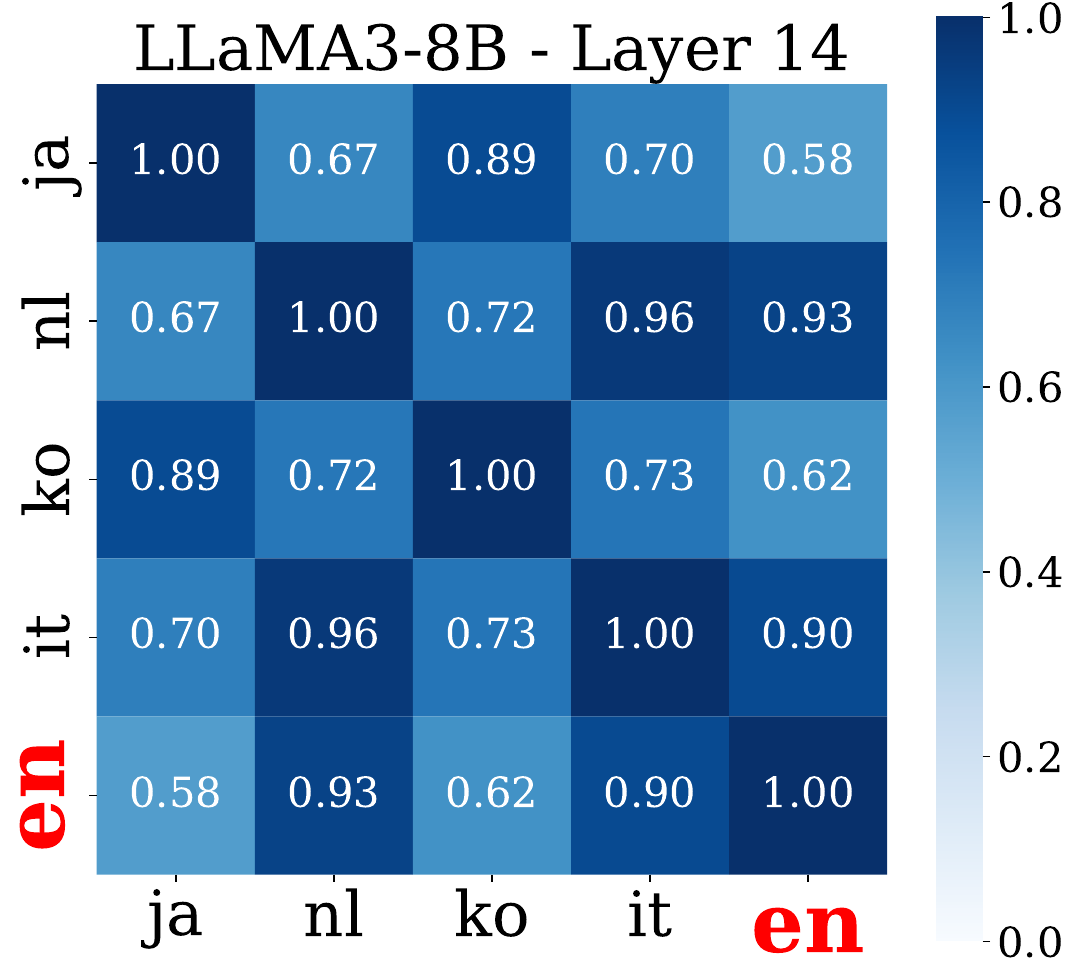}
  \includegraphics[width=0.15\linewidth]{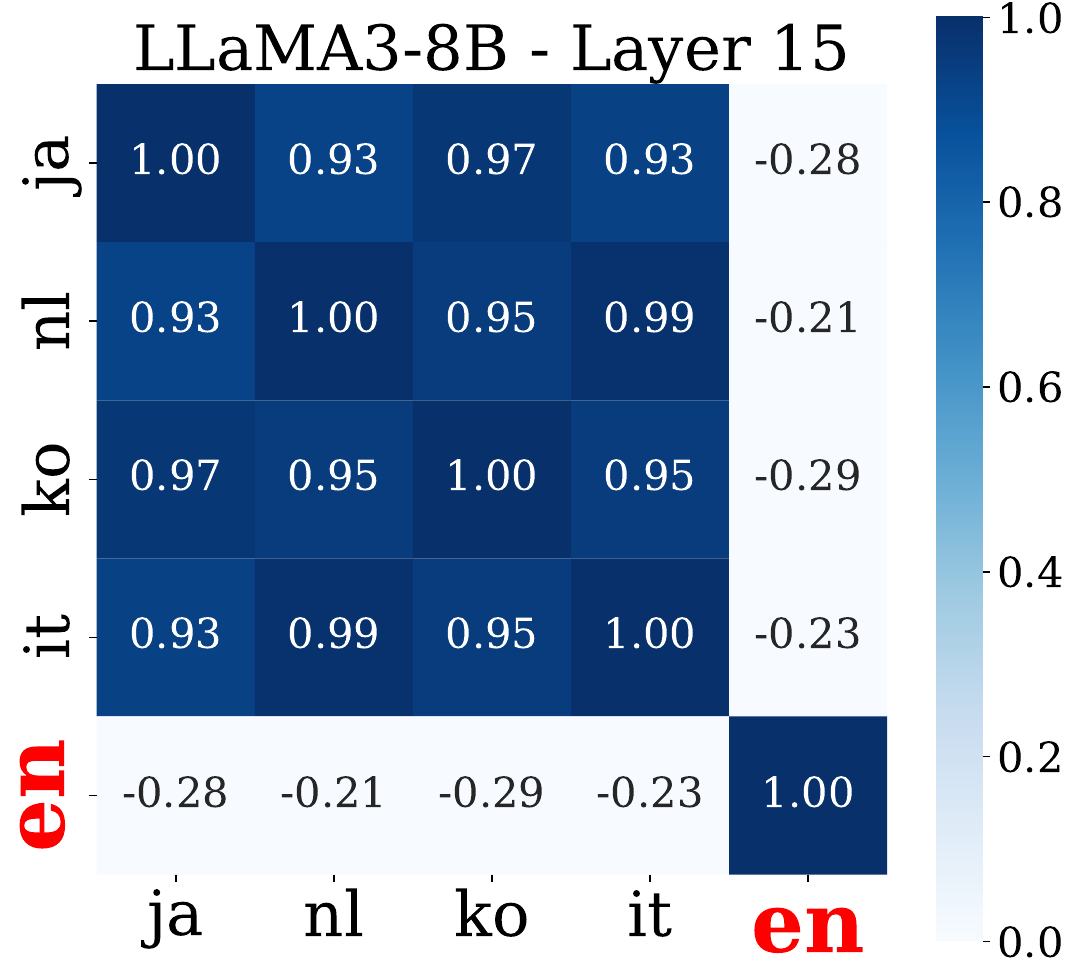}
  \includegraphics[width=0.15\linewidth]{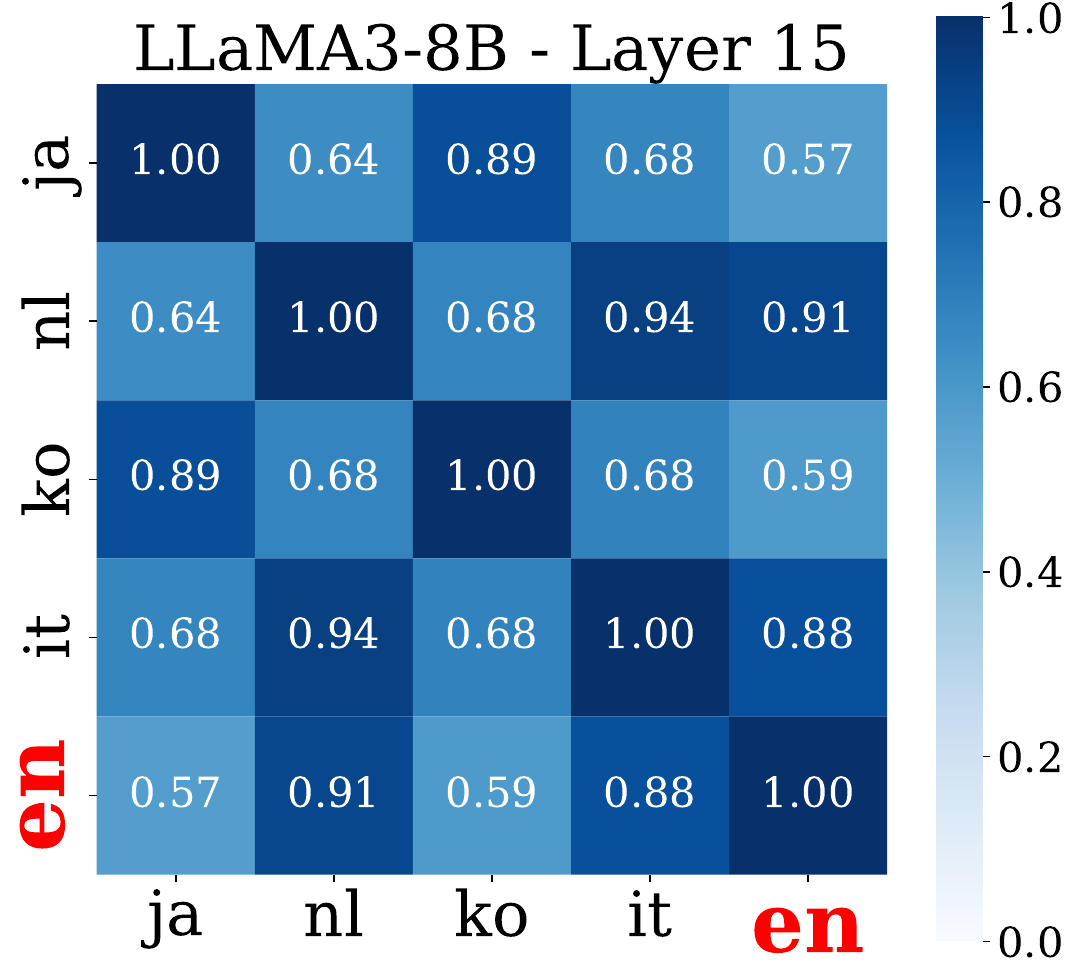}

  \begin{minipage}{0.15\linewidth}\centering \textbf{\textcolor{red}{layer 13 (Type-1)}}\end{minipage}
  \begin{minipage}{0.15\linewidth}\centering layer 13 (baseline)\end{minipage}
  \begin{minipage}{0.15\linewidth}\centering \textbf{\textcolor{red}{layer 14 (Type-1)}}\end{minipage}
  \begin{minipage}{0.15\linewidth}\centering layer 14 (baseline)\end{minipage}
  \begin{minipage}{0.15\linewidth}\centering \textbf{\textcolor{red}{layer 15 (Type-1)}}\end{minipage}
  \begin{minipage}{0.15\linewidth}\centering layer 15 (baseline)\end{minipage}

  \includegraphics[width=0.15\linewidth]{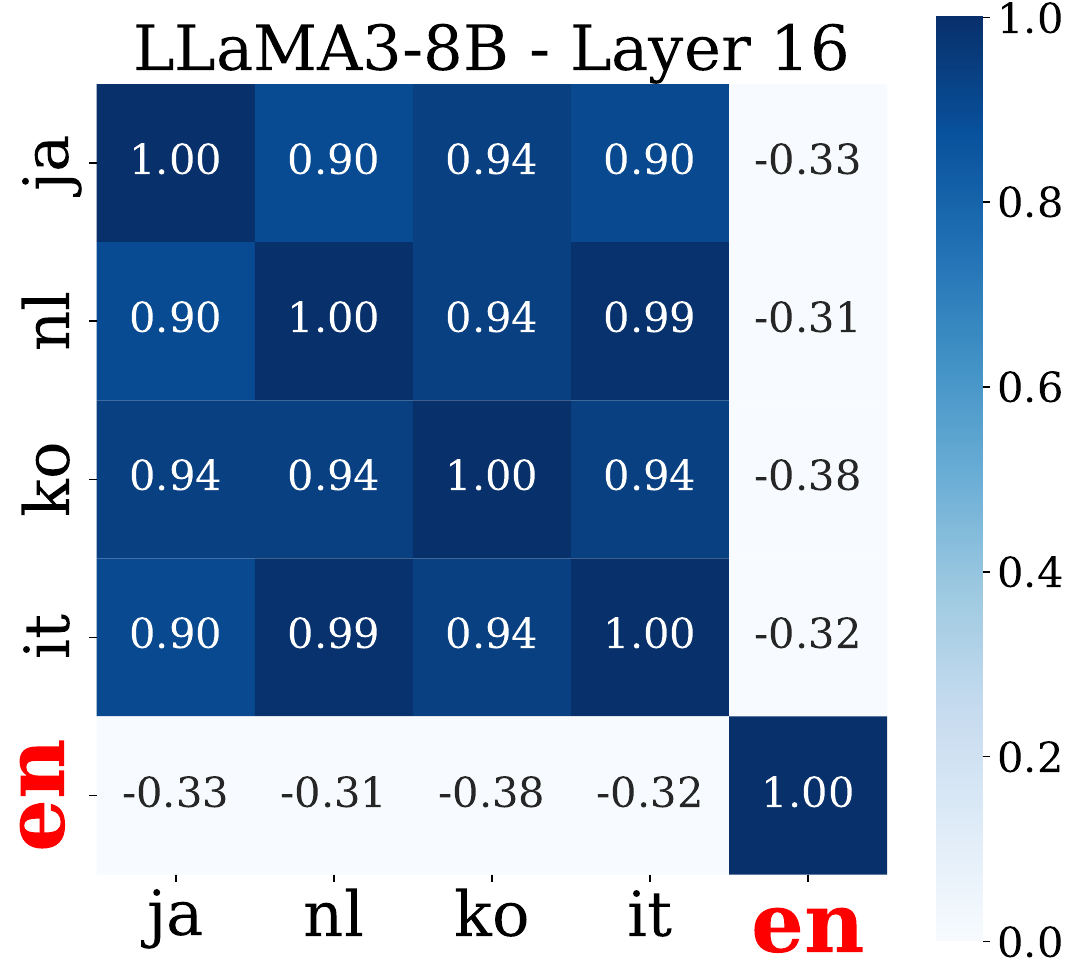}
  \includegraphics[width=0.15\linewidth]{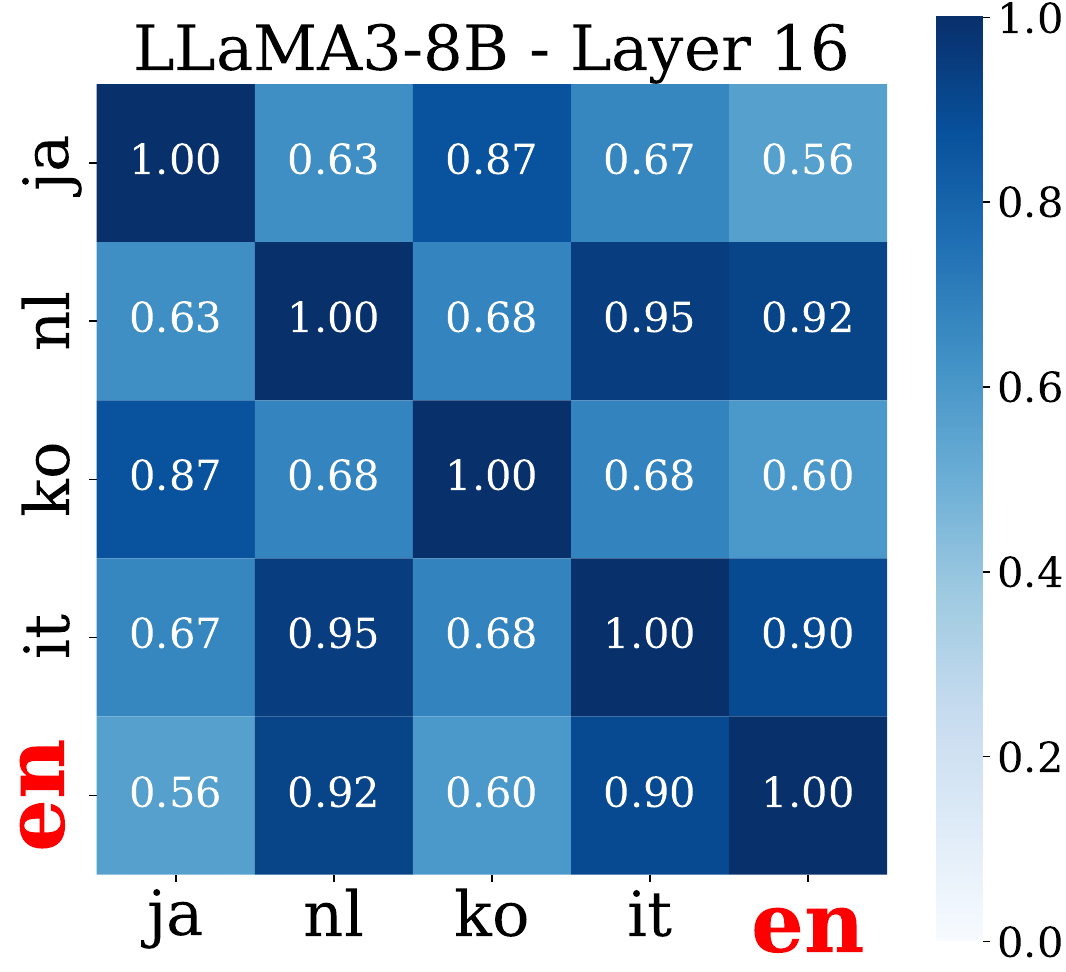}
  \includegraphics[width=0.15\linewidth]{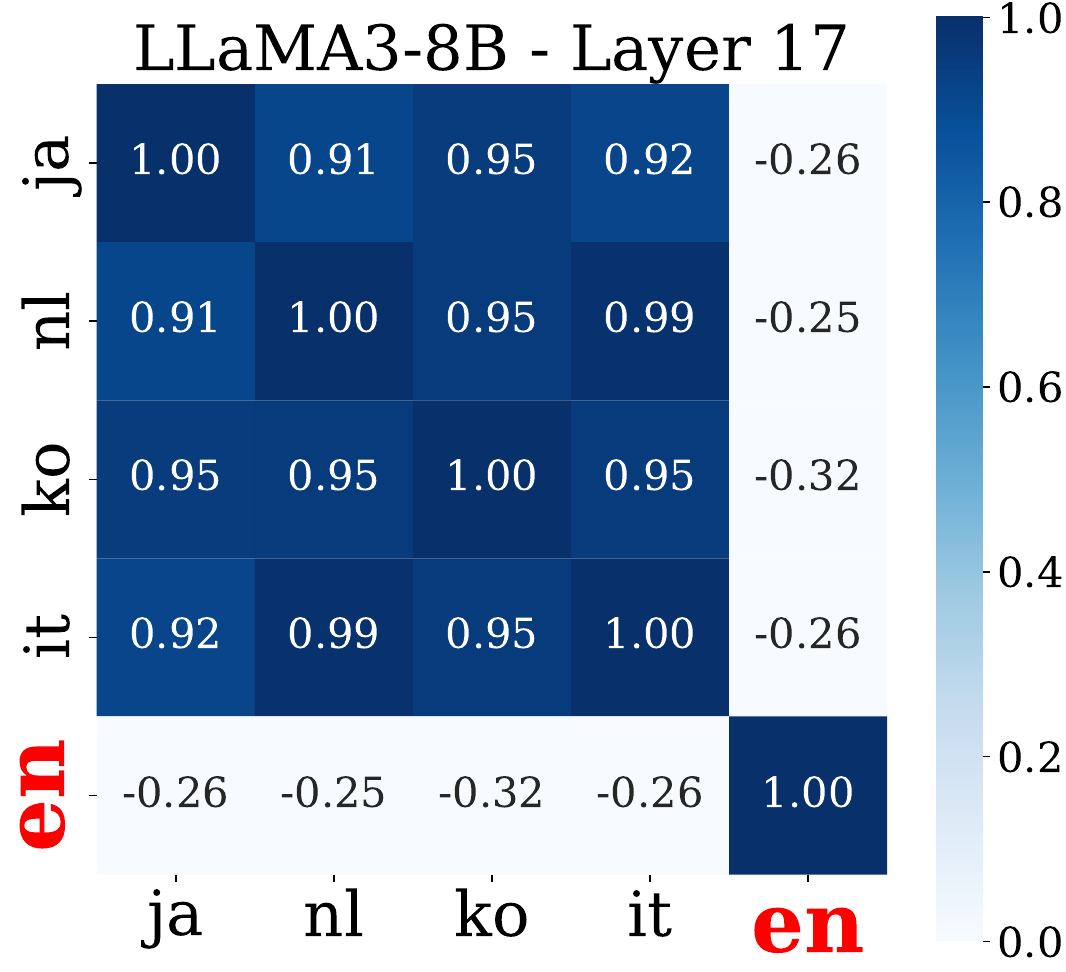}
  \includegraphics[width=0.15\linewidth]{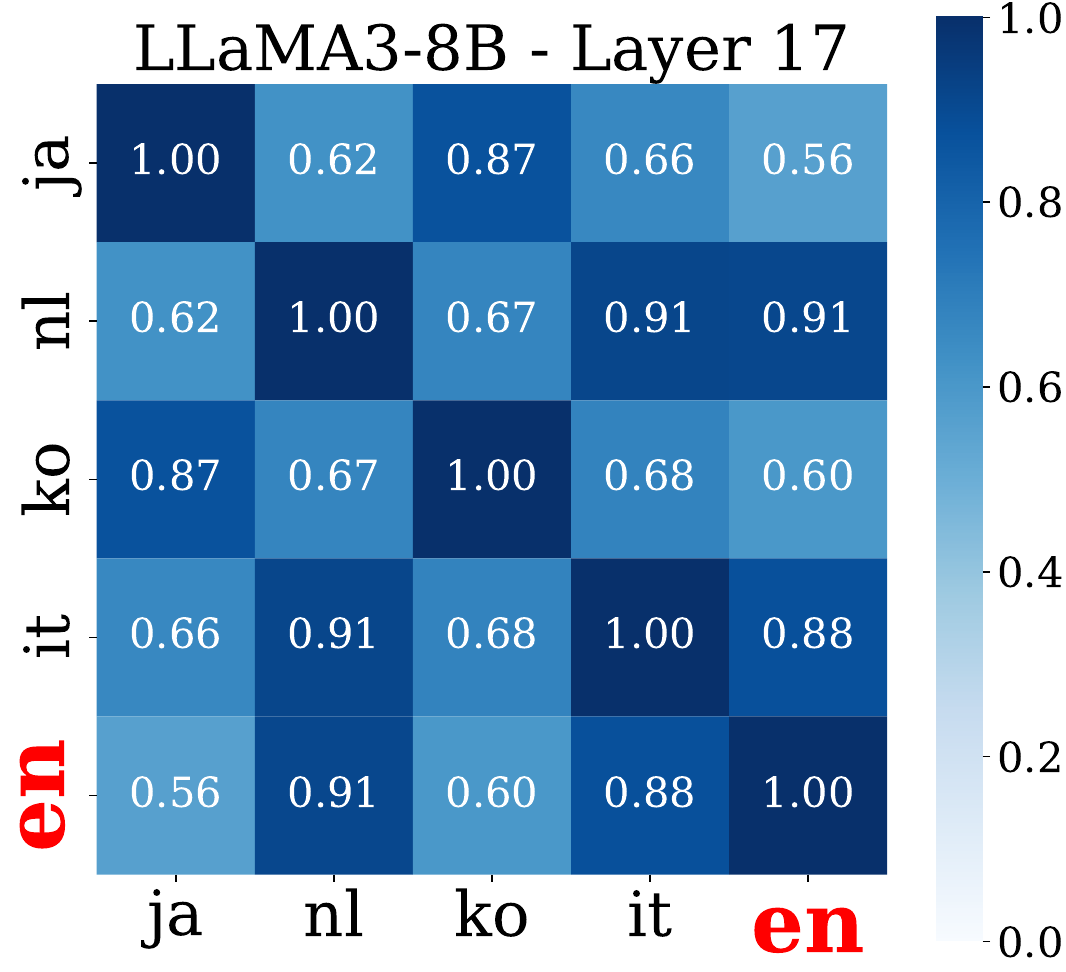}
  \includegraphics[width=0.15\linewidth]{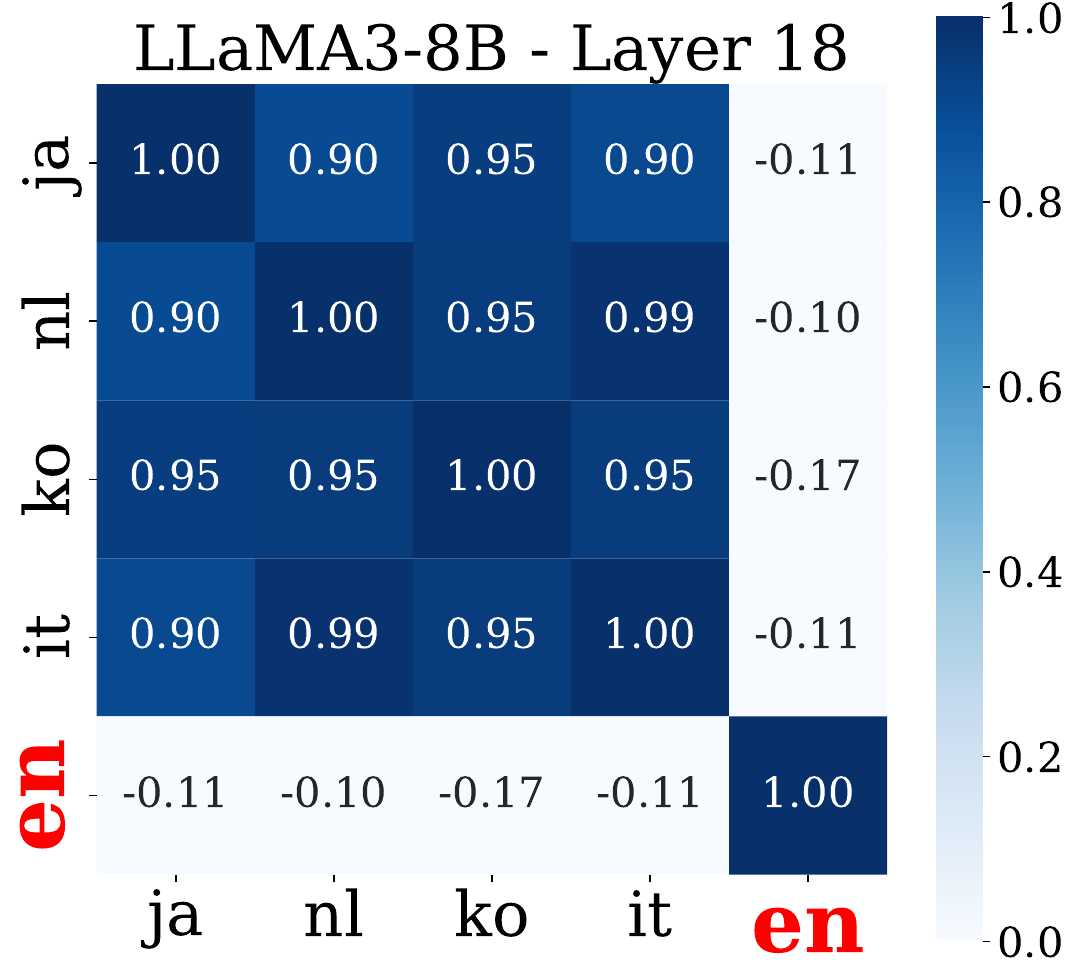}
  \includegraphics[width=0.15\linewidth]{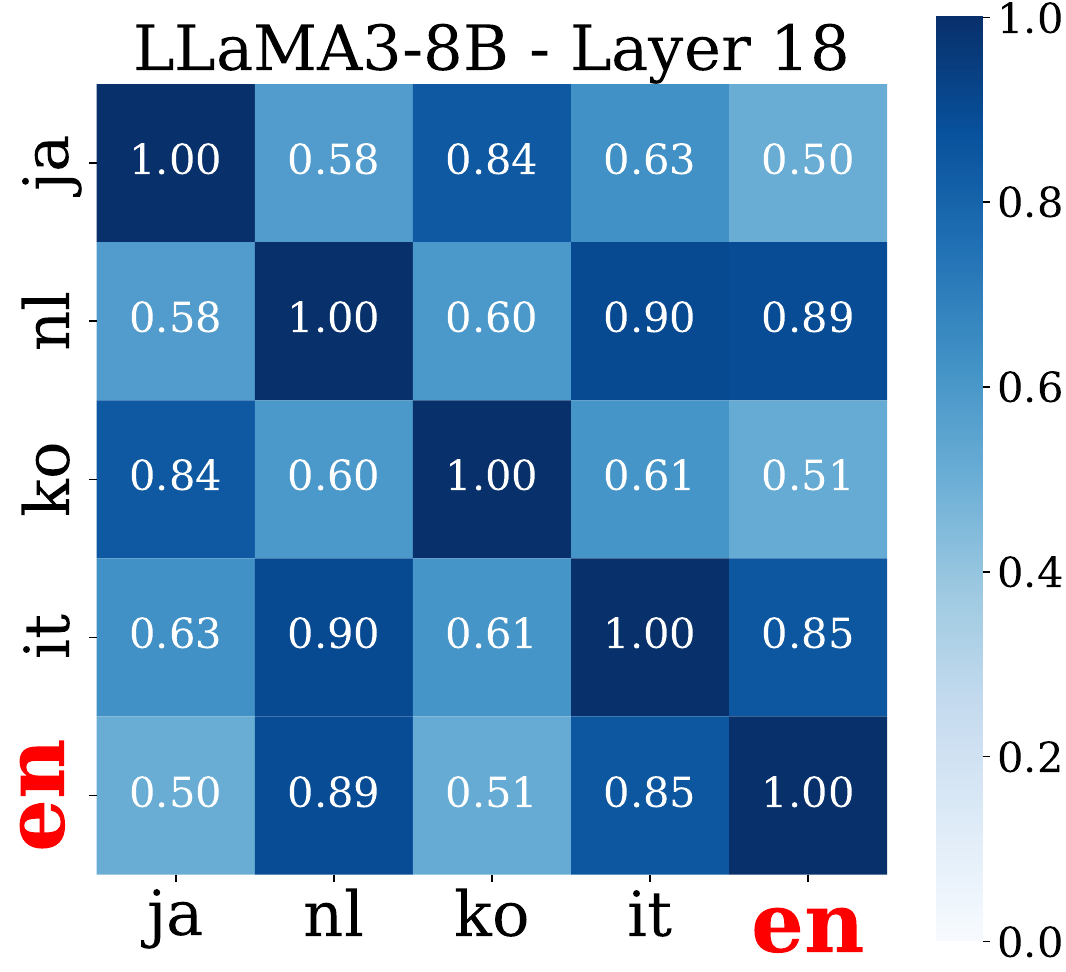}

  \begin{minipage}{0.15\linewidth}\centering \textbf{\textcolor{red}{layer 16 (Type-1)}}\end{minipage}
  \begin{minipage}{0.15\linewidth}\centering layer 16 (baseline)\end{minipage}
  \begin{minipage}{0.15\linewidth}\centering \textbf{\textcolor{red}{layer 17 (Type-1)}}\end{minipage}
  \begin{minipage}{0.15\linewidth}\centering layer 17 (baseline)\end{minipage}
  \begin{minipage}{0.15\linewidth}\centering \textbf{\textcolor{red}{layer 18 (Type-1)}}\end{minipage}
  \begin{minipage}{0.15\linewidth}\centering layer 18 (baseline)\end{minipage}

  \includegraphics[width=0.15\linewidth]{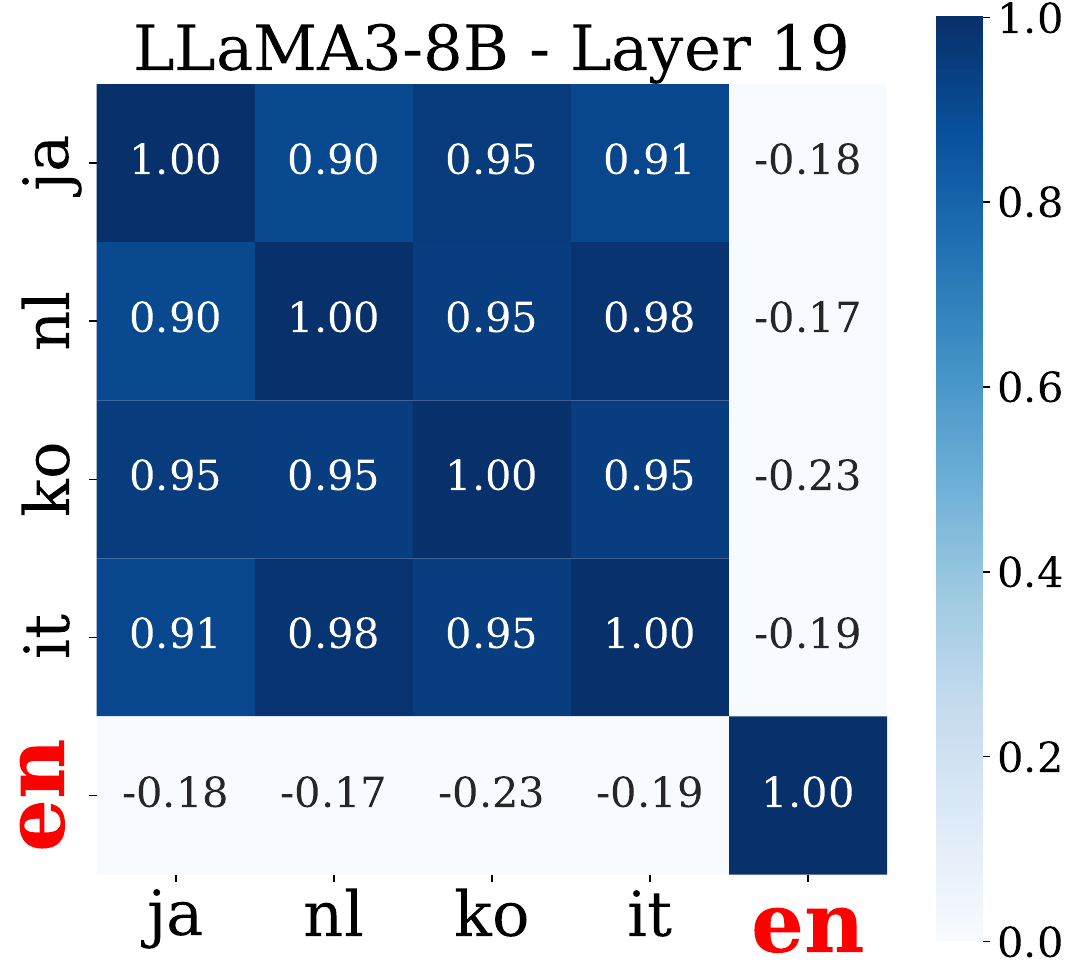}
  \includegraphics[width=0.15\linewidth]{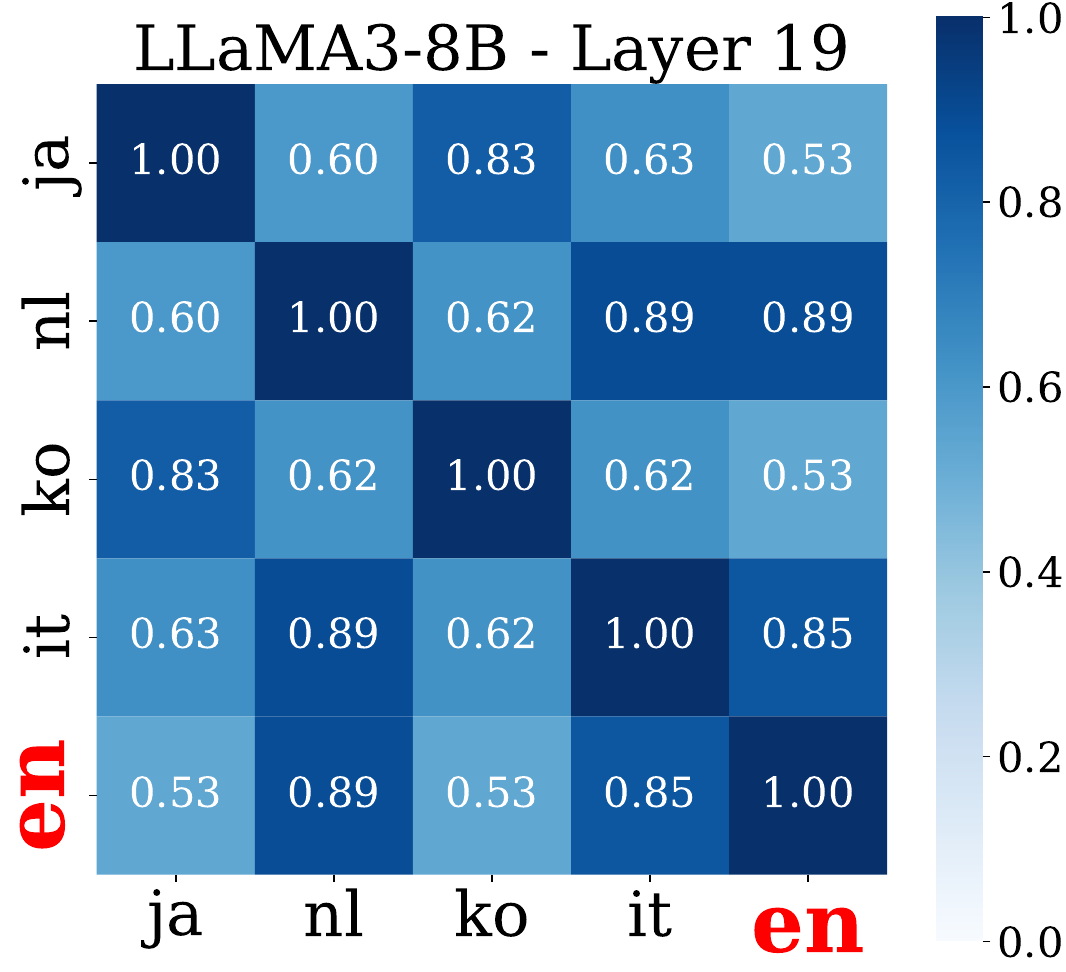}
  \includegraphics[width=0.15\linewidth]{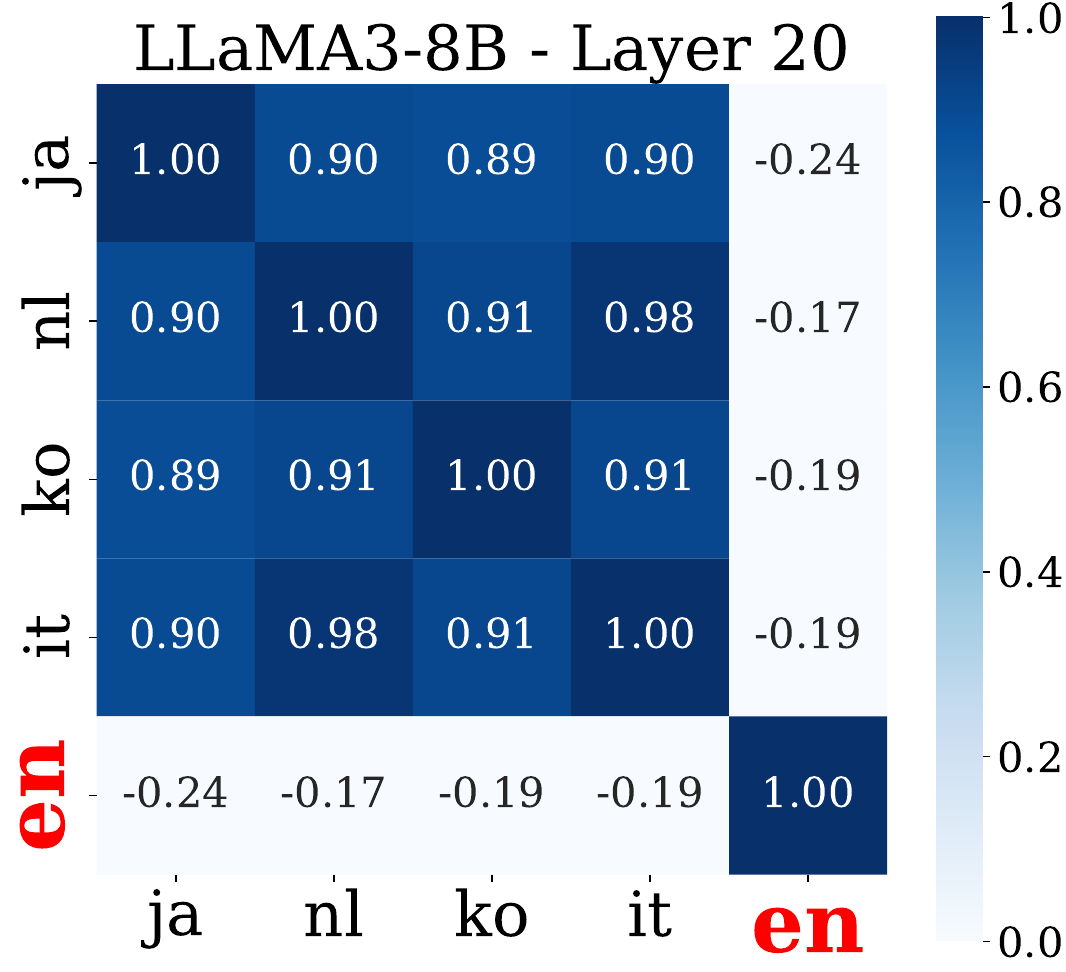}
  \includegraphics[width=0.15\linewidth]{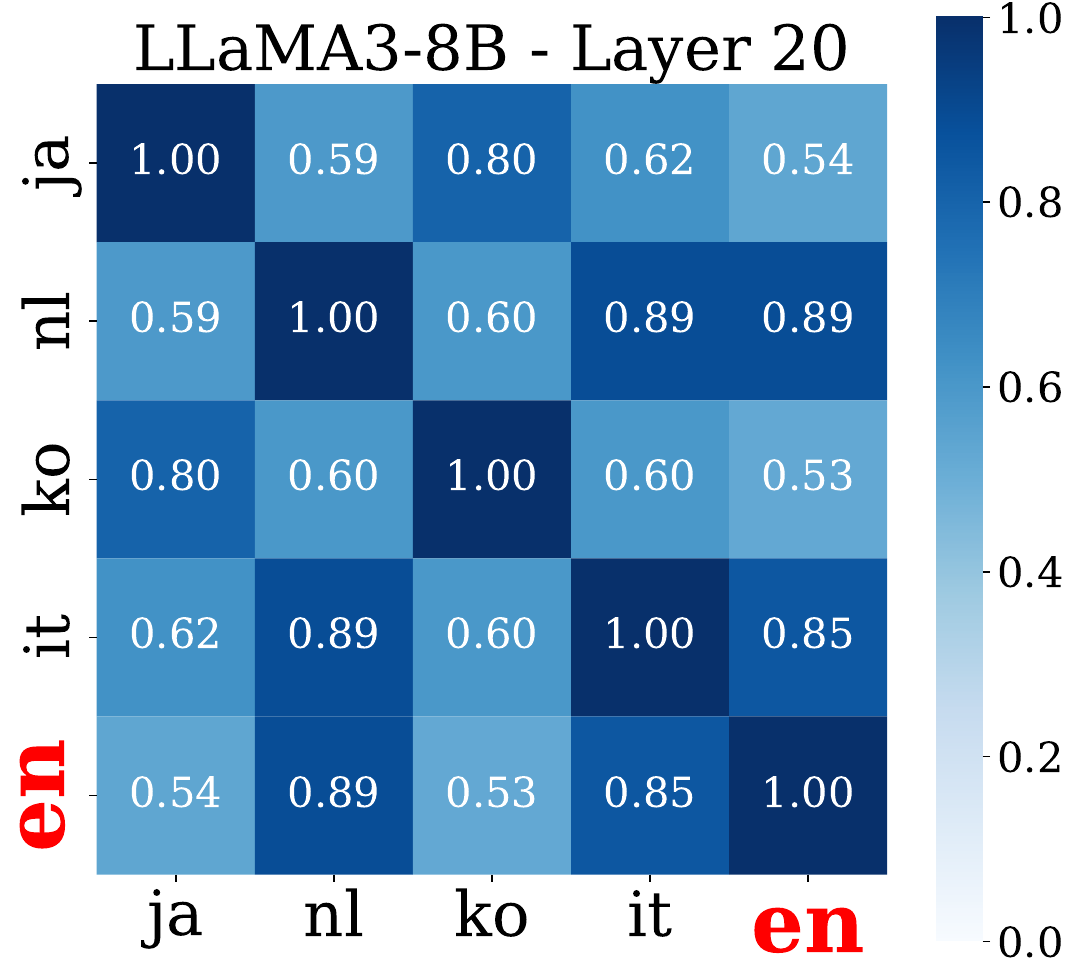}

  \begin{minipage}{0.15\linewidth}\centering \textbf{\textcolor{red}{layer 19 (Type-1)}}\end{minipage}
  \begin{minipage}{0.15\linewidth}\centering layer 19 (baseline)\end{minipage}
  \begin{minipage}{0.15\linewidth}\centering \textbf{\textcolor{red}{layer 20 (Type-1)}}\end{minipage}
  \begin{minipage}{0.15\linewidth}\centering layer 20 (baseline)\end{minipage}

  \caption{\textbf{Distance among language latent spaces while deactivating Top-1k Type-1 Transfer Neurons (LLaMA3-8B)}. \textbf{\textcolor{red}{Layer (Type-1)}} indicates the result of deactivating the top-1k Type-1 Transfer Neurons, whereas “baseline” refers to the result of deactivating 1k randomly sampled neurons from the same layers as the Type-1 Transfer Neurons.}
  \label{fig:appendix:distance centroids among langage subspaces deactivating type1 llama3}
\end{figure*}

\begin{figure*}[t]
  \centering

  \includegraphics[width=0.15\linewidth]{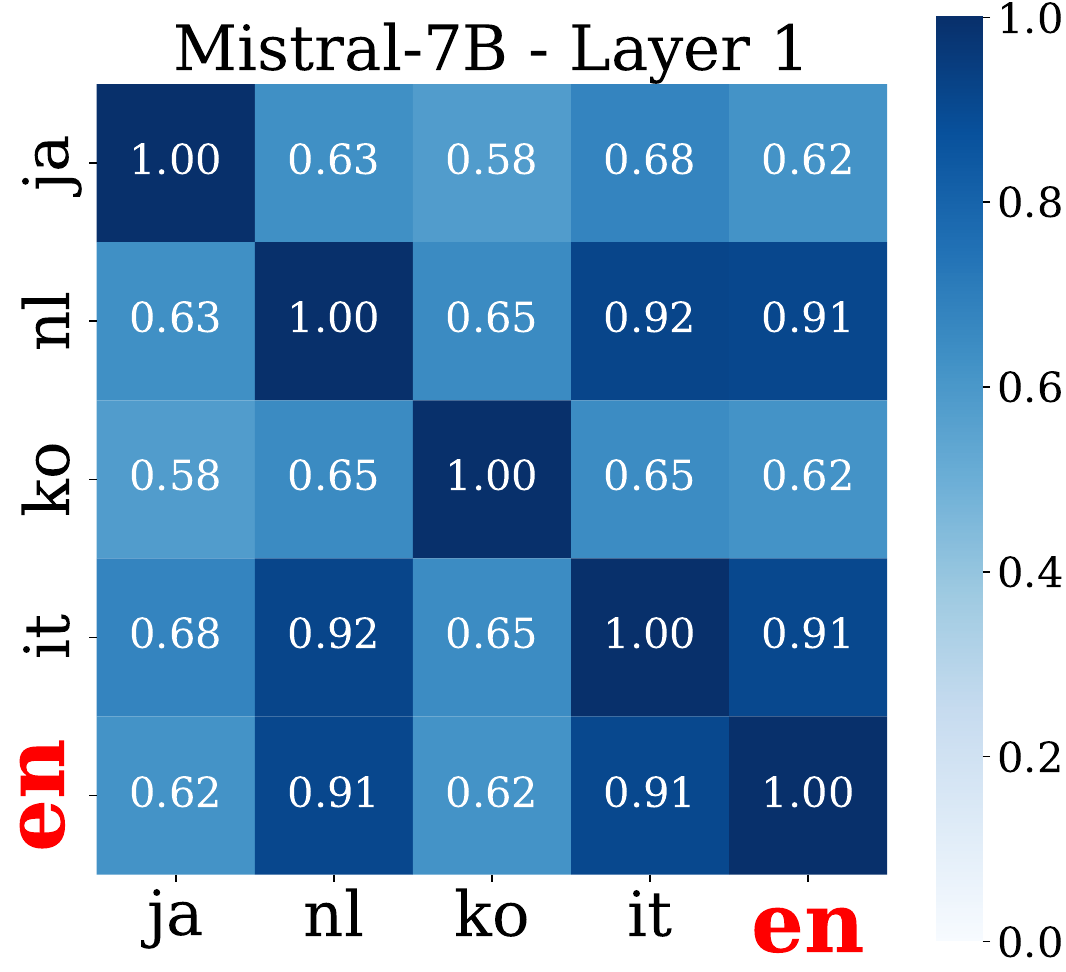}
  \includegraphics[width=0.15\linewidth]{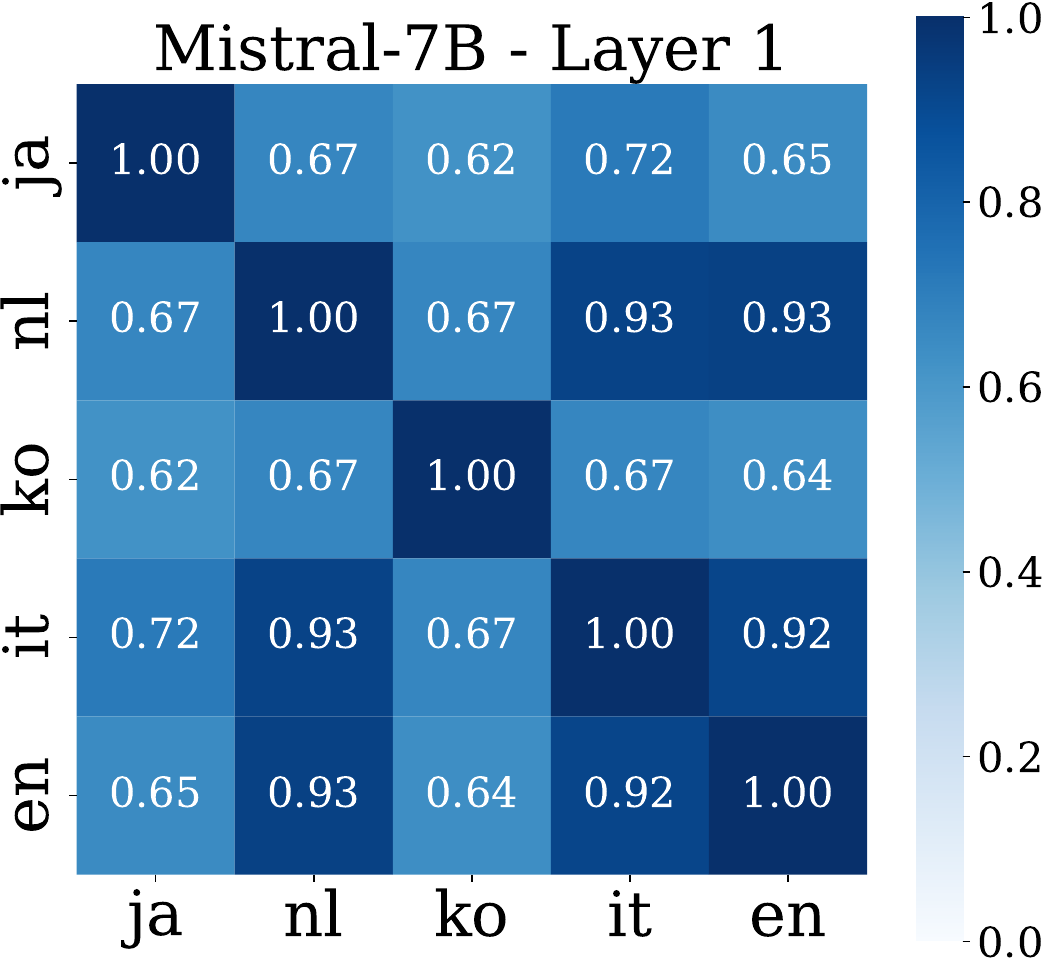}
  \includegraphics[width=0.15\linewidth]{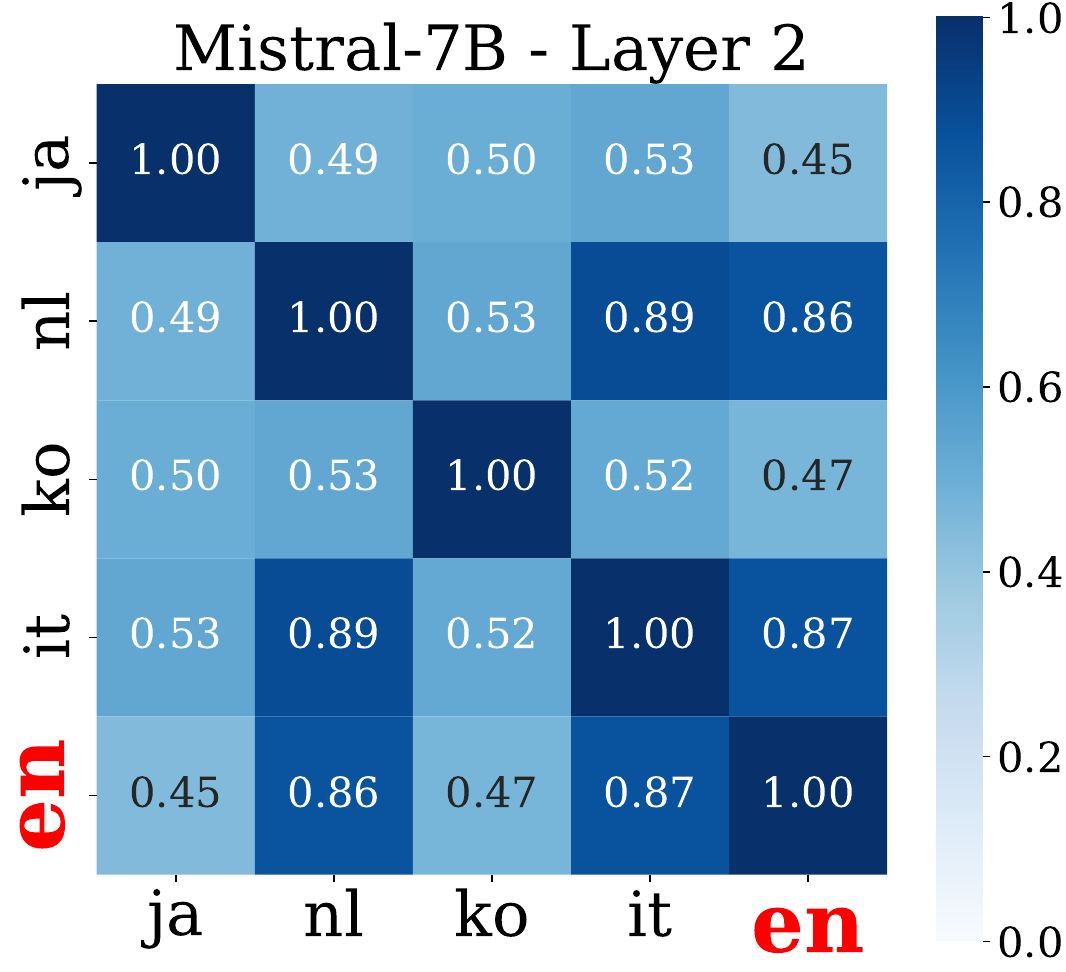}
  \includegraphics[width=0.15\linewidth]{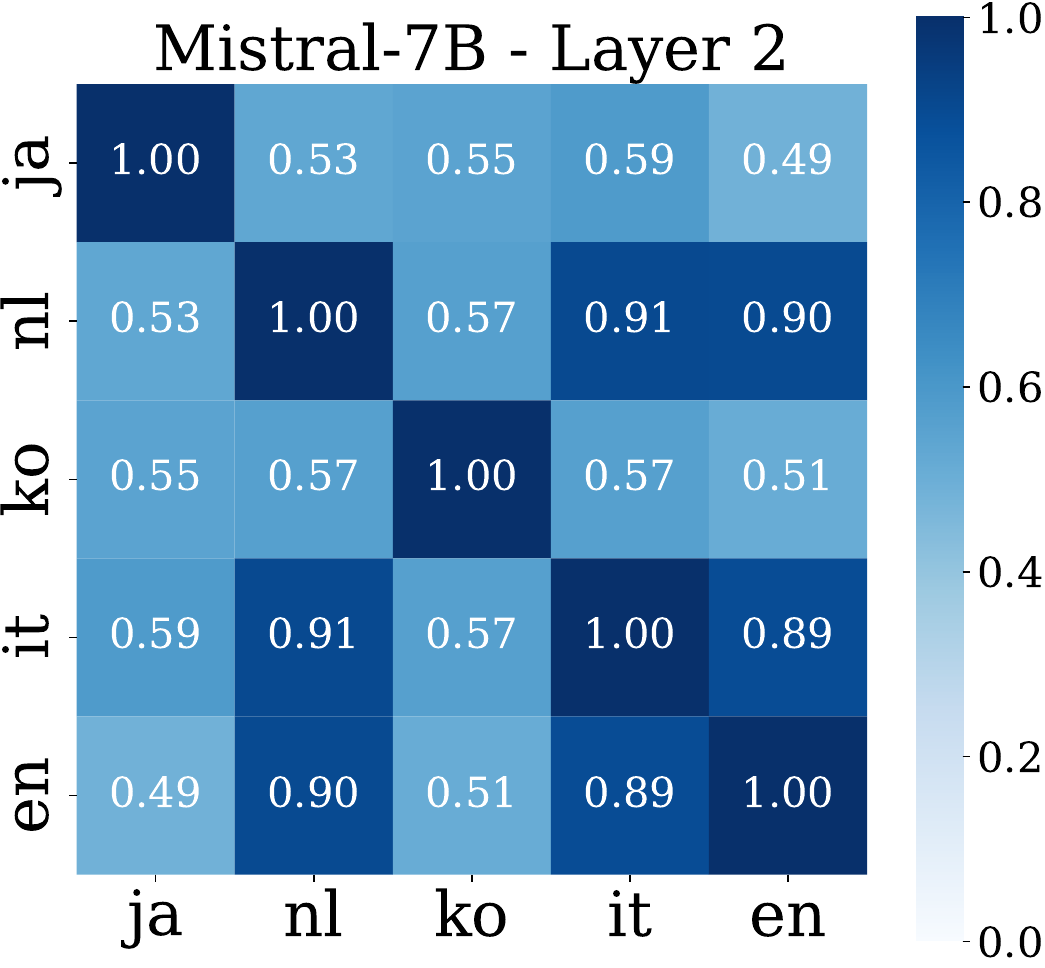}
  \includegraphics[width=0.15\linewidth]{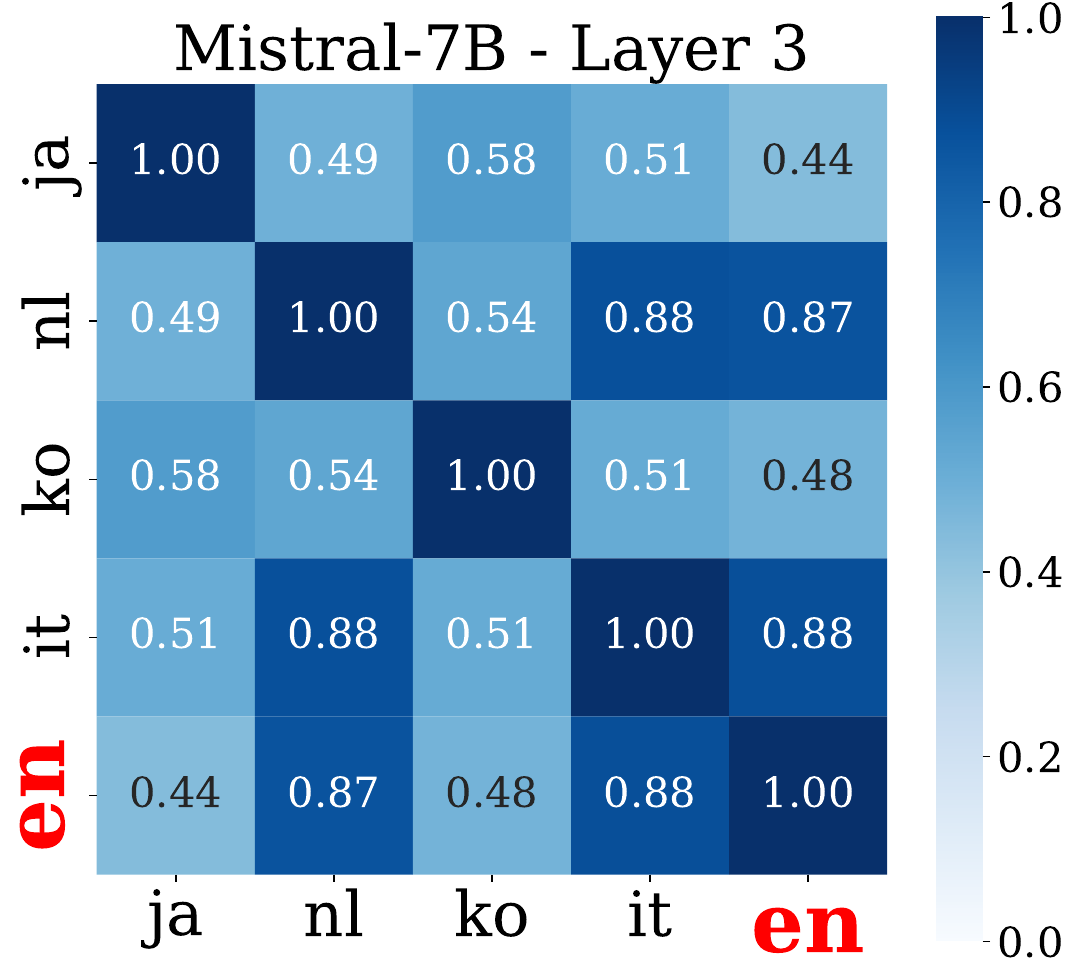}
  \includegraphics[width=0.15\linewidth]{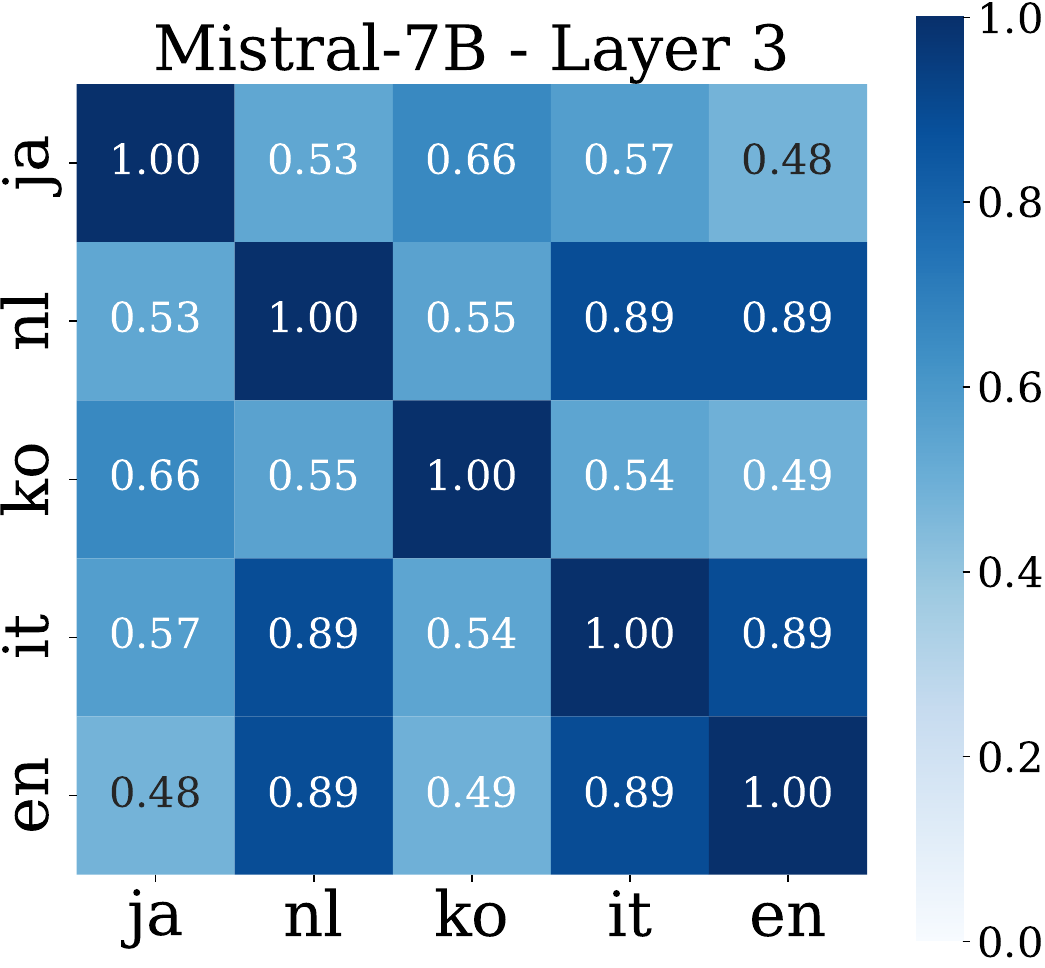}

  \begin{minipage}{0.15\linewidth}\centering \textbf{\textcolor{red}{layer 1 (Type-1)}}\end{minipage}
  \begin{minipage}{0.15\linewidth}\centering layer 1 (baseline)\end{minipage}
  \begin{minipage}{0.15\linewidth}\centering \textbf{\textcolor{red}{layer 2 (Type-1)}}\end{minipage}
  \begin{minipage}{0.15\linewidth}\centering layer 2 (baseline)\end{minipage}
  \begin{minipage}{0.15\linewidth}\centering \textbf{\textcolor{red}{layer 3 (Type-1)}}\end{minipage}
  \begin{minipage}{0.15\linewidth}\centering layer 3 (baseline)\end{minipage}

  \includegraphics[width=0.15\linewidth]{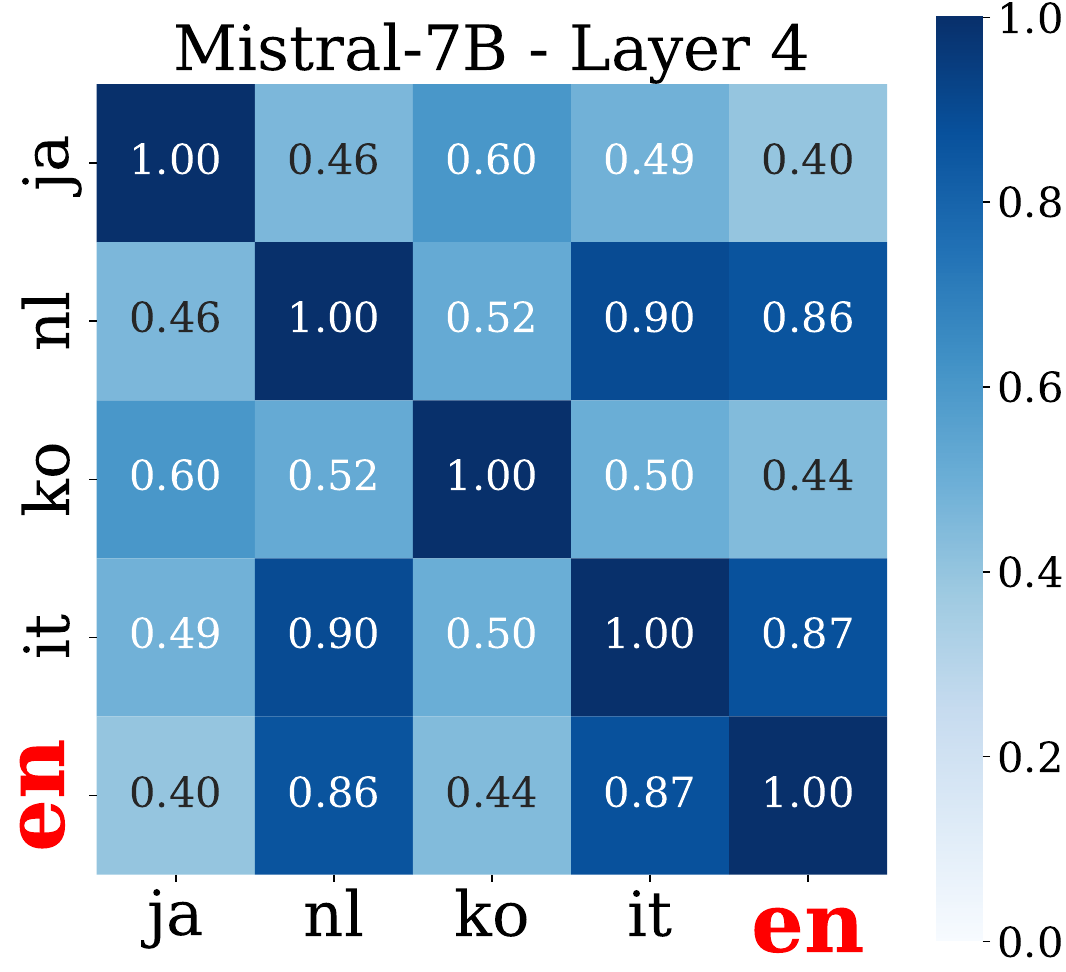}
  \includegraphics[width=0.15\linewidth]{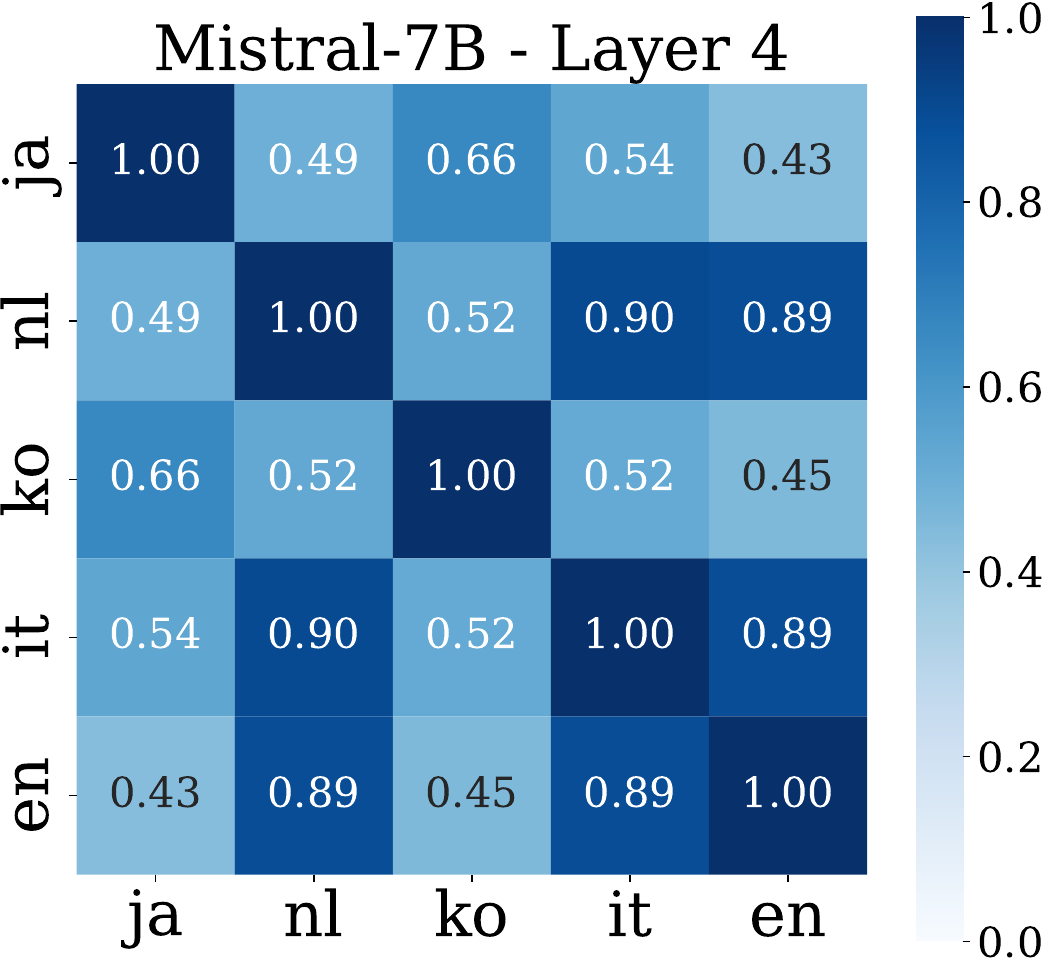}
  \includegraphics[width=0.15\linewidth]{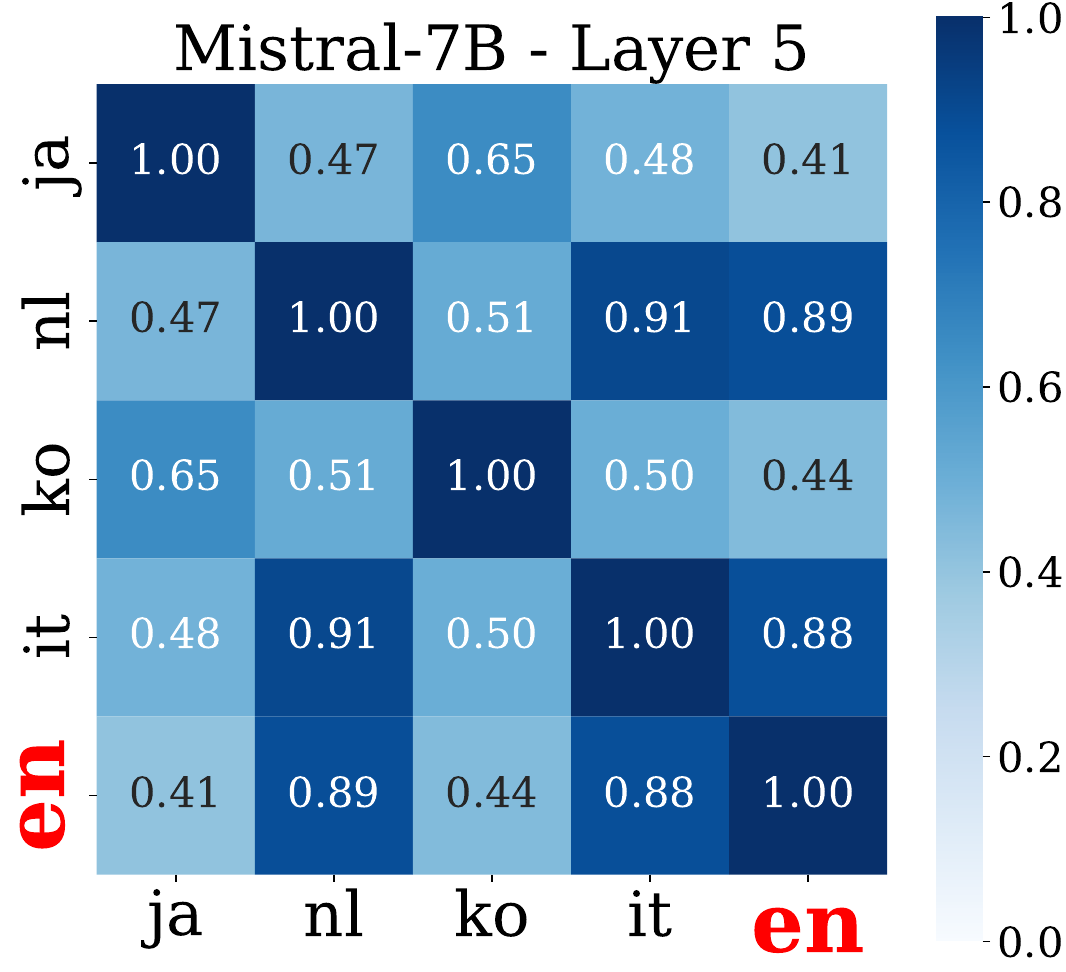}
  \includegraphics[width=0.15\linewidth]{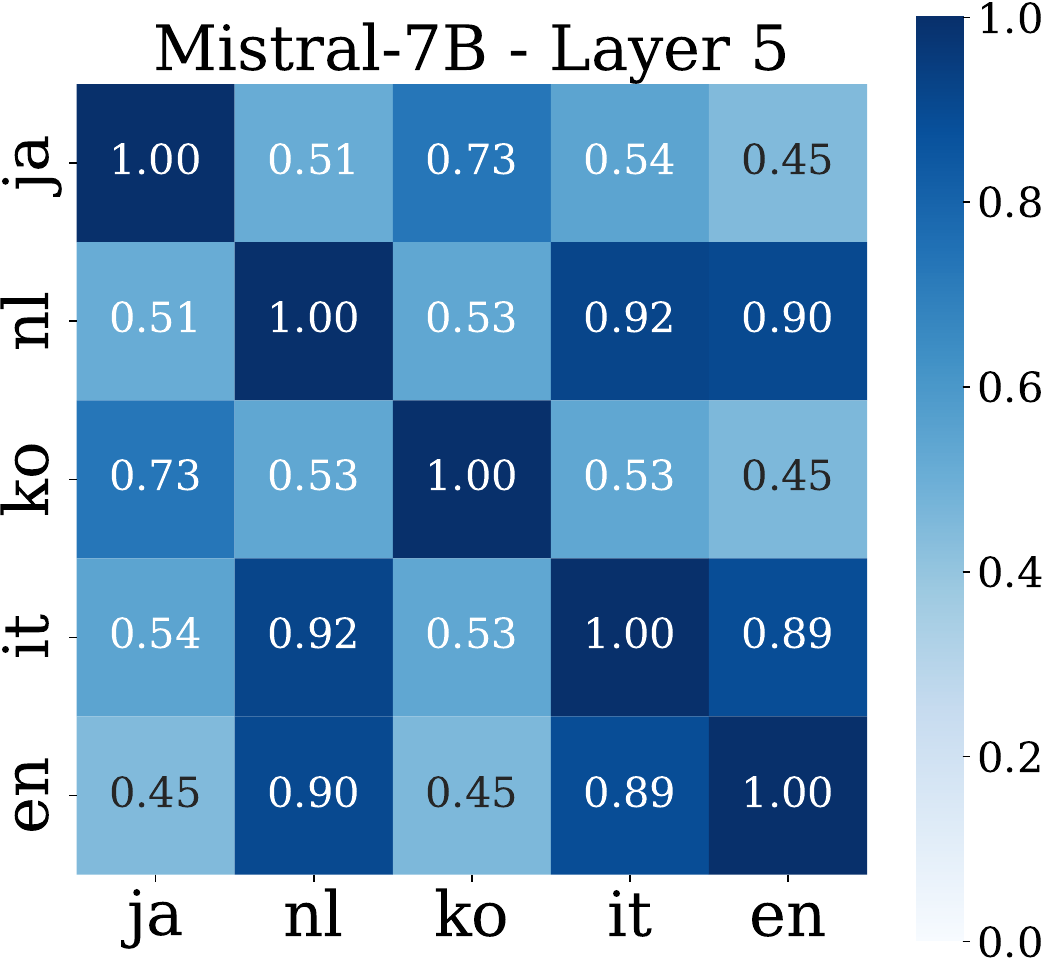}
  \includegraphics[width=0.15\linewidth]{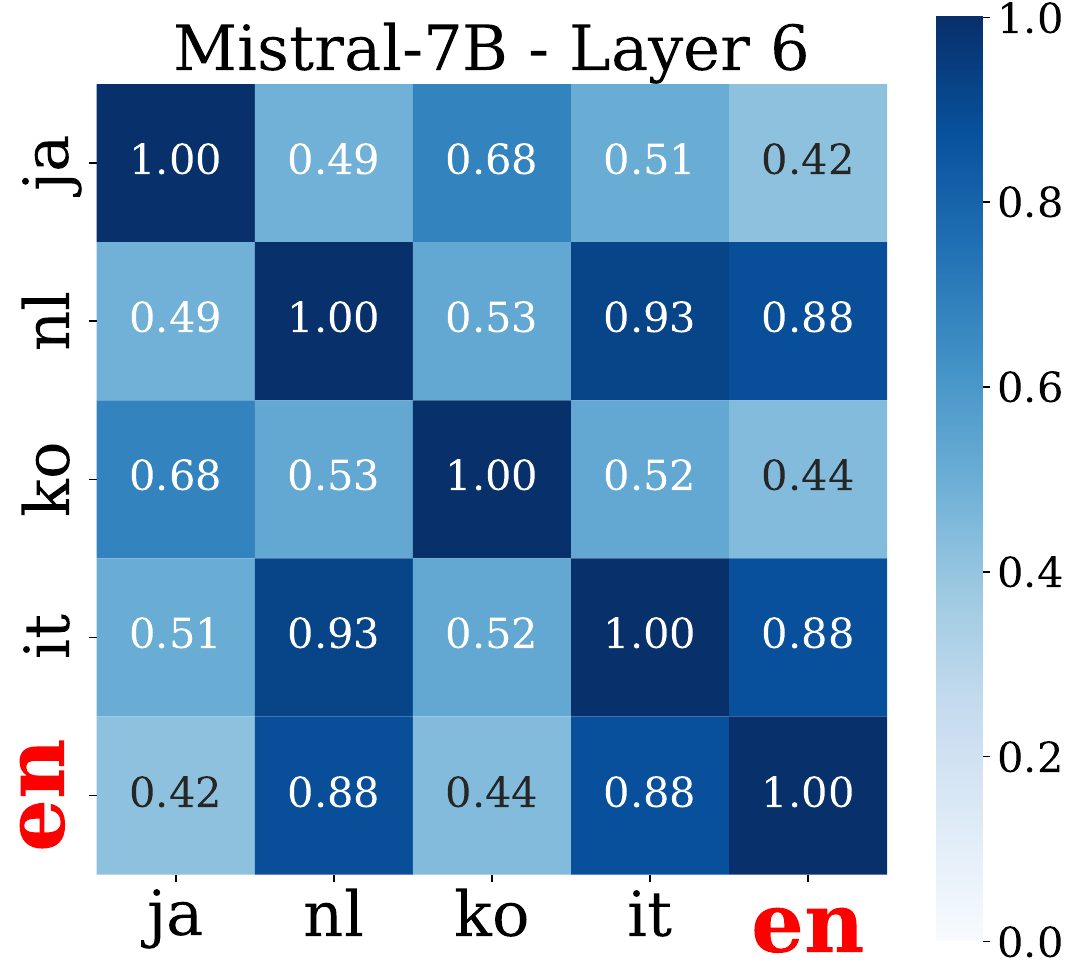}
  \includegraphics[width=0.15\linewidth]{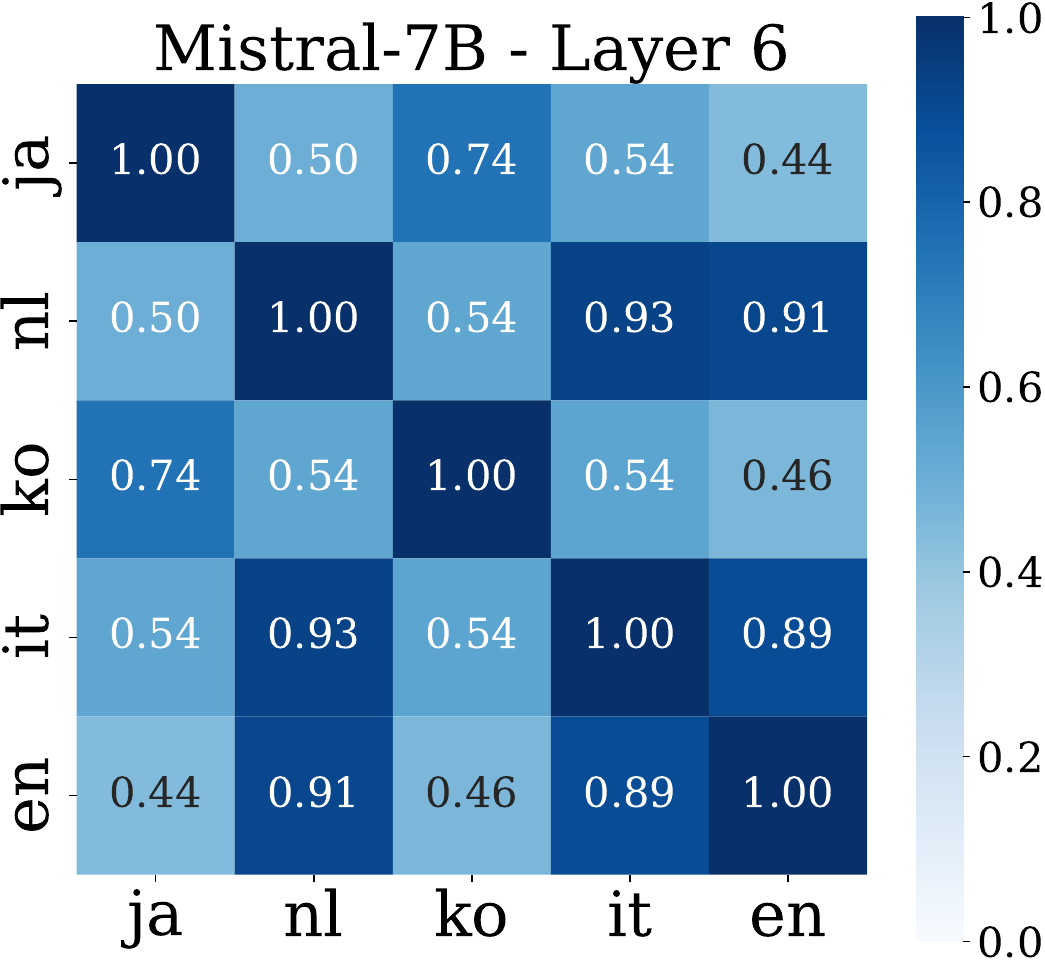}

  \begin{minipage}{0.15\linewidth}\centering \textbf{\textcolor{red}{layer 4 (Type-1)}}\end{minipage}
  \begin{minipage}{0.15\linewidth}\centering layer 4 (baseline)\end{minipage}
  \begin{minipage}{0.15\linewidth}\centering \textbf{\textcolor{red}{layer 5 (Type-1)}}\end{minipage}
  \begin{minipage}{0.15\linewidth}\centering layer 5 (baseline)\end{minipage}
  \begin{minipage}{0.15\linewidth}\centering \textbf{\textcolor{red}{layer 6 (Type-1)}}\end{minipage}
  \begin{minipage}{0.15\linewidth}\centering layer 6 (baseline)\end{minipage}

  \includegraphics[width=0.15\linewidth]{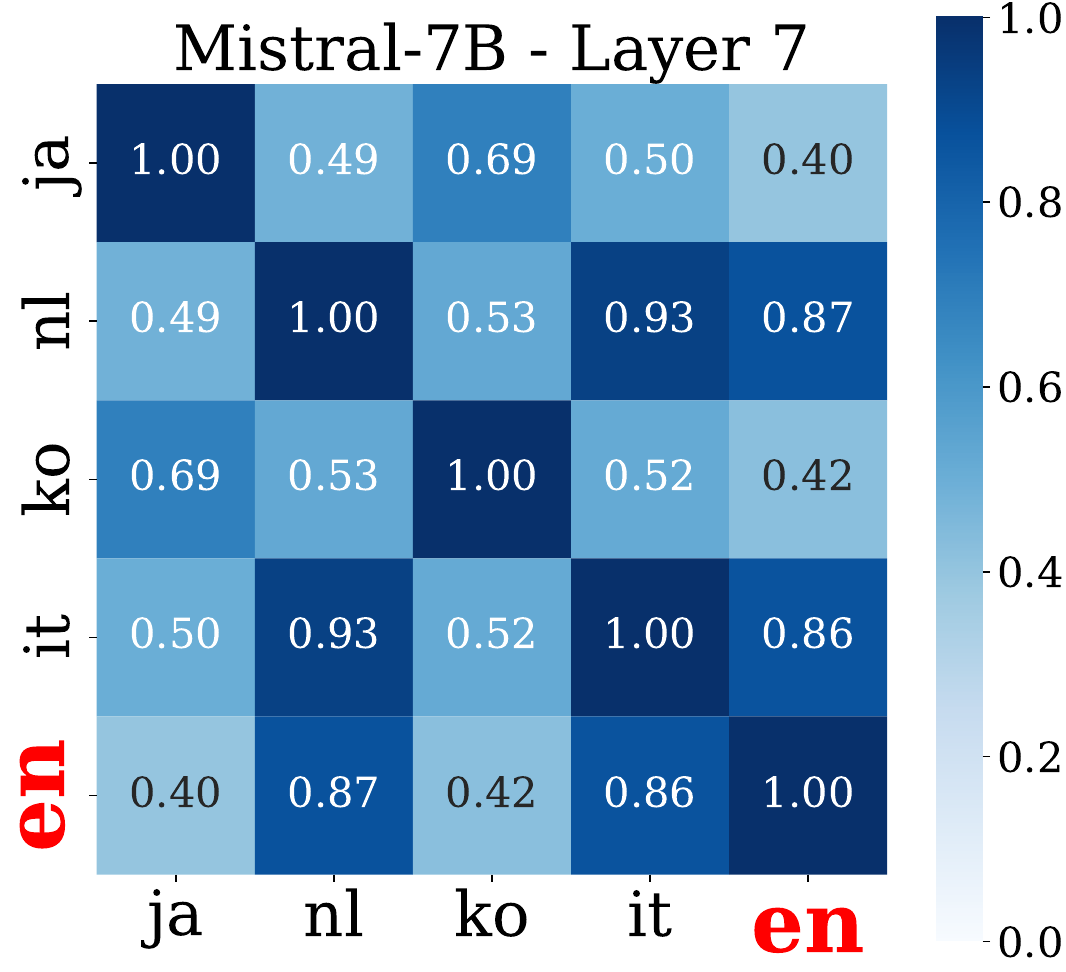}
  \includegraphics[width=0.15\linewidth]{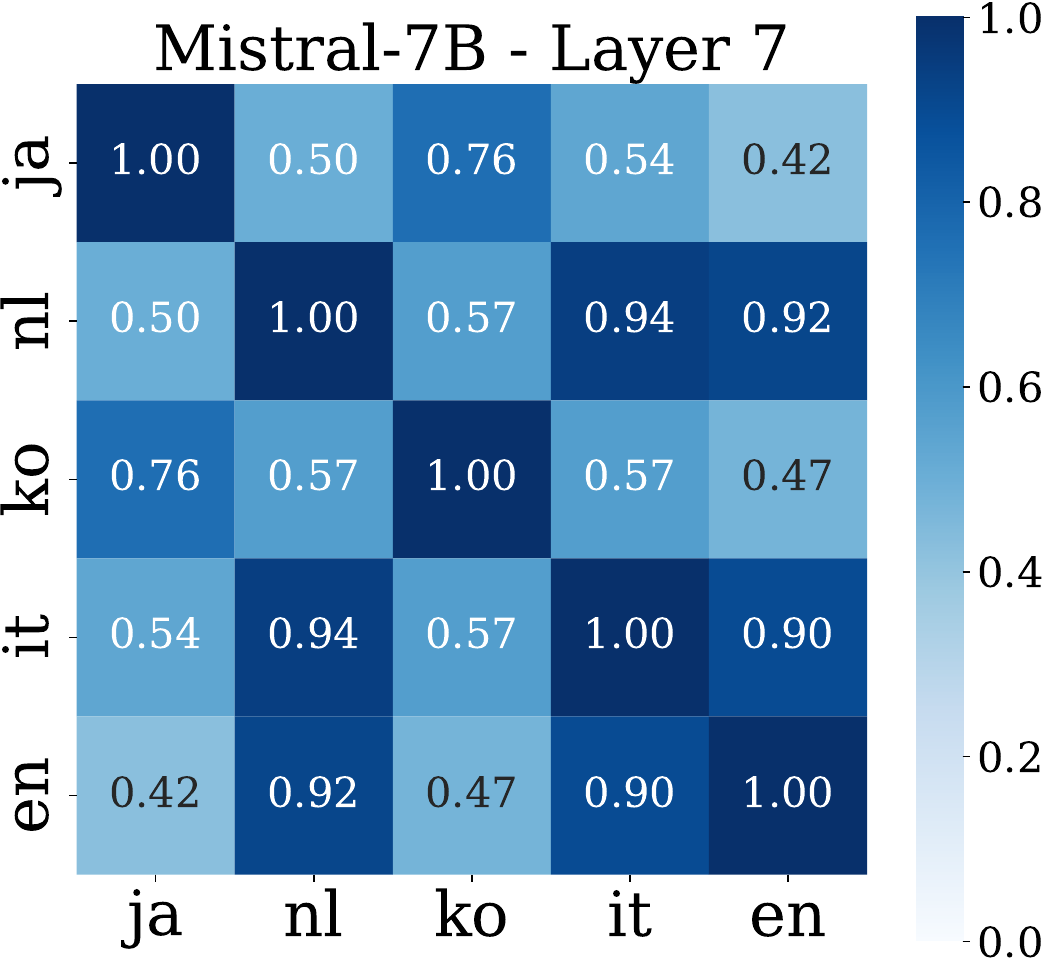}
  \includegraphics[width=0.15\linewidth]{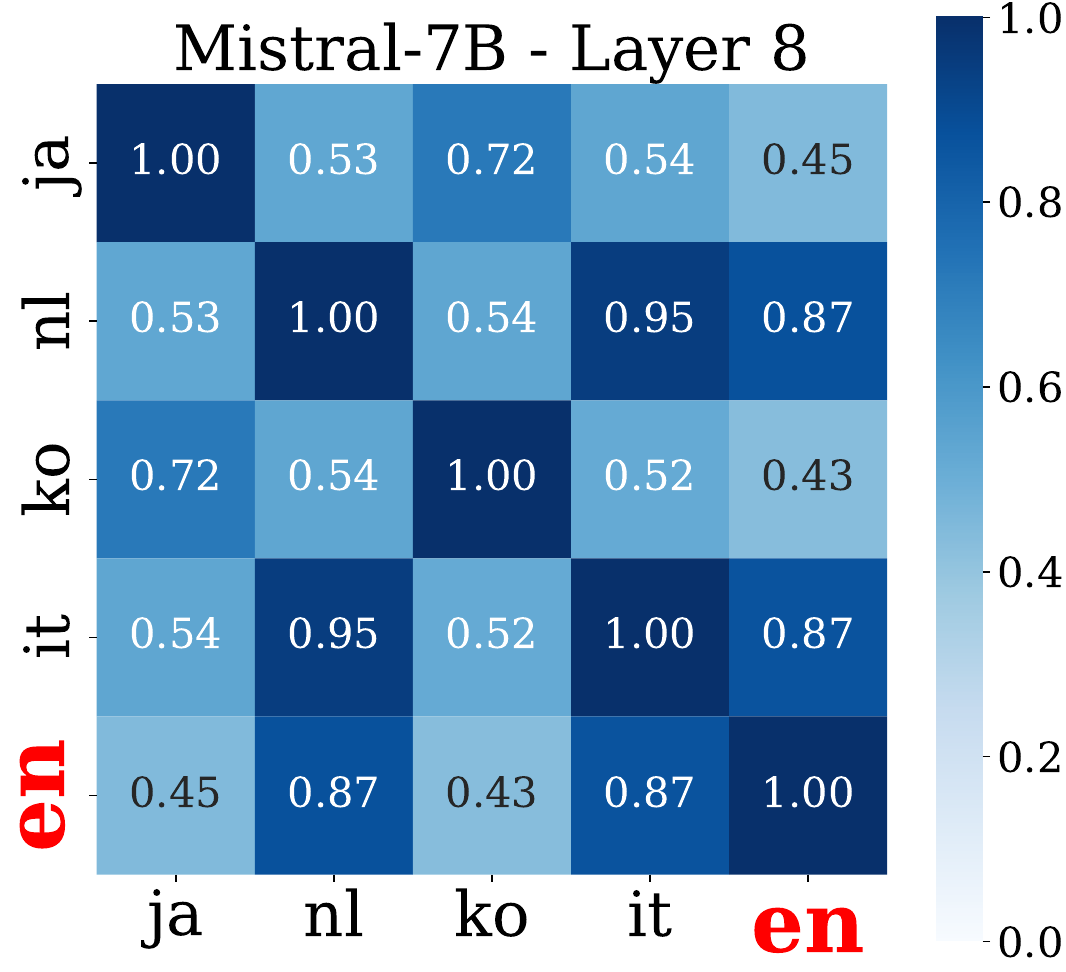}
  \includegraphics[width=0.15\linewidth]{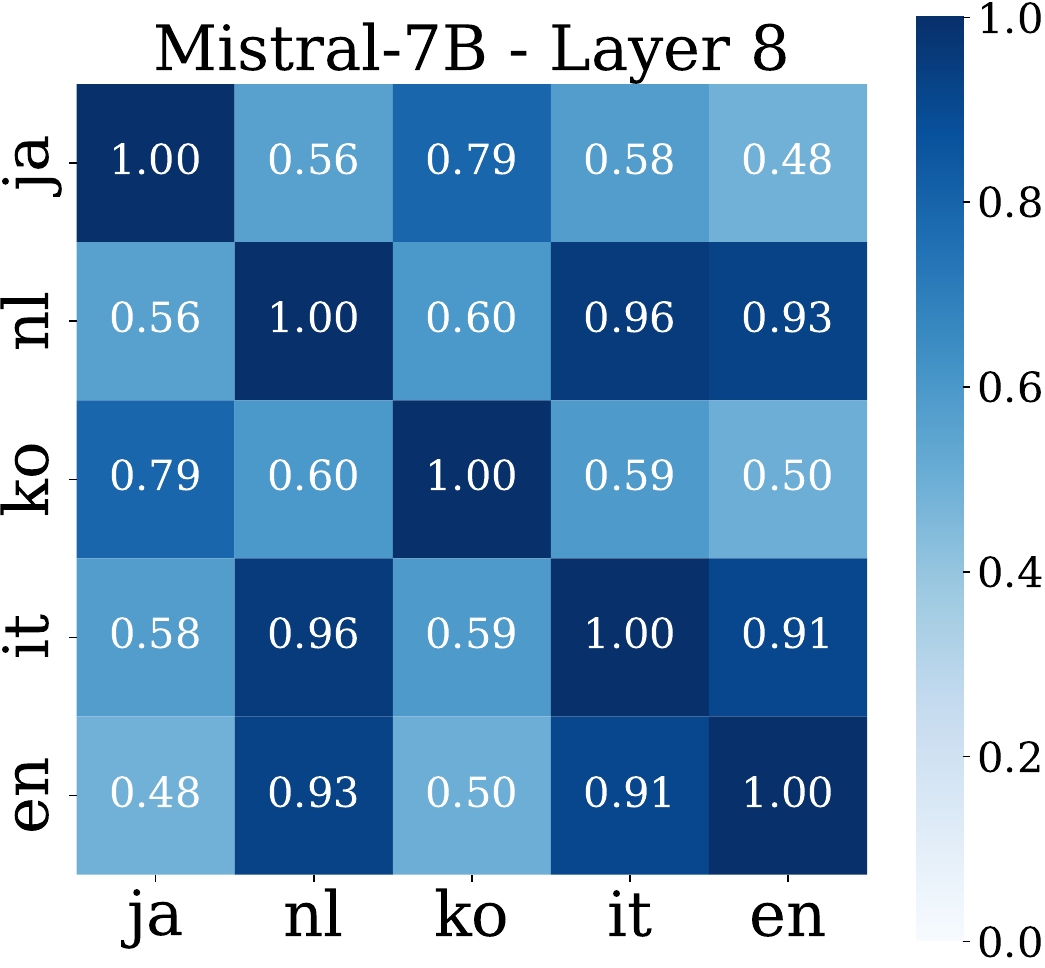}
  \includegraphics[width=0.15\linewidth]{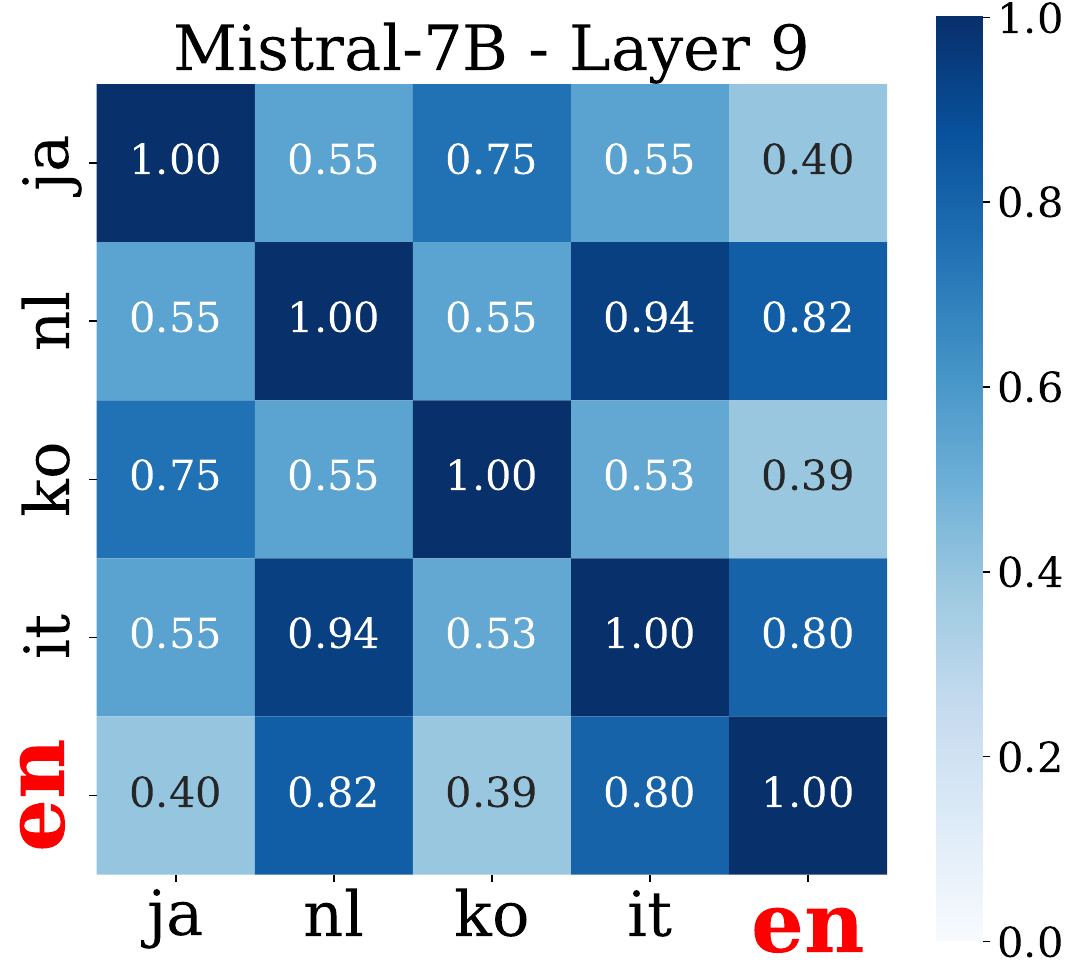}
  \includegraphics[width=0.15\linewidth]{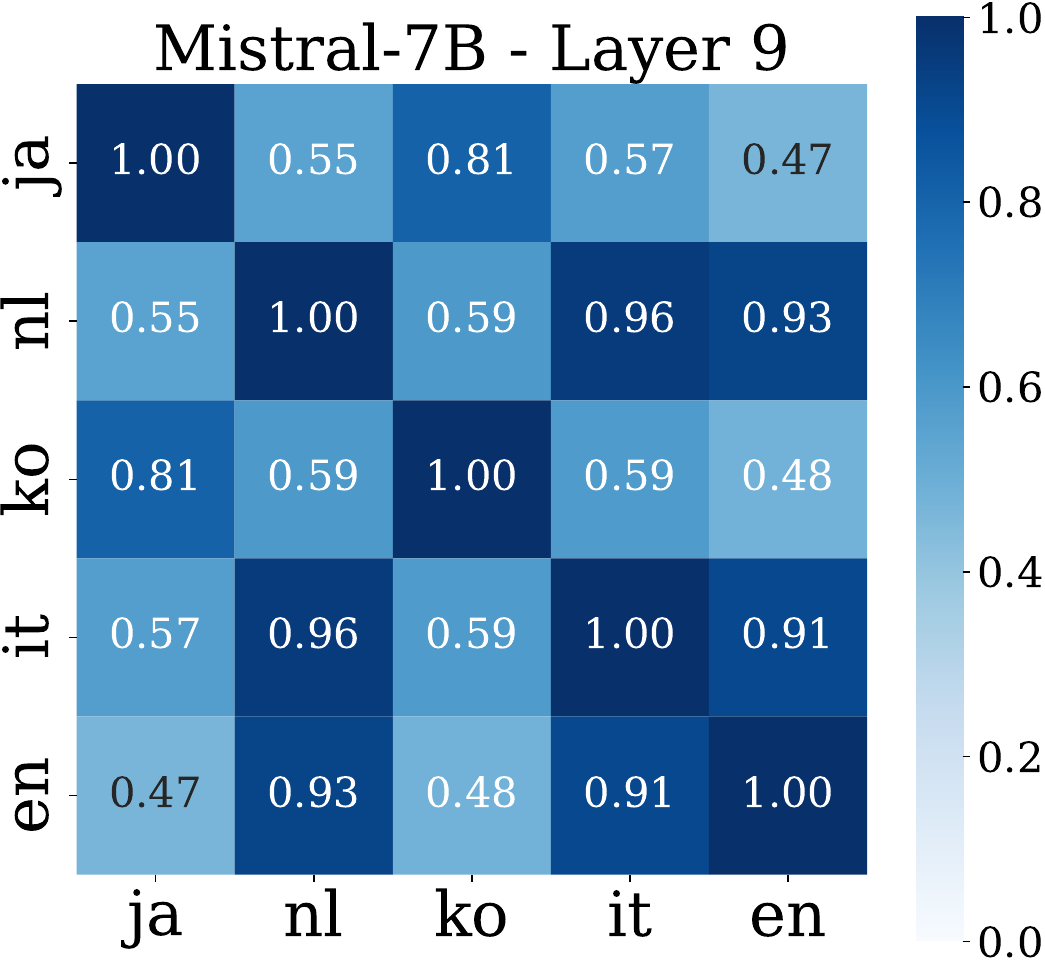}

  \begin{minipage}{0.15\linewidth}\centering \textbf{\textcolor{red}{layer 7 (Type-1)}}\end{minipage}
  \begin{minipage}{0.15\linewidth}\centering layer 7 (baseline)\end{minipage}
  \begin{minipage}{0.15\linewidth}\centering \textbf{\textcolor{red}{layer 8 (Type-1)}}\end{minipage}
  \begin{minipage}{0.15\linewidth}\centering layer 8 (baseline)\end{minipage}
  \begin{minipage}{0.15\linewidth}\centering \textbf{\textcolor{red}{layer 9 (Type-1)}}\end{minipage}
  \begin{minipage}{0.15\linewidth}\centering layer 9 (baseline)\end{minipage}

  \includegraphics[width=0.15\linewidth]{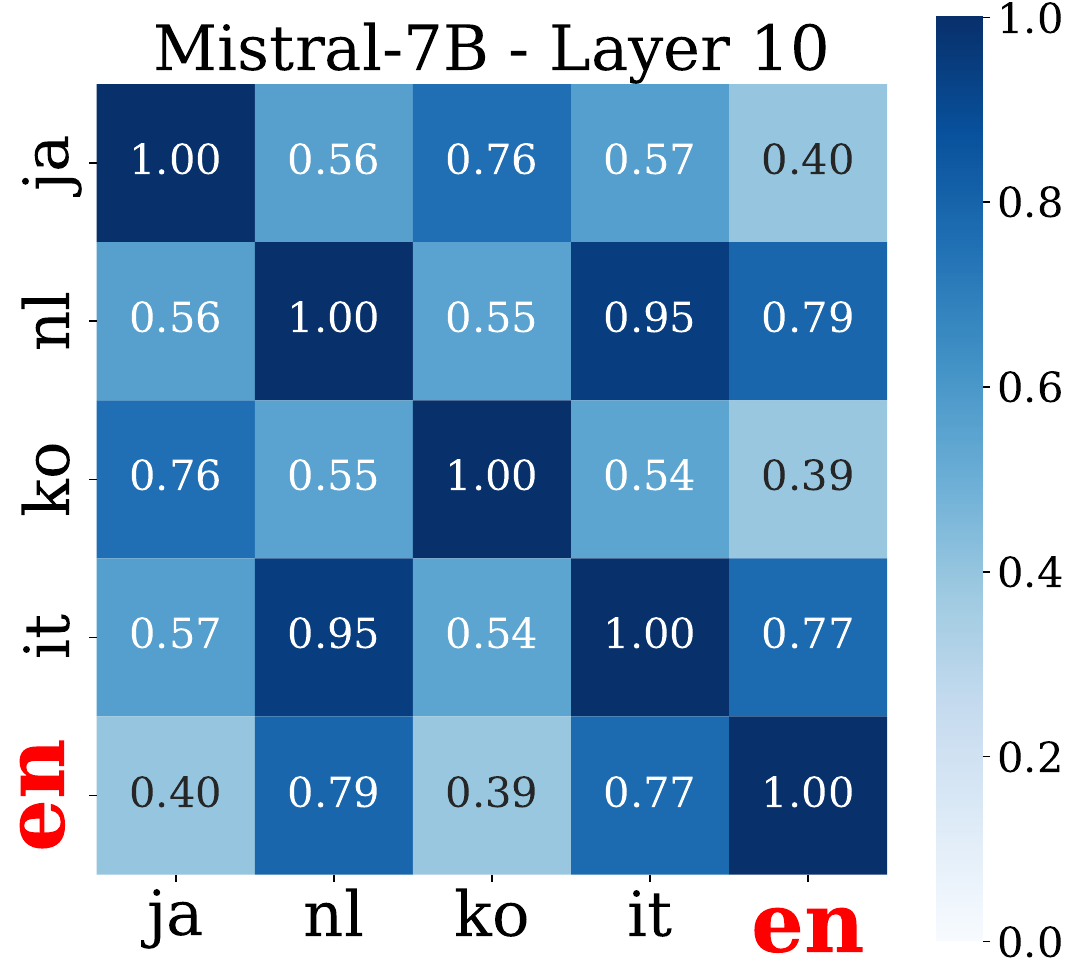}
  \includegraphics[width=0.15\linewidth]{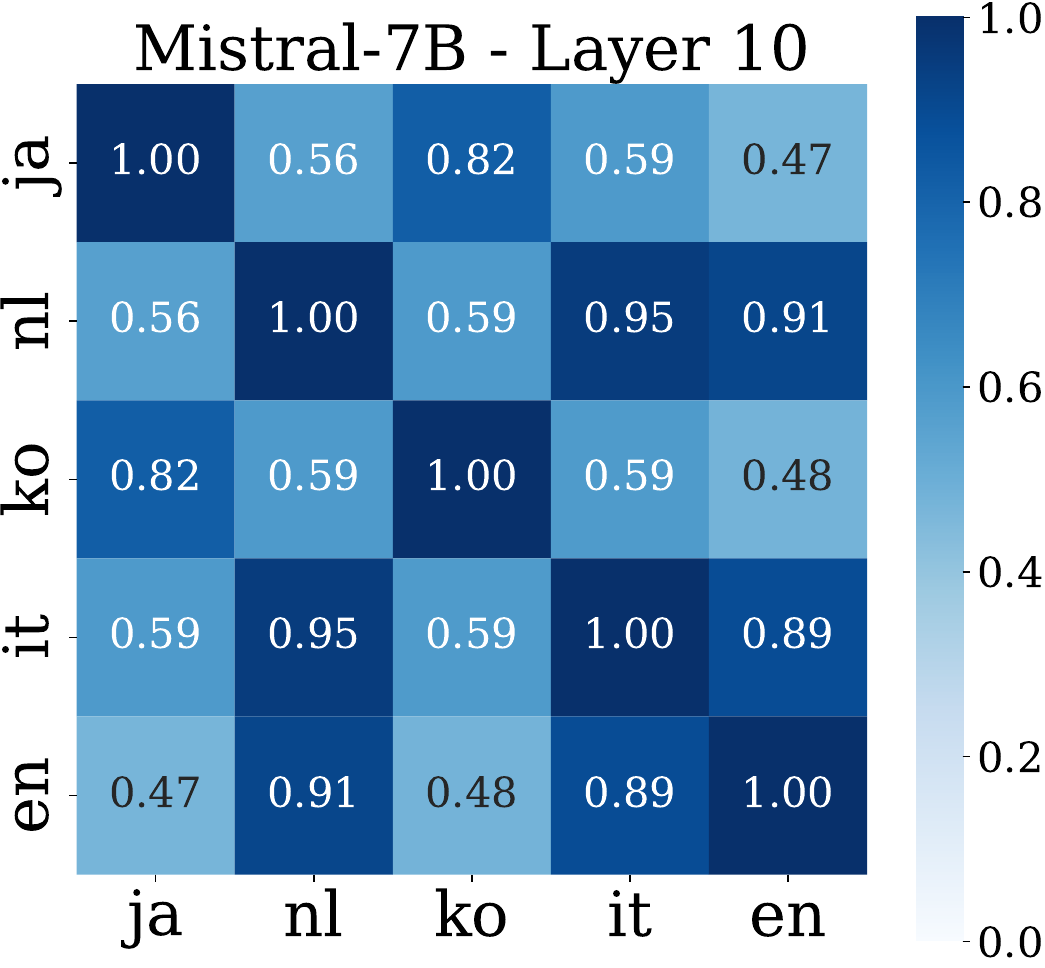}
  \includegraphics[width=0.15\linewidth]{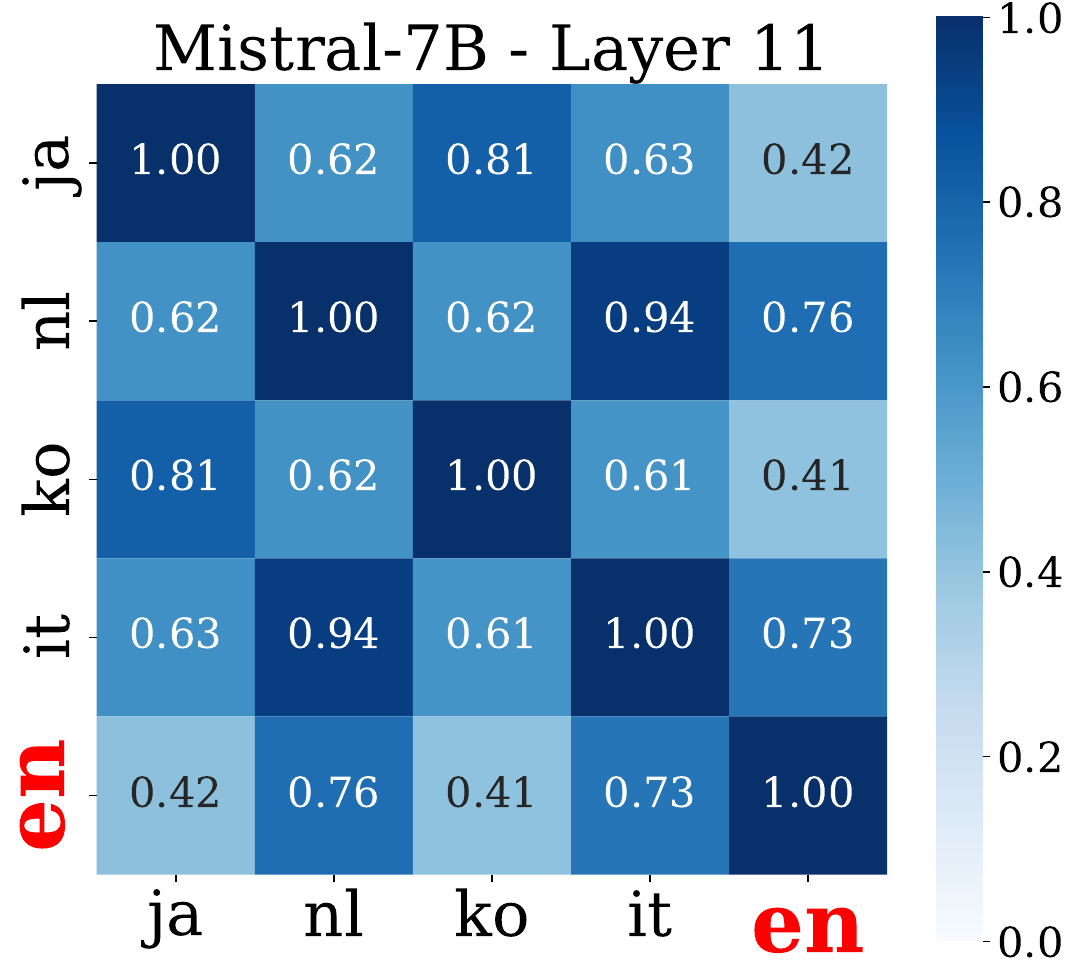}
  \includegraphics[width=0.15\linewidth]{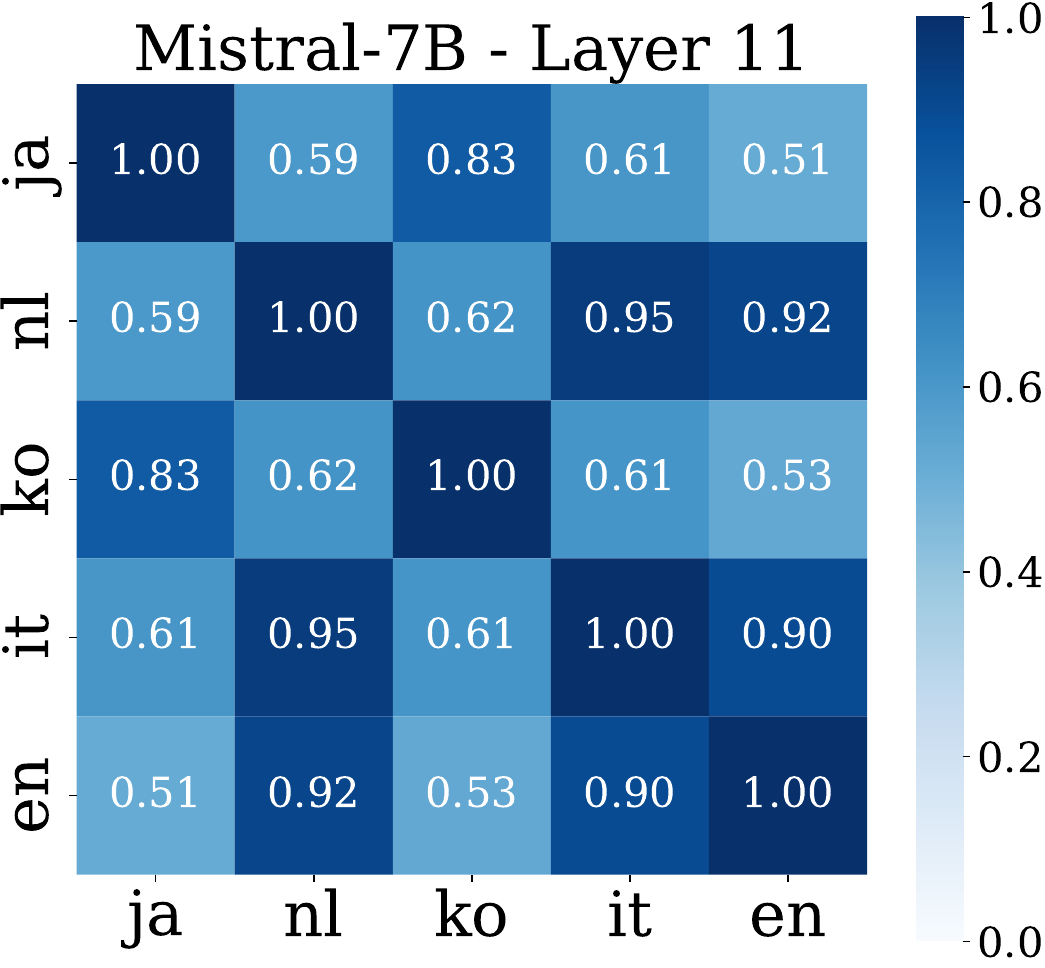}
  \includegraphics[width=0.15\linewidth]{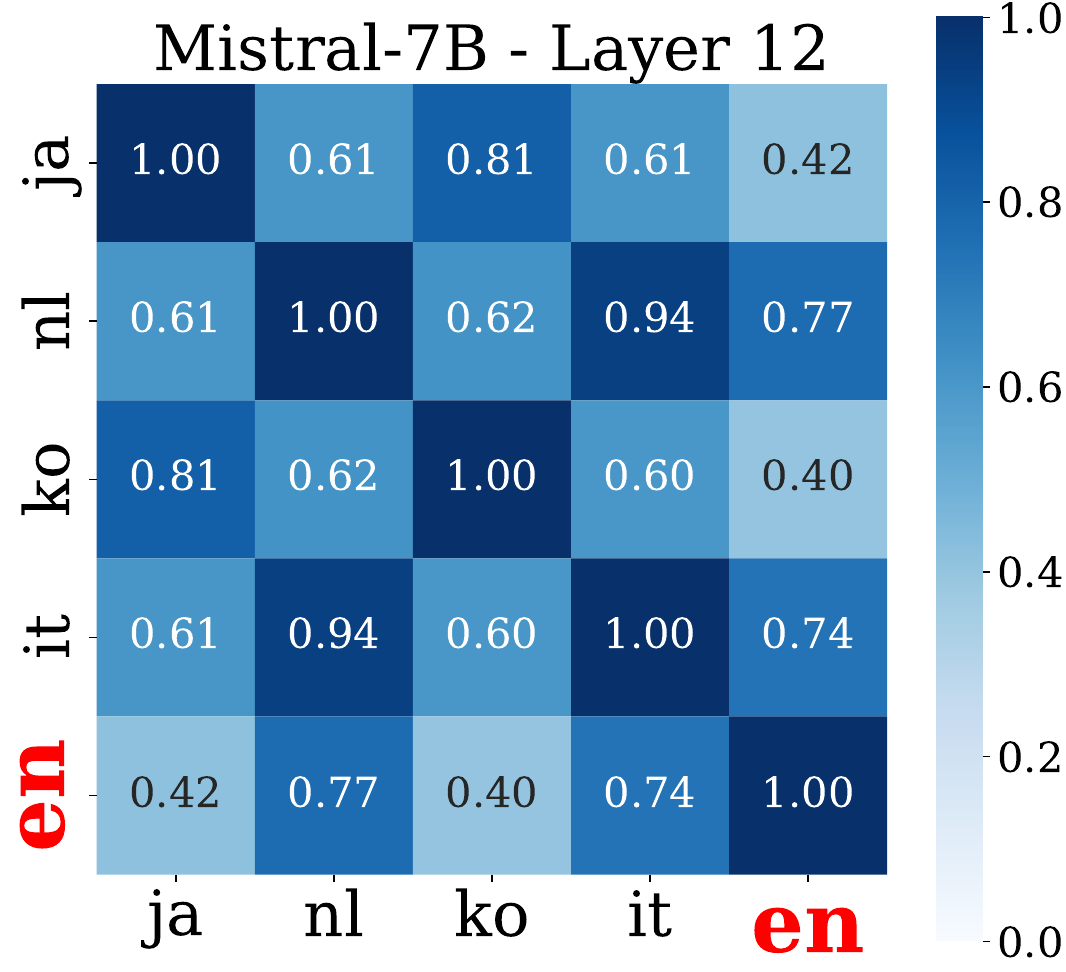}
  \includegraphics[width=0.15\linewidth]{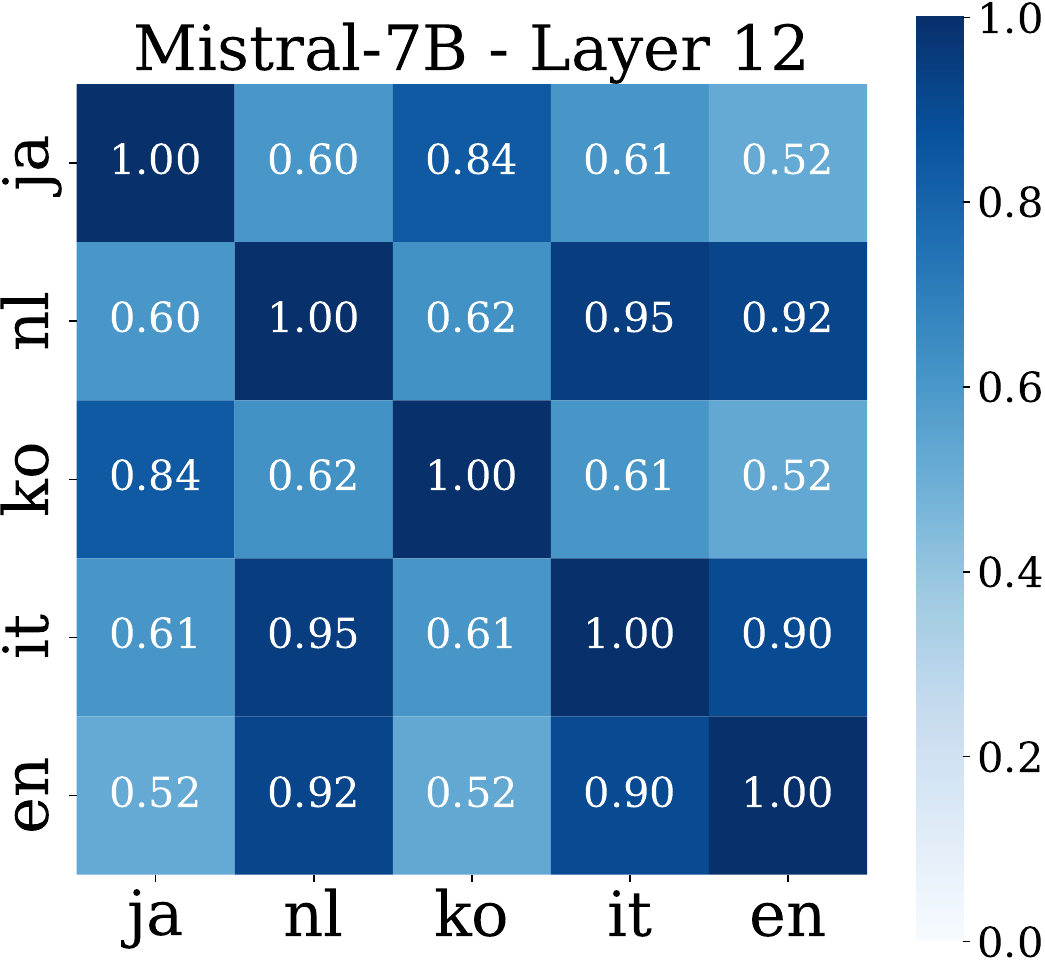}

  \begin{minipage}{0.15\linewidth}\centering \textbf{\textcolor{red}{layer 10 (Type-1)}}\end{minipage}
  \begin{minipage}{0.15\linewidth}\centering layer 10 (baseline)\end{minipage}
  \begin{minipage}{0.15\linewidth}\centering \textbf{\textcolor{red}{layer 11 (Type-1)}}\end{minipage}
  \begin{minipage}{0.15\linewidth}\centering layer 11 (baseline)\end{minipage}
  \begin{minipage}{0.15\linewidth}\centering \textbf{\textcolor{red}{layer 12 (Type-1)}}\end{minipage}
  \begin{minipage}{0.15\linewidth}\centering layer 12 (baseline)\end{minipage}

  \includegraphics[width=0.15\linewidth]{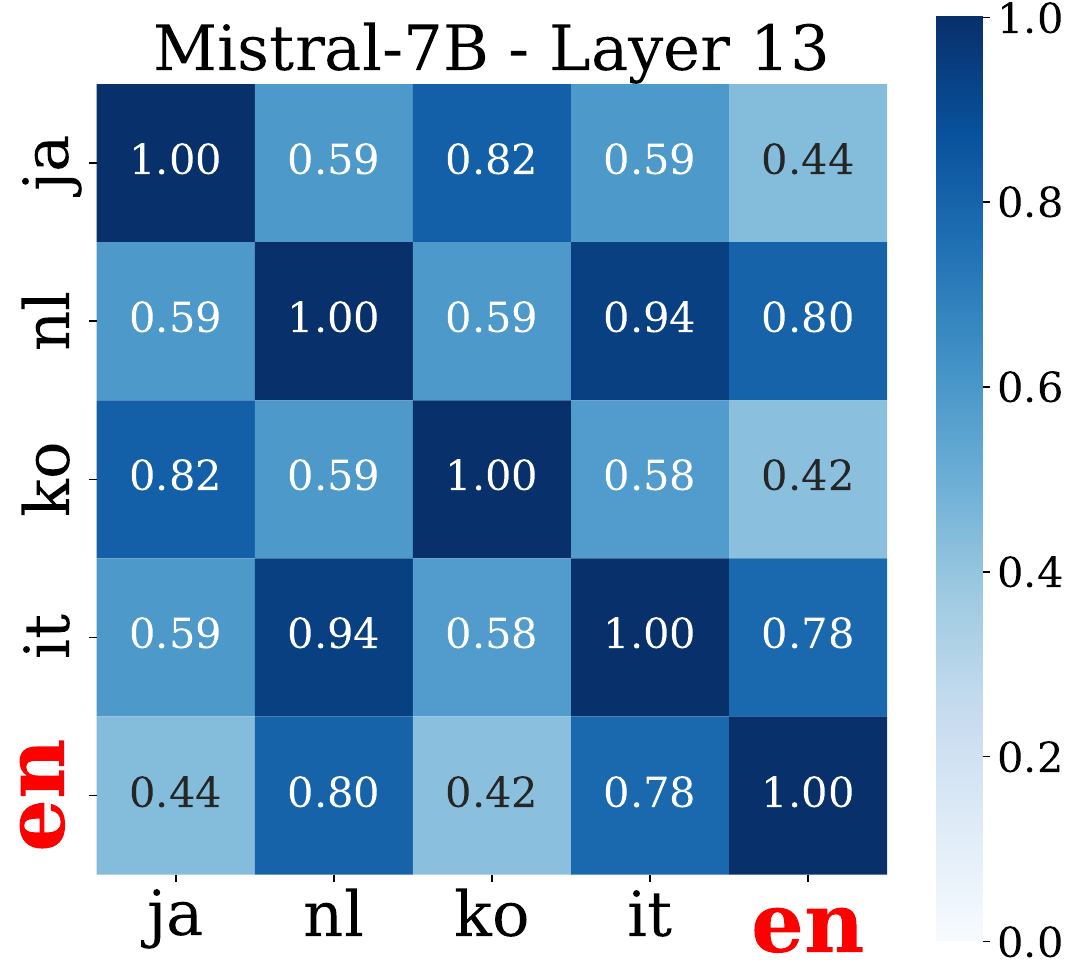}
  \includegraphics[width=0.15\linewidth]{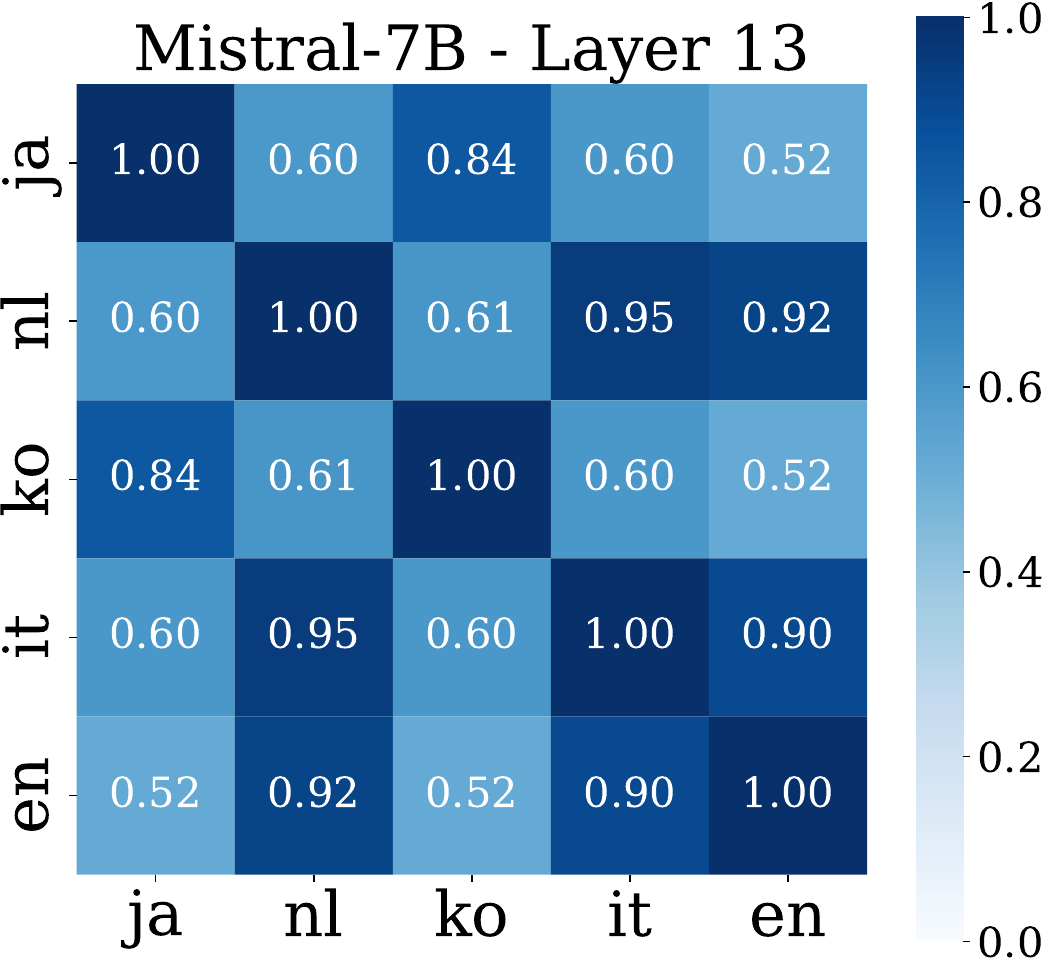}
  \includegraphics[width=0.15\linewidth]{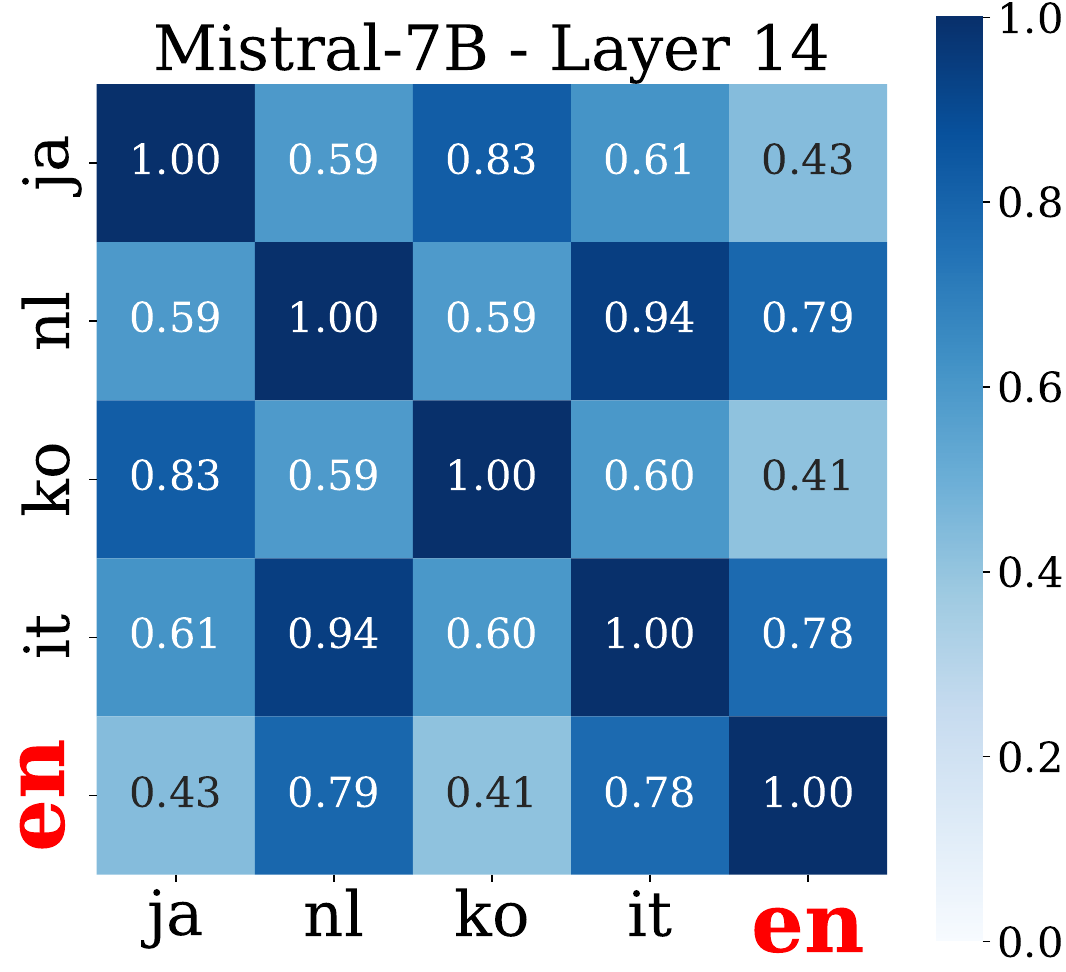}
  \includegraphics[width=0.15\linewidth]{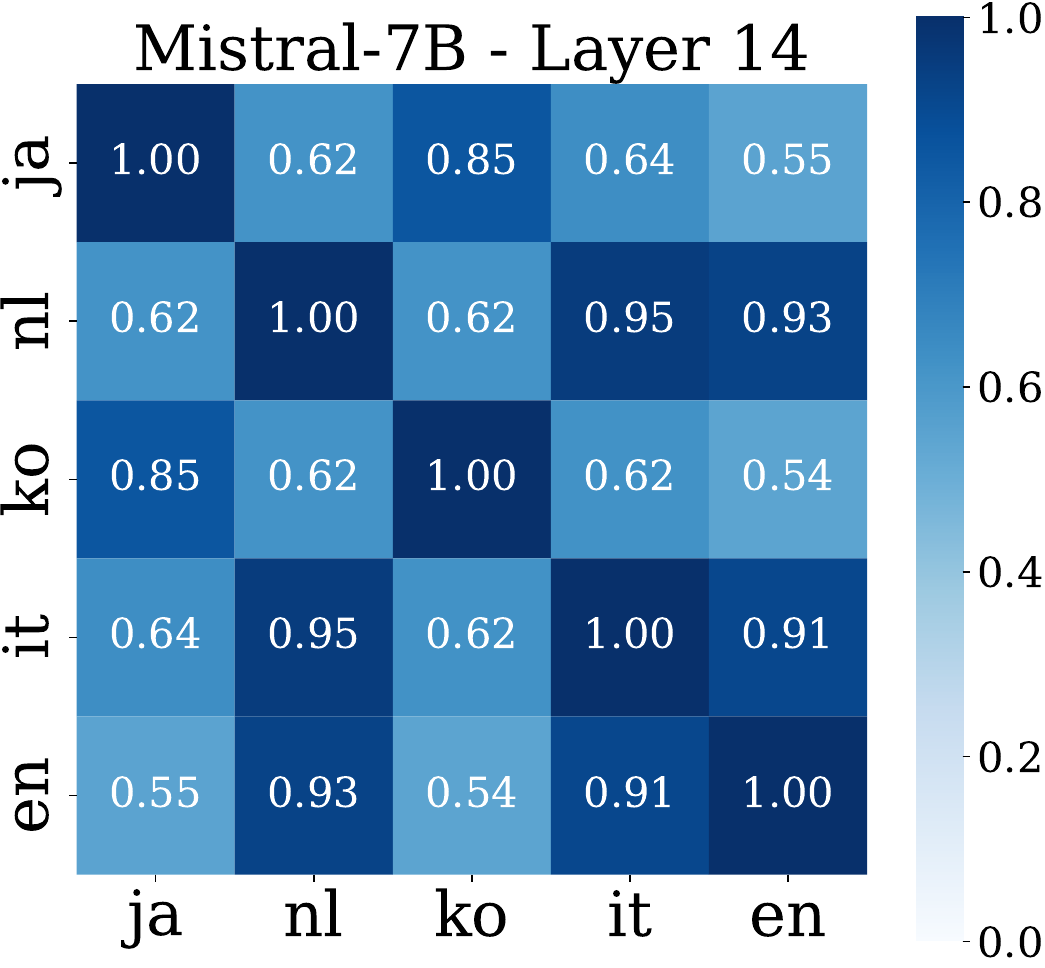}
  \includegraphics[width=0.15\linewidth]{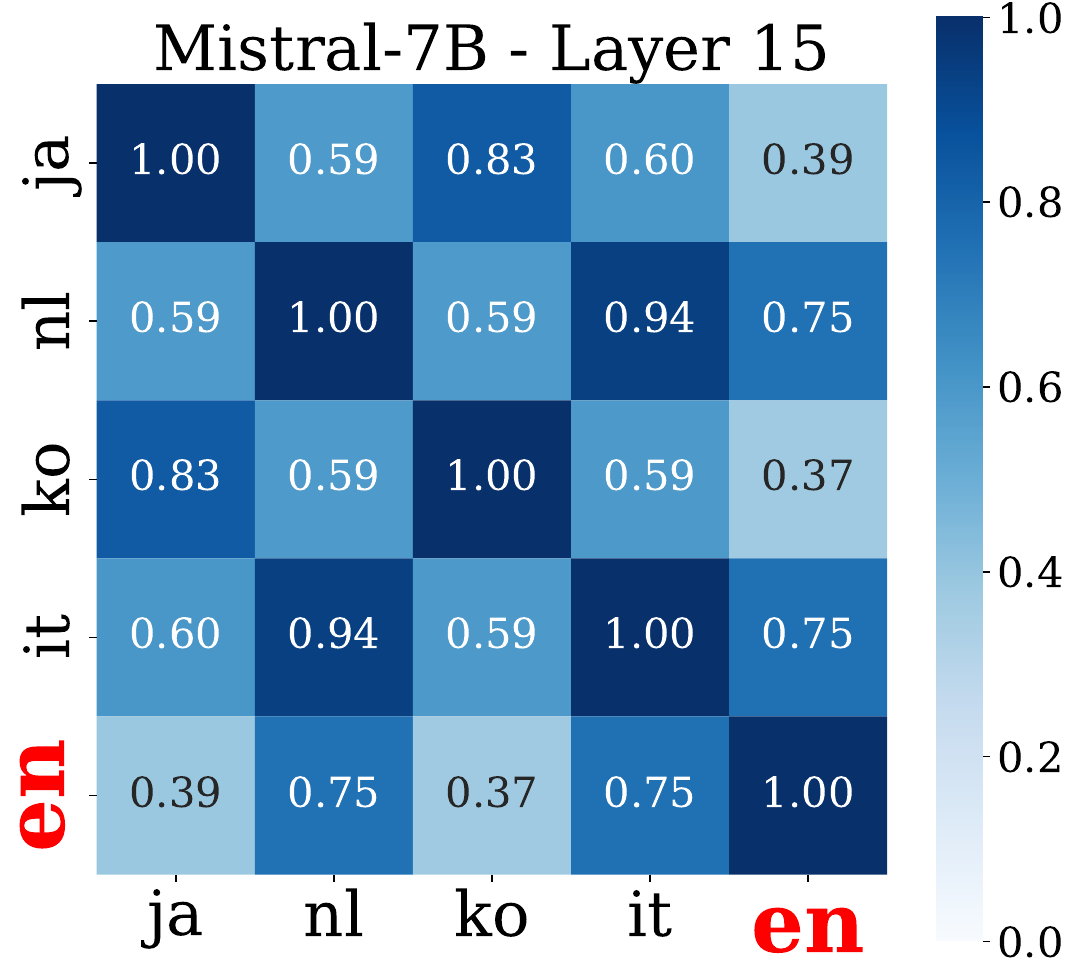}
  \includegraphics[width=0.15\linewidth]{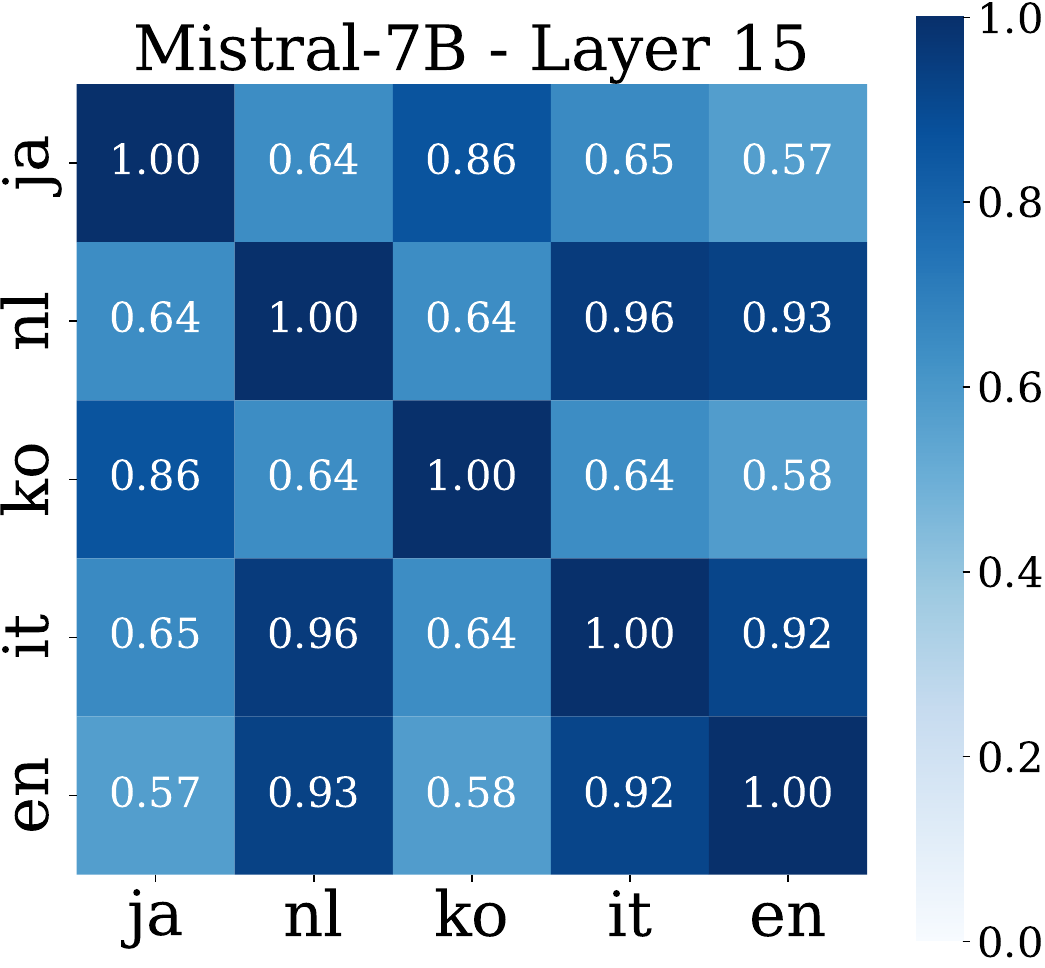}

  \begin{minipage}{0.15\linewidth}\centering \textbf{\textcolor{red}{layer 13 (Type-1)}}\end{minipage}
  \begin{minipage}{0.15\linewidth}\centering layer 13 (baseline)\end{minipage}
  \begin{minipage}{0.15\linewidth}\centering \textbf{\textcolor{red}{layer 14 (Type-1)}}\end{minipage}
  \begin{minipage}{0.15\linewidth}\centering layer 14 (baseline)\end{minipage}
  \begin{minipage}{0.15\linewidth}\centering \textbf{\textcolor{red}{layer 15 (Type-1)}}\end{minipage}
  \begin{minipage}{0.15\linewidth}\centering layer 15 (baseline)\end{minipage}

  \includegraphics[width=0.15\linewidth]{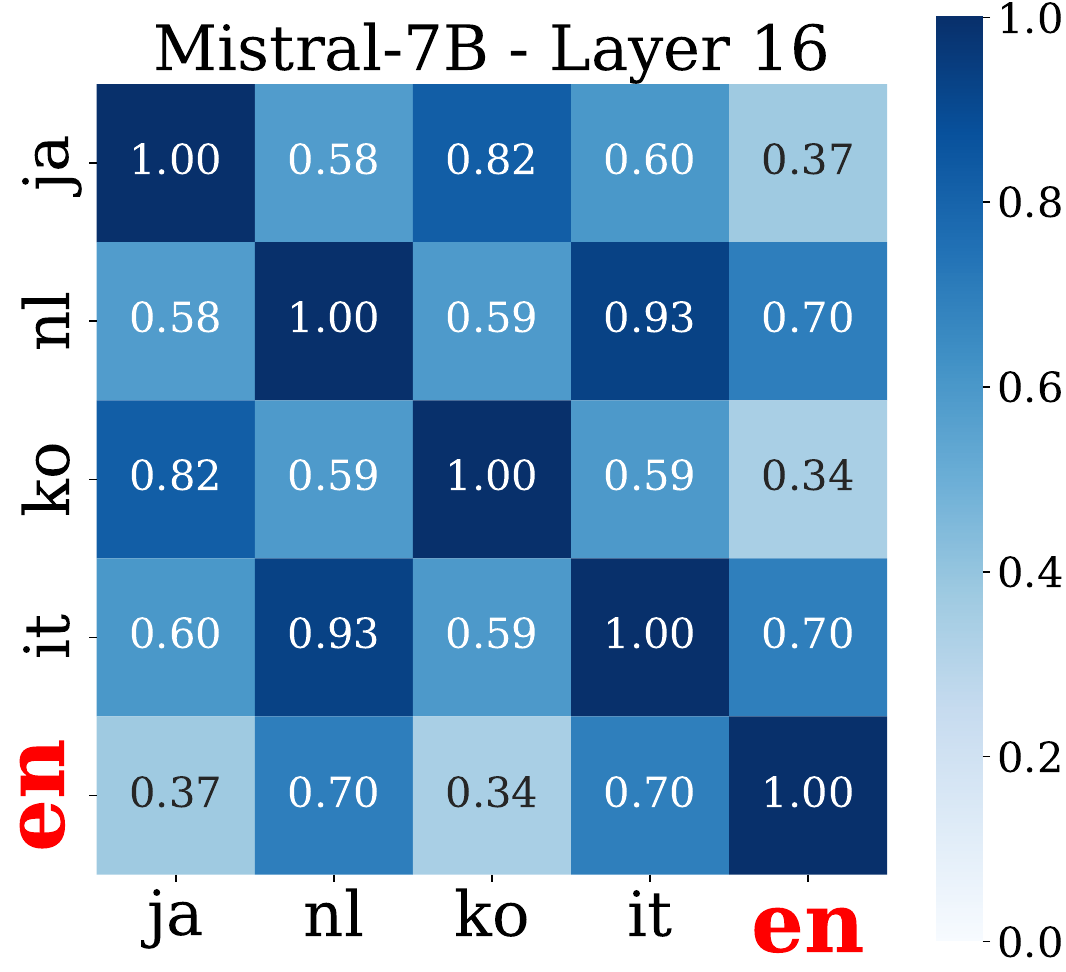}
  \includegraphics[width=0.15\linewidth]{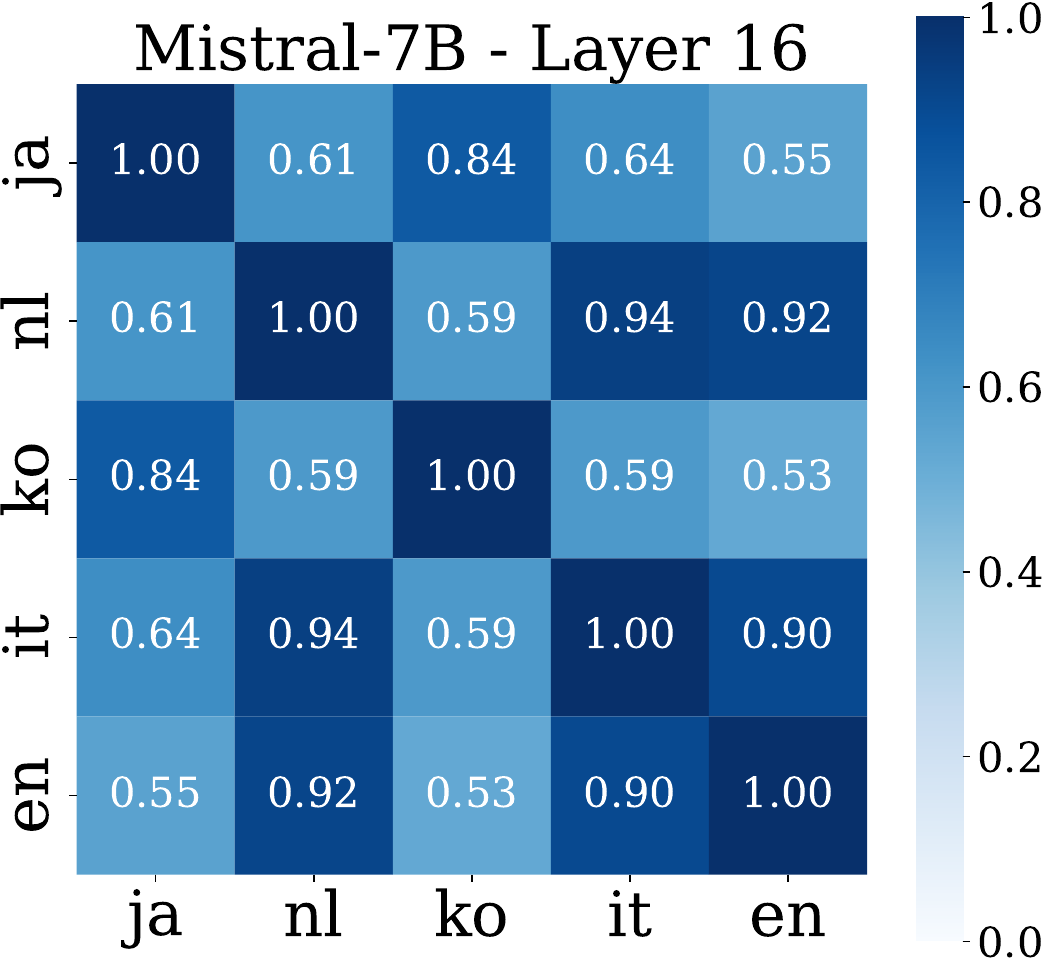}
  \includegraphics[width=0.15\linewidth]{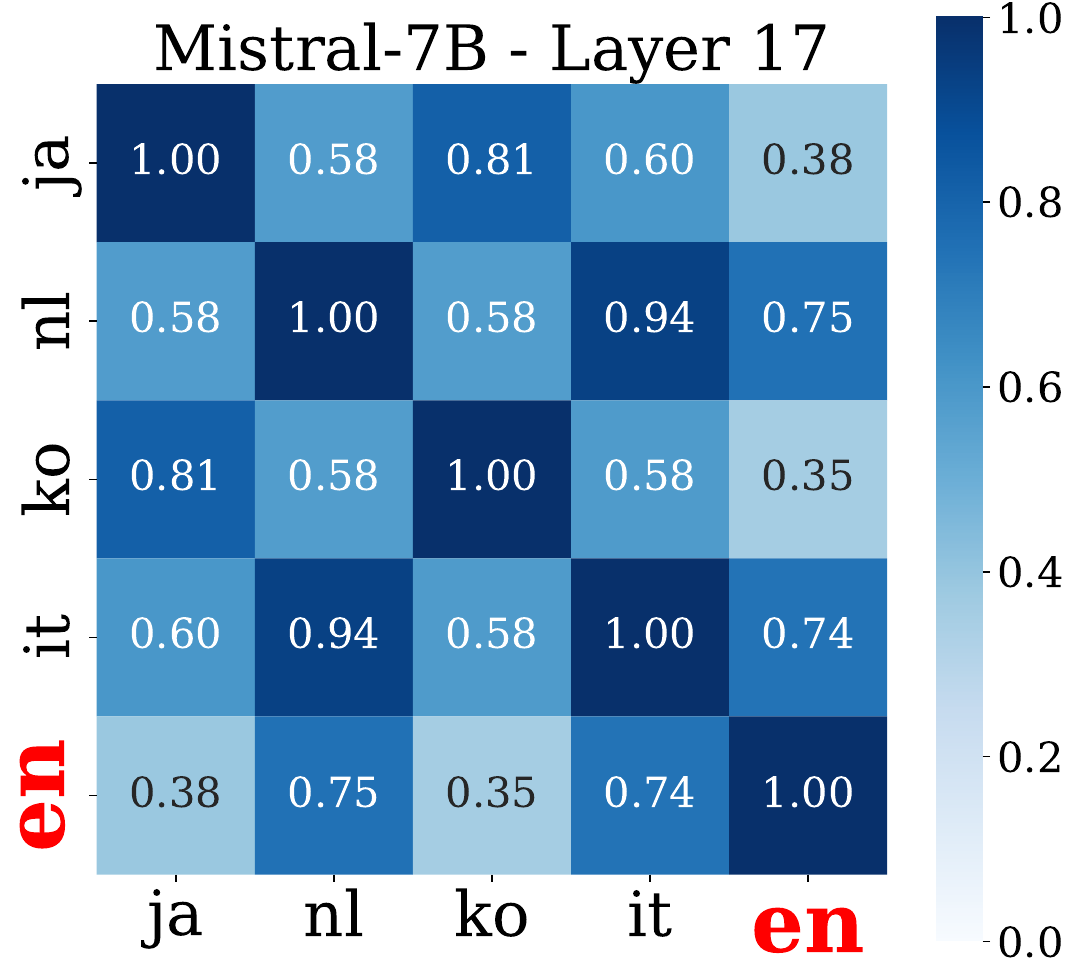}
  \includegraphics[width=0.15\linewidth]{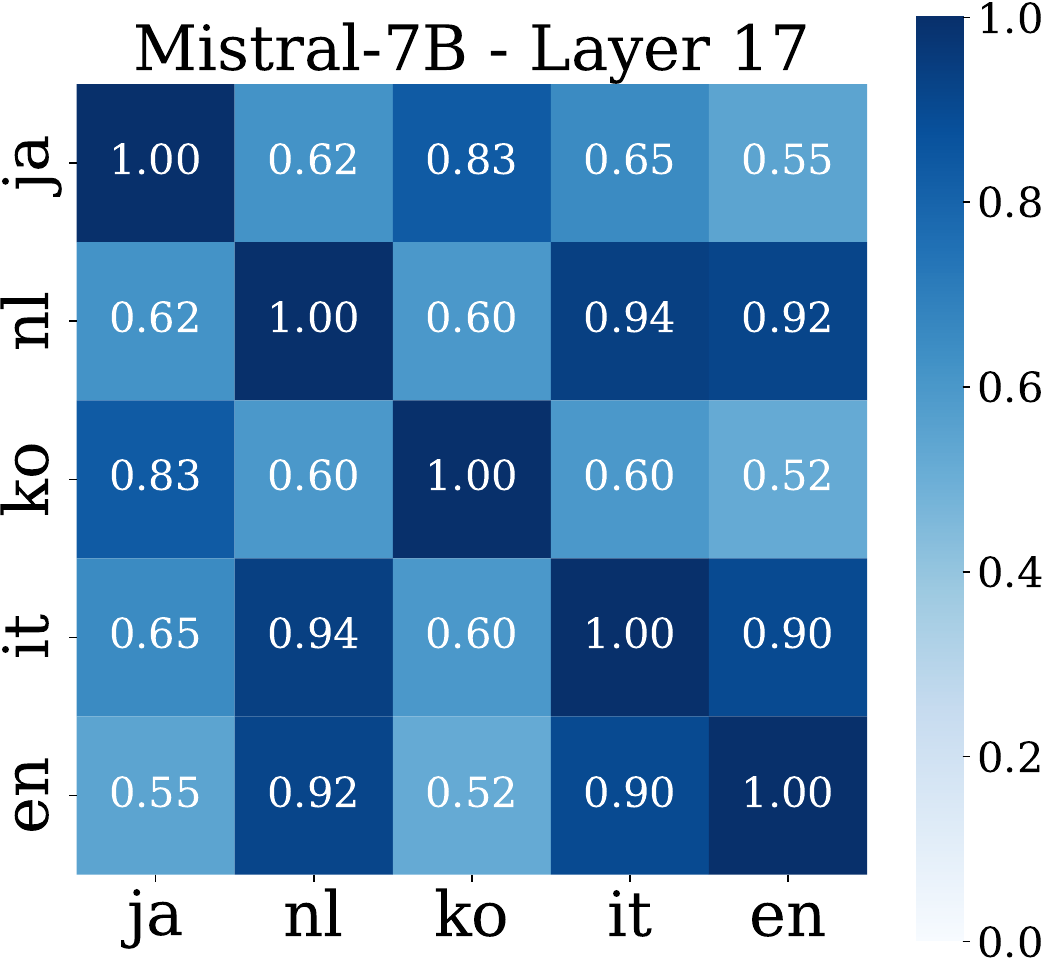}
  \includegraphics[width=0.15\linewidth]{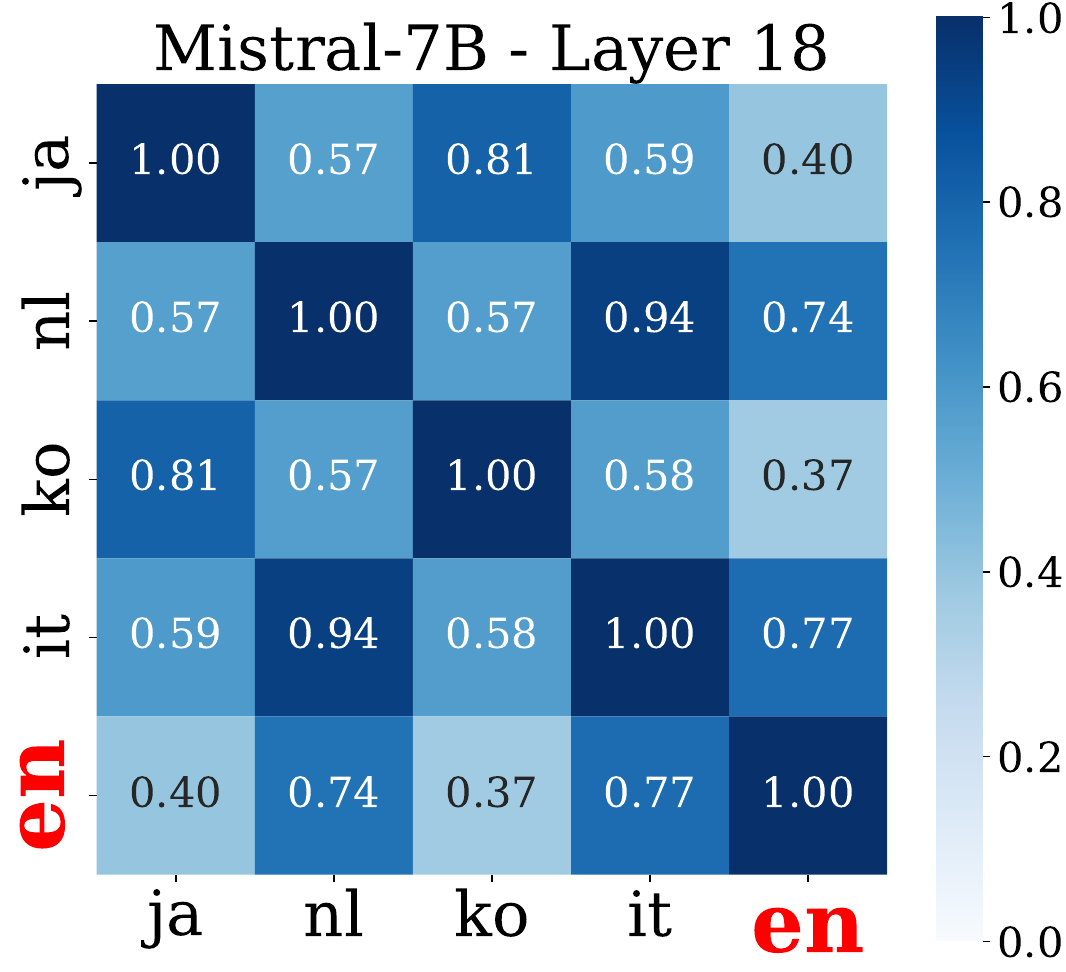}
  \includegraphics[width=0.15\linewidth]{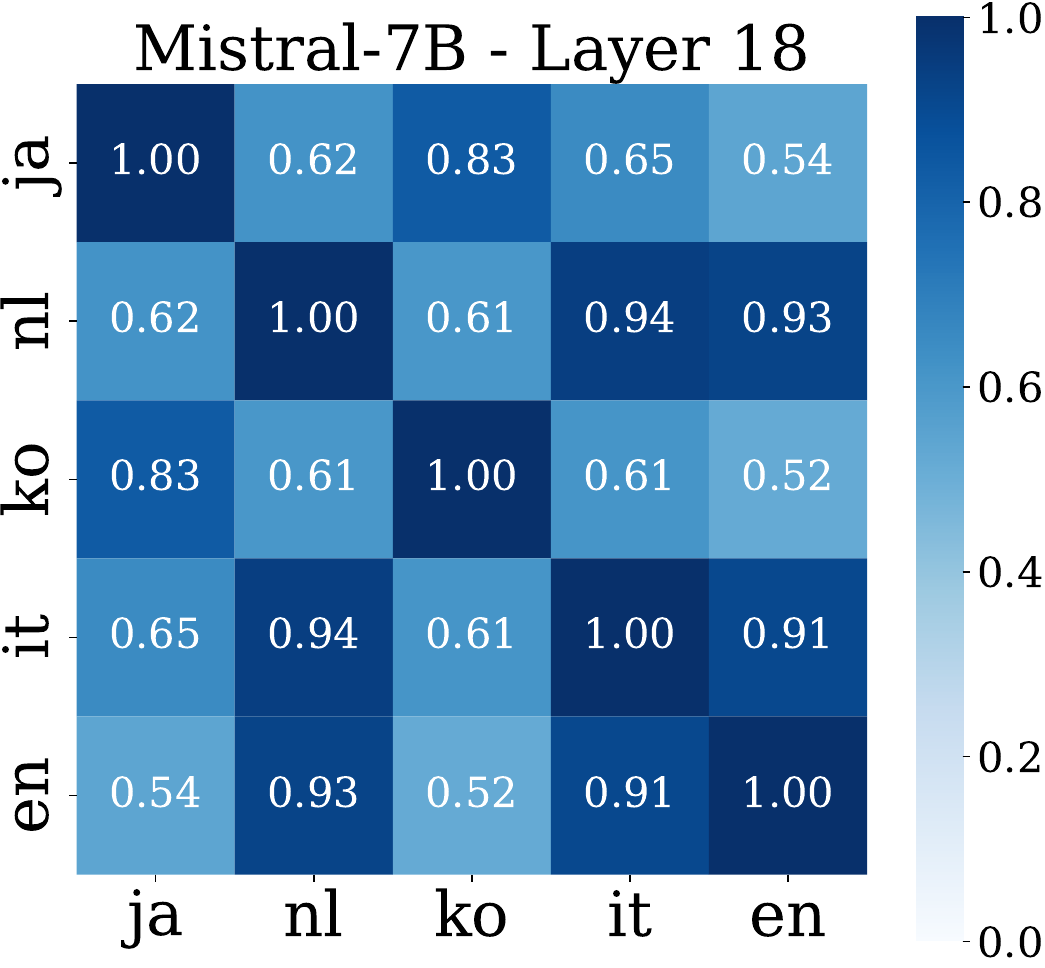}

  \begin{minipage}{0.15\linewidth}\centering \textbf{\textcolor{red}{layer 16 (Type-1)}}\end{minipage}
  \begin{minipage}{0.15\linewidth}\centering layer 16 (baseline)\end{minipage}
  \begin{minipage}{0.15\linewidth}\centering \textbf{\textcolor{red}{layer 17 (Type-1)}}\end{minipage}
  \begin{minipage}{0.15\linewidth}\centering layer 17 (baseline)\end{minipage}
  \begin{minipage}{0.15\linewidth}\centering \textbf{\textcolor{red}{layer 18 (Type-1)}}\end{minipage}
  \begin{minipage}{0.15\linewidth}\centering layer 18 (baseline)\end{minipage}

  \includegraphics[width=0.15\linewidth]{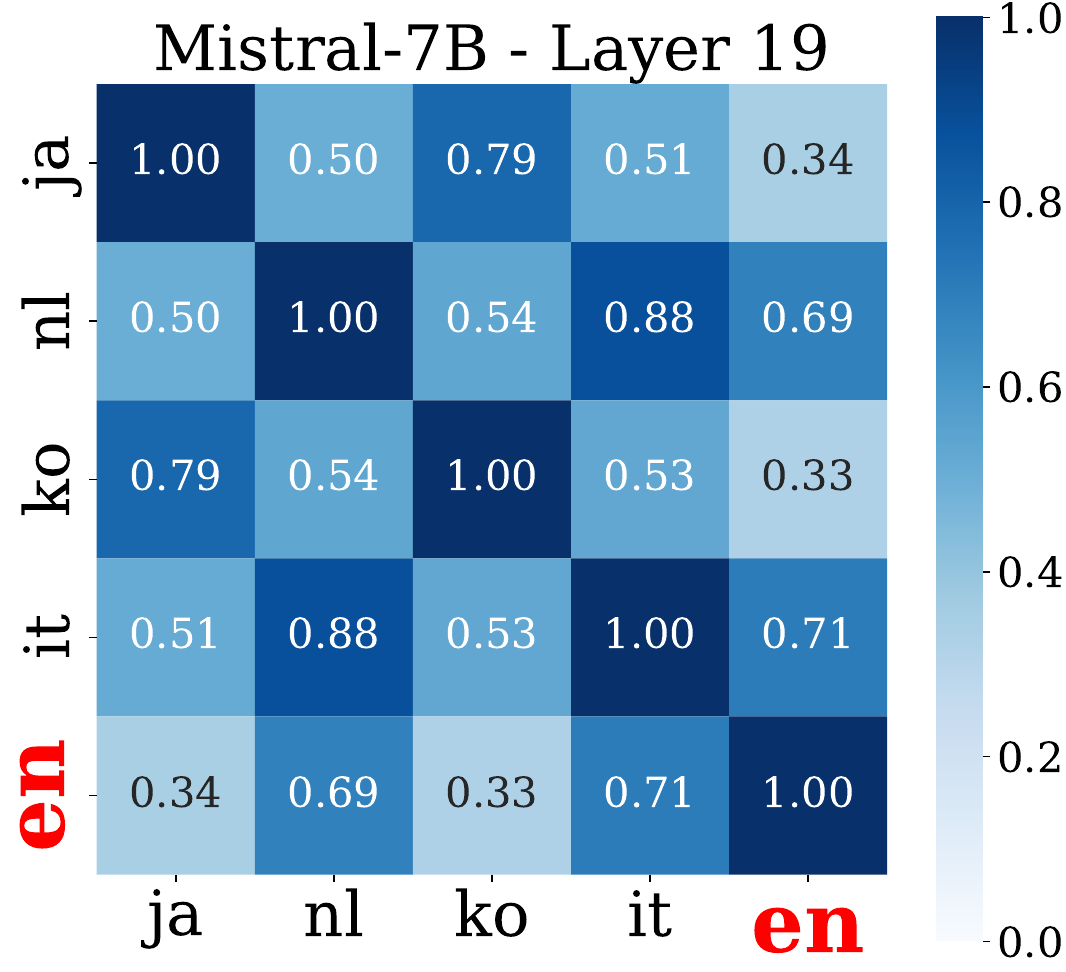}
  \includegraphics[width=0.15\linewidth]{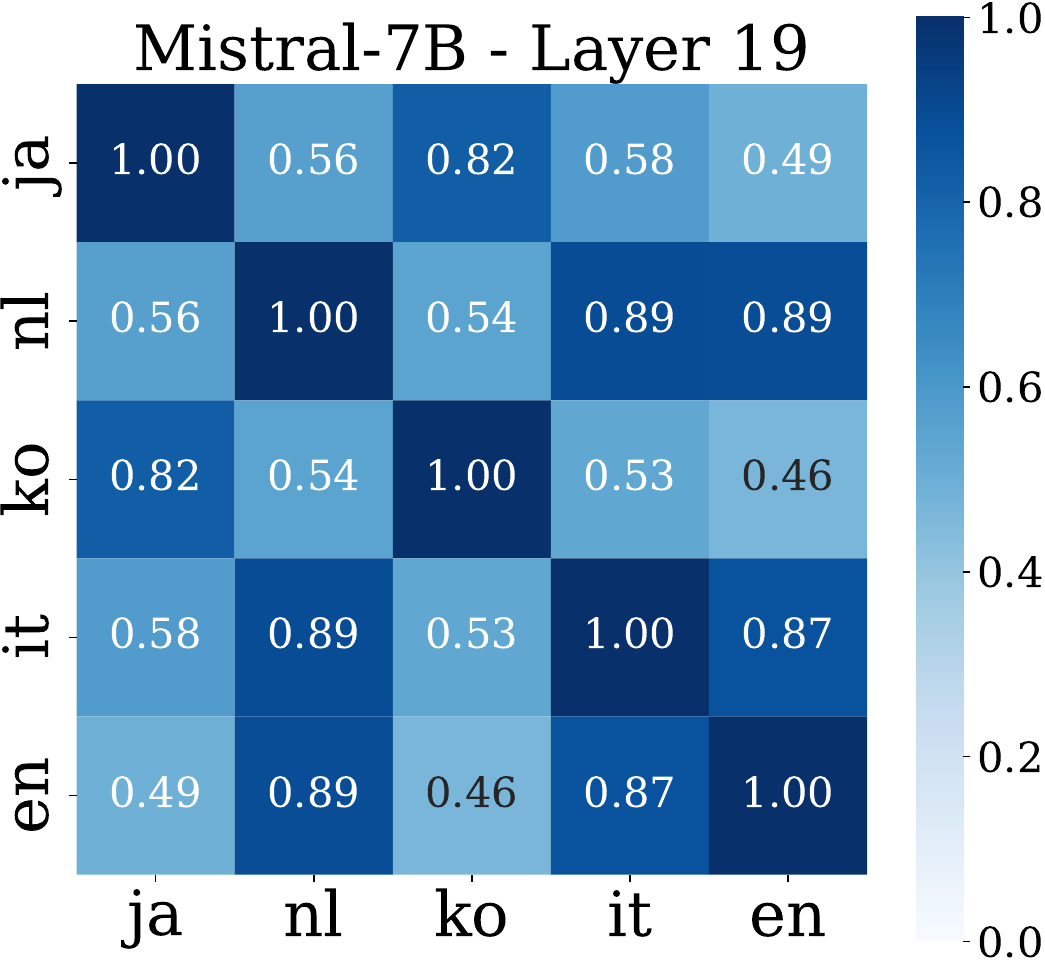}
  \includegraphics[width=0.15\linewidth]{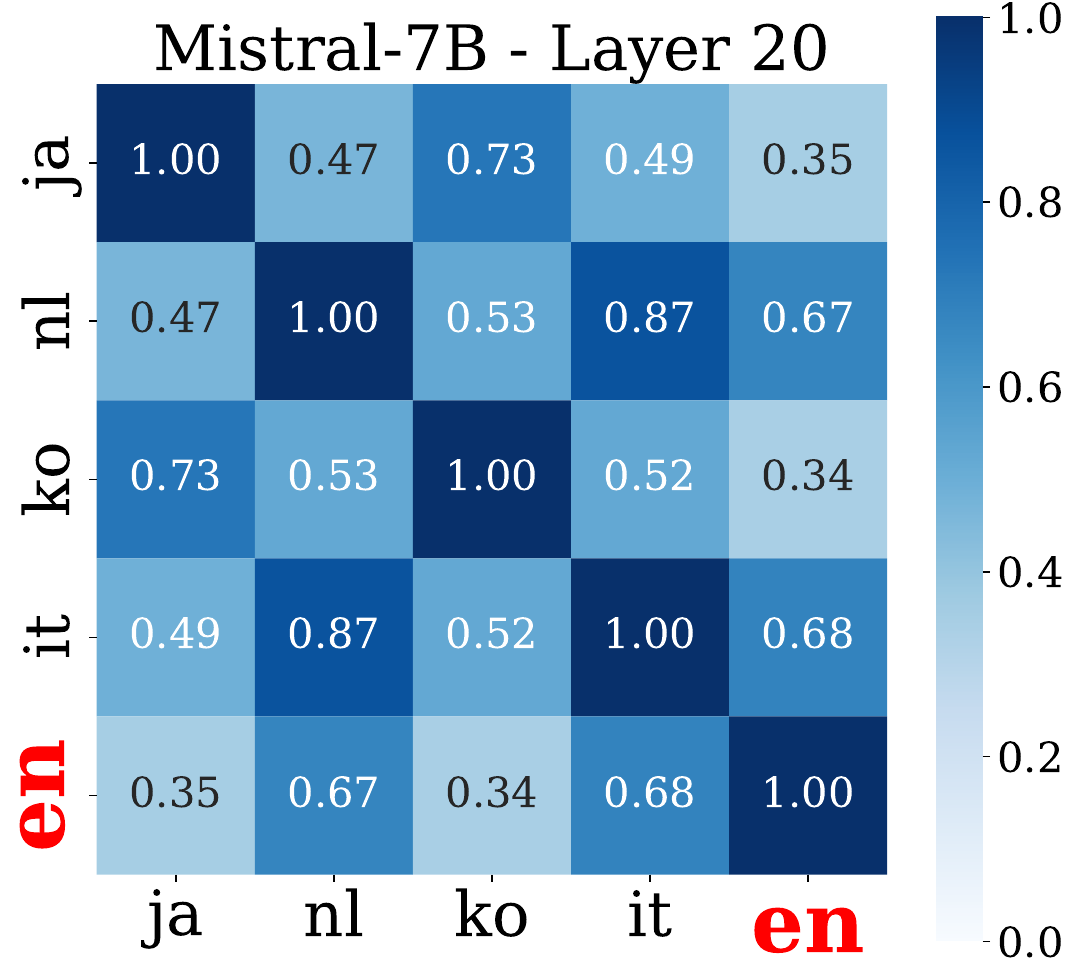}
  \includegraphics[width=0.15\linewidth]{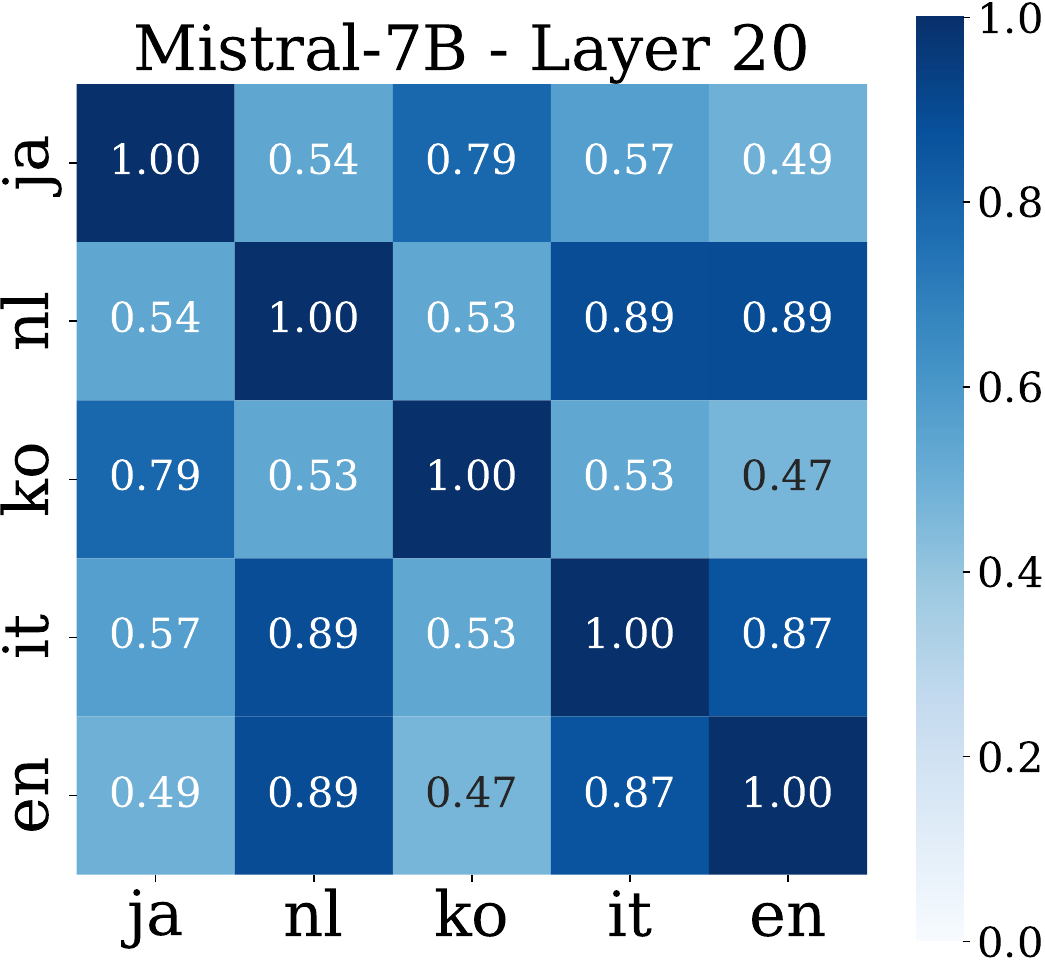}

  \begin{minipage}{0.15\linewidth}\centering \textbf{\textcolor{red}{layer 19 (Type-1)}}\end{minipage}
  \begin{minipage}{0.15\linewidth}\centering layer 19 (baseline)\end{minipage}
  \begin{minipage}{0.15\linewidth}\centering \textbf{\textcolor{red}{layer 20 (Type-1)}}\end{minipage}
  \begin{minipage}{0.15\linewidth}\centering layer 20 (baseline)\end{minipage}

  \caption{\textbf{Distance among language latent spaces while deactivating Top-1k Type-1 Transfer Neurons (Mistral-7B)}.}
  \label{fig:appendix:distance centroids among langage subspaces deactivating type1 mistral}
\end{figure*}

\begin{figure*}[t]
  \centering

  \includegraphics[width=0.15\linewidth]{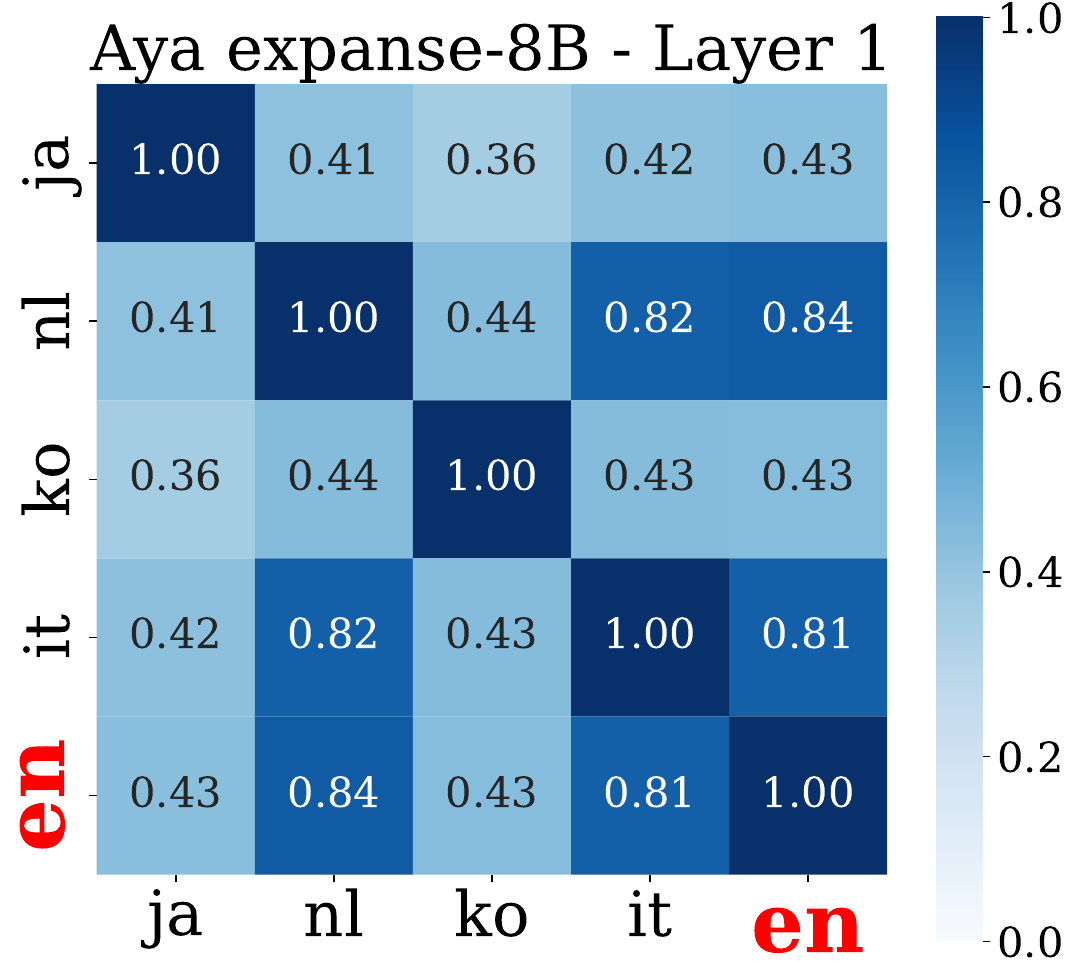}
  \includegraphics[width=0.15\linewidth]{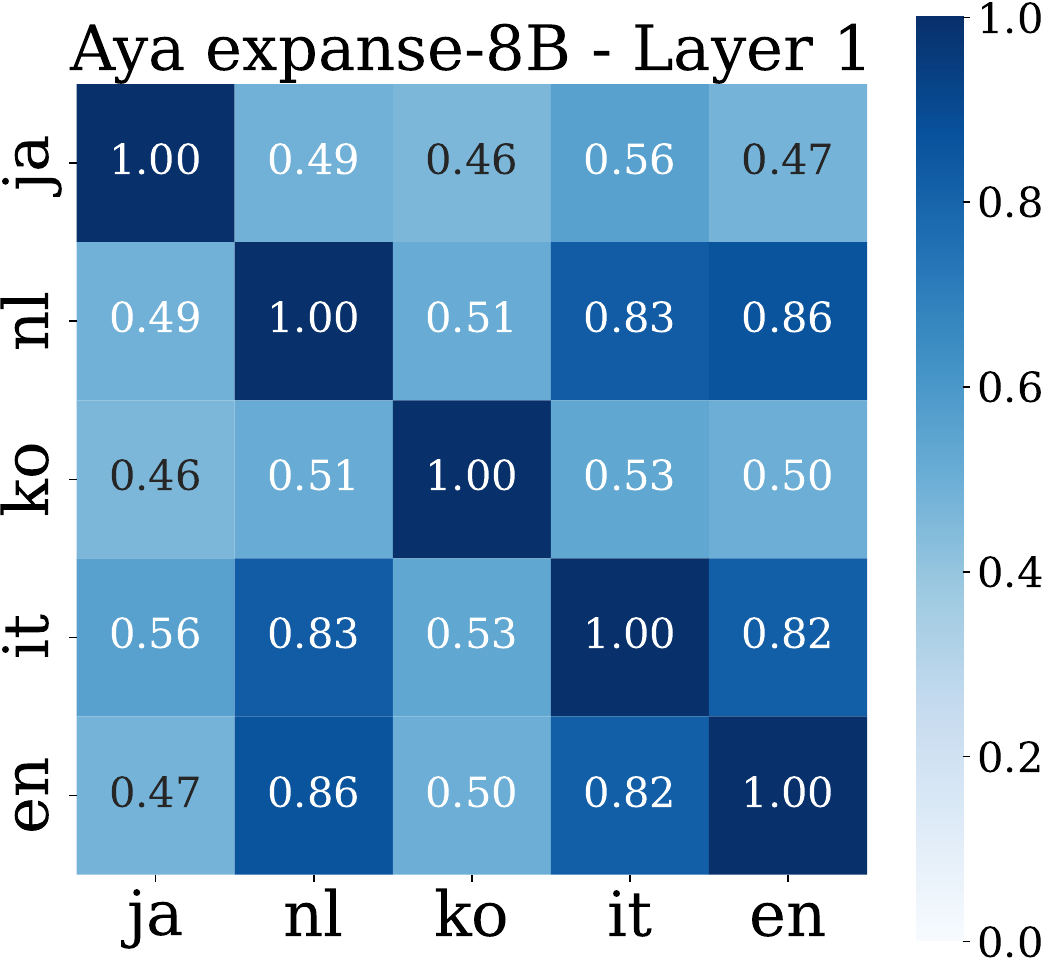}
  \includegraphics[width=0.15\linewidth]{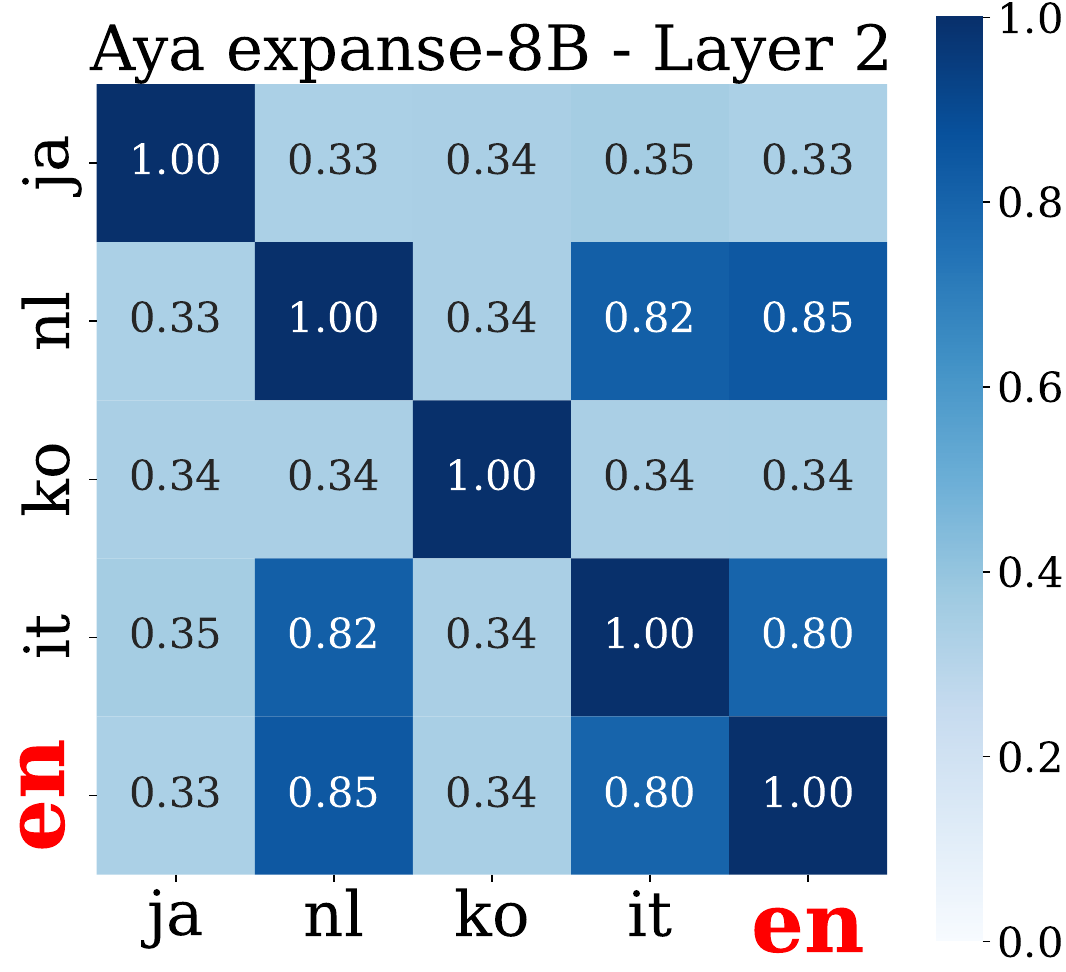}
  \includegraphics[width=0.15\linewidth]{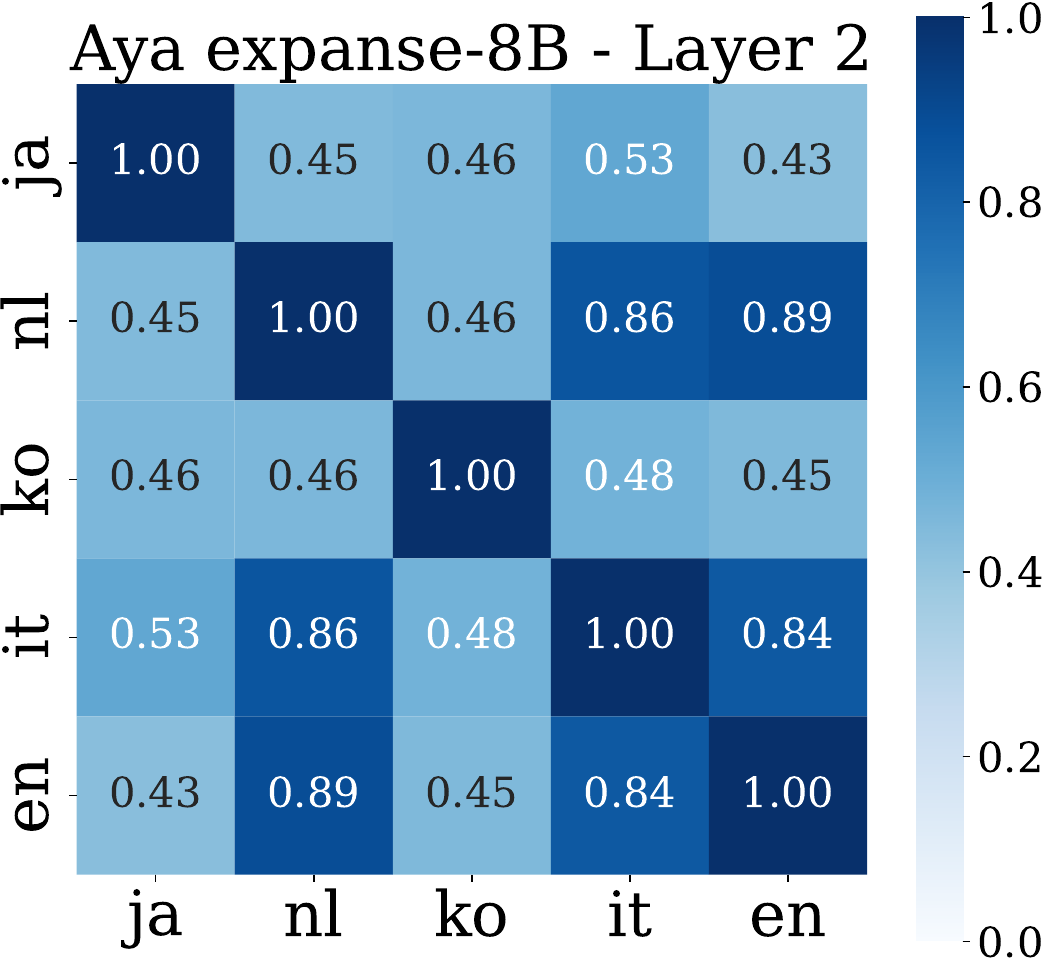}
  \includegraphics[width=0.15\linewidth]{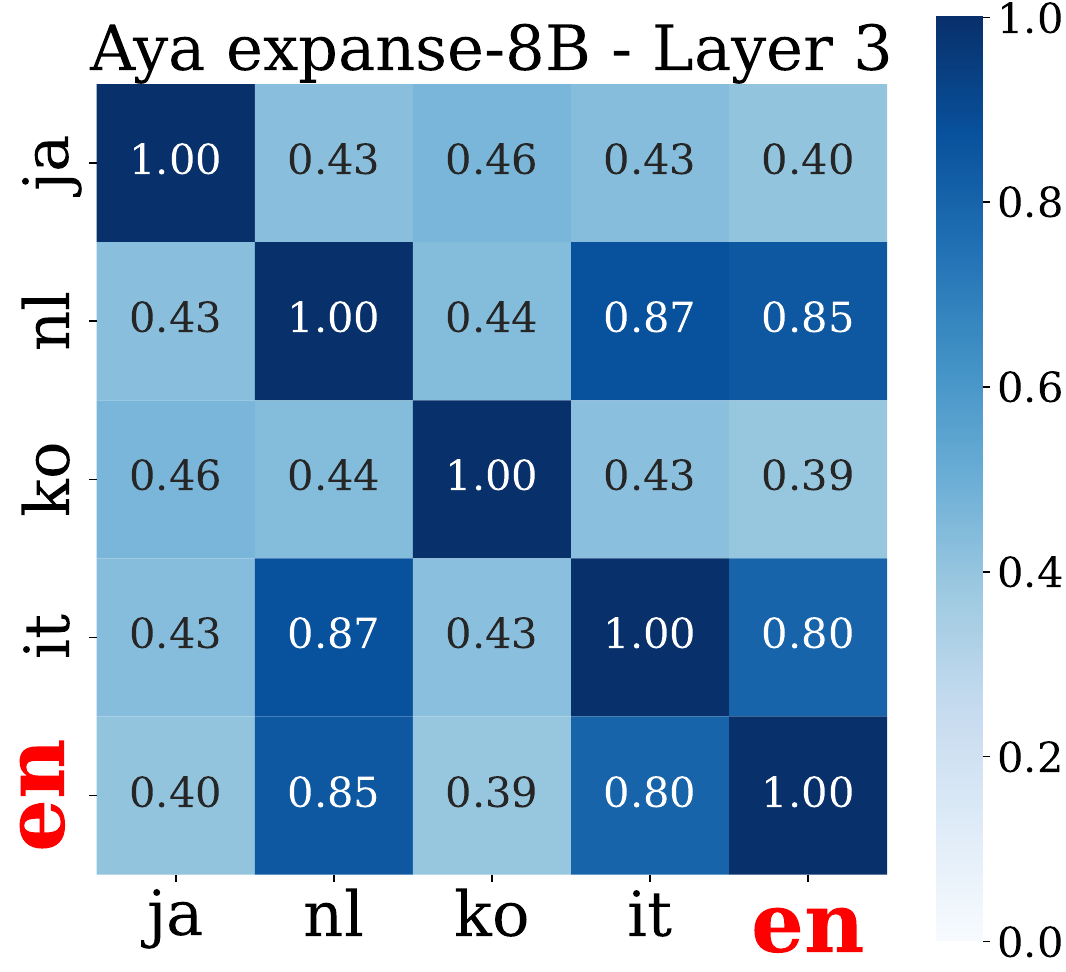}
  \includegraphics[width=0.15\linewidth]{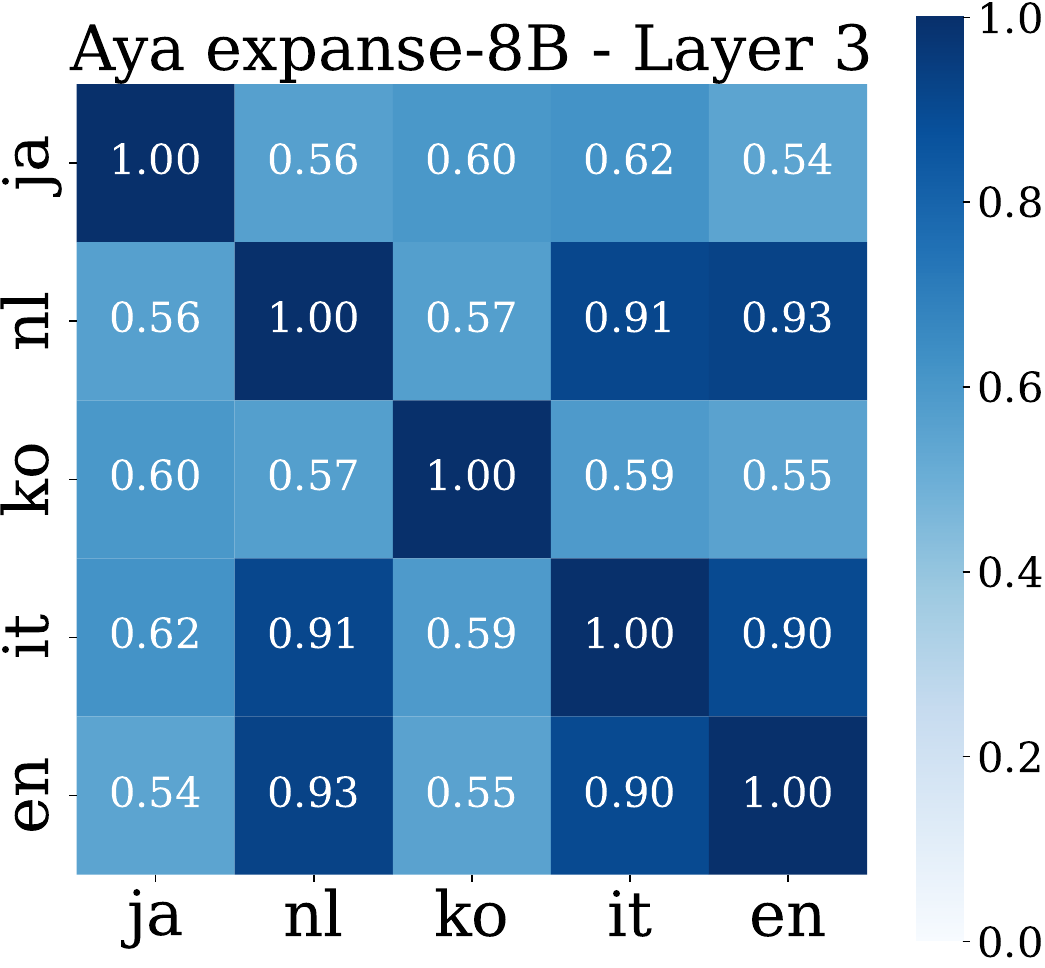}

  \begin{minipage}{0.15\linewidth}\centering \textbf{\textcolor{red}{layer 1 (Type-1)}}\end{minipage}
  \begin{minipage}{0.15\linewidth}\centering layer 1 (baseline)\end{minipage}
  \begin{minipage}{0.15\linewidth}\centering \textbf{\textcolor{red}{layer 2 (Type-1)}}\end{minipage}
  \begin{minipage}{0.15\linewidth}\centering layer 2 (baseline)\end{minipage}
  \begin{minipage}{0.15\linewidth}\centering \textbf{\textcolor{red}{layer 3 (Type-1)}}\end{minipage}
  \begin{minipage}{0.15\linewidth}\centering layer 3 (baseline)\end{minipage}

  \includegraphics[width=0.15\linewidth]{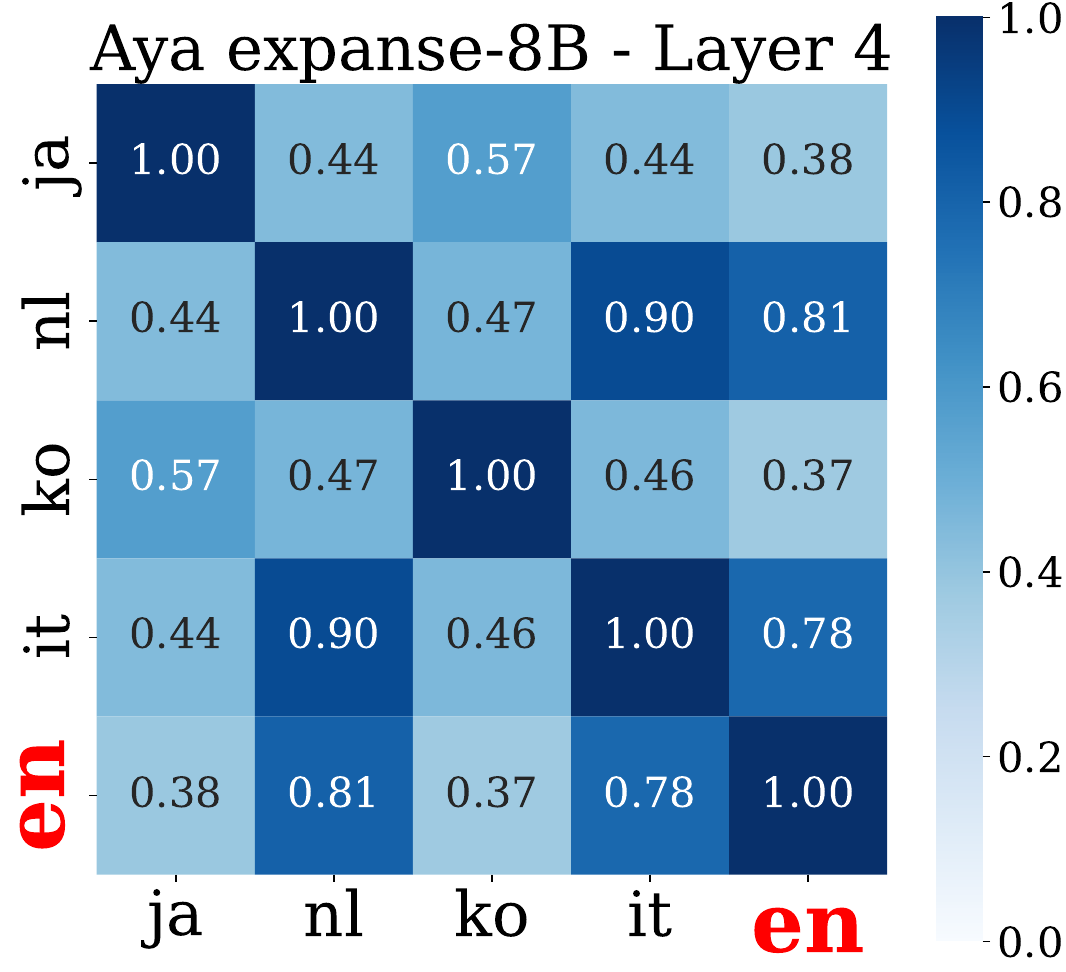}
  \includegraphics[width=0.15\linewidth]{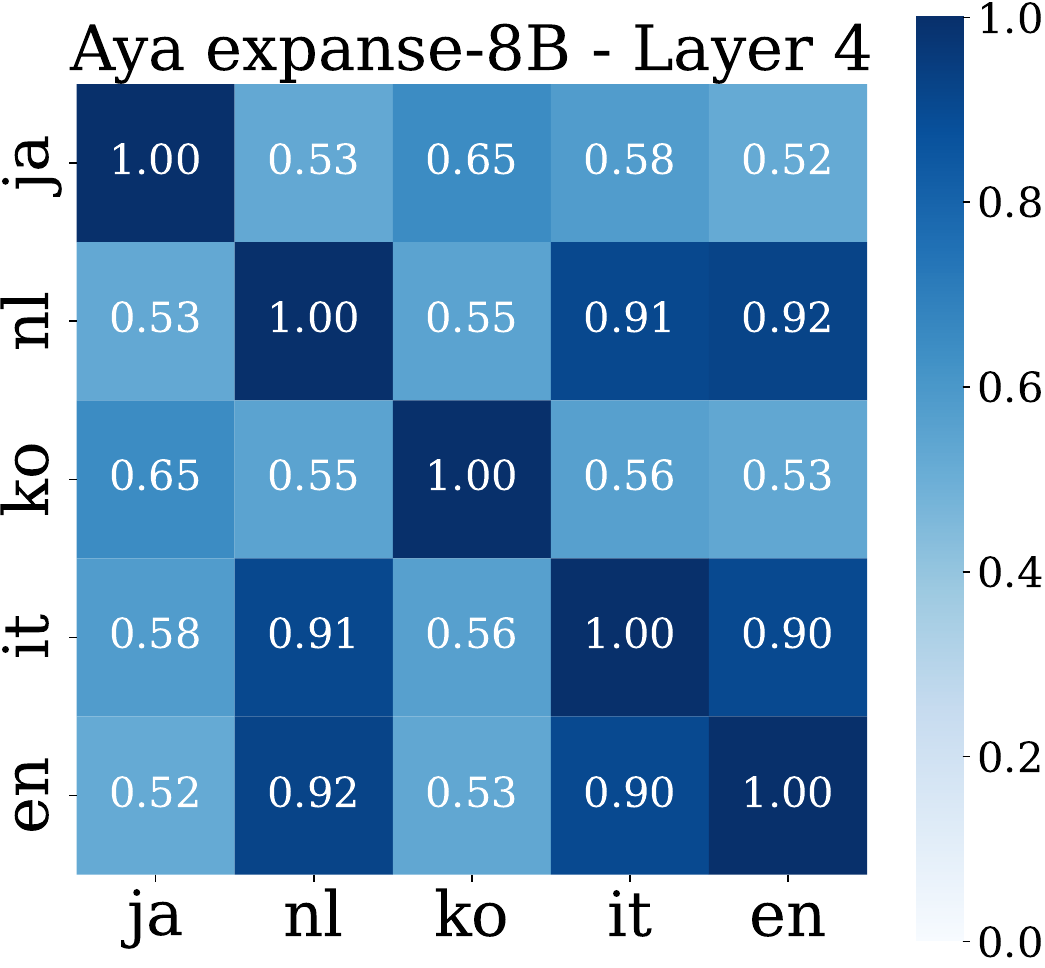}
  \includegraphics[width=0.15\linewidth]{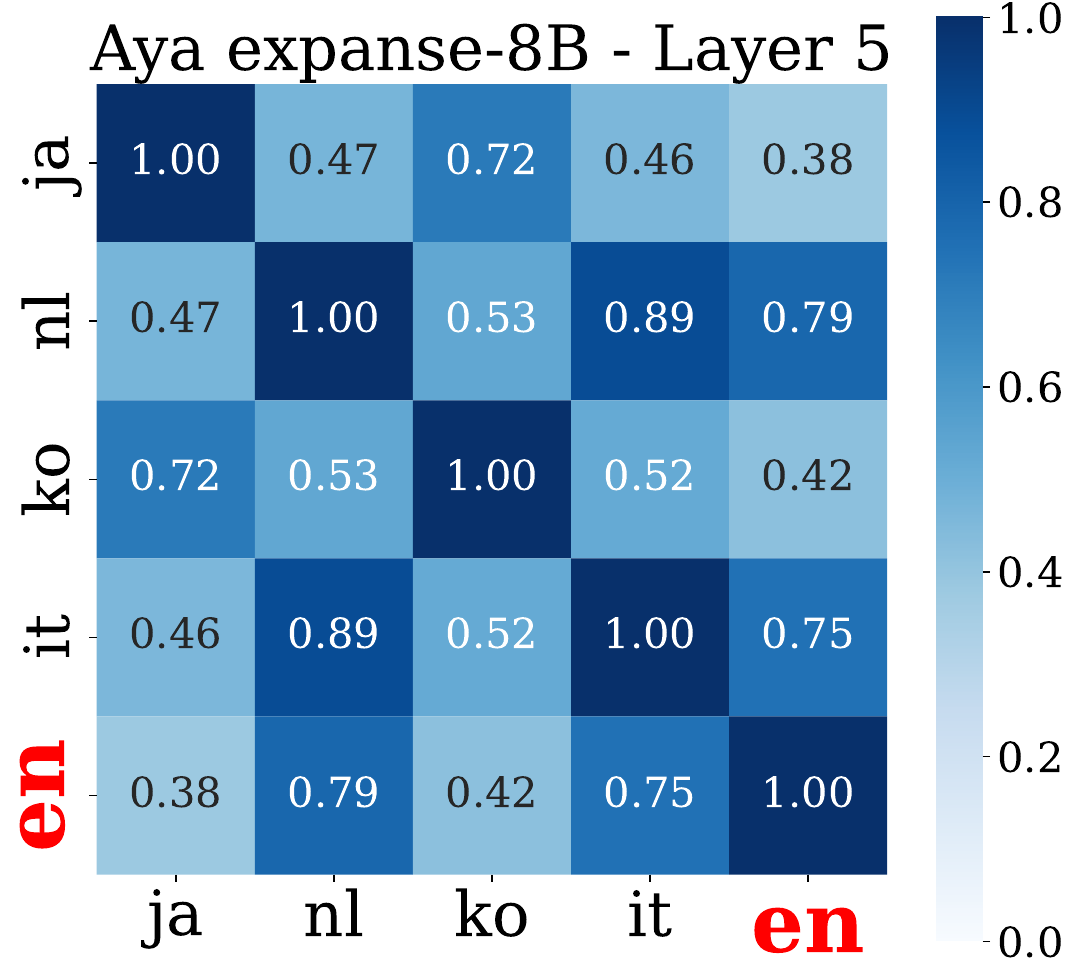}
  \includegraphics[width=0.15\linewidth]{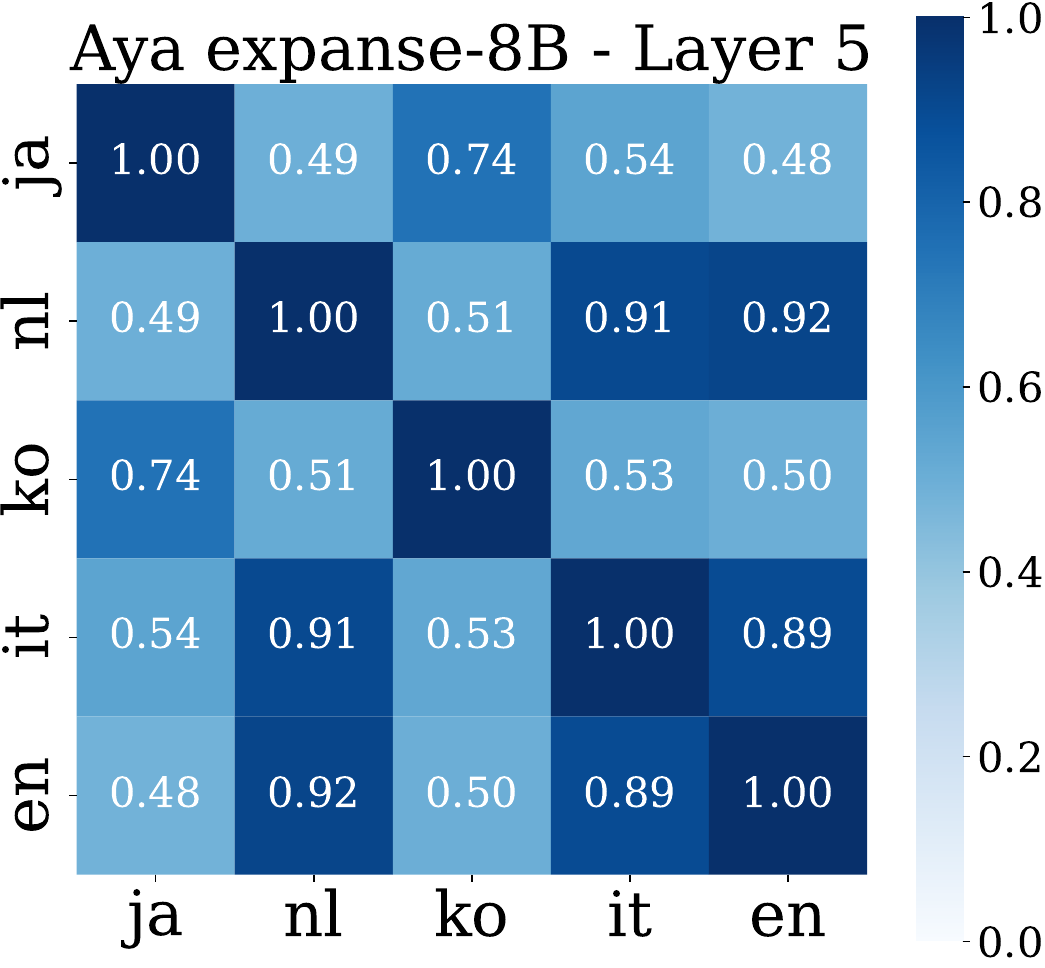}
  \includegraphics[width=0.15\linewidth]{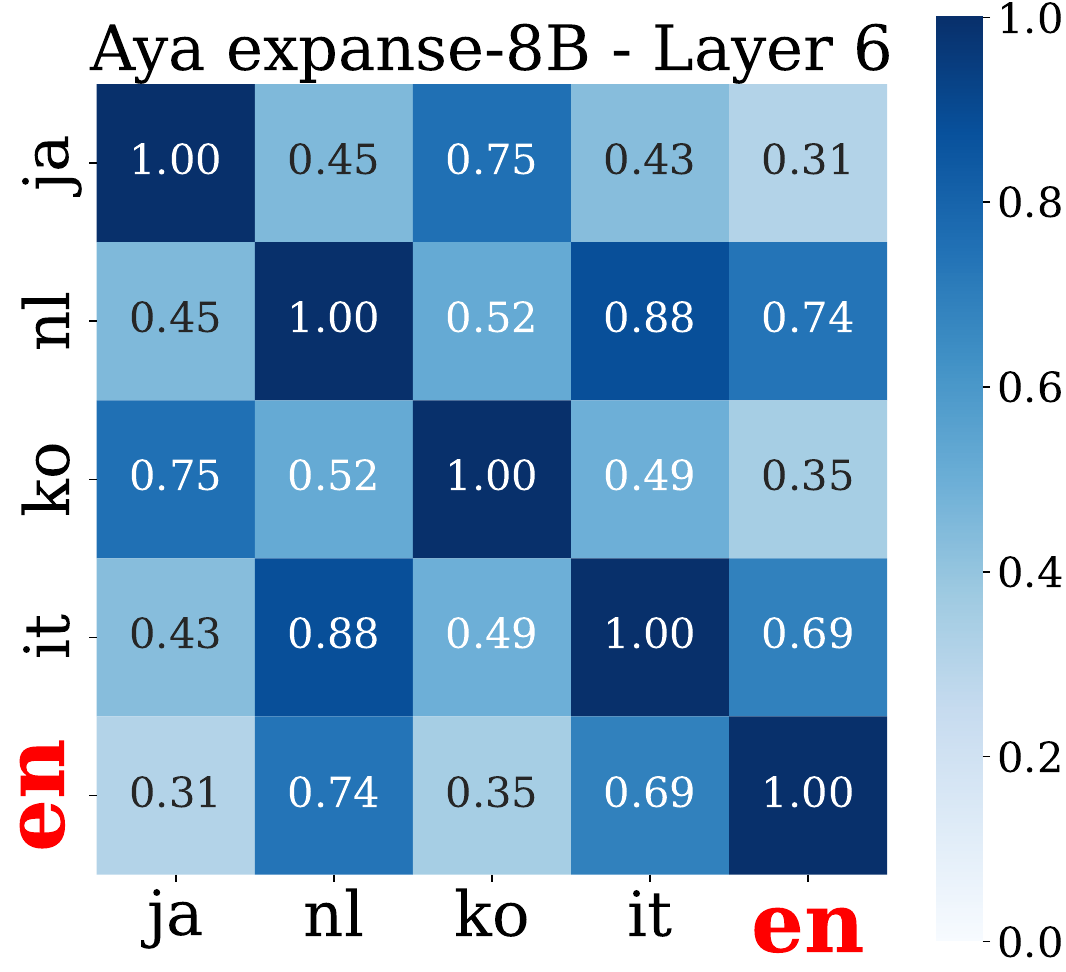}
  \includegraphics[width=0.15\linewidth]{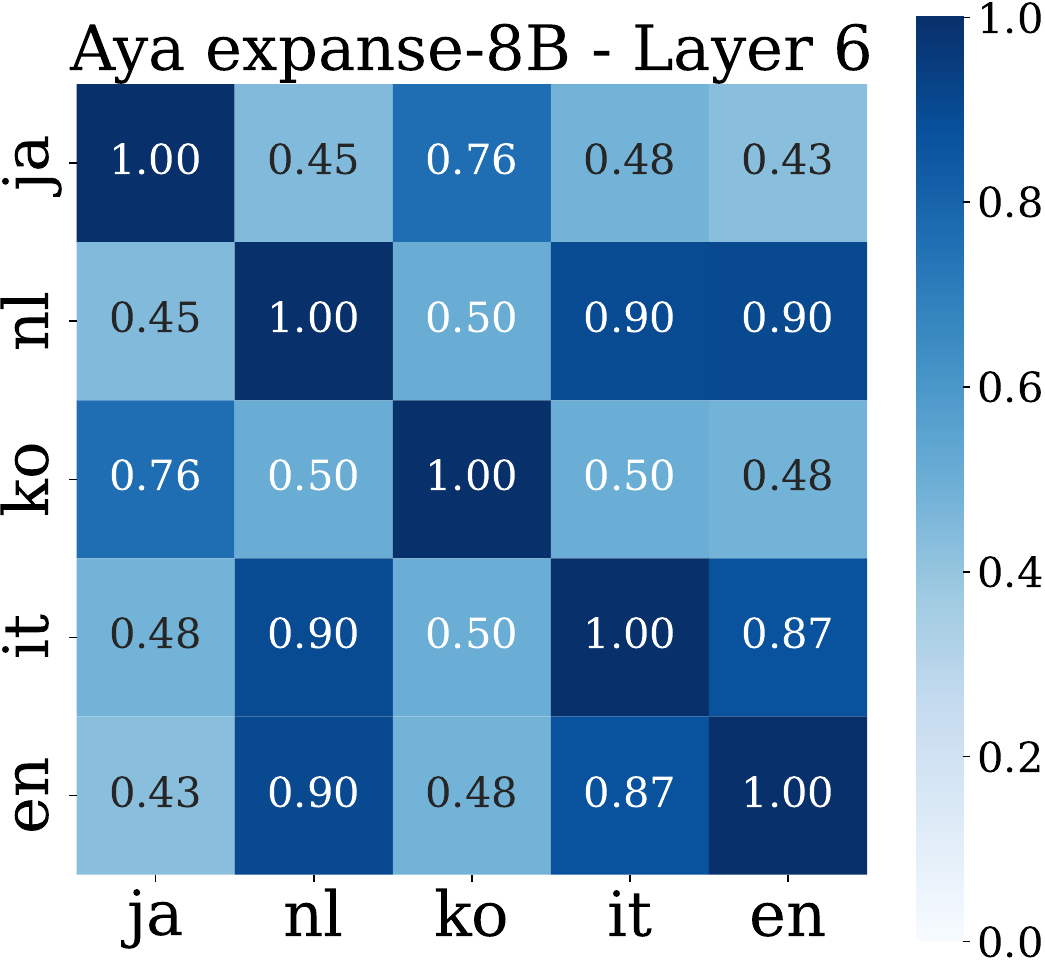}

  \begin{minipage}{0.15\linewidth}\centering \textbf{\textcolor{red}{layer 4 (Type-1)}}\end{minipage}
  \begin{minipage}{0.15\linewidth}\centering layer 4 (baseline)\end{minipage}
  \begin{minipage}{0.15\linewidth}\centering \textbf{\textcolor{red}{layer 5 (Type-1)}}\end{minipage}
  \begin{minipage}{0.15\linewidth}\centering layer 5 (baseline)\end{minipage}
  \begin{minipage}{0.15\linewidth}\centering \textbf{\textcolor{red}{layer 6 (Type-1)}}\end{minipage}
  \begin{minipage}{0.15\linewidth}\centering layer 6 (baseline)\end{minipage}

  \includegraphics[width=0.15\linewidth]{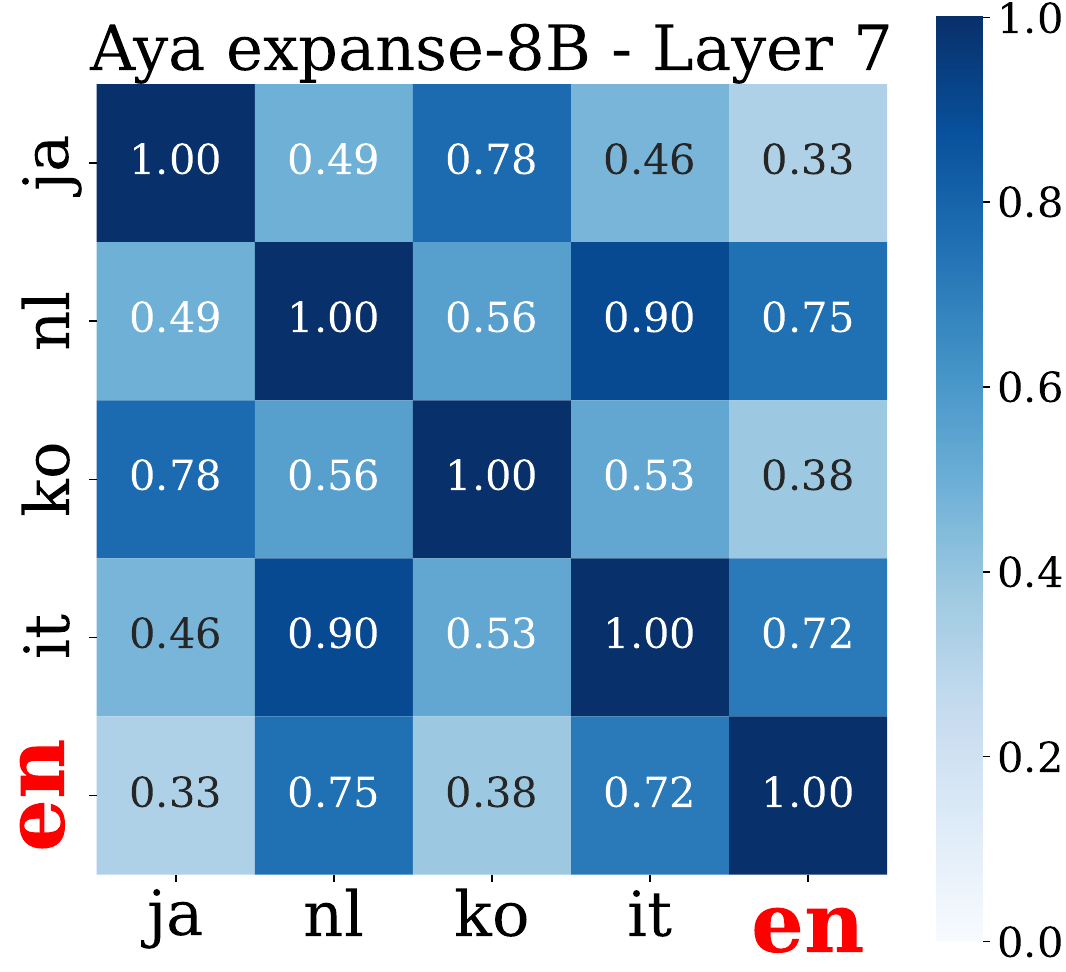}
  \includegraphics[width=0.15\linewidth]{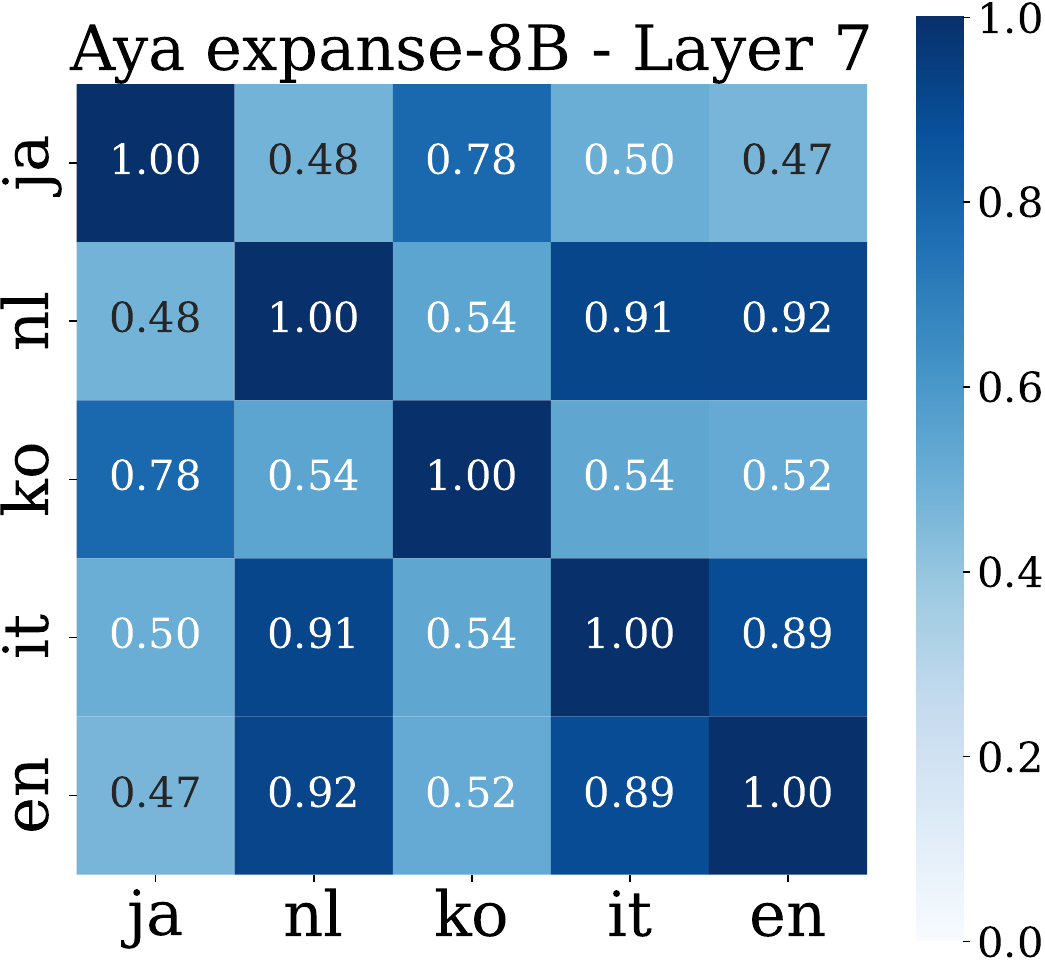}
  \includegraphics[width=0.15\linewidth]{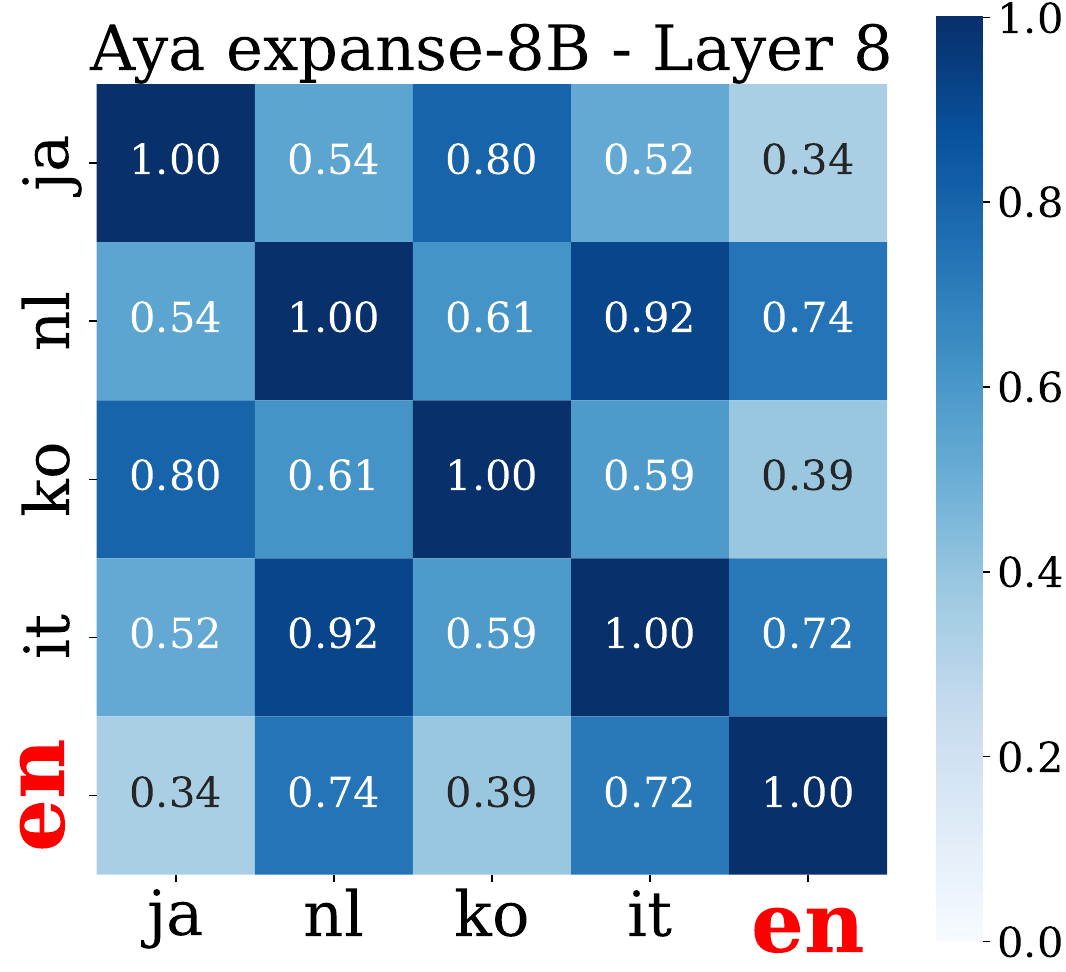}
  \includegraphics[width=0.15\linewidth]{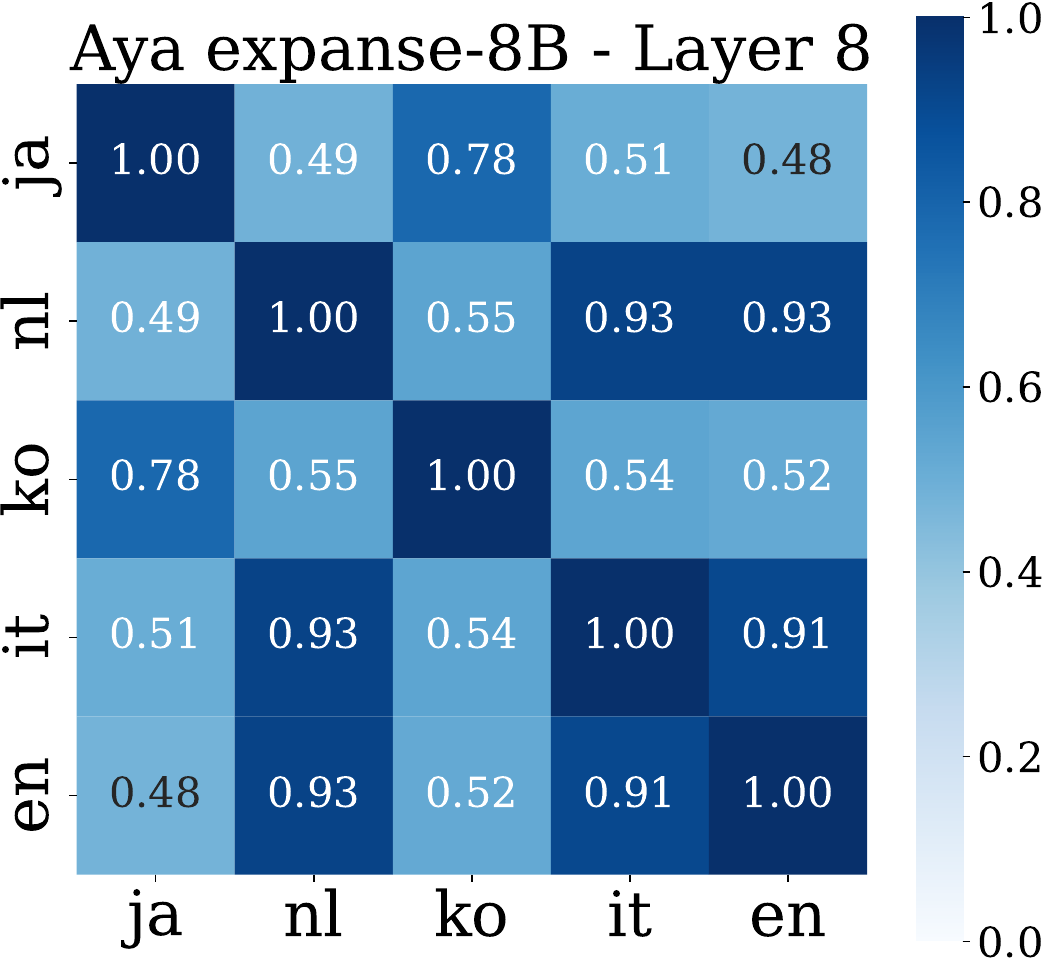}
  \includegraphics[width=0.15\linewidth]{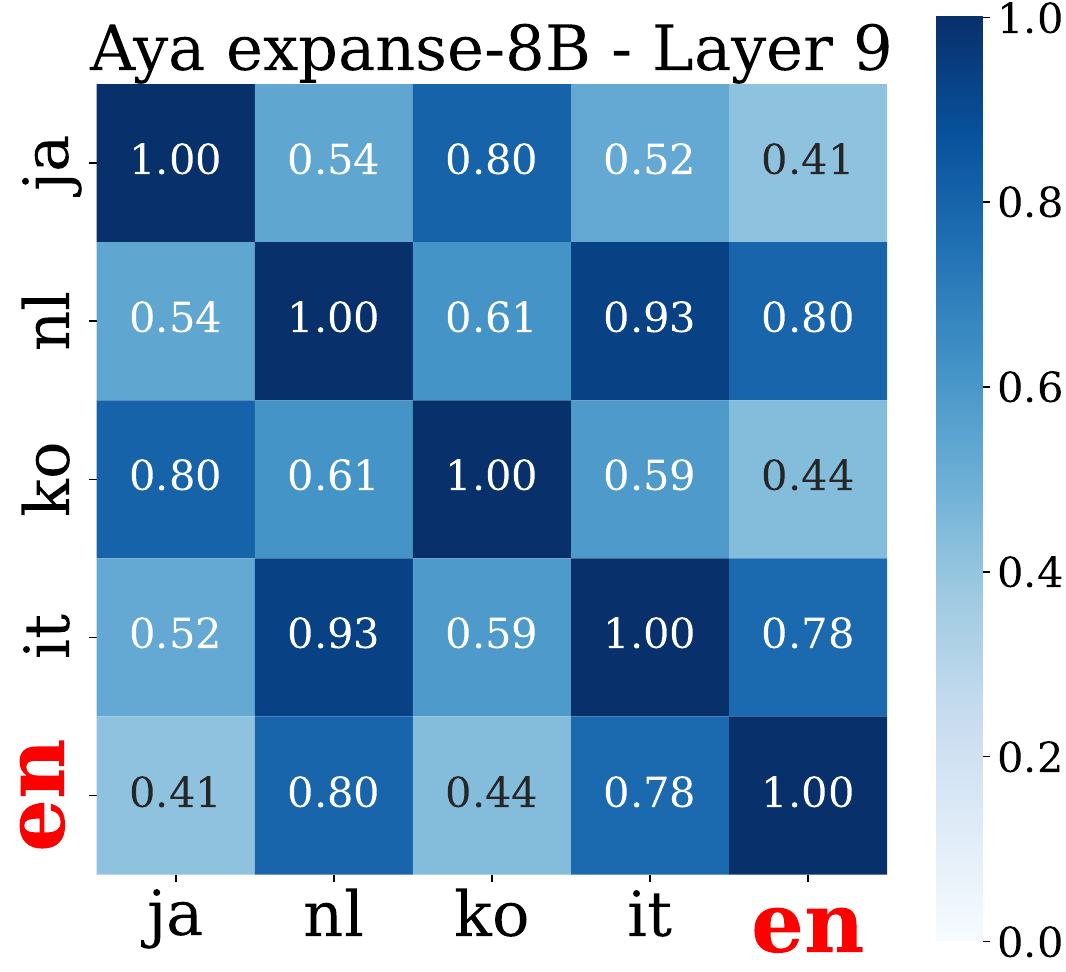}
  \includegraphics[width=0.15\linewidth]{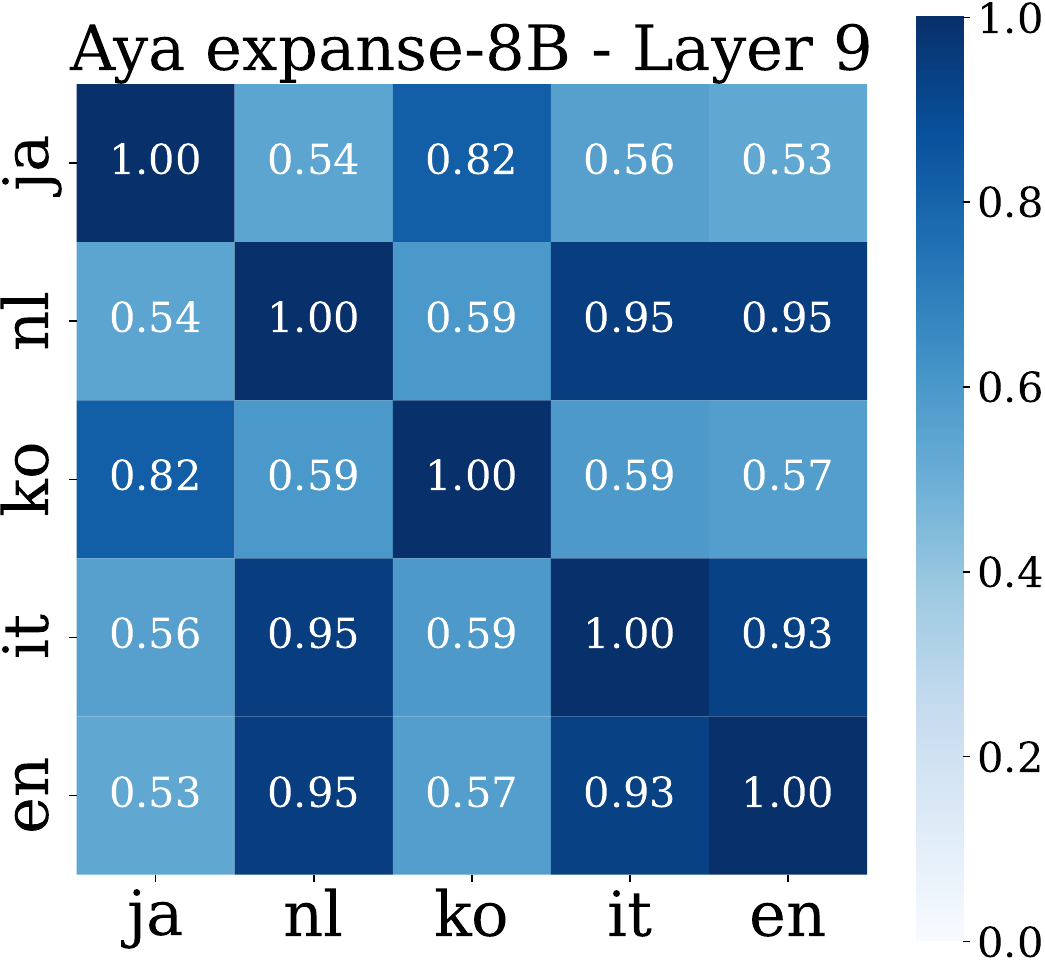}

  \begin{minipage}{0.15\linewidth}\centering \textbf{\textcolor{red}{layer 7 (Type-1)}}\end{minipage}
  \begin{minipage}{0.15\linewidth}\centering layer 7 (baseline)\end{minipage}
  \begin{minipage}{0.15\linewidth}\centering \textbf{\textcolor{red}{layer 8 (Type-1)}}\end{minipage}
  \begin{minipage}{0.15\linewidth}\centering layer 8 (baseline)\end{minipage}
  \begin{minipage}{0.15\linewidth}\centering \textbf{\textcolor{red}{layer 9 (Type-1)}}\end{minipage}
  \begin{minipage}{0.15\linewidth}\centering layer 9 (baseline)\end{minipage}

  \includegraphics[width=0.15\linewidth]{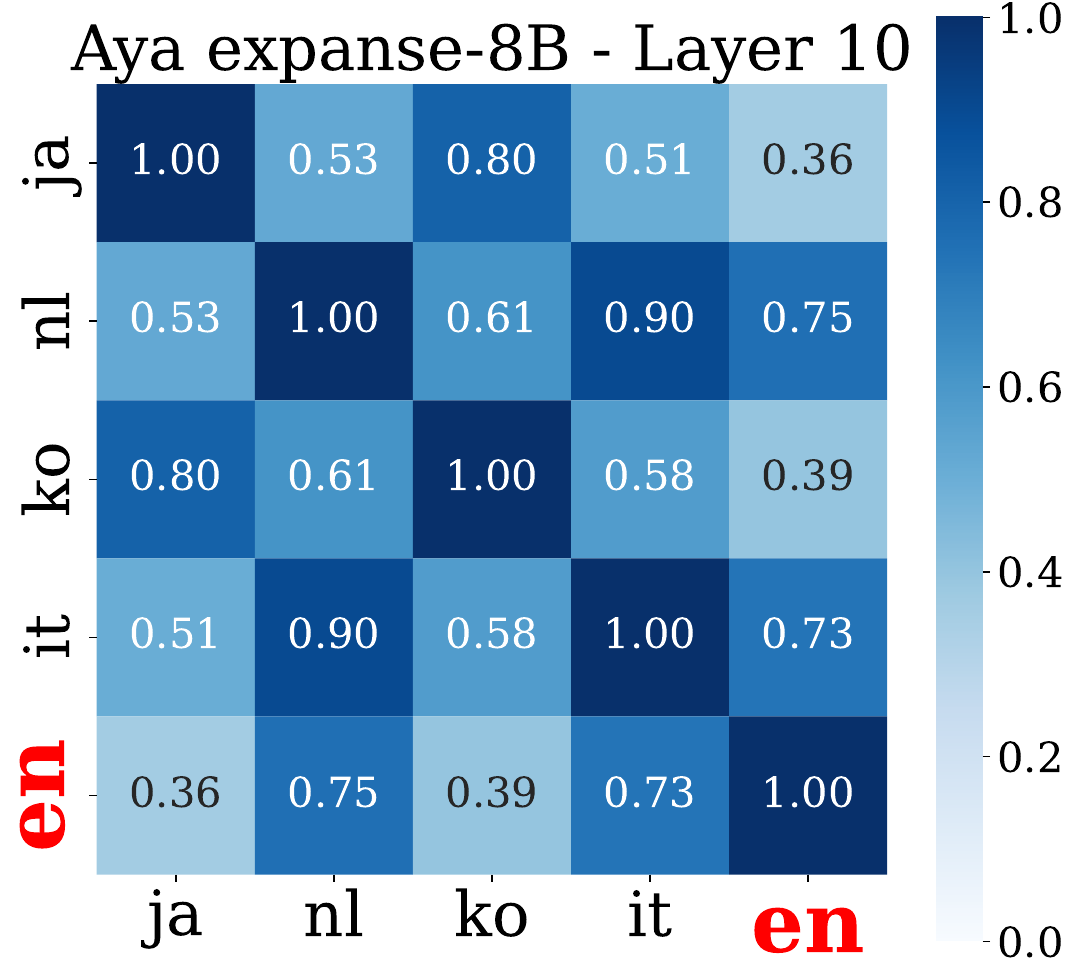}
  \includegraphics[width=0.15\linewidth]{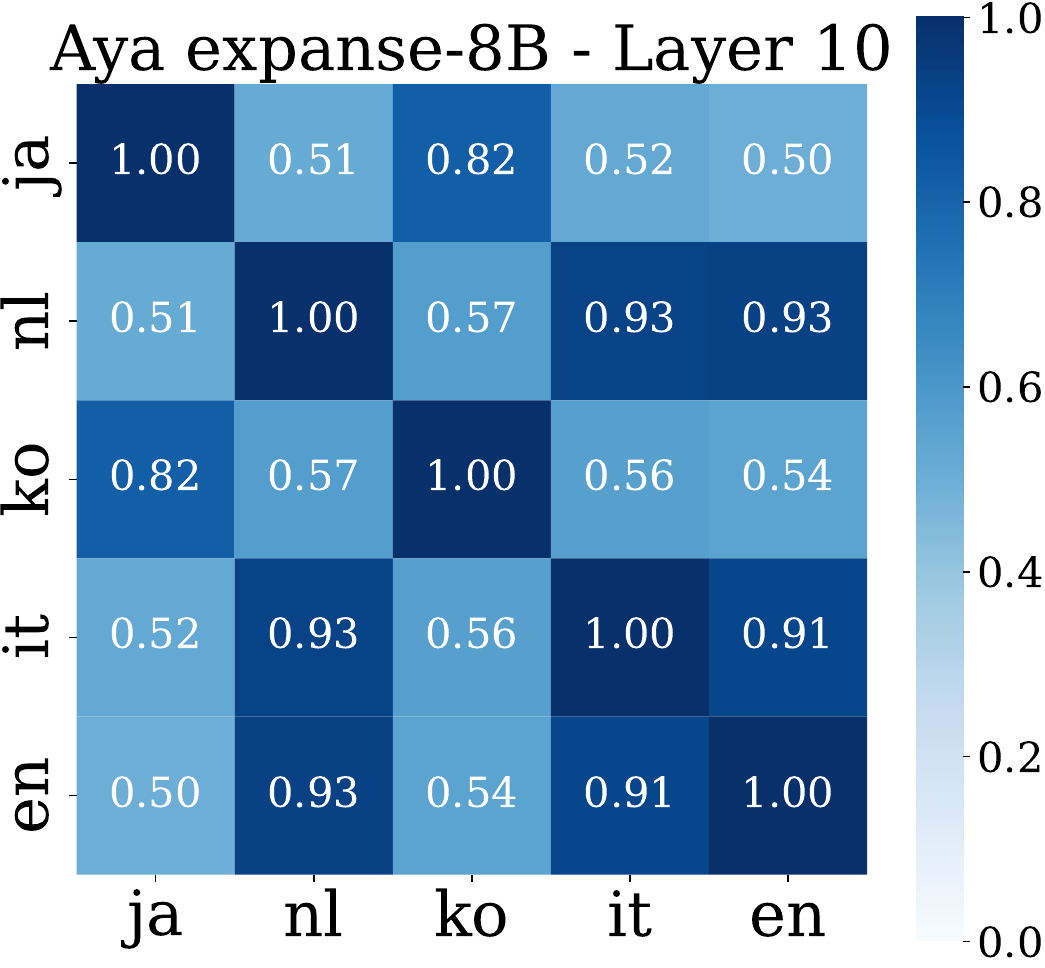}
  \includegraphics[width=0.15\linewidth]{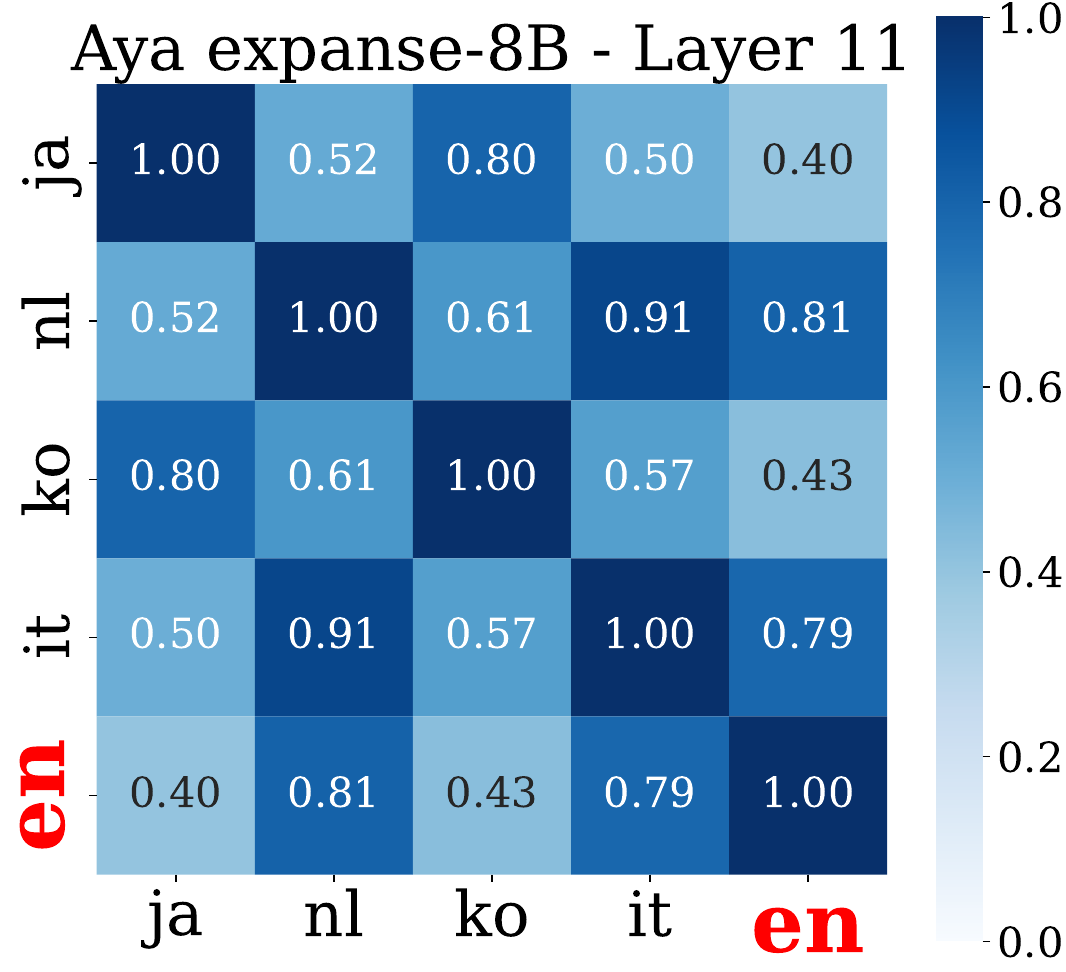}
  \includegraphics[width=0.15\linewidth]{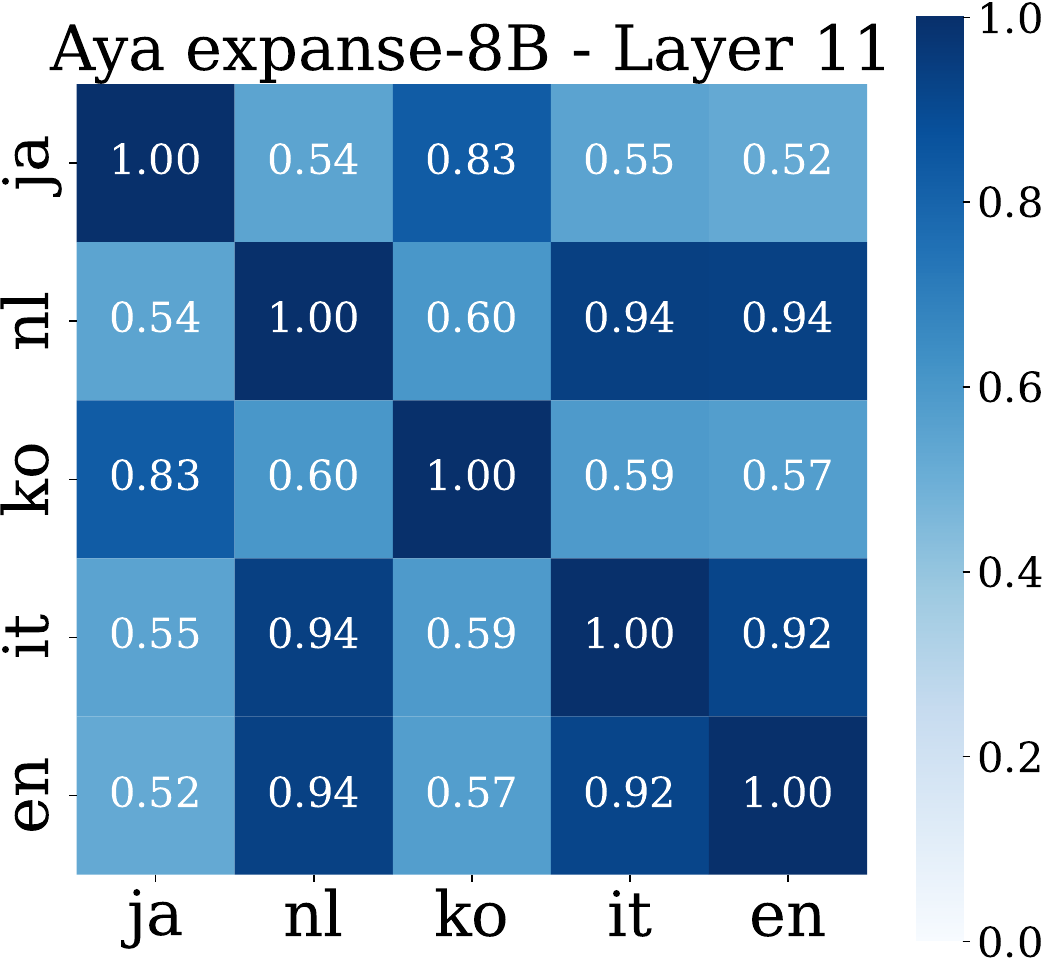}
  \includegraphics[width=0.15\linewidth]{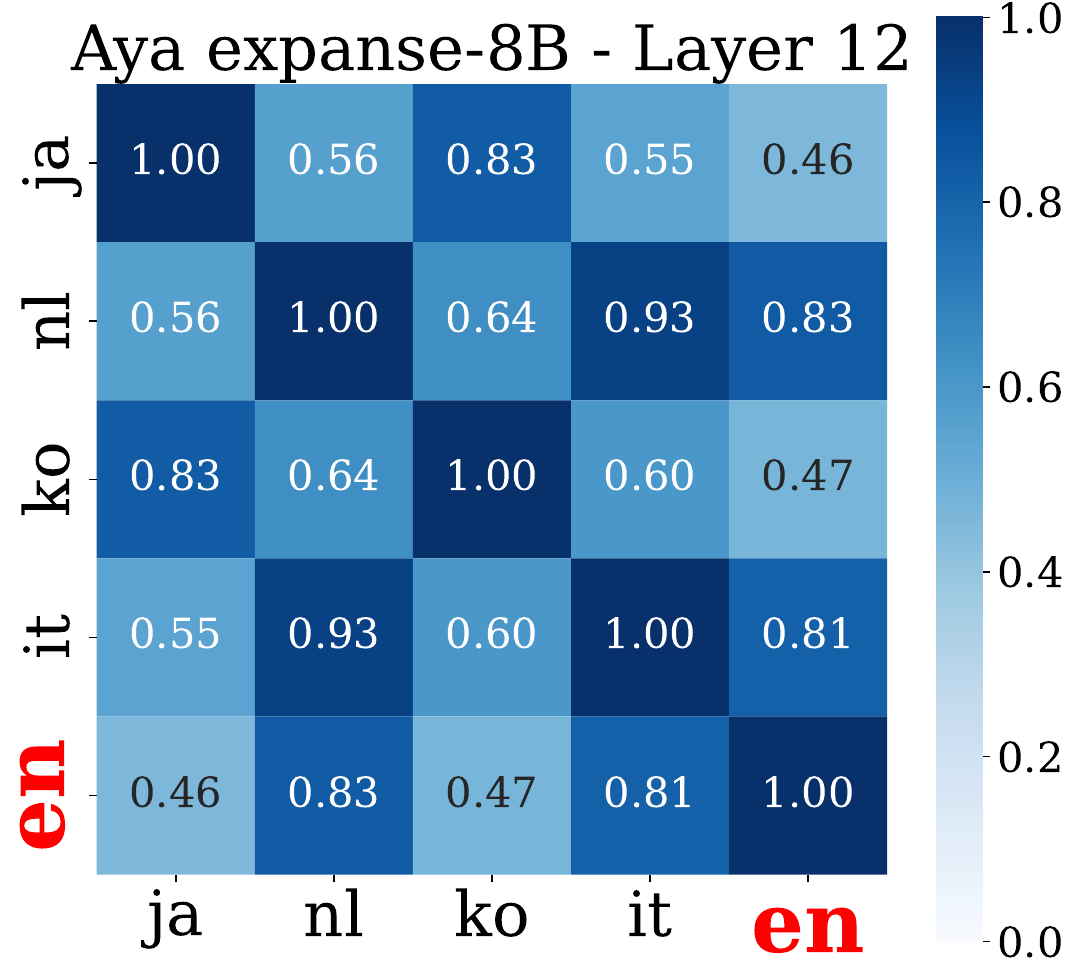}
  \includegraphics[width=0.15\linewidth]{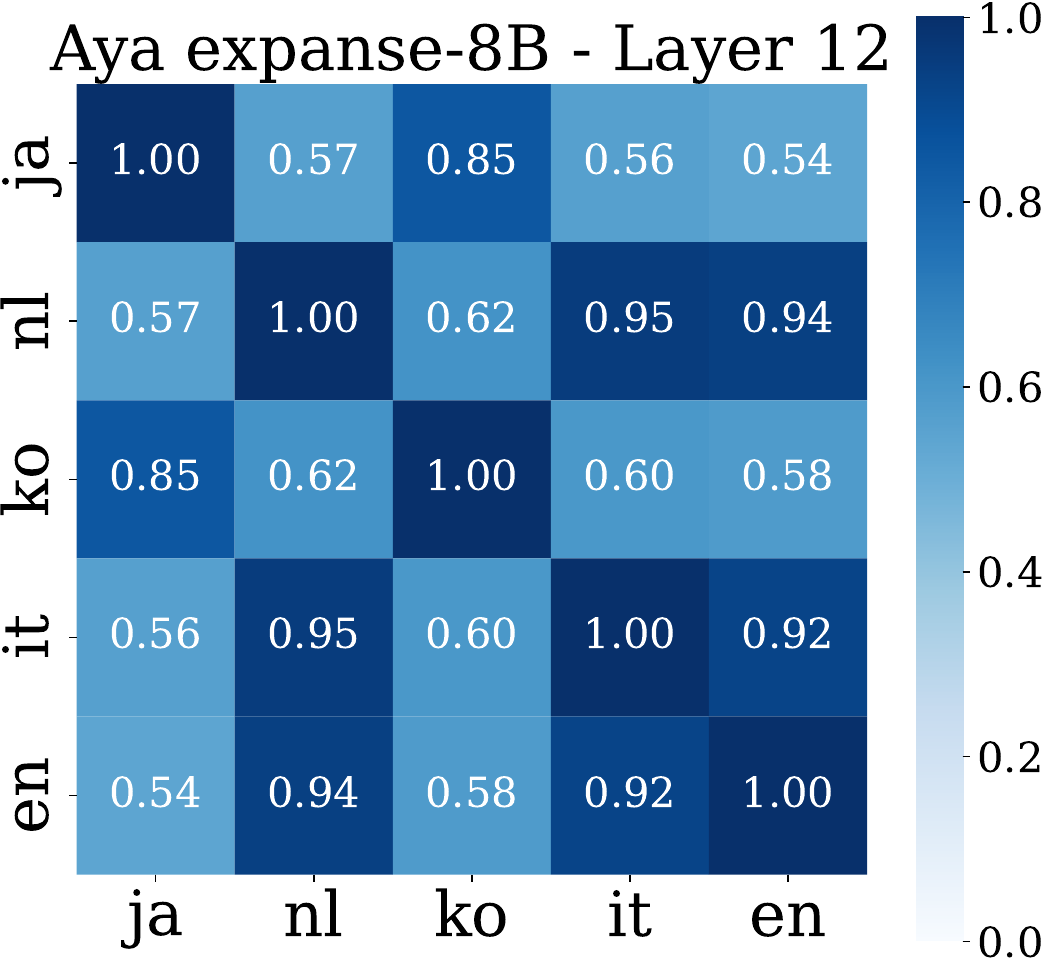}

  \begin{minipage}{0.15\linewidth}\centering \textbf{\textcolor{red}{layer 10 (Type-1)}}\end{minipage}
  \begin{minipage}{0.15\linewidth}\centering layer 10 (baseline)\end{minipage}
  \begin{minipage}{0.15\linewidth}\centering \textbf{\textcolor{red}{layer 11 (Type-1)}}\end{minipage}
  \begin{minipage}{0.15\linewidth}\centering layer 11 (baseline)\end{minipage}
  \begin{minipage}{0.15\linewidth}\centering \textbf{\textcolor{red}{layer 12 (Type-1)}}\end{minipage}
  \begin{minipage}{0.15\linewidth}\centering layer 12 (baseline)\end{minipage}

  \includegraphics[width=0.15\linewidth]{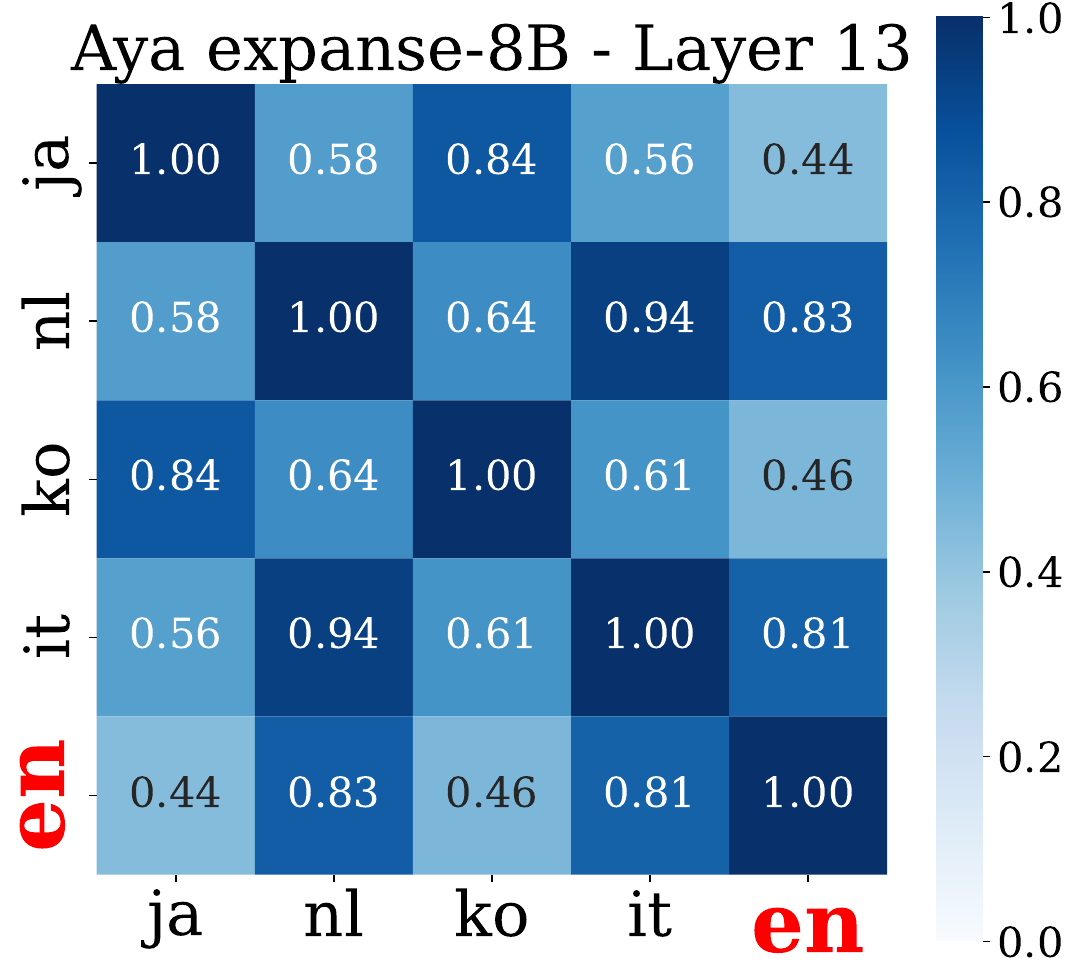}
  \includegraphics[width=0.15\linewidth]{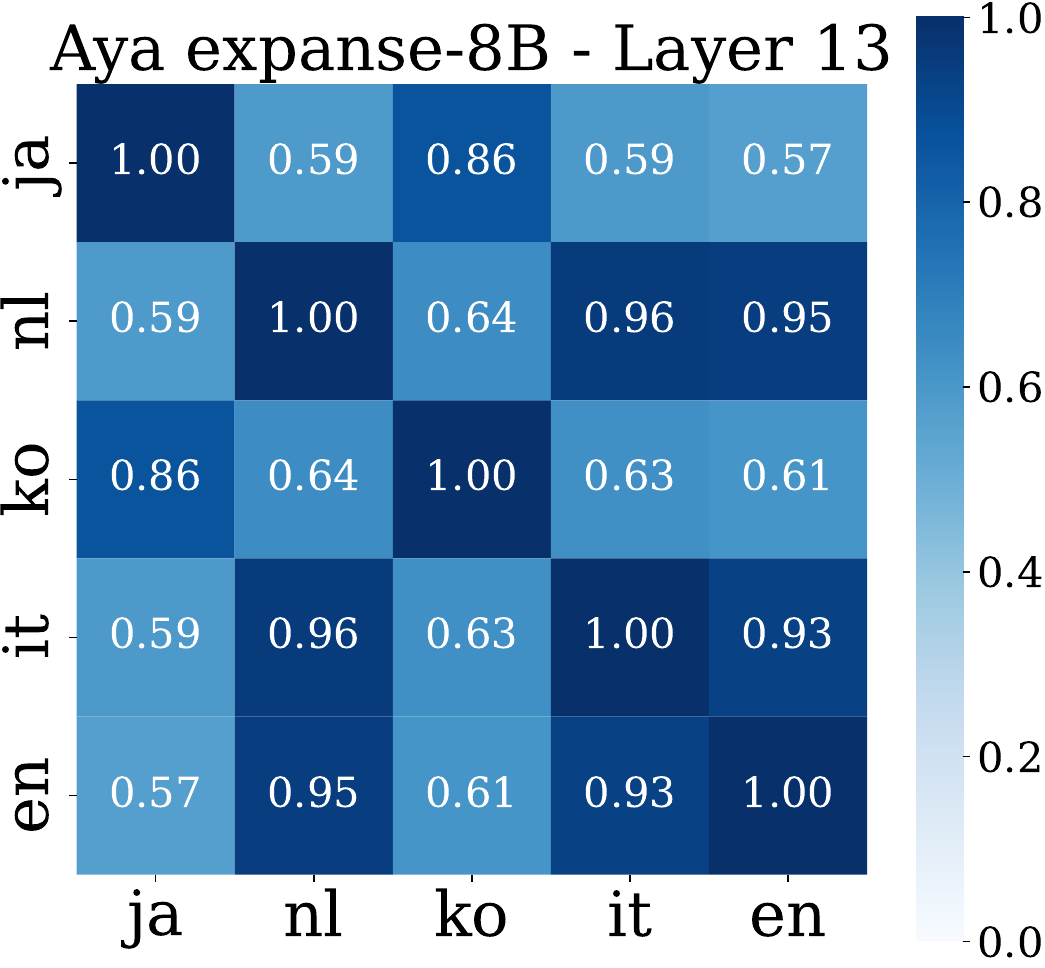}
  \includegraphics[width=0.15\linewidth]{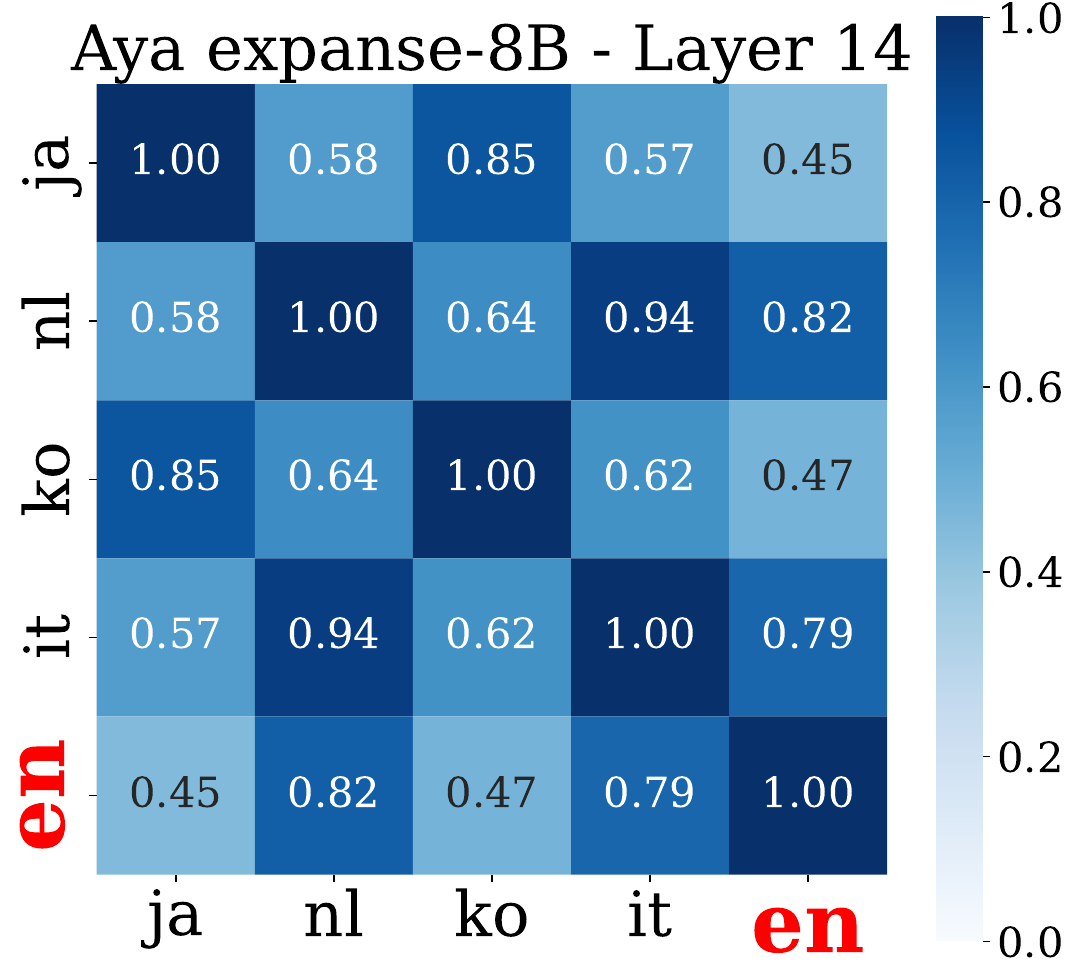}
  \includegraphics[width=0.15\linewidth]{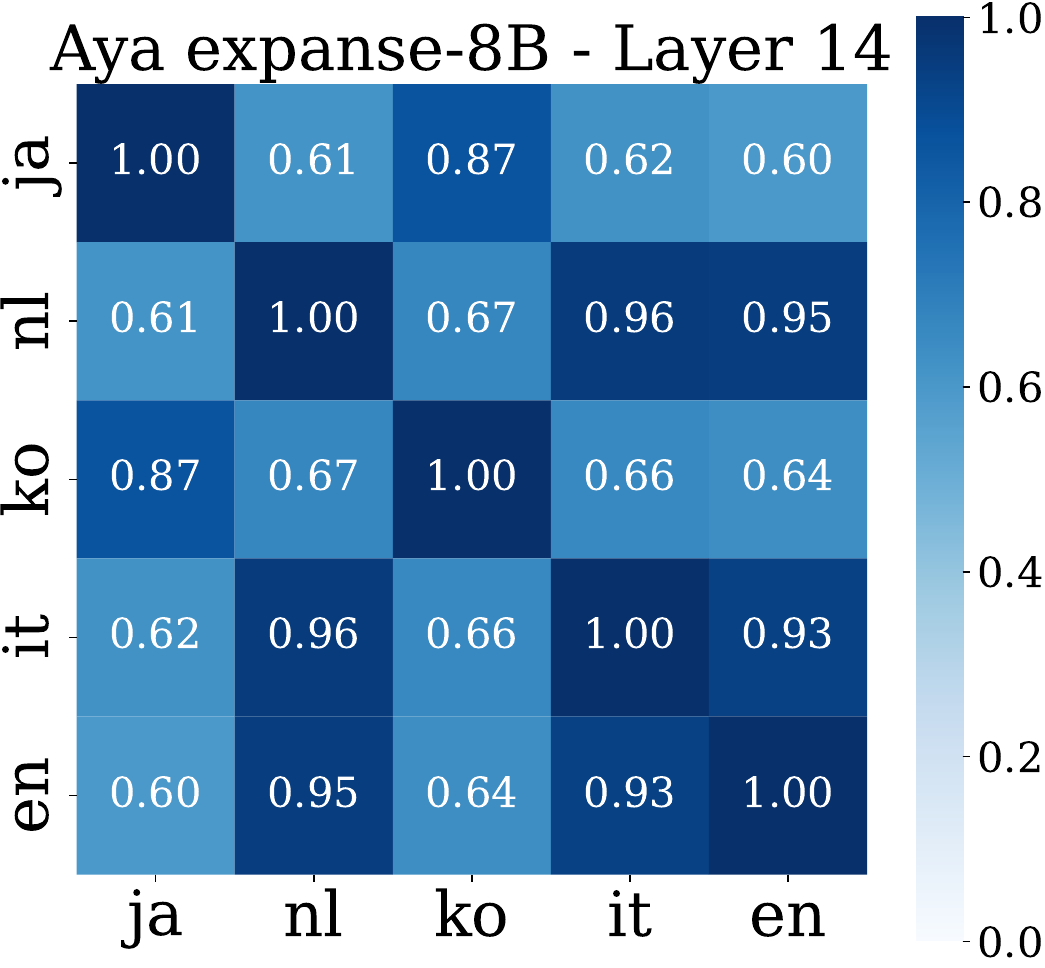}
  \includegraphics[width=0.15\linewidth]{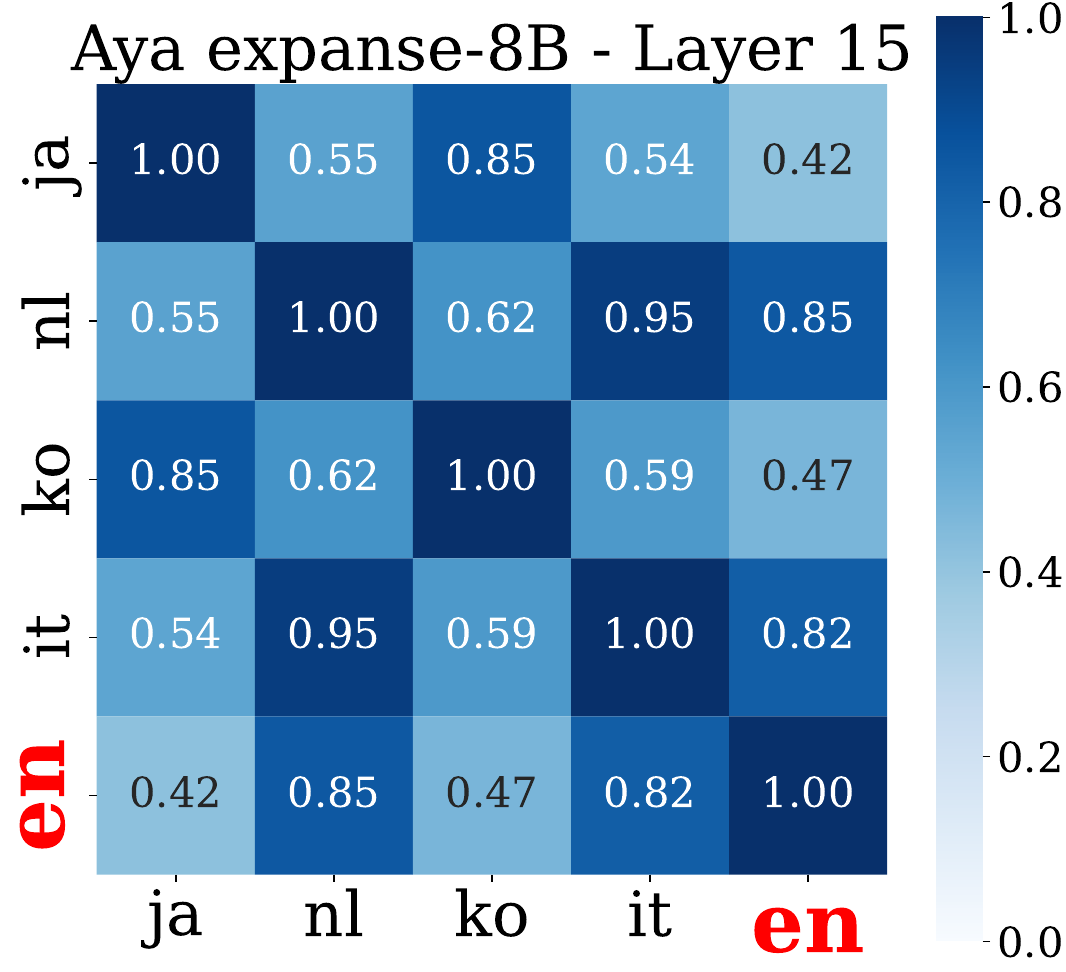}
  \includegraphics[width=0.15\linewidth]{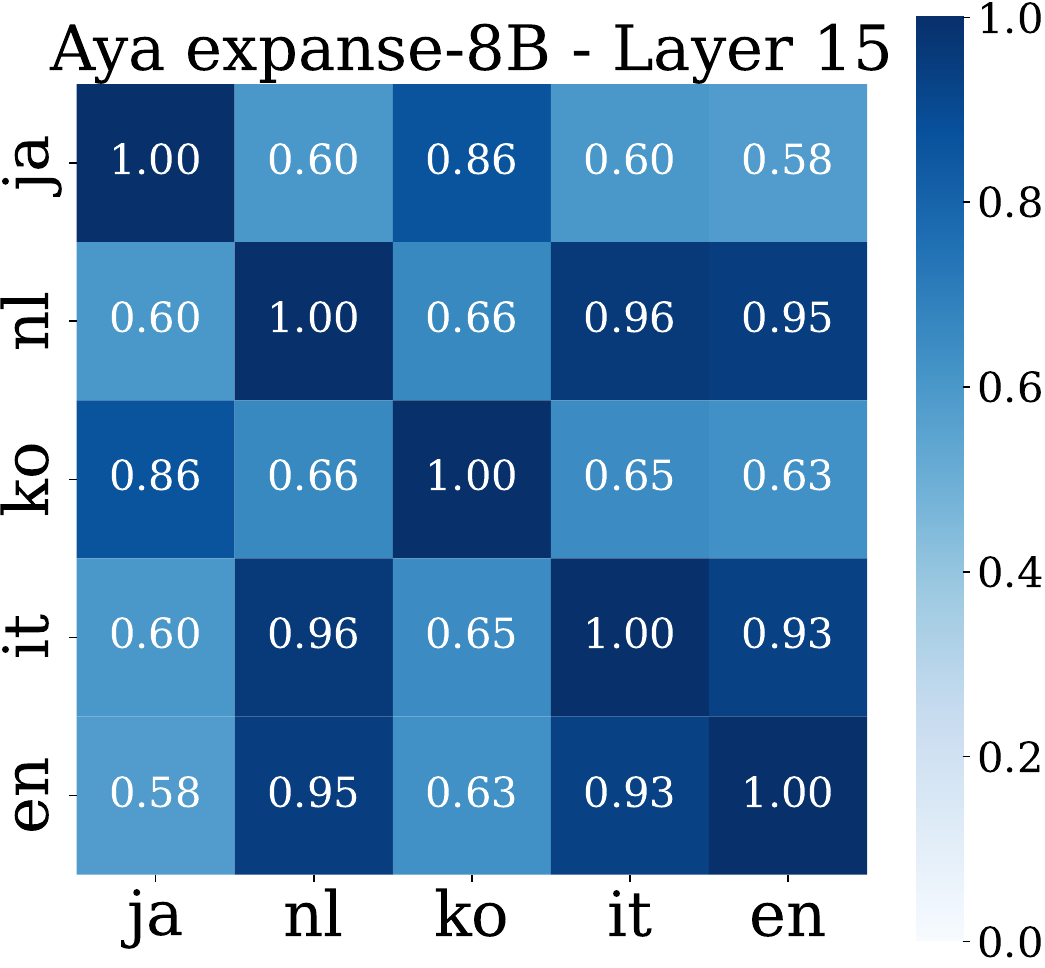}

  \begin{minipage}{0.15\linewidth}\centering \textbf{\textcolor{red}{layer 13 (Type-1)}}\end{minipage}
  \begin{minipage}{0.15\linewidth}\centering layer 13 (baseline)\end{minipage}
  \begin{minipage}{0.15\linewidth}\centering \textbf{\textcolor{red}{layer 14 (Type-1)}}\end{minipage}
  \begin{minipage}{0.15\linewidth}\centering layer 14 (baseline)\end{minipage}
  \begin{minipage}{0.15\linewidth}\centering \textbf{\textcolor{red}{layer 15 (Type-1)}}\end{minipage}
  \begin{minipage}{0.15\linewidth}\centering layer 15 (baseline)\end{minipage}

  \includegraphics[width=0.15\linewidth]{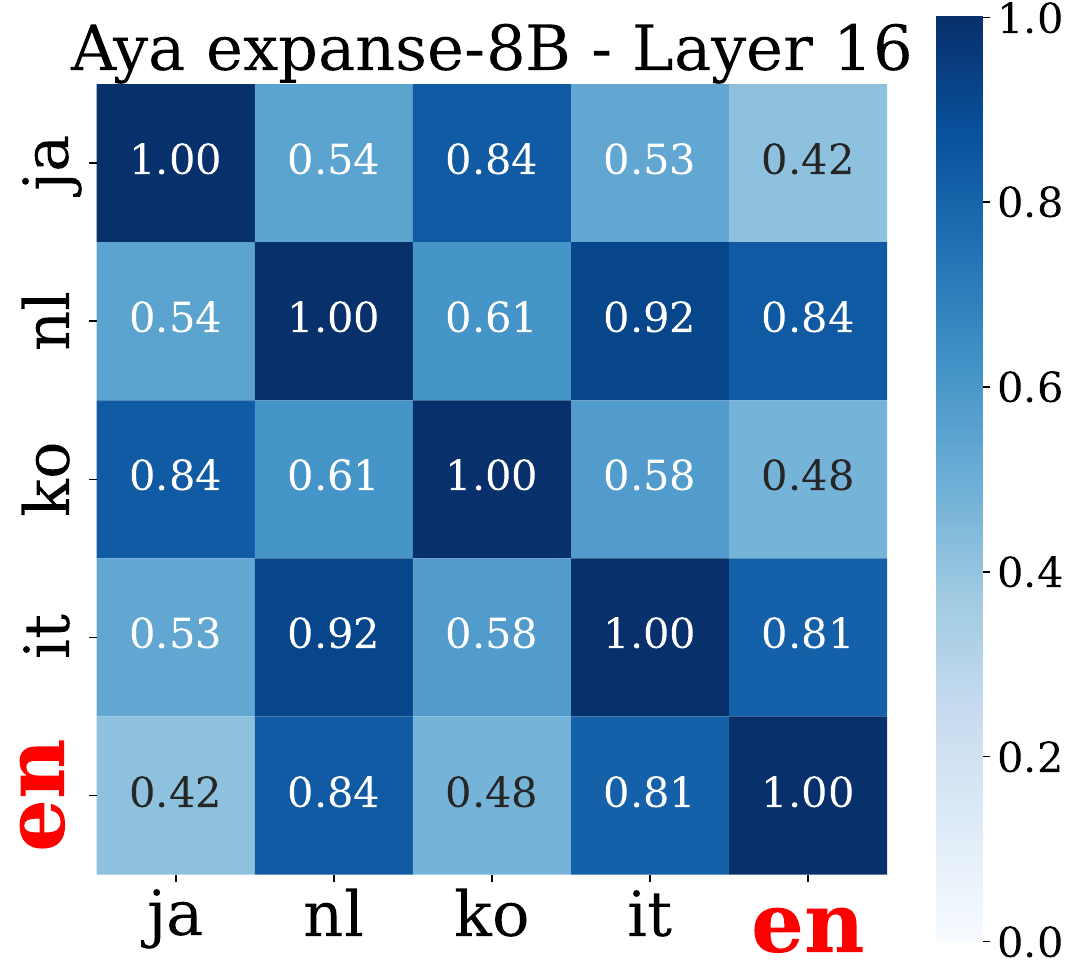}
  \includegraphics[width=0.15\linewidth]{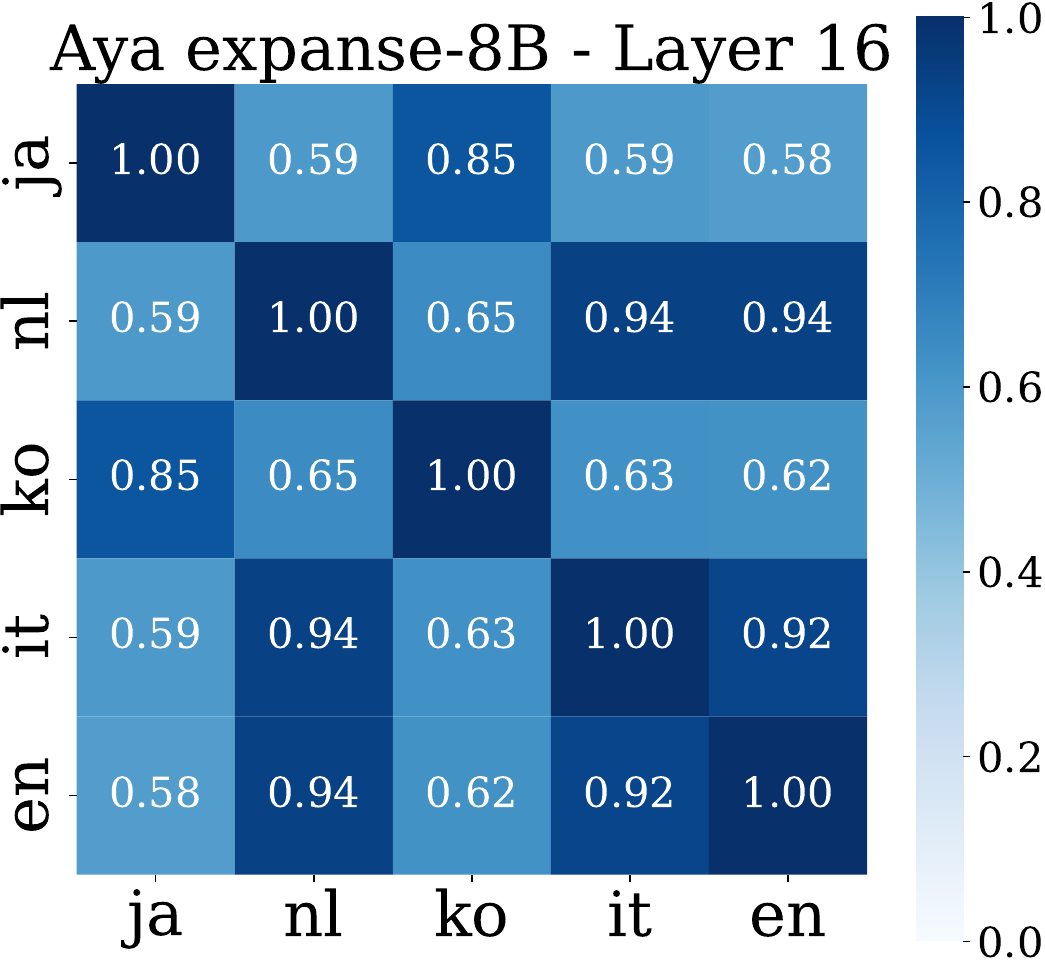}
  \includegraphics[width=0.15\linewidth]{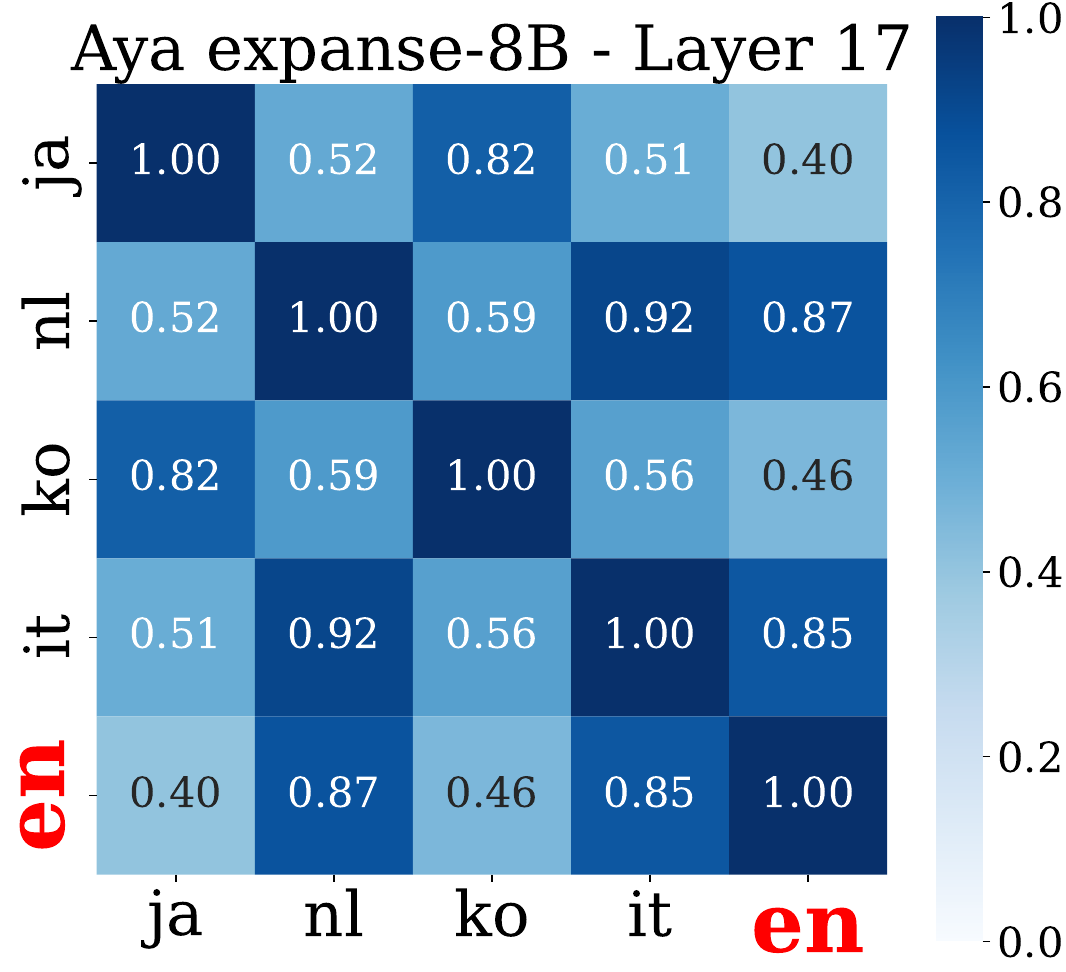}
  \includegraphics[width=0.15\linewidth]{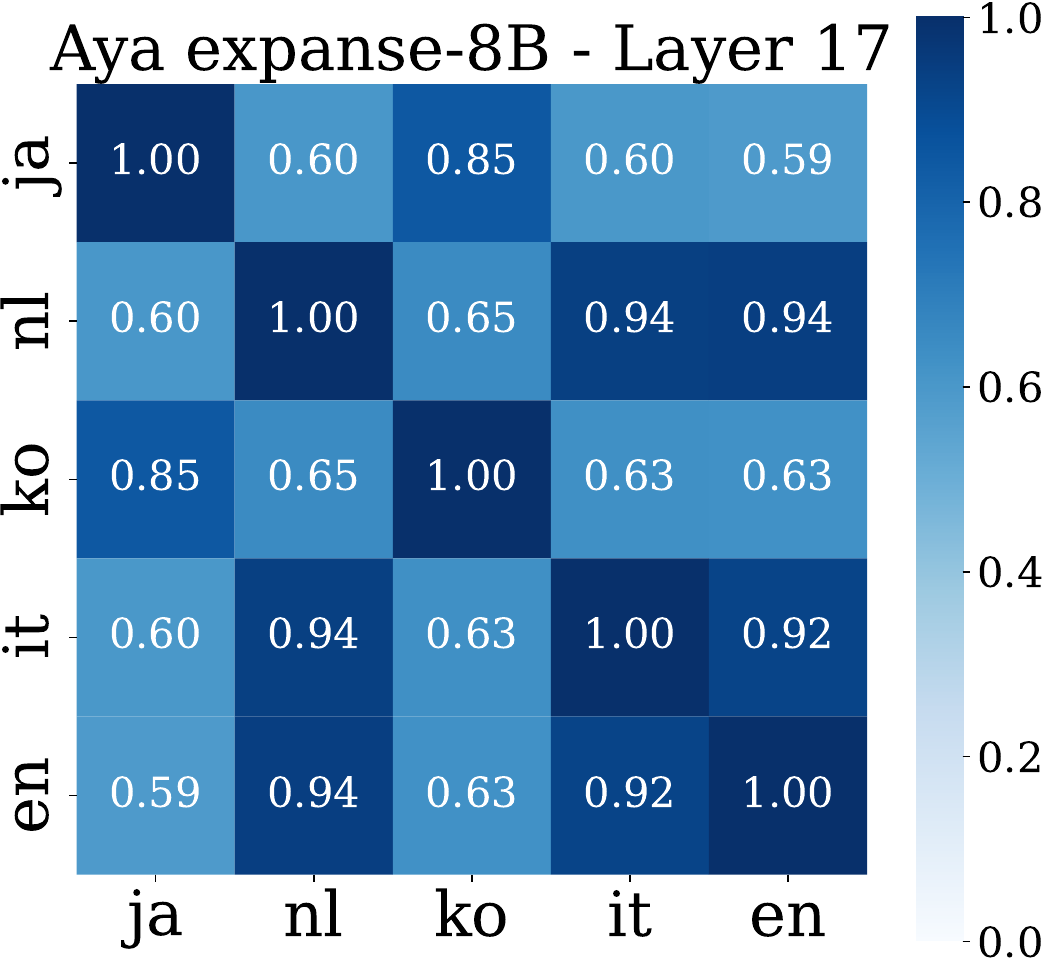}
  \includegraphics[width=0.15\linewidth]{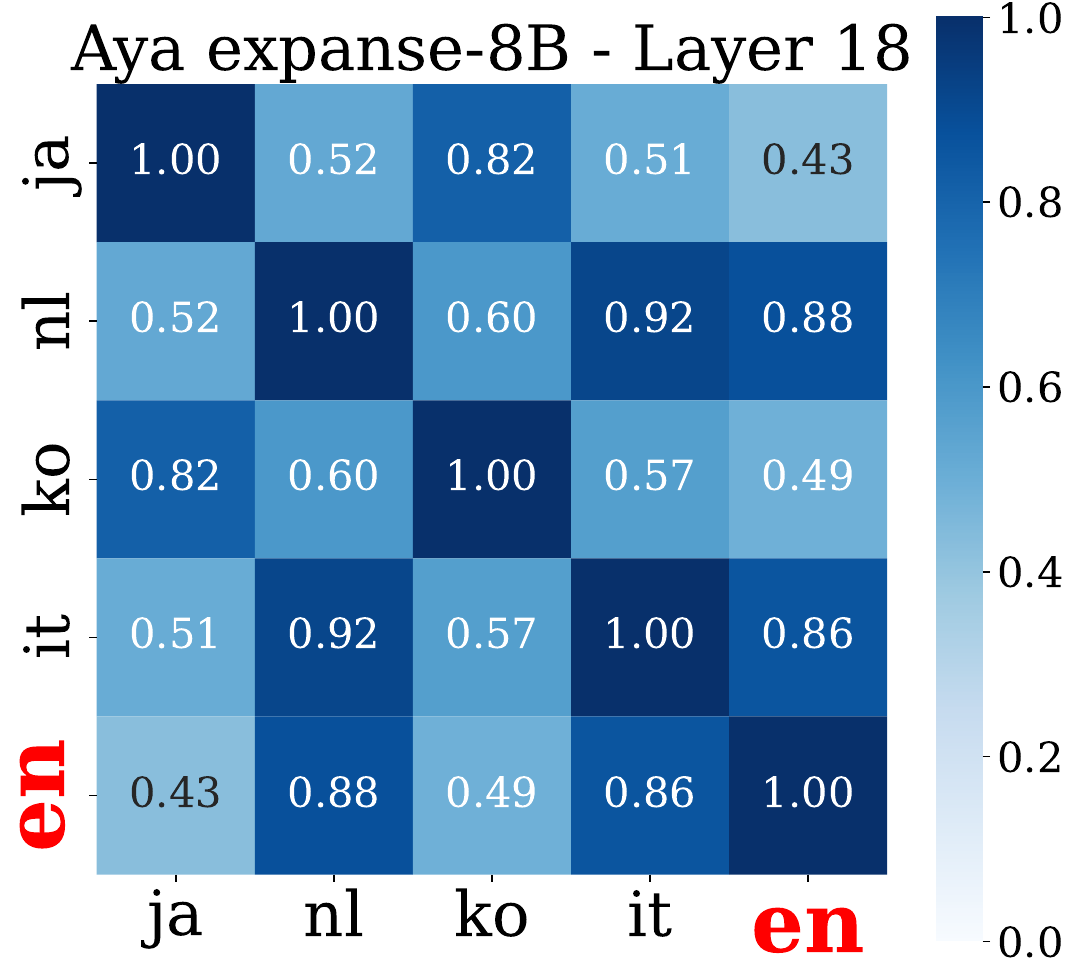}
  \includegraphics[width=0.15\linewidth]{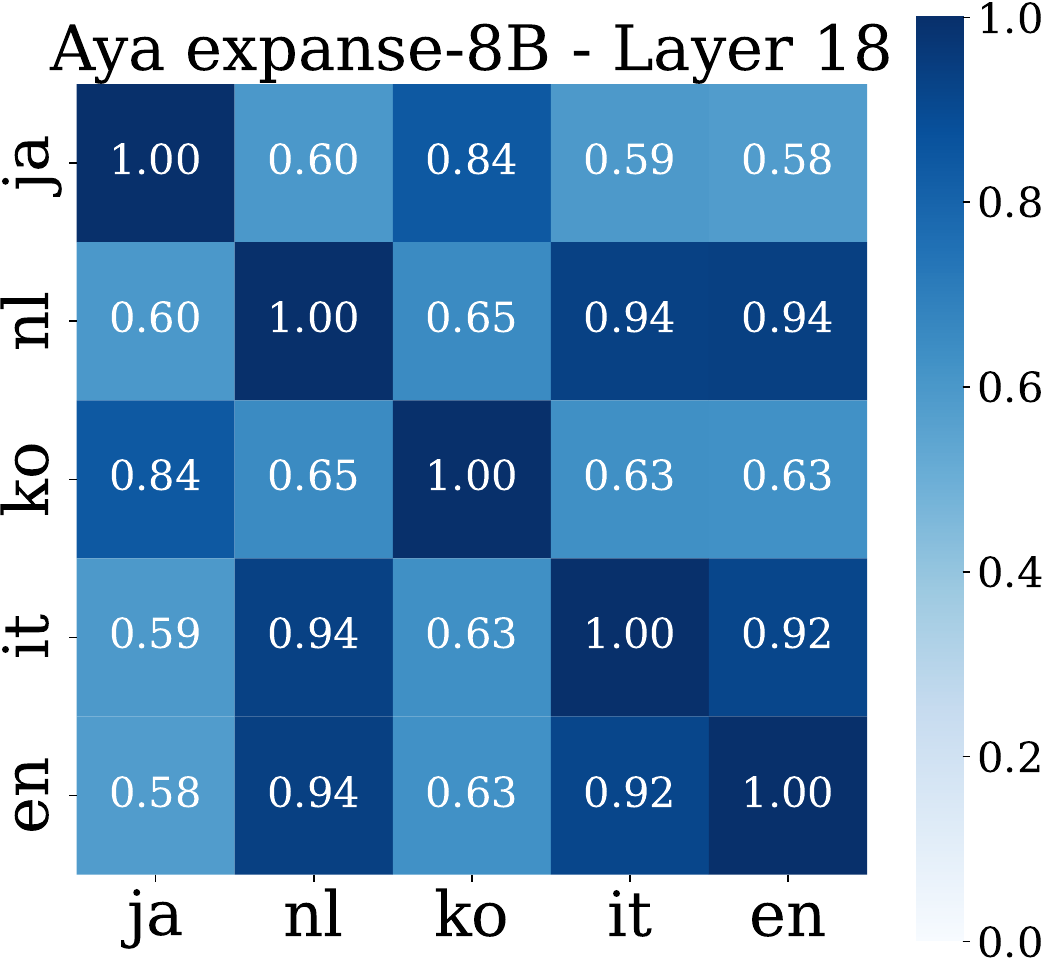}

  \begin{minipage}{0.15\linewidth}\centering \textbf{\textcolor{red}{layer 16 (Type-1)}}\end{minipage}
  \begin{minipage}{0.15\linewidth}\centering layer 16 (baseline)\end{minipage}
  \begin{minipage}{0.15\linewidth}\centering \textbf{\textcolor{red}{layer 17 (Type-1)}}\end{minipage}
  \begin{minipage}{0.15\linewidth}\centering layer 17 (baseline)\end{minipage}
  \begin{minipage}{0.15\linewidth}\centering \textbf{\textcolor{red}{layer 18 (Type-1)}}\end{minipage}
  \begin{minipage}{0.15\linewidth}\centering layer 18 (baseline)\end{minipage}

  \includegraphics[width=0.15\linewidth]{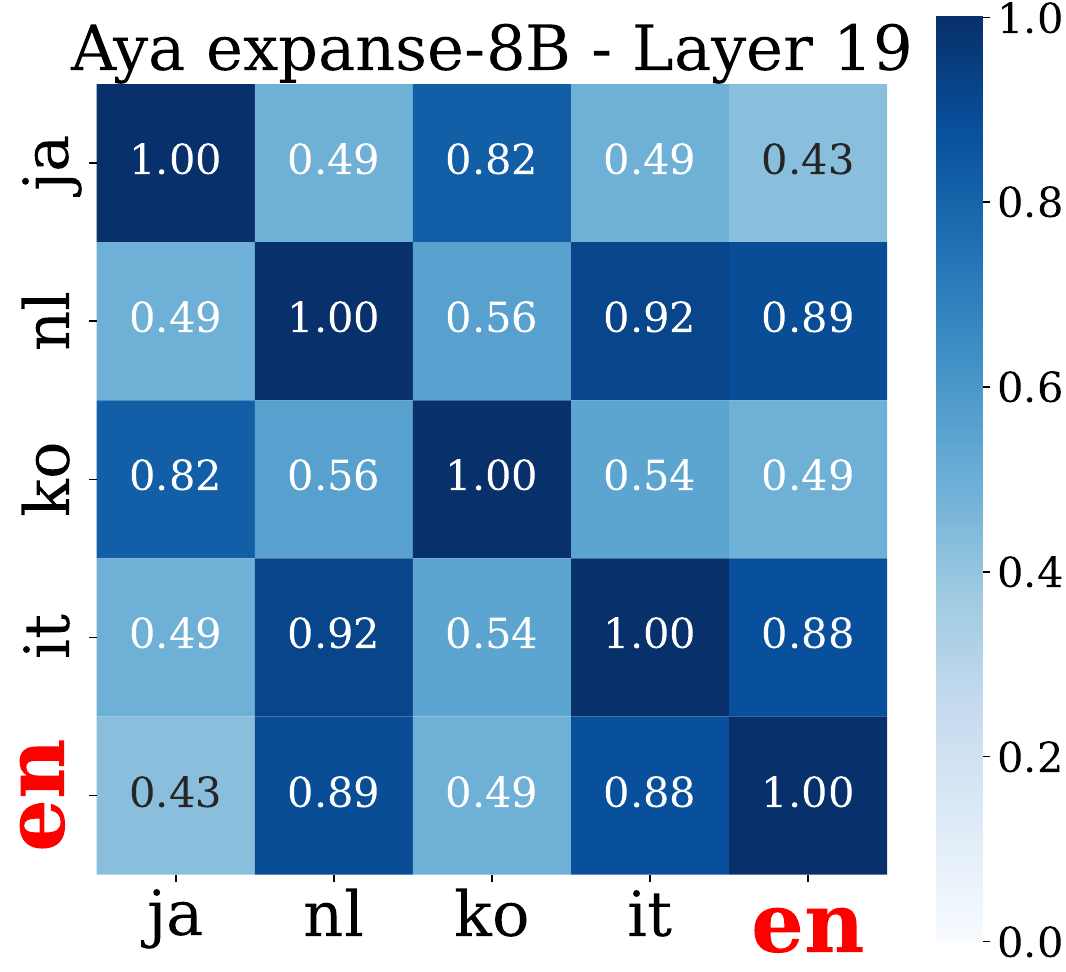}
  \includegraphics[width=0.15\linewidth]{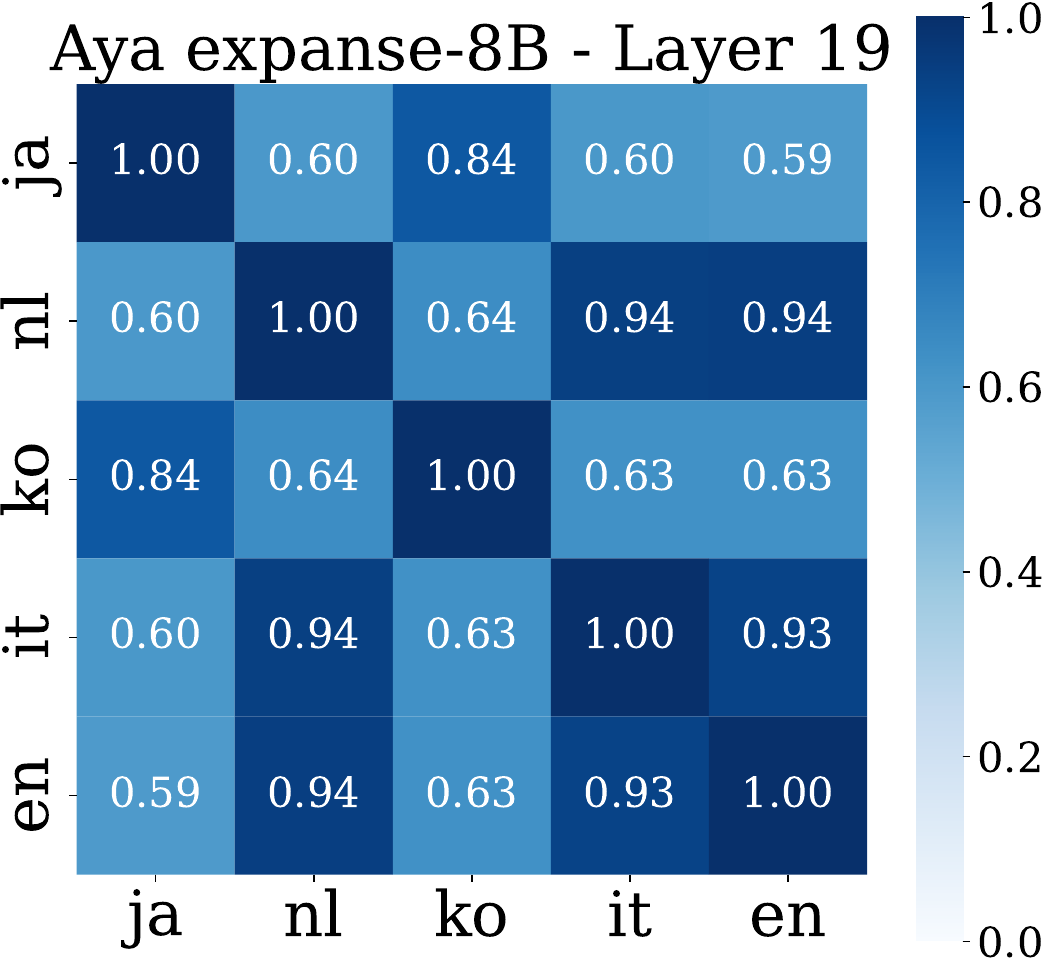}
  \includegraphics[width=0.15\linewidth]{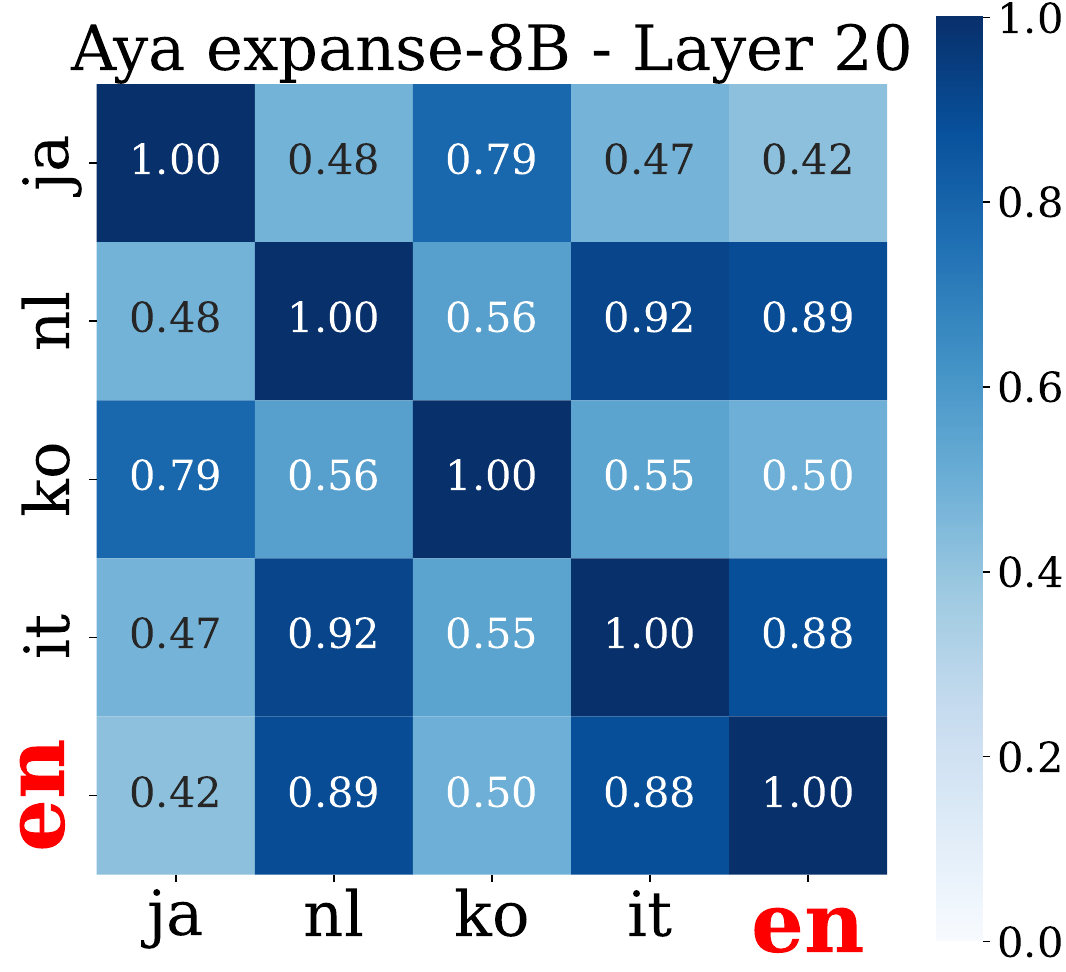}
  \includegraphics[width=0.15\linewidth]{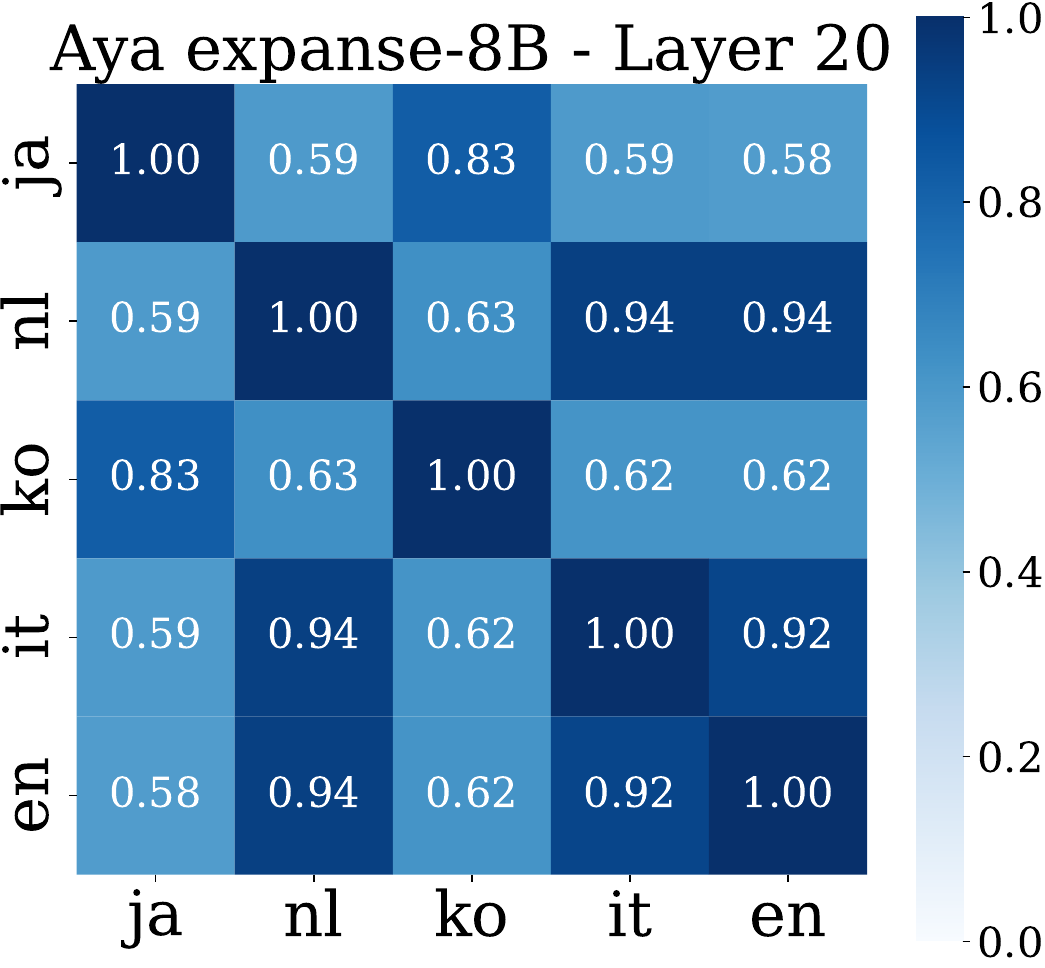}

  \begin{minipage}{0.15\linewidth}\centering \textbf{\textcolor{red}{layer 19 (Type-1)}}\end{minipage}
  \begin{minipage}{0.15\linewidth}\centering layer 19 (baseline)\end{minipage}
  \begin{minipage}{0.15\linewidth}\centering \textbf{\textcolor{red}{layer 20 (Type-1)}}\end{minipage}
  \begin{minipage}{0.15\linewidth}\centering layer 20 (baseline)\end{minipage}

  \caption{\textbf{Distance among language latent spaces while deactivating Top-1k Type-1 Transfer Neurons (Aya expanse-8B)}.}
  \label{fig:appendix:distance centroids among langage subspaces deactivating type1 aya}
\end{figure*}

Figs.~\ref{fig:appendix:distance centroids among langage subspaces deactivating type1 llama3},~\ref{fig:appendix:distance centroids among langage subspaces deactivating type1 mistral}, and~\ref{fig:appendix:distance centroids among langage subspaces deactivating type1 aya} show the distance among centroids of language latent spaces with Type-1 neurons deactivated. As shown, deactivation of Type-1 neurons hinder the movements towards English latent space significantly, compared to those of deactivation of randomly sampled neurons (baseline). This tendency is consistent across languages and models.

% centroids distance among language subspaces while deactivating type-1, llama3
% centroids distance among language subspaces while deactivating type-1, mistral
% centroids distance among language subspaces while deactivating type-1, aya

\subsection{Deactivating Type-2 Transfer Neurons}
\label{sec:sim_deactivating_Type-2_neurons_appendix}
\subsubsection{Deactivating Type-2 Transfer Neurons Significantly Inhibit Spatial Transition from the Shared Semantic Latent Space to the Language-Specific Latent Spaces}
\label{sec:appendix:Deactivating Type-2 Transfer Neurons Causes a Significant Delay in the Spatial Transition from the Shared Semantic Subspace to the Language-Specific Subspaces}
\begin{figure*}[t]
  \centering

  \includegraphics[width=0.19\linewidth]{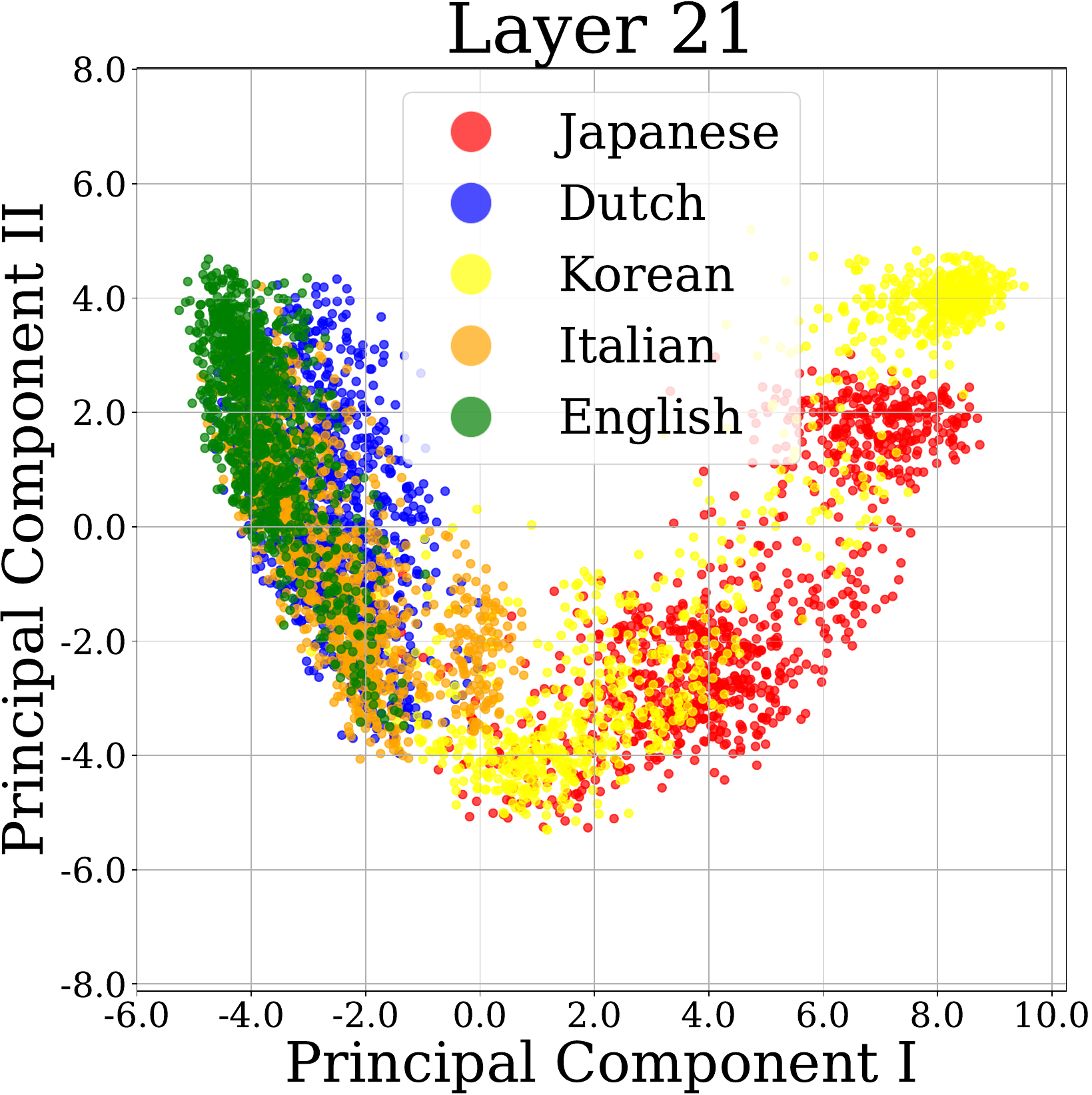}
  \includegraphics[width=0.19\linewidth]{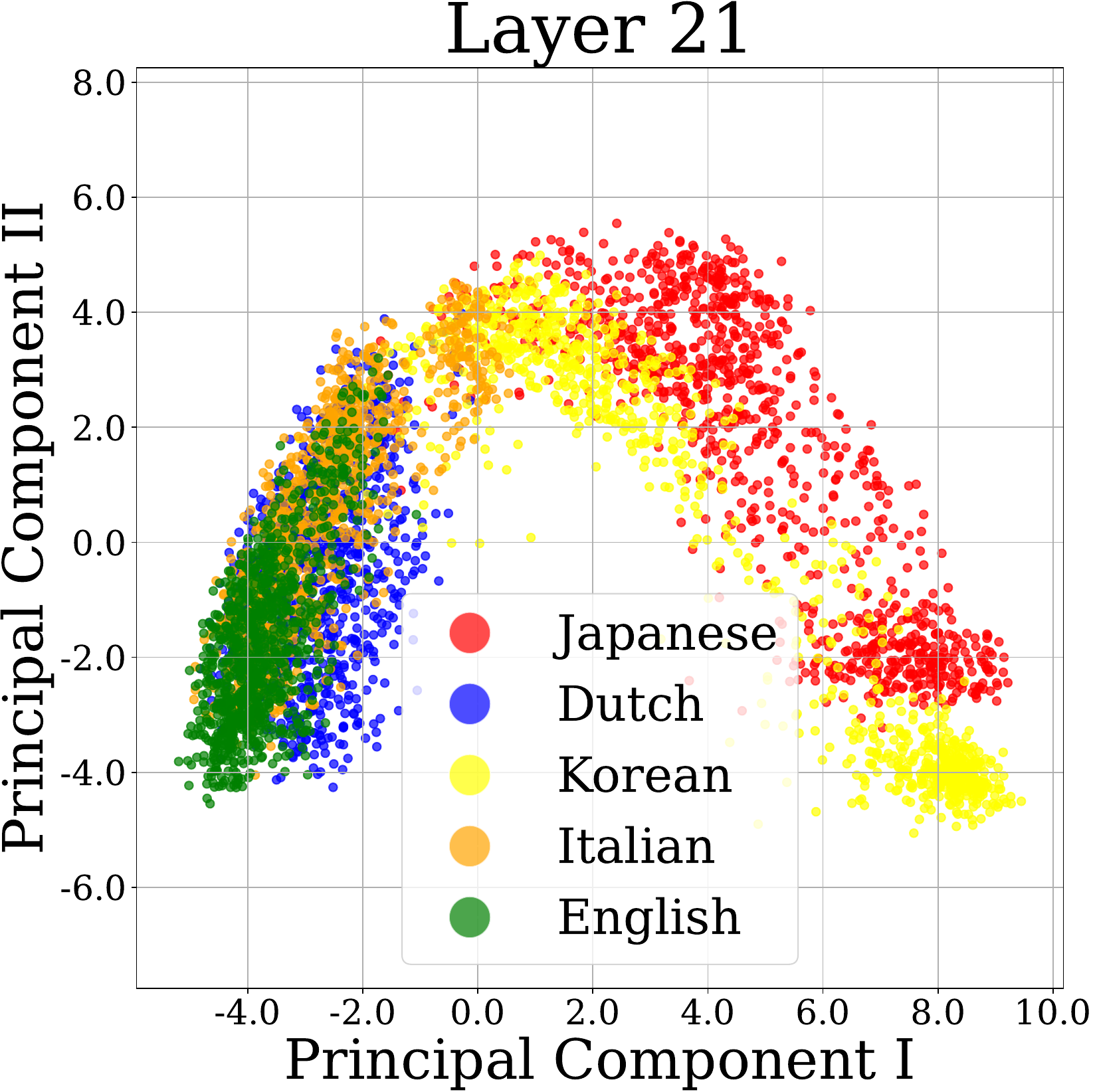}
  \includegraphics[width=0.19\linewidth]{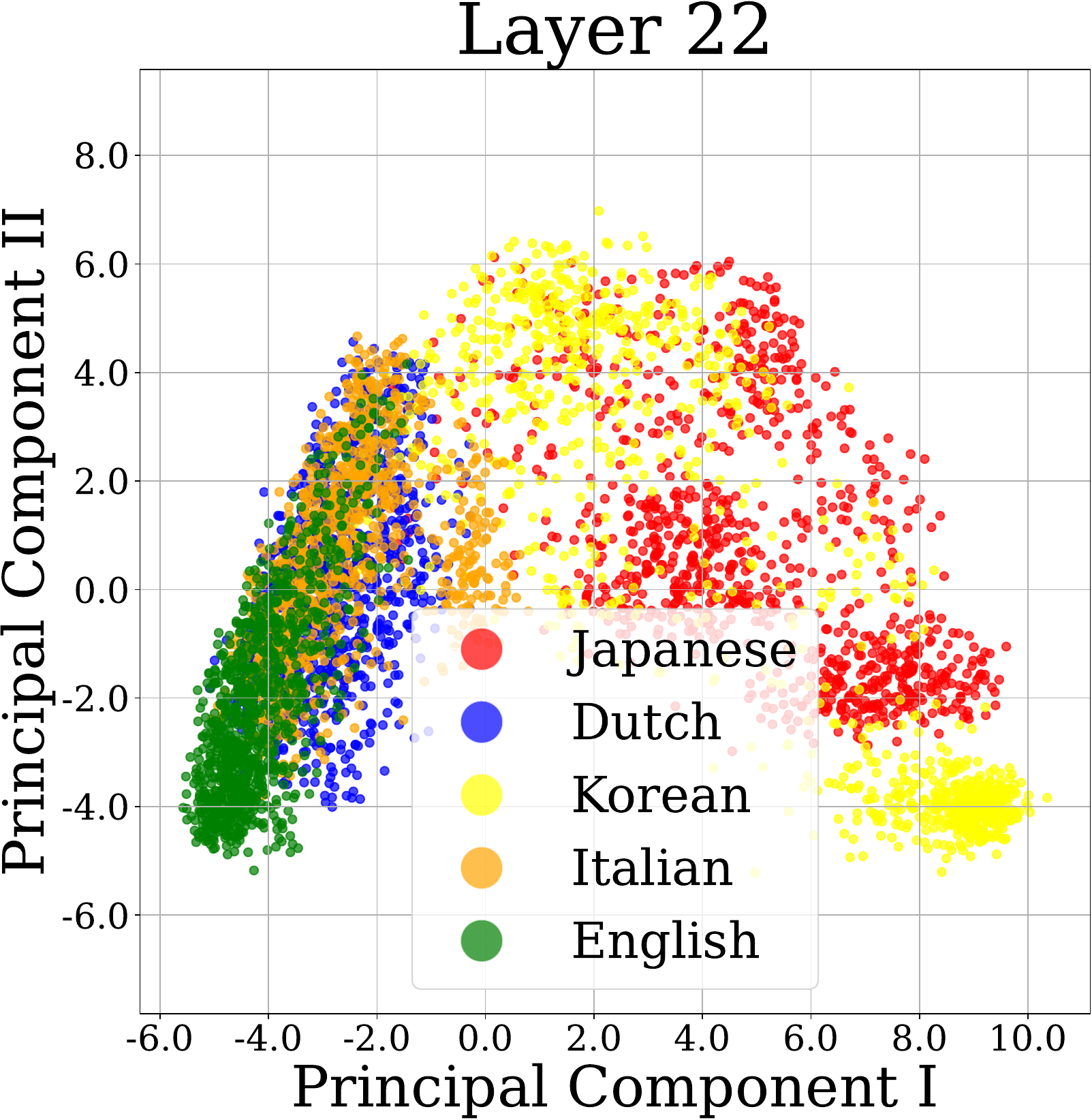}
  \includegraphics[width=0.19\linewidth]{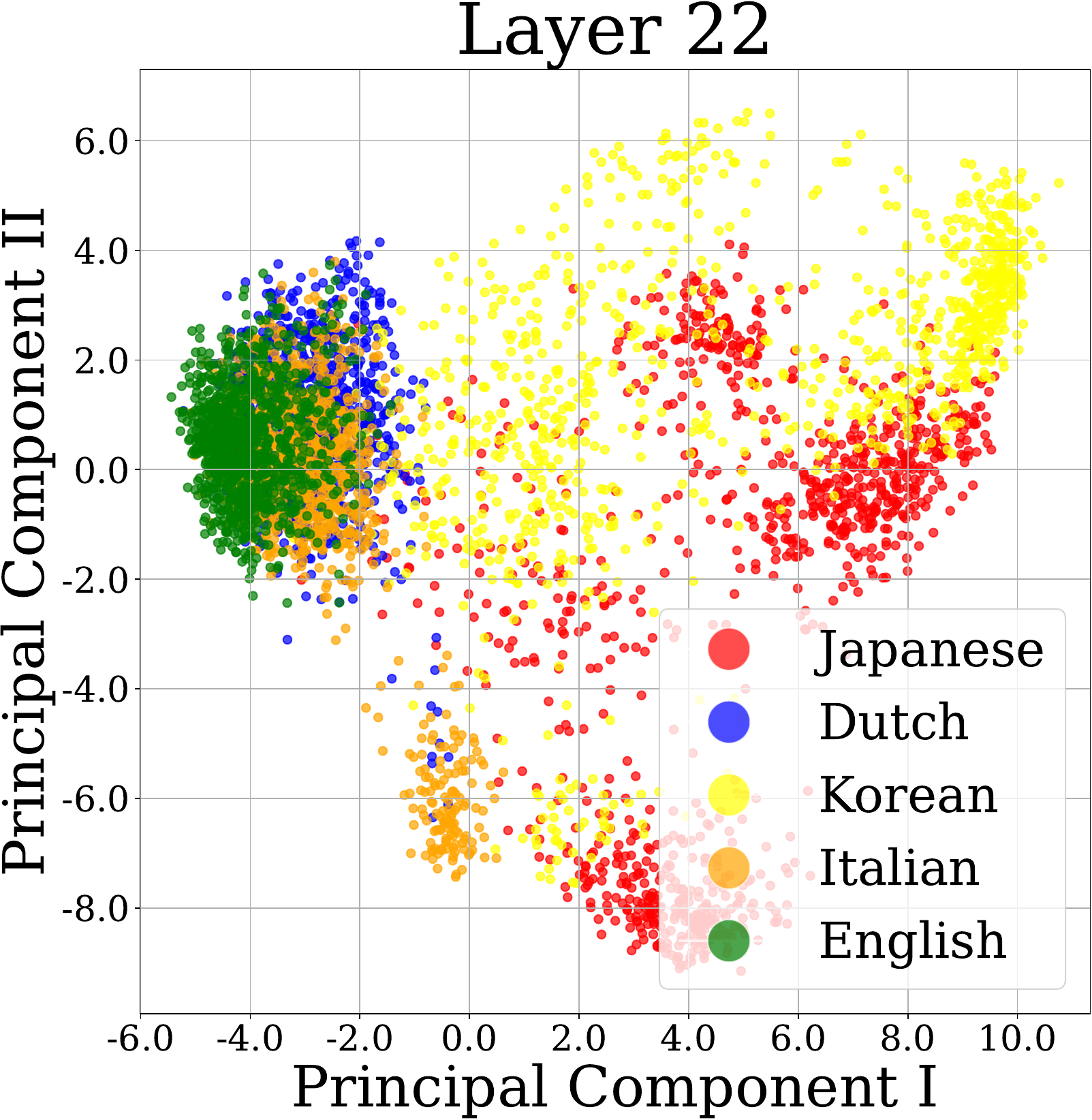}

  \begin{minipage}{0.19\linewidth}\centering \textbf{\textcolor{red}{layer 21 (Type-2)}}\end{minipage}
  \begin{minipage}{0.19\linewidth}\centering layer 21 (baseline)\end{minipage}
  \begin{minipage}{0.19\linewidth}\centering \textbf{\textcolor{red}{layer 22 (Type-2)}}\end{minipage}
  \begin{minipage}{0.19\linewidth}\centering layer 22 (baseline)\end{minipage}

  \includegraphics[width=0.19\linewidth]{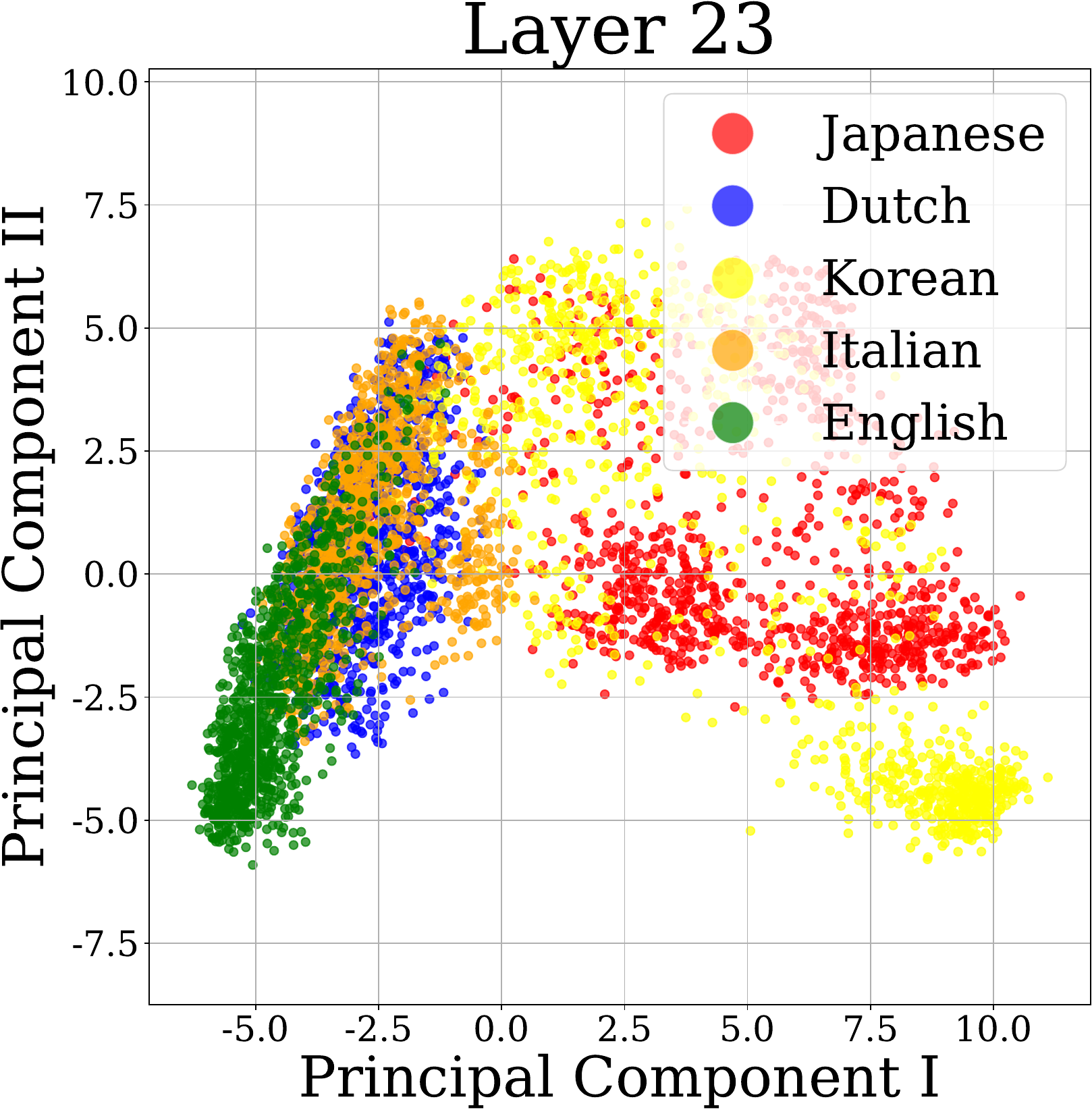}
  \includegraphics[width=0.19\linewidth]{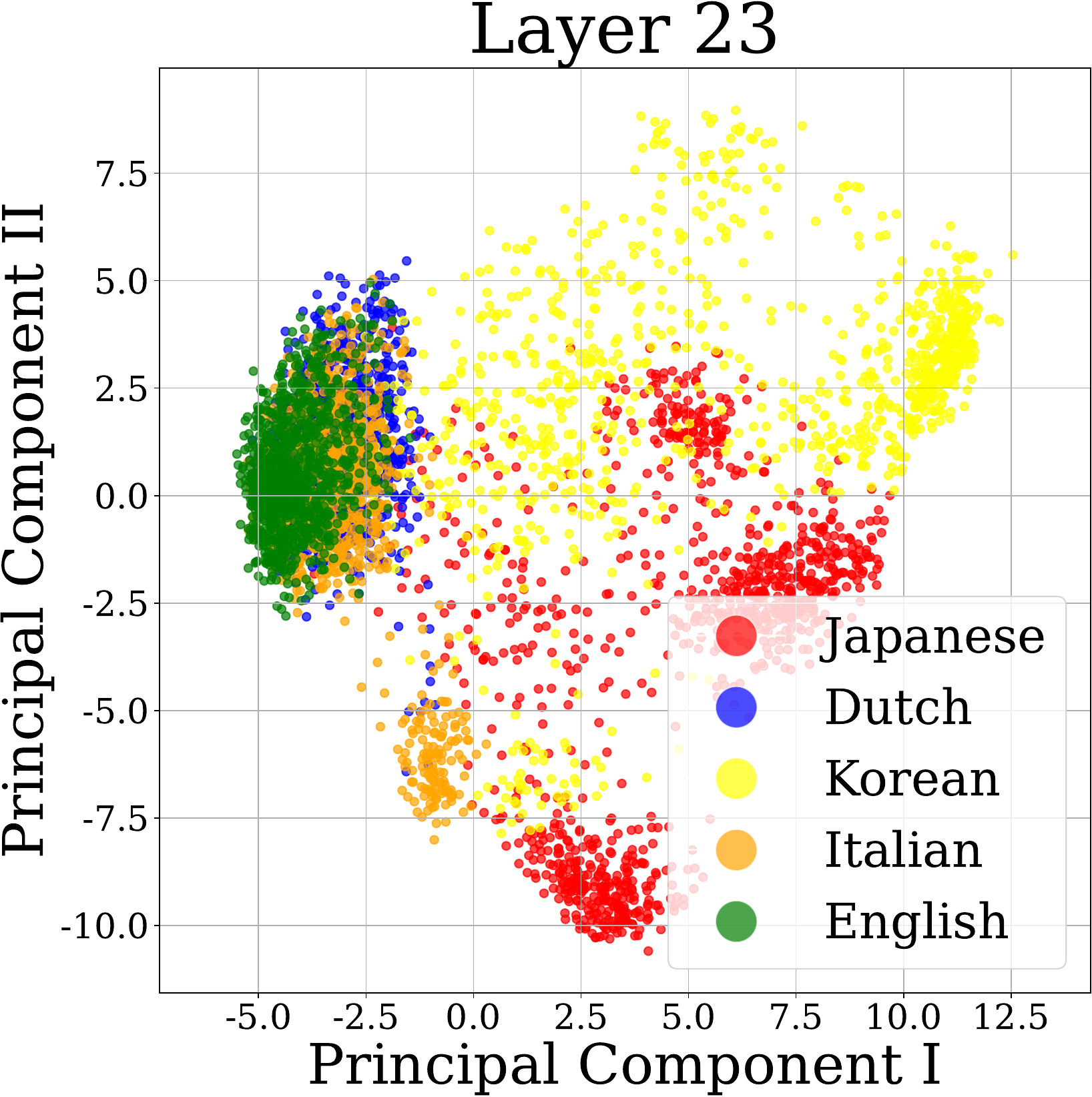}
  \includegraphics[width=0.19\linewidth]{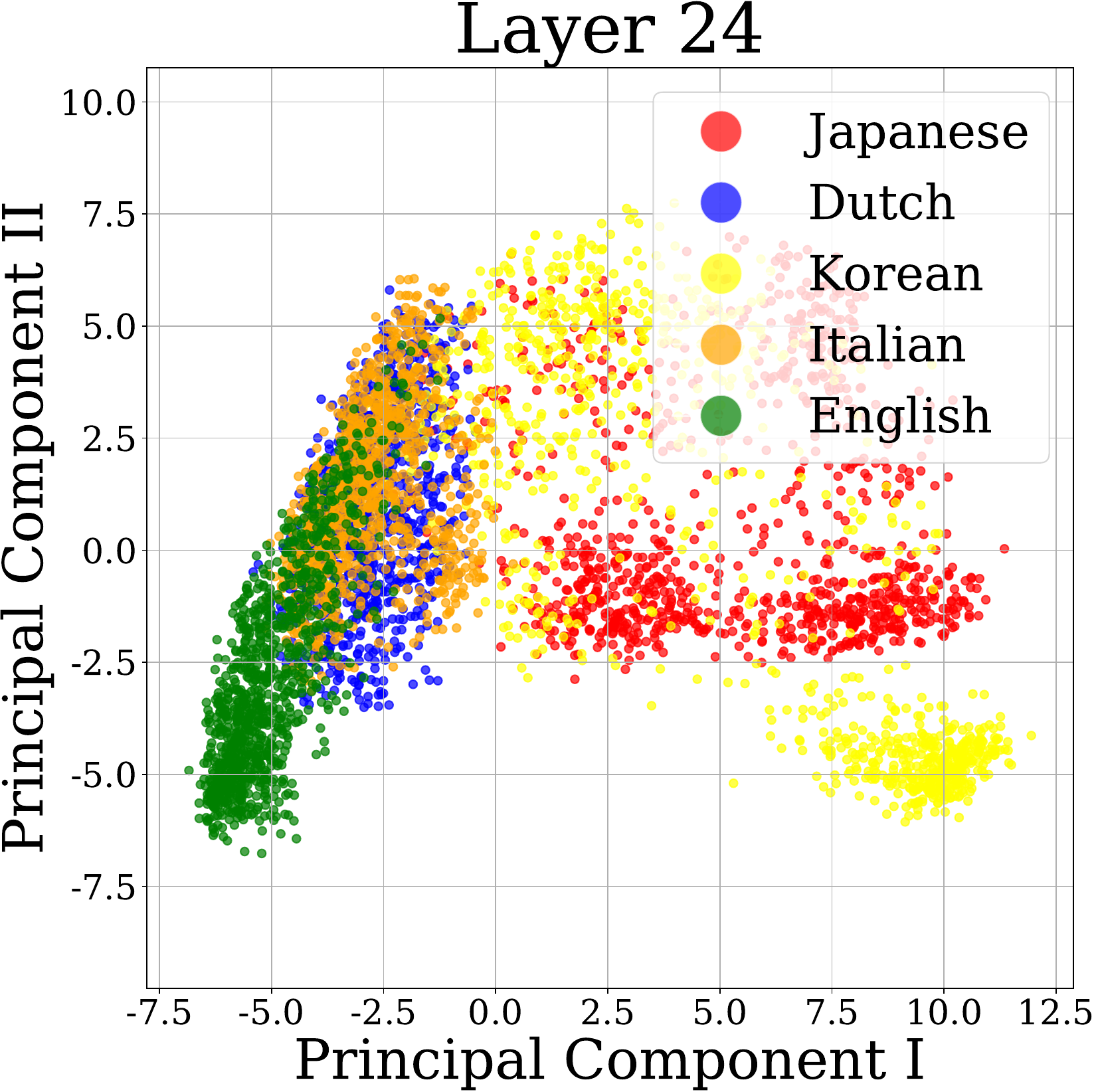}
  \includegraphics[width=0.19\linewidth]{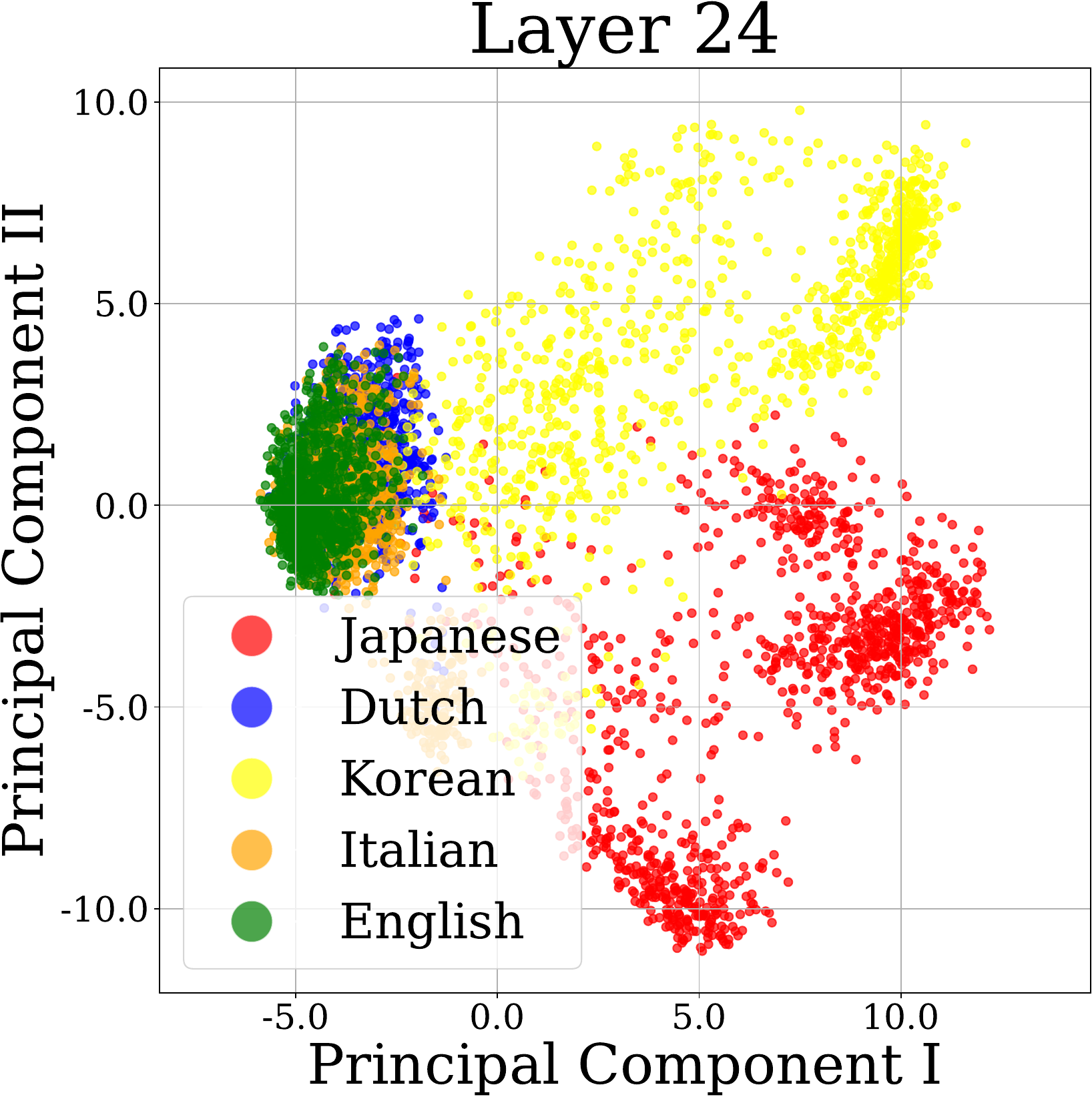}

  \begin{minipage}{0.19\linewidth}\centering \textbf{\textcolor{red}{layer 23 (Type-2)}}\end{minipage}
  \begin{minipage}{0.19\linewidth}\centering layer 23 (baseline)\end{minipage}
  \begin{minipage}{0.19\linewidth}\centering \textbf{\textcolor{red}{layer 24 (Type-2)}}\end{minipage}
  \begin{minipage}{0.19\linewidth}\centering layer 24 (baseline)\end{minipage}

  \includegraphics[width=0.19\linewidth]{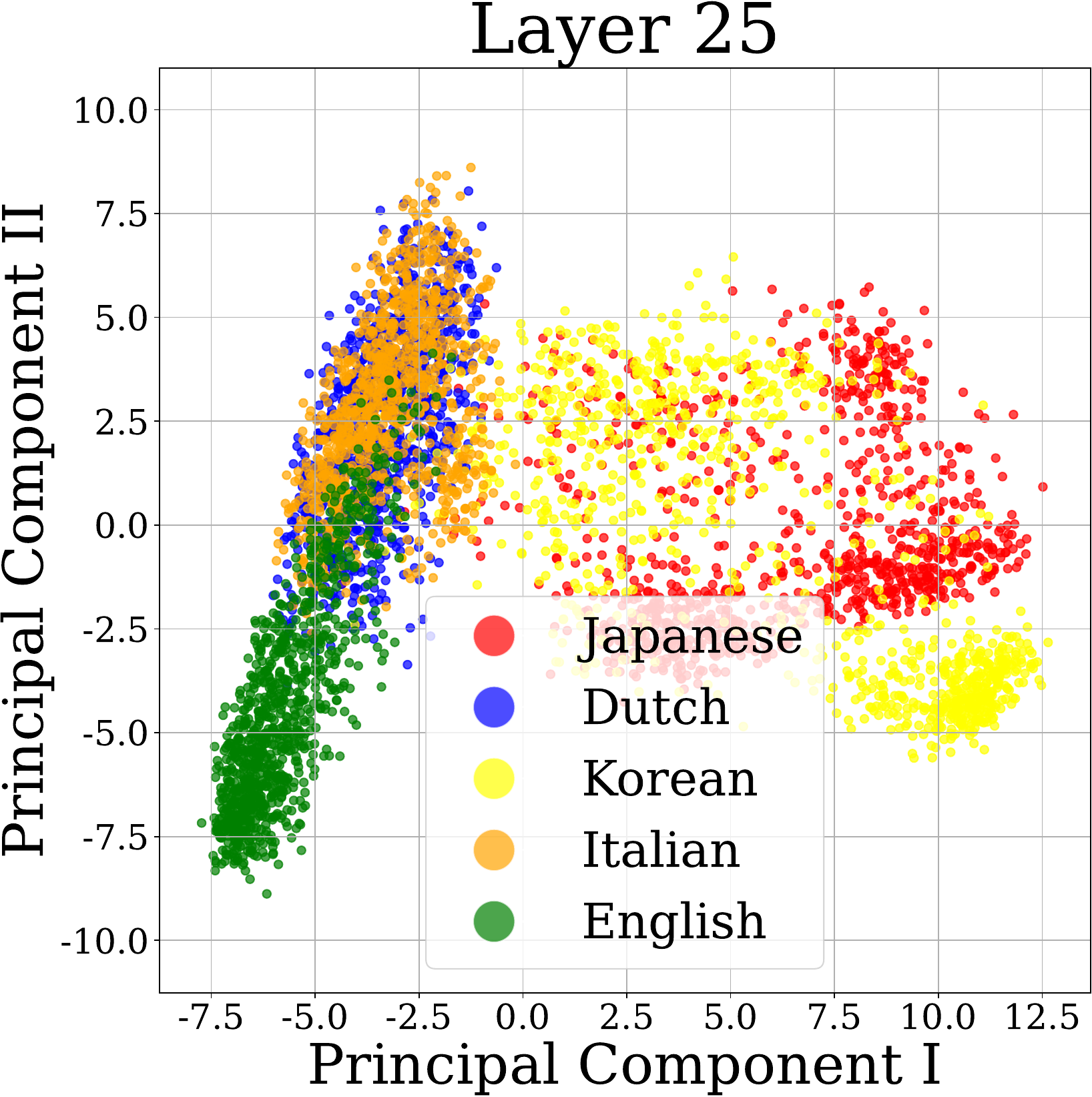}
  \includegraphics[width=0.19\linewidth]{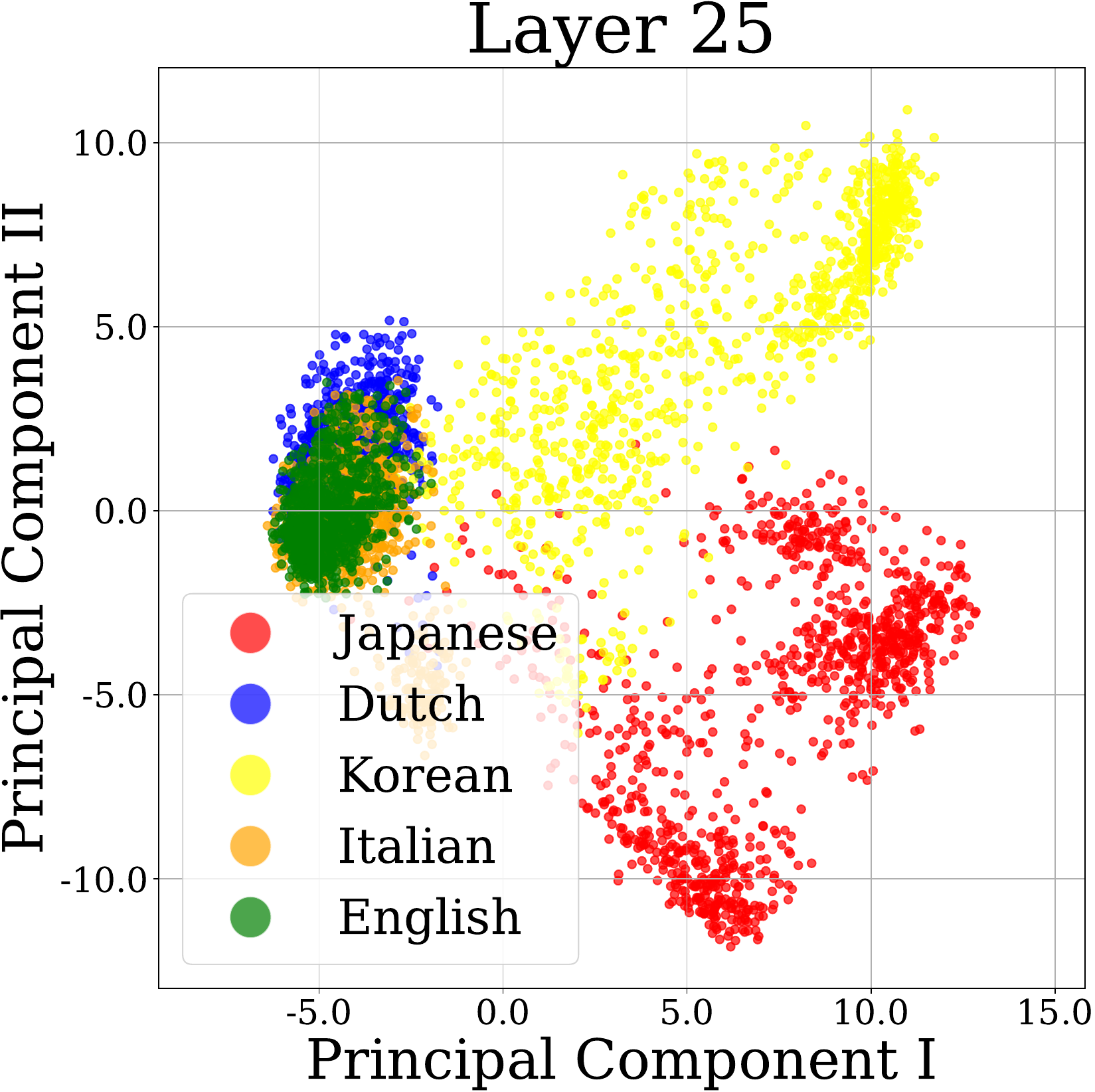}
  \includegraphics[width=0.19\linewidth]{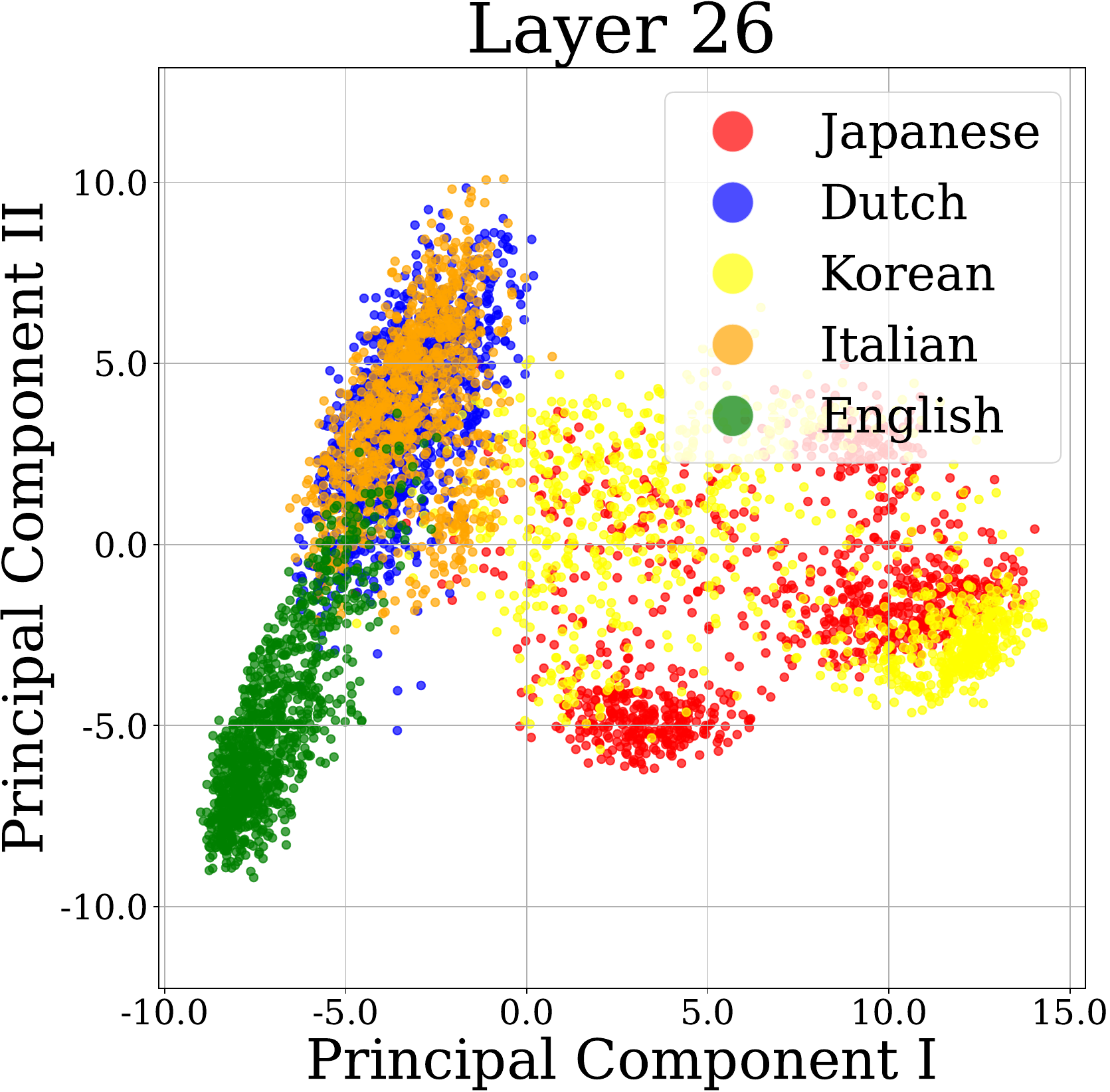}
  \includegraphics[width=0.19\linewidth]{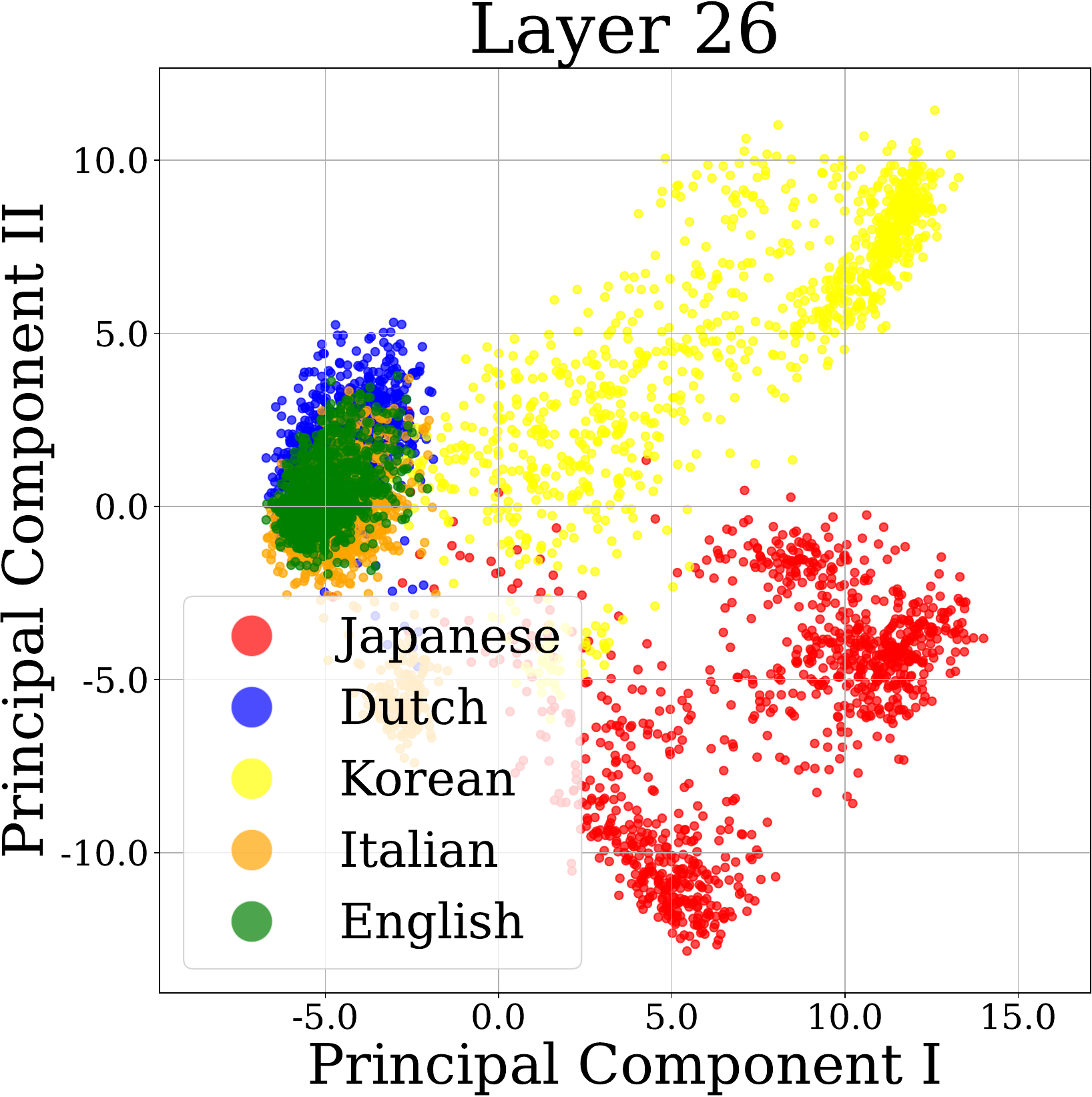}

  \begin{minipage}{0.19\linewidth}\centering \textbf{\textcolor{red}{layer 25 (Type-2)}}\end{minipage}
  \begin{minipage}{0.19\linewidth}\centering layer 25 (baseline)\end{minipage}
  \begin{minipage}{0.19\linewidth}\centering \textbf{\textcolor{red}{layer 26 (Type-2)}}\end{minipage}
  \begin{minipage}{0.19\linewidth}\centering layer 26 (baseline)\end{minipage}

  \includegraphics[width=0.19\linewidth]{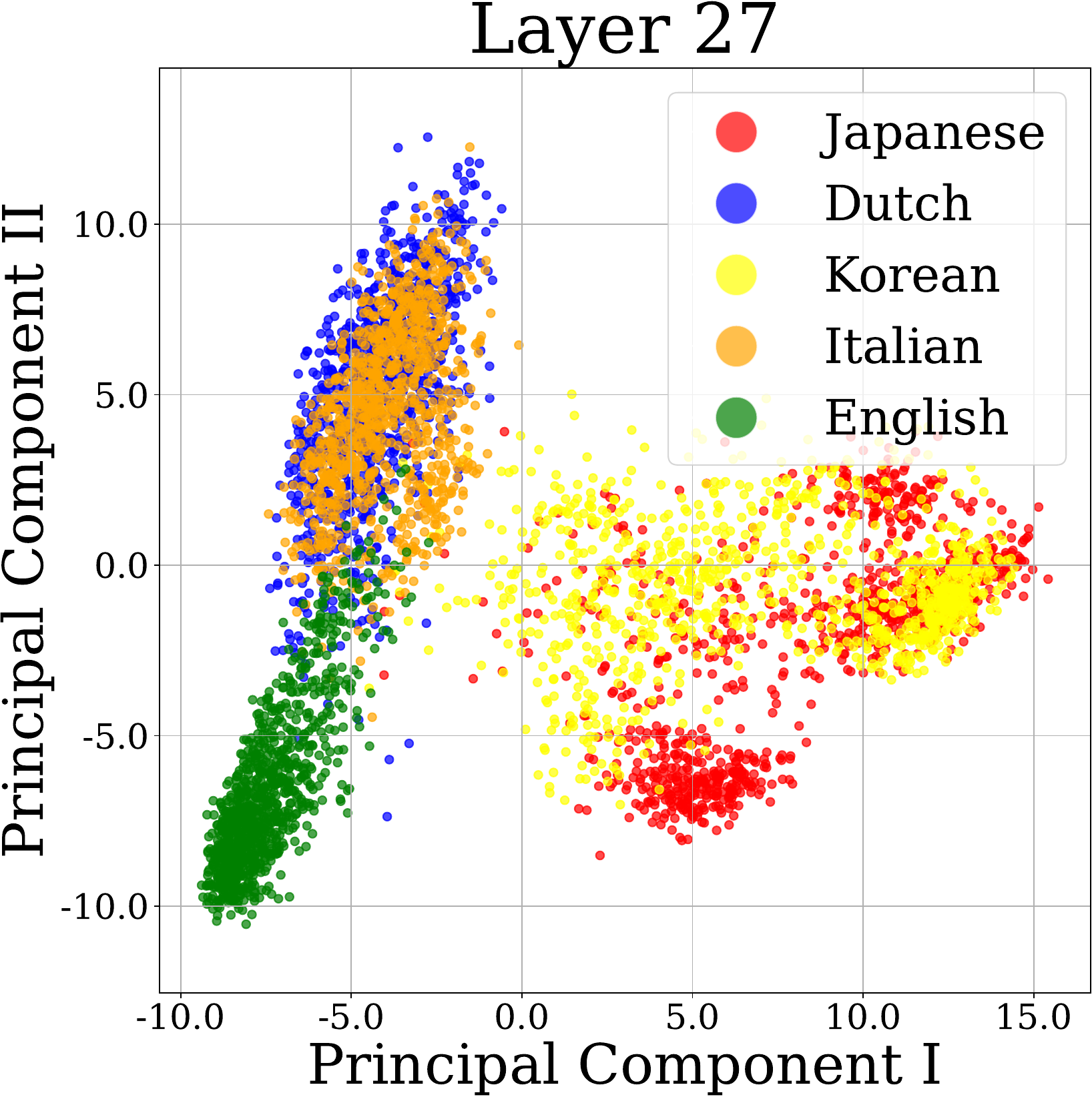}
  \includegraphics[width=0.19\linewidth]{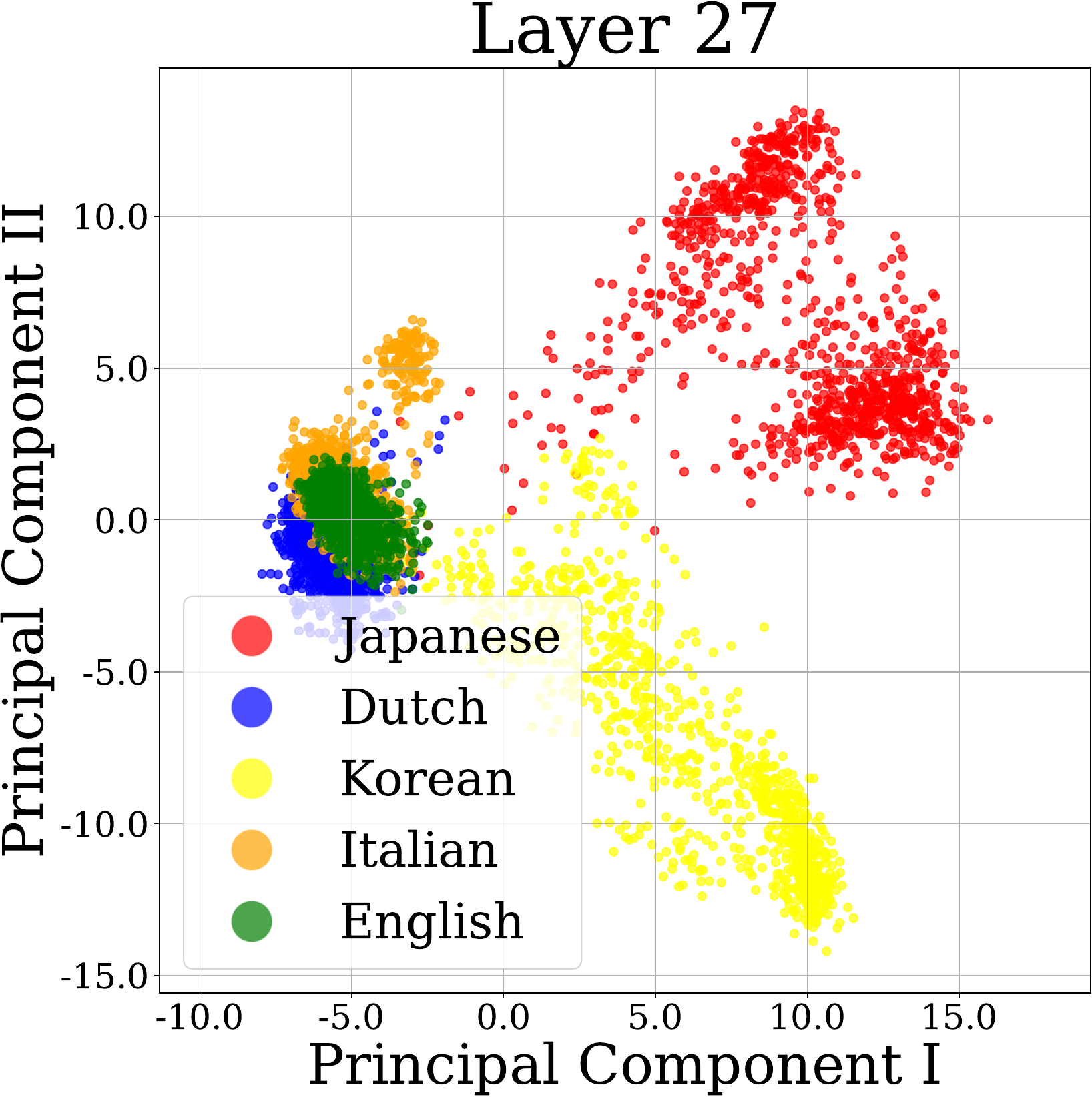}
  \includegraphics[width=0.19\linewidth]{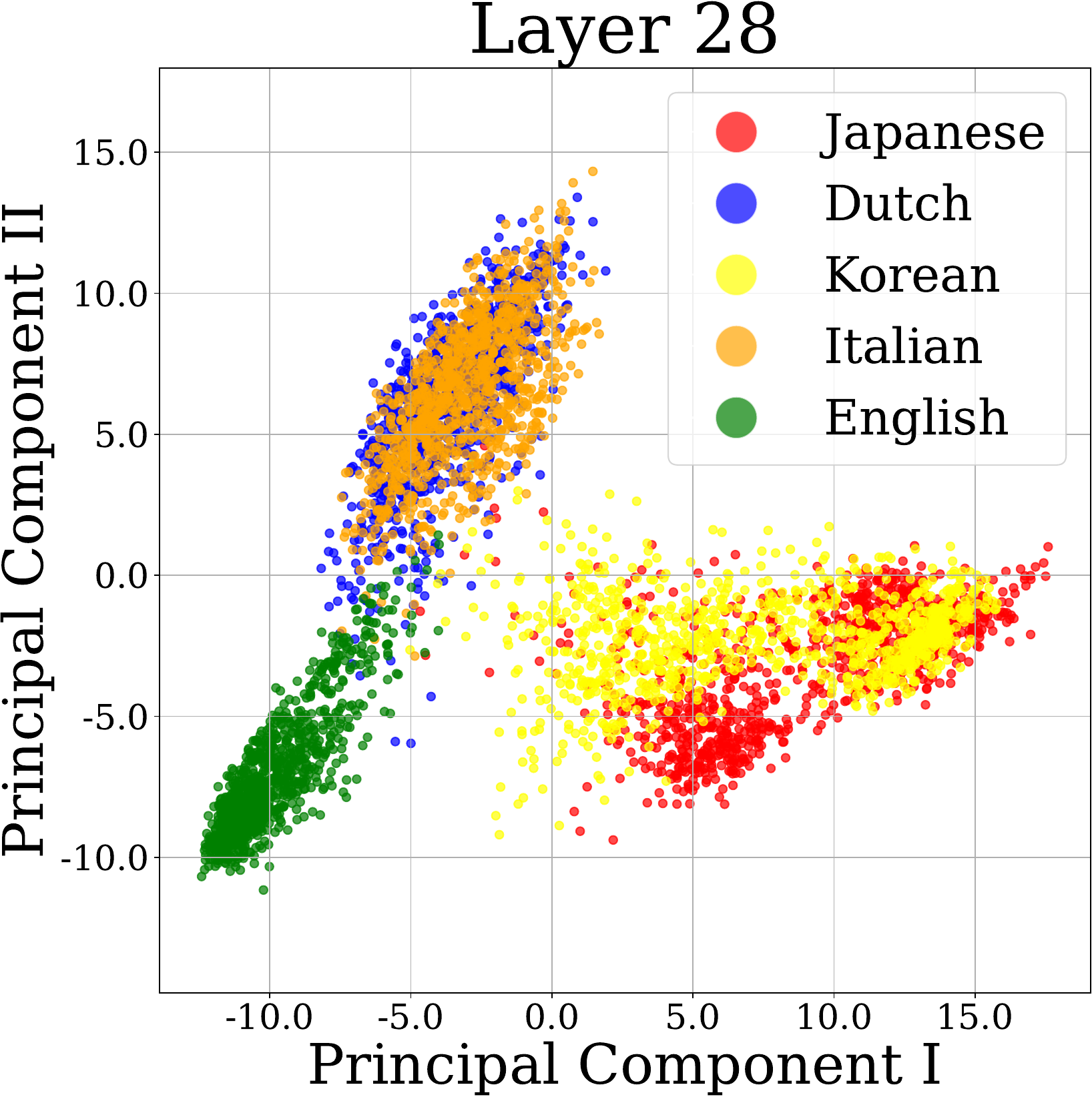}
  \includegraphics[width=0.19\linewidth]{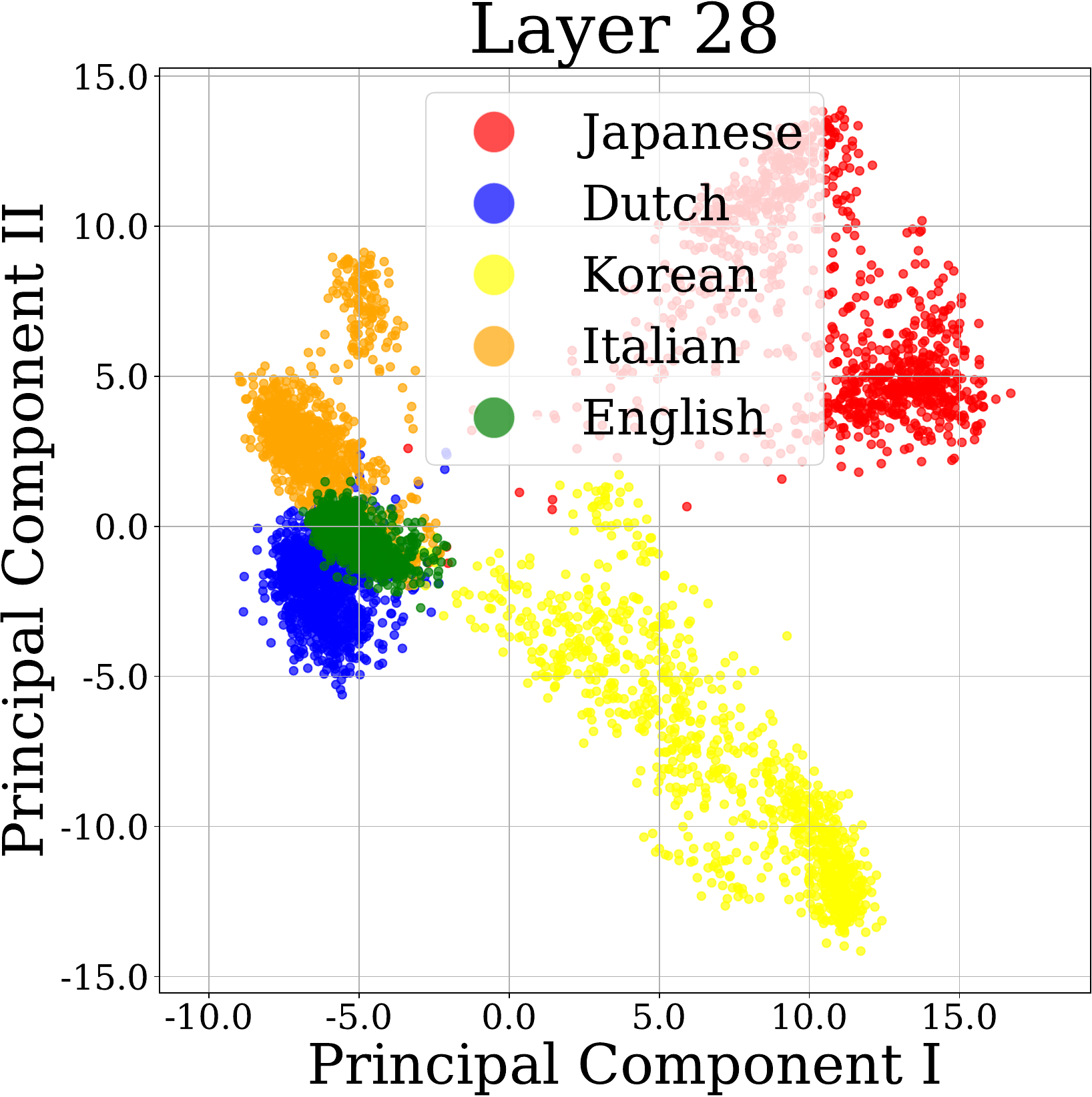}

  \begin{minipage}{0.19\linewidth}\centering \textbf{\textcolor{red}{layer 27 (Type-2)}}\end{minipage}
  \begin{minipage}{0.19\linewidth}\centering layer 27 (baseline)\end{minipage}
  \begin{minipage}{0.19\linewidth}\centering \textbf{\textcolor{red}{layer 28 (Type-2)}}\end{minipage}
  \begin{minipage}{0.19\linewidth}\centering layer 28 (baseline)\end{minipage}

  \includegraphics[width=0.19\linewidth]{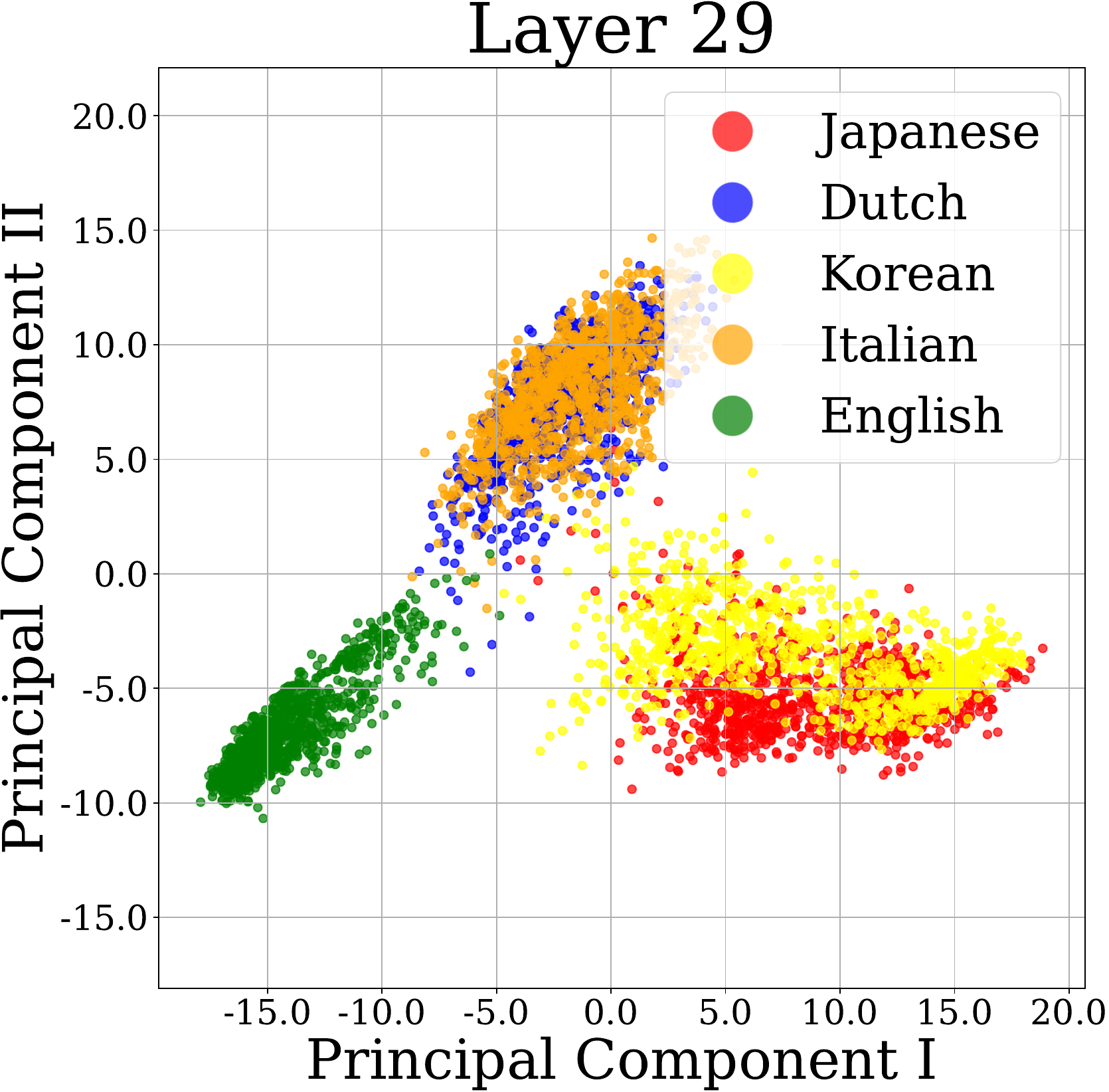}
  \includegraphics[width=0.19\linewidth]{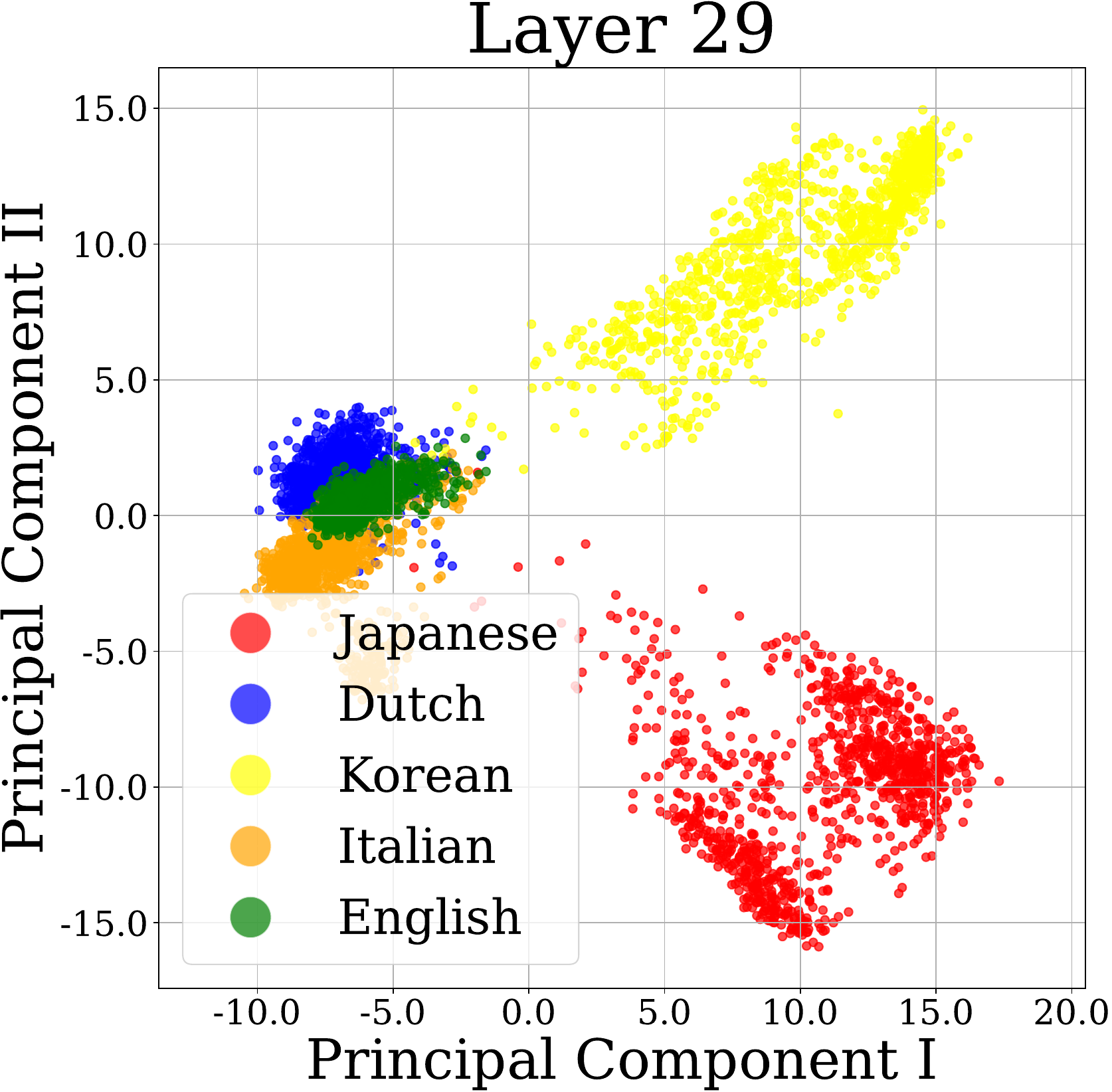}
  \includegraphics[width=0.19\linewidth]{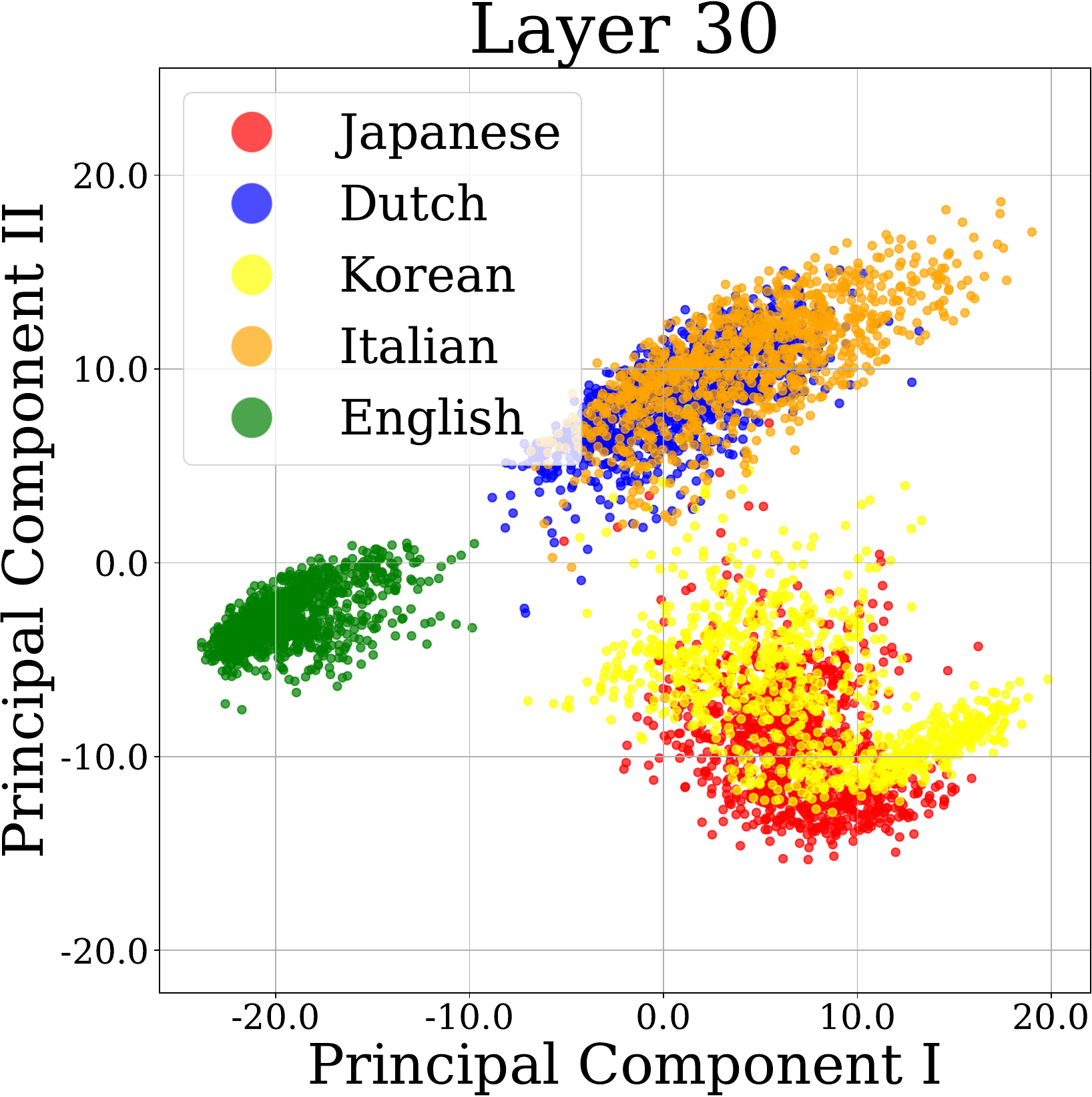}
  \includegraphics[width=0.19\linewidth]{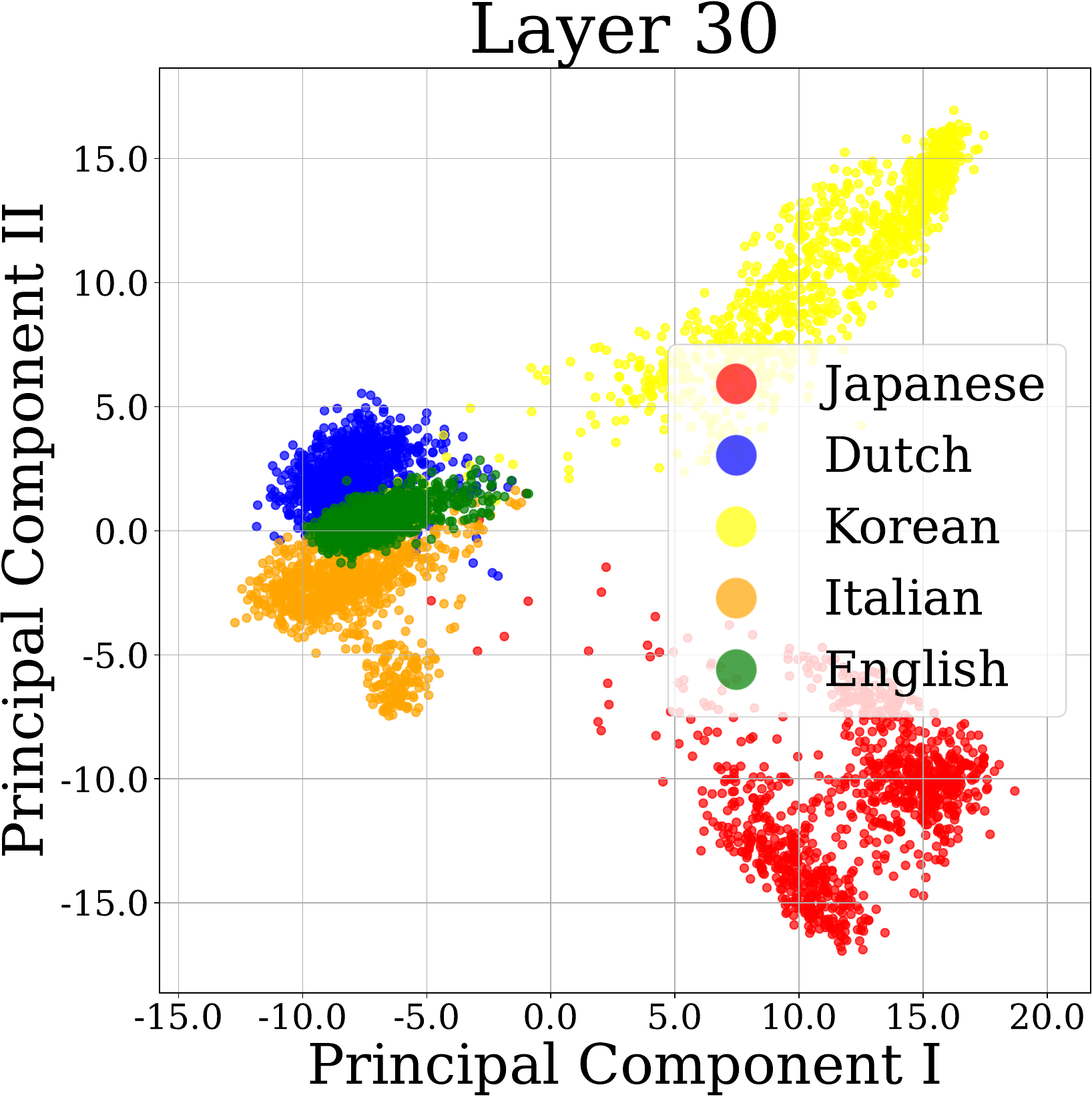}

  \begin{minipage}{0.19\linewidth}\centering \textbf{\textcolor{red}{layer 29 (Type-2)}}\end{minipage}
  \begin{minipage}{0.19\linewidth}\centering layer 29 (baseline)\end{minipage}
  \begin{minipage}{0.19\linewidth}\centering \textbf{\textcolor{red}{layer 30 (Type-2)}}\end{minipage}
  \begin{minipage}{0.19\linewidth}\centering layer 30 (baseline)\end{minipage}

  \includegraphics[width=0.19\linewidth]{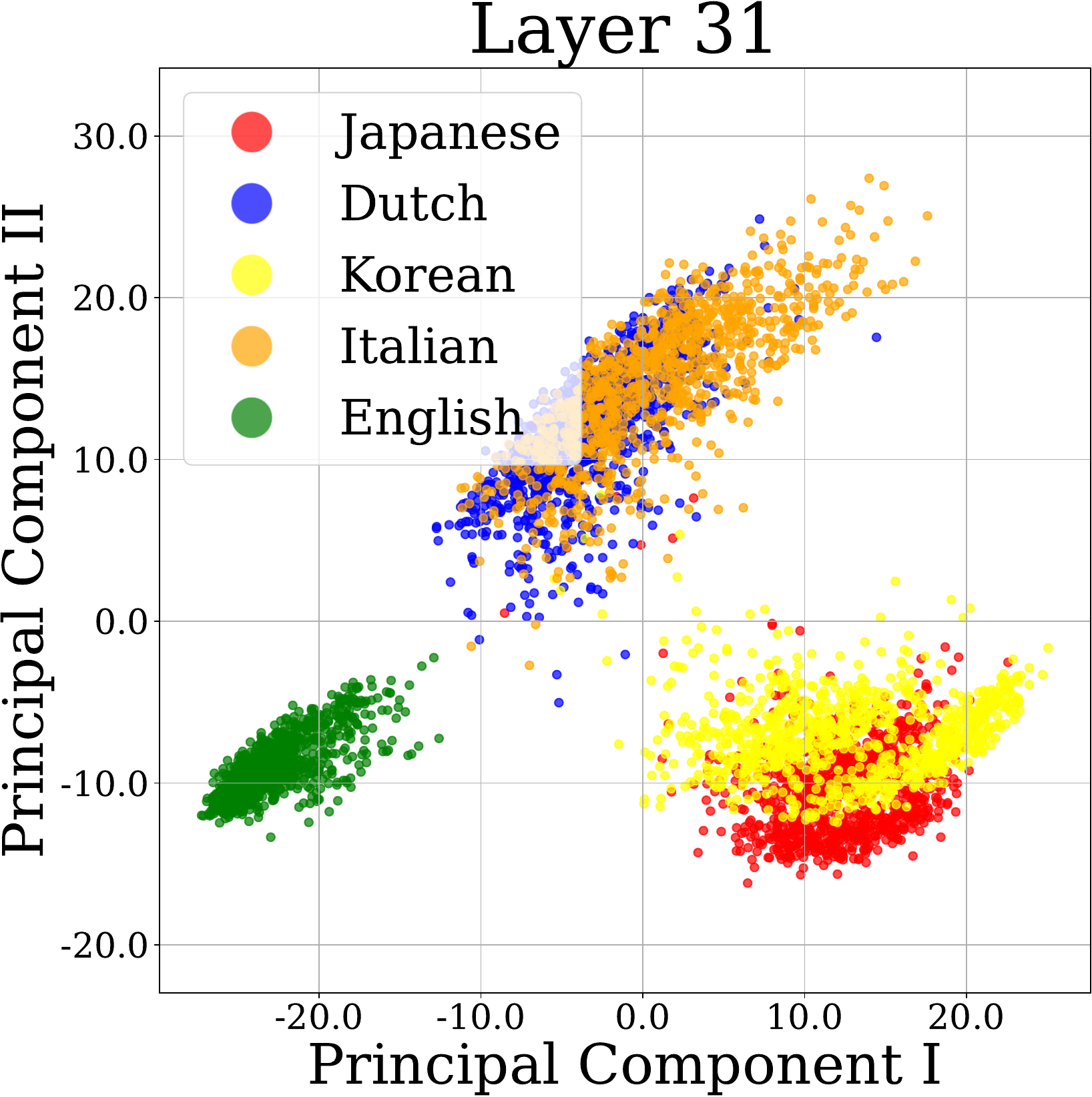}
  \includegraphics[width=0.19\linewidth]{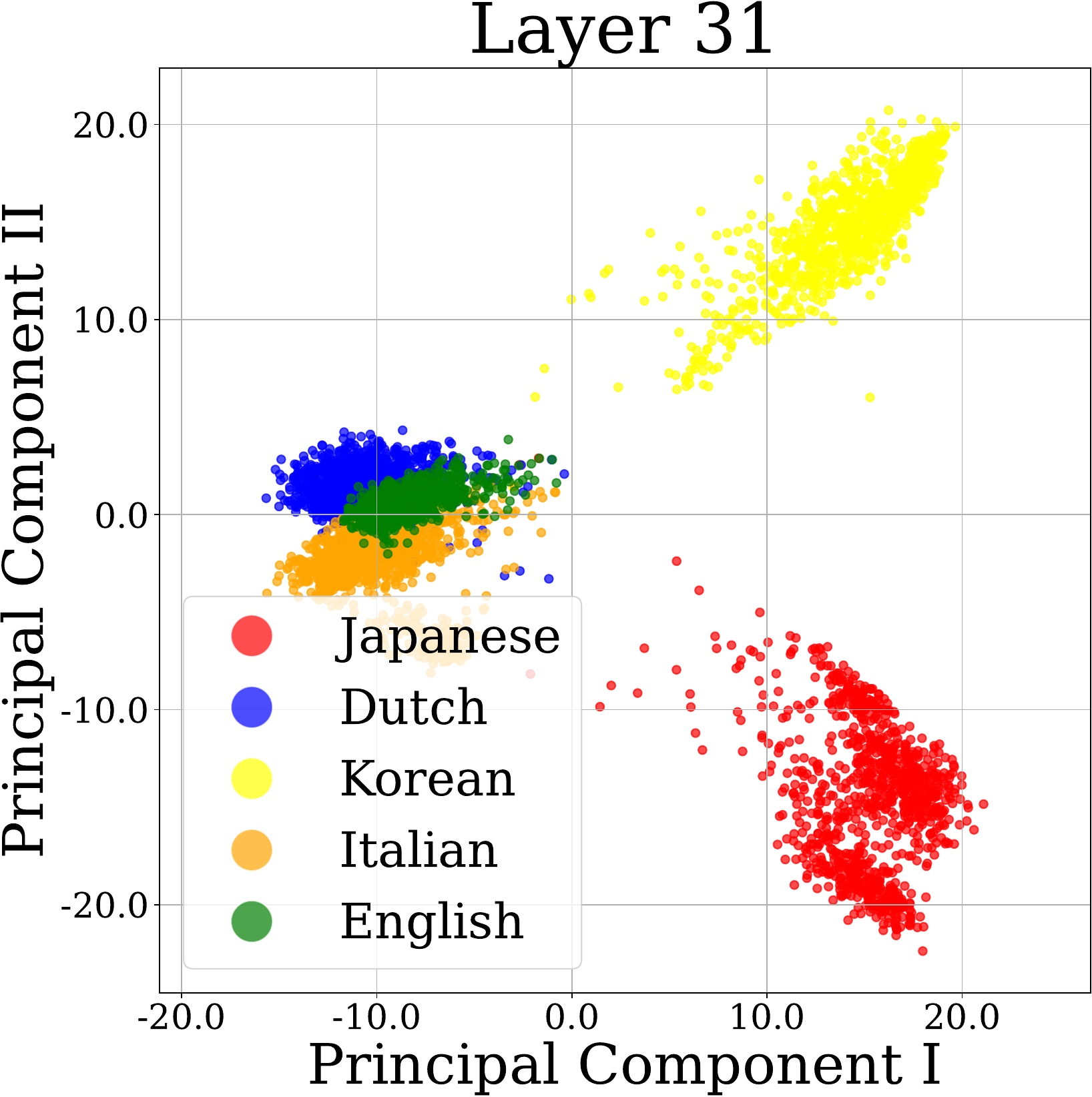}
  \includegraphics[width=0.19\linewidth]{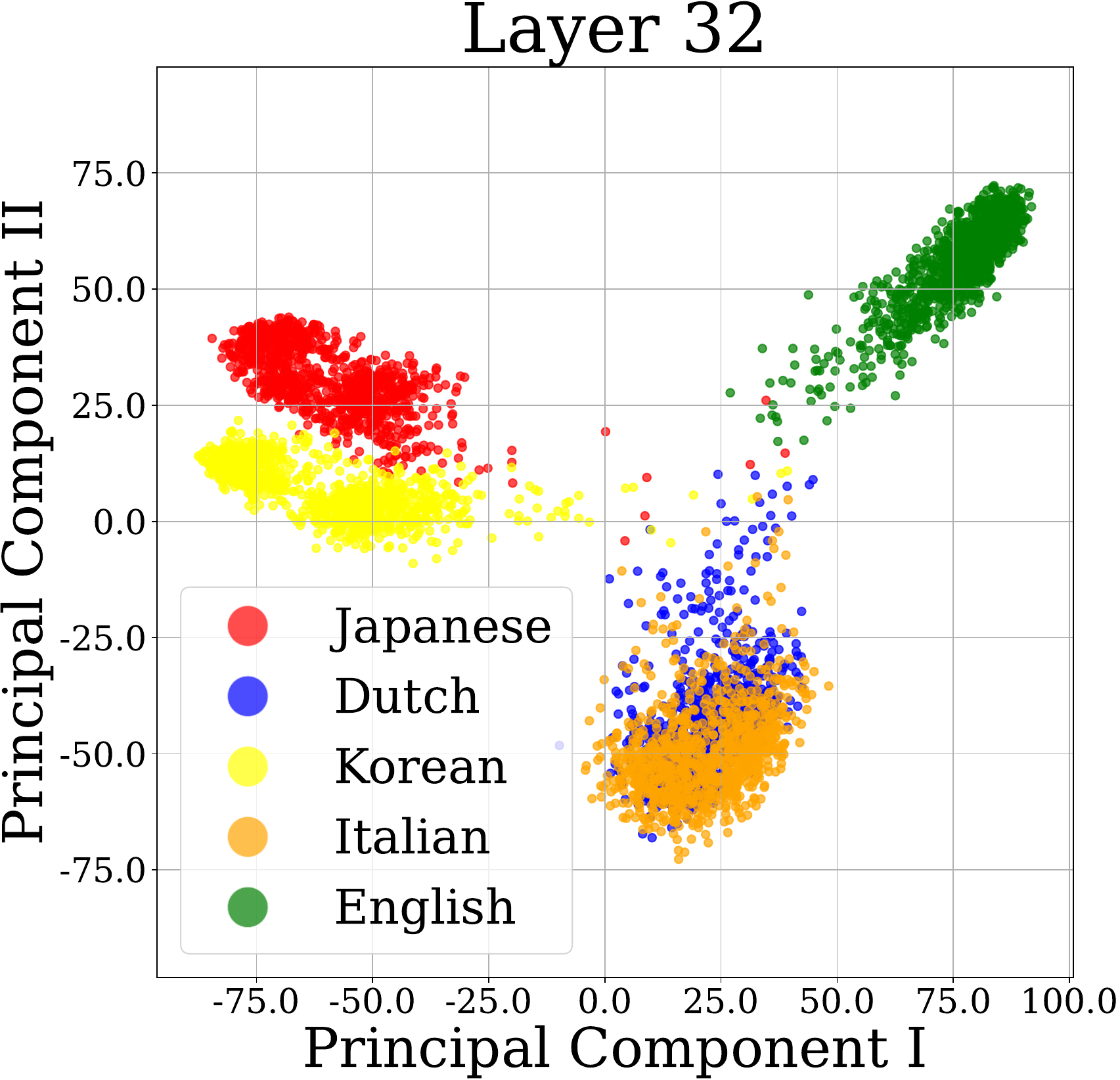}
  \includegraphics[width=0.19\linewidth]{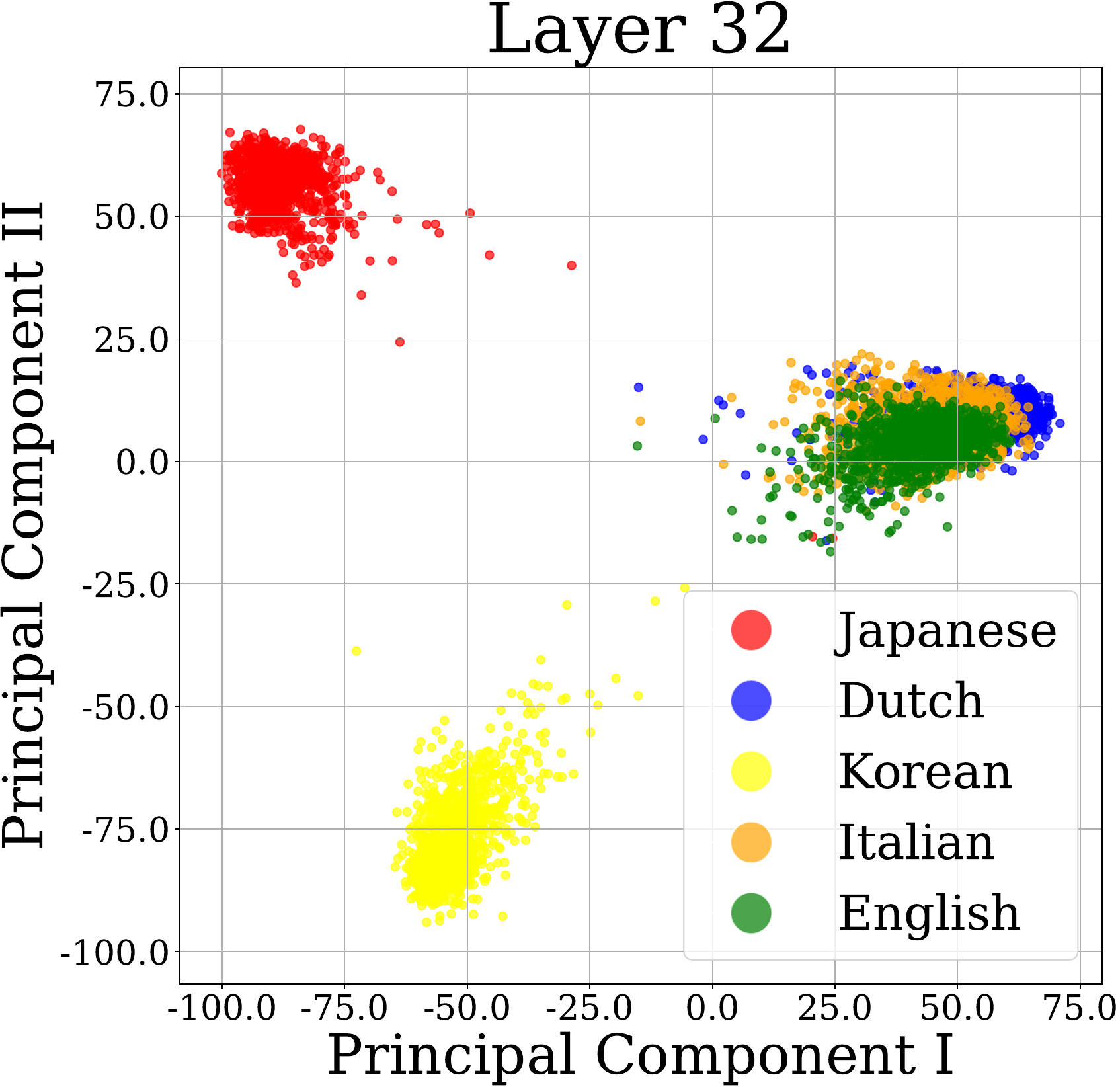}

  \begin{minipage}{0.19\linewidth}\centering \textbf{\textcolor{red}{layer 31 (Type-2)}}\end{minipage}
  \begin{minipage}{0.19\linewidth}\centering layer 31 (baseline)\end{minipage}
  \begin{minipage}{0.19\linewidth}\centering \textbf{\textcolor{red}{layer 32 (Type-2)}}\end{minipage}
  \begin{minipage}{0.19\linewidth}\centering layer 32 (baseline)\end{minipage}

  \caption{\textbf{The resutls of PCA while deactivating Top-1k Type-2 Transfer Neurons (LLaMA3-8B)}.\\ \textbf{\textcolor{red}{Layer (Type-2)}} indicates the result of deactivating the top-1k Type-2 Transfer Neurons, whereas “baseline” refers to the result of deactivating 1k randomly sampled neurons from the same layers as the Type-2 Transfer Neurons. The \textcolor[HTML]{228B22}{English} features are the only ones visualized without intervention.}
  \label{fig:appendix:pca_deactivating_type2_llama3}
\end{figure*}

\begin{figure*}[t]
  \centering

  \includegraphics[width=0.19\linewidth]{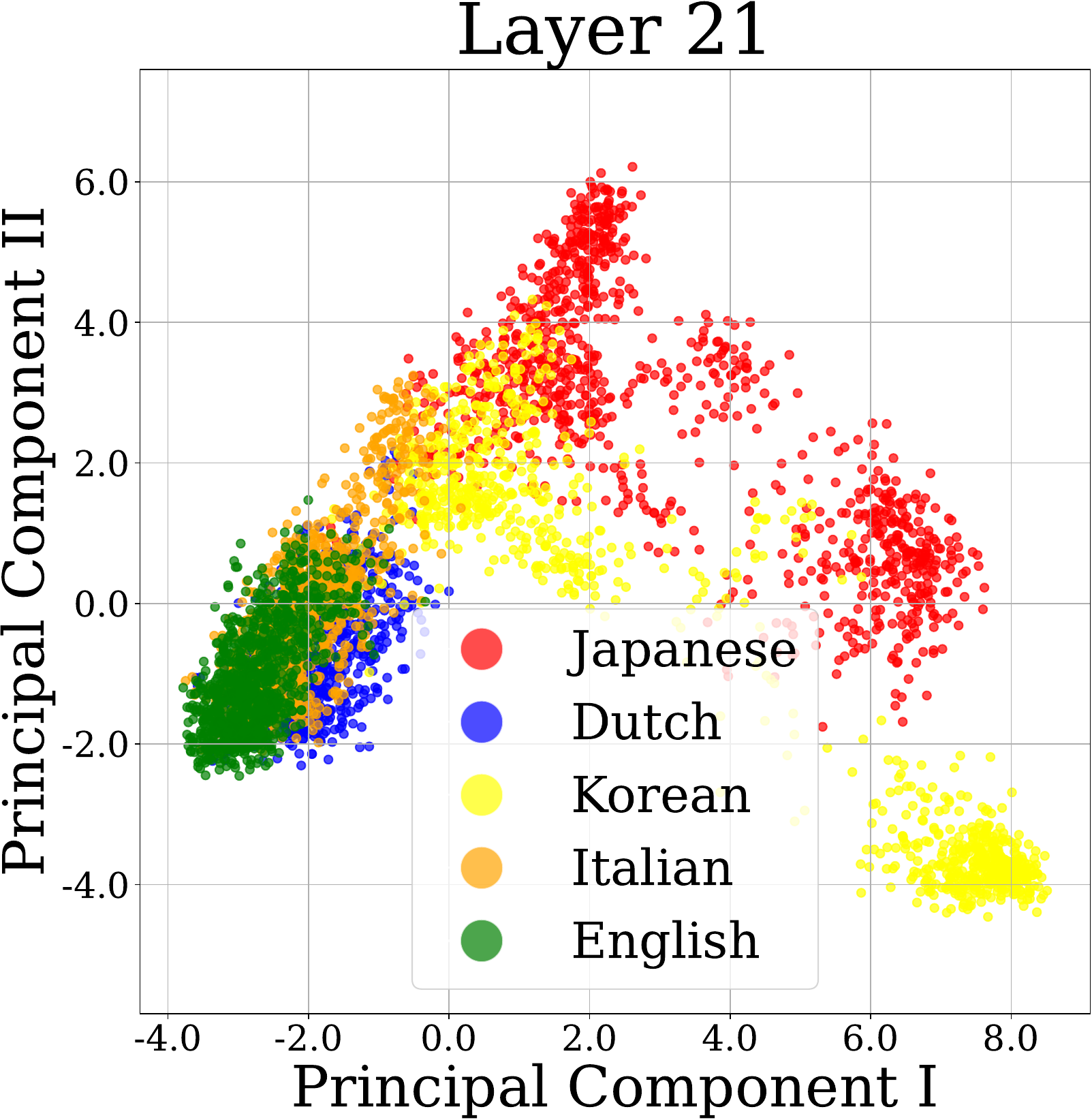}
  \includegraphics[width=0.19\linewidth]{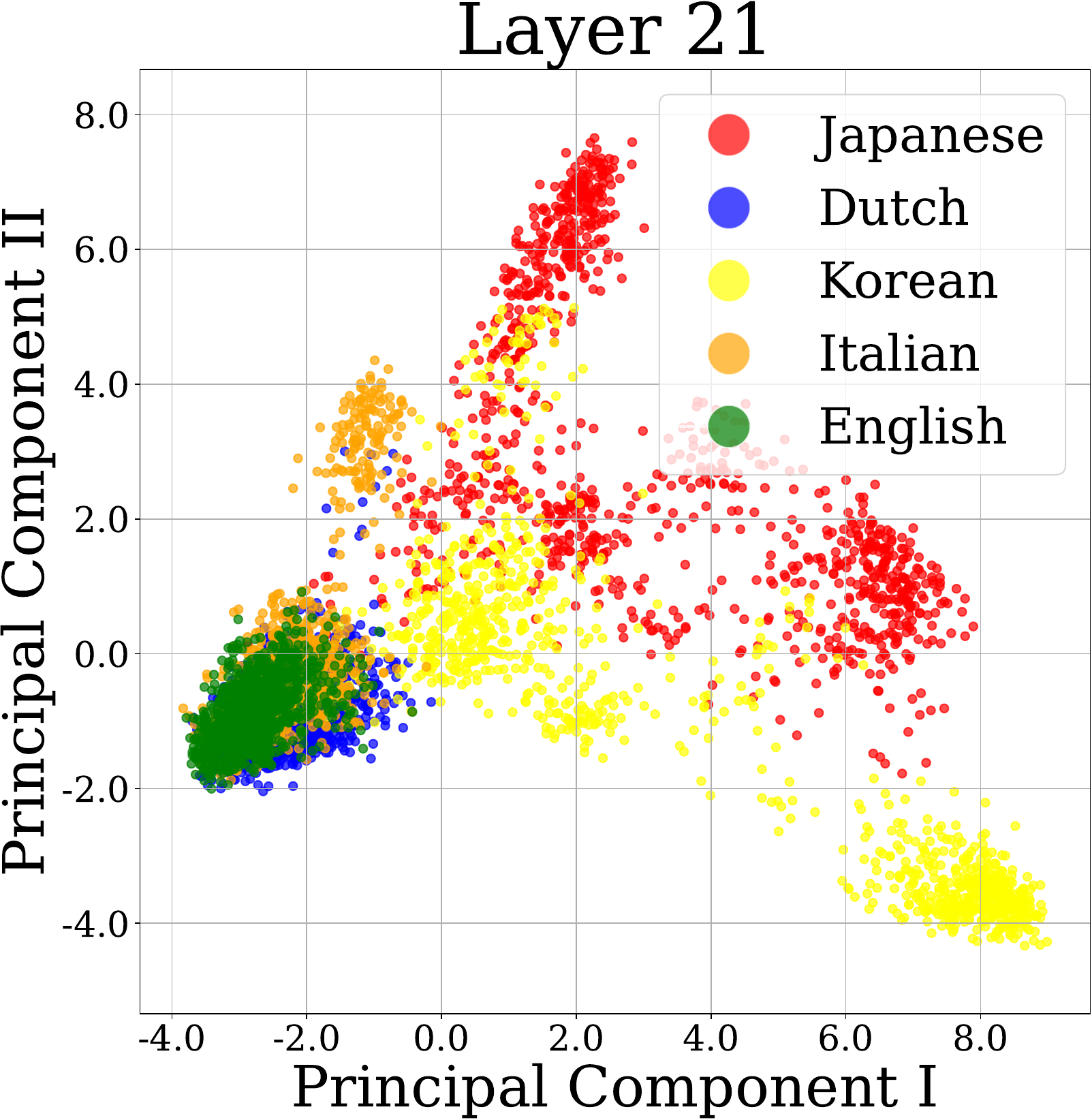}
  \includegraphics[width=0.19\linewidth]{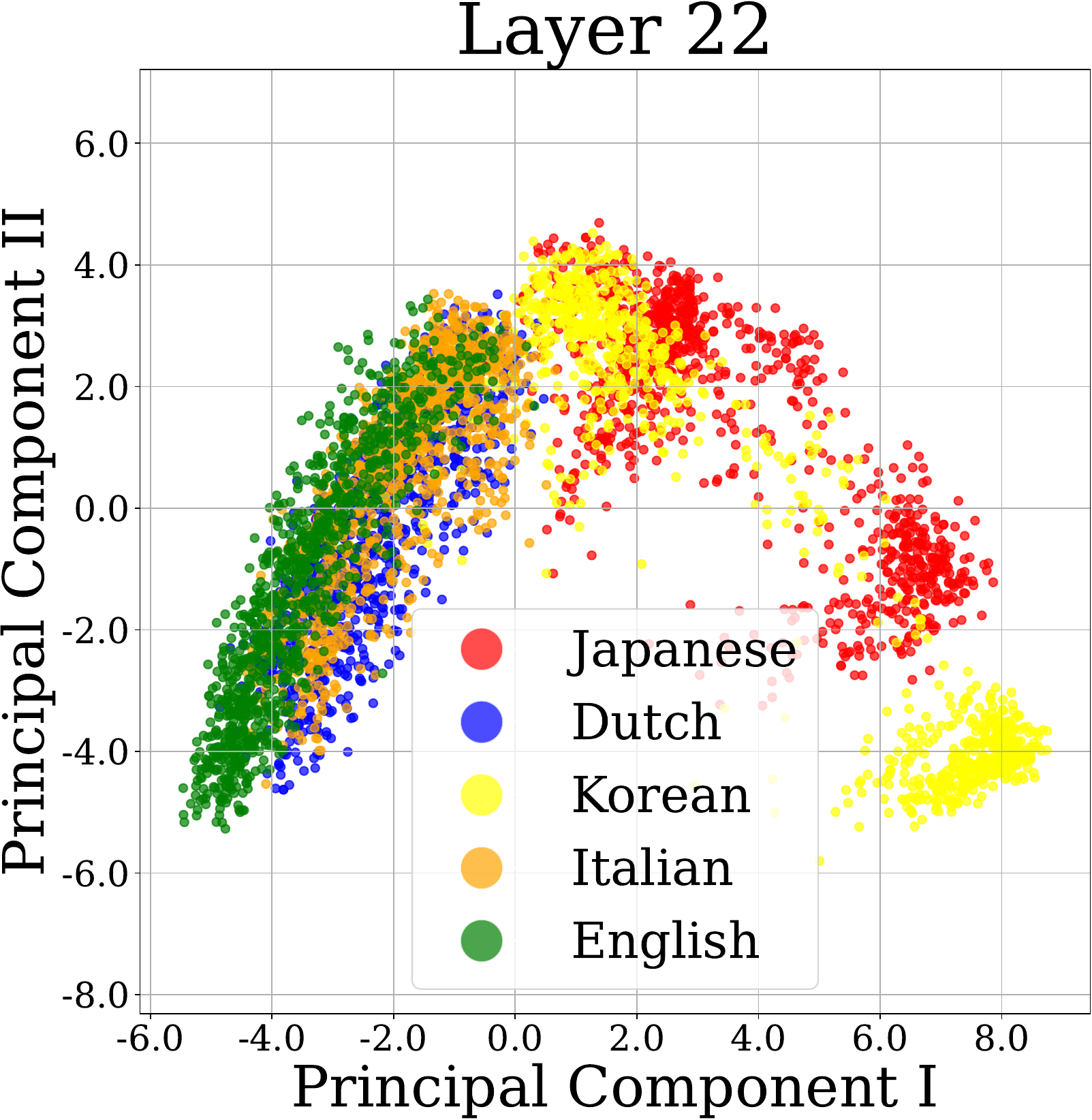}
  \includegraphics[width=0.19\linewidth]{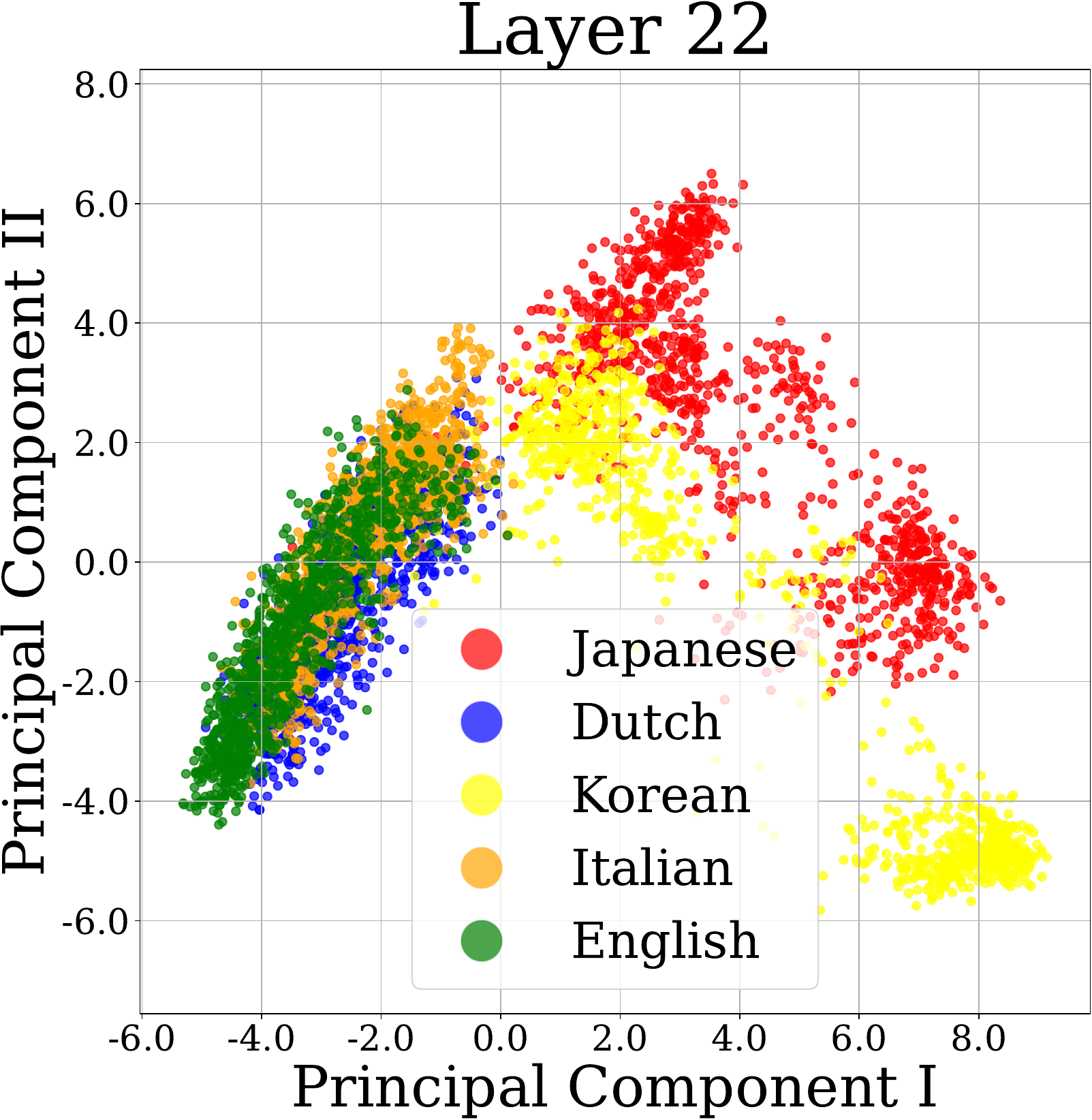}

  \begin{minipage}{0.19\linewidth}\centering \textbf{\textcolor{red}{layer 21 (Type-2)}}\end{minipage}
  \begin{minipage}{0.19\linewidth}\centering layer 21 (baseline)\end{minipage}
  \begin{minipage}{0.19\linewidth}\centering \textbf{\textcolor{red}{layer 22 (Type-2)}}\end{minipage}
  \begin{minipage}{0.19\linewidth}\centering layer 22 (baseline)\end{minipage}

  \includegraphics[width=0.19\linewidth]{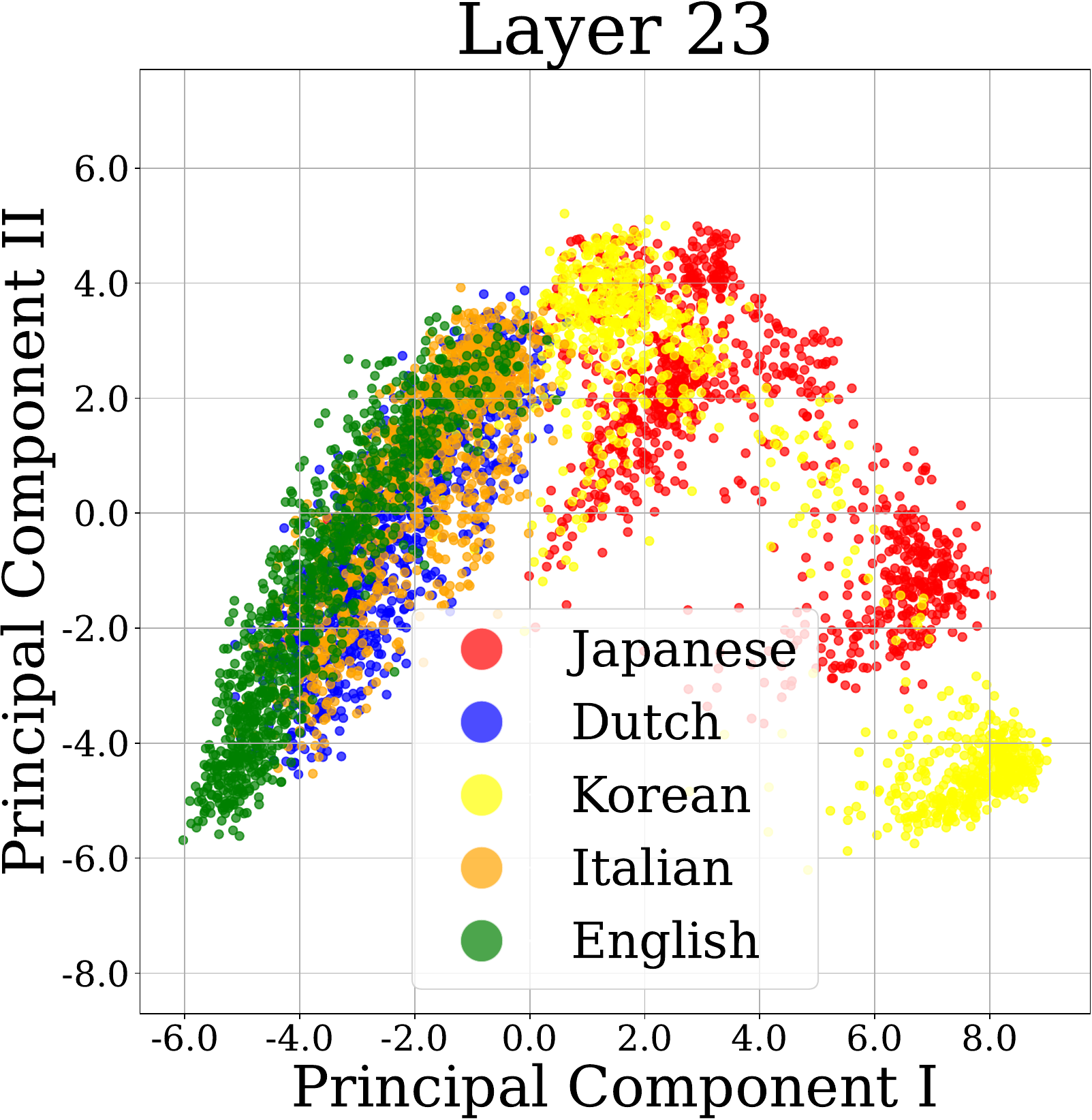}
  \includegraphics[width=0.19\linewidth]{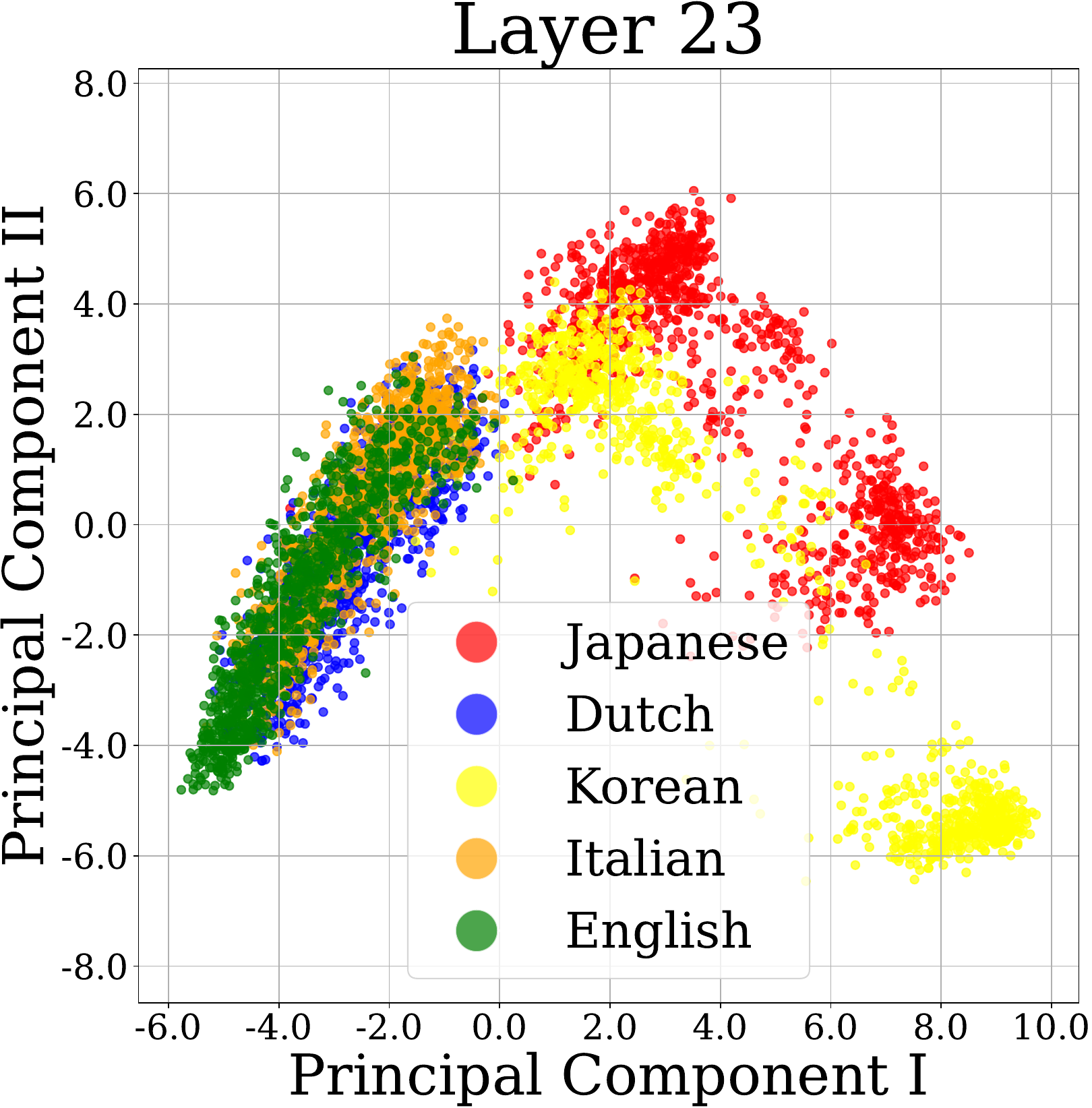}
  \includegraphics[width=0.19\linewidth]{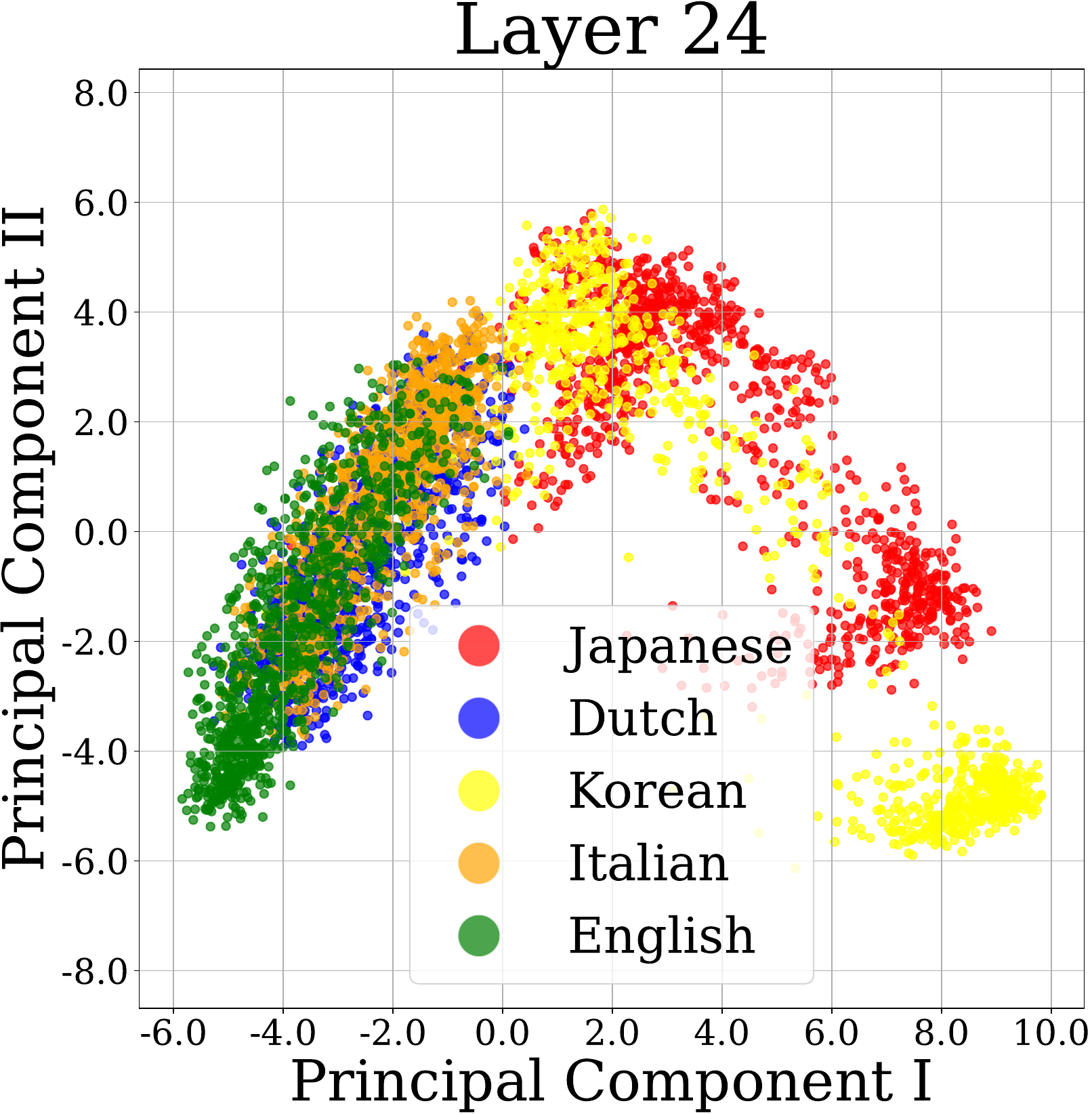}
  \includegraphics[width=0.19\linewidth]{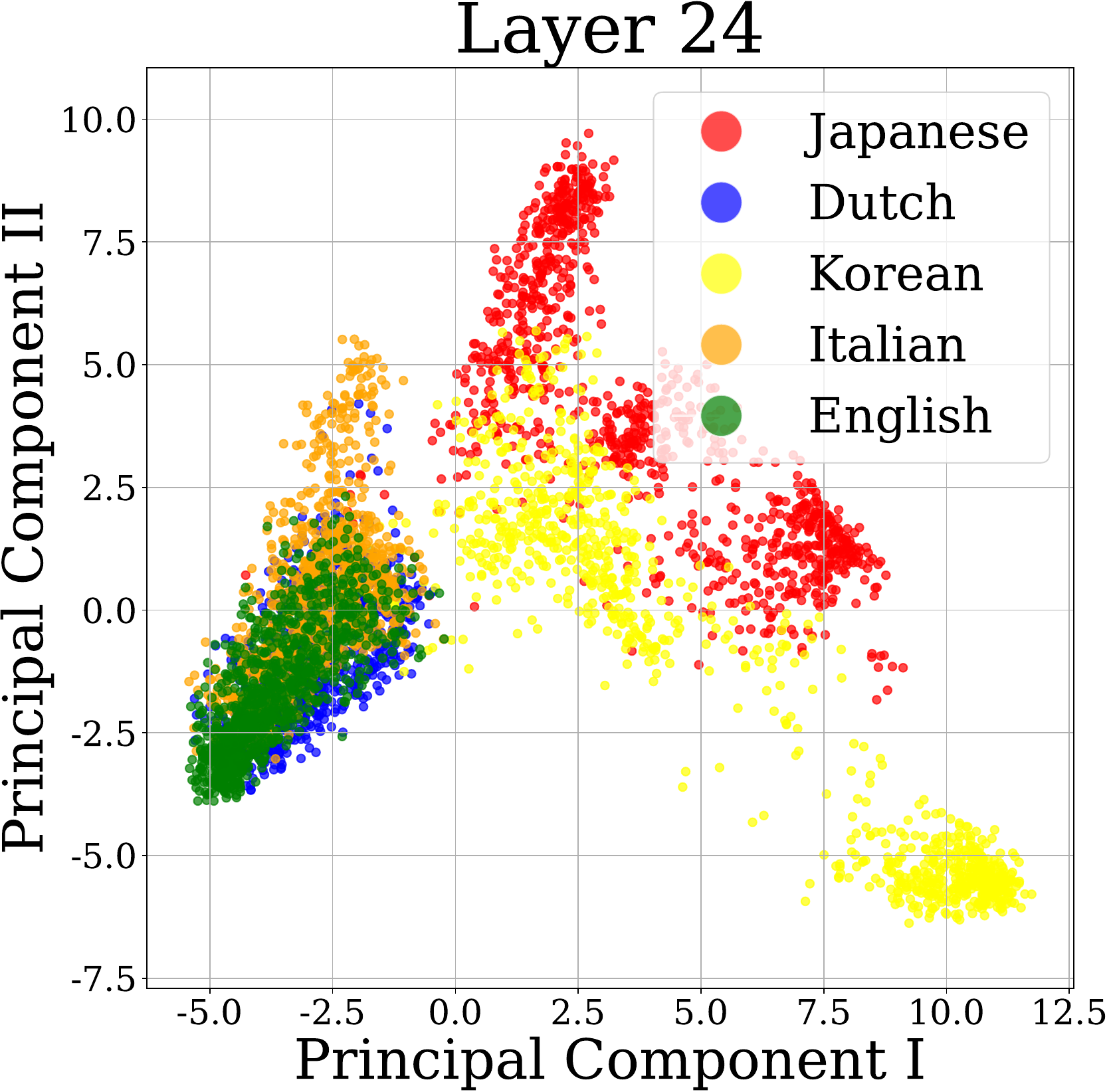}

  \begin{minipage}{0.19\linewidth}\centering \textbf{\textcolor{red}{layer 23 (Type-2)}}\end{minipage}
  \begin{minipage}{0.19\linewidth}\centering layer 23 (baseline)\end{minipage}
  \begin{minipage}{0.19\linewidth}\centering \textbf{\textcolor{red}{layer 24 (Type-2)}}\end{minipage}
  \begin{minipage}{0.19\linewidth}\centering layer 24 (baseline)\end{minipage}

  \includegraphics[width=0.19\linewidth]{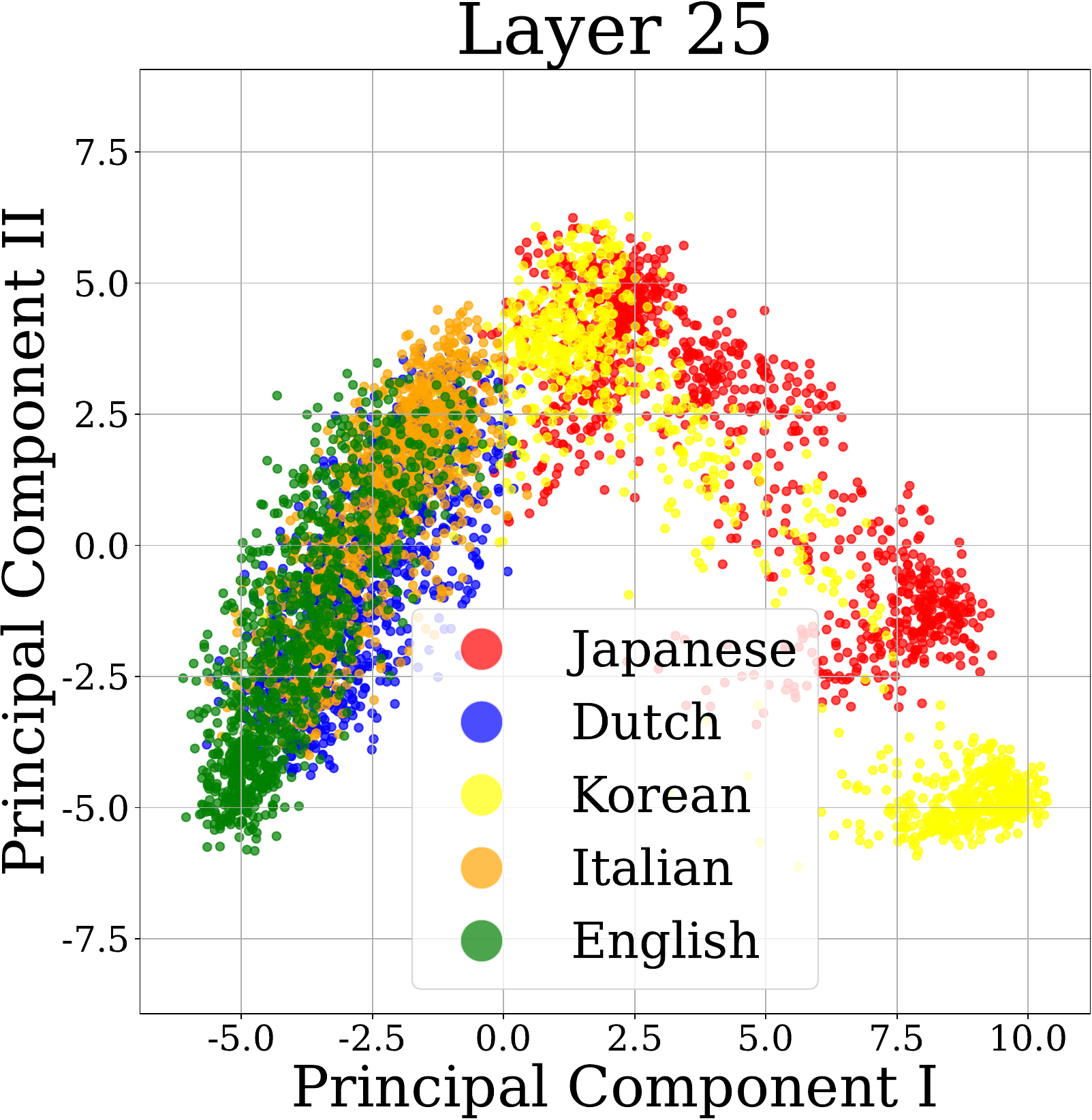}
  \includegraphics[width=0.19\linewidth]{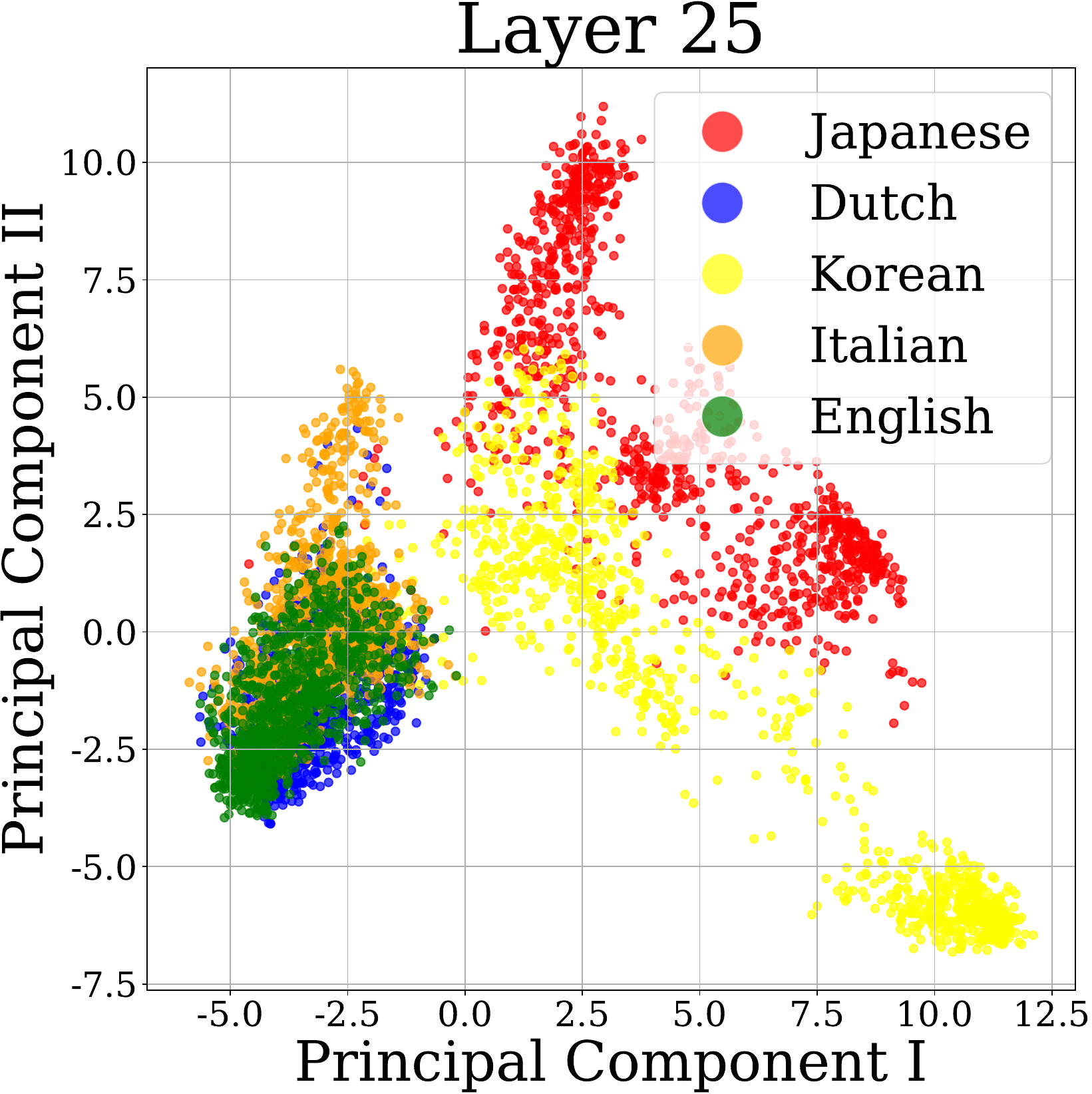}
  \includegraphics[width=0.19\linewidth]{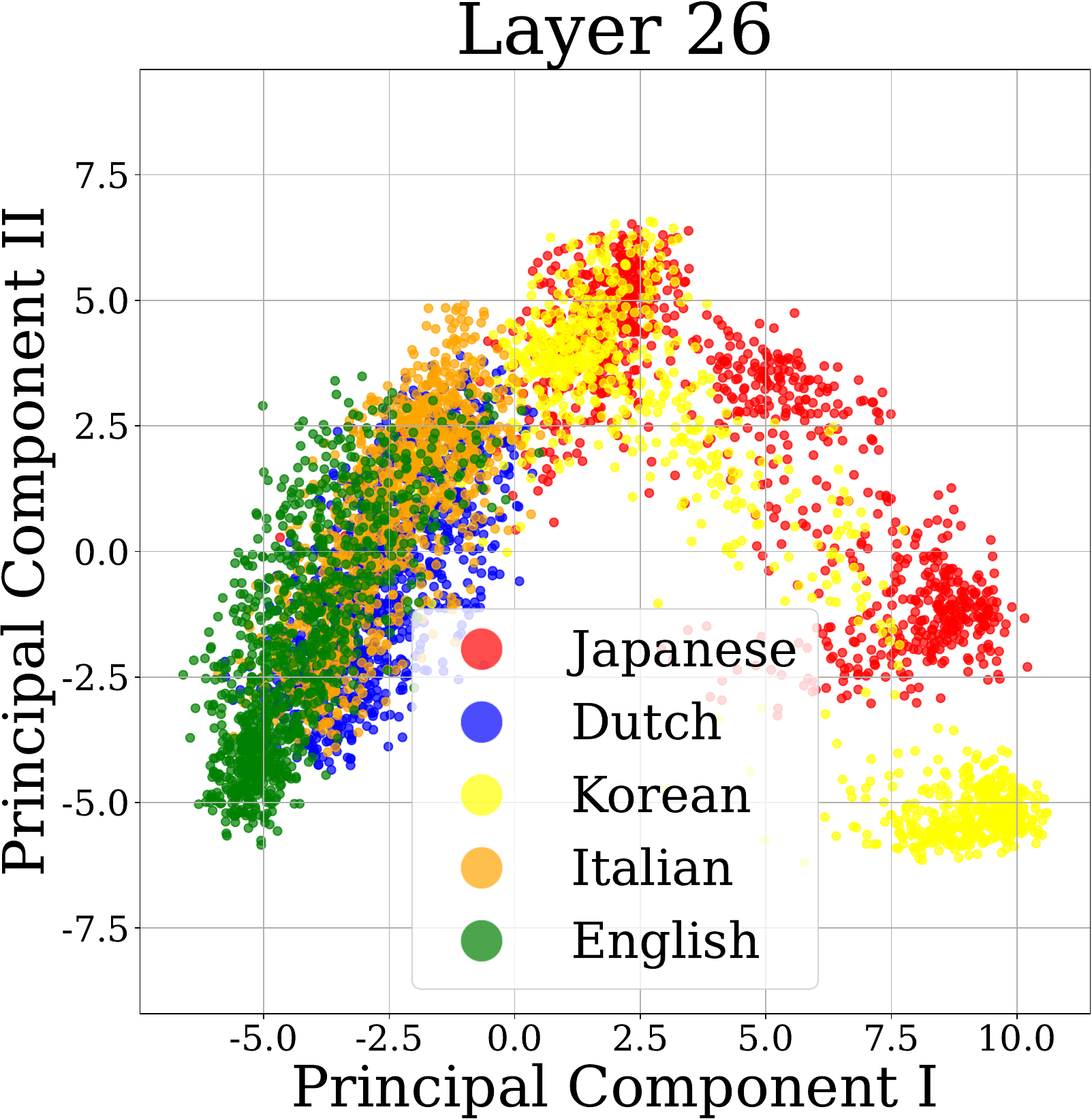}
  \includegraphics[width=0.19\linewidth]{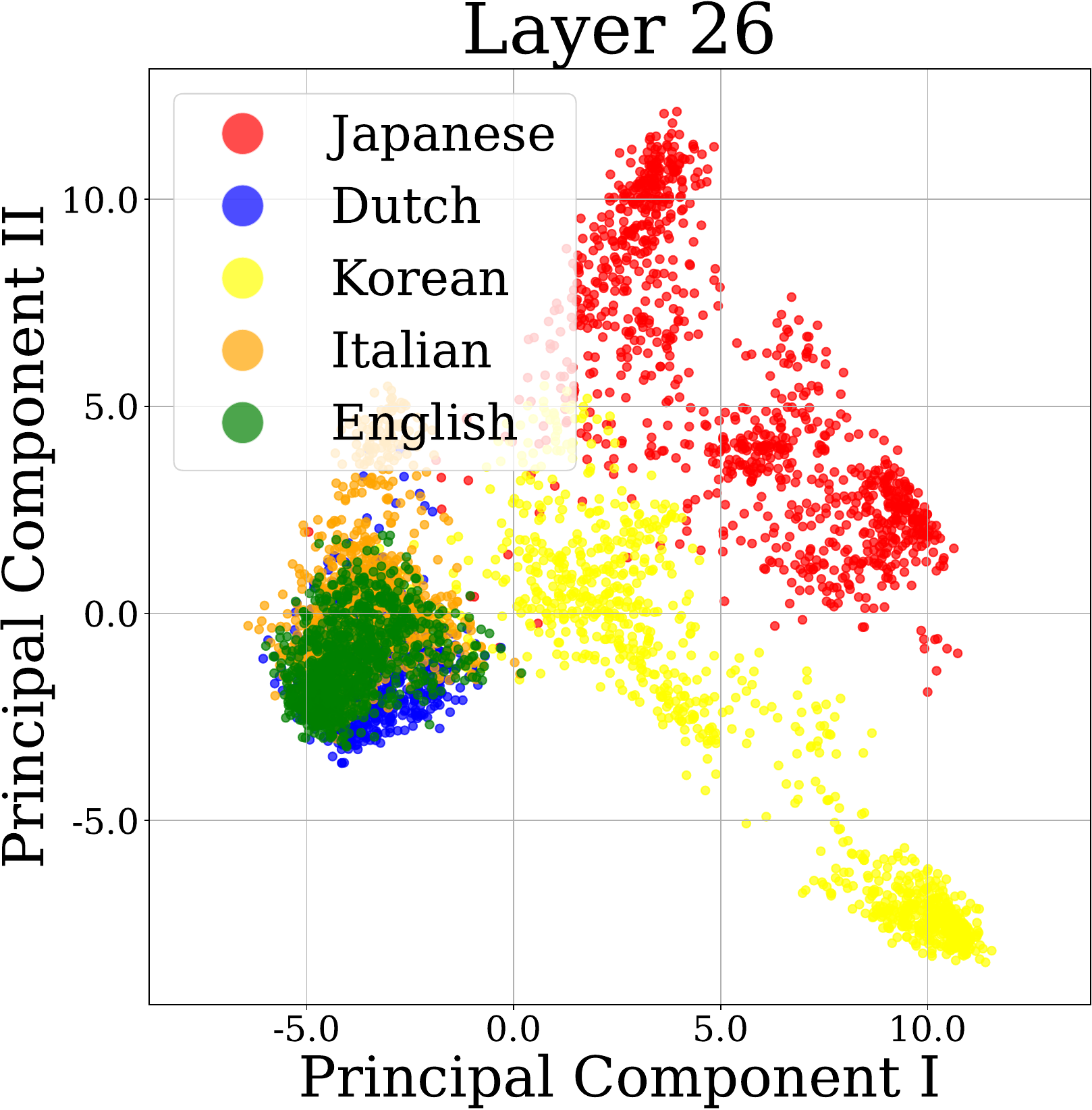}

  \begin{minipage}{0.19\linewidth}\centering \textbf{\textcolor{red}{layer 25 (Type-2)}}\end{minipage}
  \begin{minipage}{0.19\linewidth}\centering layer 25 (baseline)\end{minipage}
  \begin{minipage}{0.19\linewidth}\centering \textbf{\textcolor{red}{layer 26 (Type-2)}}\end{minipage}
  \begin{minipage}{0.19\linewidth}\centering layer 26 (baseline)\end{minipage}

  \includegraphics[width=0.19\linewidth]{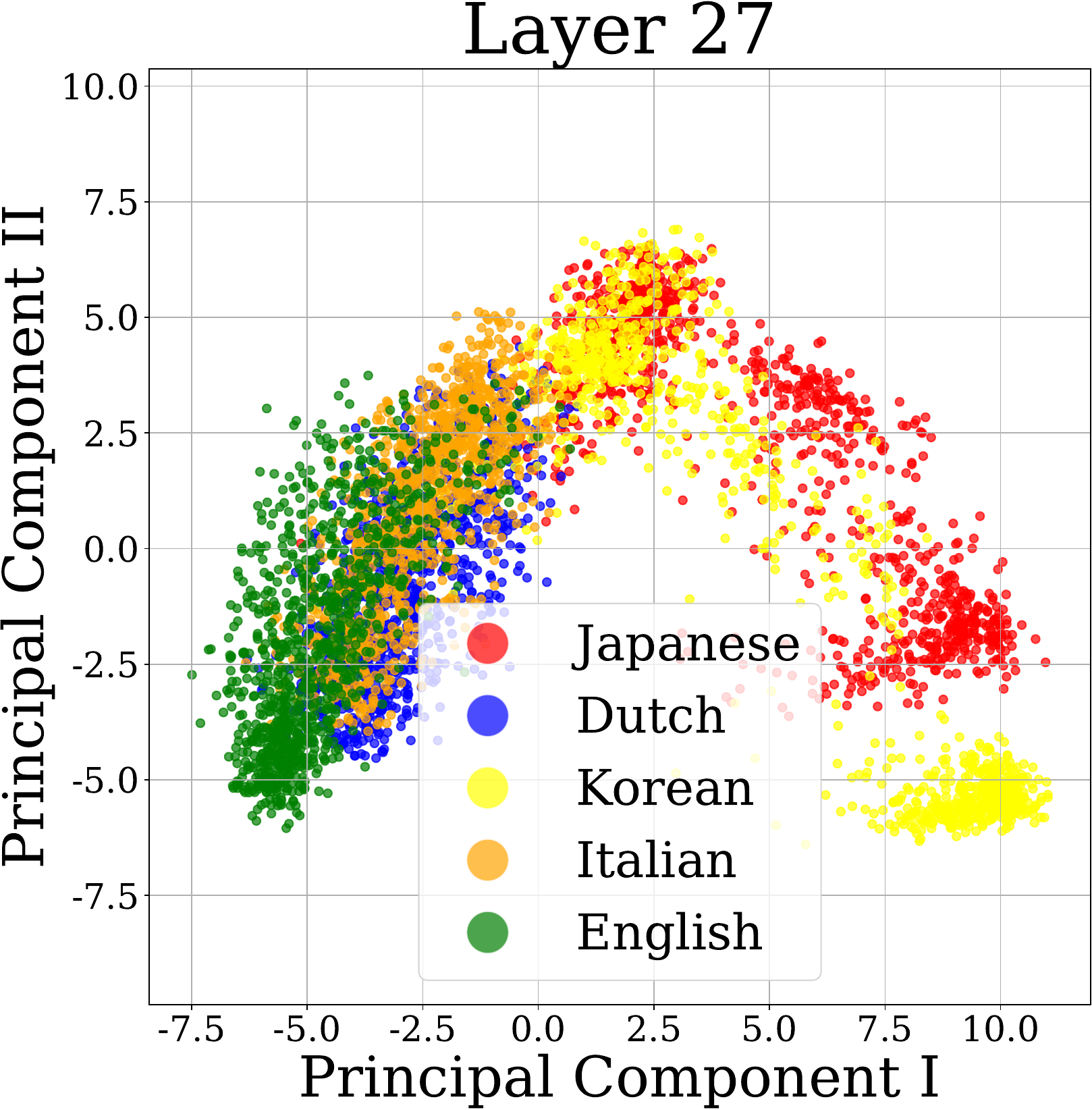}
  \includegraphics[width=0.19\linewidth]{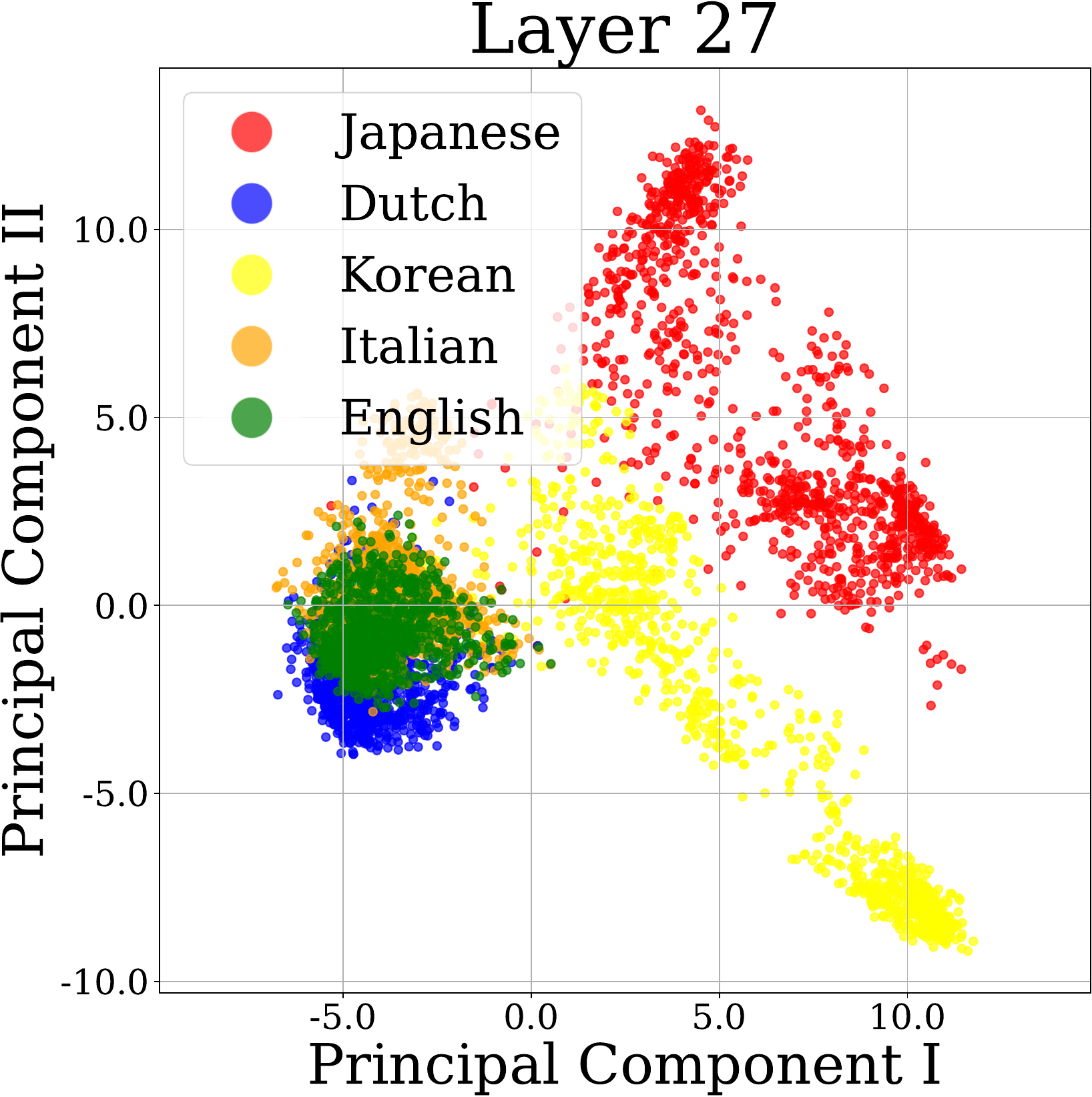}
  \includegraphics[width=0.19\linewidth]{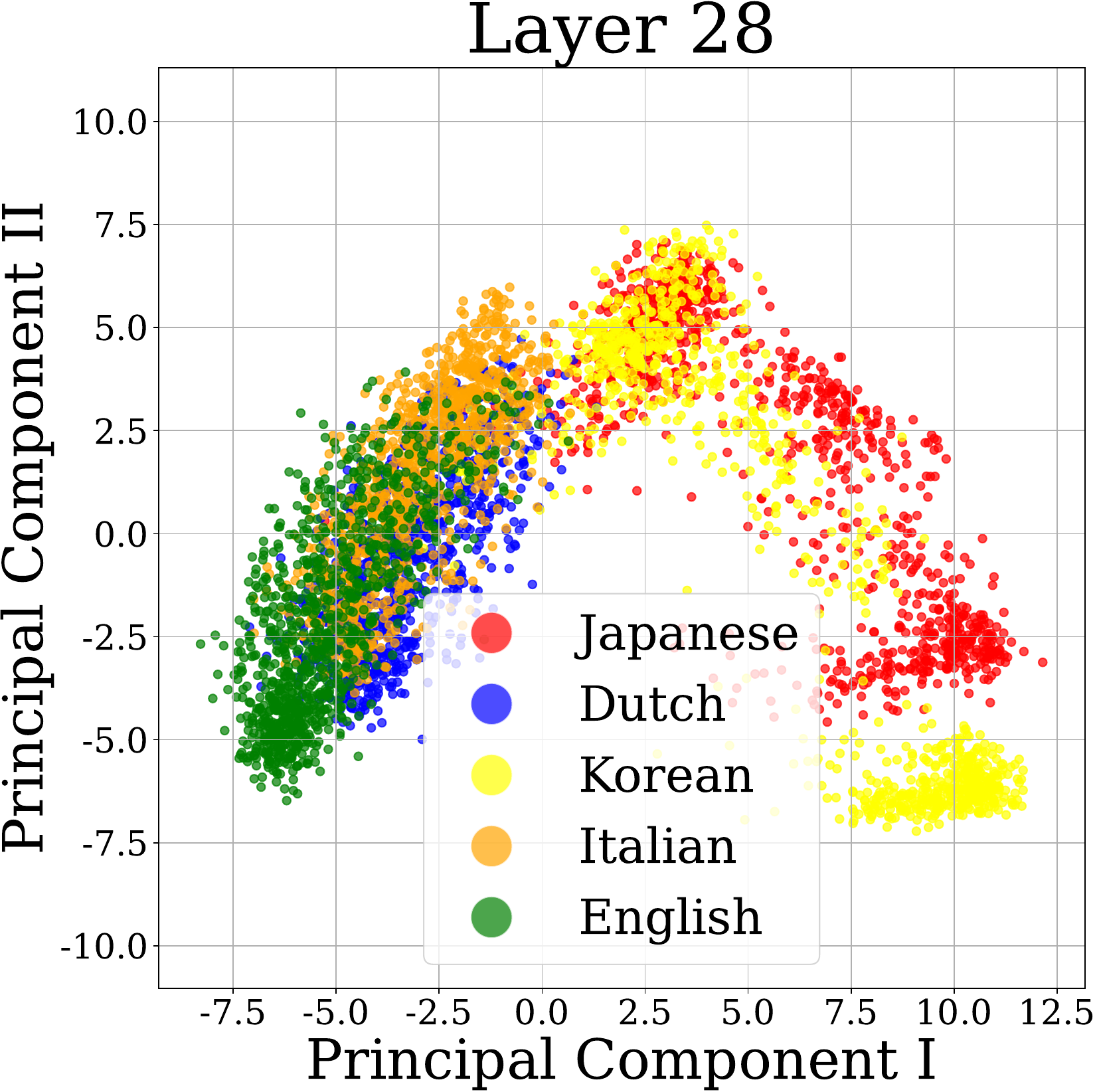}
  \includegraphics[width=0.19\linewidth]{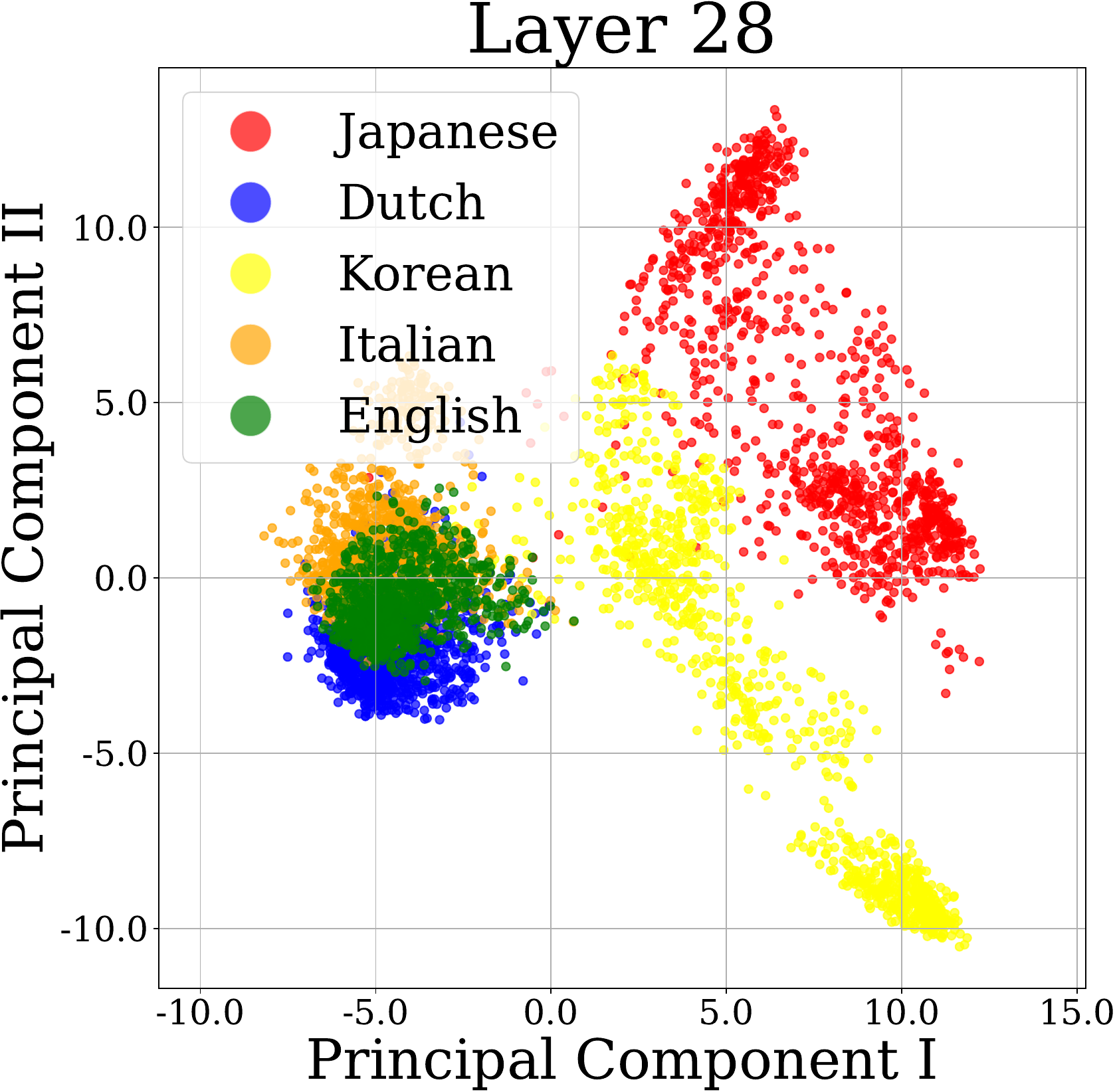}

  \begin{minipage}{0.19\linewidth}\centering \textbf{\textcolor{red}{layer 27 (Type-2)}}\end{minipage}
  \begin{minipage}{0.19\linewidth}\centering layer 27 (baseline)\end{minipage}
  \begin{minipage}{0.19\linewidth}\centering \textbf{\textcolor{red}{layer 28 (Type-2)}}\end{minipage}
  \begin{minipage}{0.19\linewidth}\centering layer 28 (baseline)\end{minipage}

  \includegraphics[width=0.19\linewidth]{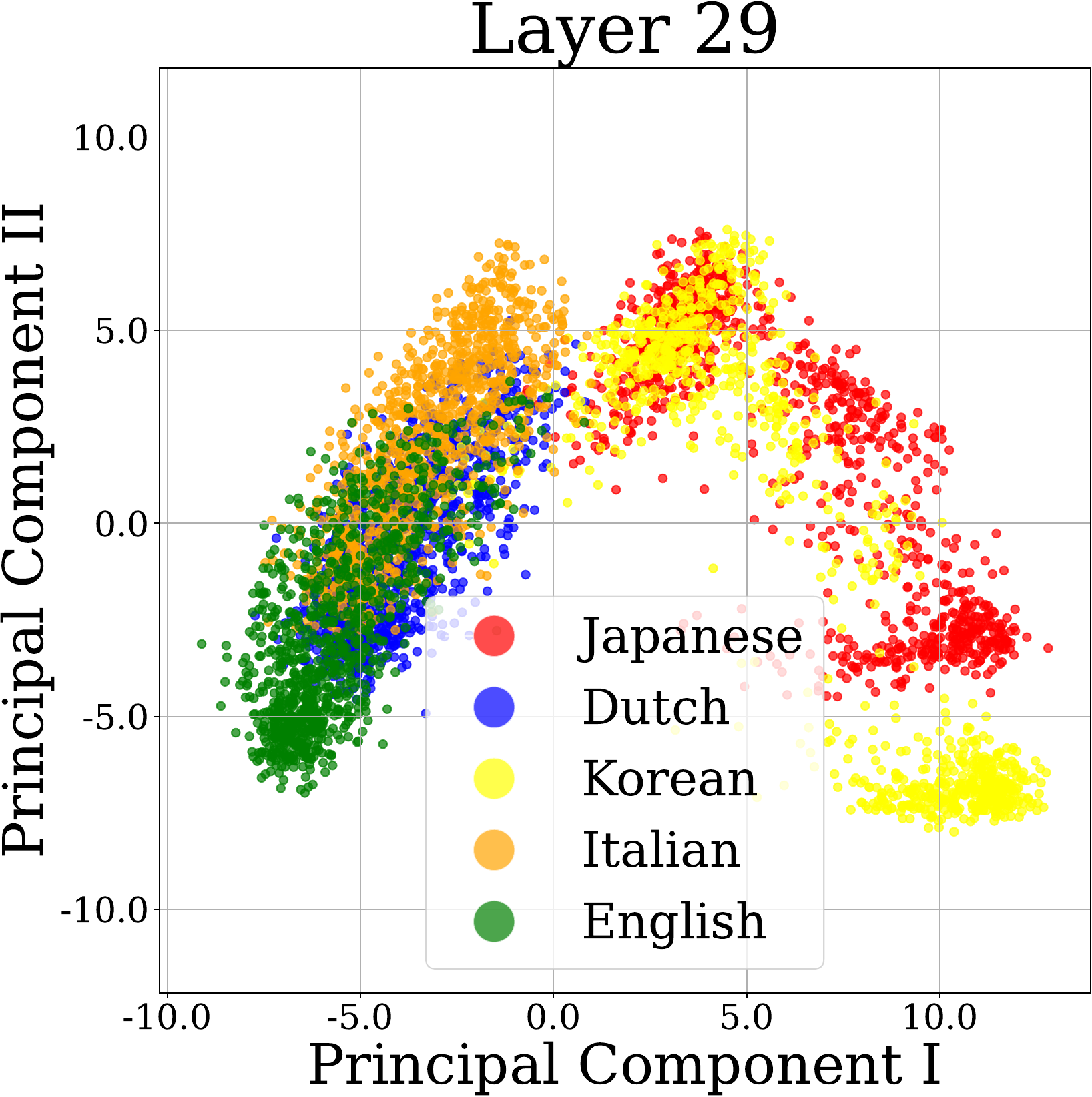}
  \includegraphics[width=0.19\linewidth]{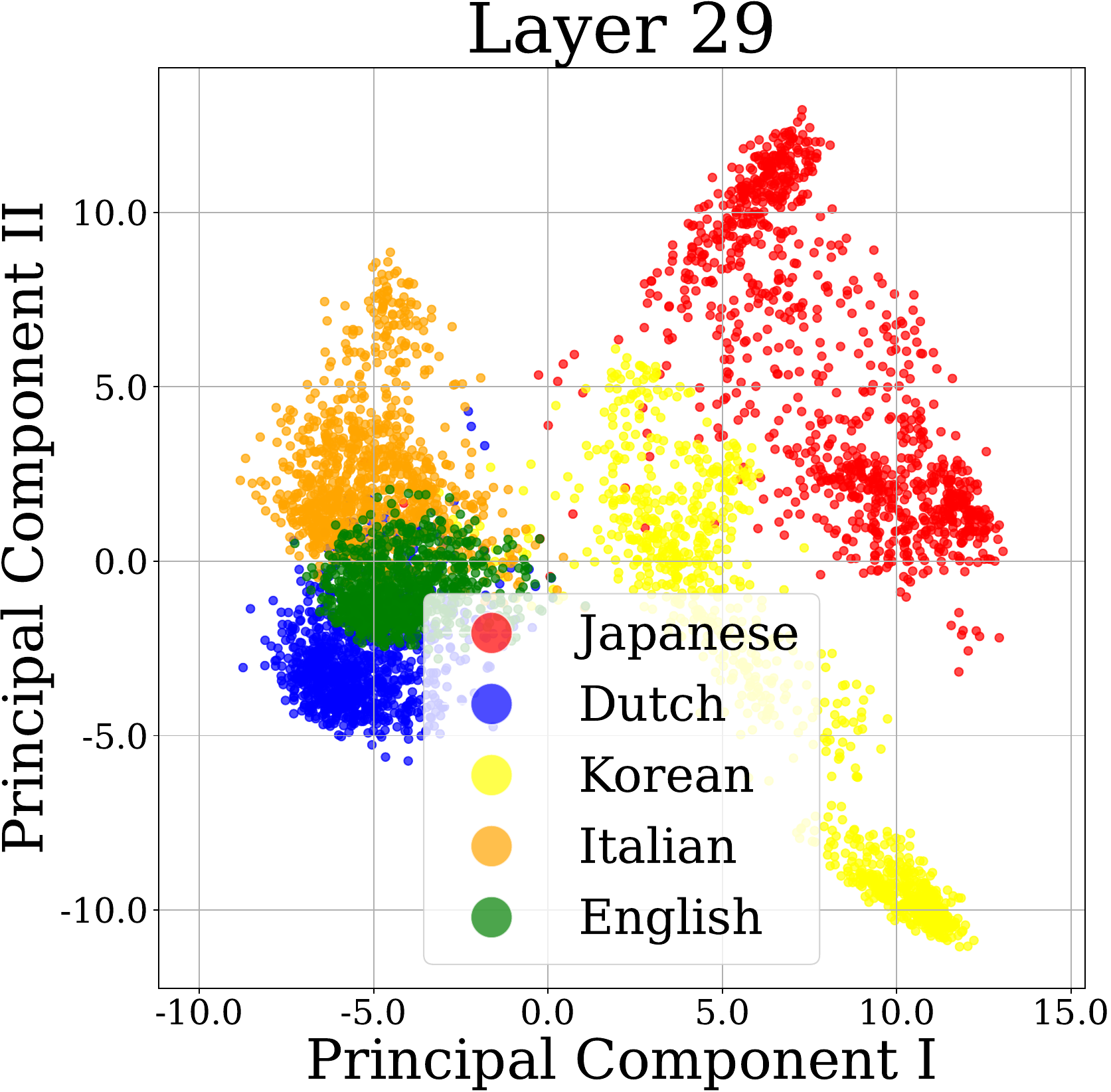}
  \includegraphics[width=0.19\linewidth]{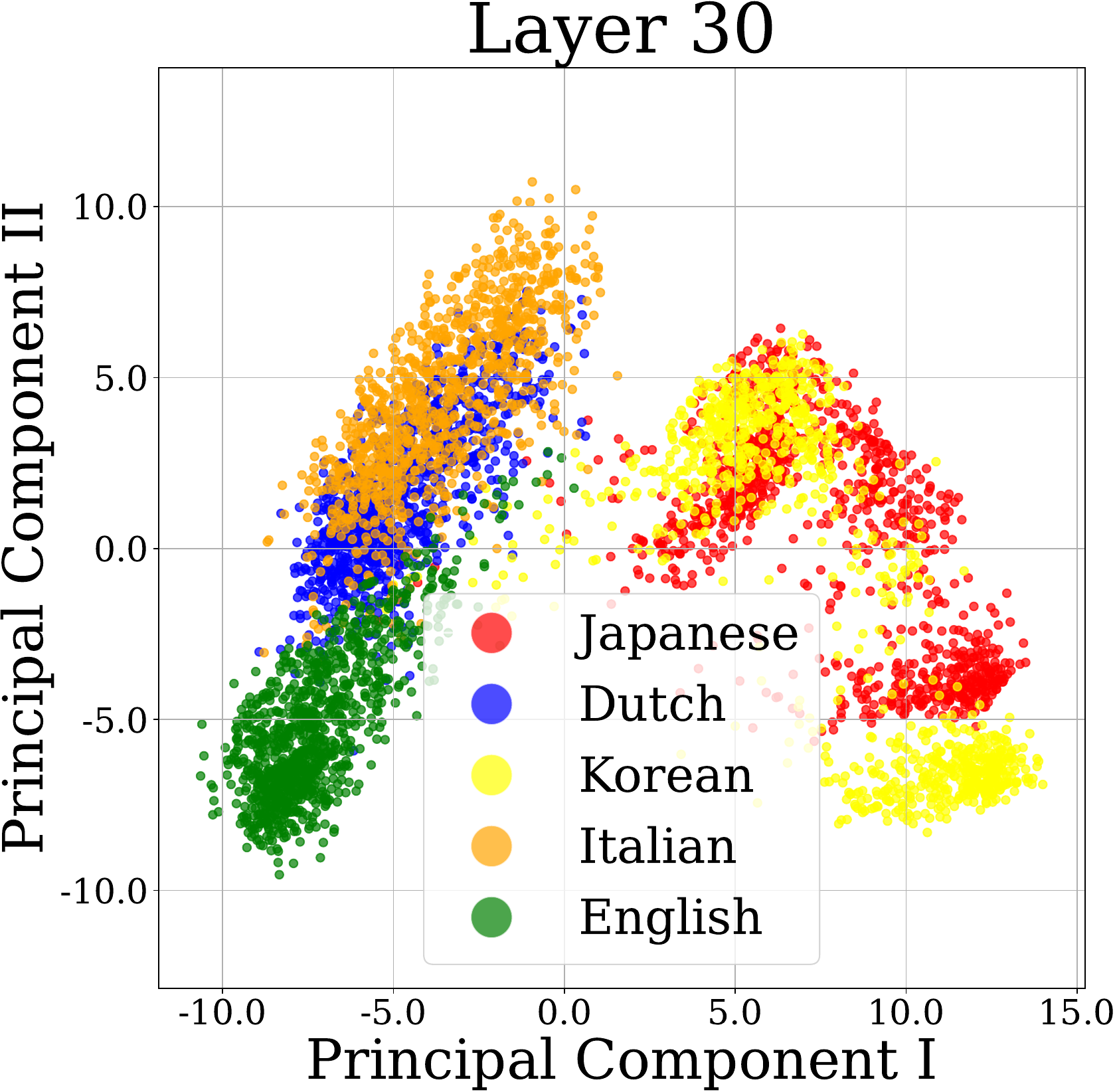}
  \includegraphics[width=0.19\linewidth]{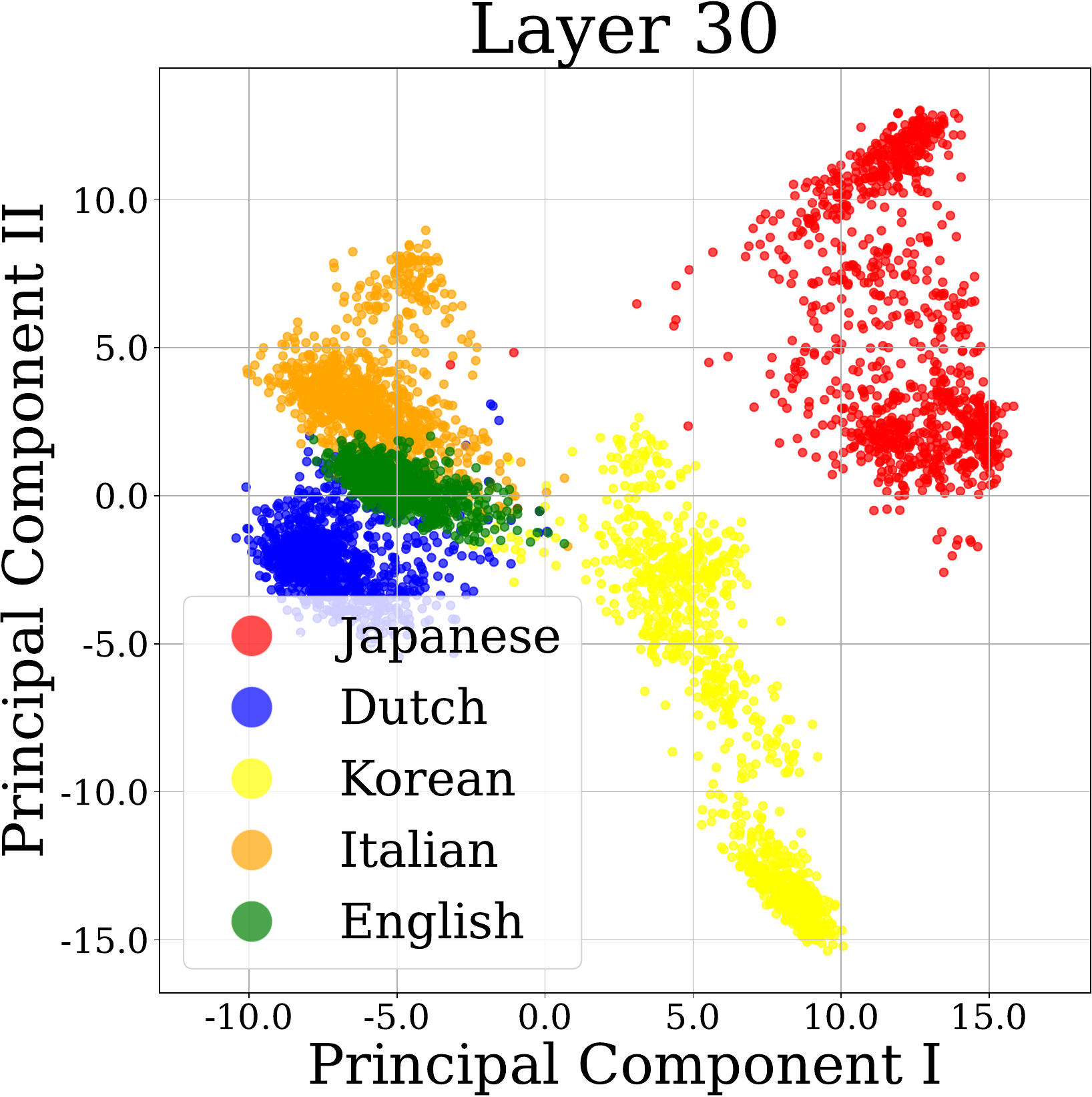}

  \begin{minipage}{0.19\linewidth}\centering \textbf{\textcolor{red}{layer 29 (Type-2)}}\end{minipage}
  \begin{minipage}{0.19\linewidth}\centering layer 29 (baseline)\end{minipage}
  \begin{minipage}{0.19\linewidth}\centering \textbf{\textcolor{red}{layer 30 (Type-2)}}\end{minipage}
  \begin{minipage}{0.19\linewidth}\centering layer 30 (baseline)\end{minipage}

  \includegraphics[width=0.19\linewidth]{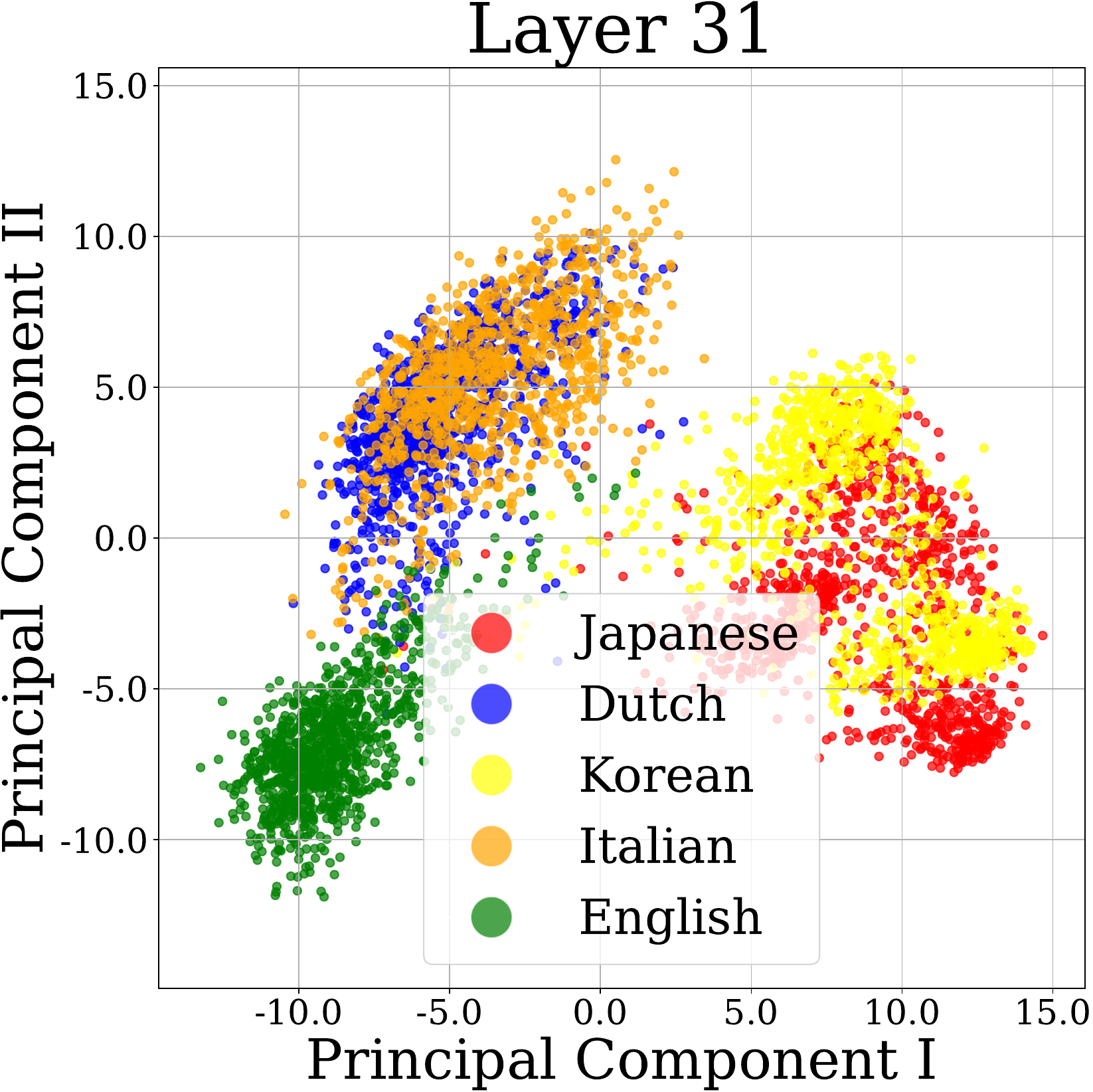}
  \includegraphics[width=0.19\linewidth]{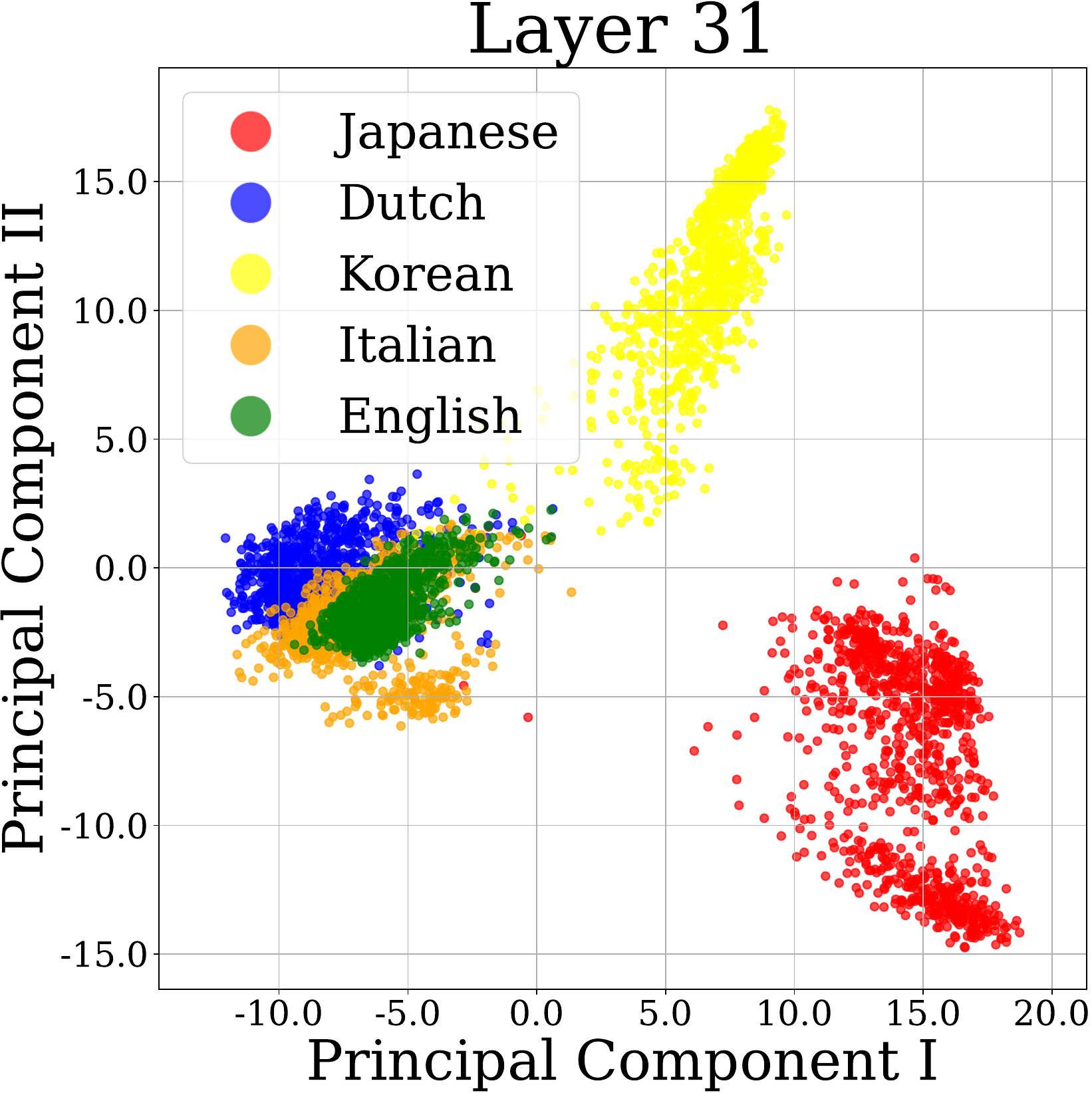}
  \includegraphics[width=0.19\linewidth]{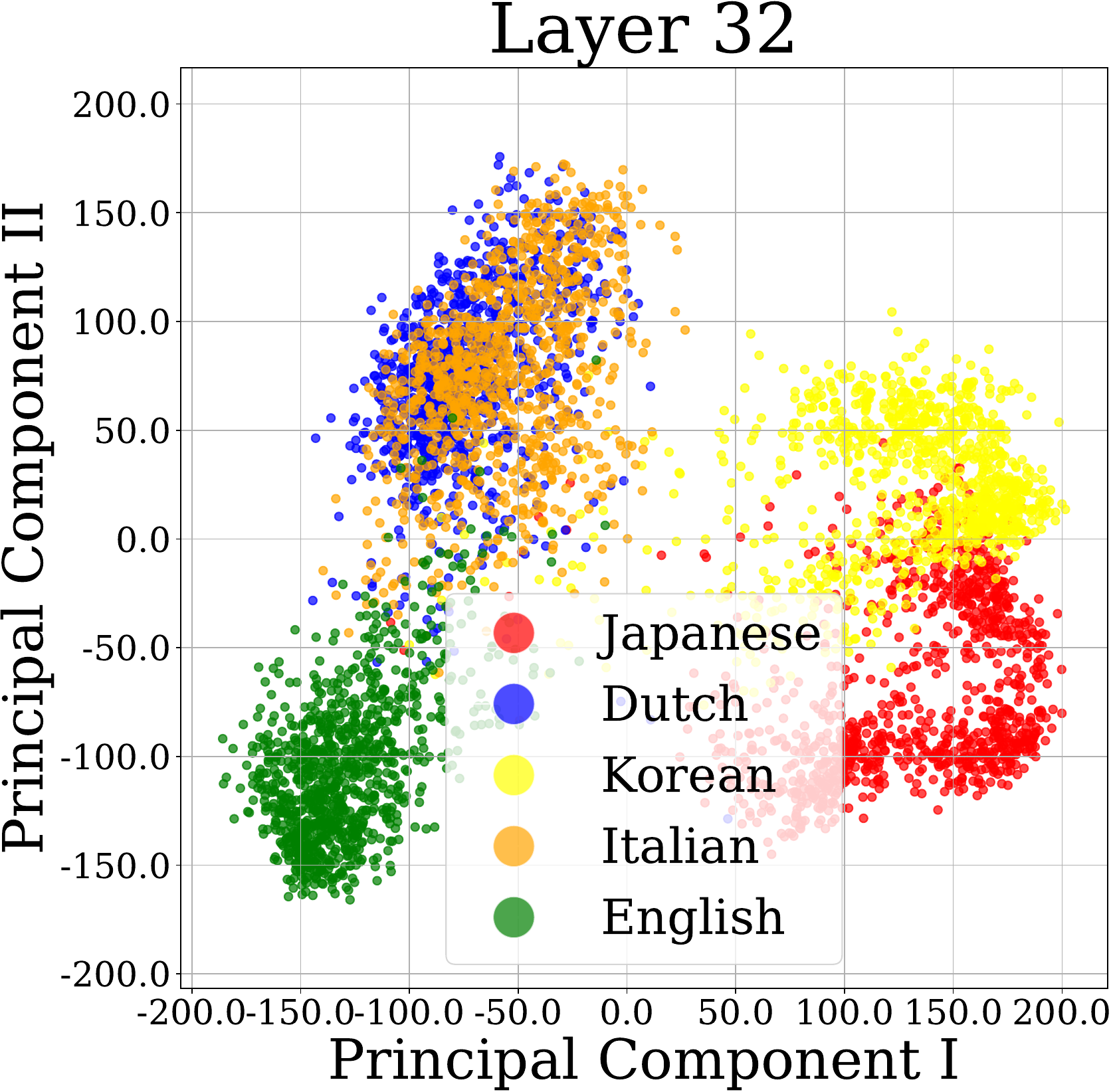}
  \includegraphics[width=0.19\linewidth]{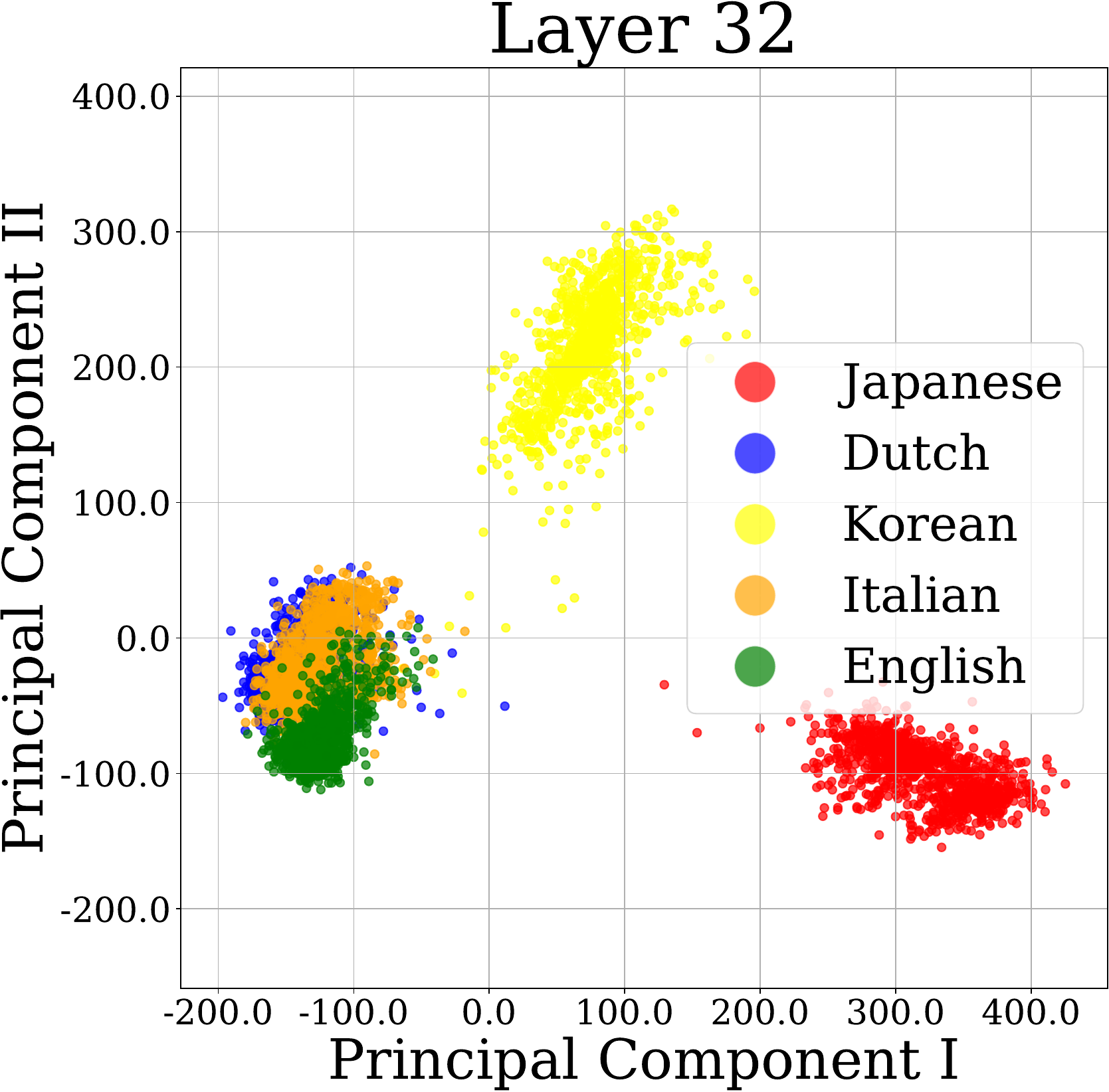}

  \begin{minipage}{0.19\linewidth}\centering \textbf{\textcolor{red}{layer 31 (Type-2)}}\end{minipage}
  \begin{minipage}{0.19\linewidth}\centering layer 31 (baseline)\end{minipage}
  \begin{minipage}{0.19\linewidth}\centering \textbf{\textcolor{red}{layer 32 (Type-2)}}\end{minipage}
  \begin{minipage}{0.19\linewidth}\centering layer 32 (baseline)\end{minipage}

  \caption{\textbf{The resutls of PCA while deactivating Top-1k Type-2 Transfer Neurons (Mistral-7B)}.\\.}
  \label{fig:appendix:pca_deactivating_type2_mistral}
\end{figure*}

\begin{figure*}[t]
  \centering

  \includegraphics[width=0.19\linewidth]{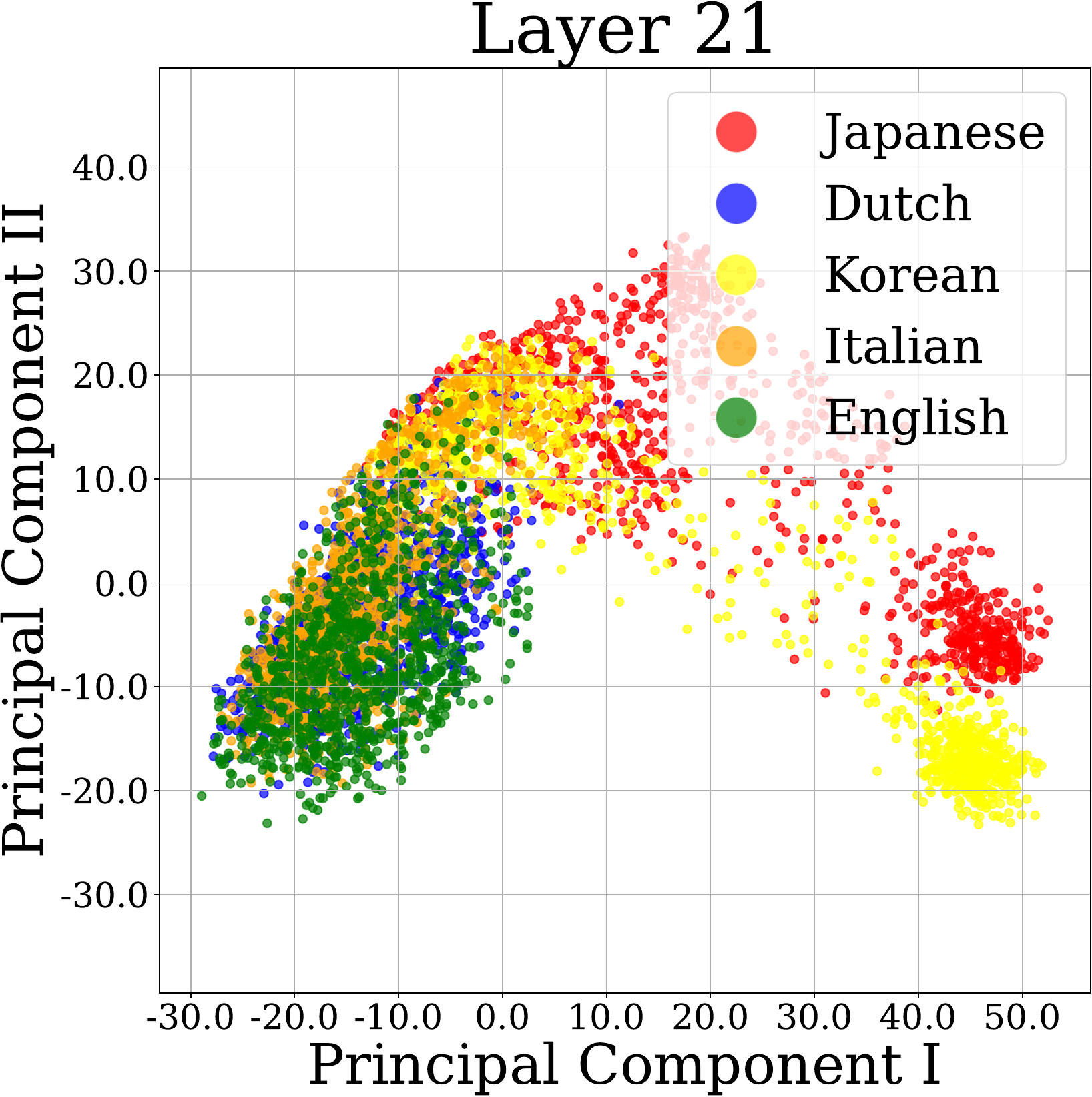}
  \includegraphics[width=0.19\linewidth]{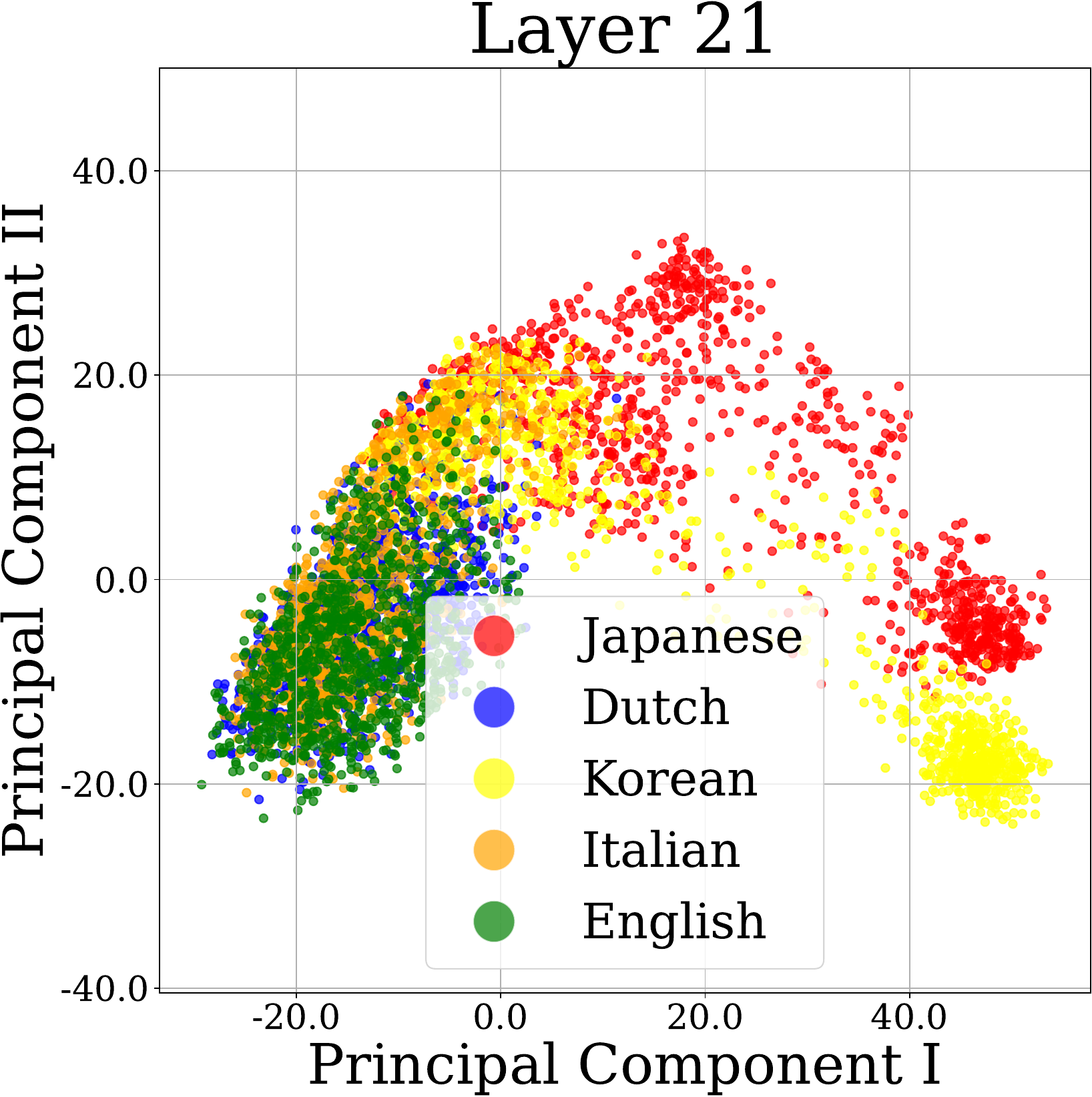}
  \includegraphics[width=0.19\linewidth]{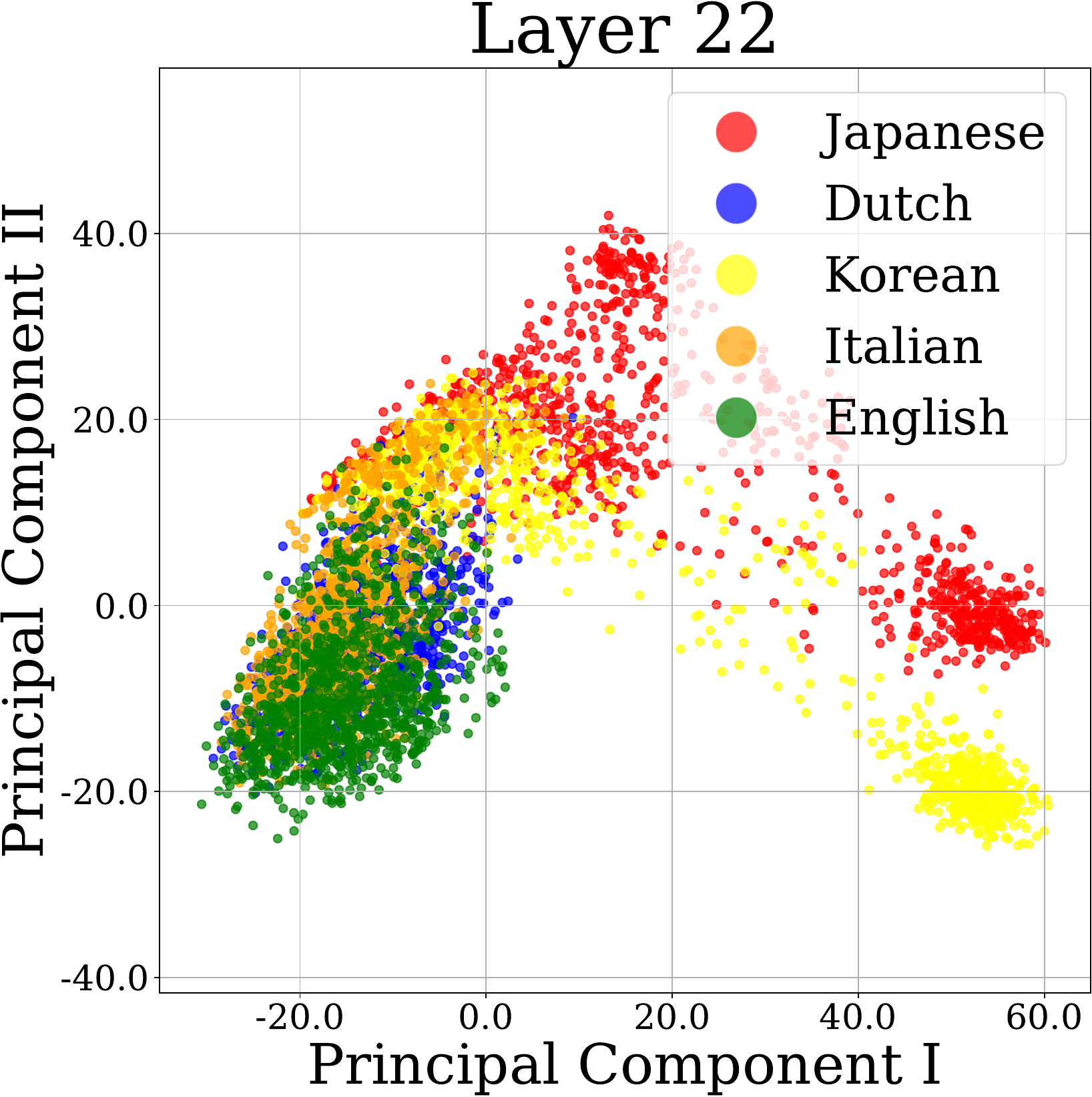}
  \includegraphics[width=0.19\linewidth]{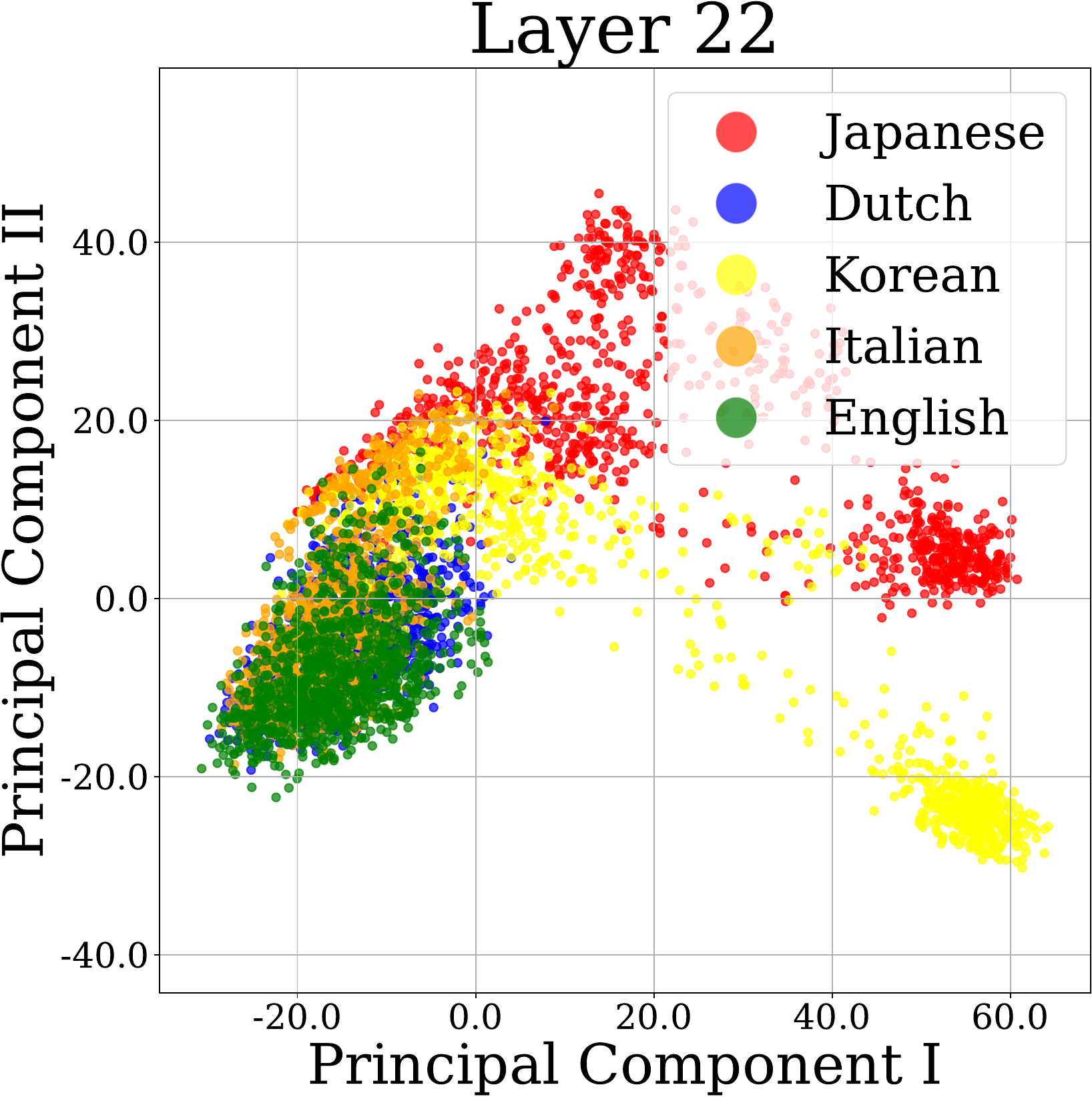}

  \begin{minipage}{0.19\linewidth}\centering \textbf{\textcolor{red}{layer 21 (Type-2)}}\end{minipage}
  \begin{minipage}{0.19\linewidth}\centering layer 21 (baseline)\end{minipage}
  \begin{minipage}{0.19\linewidth}\centering \textbf{\textcolor{red}{layer 22 (Type-2)}}\end{minipage}
  \begin{minipage}{0.19\linewidth}\centering layer 22 (baseline)\end{minipage}

  \includegraphics[width=0.19\linewidth]{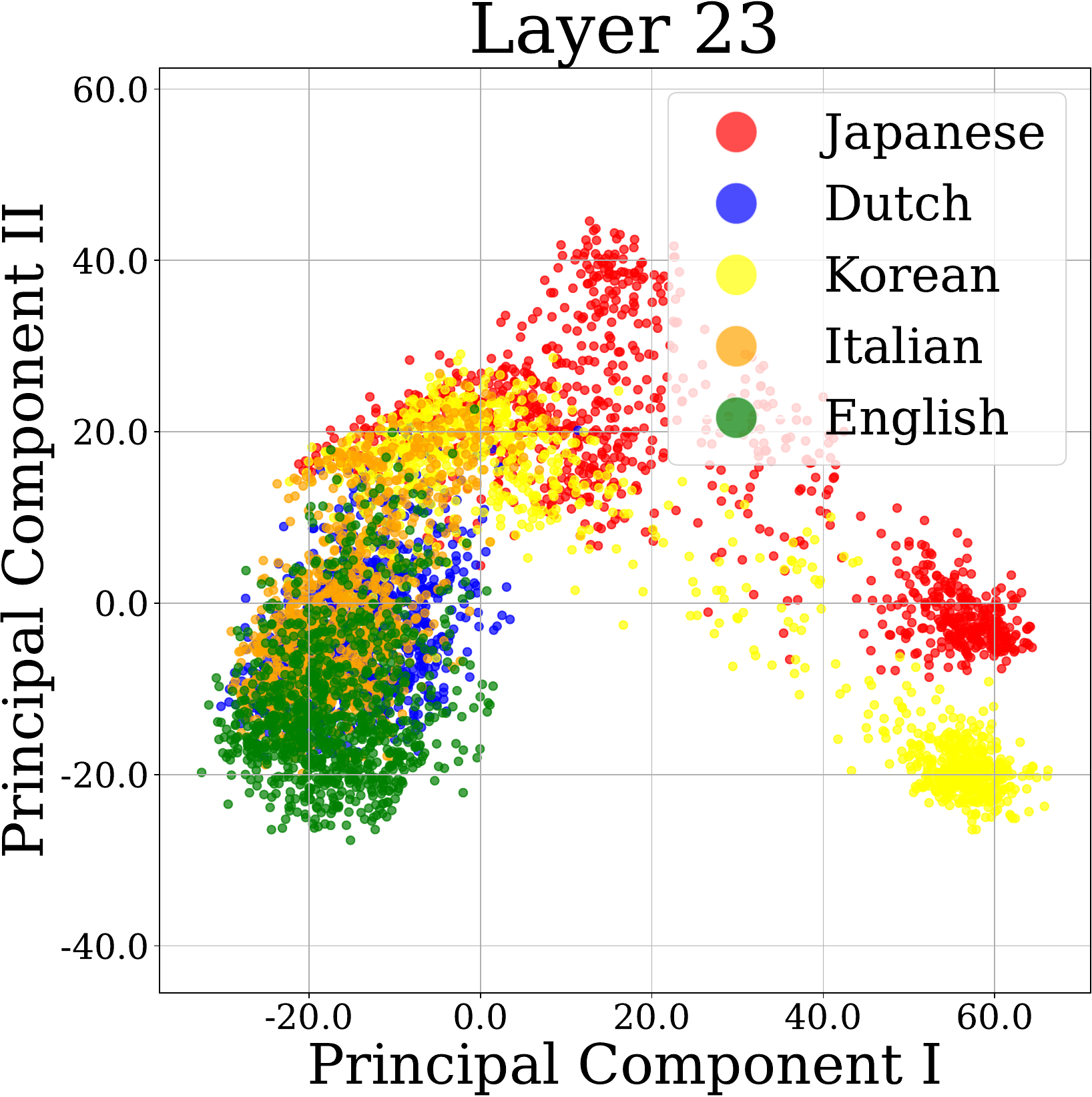}
  \includegraphics[width=0.19\linewidth]{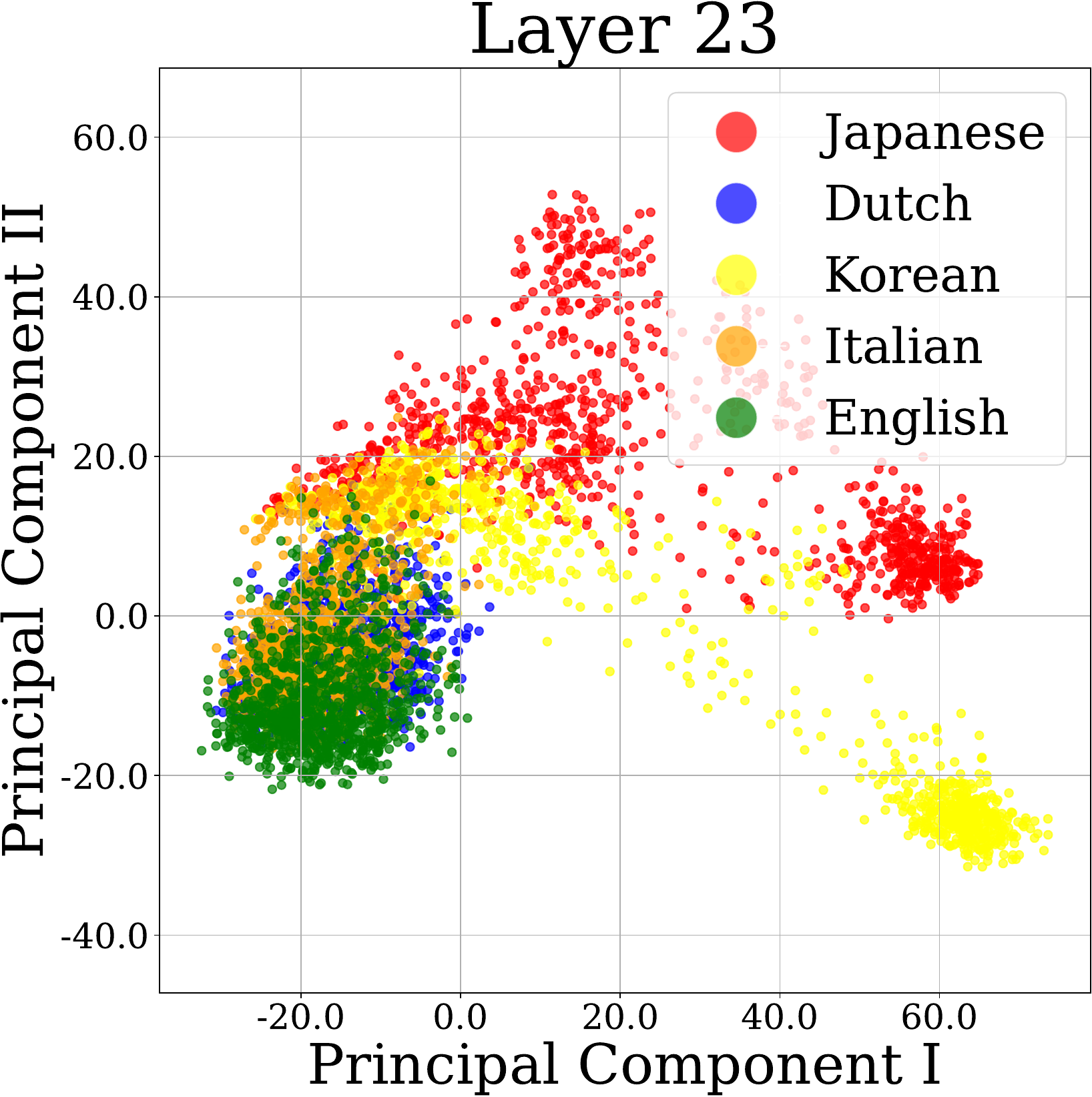}
  \includegraphics[width=0.19\linewidth]{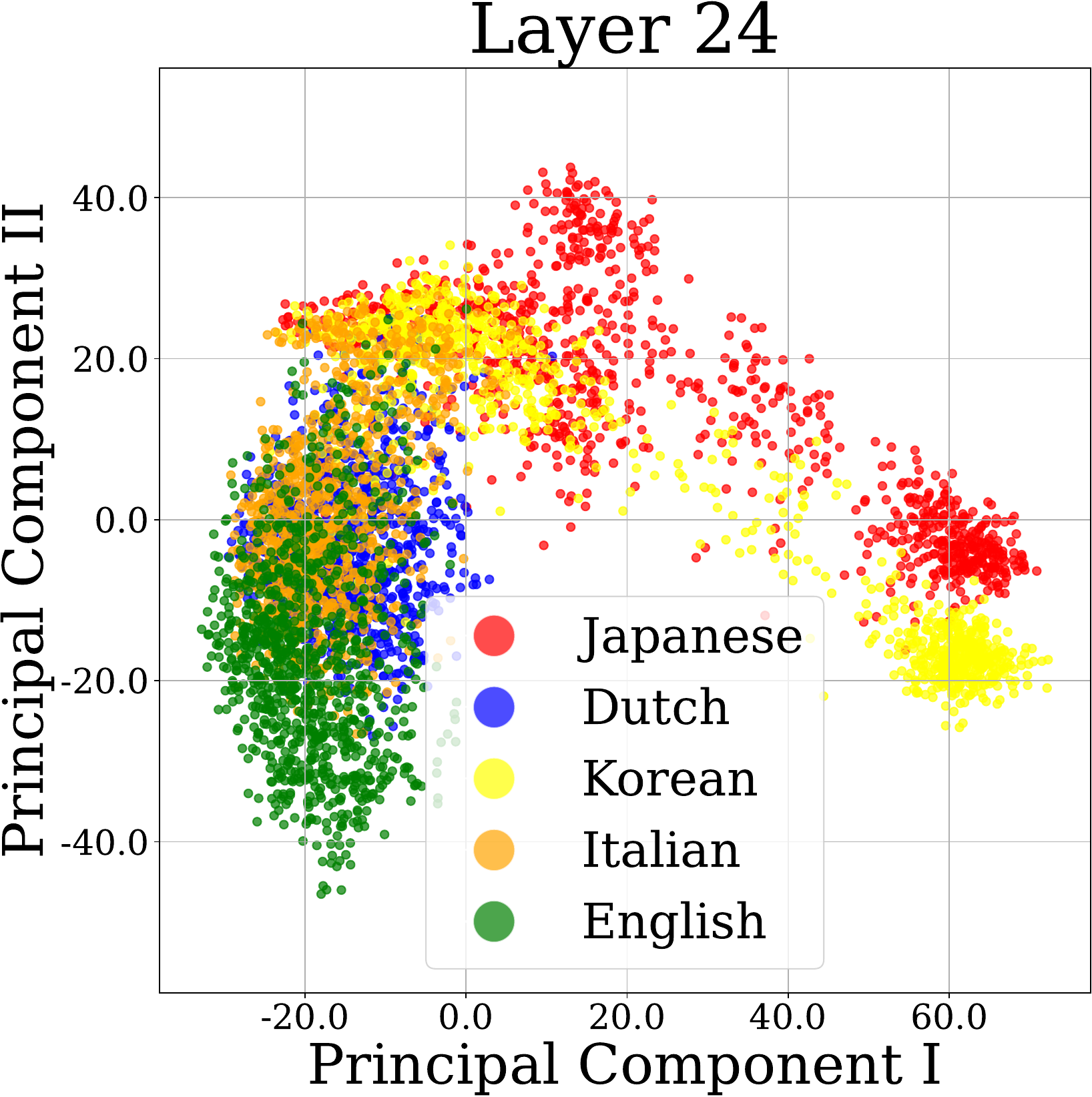}
  \includegraphics[width=0.19\linewidth]{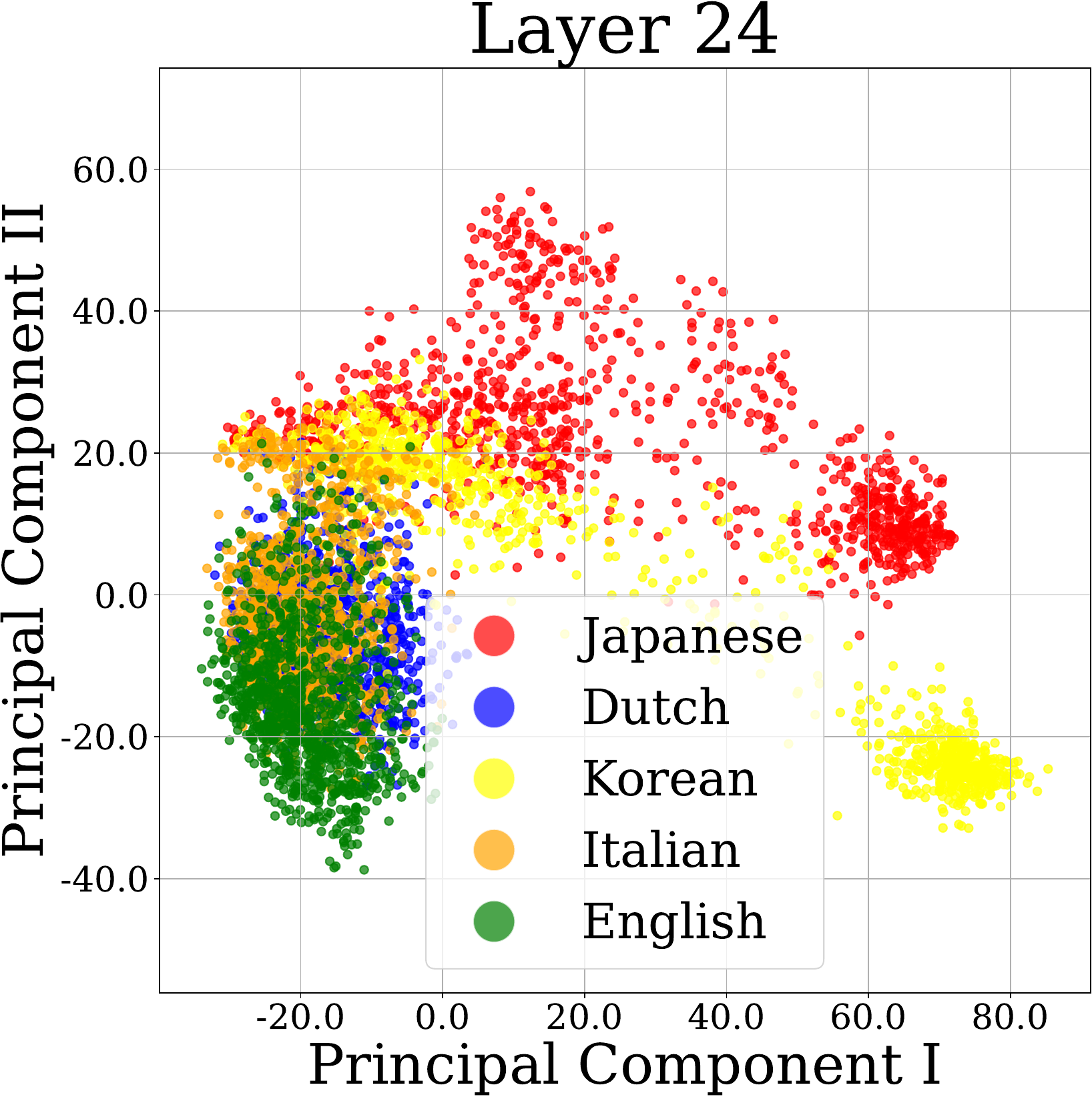}

  \begin{minipage}{0.19\linewidth}\centering \textbf{\textcolor{red}{layer 23 (Type-2)}}\end{minipage}
  \begin{minipage}{0.19\linewidth}\centering layer 23 (baseline)\end{minipage}
  \begin{minipage}{0.19\linewidth}\centering \textbf{\textcolor{red}{layer 24 (Type-2)}}\end{minipage}
  \begin{minipage}{0.19\linewidth}\centering layer 24 (baseline)\end{minipage}

  \includegraphics[width=0.19\linewidth]{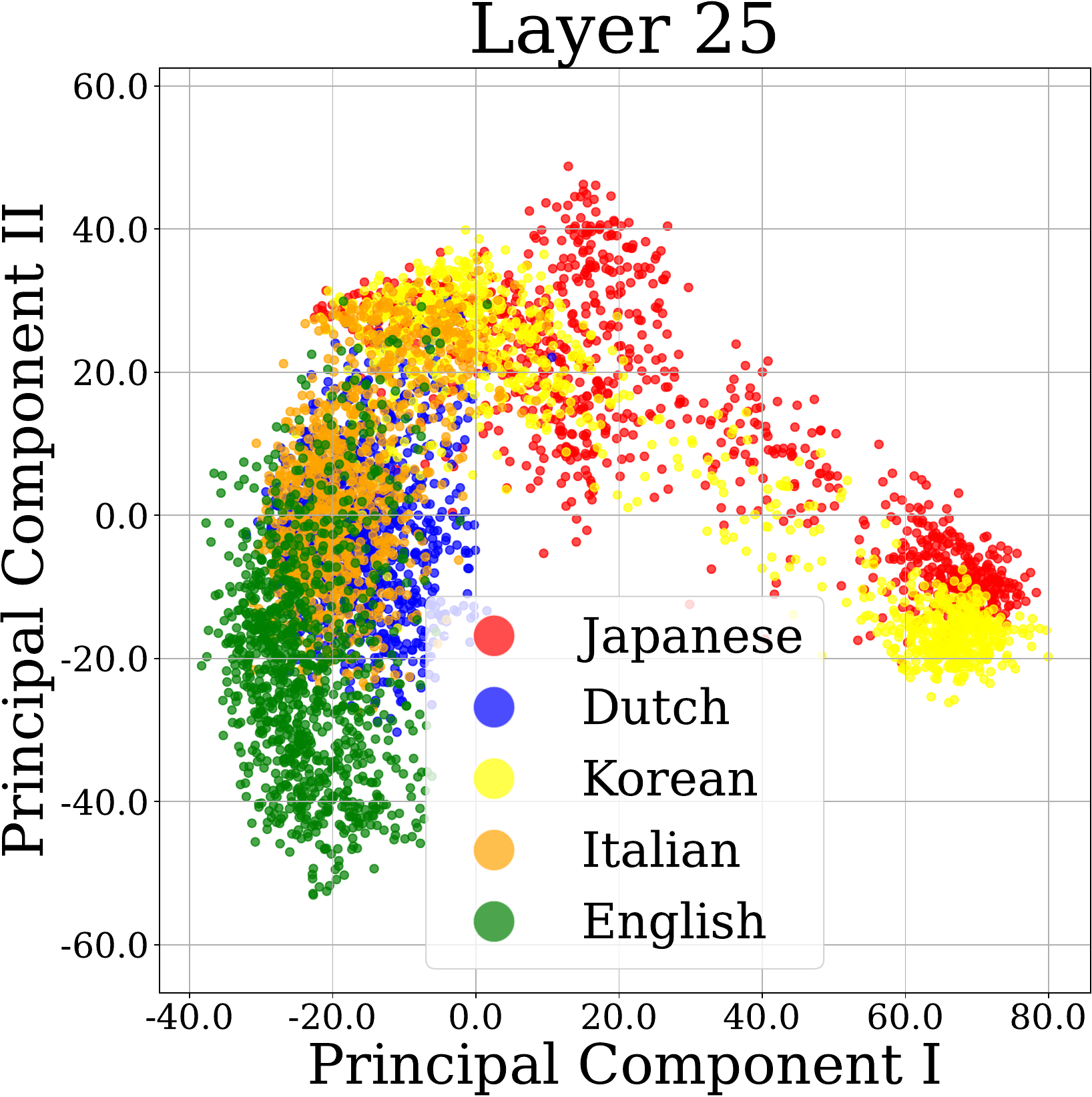}
  \includegraphics[width=0.19\linewidth]{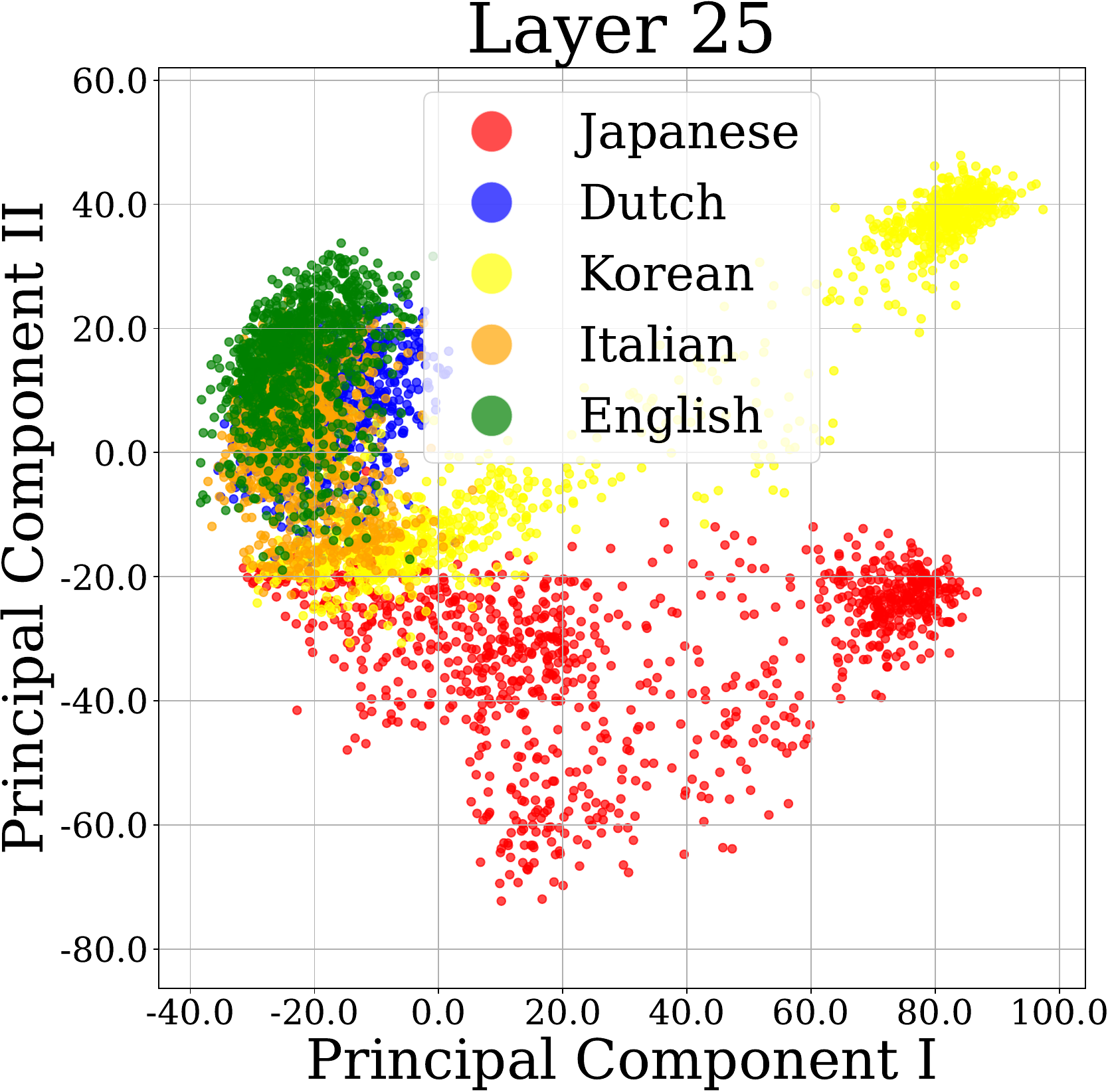}
  \includegraphics[width=0.19\linewidth]{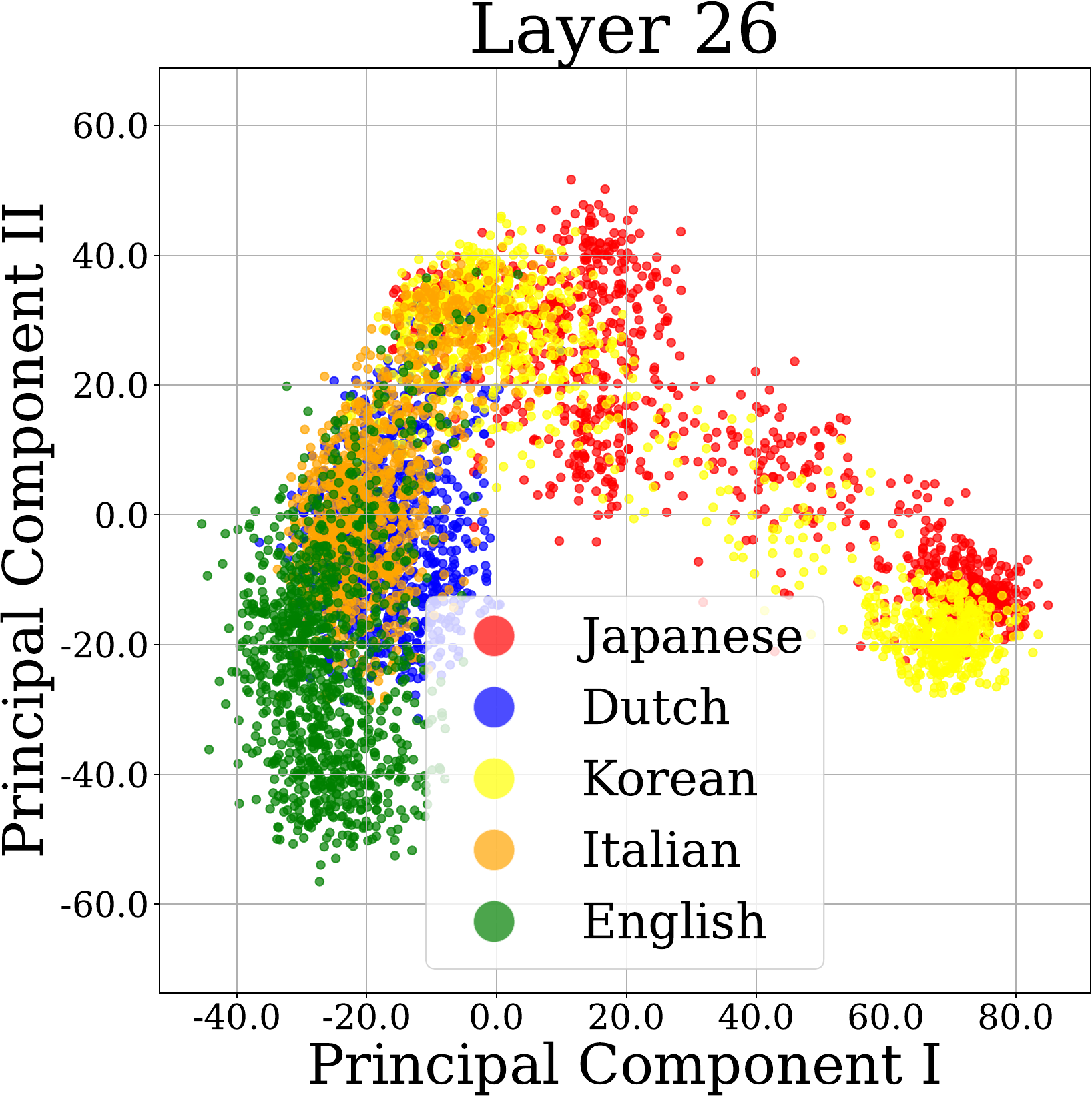}
  \includegraphics[width=0.19\linewidth]{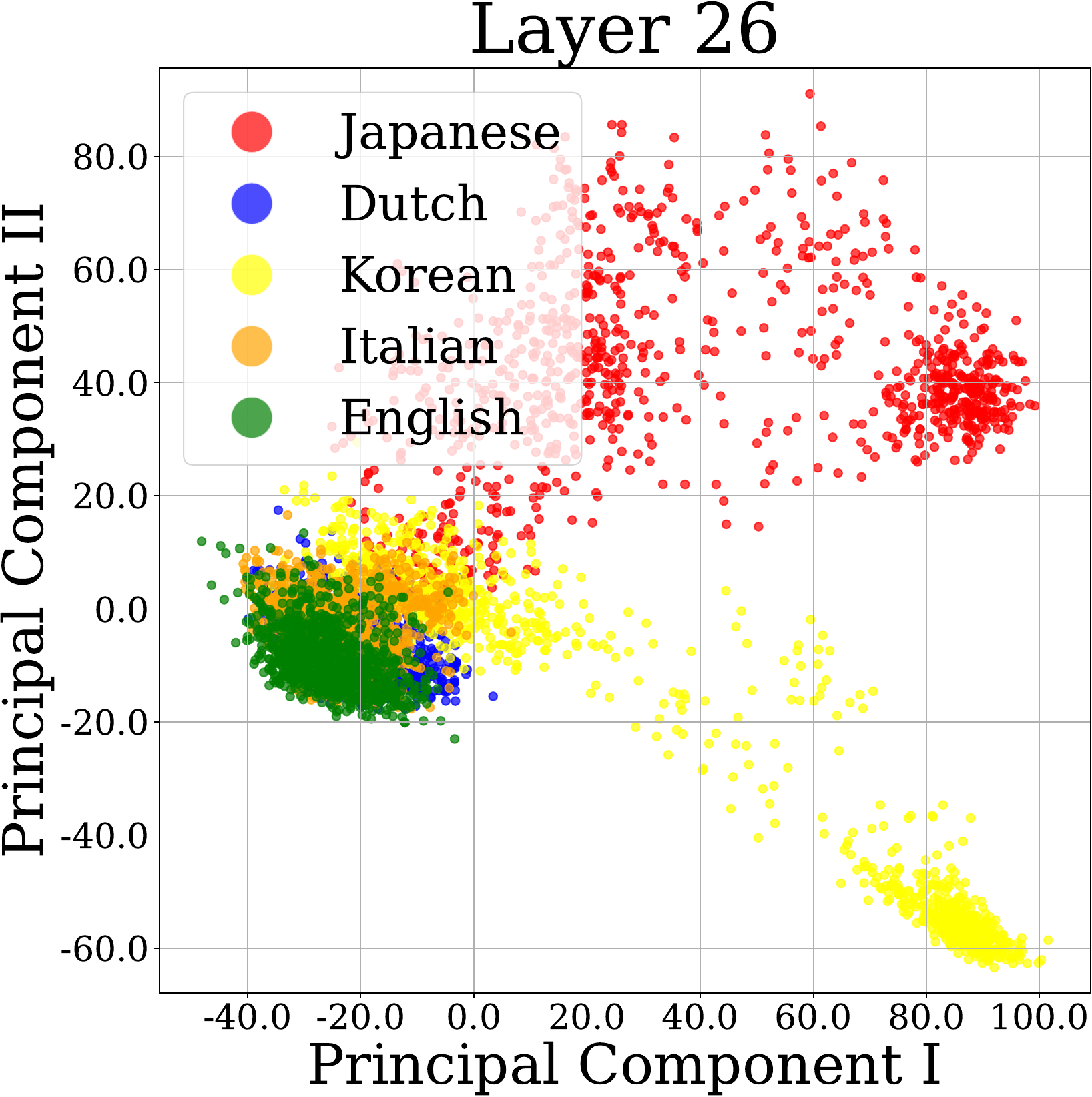}

  \begin{minipage}{0.19\linewidth}\centering \textbf{\textcolor{red}{layer 25 (Type-2)}}\end{minipage}
  \begin{minipage}{0.19\linewidth}\centering layer 25 (baseline)\end{minipage}
  \begin{minipage}{0.19\linewidth}\centering \textbf{\textcolor{red}{layer 26 (Type-2)}}\end{minipage}
  \begin{minipage}{0.19\linewidth}\centering layer 26 (baseline)\end{minipage}

  \includegraphics[width=0.19\linewidth]{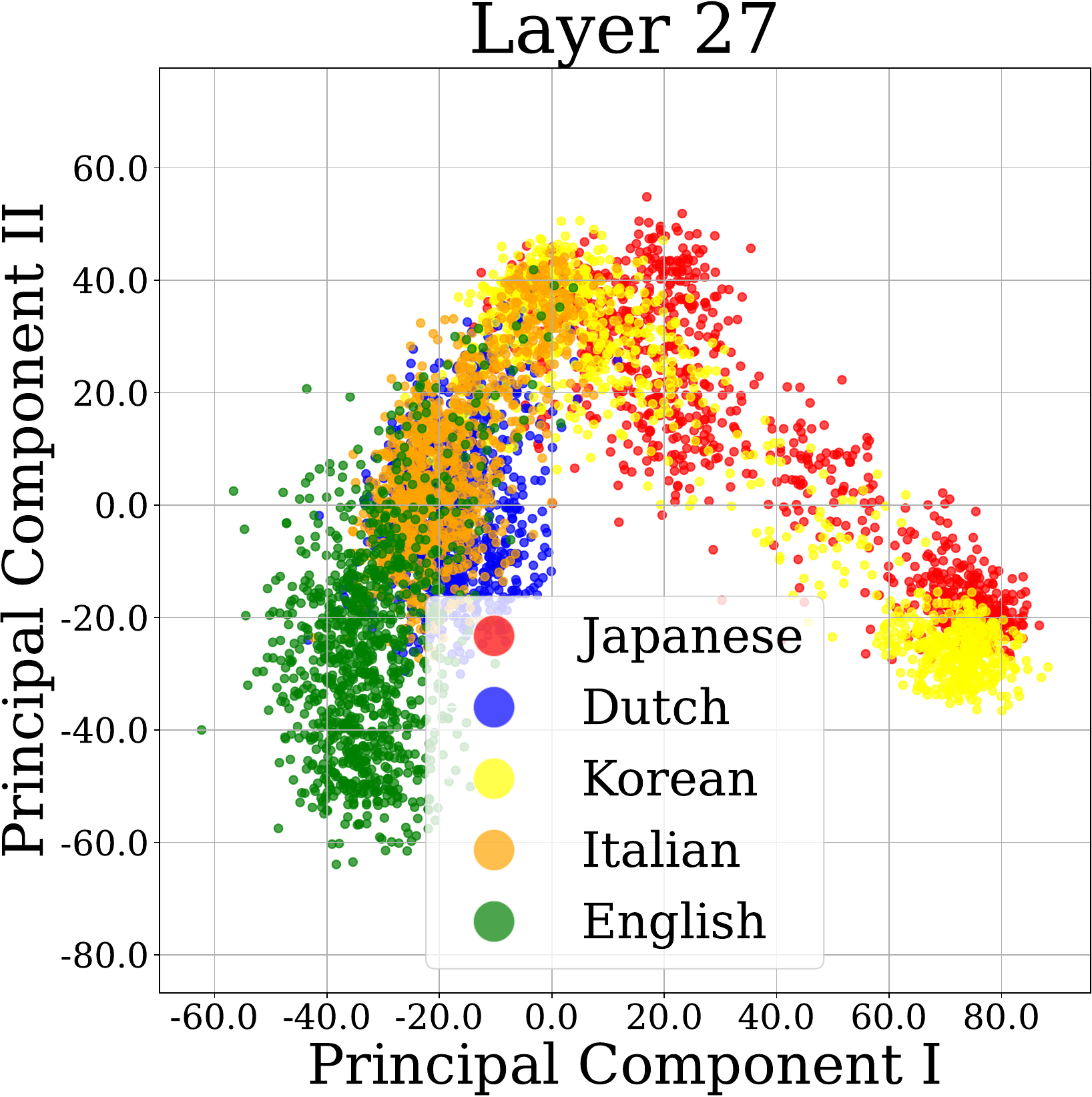}
  \includegraphics[width=0.19\linewidth]{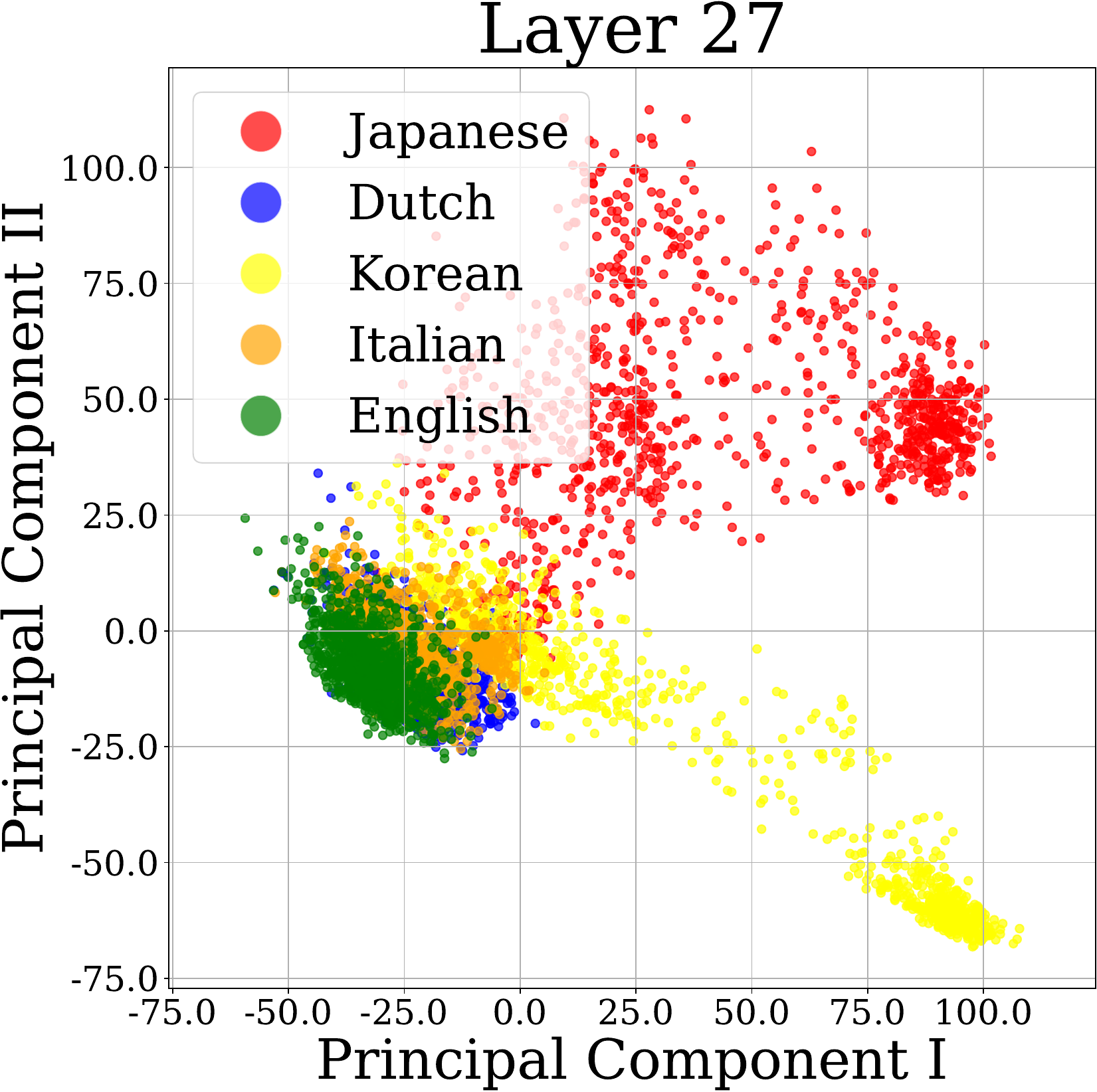}
  \includegraphics[width=0.19\linewidth]{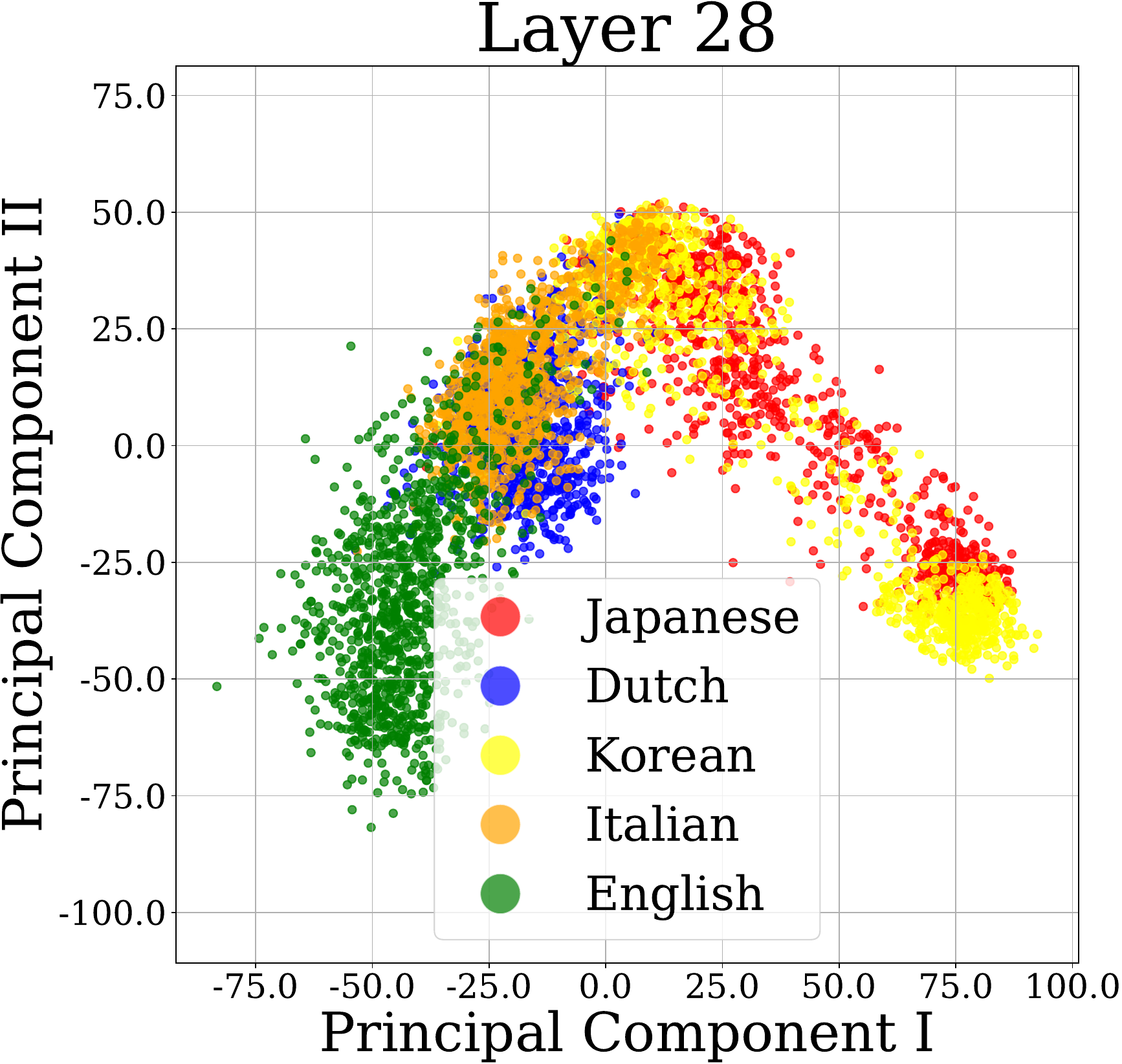}
  \includegraphics[width=0.19\linewidth]{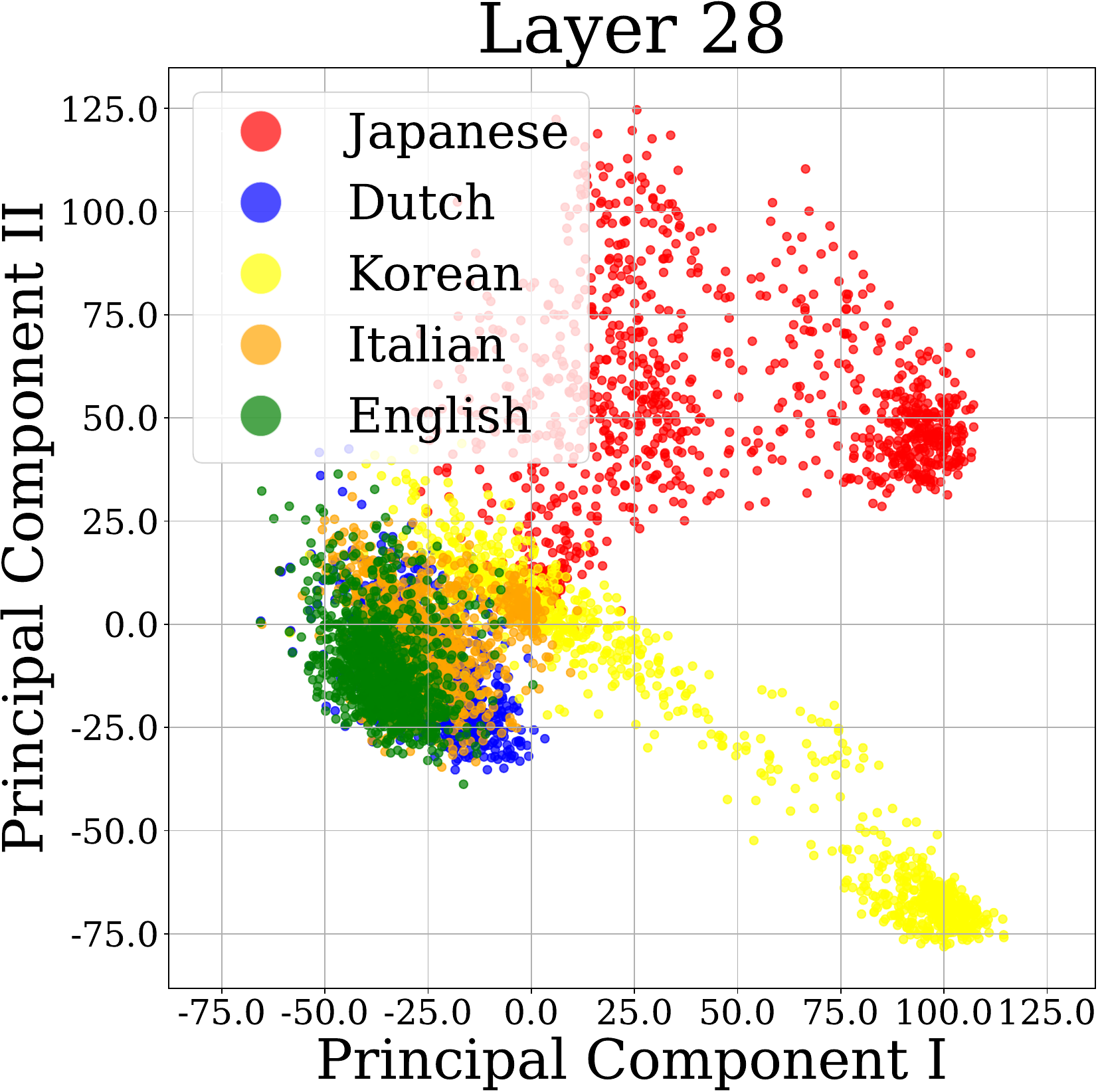}

  \begin{minipage}{0.19\linewidth}\centering \textbf{\textcolor{red}{layer 27 (Type-2)}}\end{minipage}
  \begin{minipage}{0.19\linewidth}\centering layer 27 (baseline)\end{minipage}
  \begin{minipage}{0.19\linewidth}\centering \textbf{\textcolor{red}{layer 28 (Type-2)}}\end{minipage}
  \begin{minipage}{0.19\linewidth}\centering layer 28 (baseline)\end{minipage}

  \includegraphics[width=0.19\linewidth]{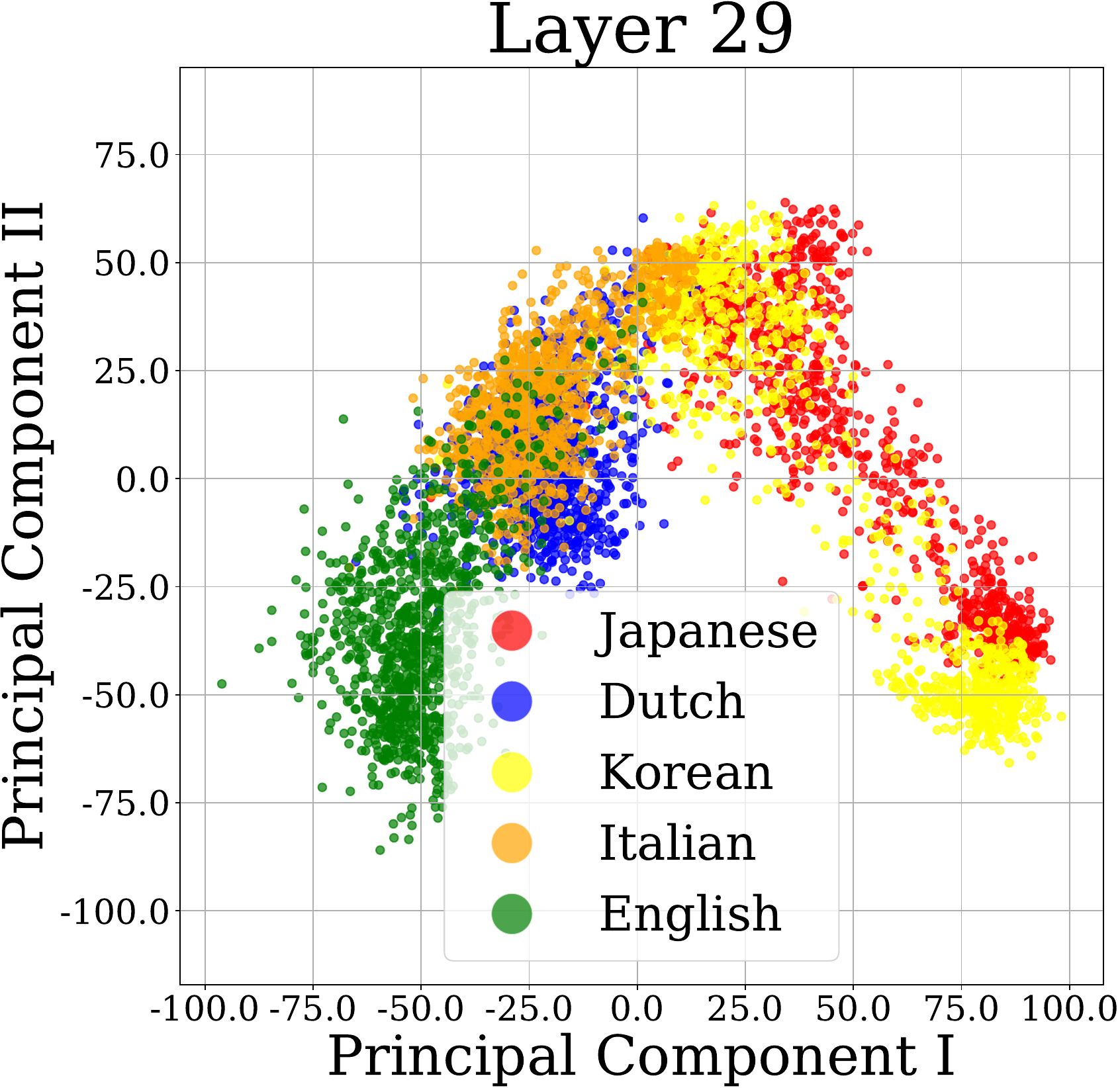}
  \includegraphics[width=0.19\linewidth]{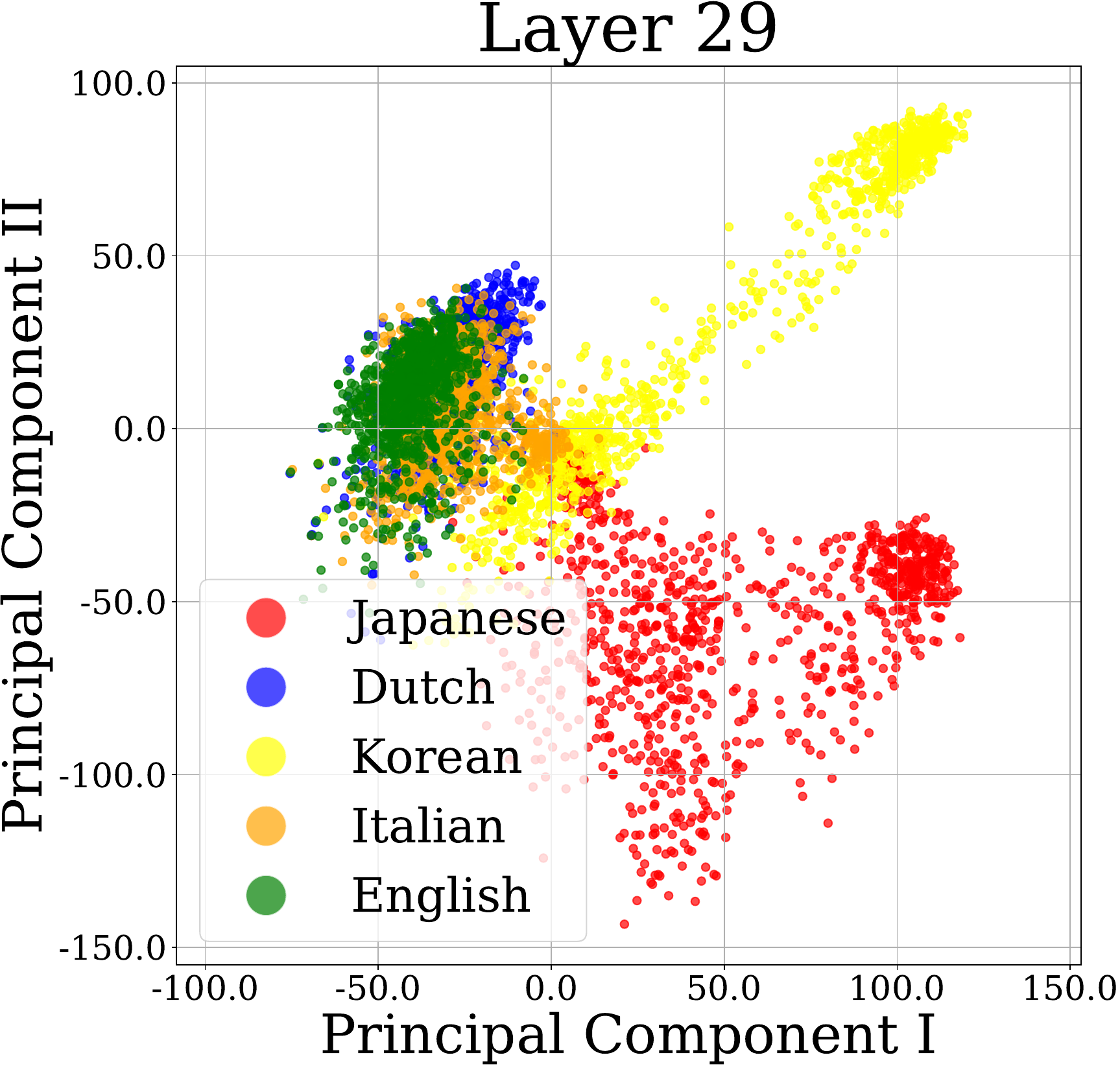}
  \includegraphics[width=0.19\linewidth]{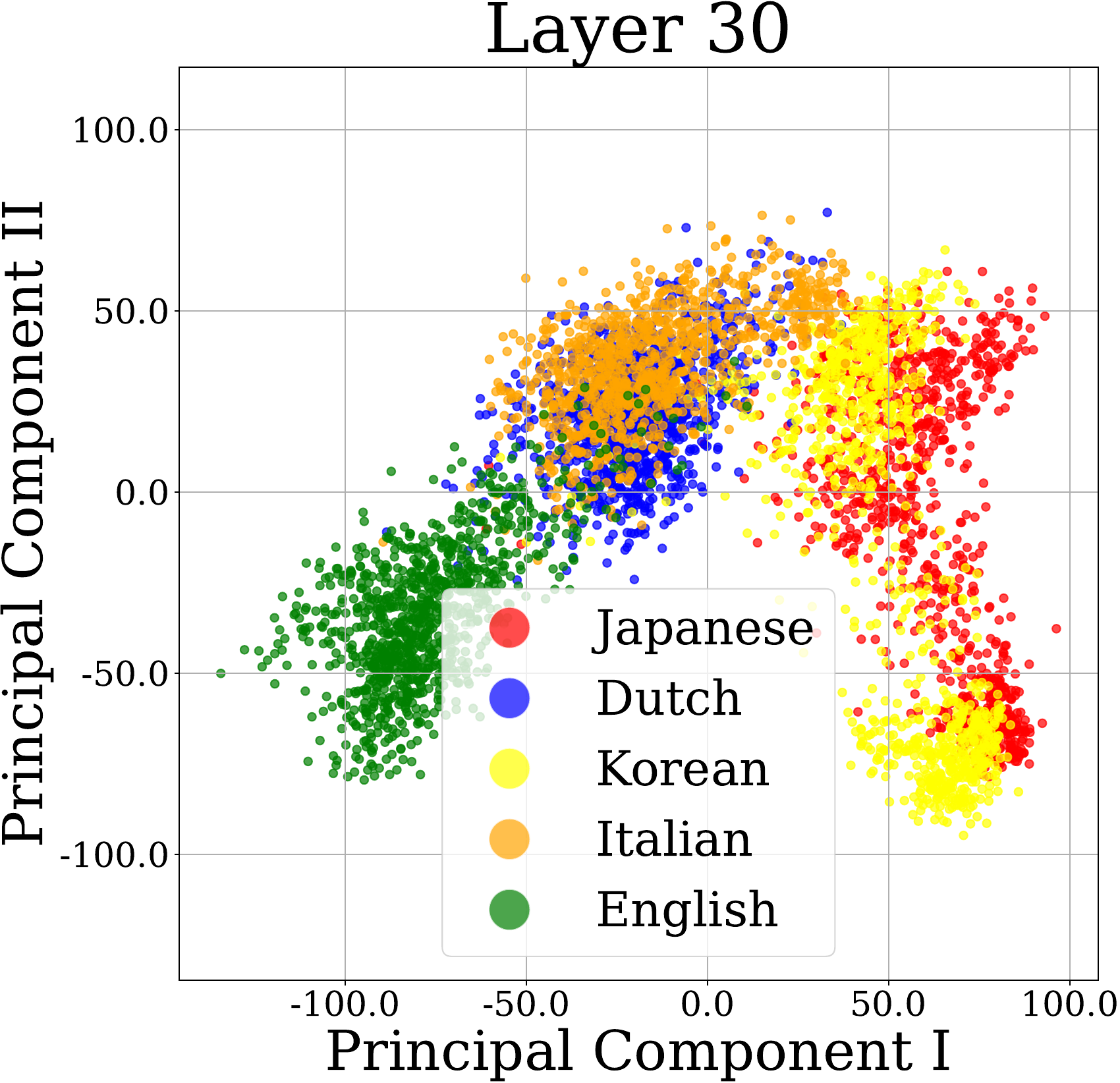}
  \includegraphics[width=0.19\linewidth]{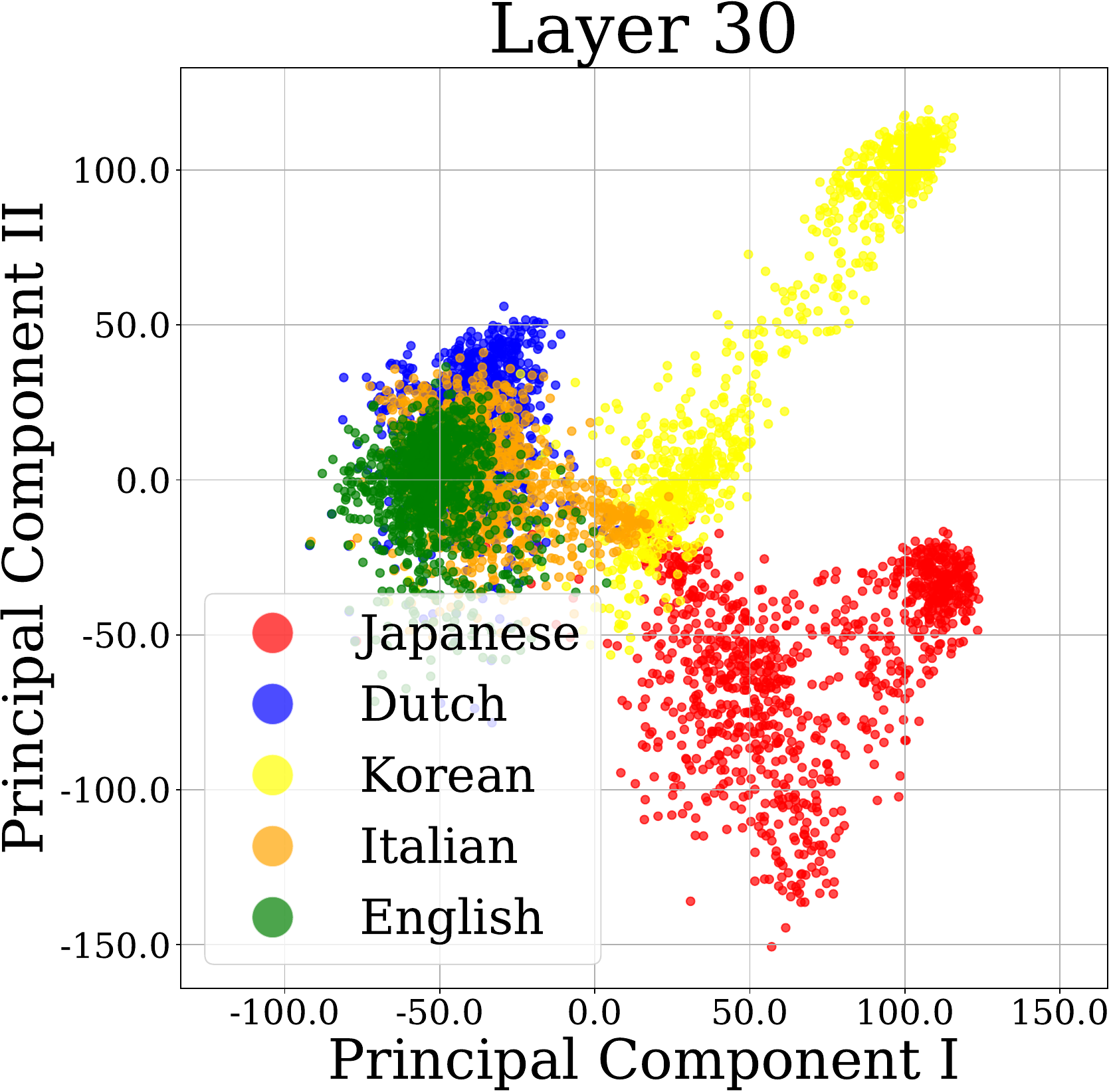}

  \begin{minipage}{0.19\linewidth}\centering \textbf{\textcolor{red}{layer 29 (Type-2)}}\end{minipage}
  \begin{minipage}{0.19\linewidth}\centering layer 29 (baseline)\end{minipage}
  \begin{minipage}{0.19\linewidth}\centering \textbf{\textcolor{red}{layer 30 (Type-2)}}\end{minipage}
  \begin{minipage}{0.19\linewidth}\centering layer 30 (baseline)\end{minipage}

  \includegraphics[width=0.19\linewidth]{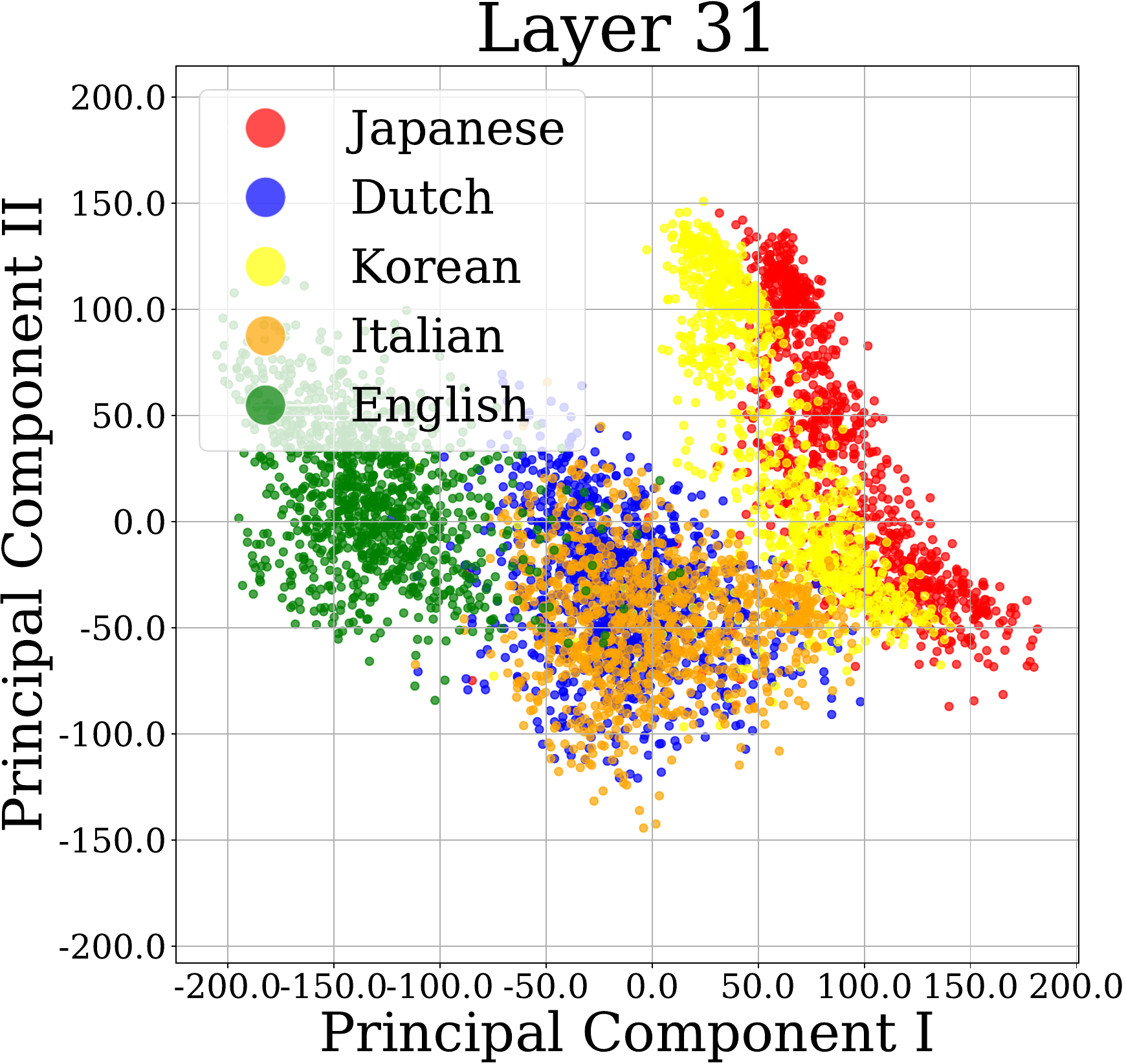}
  \includegraphics[width=0.19\linewidth]{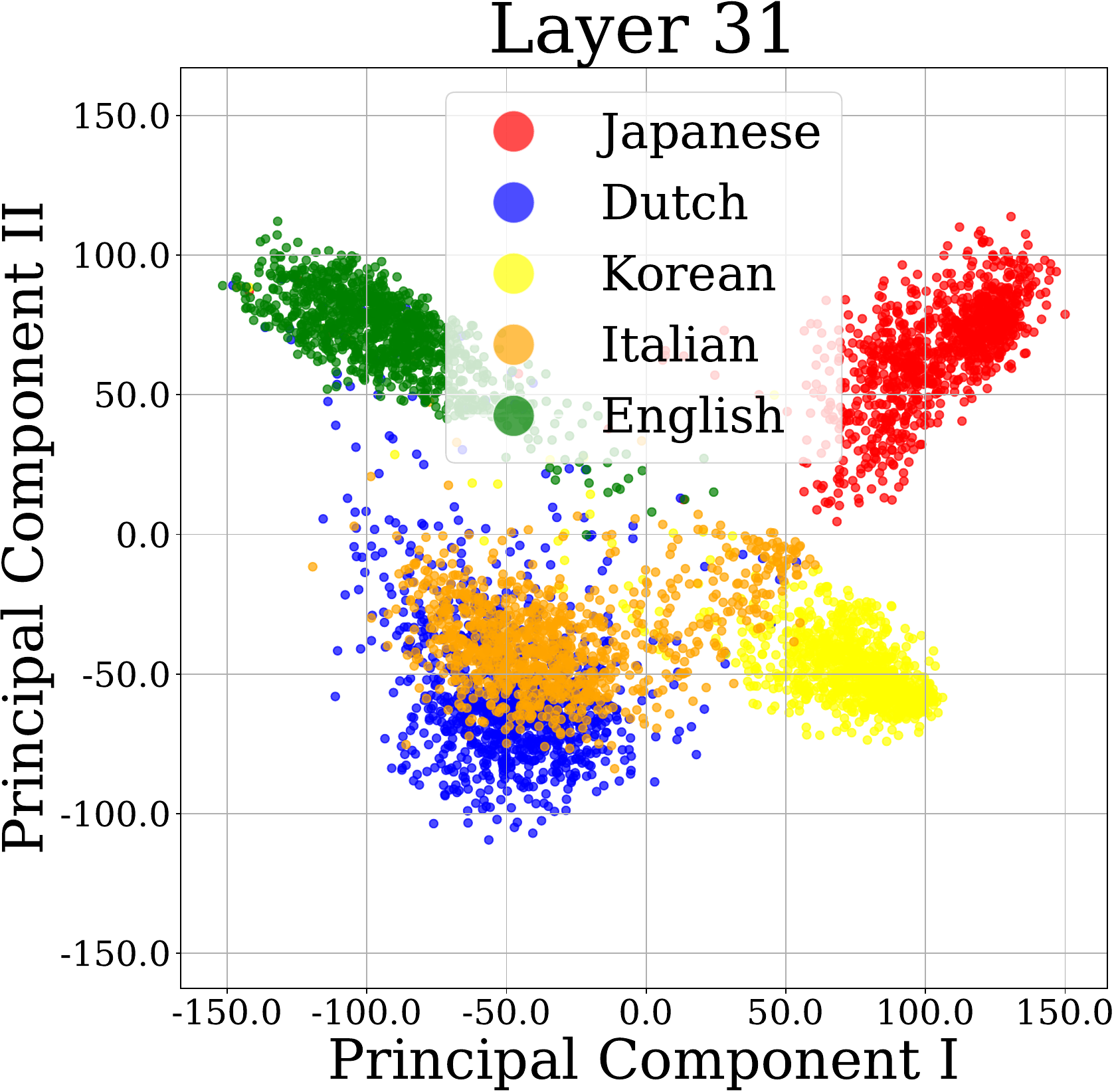}
  \includegraphics[width=0.19\linewidth]{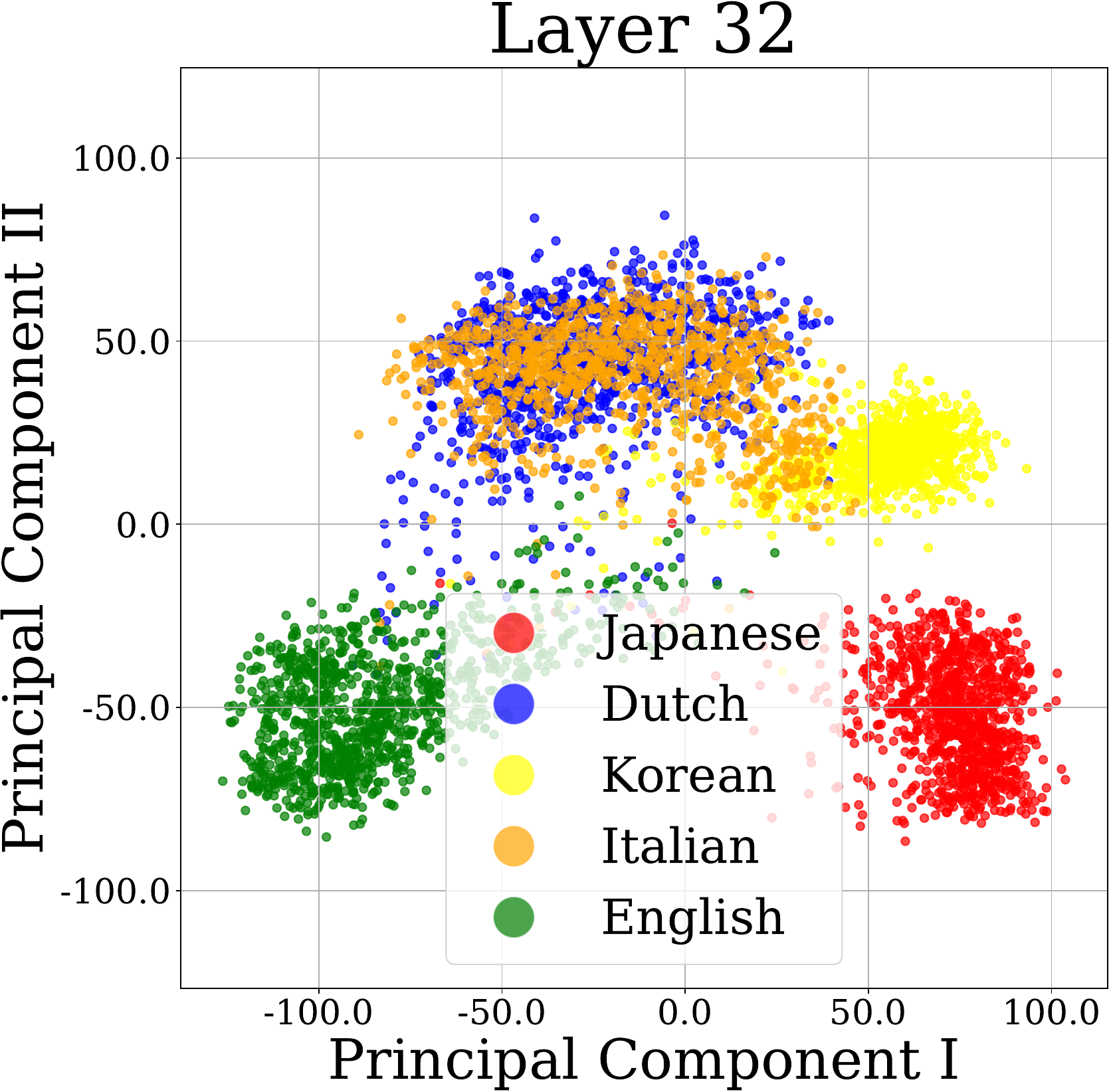}
  \includegraphics[width=0.19\linewidth]{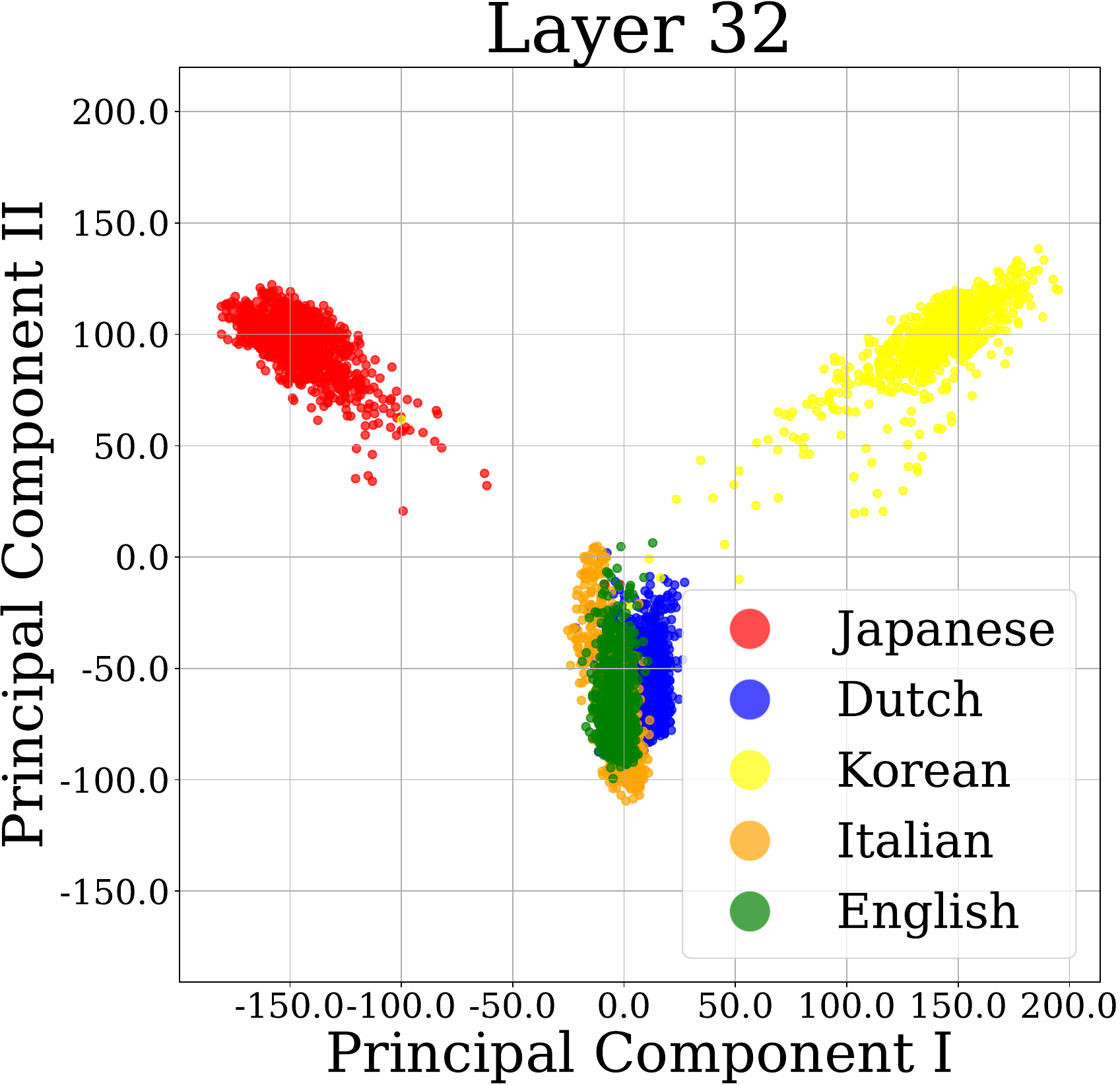}

  \begin{minipage}{0.19\linewidth}\centering \textbf{\textcolor{red}{layer 31 (Type-2)}}\end{minipage}
  \begin{minipage}{0.19\linewidth}\centering layer 31 (baseline)\end{minipage}
  \begin{minipage}{0.19\linewidth}\centering \textbf{\textcolor{red}{layer 32 (Type-2)}}\end{minipage}
  \begin{minipage}{0.19\linewidth}\centering layer 32 (baseline)\end{minipage}

  \caption{\textbf{The resutls of PCA while deactivating Top-1k Type-2 Transfer Neurons (Aya expanse-8B)}.\\.}
  \label{fig:appendix:pca_deactivating_type2_aya}
\end{figure*}

Fig.~\ref{fig:appendix:pca_deactivating_type2_llama3},~\ref{fig:appendix:pca_deactivating_type2_mistral}, and, ~\ref{fig:appendix:pca_deactivating_type2_aya} show the results of PCA applied to hidden representations specific to each language, while deactivating top-1k Type-2 neurons (\textcolor[HTML]{228B22}{English} features are the only ones visualized without intervention). As shown, although only 0.2\% of all neurons in the model were deactivated, the deactivation caused a significant delay in the movement of representations towards each language-specific latent space — especially in the final layers where most Type-2 neurons reside — compared to the baseline (i.e., deactivating the 1k neurons randomly sampled from the same layers as the Type-2 neurons). These results support the functional impact of the detected Type-2 neurons.

% PCA results, deactivating Type-2, llama3
% PCA results, deactivating type2, mistral
% PCA results, deactivating type2, aya

\subsubsection{Quantitative Distance among Language Latent Spaces}
\label{sec:appendix:quantitative distance among language subspaces while deactivating Type-2}

\begin{figure*}[t]
  \centering

  \includegraphics[width=0.19\linewidth]{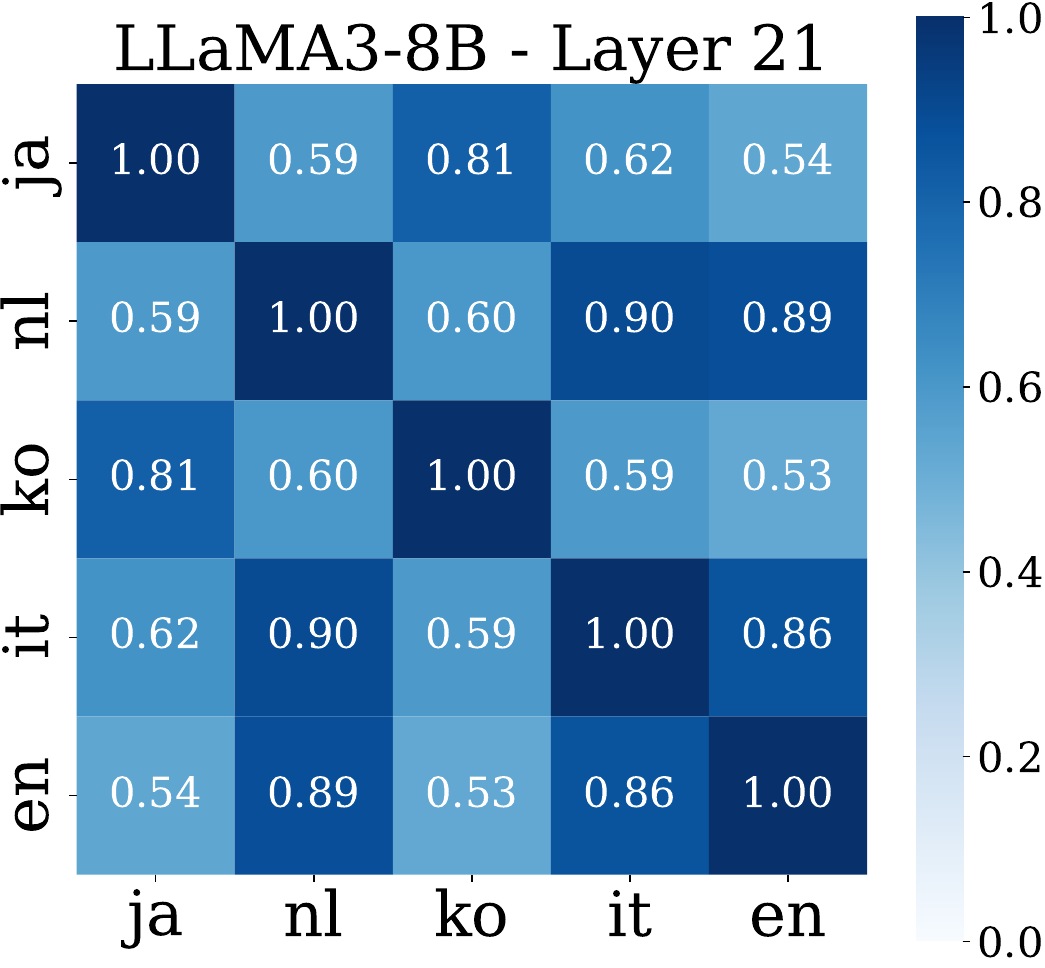}
  \includegraphics[width=0.19\linewidth]{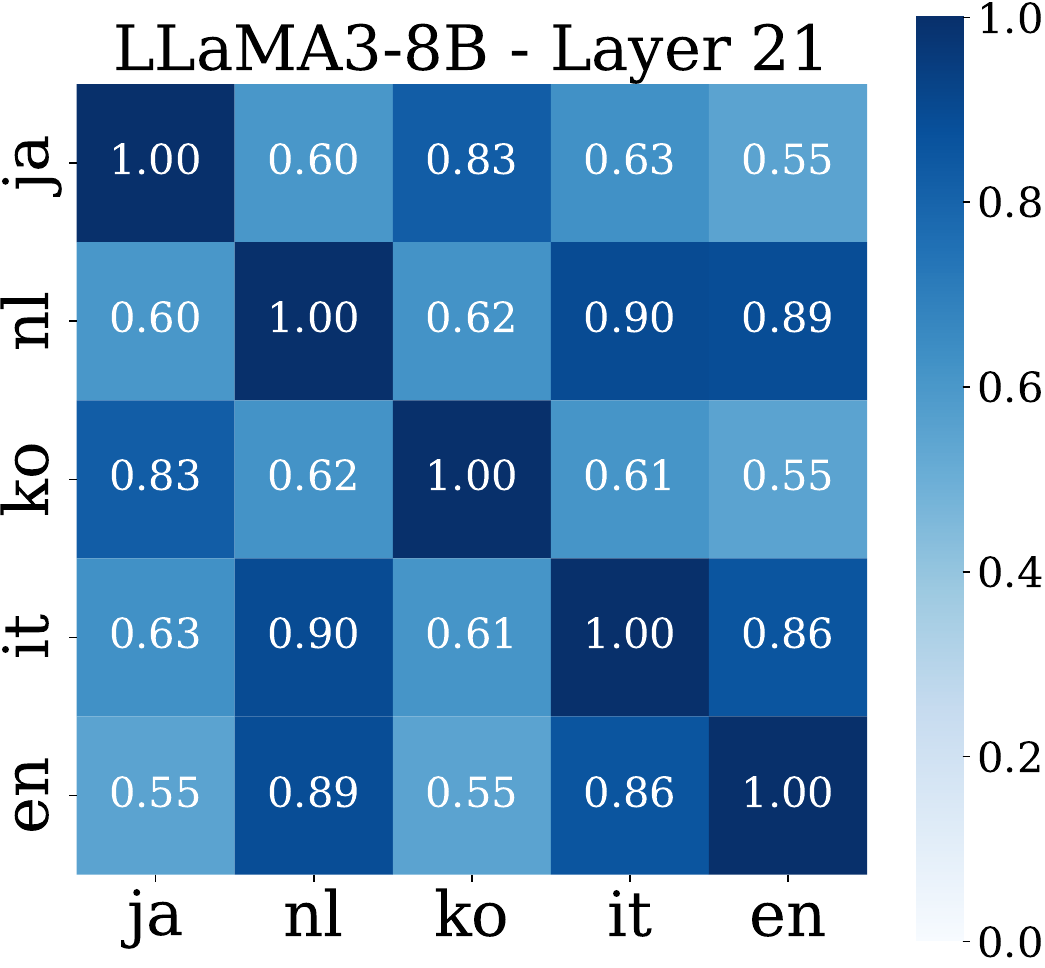}
  \includegraphics[width=0.19\linewidth]{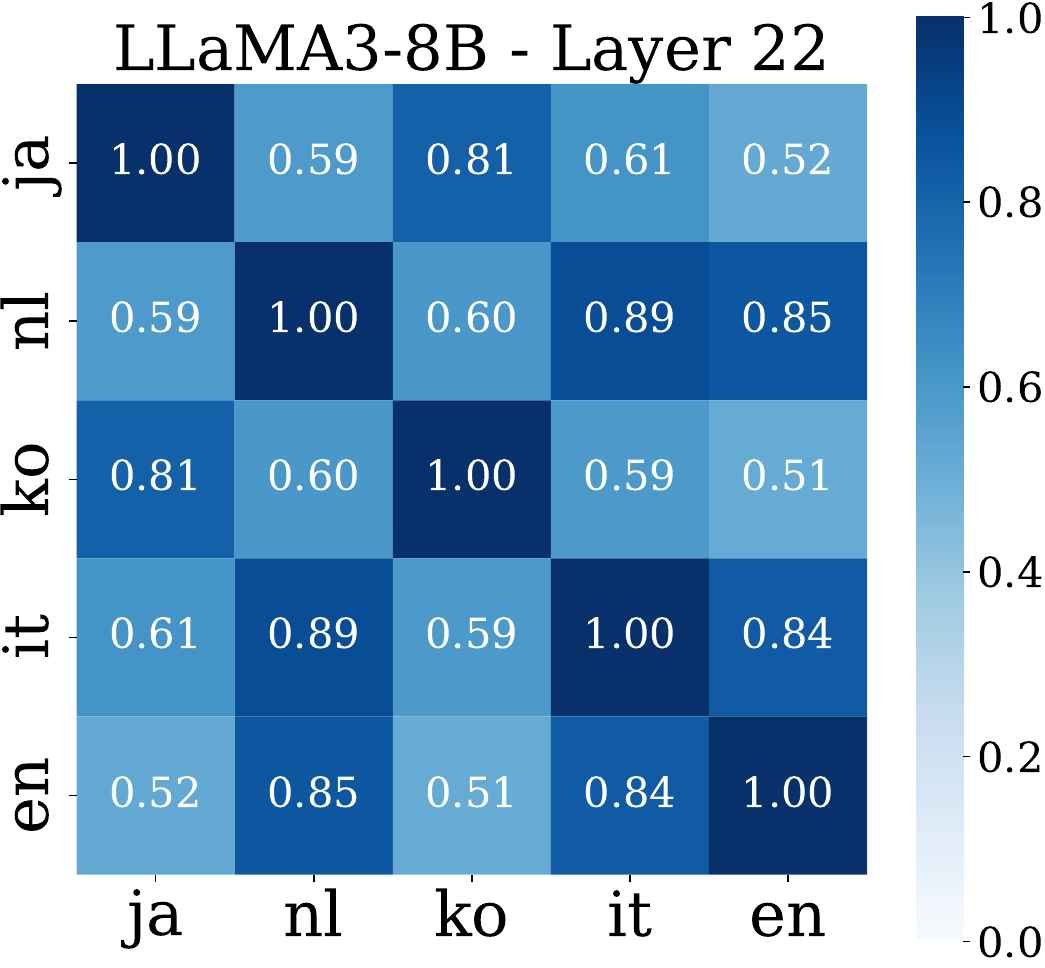}
  \includegraphics[width=0.19\linewidth]{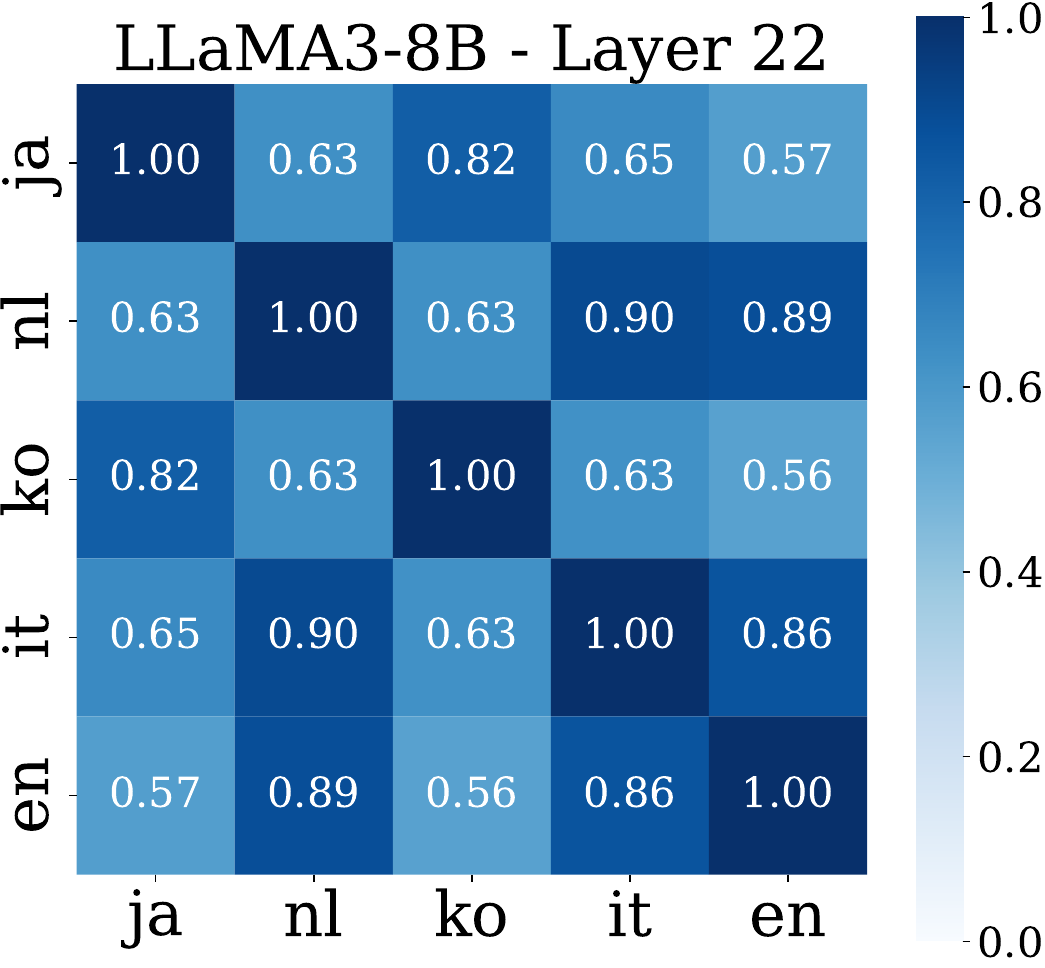}

  \begin{minipage}{0.19\linewidth}\centering \textbf{\textcolor{red}{layer 21 (Type-2)}}\end{minipage}
  \begin{minipage}{0.19\linewidth}\centering layer 21 (baseline)\end{minipage}
  \begin{minipage}{0.19\linewidth}\centering \textbf{\textcolor{red}{layer 22 (Type-2)}}\end{minipage}
  \begin{minipage}{0.19\linewidth}\centering layer 22 (baseline)\end{minipage}

  \includegraphics[width=0.19\linewidth]{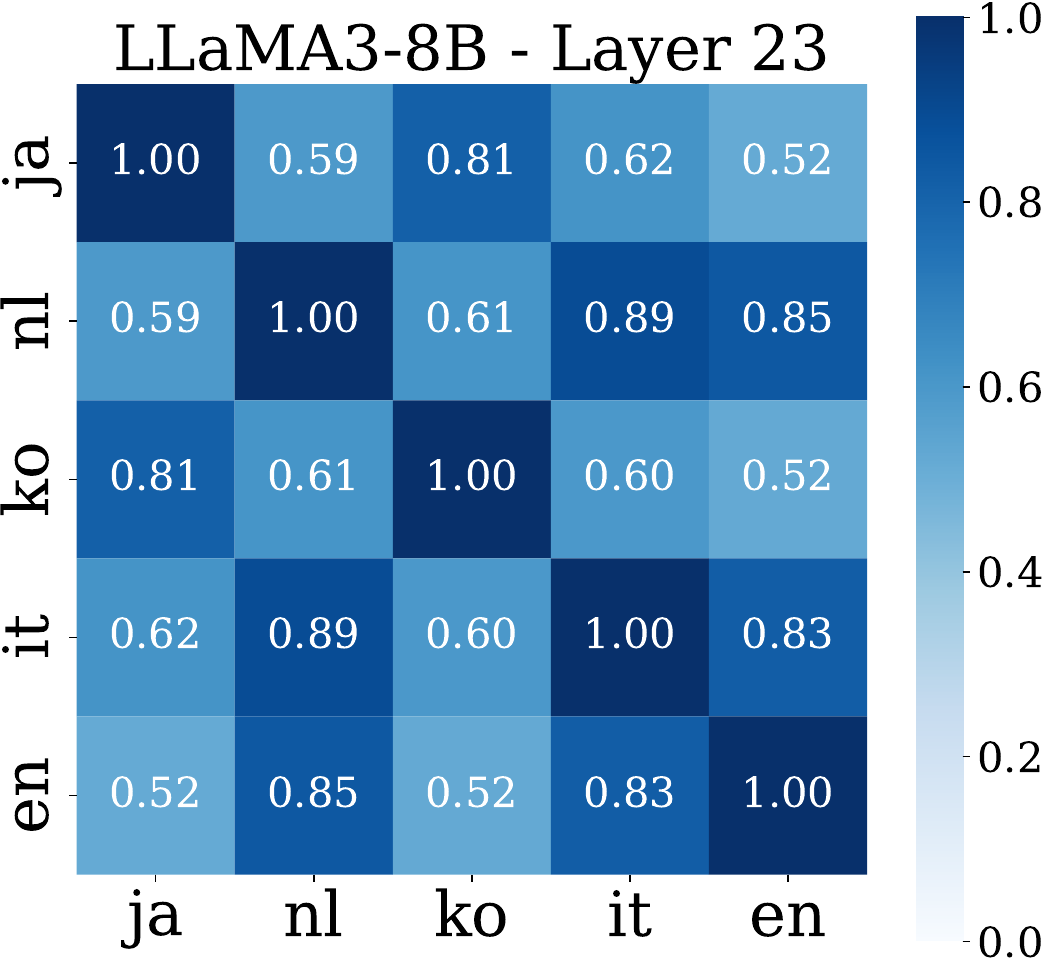}
  \includegraphics[width=0.19\linewidth]{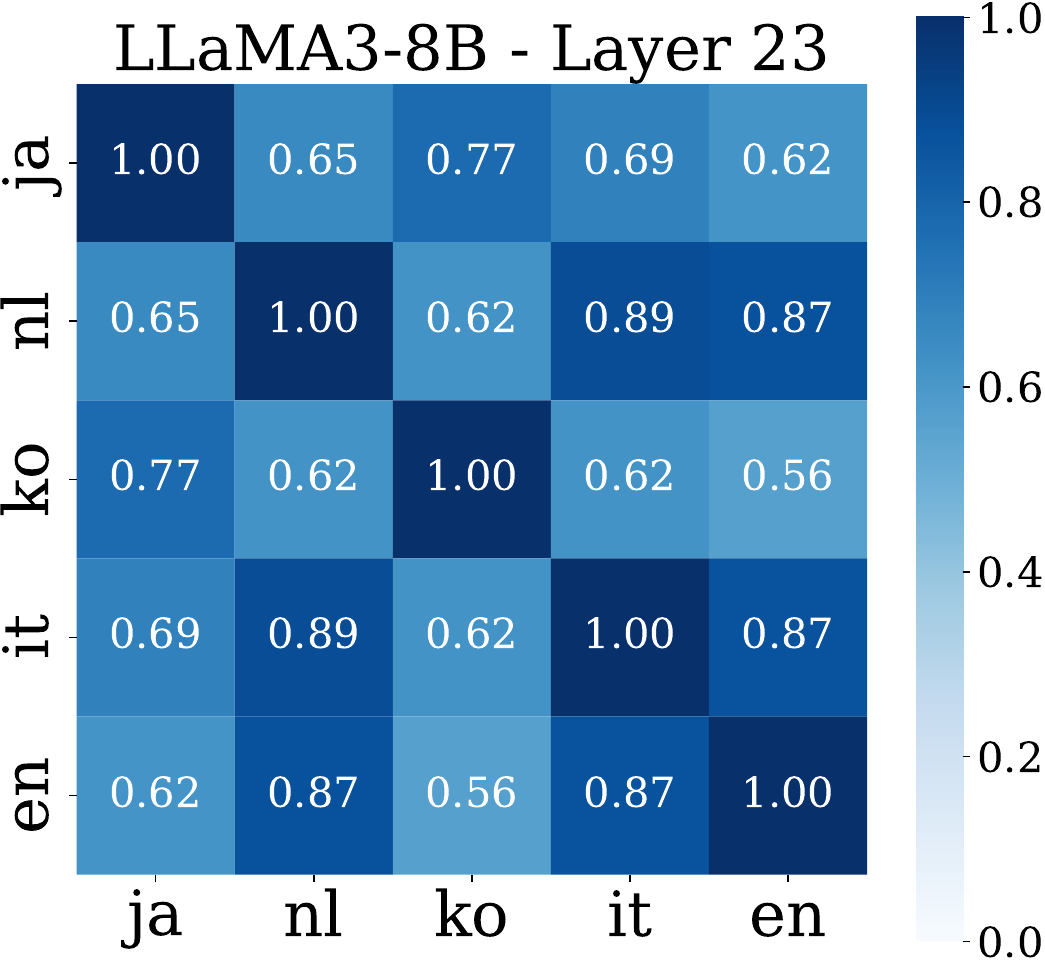}
  \includegraphics[width=0.19\linewidth]{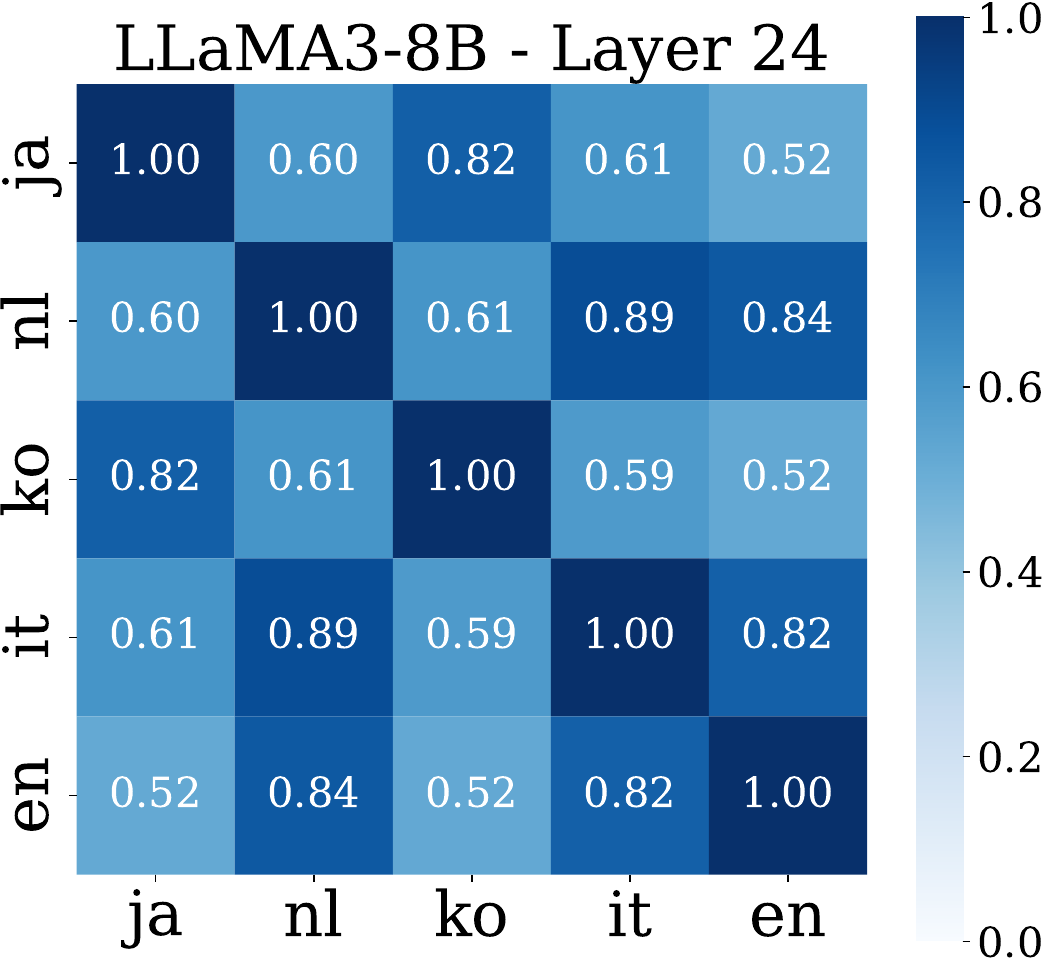}
  \includegraphics[width=0.19\linewidth]{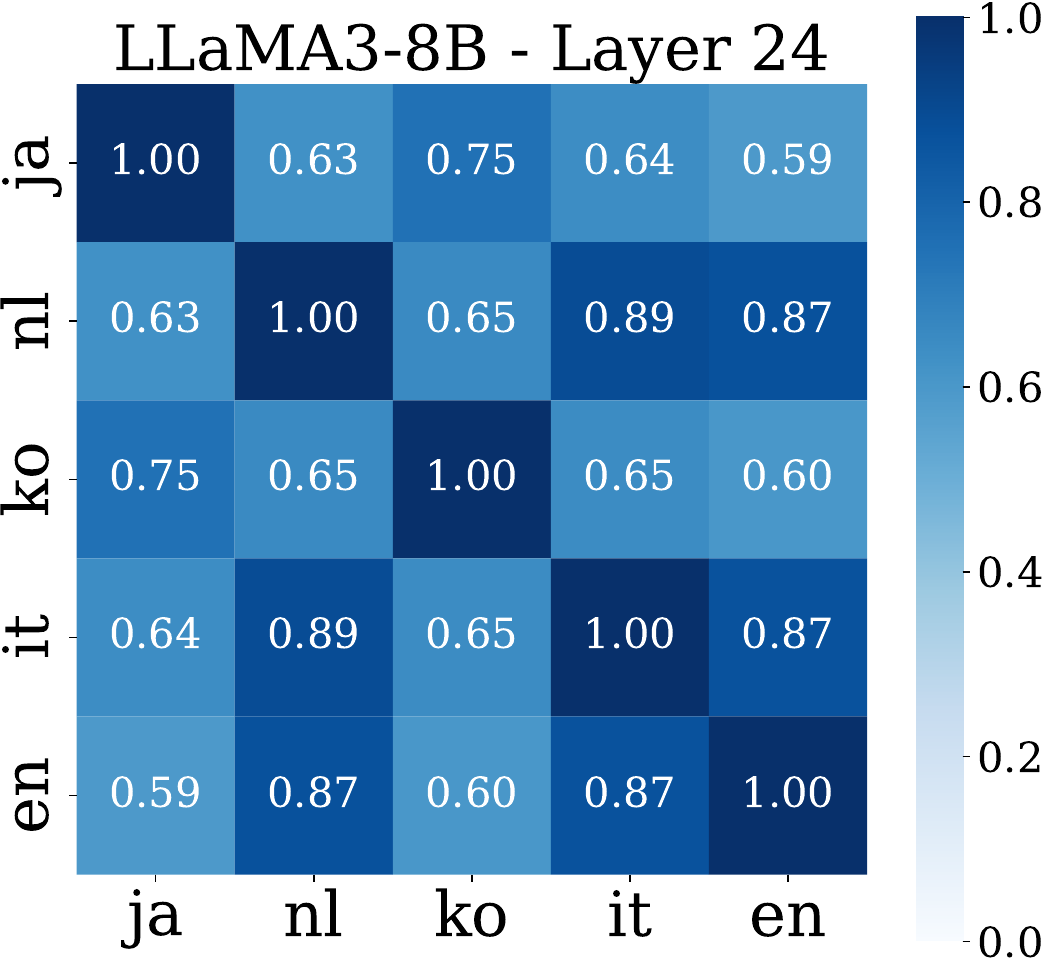}

  \begin{minipage}{0.19\linewidth}\centering \textbf{\textcolor{red}{layer 23 (Type-2)}}\end{minipage}
  \begin{minipage}{0.19\linewidth}\centering layer 23 (baseline)\end{minipage}
  \begin{minipage}{0.19\linewidth}\centering \textbf{\textcolor{red}{layer 24 (Type-2)}}\end{minipage}
  \begin{minipage}{0.19\linewidth}\centering layer 24 (baseline)\end{minipage}

  \includegraphics[width=0.19\linewidth]{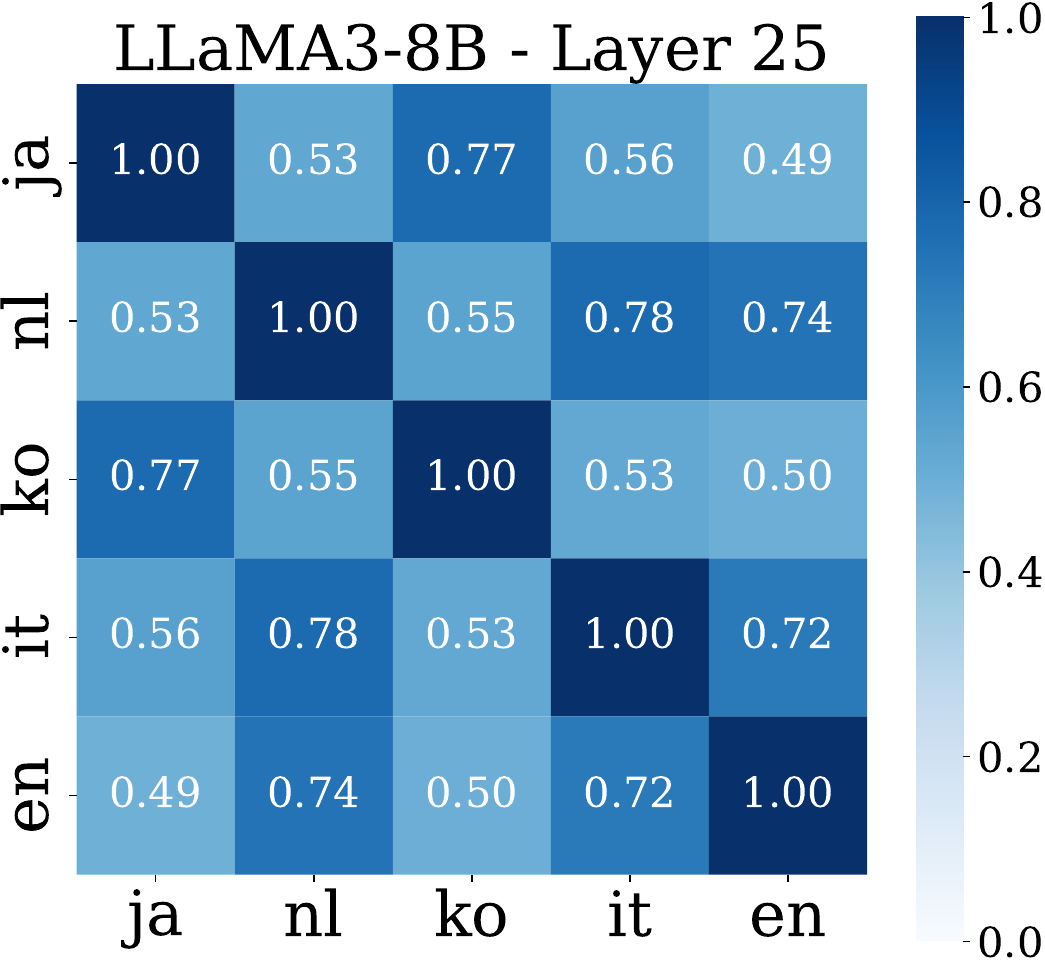}
  \includegraphics[width=0.19\linewidth]{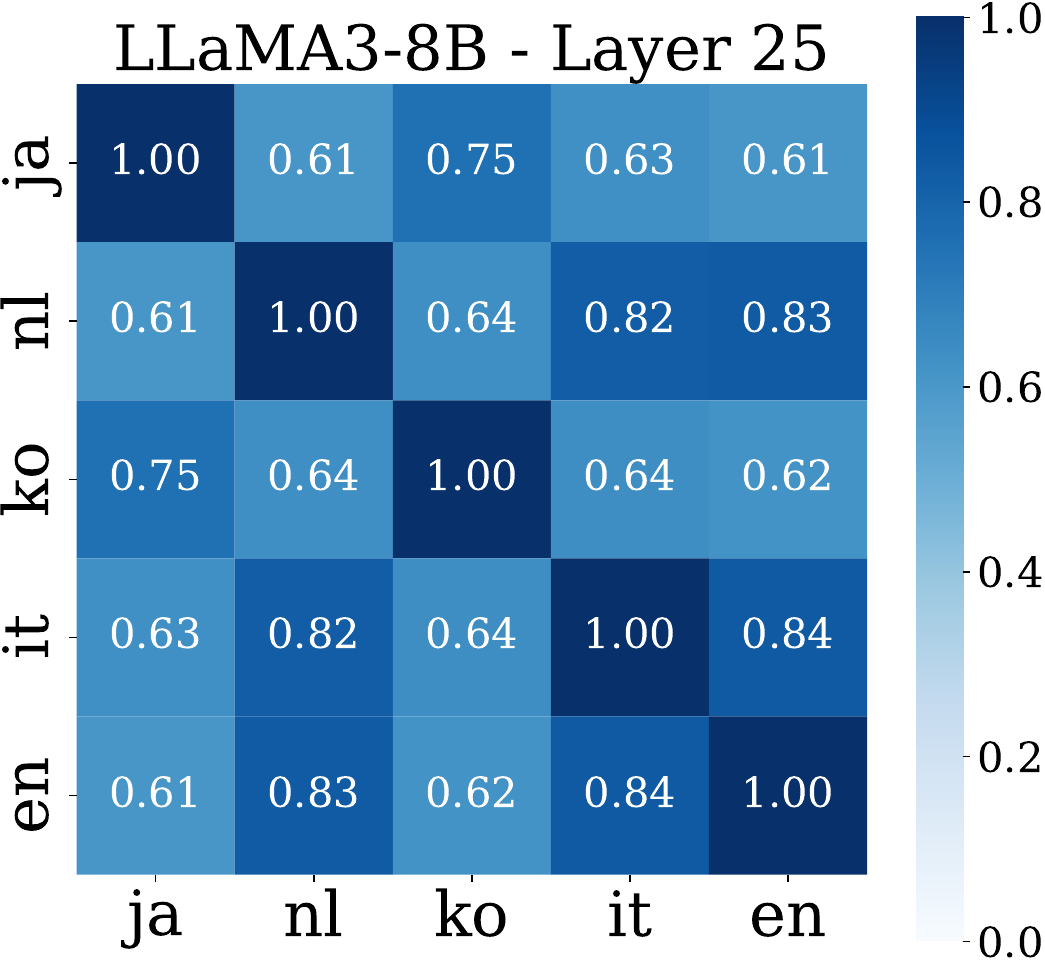}
  \includegraphics[width=0.19\linewidth]{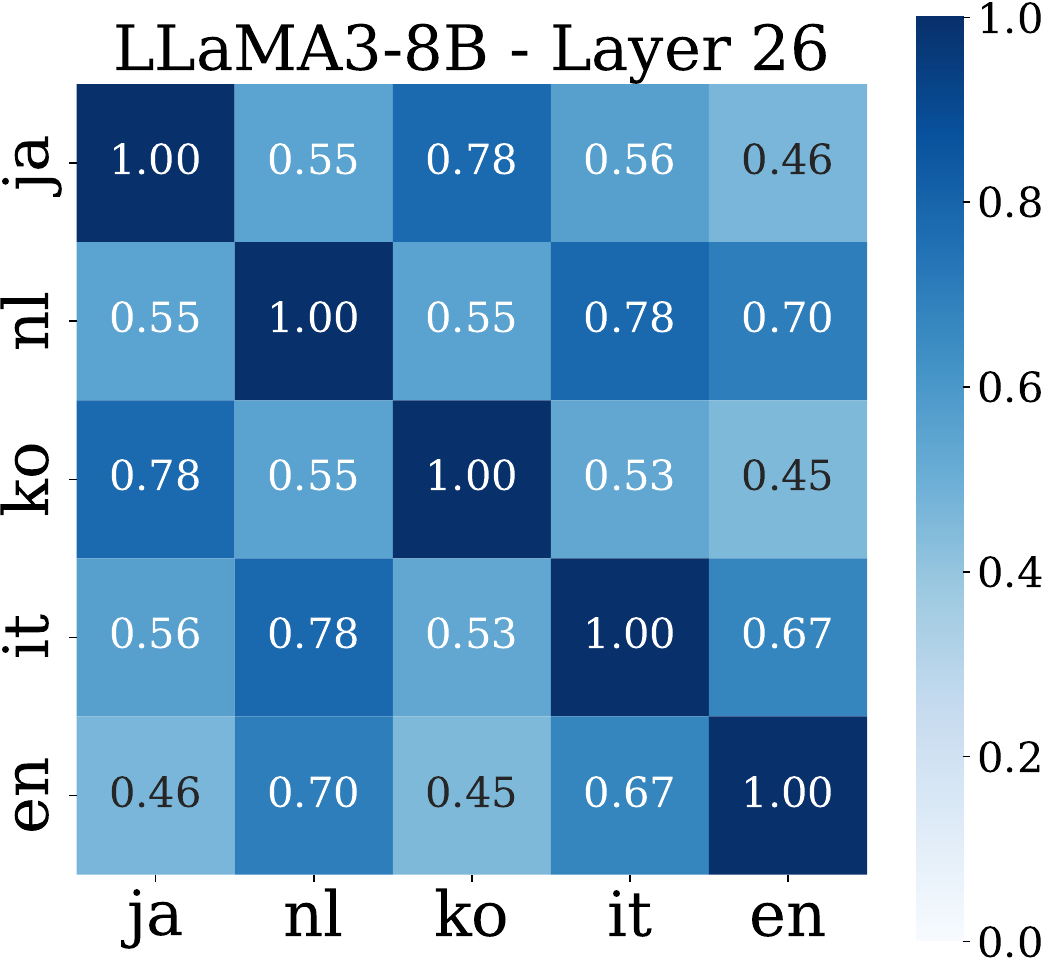}
  \includegraphics[width=0.19\linewidth]{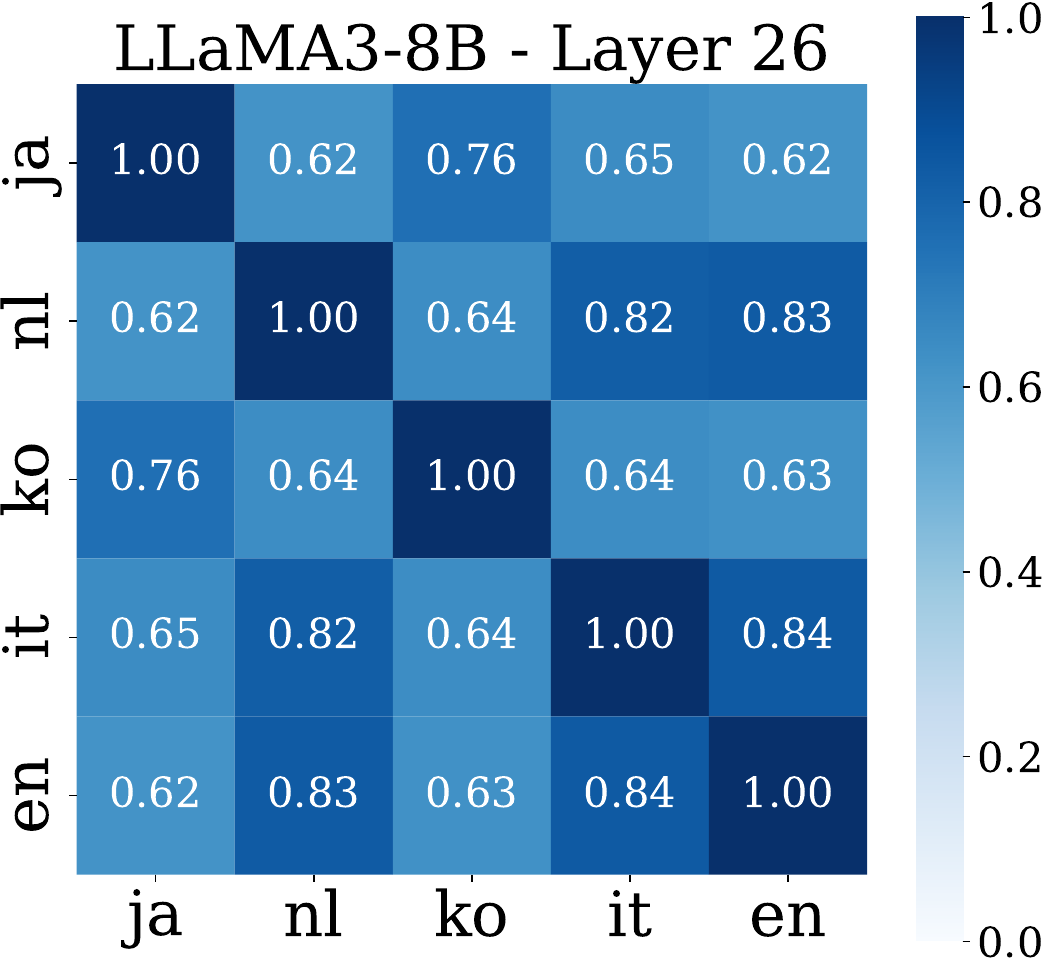}
  
  \begin{minipage}{0.19\linewidth}\centering \textbf{\textcolor{red}{layer 25 (Type-2)}}\end{minipage}
  \begin{minipage}{0.19\linewidth}\centering layer 25 (baseline)\end{minipage}
  \begin{minipage}{0.19\linewidth}\centering \textbf{\textcolor{red}{layer 26 (Type-2)}}\end{minipage}
  \begin{minipage}{0.19\linewidth}\centering layer 26 (baseline)\end{minipage}

  \includegraphics[width=0.19\linewidth]{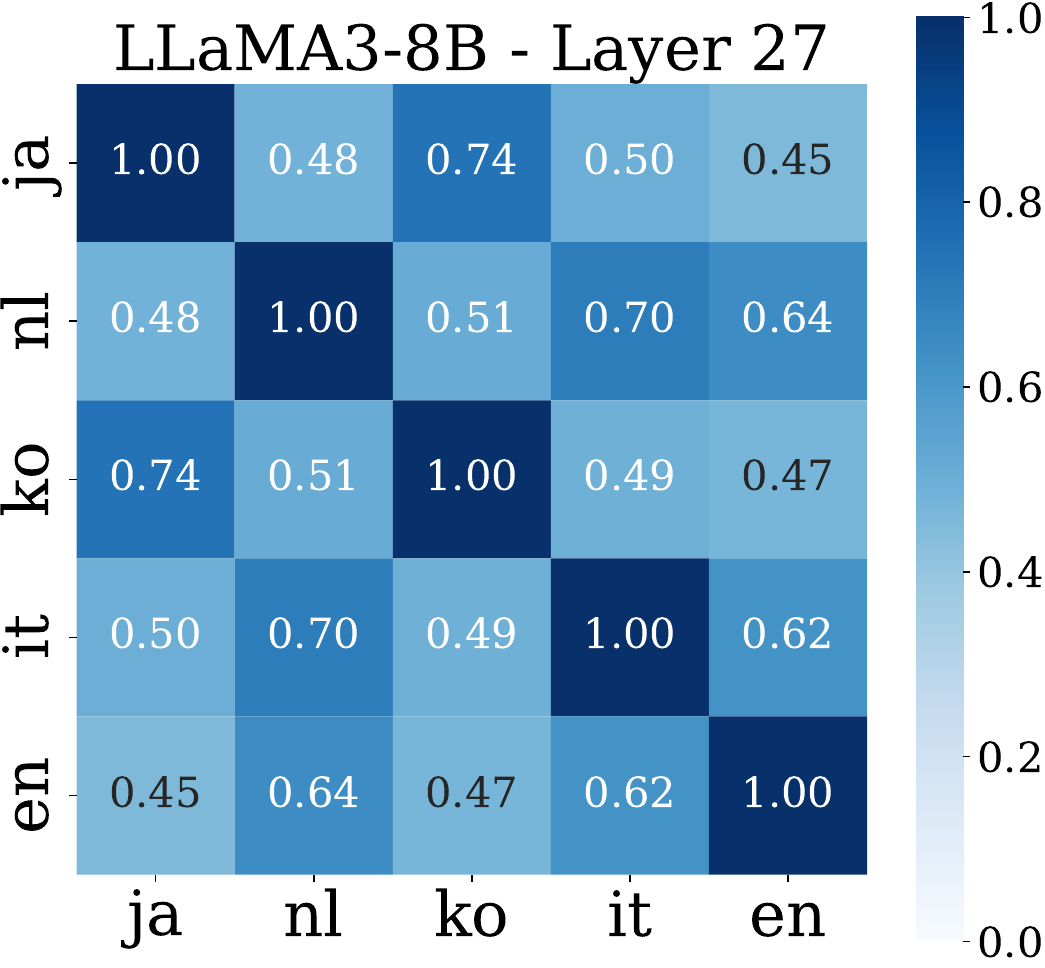}
  \includegraphics[width=0.19\linewidth]{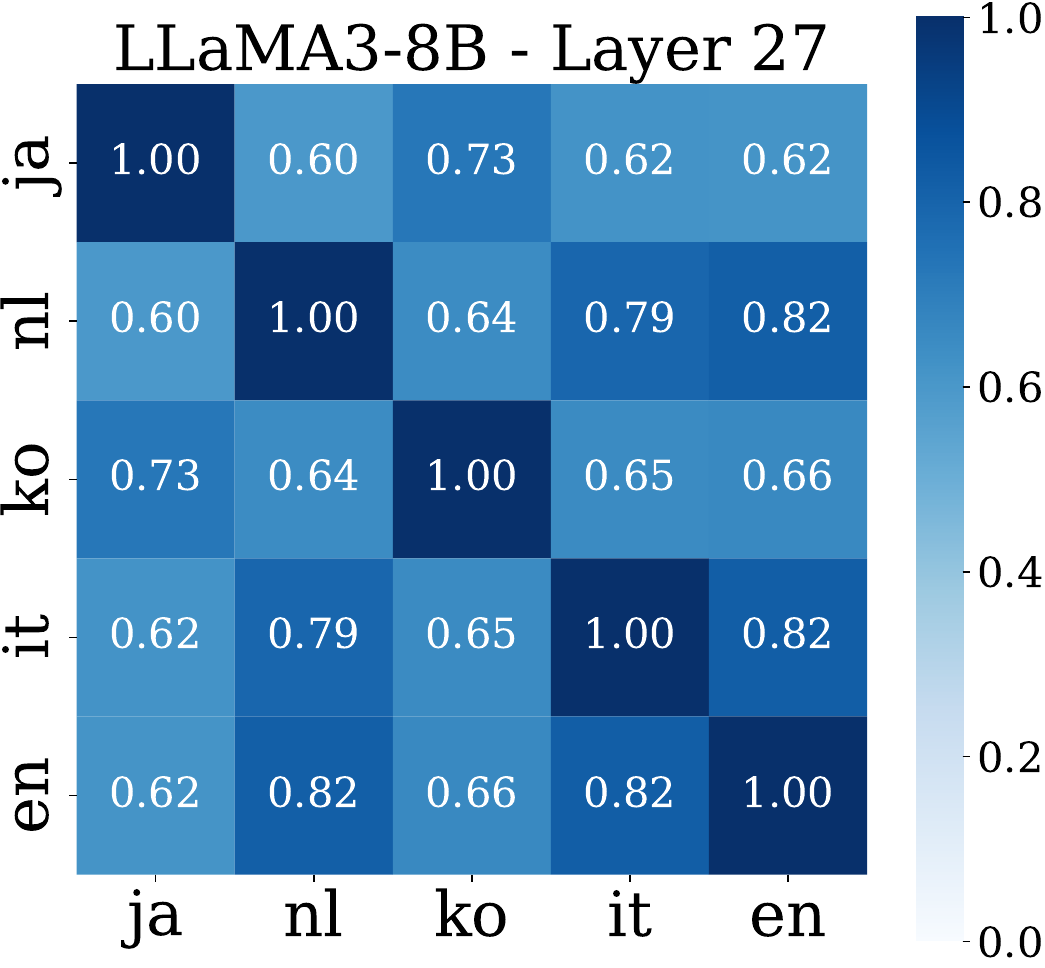}
  \includegraphics[width=0.19\linewidth]{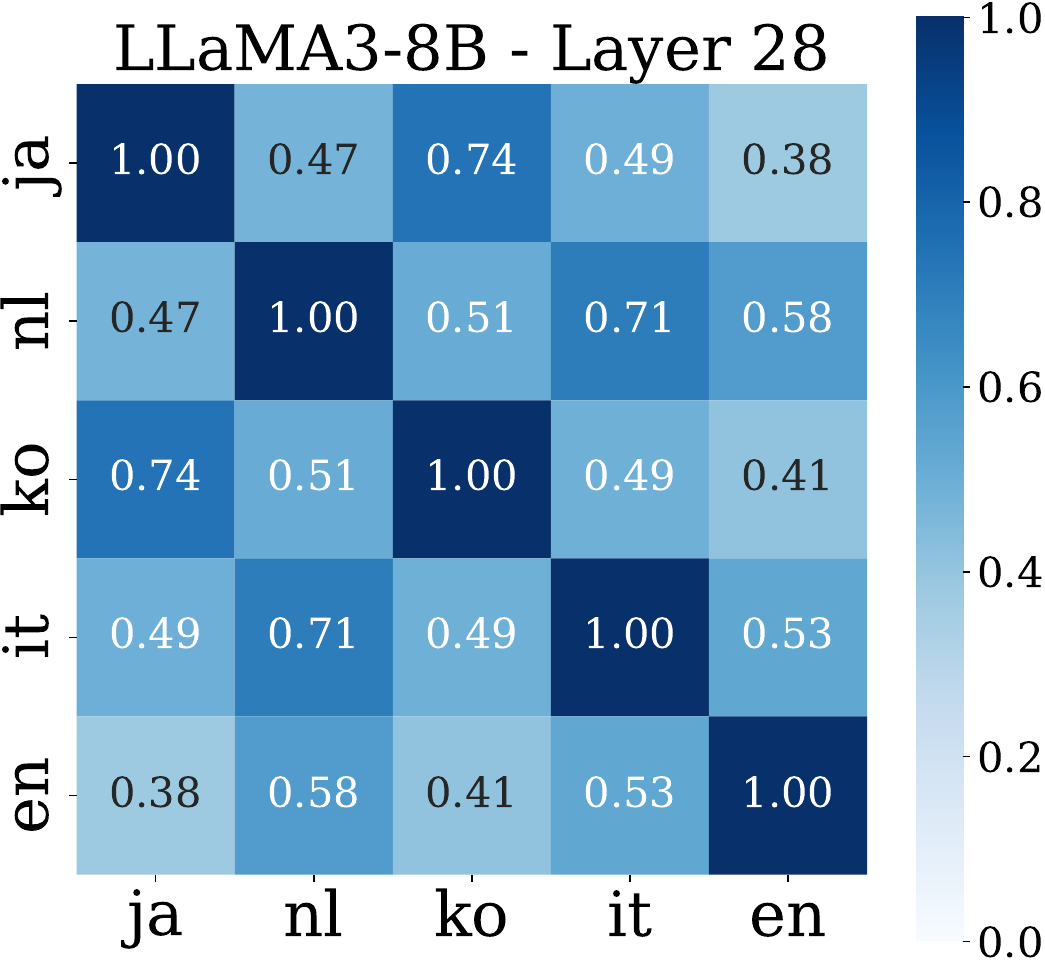}
  \includegraphics[width=0.19\linewidth]{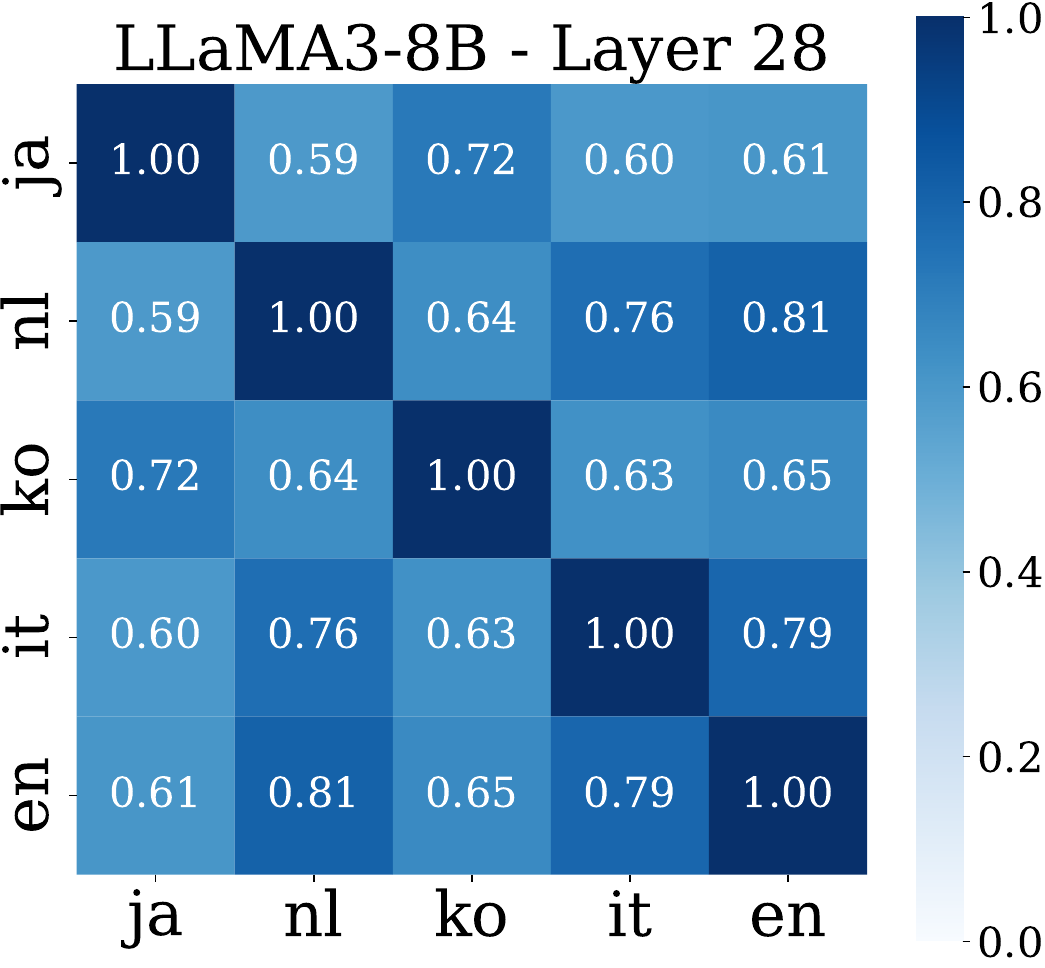}

  \begin{minipage}{0.19\linewidth}\centering \textbf{\textcolor{red}{layer 27 (Type-2)}}\end{minipage}
  \begin{minipage}{0.19\linewidth}\centering layer 27 (baseline)\end{minipage}
  \begin{minipage}{0.19\linewidth}\centering \textbf{\textcolor{red}{layer 28 (Type-2)}}\end{minipage}
  \begin{minipage}{0.19\linewidth}\centering layer 28 (baseline)\end{minipage}

  \includegraphics[width=0.19\linewidth]{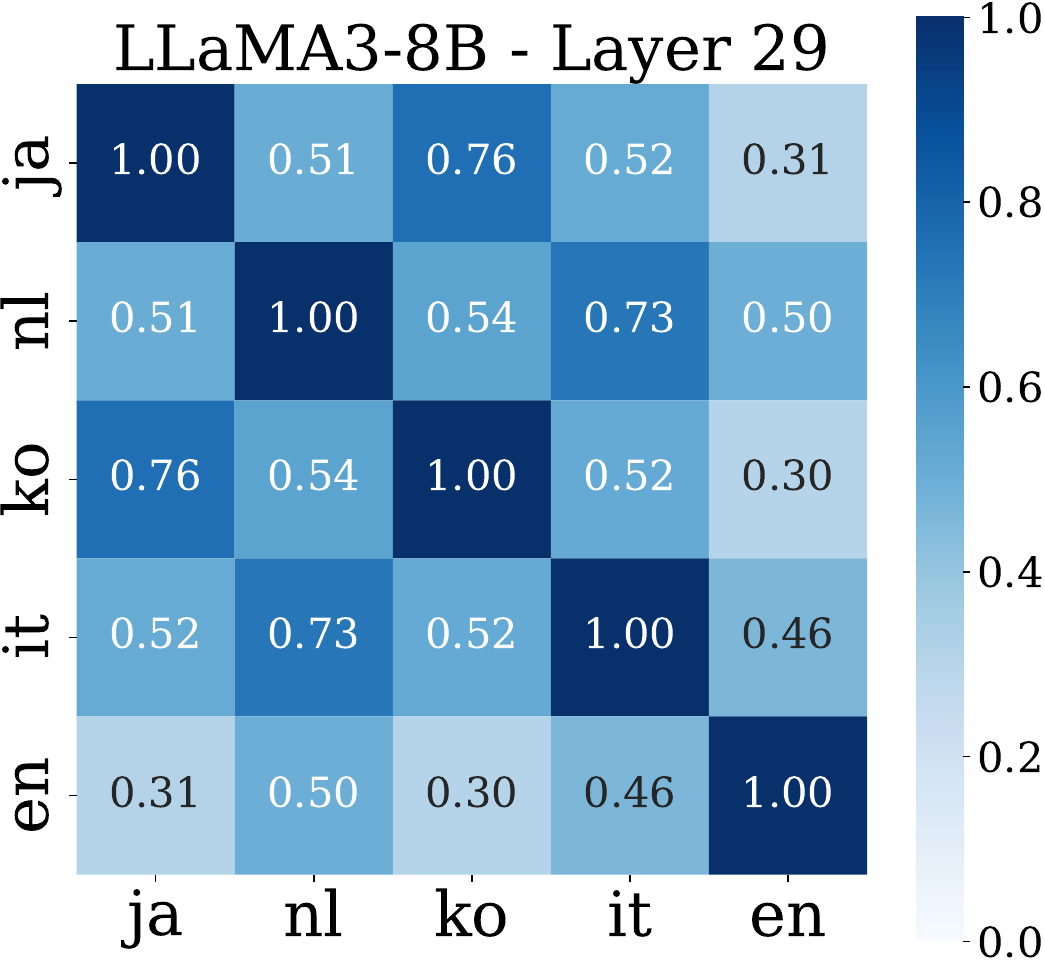}
  \includegraphics[width=0.19\linewidth]{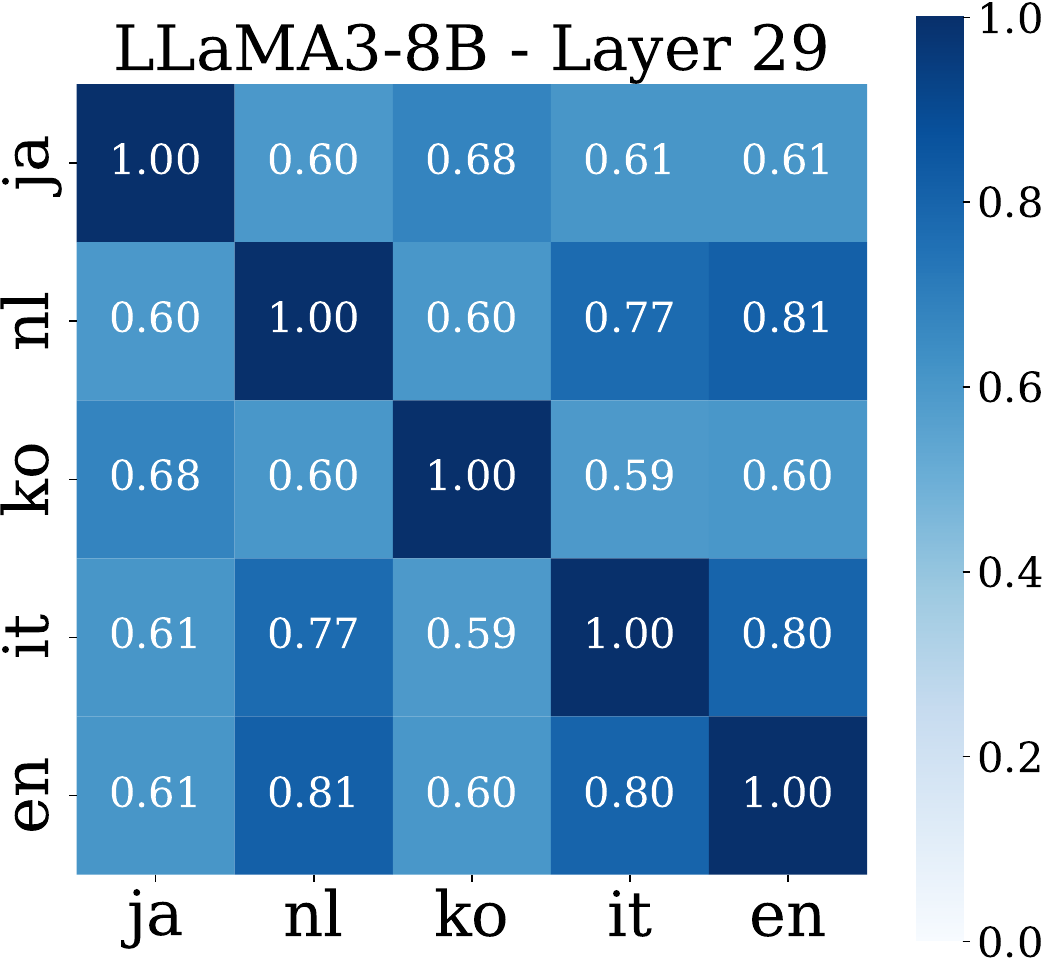}
  \includegraphics[width=0.19\linewidth]{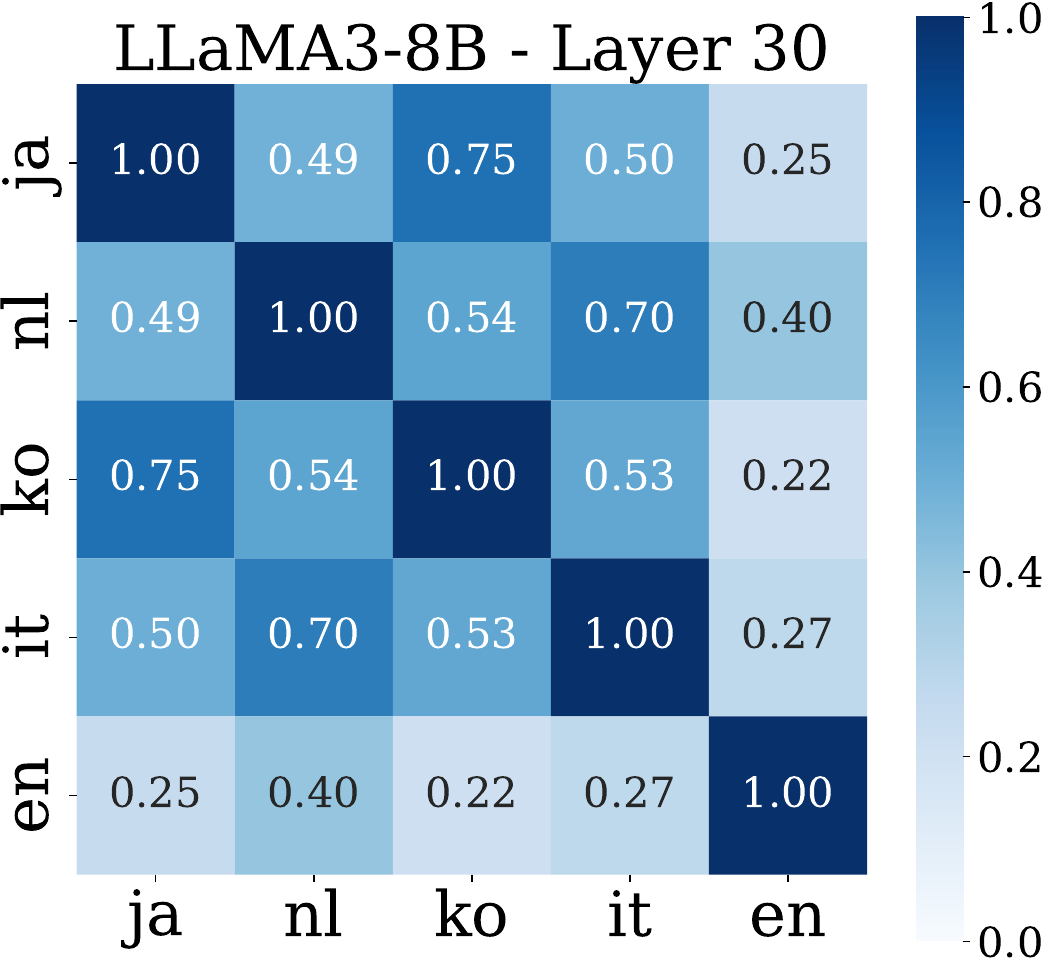}
  \includegraphics[width=0.19\linewidth]{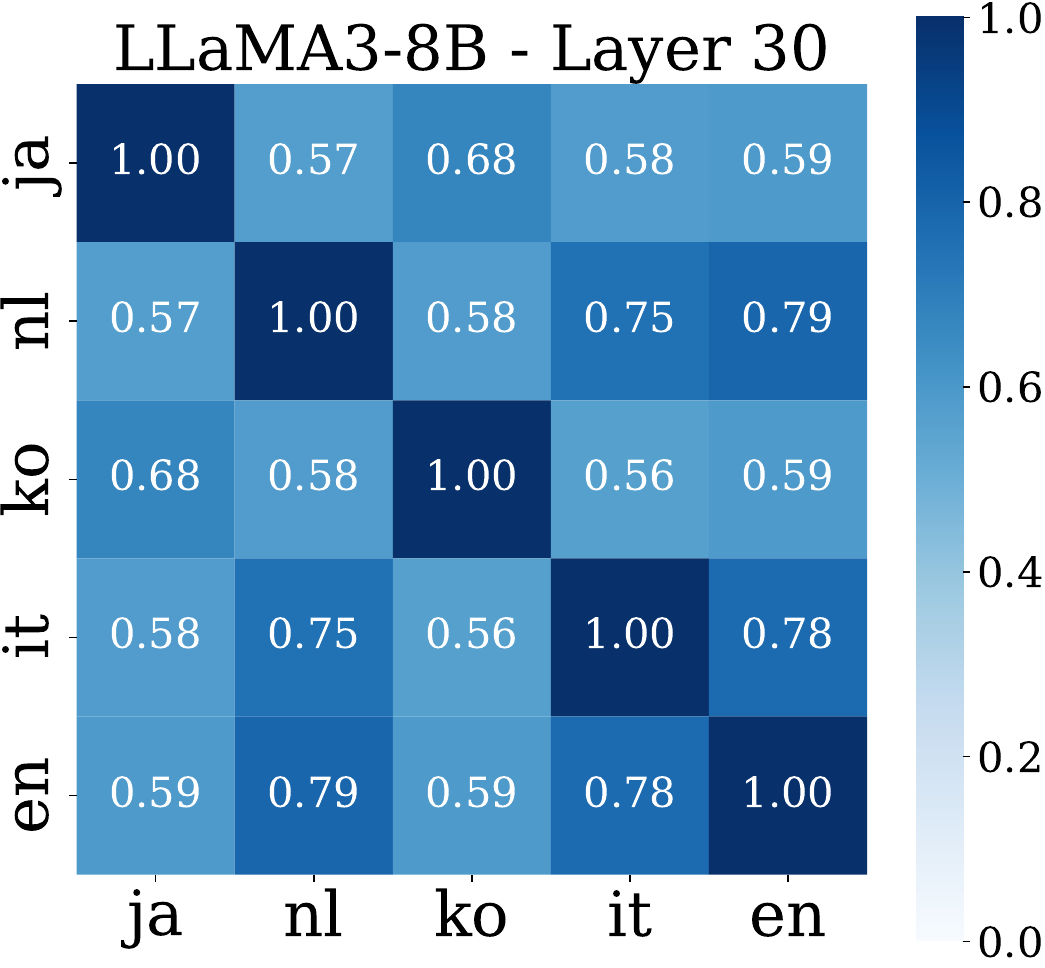}

  \begin{minipage}{0.19\linewidth}\centering \textbf{\textcolor{red}{layer 29 (Type-2)}}\end{minipage}
  \begin{minipage}{0.19\linewidth}\centering layer 29 (baseline)\end{minipage}
  \begin{minipage}{0.19\linewidth}\centering \textbf{\textcolor{red}{layer 30 (Type-2)}}\end{minipage}
  \begin{minipage}{0.19\linewidth}\centering layer 30 (baseline)\end{minipage}

  \includegraphics[width=0.19\linewidth]{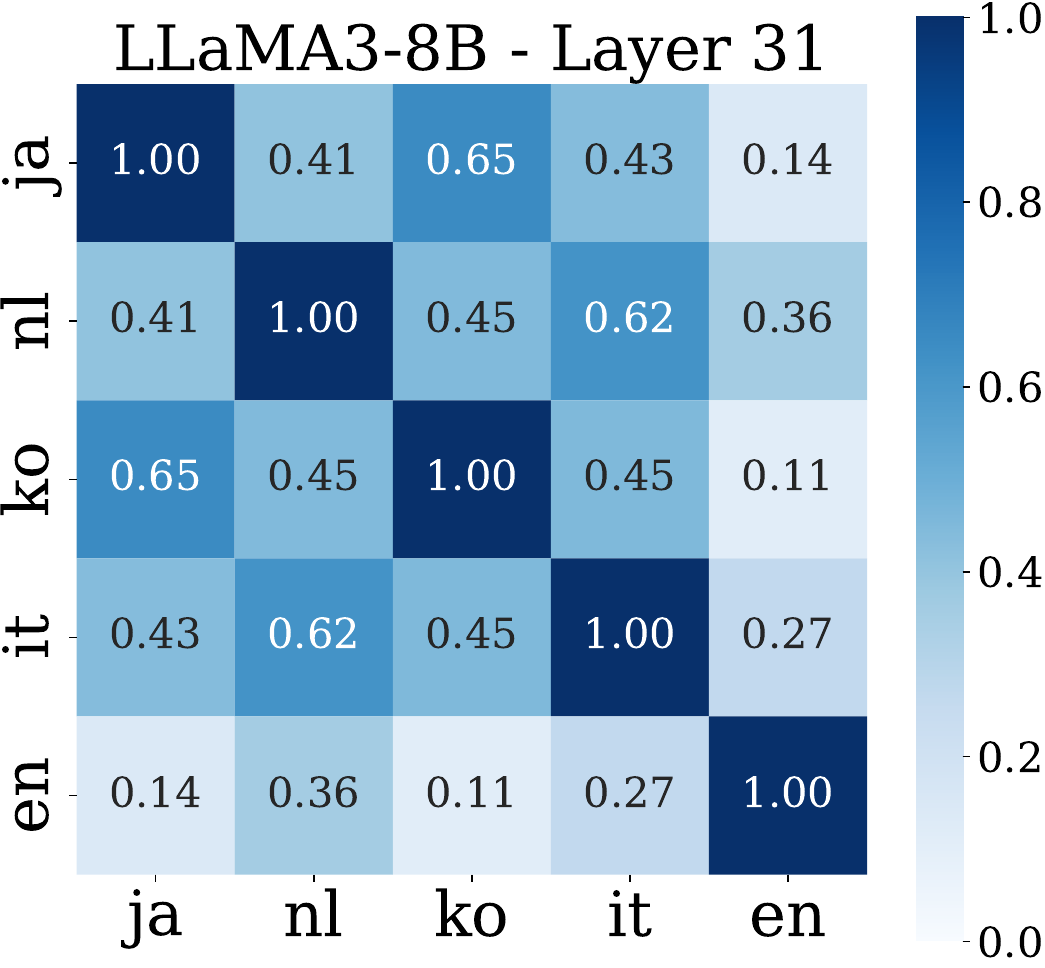}
  \includegraphics[width=0.19\linewidth]{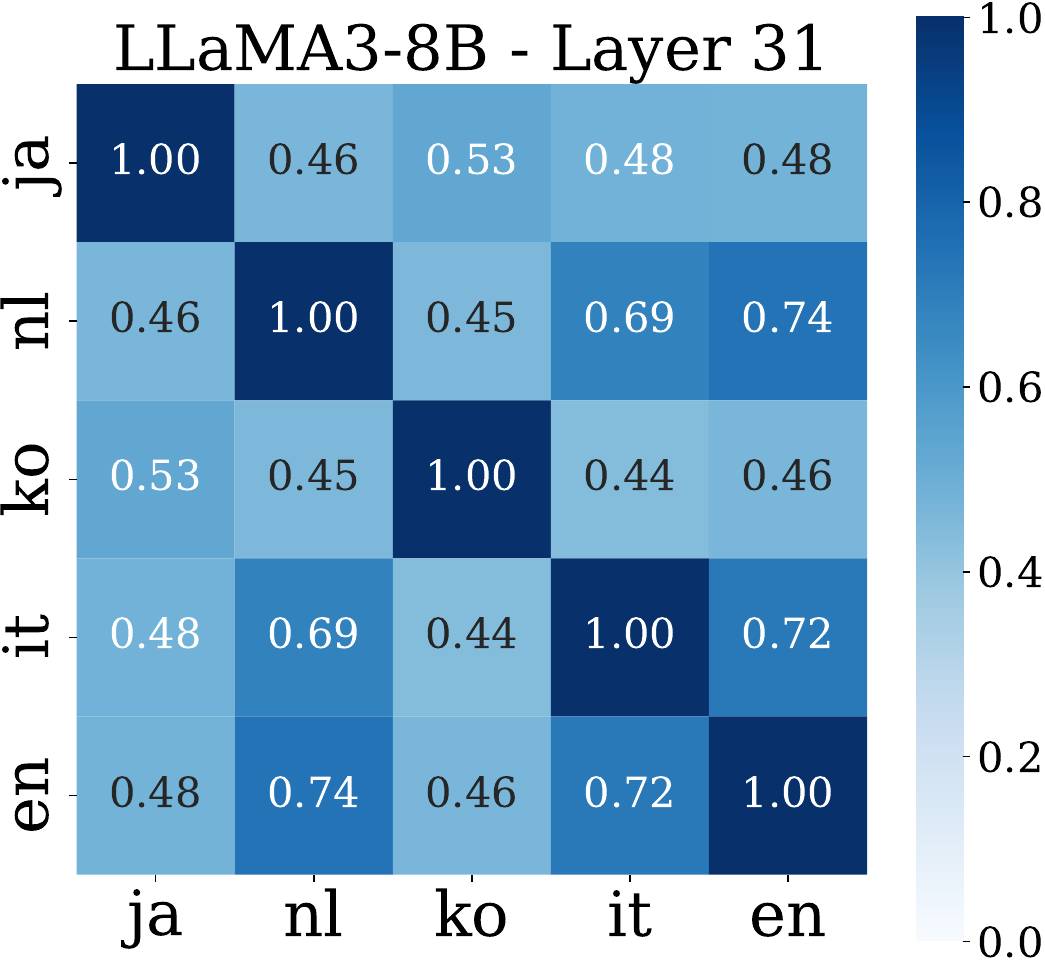}
  \includegraphics[width=0.19\linewidth]{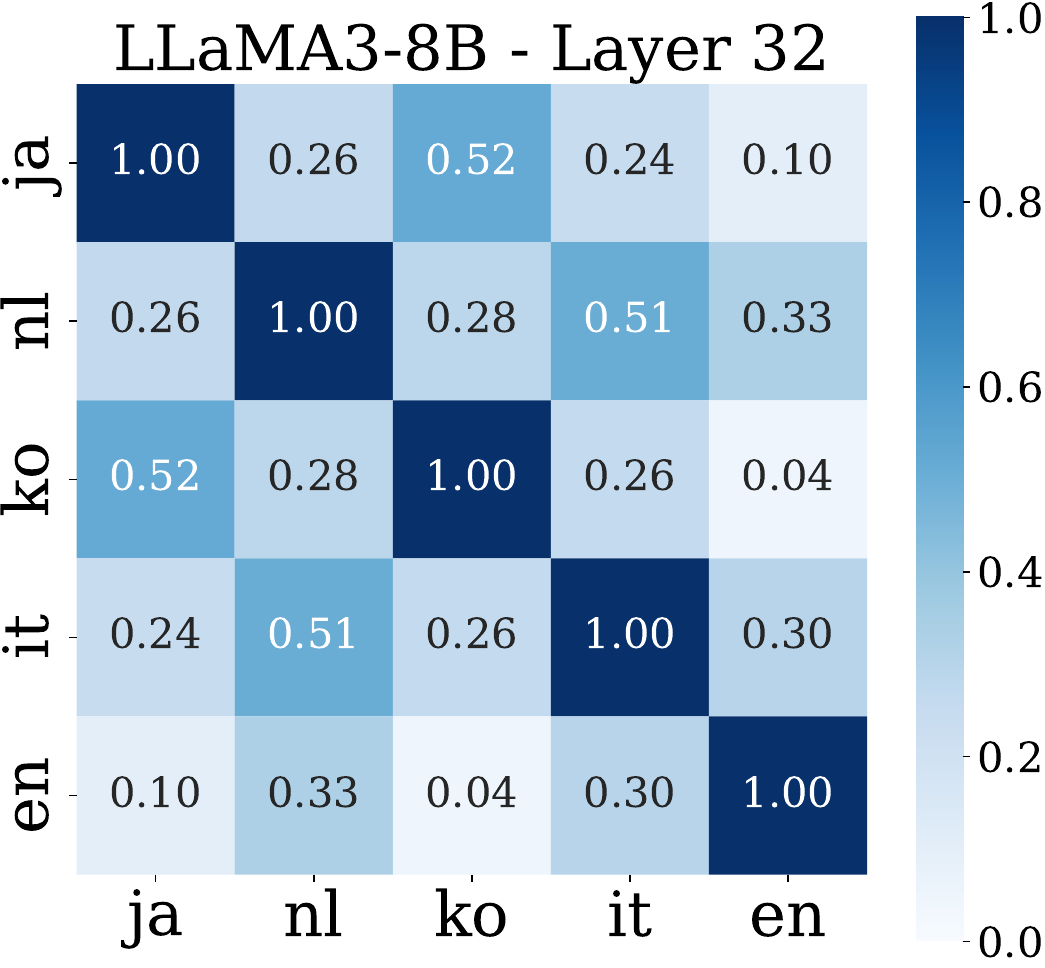}
  \includegraphics[width=0.19\linewidth]{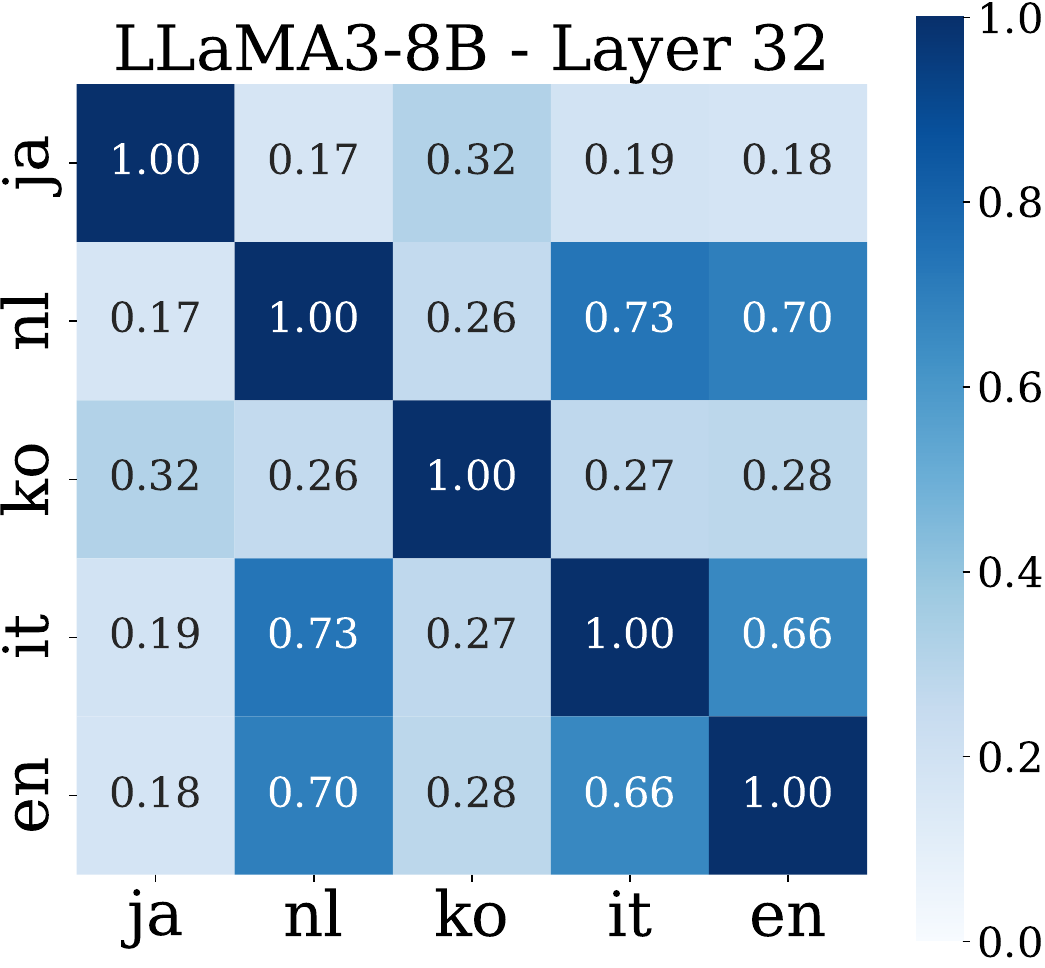}

  \begin{minipage}{0.19\linewidth}\centering \textbf{\textcolor{red}{layer 31 (Type-2)}}\end{minipage}
  \begin{minipage}{0.19\linewidth}\centering layer 31 (baseline)\end{minipage}
  \begin{minipage}{0.19\linewidth}\centering \textbf{\textcolor{red}{layer 32 (Type-2)}}\end{minipage}
  \begin{minipage}{0.19\linewidth}\centering layer 32 (baseline)\end{minipage}

  \caption{\textbf{Distance among centroids of language latent spaces while deactivating top-1k Type-2 Transfer Neurons (LLaMA3-8B)}. \textbf{\textcolor{red}{Layer (Type-2)}} indicates the result of deactivating Type-2 neurons, whereas "baseline" refers tot the result of deactivating randomly sampled neurons.}
  \label{fig:appendix:centroids distance among language subspaces while deactivating type2 llama3}
\end{figure*}

\begin{figure*}[t]
  \centering

  \includegraphics[width=0.19\linewidth]{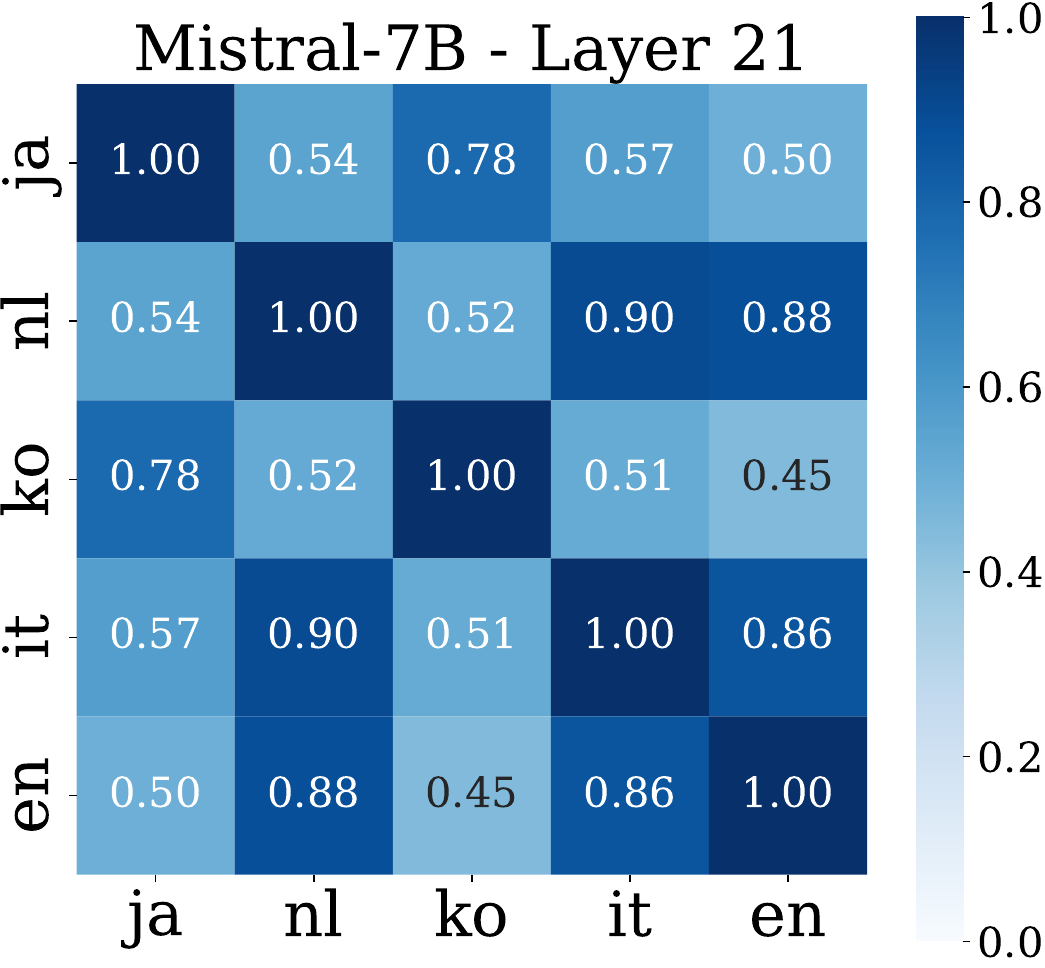}
  \includegraphics[width=0.19\linewidth]{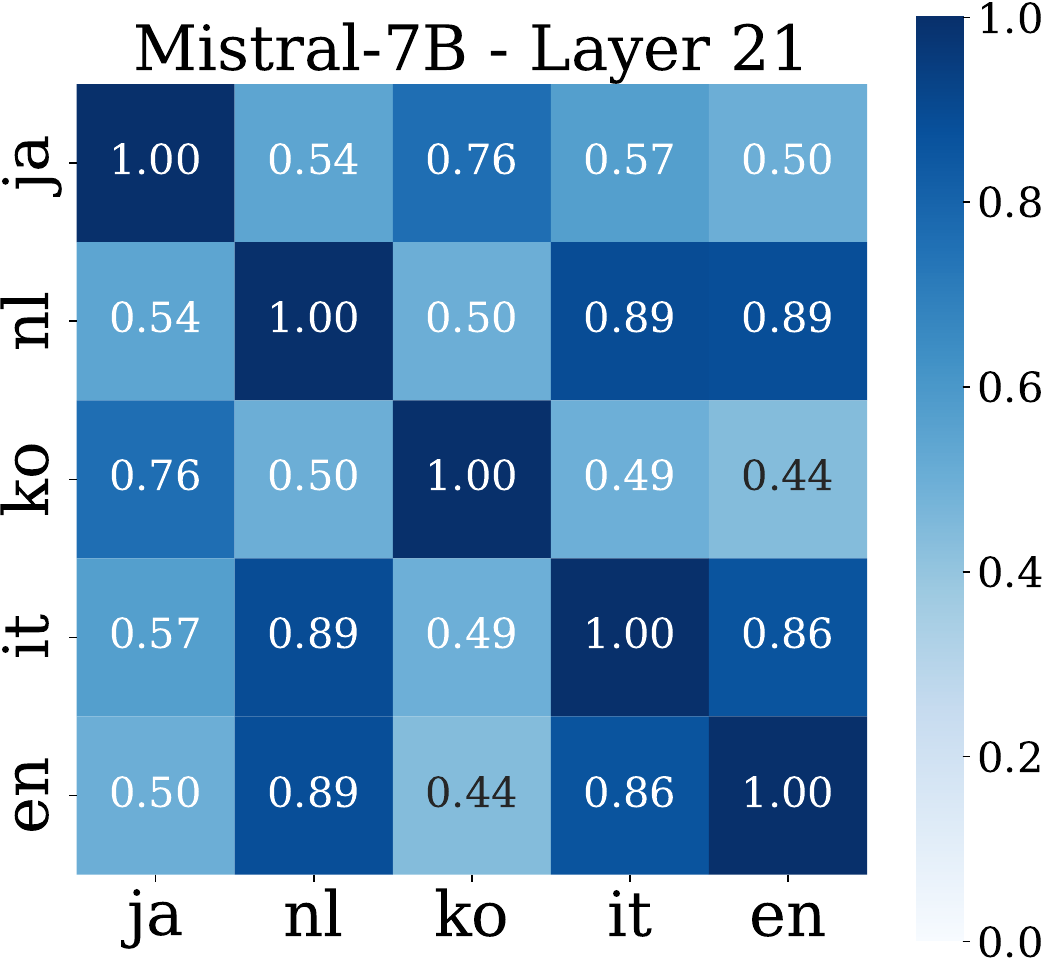}
  \includegraphics[width=0.19\linewidth]{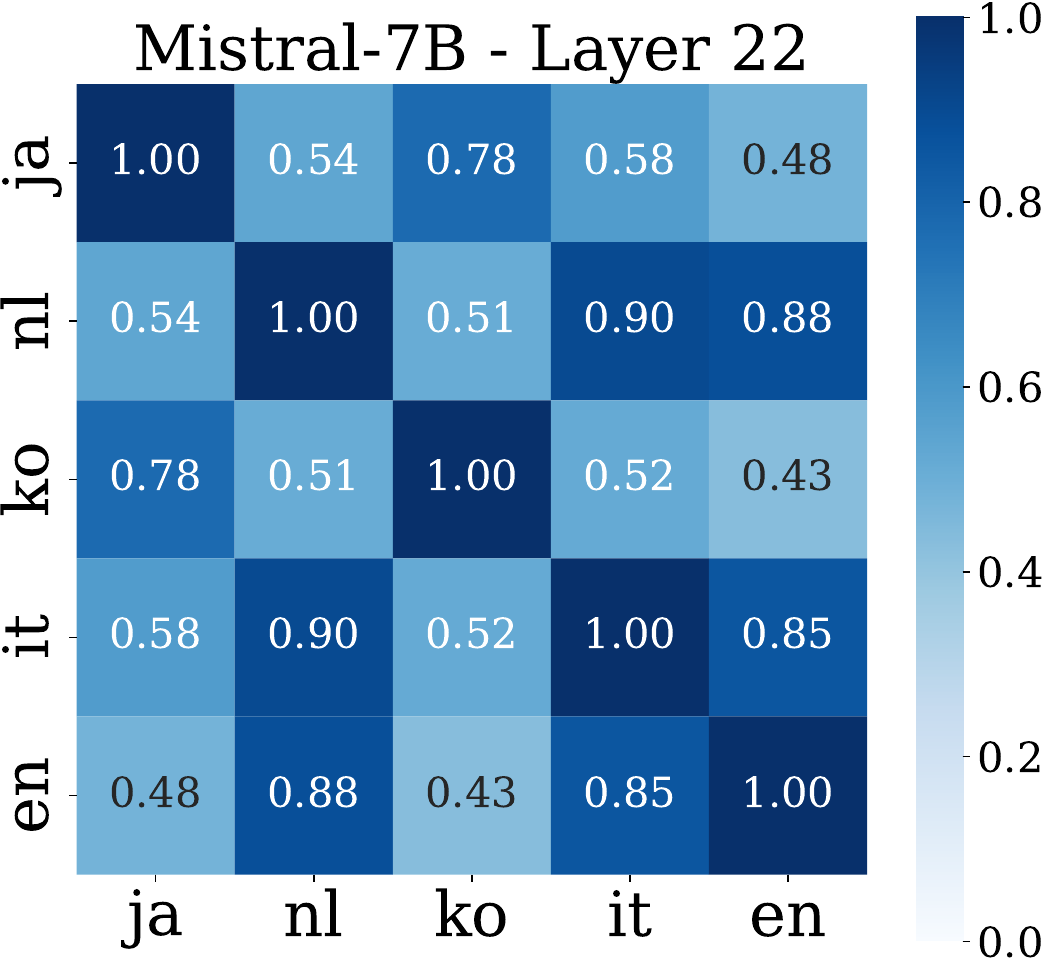}
  \includegraphics[width=0.19\linewidth]{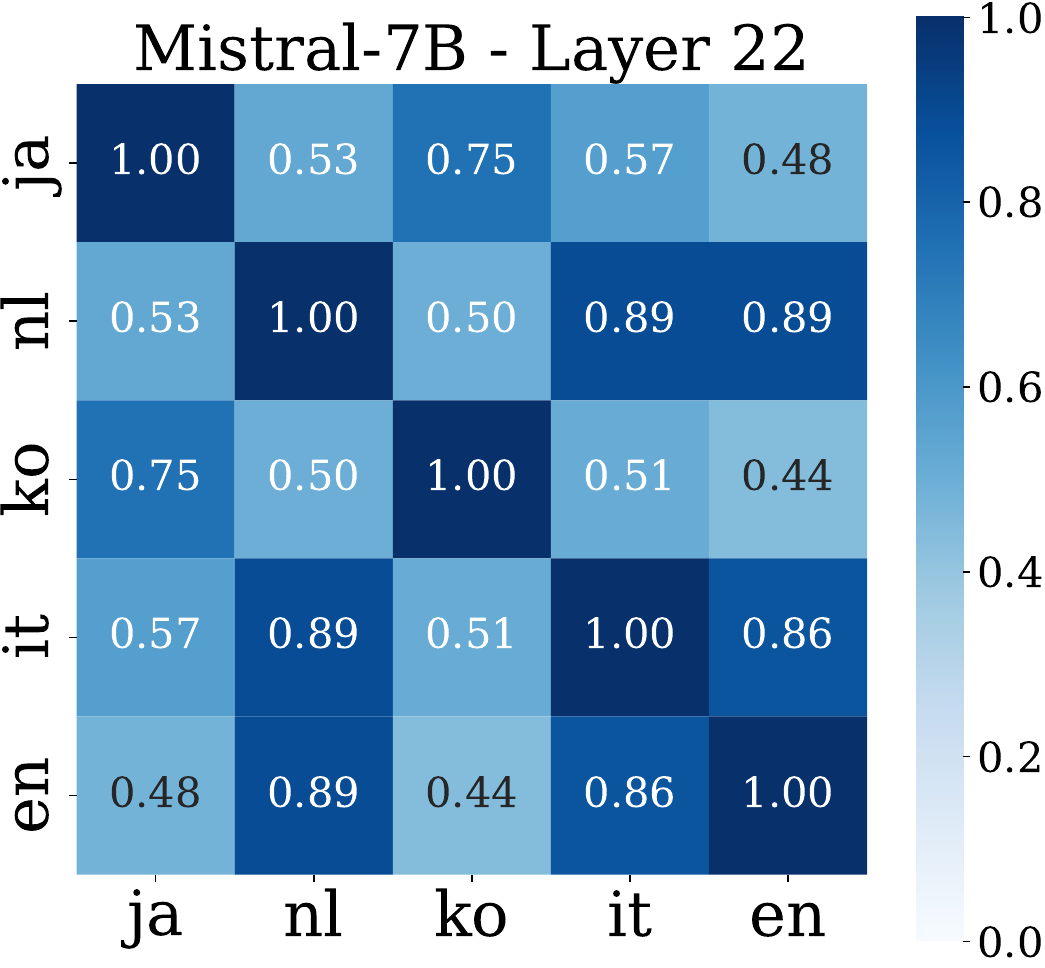}

  \begin{minipage}{0.19\linewidth}\centering \textbf{\textcolor{red}{layer 21 (Type-2)}}\end{minipage}
  \begin{minipage}{0.19\linewidth}\centering layer 21 (baseline)\end{minipage}
  \begin{minipage}{0.19\linewidth}\centering \textbf{\textcolor{red}{layer 22 (Type-2)}}\end{minipage}
  \begin{minipage}{0.19\linewidth}\centering layer 22 (baseline)\end{minipage}

  \includegraphics[width=0.19\linewidth]{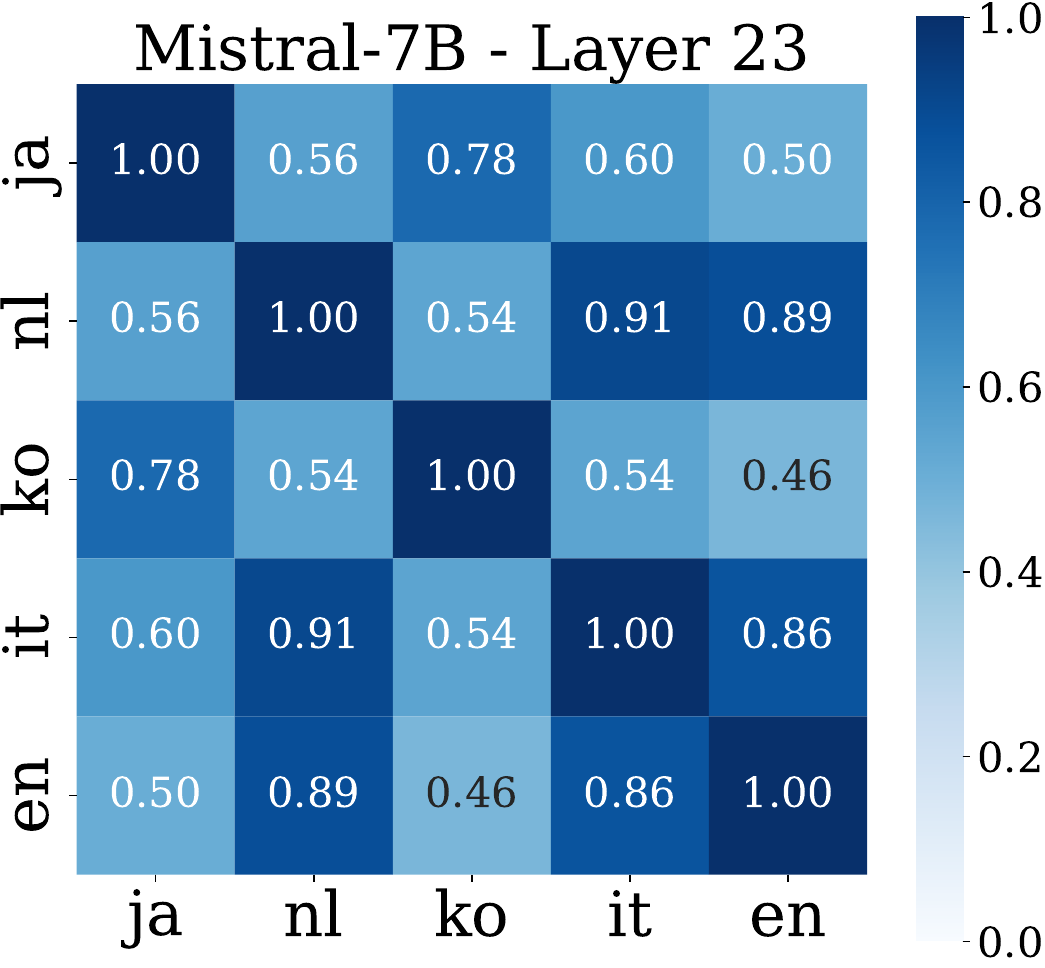}
  \includegraphics[width=0.19\linewidth]{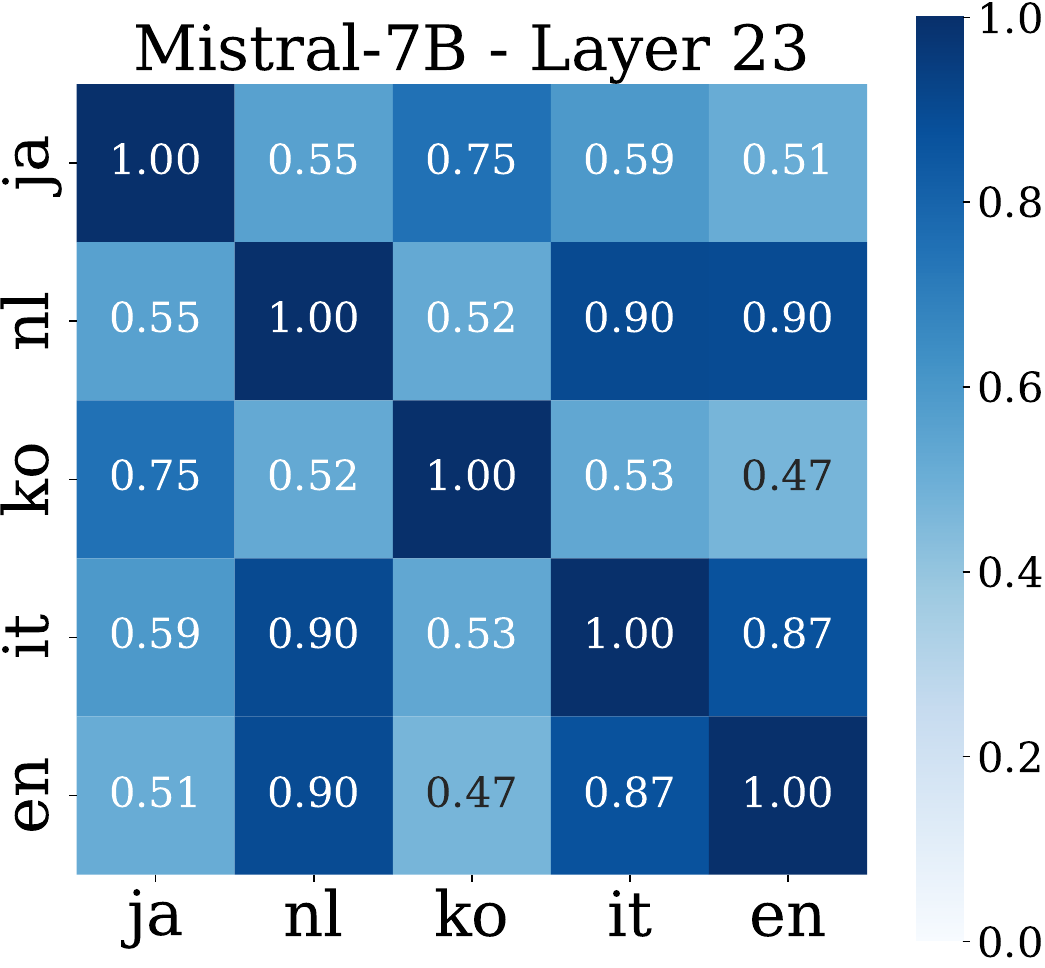}
  \includegraphics[width=0.19\linewidth]{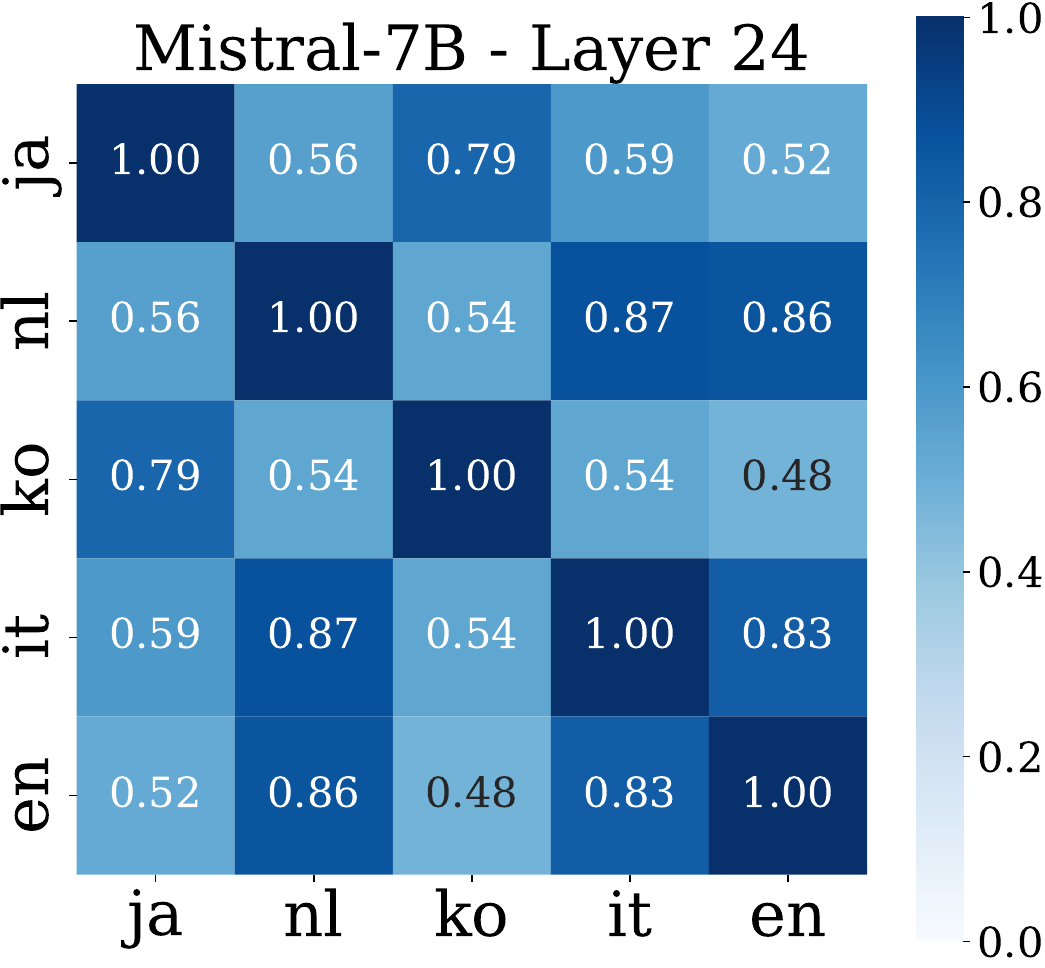}
  \includegraphics[width=0.19\linewidth]{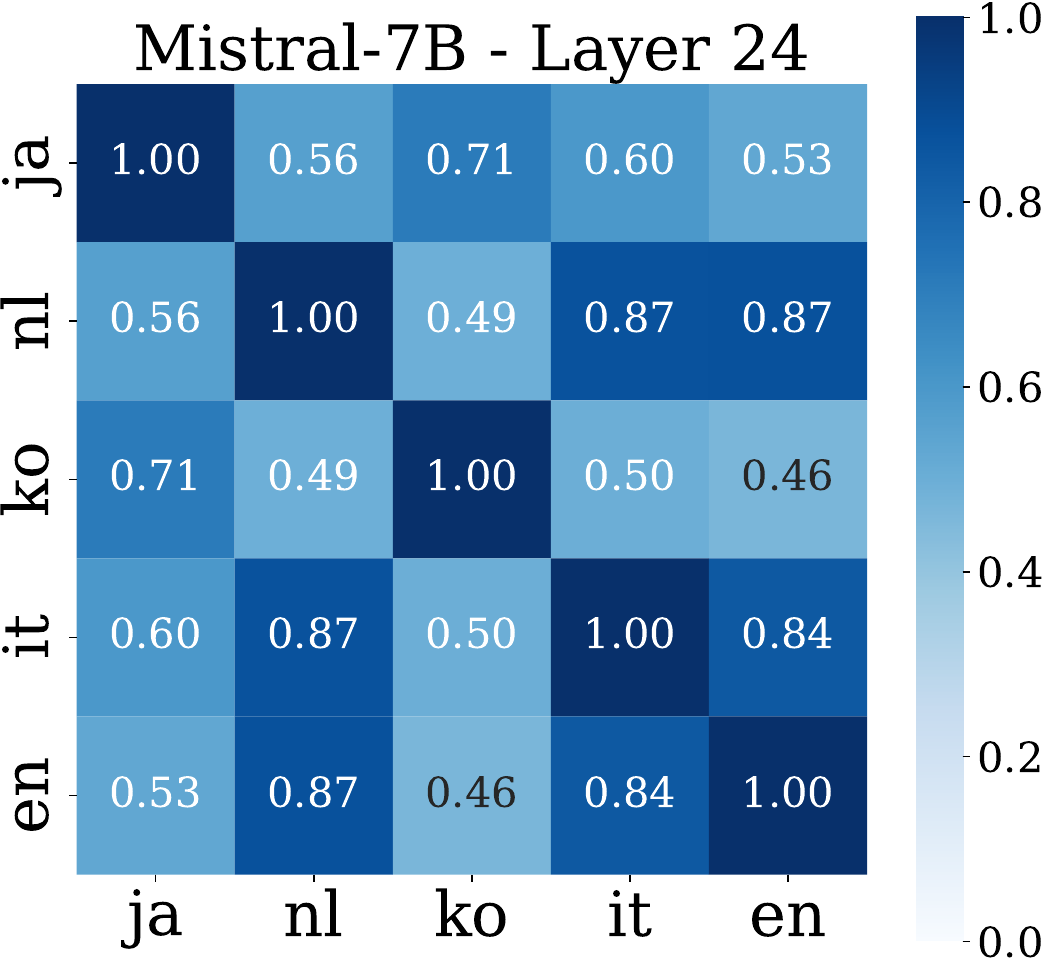}

  \begin{minipage}{0.19\linewidth}\centering \textbf{\textcolor{red}{layer 23 (Type-2)}}\end{minipage}
  \begin{minipage}{0.19\linewidth}\centering layer 23 (baseline)\end{minipage}
  \begin{minipage}{0.19\linewidth}\centering \textbf{\textcolor{red}{layer 24 (Type-2)}}\end{minipage}
  \begin{minipage}{0.19\linewidth}\centering layer 24 (baseline)\end{minipage}

  \includegraphics[width=0.19\linewidth]{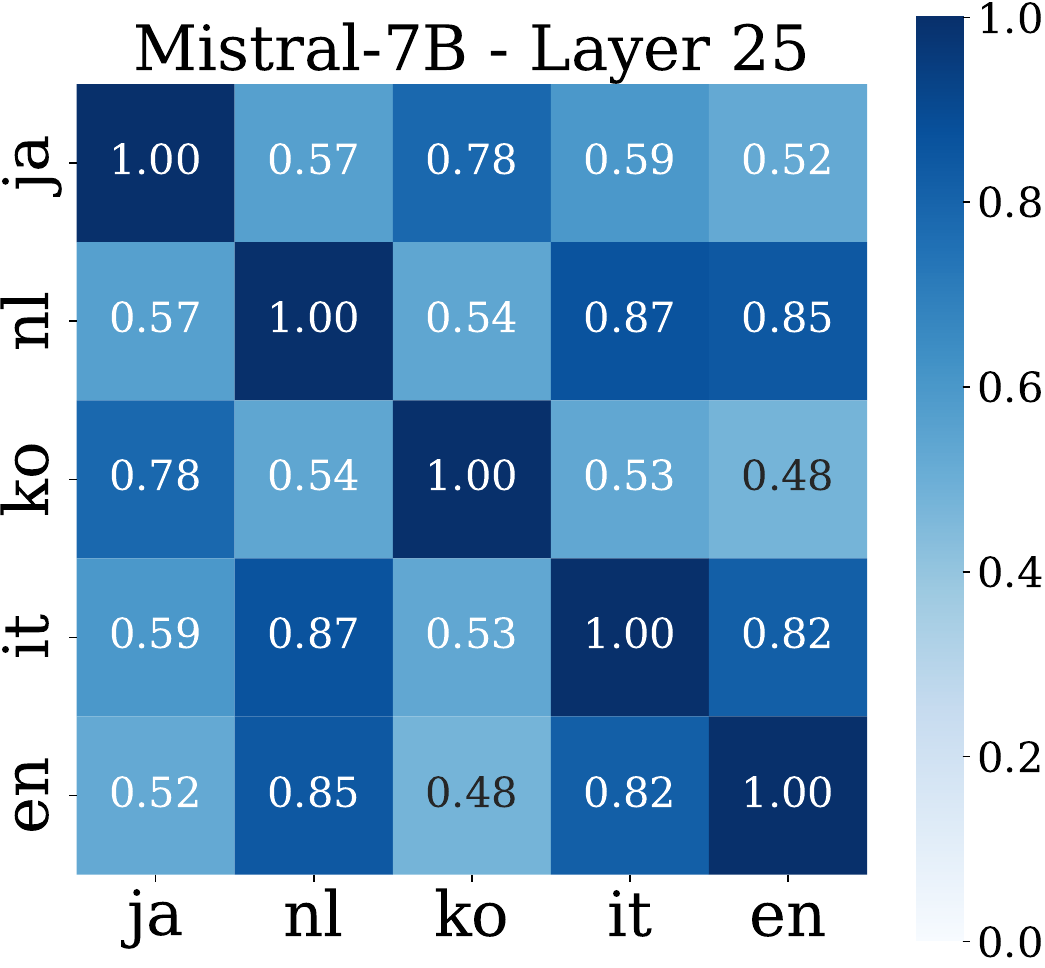}
  \includegraphics[width=0.19\linewidth]{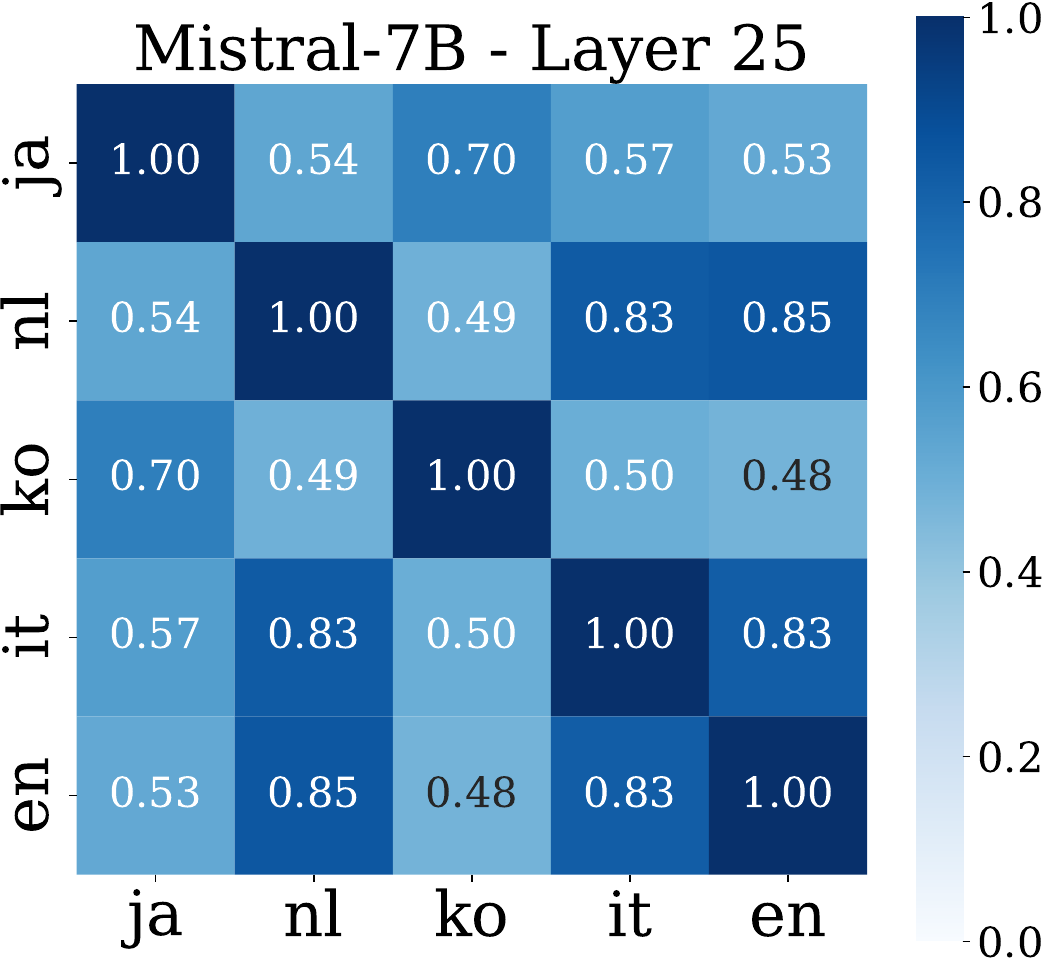}
  \includegraphics[width=0.19\linewidth]{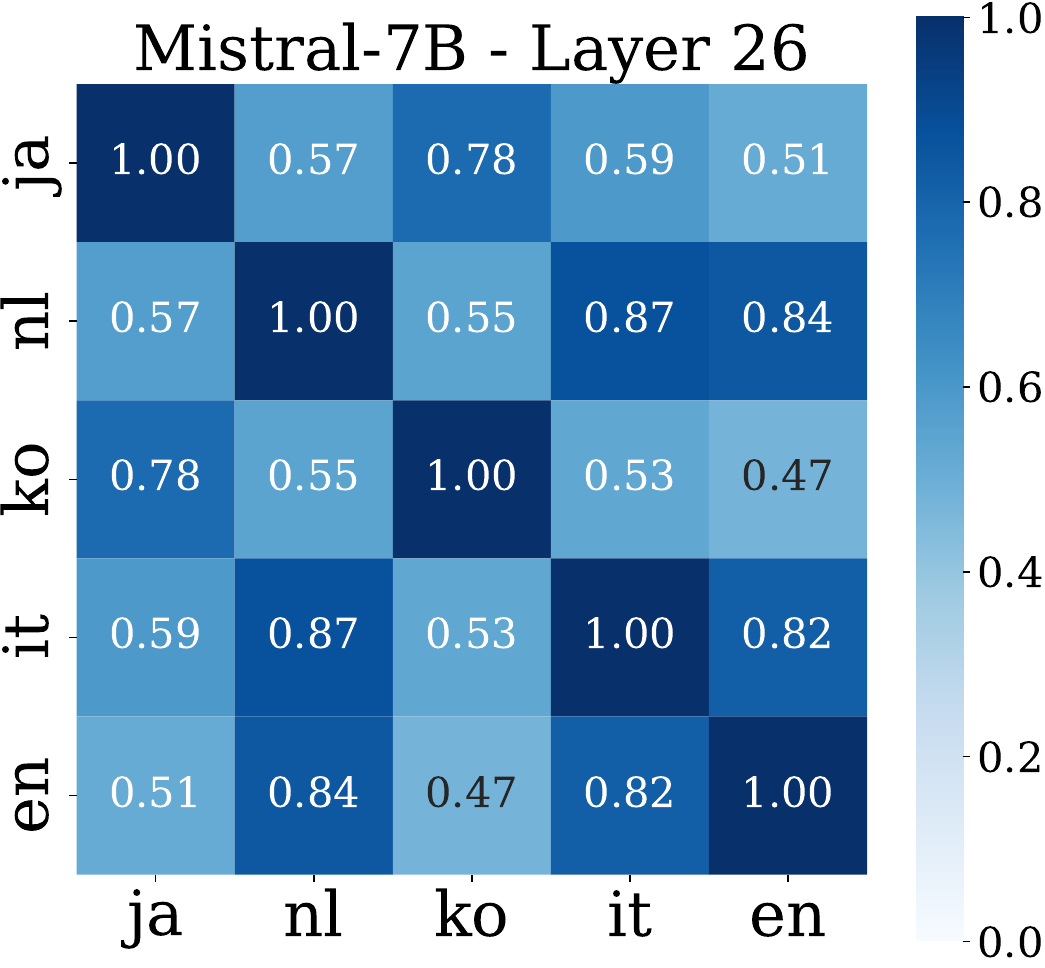}
  \includegraphics[width=0.19\linewidth]{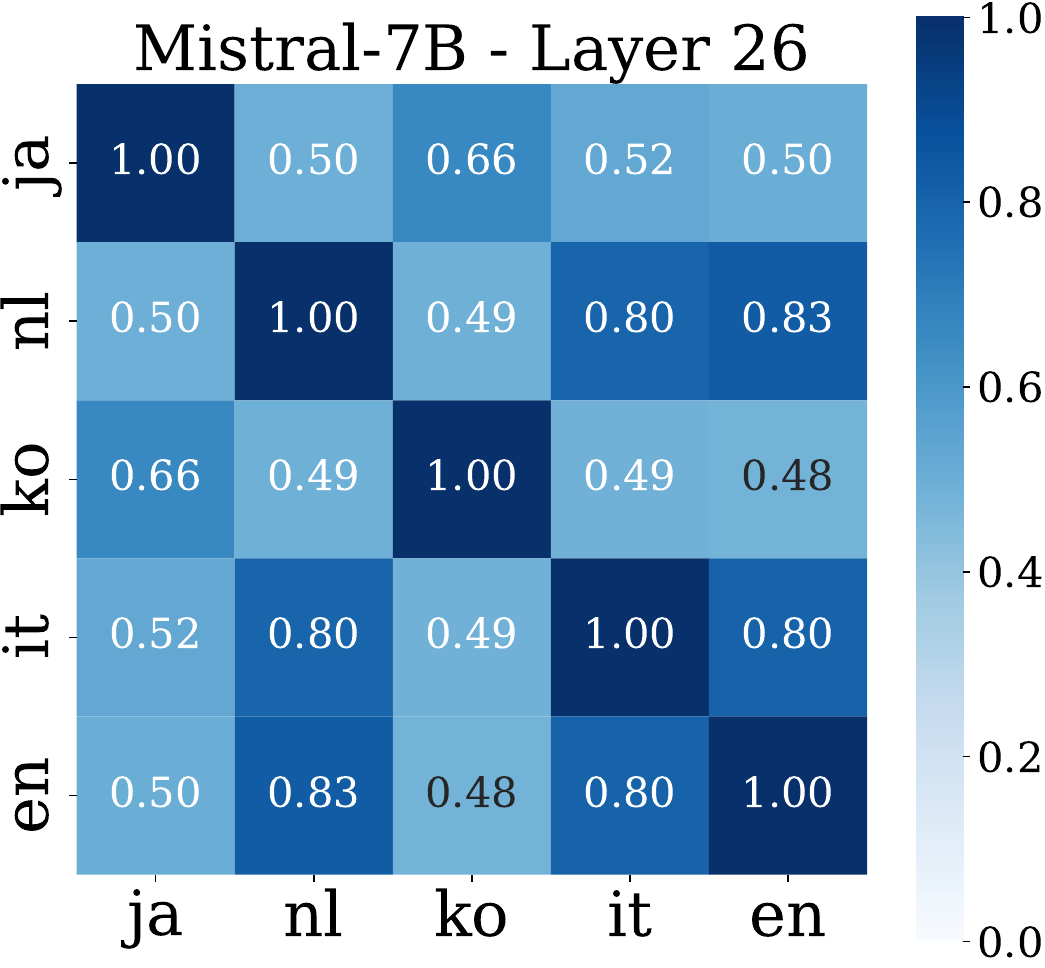}
  
  \begin{minipage}{0.19\linewidth}\centering \textbf{\textcolor{red}{layer 25 (Type-2)}}\end{minipage}
  \begin{minipage}{0.19\linewidth}\centering layer 25 (baseline)\end{minipage}
  \begin{minipage}{0.19\linewidth}\centering \textbf{\textcolor{red}{layer 26 (Type-2)}}\end{minipage}
  \begin{minipage}{0.19\linewidth}\centering layer 26 (baseline)\end{minipage}

  \includegraphics[width=0.19\linewidth]{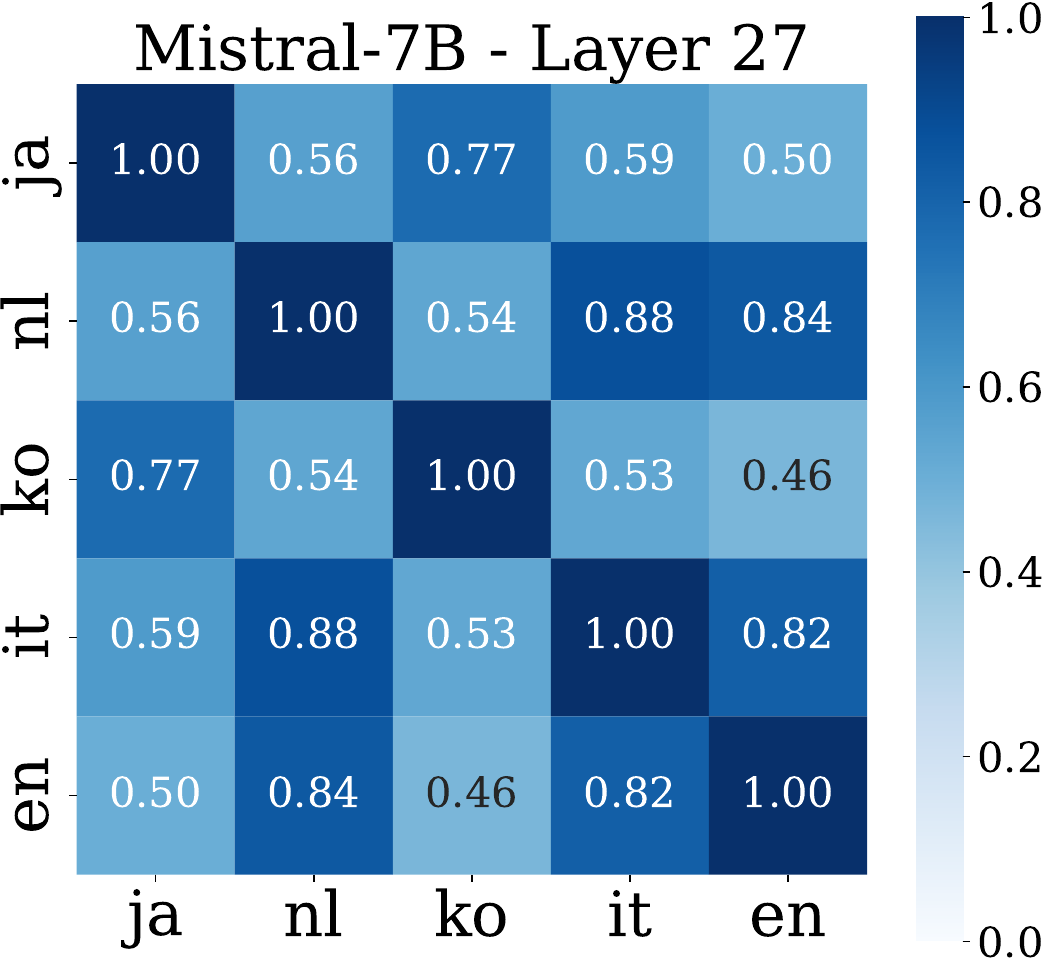}
  \includegraphics[width=0.19\linewidth]{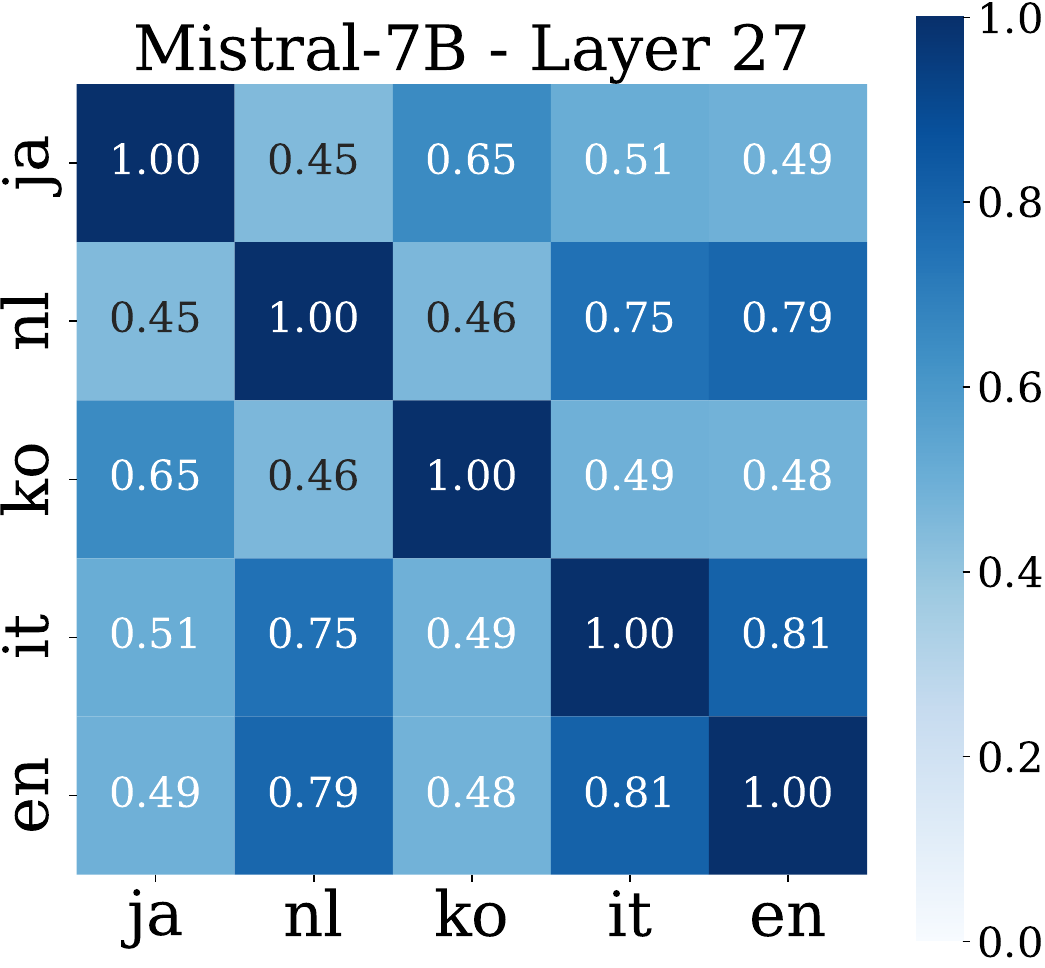}
  \includegraphics[width=0.19\linewidth]{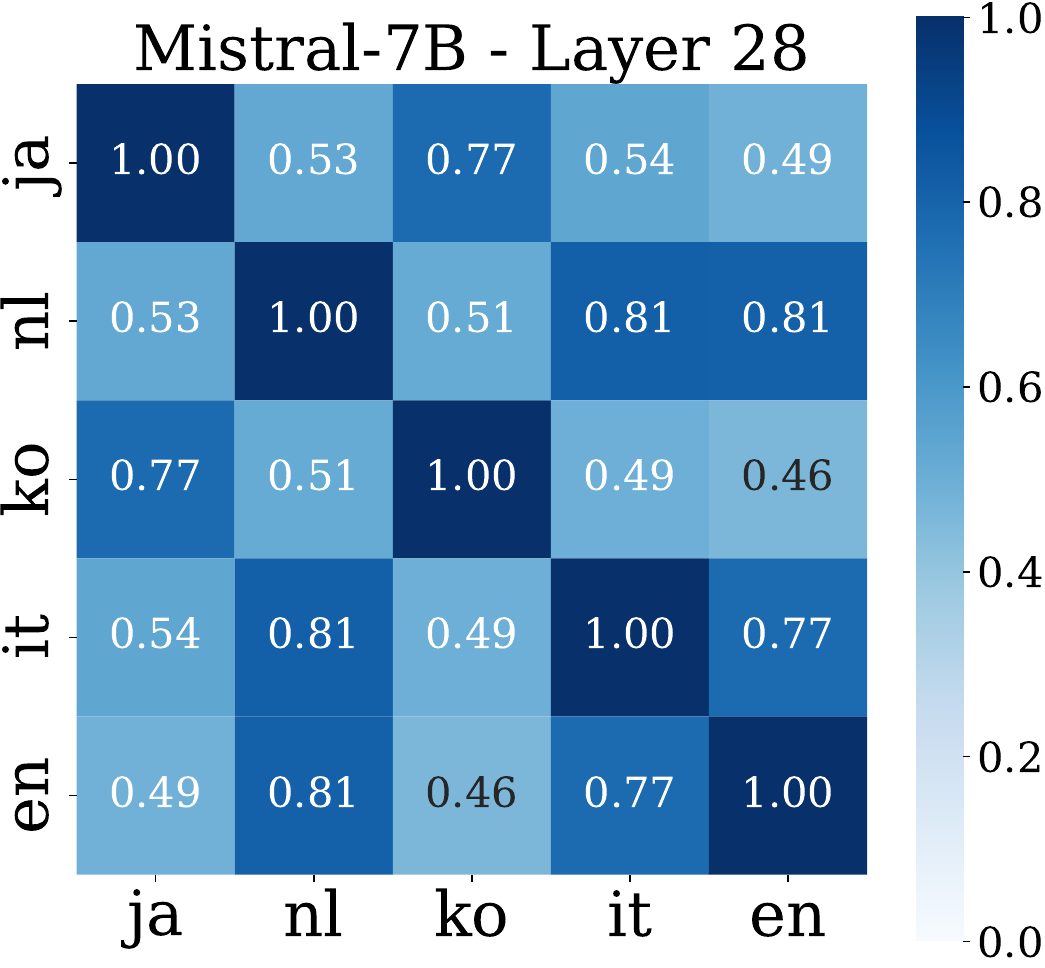}
  \includegraphics[width=0.19\linewidth]{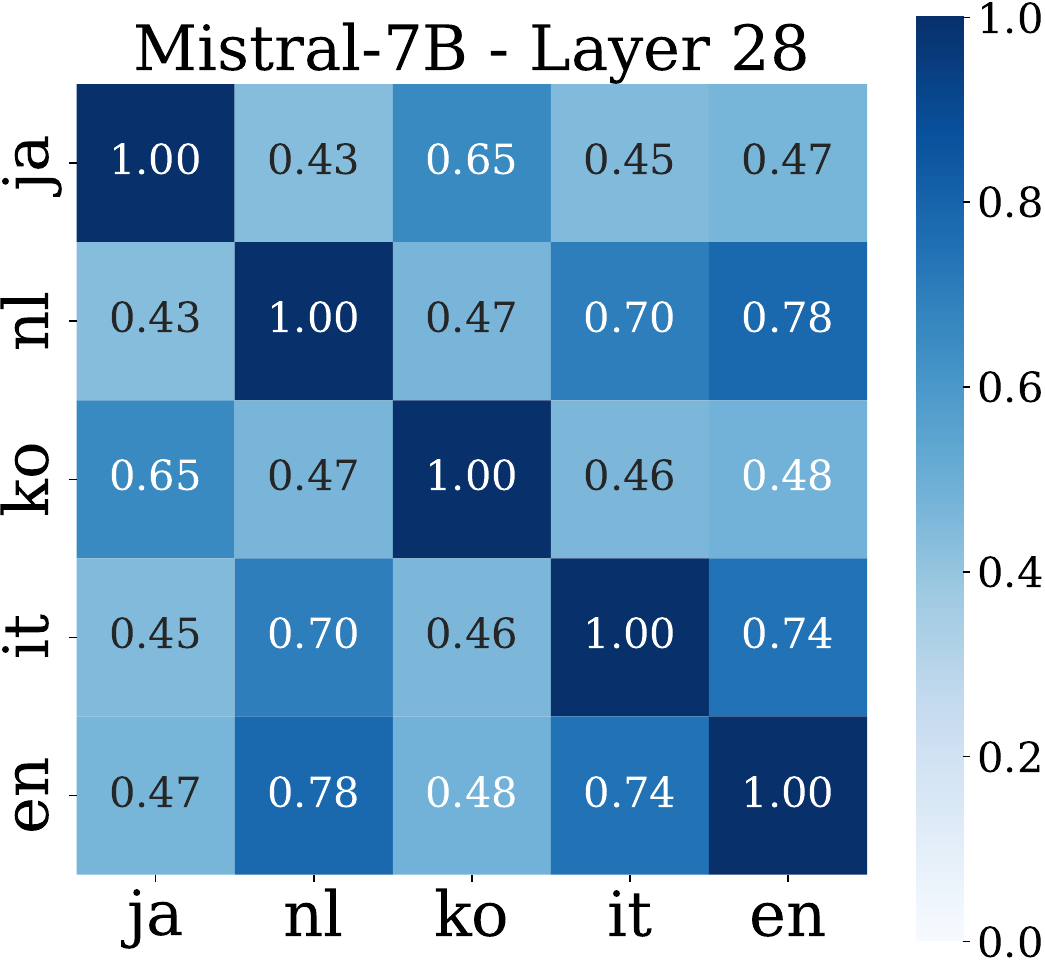}

  \begin{minipage}{0.19\linewidth}\centering \textbf{\textcolor{red}{layer 27 (Type-2)}}\end{minipage}
  \begin{minipage}{0.19\linewidth}\centering layer 27 (baseline)\end{minipage}
  \begin{minipage}{0.19\linewidth}\centering \textbf{\textcolor{red}{layer 28 (Type-2)}}\end{minipage}
  \begin{minipage}{0.19\linewidth}\centering layer 28 (baseline)\end{minipage}

  \includegraphics[width=0.19\linewidth]{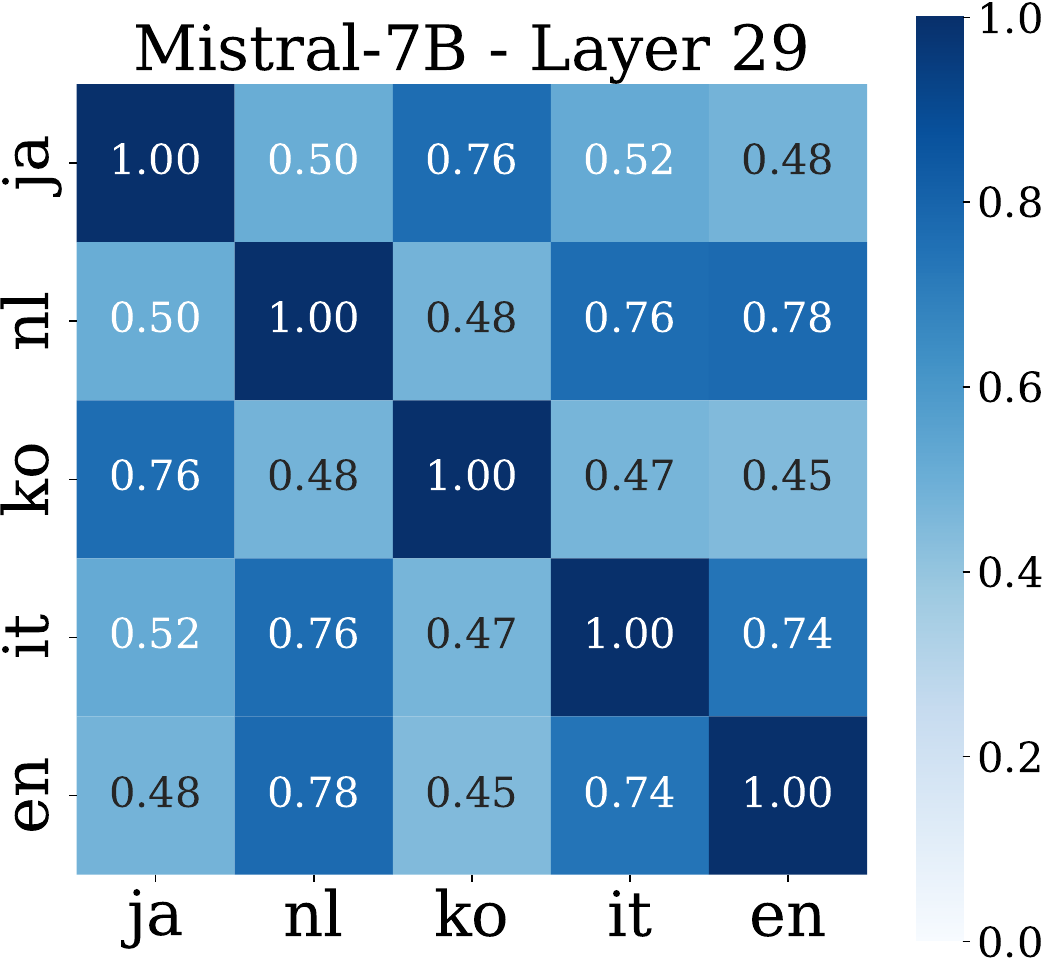}
  \includegraphics[width=0.19\linewidth]{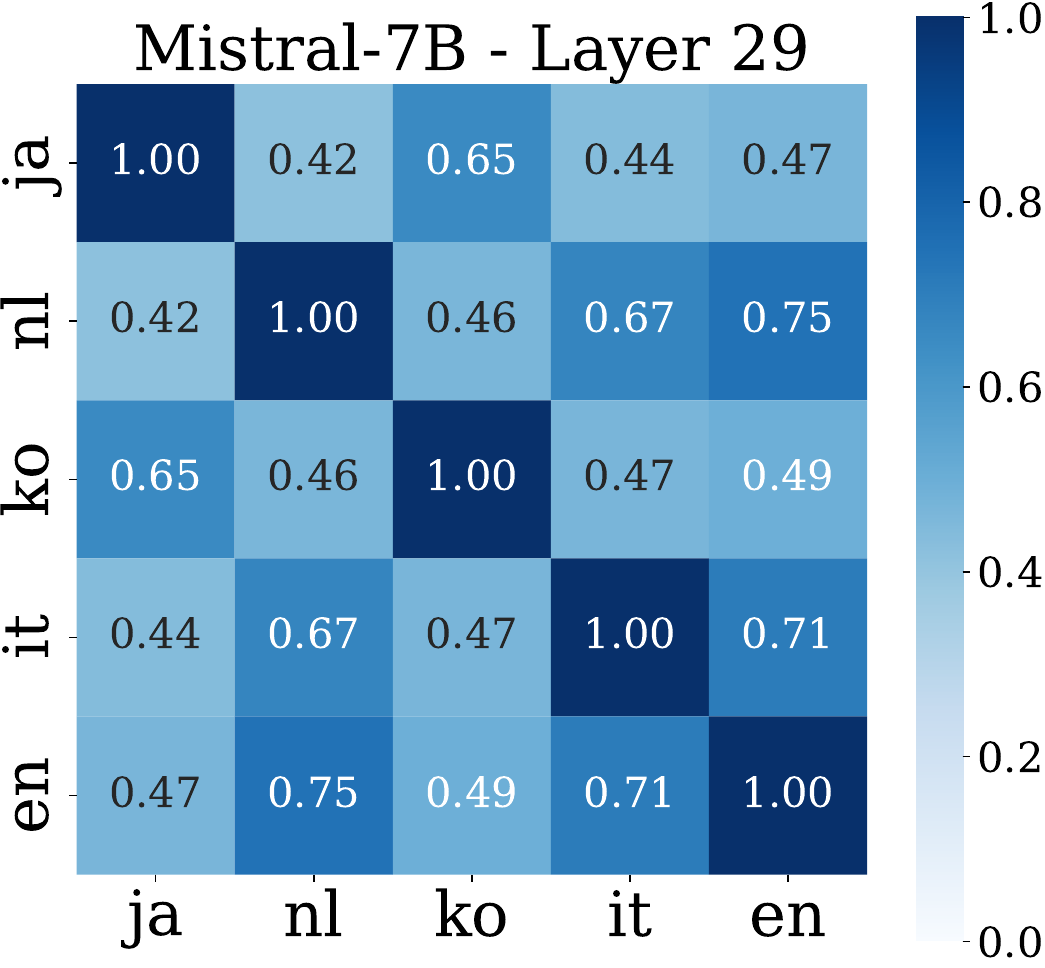}
  \includegraphics[width=0.19\linewidth]{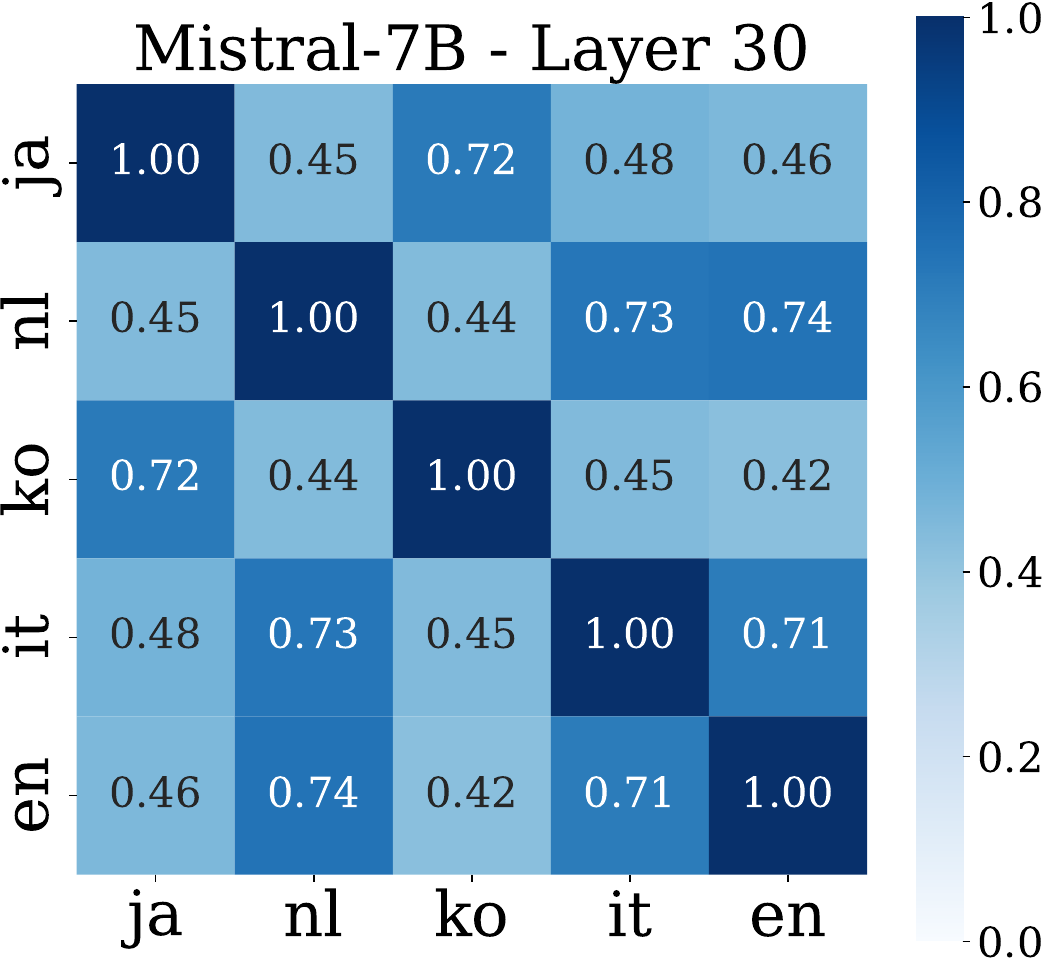}
  \includegraphics[width=0.19\linewidth]{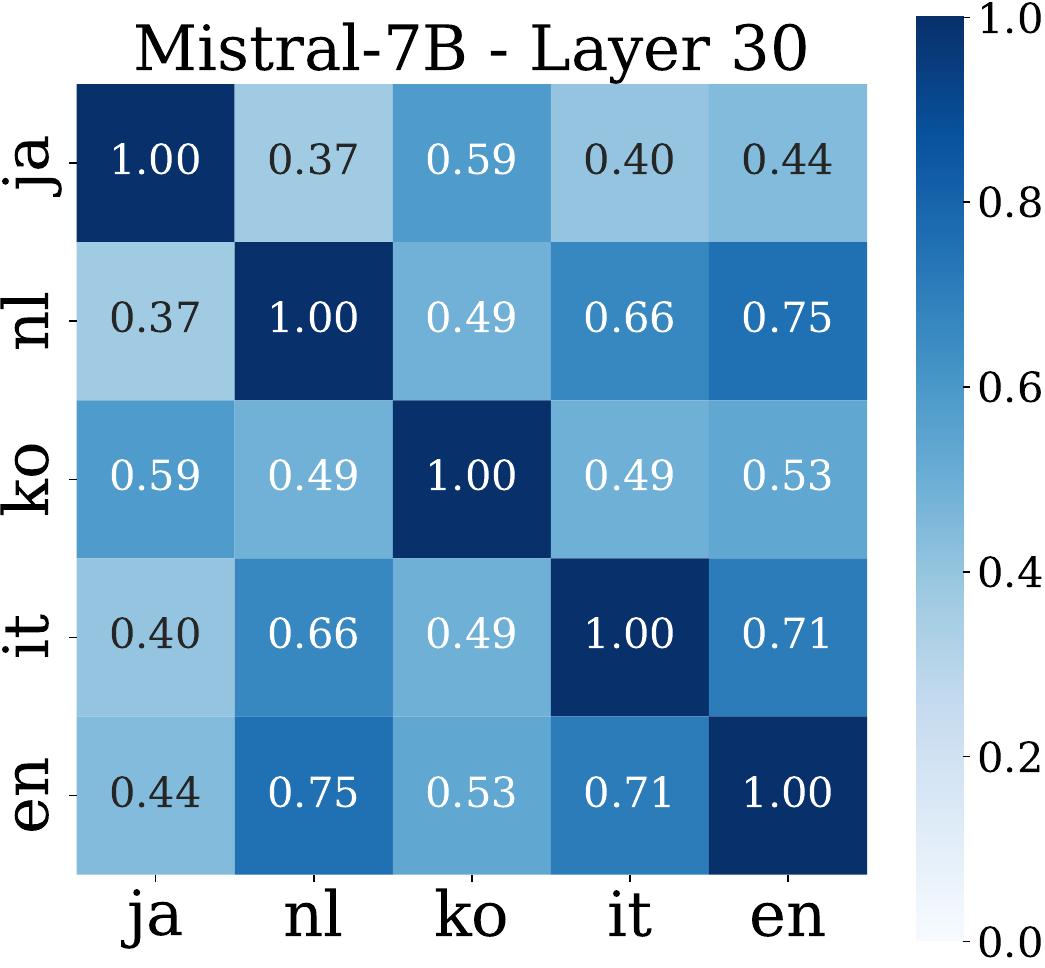}

  \begin{minipage}{0.19\linewidth}\centering \textbf{\textcolor{red}{layer 29 (Type-2)}}\end{minipage}
  \begin{minipage}{0.19\linewidth}\centering layer 29 (baseline)\end{minipage}
  \begin{minipage}{0.19\linewidth}\centering \textbf{\textcolor{red}{layer 30 (Type-2)}}\end{minipage}
  \begin{minipage}{0.19\linewidth}\centering layer 30 (baseline)\end{minipage}

  \includegraphics[width=0.19\linewidth]{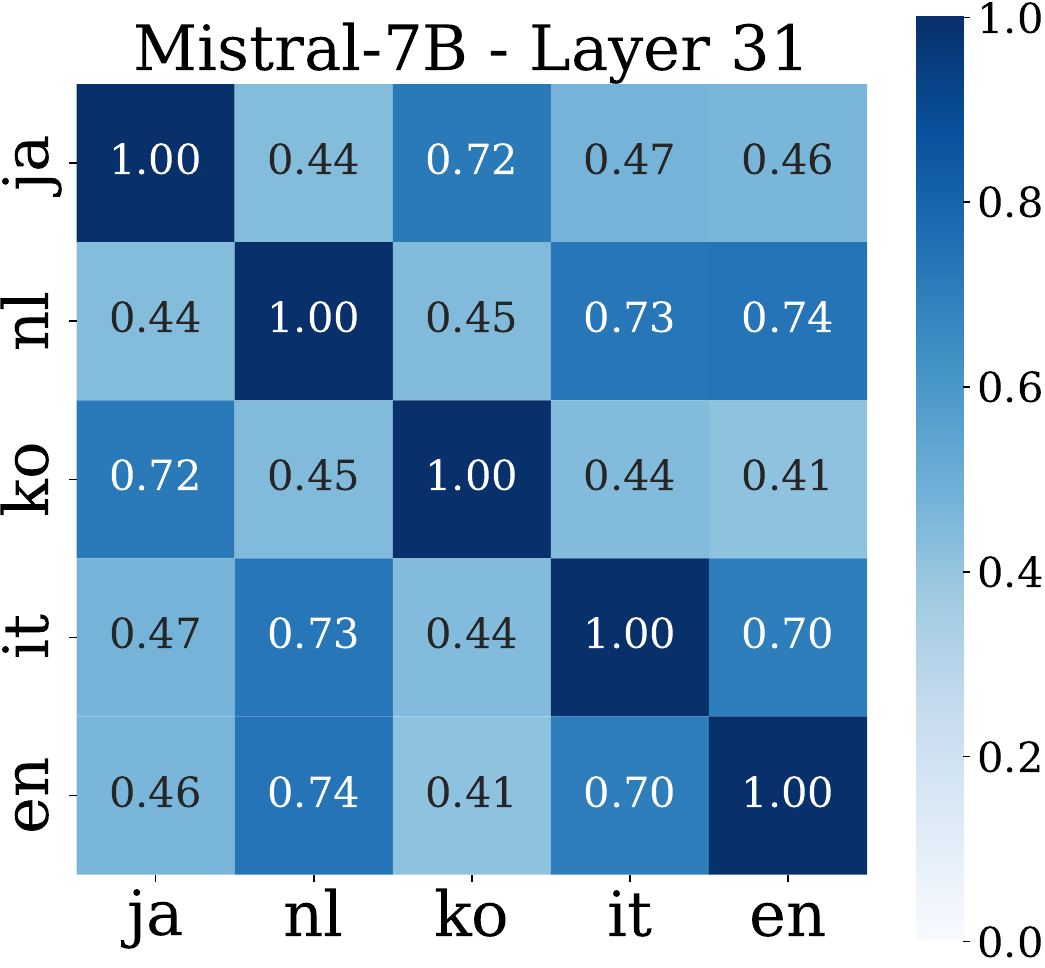}
  \includegraphics[width=0.19\linewidth]{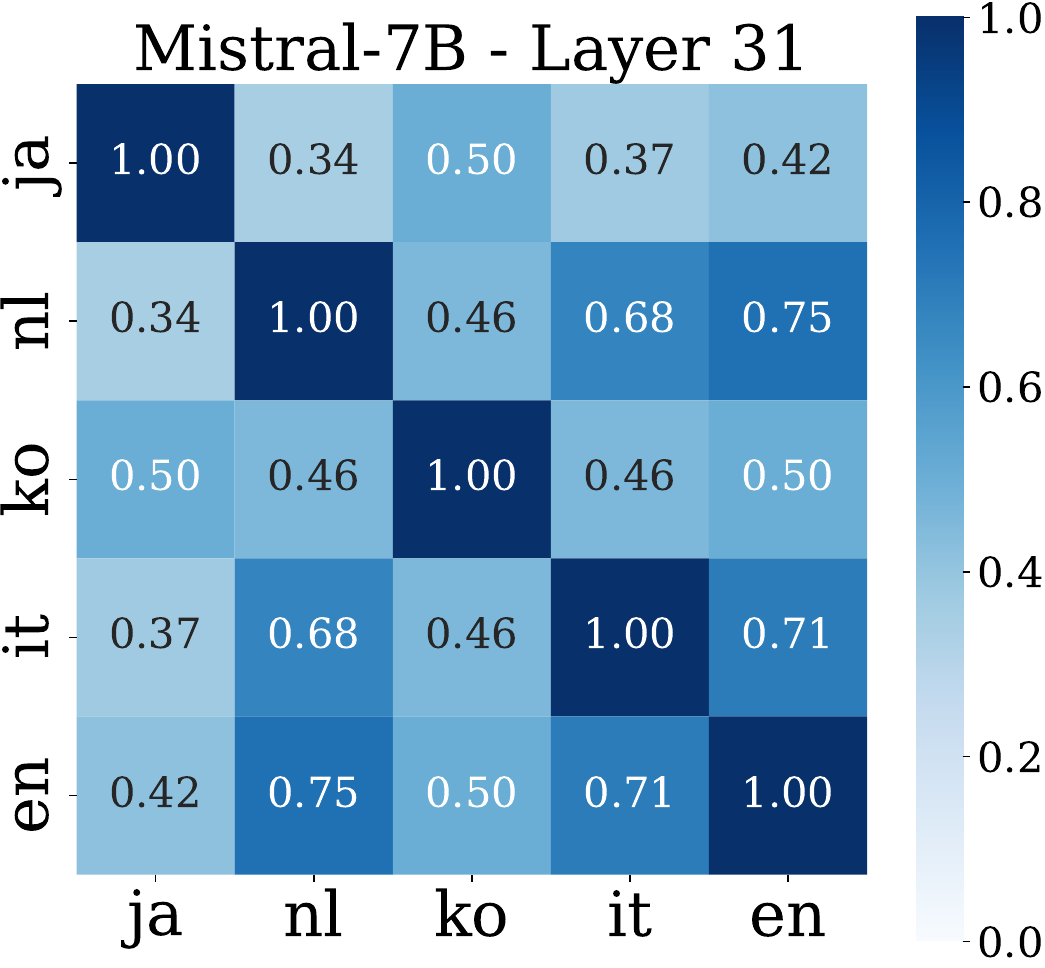}
  \includegraphics[width=0.19\linewidth]{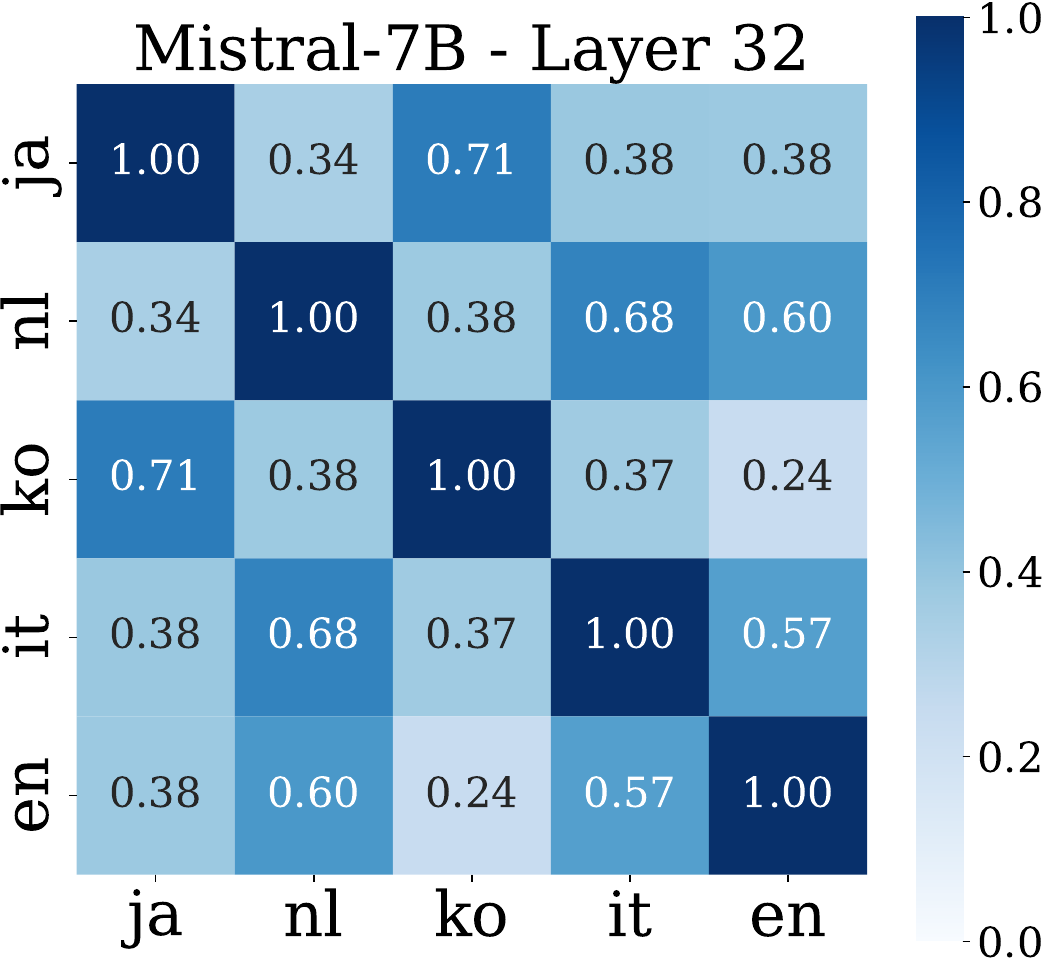}
  \includegraphics[width=0.19\linewidth]{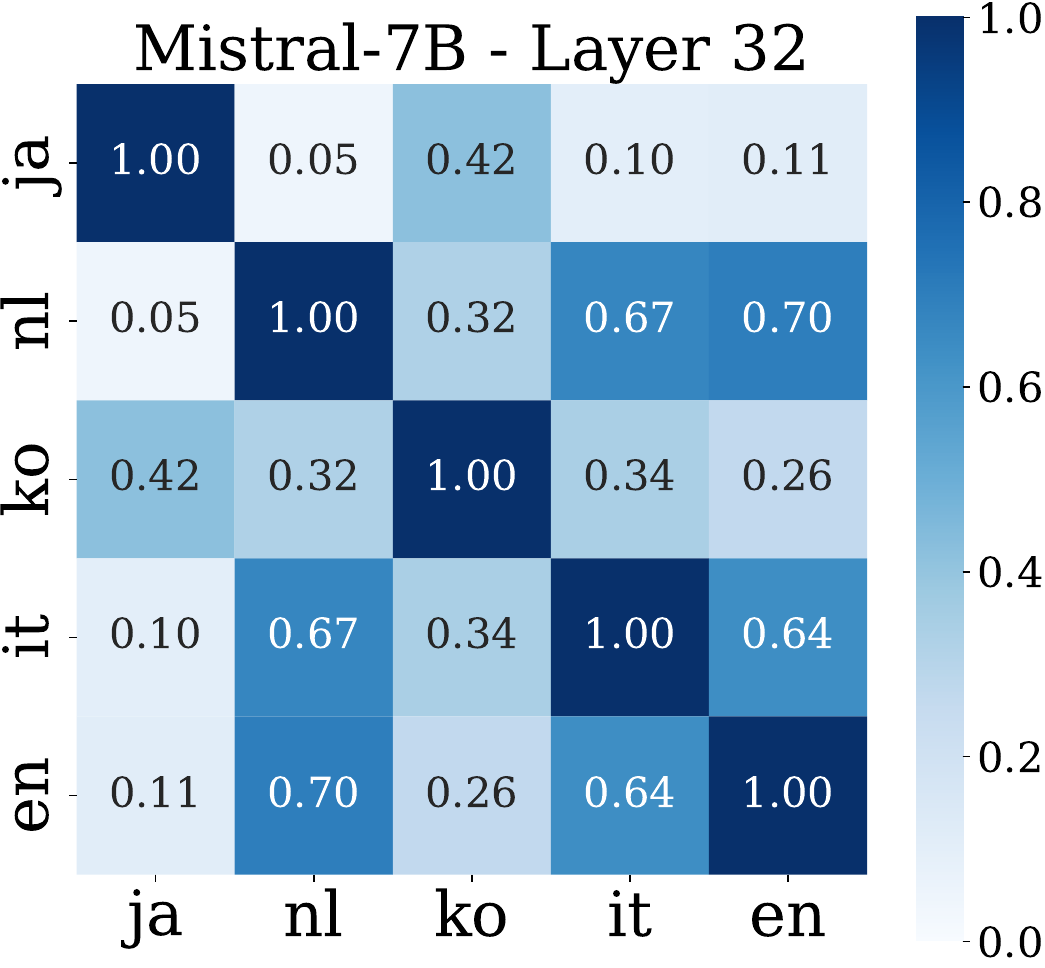}

  \begin{minipage}{0.19\linewidth}\centering \textbf{\textcolor{red}{layer 31 (Type-2)}}\end{minipage}
  \begin{minipage}{0.19\linewidth}\centering layer 31 (baseline)\end{minipage}
  \begin{minipage}{0.19\linewidth}\centering \textbf{\textcolor{red}{layer 32 (Type-2)}}\end{minipage}
  \begin{minipage}{0.19\linewidth}\centering layer 32 (baseline)\end{minipage}

  \caption{\textbf{Distance among centroids of language latent spaces while deactivating top-1k Type-2 Transfer Neurons (Mistral-7B)}.}
  \label{fig:appendix:centroids distance among language subspaces while deactivating type2 mistral}
\end{figure*}

\begin{figure*}[t]
  \centering

  \includegraphics[width=0.19\linewidth]{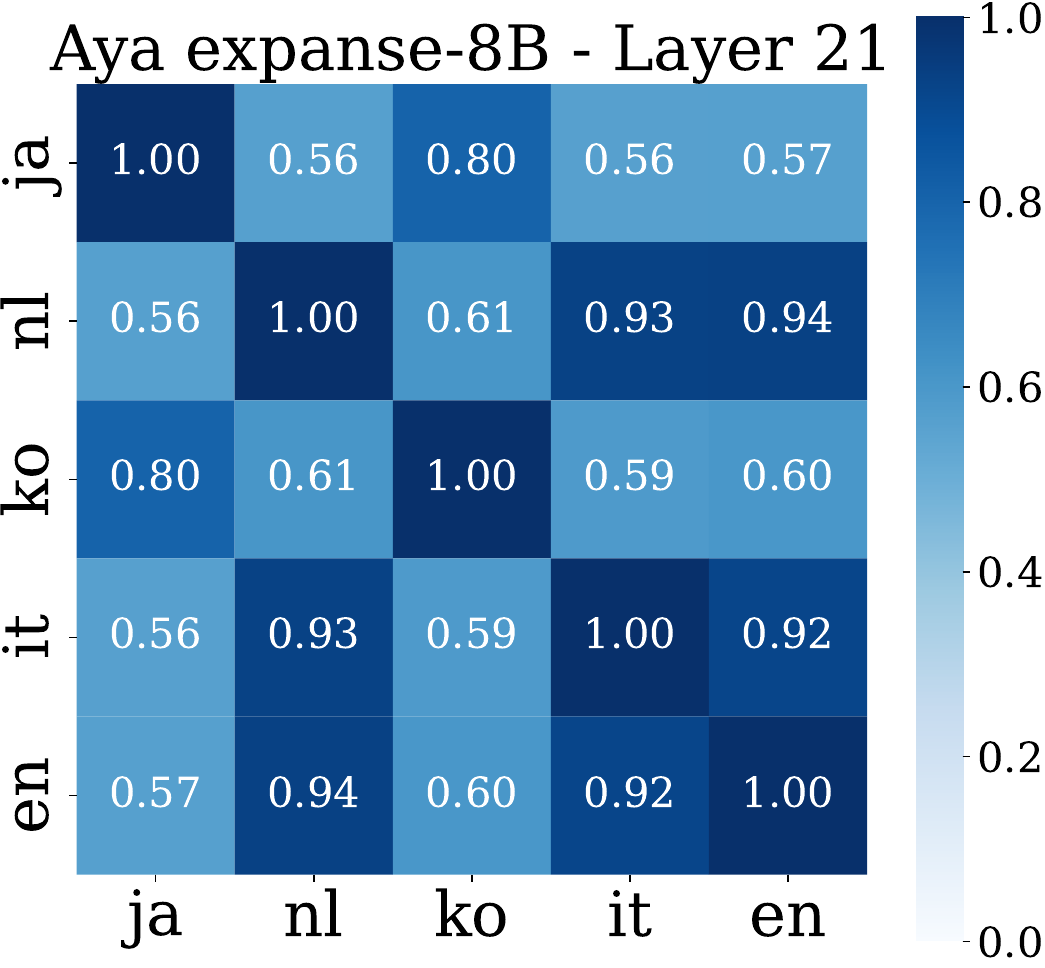}
  \includegraphics[width=0.19\linewidth]{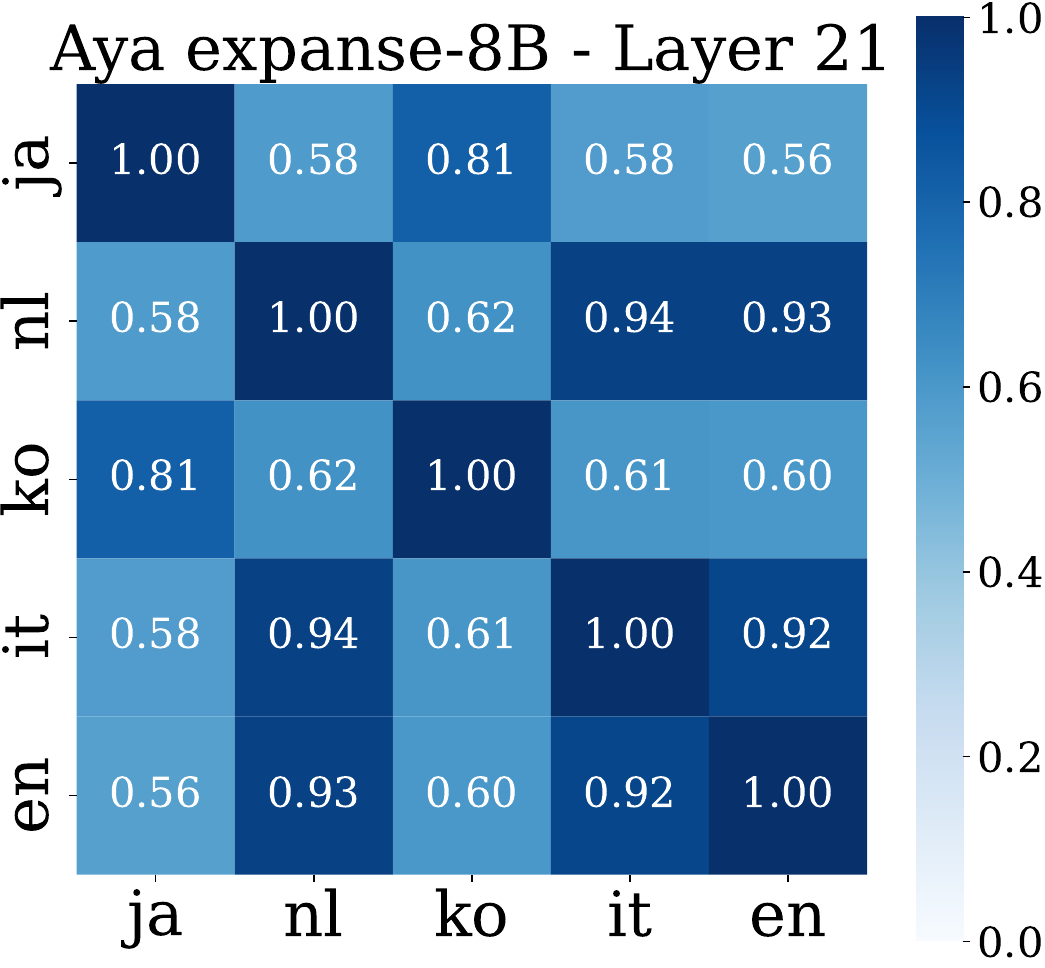}
  \includegraphics[width=0.19\linewidth]{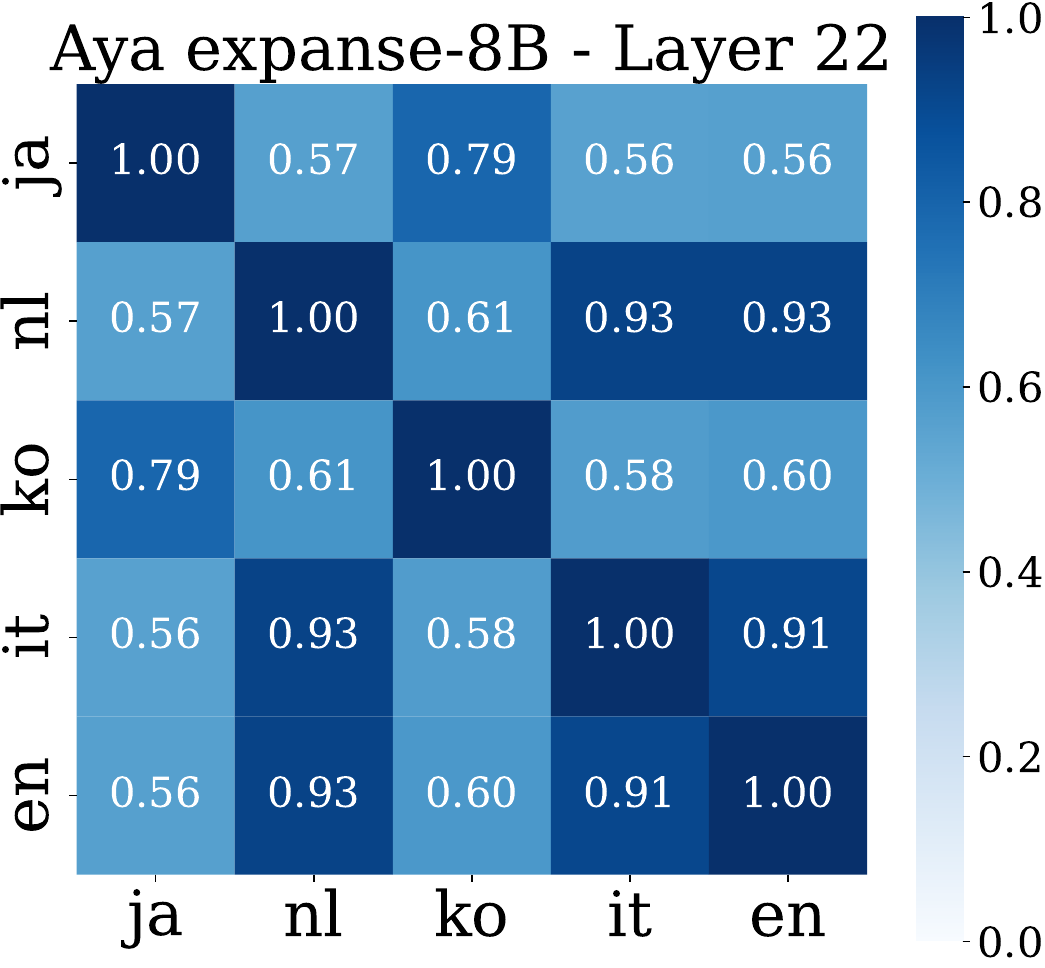}
  \includegraphics[width=0.19\linewidth]{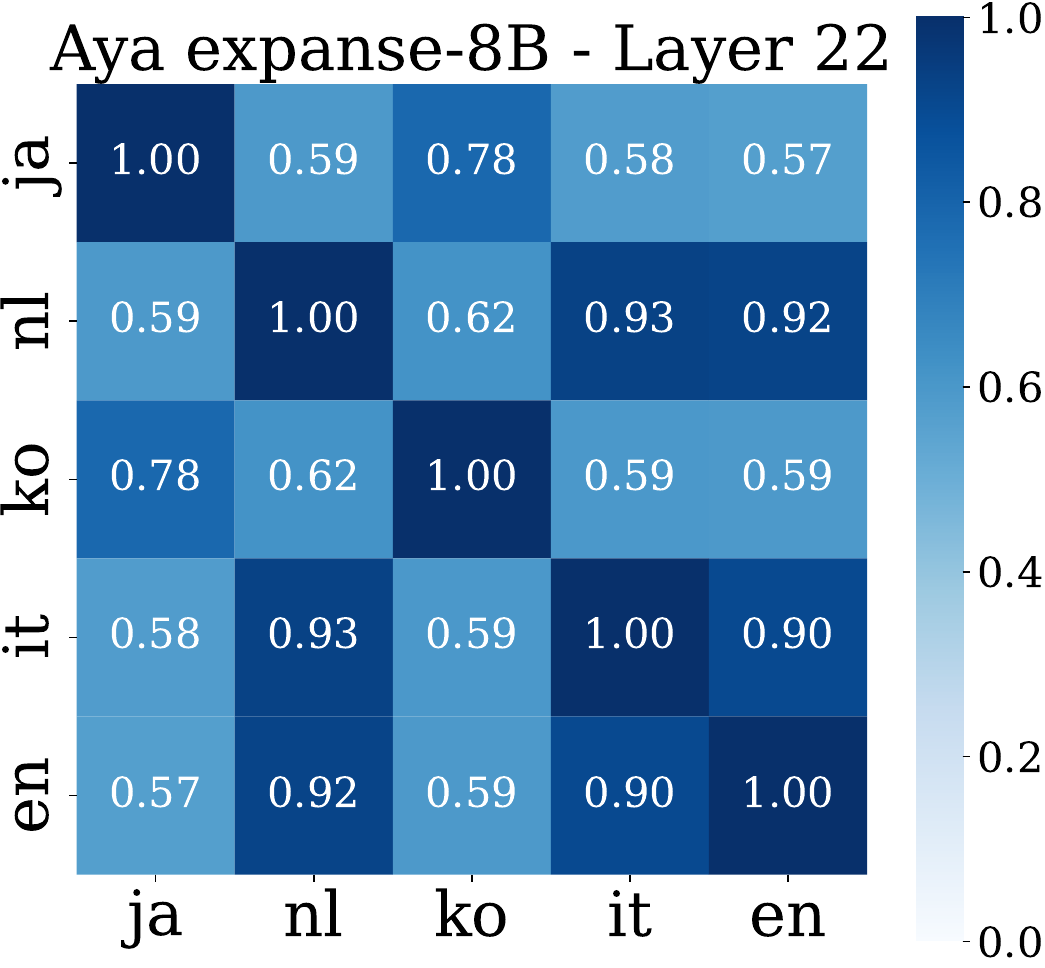}

  \begin{minipage}{0.19\linewidth}\centering \textbf{\textcolor{red}{layer 21 (Type-2)}}\end{minipage}
  \begin{minipage}{0.19\linewidth}\centering layer 21 (baseline)\end{minipage}
  \begin{minipage}{0.19\linewidth}\centering \textbf{\textcolor{red}{layer 22 (Type-2)}}\end{minipage}
  \begin{minipage}{0.19\linewidth}\centering layer 22 (baseline)\end{minipage}

  \includegraphics[width=0.19\linewidth]{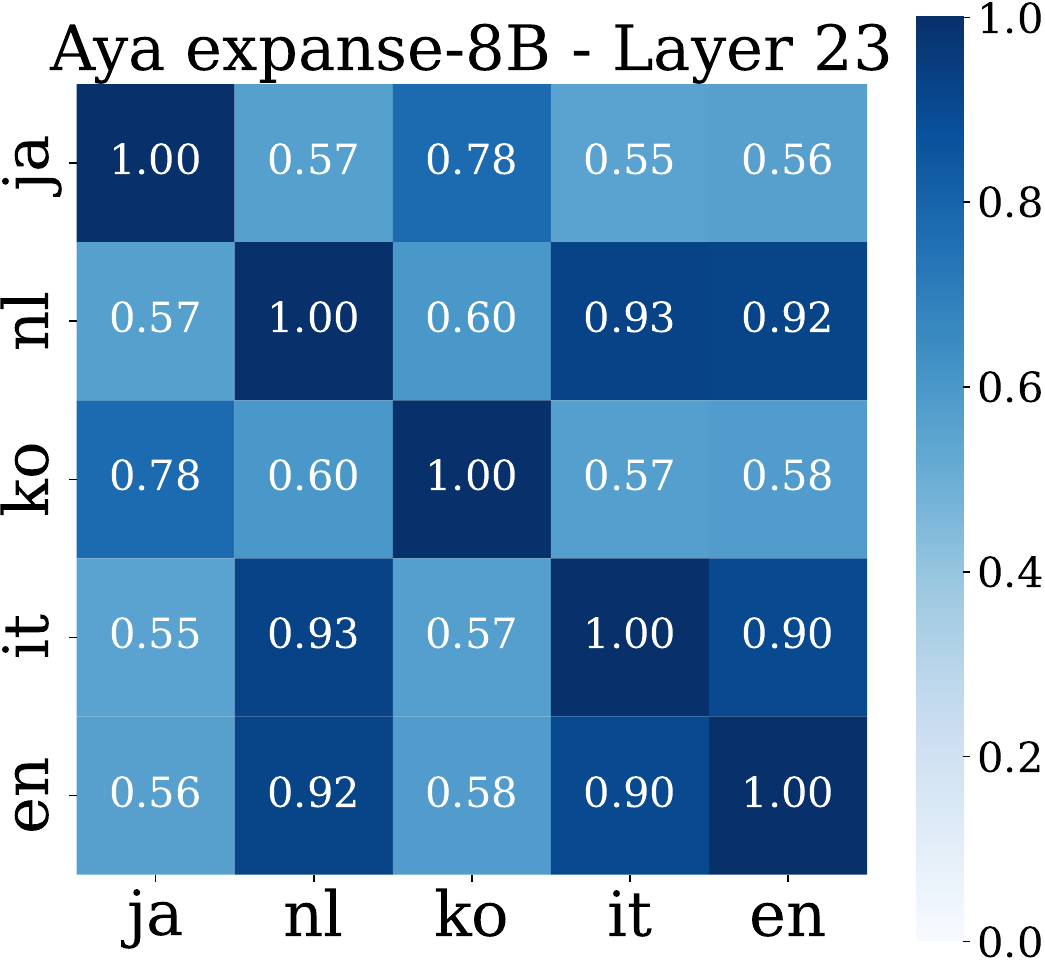}
  \includegraphics[width=0.19\linewidth]{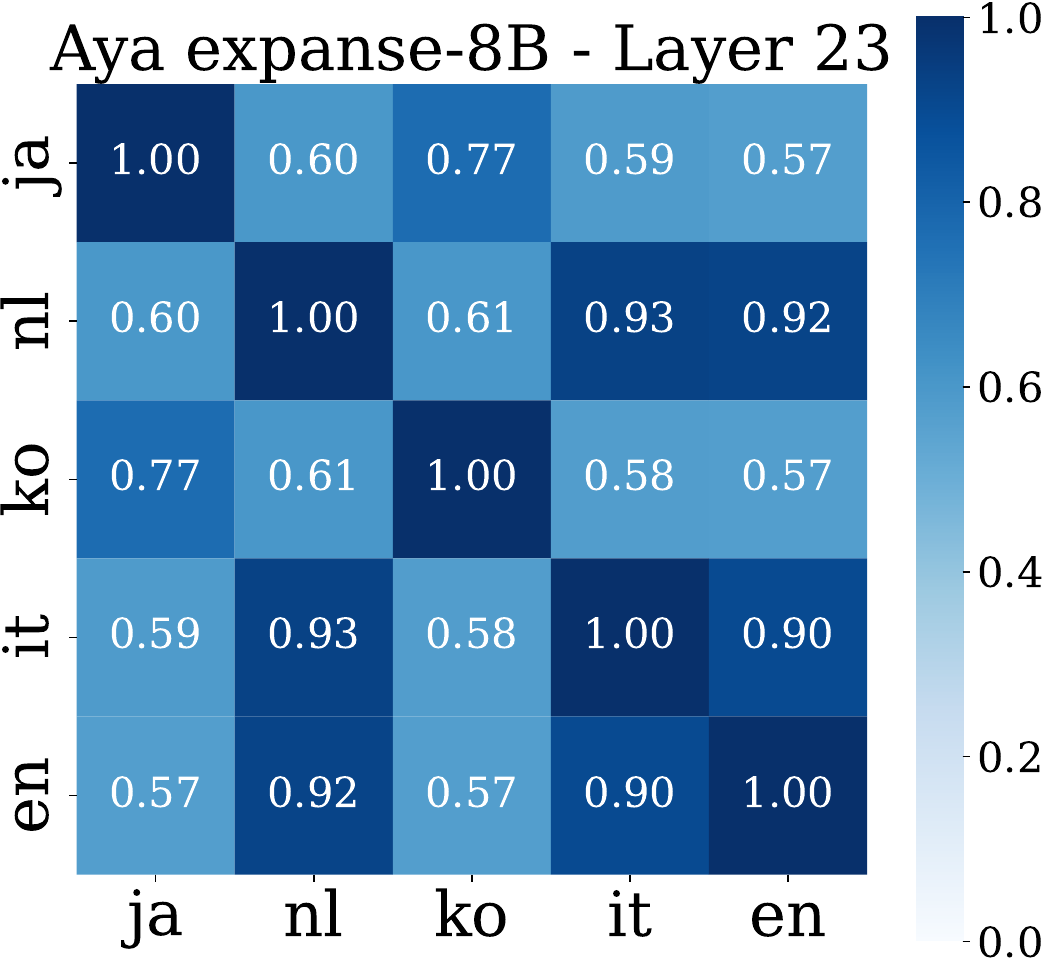}
  \includegraphics[width=0.19\linewidth]{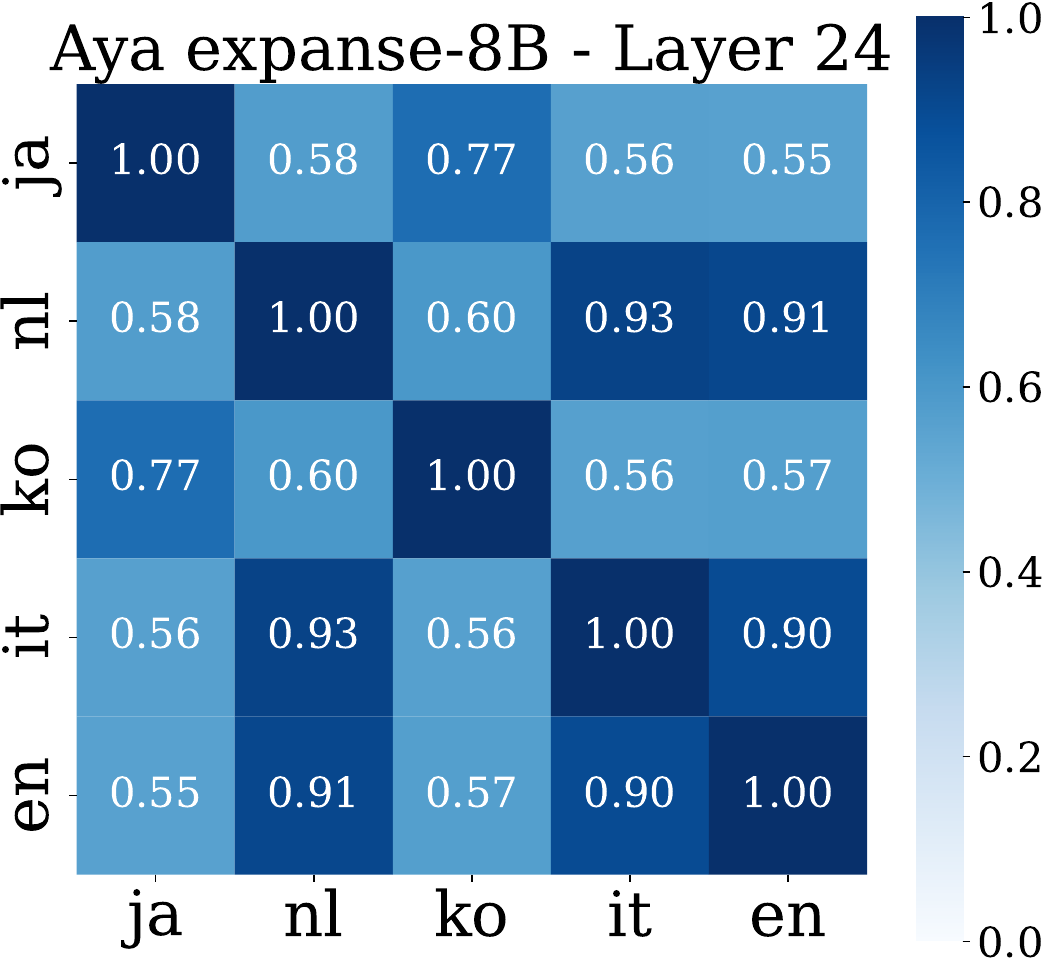}
  \includegraphics[width=0.19\linewidth]{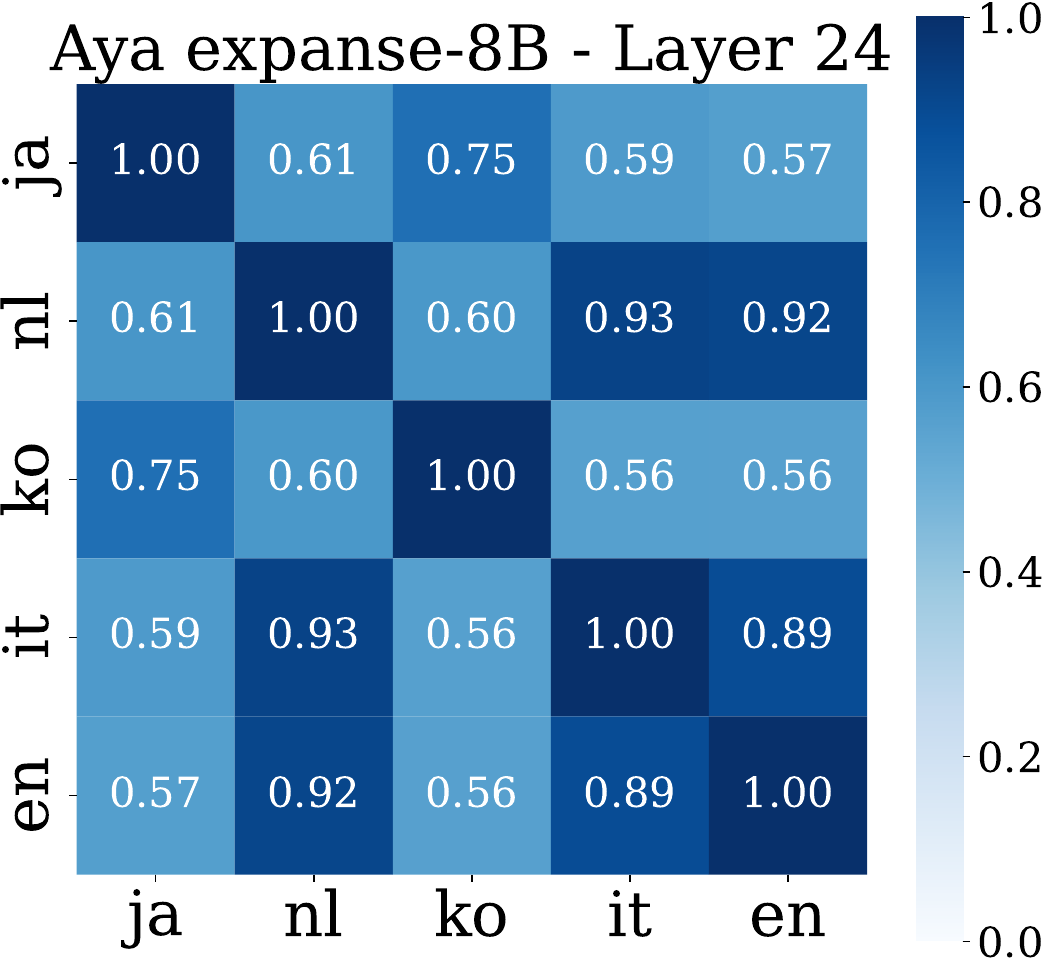}

  \begin{minipage}{0.19\linewidth}\centering \textbf{\textcolor{red}{layer 23 (Type-2)}}\end{minipage}
  \begin{minipage}{0.19\linewidth}\centering layer 23 (baseline)\end{minipage}
  \begin{minipage}{0.19\linewidth}\centering \textbf{\textcolor{red}{layer 24 (Type-2)}}\end{minipage}
  \begin{minipage}{0.19\linewidth}\centering layer 24 (baseline)\end{minipage}

  \includegraphics[width=0.19\linewidth]{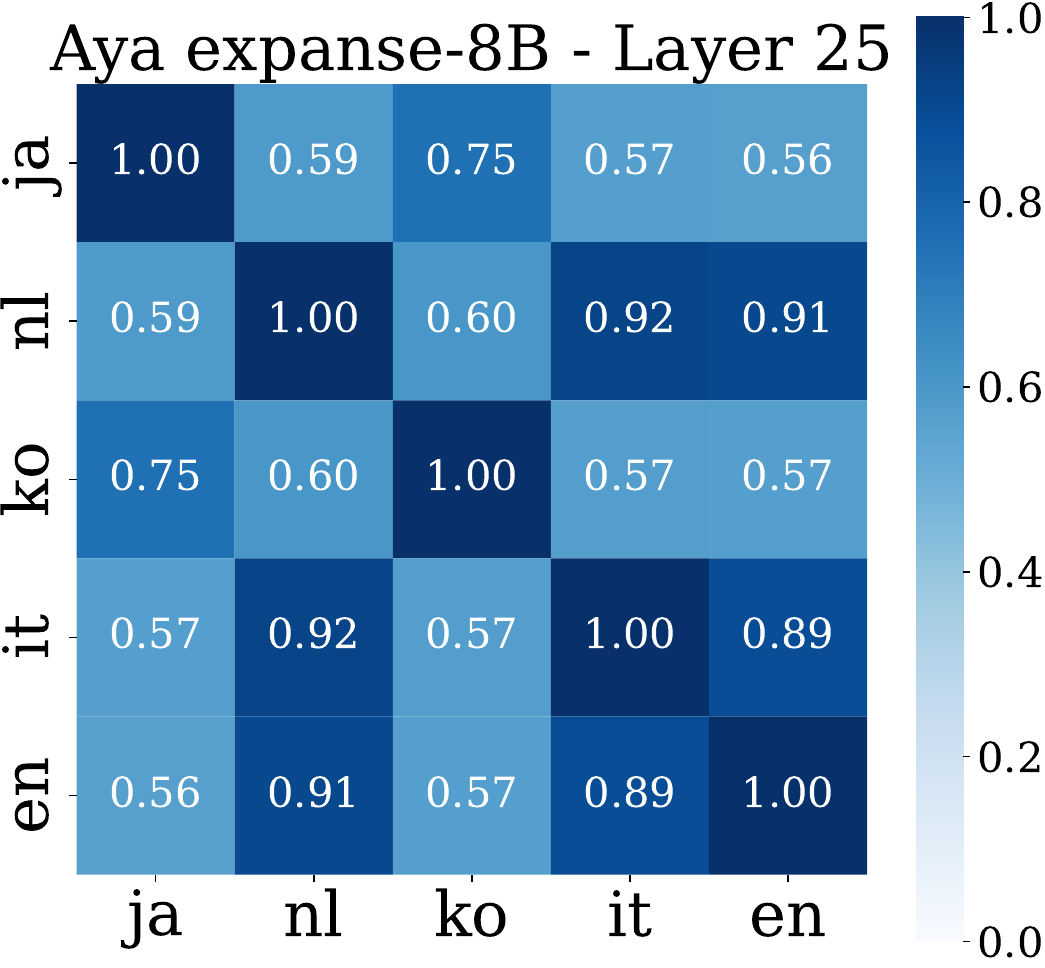}
  \includegraphics[width=0.19\linewidth]{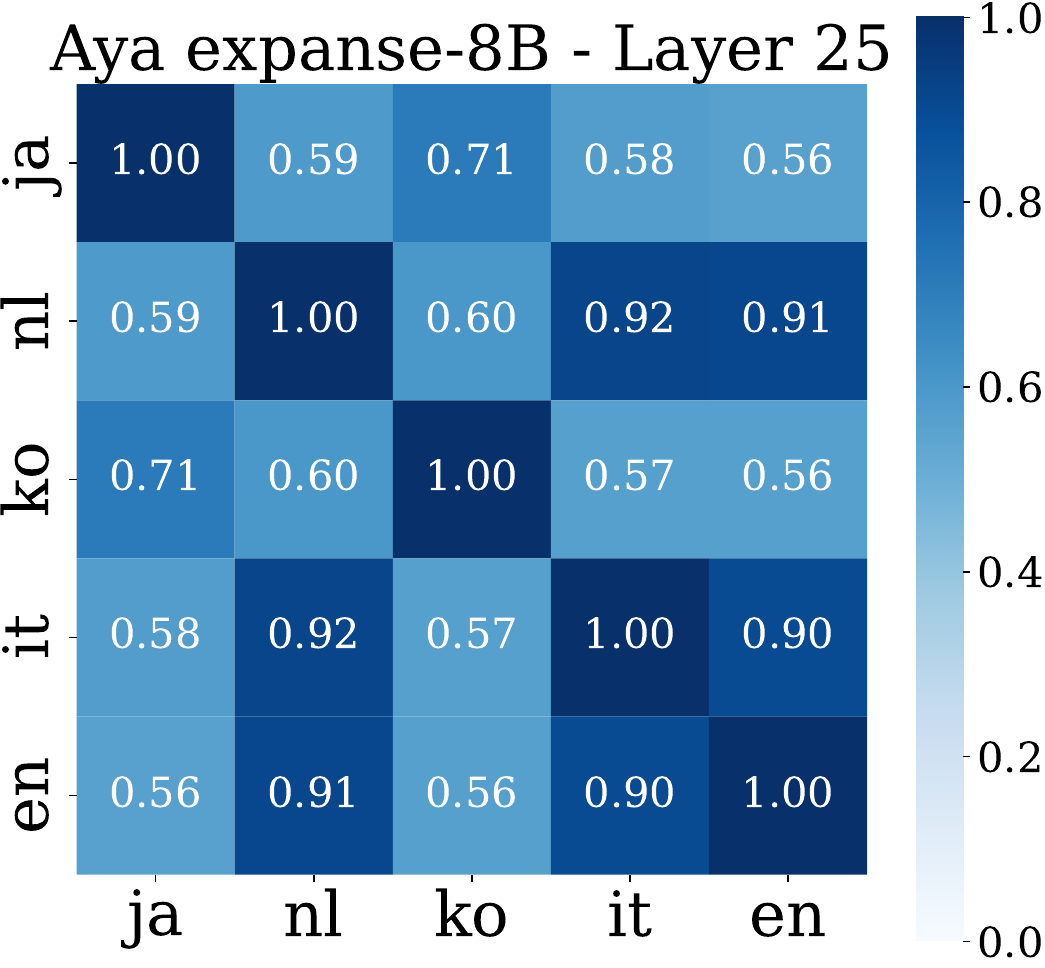}
  \includegraphics[width=0.19\linewidth]{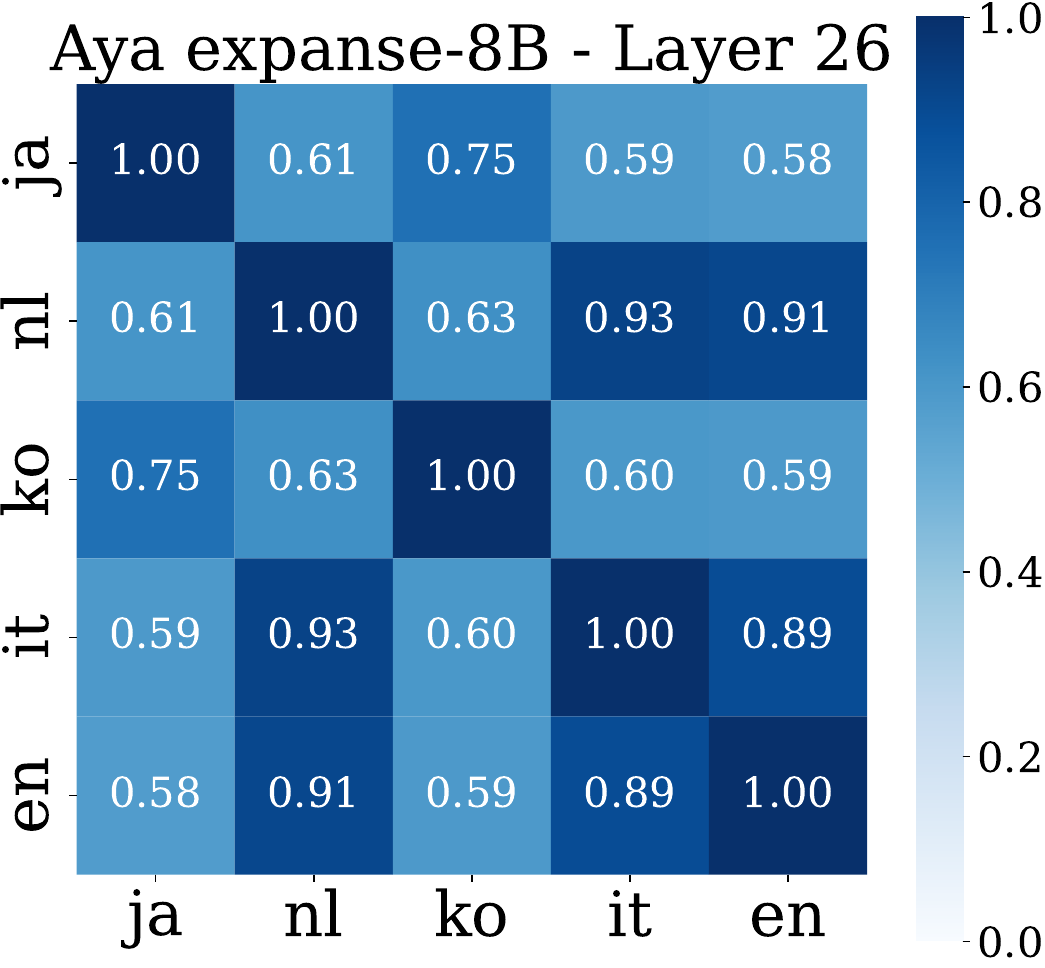}
  \includegraphics[width=0.19\linewidth]{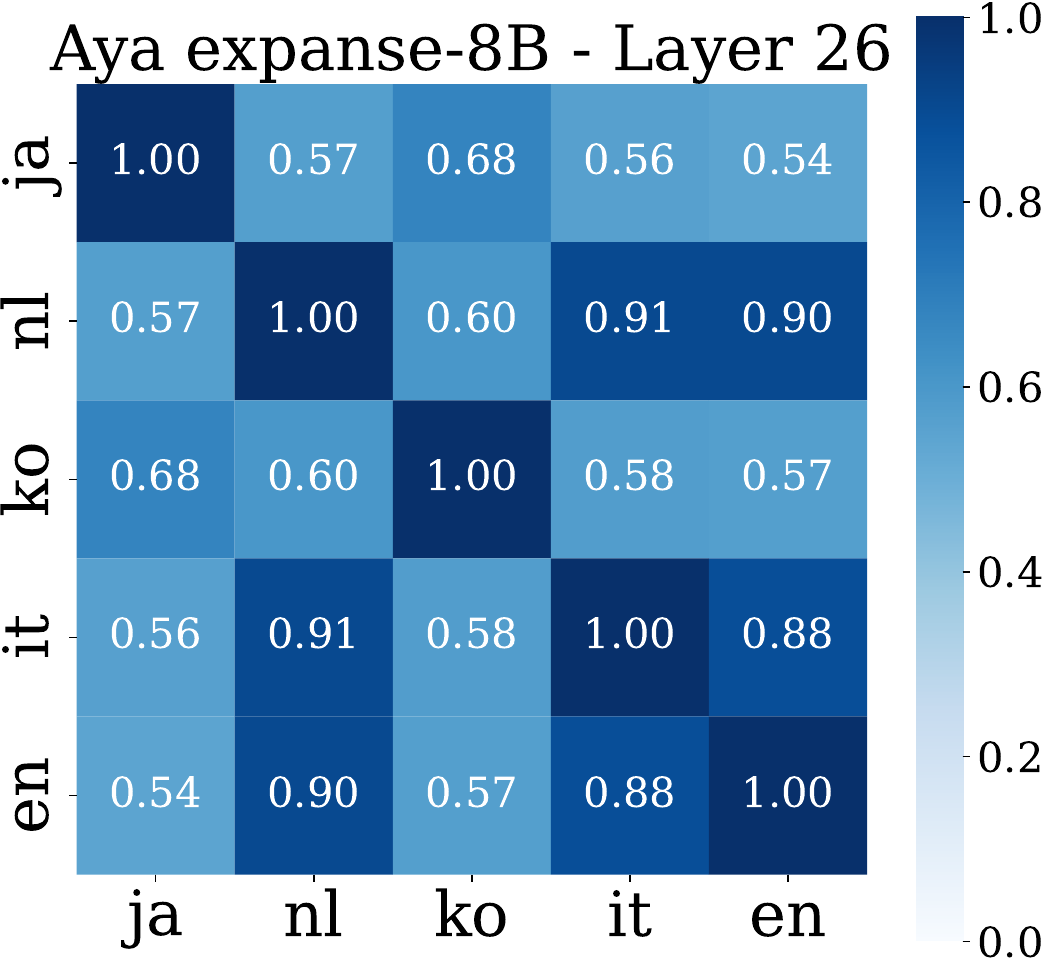}
  
  \begin{minipage}{0.19\linewidth}\centering \textbf{\textcolor{red}{layer 25 (Type-2)}}\end{minipage}
  \begin{minipage}{0.19\linewidth}\centering layer 25 (baseline)\end{minipage}
  \begin{minipage}{0.19\linewidth}\centering \textbf{\textcolor{red}{layer 26 (Type-2)}}\end{minipage}
  \begin{minipage}{0.19\linewidth}\centering layer 26 (baseline)\end{minipage}

  \includegraphics[width=0.19\linewidth]{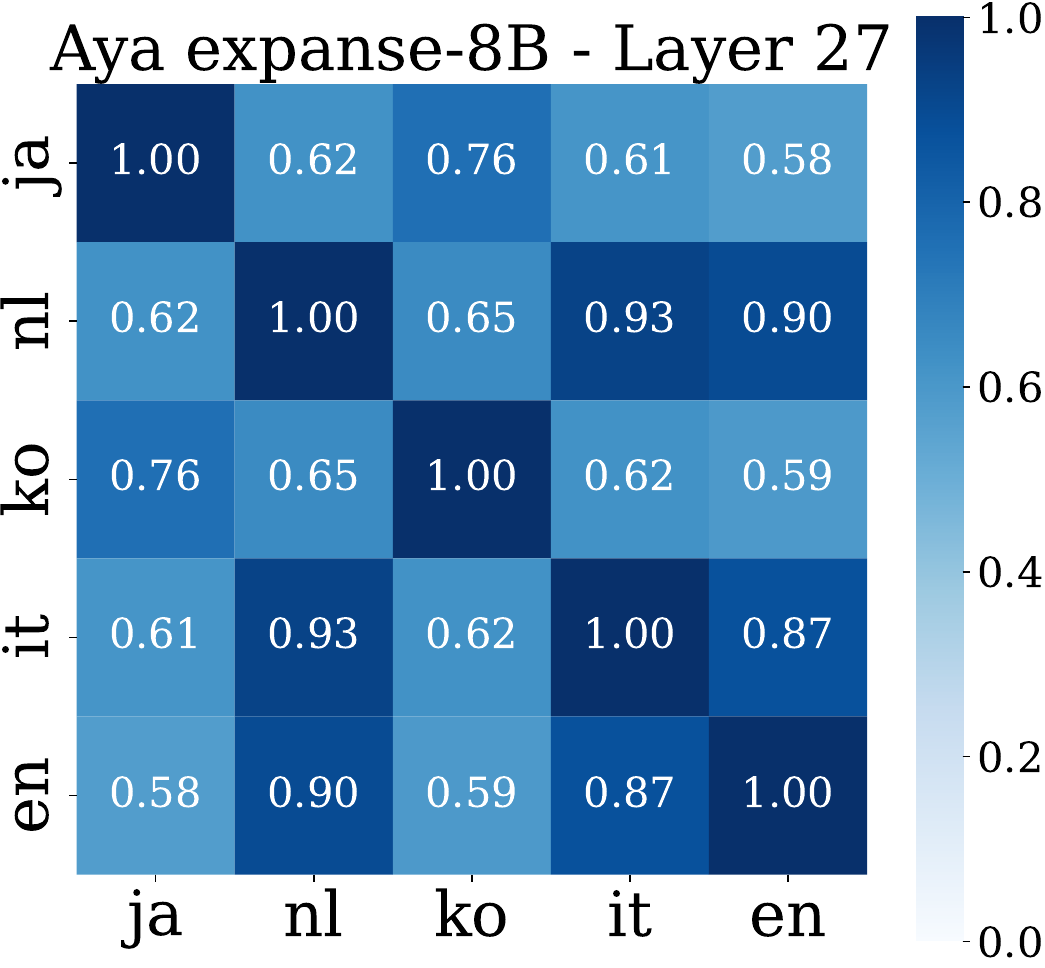}
  \includegraphics[width=0.19\linewidth]{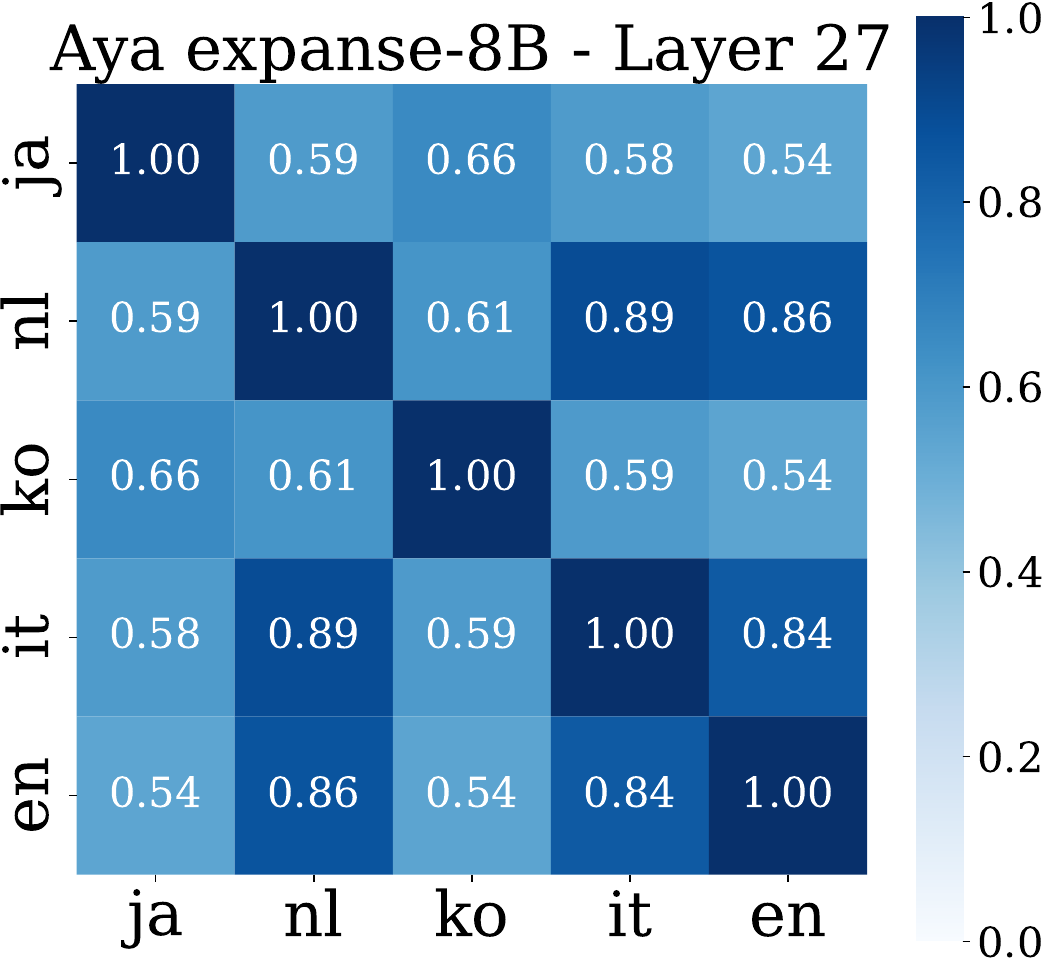}
  \includegraphics[width=0.19\linewidth]{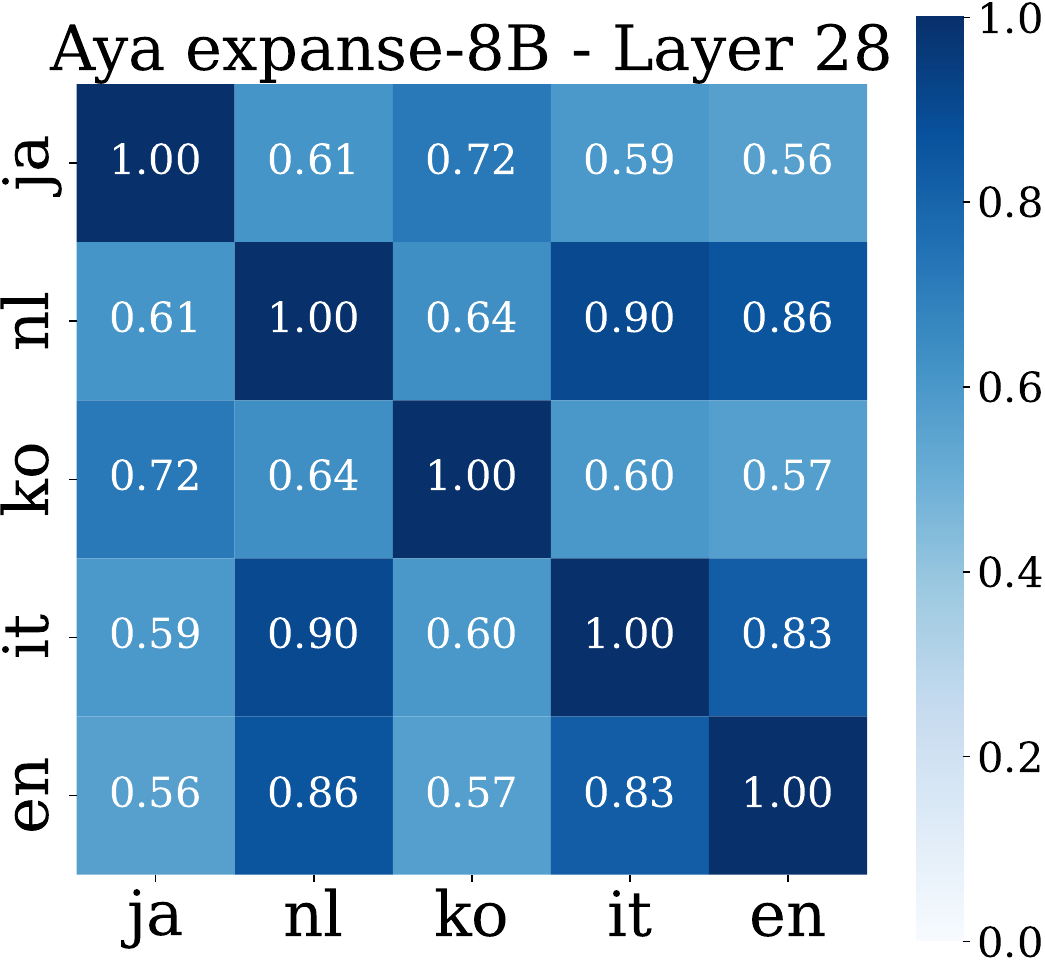}
  \includegraphics[width=0.19\linewidth]{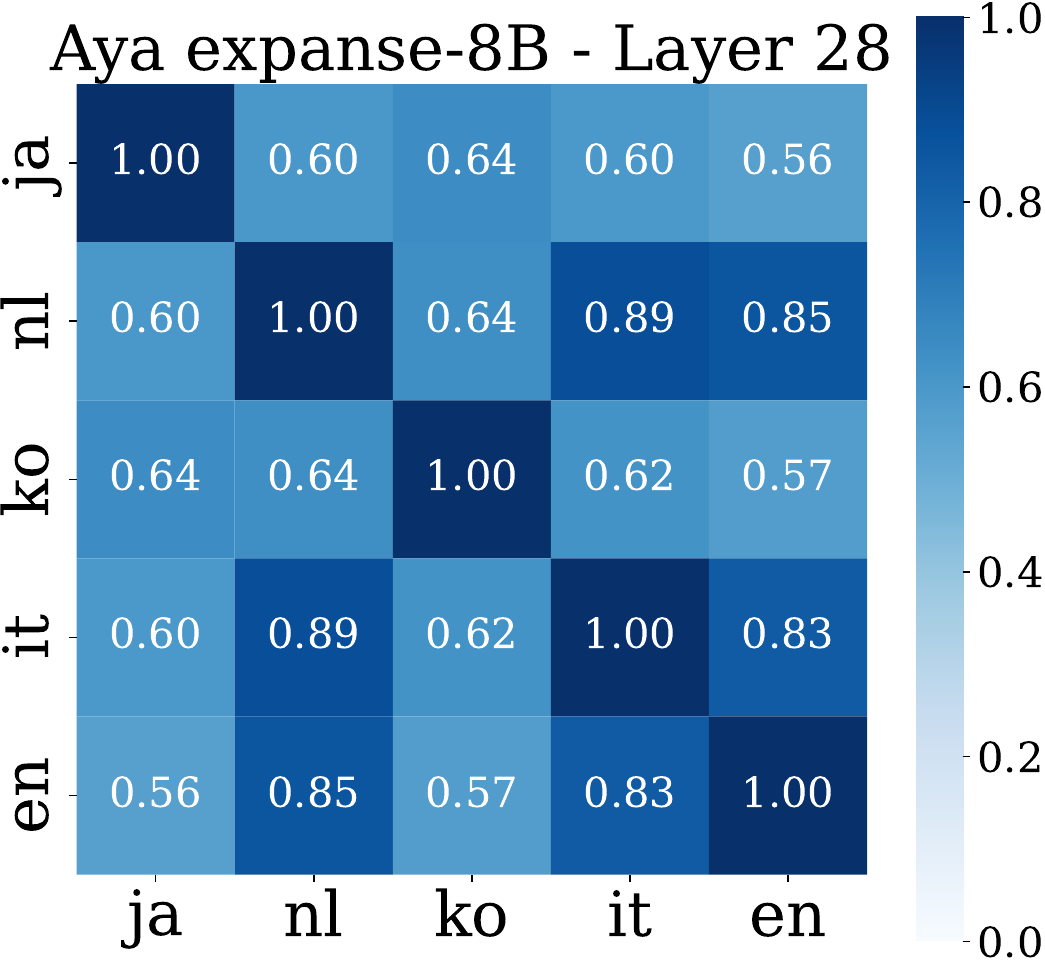}

  \begin{minipage}{0.19\linewidth}\centering \textbf{\textcolor{red}{layer 27 (Type-2)}}\end{minipage}
  \begin{minipage}{0.19\linewidth}\centering layer 27 (baseline)\end{minipage}
  \begin{minipage}{0.19\linewidth}\centering \textbf{\textcolor{red}{layer 28 (Type-2)}}\end{minipage}
  \begin{minipage}{0.19\linewidth}\centering layer 28 (baseline)\end{minipage}

  \includegraphics[width=0.19\linewidth]{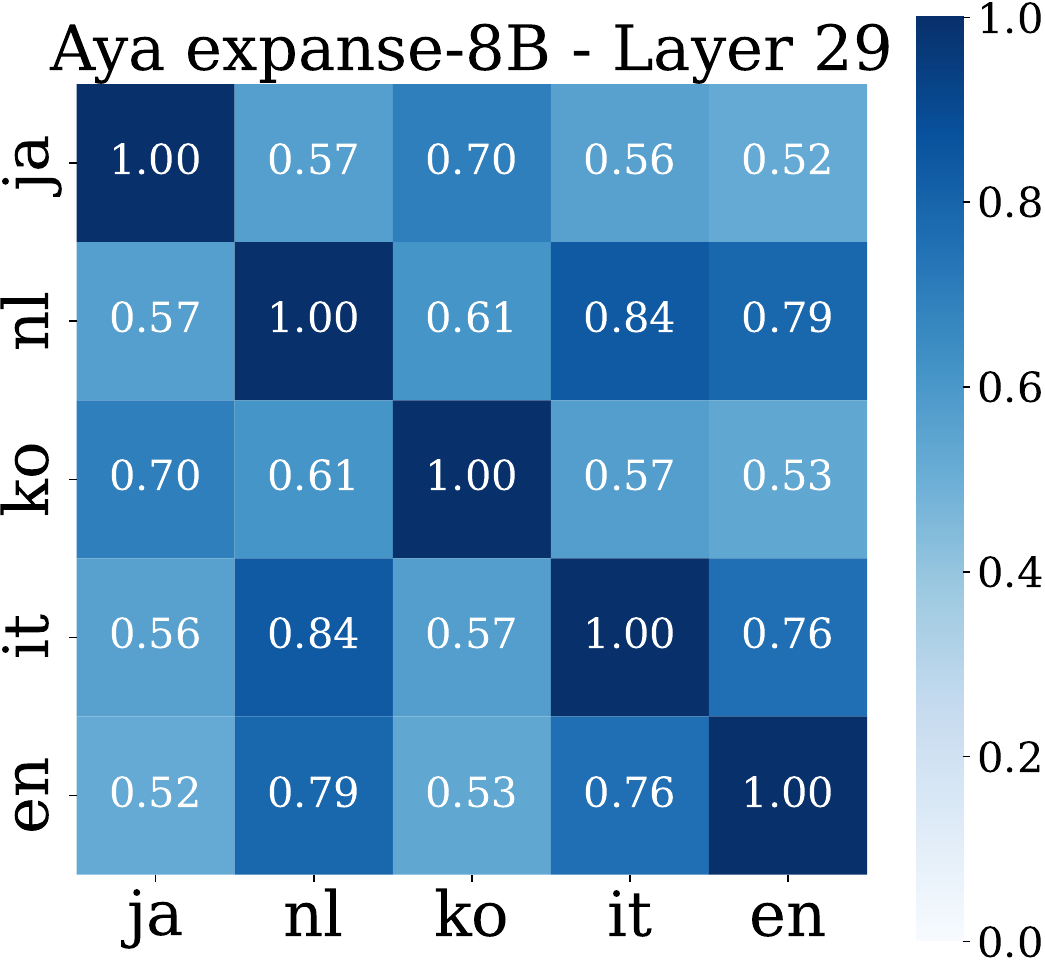}
  \includegraphics[width=0.19\linewidth]{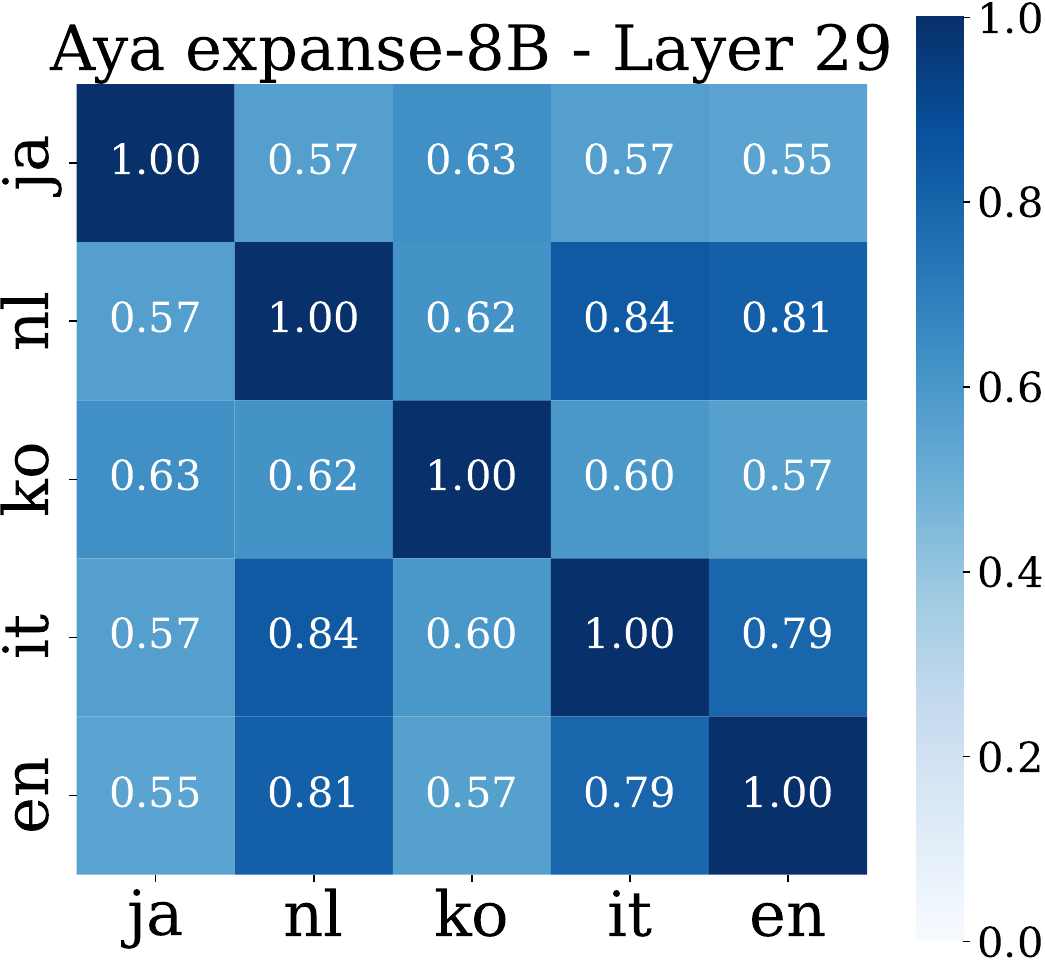}
  \includegraphics[width=0.19\linewidth]{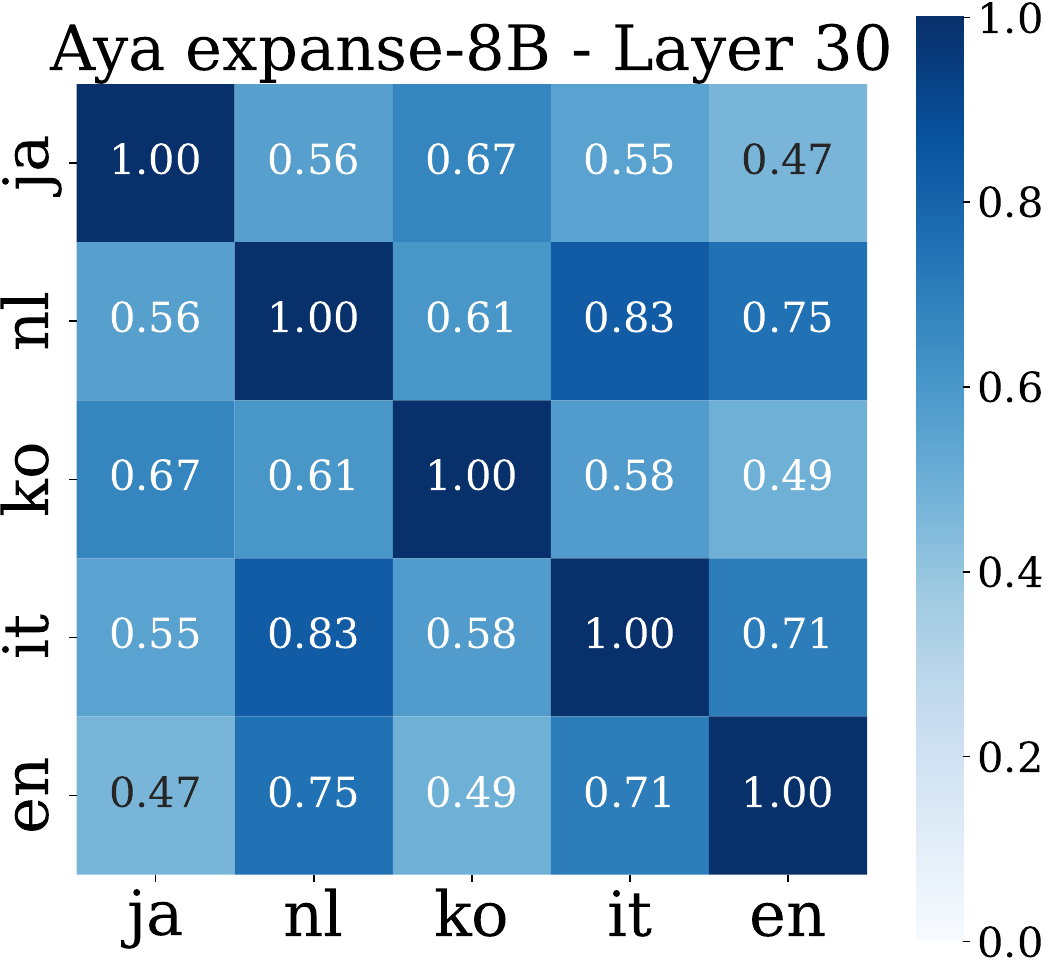}
  \includegraphics[width=0.19\linewidth]{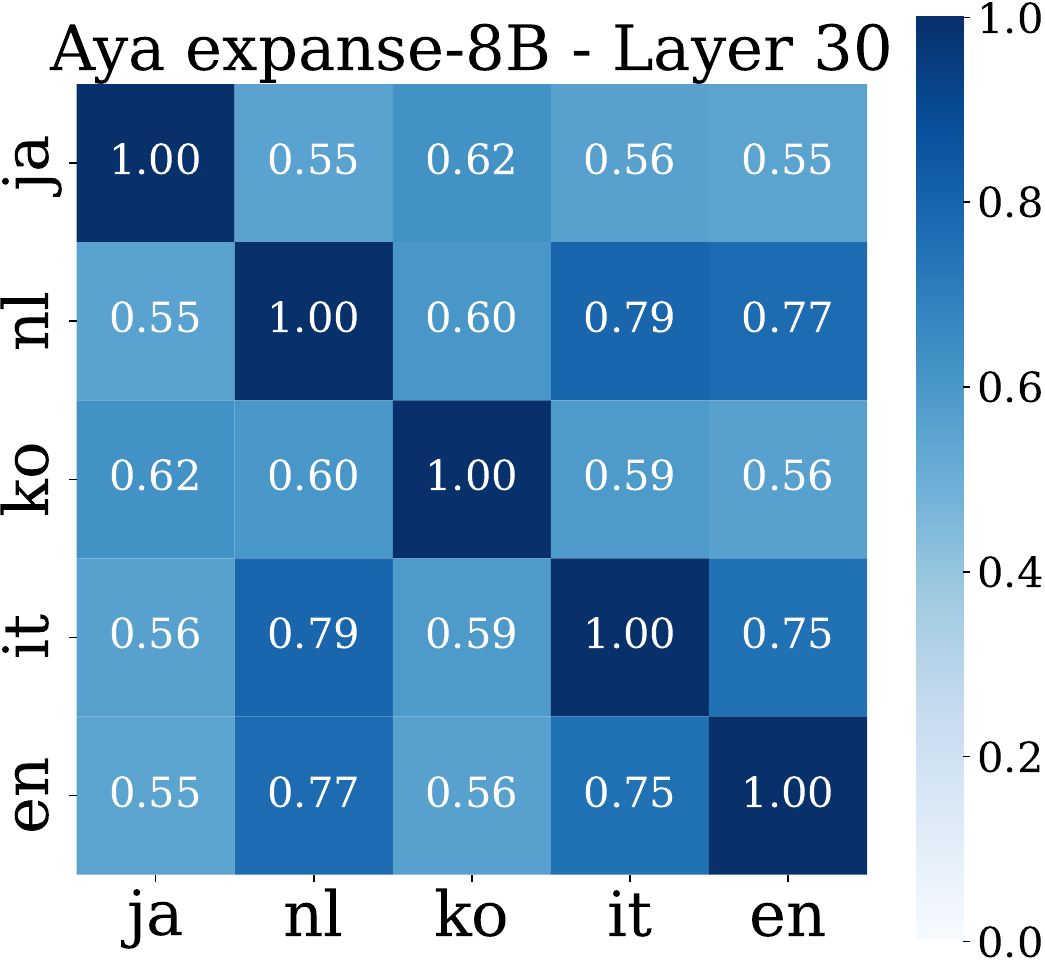}

  \begin{minipage}{0.19\linewidth}\centering \textbf{\textcolor{red}{layer 29 (Type-2)}}\end{minipage}
  \begin{minipage}{0.19\linewidth}\centering layer 29 (baseline)\end{minipage}
  \begin{minipage}{0.19\linewidth}\centering \textbf{\textcolor{red}{layer 30 (Type-2)}}\end{minipage}
  \begin{minipage}{0.19\linewidth}\centering layer 30 (baseline)\end{minipage}

  \includegraphics[width=0.19\linewidth]{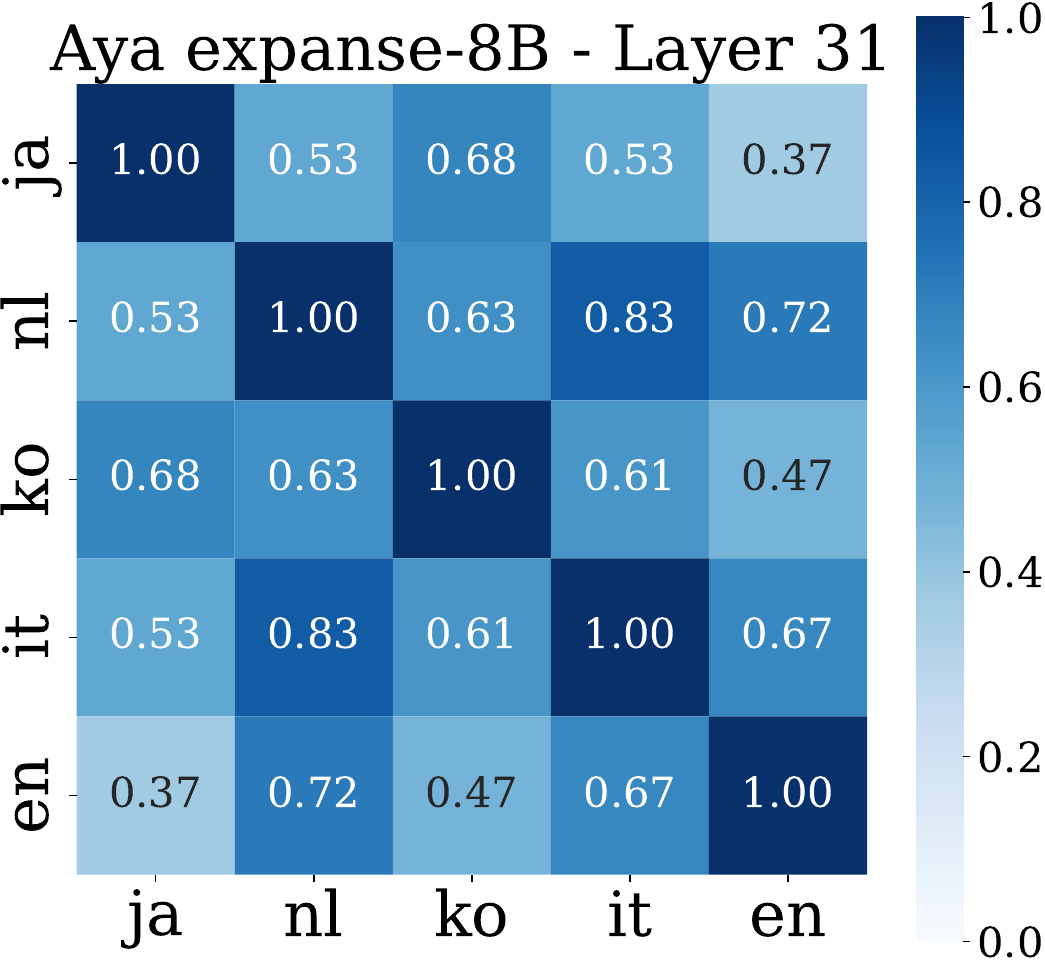}
  \includegraphics[width=0.19\linewidth]{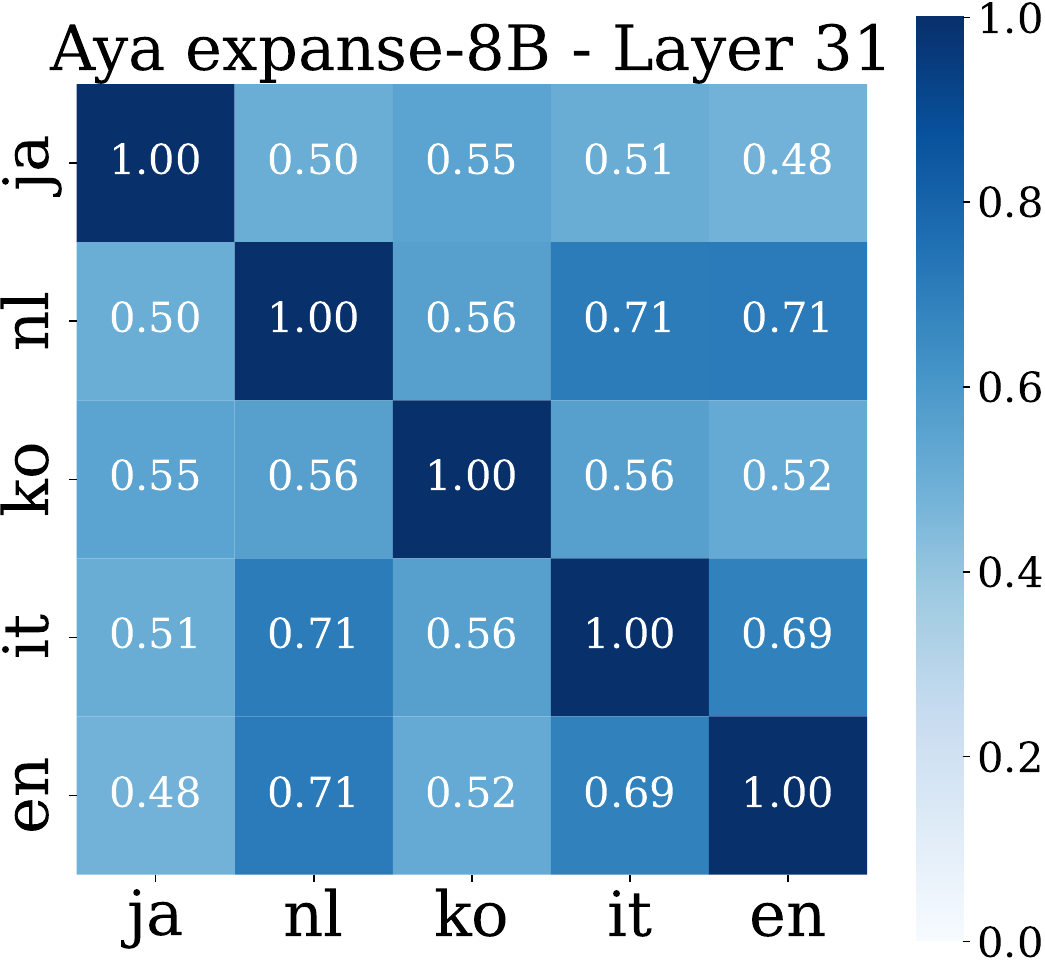}
  \includegraphics[width=0.19\linewidth]{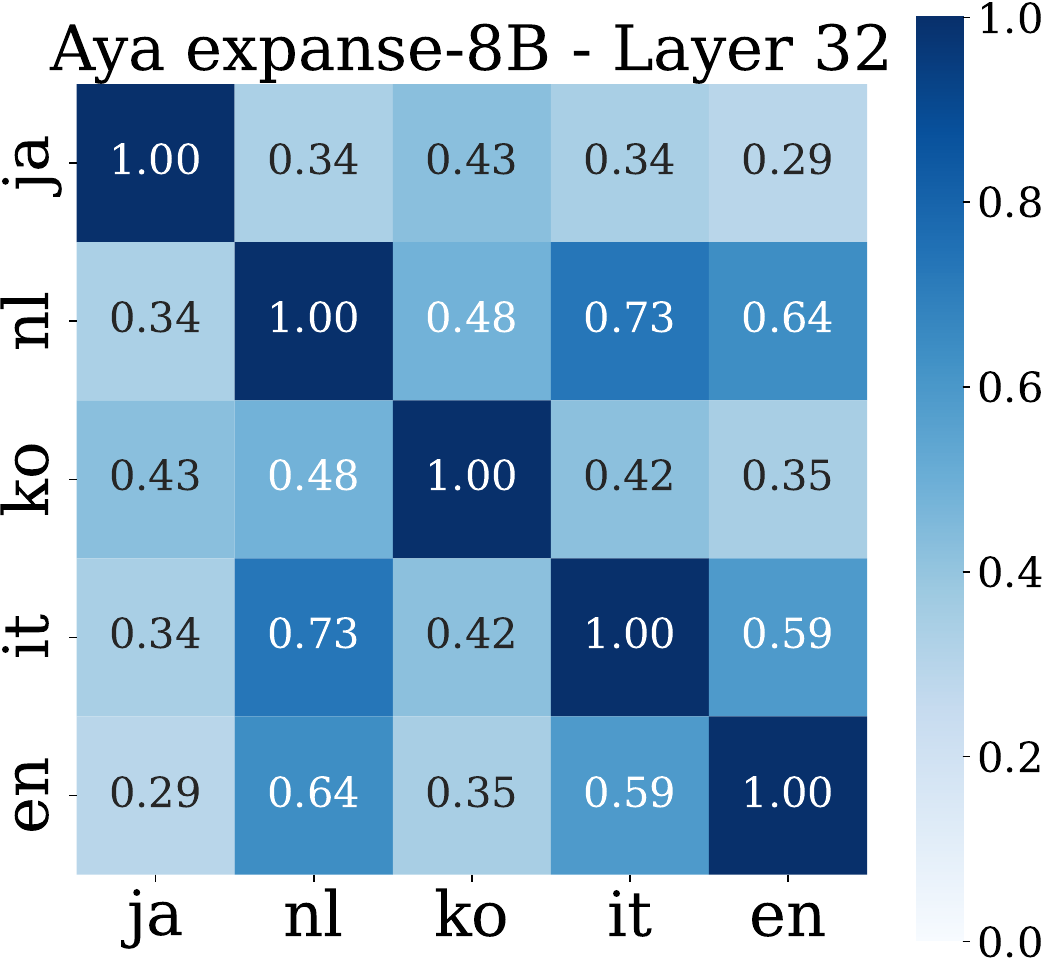}
  \includegraphics[width=0.19\linewidth]{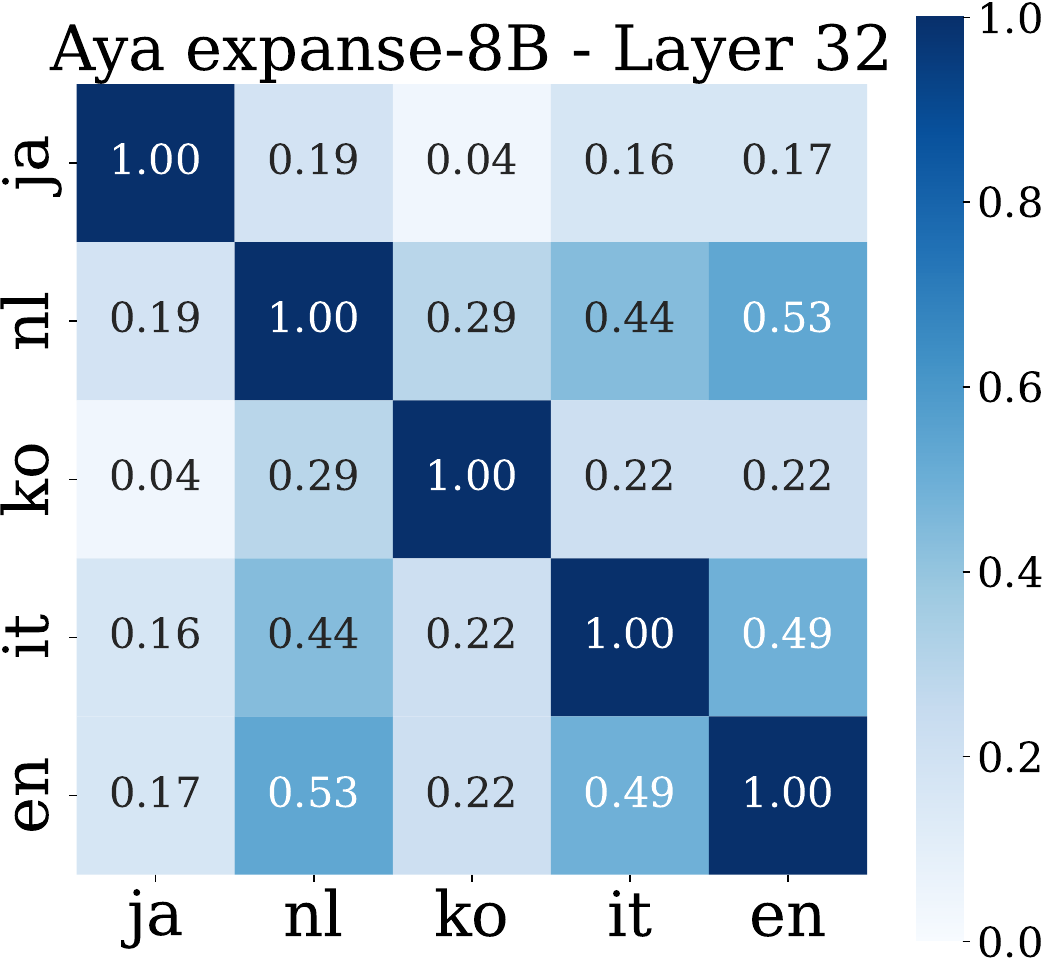}

  \begin{minipage}{0.19\linewidth}\centering \textbf{\textcolor{red}{layer 31 (Type-2)}}\end{minipage}
  \begin{minipage}{0.19\linewidth}\centering layer 31 (baseline)\end{minipage}
  \begin{minipage}{0.19\linewidth}\centering \textbf{\textcolor{red}{layer 32 (Type-2)}}\end{minipage}
  \begin{minipage}{0.19\linewidth}\centering layer 32 (baseline)\end{minipage}

  \caption{\textbf{Distance among centroids of language latent spaces while deactivating Top-1k Type-2 Transfer Neurons (Aya expanse-8B)}.}
  \label{fig:appendix:centroids distance among language subspaces while deactivating type2 aya}
\end{figure*}

Figs.~\ref{fig:appendix:centroids distance among language subspaces while deactivating type2 llama3},~\ref{fig:appendix:centroids distance among language subspaces while deactivating type2 mistral}, and~\ref{fig:appendix:centroids distance among language subspaces while deactivating type2 aya} show the distance among centroids of language latent spaces while deactivating Type-2 neurons. As indicated, deactivating Type-2 neurons significantly inhibit the movements towards each language specific latent space in final layers. This observation is well aligned with the PCA visualization while deactivating top-1k Type-2 neurons, as presented in Appendix~\ref{sec:appendix:Deactivating Type-2 Transfer Neurons Causes a Significant Delay in the Spatial Transition from the Shared Semantic Subspace to the Language-Specific Subspaces} above.

% Distance among language-subspaces, deactivating Type-2, llama3
% Distance among language-subspaces, deactivating Type-2, mistral
% Distance among language-subspaces, deactivating Type-2, aya

\section{The Nature of Transfer Neurons}
\label{sec:appendix:the nature of transfer neurons}
\subsection{Distribution}
\label{sec:appendix:distribution}
\begin{figure*}[t]
    \centering

    \begin{minipage}{0.23\linewidth}
      \centering
      \includegraphics[width=\linewidth]{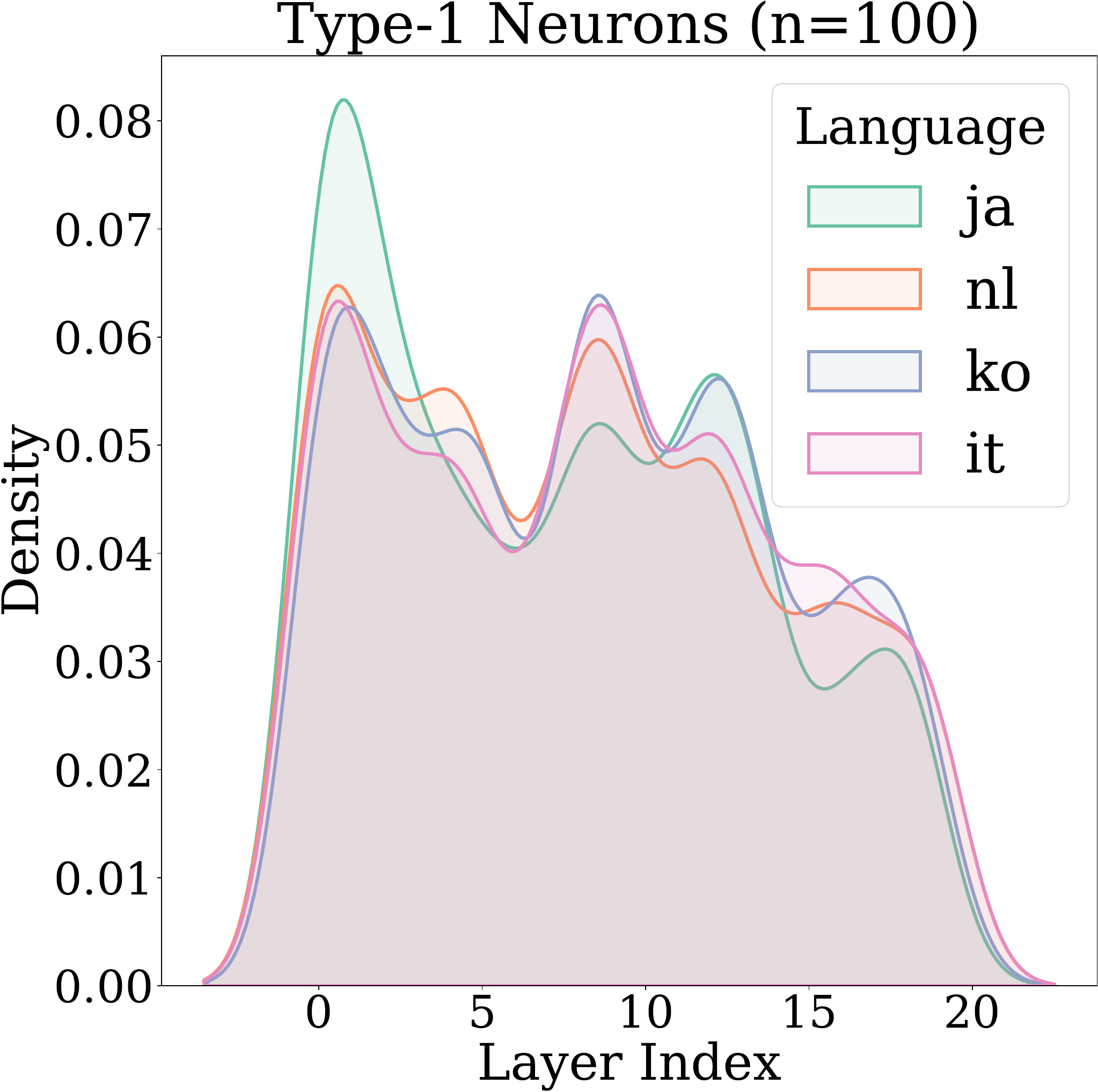}
      \subcaption{llama3, top-100}
    \end{minipage}
    \begin{minipage}{0.23\linewidth}
      \centering
      \includegraphics[width=\linewidth]{figures/llama3/distribution/to_shared/cos_sim_allLangs_n1000.pdf}
      \subcaption{llama3, top-1000}
    \end{minipage}
    \begin{minipage}{0.23\linewidth}
      \centering
      \includegraphics[width=\linewidth]{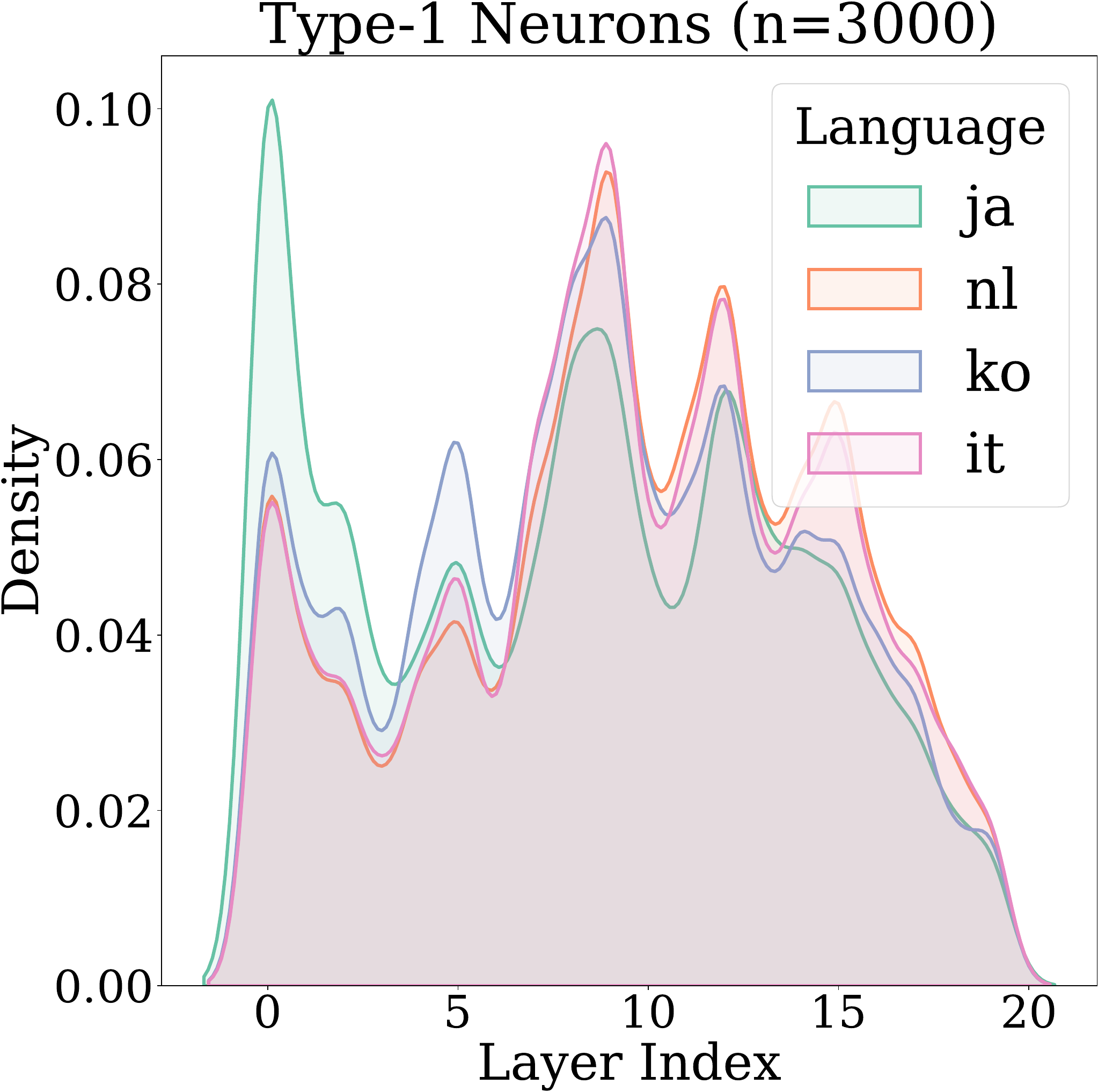}
      \subcaption{llama3, top-3000}
    \end{minipage}
    \begin{minipage}{0.23\linewidth}
      \centering
      \includegraphics[width=\linewidth]{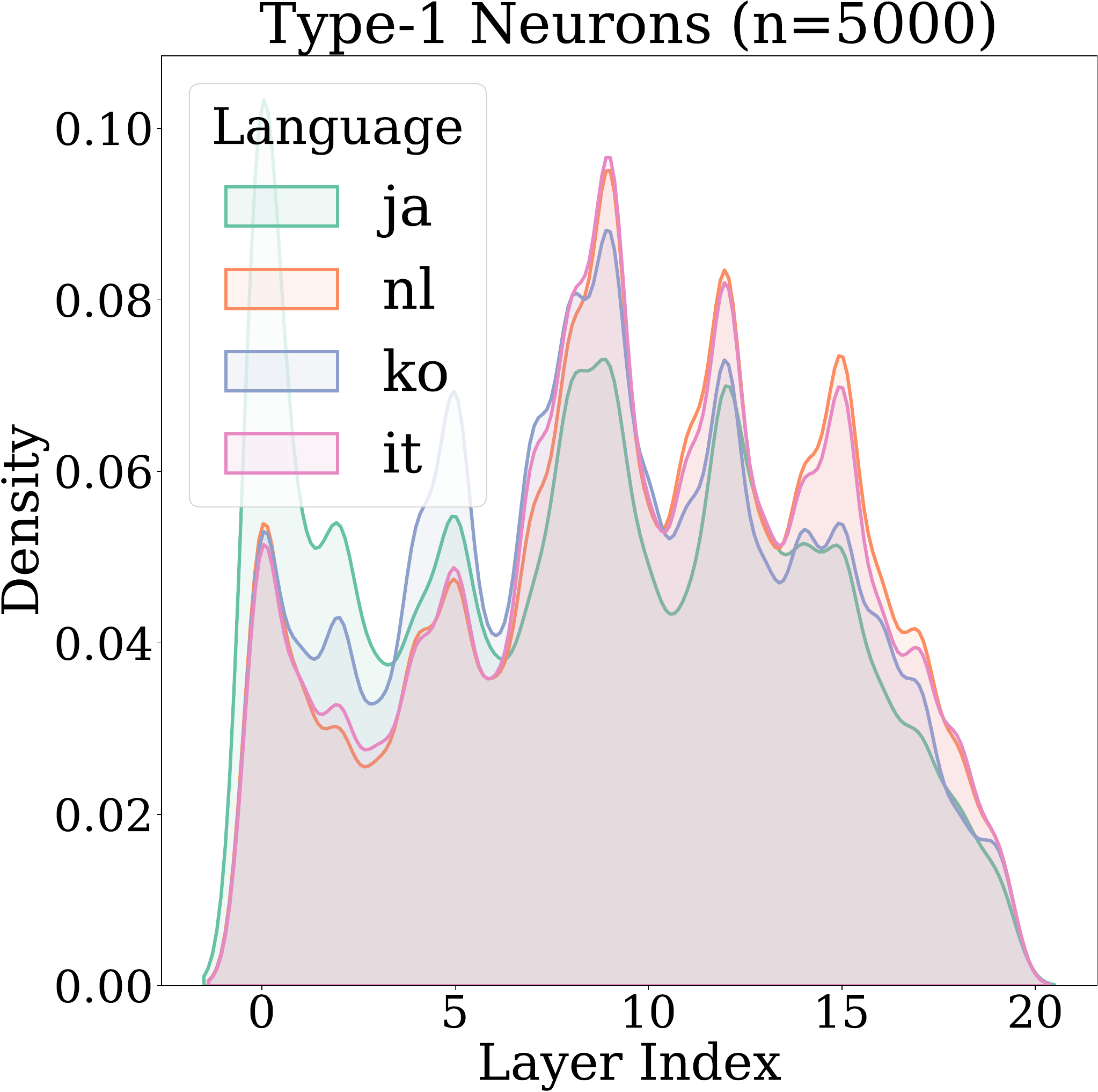}
      \subcaption{llama3, top-5000}
    \end{minipage}

    \begin{minipage}{0.23\linewidth}
      \centering
      \includegraphics[width=\linewidth]{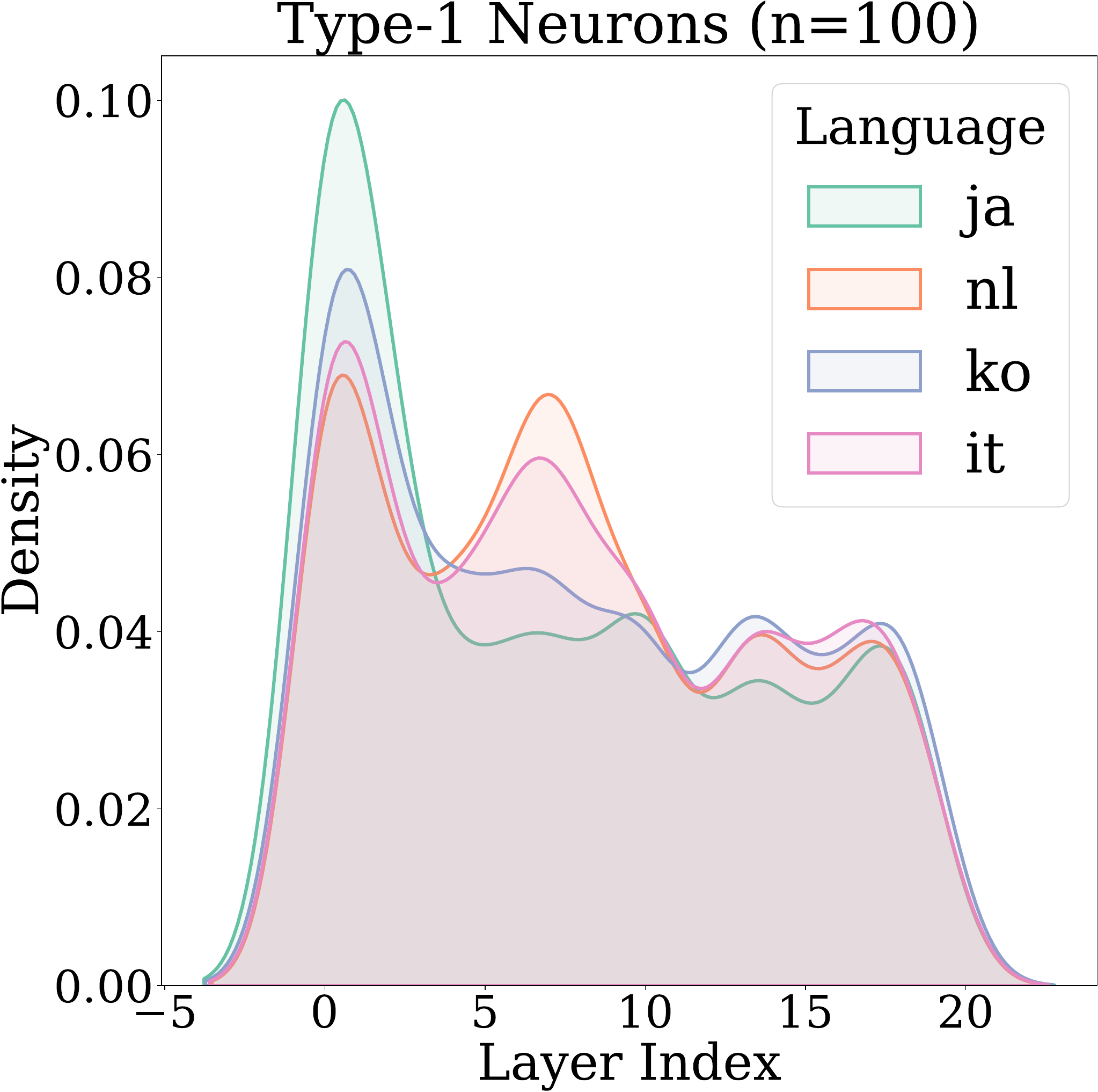}
      \subcaption{mistral, top-100}
    \end{minipage}
    \begin{minipage}{0.23\linewidth}
      \centering
      \includegraphics[width=\linewidth]{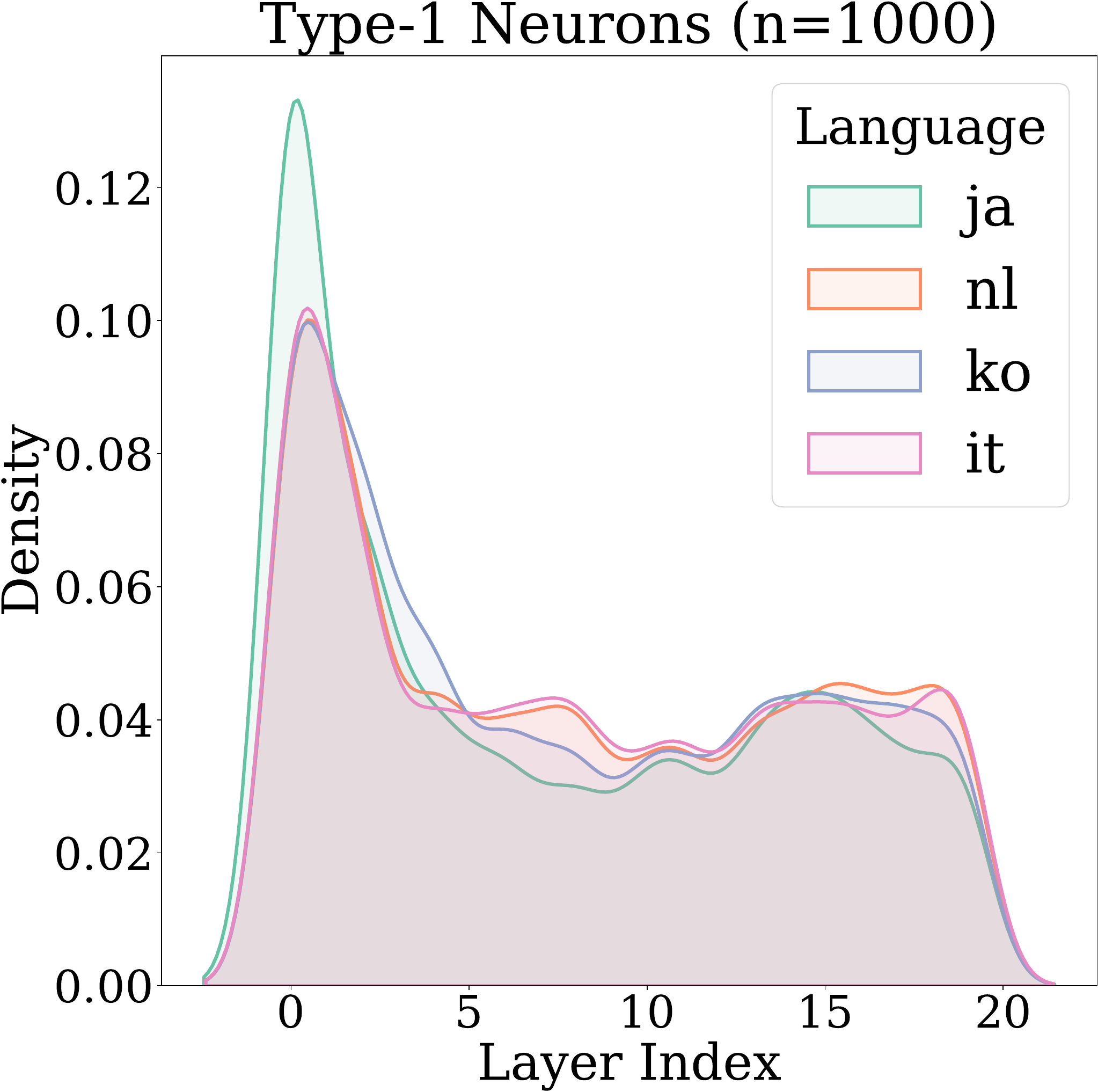}
      \subcaption{mistral, top-1000}
    \end{minipage}
    \begin{minipage}{0.23\linewidth}
      \centering
      \includegraphics[width=\linewidth]{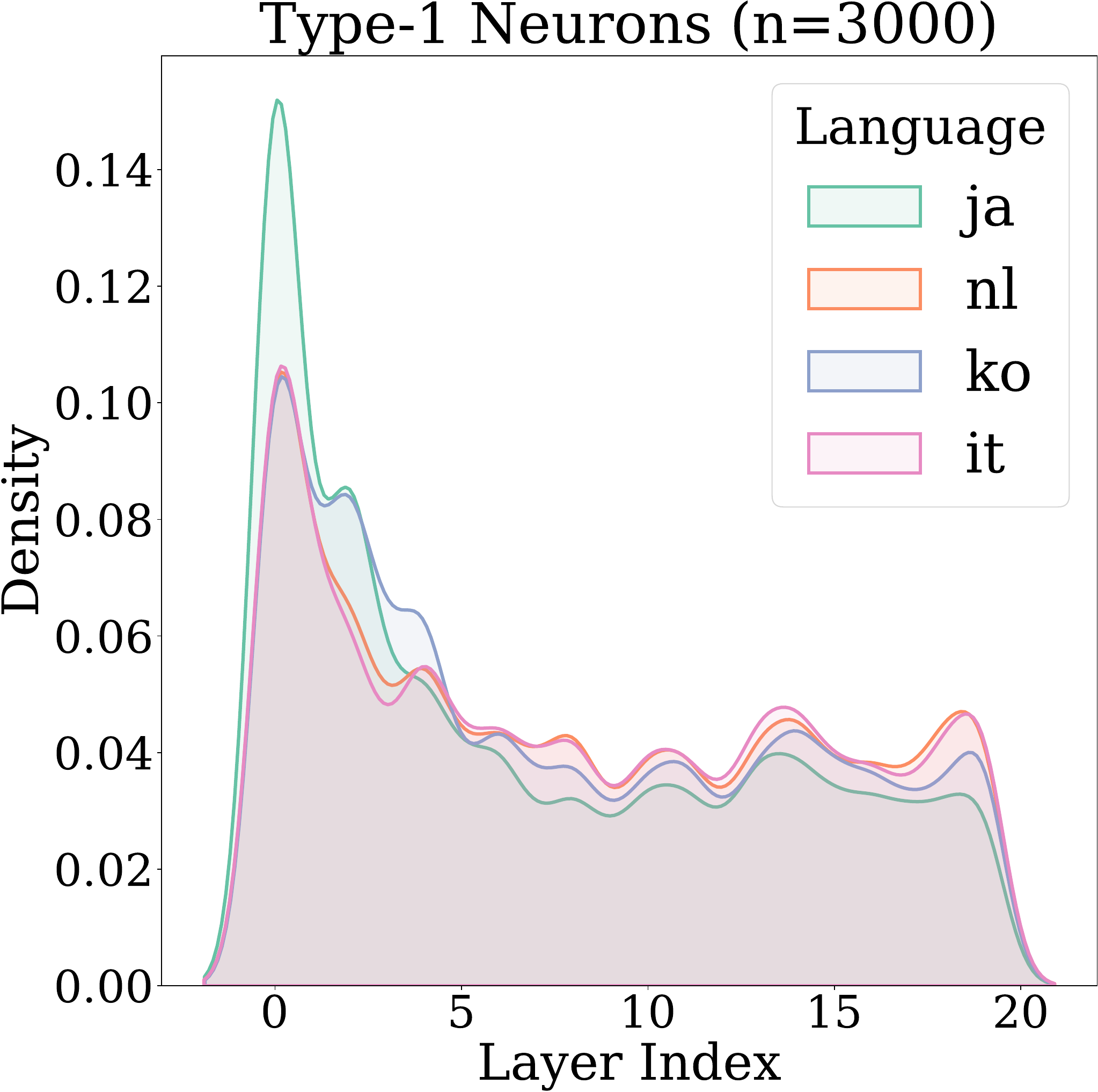}
      \subcaption{mistral, top-3000}
    \end{minipage}
    \begin{minipage}{0.23\linewidth}
      \centering
      \includegraphics[width=\linewidth]{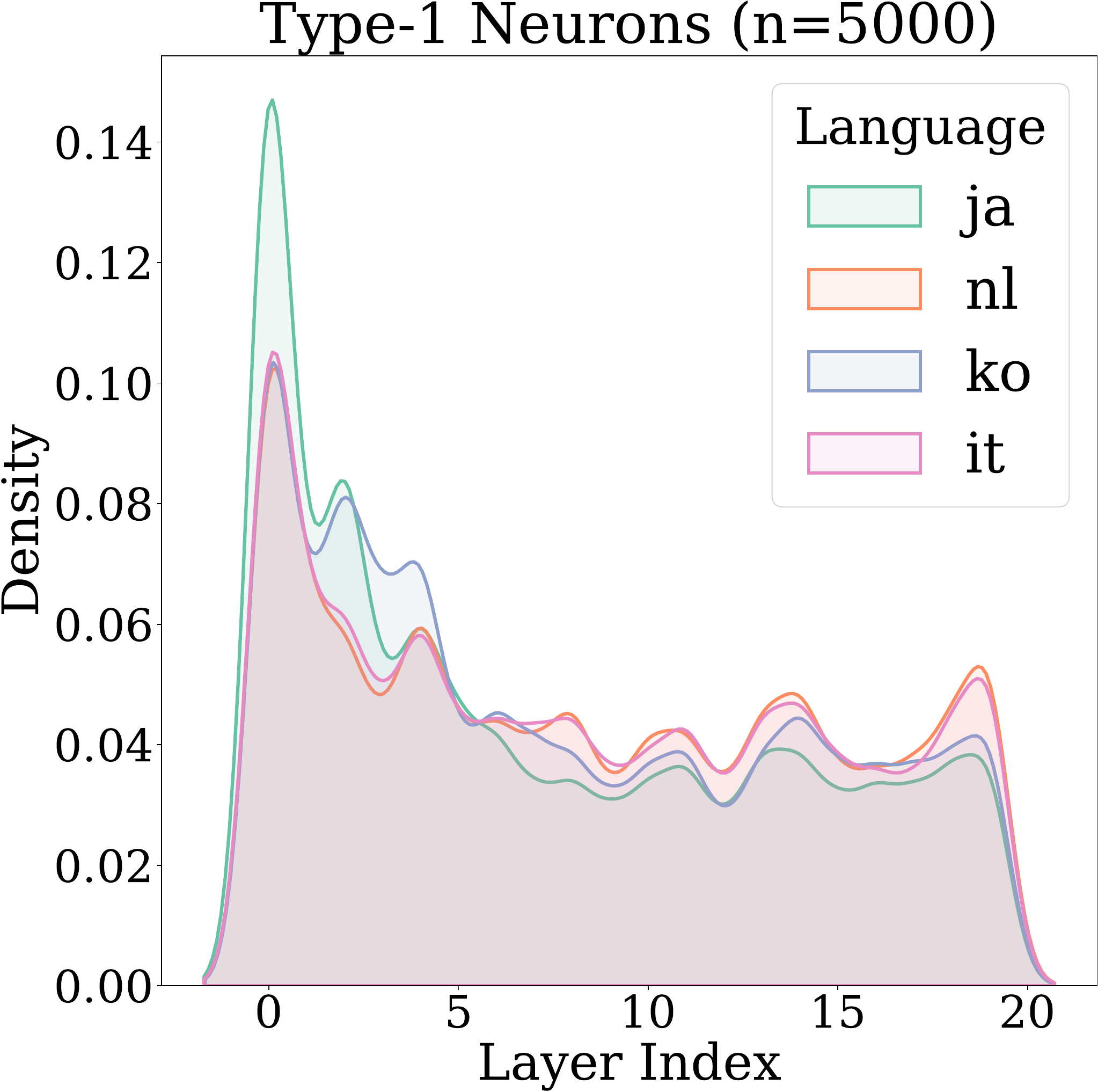}
      \subcaption{mistral, top-5000}
    \end{minipage}

    \begin{minipage}{0.23\linewidth}
      \centering
      \includegraphics[width=\linewidth]{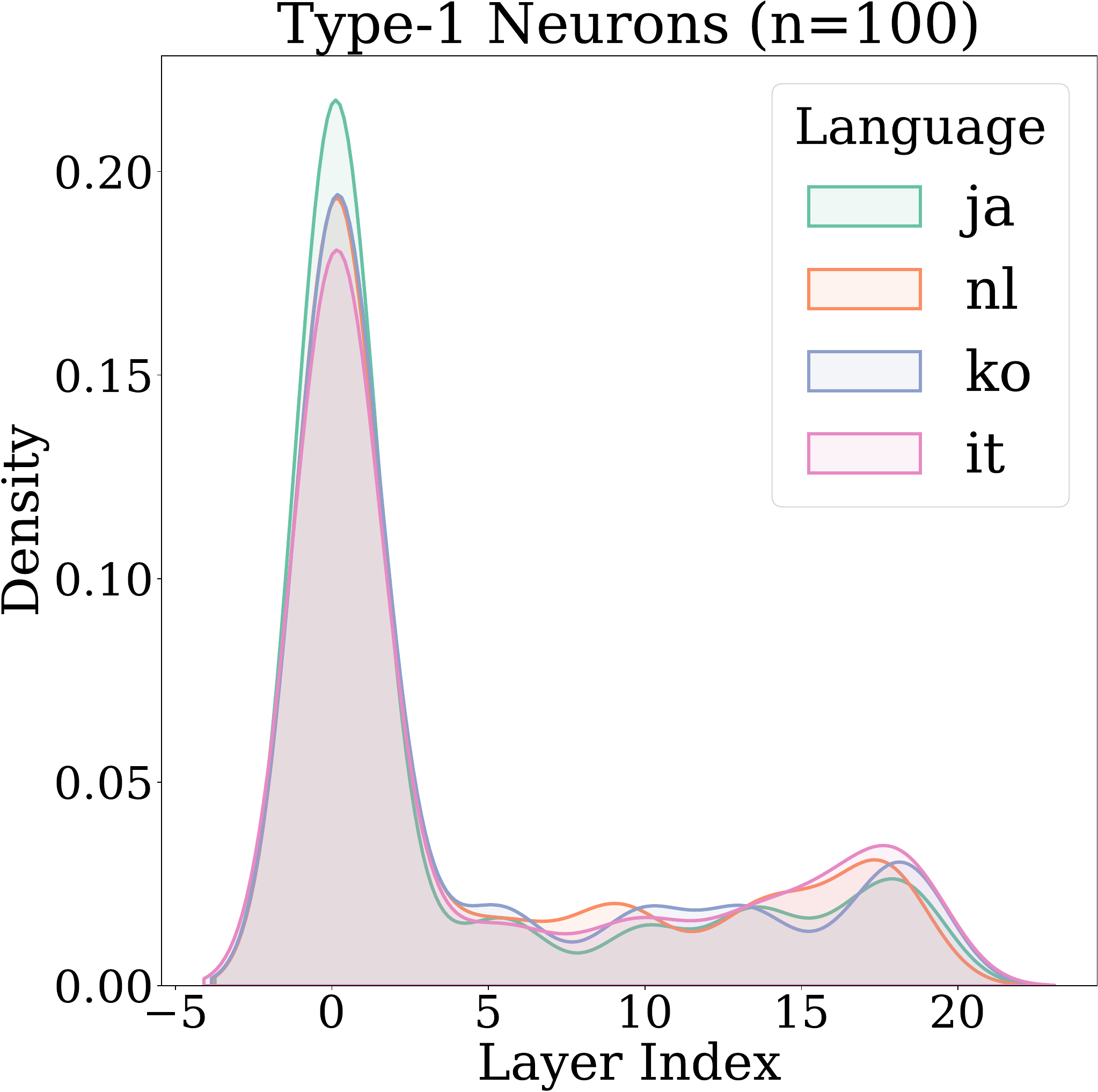}
      \subcaption{aya, top-100}
    \end{minipage}
    \begin{minipage}{0.23\linewidth}
      \centering
      \includegraphics[width=\linewidth]{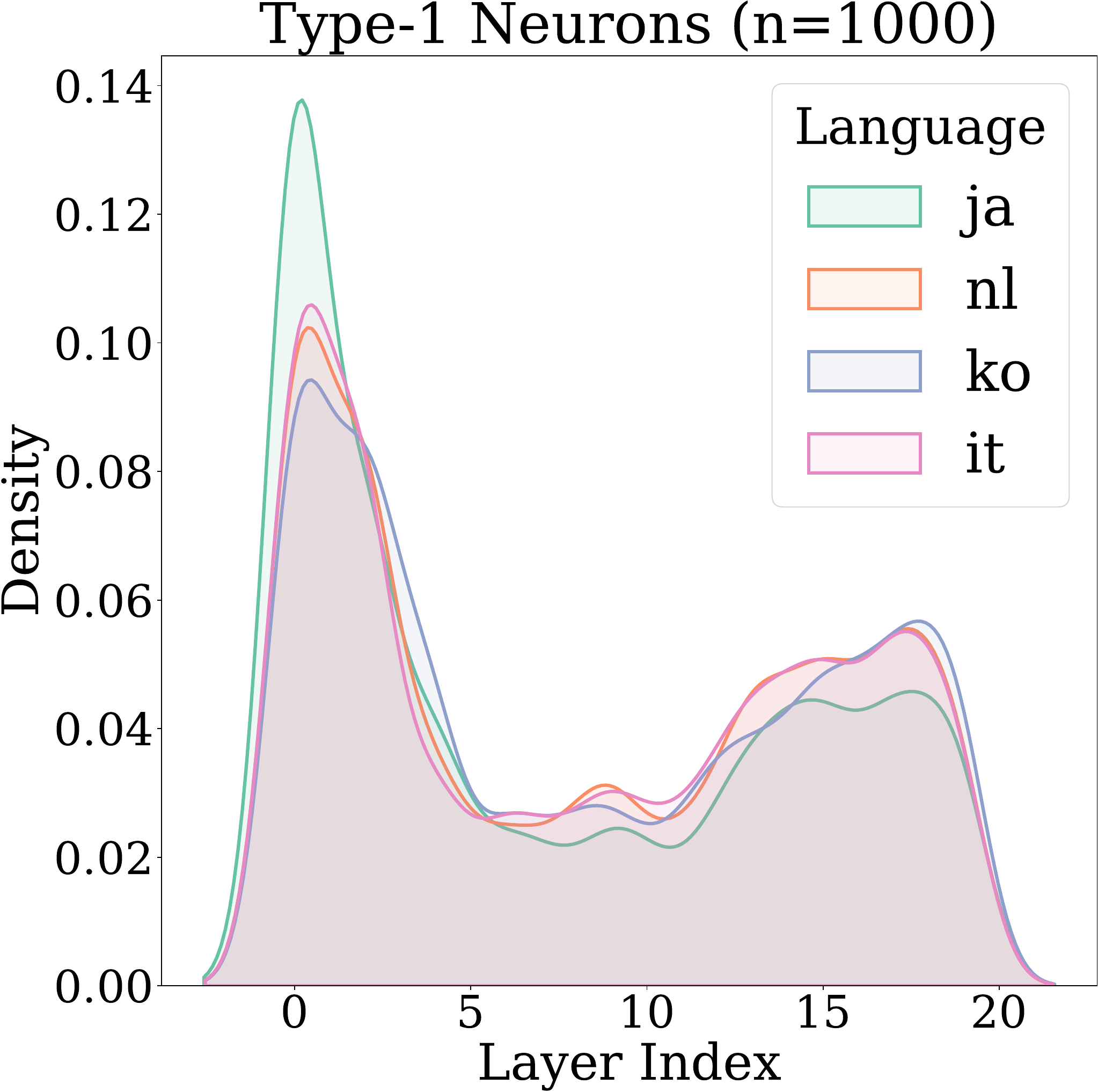}
      \subcaption{aya, top-1000}
    \end{minipage}
    \begin{minipage}{0.23\linewidth}
      \centering
      \includegraphics[width=\linewidth]{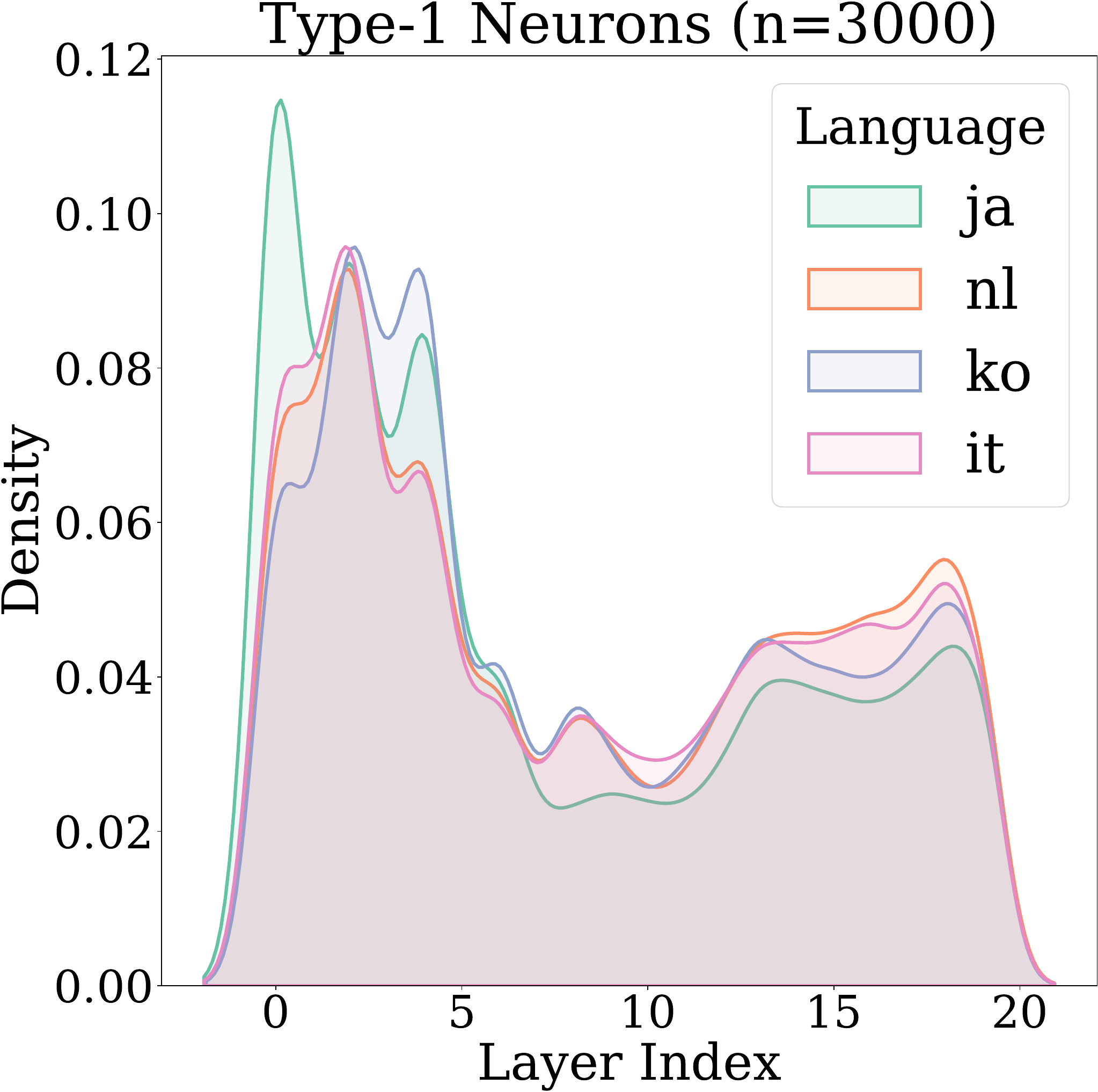}
      \subcaption{aya, top-3000}
    \end{minipage}
    \begin{minipage}{0.23\linewidth}
      \centering
      \includegraphics[width=\linewidth]{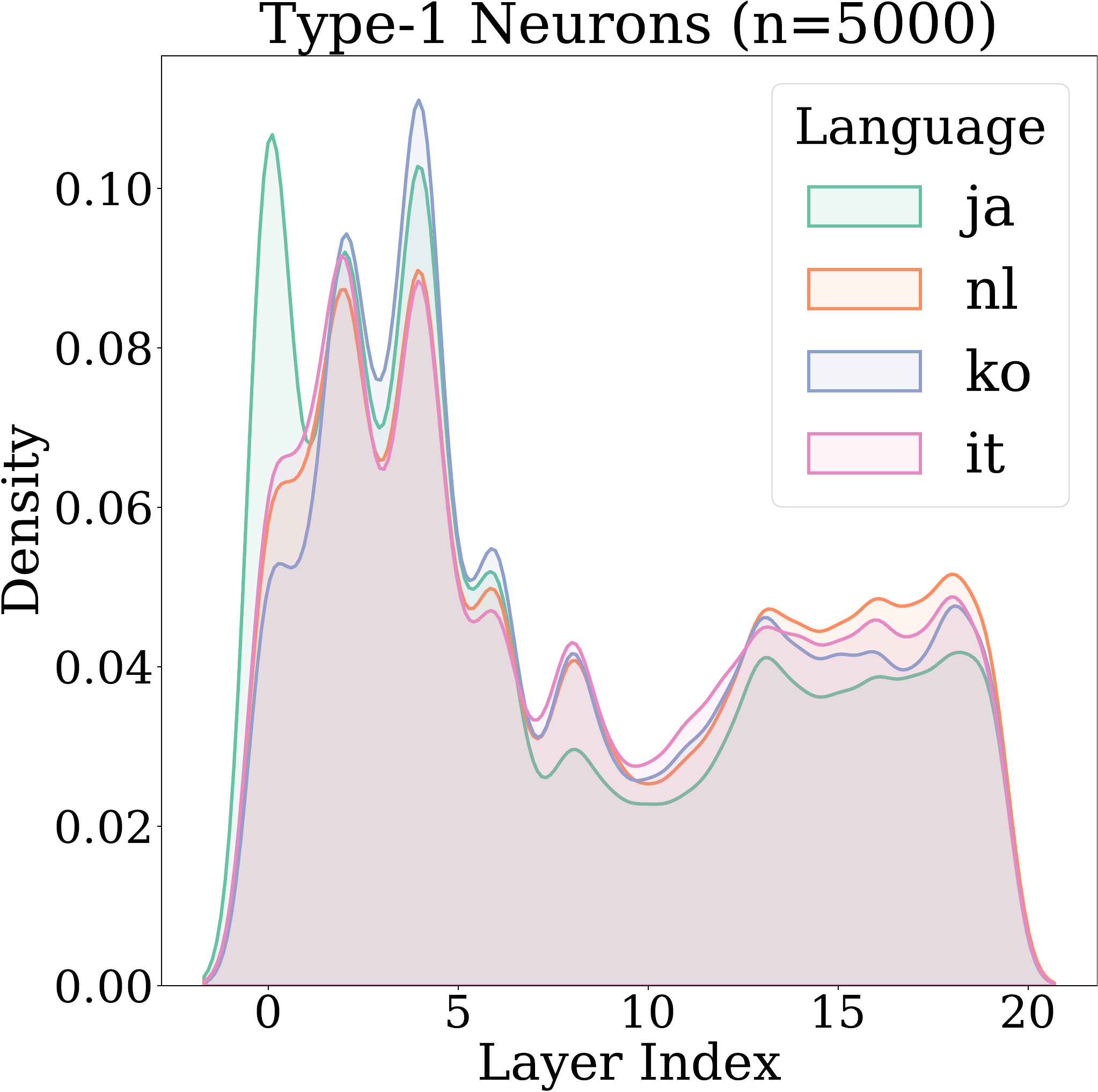}
      \subcaption{aya, top-5000}
    \end{minipage}

  \caption{\textbf{Distribution of Type-1 Transfer Neurons (LLaMA3-8B, Mistral-7B, and Aya expanse-8B).} 1-20 layers.}
  \label{fig:appendix:distribution_Type-1}
\end{figure*}

\begin{figure*}[t]
    \centering

    \begin{minipage}{0.23\linewidth}
      \centering
      \includegraphics[width=\linewidth]{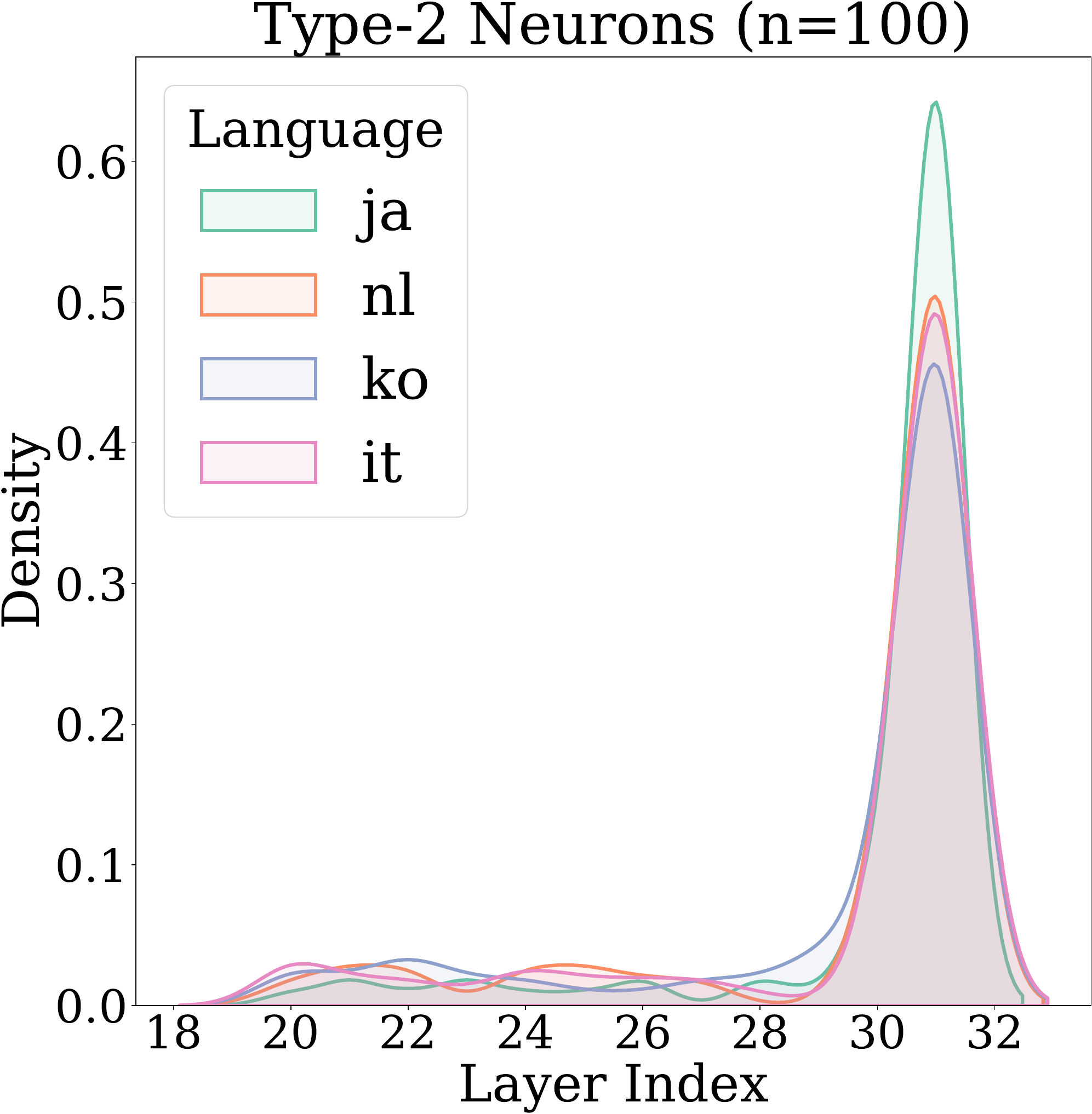}
      \subcaption{llama3, top-100}
    \end{minipage}
    \begin{minipage}{0.23\linewidth}
      \centering
      \includegraphics[width=\linewidth]{figures/llama3/distribution/to_lang_specific/cos_sim_allLangs_n1000.pdf}
      \subcaption{llama3, top-1000}
    \end{minipage}
    \begin{minipage}{0.23\linewidth}
      \centering
      \includegraphics[width=\linewidth]{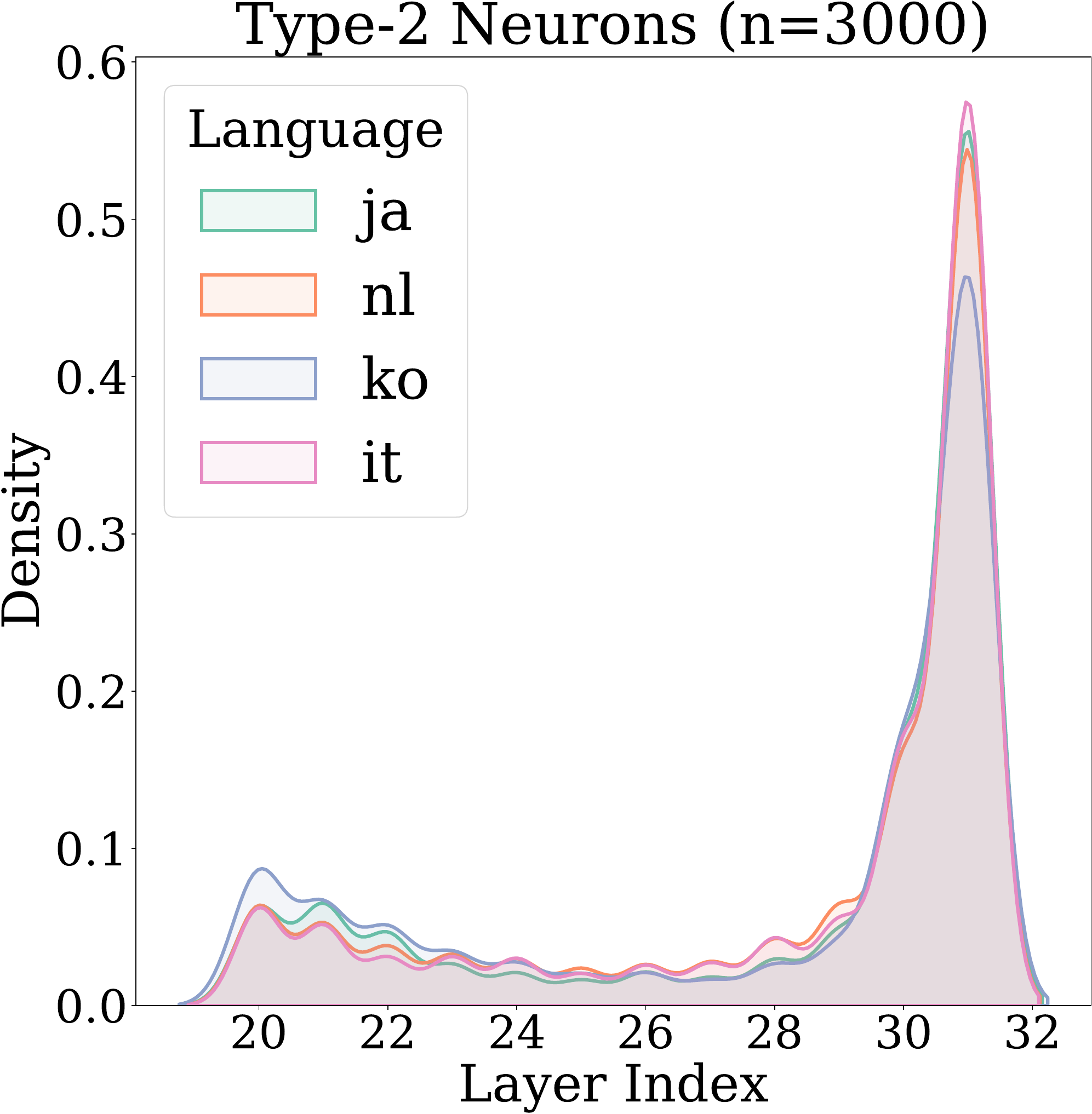}
      \subcaption{llama3, top-3000}
    \end{minipage}
    \begin{minipage}{0.23\linewidth}
      \centering
      \includegraphics[width=\linewidth]{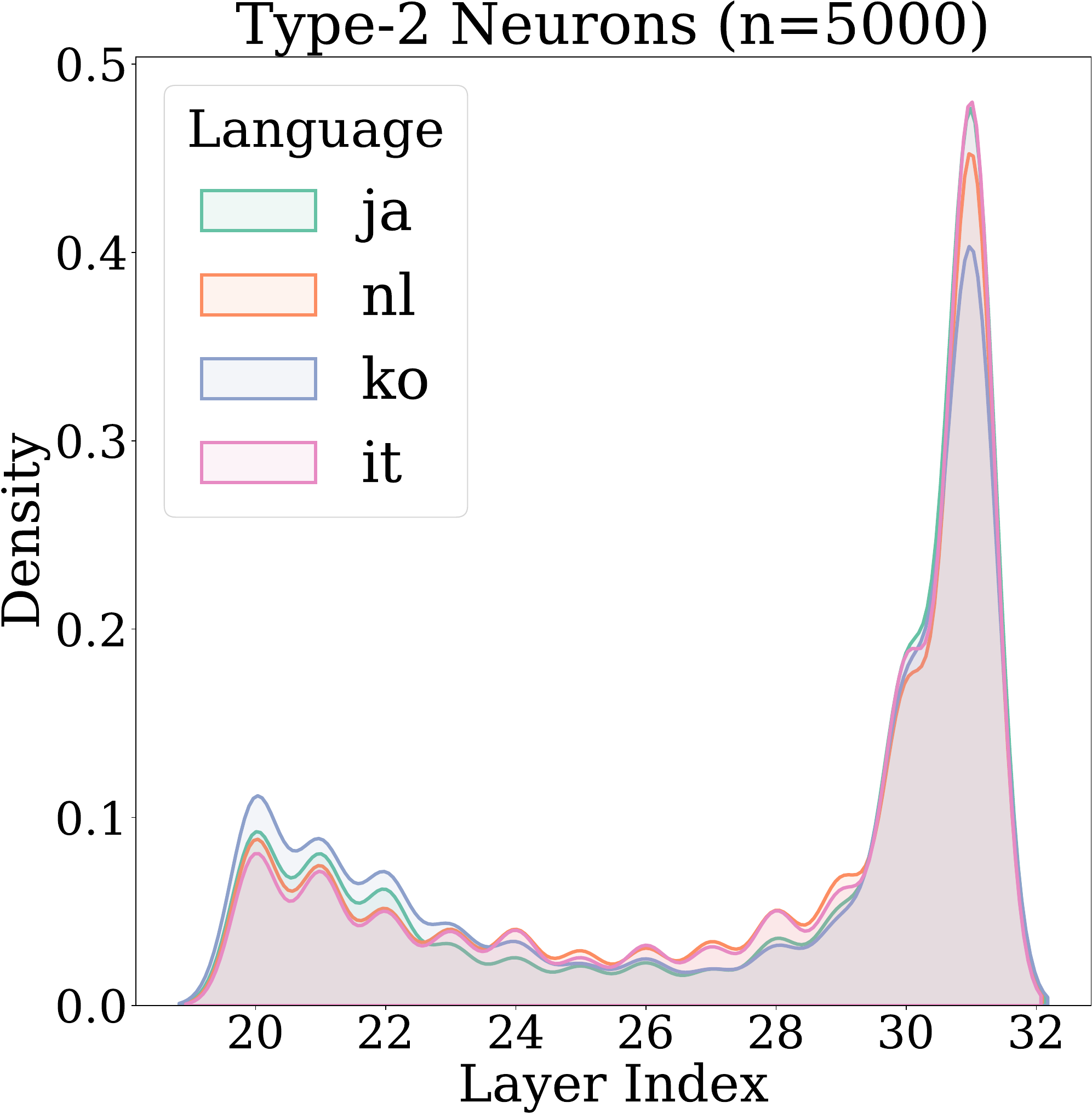}
      \subcaption{llama3, top-5000}
    \end{minipage}

    \begin{minipage}{0.23\linewidth}
      \centering
      \includegraphics[width=\linewidth]{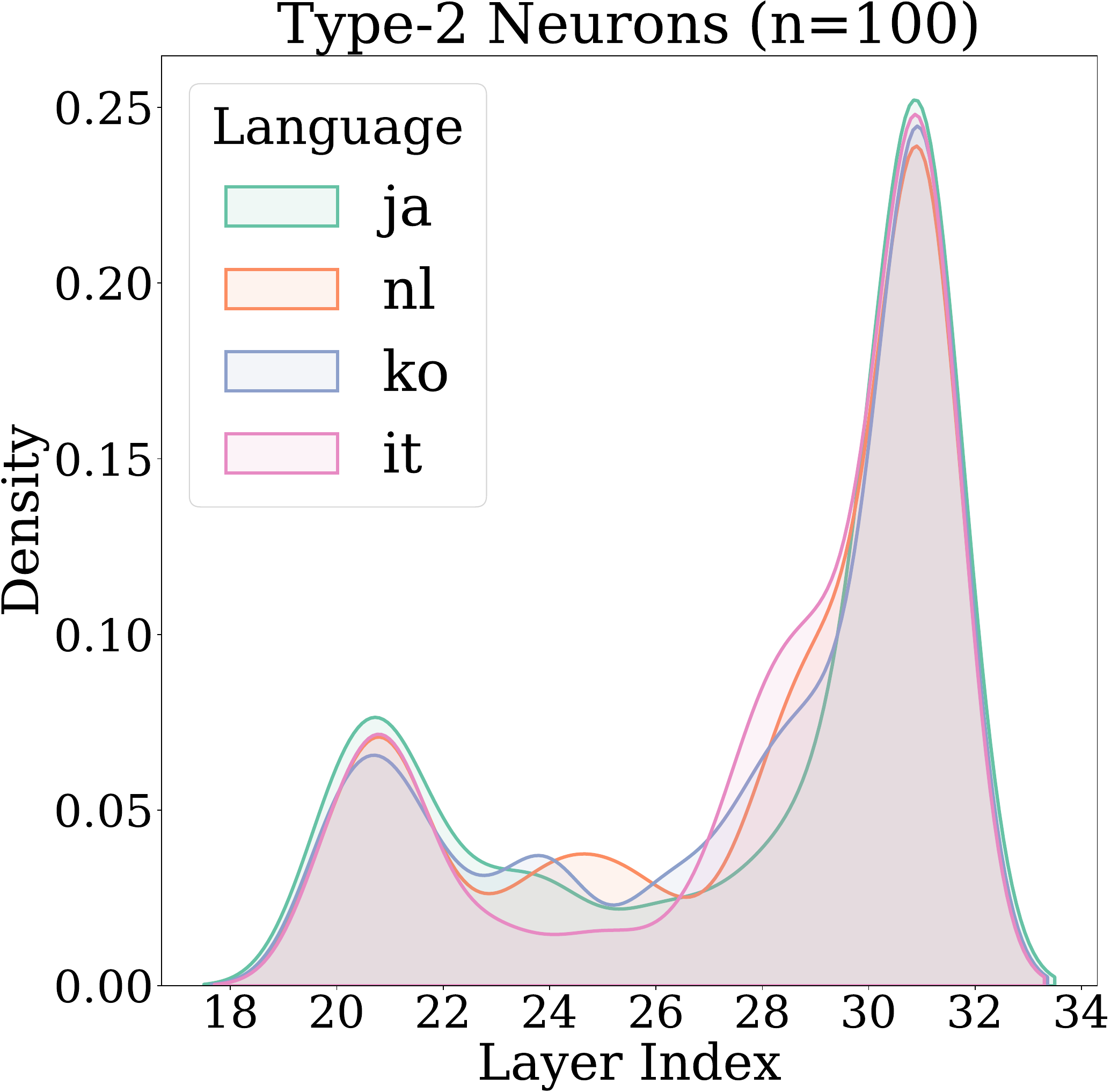}
      \subcaption{mistral, top-100}
    \end{minipage}
    \begin{minipage}{0.23\linewidth}
      \centering
      \includegraphics[width=\linewidth]{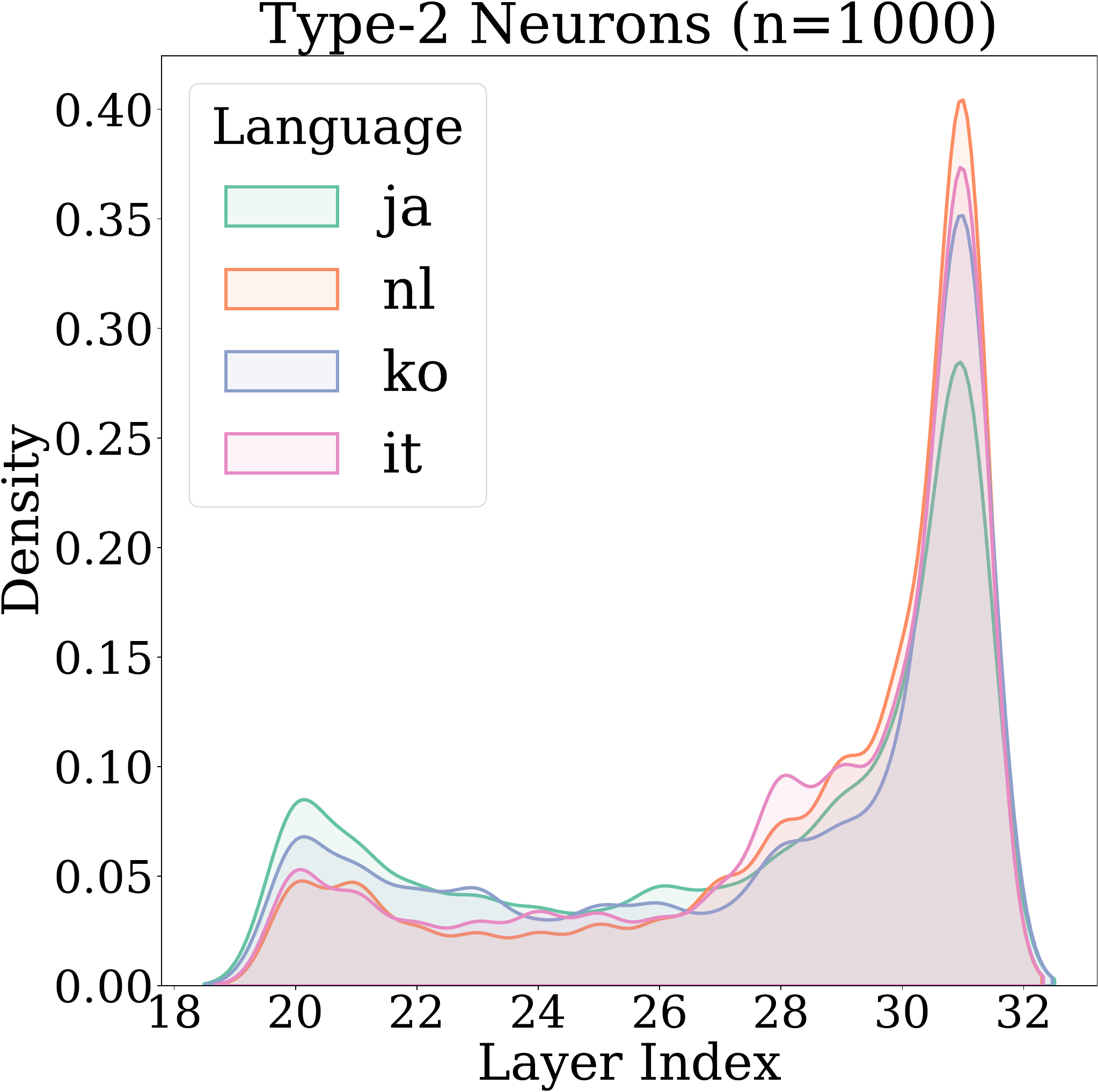}
      \subcaption{mistral, top-1000}
    \end{minipage}
    \begin{minipage}{0.23\linewidth}
      \centering
      \includegraphics[width=\linewidth]{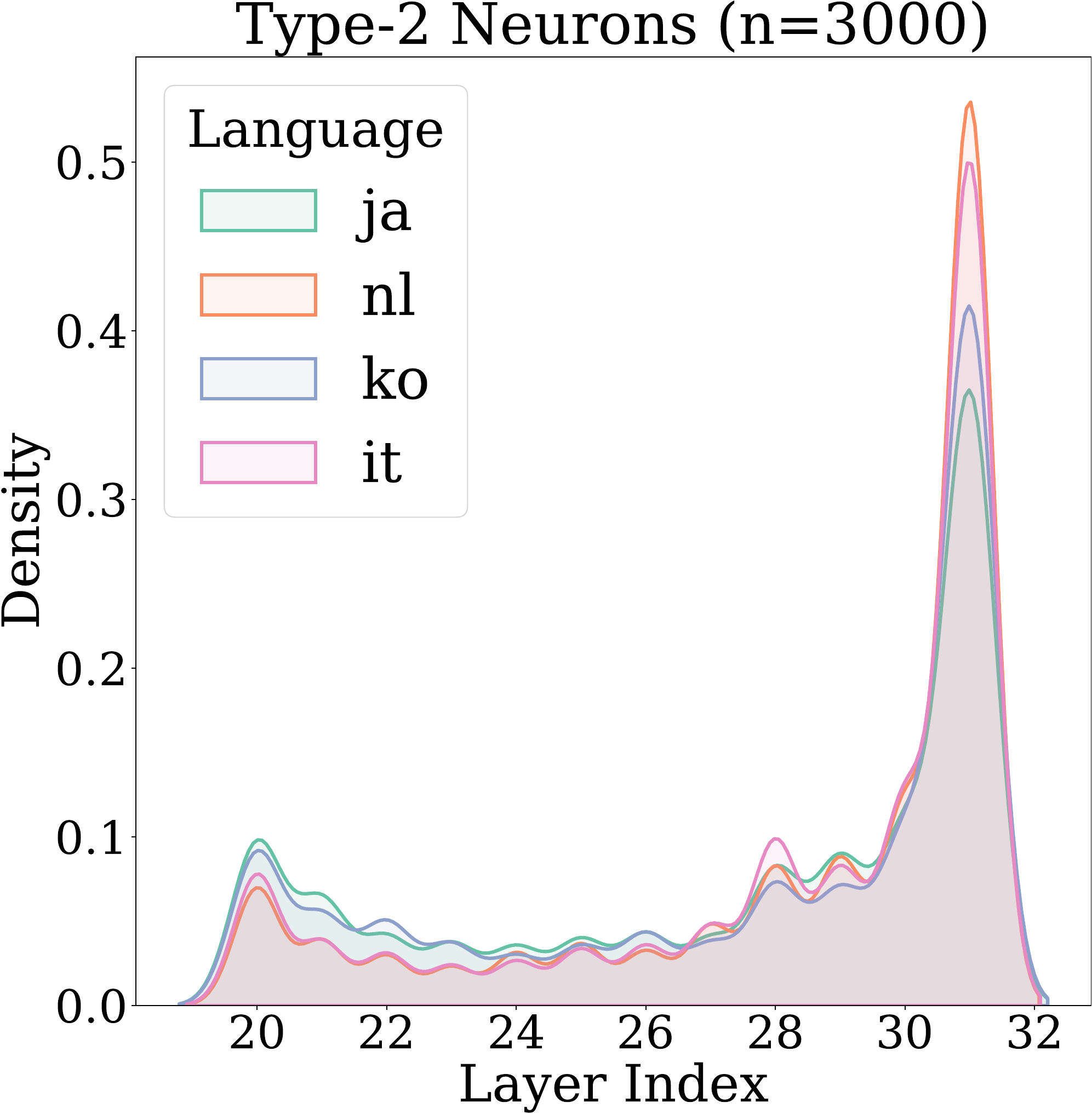}
      \subcaption{mistral, top-3000}
    \end{minipage}
    \begin{minipage}{0.23\linewidth}
      \centering
      \includegraphics[width=\linewidth]{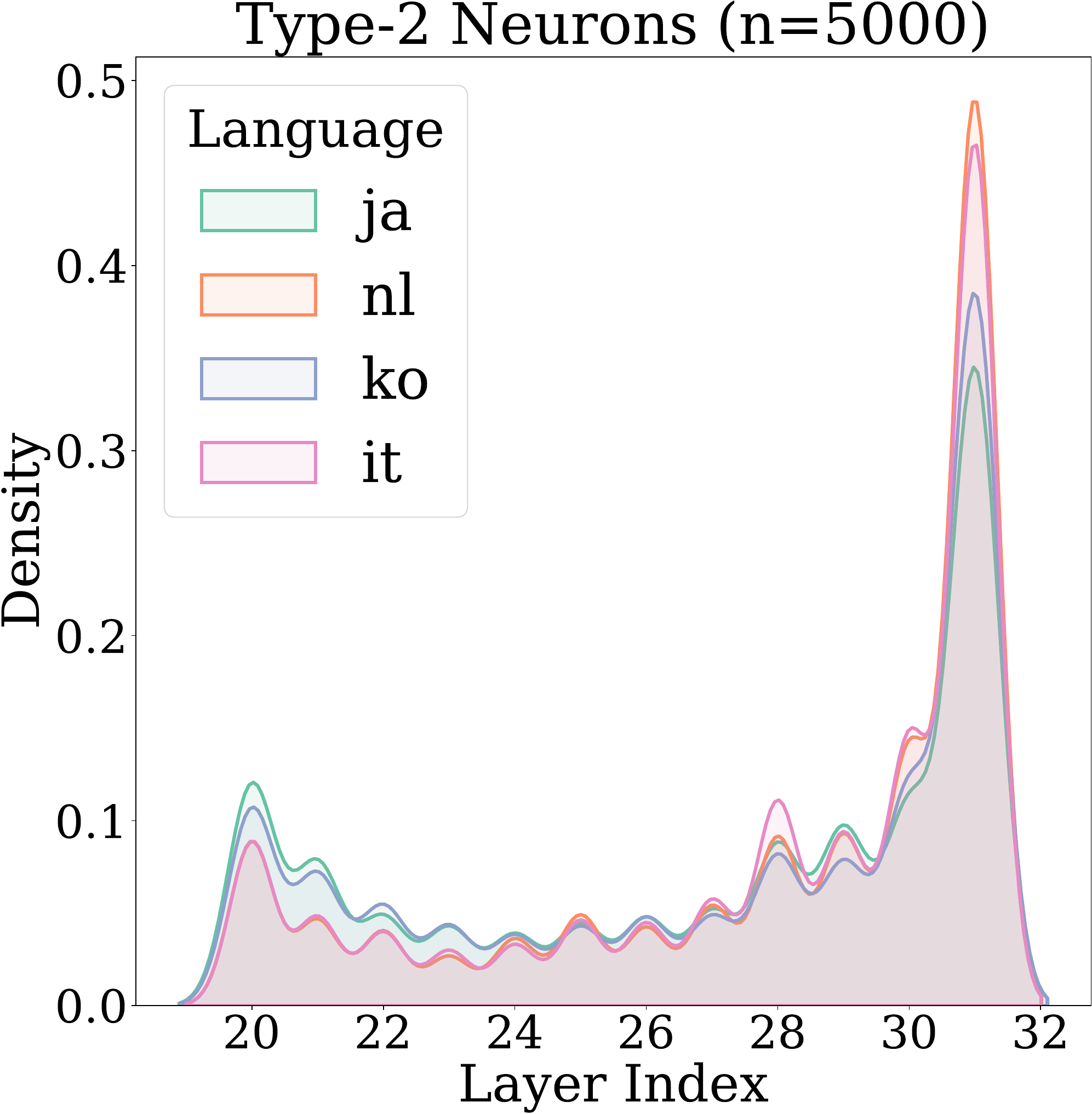}
      \subcaption{mistral, top-5000}
    \end{minipage}

    \begin{minipage}{0.23\linewidth}
      \centering
      \includegraphics[width=\linewidth]{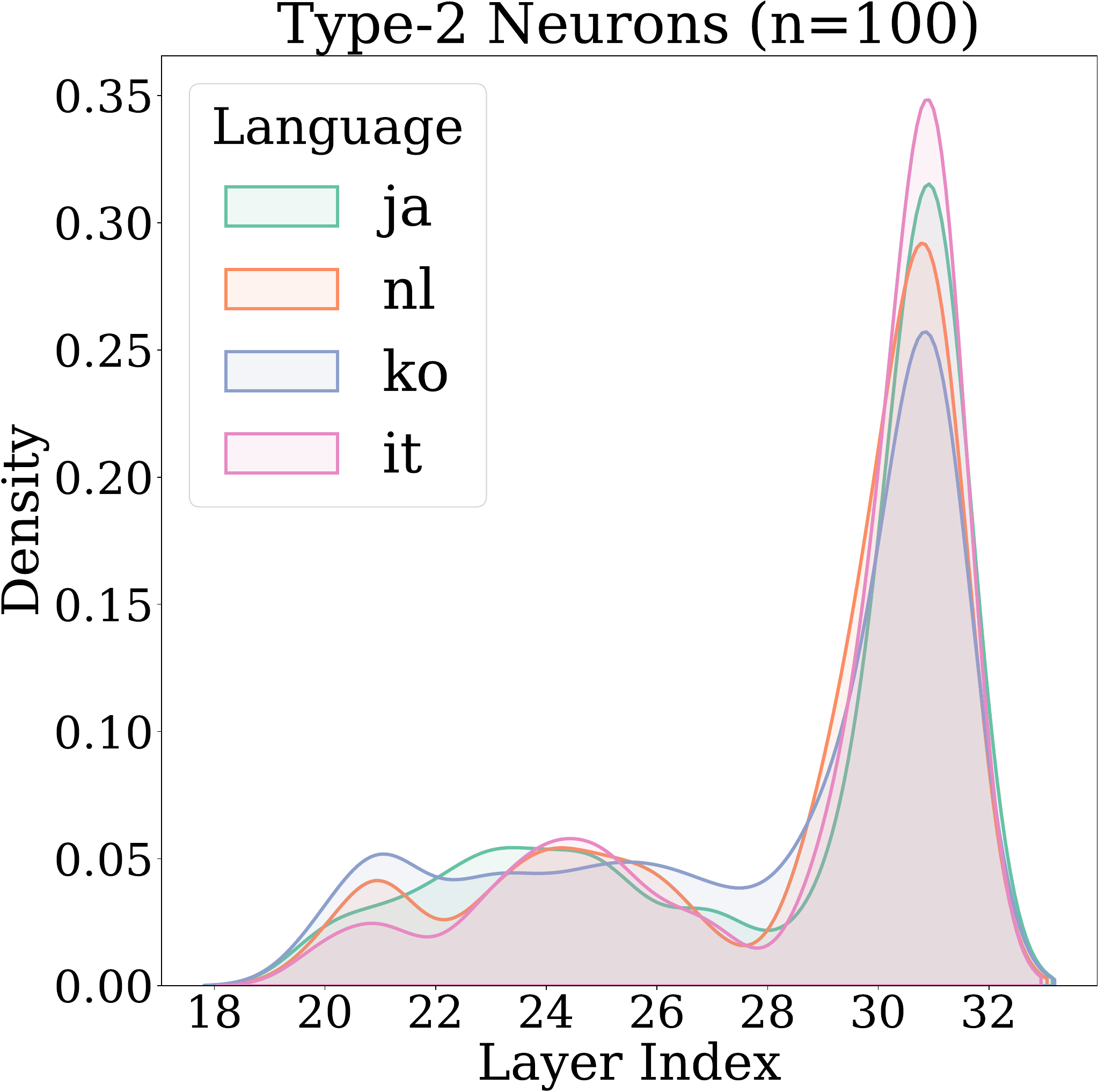}
      \subcaption{aya, top-100}
    \end{minipage}
    \begin{minipage}{0.23\linewidth}
      \centering
      \includegraphics[width=\linewidth]{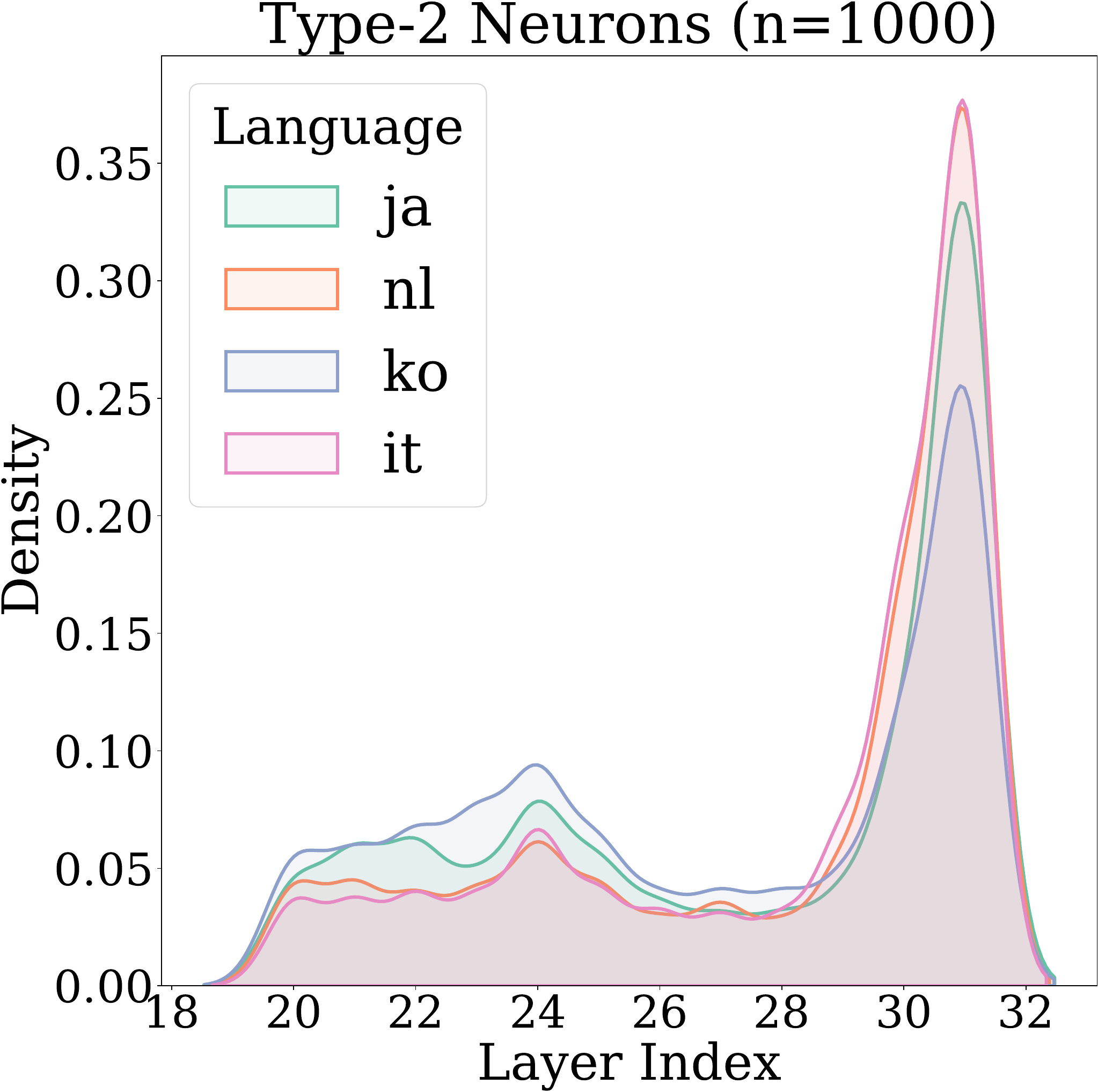}
      \subcaption{aya, top-1000}
    \end{minipage}
    \begin{minipage}{0.23\linewidth}
      \centering
      \includegraphics[width=\linewidth]{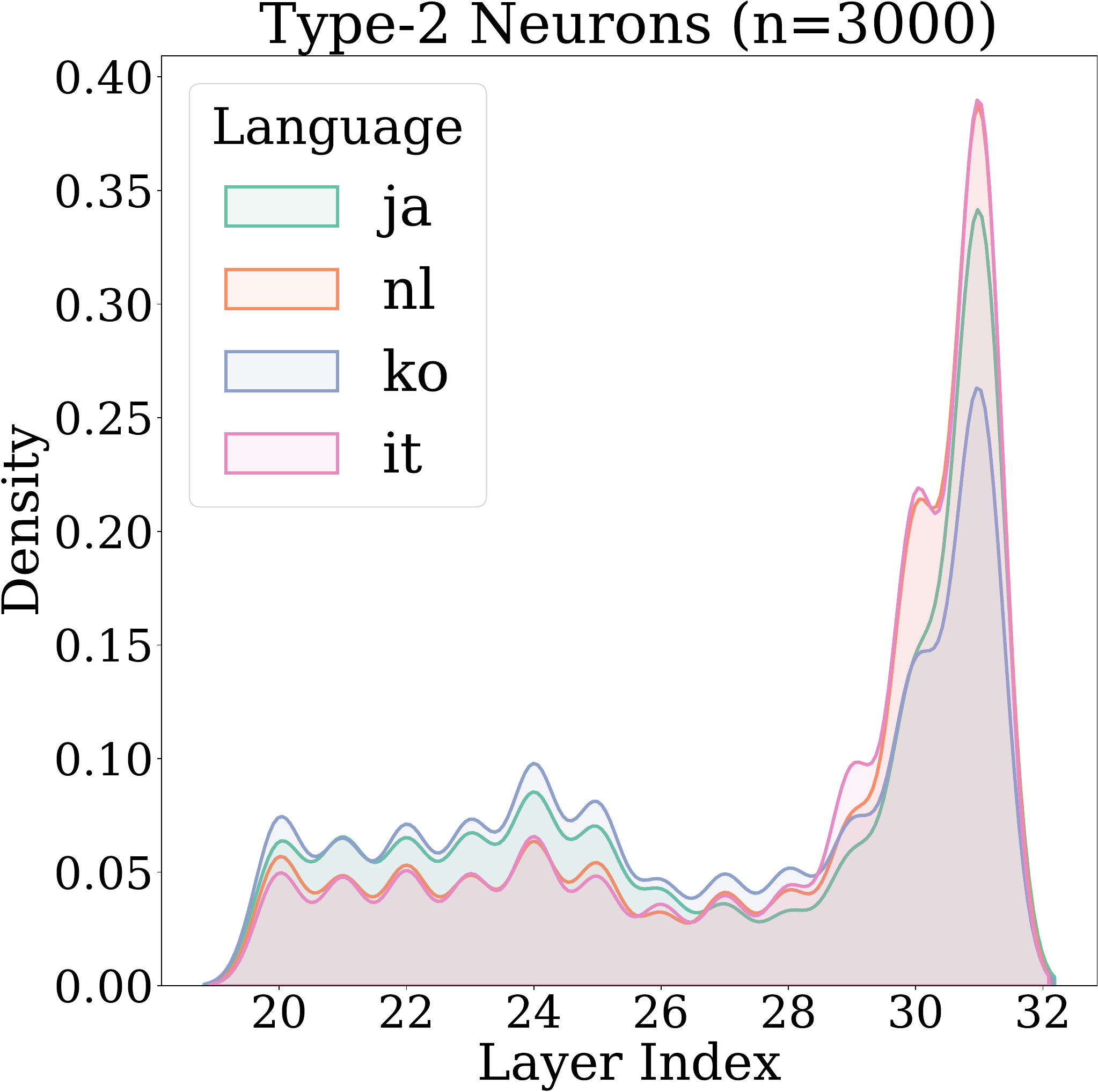}
      \subcaption{aya, top-3000}
    \end{minipage}
    \begin{minipage}{0.23\linewidth}
      \centering
      \includegraphics[width=\linewidth]{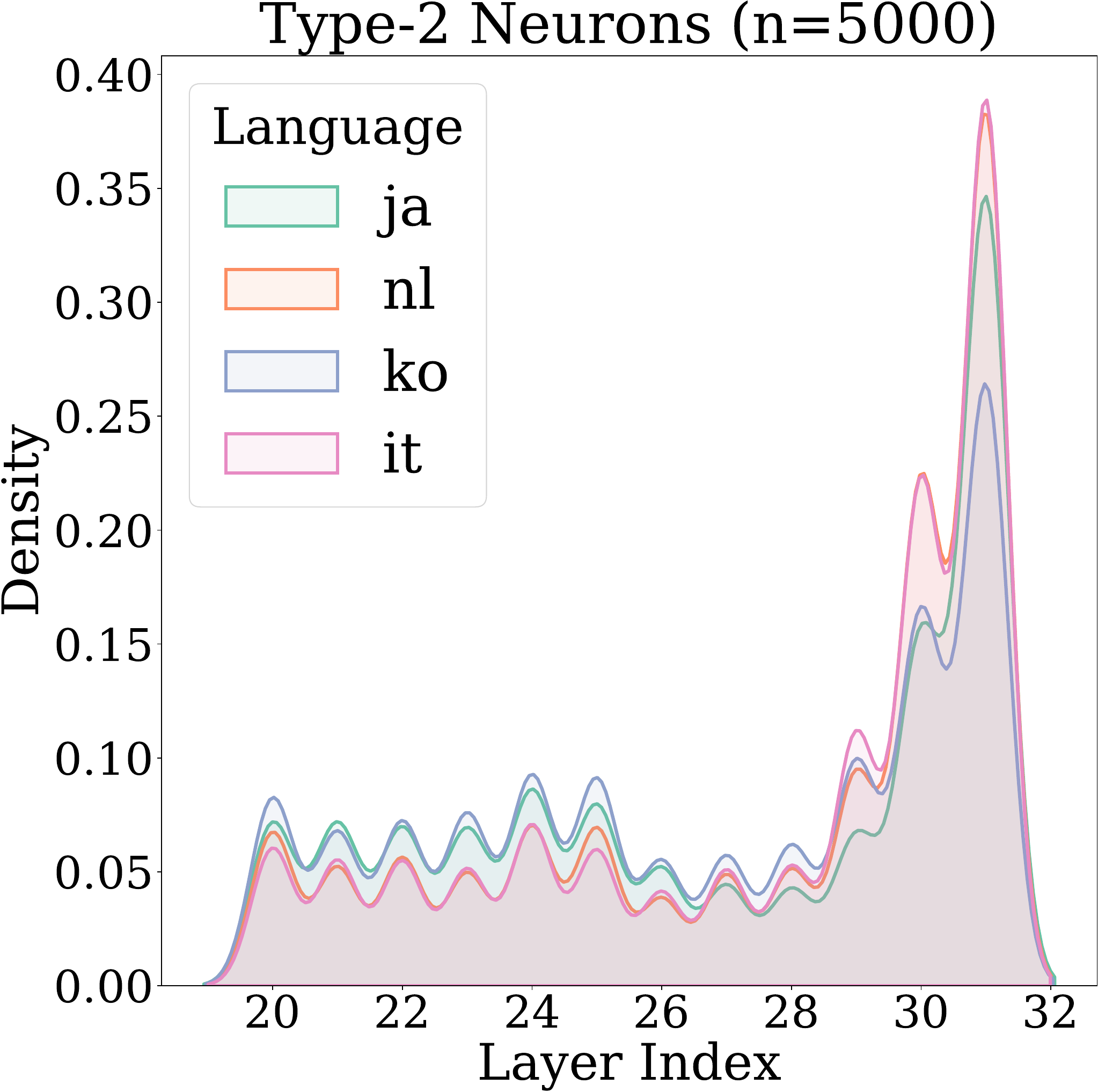}
      \subcaption{aya, top-5000}
    \end{minipage}

  \caption{\textbf{Distribution of Type-2 Transfer Neurons (LLaMA3-8B, Mistral-7B, and Aya expanse-8B).} 21-32 layers.}
  \label{fig:appendix:distribution_Type-2}
\end{figure*}

Figs.~\ref{fig:appendix:distribution_Type-1} and~\ref{fig:appendix:distribution_Type-2} show the distribution of Type-1 and Type-2 neurons.

% distribution, Type-1, n100,1k,3k,5k,10k, all models
% distribution, Type-2, n100,1k,3k,5k,10k, all models

\subsection{Overlap Ratio of Transfer Neurons}
\begin{figure*}[t]
  \centering
  \includegraphics[width=0.49\linewidth]{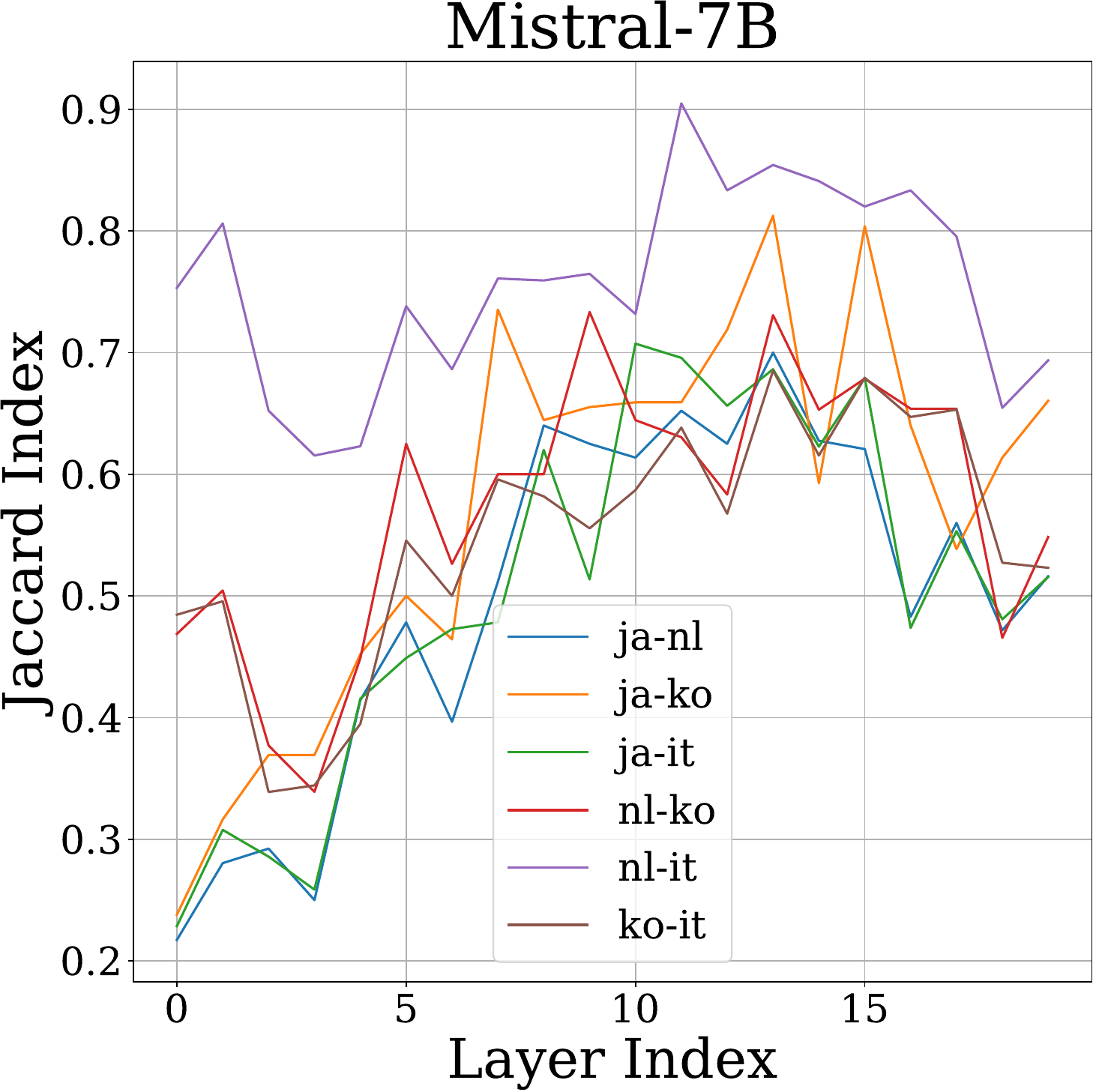}
  \includegraphics[width=0.49\linewidth]{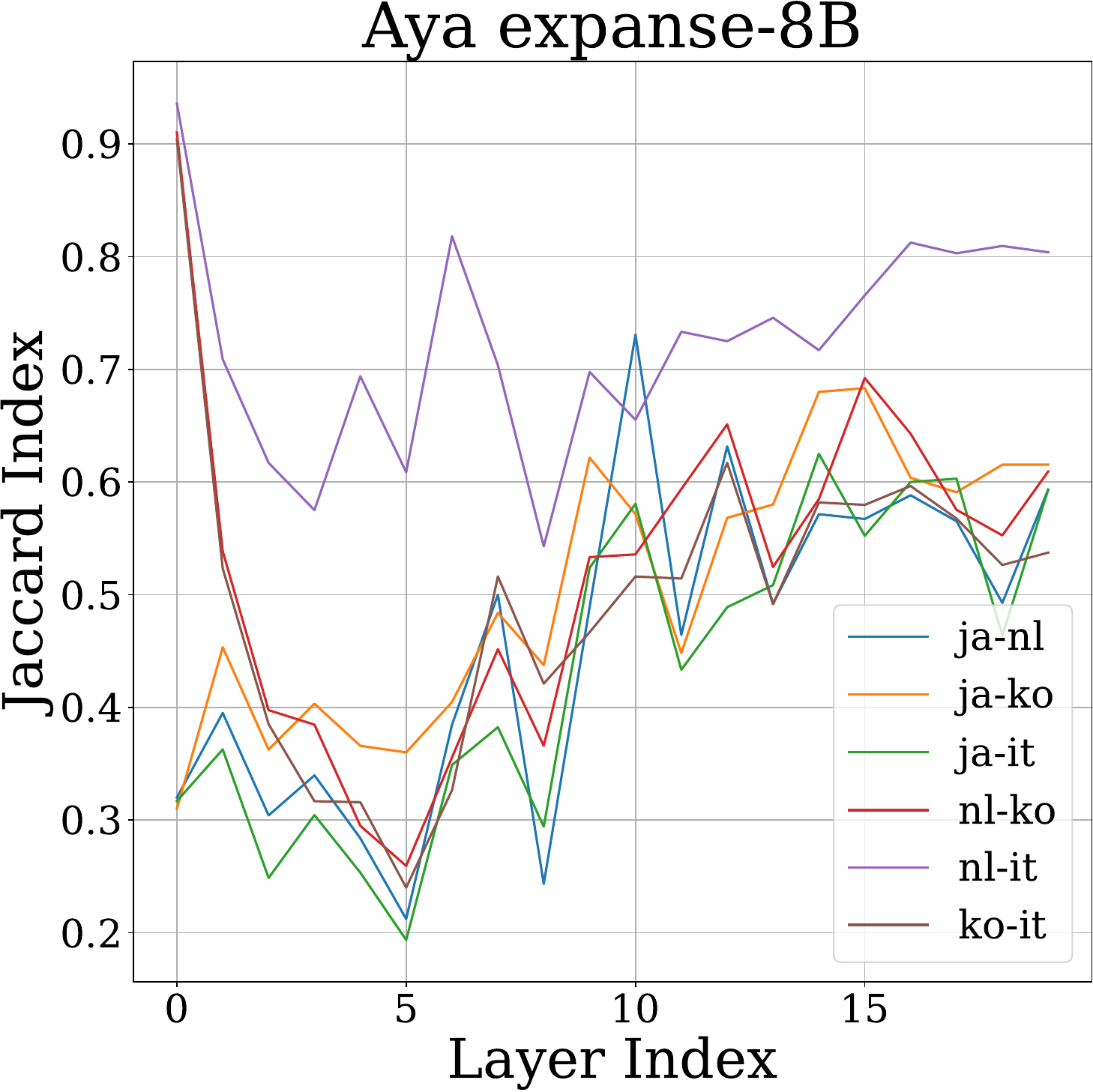}

  \caption{\textbf{Overlap ratio (Jaccard Index) of Type-1 Transfer Neurons across language pairs and decoder layers (Mistral-7B, and Aya expanse-8B)}. Higher values on the vertical axis indicate greater overlap of neurons across languages.}
  \label{fig:appendix:type1_overlap_mistral_aya}
\end{figure*}

\begin{figure*}[t]
  \centering
  \includegraphics[width=0.32\linewidth]{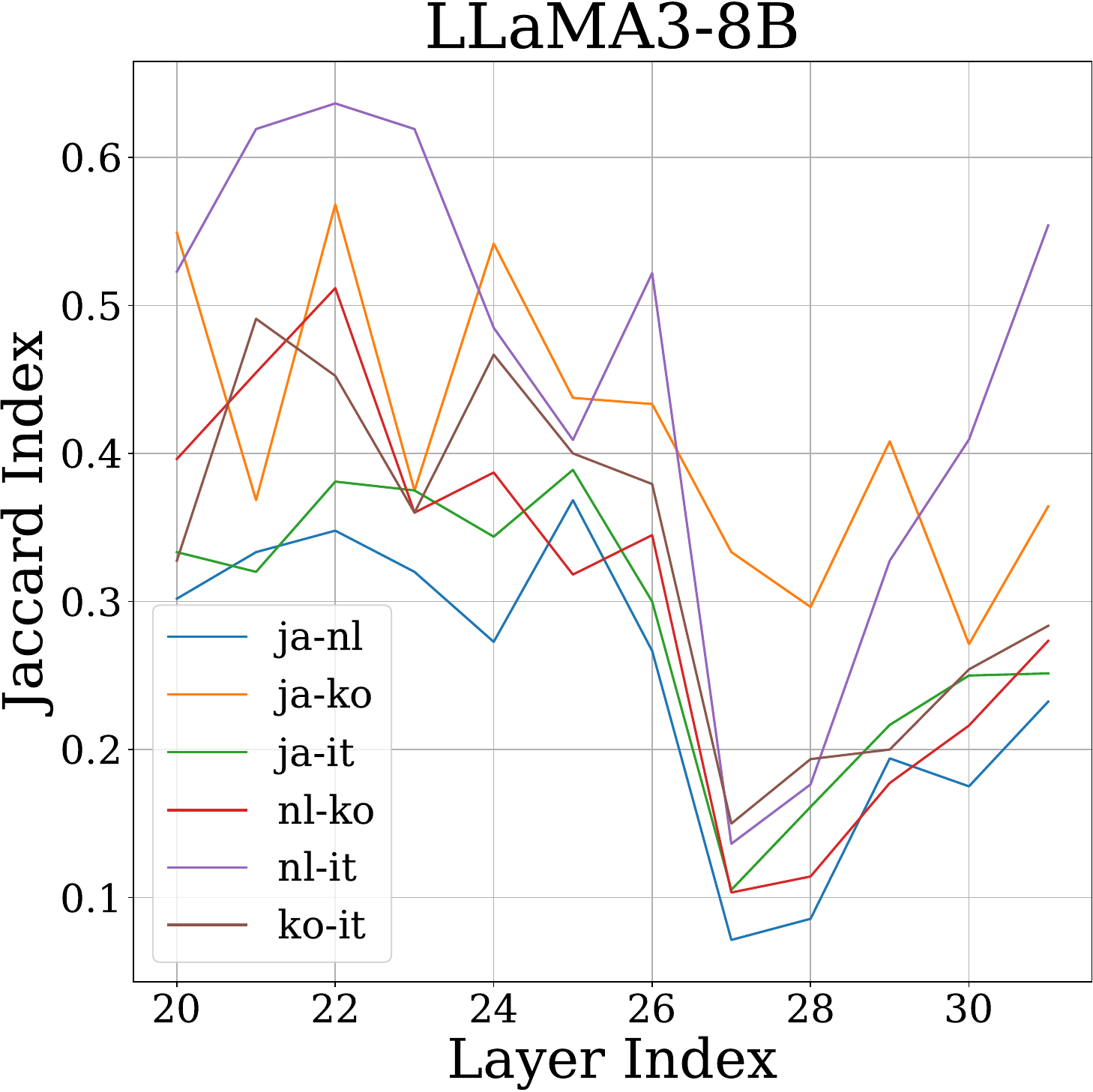}
  \includegraphics[width=0.32\linewidth]{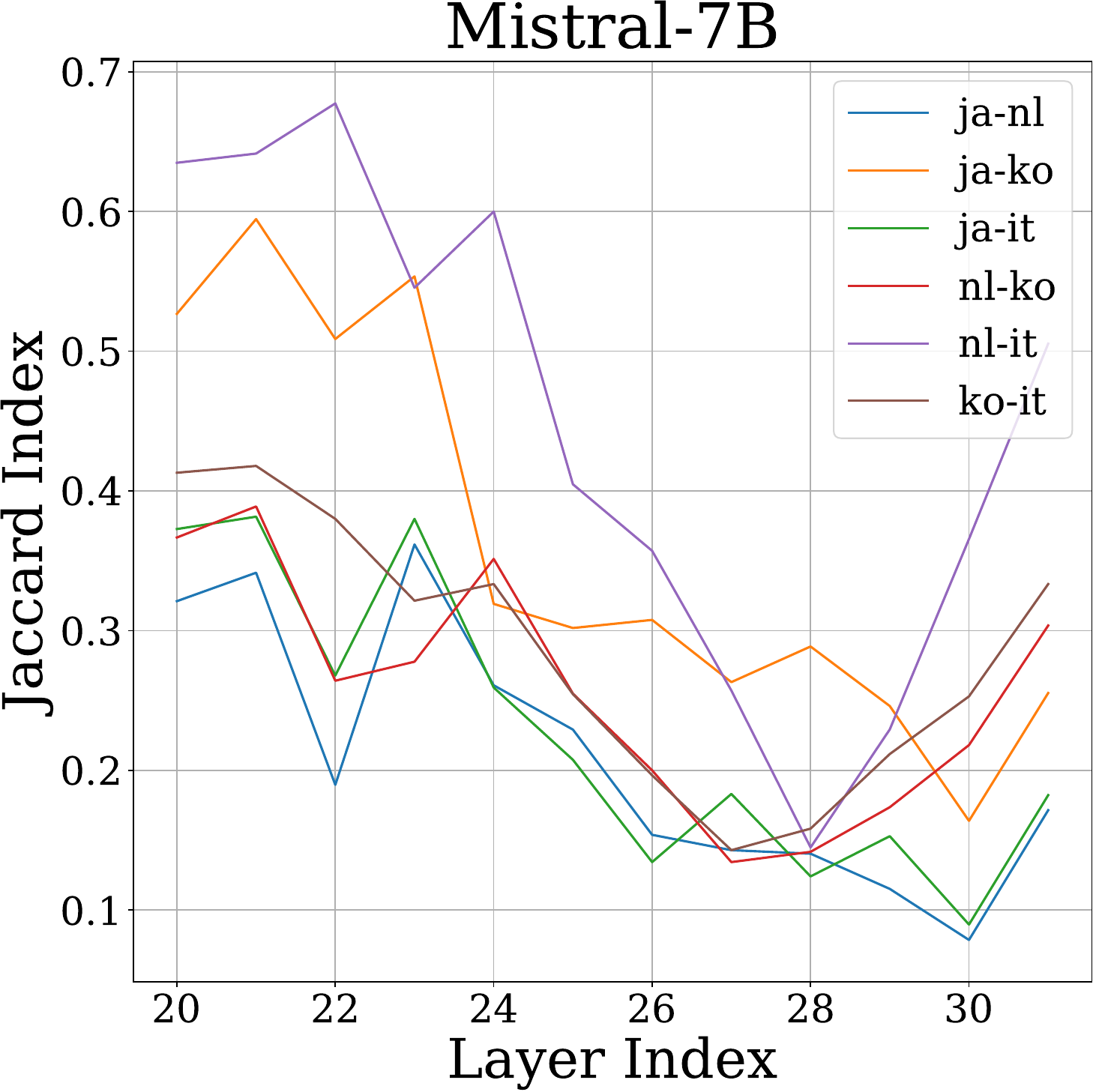}
  \includegraphics[width=0.32\linewidth]{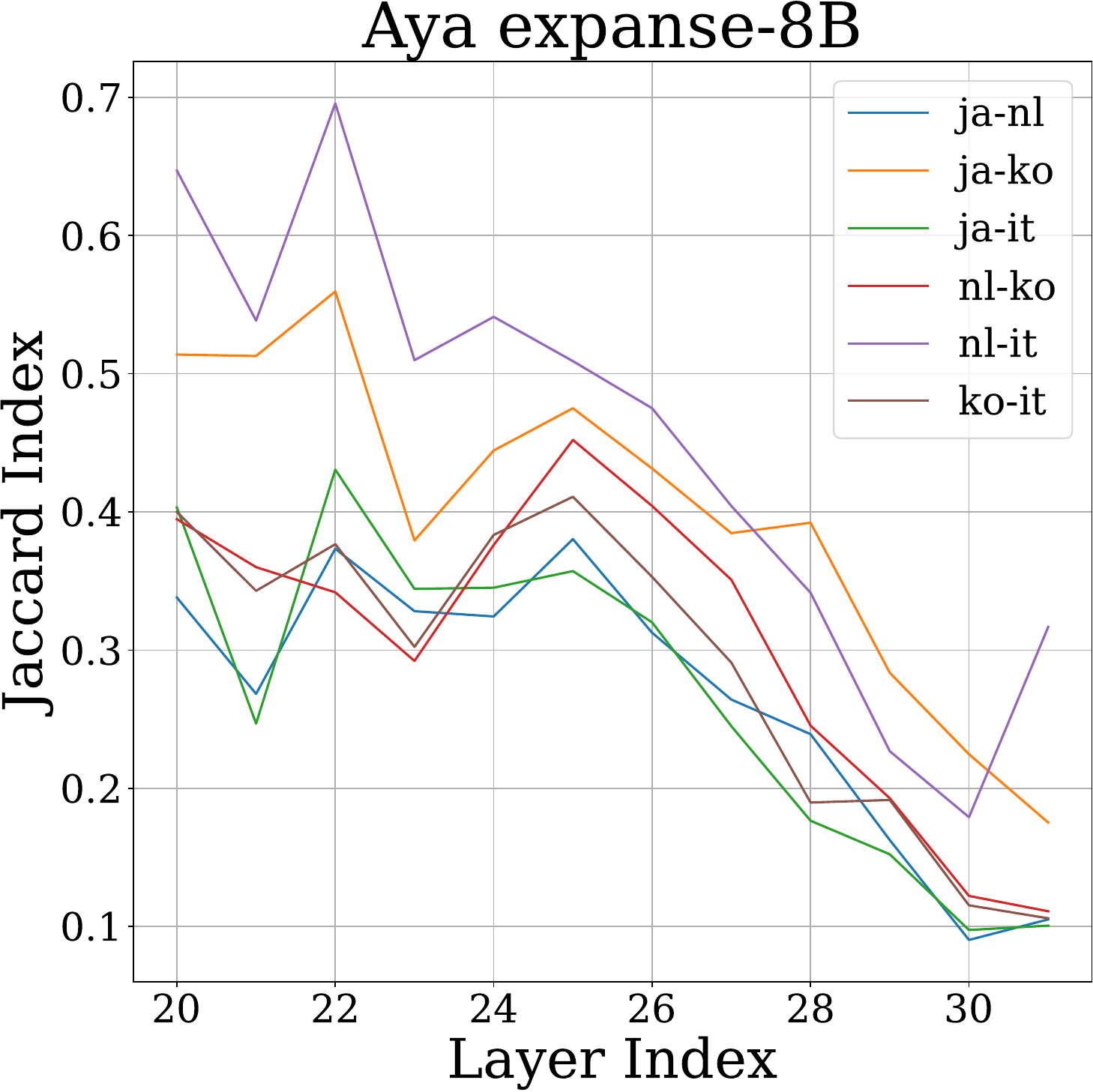}

  \caption{\textbf{Overlap ratio (Jaccard Index) of Type-2 Transfer Neurons across language pairs and decoder layers (LLaMA3-8B, Mistral-7B, and Aya expanse-8B)}.}
  \label{fig:appendix:type2_overlap_llama_mistral_aya}
\end{figure*}

\paragraph{Type-1 Transfer Neurons.}Fig.~\ref{fig:appendix:type1_overlap_mistral_aya} shows the overlap ratio (i.e., Jaccard Index) for Type-1 neurons across language pairs for Mistral-7B and Aya expanse-8B. As indicated, their overlap ratio increase as language representations move closer to each other towards the shared latent space in middle layers.
\paragraph{Type-2 Transfer Neurons.}Fig.~\ref{fig:appendix:type2_overlap_llama_mistral_aya} shows the overlap ratio (i.e., Jaccard Index) for Type-2 neurons across language pairs for all the models. As shown, their overlap ratio decrease as language representations move to each language-specific latent space for output generation.

\subsection{Detecting Language-Specific Neurons}
\label{sec:appendix:language specific neurons}
Building on the method proposed by \citet{kojima2024}, we independently identify language-specific neurons. \citet{kojima2024} detected such neurons by computing the \textit{Average Precision} (i.e., \textit{Area Under the Precision-Recall Curve}) between each neuron’s activation — defined as the output of the non-linear activation function in the MLP module — and language-labeled sentences\footnote{This method was originally proposed by \citet{suau2022selfcond}.}. Sentences in the target language were labeled as \texttt{1}, and those in other languages were labeled as \texttt{0}. They defined both the top-1k and bottom-1k (2k neurons in total) neurons in the score ranking as language-specific neurons, considering not only those positively correlated with the target language but also those negatively correlated.

However, their method is ambiguous regarding how many neurons should be selected from the top and bottom rankings. Therefore, in our approach, we adopt the \textit{Correlation Ratio} (Appendix~\ref{sec:appendix:corr_ratio}) instead of Average Precision as a metric, as it allows us to capture both positive and negative correlations simultaneously, with a single explicit value for each neuron that reflects the strength of the correlation.

Let $S$ be the set of all sentences across all languages, i.e., the union of English, Japanese, Dutch, Korean, and Italian sentence sets:

\begin{equation}
    S = \{S_{\mathrm{en}}, S_{\mathrm{ja}}, S_{\mathrm{nl}}, S_{\mathrm{ko}}, S_{\mathrm{it}}\}
    \label{eq:appendix:all_sentences_all_languages}
\end{equation}

Each language contains 1k sentence samples (5k in total). For instance, to detect Japanese-specific neurons, we assign \texttt{label1} to all 1k Japanese sentences and \texttt{label0} to sentences from all other languages. Then, for each neuron, we compute the correlation ratio between its activation values and the labels of the 5k sentence samples.

\begin{figure*}[t]
  \centering

  \includegraphics[width=0.19\linewidth]{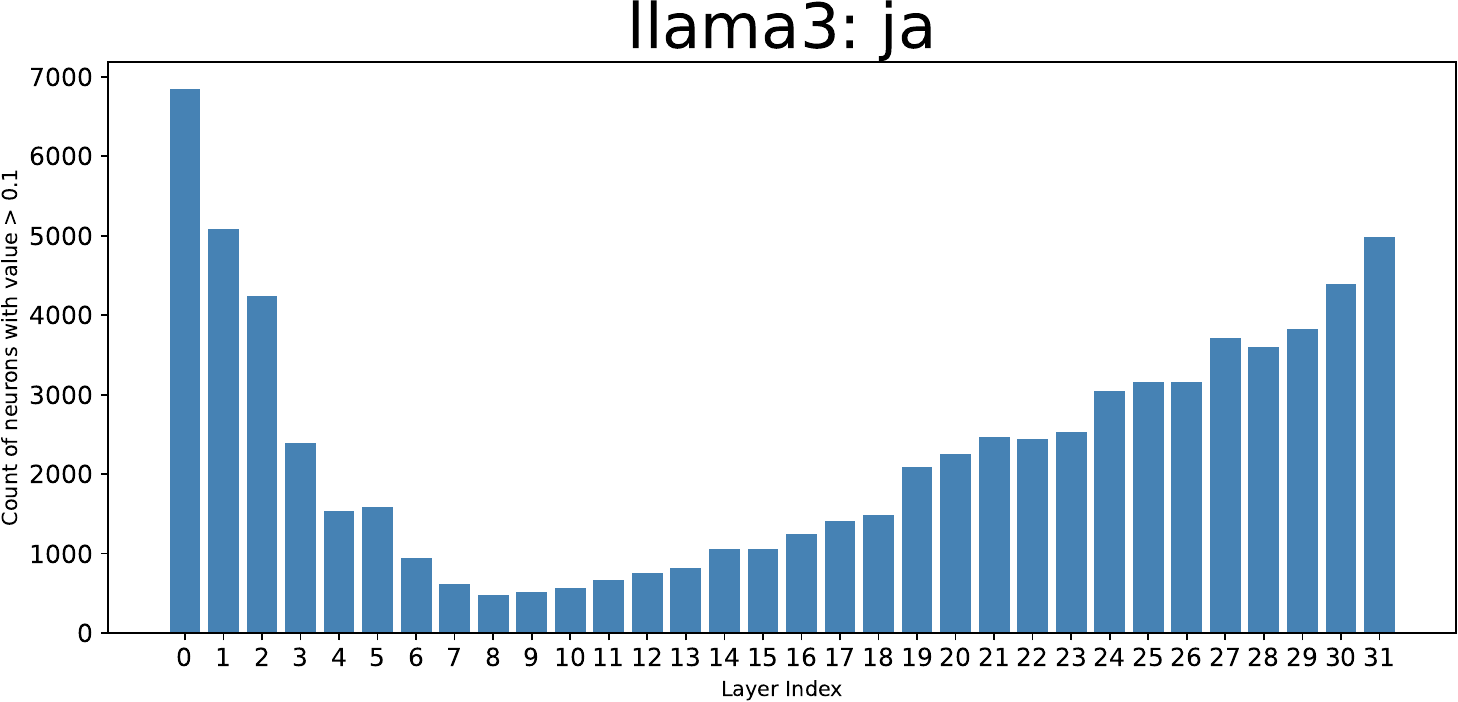}
  \includegraphics[width=0.19\linewidth]{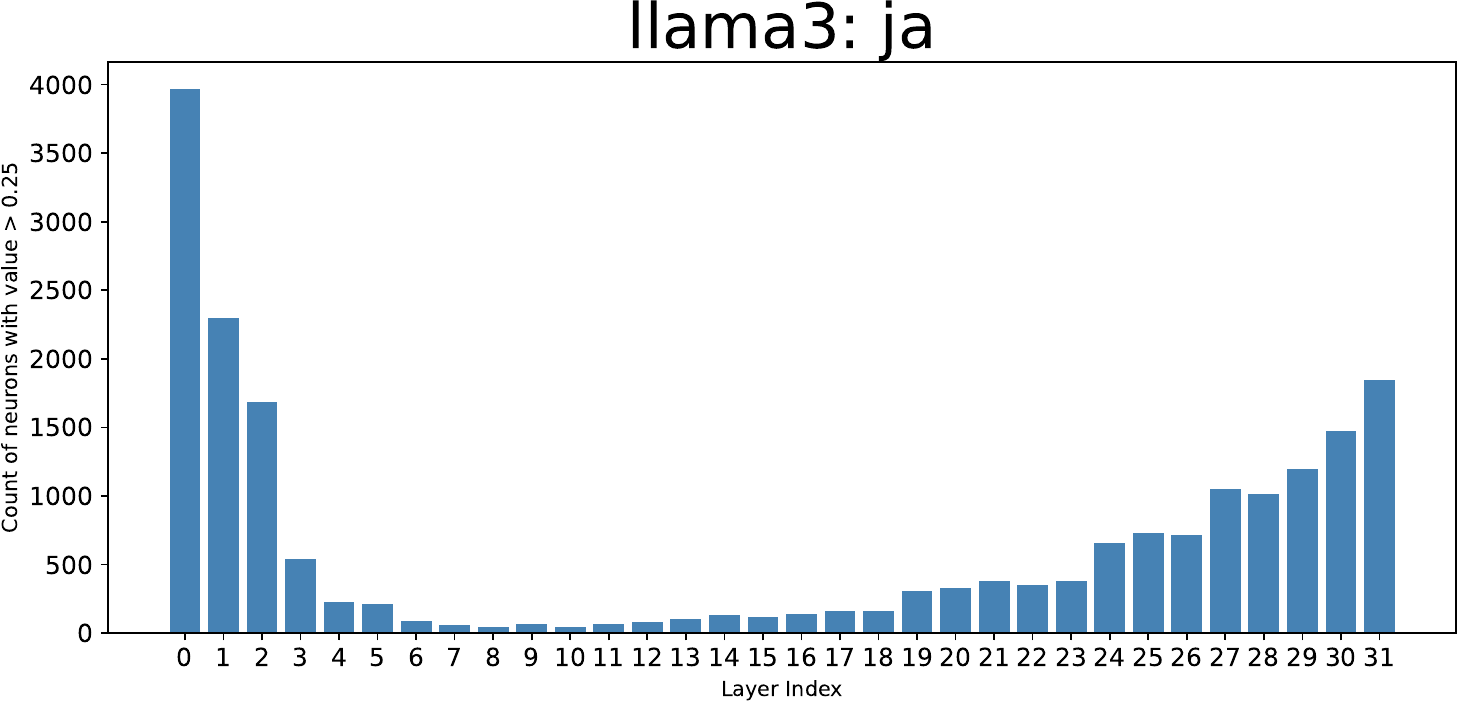}
  \includegraphics[width=0.19\linewidth]{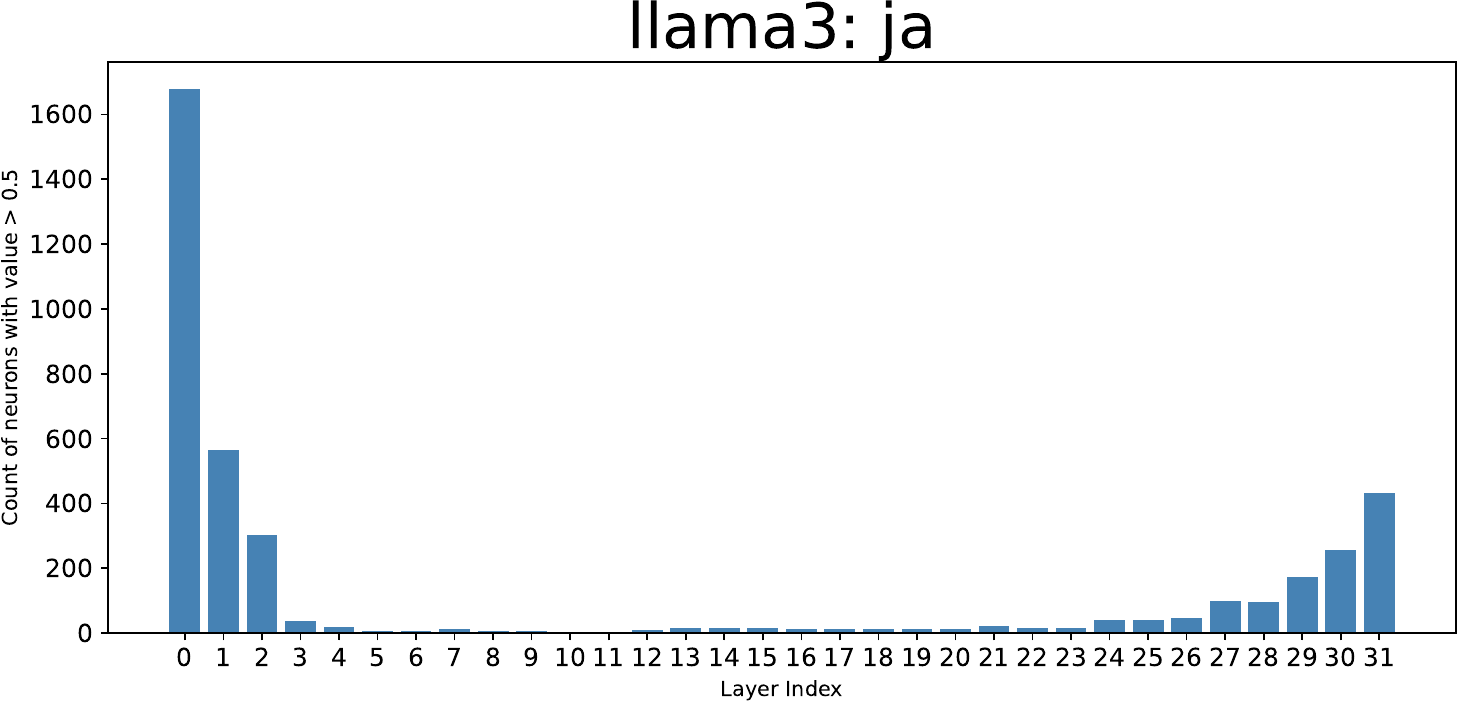}
  \includegraphics[width=0.19\linewidth]{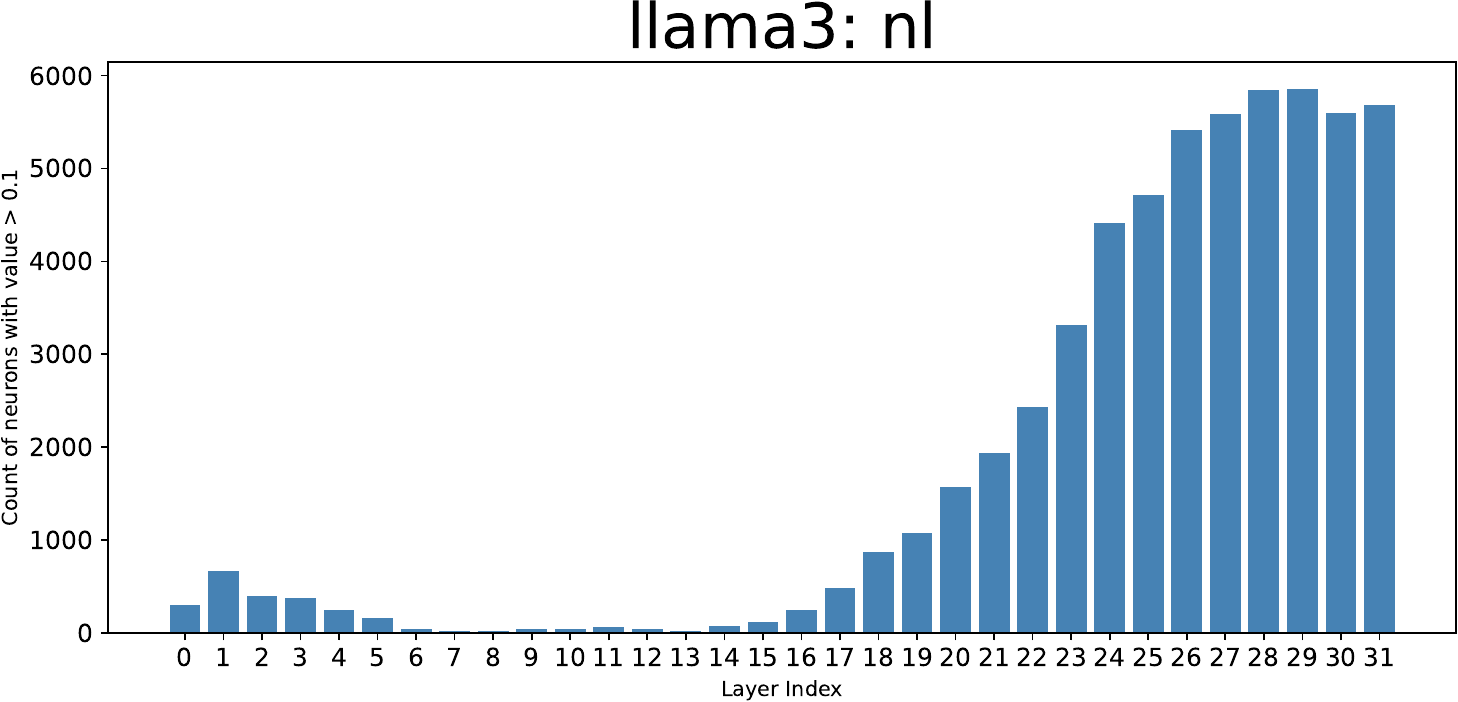}
  \includegraphics[width=0.19\linewidth]{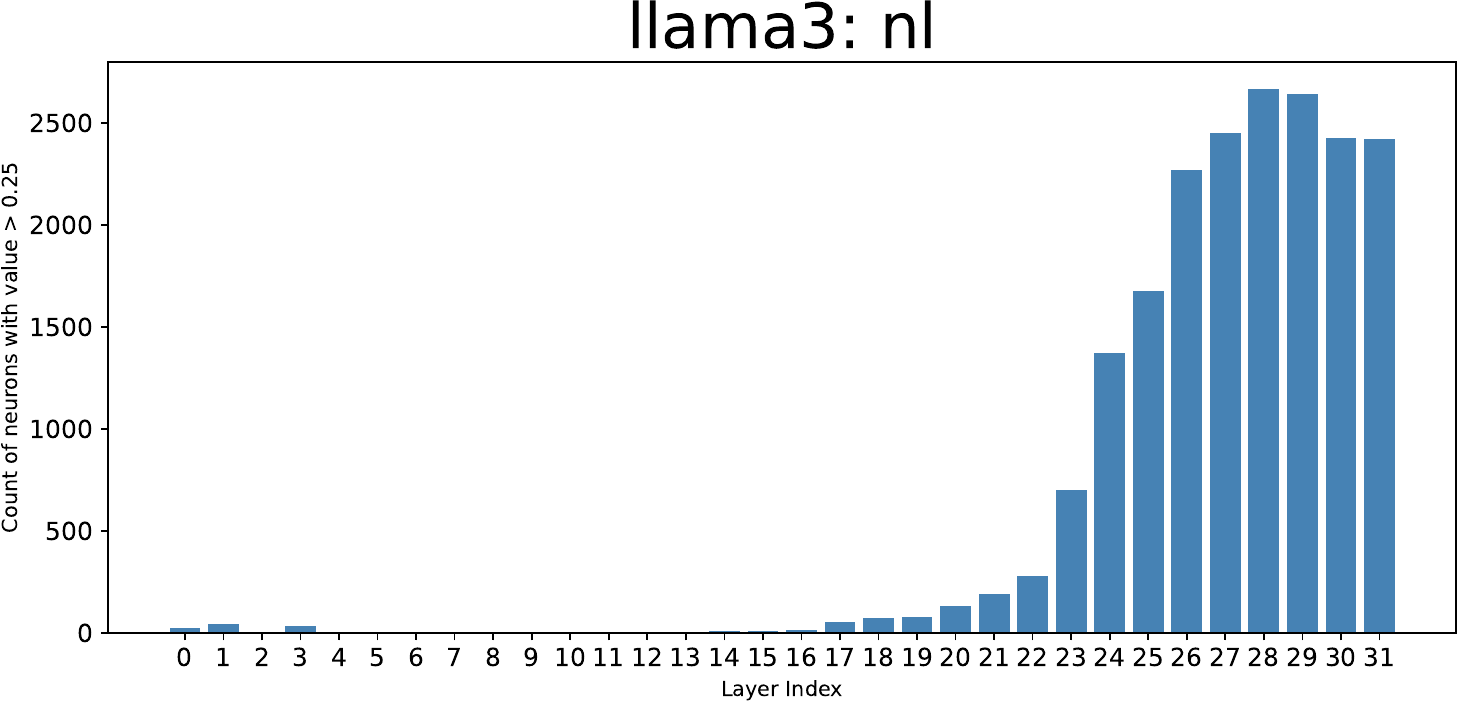}

  \begin{minipage}{0.19\linewidth}\centering ja, $\geq0.1$\end{minipage}
  \begin{minipage}{0.19\linewidth}\centering ja, $\geq0.25$\end{minipage}
  \begin{minipage}{0.19\linewidth}\centering ja, $\geq0.5$\end{minipage}
  \begin{minipage}{0.19\linewidth}\centering nl, $\geq0.1$\end{minipage}
  \begin{minipage}{0.19\linewidth}\centering nl, $\geq0.25$\end{minipage}

  \includegraphics[width=0.19\linewidth]{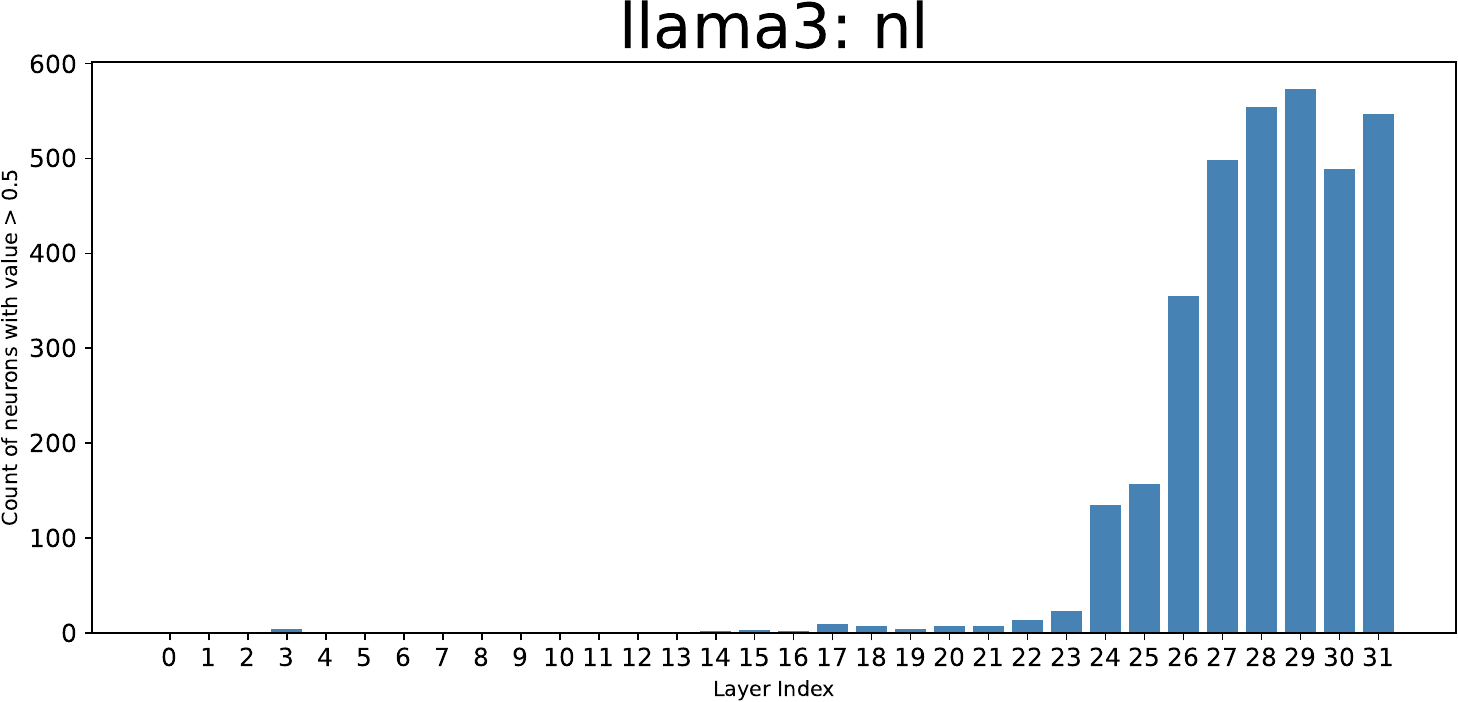}
  \includegraphics[width=0.19\linewidth]{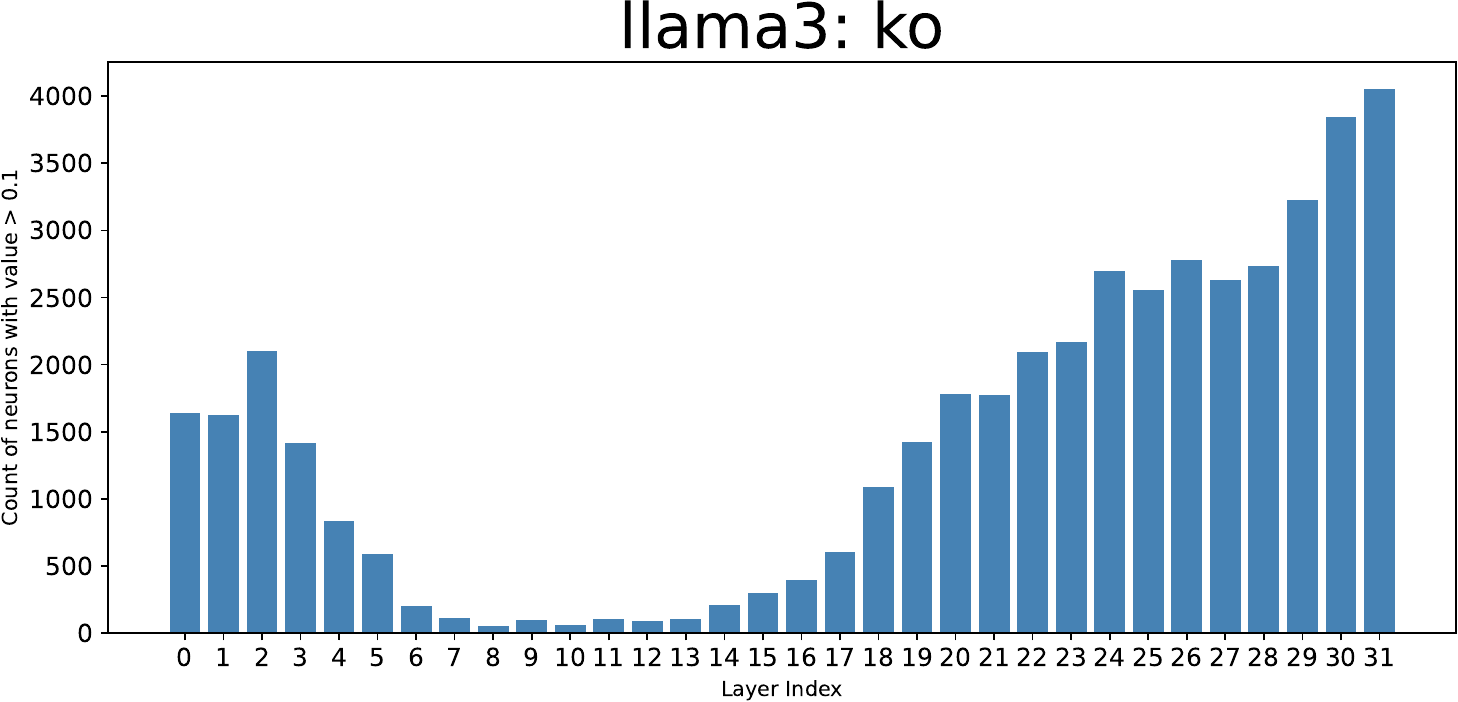}
  \includegraphics[width=0.19\linewidth]{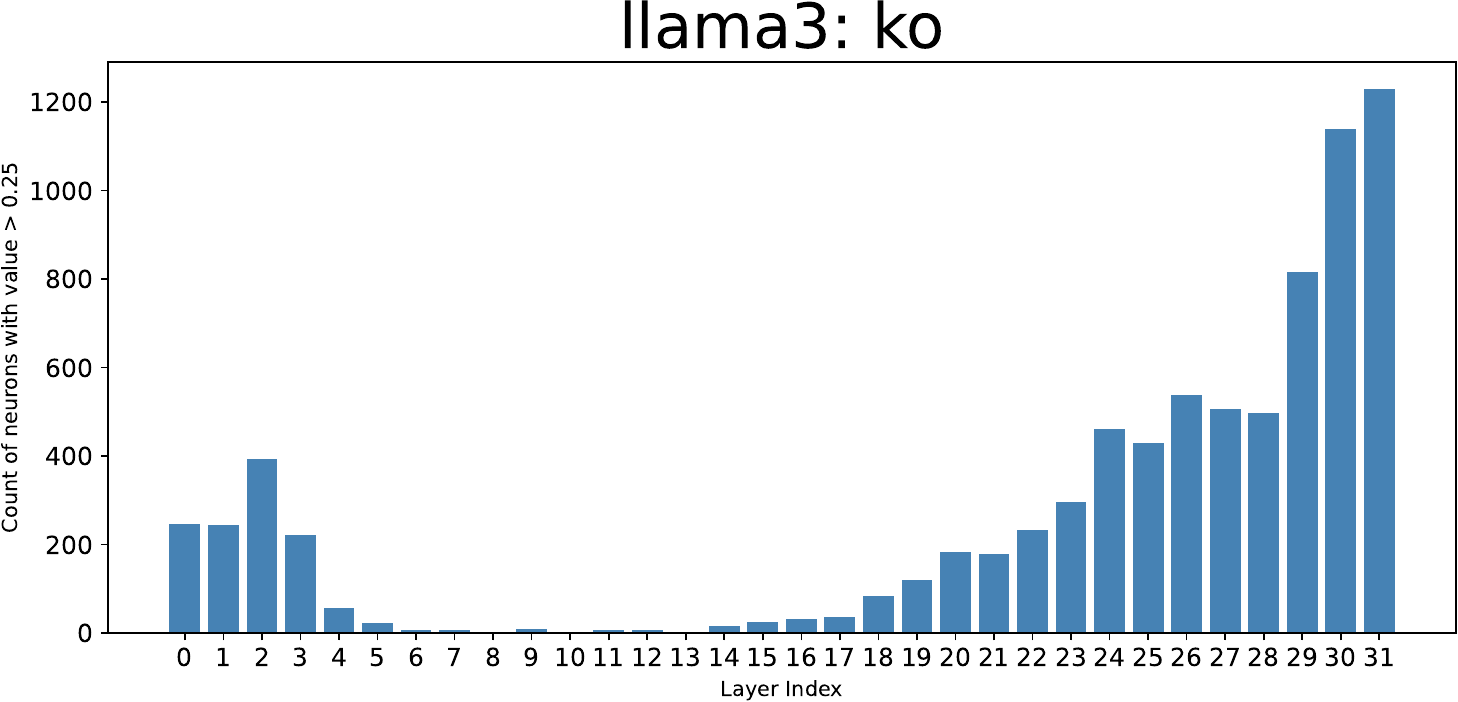}
  \includegraphics[width=0.19\linewidth]{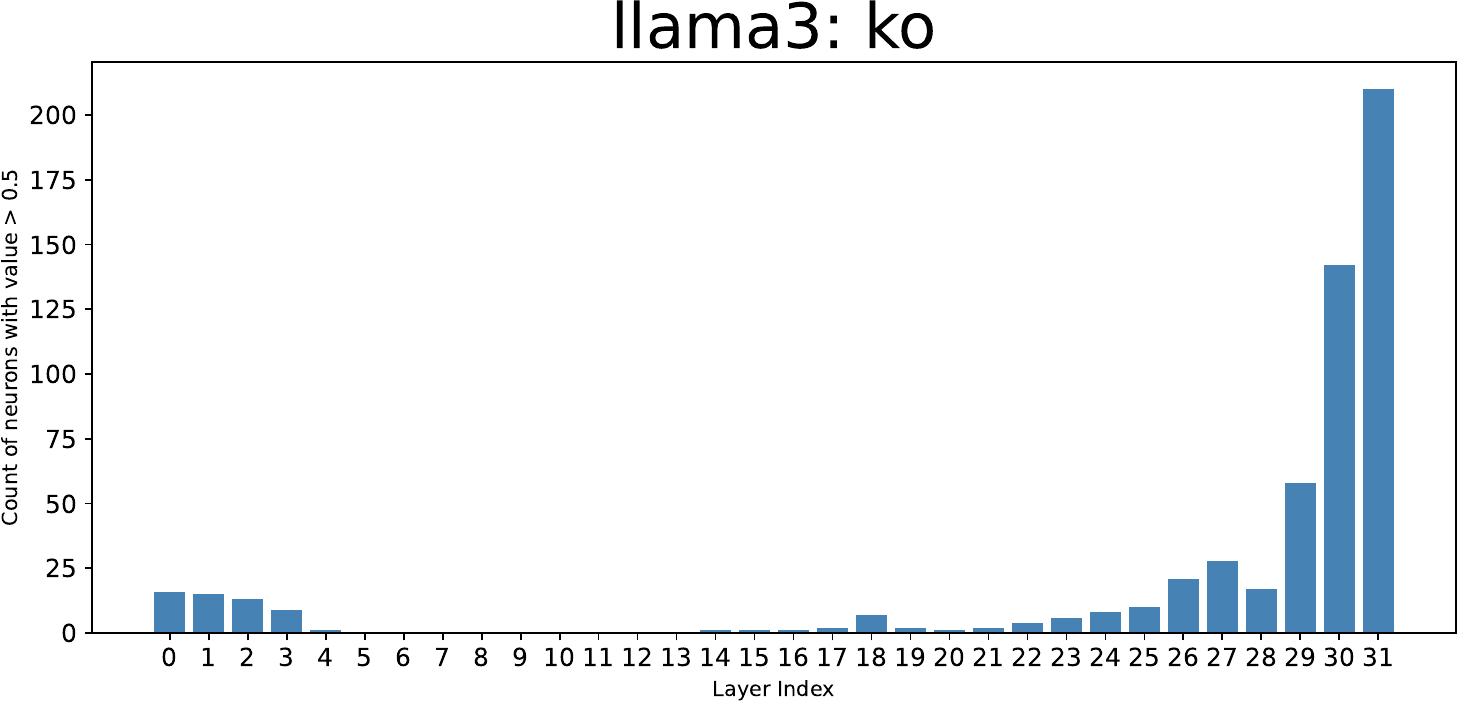}
  \includegraphics[width=0.19\linewidth]{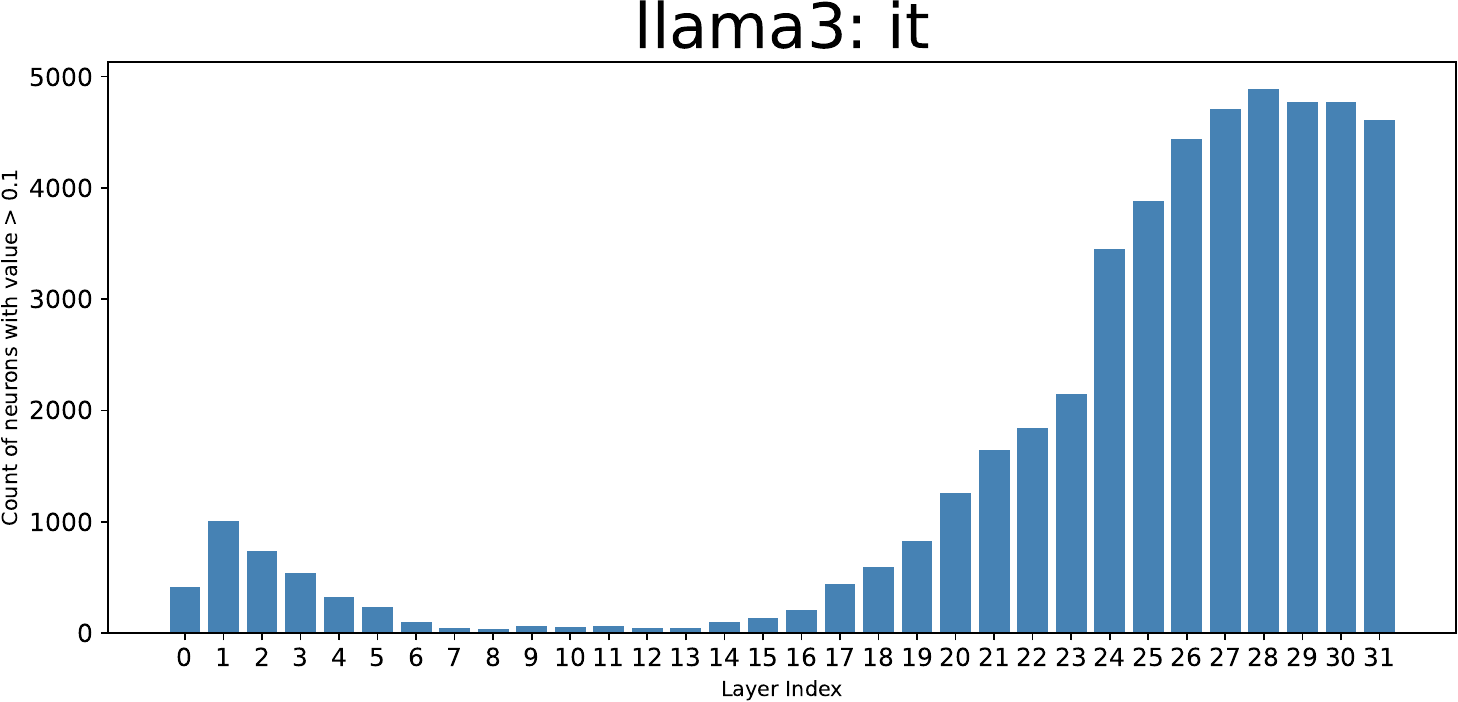}

  \begin{minipage}{0.19\linewidth}\centering nl, $\geq0.5$\end{minipage}
  \begin{minipage}{0.19\linewidth}\centering ko, $\geq0.1$\end{minipage}
  \begin{minipage}{0.19\linewidth}\centering ko, $\geq0.25$\end{minipage}
  \begin{minipage}{0.19\linewidth}\centering ko, $\geq0.5$\end{minipage}
  \begin{minipage}{0.19\linewidth}\centering it, $\geq0.1$\end{minipage}

  \includegraphics[width=0.19\linewidth]{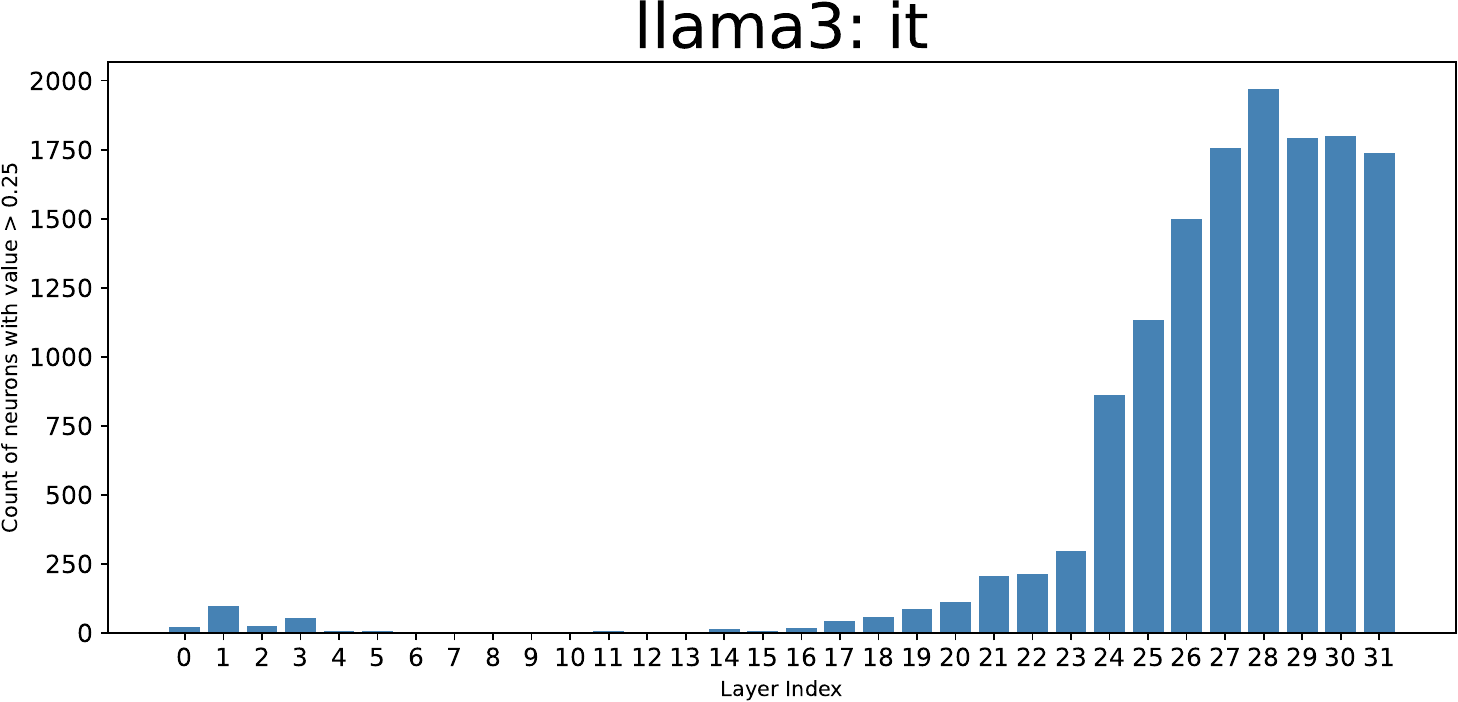}
  \includegraphics[width=0.19\linewidth]{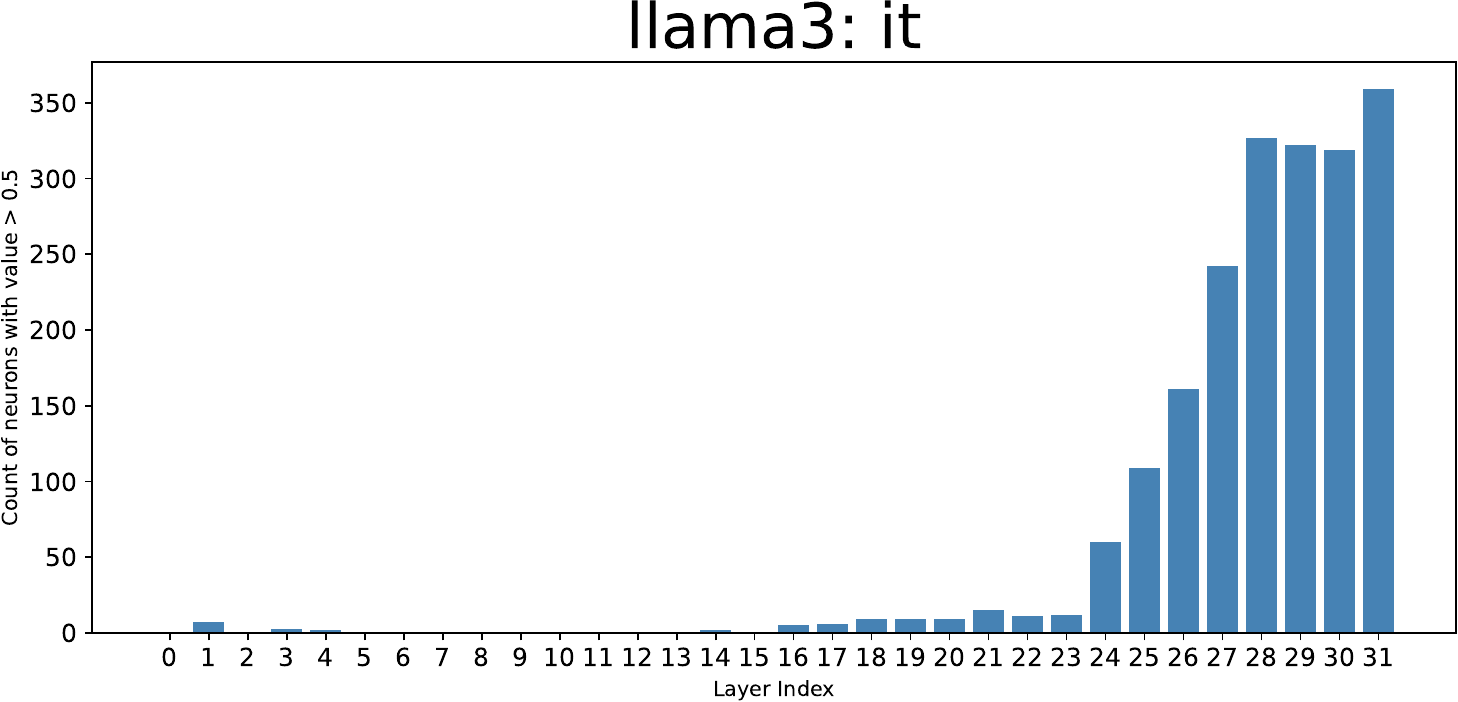}

  \begin{minipage}{0.19\linewidth}\centering it, $\geq0.25$\end{minipage}
  \begin{minipage}{0.19\linewidth}\centering it, $\geq0.5$\end{minipage}

\caption{\textbf{Distribution of language-specific neurons (LLaMA3-8B).} Each label below the figure indicates the target language and the threshold for correlation ratio. Typically, a correlation ratio above \textbf{0.1} indicates a correlation, above \textbf{0.25} indicates a moderately strong correlation, and above \textbf{0.5} indicates a strong correlation. horizontal-axis denotes the layer indices, and vertical-axis represents the number of neurons.}
  \label{fig:appendix:distribution_lang_specific_neurons_llama}
\end{figure*}

\begin{figure*}[t]
  \centering

  \includegraphics[width=0.19\linewidth]{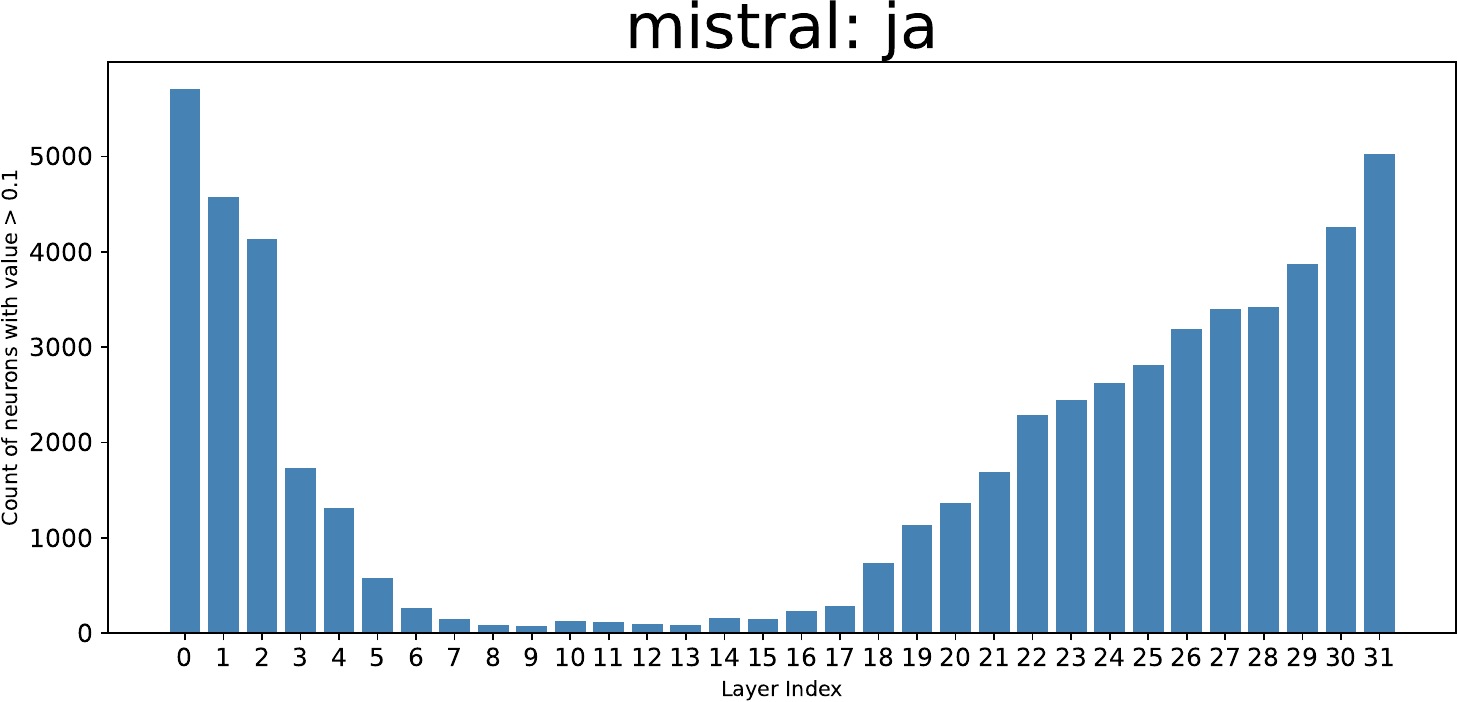}
  \includegraphics[width=0.19\linewidth]{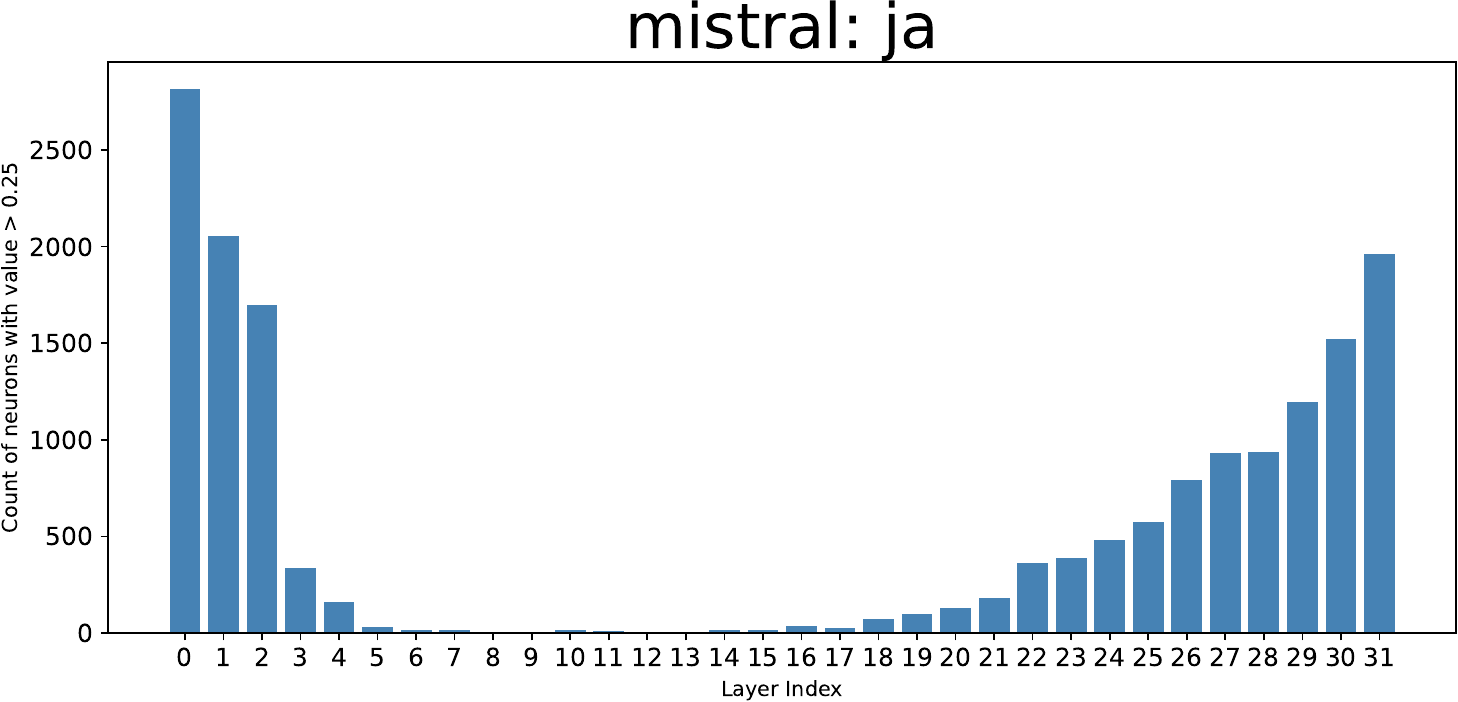}
  \includegraphics[width=0.19\linewidth]{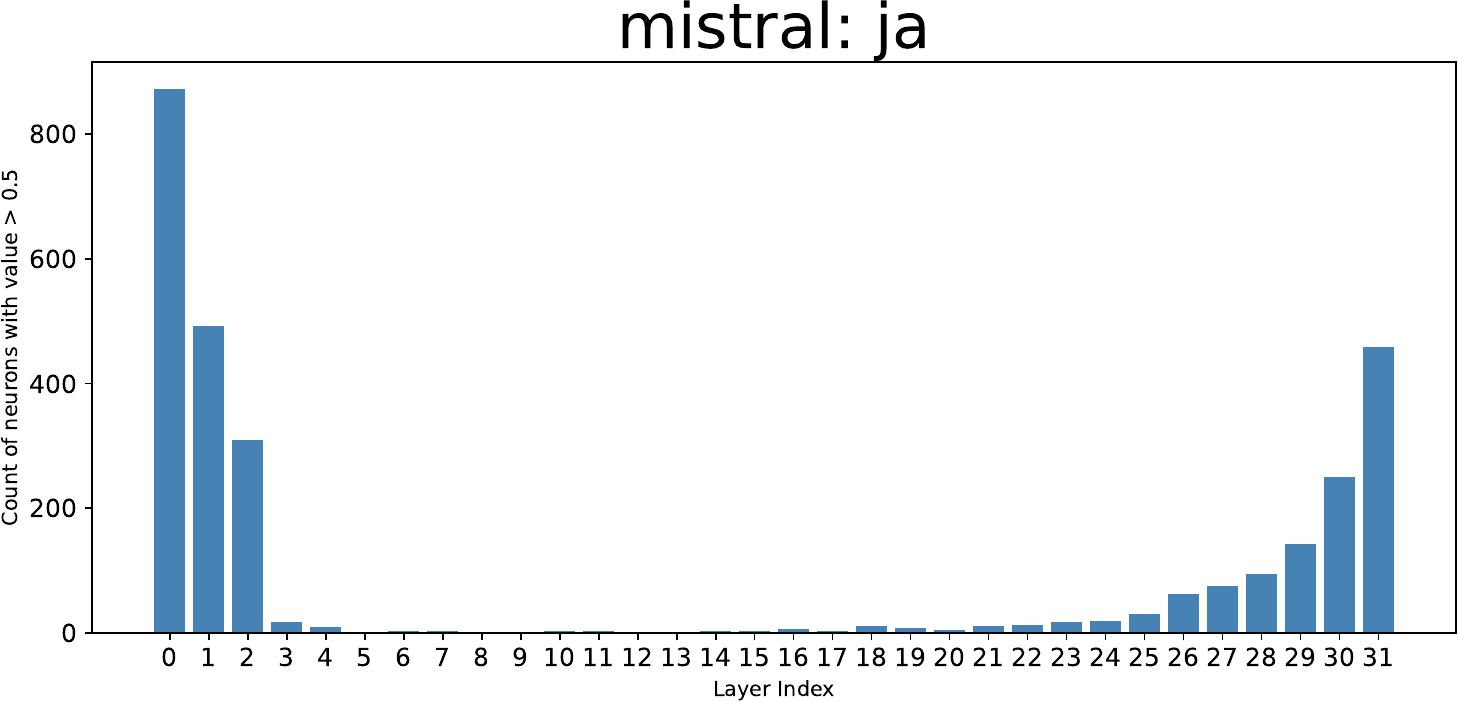}
  \includegraphics[width=0.19\linewidth]{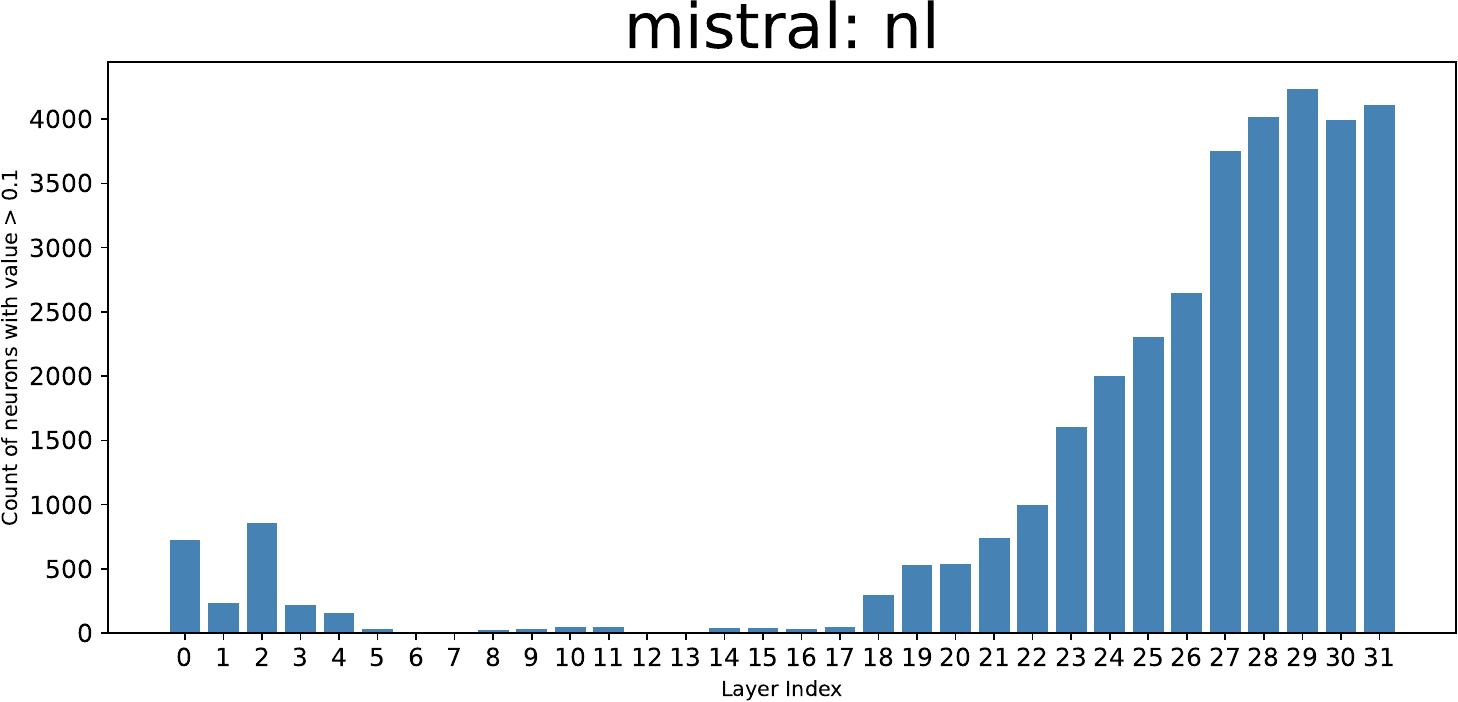}
  \includegraphics[width=0.19\linewidth]{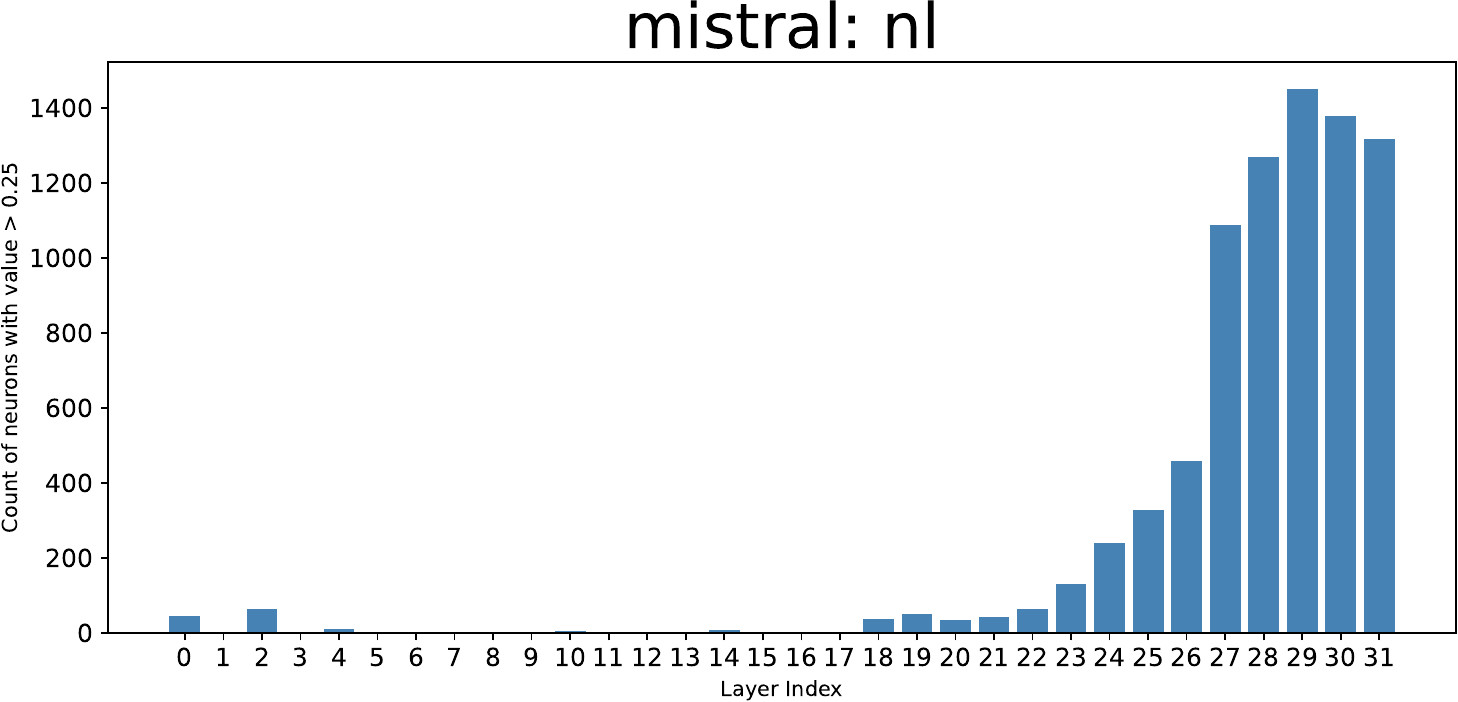}

  \begin{minipage}{0.19\linewidth}\centering ja, $\geq0.1$\end{minipage}
  \begin{minipage}{0.19\linewidth}\centering ja, $\geq0.25$\end{minipage}
  \begin{minipage}{0.19\linewidth}\centering ja, $\geq0.5$\end{minipage}
  \begin{minipage}{0.19\linewidth}\centering nl, $\geq0.1$\end{minipage}
  \begin{minipage}{0.19\linewidth}\centering nl, $\geq0.25$\end{minipage}

  \includegraphics[width=0.19\linewidth]{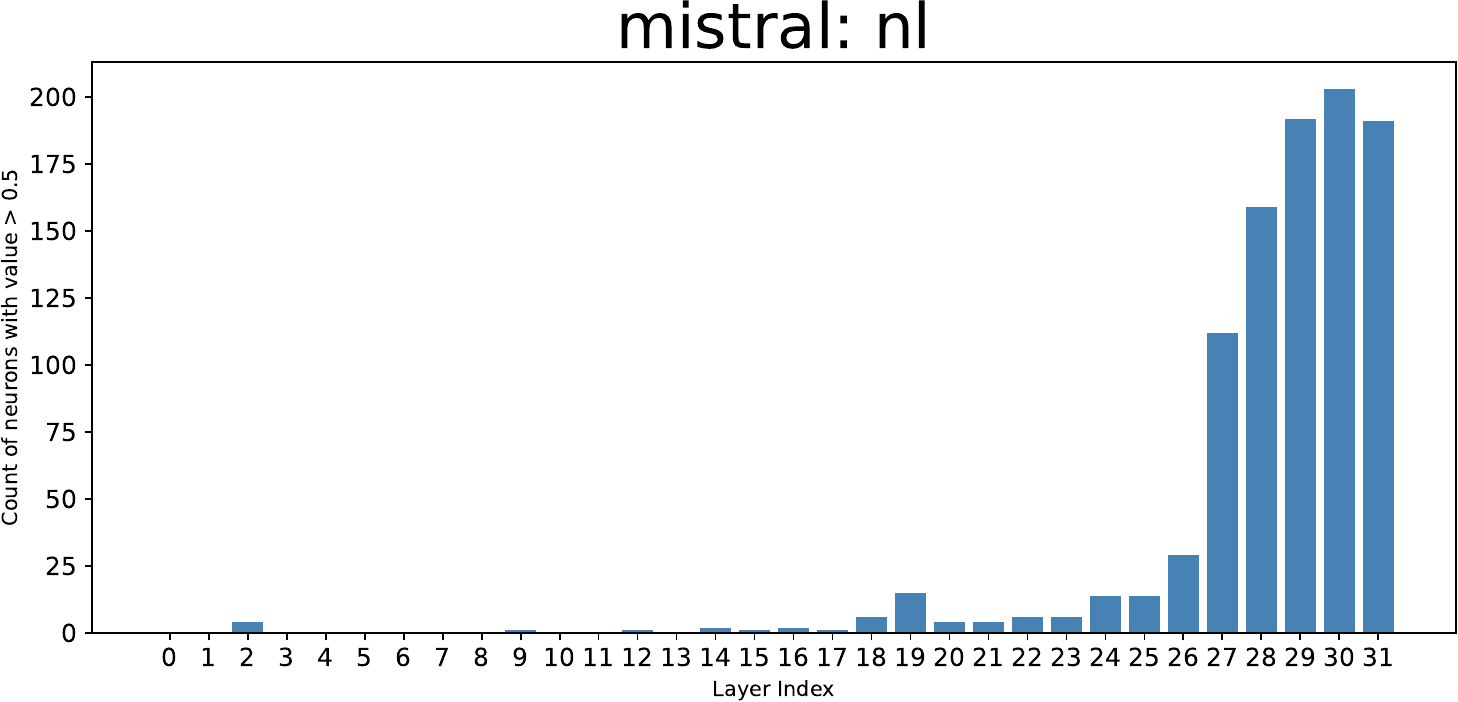}
  \includegraphics[width=0.19\linewidth]{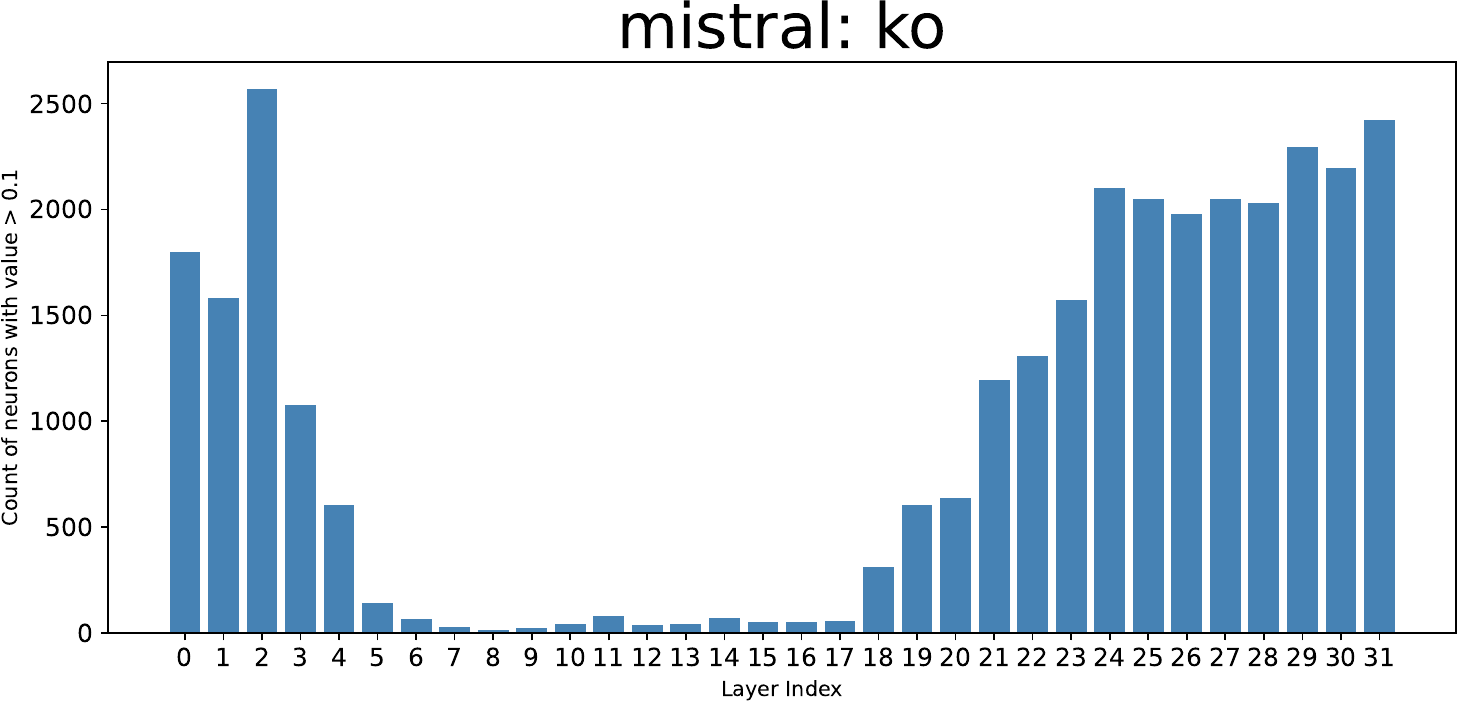}
  \includegraphics[width=0.19\linewidth]{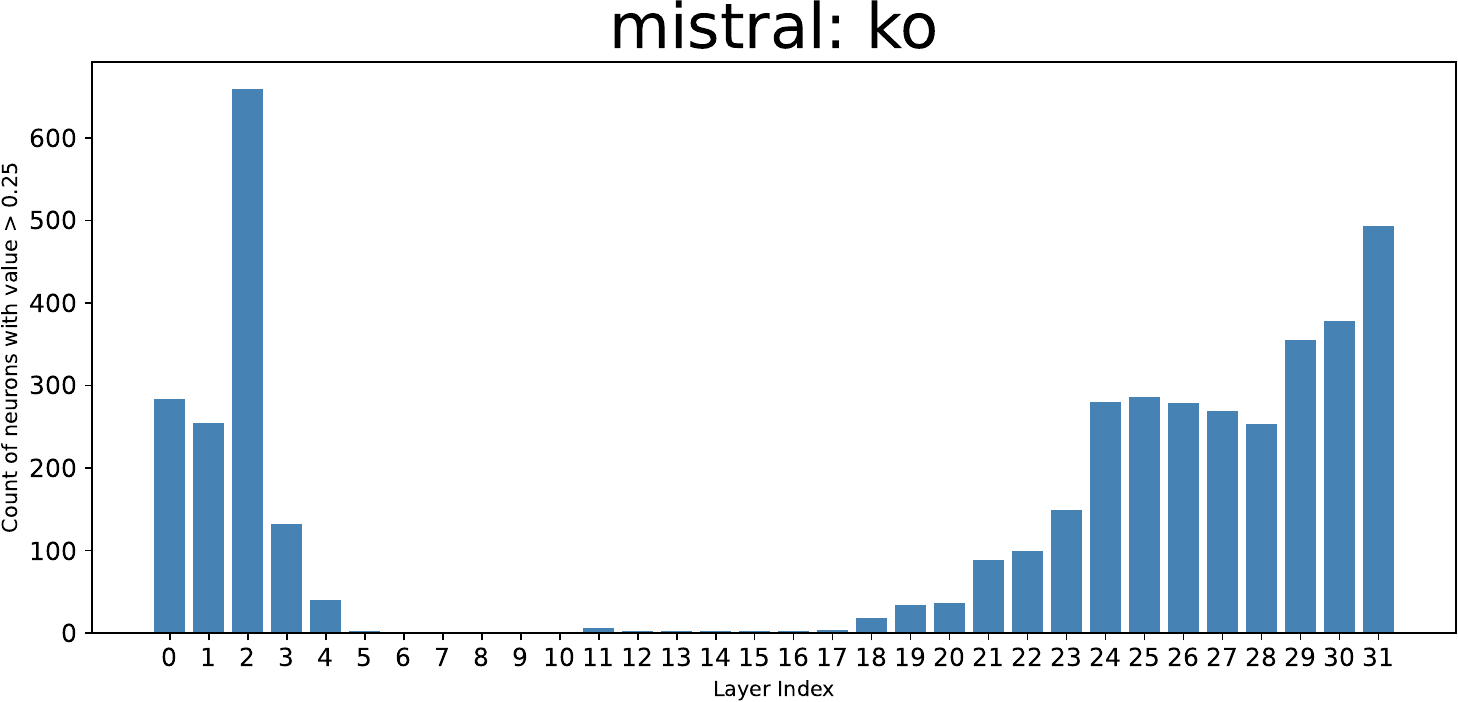}
  \includegraphics[width=0.19\linewidth]{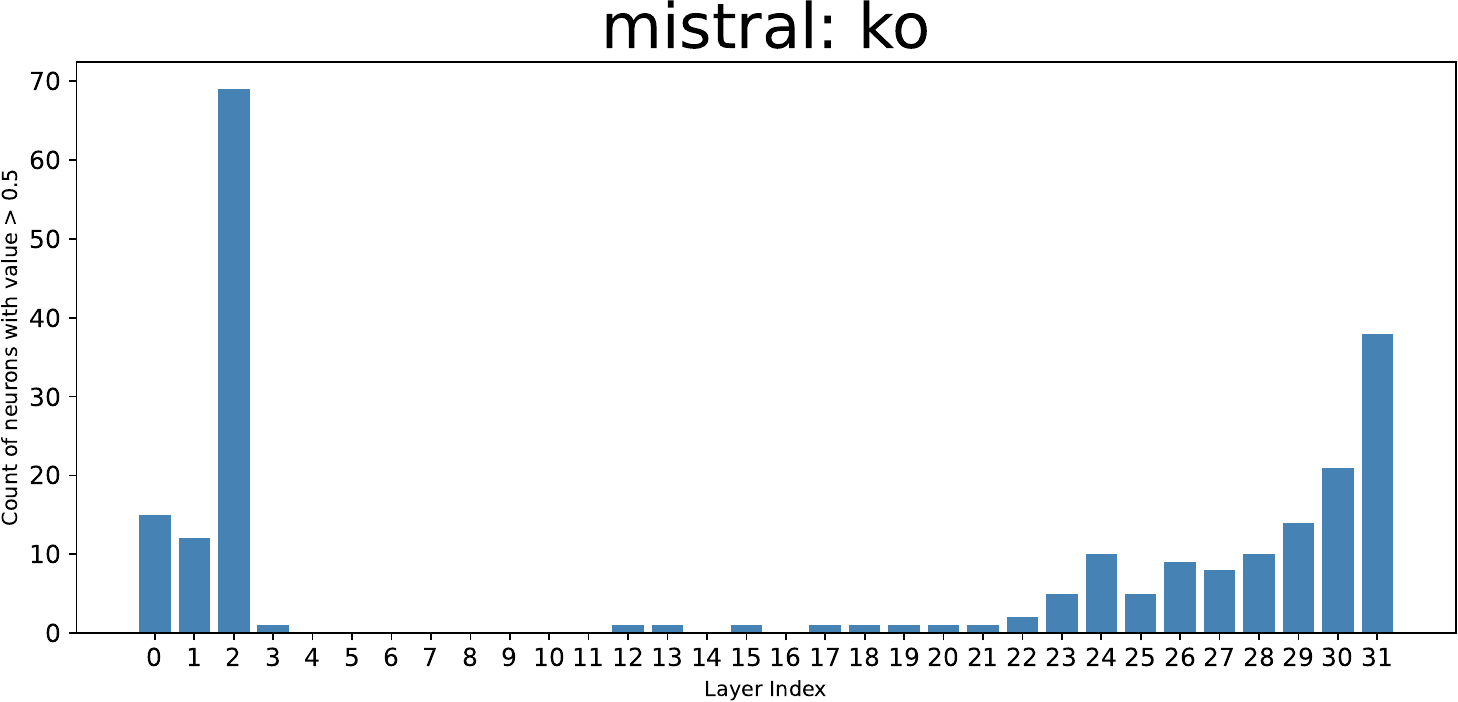}
  \includegraphics[width=0.19\linewidth]{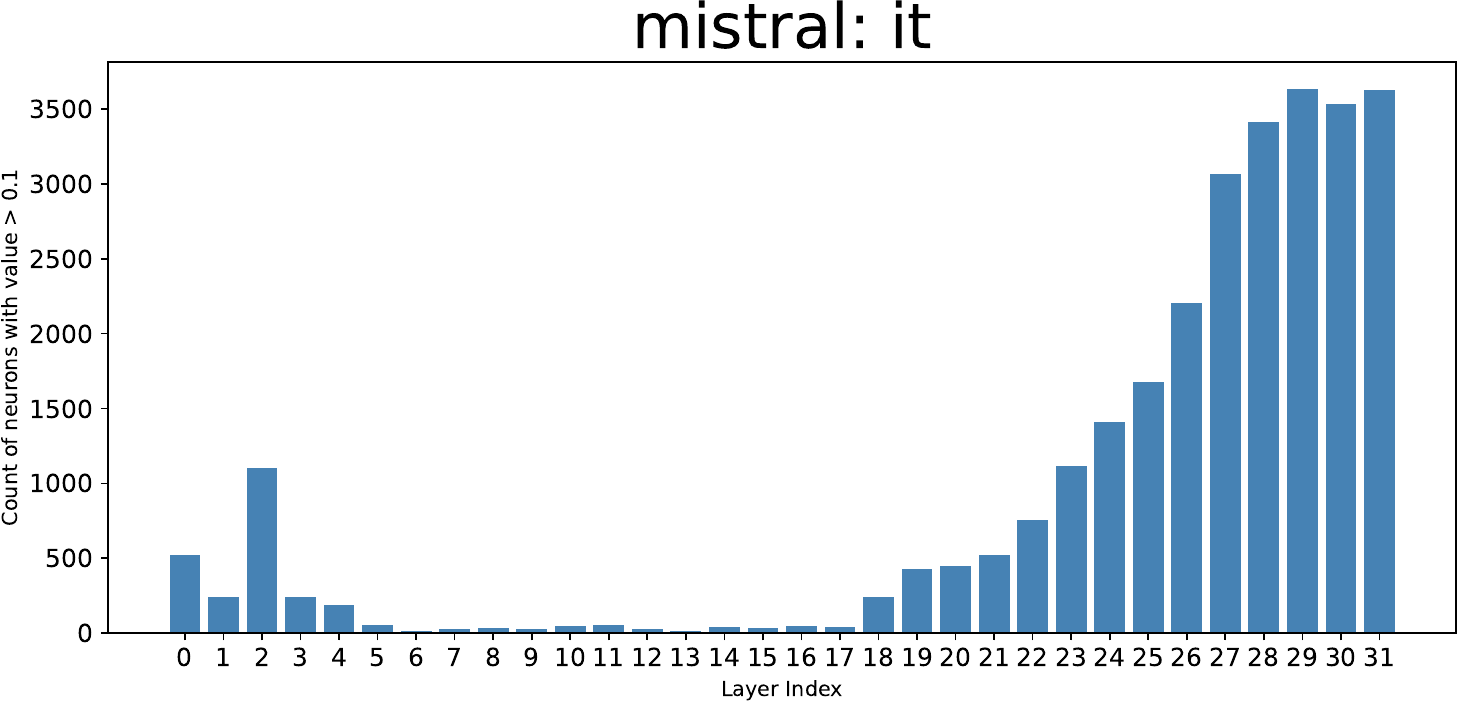}

  \begin{minipage}{0.19\linewidth}\centering nl, $\geq0.5$\end{minipage}
  \begin{minipage}{0.19\linewidth}\centering ko, $\geq0.1$\end{minipage}
  \begin{minipage}{0.19\linewidth}\centering ko, $\geq0.25$\end{minipage}
  \begin{minipage}{0.19\linewidth}\centering ko, $\geq0.5$\end{minipage}
  \begin{minipage}{0.19\linewidth}\centering it, $\geq0.1$\end{minipage}

  \includegraphics[width=0.19\linewidth]{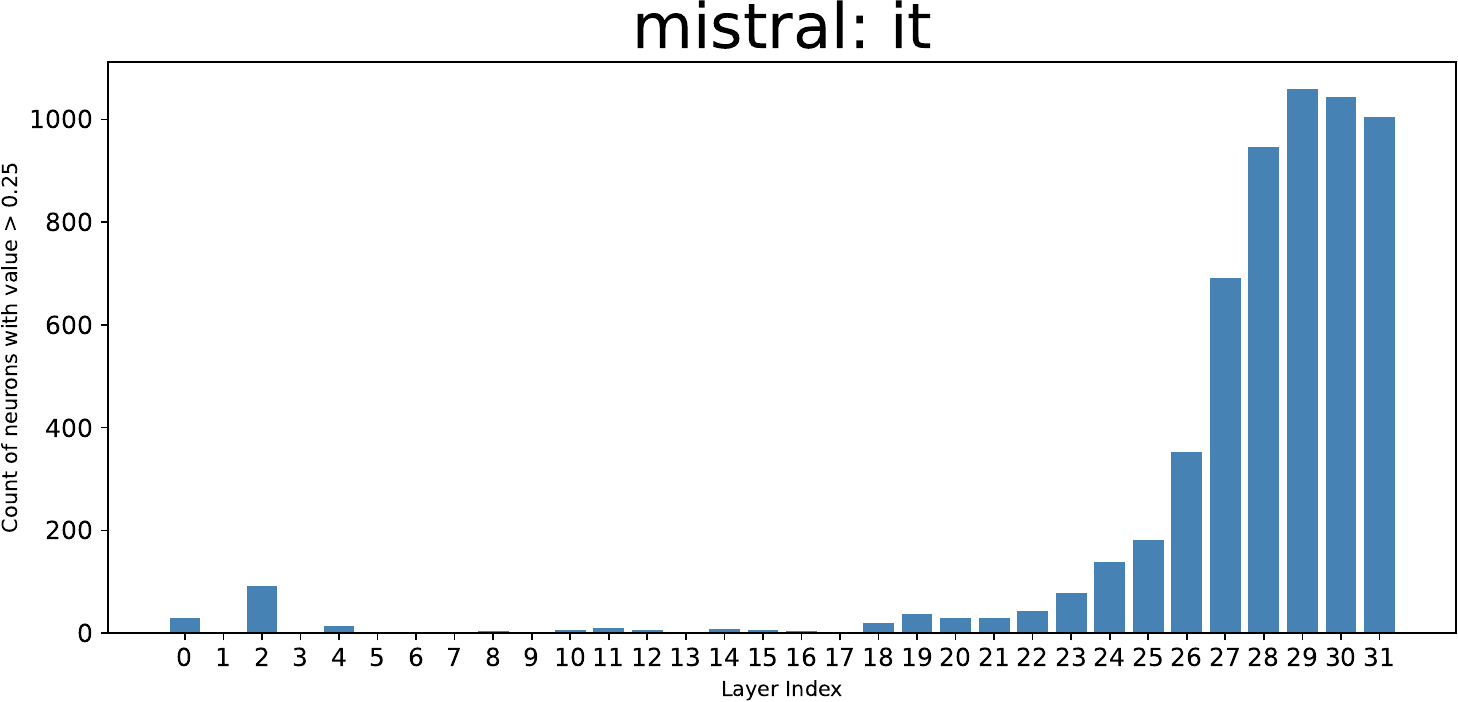}
  \includegraphics[width=0.19\linewidth]{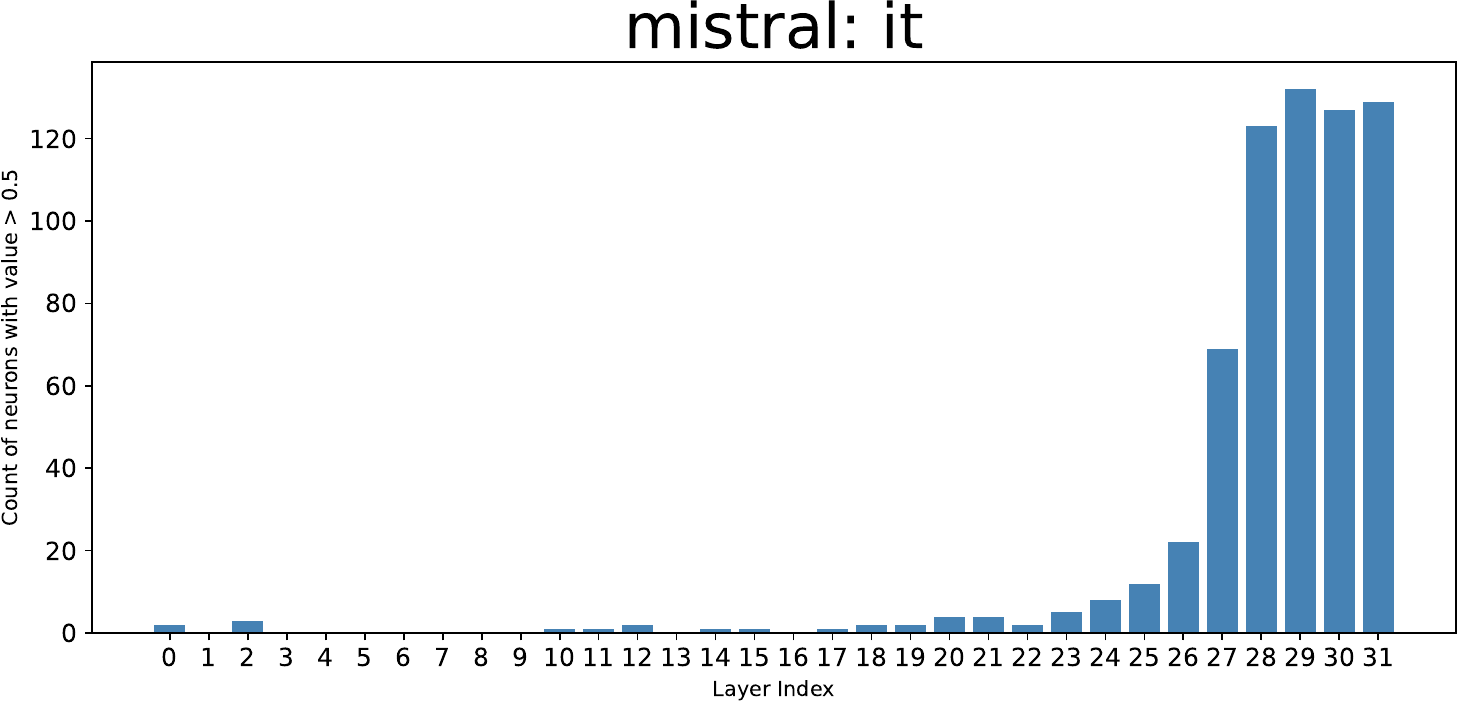}

  \begin{minipage}{0.19\linewidth}\centering it, $\geq0.25$\end{minipage}
  \begin{minipage}{0.19\linewidth}\centering it, $\geq0.5$\end{minipage}

\caption{\textbf{Distribution of language-specific neurons (Mistral-7B).}}
  \label{fig:appendix:distribution_lang_specific_neurons_mistral}
\end{figure*}

\begin{figure*}[t]
  \centering

  \includegraphics[width=0.19\linewidth]{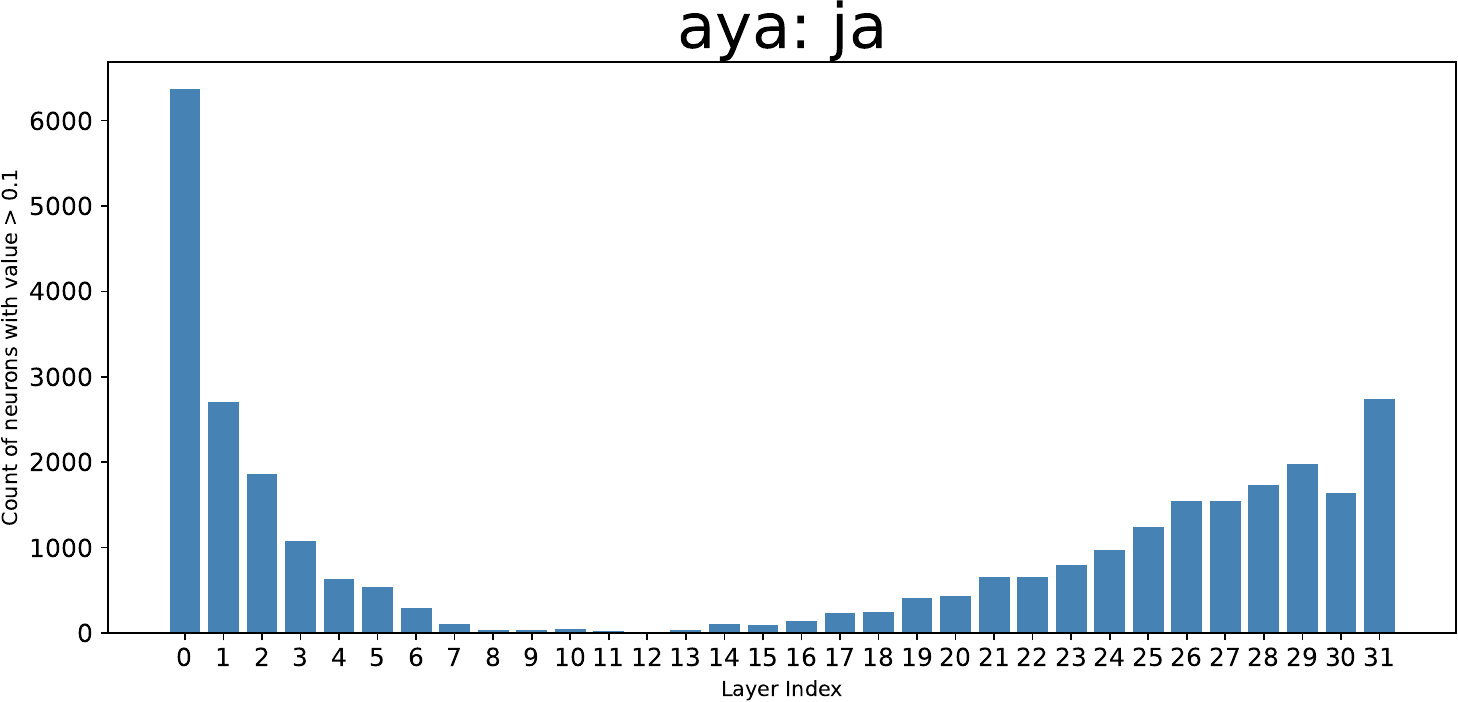}
  \includegraphics[width=0.19\linewidth]{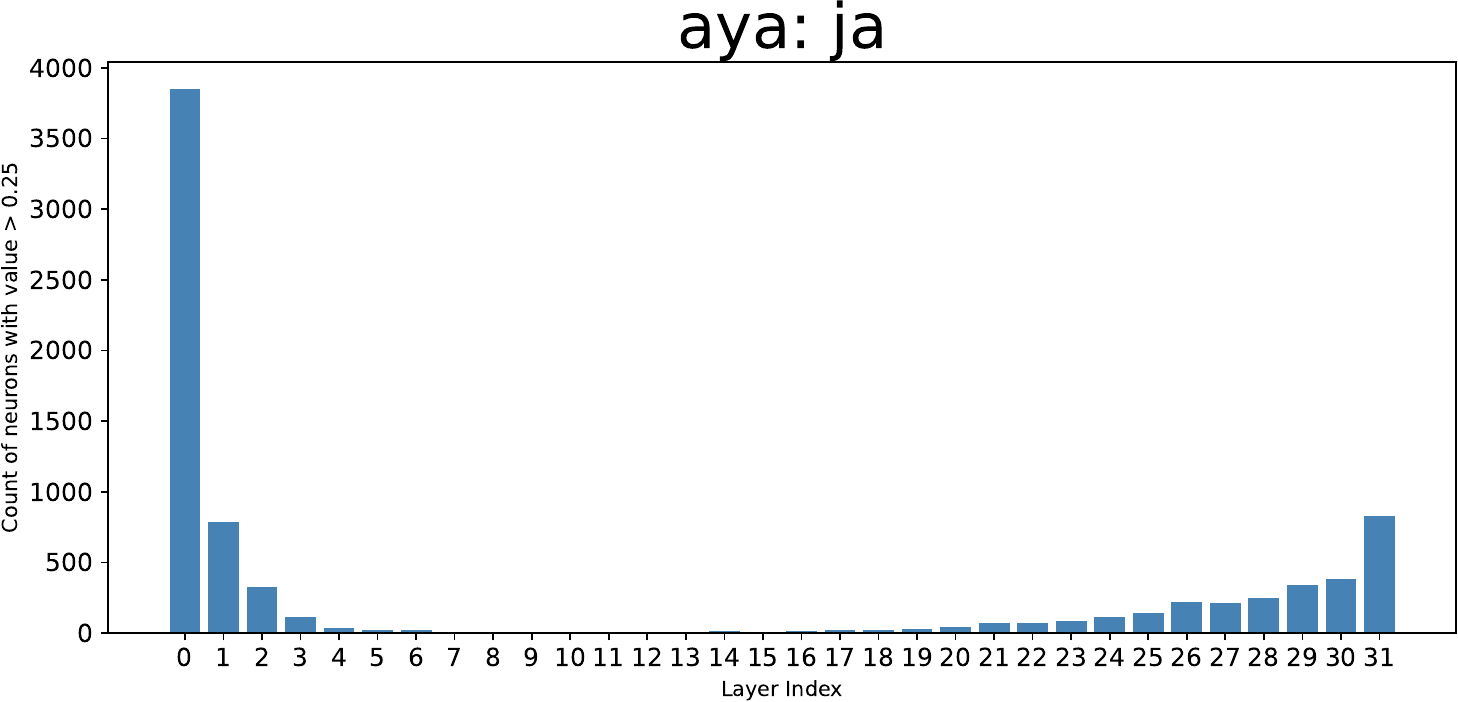}
  \includegraphics[width=0.19\linewidth]{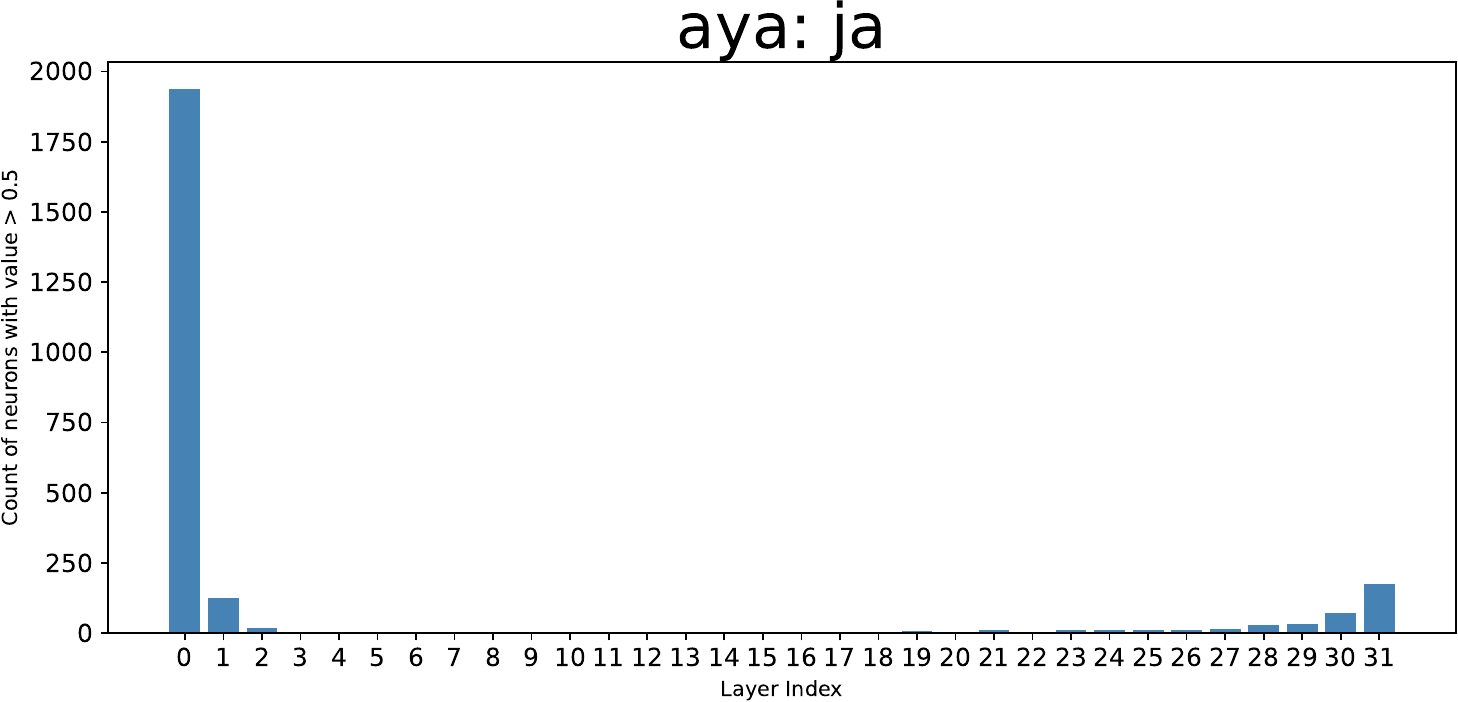}
  \includegraphics[width=0.19\linewidth]{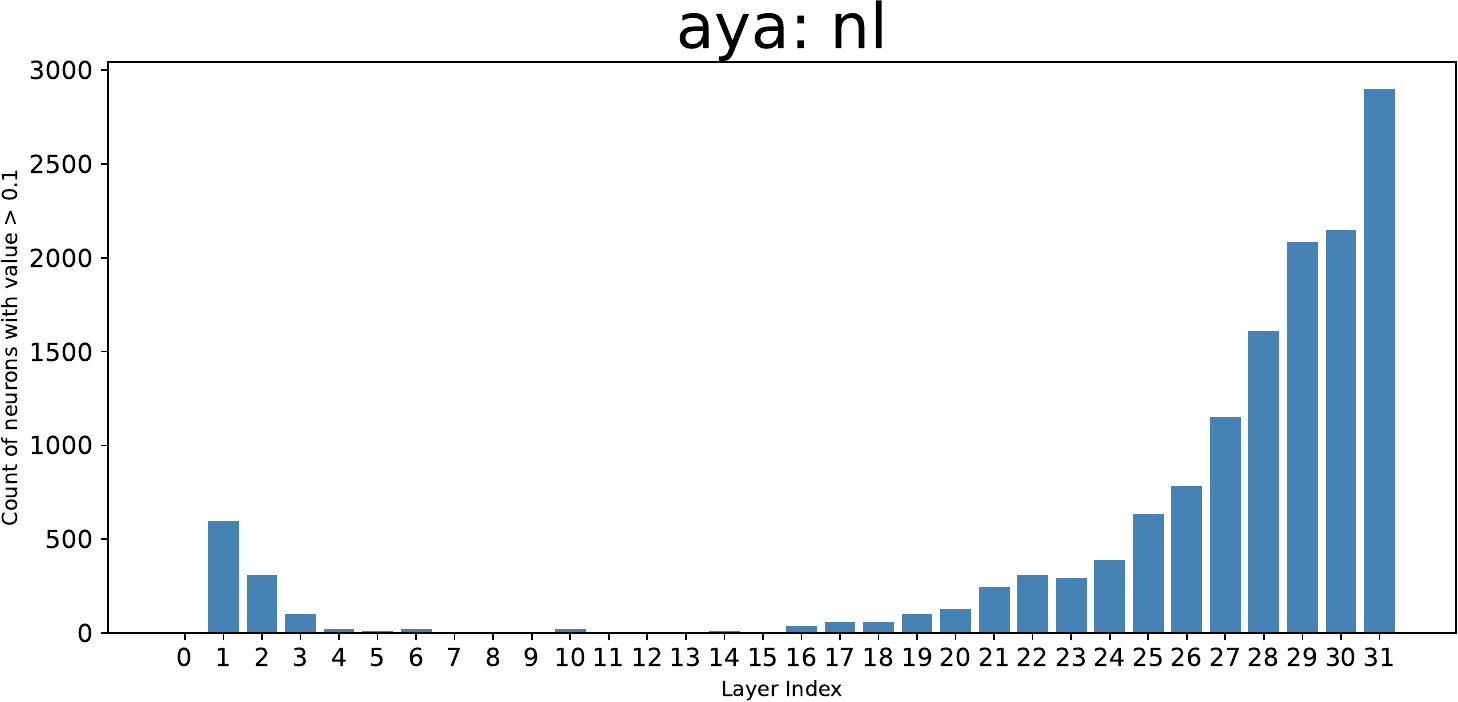}
  \includegraphics[width=0.19\linewidth]{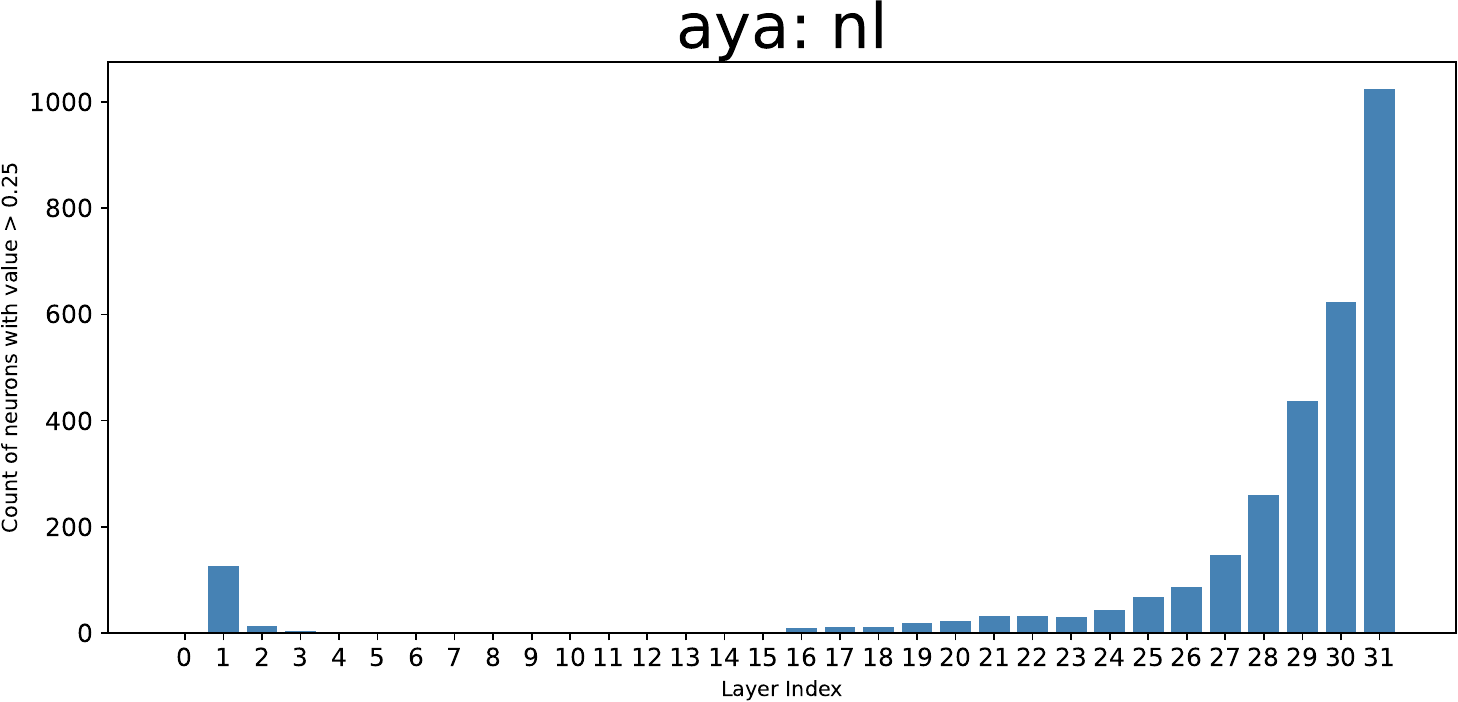}

  \begin{minipage}{0.19\linewidth}\centering ja, $\geq0.1$\end{minipage}
  \begin{minipage}{0.19\linewidth}\centering ja, $\geq0.25$\end{minipage}
  \begin{minipage}{0.19\linewidth}\centering ja, $\geq0.5$\end{minipage}
  \begin{minipage}{0.19\linewidth}\centering nl, $\geq0.1$\end{minipage}
  \begin{minipage}{0.19\linewidth}\centering nl, $\geq0.25$\end{minipage}

  \includegraphics[width=0.19\linewidth]{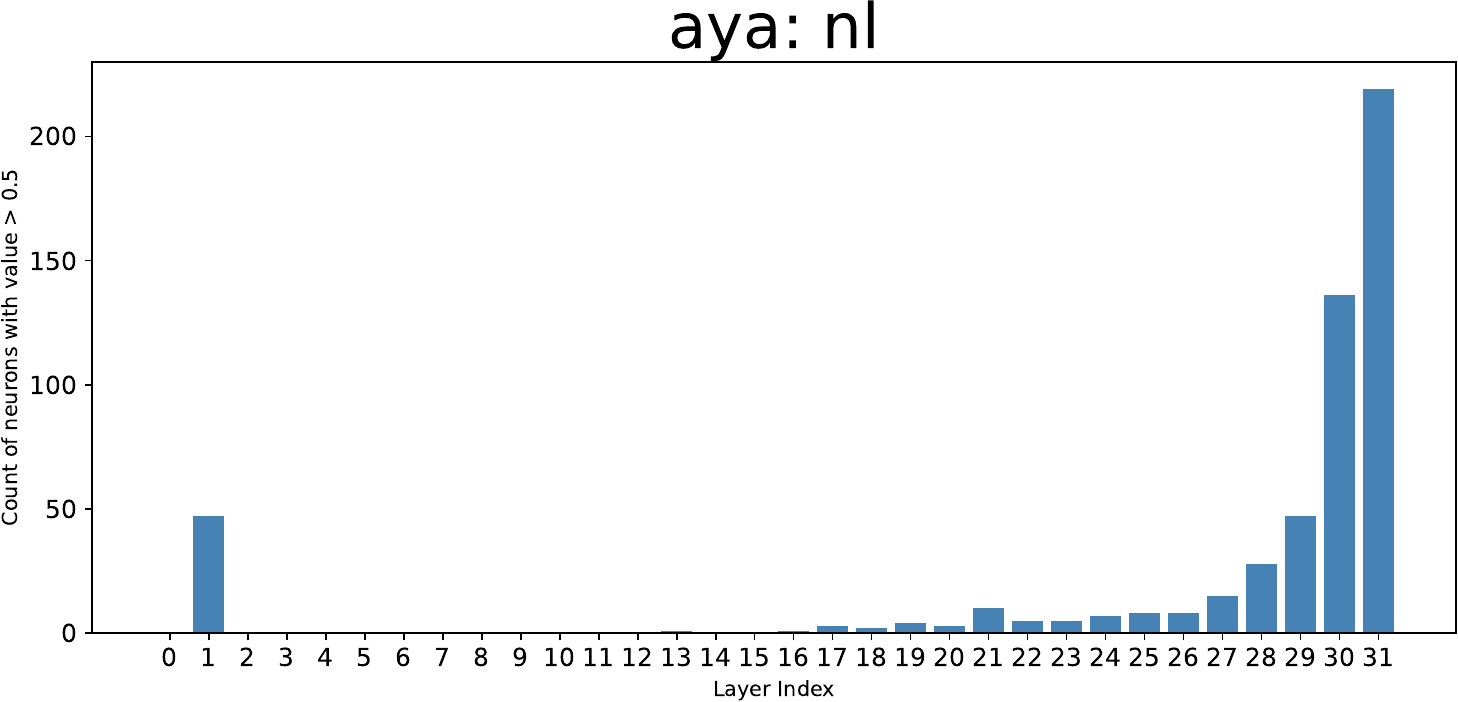}
  \includegraphics[width=0.19\linewidth]{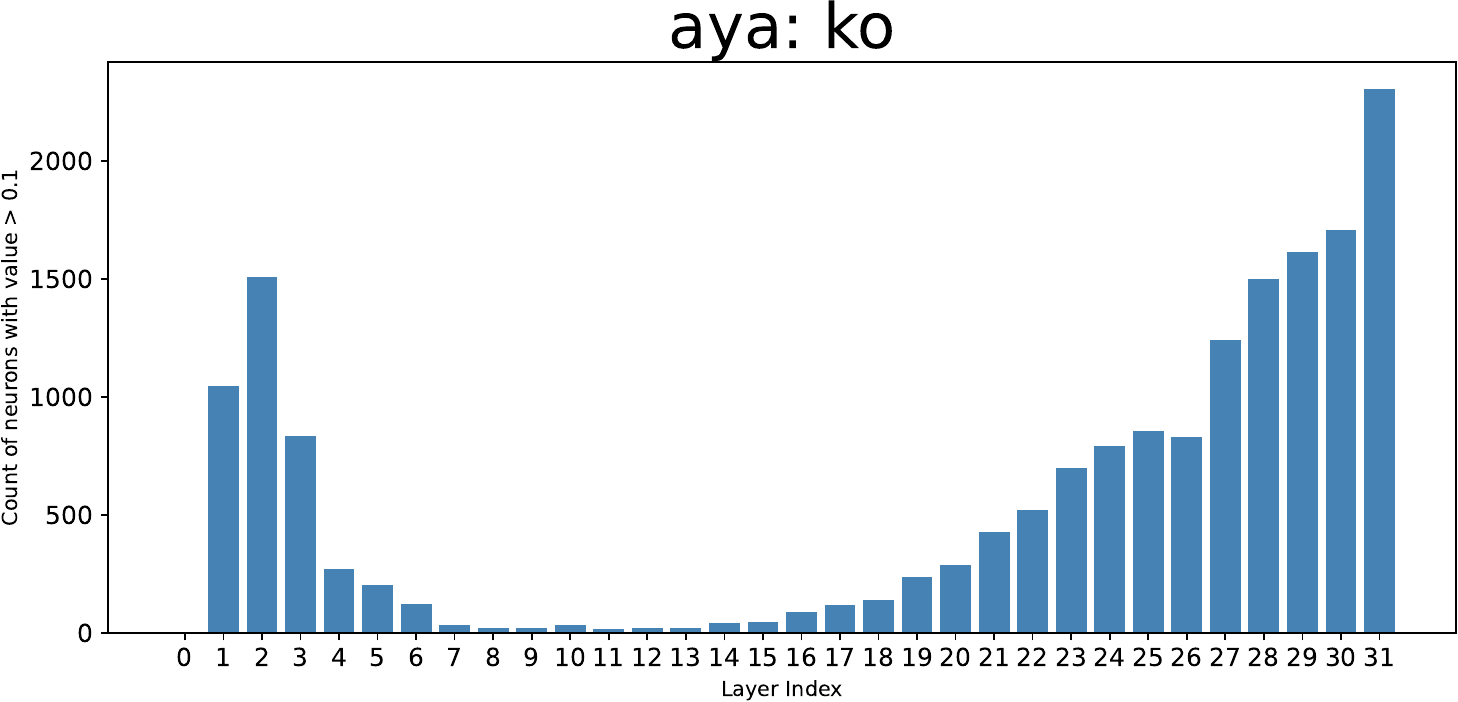}
  \includegraphics[width=0.19\linewidth]{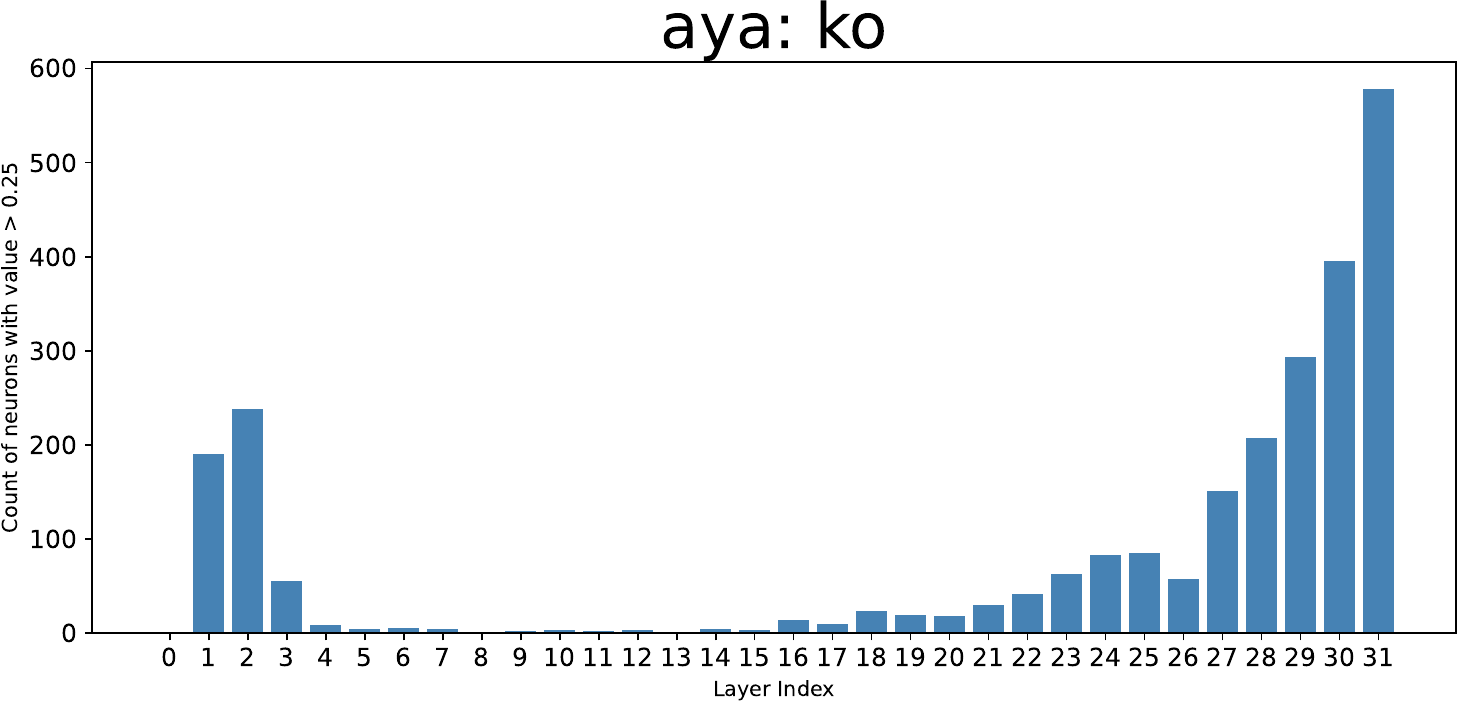}
  \includegraphics[width=0.19\linewidth]{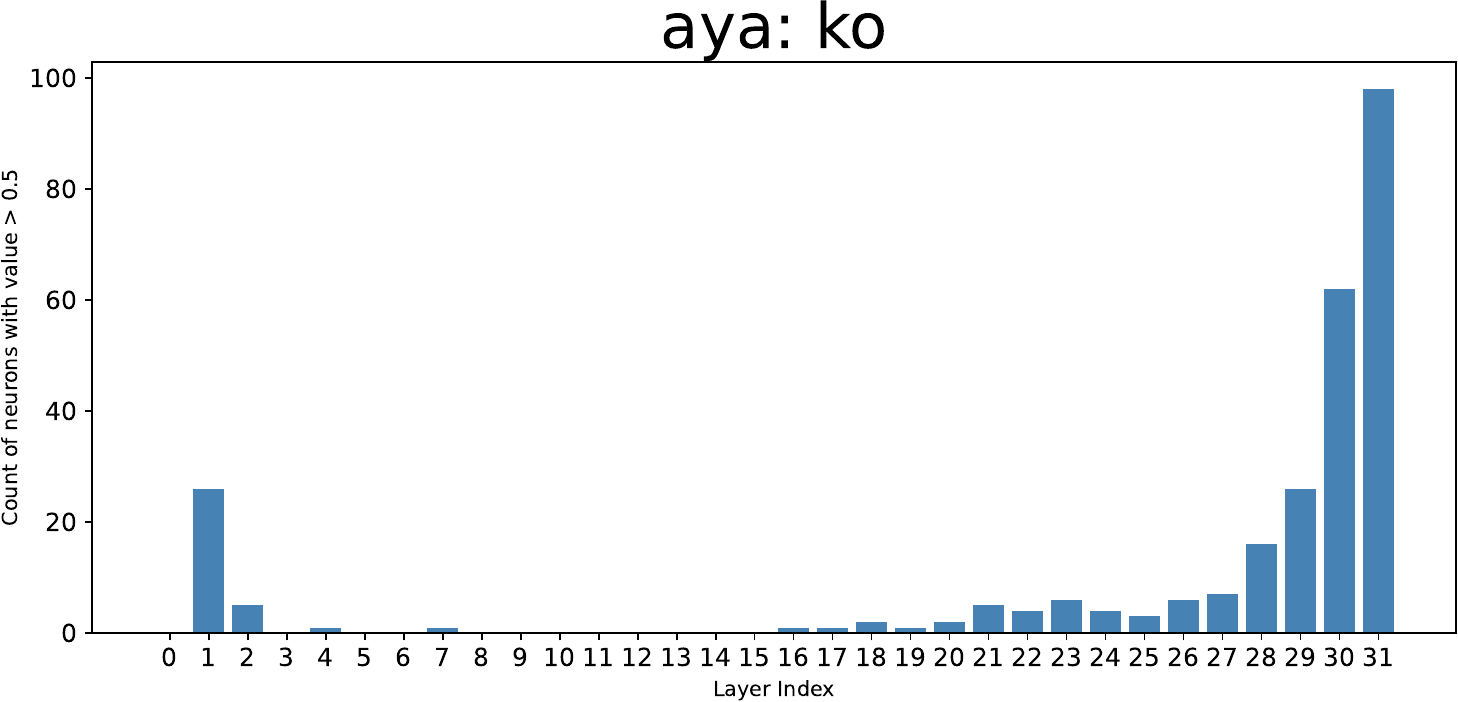}
  \includegraphics[width=0.19\linewidth]{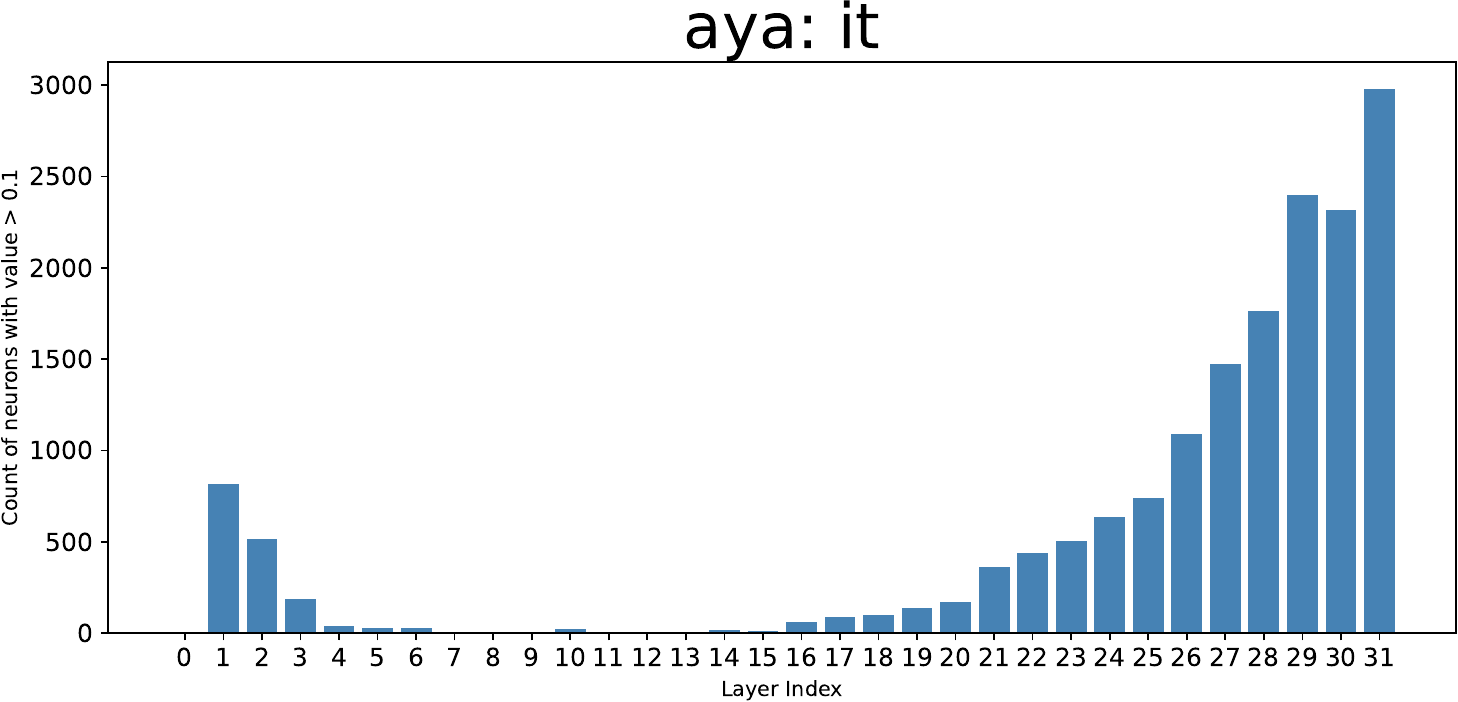}

  \begin{minipage}{0.19\linewidth}\centering nl, $\geq0.5$\end{minipage}
  \begin{minipage}{0.19\linewidth}\centering ko, $\geq0.1$\end{minipage}
  \begin{minipage}{0.19\linewidth}\centering ko, $\geq0.25$\end{minipage}
  \begin{minipage}{0.19\linewidth}\centering ko, $\geq0.5$\end{minipage}
  \begin{minipage}{0.19\linewidth}\centering it, $\geq0.1$\end{minipage}

  \includegraphics[width=0.19\linewidth]{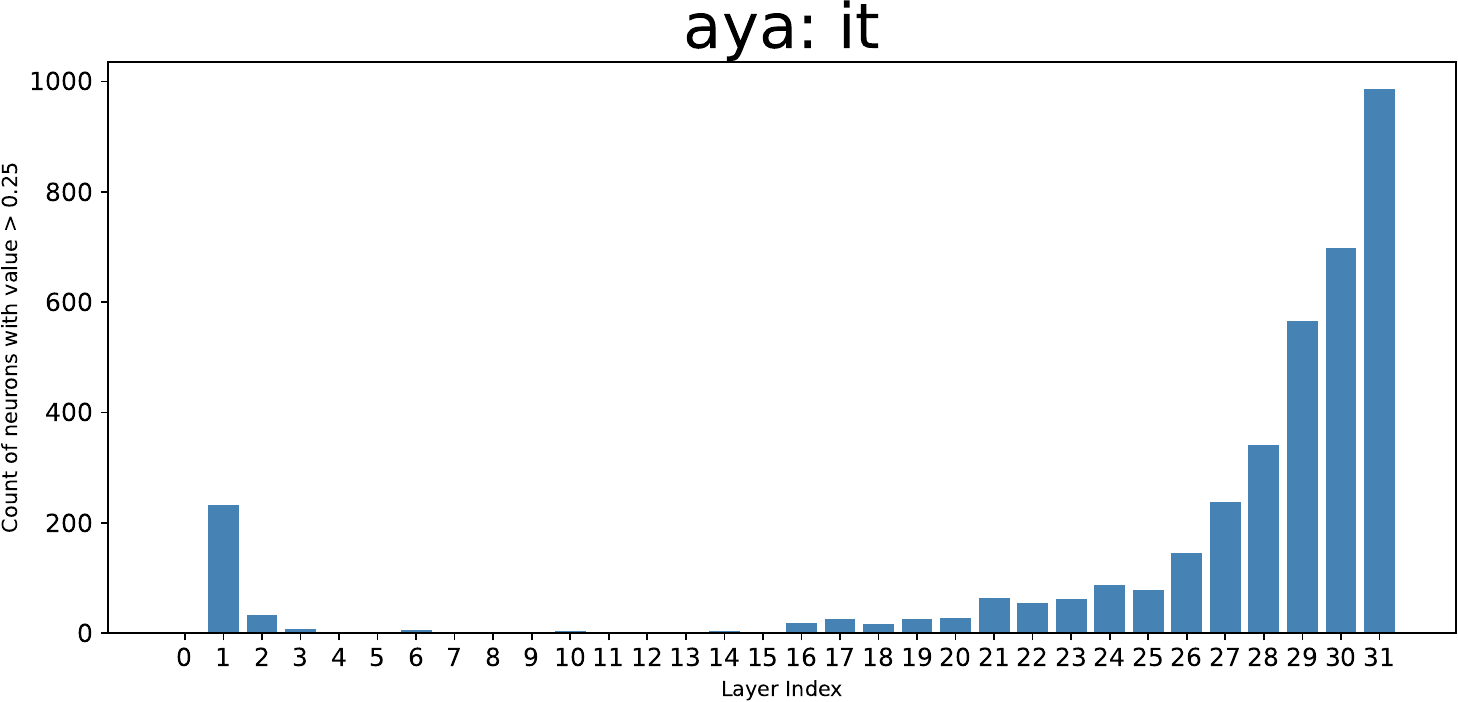}
  \includegraphics[width=0.19\linewidth]{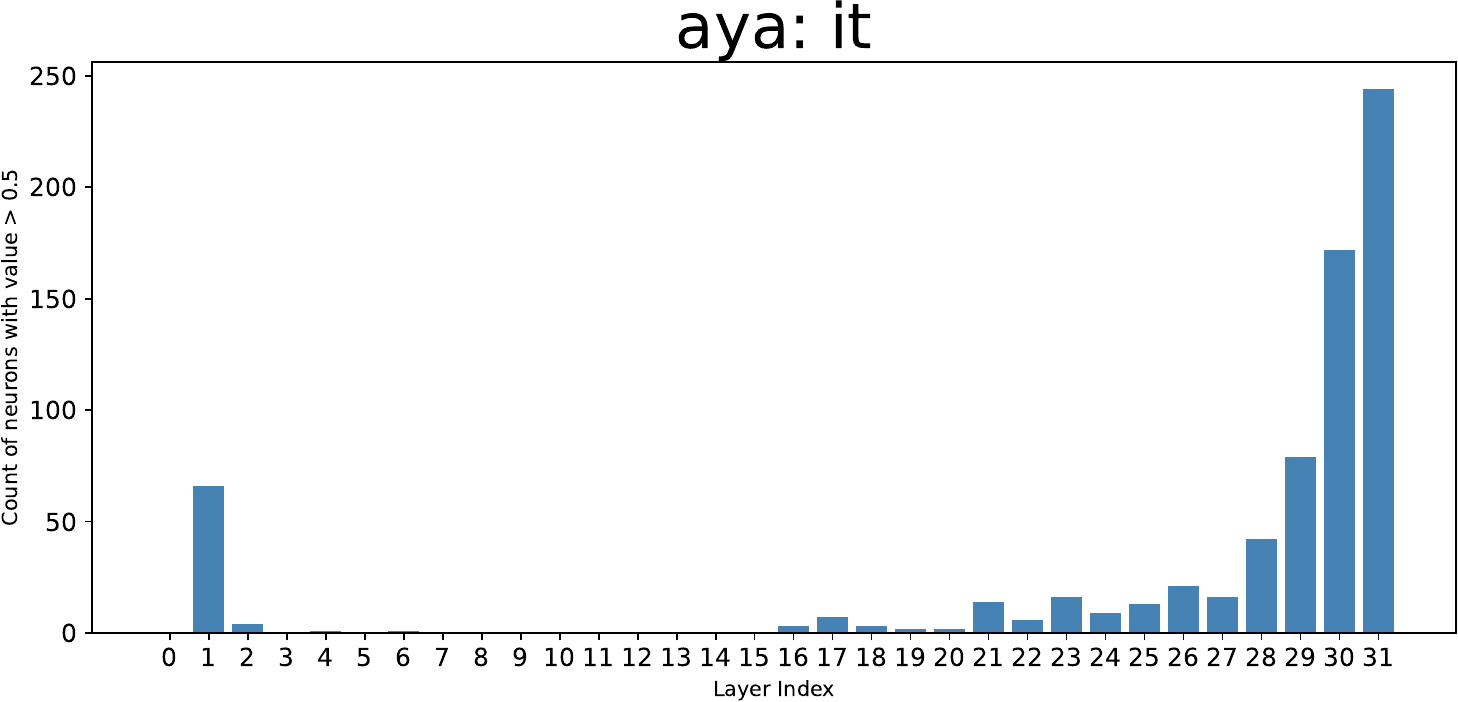}

  \begin{minipage}{0.19\linewidth}\centering it, $\geq0.25$\end{minipage}
  \begin{minipage}{0.19\linewidth}\centering it, $\geq0.5$\end{minipage}

\caption{\textbf{Distribution of language-specific neurons (Aya expanse-8B).}}
  \label{fig:appendix:distribution_lang_specific_neurons_aya}
\end{figure*}

Figs.~\ref{fig:appendix:distribution_lang_specific_neurons_llama},~\ref{fig:appendix:distribution_lang_specific_neurons_mistral}, and~\ref{fig:appendix:distribution_lang_specific_neurons_aya} show the distribution of detected language-specific neurons. As shown, the stronger the correlation between a neuron’s activations and the target language labels, the more it tends to be located in the initial and final layers. This tendency is well aligned with the findings of previous studies. Also, as we described in \S\ref{sec:distribution of transfer neurons}, their distribution is similar to those of transfer neurons.

% figure: distribution of lang-specific neurons, llama3.
% figure: distribution of lang-specific neurons, mistral.
% figure: distribution of lang-specific neurons, aya.

\subsection{Language Specificity}
\label{sec:appendix:language specificity}
Tabs.~\ref{table:appendix:corr-ratio_llama},~\ref{table:appendix:corr-ratio_mistral}, and~\ref{table:appendix:corr-ratio_aya} present the correlation ratios of transfer neurons with respect to language specificity across all models. The computation settings follow those described in \S\ref{sec:language-specific nature of transfer neurons}.
% corr ratio, llama3
\begin{table*}[t]
    \small
    \centering
    \setlength{\tabcolsep}{4pt}
    \renewcommand{\arraystretch}{1.2}
    \setContinuousHeatMapMax{0.3}
    \setContinuousHeatMapMin{0.0}
    
    \begin{tabularx}{\linewidth}{l c c c c  l c c c c}
        \hline
        {\textbf{Top-1000}} & \textbf{ja} & \textbf{nl} & \textbf{ko} & \textbf{it} &
        {\textbf{Top-100}} & \textbf{ja} & \textbf{nl} & \textbf{ko} & \textbf{it} \\
        \hline
        \textbf{Type-1 TN} & 
        \begin{tabular}{V} 0.16 \end{tabular} & 
        \begin{tabular}{V} 0.03 \end{tabular} & 
        \begin{tabular}{V} 0.05 \end{tabular} & 
        \begin{tabular}{V} 0.03 \end{tabular} & 
        \textbf{Type-1 TN} & 
        \begin{tabular}{V} 0.16 \end{tabular} & 
        \begin{tabular}{V} 0.02 \end{tabular} & 
        \begin{tabular}{V} 0.05 \end{tabular} & 
        \begin{tabular}{V} 0.03 \end{tabular} \\
        \textbf{Type-2 TN} & 
        \begin{tabular}{V} 0.25 \end{tabular} & 
        \begin{tabular}{V} 0.22 \end{tabular} & 
        \begin{tabular}{V} 0.17 \end{tabular} & 
        \begin{tabular}{V} 0.16 \end{tabular} & 
        \textbf{Type-2 TN} & 
        \begin{tabular}{V} 0.40 \end{tabular} & 
        \begin{tabular}{V} 0.27 \end{tabular} & 
        \begin{tabular}{V} 0.33 \end{tabular} & 
        \begin{tabular}{V} 0.19 \end{tabular} \\
        \hline
        {\textbf{Top-10}} & \textbf{ja} & \textbf{nl} & \textbf{ko} & \textbf{it} & & & & & \\
        \cline{1-5}
        \textbf{Type-1 TN} & 
        \begin{tabular}{V} 0.38 \end{tabular} & 
        \begin{tabular}{V} 0.01 \end{tabular} & 
        \begin{tabular}{V} 0.01 \end{tabular} & 
        \begin{tabular}{V} 0.03 \end{tabular} & 
        & & & & \\
        \textbf{Type-2 TN} & 
        \begin{tabular}{V} 0.54 \end{tabular} & 
        \begin{tabular}{V} 0.32 \end{tabular} & 
        \begin{tabular}{V} 0.76 \end{tabular} & 
        \begin{tabular}{V} 0.30 \end{tabular} & 
        & & & & \\
        \hline
    \end{tabularx}

    \caption{\textbf{Correlation ratio of top-10, top-100, and top-1k Transfer Neurons for language specificity (LLaMA3-8B).}
    Typically, a correlation ratio above \textbf{0.1} indicates a correlation, above \textbf{0.25} indicates a moderately strong correlation, and above \textbf{0.5} indicates a strong correlation.}
    \label{table:appendix:corr-ratio_llama}
\end{table*}
% corr ratio, mistral
\begin{table*}[t]
    \small
    \centering
    \setlength{\tabcolsep}{4pt}
    \renewcommand{\arraystretch}{1.2}
    \setContinuousHeatMapMax{0.3}
    \setContinuousHeatMapMin{0.0}
    
    \begin{tabularx}{\linewidth}{l c c c c  l c c c c}
        \hline
        {\textbf{Top-1000}} & \textbf{ja} & \textbf{nl} & \textbf{ko} & \textbf{it} &
        {\textbf{Top-100}} & \textbf{ja} & \textbf{nl} & \textbf{ko} & \textbf{it} \\
        \hline
        \textbf{Type-1 TN} & 
        \begin{tabular}{V} 0.17 \end{tabular} & 
        \begin{tabular}{V} 0.03 \end{tabular} & 
        \begin{tabular}{V} 0.06 \end{tabular} & 
        \begin{tabular}{V} 0.03 \end{tabular} & 
        \textbf{Type-1 TN} & 
        \begin{tabular}{V} 0.18 \end{tabular} & 
        \begin{tabular}{V} 0.03 \end{tabular} & 
        \begin{tabular}{V} 0.08 \end{tabular} & 
        \begin{tabular}{V} 0.03 \end{tabular} \\
        \textbf{Type-2 TN} & 
        \begin{tabular}{V} 0.28 \end{tabular} & 
        \begin{tabular}{V} 0.22 \end{tabular} & 
        \begin{tabular}{V} 0.11 \end{tabular} & 
        \begin{tabular}{V} 0.18 \end{tabular} & 
        \textbf{Type-2 TN} & 
        \begin{tabular}{V} 0.50 \end{tabular} & 
        \begin{tabular}{V} 0.27 \end{tabular} & 
        \begin{tabular}{V} 0.21 \end{tabular} & 
        \begin{tabular}{V} 0.24 \end{tabular} \\
        \hline
        {\textbf{Top-10}} & \textbf{ja} & \textbf{nl} & \textbf{ko} & \textbf{it} & & & & & \\
        \cline{1-5}
        \textbf{Type-1 TN} & 
        \begin{tabular}{V} 0.37 \end{tabular} & 
        \begin{tabular}{V} 0.02 \end{tabular} & 
        \begin{tabular}{V} 0.07 \end{tabular} & 
        \begin{tabular}{V} 0.03 \end{tabular} & 
        & & & & \\
        \textbf{Type-2 TN} & 
        \begin{tabular}{V} 0.80 \end{tabular} & 
        \begin{tabular}{V} 0.43 \end{tabular} & 
        \begin{tabular}{V} 0.52 \end{tabular} & 
        \begin{tabular}{V} 0.41 \end{tabular} & 
        & & & & \\
        \hline
    \end{tabularx}

    \caption{\textbf{Correlation ratio of top-10, top-100, and top-1k Transfer Neurons for language specificity (Mistral-7B).}}
    \label{table:appendix:corr-ratio_mistral}
\end{table*}
% corr ratio, aya
\begin{table*}[t]
    \small
    \centering
    \setlength{\tabcolsep}{4pt}
    \renewcommand{\arraystretch}{1.2}
    \setContinuousHeatMapMax{0.3}
    \setContinuousHeatMapMin{0.0}
    
    \begin{tabularx}{\linewidth}{l c c c c  l c c c c}
        \hline
        {\textbf{Top-1000}} & \textbf{ja} & \textbf{nl} & \textbf{ko} & \textbf{it} &
        {\textbf{Top-100}} & \textbf{ja} & \textbf{nl} & \textbf{ko} & \textbf{it} \\
        \hline
        \textbf{Type-1 TN} & 
        \begin{tabular}{V} 0.19 \end{tabular} & 
        \begin{tabular}{V} 0.02 \end{tabular} & 
        \begin{tabular}{V} 0.05 \end{tabular} & 
        \begin{tabular}{V} 0.03 \end{tabular} & 
        \textbf{Type-1 TN} & 
        \begin{tabular}{V} 0.32 \end{tabular} & 
        \begin{tabular}{V} 0.02 \end{tabular} & 
        \begin{tabular}{V} 0.03 \end{tabular} & 
        \begin{tabular}{V} 0.02 \end{tabular} \\
        \textbf{Type-2 TN} & 
        \begin{tabular}{V} 0.24 \end{tabular} & 
        \begin{tabular}{V} 0.22 \end{tabular} & 
        \begin{tabular}{V} 0.17 \end{tabular} & 
        \begin{tabular}{V} 0.24 \end{tabular} & 
        \textbf{Type-2 TN} & 
        \begin{tabular}{V} 0.43 \end{tabular} & 
        \begin{tabular}{V} 0.37 \end{tabular} & 
        \begin{tabular}{V} 0.40 \end{tabular} & 
        \begin{tabular}{V} 0.36 \end{tabular} \\
        \hline
        {\textbf{Top-10}} & \textbf{ja} & \textbf{nl} & \textbf{ko} & \textbf{it} & & & & & \\
        \cline{1-5}
        \textbf{Type-1 TN} & 
        \begin{tabular}{V} 0.27 \end{tabular} & 
        \begin{tabular}{V} 0.03 \end{tabular} & 
        \begin{tabular}{V} 0.03 \end{tabular} & 
        \begin{tabular}{V} 0.02 \end{tabular} & 
        & & & & \\
        \textbf{Type-2 TN} & 
        \begin{tabular}{V} 0.80 \end{tabular} & 
        \begin{tabular}{V} 0.50 \end{tabular} & 
        \begin{tabular}{V} 0.70 \end{tabular} & 
        \begin{tabular}{V} 0.33 \end{tabular} & 
        & & & & \\
        \hline
    \end{tabularx}

    \caption{\textbf{Correlation ratio of top-10, top-100, and top-1k Transfer Neurons for language specificity (Aya expanse-8B).}}
    \label{table:appendix:corr-ratio_aya}
\end{table*}

\subsection{Language-Family Specificity}
\subsubsection{Jaccard Index across Language Pairs}
\label{sec:appendix:jaccard index}
Tabs.~\ref{table:appendix:jaccard_index_llama3},~\ref{table:appendix:jaccard_index_mistral}, and~\ref{table:appendix:jaccard_index_aya} show the Jaccard Index of transfer neurons for each language pair, representing the degree of overlap in neurons between pairs.
%--- Jaccard Index for Language-Family Specificity ---%
% tab: jaccard index for language-family specificity, llama3
\begin{table*}[t]
    \small
    \centering
    \renewcommand{\arraystretch}{1.2}
    
    \begin{tabular}{l c c c c c c c}
        \hline
          & \textbf{ALL} & \textbf{ja-nl} & \textbf{ja-ko} & \textbf{nl-it} & \textbf{ja-it} & \textbf{nl-ko} & \textbf{it-ko}\\
        \hline
        \textbf{Type-1 TN} & 0.30 & 0.39 & 0.51 & 0.75 & 0.41 & 0.50 & 0.51\\
        \textbf{Type-2 TN} & 0.12 & 0.23 & 0.37 & 0.51 & 0.26 & 0.28 & 0.30\\
        \hline
    \end{tabular}

    \caption{\textbf{Jaccard index for top-1k Transfer Neurons across language pairs. (LLaMA3-8B).}}
    \label{table:appendix:jaccard_index_llama3}
\end{table*}
% tab: jaccard index for language-family specificity, mistral
\begin{table*}[t]
    \small
    \centering
    \renewcommand{\arraystretch}{1.2}
    
    \begin{tabular}{l c c c c c c c}
        \hline
          & \textbf{ALL} & \textbf{ja-nl} & \textbf{ja-ko} & \textbf{nl-it} & \textbf{ja-it} & \textbf{nl-ko} & \textbf{it-ko}\\
        \hline
        \textbf{Type-1 TN} & 0.30 & 0.42 & 0.48 & 0.75 & 0.43 & 0.54 & 0.52\\
        \textbf{Type-2 TN} & 0.09 & 0.18 & 0.31 & 0.42 & 0.19 & 0.26 & 0.29\\
        \hline
    \end{tabular}

    \caption{\textbf{Jaccard index for top-1k Transfer Neurons across language pairs. (Mistral-7B).}}
    \label{table:appendix:jaccard_index_mistral}
\end{table*}
% tab: jaccard index for language-family specificity, aya
\begin{table*}[t]
    \small
    \centering
    \renewcommand{\arraystretch}{1.2}
    
    \begin{tabular}{l c c c c c c c}
        \hline
          & \textbf{ALL} & \textbf{ja-nl} & \textbf{ja-ko} & \textbf{nl-it} & \textbf{ja-it} & \textbf{nl-ko} & \textbf{it-ko}\\
        \hline
        \textbf{Type-1 TN} & 0.30 & 0.42 & 0.46 & 0.75 & 0.40 & 0.55 & 0.53\\
        \textbf{Type-2 TN} & 0.08 & 0.18 & 0.31 & 0.35 & 0.18 & 0.22 & 0.21\\
        \hline
    \end{tabular}

    \caption{\textbf{Jaccard index for top-1k Transfer Neurons across language pairs. (Aya expanse-8B).}}
    \label{table:appendix:jaccard_index_aya}
\end{table*}

\subsubsection{Correlation Ratio for Language-Family Specificity}
\label{sec:appendix:language-family specificity}
To investigate language-family specificity from another perspective of view, we computed the correlation ratio by assigning \texttt{label1} to sentence pairs that largely belong to the same language family, and \texttt{label0} to all others, following the settings described in Appendix~\ref{sec:appendix:language specific neurons}. In this experiment, we consider English, Dutch, and Italian as belonging to the same language-family, whereas Japanese and Korean are grouped into another language-family.
Tabs.~\ref{table:appendix:corr-ratio_for_language_family_specificity_llama},~\ref{table:appendix:corr-ratio_for_language_family_specificity_mistral}, and~\ref{table:appendix:corr-ratio_for_language_family_specificity_aya} present the results. Language-family specificity was particularly evident in Type-2 neurons.

%--- Corr Ratio for Language-Family Specificity ---%
% tab: corr ratio for language-family specificity, llama3
\begin{table*}[t]
    \small
    \centering
    \setlength{\tabcolsep}{4pt}
    \renewcommand{\arraystretch}{1.2}
    \setContinuousHeatMapMax{0.3}
    \setContinuousHeatMapMin{0.0}
    
    \begin{tabularx}{\linewidth}{l c c c c  l c c c c}
        \hline
        {\textbf{Top-1000}} & \textbf{ja} & \textbf{nl} & \textbf{ko} & \textbf{it} &
        {\textbf{Top-100}} & \textbf{ja} & \textbf{nl} & \textbf{ko} & \textbf{it} \\
        \hline
        \textbf{Type-1 TN} & 
        \begin{tabular}{V} 0.14 \end{tabular} & 
        \begin{tabular}{V} 0.08 \end{tabular} & 
        \begin{tabular}{V} 0.11 \end{tabular} & 
        \begin{tabular}{V} 0.08 \end{tabular} & 
        \textbf{Type-1 TN} & 
        \begin{tabular}{V} 0.15 \end{tabular} & 
        \begin{tabular}{V} 0.11 \end{tabular} & 
        \begin{tabular}{V} 0.12 \end{tabular} & 
        \begin{tabular}{V} 0.11 \end{tabular} \\
        \textbf{Type-2 TN} & 
        \begin{tabular}{V} 0.25 \end{tabular} & 
        \begin{tabular}{V} 0.23 \end{tabular} & 
        \begin{tabular}{V} 0.22 \end{tabular} & 
        \begin{tabular}{V} 0.23 \end{tabular} & 
        \textbf{Type-2 TN} & 
        \begin{tabular}{V} 0.31 \end{tabular} & 
        \begin{tabular}{V} 0.25 \end{tabular} & 
        \begin{tabular}{V} 0.27 \end{tabular} & 
        \begin{tabular}{V} 0.25 \end{tabular} \\
        \hline
        {\textbf{Top-10}} & \textbf{ja} & \textbf{nl} & \textbf{ko} & \textbf{it} & & & & & \\
        \cline{1-5}
        \textbf{Type-1 TN} & 
        \begin{tabular}{V} 0.25 \end{tabular} & 
        \begin{tabular}{V} 0.10 \end{tabular} & 
        \begin{tabular}{V} 0.04 \end{tabular} & 
        \begin{tabular}{V} 0.13 \end{tabular} & 
        & & & & \\
        \textbf{Type-2 TN} & 
        \begin{tabular}{V} 0.40 \end{tabular} & 
        \begin{tabular}{V} 0.45 \end{tabular} & 
        \begin{tabular}{V} 0.38 \end{tabular} & 
        \begin{tabular}{V} 0.43 \end{tabular} & 
        & & & & \\
        \hline
    \end{tabularx}

    \caption{\textbf{Correlation ratio of top-10, top-100, and top-1k Transfer Neurons for language-family specificity (LLaMA3-8B).}
    Typically, a correlation ratio above \textbf{0.1} indicates a correlation, above \textbf{0.25} indicates a moderately strong correlation, and above \textbf{0.5} indicates a strong correlation.}
    \label{table:appendix:corr-ratio_for_language_family_specificity_llama}
\end{table*}
% tab: corr ratio for language-family specificity, mistral
\begin{table*}[t]
    \small
    \centering
    \setlength{\tabcolsep}{4pt}
    \renewcommand{\arraystretch}{1.2}
    \setContinuousHeatMapMax{0.3}
    \setContinuousHeatMapMin{0.0}
    
    \begin{tabularx}{\linewidth}{l c c c c  l c c c c}
        \hline
        {\textbf{Top-1000}} & \textbf{ja} & \textbf{nl} & \textbf{ko} & \textbf{it} &
        {\textbf{Top-100}} & \textbf{ja} & \textbf{nl} & \textbf{ko} & \textbf{it} \\
        \hline
        \textbf{Type-1 TN} & 
        \begin{tabular}{V} 0.13 \end{tabular} & 
        \begin{tabular}{V} 0.08 \end{tabular} & 
        \begin{tabular}{V} 0.10 \end{tabular} & 
        \begin{tabular}{V} 0.07 \end{tabular} & 
        \textbf{Type-1 TN} & 
        \begin{tabular}{V} 0.16 \end{tabular} & 
        \begin{tabular}{V} 0.10 \end{tabular} & 
        \begin{tabular}{V} 0.15 \end{tabular} & 
        \begin{tabular}{V} 0.10 \end{tabular} \\
        \textbf{Type-2 TN} & 
        \begin{tabular}{V} 0.21 \end{tabular} & 
        \begin{tabular}{V} 0.17 \end{tabular} & 
        \begin{tabular}{V} 0.17 \end{tabular} & 
        \begin{tabular}{V} 0.16 \end{tabular} & 
        \textbf{Type-2 TN} & 
        \begin{tabular}{V} 0.35 \end{tabular} & 
        \begin{tabular}{V} 0.23 \end{tabular} & 
        \begin{tabular}{V} 0.27 \end{tabular} & 
        \begin{tabular}{V} 0.22 \end{tabular} \\
        \hline
        {\textbf{Top-10}} & \textbf{ja} & \textbf{nl} & \textbf{ko} & \textbf{it} & & & & & \\
        \cline{1-5}
        \textbf{Type-1 TN} & 
        \begin{tabular}{V} 0.21 \end{tabular} & 
        \begin{tabular}{V} 0.16 \end{tabular} & 
        \begin{tabular}{V} 0.14 \end{tabular} & 
        \begin{tabular}{V} 0.14 \end{tabular} & 
        & & & & \\
        \textbf{Type-2 TN} & 
        \begin{tabular}{V} 0.52 \end{tabular} & 
        \begin{tabular}{V} 0.34 \end{tabular} & 
        \begin{tabular}{V} 0.42 \end{tabular} & 
        \begin{tabular}{V} 0.31 \end{tabular} & 
        & & & & \\
        \hline
    \end{tabularx}

    \caption{\textbf{Correlation ratio of top-10, top-100, and top-1k Transfer Neurons for language-family specificity (Mistral-7B).}}
    \label{table:appendix:corr-ratio_for_language_family_specificity_mistral}
\end{table*}
% tab: corr ratio for language-family specificity, aya
\begin{table*}[t]
    \small
    \centering
    \setlength{\tabcolsep}{4pt}
    \renewcommand{\arraystretch}{1.2}
    \setContinuousHeatMapMax{0.3}
    \setContinuousHeatMapMin{0.0}
    
    \begin{tabularx}{\linewidth}{l c c c c  l c c c c}
        \hline
        {\textbf{Top-1000}} & \textbf{ja} & \textbf{nl} & \textbf{ko} & \textbf{it} &
        {\textbf{Top-100}} & \textbf{ja} & \textbf{nl} & \textbf{ko} & \textbf{it} \\
        \hline
        \textbf{Type-1 TN} & 
        \begin{tabular}{V} 0.11 \end{tabular} & 
        \begin{tabular}{V} 0.07 \end{tabular} & 
        \begin{tabular}{V} 0.08 \end{tabular} & 
        \begin{tabular}{V} 0.07 \end{tabular} & 
        \textbf{Type-1 TN} & 
        \begin{tabular}{V} 0.14 \end{tabular} & 
        \begin{tabular}{V} 0.08 \end{tabular} & 
        \begin{tabular}{V} 0.08 \end{tabular} & 
        \begin{tabular}{V} 0.08 \end{tabular} \\
        \textbf{Type-2 TN} & 
        \begin{tabular}{V} 0.19 \end{tabular} & 
        \begin{tabular}{V} 0.15 \end{tabular} & 
        \begin{tabular}{V} 0.17 \end{tabular} & 
        \begin{tabular}{V} 0.15 \end{tabular} & 
        \textbf{Type-2 TN} & 
        \begin{tabular}{V} 0.28 \end{tabular} & 
        \begin{tabular}{V} 0.24 \end{tabular} & 
        \begin{tabular}{V} 0.26 \end{tabular} & 
        \begin{tabular}{V} 0.23 \end{tabular} \\
        \hline
        {\textbf{Top-10}} & \textbf{ja} & \textbf{nl} & \textbf{ko} & \textbf{it} & & & & & \\
        \cline{1-5}
        \textbf{Type-1 TN} & 
        \begin{tabular}{V} 0.10 \end{tabular} & 
        \begin{tabular}{V} 0.15 \end{tabular} & 
        \begin{tabular}{V} 0.15 \end{tabular} & 
        \begin{tabular}{V} 0.15 \end{tabular} & 
        & & & & \\
        \textbf{Type-2 TN} & 
        \begin{tabular}{V} 0.31 \end{tabular} & 
        \begin{tabular}{V} 0.23 \end{tabular} & 
        \begin{tabular}{V} 0.30 \end{tabular} & 
        \begin{tabular}{V} 0.28 \end{tabular} & 
        & & & & \\
        \hline
    \end{tabularx}

    \caption{\textbf{Correlation ratio of top-10, top-100, and top-1k Transfer Neurons for language-family specificity (Aya expanse-8B).}}
    \label{table:appendix:corr-ratio_for_language_family_specificity_aya}
\end{table*}

\subsection{Correlation Ratio}
\label{sec:appendix:corr_ratio}
Correlation ratio is a well established statistical measure of association between categorical and quantitative variables. It is computed as follows:
\begin{equation}
    \text{corr ratio}(\eta^2) = \frac{S_B}{S_W + S_B}
\end{equation}
where $S_B$ (between-group sum of squares) measures the variance between label means, and $S_W$ (within-group sum of squares) measures the variance within each label.

That is, in our case, neurons that tend to exhibit similar activation patterns for sentences within the same label (i.e., language or language-family), and distinct patterns for sentences with different labels, receive higher correlation ratio scores.

\subsection{Hypothesis Testing for Reported Correlation Values}
\label{sec:appendix:hypothesis testing for correlation metric}
To verify the reliability of the correlation ratio scores reported in the series of experiments in this study, we perform hypothesis testing using \textit{\textbf{An}alysis \textbf{o}f \textbf{Va}riance} (\textit{ANOVA}). ANOVA is a statistical method used to determine whether there are significant differences between the means of multiple groups. 

That is, in our context, we test whether there are significant differences in the activation values of each transfer neuron depending on the label (i.e., language or language family). Note that each transfer neuron has \textit{p value}.

We present proportions for both Type-1 and Type-2 neurons whose activations meet following conditions: (i) statistically significant ($p < 0.05$) and (ii) correlation ratio exceeds 0.1 (meaningful correlation) and 0.25 (moderately strong correlation). The results for language specific correlation are shown in Tabs.~\ref{tab:anova_type1_01},~\ref{tab:anova_type2_01},~\ref{tab:anova_type1_025}, and~\ref{tab:anova_type2_0.25}. As presented, the proportions are well aligned with the results of correlation analysis in this paper, which shows Type-1 neurons generally do not exhibit correlation (except for Japanese ones), whereas most Type-2 neurons exhibit correlation with input languages.

We also present the results for the same testing method for the correlation with language-family specificity in Tabs.~\ref{tab:anova_type1_01_language_family},~\ref{tab:anova_type2_01_language_family},~\ref{tab:anova_type1_025_language_family}, and~\ref{tab:anova_type2_025_language_family}.

\begin{figure*}[t]
  \centering
  \begin{subfigure}{0.49\linewidth}
    \includegraphics[width=\linewidth]{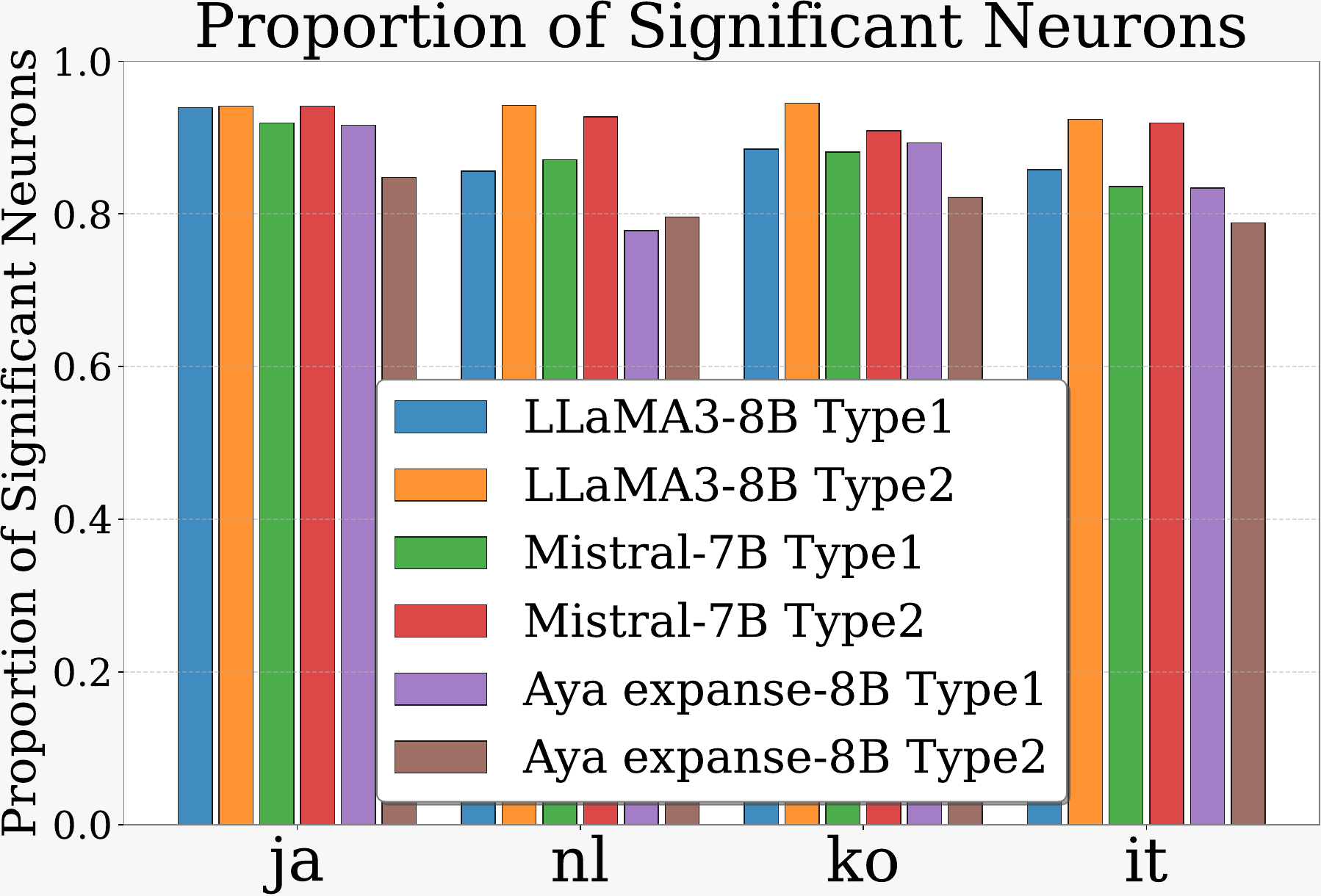}
    \caption{Language specificity}
  \end{subfigure}
  \begin{subfigure}{0.49\linewidth}
    \includegraphics[width=\linewidth]{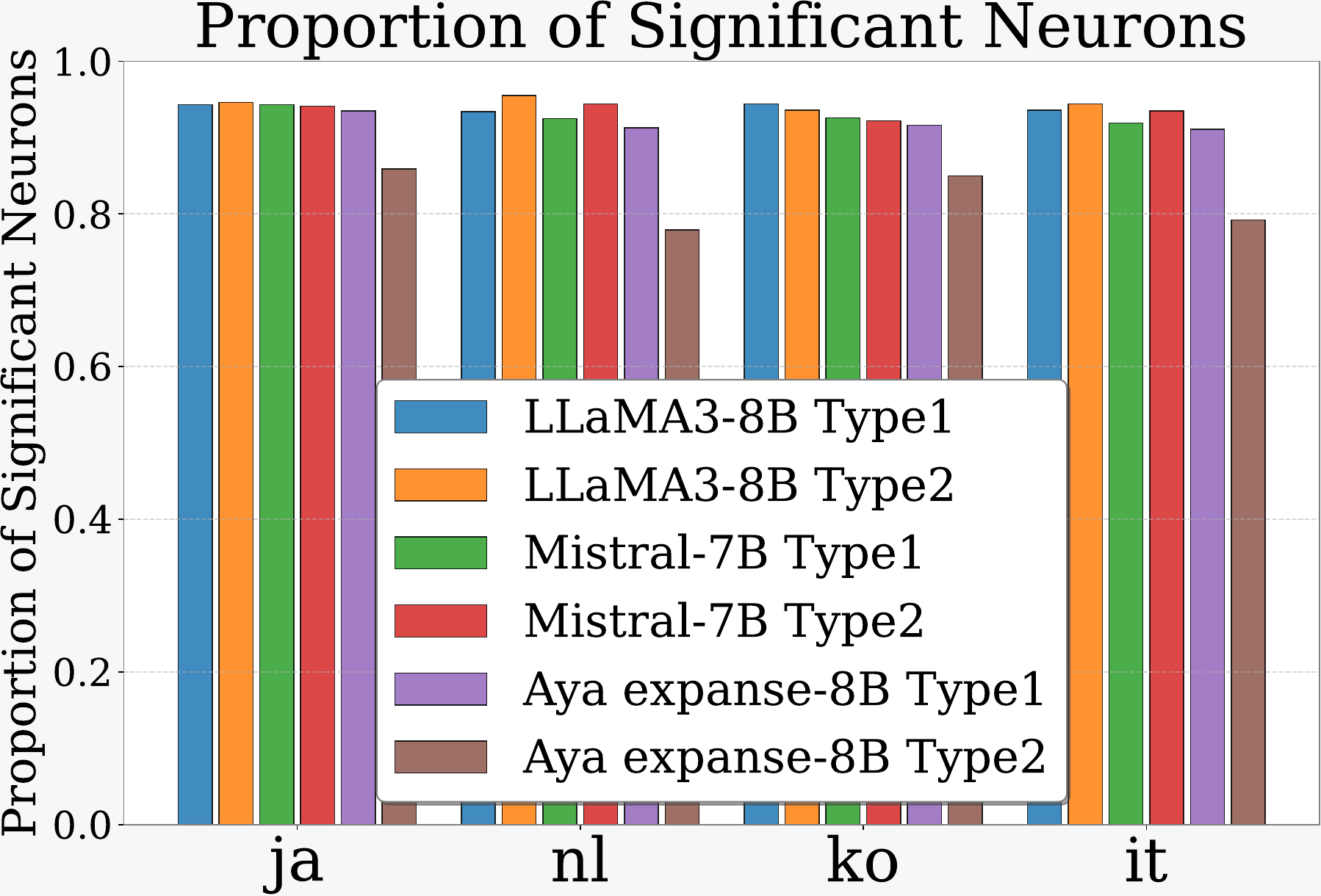}
    \caption{Language-family specificity}
  \end{subfigure}
  \caption{Proportion of statistically significant neurons among top-1k Transfer Neurons.}
  \label{fig:appendix:proportions of significant neurons}
\end{figure*}

Additionally, as shown in Fig.~\ref{fig:appendix:proportions of significant neurons}, it turned out that approximately 80-95\% of top-1k transfer neurons are statistically significant, which further strengthen our correlation analysis.

Furthermore, we conducted a \textit{Mann–Whitney U test} as a non-parametric test under the same settings described above. While ANOVA relies on the \textit{Central Limit Theorem}, assuming that, in our case, the distribution of the mean activation values of transfer neurons converges to a Gaussian distribution. Although it is likely that the mean activations follow the theorem, we also performed the non-parametric Mann–Whitney U test to validate our findings since activation values are outputs of non-linear functions. As a result, it yielded results that were highly consistent with those of ANOVA, further enhancing the reliability of our analysis.

% figure: proportion of significant neurons.

% Type-1, >= 0.1, p < 0.5, language label
\begin{table*}[t]
\small
\centering
\begin{tabular}{lccc}
\toprule
\textbf{} & LLaMA3-8B & Mistral-7B & Aya expanse-8B \\
\midrule
Japanese & 0.433 & 0.412 & 0.386 \\
Dutch    & 0.300 & 0.261 & 0.245 \\
Korean   & 0.337 & 0.305 & 0.287 \\
Italian  & 0.297 & 0.253 & 0.254 \\
\bottomrule
\end{tabular}
\caption{\textbf{\% of Type-1 Transfer Neurons with correlation ratio $> 0.1$ and $p < 0.05$}}
\label{tab:anova_type1_01}
\end{table*}
% Type-2 >= 0.1, p < 0.5, language label
\begin{table*}[t]
\small
\centering
\begin{tabular}{lccc}
\toprule
\textbf{} & LLaMA3-8B & Mistral-7B & Aya expanse-8B \\
\midrule
Japanese & 0.624 & 0.615 & 0.530 \\
Dutch    & 0.635 & 0.559 & 0.379 \\
Korean   & 0.556 & 0.484 & 0.471 \\
Italian  & 0.631 & 0.507 & 0.390 \\
\bottomrule
\end{tabular}
\caption{\textbf{\% of Type-2 Transfer Neurons with correlation ratio $> 0.1$ and $p < 0.05$}}
\label{tab:anova_type2_01}
\end{table*}
% Type-1, >= 0.25, p < 0.5, language label
\begin{table*}[t]
\small
\centering
\begin{tabular}{lccc}
\toprule
\textbf{} & LLaMA3-8B & Mistral-7B & Aya expanse-8B \\
\midrule
Japanese & 0.203 & 0.188 & 0.155 \\
Dutch    & 0.094 & 0.076 & 0.064 \\
Korean   & 0.136 & 0.125 & 0.098 \\
Italian  & 0.089 & 0.066 & 0.065 \\
\bottomrule
\end{tabular}
\caption{\textbf{\% of Type-1 Transfer Neurons with correlation ratio $> 0.25$ and $p < 0.05$}}
\label{tab:anova_type1_025}
\end{table*}
% Type-2 >= 0.25, p < 0.5, language label
\begin{table*}[t]
\small
\centering
\begin{tabular}{lccc}
\toprule
\textbf{} & LLaMA3-8B & Mistral-7B & Aya expanse-8B \\
\midrule
Japanese & 0.413 & 0.363 & 0.282 \\
Dutch    & 0.377 & 0.232 & 0.165 \\
Korean   & 0.345 & 0.254 & 0.233 \\
Italian  & 0.383 & 0.221 & 0.144 \\
\bottomrule
\end{tabular}
\caption{\textbf{\% of Type-2 Transfer Neurons with correlation ratio $> 0.25$ and $p < 0.05$}}
\label{tab:anova_type2_0.25}
\end{table*}
% Type-1, >= 0.1, p < 0.05, language-family label
\begin{table*}[t]
\small
\centering
\begin{tabular}{lccc}
\toprule
\textbf{} & LLaMA3-8B & Mistral-7B & Aya expanse-8B \\
\midrule
Japanese & 0.433 & 0.412 & 0.386 \\
Dutch    & 0.300 & 0.261 & 0.245 \\
Korean   & 0.337 & 0.305 & 0.287 \\
Italian  & 0.297 & 0.253 & 0.254 \\
\bottomrule
\end{tabular}
\caption{\textbf{\% of Type-1 Transfer Neurons with correlation ratio $> 0.1$ and $p < 0.05$, Language-family label}}
\label{tab:anova_type1_01_language_family}
\end{table*}
% Type-2, >= 0.1, p < 0.05, language-family label
\begin{table*}[t]
\small
\centering
\begin{tabular}{lccc}
\toprule
\textbf{} & LLaMA3-8B & Mistral-7B & Aya expanse-8B \\
\midrule
Japanese & 0.624 & 0.615 & 0.530 \\
Dutch    & 0.635 & 0.559 & 0.379 \\
Korean   & 0.556 & 0.484 & 0.471 \\
Italian  & 0.631 & 0.507 & 0.390 \\
\bottomrule
\end{tabular}
\caption{\textbf{\% of Type-2 Transfer Neurons with correlation ratio $> 0.1$ and $p < 0.05$, Language-family label}}
\label{tab:anova_type2_01_language_family}
\end{table*}
% Type-2, >= 0.1, p < 0.05, language-family label
\begin{table*}[t]
\small
\centering
\begin{tabular}{lccc}
\toprule
\textbf{} & LLaMA3-8B & Mistral-7B & Aya expanse-8B \\
\midrule
Japanese & 0.203 & 0.188 & 0.155 \\
Dutch    & 0.094 & 0.076 & 0.064 \\
Korean   & 0.136 & 0.125 & 0.098 \\
Italian  & 0.089 & 0.066 & 0.065 \\
\bottomrule
\end{tabular}
\caption{\textbf{\% of Type-1 Transfer Neurons with correlation ratio $> 0.25$ and $p < 0.05$, Language-family label}}
\label{tab:anova_type1_025_language_family}
\end{table*}
% Type-2, >= 0.25, p < 0.05, language-family label
\begin{table*}[t]
\small
\centering
\begin{tabular}{lccc}
\toprule
\textbf{} & LLaMA3-8B & Mistral-7B & Aya expanse-8B \\
\midrule
Japanese & 0.413 & 0.363 & 0.282 \\
Dutch    & 0.377 & 0.232 & 0.165 \\
Korean   & 0.345 & 0.254 & 0.233 \\
Italian  & 0.383 & 0.221 & 0.144 \\
\bottomrule
\end{tabular}
\caption{\textbf{\% of Type-2 Transfer Neurons with correlation ratio $> 0.25$ and $p < 0.05$, Language-family label}}
\label{tab:anova_type2_025_language_family}
\end{table*}

\section{The Effects of Transfer Neurons on Internal Representations under Cross-Lingual Deactivation}
\paragraph{Experimental Setup.}To analyze cross-lingual effects of transfer neurons in hidden state space, we compute the centroids of L2 hidden states (L2 latent space) at each layer: (i) $\bm{C}^l_\mathrm{L2}$: without any deactivation, (ii) $\bm{C}^{l-\mathrm{L1TN}}_{\mathrm{L2}}$: with L1 transfer neurons deactivated, and (iii) $\bm{C}^{l-\mathrm{L2TN}}_\mathrm{L2}$: with L2 transfer neurons deactivated. We then compute $\mathrm{cos}(\bm{C}^l_\mathrm{L2}, \bm{C}^{l-\mathrm{L1TN}}_{\mathrm{L2}})$ to assess the effect of cross-lingual deactivation on both types of neurons, using $\mathrm{cos}(\bm{C}^l_{\mathrm{L2}}, \bm{C}^{l-\mathrm{L2TN}}_{\mathrm{L2}})$ as a reference point.
\paragraph{Results.}
We report the results for final layer of each type of neurons (i.e., 20th layer for Type-1 neurons and final layer for Type-2 neurons) for LLaMA3-8B, Mistral-7B, and Aya expanse-8B across all the languages in Tabs.~\ref{tab:appendix:cross-lingual_hs_llama},~\ref{tab:appendix:cross-lingual_hs_mistral}, and~\ref{tab:appendix:cross-lingual_hs_aya}.
Deactivating Type-1 neurons for L1 had a similar impact as the L2 deactivation baseline, indicating that these neurons have a noticeable effect on the hidden state dynamics of L2 inputs. This aligns with our finding in \S\ref{sec:language-specific nature of transfer neurons} that Type-1 neurons are less language-specific. On the other hand, deactivating Type-2 neurons for L1 had a much smaller impact compared to the L2 deactivation baseline, indicating that these neurons have little to no influence on the hidden state dynamics of L2 inputs. A limited degree of influence was observed in cases where the target latent spaces of L1 and L2 were spatially close (e.g., L1 = Dutch and L2 = Italian). This is consistent with our finding in \S\ref{sec:language-specific nature of transfer neurons} that Type-2 neurons are more language-specific.

\section{Downstream Reasoning Tasks while Deactivating Transfer Neurons}
\label{sec:appendix:downstream task while deactivating Type-1}

\subsection{Multilingual Knowledge QA (MKQA)}
\label{sec:appendix: mkqa}
\begin{figure*}[t]
  \centering

  % F1 >= 0.8
  \includegraphics[width=0.24\linewidth]{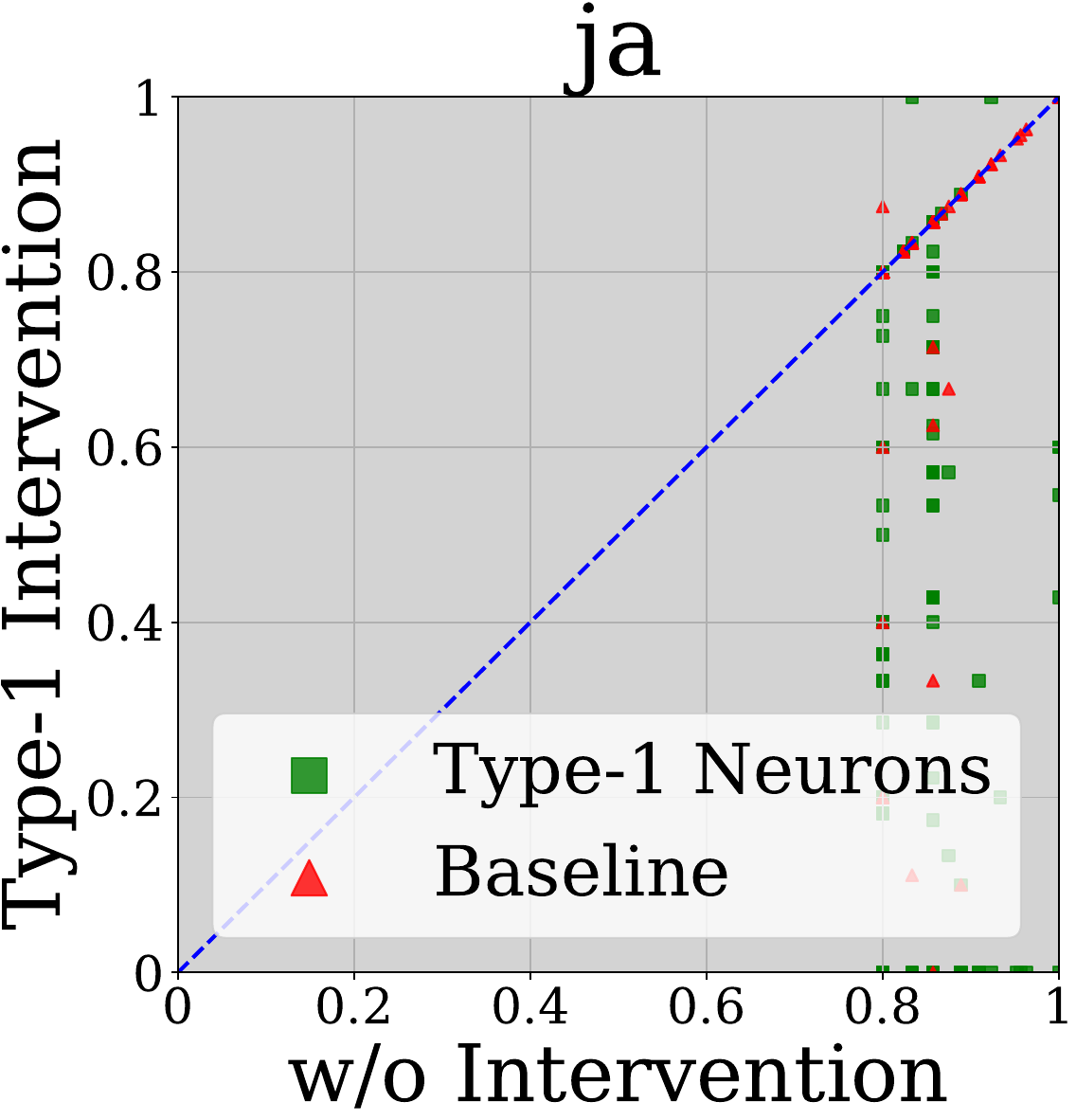}
  \includegraphics[width=0.24\linewidth]{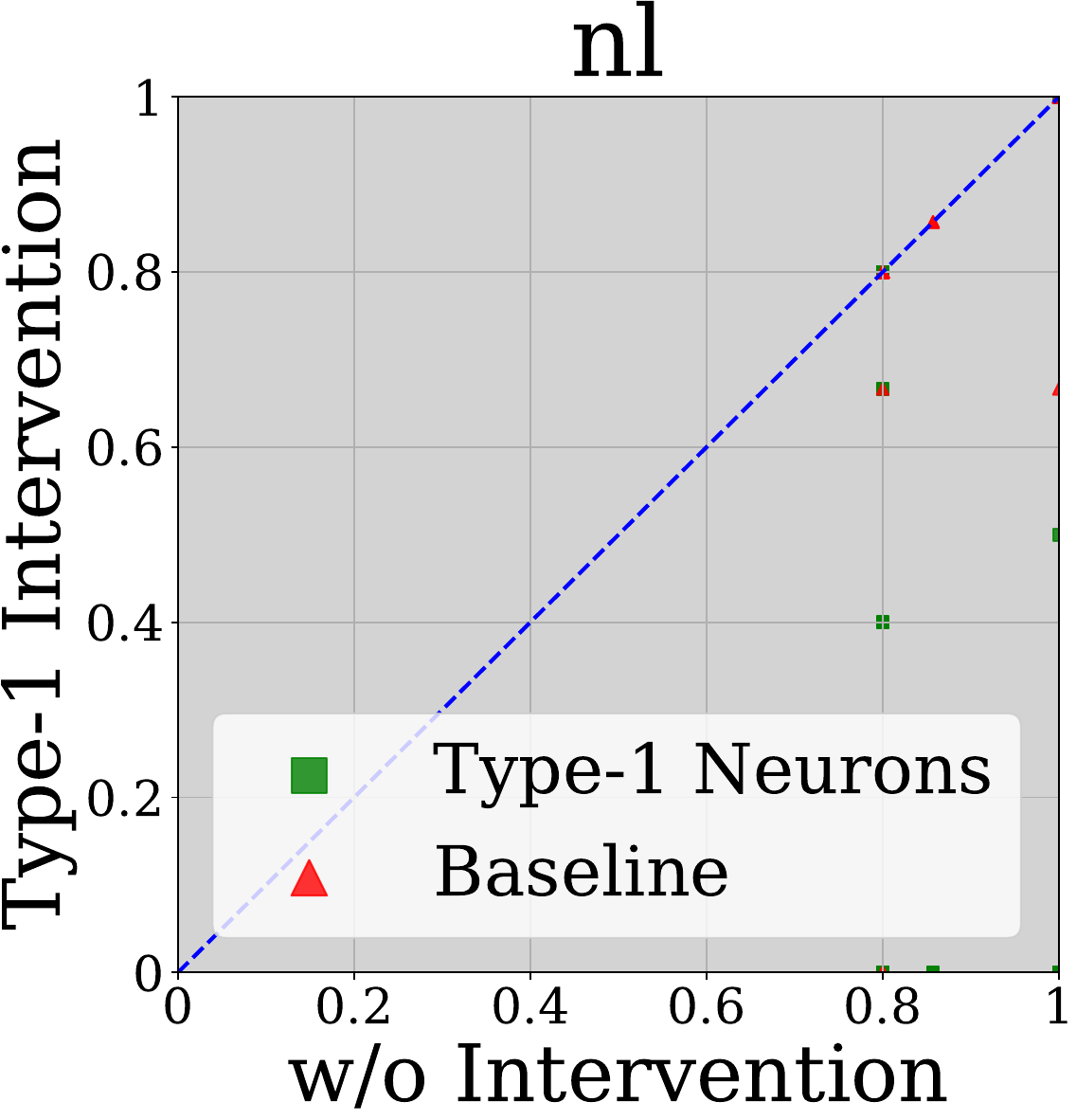}
  \includegraphics[width=0.24\linewidth]{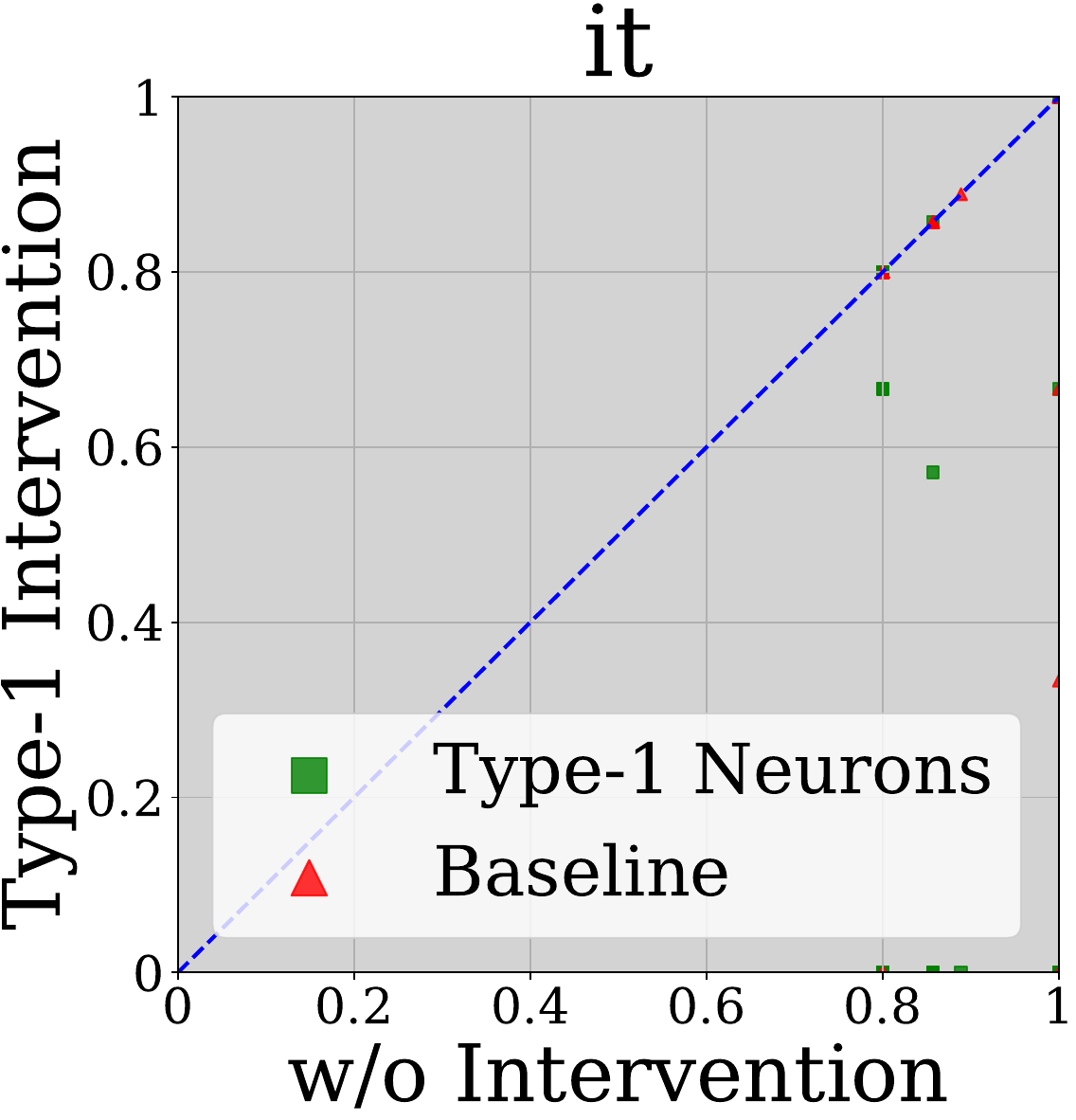}
  \includegraphics[width=0.24\linewidth]{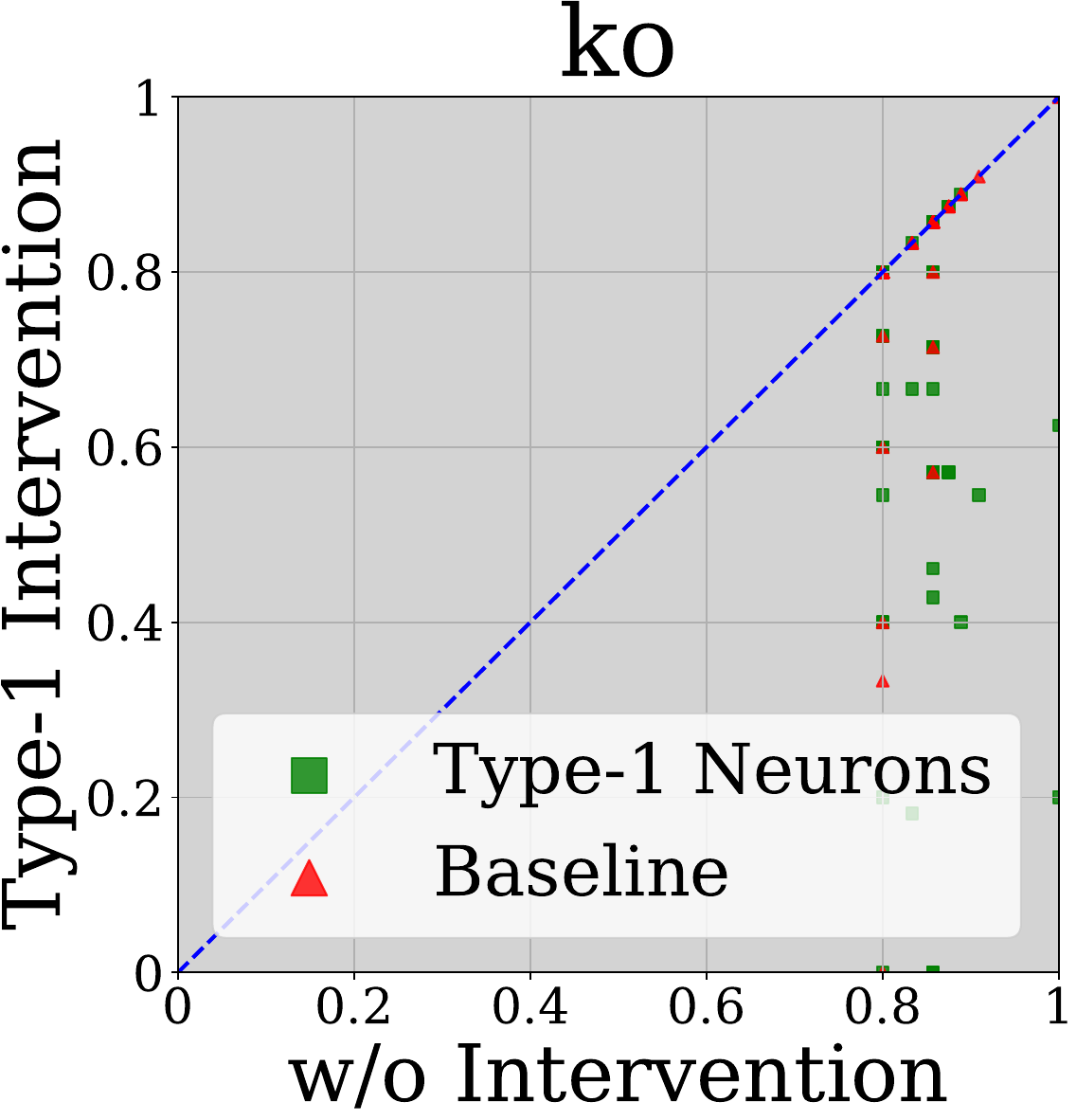}
  \makebox[\linewidth][c]{\small F1 $\geq$ 0.8}

  \caption{\textbf{Token-based F1 score when deactivating top-1k Type-1 transfer neurons (LLaMA3-8B).}
  \textcolor[HTML]{228B22}{Green} denotes the score with intervention in \textcolor[HTML]{228B22}{the Type-1 neurons}, while \textcolor{red}{red} represents the score with intervention in \textcolor{red}{the randomly sampled neurons}. Points below the \textcolor{blue}{$y=x$ line} indicate a decrease in the score due to the intervention, whereas points above the $y=x$ line indicate an increase in the score. Points on the $y=x$ line denote no change in the score before and after the intervention.}
  \label{fig:appendix:mkqa_llama3_above}
\end{figure*}

\begin{figure*}[t]
  \centering

  % F1 >= 0.5
  \includegraphics[width=0.24\linewidth]{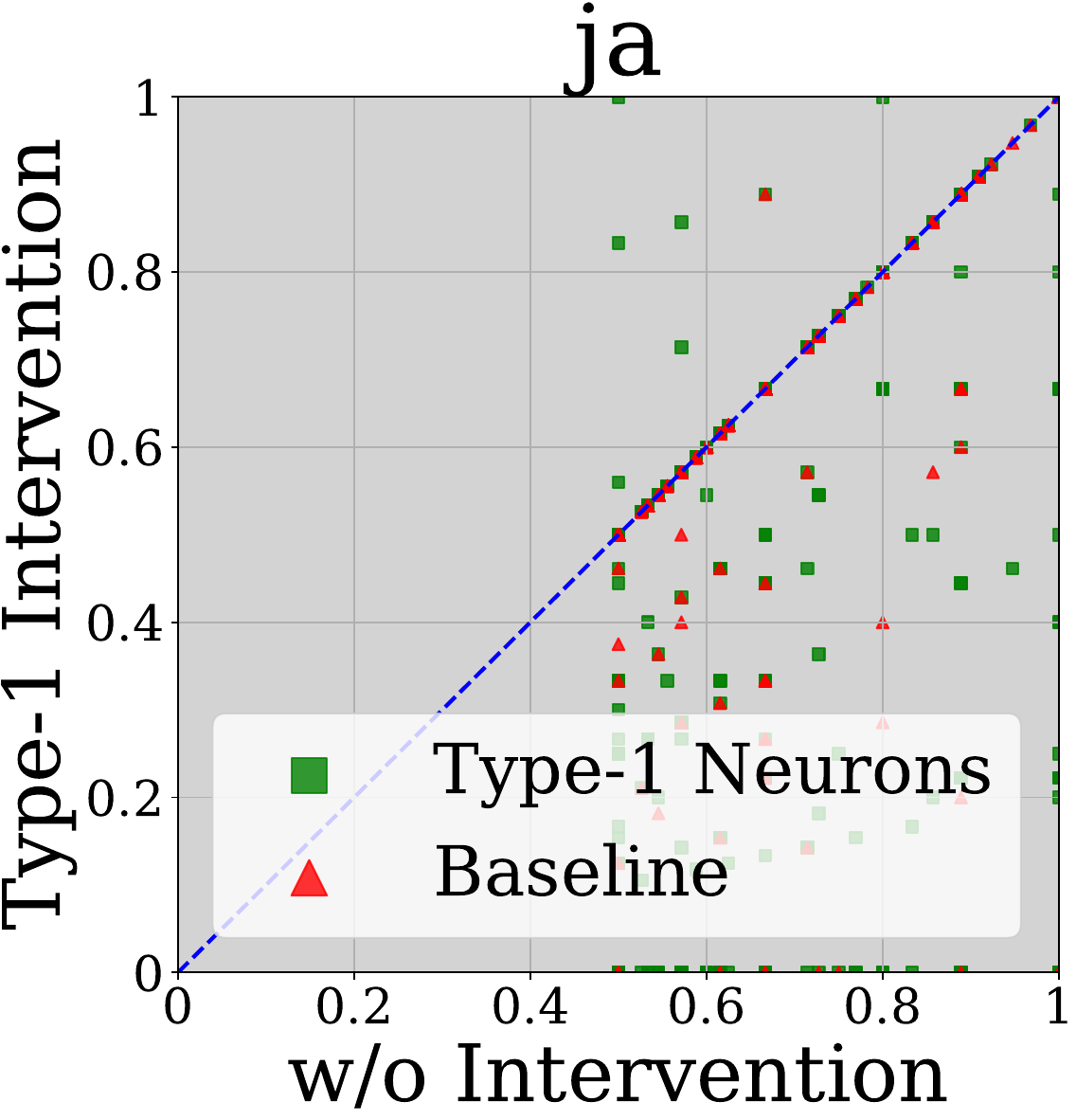}
  \includegraphics[width=0.24\linewidth]{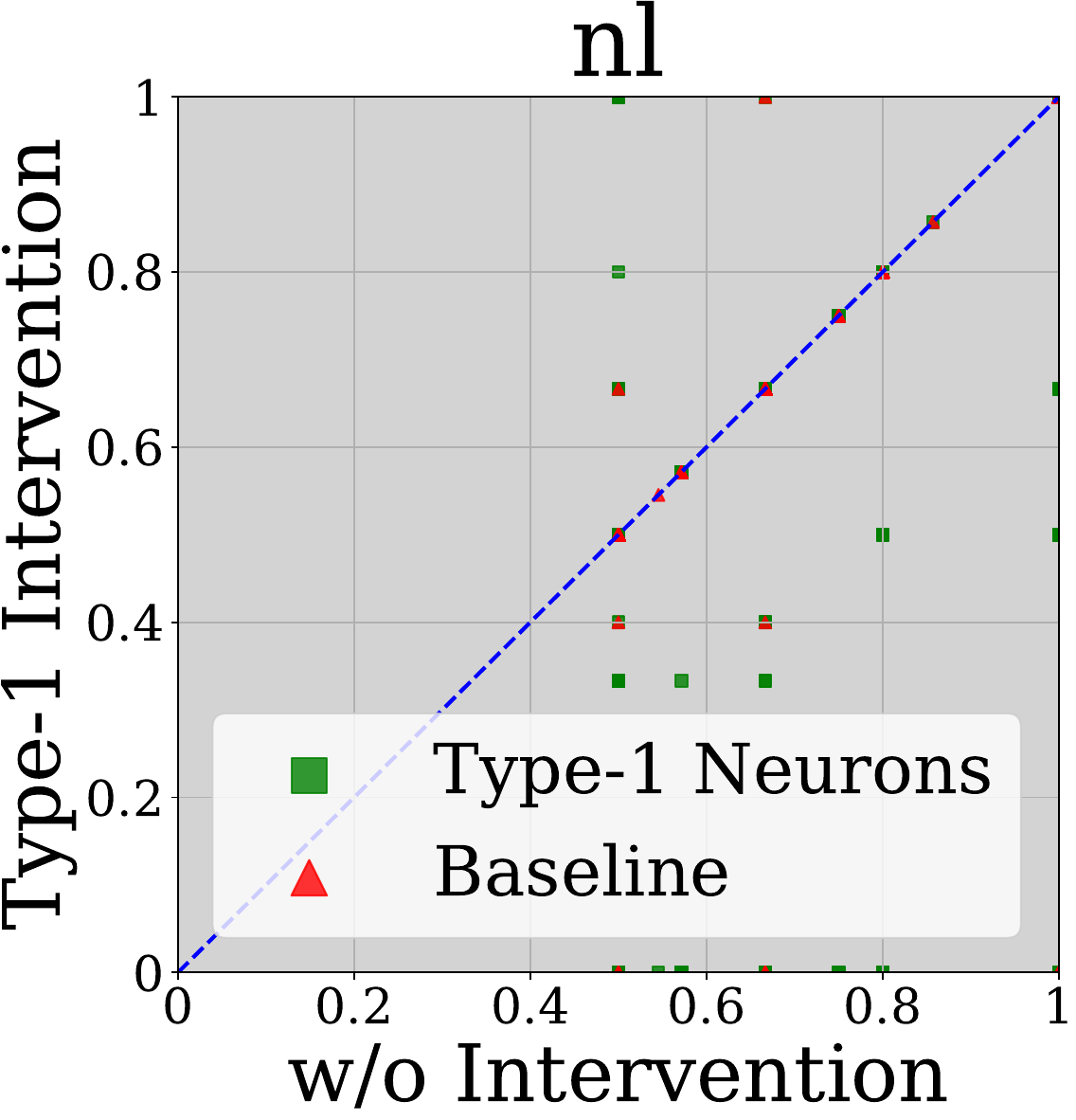}
  \includegraphics[width=0.24\linewidth]{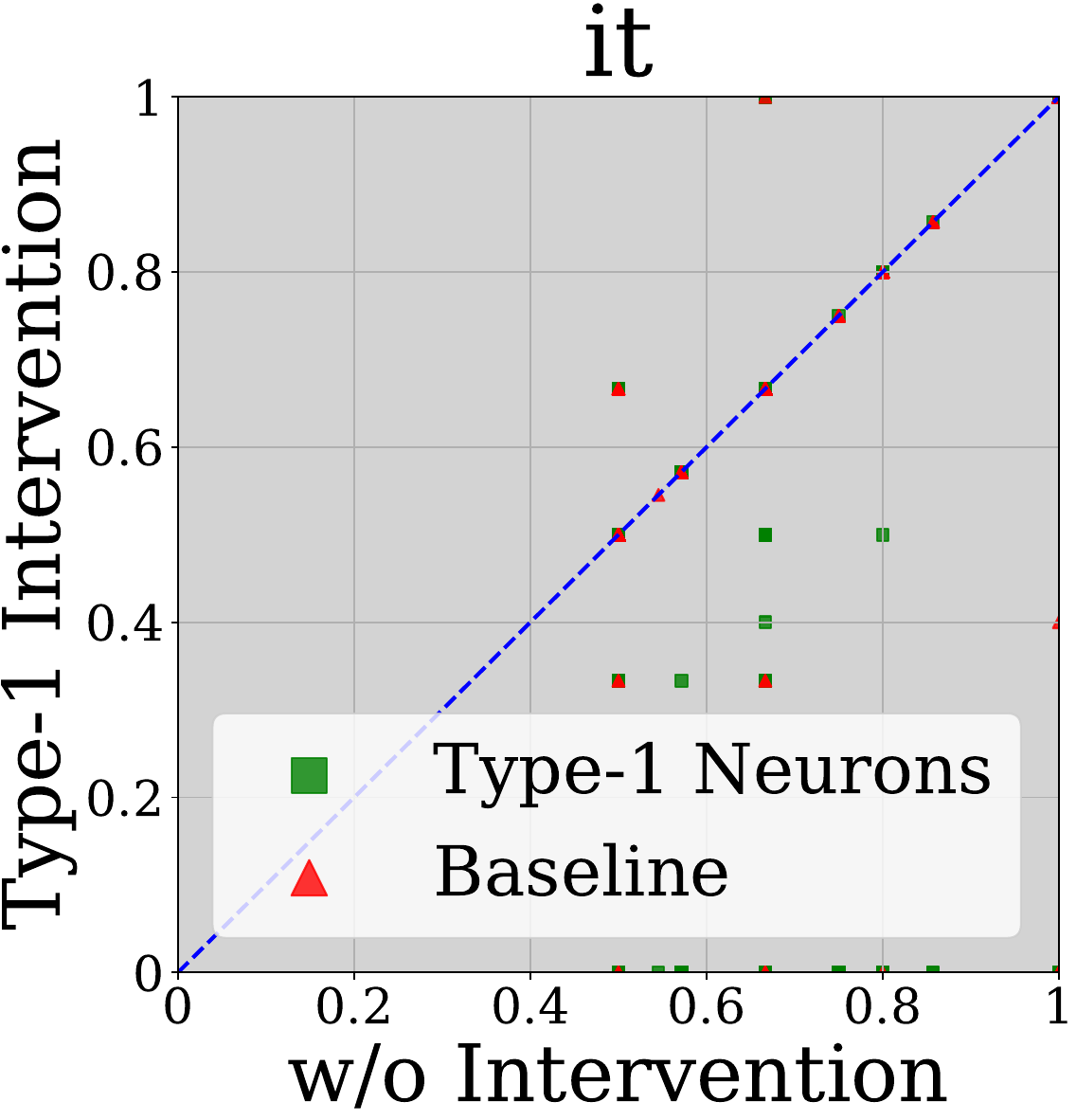}
  \includegraphics[width=0.24\linewidth]{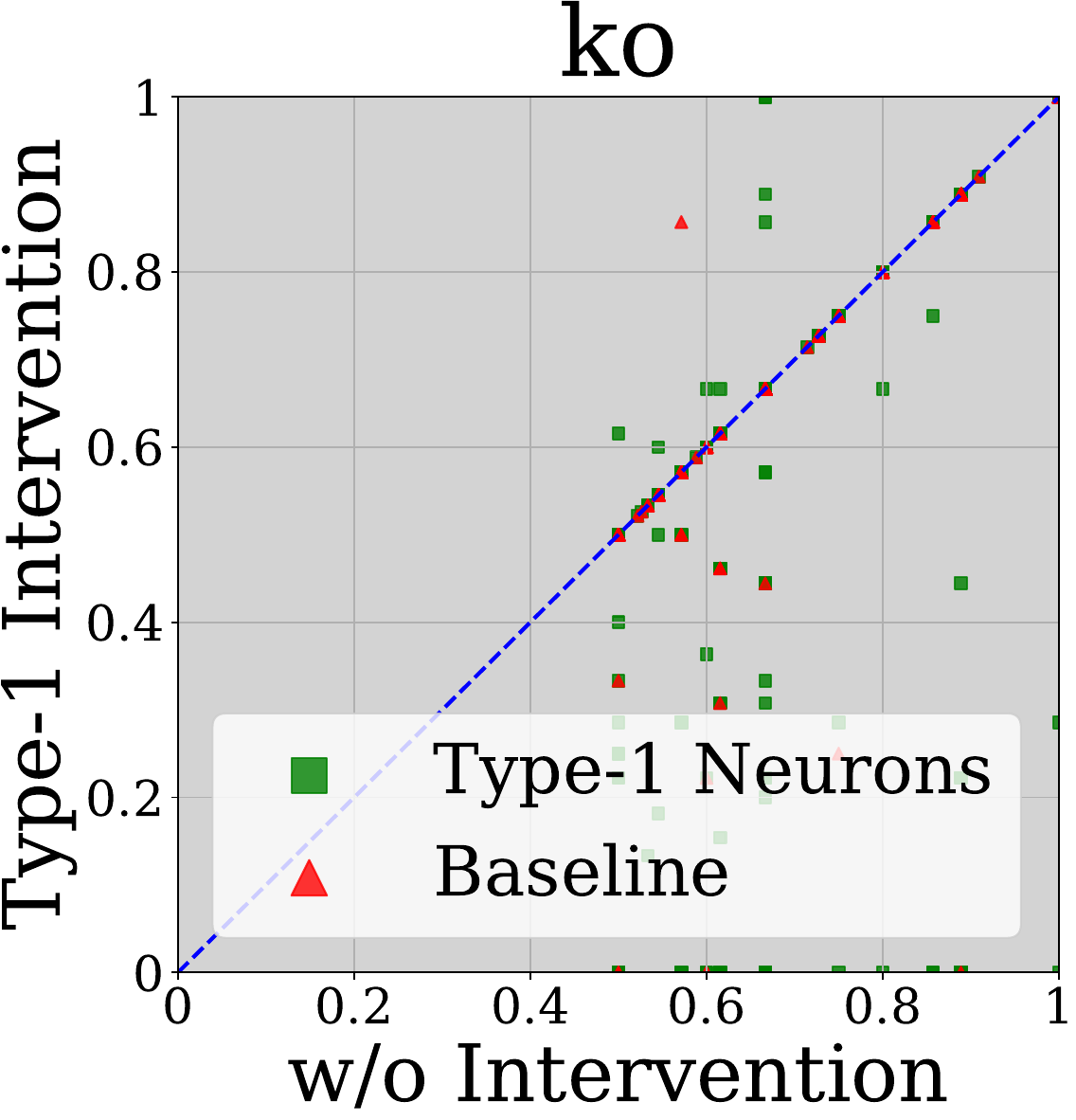}
  \makebox[\linewidth][c]{\small F1 $\geq$ 0.5}

  % F1 >= 0.8
  \includegraphics[width=0.24\linewidth]{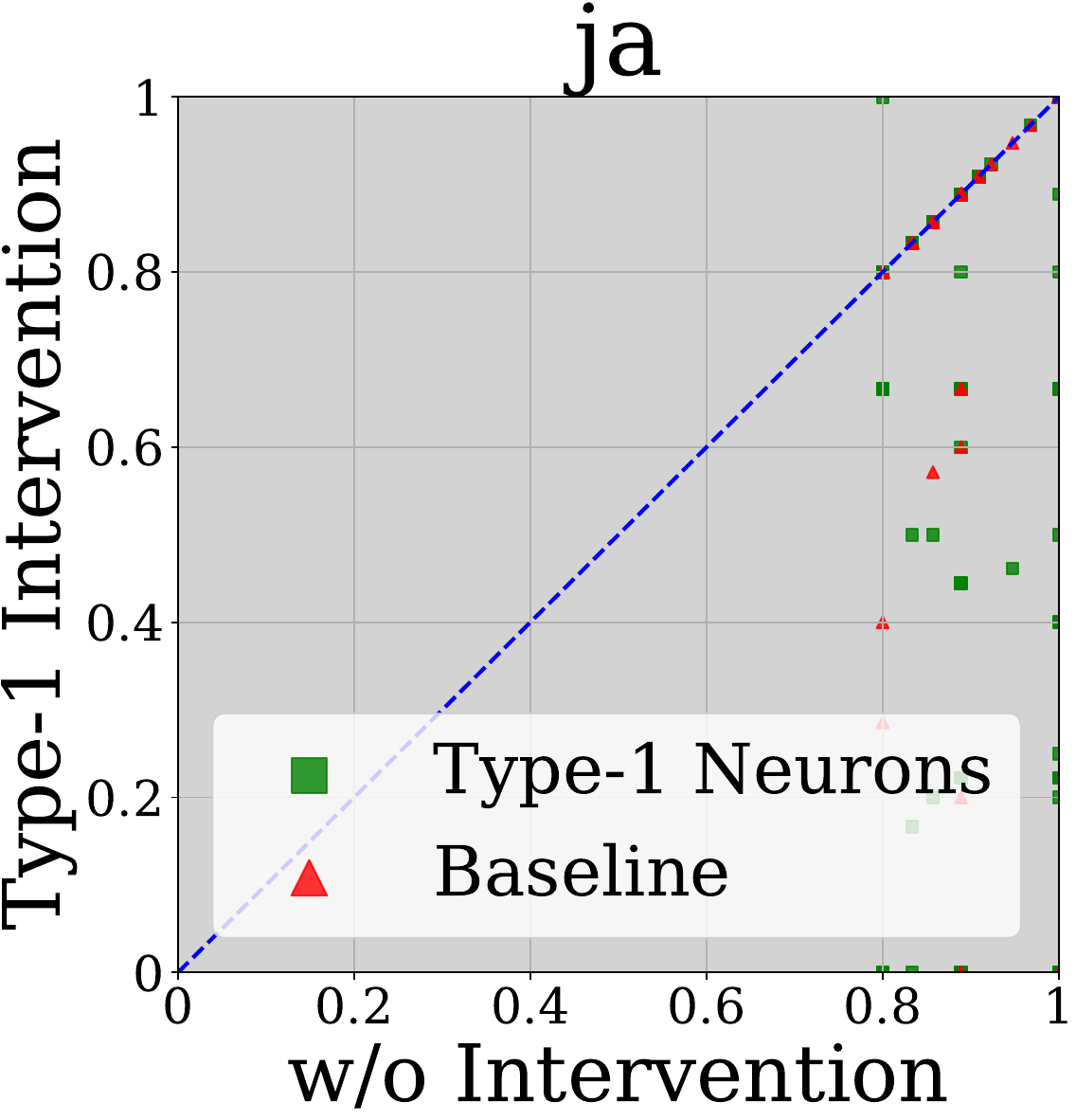}
  \includegraphics[width=0.24\linewidth]{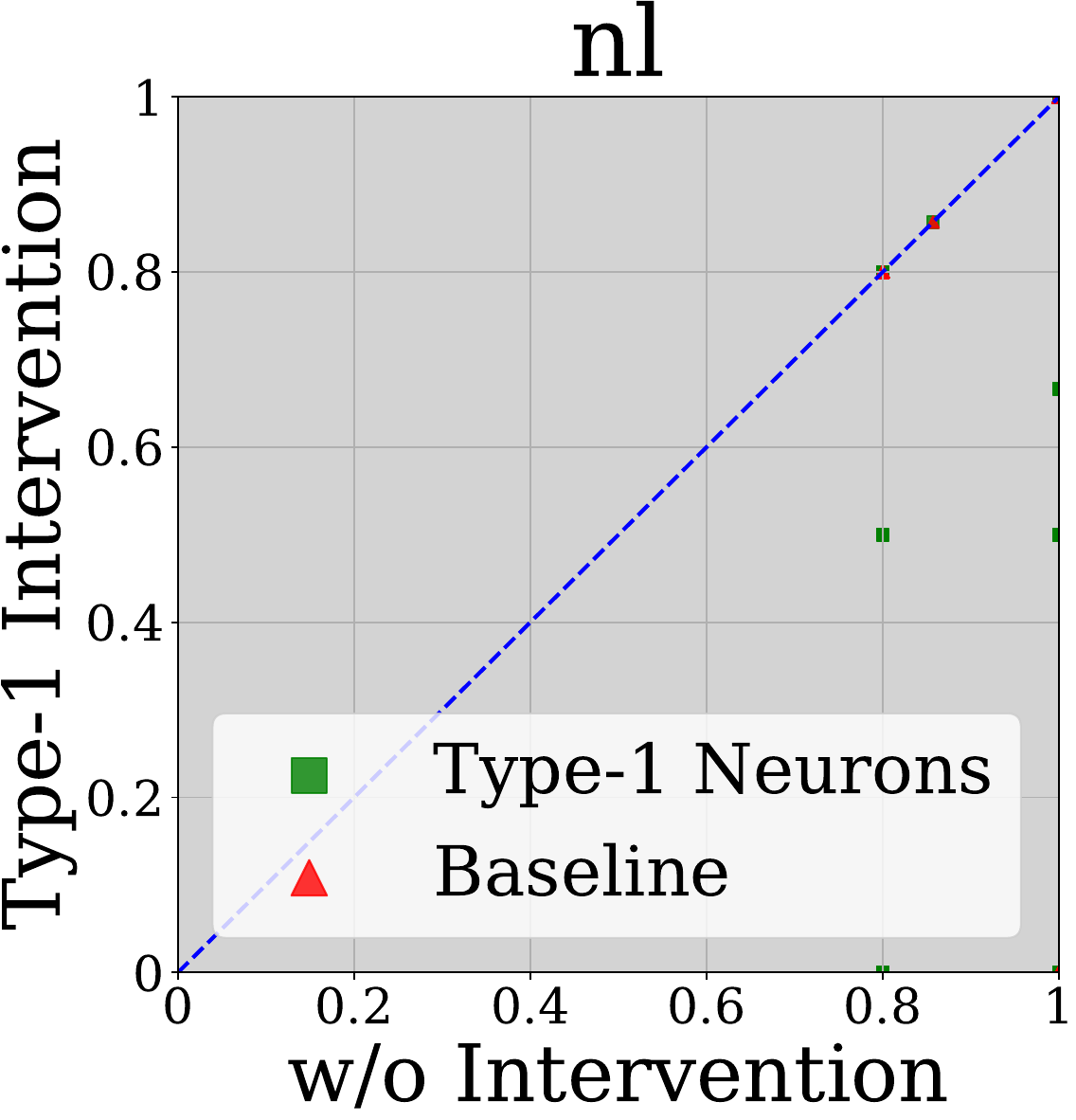}
  \includegraphics[width=0.24\linewidth]{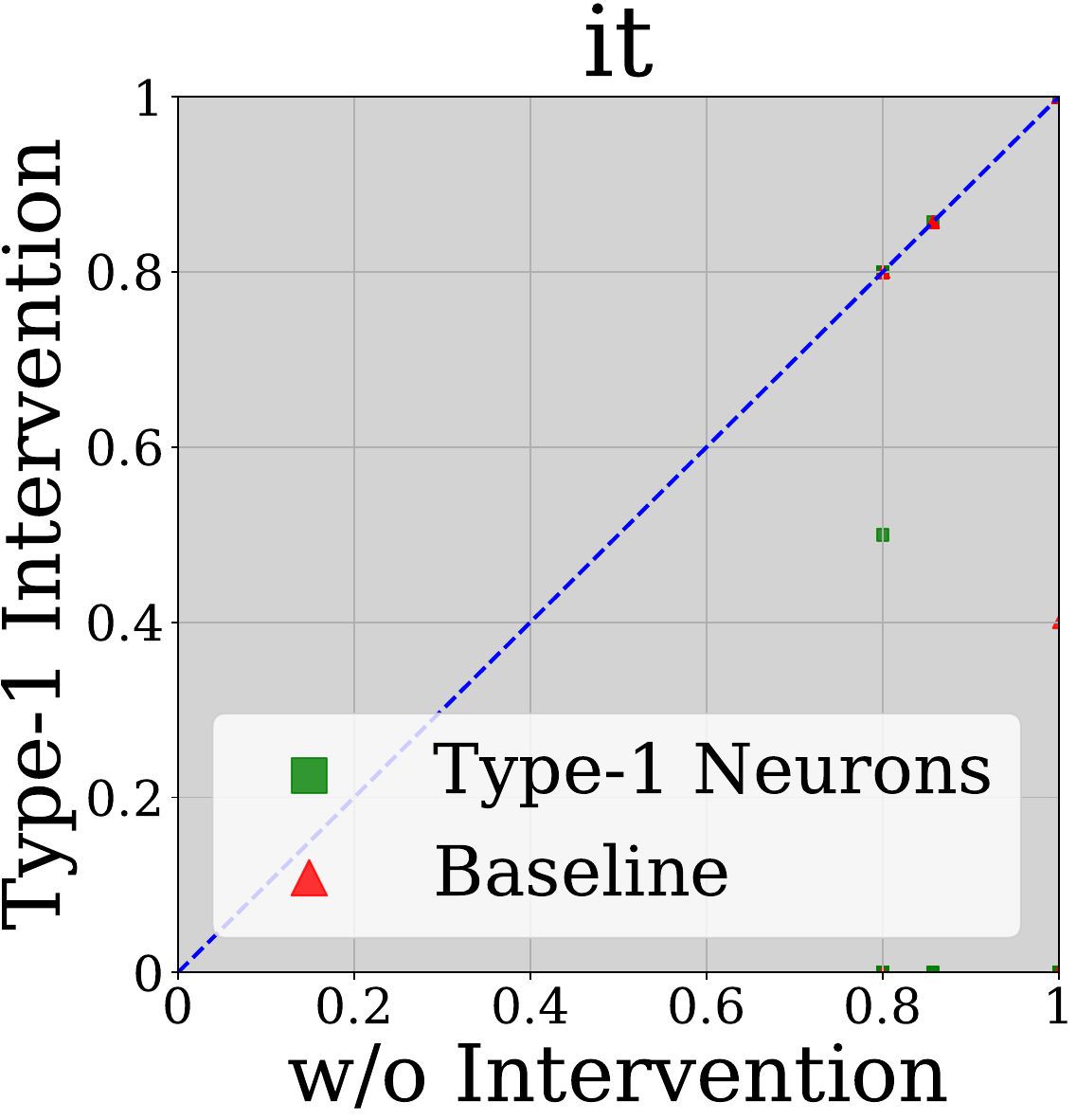}
  \includegraphics[width=0.24\linewidth]{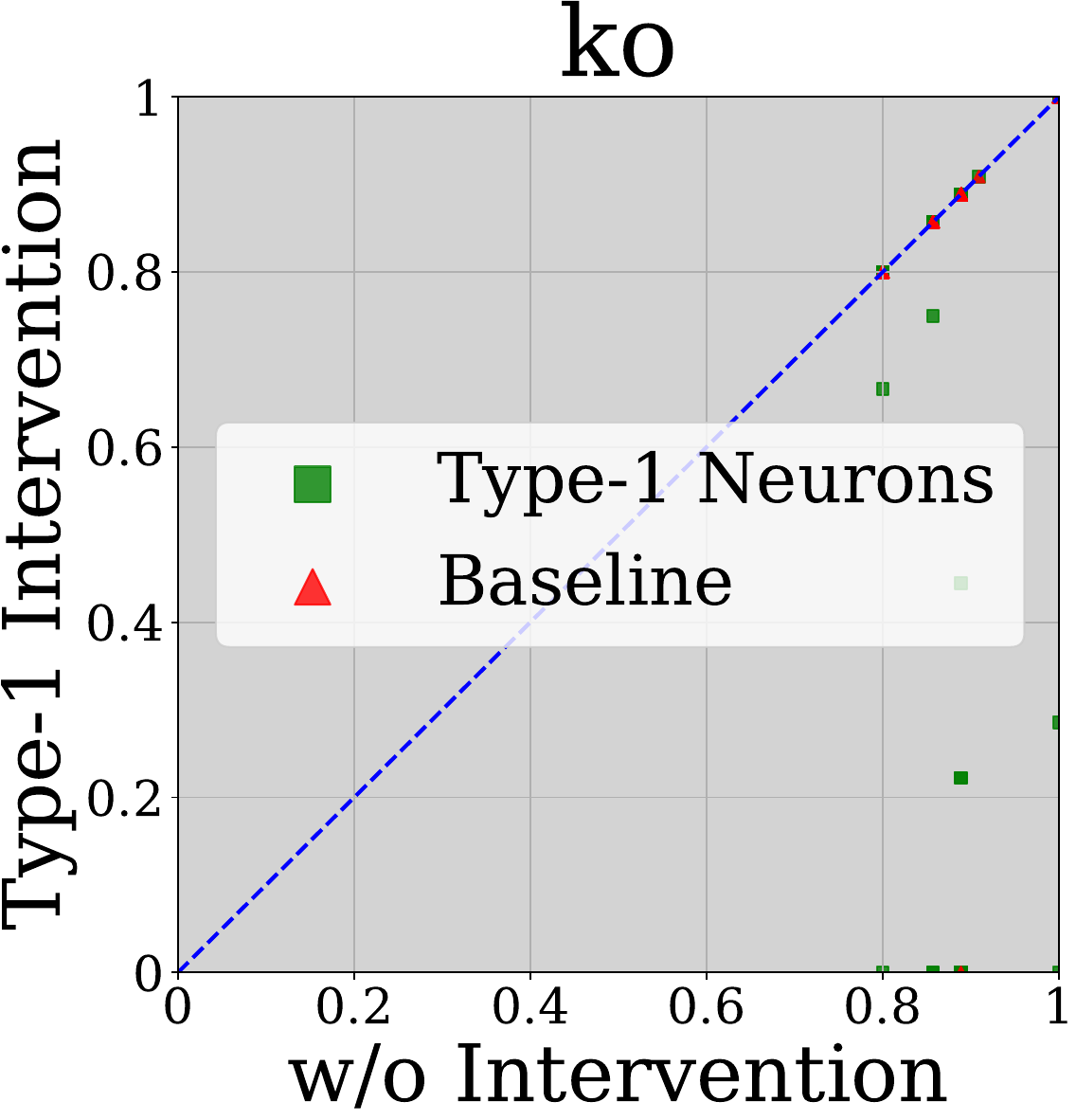}
  \makebox[\linewidth][c]{\small F1 $\geq$ 0.8}

  \caption{\textbf{Token-based F1 score when deactivating top-1k Type-1 transfer neurons (Mistral-7B).}}
  \label{fig:appendix:mkqa_mistral_above}
\end{figure*}

\begin{figure*}[t]
  \centering

  % F1 >= 0.5
  \includegraphics[width=0.24\linewidth]{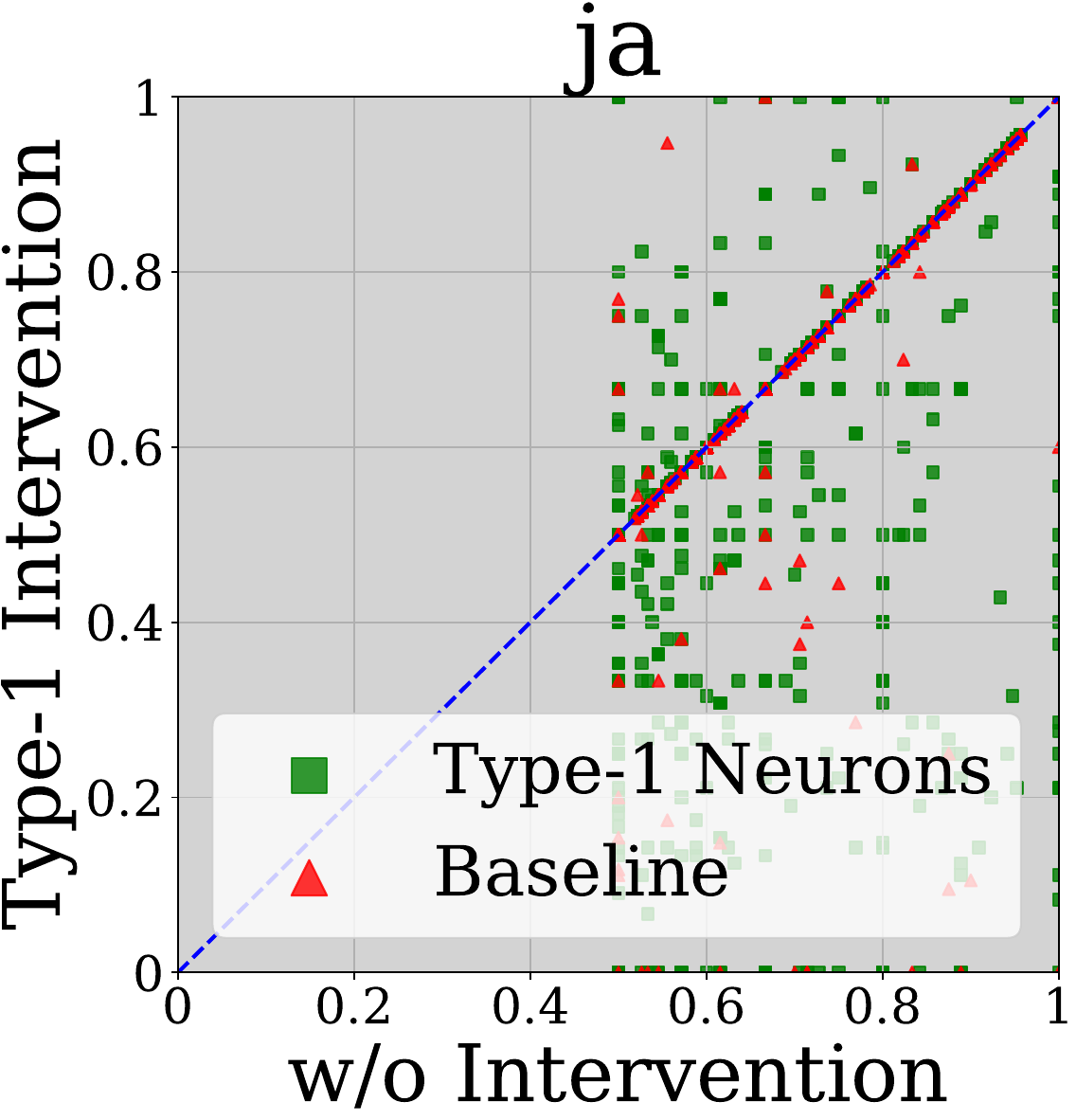}
  \includegraphics[width=0.24\linewidth]{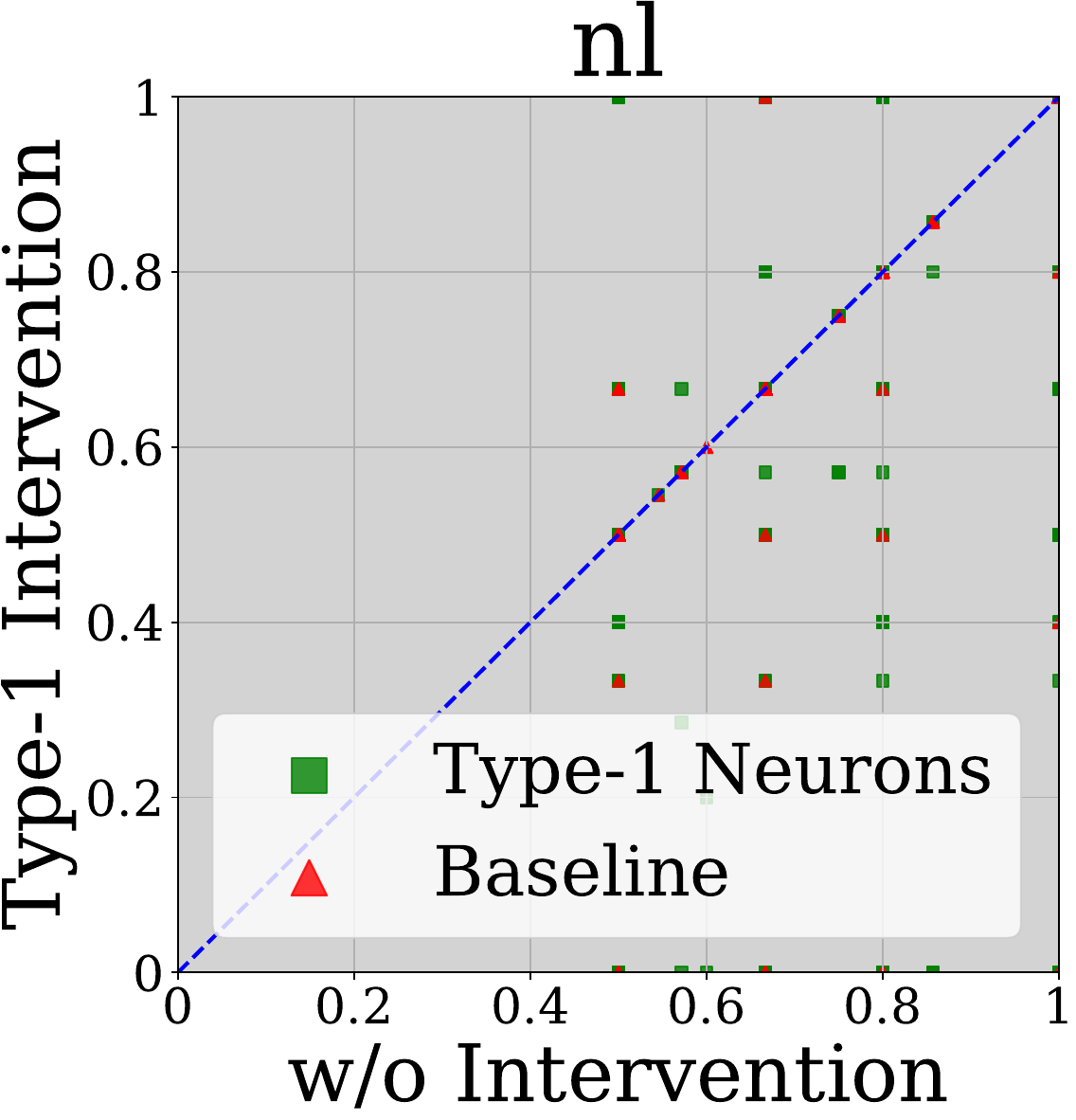}
  \includegraphics[width=0.24\linewidth]{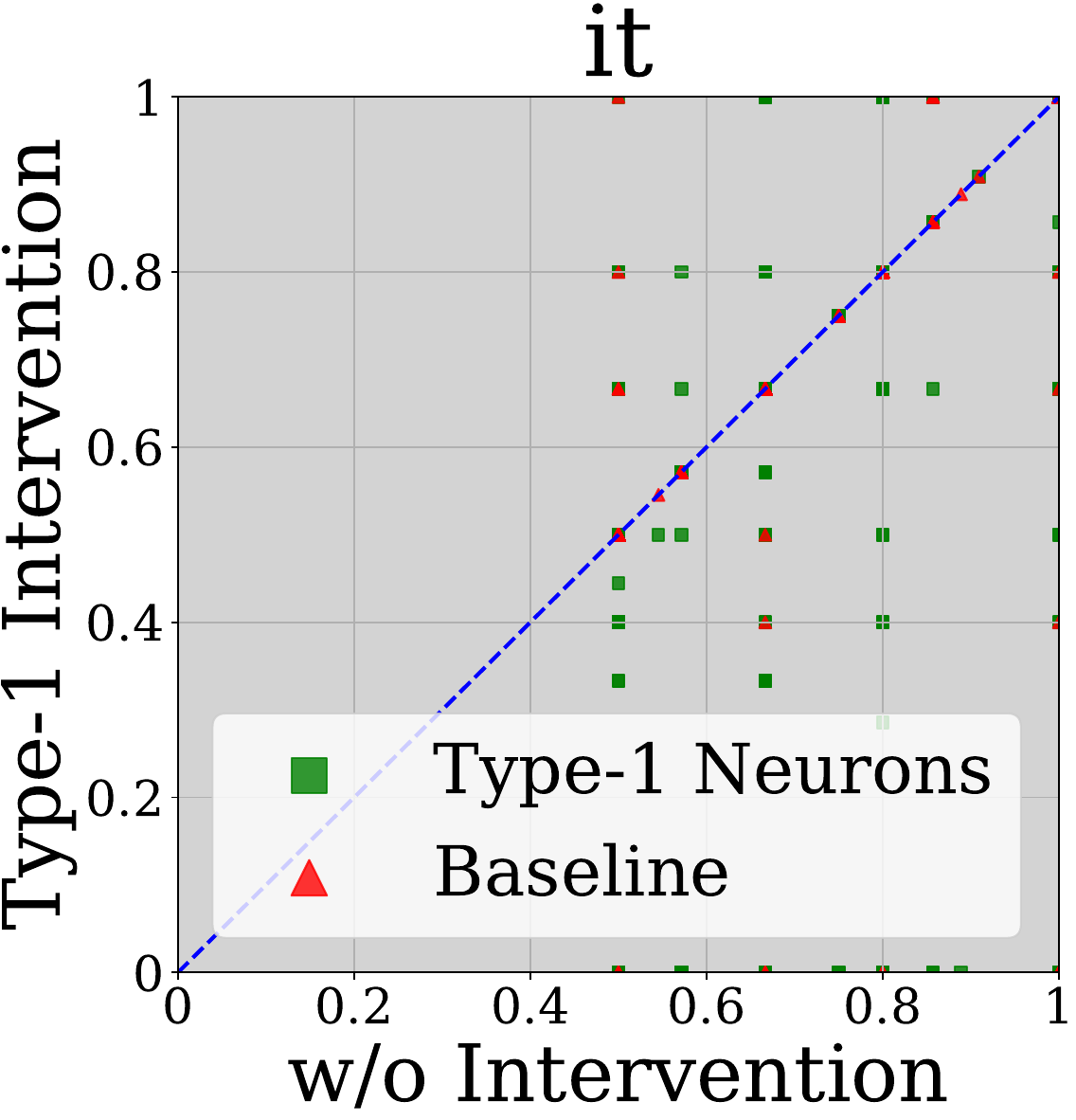}
  \includegraphics[width=0.24\linewidth]{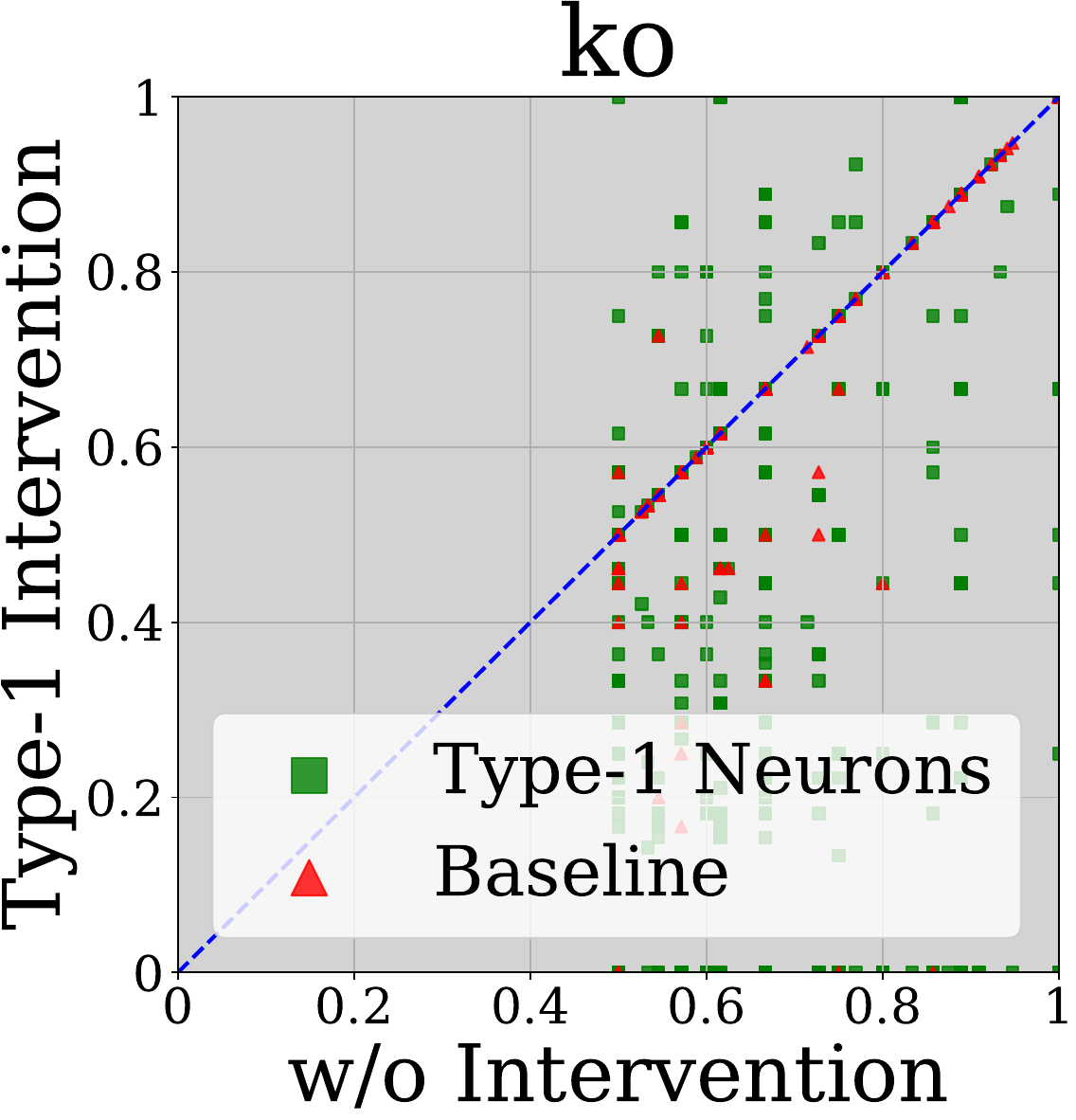}
  \makebox[\linewidth][c]{\small F1 $\geq$ 0.5}

  % F1 >= 0.8
  \includegraphics[width=0.24\linewidth]{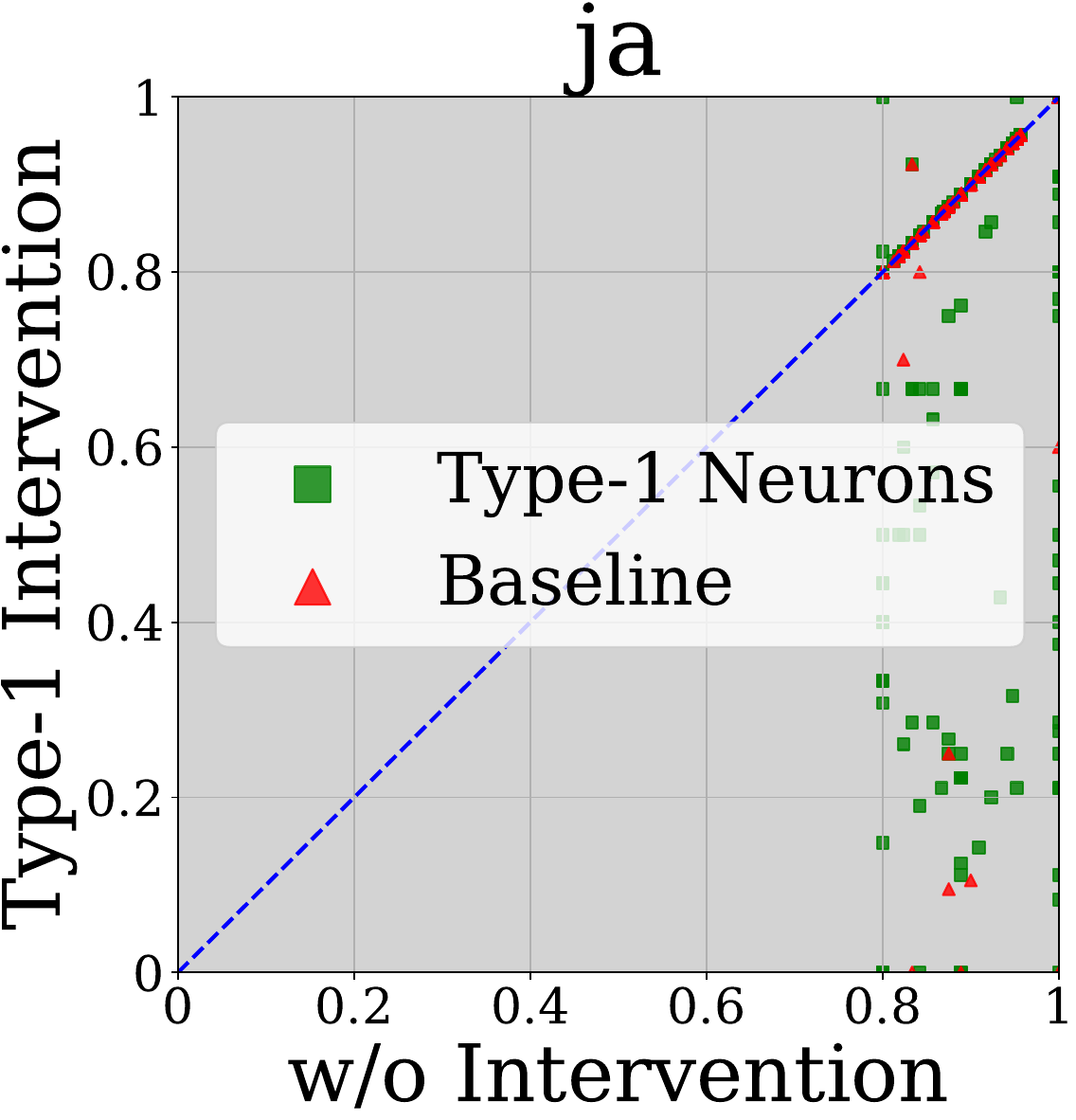}
  \includegraphics[width=0.24\linewidth]{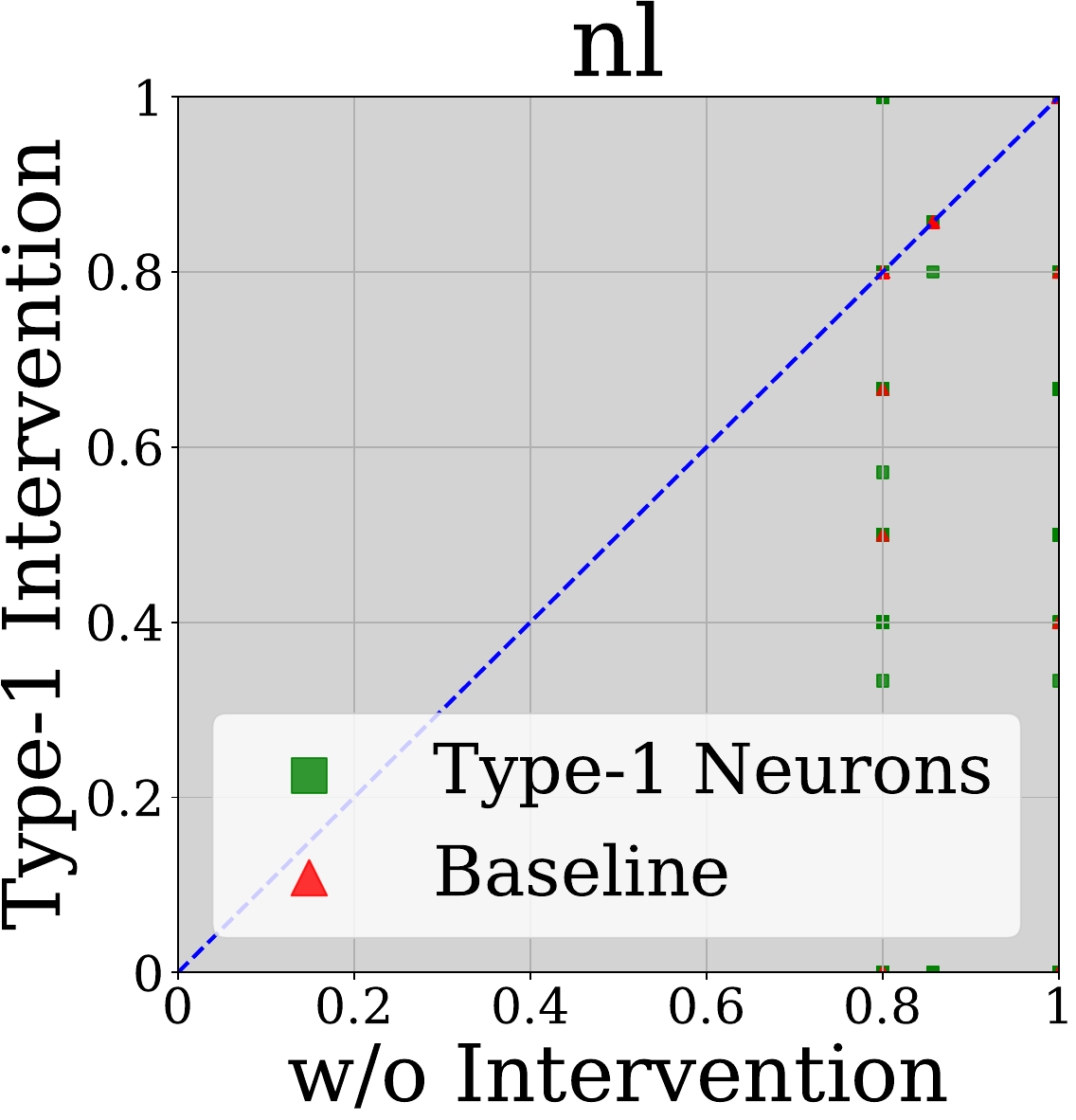}
  \includegraphics[width=0.24\linewidth]{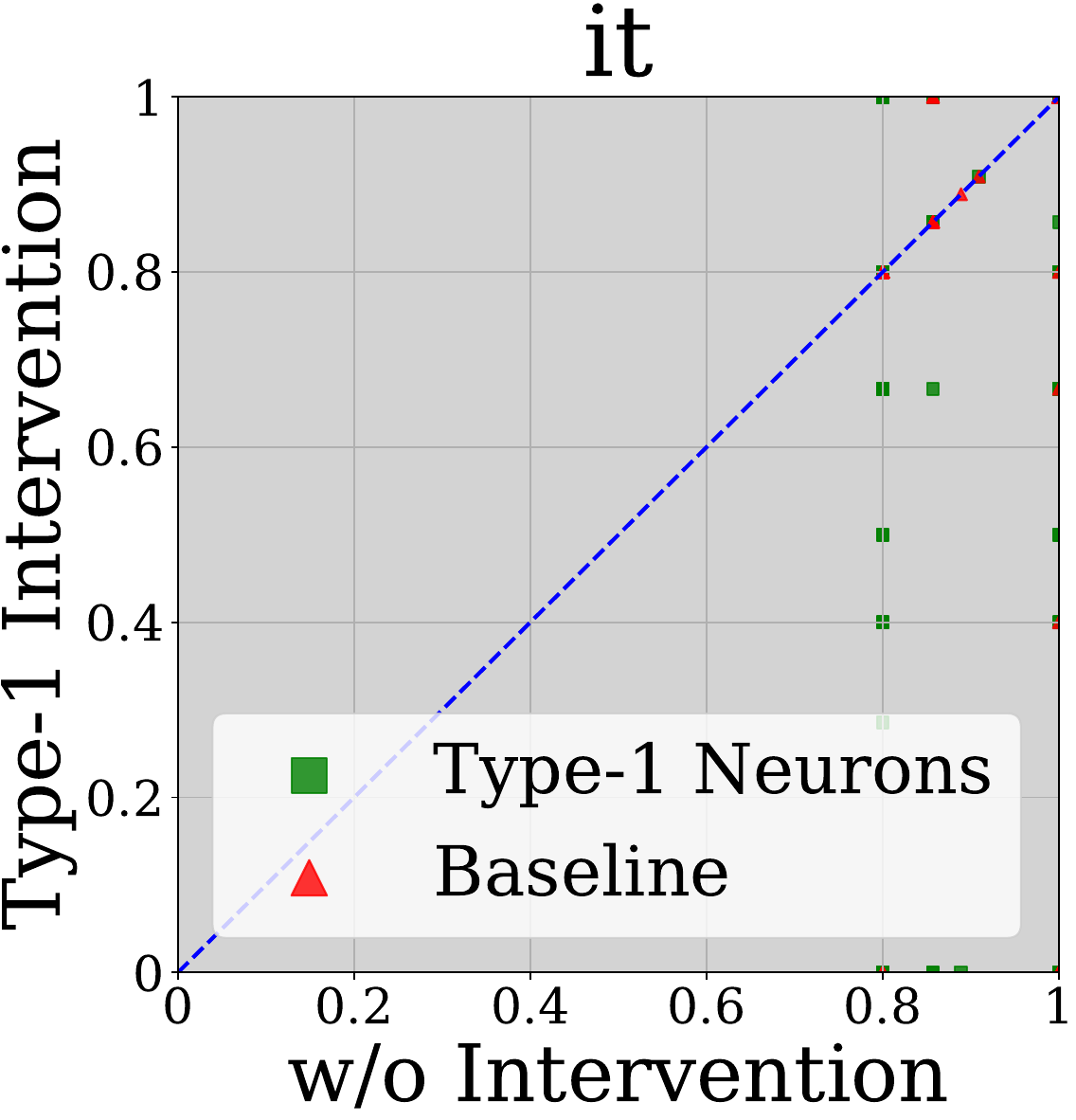}
  \includegraphics[width=0.24\linewidth]{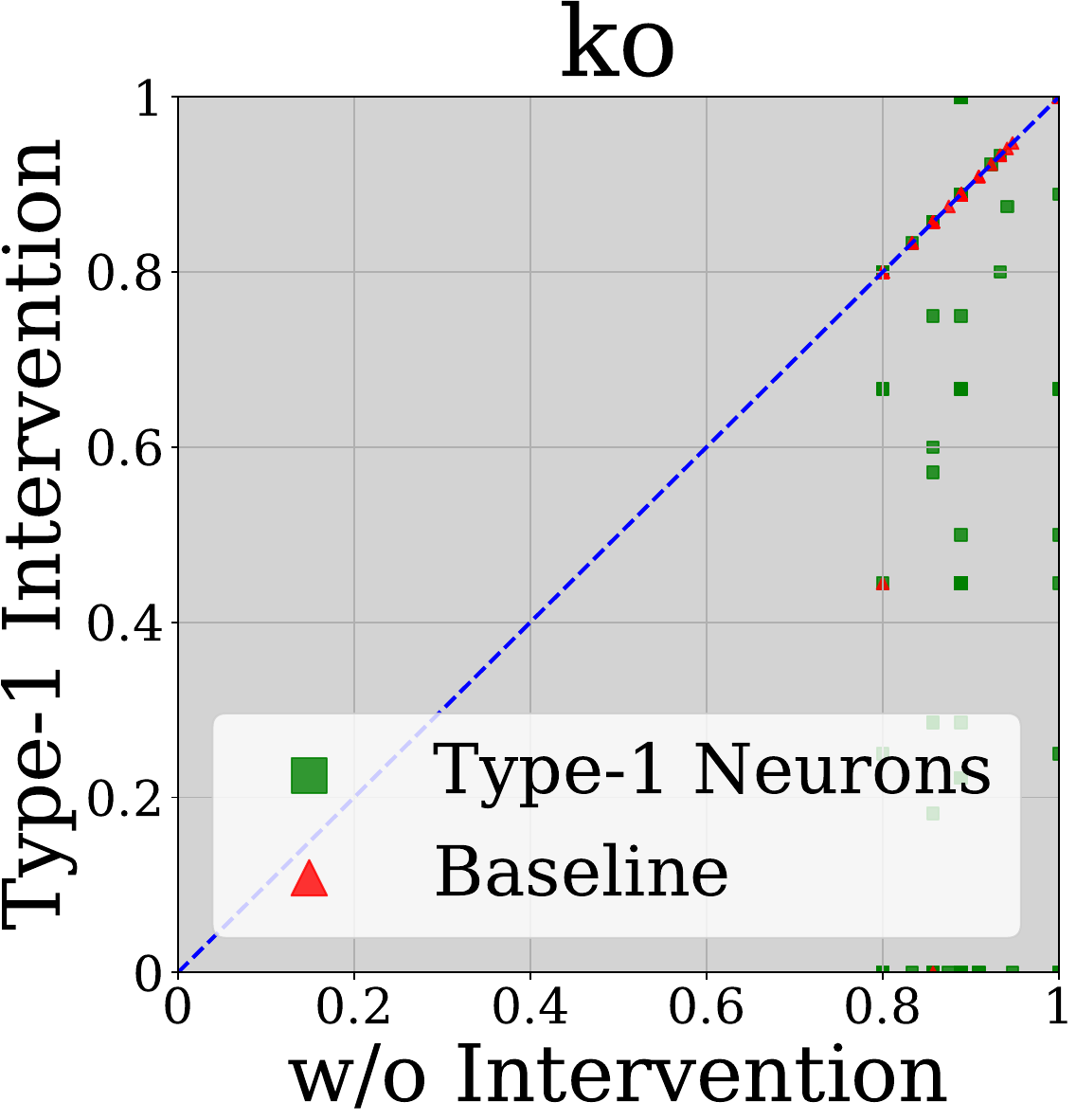}
  \makebox[\linewidth][c]{\small F1 $\geq$ 0.8}

  \caption{\textbf{Token-based F1 score when deactivating top-1k Type-1 transfer neurons (Aya expanse-8B).}}
  \label{fig:appendix: mkqa_aya_above}
\end{figure*}

Figs.~\ref{fig:appendix:mkqa_llama3_above},~\ref{fig:appendix:mkqa_mistral_above}, and~\ref{fig:appendix: mkqa_aya_above} and Tabs~\ref{tab:mkqa_f1_llama_0.8},~\ref{tab:mkqa_f1_mistral_0.5},~\ref{tab:mkqa_f1_mistral_0.8},~\ref{tab:mkqa_f1_aya_0.5}, and~\ref{tab:mkqa_f1_aya_0.8} show all the results for MKQA task described in \S\ref{sec:reasoning}. As shown, the results are highly consistent with those reported in \S\ref{sec:reasoning}.

\subsection{MMLU-ProX}
\label{mmlu-prox}
Tabs.~\ref{mmluprox_type1_baseline_woi_llama3_ja},~\ref{mmluprox_type1_baseline_woi_llama3_ko},~\ref{mmluprox_type1_baseline_woi_llama3_fr},~\ref{mmluprox_type1_baseline_woi_mistral_ja},~\ref{mmluprox_type1_baseline_woi_mistral_ko},~\ref{mmluprox_type1_baseline_woi_mistral_fr},~\ref{mmluprox_type1_baseline_woi_aya_ja},~\ref{mmluprox_type1_baseline_woi_aya_ko}, and~\ref{mmluprox_type1_baseline_woi_aya_fr} present the results of MMLU-ProX under the same settings as MKQA task, as described in \S\ref{sec:reasoning} and Appendix~\ref{sec:appendix: mkqa}. These are the results for Japanese, Korean and French. While our original experiments used Japanese, Korean, Italian, and Dutch, the evaluation tool used in the original paper~\citep{mmluprox}\footnote{\url{https://github.com/EleutherAI/lm-evaluation-harness}} have limited language coverage. Therefore, we report results for Japanese and Korean, and additionally include French as a linguistically closer language to English such as Italian and Dutch. As shown, deactivating top-1k Type-1 neurons significantly degrade performance, whereas deactivating a baseline set of 1k randomly sampled neurons (from the same layers as the Type-1 neurons) results in negligible change compared to the original states (w/o intervention). These results are consistent with the results of MKQA task, further highlighting the critical role of Type-1 neurons in reasoning.

\subsection{Open-Ended Text Generations while Deactivating Transfer Neurons}
\label{sec:appendix:text_generations}
To evaluate the models’ generation quality while nullifying the influence of top-1k transfer neurons during the generation process, we use PolyWrite~\cite{polywrite}, a multilingual open-ended generation task with no single correct answer. The following are some example questions in English.
\begin{itemize}
    \item \texttt{Describe a day on Earth where gravity suddenly reverses for a few hours. How do people and animals react?}
    \item \texttt{Create a character who was once a hero but has become a villain due to a tragic misunderstanding.}
    \item \texttt{Write an email to your supervisor asking for feedback on a recent report you submitted. Express your interest in improving your work and ask for specific suggestions.}
\end{itemize}
We manually evaluate the quality of 50 generated texts for each language under the following settings: (a) without any intervention, (b) with top-1k Type-1 neurons intervention, and (c) with top-1k Type-2 neurons intervention. The evaluation, conducted by the authors, focus on how interventions in settings (b) and (c) alter the generated outputs compared to setting (a)\footnote{The authors are not proficient in Dutch, Korean, and Italian; However, we still made an effort to assess outputs of these languages by using machine translation.}. We conduct generations under greedy decoding to compare outputs before and after the intervention.

\paragraph{Top-1k Type-1 Neurons Deactivated.}
\begin{itemize}
    \item LLaMA3-8B: Generated gibberish, suggesting that the model is no longer capable of reasoning or producing coherent output. This effect was consistent across languages.
    \item Mistral-7B: Fluency degrades, but not to full gibberish. In Italian, repetition spikes markedly and unique sentence variety drops, indicating looping and reduced coherence. In Dutch, outputs shorten with more line breaks and fewer unique sentences, suggesting fragmenting/early truncation. In Japanese and Korean, mild increase in repetition with overall structure mostly intact.
    \item Aya expanse-8B: Generally stable and coherent across languages with mild regression. Italian and Dutch keep length and repetition near normal but lose sentence variety. Japanese shows more line breaks and higher repetition, less fluent. Korean is close to normal with a small drop in variety.
\end{itemize}

\paragraph{Top-1k Type-2 Neurons Deactivated.}
\begin{itemize}
    \item LLaMA3-8B: Although it produced gibberish for many samples (similar to when Type-1 neurons were deactivated), in some cases it generated coherent text in other languages (e.g., English), different from the input language, yet still aligned with the question context. We show an example in Tab.~\ref{tab:appendix:generation_ja_en_llama}. This phenomena support the fact that Type-2 neurons critically move internal reasoned representations to the latent space of output language in model space.
    \item Mistral-7B: A moderate decline in fluency is observed. In Italian, this manifests as increased looping and reduced sentence variety, while in Dutch the generations tend to be shorter, more fragmented, and frequently cut off prematurely. Nevertheless, the model occasionally produces coherent text in another language. For instance, Table~\ref{tab:appendix:generation_it_en_mistral} presents a case where the output switches from Italian to English.
    \item Aya expanse-8B: Although severely broken sentences are less likely to be generated compared to the other two models, the outputs tend to contain many repetitions. However, in a subset of samples, the model (1) generated answers in another language that were well aligned with the input question, or (2) produced answers in a code-switching–like form, combining the input language with another language. Some examples are shown in in Tabs.~\ref{tab:appendix:generation_ja_chi_aya} and~\ref{tab:appendix:ko_ja_aya_code_switch}.
\end{itemize}

Overall, for Type-2 neurons, certain samples yielded answers in a different language than those in the without-intervention setting, yet the responses remained coherent with the input question. In particular, the generated language tended to be linguistically proximate to the input language (e.g., input: Italian → output: English; input: Japanese → output: Chinese). These results in line with the PCA visualization of the geometry of language latent spaces in the hidden state space, where latent spaces corresponding to linguistically related languages remain closely clustered when the top-1k Type-2 neurons are deactivated (see Figs.~\ref{fig:appendix:pca_deactivating_type2_llama3},~\ref{fig:appendix:pca_deactivating_type2_mistral}, and~\ref{fig:appendix:pca_deactivating_type2_aya}).

\section{Statements}
\subsection{License for Artifacts}
\paragraph{Models.}
\begin{itemize}
    \item LLaMA3-8B: \texttt{License for Llama family}
    \item Mistral-7B: \texttt{apache-2.0}
    \item Aya expanse-8B: \texttt{cc-by-nc-4.0}
\end{itemize}
\paragraph{Datasets.}
\begin{itemize}
    \item Tatoeba: \texttt{cc-by-2.0}
    \item MKQA: \texttt{cc-by-3.0}
    \item MMLU-ProX: \texttt{MIT License}
    \item PolyWrite: \texttt{odc-by-1.0}
\end{itemize}
\paragraph{Evaluation Tools.}
\begin{itemize}
    \item lm-evaluation-harness: \texttt{MIT License}
\end{itemize}

\paragraph{Consistency of Usage.} All models, datasets and a evaluation tool were used in accordance with their original intended usage.

\subsection{AI Agent Usage}
AI agents were used for grammar checking during the writing of this paper.

\clearpage
% Llama3
\begin{table*}[t]
\centering
\small
\begin{tabular}{lrrrrrrrrrr}
\hline
Lang & T1 Same & T1: ja & T1: nl & T1: ko & T1: it & T2 Same & T2: ja & T2: nl & T2: ko & T2: it \\
\hline
ja & \textbf{-0.021} & -- & 0.051 & 0.042 & 0.053 & \textbf{0.386} & -- & 0.777 & 0.520 & 0.742 \\
nl & \textbf{-0.090} & -0.128 & -- & -0.092 & -0.089 & \textbf{0.542} & 0.710 & -- & 0.744 & 0.600 \\
ko & \textbf{0.055} & 0.004 & 0.085 & -- & 0.087 & \textbf{0.344} & 0.628 & 0.788 & -- & 0.805 \\
it & \textbf{-0.150} & -0.187 & -0.150 & -0.154 & -- & \textbf{0.520} & 0.675 & 0.586 & 0.702 & -- \\
\hline
\end{tabular}
\caption{\textbf{Hidden states similarity under cross-lingual deactivation across layers (LLaMA3-8B).} T1 refers to Type-1 neurons, while T2 refers to Type-2 neurons. The “Same” columns denote results obtained under non-cross-lingual deactivation, i.e., when the Transfer Neurons correspond to the same language as the input.}
\label{tab:appendix:cross-lingual_hs_llama}
\end{table*}

% Mistral
\begin{table*}[t]
\centering
\small
\begin{tabular}{lrrrrrrrrrr}
\hline
Lang & T1 Same & T1: ja & T1: nl & T1: ko & T1: it & T2 Same & T2: ja & T2: nl & T2: ko & T2: it \\
\hline
ja & \textbf{0.875} & -- & 0.879 & 0.852 & 0.884 & \textbf{0.754} & -- & 0.981 & 0.912 & 0.960 \\
nl & \textbf{0.845} & 0.875 & -- & 0.843 & 0.864 & \textbf{0.880} & 0.955 & -- & 0.928 & 0.906 \\
ko & \textbf{0.829} & 0.856 & 0.869 & -- & 0.874 & \textbf{0.805} & 0.864 & 0.949 & -- & 0.897 \\
it & \textbf{0.840} & 0.857 & 0.825 & 0.821 & -- & \textbf{0.901} & 0.962 & 0.919 & 0.932 & -- \\
\hline
\end{tabular}
\caption{\textbf{Hidden states similarity under cross-lingual deactivation across layers (Mistral-7B).}}
\label{tab:appendix:cross-lingual_hs_mistral}
\end{table*}

% Aya
\begin{table*}[t]
\centering
\small
\begin{tabular}{lrrrrrrrrrr}
\hline
Lang & T1 Same & T1: ja & T1: nl & T1: ko & T1: it & T2 Same & T2: ja & T2: nl & T2: ko & T2: it \\
\hline
ja & \textbf{0.885} & -- & 0.896 & 0.890 & 0.891 & \textbf{0.699} & -- & 0.945 & 0.936 & 0.947 \\
nl & \textbf{0.951} & 0.955 & -- & 0.956 & 0.949 & \textbf{0.755} & 0.922 & -- & 0.909 & 0.899 \\
ko & \textbf{0.923} & 0.933 & 0.930 & -- & 0.928 & \textbf{0.718} & 0.886 & 0.936 & -- & 0.958 \\
it & \textbf{0.946} & 0.952 & 0.947 & 0.952 & -- & \textbf{0.774} & 0.944 & 0.919 & 0.926 & -- \\
\hline
\end{tabular}
\caption{\textbf{Hidden states similarity under cross-lingual deactivation across layers (Aya expanse-8B).}}
\label{tab:appendix:cross-lingual_hs_aya}
\end{table*}

% LLaMA3-8B
% Mistral-7B
% Aya expanse-8B

% table: qa, llama3, 0.8
\begin{table*}
    \small
    \centering
    \renewcommand{\arraystretch}{1.2}
    \begin{tabularx}{\linewidth}{l *{4}{>{\centering\arraybackslash}X}}
        \hline
        \textbf{MKQA (F1)} & \textbf{ja} & \textbf{nl} & \textbf{ko} & \textbf{it} \\
        \hline
        \textbf{(a) normal} & 0.84 & 0.83 & 0.84 & 0.82 \\
        \textbf{(b) Type-1 $\Delta$} & -0.30 & -0.41 & -0.15 & -0.62 \\
        \textbf{(c) baseline $\Delta$} & -0.03 & -0.09 & -0.04 & -0.07 \\
        \hline
    \end{tabularx}
    \caption{\textbf{Changes in token-based F1 scores for questions with original scores above 0.8 (LLaMA3-8B).}}
    \label{tab:mkqa_f1_llama_0.8}
\end{table*}
% table: qa, mistral, 0.5
\begin{table*}
    \small
    \centering
    \renewcommand{\arraystretch}{1.2}
    \begin{tabularx}{\linewidth}{l *{4}{>{\centering\arraybackslash}X}}
        \hline
        \textbf{MKQA (F1)} & \textbf{ja} & \textbf{nl} & \textbf{ko} & \textbf{it} \\
        \hline
        \textbf{(a) normal} & 0.68 & 0.72 & 0.68 & 0.72 \\
        \textbf{(b) Type-1 $\Delta$} & -0.06 & -0.13 & -0.05 & -0.21 \\
        \textbf{(c) baseline $\Delta$} & -0.01 & -0.01 & $\pm$0 & -0.02 \\
        \hline
    \end{tabularx}
    \caption{\textbf{Changes in token-based F1 scores for questions with original scores above 0.5 (Mistral-7B).}}
    \label{tab:mkqa_f1_mistral_0.5}
\end{table*}
% table: qa, mistral, 0.8
\begin{table*}
    \small
    \centering
    \renewcommand{\arraystretch}{1.2}
    \begin{tabularx}{\linewidth}{l *{4}{>{\centering\arraybackslash}X}}
        \hline
        \textbf{MKQA (F1)} & \textbf{ja} & \textbf{nl} & \textbf{ko} & \textbf{it} \\
        \hline
        \textbf{(a) normal} & 0.90 & 0.96 & 0.90 & 0.96 \\
        \textbf{(b) Type-1 $\Delta$} & -0.09 & -0.06 & -0.05 & -0.28 \\
        \textbf{(c) baseline $\Delta$} & -0.03 & -0.02 & -0.01 & -0.02 \\
        \hline
    \end{tabularx}
    \caption{\textbf{Changes in token-based F1 scores for questions with original scores above 0.8 (Mistral-7B).}}
    \label{tab:mkqa_f1_mistral_0.8}
\end{table*}
% table: qa, aya, 0.5
\begin{table*}
    \small
    \centering
    \renewcommand{\arraystretch}{1.2}
    \begin{tabularx}{\linewidth}{l *{4}{>{\centering\arraybackslash}X}}
        \hline
        \textbf{MKQA (F1)} & \textbf{ja} & \textbf{nl} & \textbf{ko} & \textbf{it} \\
        \hline
        \textbf{(a) normal} & 0.71 & 0.72 & 0.69 & 0.73 \\
        \textbf{(b) Type-1 $\Delta$} & -0.05 & -0.16 & -0.11 & -0.28 \\
        \textbf{(c) baseline $\Delta$} & $\pm$0 & -0.02 & $\pm$0 & -0.01 \\
        \hline
    \end{tabularx}
    \caption{\textbf{Changes in token-based F1 scores for questions with original scores above 0.5 (Aya expanse-8B).}}
    \label{tab:mkqa_f1_aya_0.5}
\end{table*}
% table: qa, aya, 0.8
\begin{table*}
    \small
    \centering
    \renewcommand{\arraystretch}{1.2}
    \begin{tabularx}{\linewidth}{l *{4}{>{\centering\arraybackslash}X}}
        \hline
        \textbf{MKQA (F1)} & \textbf{ja} & \textbf{nl} & \textbf{ko} & \textbf{it} \\
        \hline
        \textbf{(a) normal} & 0.90 & 0.93 & 0.89 & 0.94 \\
        \textbf{(b) Type-1 $\Delta$} & -0.06 & -0.19 & -0.14 & -0.26 \\
        \textbf{(c) baseline $\Delta$} & $\pm$0 & -0.02 & $\pm$0 & -0.02 \\
        \hline
    \end{tabularx}
    \caption{\textbf{Changes in token-based F1 scores for questions with original scores above 0.8 (Aya expanse-8B).}}
    \label{tab:mkqa_f1_aya_0.8}
\end{table*}

%- MMLU-ProX -%

%- LLaMA3-8B -%
% mmluprox, llama3, ja
\begin{table*}[ht]
\centering
\small
\caption{MMLU-ProX: Type-1 deactivated vs Baseline deactivated vs w/o intervention (LLaMA3-8B, \textbf{Japanese})}
\begin{tabular}{lccc}
\toprule
\textbf{Tasks} & \textbf{Type-1 Score} & \textbf{Baseline Score} & \textbf{w/o intervention} \\
\midrule
\textbf{OVERALL}         & \textbf{0}. & 0.1996 & 0.2033 \\
\addlinespace
\quad - biology          & 0. & 0.2873 & 0.2887 \\
\quad - business         & 0. & 0.2180 & 0.2294 \\
\quad - chemistry        & 0. & 0.1590 & 0.1564 \\
\quad - computer\_science & 0. & 0.2756 & 0.2707 \\
\quad - economics        & 0. & 0.2915 & 0.3009 \\
\quad - engineering      & 0. & 0.1362 & 0.1414 \\
\quad - health           & 0. & 0.2052 & 0.2038 \\
\quad - history          & 0. & 0.1942 & 0.2152 \\
\quad - law              & 0. & 0.1043 & 0.1178 \\
\quad - math             & 0. & 0.1887 & 0.1828 \\
\quad - other            & 0. & 0.1959 & 0.2045 \\
\quad - philosophy       & 0. & 0.2285 & 0.2265 \\
\quad - physics          & 0. & 0.1316 & 0.1355 \\
\quad - psychology       & 0. & 0.3283 & 0.3308 \\
\bottomrule
\end{tabular}
\label{mmluprox_type1_baseline_woi_llama3_ja}
\end{table*}
% mmluprox, llama3, ko
\begin{table*}[ht]
\centering
\small
\caption{MMLU-ProX: Type-1 deactivated vs Baseline deactivated vs w/o intervention (LLaMA3-8B, \textbf{Korean})}
\begin{tabular}{lccc}
\toprule
\textbf{Tasks} & \textbf{Type-1 Score} & \textbf{Baseline Score} & \textbf{w/o intervention} \\
\midrule
\textbf{OVERALL}         & \textbf{0}. & 0.2097 & 0.2115 \\
\addlinespace
\quad - biology          & 0. & 0.2580 & 0.2524 \\
\quad - business         & 0. & 0.2408 & 0.2598 \\
\quad - chemistry        & 0. & 0.1793 & 0.1661 \\
\quad - computer\_science & 0. & 0.2829 & 0.2561 \\
\quad - economics        & 0. & 0.2867 & 0.2950 \\
\quad - engineering      & 0. & 0.1754 & 0.1816 \\
\quad - health           & 0. & 0.1747 & 0.1921 \\
\quad - history          & 0. & 0.1890 & 0.2073 \\
\quad - law              & 0. & 0.1283 & 0.1220 \\
\quad - math             & 0. & 0.2095 & 0.2250 \\
\quad - other            & 0. & 0.2348 & 0.2424 \\
\quad - philosophy       & 0. & 0.2084 & 0.1944 \\
\quad - physics          & 0. & 0.1486 & 0.1470 \\
\quad - psychology       & 0. & 0.3108 & 0.2995 \\
\bottomrule
\end{tabular}
\label{mmluprox_type1_baseline_woi_llama3_ko}
\end{table*}
% mmluprox, llama3, fr
\begin{table*}[ht]
\centering
\small
\caption{MMLU-ProX: Type-1 deactivated vs Baseline deactivated vs w/o intervention (LLaMA3-8B, \textbf{French})}
\begin{tabular}{lccc}
\toprule
\textbf{Tasks} & \textbf{Type-1 Score} & \textbf{Baseline Score} & \textbf{w/o intervention} \\
\midrule
\textbf{OVERALL}         & \textbf{0}. & 0.2728 & 0.2705 \\
\addlinespace
\quad - biology          & 0. & 0.4170 & 0.4338 \\
\quad - business         & 0. & 0.2738 & 0.2801 \\
\quad - chemistry        & 0. & 0.1820 & 0.1988 \\
\quad - computer\_science & 0. & 0.2683 & 0.2390 \\
\quad - economics        & 0. & 0.3744 & 0.3756 \\
\quad - engineering      & 0. & 0.2054 & 0.2002 \\
\quad - health           & 0. & 0.2780 & 0.2838 \\
\quad - history          & 0. & 0.3123 & 0.2992 \\
\quad - law              & 0. & 0.1512 & 0.1293 \\
\quad - math             & 0. & 0.2494 & 0.2428 \\
\quad - other            & 0. & 0.3202 & 0.3139 \\
\quad - philosophy       & 0. & 0.3026 & 0.2926 \\
\quad - physics          & 0. & 0.2271 & 0.2248 \\
\quad - psychology       & 0. & 0.4110 & 0.4085 \\
\bottomrule
\end{tabular}
\label{mmluprox_type1_baseline_woi_llama3_fr}
\end{table*}
%- Mistral-7B -%
% mmluprox, mistral, ja
\begin{table*}[ht]
\centering
\small
\caption{MMLU-ProX: Type-1 deactivated vs Baseline deactivated vs w/o intervention (Mistral-7B, \textbf{Japanese})}
\begin{tabular}{lccc}
\toprule
\textbf{Tasks} & \textbf{Type-1 Score} & \textbf{Baseline Score} & \textbf{w/o intervention} \\
\midrule
\textbf{OVERALL}         & \textbf{0.0717} & 0.1712 & 0.1750 \\
\addlinespace
\quad - biology          & 0.0293 & 0.2594 & 0.2664 \\
\quad - business         & 0.1179 & 0.2155 & 0.2231 \\
\quad - chemistry        & 0.0221 & 0.1210 & 0.1193 \\
\quad - computer\_science & 0.1488 & 0.2341 & 0.2366 \\
\quad - economics        & 0.1197 & 0.2393 & 0.2512 \\
\quad - engineering      & 0.0485 & 0.1620 & 0.1465 \\
\quad - health           & 0.0815 & 0.1718 & 0.1659 \\
\quad - history          & 0.0919 & 0.1470 & 0.1575 \\
\quad - law              & 0.0083 & 0.0949 & 0.1116 \\
\quad - math             & 0.0607 & 0.1392 & 0.1540 \\
\quad - other            & 0.1061 & 0.1753 & 0.1742 \\
\quad - philosophy       & 0.1062 & 0.1764 & 0.1723 \\
\quad - physics          & 0.0370 & 0.1055 & 0.070 \\
\quad - psychology       & 0.1441 & 0.2820 & 0.2882 \\
\bottomrule
\end{tabular}
\label{mmluprox_type1_baseline_woi_mistral_ja}
\end{table*}
% mmluprox, mistral, ko
\begin{table*}[ht]
\centering
\small
\caption{MMLU-ProX: Type-1 deactivated vs Baseline deactivated vs w/o intervention (Mistral-7B, \textbf{Korean})}
\begin{tabular}{lccc}
\toprule
\textbf{Tasks} & \textbf{Type-1 Score} & \textbf{Baseline Score} & \textbf{w/o intervention} \\
\midrule
\textbf{OVERALL}         & \textbf{0.0454} & 0.1606 & 0.1593 \\
\addlinespace
\quad - biology          & 0.0014 & 0.1520 & 0.1632 \\
\quad - business         & 0.0963 & 0.2243 & 0.2281 \\
\quad - chemistry        & 0.0053 & 0.1042 & 0.0989 \\
\quad - computer\_science & 0.0829 & 0.2561 & 0.2439 \\
\quad - economics        & 0.1126 & 0.2192 & 0.2121 \\
\quad - engineering      & 0.0444 & 0.1486 & 0.1465 \\
\quad - health           & 0.0757 & 0.1645 & 0.1587 \\
\quad - history          & 0.0367 & 0.1286 & 0.1312 \\
\quad - law              & 0.0083 & 0.1137 & 0.0980 \\
\quad - math             & 0.0200 & 0.1340 & 0.1295 \\
\quad - other            & 0.0714 & 0.1634 & 0.1688 \\
\quad - philosophy       & 0.0641 & 0.1383 & 0.1242 \\
\quad - physics          & 0.0200 & 0.1162 & 0.1309 \\
\quad - psychology       & 0.0677 & 0.2845 & 0.2845 \\
\bottomrule
\end{tabular}
\label{mmluprox_type1_baseline_woi_mistral_ko}
\end{table*}
% mmluprox, mistral, fr
\begin{table*}[ht]
\centering
\small
\caption{MMLU-ProX: Type-1 deactivated vs Baseline deactivated vs w/o intervention (Mistral-7B, \textbf{French})}
\begin{tabular}{lccc}
\toprule
\textbf{Tasks} & \textbf{Type-1 Score} & \textbf{Baseline Score} & \textbf{w/o intervention} \\
\midrule
\textbf{OVERALL}         & \textbf{0.1477} & 0.2462 & 0.2460 \\
\addlinespace
\quad - biology          & 0.1311 & 0.4254 & 0.4045 \\
\quad - business         & 0.1584 & 0.2459 & 0.2446 \\
\quad - chemistry        & 0.0972 & 0.1572 & 0.1705 \\
\quad - computer\_science & 0.1488 & 0.2488 & 0.2537 \\
\quad - economics        & 0.1765 & 0.3436 & 0.3389 \\
\quad - engineering      & 0.0939 & 0.1806 & 0.1734 \\
\quad - health           & 0.2227 & 0.2693 & 0.2780 \\
\quad - history          & 0.2283 & 0.2625 & 0.2756 \\
\quad - law              & 0.0667 & 0.1397 & 0.1335 \\
\quad - math             & 0.0888 & 0.1969 & 0.1895 \\
\quad - other            & 0.1861 & 0.2825 & 0.2955 \\
\quad - philosophy       & 0.2204 & 0.2565 & 0.2545 \\
\quad - physics          & 0.1532 & 0.2009 & 0.1971 \\
\quad - psychology       & 0.2531 & 0.3960 & 0.4048 \\
\bottomrule
\end{tabular}
\label{mmluprox_type1_baseline_woi_mistral_fr}
\end{table*}
%- Aya expanse-8B -%
% mmluprox, aya, ja
\begin{table*}[ht]
\centering
\small
\caption{MMLU-ProX: Type-1 deactivated vs Baseline deactivated vs w/o intervention (Aya expanse-8B, \textbf{Japanese})}
\begin{tabular}{lccc}
\toprule
\textbf{Tasks} & \textbf{Type-1 Score} & \textbf{Baseline Score} & \textbf{w/o intervention} \\
\midrule
\textbf{OVERALL}         & \textbf{0.1685} & 0.2636 & 0.2648 \\
\addlinespace
\quad - biology          & 0.2734 & 0.4184 & 0.4017 \\
\quad - business         & 0.1762 & 0.2801 & 0.2842 \\
\quad - chemistry        & 0.1166 & 0.2147 & 0.2138 \\
\quad - computer\_science & 0.1610 & 0.2585 & 0.2610 \\
\quad - economics        & 0.1339 & 0.2133 & 0.2097 \\
\quad - engineering      & 0.1589 & 0.2116 & 0.2260 \\
\quad - health           & 0.1426 & 0.2402 & 0.2402 \\
\quad - history          & 0.1706 & 0.2808 & 0.2756 \\
\quad - law              & 0.1074 & 0.1616 & 0.1814 \\
\quad - math             & 0.1651 & 0.3168 & 0.3101 \\
\quad - other            & 0.1926 & 0.2587 & 0.2554 \\
\quad - philosophy       & 0.1583 & 0.2325 & 0.2305 \\
\quad - physics          & 0.1463 & 0.2325 & 0.2394 \\
\quad - psychology       & 0.3070 & 0.4173 & 0.4148 \\
\bottomrule
\end{tabular}
\label{mmluprox_type1_baseline_woi_aya_ja}
\end{table*}
% mmluprox, aya, ko
\begin{table*}[ht]
\centering
\small
\caption{MMLU-ProX: Type-1 deactivated vs Baseline deactivated vs w/o intervention (Aya expanse-8B, \textbf{Korean})}
\begin{tabular}{lccc}
\toprule
\textbf{Tasks} & \textbf{Type-1 Score} & \textbf{Baseline Score} & \textbf{w/o intervention} \\
\midrule
\textbf{OVERALL}         & \textbf{0.1854} & 0.2696 & 0.2702 \\
\addlinespace
\quad - biology          & 0.2148 & 0.3250 & 0.3543 \\
\quad - business         & 0.1952 & 0.2864 & 0.2864 \\
\quad - chemistry        & 0.1466 & 0.2094 & 0.2191 \\
\quad - computer\_science & 0.1244 & 0.3000 & 0.30981 \\
\quad - economics        & 0.2666 & 0.3318 & 0.3412 \\
\quad - engineering      & 0.1476 & 0.2652 & 0.2487 \\
\quad - health           & 0.1587 & 0.2213 & 0.2183 \\
\quad - history          & 0.1496 & 0.2598 & 0.2415 \\
\quad - law              & 0.1241 & 0.1814 & 0.1700 \\
\quad - math             & 0.1858 & 0.3198 & 0.3183 \\
\quad - other            & 0.1991 & 0.2630 & 0.2727 \\
\quad - philosophy       & 0.1743 & 0.1924 & 0.1864 \\
\quad - physics          & 0.1586 & 0.2232 & 0.2186 \\
\quad - psychology       & 0.3434 & 0.4110 & 0.4123 \\
\bottomrule
\end{tabular}
\label{mmluprox_type1_baseline_woi_aya_ko}
\end{table*}
% mmluprox, aya, fr
\begin{table*}[ht]
\centering
\small
\caption{MMLU-ProX: Type-1 deactivated vs Baseline deactivated vs w/o intervention (Aya expanse-8B, \textbf{French})}
\begin{tabular}{lccc}
\toprule
\textbf{Tasks} & \textbf{Type-1 Score} & \textbf{Baseline Score} & \textbf{w/o intervention} \\
\midrule
\textbf{OVERALL}         & \textbf{0.2371} & 0.3191 & 0.3207 \\
\addlinespace
\quad - biology          & 0.3501 & 0.5397 & 0.5300 \\
\quad - business         & 0.2383 & 0.3131 & 0.3105 \\
\quad - chemistry        & 0.1590 & 0.2182 & 0.2279 \\
\quad - computer\_science & 0.2220 & 0.3049 & 0.3000 \\
\quad - economics        & 0.3235 & 0.4218 & 0.4336 \\
\quad - engineering      & 0.1827 & 0.2693 & 0.2869 \\
\quad - health           & 0.2606 & 0.3290 & 0.3115 \\
\quad - history          & 0.1444 & 0.3360 & 0.3150 \\
\quad - law              & 0.1470 & 0.1741 & 0.18777 \\
\quad - math             & 0.2865 & 0.3434 & 0.3501 \\
\quad - other            & 0.2522 & 0.3290 & 0.3323 \\
\quad - philosophy       & 0.2064 & 0.2866 & 0.2786 \\
\quad - physics          & 0.1964 & 0.2594 & 0.2594 \\
\quad - psychology       & 0.3446 & 0.4511 & 0.4398 \\
\bottomrule
\end{tabular}
\label{mmluprox_type1_baseline_woi_aya_fr}
\end{table*}

% output language change, llama
\begin{CJK}{UTF8}{min} % ← テーブル全体を日本語/中国語OKにする
\begin{table*}[t]
\footnotesize
\setlength{\tabcolsep}{6pt}
\renewcommand{\arraystretch}{1.2}
\begin{tabularx}{\linewidth}{lX}
\toprule
\textbf{Question (Japanese)} & プロジェクトの期限に間に合わなかったことを上司に謝罪するメールを作成します。遅延の理由を説明し、同じことが二度と起こらないようにするための手順を説明します。\\
& \textcolor{gray}{\textit{Write an email to your supervisor apologizing for not meeting the project deadline. Explain the reasons for the delay and describe the steps that will be taken to ensure that the same issue does not occur again.}} \\
\midrule
\textbf{Output without any intervention (Japanese)} &
いつもお世話になっております。\textbackslash n     今日は、プロジェクトの期限に間に合わなかったことをお詫び申し上げます。\textbackslash n     その理由は、... \\
& \textcolor{gray}{\textit{Thank you always for your support.\textbackslash n     Today, I would like to apologize for not meeting the project deadline.\textbackslash n     The reason is, ... ...}} \\
\midrule
\textbf{Output with top-1k Type-2 neurons deactivated (English)} &
Dear Mr. Smith,\textbackslash n     \textbackslash n     I'm writing to apologize for missing the project's deadline. The reason for the delay was ... \\
\bottomrule
\end{tabularx}
\caption{\textbf{Example of generated text in which the output language changed from the input language after deactivating the top-1k Type-2 neurons (LLaMA3-8B).}}
\label{tab:appendix:generation_ja_en_llama}
\end{table*}
\end{CJK}
% output language change, mistral
\begin{table*}[t]
\footnotesize
\setlength{\tabcolsep}{6pt}
\renewcommand{\arraystretch}{1.2}
\begin{tabularx}{\linewidth}{lX}
\toprule
\textbf{Question (Italian)} & Scrivi un\&\#39;email a un venditore che ha consegnato prodotti errati o danneggiati. Spiega educatamente il problema, fornisci dettagli e richiedi un rimborso o una sostituzione.\\
& \textcolor{gray}{\textit{Write an email to a seller who has delivered incorrect or damaged products. Politely explain the problem, provide details, and request a refund or a replacement.}} \\
\midrule
\textbf{Output without any intervention (Italian)} &
Grazie per aver contattato il nostro team di supporto.\textbackslash n\textbackslash n    Siamo molto scioccati per il problema che hai avuto con il nostro prodotto. Siamo dispiaciuti per il disagio che hai subito e siamo pronti a fornire una soluzione. ... \\
& \textcolor{gray}{\textit{Thank you for contacting our support team. We are very shocked about the issue you experienced with our product. We are sorry for the inconvenience you have suffered, and we are ready to provide a solution. ...}} \\
\midrule
\textbf{Output with top-1k Type-2 neurons deactivated (English)} &
Hello,\textbackslash n\textbackslash nI recently purchased a product from your store and unfortunately, it arrived damaged. I'm writing to request a replacement or a refund for the item.\textbackslash n\textbackslash nHere are the details of the issue: ... \\
\bottomrule
\end{tabularx}
\caption{\textbf{Example of generated text in which the output language changed from the input language after deactivating the top-1k Type-2 neurons (Mistral-7B).}}
\label{tab:appendix:generation_it_en_mistral}
\end{table*}
% output language change, aya
\begin{CJK}{UTF8}{min} % ← テーブル全体を日本語/中国語OKにする
\begin{table*}[t]
\footnotesize
\setlength{\tabcolsep}{6pt}
\renewcommand{\arraystretch}{1.2}
\begin{tabularx}{\linewidth}{lX}
\toprule
\textbf{Question (Japanese)} & 気候変動は地球に大きな影響を与えています。気候変動の主な原因は何だと思いますか。また、気候変動は環境にどのような影響を与える可能性がありますか。\\
& \textcolor{gray}{\textit{Climate change is having a significant impact on the Earth. What do you think are the main causes of climate change, and what potential impacts might it have on the environment?}} \\
\midrule
\textbf{Output without any intervention (Japanese)} &
気候変動の主な原因は、人間の活動による温室効果ガスの排出です。特に、化石燃料の燃焼（石炭、石油、天然ガスなど）や森林破壊が大きな要因となっています。これらの活動によって二酸化炭素（CO2）、メタン（CH4）、亜酸化窒素（N2O）などの温室効果ガスが大量に放出され、地球の大気中に蓄積しています。... \\
& \textcolor{gray}{\textit{The main cause of climate change is the emission of greenhouse gases from human activities. In particular, the burning of fossil fuels (such as coal, oil, and natural gas) and deforestation are major factors. These activities release large amounts of greenhouse gases, including carbon dioxide (CO₂), methane (CH₄), and nitrous oxide (N₂O), which accumulate in the Earth’s atmosphere. ...}} \\
\midrule
\textbf{Output with top-1k Type-2 neurons deactivated (Chinese)} &
氣候變動的主要原因之一是人類活動，特別是燃燒化石燃料（如煤、石油和天然氣）和森林砍伐。這些活動釋放出大量的溫室氣體（如二氧化碳和甲烷），導致地球大氣層中的溫室效應增強，進而導致全球氣溫上升。... \\
& \textcolor{gray}{\textit{One of the main causes of climate change is human activity, particularly the burning of fossil fuels (such as coal, oil, and natural gas) and deforestation. These activities release large amounts of greenhouse gases (such as carbon dioxide and methane), which intensify the greenhouse effect in the Earth’s atmosphere and, in turn, lead to a rise in global temperatures. ...}} \\
\bottomrule
\end{tabularx}
\caption{\textbf{Example of generated text in which the output language changed from the input language after deactivating the top-1k Type-2 neurons (Aya expanse-8B).}}
\label{tab:appendix:generation_ja_chi_aya}
\end{table*}
\end{CJK}
% output like code-switch, aya
\begin{CJK}{UTF8}{mj} % ← 韓国語を表示可能に
\begin{table*}[t]
\footnotesize
\setlength{\tabcolsep}{6pt}
\renewcommand{\arraystretch}{1.2}
\begin{tabularx}{\linewidth}{lX}
\toprule
\textbf{Question (Korean)} & 인구 과잉은 많은 도시에서 심각한 문제가 되고 있습니다. 인구 과잉의 원인은 무엇이며 도시 생활에 미치는 영향은 무엇입니까? \\
& \textcolor{gray}{\textit{Overpopulation is becoming a serious problem in many cities. What are the causes of overpopulation, and what effects does it have on urban life?}} \\
\midrule
\textbf{Output without any intervention (Korean)} &
21세기의 많은 도시들은 급속한 인구 증가로 인해 인구 과잉 문제를 겪고 있습니다. 인구 과잉은 특정 지역이나 도시에 너무 많은 사람들이 모여 사는 현상을 말합니다. 이 현상의 원인과 영향에 대해 살펴보겠습니다. ... \\
& \textcolor{gray}{\textit{Many cities in the 21st century are experiencing problems of overpopulation due to rapid population growth. Overpopulation refers to the phenomenon in which too many people live concentrated in a particular area or city. Let us examine the causes and effects of this phenomenon. ...}} \\
\midrule
\textbf{Output with top-1k Type-2 neurons deactivated (Korean and Japanese)} &
人口過剩, 즉 인구 과잉은 한 지역에서나 국가 전체에서, 그 지역이나 국가의 생태학적 용량을 초과하는 인구 수를 의미합니다. 人口過剩의 원인과 도시 생활에 미치는 영향은 다음과 같습니다.\textbackslash n\textbackslash n 1. 人口增加率과 資源有限性: ... \\
& \textcolor{gray}{\textit{Overpopulation, that is, excessive population, refers to a number of people in a region or an entire country that exceeds the ecological capacity of that region or country. The causes of overpopulation and its effects on urban life are as follows.\textbackslash n\textbackslash n
1. Population growth rate and limited resources: ...}} \\
\bottomrule
\end{tabularx}
\caption{\textbf{Example of generated text in code-switching–like form in Korean and Japanese after deactivating the top-1k Type-2 neurons (Aya expanse-8B).}}
\label{tab:appendix:ko_ja_aya_code_switch}
\end{table*}
\end{CJK}

\end{document}